\documentclass{book}
\usepackage{latexsym}
\usepackage{lingmacros,xypic,prooftree,amssymb,lscape,a4wide}
\usepackage{pxfonts}
\usepackage{stmaryrd}
\usepackage{times}
\usepackage{wasysym}
\title{Logical Computational Linguistics}
\author{Glyn V.~Morrill\footnote{glyn.morrill@upc.edu} \& Oriol Valent\'{\i}n\footnote{oriol.valentin@upc.edu}}
\date{17th April 2026\footnote{In citing, please refer to this draft as version v.1.0--0026.04.17}\\
\rotatebox{-62.85
}{$\tiny
		\begin{array}{r|ccc|ccc|c|c|c|c|c|c|c|c|}
			\cline{2-14}
			& \multicolumn{3}{c|}{\begin{tabular}{c}concatenative\\multiplicatives\end{tabular}}&
			\multicolumn{3}{c|}{\begin{tabular}{c}intercalative\\ multiplicatives\end{tabular}}  &
			\mathrm{additives} & \begin{tabular}{c}1st~order\\quantifiers\end{tabular} & 
			\begin{tabular}{c}normal\\modalities\end{tabular} & 
			\begin{tabular}{c}bracket\\modalities\end{tabular} & \begin{tabular}{c}soft\\exponentials\end{tabular} 
			&\multicolumn{1}{c|}{\begin{tabular}{c}limited\\contraction\\\& limited\\expansion\end{tabular}} & 
			\mathrm{difference}
			\\
			\cline{2-14}
			&&&&&&&&&&&&&\\
			&/ & & \bsl & \scircum{} & & \sinfix{} & \aconj_{} & \bigwedge_{} & \mymod_{} & \abrack_{} & \univexp_{} &|_{}&\\
			& & \product  & & & \swprod{}{} &&&&&& && -  \\
			& & I &&& J_{}&& \adisj_{} & \bigvee_{} & \emod_{} & \mybrack_{} & \exstexp_{} &+ & \\
			&&&&&&&&&&&&&\\
			\cline{2-14}
			&\multicolumn{3}{c|}{\rotatebox{90}{\begin{tabular}[t]{l}Continuity\end{tabular}}}  &
			\multicolumn{3}{c|}{\rotatebox{90}{\begin{tabular}[t]{l}Discontinuity\end{tabular}}}&
			\rotatebox{90}{\begin{tabular}[t]{l}Polymorphism\end{tabular}}&
			\rotatebox{90}{\begin{tabular}[t]{l}Features\end{tabular}}&
			\rotatebox{90}{\begin{tabular}[t]{l}Intensionality\end{tabular}}&
			\rotatebox{90}{\begin{tabular}[t]{l}Prosodic\\domains\end{tabular}}&
			\rotatebox{90}{\begin{tabular}[t]{l}Non-linearity\end{tabular}}&
			\rotatebox{90}{\begin{tabular}[t]{l}Anaphora\\\& Words-as-types\end{tabular}}&
			\rotatebox{90}{\begin{tabular}[t]{l}Exceptions\end{tabular}}\\
			\cline{2-14}
		\end{array}
		$}}

\newcommand{\techterm}[1]{\mbox{\it #1}}

\newcommand{\disp}[1]{\enumsentence{#1}}

\newcommand{\verbquote}[1]{``#1''}

\newcommand{\tb}{\hspace*{0.25in}}

\newcommand{\tab}{\hspace*{0.5in}}

\newcommand{\oscott}{[[}
\newcommand{\cscott}{]]}
\newcommand{\vtab}{\ \vspace{2.5ex}\\}

\newcommand{\yields}{\mbox{\ $\Rightarrow$\ }}

\newcommand{\add}{\mbox{$+$}}
\newcommand{\nil}{\mbox{$0$}}
\newcommand{\one}{\mbox{$1$}}

\newcommand{\wrap}{\mbox{$\times$}}
\newcommand{\bsl}{\mbox{$\backslash$}}
\newcommand{\product}{\mbox{$\bullet$}}
\newcommand{\infix}{\mbox{$\downarrow$}}

\newcommand{\circum}{\mbox{$\uparrow$}}
\newcommand{\wprod}{\mbox{$\odot$}}
\newcommand{\sinfix}[1]{\mbox{$\downarrow_{#1}$}}
\newcommand{\swrap}[1]{\mbox{$\wrap{_{#1}}$}}
\newcommand{\smwrap}[1]{\mbox{$\,|{_{#1}}\,$}}
\newcommand{\mwrap}[1]{\mbox{$|$}}
\newcommand{\scircum}[1]{\mbox{$\uparrow_{#1}$}}
\newcommand{\swprod}[1]{\mbox{$\odot{_{#1}}$}}
\newcommand{\D}{\mbox{\bf D}}
\newcommand{\abrack}{\mbox{$[\,]^{-1}$}}
\newcommand{\mybrack}{\mbox{$\langle\rangle$}}
\newcommand{\sep}{\mbox{$1$}}

\newcommand{\vect}[1]{\overrightarrow{#1}}
\newcommand{\aconj}{\mbox{$\&$}}
\newcommand{\adisj}{\mbox{$\oplus$}}
\newcommand{\iaconj}{\mbox{$\sqcap$}}
\newcommand{\iadisj}{\mbox{$\sqcup$}}
\newcommand{\dn}{\mbox{$^\vee$}}
\newcommand{\up}{\mbox{$^\wedge$}}
\newcommand{\dnc}{\mbox{$^\cup$}}
\newcommand{\upc}{\mbox{$^\cap$}}
\newcommand{\inp}{{\mbox{$^{\bullet}$}}}
\newcommand{\out}{{\mbox{$^{\circ}$}}}
\newcommand{\mymod}{\mbox{$\Box$}}
\newcommand{\emod}{\mbox{$\Diamond$}}
\newcommand{\imod}{\mbox{$\blacksquare$}}
\newcommand{\iemod}{\mbox{$\blacklozenge$}}

\newcommand{\univexp}{\mbox{!}}
\newcommand{\exstexp}{\mbox{?}}
\newcommand{\rightproj}{\mbox{$\triangleright^{-1}$}}
\newcommand{\leftproj}{\mbox{$\triangleleft^{-1}$}}
\newcommand{\rightinj}{\mbox{$\triangleright$}}
\newcommand{\leftinj}{\mbox{$\triangleleft$}}
\newcommand{\nproj}{\mbox{$\triangleleft\triangleright^{-1}$}}
\newcommand{\ninj}{\mbox{$\triangleleft\triangleright$}}
\newcommand{\mysplit}{\mbox{$\check{\ }$}}
\newcommand{\bridge}{\mbox{$\hat{\ }$}}
\newcommand{\ssplit}[1]{\mbox{$\check{\ }^{_{#1}}$}}
\newcommand{\sbridge}[1]{\mbox{$\hat{\ }^{_{#1}}$}}
\newcommand{\nprod}{\mbox{$\circ$}}
\newcommand{\ndiv}{\mbox{$\div$}}
\newcommand{\ninfix}{\mbox{$\stackrel{\vee}{=}$}}
\newcommand{\nextract}{\mbox{$\stackrel{\wedge}{=}$}}
\newcommand{\ndprod}{\mbox{$\diamond$}}
\newcommand{\nbridge}{\mbox{$\hat{\hat{\ }}$}}
\newcommand{\nsplit}{\mbox{$\check{\check{\ }}$}}
\newcommand{\ass}{\mbox{$:\,$}}
\newcommand{\sass}{\mbox{$;\,$}}
\newcommand{\mun}{\mbox{$\,\uplus\,$}}
\newcommand{\subst}[1]{\mbox{$\{#1\}$}}
\newcommand{\zero}{\mbox{$0$}}
\newcommand{\commentout}[1]{}

\newcommand{\fununder}{\mbox{$\multimap$}}
\newcommand{\argunder}{\mbox{$\multimapdot$}}
\newcommand{\argover}{\mbox{$\multimapdotinv$}}
\newcommand{\funover}{\mbox{$\multimapinv$}}
\newcommand{\arginfix}[1]{\mbox{$
		\begin{picture}(6,8)(0,-2)\put(0.7,7){\rotatebox{-90}{$\multimap$}}
		\end{picture}_{#1}$}}
\newcommand{\funcircum}[1]{\mbox{$
		\begin{picture}(6,7)(0,-2)\put(1.5,-3){\rotatebox{90}{$\multimapdot$}}
		\end{picture}_{#1}$}}
\newcommand{\funinfix}[1]{\mbox{$
		\begin{picture}(6,8)(0,-2)\put(0.7,7){\rotatebox{-90}{$\multimapdot$}}
		\end{picture}_{#1}$}}
\newcommand{\argcircum}[1]{\mbox{$
		\begin{picture}(6,7)(0,-2)\put(1.5,-3){\rotatebox{90}{$\multimap$}}
		\end{picture}_{#1}$}}
\newcommand{\leftiprod}{\mbox{$\LEFTcircle$}}
\newcommand{\rightiprod}{\mbox{$\RIGHTcircle$}}
\newcommand{\upperiwprod}[1]{\mbox{$\begin{picture}(7,7)(0,0)\put(0,7){\rotatebox{-90}{$\LEFTcircle$}}\end{picture}_{#1}$}}
\newcommand{\loweriwprod}[1]{\mbox{$\begin{picture}(7,7)(0,0)\put(0,7){\rotatebox{-90}{$\RIGHTcircle$}}\end{picture}_{#1}$}}
\newcommand{\nlif}{\mbox{$>$}}
\newcommand{\nlip}{\mbox{$\ogreaterthan$}}

\usepackage{natbib}

\newcommand{\syncnst}[1]{\mbox{\bf #1}}
\newcommand{\pcnst}[1]{\mbox{\bf #1}}
\newcommand{\CN}{\mbox{$\mathit{CN}$}}

\newcommand{\VP}{\mbox{$\mathit{VP}$}}
\newcommand{\TV}{\mbox{$\mathit{TV}$}}
\newcommand{\Det}{\mbox{$\mathit{Det}$}}
\newcommand{\PP}{\mbox{$\mathit{PP}$}}
\newcommand{\CP}{\mbox{$\mathit{CP}$}}

\newcommand{\lyields}{\mbox{$\Longrightarrow$}} 
\newcommand{\lyieldsw}{\mbox{$\Longrightarrow_w$}} 
\newcommand{\lyieldsws}{\mbox{$\Longrightarrow$}} 
\newcommand{\ncut}{\mbox{\it{n-Cut}}} 
\newcommand{\foc}{\mbox{\it{foc}}} 
\newcommand{\pcut}{\mbox{\it{p-Cut}}} 
\newcommand{\pf}{\mbox{ $\Diamond$ foc}}

\newcommand{\prf}[1]{\noindent{\bf Proof}. #1 This completes the proof.}

\newcommand{\mymin}{{\it min}}
\newcommand{\mymax}{{\it max}}

\newcommand{\diagmod}{\mbox{$\Box\hspace{-1.4ex}\times$}}
\newcommand{\diagemod}{\mbox{${\large \Diamond}\hspace{-1.9ex}{}{+}{}$}}
\newcommand{\mini}{\footnotesize}
\newcommand{\lingform}[1]{\mbox{\bf #1}} 
\newcommand{\scare}[1]{`#1'}
\newcommand{\alphaplus}{\mbox{\bf $\mathbf{\alpha^+}$}}
\newcommand{\Phinplus}{\mbox{\bf $\mathbf{\Phi^{n+}}$}}

\newcommand{\swraptwo}[1]{\mbox{$^{\wedge_{#1}}$}}
\newcommand{\scircumtwo}[1]{\mbox{$\Uparrow_{#1}$}}
\newcommand{\swprodtwo}[1]{\mbox{$\circledcirc{_{#1}}$}}
\newcommand{\sinfixtwo}[1]{\mbox{$\Downarrow_{#1}$}}
\newcommand{\Tp}{\mbox{\bf Tp}}
\newcommand{\Config}{\mbox{\bf Config}}
\newcommand{\Tterm}{\mbox{\bf TreeTerm}}
\newcommand{\Zone}{\mbox{\bf Zone}}
\newcommand{\Stoup}{\mbox{\bf Stoup}}
\newcommand{\smwwrap}[1]{\mbox{$\,||{_{#1}}\,$}}
\newcommand{\circumtwo}{\mbox{$\Uparrow$}}
\newcommand{\infixtwo}{\mbox{$\Downarrow$}}
\newcommand{\wprodtwo}{\mbox{$\circledcirc$}}
\newcommand{\semcnst}[1]{\mbox{\it #1}}
\newcommand{\unacc}{\mbox{\begin{picture}(0,0)\put(-5,-2.5){*}\end{picture}}}
\newcommand{\diff}{\mbox{$-$}}
\newcommand{\sextract}[1]{\mbox{$\uparrow_{#1}$}}

\newcommand{\sextracttwo}[1]{\mbox{$\Uparrow_{#1}$}}
\newcommand{\sdprod}[1]{\mbox{$\odot{_{#1}}$}}
\newcommand{\sdprodtwo}[1]{\mbox{$\circledcirc{_{#1}}$}}
\newcommand{\funextract}[1]{\mbox{$
		\begin{picture}(6,7)(0,-2)\put(1.5,-3){\rotatebox{90}{$\multimapdot$}}
		\end{picture}_{#1}$}}
\newcommand{\argextract}[1]{\mbox{$
		\begin{picture}(6,7)(0,-2)\put(1.5,-3){\rotatebox{90}{$\multimap$}}
		\end{picture}_{#1}$}}
\newcommand{\funextracttwo}[1]{\mbox{$
		\begin{picture}(6,7)(0,-2)\put(1.5,-3){\rotatebox{90}{\mbox{$=\!\!\!\bullet$}}}
		\end{picture}_{#1}$}}
\newcommand{\argextracttwo}[1]{\mbox{$
		\begin{picture}(6,7)(0,-2)\put(1.5,-3){\rotatebox{90}{$=\!\!\!\circ$}}
		\end{picture}_{#1}$}}
\newcommand{\arginfixtwo}[1]{\mbox{$
		\begin{picture}(6,8)(0,-2)\put(0.7,7){\rotatebox{-90}{\mbox{$=\!\!\!\circ$}}}
		\end{picture}_{#1}$}}
\newcommand{\funinfixtwo}[1]{\mbox{$
		\begin{picture}(6,8)(0,-2)\put(0.7,7){\rotatebox{-90}{\mbox{$=\!\!\!\bullet$}}}
		\end{picture}_{#1}$}}
\newcommand{\upperiwprodtwo}[1]{\mbox{\begin{picture}(7,7)(0,0)\put(0,-3){$\stackrel{\rotatebox{-90}{$\LEFTcircle$}}{=}$}\end{picture}$_{#1}$}}
\newcommand{\loweriwprodtwo}[1]{\mbox{\begin{picture}(7,7)(0,0)\put(0,-3){$\stackrel{\rotatebox{-90}{$\RIGHTcircle$}}{=}$}\end{picture}$_{#1}$}}
\newcommand{\ssplittwo}[1]{\mbox{$\check{\check{\ }}^{_{#1}}$}}
\newcommand{\sbridgetwo}[1]{\mbox{$\hat{\hat{\ }}^{_{#1}}$}}
\newcommand{\casearrow}{\mbox{$-\!\!>$}}

\newcommand{\AL}{\mbox{\bf L}}
\newcommand{\DA}{\mbox{\bf DA}}
\newcommand{\DAf}{\mbox{$\mathbf{DA_{foc}}$}} 
\newcommand{\DAF}{\mbox{$\mathbf{DA_{Foc}}$}}
\newcommand{\mybot}{\mbox{$\bot$}}
\newcommand{\mytop}{\mbox{$\top$}}
\newcommand{\topbot}{\mbox{$*$}}
\newcommand{\posex}{\mbox{$X^+$}}
\newcommand{\negex}{\mbox{$X^-$}}
\newcommand{\PrimTypes}{\mbox{$\cal P$}}
\newcommand{\mle}{\mbox{\ $\le$}\ }
\newcommand{\mge}{\mbox{\ $\ge$}\ }

\newtheorem{theorem}{Theorem}[section]

\newcommand{\ih}{\mbox{i.h.}}
\newcommand{\id}{\mbox{\it id}}
\newcommand{\foreign}[1]{\mbox{\it #1 \/}}

\newcommand{\LC}{\mbox{\bf L}}
\newcommand{\DC}{\mbox{\bf D}}
\newcommand{\DCA}{\mbox{\bf DA}}
\newcommand{\EDL}{\mbox{\bf EDL}}
\newcommand{\FDL}{\mbox{\bf FDL}}
\newcommand{\FDLplusEx}{\mbox{\bf FDLb}+}
\newcommand{\FDLb}{\mbox{\bf FDLb}}
\newcommand{\FDLbplusEx}{\mbox{\bf FDLb}+}
\newcommand{\FDLbplusExplusSi}{\mbox{\bf FDLb}++}

\newcommand{\GDLminusLmminusNd}{\mbox{\bf GDL}--}

\newcommand{\GDLminusLm}{\mbox{\bf GDL}-}

\newcommand{\GDL}{\mbox{\bf GDL}}
\newcommand{\GDLprime}{\mbox{\bf GDL$'$}}

\begin{document}

\maketitle

\noindent
\foreign{Caveat lector}\\

\noindent
\copyright~Glyn V.~Morrill \& Oriol Valent\'{\i}n 2026

\nocite{fgea:22}
\nocite{abrusci:comparison}

\newpage

\noindent
\verbquote{This study deals with syntactic structure both in the broad sense (as opposed to semantics) and the narrow sense (as opposed to phonemics and morphology). It forms part of an attempt to construct a formalized general theory of linguistic structure and to explore the foundations of such a theory. The search for rigorous formulation in linguistics has a much more serious motivation than mere concern for logical niceties or the desire to purify well-established methods of linguistic analysis. Precisely constructed models for linguistic structure can play an important role, both negative and positive, in the process of discovery itself. By pushing a precise but inadequate formulation to an unacceptable conclusion, we can
often expose the exact source of this inadequacy and, consequently, gain a deeper understanding of the linguistic data. More positively, a formalized theory may automatically provide solutions for many problems other than those for which it was explicitly designed. Obscure and intuition-bound notions can neither lead to absurd conclusions nor provide new and correct ones, and hence they fail to be useful in two important respects. I think that some of those
linguists who have questioned the value of precise and technical development of linguistic theory may have failed to recognize the productive potential in the method of rigorously stating a proposed theory and applying it strictly to linguistic material with no attempt to avoid unacceptable conclusions by ad hoc adjustments or loose formulation. The results reported below were obtained by a conscious attempt to follow this course systematically. Since this fact may be obscured by the informality of the presentation, it is important to emphasize it here.}\\

\hfill Noam Chomsky (1957) {\it Syntactic Structures},
Preface\nocite{chomsky:synstruct}\\\ \\

\noindent
\verbquote{I reject the contention that an important theoretical difference
exists between formal and natural languages. On the other hand,
I do not regard as successful the formal treatments of natural
languages attempted by certain contemporary linguists. Like
Donald Davidson[$^1$] I regard the construction of a theory of
truth---or rather, of the more general notion of truth under an
arbitrary interpretation---as the basic goal of serious syntax
and semantics; and the developments emanating from the
Massachusetts Institute of Technology offer little promise towards
that end.}\\

\hfill Richard Montague (1970) \verbquote{English as a Formal Language}, 
opening paragraph\nocite{montague:efl}

\newpage

\section*{Preface}

\hfill{\it The book of nature is written in the language of mathematics.}\\

\hfill{Galileo\ Galilei}\\

\noindent
We agree with Montague that the construction of a theory of truth
under an interpretation is the basic goal of syntax and semantics, and 
we think that
Chomsky has not adhered to the formal methodology
so eloquently advocated in his Preface to {\em Syntactic Structures}.
But we disagree with Montague that there is no important difference
between formal and natural languages: the latter have
sentimental qualities which the former do not.
This book on syntax and semantics extends
to logical syntax and formal grammar
Montague's original program of logical semantics and extends to logical
semantics and formal grammar Chomsky's original
program of formal syntax.  

Logic,
Computation,
and Linguistics form a triangle of disciplines of informatics.
Logic and theoretical computer science,
ever since their modern conception,
about a century ago,
have formed branches of mathematics.
But linguistics,
until now,
has had a poorer relation with mathematics;
rather,
it has struggled to be even formal. There has been
some aspiration to rigour,
but the result has fallen short of generating structures which are of interest mathematically:
there has lacked this sign that the science has come of age.

Chomsky (1957\cite{chomsky:synstruct}) made an articulate plea for
formal syntax.
He later reneged on this methodology,
for example in his Mimimalist Program
(Chomsky 1996\cite{chomsky:minimalism}),
with
desparaging remarks about formalisation for formalisation's sake
and formalisation being premature,
but the seeds of formal grammar had been planted.
Montague (1973\cite{montague:ptq}) inaugurated formal semantics,
and formal syntax and semantics enjoyed a golden era
in the 1980s and 1990s.
At that time there was a strong emphasis on {interdisciplinary}
informatics at institutions like the Centre for Cognitive Science in Edinburgh
and the ILLC in Amsterdam
and in projects such as the DYANA and DYANA-2 European Basic
Research Actions. Our goal now is to promote \emph{mathematical\/} linguistics.
Our thesis is that what we present here is not just \emph{interdisciplinary\/},
but is promising of a \emph{paradigm shift}.

In 1992 one of us (G.V.M.) asked Noam Chomsky what he thought of Montague Grammar.
He replied that Montague believed grammar was a branch of mathematics and that
mathematicians were interested in theorems,
and he asked, rhetorically, whether there were any interesting theorems in Montague
Grammar; of course there were none.

Logical syntax and semantics, however, \emph{has\/} generated interesting theorems,
for example the proof by Pentus (1992\cite{pentus:lccf}) of context-free equivalence
of Lambek calculus or, we suggest, Valent\'{\i}n's (2012\cite{valentin:phd}) analysis
of Cut-elimination for displacement calculus.
On our view, from a steep climb of formalisation,
this branch of linguistics has hauled itself up onto the plateau of mathematics:\\

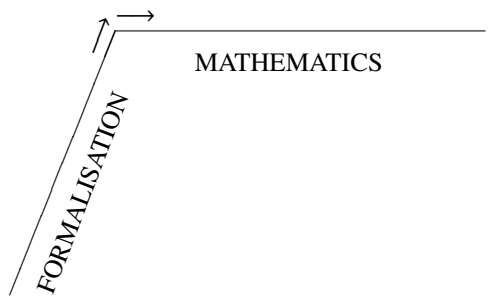
\begin{center}
\begin{picture}(100, 100)(50, 0)
\put(0, 0){\line(2,5){40}}
\put(40,100){\line(1,0){140}}
\put(10,0){\rotatebox{-291}{\mbox{FORMALISATION}}}
\put(70, 85){\mbox{MATHEMATICS}}
\put(30, 90){\rotatebox{-291}{\mbox{$\longrightarrow$}}}
\put(40, 103){\mbox{$\longrightarrow$}}
\end{picture}
\end{center}

\newpage

\noindent
Thus we feel that recent years tend towards a paradigm shift whereby (categorial)
linguistics becomes a fully-fledged branch of mathematics in the informatics
triangle:\\

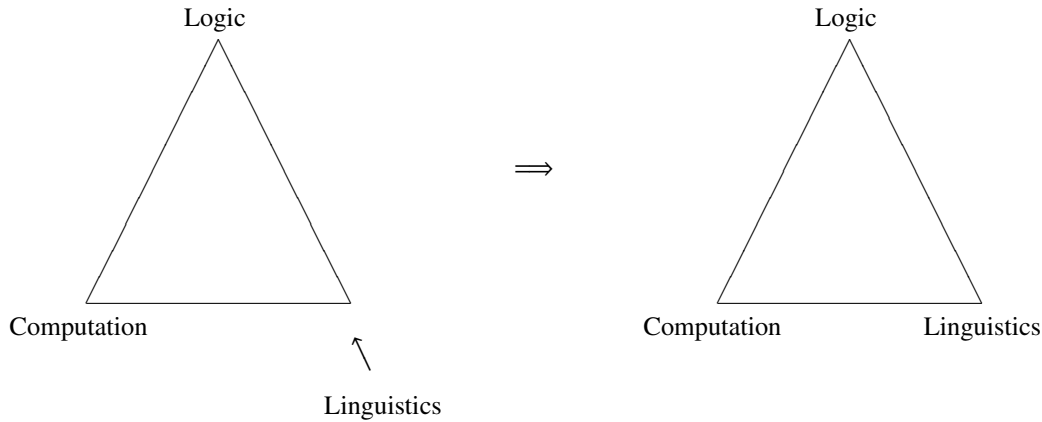
\begin{center}
\begin{picture}(100, 100)(0, 0)
\put(0, 0){\line(1, 0){100}}
\put(0, 0){\line(1,2){50}}
\put(50,100){\line(1,-2){50}}
\put(-29,-12){\mbox{Computation}}
\put(37, 105){\mbox{Logic}}
\put(90,-42){\mbox{Linguistics}}
\put(100, -25){\rotatebox{-245}{\mbox{$\longrightarrow$}}}
\end{picture}
$\tb\tab\begin{array}{c}\Longrightarrow\\\\\\\\\\\\\\\\\\\end{array}\tb\tab$
\begin{picture}(100, 100)(0, 0)
\put(0, 0){\line(1, 0){100}}
\put(0, 0){\line(1,2){50}}
\put(50,100){\line(1,-2){50}}
\put(-28,-12){\mbox{Computation}}
\put(37, 105){\mbox{Logic}}
\put(78,-12){\mbox{Linguistics}}
\end{picture}
\end{center}

\noindent
If we are right, linguistics is becoming mathematics categorially.
The present work pursues and develops this tendency. 
The book is divided into five parts. 
Part~I, SYNTAX, presents a fragment of categorial logic
with algebraic semantics,
explaining how it forms a type logic of syntax.
Part~II, SEMANTICS, presents this fragment of categorial logic with the so-called
Curry-Howard formulas-as-types and proofs-as-programs semantic labelling,
explaining how the type logic of syntax is also a logic of semantics.
Part~III, PROCESSING, explains focusing and count-invariance for our
categorial logic, these forming the basis of the efficient computer implementation
CatLog3 of parsing/theorem proving for the categorial logic.
Part~IV, GRAMMAR, gives examples of grammatical analysis in the framework,
with derivations produced by CatLog3.
Part~V, CONCLUSION, looks into future prospects.

Chapter~\ref{semCut} of Part~I is based on the communications of Valent\'{\i}n 
{\it Algebraic Semantics for Full Displacement Calculus
  with Linguistic Subexponentials and Bracket Modalities\/},
  Logical Aspects of Computational Linguistics LACL21 (Montpellier 2021), and Valent\'{\i}n
  {\it Algebraic Completeness and Semantic Cut-Elimination of a Multimodal Formulation of Morrrill's Categorial Logic\/}, NCL'24 Non-Classical Logics: Theory and Applications ({\L}\'od\'z 2024).
Chapters~\ref{focchap} and~\ref{countchap} of Part~III are based on Morrill and~Valent\'{\i}n (2018\cite{mv:spurambfoc}) {\it Spurious Ambiguity and Focalization\/}
and~Kuznetsov, Morrill and~Valent\'{\i}n
(2017\cite{mol16count}) {\it Count-Invariance Including Exponentials\/} respectively.
Chapters~\ref{monfrag}, \ref{mvfchap}, and~\ref{relchap} of Part~IV
are based on Morrill 
and~Valent\'{\i}n (2016\cite{mv:paris}) {\it Computational Coverage of Type Logical Grammar: The Montague Test}, Morrill, Valent\'{\i}n and~Fadda 
(2011\cite{mvf:tdc}) {\it The Displacement Calculus},
Chapter~\ref{relchap} is based on Morrill (2017\cite{gl:rel}) {\it Grammar Logicised: Relativisation}. And
Chapter~\ref{concchap} of Part~V is partially based on Morrill and Valent\'{\i}n (2017[\cite{mv:nllt}) {\it A Reply to Kubota and Levine on Gapping}.
We thank the publishers for permission to reuse this material;
the rest of the material is new expression of the literature and the
folklaw of the field.\\

\hfill{
\begin{tabular}[t]{r}
G.V.M.~\&~O.V.G.\\
Barcelona\\
2026
\end{tabular}}

\begin{figure}
\begin{center}
\section*{Primitive Categorial Connectives}\ \\\ \\\ \\\ \\
$
\xymatrix{
\mbox{Multiplicatives}   & / && \bsl &   \mbox{Lambek (1958\cite{lambek:mathematics}; 1988\cite{lambek:88})}\\
 & &  \product \\
    &   & I & & \\
 &  \scircum{} &  & \sinfix{}  &  \mbox{Morrill, Valent\'{\i}n \& Fadda (2011\cite{mvf:tdc})}\\
  & & \swprod{}\\
   &        & J & &\\
\mbox{Additives} & \aconj & & \adisj & \mbox{Lambek (1961\cite{lambek:61}), Girard (1987\cite{girard:87})}\\
\mbox{Quantifiers} & \bigwedge & & \bigvee & \mbox{Morrill (1994\cite{morrill:tlg})}\\
\mbox{Normal modalities} & \mymod & & \emod & \mbox{Morrill (1990\cite{morrill:ib90}; 1994\cite{morrill:tlg}), Moortgat (1997\cite{moortgat:handbook})}\\
\mbox{Bracket modalities} & \abrack & & \mybrack & \mbox{Morrill (1992\cite{morrill:92}), Moortgat (1995\cite{moortgat:95})}\\
\mbox{(Sub)exponentials} & \univexp & & \exstexp & \mbox{Girard (1987\cite{girard:87})}\\
\mbox{Limited contr. \& limited expan.} &  | & & + & \mbox{J\"ager (2005\cite{jaeger:2005}); 
Morrill \& Valent\'{\i}n (2014\cite{mv:words}})\\
\mbox{Difference} & & - & & \mbox{Morrill \& Valent\'{\i}n (2014\cite{mv:jcss13})}
}
$
\end{center}
\end{figure}

\clearpage

\begin{center}
\section*{Systems of Generalised Displacement Logic \GDL}
\hspace*{2,5cm}\mbox
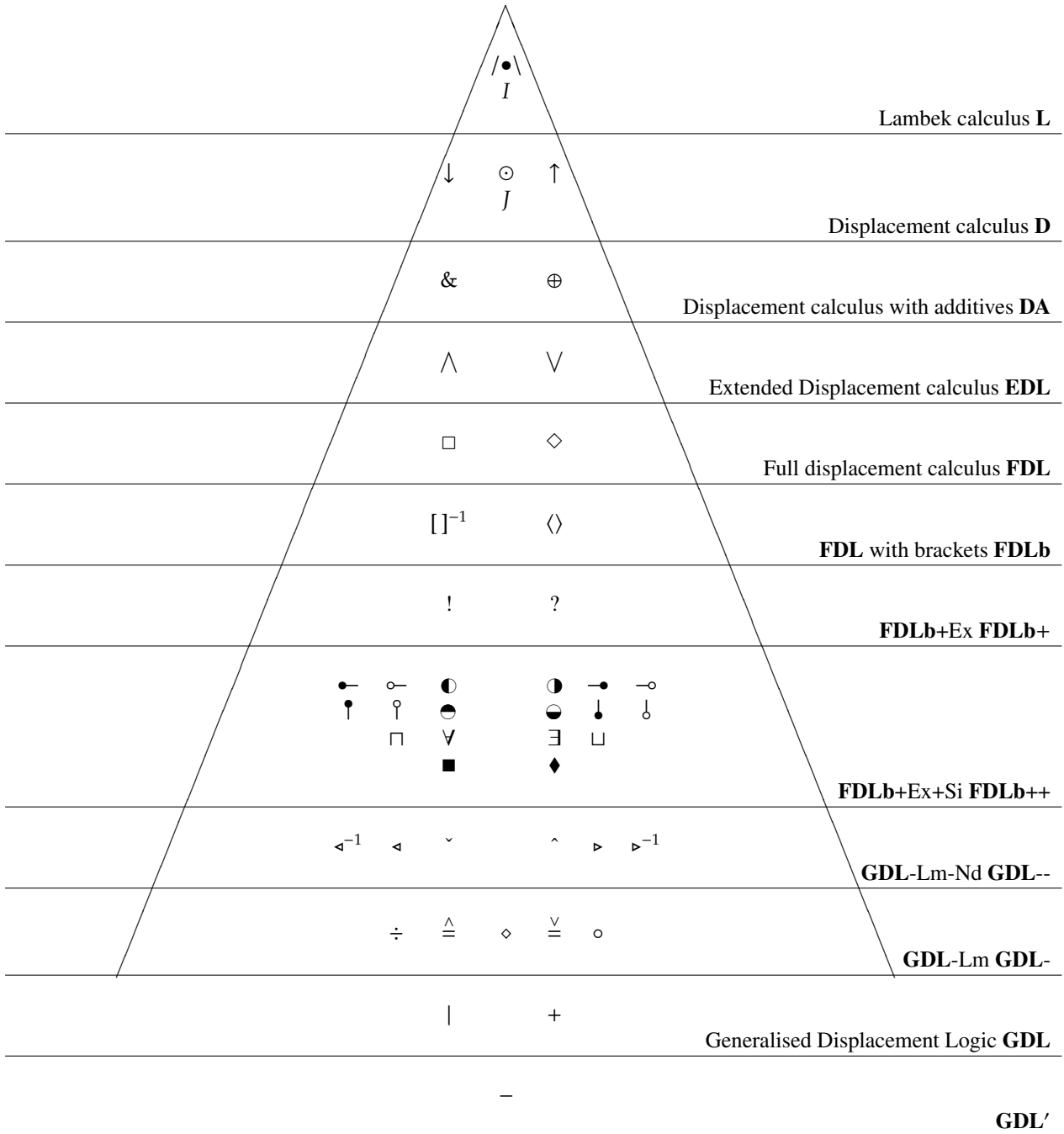
{
\begin{picture}(350,525)(-80,-80)
\put(-130,0){\line(2,5){175}}
\put(220,0){\line(-2,5){175}}
\put(55, 190){\makebox(5,00){$
\begin{array}[t]{lccccccccr}\\\\\\
\hspace*{1in}&&&&& /\product\bsl \\
&&&&& I && &&\\
&&&&&&&&&\mbox{Lambek calculus }\LC\\
\hline\\
&&&&\infix{} & \wprod{} & \circum{}\\
&&&&& J &&&& \\
&&&&&&&&&\mbox{Displacement calculus }\DC\\
\hline\\
&&&& \aconj && \adisj\\
&&&&&&&&&\mbox{Displacement calculus with additives }\DCA\\
\hline\\
&& &&\bigwedge && \bigvee &\\
&&&&&&&&&\mbox{Extended Displacement calculus }\EDL\\
\hline\\
& &&& \mymod && \emod\\
&&&&&&&&&\mbox{Full displacement calculus }\FDL\\
\hline\\
& &&& \abrack && \mybrack && \\
& &&&  &  && &&\FDL\mbox{ with brackets }\FDLb\\
\hline\\
&& && \univexp  && \exstexp&&&\\
& &  &   &   & &&&& \mbox{\FDLb+Ex} ~\FDLplusEx\\
\hline\\
&&\argover{} & \funover{} & \leftiprod{} &&  \rightiprod{} & \argunder{} & \fununder{} \\
&&\funcircum{}  & \argcircum{} & \upperiwprod{} && \loweriwprod{} & \funinfix{} & \arginfix{}  \\
&&&\iaconj & \forall &  & \exists & \iadisj&\\
&&&& \imod && \iemod\\
&&&&&&&&&\mbox{\FDLb+Ex+Si \FDLb++}\\
\hline\\
&&\leftproj & \leftinj & \mysplit{} && \bridge{} & \rightinj & \rightproj\\
&&&&&&&&&\mbox{\GDLminusLm{Lm}-Nd \GDL-{}-}\\
\hline\\
\hspace*{1.7in}&&&\ndiv & \nextract & \ndprod & \ninfix & \nprod  \\
&&&&&&&&&\mbox{\GDLminusLm{Lm} \GDLminusLm}\\
\hline\\
&&&& | && +\\
&&&&&&&&&\mbox{Generalised Displacement Logic }\GDL\\
\hline\\
&&&&&-&&\\
&&&&&&&&&\GDLprime\\
\hline
\end{array}$}}
\end{picture}}
\end{center}

\tableofcontents

\clearpage

\begin{figure}
\begin{center}
\section*{Table of Numbered Categorial Connectives}

\ 

\rotatebox{-0}{
\scriptsize
$
\begin{array}{r|ccc|ccc|c|c|c|c||c|c|c|cl}
& \multicolumn{3}{c|}{\begin{tabular}{c}cont.\\mult.\end{tabular}}&
\multicolumn{3}{c|}{\begin{tabular}{c}disc.\\ mult.\end{tabular}}  &
\mathrm{add.} & \mathrm{qu.} & \begin{tabular}{c}norm.\\ mod.\end{tabular} & 
\begin{tabular}{c}brack.\\ mod.\end{tabular} & \mathrm{subexp.} &
 &\multicolumn{1}{c|}{\begin{tabular}{c}limited\\contr.\\\& expan.\end{tabular}}
\\
 \cline{1-12}\cline{14-14}
&&&&&&&&&&&&&\\
&/_{1} & & \bsl_{2} & \scircum{}_{5} & & \sinfix{}_{6} & \aconj_{9} & \bigwedge_{11} & \mymod_{13} & \abrack_{15} & \univexp_{17} &&|_{48}\\
\mathrm{primitive} & & \product_{3}  & & & \swprod{}_{7} &&&&&&& &   \\
& & I_{4} &&& J_{8}&& \adisj_{10} & \bigvee_{12} & \emod_{14} & \mybrack_{16} & \exstexp_{{18}} &&+_{49}\\
&&&&&&&&&&&&&\\
 \cline{1-12}\cline{14-14}
  &&&&&&&&&\\
 \mathrm{sem.} &\argover_{19}\ \fununder_{20}&&\funover_{22}\ \argunder_{23}&\funcircum{}_{25}\ \arginfix{}_{26} &&\argcircum{}_{28}\ \funinfix{}_{29}& \iaconj_{31} & \forall_{33} & \imod_{35} 
 \\
 \mathrm{inactive}&&&&&&&&&\\
  \mathrm{variants}&\leftiprod{}_{21}&&\rightiprod{}_{24}&\upperiwprod{}_{27}&&\loweriwprod{}_{30}&\iadisj_{32}&\exists_{34}&\iemod_{36}&\multicolumn{1}{c}{}
 \\
 &&&&&&&&&\\
 \cline{1-10}
 & &&&&&\\
\mathrm{det.}&\leftproj_{37} && \rightproj_{39} && \ssplit{}_{41} &\\
&&&&&&&\multicolumn{6}{c|}{}&&& \mbox{}\\
\mathrm{synth.}&\leftinj_{38} && \rightinj_{40} && \sbridge{42} &&\multicolumn{6}{c|}{}&\mbox{diff.}\\
&&&&&&&\multicolumn{6}{c|}{}&&& \mbox{}\\
 \cline{1-7}\cline{14-14}
 &&&&&&&\multicolumn{1}{l}{}&\multicolumn{5}{c|}{}&\\
\mathrm{non-det.} &  & \ndiv_{43} && \nextract_{45} && \ninfix_{46}&\multicolumn{6}{c|}{}&\\
 &&&&&&& \multicolumn{1}{l}{}&\multicolumn{5}{c|}{}&-_{50}\\
\mathrm{synth.}  && \nprod_{44} &&& \ndprod_{47} &&\multicolumn{6}{c|}{}& \\
 &&&&&&&\multicolumn{6}{c|}{}&\\
 \cline{1-7}\cline{14-14}
\end{array}$}
\end{center}
\end{figure}

\part{SYNTAX}

\noindent
In this part we present the syntactic calculus of our logical grammar.
In Chapter~\ref{introchap} we comment on the conceptual and historical contexts
of syntactic types and syntactic calculus: Lambek calculus $\LC$
and displacement calculus $\DC$.
In Chapter~\ref{CL-primtypeschap} we define the types of the latter,
establishing the core connectives for the rest of the book,
and we define the syntactical models (phase semantics) for $\DC$.
In Chapter~\ref{seqcalcchap} we present the sequent calculus for
$DC$, note soundness, and illustrate linguistically.
We go on to define further connectives, syntactical models, 
(sound) sequent calculus, and linguistic illustration
in Chapter~\ref{CL-remtypeschap}.
In Chapter~\ref{semCut} we present completeness results, which have as a corollary semantic Cut-elimination.

\clearpage

\chapter{Introduction}

\label{introchap}

\section{Logical grammar}

`Logic' is synonymous with the organisation of information
and verification, and forms a basis of management
of the large amounts of information making up natural language grammar.
In the course
of the 20th century the term has acquired a theoretical and 
technical refinement of the colloquial use. The relevance of
such mathematical logic to the logical semantics of natural language
seems clear enough when the logic is the semantic logic of the object natural language itself. However, 
why should \emph{syntax\/}
be logical?

A grammar characterises
as a subset of a set of expressions those which are grammatical.
A logic characterises as a subset of a set of statements those which
are logical. Equate grammar with logic, and grammatical derivations
are proofs.
Standard logic deals with resources which are mental: 
ideas,
which are persistent; they have no weight and are timeless,
and such logic is appropriate for logical semantics, the mental semantics
of the object natural language.
For logical syntax we require a logic of resources which are physical and temporal:
things,
which are material; that have multiplicity and location in time, the physical prosodics
of the object natural language.
Once we accept the possibility of such resource-conscious logic,
a logic of mental ideas can be complemented by a logic of physical things
in grammar, facilitating the description of the vast mind/body association that is language.

If this explains how grammar \emph{could\/} be logical,
can we say something about why it \emph{should}? 
The following two views were expressed to one of us (G.V.M.) by
Andre Scedrov and Stefan Kuznetsov, p.c.
Firstly, if we model grammar as logic \emph{and\/} prove
Cut-elimination then we have not just modelled grammar as a theory
in a logic technically but have also identified the logic of the empirical domain
modelled, grammar, itself in reality (A.~Scedrov).
Secondly, we think that logic holds out the prospect of \emph{verification\/} of grammar; imagine that natural language processing was life-critical
(S.~Kuznetsov).

\section{Syntactic calculus}

The syntactic and semantic structures of logic are generally
well-nested. But from the inception of modern linguistics
it has been emphasized that the central question in grammar is
how to treat the pervasive syntactic/semantic mismatch in natural
language, a question to which Chomsky proposed the answer of
transformations. We think that although this is the
right question, transformations are not the right answer.
Rather we propose to treat syntactic/semantic `disorder' by logical
syntax and semantics based on displacement calculus, a discontinuous
sublinear intuitionistic logic in which `displacement' emerges from an underlyingly
well-nested structure. Thus apparent disorder is, eventually,
order again.


Lambek (1958\cite{lambek:mathematics}) defined a calculus of syntactic types.
This {\it Lambek calculus\/}
is free of structural rules, and Lambek proved (constructively) that it enjoys the Cut-elimination theorem,
i.e.\ that every theorem has a Cut-free proof.
This Cut-Elimination property has as corollaries the subformula property (that every
theorem has a proof containing only its subformulas ---
in this case any one of its Cut-free proofs),
the finite proof property (that every theorem has only a finite number of Cut-free
proofs, since the Cut-free proof search-space is finite) and decidability (that whether or not a statement
is provable can be determined by an effective procedure ---
for example Cut-free backward chaining proof search).

Retrospectively the Lambek calculus can be recognised as the multiplicative fragment
of intuitionistic non-commutative linear logic
(Girard 1987\cite{girard:87};
Abrusci 1990\cite{abrusci:90},
1991\cite{abrusci:91})
without empty antecedents.
The Lambek calculus was predated by Ajdukiewicz (1935\cite{ajdukiewicz}) and
Bar-Hillel (1953\cite{bar-hillel:quasi}) but was developed independently of these;
Bar-Hillel, Gaifman and~Shamir (1961\cite{bar-hilleletal}) introduced
the term `categorial grammar' to refer to such systems of {\it logical syntax}.

Montague (1973\cite{montague:ptq}) defined a {\it logical semantics\/}
for a significant formal fragment of English,
a remarkable achievement at a time when it was widely believed
that semantics was beyond the reaches of formalisation.
As a demonstration of the power of our approach we have given a
computational cover grammar of the Montague fragment in Chapter~\ref{monfrag}, 
and we
propose that doing so be adopted as a standard criterion or
`Montague test' in computational syntax and semantics.

The Lambek calculus remained largely unknown until its rediscovery
in the 1980s when it was observed that there is a perfect matching of
the logical syntax of Lambek and the logical semantics of Montague;
see van Benthem (1983\cite{benthem:variety}).
(In this context the finite proof property has a further corollary: the finite
reading property whereby every expression is at most 
finitely ambiguous, a finiteness which appears to be an empirical universal of
the natural languages of the world; it follows
from semantic Cut-elimination, Hendriks (1993\cite{hendriksH:phd}).)
Monograph and reference article studies developing this {\it logical grammar\/}
include Moortgat (1988\cite{moortgat:phd}; 1997\cite{moortgat:handbook}), 
Morrill (1994\cite{morrill:tlg}; 2011\cite{morrill:oxford}; 2012\cite{morrill:loggram}),
Carpenter (1997\cite{carpenter:96}),
J\"ager (2005\cite{jaeger:2005}) and Moot and Retor\'e (2012\cite{mootret}).

Probably one reason why Lambek calculus $\LC$ was largely ignored for so long
was the suspicion that it was context-free in generative
power together with the belief, deriving from Chomsky (1957\cite{chomsky:synstruct}), that context-free grammar is observationally
inadequate for natural language.
And indeed, Shieber (1985\cite{shieber:85}) proves that Swiss-German
is non-context free, and Pentus (1993\cite{pentus:lccf}) proves that
the Lambek calculus generates exactly the context free languages
without the empty string.
Thus although the categorial syntax/semantics interface is philosophically
and technically impeccable, 
there remained a justified doubt
about it's linguistic adequacy.
Morrill, Valent\'{\i}n and Fadda (2011\cite{mvf:tdc}) address this concern
by defining a conservative extension of the Lambek calculus which
allows displacement while remaining free of structural rules
and preserving the syntax/semantics interface and Cut-elimination of the Lambek calculus, and all of its metatheoretic 
corollaries: the subformula property, the finite proof property, decidability,
and the finite reading property.
The displacement calculus characterises, for example, both covert movement such as quantifier
scoping and overt movement such as cross-serial dependencies.
This book develops logical grammar built over the displacement calculus $\DC$.
Linguistics motivates additional sensitivity and expressivity.
 
 The \emph{displacement calculus with additives\/},
 $\DA$,
 is the displacement calculus plus the
 additives (\aconj, \adisj) 9--10,
 which have application to polymorphism.
 \emph{Extended displacement calculus},
 $\EDL$,
 is $\DA$ extended with 11--12:
 1st order quantifiers ($\bigwedge$, $\bigvee$),
 which have application to features.
 {\em Full displacement calculus} $\FDL$ is Extended displacement calculus $\EDL$ plus normal modalities.
 {\em Full displacement calculus with brackets\/} $\FDLb$ is Full displacement calculus $\FDL$ plus brackets;
 this also requires a change in the notion of sequent, to include brackets in antecedents.
 
 \techterm{Full displacement calculus with brackets and soft subexponentials\/}
 \mbox{$\FDLbplusEx$} is Full displacement calculus with brackets $\FDLb$ plus soft subexponentials (\univexp, \exstexp) (term of
 Kanovich et al.~(2018\cite{2018subexp}).
 
 \techterm{Full displacement calculus with brackets and soft subexponentials and semantically inactive variants}\\
 $\FDLbplusExplusSi$ is Full displacement calculus with brackets and soft subexponentials $\FDLplusEx$ plus
the semantically inactive variants 19--47.
We call the connectives of $\FDLbplusExplusSi$ the \emph{negation-free primary\/} connectives.

\techterm{Generalised displacement logic minus limited contraction and expansion and minus non-determinism}\\
$\GDLminusLmminusNd$ is \techterm{Generalised displacement logic minus limited contraction and expansion\/}
$\FDLbplusExplusSi$
without non-determinism.
We call the connectives of $\GDLminusLmminusNd$ the \emph{negation-free secondary\/} connectives.
\techterm{Generalised displacement calculus without limited contraction and expansion} $\GDLminusLm$ 
is Generalised displacement calculus $\GDL$ without limited contraction and expansion plus
the \scare{synthetic}
\emph(defined\/) connectives .
We call these the tercary \emph{negation-free primitive\/} connectives.

\techterm{Generalised displacement calculus with difference} $\GDLprime$ is Generalised displacement calculus $\GDL$ is
plus the \scare{metalogical negation}~50 difference.
 
\chapter{The types of Displacement Logic {\bf DL}}

\label{CL-primtypeschap}

\section{Syntactic types of {\bf DL}}

The syntactic types $\{\Tp{}_i\}_{i\in{\cal N}}$ of our categorial logic are sorted
according to the number of points $i\in{\cal N}$ of discontinuity that their expressions contain. $i'=i+1$. 
Each \techterm{type predicate letter $P$} will have a sort and an arity which
are naturals, and a corresponding semantic type. Assuming to be already given ordinary (feature) terms 
interpreted in a domain $F$,
where $P$ is a type predicate letter of sort $i$, arity $n$, and semantic type $\sigma$
and  $t_1, \ldots, t_n$ are feature terms,
$Pt_1\ldots t_n$ is an atomic type of sort $i$ and of semantic type $\sigma$.
Compound types of Displacement Logic {\bf DL} are formed from these by connectives as indicated in Figure~\ref{DL-types},
and the structure preserving prosodic type map $s$ associates these with sorts. 
The sort $s(A)$ ($sA$) of a type $A$ is the $i\in{\cal N}$ such that $A\in\Tp{}_i$.
For example, if $s(N)=s(S)=0$, we have $s((S\scircum{1}N)\scircum{0}N)=
s((S\scircum{0}N)\scircum{1}N)=2$, and if $s(\VP)=1$, 
$s((J\bsl(N\bsl\VP))\scircum{0}N)=2$.

\begin{figure}[ht]
\small$$
\begin{array}{lrclrclll}
0. & \Tp{}_i & ::= & Pt_1\ldots t_n & s(A) & = & \sigma\\
1. & \Tp{}_i & ::= & \Tp{}_{i{+}j}/\Tp{}_j & s(C/B) & = & s(C)-s(B) & 
\mbox{over \cite{lambek:mathematics}}\\
2. & \Tp{}_j & ::= & \Tp{}_i\bsl\Tp{}_{i{+}j} & s(A\bsl C) & = & s(C)-s(A) & 
\mbox{over \cite{lambek:mathematics}}\\
3. & \Tp{}_{i{+}j} & ::= & \Tp{}_i\product\Tp{}_j & s(A\product B) & = & s(A)+s(B) 
&
\mbox{continuous product \cite{lambek:mathematics}}\\
4. & \Tp{}_0 & ::= & I  & s(I) & = & 0 
& \mbox{continuous unit \cite{lambek:88}}\\
5, k. & \Tp{}_{i'} & ::= & \Tp{}_{i{+}j}\scircum{k}\Tp{}_j, 0< k< i' & s(C\scircum{k} B) & = & s(C)'-s(B) 
& \mbox{circumfix \cite{mvf:tdc}}\\
6, k. & \Tp{}_j & ::= & \Tp{}_{i'}\sinfix{k}\Tp{}_{i{+}j}, 0< k< i' & s(A\sinfix{k} C) & = & s(C)'-s(A) 
& \mbox{infix \cite{mvf:tdc}}\\
7, k. & \Tp{}_{i{+}j} & ::= & \Tp{}_{i'}\swprod{k}\Tp{}_j, 0< k< i' & s(A\swprod{k} B) & = & s(A)+s(B)-1
& \mbox{discontinuous product \cite{mvf:tdc}}\\
8. & \Tp{}_1 & ::= & J & s(J) & = & 1 & 
\mbox{discontinuous unit \cite{mvf:tdc}}\\
9. & \Tp{}_i & ::= & \Tp{}_i\aconj\Tp{}_i & s(A\aconj B) & = & s(A)=s(B) 
& \mbox{additive conjunction \cite{lambek:61, morrill:galt}}\\
10. & \Tp{}_i & ::= & \Tp{}_i\adisj\Tp{}_i & s(A\adisj B) & = & s(A)=s(B) 
& \mbox{additive
disjunction  \cite{lambek:61, morrill:galt}}\\
11. & \Tp{}_i & ::= & \bigwedge V\Tp{}_i & s(\bigwedge vA) & = & s(A) 
& \mbox{1st order univ.\ qu.\ 
\cite{morrill:tlg}}\\
12. & \Tp{}_i & ::= & \bigvee V\Tp{}_i & s(\bigvee vA) & = & s(A) 
& \mbox{1st order exist.\ qu.\ 
\cite{morrill:tlg}}\\
13. & \Tp{}_i & ::= & \mymod\Tp{}_i & s(\mymod A) & = & s(A) 
& \mbox{univ.\ modality \cite{morrill:ib90}}\\
14. & \Tp{}_i & ::= & \emod\Tp{}_i & s(\emod A) & = & s(A)  & 
\mbox{exist.\ 
modality \cite{moortgat:handbook}}\\
15. & \Tp{}_i & ::= & \abrack{}\Tp{}_i & s(\abrack A) & = & s(A) 
& \mbox{univ.\ bracket modality 
\cite{morrill:92, moortgat:95}}\\
16. & \Tp{}_i & ::= & \mybrack{}\Tp{}_i & s(\mybrack A) & = & s(A) 
& \mbox{exist.\ bracket modality
\cite{morrill:92, moortgat:95}}\\
17. & \Tp{}_0 & ::= & \univexp\Tp{}_0 & s(\univexp A) & = & s(A)=0 
& \mbox{univ.\ subexponential
\cite{bhlm:91}}\\
18. & \Tp{}_0 & ::= & \exstexp\Tp{}_0 & s(\exstexp A) & = & s(A)=0
& \mbox{exist.\ subexponential
\cite{morrill:tlg}}\\
\end{array}
$$
\caption{The types of Displacement Logic {\bf DL}}
\label{DL-types}
\end{figure}

The family of types 1--4: $\{/, \bsl, \product, I\}$ are the Lambek connectives of
 Lambek 1958\cite{lambek:mathematics} and Lambek 1988\cite{lambek:88}. 
 They are defined in relation to concatenation, with left
 residual $\bsl$ (\scare{under}),
 right residual $/$ (\scare{over}), language concatenation $\product$ (\scare{product}) and language concatenation
 unit $I$ (\scare{product unit}). The structure $(\Tp, \bsl, \product, /, I; \subseteq)$ forms a residuated monoid.
 The calculus with only these connectives is the \emph{Lambek calculus (with unit)}, {\bf L}.
 
 The family of types 5--8: $\{\scircum{k}, \sinfix{k}, \sdprod{k}, J\}$ are the displacement connectives of
 Morrill 2011\cite{morrill:oxford},
 Mor\-rill and Valent\'{\i}n 2010\cite{mv:disp} and
 Morrill, Valent\'{\i}n and Fadda 2011\cite{mvf:tdc}. They are defined in relation to intercalation,
 i.e.\ the replacement by a second operand of a point of discontinuity (\scare{separator}) in a first operand;
 the subscript $k$ indexes which separator is replaced: 0~for the first from the left, 1~for the second from the left, 2 for the third from the left, as so on. 
 There is the left residual $\sinfix{k}$ (\scare{infix}), right residual $\scircum{k}$ (\scare{circumfix}), language intercalation $\sdprod{k}$ (\scare{discontinuous product})
 and language intercalation unit $J$ (\scare{discontinuous product unit}).
 The structure $(\Tp, \sinfix{k}, \sdprod{k}, \scircum{k}, J; \subseteq)$ also forms a (sorted) residuated monoid.
 The calculus with the connectives 1--8 is the \emph{displacement calculus}, {\bf D}.
 
 Displacement calculus {\bf D}, like Lambek calculus {\bf L}, is a \scare{multiplicative}
 system in the terminology of linear logic, meaning that resources
 (like words) have multiplicity
 of occurrence and do not preserve grammatical properties like well-formedness
 and meaning under either fission (contraction) or fusion (expansion),
 but must be perfectly balanced between production and consumption in reasoning.

\section{Syntactical models of {\bf DL}}

In standard logic information does not have multiplicity. Thus where $\add$
is the notion of addition of information and $\le$ is the notion of inclusion
of information we have \scare{sharing} $x\add x\le x$ and $x\le x\add x$; together these
two properties amount to \techterm{idempotency}: $x\add x=x$.
The properties are expressed by the rules of inference Contraction and
Expansion:
\disp{$
\prooftree
\Delta(A, A)\yields B
\justifies
\Delta(A)\yields B
\using \mbox{Contraction}
\endprooftree
\tb
\prooftree
\Delta(A)\yields B
\justifies
\Delta(A, A)\yields B
\using \mbox{Expansion}
\endprooftree
$}
Expansion is a special case of Weakening:
\disp{$
\prooftree
\Delta()\yields B
\justifies
\Delta(A)\yields B
\using \mbox{Weakening}
\endprooftree
$}
Linguistic resources do not have these properties: 
grammaticality is not generally preserved under addition or removal
of copies of words or expressions.
However, there are some constructions manifesting something
similiar.
Parasitic gaps allow a kind of controlled contraction:
\disp{\begin{tabular}[t]{ll}
a. & patient that$_i$ John convinced the friends of Mary to visit $t_i$\\
b. & patient that$_i$ John convinced the friends of $t_i$ to visit Mary\\
c. & patient that$_i$ John convinced the friends of $t_i$ to visit $t_i$\\
\end{tabular}}
And iterated coordination allows a kind of controlled expansion: 
\disp{
John, Fred, Bill, \ldots~and Mary.}
The left hand conjunct can be expanded indefinitely by like-type
conjuncts, preserving grammaticality.

Anaphora and expletive pronouns allow kinds of limited contraction
and limited expansion:

\disp{\begin{tabular}[t]{ll}
a. & Near him/Dan, Dan/he saw a snake.\\
b. & It rains.
\end{tabular}}
That is, in logical grammar idempotency, or sharing, is the exception
rather than the rule.

Idempotency introduces complexity in the syntactical model theory
of categorial grammar.
We include here syntactical interpretation for controlled and limited
contraction and
expansion.

\chapter{Hedge sequent calculus}

\label{seqcalcchap}

In this chapter we present Gentzen sequent calculus for Displacement Logic.
Gentzen sequent format is a kind of linguafranca of logic,
in which all rules are of the form $$\frac{\Sigma_1\ \ldots\ \Sigma_n}{\Sigma_0}$$
where the $\Sigma_i$ are statements of consequence called \techterm{sequents}.
Typically in (sub)linear logics,
once the {\em Cut-elimination\/} theorem is proved,
Gentzen sequent calculus yields a finite search space and hence a decision
procedure for theoremhood.\footnote{Other formats, such as natural deduction,
do not necessarily do the same, and usually adapting them to yield a decision procedure
amounts, in effect, to converting them to Gentzen sequent format.}
We employ a particular form of sequent calculus to deal with the discontinuity
in displacement logic which we will refer to as
{\em hedge\/} sequent calculus.
In addition,
the sequent calculus contains bracketing to deal with the bracket modalities (\scare{structural inhibition})
and \emph{stoups\/} (Girard 2011\cite{girard:blind}) to deal with subexponentials (\scare{structural facilitation}). 
A stoup is a reserved location containing types which can undergo structural rules
such as contraction. 

\section{Hedge sequent calculus}

The sets $\Zone$ of \techterm{zones},
$\Stoup$ of \techterm{stoups},
$\Config$ of \techterm{configurations\/} and
$\Tterm$ of \techterm{tree terms\/} of hedge sequent calculus {\bf hDL}
for Displacement Logic are defined by mutual recursion as follows,
where $\emptyset$ is the metalinguistic empty stoup,
$\Lambda$ is the metalinguistic empty configuration and {\sep} is a metalinguistic
placeholder called the \techterm{separator}:
\disp{$
\begin{array}[t]{rcl}
\Zone & ::= & \Stoup; \Config\\
\Stoup & ::= & \emptyset\ |\ \Tp{}_0, \Stoup\\
\Config{} & ::= & \Lambda\ |\ \Tterm, \Config\\
\Tterm & ::= & \sep\ |\ \Tp{}_0\ |\ 
\Tp{}_{{i>}0}\{\underbrace{\Config{}: \ldots: \Config{}}_{i\ \mbox{\scriptsize{\bf Config}'s}}\}
\ |\ [\Zone]
\end{array}$
\label{configdef}}
For example, there is the configuration $\gamma=
A\{B, \sep: C\{\sep\}, D\}, E, \sep$ where $s(A)=2$ and $s(C)=1$ and
$s(B)=s(D)=s(E)=0$. 
Diagramatically this configuration $\gamma$ is:
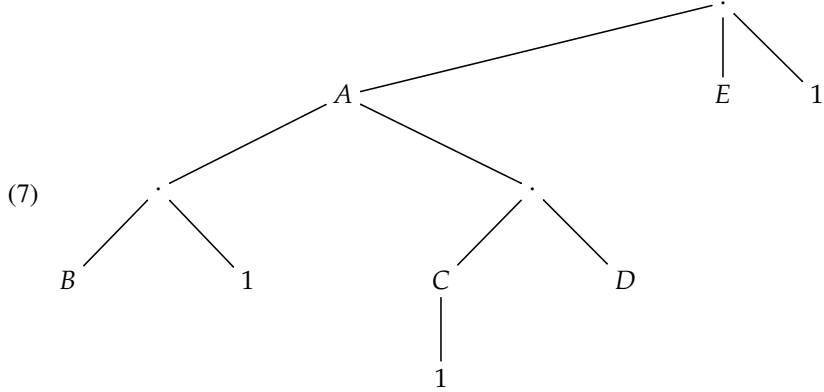
\disp{
$\diagram
&&&&&&&\cdot&\\
&&&A\xline[-1,4]&&&&E\uline&\sep\ulline\\
&\cdot\urrline&&&&\cdot\ullline&&&\\
B\urline&&\sep\ulline&&C\urline&&D\ulline&&\\
&&&& \sep\uline
\enddiagram$
}
The intuition is the following.
Dotted nodes signify unbounded arity concatenations and a type labelling a
mother node signifies a discontinuous type intercalated by its daughter configurations.
Leaf types are continuous, and a leaf $\sep$ marks a point of discontinuity.
The sort $s(\Gamma)$ of a configuration $\Gamma$ is the number of separators \sep{} that  $\Gamma$  contains.
For example, $s(\gamma)=3$.

Note that a stoup can only contain types of sort $0$; types of other sort would
not preserve sort-equality under contraction.\footnote{For sort $i$ a natural, $i=i+i$ only when $i=0$.}
The sort of a zone is the sort of its configuration. A \emph{sequent} is of the form:
\disp{
$\Zone\yields\Tp$ \tb where $s(\Zone) = s(\Tp)$}

The \techterm{figure} $\vect{A}$ of a type $A$ is defined by:
\disp{$
\vect{A} = \left\{
\begin{array}{ll}
A & \mbox{if\ } s(A)=0\\
A\{\underbrace{\sep: \ldots: \sep}_{sA\ 1\mathit{'s}}\} & \mbox{if\ } s(A)>0
\end{array}\right.$}
For example, $\vect{(S\scircum{1}N)\scircum{2}N} = (S\scircum{1}N)\scircum{2}N\{\sep:\sep\}$.
Where $\Gamma$ is a configuration of sort $i$ and $\Delta_1, \ldots, \Delta_i$
are configurations,
the \emph{fold} $\Gamma\otimes\langle\Delta_1: \ldots: \Delta_i\rangle$
is the result of replacing the successive \sep{}'s in $\Gamma$
by $\Delta_1, \ldots, \Delta_i$ respectively.
Thus for our running example, if $\gamma_1$, $\gamma_2$ and $\gamma_3$ are configurations, the fold
$\gamma\otimes\langle\gamma_1: \gamma_2: \gamma_3\rangle =
A\{C, \gamma_1: B\{\gamma_2\}, D\}, F, \gamma_3$.
Where $\Delta$ is a configuration of sort $i>0$ and $\Gamma$ is a configuration, the
$k$\/th metalinguistic wrap, $1\le k\le i$, $\Delta\smwrap{k}\Gamma$ is given by
\disp{$
\Delta\smwrap{k}\Gamma =_{df} \Delta\otimes\langle\underbrace{\sep: \ldots: \sep}_{k{-}1\ 1\mbox{\footnotesize 's}}: \Gamma: \underbrace{\sep: \ldots: \sep}_{i{-}k\ 1\mbox{\footnotesize 's}}\rangle$
}
i.e.\ $\Delta\smwrap{k}\Gamma$ is the configuration
resulting from replacing by $\Gamma$ the $k$\/th separator
in $\Delta$.
Thus for our running example, where $\gamma'$ is a configuration,
$\gamma\smwrap{2}\gamma' = A\{C, \sep: B\{\gamma'\}, D\}, F, \sep$.

$\Delta(\Gamma)$ signifies a configuration or zone $\Delta$ with a distinguished
subconfiguration or subzone $\Gamma$.
Where $\Gamma$ is a configuration of sort $i$, $\langle\Gamma\rangle$ signifies
$\Gamma\otimes\langle\Delta_1, \ldots, \Delta_n\rangle$; and
$\Delta\langle\Gamma\rangle$ signifies $\Delta_0(\langle\Gamma\rangle)$ i.e.\ 
$\Delta_0(\Gamma\otimes\langle \Delta_1, \ldots, \Delta_i\rangle)$,
i.e.\ a configuration $\Delta$ with a potentially
discontinuous distinguished subconfiguration $\Gamma$.

The hedge sequent calculus {\bf hDL} for the {\bf DL} of 
Chapter~\ref{CL-primtypeschap} 
has the following identity axiom and Cut rule; $\Xi$ denotes a zone; $\Gamma$ and $\Delta$, possibly with
subscripts, denote configurations; $\zeta$, possibly with subscripts, denotes a stoup. 
\disp{$\prooftree
\justifies
\emptyset\sass\vect{P}\yields P
\using \mathit{id}
\endprooftree \tab
\prooftree
\zeta_1\sass\Gamma\yields A \tb
\Xi(\zeta_2\sass \Delta_1, \langle \vect{A}\rangle, \Delta_2)\yields B
\justifies
\Xi(\zeta_1\mun\zeta_2\sass\Delta_1, \langle\Gamma\rangle, \Delta_2)\yields B
\using \mathit{Cut}
\endprooftree
$}
Note that the conclusion stoup is empty in the
axiomatic id rule and is partitioned in the binary Cut rule.
The identity axiom asserts the reflexivity of the derivability relation
for atomic types $P$. In general it will be the case that for all types $A$,
$\vect{A}\yields A$.
The Cut rule asserts the transitivity of the derivability relation,
which is a property that is usually desired. 
However Cut is problematic computationally because the Cut formula
$A$ is a new unknown reading from conclusion to premises.
A part of the art of Gentzen sequent calculus is to formulate rules
which are partially executed with respect to Cut in such a way that
all theorems have Cut-free proofs,
that is: to have the effect of Cut without using Cut. Gentzen called the proof
of such Cut-elimination his \emph{Haupsatz}.

\section{Rules for the primary connectives of {\bf hDL}}

The logical rules are listed in what follows. 
Linguistic applications are given.

\subsection{Multiplicatives}

The continuous multiplicatives  \{$/$, $\bsl$, $\product$, $I$\}
the Lambek connectives
of 
Lambek (1958\cite{lambek:mathematics};
1988\cite{lambek:88}),
are the basic means of categorial (sub)categorization.
Their Gentzen sequent rules are as in Figure~\ref{cmult}.

\begin{figure}[ht]
\begin{center}
$
\begin{array}[t]{lc}
1. &\prooftree
\zeta_1\sass \Gamma\yields B \tb
\Xi(\zeta_2\sass \Delta_1, \langle\vect{C}\rangle, \Delta_2)\yields D
\justifies
\Xi(\zeta_1\mun\zeta_2\sass \Delta_1, \langle\vect{C/B}, \Gamma\rangle, \Delta_2)\yields D
\using / L
\endprooftree \tb
\prooftree
\zeta\sass\Gamma, \vect{B}\yields C
\justifies
\zeta\sass \Gamma\yields C/B
\using / R
\endprooftree
\\\\
2. &\prooftree
\zeta_1\sass\Gamma\yields A \tb
\Xi(\zeta_2\sass\Delta_1, \langle\vect{C}\rangle, \Delta_2)\yields D
\justifies
\Xi(\zeta_1\mun\zeta_2\sass \Delta_1, \langle\Gamma, \vect{A\bsl C}\rangle, \Delta_2)\yields D
\using \bsl L
\endprooftree \tb
\prooftree
\zeta\sass \vect{A}, \Gamma\yields C
\justifies
\zeta\sass \Gamma\yields A\bsl C
\using \bsl R
\endprooftree
\\\\
3. & \prooftree
\Xi\langle\vect{A}, \vect{B}\rangle\yields D
\justifies
\Xi\langle\vect{A\product B}\rangle\yields D
\using \product L
\endprooftree \tb
\prooftree
\zeta_1\sass\Gamma_1\yields A\tb\zeta_2\sass\Gamma_2\yields B
\justifies
\zeta_1\mun\zeta_2\sass\Gamma_1, \Gamma_2\yields A\product B
\using \product R
\endprooftree
\\\\
4. &\prooftree
\Xi\langle\Lambda\rangle\yields A
\justifies
\Xi\langle\vect{I}\rangle\yields A
\using IL
\endprooftree\tb
\prooftree
\justifies
\emptyset\sass\Lambda\yields I
\using IR
\endprooftree
\end{array}
$
\end{center}
\caption{Continuous multiplicative rules}
\label{cmult}
\end{figure}
\noindent
Notice how in the multiplicative rules the conclusion stoup is partitioned between premises
in binary rules, copied to the premise in unary rules, and is empty in the axiomatic $IR$ rule.
When stoups are empty, both the empty stoup and \scare{$\sass$} may be omitted in derivations.

The directional divisions \verbquote{over}, $/$, and 
\verbquote{under}, $\bsl$, are exemplified
by assignments such as $\syncnst{the}\ass N/\CN$ for the noun phrase
$\syncnst{the man}\ass N$ and $\syncnst{sings}\ass N\bsl S$ for the sentence
$\syncnst{John sings}\ass S$
and $\syncnst{loves}\ass (N\bsl S)/N$ for the sentence $\syncnst{John loves Mary}\ass S$. 
Hence, for \syncnst{the man}:
\disp{
\prooftree
\CN\yields\CN\tb N\yields N
\justifies
N/\CN, \CN\yields N
\using /L
\endprooftree
}
And for \syncnst{John sings} and \syncnst{John loves Mary}:
\disp{$
\prooftree
N\yields N\tb S\yields S
\justifies
N, N\bsl S\yields S
\using \bsl L
\endprooftree
\tab
\prooftree
N\yields N
\prooftree
N\yields N\tb S\yields S
\justifies
N, N\bsl S\yields S
\using \bsl L
\endprooftree
\justifies
N, (N\bsl S)/N, N\yields S
\using /L
\endprooftree
$}

The continuous product \verbquote{times}, $\product$, 
is exemplified by a \scare{small clause}
assignment like $\syncnst{considers}\ass$ $(N\bsl S)/(N\product(\CN/\CN))$
for $\syncnst{John considers Mary socialist}\ass S$:
\disp{$
\prooftree
\prooftree
N\yields N\tb 
\prooftree
\prooftree
\CN\yields\CN\tb\CN\yields\CN
\justifies
\CN/\CN, \CN\yields \CN
\using /L
\endprooftree
\justifies
\CN/\CN\yields \CN/\CN
\using /R
\endprooftree
\justifies
N, \CN/\CN\yields N\product(\CN/\CN)
\using \product R
\endprooftree
\prooftree
N\yields N\tb S\yields S
\justifies
N, N\bsl S\yields S
\using \bsl L
\endprooftree
\justifies
N, (N\bsl S)/(N\product(\CN/\CN)), N, \CN/\CN\yields S
\using /L
\endprooftree$
}
Of course this positive/succedent/right use of times is not essential: 
we could just as well
have used  $((N\bsl S)/(\CN/\CN))/N$ since in general we have both
$$A/(C\product B)\yields (A/B)/C~\mbox{(\scare{currying})}$$
and
$$(A/B)/C\yields A/(C\product B)~\mbox{(\scare{uncurrying})}.$$
For a 
negative/antecedent/left essential use
of times, for past participles, see Morrill (2000\cite{morrill:complexity}, section~2)).

The discontinuous multiplicatives \{$\scircum{}$, $\sinfix{}$, $\swprod{}$, $J$\},
the displacement connectives, of
Morrill and Valent\'{\i}n (2010\cite{mv:disp}),
Morrill, Valent\'{\i}n \& Fadda (2011\cite{mvf:tdc}) and
Morrill (2011\cite{morrill:oxford}\commentout{, chapter~ZZZ}),
are defined in relation to intercalation (wrapping). 
Their Gentzen sequent rules are given in Figure~\ref{dmult}.

\begin{figure}[ht]
\begin{center}
$
\begin{array}[t]{lc}
5, k. &\prooftree
\zeta_1\sass\Gamma\yields B \tb
\Xi(\zeta_2\sass\Delta_1, \langle\vect{C}\rangle, \Delta_2)\yields D
\justifies
\Xi(\zeta_1\mun\zeta_2\sass\Delta_1, \langle\vect{C\scircum{k} B}\smwrap{k}\Gamma\rangle, \Delta_2)\yields D
\using \scircum{k} L
\endprooftree \tb
\prooftree
\zeta\sass \Gamma\smwrap{k}\vect{B}\yields C
\justifies
\zeta\ass \Gamma\yields C\scircum{k} B
\using \scircum{k} R
\endprooftree
\\\\
6, k. &\prooftree
\zeta_1\sass\Gamma\yields A \tb
\Xi(\zeta_2\sass\Delta_1, \langle\vect{C}\rangle, \Delta_2)\yields D
\justifies
\Xi(\zeta_1\mun\zeta_2\sass\Delta_1, \langle\Gamma\smwrap{k}\vect{A\sinfix{k} C}\rangle, \Delta_2)\yields D
\using \sinfix{k} L
\endprooftree \tb
\prooftree
\zeta\sass\vect{A}\smwrap{k}\Gamma\yields C
\justifies
\zeta\sass\Gamma\yields A\sinfix{k} C
\using \sinfix{k} R
\endprooftree
\\\\
7, k. &\prooftree
\Xi\langle\vect{A}\smwrap{k}\vect{B}\rangle\yields D
\justifies
\Xi\langle\vect{A\swprod{k} B}\rangle\yields D
\using \swprod{k} L
\endprooftree \tb
\prooftree
\zeta_1\sass\Gamma_1\yields A\tb\zeta_2\sass\Gamma_2\yields B
\justifies
\zeta_1\mun\zeta_2\sass\Gamma_1\smwrap{k}\Gamma_2\yields A\swprod{k} B
\using \swprod{k} R
\endprooftree
\\\\
8. &\prooftree
\Xi\langle\sep\rangle\yields A
\justifies
\Xi\langle\vect{J}\rangle\yields A
\using JL
\endprooftree\tb
\prooftree
\justifies
\emptyset\sass\sep\yields J
\using JR
\endprooftree
\end{array}
$
\end{center}
\caption{Discontinuous multiplicative rules}
\label{dmult}
\end{figure}
\noindent
Notice how the discontinuous multiplicative rules have exactly the same form as the continuous
multiplicative rules but with metalinguistic intercalation 
\verbquote{$\smwrap{k}$} in place of the metalinguistic concatenation
\verbquote{$,$}; the stoups distribute as before.

When the value of the $k$ subscript is one it may be omitted,
i.e.\ it defaults to one.
\verbquote{Circumfixation}, 
$\scircum{}$,
is exemplified by a particle verb assignment
$\syncnst{calls}\add\sep\add\syncnst{up}\ass$ 
$(N\bsl S)\scircum{}N$ for
$\syncnst{Mary calls the man}$ $\syncnst{up}$ $\ass S$:
\disp{$
\prooftree
\prooftree
\CN\yields\CN\tb N\yields N
\justifies
N/\CN, \CN\yields N
\using /L
\endprooftree
\prooftree
N\yields N\tb S\yields S
\justifies
N, N\bsl S\yields S
\using \bsl L
\endprooftree
\justifies
N, (N\bsl S)\circum{}N\{N/\CN, \CN\}\yields S
\using \circum{}L
\endprooftree$}
\scare{Infixation}, 
$\sinfix{}$,
and circumfixation
together are exemplified by a quantifier assignment
$\syncnst{everyone}\ass$ $(S\scircum{} N)\sinfix{}S$
simulating Montague's S14 quantifying in:
\disp{$
\prooftree
\prooftree
\ldots, N, \ldots\yields S
\justifies
\ldots, \sep, \ldots\yields S\circum{}N
\using \circum{}R
\endprooftree
\prooftree
\justifies
S\yields S
\using \mbox{\it id}
\endprooftree
\justifies
\ldots, (S\circum{} N)\infix{} S, \ldots\yields S
\using \infix{}L
\endprooftree$}

Circumfixation and discontinuous product, \verbquote{wrap},
$\swprod{}$,
are illustrated in an assignment to a relative pronoun
$\syncnst{that}\ass(\CN\bsl\CN)/((S\circum{}N)\swprod{}I)$
allowing both peripheral and medial extraction,
$\syncnst{that John likes}\ass$ $\CN\bsl\CN$ and
$\syncnst{that John saw today}\ass\CN\bsl\CN$:
\disp{$
\prooftree
\prooftree
\prooftree
N, (N\bsl S)/N, N\yields S
\justifies
N, (N\bsl S)/N, \sep\yields S\circum{}N
\using \circum{}R
\endprooftree
\prooftree
\justifies
\yields I
\using IL
\endprooftree
\justifies
N, (N\bsl S)/N\yields (S\circum{}N)\wprod I
\using \wprod R
\endprooftree
\CN\bsl\CN\yields \CN\bsl\CN
\justifies
(\CN\bsl\CN)/((S\circum N)\wprod I), N, (N\bsl S)/N\yields \CN\bsl\CN
\using /L
\endprooftree$}

\disp{$
\prooftree
\prooftree
\prooftree
N, (N\bsl S)/N, N, S\bsl S\yields S
\justifies
N, (N\bsl S)/N, \sep, S\bsl S\yields S\circum{}N
\using \circum{}R
\endprooftree
\prooftree
\justifies
\yields I
\using IL
\endprooftree
\justifies
N, (N\bsl S)/N, S\bsl S\yields (S\circum{}N)\wprod I
\using \wprod R
\endprooftree
\CN\bsl\CN\yields \CN\bsl\CN
\justifies
(\CN\bsl\CN)/((S\circum N)\wprod I), N, (N\bsl S)/N, S\bsl S\yields \CN\bsl\CN
\using /L
\endprooftree
$}

\subsection{Additives}

The additive conjunction and disjunction \{$\aconj$, $\adisj$\}
of 
Lambek (1961\cite{lambek:61}), 
Morrill (1990\cite{morrill:galt90}), and Kanazawa (1992\cite{kanazawa:additives}),
capture polymorphism. Their rules are given in Figure~\ref{add}.

\begin{figure}[ht]
\begin{center}
$
\begin{array}{lc}
9. &
\prooftree
\Xi\langle\vect{A}\rangle\yields C
\justifies
\Xi\langle\vect{A\aconj B}\rangle\yields C
\using \aconj L_1
\endprooftree\tb
\prooftree
\Xi\langle\vect{B}\rangle\yields C
\justifies
\Xi\langle\vect{A\aconj B}\rangle\yields C
\using \aconj L_2
\endprooftree
\\\\
&\prooftree
\Xi\yields A\tb\Xi\yields B
\justifies
\Xi\yields A\aconj B
\using \aconj R
\endprooftree
\\\\
10. &
\prooftree
\Xi\langle\vect{A}\rangle\yields C\tb\Xi\langle\vect{B}\rangle\yields C
\justifies
\Xi\langle\vect{A\adisj B}\rangle\yields C
\using \adisj L
\endprooftree
\\\\
&\prooftree
\Xi\yields A
\justifies
\Xi\yields A\adisj B
\using \adisj R_1
\endprooftree\tb
\prooftree
\Xi\yields B
\justifies
\Xi \yields A\adisj B
\using \adisj R_2
\endprooftree
\end{array}
$
\end{center}
\caption{Additive rules}
\label{add}
\end{figure}
\noindent
Notice how the stoup is shared between premises and conclusion in the additive rules.
By way of example of the additives,
the additive conjunction \verbquote{with}, $\aconj$, can be used for the polymorphism of 
a mass noun
$\syncnst{rice}\ass N\aconj\CN$ as in $\syncnst{rice grows}\ass S$
and $\syncnst{the rice grows}\ass S$:
\disp{$
\prooftree
\prooftree
N\yields N
\justifies
N\aconj\CN\yields N
\using \aconj L_1
\endprooftree
S\yields S
\justifies
N\aconj\CN, N\bsl S\yields S
\using \bsl L
\endprooftree
\tab
\prooftree
N/\CN, \CN, N\bsl S\yields S
\justifies
N/\CN, N\aconj\CN, N\bsl S\yields S
\using \aconj L_2
\endprooftree
$}

The additive disjunction \verbquote{plus}, $\adisj$, 
can be used for the polymorphism
of the copula 
$$\syncnst{is}\ass(N\bsl S)/(N\adisj(\CN/\CN))$$
as in $\syncnst{Tully is Cicero}\ass S$ and $\syncnst{Tully is humanist}\ass S$:

\disp{$
\prooftree
\prooftree
N\yields N
\justifies
N\yields N\adisj(\CN/\CN)
\using \adisj R_1
\endprooftree
N\bsl S\yields N\bsl S
\justifies
(N\bsl S)/(N\adisj(\CN/\CN)), N\yields N\bsl S
\using /L
\endprooftree
\tab
\prooftree
\prooftree
\CN/\CN\yields \CN/\CN
\justifies
\CN/\CN\yields N\adisj(\CN/\CN)
\using \adisj R_2
\endprooftree
N\bsl S\yields N\bsl S
\justifies
(N\bsl S)/(N\adisj(\CN/\CN)), \CN/\CN\yields N\bsl S
\using /L
\endprooftree
$}

\subsection{Quantifiers}

The 1st order quantifiers \{$\bigwedge$, $\bigvee$\}
of 
Morrill (1994\cite{morrill:tlg}),
have application to features.
The rules are given in Figure~\ref{quant}.

\clearpage

\begin{figure}[ht]
\begin{center} 
 $
 \begin{array}{lc}
11. & \prooftree
 \Xi\langle \vect{A[t/v]}\rangle\yields B
 \justifies
 \Xi\langle\vect{\bigwedge vA}\rangle\yields B
 \using \bigwedge L
 \endprooftree\tb
 \prooftree
 \Xi\yields A[a/v]
 \justifies
 \Xi\yields \bigwedge vA
 \using \bigwedge R^\dagger
 \endprooftree
\\\\
12. &
\prooftree
\Xi\langle\vect{A[a/v]}\rangle\yields B
\justifies
\Xi\langle\vect{\bigvee vA}\rangle\yields B
\using \bigvee L^\dagger
\endprooftree\tb
\prooftree
\Xi\yields A[t/v]
\justifies
\Xi\yields\bigvee vA
\using \bigvee R
\endprooftree
\end{array}
$
\end{center}
\caption{Quantifier rules, where $^\dagger$ indicates that there is no $a$ in the conclusion}
\label{quant}
\end{figure}
\noindent
Notice how the stoup is identical on premises and conclusions in the quantifier rules.
By way of example of the quantifiers, we can generalise over singular and plural number in
$\syncnst{sheep}\ass$ $\bigwedge{}n\CN n$ for
$\syncnst{the sheep grazes}\ass S$ and $\syncnst{the sheep graze}\ass S$:
\disp{$
\prooftree
\CN\mbox{\it sg}\yields \CN\mbox{\it sg}
\justifies
\bigwedge n\CN n\yields \CN\mbox{\it sg}
\using \bigwedge L
\endprooftree
\tab
\prooftree
\CN\mbox{\it pl}\yields \CN\mbox{\it pl}
\justifies
\bigwedge n\CN n\yields \CN\mbox{\it pl}
\using \bigwedge L
\endprooftree
$}
And we can express a past, present or future tense finite sentence complement:
$\syncnst{said}\ass (N\bsl S)/$$\bigvee{}tSf(t)$ in
$\syncnst{John said Mary walked}\ass S$,
$\syncnst{John said Mary walks}\ass S$ and
$\syncnst{John said Mary}$ $\syncnst{will walk}\ass S$:
\disp{$
\prooftree
Sf(\mbox{\it past})\yields Sf(\mbox{\it past})
\justifies
Sf(\mbox{\it past})\yields \bigvee t Sf(t)
\using \bigvee R
\endprooftree
\tb
\prooftree
Sf(\mbox{\it pres})\yields Sf(\mbox{\it pres})
\justifies
Sf(\mbox{\it pres})\yields \bigvee t Sf(t)
\using \bigvee R
\endprooftree
\tb
\prooftree
Sf(\mbox{\it fut})\yields Sf(\mbox{\it fut})
\justifies
Sf(\mbox{\it fut})\yields \bigvee t Sf(t)
\using \bigvee R
\endprooftree
$}

\subsection{Modalities}

\subsubsection{Normal modalities}

With respect to the normal modalities \{$\mymod$, $\emod$\} of 
Morrill (1990\cite{morrill:ib90}, 1994\cite{morrill:tlg}), Hepple 1990\cite{hepple:phd}, and
Moortgat (1996\cite{moortgat:95}, 1997\cite{moortgat:handbook}),
the universal has application to semantic intensionality (Morrill \cite{morrill:ib90}).
The rules are given in Figure~\ref{smod}.

\begin{figure}[ht]
\begin{center}
$
\begin{array}{lcc}
13. & \prooftree
\Xi\langle\vect{A}\rangle\yields B
 \justifies
\Xi\langle\vect{\mymod A}\rangle\yields B
 \using \mymod L
 \endprooftree 
 &
\prooftree
 \diagmod\Xi\yields A
 \justifies
 \diagmod\Xi\yields \mymod A
 \using \mymod R
 \endprooftree
 \\\\
 14.\tb &
 \prooftree
\diagmod\Xi\langle\vect{A}\rangle\yields \diagemod B
 \justifies
\diagmod\Xi\langle\vect{\emod A}\rangle\yields \diagemod B
 \using \emod L
 \endprooftree 
 &
\prooftree
 \Xi\yields A
 \justifies
 \Xi\yields \emod A
 \using \emod R
 \endprooftree
 \end{array}
 $
 \end{center}
\caption{Normal modality rules; $\diagmod/\diagemod$ marks a structure all the types of
which have principal connective a box/diamond}
\label{smod}
\end{figure}
\noindent
Note how the stoup is identical in premises and conclusions in the normal modality rules.
With respect to the ({\bf S4}) normal modalities
the universal (Morrill 1990\cite{morrill:ib90}) has application to intensionality.
For example,
for a propositional attitude verb such as
$\syncnst{believes}$ we can assign type $\mymod((N\bsl S)/\mymod S)$
with a modality outermost since the word has a sense,
and a modality on the first argument but not the second,
since the sentential complement is an intensional domain,
but not the subject.
The modalities are in the categorial type, distinctly from, but in relation
to, the logical interpretation of the propositional attitude verb.

\subsubsection{Bracket modalities}

The bracket modalities \{$\abrack$, $\mybrack$\} of 
Morrill (1992\cite{morrill:92}) and
Moortgat (1995\cite{moortgat:95}),
have application to prosodic/syntactic domains
such as prosodic phrases and extraction
islands or intonational domains.
The rules are given in Figure~\ref{brmod}.

\begin{figure}[ht]
\begin{center}
$
\begin{array}{lc}
15. &
\prooftree
\Xi\langle \vect{A}\rangle\yields B
\justifies
\Xi\langle[\vect{\abrack A}]\rangle\yields B
\using \abrack L
\endprooftree \tb
\prooftree
[\Xi]\yields A
\justifies
\Xi\yields \abrack A
\using \abrack R
\endprooftree
\\\\
16. & 
\prooftree
\Xi\langle[\vect{A}]\rangle\yields B
\justifies
\Xi\langle\vect{\mybrack A}\rangle\yields B
\using \mybrack L
\endprooftree \tb
\prooftree
\Xi\yields A
\justifies
[\Xi]\yields \mybrack A
\using \mybrack R
\endprooftree
\end{array}
$
 \end{center}
\caption{Bracket modality rules}
\label{brmod}
\end{figure}
\noindent
Notice how the stoup is identical in conclusions and premises of bracket modality rules.

By way of example of bracket modalities, we may assign
$\syncnst{walks}\ass\mybrack{}N\bsl S$ for the subject condition
(Chomsky 1973\cite{chomsky:73}),
and
$\syncnst{before}\ass$ $\abrack{}(\VP\bsl\VP)/\VP$ for the adverbial island constraint,
which are weak islands, 
and can contain parasitic gaps;
for a strong island such as a coordinate structure, which cannot contain a parasitic gap,
we define doubly bracketed strong islands --- $\syncnst{and}\ass(S\bsl \abrack\abrack S)/S$.
\disp{$ 
\prooftree
\prooftree
N\yields N
\justifies
[N]\yields\mybrack N
\using \mybrack R
\endprooftree
S\yields S
\justifies
[N], \mybrack N\bsl S\yields S
\using \bsl L
\endprooftree
\tab
\prooftree
S\yields S\tb
\prooftree
S\yields S\tb
\prooftree
\prooftree
S\yields S
\justifies
[\abrack S]\yields S
\using \abrack L
\endprooftree
\justifies
[[\abrack\abrack S]]\yields S
\using \abrack L
\endprooftree
\justifies
[[S, S\bsl\abrack\abrack S]]\yields S
\using \bsl S
\endprooftree
\justifies
[[S, (S\bsl\abrack\abrack S)/S, S]]\yields S
\using / S
\endprooftree$}

\subsection{Subexponentials}

The subexponentials \{$\univexp$, $\exstexp$\} of
Girard (1987\cite{girard:87}),
Barry et al.\ (1991\cite{bhlm:91}),
Morrill (1994\cite{morrill:tlg}), Morrill (2011\cite{morrill:oxford}),
Morrill and Valent\'{\i}n (2015\cite{cctlg:nl}), 
Morrill and Valent\'{\i}n (2016\cite{mv:resp}),
and Morrill (2017\cite{gl:rel}
have application to nonlinearity.
The rules are given in Figure~\ref{exp}.
Note that the rule for $\exstexp L$ is infinitary (we know of no linguistic need for it).

\begin{figure}[ht]
\begin{center}
$
\begin{array}{ll}
17.\tb & 
 \prooftree
\Xi(\zeta\mun\{A\}\sass\Gamma_1, \Gamma_2)\yields B
 \justifies
\Xi(\zeta\sass\Gamma_1, \univexp A, \Gamma_2)\yields B
 \using \univexp L
 \endprooftree \tb
\prooftree
\zeta\sass \yields A
 \justifies
\zeta\sass \yields \univexp A
 \using \univexp R
 \endprooftree \\\\
&  \prooftree
\Xi(\zeta\sass\Gamma_1, A, \Gamma_2)\yields B
 \justifies
\Xi(\zeta\mun\{A\}\sass\Gamma_1, \Gamma_2)\yields B
 \using \univexp P
 \endprooftree
 \tb
  \commentout{ \prooftree
 \Xi(\zeta\mun\{\mbox{\fbox{A}}\}; \Gamma_1, [\{A\}; \Gamma_2], \Gamma_3)\yields B
 \justifies
 \Xi(\zeta\mun\{\mbox{\fbox{A}}\}; \Gamma_1, [\nil; [\nil; \Gamma_2]], \Gamma_3)\yields B\ass\psi
 \using C
 \endprooftree}
 \prooftree
 \Xi(\zeta\mun\{A\}; \Gamma_1, [\{A\}; \Gamma_2], \Gamma_3)\yields B
 \justifies
 \Xi(\zeta\mun\{A\}; \Gamma_1, \Gamma_2, \Gamma_3)\yields B
 \using \univexp C
 \endprooftree
  \\\\
 18. &
 \prooftree
 \Xi(A)\yields D\tb\Xi(A, A)\yields D\tb\cdots
 \justifies
 \Xi(\exstexp A)\yields D
 \using \exstexp L
\endprooftree
\tb
\prooftree
\Xi\yields A
\justifies
\Xi\yields \exstexp A
\using \exstexp R
\endprooftree\\\\
&
\prooftree
\Xi\yields A\tb\Xi'\yields \exstexp A
\justifies
\Xi, \Xi'\yields \exstexp A
\using \exstexp M
\endprooftree
\end{array}
$
 \end{center}
\caption{Subexponential rules}
\label{exp}
\end{figure}
\noindent
Using the universal subexponential, 
\univexp,
we can assign a relative pronoun type
$$\syncnst{that}\ass(\CN\bsl\CN)/(S/\univexp N)$$ allowing both medial extraction (via the permutation rule)
and parasitic extraction (via the contraction rule),
Morrill (2011\cite{morrill:oxford}),
Morrill and Valent\'{\i}n (2015\cite{cctlg:nl}), and Morrill (2017\cite{gl:rel}),
such as
$\syncnst{paper that}$ $\syncnst{John filed without reading}\ass\CN$,
where parasitic gaps can appear only in (weak) islands,
but can be iterated in (weak) subislands, subsubislands, and so on.
Using the existential exponential, 
\exstexp,
we can assign a coordinator type
$\syncnst{and}\ass(\exstexp N\bsl N)/N$ allowing iterated coordination
as in $\syncnst{John, Bill, Mary and Suzy}\ass N$.

\section{Soundness of {\bf hDL} for {\bf DL}}

\chapter{Further Types}

\label{CL-remtypeschap}

In this chapter we present further types. 
There are
semantically inactive primitive types and
defined synthetic connectives, 
limited contraction and limited expansion,
and the primitive difference operator.
Including the {\bf DL} primitive connectives from the previous chapter,
all the connectives of Generalised Displacement Logic {\bf GDL} are given in Figure~\ref{alltypes}.

\begin{figure}[ht]
\small$$
\begin{array}{lrclrclll}
1. & \Tp{}_i & ::= & \Tp{}_{i{+}j}/\Tp{}_j & s(C/B) & = & s(C)-s(B) & 
\mbox{over \cite{lambek:mathematics}}\\
2. & \Tp{}_j & ::= & \Tp{}_i\bsl\Tp{}_{i{+}j} & s(A\bsl C) & = & s(C)-s(A) & 
\mbox{under \cite{lambek:mathematics}}\\
3. & \Tp{}_{i{+}j} & ::= & \Tp{}_i\product\Tp{}_j & s(A\product B) & = & s(A)+s(B) 
&
\mbox{times \cite{lambek:mathematics}}\\
4. & \Tp{}_0 & ::= & I  & s(I) & = & 0 
& \mbox{continuous unit \cite{lambek:88}}\\
5, k. & \Tp{}_{i'} & ::= & \Tp{}_{i{+}j}\scircum{k}\Tp{}_j, 1\le k\le i' & s(C\scircum{k} B) & = & s(C)'-s(B) 
& \mbox{circumfix \cite{mvf:tdc}}\\
6, k. & \Tp{}_j & ::= & \Tp{}_{i'}\sinfix{k}\Tp{}_{i{+}j}, 1\le k\le i' & s(A\sinfix{k} C) & = & s(C)'-s(A) 
& \mbox{infix \cite{mvf:tdc}}\\
7, k. & \Tp{}_{i{+}j} & ::= & \Tp{}_{i'}\swprod{k}\Tp{}_j, 1\le k\le i' & s(A\swprod{k} B) & = & s(A)+s(B)-1
& \mbox{wrap\cite{mvf:tdc}}\\
8. & \Tp{}_1 & ::= & J & s(J) & = & 1 & 
\mbox{discontinuous unit \cite{mvf:tdc}}\\
9. & \Tp{}_i & ::= & \Tp{}_i\aconj\Tp{}_i & s(A\aconj B) & = & s(A)=s(B) 
& \mbox{with \cite{lambek:61, morrill:galt}}\\
10. & \Tp{}_i & ::= & \Tp{}_i\adisj\Tp{}_i & s(A\adisj B) & = & s(A)=s(B) 
& \mbox{plus  \cite{lambek:61, morrill:galt}}\\
11. & \Tp{}_i & ::= & \bigwedge V\Tp{}_i & s(\bigwedge vA) & = & s(A) 
& \mbox{1st order univ.\ qu.\ 
\cite{morrill:tlg}}\\
12. & \Tp{}_i & ::= & \bigvee V\Tp{}_i & s(\bigvee vA) & = & s(A) 
& \mbox{1st order exist.\ qu.\ 
\cite{morrill:tlg}}\\
13. & \Tp{}_i & ::= & \mymod\Tp{}_i & s(\mymod A) & = & s(A) 
& \mbox{univ.~norm.~modality \cite{morrill:ib90}}\\
14. & \Tp{}_i & ::= & \emod\Tp{}_i & s(\emod A) & = & s(A)  & 
\mbox{exist.~norm.~modality \cite{moortgat:handbook}}\\
15. & \Tp{}_i & ::= & \abrack{}\Tp{}_i & s(\abrack A) & = & s(A) 
& \mbox{univ.~bracket modality 
\cite{morrill:92, moortgat:95}}\\
16. & \Tp{}_i & ::= & \mybrack{}\Tp{}_i & s(\mybrack A) & = & s(A) 
& \mbox{exist.~bracket modality
\cite{morrill:92, moortgat:95}}\\
17. & \Tp{}_0 & ::= & \univexp\Tp{}_0 & s(\univexp A) & = & s(A)=0 
& \mbox{universal subexponential
\cite{bhlm:91}}\\
18. & \Tp{}_0 & ::= & \exstexp\Tp{}_0 & s(\exstexp A) & = & s(A)=0
& \mbox{existential subexponential
\cite{morrill:tlg}}\\
19. & \Tp{}_i & := & \Tp{}_{i{+}j}\argover \Tp{}_j & s(C\argover B) & = & s(C)-s(B)
& \mbox{left sem.~inactive over \cite{mv:words}} \\ 
20. & \Tp{}_j & := & \Tp{}_i\fununder\Tp{}_{i{+}j} & s(A\fununder C) & = & s(C)-s(A) 
& \mbox{left sem.~inactive under \cite{mv:words}}\\ 
21. & \Tp{}_{i{+}j} & ::= & \Tp{}_i\leftiprod\Tp{}_j & s(A\leftiprod B) & = & s(A)+s(B) 
&
\mbox{left sem.~inactive times \cite{mv:words}}\\
22. & \Tp{}_i & := & \Tp{}_{i{+}j}\funover \Tp{}_j & s(C\funover B) & = & s(C)-s(B) 
& \mbox{right sem.~inactive over \cite{mv:words}}\\ 
23. & \Tp{}_j & := & \Tp{}_i\argunder\Tp{}_{i{+}j} & s(A\argunder C) & = & s(C)-s(A) 
& \mbox{right sem.~inactive under \cite{mv:words}}\\ 
24. & \Tp{}_{i{+}j} & ::= & \Tp{}_i\rightiprod\Tp{}_j & s(A\rightiprod B) & = & s(A)+s(B) 
&
\mbox{right sem.~inactive times \cite{mv:words}}\\
25, k. & \Tp{}_{i'} & := & \Tp{}_{i{+}j}\funcircum{k} \Tp{}_j, 1\le k\le i{+}j & s(C\funcircum{k} B) & = & s(C)'-s(B) 
& \mbox{upper sem.~inactive circumfix \cite{mv:words}}\\ 
26, k. & \Tp{}_j & := & \Tp{}_{i'}\arginfix{k}\Tp{}_{i{+}j}, 1\le k\le i' & s(A\arginfix{k} C) & = & s(C)'-s(A) 
& \mbox{upper sem.~inactive infix \cite{mv:words}}\\ 
27, k. & \Tp{}_{i{+}j} & ::= & \Tp{}_{i'}\upperiwprod{k}\Tp{}_j, 1\le k\le i' & s(A\upperiwprod{k} B) & = & s(A)+s(B)-1
& \mbox{upper sem.~inactive wrap \cite{mv:words}}\\
28, k. & \Tp{}_{i'} & := & \Tp{}_{i{+}j}\argcircum{k} \Tp{}_j, 1\le k\le i' & T(C\argcircum{k} B) & = & s(C)'-s(B)
& \mbox{lower sem.~inactive circumfix \cite{mv:words}}\\ 
29, k. & \Tp{}_j & := & \Tp{}_{i'}\funinfix{k}\Tp{}_{i{+}j}, 1\le k\le i' & s(A\funinfix{k} C) & = & s(C)'-s(A) 
& \mbox{lower sem.~inactive infix \cite{mv:words}}\\ 
30, k. & \Tp{}_{i{+}j} & ::= & \Tp{}_{i'}\loweriwprod{k}\Tp{}_j, 1\le k\le i' & s(A\loweriwprod{k} B) & = & s(A)+s(B)-1
&
\mbox{lower sem.~inactive wrap\cite{mv:words}}\\
31. & \Tp{}_i & ::= & \Tp{}_i\iaconj\Tp{}_i & s(A\iaconj B) & = & s(A)=s(B) 
& \mbox{sem.~inactive with \cite{morrill:tlg}}\\
32. & \Tp{}_i & ::= & \Tp{}_i\iadisj\Tp{}_i & s(A\iadisj B) & = & s(A)=s(B) 
& \mbox{sem.\
inactive plus \cite{morrill:tlg}}\\
33. & \Tp{}_i & ::= & \forall V\Tp{}_i & s(\forall vA) & = & s(A) 
& \mbox{sem.~inactive 1st order univ.~qu.~\cite{morrill:tlg}}\\
34. & \Tp{}_i & ::= & \exists V\Tp{}_i & s(\exists vA) & = & s(A) 
& \mbox{sem.\ inactive 1st order exist.~qu.~\cite{morrill:tlg}}\\
35. & \Tp{}_i & ::= & \imod\Tp{}_i & s(\imod A) & = & s(A) 
& \mbox{sem.~inactive univ.~norm.~modality \cite{morrill:ib90}}\\
36. & \Tp{}_i & ::= & \iemod\Tp{}_i & s(\iemod A) & = & s(A) 
& \mbox{sem.~inactive exist.~norm. modality \cite{morrill:tlg}}\\
37. & \Tp{}_i & ::= & \leftproj\Tp{}_{i'} &
s(\leftproj A) & = & s(A)-1 &
\mbox{left projection
\cite{mvf:tbilisi}}\\
38. & \Tp{}_{i'} & ::= & \leftinj\Tp{}_i & 
s(\leftinj A) & = & s(A)' &
\mbox{left injection
\cite{mvf:tbilisi}}\\
39. & \Tp{}_i & ::= & \rightproj\Tp{}_{i'} &
s(\rightproj A) & = & s(A)-1 &
\mbox{right projection
\cite{mvf:tbilisi}}\\
39. & \Tp{}_{i'} & ::= & \leftinj\Tp{}_i & 
s(\leftinj A) & = & s(A)' &
\mbox{left injection
\cite{mvf:tbilisi}}\\
40. & \Tp{}_{i'} & ::= & \rightinj\Tp{}_i & 
s(\rightinj A) & = & s(A)' &
\mbox{right injection
\cite{mvf:tbilisi}}\\
41, k. & \Tp{}_{i'} & ::= & \ssplit{k}\Tp{}_i & 
s(\ssplit{k}A) & = & s(A)' &
\mbox{split
\cite{morrill:merenciano}}\\
42, k. & \Tp{}_i & ::= & \sbridge{k}\Tp{}_{i'} &
s(\sbridge{k}A) & = & s(A)-1 &
\mbox{bridge
\cite{morrill:merenciano}}\\
43. & \Tp{}_i & ::= & \Tp{}_{i{+}j}\ndiv\Tp{}_j^{\overline p} &  
s(B\ndiv A) & = & s(B)-s(A)  & 
 \mbox{non-det.~division \cite{mvf:tdc}}\\
44. & \Tp{}_{i{+}j} & ::= & \Tp{}_i\nprod\Tp{}_j & 
s(A\nprod B) & = & s(A)+s(B) & 
\mbox{non-det.~times  \cite{mvf:tdc}}\\
45. & \Tp{}_{i'} & ::= &  \Tp{}_{i{+}j}\nextract\Tp{}_j^{\overline p} &
s(C\nextract B) & = & s(C)'-s(B) & 
\mbox{non-det.~circumfix \cite{mvf:tdc}}\\
46. & \Tp{}_j & ::= & \Tp{}_{i'}^{\overline p}\ninfix\Tp{}_{i{+}j} &
s(A\ninfix C) & = & s(C)'-s(A) & 
\mbox{non-det.~infix  \cite{mvf:tdc}}\\
47. & \Tp{}_{i{+}j} & ::= & \Tp{}_{i'}\ndprod\Tp{}_j & 
s(A\ndprod B) & = & s(A)+s(B)-1 & 
\mbox{non-det.~wrap \cite{mvf:tdc}}\\
48. & \Tp{}_{i{+}j} & ::= & \Tp{}_{i{+}j}|\Tp{}_j & s(B|A) & = & s(B)\ge T(A) 
&
\mbox{limited contraction \cite{jaeger:2005}}\\ 
49. & \Tp{}_0 & ::= & w & s(w) & = & 0 
& \mbox{limited expansion \cite{mv:words}}\\
50. & \Tp{}_i & ::= & \Tp{}_i-\Tp{}_i & 
 s(A-B) & = & s(A)=s(B) & 
\mbox{difference
\cite{mv:jcss13}}
\end{array}
$$
\caption{The types of Generalised Displacement Logic {\bf GDL}}
\label{alltypes}
\end{figure}

\clearpage

\section{Semantically inactive primitives of {\bf EDL}}

The semantically inactive connectives divide into the semantically inactive multiplicatives
(continuous and discontinuous),
additives,
quantifers,
and (normal) modalities:

\begin{center}
\footnotesize$
\diagram
\mbox{Semantically inactive multiplicatives} & \argover\ \fununder &&
\funover\ \argunder & \mbox{Morrill \& Valent\'{\i}n (2014\cite{mv:words})}\\
 & \leftiprod\ && \rightiprod\\
 &  \funcircum{k}~\arginfix{k} && \argcircum{k}~\funinfix{k}\\
 & \upperiwprod{k}&&   \loweriwprod{k}\\
\mbox{Semantically inactive additives} & \iaconj & & \iadisj & \mbox{Morrill (1994\cite{morrill:tlg})}\\
\mbox{Semantically inactive quantifiers} & \forall & & \exists & \mbox{Morrill (1994\cite{morrill:tlg})}\\
\mbox{Semantically inactive normal modalities} & \imod & & \iemod & \mbox{Hepple (1990\cite{hepple:phd}),
Morrill (1994\cite{morrill:tlg})}\\
\enddiagram
$
\end{center}

\subsubsection{Multiplicatives}

Syntactically, the rules for the semantically inactive connectives are exactly the same
as for their semantically active counterparts; it is in their semantic labelling,
given in Part~II, that the semantically active and inactive rules differ.
Thus the semantically inactive continuous multiplicative rules are as given in
Figure~\ref{unconcmult}
and the semantically inactive discontinuous multiplicative rules are as given in
Figure~\ref{unimult}.


\begin{figure}[ht]
\begin{center}
$
\begin{array}{lc}
19. &
\prooftree
\zeta_1\sass\Gamma\yields B \tb
\Xi(\zeta_2\sass\Delta_1, \langle\vect{C}\rangle, \Delta_2)\yields D
\justifies
\Xi(\zeta_1\mun\zeta_2\sass\Delta_1, \langle\vect{C\argover B}, \Gamma\rangle, \Delta_2)\yields D
\using \argover L
\endprooftree \tb
\prooftree
\zeta\sass\Gamma, \vect{B}\yields C
\justifies
\zeta\ass\Gamma\yields C\argover B
\using \argover R
\endprooftree
\\\\
20. &
\prooftree
\zeta_1\sass\Gamma\yields A \tb
\Xi(\zeta_2\sass\Delta_1, \langle\vect{C}\rangle, \Delta_2)\yields D
\justifies
\Xi(\zeta_1\mun\zeta_2\sass\Delta_1, \langle\Gamma, \vect{A\fununder C}\rangle, \Delta_2)\yields D
\using \fununder L
\endprooftree \tb
\prooftree
\zeta\sass\vect{A}, \Gamma\yields C
\justifies
\zeta\sass\Gamma\yields A\fununder C
\using \fununder R
\endprooftree
\\\\
21. & \prooftree
\Xi\langle\vect{A}, \vect{B}\rangle\yields D
\justifies
\Xi\langle\vect{A\leftiprod B}\rangle\yields D
\using \leftiprod L
\endprooftree \tb
\prooftree
\zeta_1\sass\Delta\yields A\tb\zeta_2\sass\Gamma\yields B
\justifies
\zeta_1\mun\zeta_2\sass\Delta, \Gamma\yields A\leftiprod B
\using \leftiprod R
\endprooftree
\\\\
22. &
\prooftree
\zeta_1\sass\Gamma\yields B \tb
\Xi(\zeta_2\sass\Delta_1, \langle\vect{C}\rangle, \Delta_2)\yields D
\justifies
\Xi(\zeta_1\mun\zeta_2\sass\Delta_1, \langle\vect{C\funover B}, \Gamma\rangle, \Delta_2)\yields D
\using \funover L
\endprooftree \tb
\prooftree
\zeta\sass\Gamma, \vect{B}\yields C
\justifies
\zeta\sass\Gamma\yields C\funover B
\using \funover R
\endprooftree
\\\\
23. &
\prooftree
\zeta_1\sass\Gamma\yields A \tb
\Xi(\zeta_2\sass\Delta_1, \langle\vect{C}\rangle, \Delta_2)\yields D
\justifies
\Xi(\zeta_1\mun\zeta_2\sass\Delta_1, \langle\Gamma, \vect{A\argunder C}\rangle, \Delta_2)\yields D
\using \argunder L
\endprooftree \tb
\prooftree
\zeta\sass\vect{A}, \Gamma\yields C
\justifies
\zeta\sass\Gamma\yields A\argunder C
\using \argunder R
\endprooftree
\\\\
24. &
\prooftree
\Xi\langle\vect{A}, \vect{B}\rangle\yields D
\justifies
\Xi\langle\vect{A\rightiprod B}\rangle\yields D
\using \rightiprod L
\endprooftree \tb
\prooftree
\zeta_1\sass\Delta\yields A\tb\zeta_2\sass\Gamma\yields B
\justifies
\zeta_1\mun\zeta_2\sass\Delta, \Gamma\yields A\rightiprod B
\using \rightiprod R
\endprooftree
\end{array}
$
\end{center}
\caption{Semantically inactive continuous multiplicative rules}
\label{unconcmult}
\end{figure}

\begin{figure}[ht]
\begin{center}
$
\begin{array}{lc}
25, k. &
\prooftree
\zeta_1\sass\Gamma\yields B \tb
\Xi(\zeta_2\sass\Delta_1, \langle\vect{C}\rangle, \Delta_2)\yields D
\justifies
\Xi(\zeta_1\mun\zeta_2\sass\Delta_1, \langle\vect{C\funcircum{k} B}\smwrap{k}\Gamma\rangle, \Delta_2)\yields D
\using \funcircum{k} L
\endprooftree \tb
\prooftree
\zeta\sass\Gamma\smwrap{k}\vect{B}\yields C
\justifies
\zeta\sass\Gamma\yields C\funcircum{k} B
\using \funcircum{k} R
\endprooftree
\\\\
26, k. & 
\prooftree
\zeta_1\sass\Gamma\yields A \tb
\Xi(\zeta_2\sass\Delta_1, \langle\vect{C}\rangle, \Delta_2)\yields D
\justifies
\Xi(\zeta_1\mun\zeta_2\sass\Delta_1, \langle\Gamma\smwrap{k}\vect{A\arginfix{k} C}\rangle, \Delta_2)\yields D
\using \arginfix{k} L
\endprooftree \tb
\prooftree
\zeta\sass\vect{A}\smwrap{k}\Gamma\yields C
\justifies
\zeta\sass\Gamma\yields A\arginfix{k} C
\using \arginfix{k} R
\endprooftree
\\\\
27, k. &
\prooftree
\Xi\langle\vect{A}\smwrap{k}\vect{B}\rangle\yields D
\justifies
\Xi\langle\vect{A\upperiwprod{k} B}\rangle\yields D
\using \upperiwprod{k} L
\endprooftree \tb
\prooftree
\zeta_1\sass\Delta\yields A\tb\zeta_2\sass\Gamma\yields B
\justifies
\zeta_1\mun\zeta_2\sass\Delta\smwrap{k}\Gamma\yields A\upperiwprod{k} B
\using \upperiwprod{k} R
\endprooftree\\\\
28, k. &
\prooftree 
\zeta_1\sass\Gamma\yields B \tb
\Xi(\zeta_2\sass\Delta_1, \langle\vect{C}\rangle, \Delta_2)\yields D
\justifies
\Xi(\zeta_1\mun\zeta_2\sass\Delta_1, \langle\vect{C\argcircum{k} B}\smwrap{k}\Gamma\rangle, \Delta_2)\yields D
\using \argcircum{k} L
\endprooftree \tb
\prooftree
\zeta\sass\Gamma\smwrap{k}\vect{B}\yields C
\justifies
\zeta\sass\Gamma\yields C\argcircum{k} B
\using \argcircum{k} R
\endprooftree
\\\\
29, k. &
\prooftree
\zeta_1\sass\Gamma\yields A \tb
\Xi(\zeta_2\sass\Delta_1, \langle\vect{C}\rangle, \Delta_2)\yields D
\justifies
\Xi(\zeta_1\mun\zeta_2\sass\Delta_1,\langle\Gamma\smwrap{k}\vect{A\funinfix{k} C}\rangle, \Delta_2)\yields D
\using \funinfix{k} L
\endprooftree \tb
\prooftree
\zeta\sass\vect{A}\smwrap{k}\Gamma\yields C
\justifies
\zeta\sass\Gamma\yields A\funinfix{k} C
\using \funinfix{k} R
\endprooftree
\\\\
30, k. &
\prooftree
\Xi\langle\vect{A}\smwrap{k}\vect{B}\rangle\yields D
\justifies
\Xi\langle\vect{A\loweriwprod{k} B}\rangle\yields D
\using \loweriwprod{k} L
\endprooftree \tb
\prooftree
\zeta_1\sass\Delta\yields A\tb\zeta_2\sass\Gamma\yields B
\justifies
\zeta_1\mun\zeta_2\sass\Delta\smwrap{k}\Gamma\yields A\loweriwprod{k} B
\using \loweriwprod{k} R
\endprooftree
\end{array}$
\end{center}
\caption{Semantically inactive discontinuous multiplicative rules}
\label{unimult}
\end{figure}

\clearpage

\subsubsection{Additives, quantifiers and normal modalities}

The semantically inactive additives, quantifiers and normal
modalities are as shown in Figures~\ref{unadd},
\ref{unquant}, and~\ref{unmod} respectively.

\begin{figure}[ht]
\begin{center}
$
\begin{array}{lc}
31. &
\prooftree
\Xi\langle\vect{A}\rangle\yields C
\justifies
\Xi\langle\vect{A\iaconj B}\rangle\yields C
\using \iaconj L_1
\endprooftree\tb
\prooftree
\Xi\langle\vect{B}\rangle\yields C
\justifies
\Xi\langle\vect{A\iaconj B}\rangle\yields C
\using \iaconj L_2
\endprooftree
\\\\
&\prooftree
\Xi\yields A\tb\Xi\yields B
\justifies
\Xi\yields A\iaconj B
\using \iaconj R
\endprooftree
\\\\
32. &
\prooftree
\Xi\langle\vect{A}\rangle\yields C\tb\Xi\langle\vect{B}\rangle\yields C
\justifies
\Xi\langle\vect{A\iadisj B}\rangle\yields C
\using \iadisj L
\endprooftree
\\\\
&\prooftree
\Xi\yields A
\justifies
\Xi\yields A\iadisj B
\using \iadisj R_1
\endprooftree\tb
\prooftree
\Xi\yields B
\justifies
\Xi \yields A\iadisj B
\using \iadisj R_2
\endprooftree
\end{array}
$
\end{center}
\caption{Semantically inactive additive rules}
\label{unadd}
\end{figure}

\begin{figure}[ht]
\begin{center} 
 $
\begin{array}{lc}
33. &  \prooftree
 \Xi\langle \vect{A[t/v]}\rangle\yields B
 \justifies
\Xi\langle\vect{\forall vA}\rangle\yields B
 \using \forall L
 \endprooftree\tb
 \prooftree
\Xi\yields A[a/v]
 \justifies
\Xi\yields \forall vA
 \using \forall R^\dagger
 \endprooftree
\\\\
34. &
\prooftree
\Xi\langle\vect{A[a/v]}\rangle\yields B
\justifies
\Xi\langle\vect{\exists vA}\rangle\yields B
\using \exists L^\dagger
\endprooftree\tb
\prooftree
\Xi\yields A[t/v]
\justifies
\Xi\yields\exists vA
\using \exists R
\endprooftree
\end{array}$
\end{center}
\caption{Semantically inactive quantifier rules, 
where $^\dagger$ indicates that there is no $a$ in the conclusion}
\label{unquant}
\end{figure}

\begin{figure}[ht]
\begin{center}
$
\begin{array}{lcc}
35. & \prooftree
\Xi\langle\vect{A}\rangle\yields B
 \justifies
\Xi\langle\vect{\imod A}\rangle\yields B
 \using \imod L
 \endprooftree 
 &
\prooftree
 \diagmod\Xi\yields A
 \justifies
 \diagmod\Xi\yields \imod A
 \using \imod R
 \endprooftree
 \\\\
 36.\tb &
 \prooftree
\diagmod\Xi\langle\vect{A}\rangle\yields \diagemod B
 \justifies
\diagmod\Xi\langle\vect{\iemod A}\rangle\yields \diagemod B
 \using \iemod L
 \endprooftree 
 &
\prooftree
 \Xi\yields A
 \justifies
 \Xi\yields \iemod A
 \using \iemod R
 \endprooftree
 \end{array}
 $
 \end{center}
\caption{Semantically inactive normal modality rules; $\diagmod/\diagemod$ marks a structure all the types of
which have principal connective a box/diamond}
\label{unmod}
\end{figure}

\clearpage

\section{Synthetic connectives of {\bf FDL}}

We consider in the following subsection unary defined synthetic connectives and binary
defined synthetic connectives.

\subsection{Unary synthetic multiplicatives}

The unary (or deterministic) synthetic multiplicatives divide into projection and injection,
which are continuous,
and split and bridge, which are discontinuous:
\begin{center}
\footnotesize$
\diagram
\mbox{Left and right projection and injection} & 
\leftproj &  \leftinj & \rightinj &\rightproj  & \mbox{Morrill, Valent\'{\i}n \& Fadda (2009\cite{mvf:tbilisi})}\\
\mbox{Split and bridge}  &  & \ssplit{k}  & \sbridge{k} & &
\mbox{Morrill \& Merenciano (1996\cite{morrill:merenciano})}
\enddiagram
$
\end{center}
They are defined as shown in Figure~\ref{unarysyn}.

\begin{figure}[ht]
$$
\begin{array}[t]{rcl
rcllll}
\rightproj A & =_{df} & J\fununder A & 
s(\rightproj A) & = & s(A)-1 &
  \multicolumn{3}{l}{\mbox{right projection
\cite{mvf:tbilisi}}}\\
\leftproj A & =_{df} & A\funover J & 
s(\leftproj A) & = & s(A)-1 &
\multicolumn{3}{l}{\mbox{left projection
\cite{mvf:tbilisi}}}\\
\rightinj A & =_{df} & J\leftiprod A & 
s(\rightinj A) & = & s(A)' &
 \multicolumn{3}{l}{\mbox{right injection
\cite{mvf:tbilisi}}}\\
\leftinj A & =_{df} & A\rightiprod J & 
s(\leftinj A) & = & s(A)' &
\multicolumn{3}{l}{\mbox{left injection
\cite{mvf:tbilisi}}}\\
\ssplit{k}A & =_{df} & A\argcircum{k} I & 
s(\ssplit{k}A) & = & s(A)' &
\multicolumn{3}{l}{\mbox{leftmost and rightmost split
\cite{morrill:merenciano}}}\\
\sbridge{k} & =_{df} & A\loweriwprod{k} I & 
s(\sbridge{k}A) & = & s(A)-1 &
 \multicolumn{3}{l}{\mbox{leftmost and rightmost bridge
\cite{morrill:merenciano}}}
\end{array}
$$
\caption{Unary synthetic multiplicatives}
\label{unarysyn}
\end{figure}

\subsubsection{Rules for unary synthetic multiplicatives}

\begin{figure}[ht]
\begin{center}
$
\begin{array}{lc}
37. &
\prooftree
\Xi\langle\vect{A}\rangle\yields B
\justifies
\Xi\langle\vect{\leftproj A}, \sep\rangle\yields B
\using \leftproj L
\endprooftree\tb
\prooftree
\zeta\sass\Gamma, \sep\yields A
\justifies
\zeta\sass\Gamma\yields \leftproj A
\using \leftproj R
\endprooftree
\\\\
38. &
\prooftree
\Xi\langle\vect{A}\rangle\yields B
\justifies
\Xi\langle\sep, \vect{\rightproj A}\rangle\yields B
\using \rightproj L
\endprooftree\tb
\prooftree
\zeta\sass\sep, \Gamma\yields A
\justifies
\zeta\sass\Gamma\yields\rightproj A
\using \rightproj R
\endprooftree
\\\\
39. &
\prooftree
\Xi\langle \vect{A}, \sep\rangle\yields B
\justifies
\Xi\langle\vect{\leftinj A}\rangle\yields B
\using \leftinj L
\endprooftree\tb
\prooftree
\zeta\sass\Gamma\yields A
\justifies
\zeta\sass\Gamma, \sep\yields\leftinj A
\using \leftinj R
\endprooftree
\\\\
40. &
\prooftree
\Xi\langle\sep, \vect{A}\rangle\yields B
\justifies
\Xi\langle\vect{\rightinj A}\rangle\yields B
\using \rightinj L
\endprooftree\tb
\prooftree
\zeta\sass\Gamma\yields A
\justifies
\zeta\sass\sep, \Gamma\yields\rightinj A
\using \rightinj R
\endprooftree
\\\\
41, k. &
\prooftree
\Xi\langle \vect{B}\rangle\yields C
\justifies
\Xi\langle\vect{\ssplit{k} B}\smwrap{k}\Lambda\rangle\yields C
\using \ssplit{k} L
\endprooftree\tb
\prooftree
\zeta\sass\Delta\smwrap{k}\Lambda\yields B
\justifies
\zeta\sass\Delta\yields\ssplit{k} B
\using \ssplit{k} R
\endprooftree
\\\\
42, k. &
\prooftree
\Xi\langle\vect{B}\smwrap{k}\Lambda\rangle\yields C
\justifies
\Xi\langle\vect{\sbridge{k} B}\rangle\yields C
\using \sbridge{k} L
\endprooftree\tb
\prooftree
\zeta\sass \Delta\yields B
\justifies
\zeta\sass\Delta\smwrap{k}\Lambda\yields\sbridge{k} B
\using \sbridge{k} R
\endprooftree
\end{array}
$
\end{center}
\caption{Unary synthetic multiplicative rules}
\label{ununder}
\end{figure}

\subsection{Binary synthetic multiplicatives}

The binary (or non-deterministic) synthetic connectives divide into non-deterministic continuous division and times,
and non-deterministic discontinuous circumfix, infix and wrap:
\begin{center}
\footnotesize$
\diagram
\mbox{Non-det.~division and times} & \ndiv  & & \nprod & \mbox{Morrill, Valent\'{\i}n \& Fadda (2011\cite{mvf:tdc})}\\
\mbox{Non-det.\ infix,  wrap and circumfix} & \ninfix & \ndprod & \nextract & \mbox{Morrill, Valent\'{\i}n \& Fadda (2011\cite{mvf:tdc})}\\
\commentout{\mbox{Non-det.\ projection and injection} & \nproj && \ninj\\
\mbox{Non-det.\ split and bridge} & \nsplit & & \nbridge & 
\mbox{Semantically inactive non-det.\ division and product} & \nlif && \nlip}
\enddiagram
$
\end{center}
These are defined as shown in Figure~\ref{binarysyn}.
\begin{figure}[ht]
$$\scriptsize
\rotatebox{-0}{
\begin{array}[t]{rcc
rclllll}
B\ndiv A & =_{df} & (A\bsl B)\iaconj(B/A) & 
\{s|\ \forall s'\in A, s_3, +(s, s', s_3)\  \Rightarrow\ s_3\in B\} & 
s(B\ndiv A) & = & 
s(B)-s(A) 
 &  \multicolumn{3}{l}{\mbox{non-det.~division}}\\
A\nprod B & =_{df} & (A\product B)\iadisj(B\product A) & 
\{s_3|\ \exists s_1\in A, s_2\in B, +(s_1, s_2, s_3)\} & 
s(A\nprod B) & = & s(A)+s(B)
 & \multicolumn{3}{l}{\mbox{nond-et.~times}}\\
A\ninfix C  & =_{df} & (A\sinfix{1} C)\iaconj\cdots\iaconj(A\sinfix{s(A)} C) & 
\{s_2|\ \forall s_1\in A, s_3, \times(s_1, s_2, s_3)\  \Rightarrow\ s_3\in C\} &
s(A\ninfix C) & = & s(C)'-s(A)
 & \multicolumn{3}{l}{\mbox{non-det.~infix}}\\
C\nextract B & =_{df} & (C\scircum{1} B)\iaconj\cdots\iaconj(C\scircum{s(C)'-s(B)} B) & 
\{s_1|\ \forall s_2\in B, s_3, \times(s_1, s_2, s_3)\  \Rightarrow\ s_3\in C\} &
s(C\nextract B) & = & s(C)'-T(B)
 & \multicolumn{3}{l}{\mbox{non-det.~circumfix}}\\
A\ndprod B & =_{df} & (A\swprod{1} B)\iadisj\cdots\iadisj(A\swprod{s(A)} B) & 
\{s_3|\ \exists s_1\in A, s_2\in B, \times(s_1, s_2, s_3)\} &
s(A\ndprod B) & = & s(A)'-s(B)
 & \multicolumn{3}{l}{\mbox{non-det.~wrap}}\\
 \commentout{
 \nproj A & =_{df} & \leftproj A\iaconj\rightproj A & T(\nproj A) & = & T(A) & \multicolumn{3}{l}{\mbox{non-det.~projection}}\\
 \ninj A & =_{df} & \leftinj A\iadisj\rightinj A & T(\ninj A) & = & T(A) & \multicolumn{3}{l}{\mbox{non-det.\ injection}}\\
  \nsplit A & =_{df} & \ssplit{+}A\iaconj\ssplit{-}A & T(\nsplit A) & = & T(A) & \multicolumn{3}{l}{\mbox{non-det.~split}}\\ 
 \nbridge A & =_{df} & \sbridge{+}A\iadisj\sbridge{-}A & T(\nbridge A) & = & T(A) & \multicolumn{3}{l}{\mbox{non-det.~bridge}}\\ 
 A\nlif B & =_{df} & (A\fununder B)\iaconj(B\funover A) & T(A\nlif B) & = & T(B) \mbox{\ if\ }T(A)=\top
 & \mbox{non-det.\ sem.~inactive division}\\
 A\nlip B & =_{df} & (A\leftiprod B)\iadisj(B\rightiprod A) & T(A\nlip B) & = & T(B) \mbox{\ if\ }
 T(A)=\top & \mbox{non-det.~sem.~inactive times}\\}
 \end{array}
}
$$
\caption{Binary synthetic multiplicatives}
\label{binarysyn}
\end{figure}

\subsubsection{Rules for binary synthetic multiplicatives}

\begin{figure}[ht]
\begin{center}
$
\begin{array}{lc}
43. &
\prooftree
\zeta_1\sass\Gamma\yields A\tb
\Xi(\zeta_2\sass\Delta_1, \langle\vect{C}\rangle, \Delta_2)\yields D
\justifies
\Xi(\zeta_1\mun\zeta_2\sass\Delta_1, \langle\Gamma, \vect{C\ndiv A}\rangle, \Delta_2)\yields D
\using \ndiv L_1
\endprooftree \tb
\prooftree
\zeta_1\sass \Gamma\yields A \tb
\Xi(\zeta_2\sass \Delta_1, \langle\vect{C}\rangle, \Delta_2)\yields D
\justifies
\Xi(\zeta_1\mun\zeta_2\sass\Delta_1, \langle\vect{C\ndiv A}, \Gamma\rangle, \Delta_2)\yields D
\using \ndiv L_2
\endprooftree
\\\\
&\prooftree
\zeta\sass\vect{A}, \Gamma\yields C\tb
\zeta\sass\Gamma, \vect{A}\yields C
\justifies
\zeta\sass\Gamma\yields C\ndiv A
\using \ndiv R
\endprooftree
\\\\
44. &
\prooftree
\Xi\langle\vect{A}, \vect{B}\rangle\yields D\tb
\Xi\langle\vect{B}, \vect{A}\rangle\yields D
\justifies
\Xi\langle\vect{A\nprod B}\rangle\yields D\ass
\using \nprod L
\endprooftree
\\\\
&\prooftree
\zeta_1\sass\Delta\yields A\tb\zeta_2\sass\Gamma\yields B
\justifies
\zeta_1\mun\zeta_2\sass\Delta, \Gamma\yields A\nprod B
\using \nprod R_1
\endprooftree
\tb
\prooftree
\zeta_1\sass \Delta\yields B\tb\zeta_2\sass\Gamma\yields A
\justifies
\zeta_1\mun\zeta_2\sass\Delta, \Gamma\yields A\nprod B
\using \nprod R_2
\endprooftree
\\\\
45. &
\prooftree
\zeta_1\sass\Gamma\yields B\tb
\Xi(\zeta_2\sass\Delta_1, \langle\vect{C}\rangle, \Delta_2)\yields D
\justifies
\Xi(\zeta_1\mun\zeta_2\sass\Delta_1, \langle\vect{C\nextract B}\smwrap{k}\Gamma\rangle, \Delta_2)\yields D
\using \nextract L
\endprooftree \tb
\prooftree
\zeta\sass\Gamma\smwrap{1}\vect{B}\ass y\yields C\quad\cdots\quad
\zeta\sass\Gamma\smwrap{sC'-sB}\vect{B}\ass y\yields C
\justifies
\zeta\sass\Gamma\yields C\nextract B
\using \nextract R
\endprooftree
\\\\
46. & 
\prooftree
\zeta_1\sass\Gamma\yields A \tb
\Xi(\zeta_2\sass\Delta_1, \langle\vect{C}\rangle, \Delta_2)\yields D\ass
\justifies
\Xi(\zeta_1\mun\zeta_2\sass\Delta_1, \langle\Gamma\smwrap{k}\vect{A\ninfix C}\rangle, \Delta_2)\yields D
\using \ninfix L
\endprooftree \tb
\prooftree
\zeta\sass\vect{A}\smwrap{1}\Gamma\yields C\quad\cdots\quad
\zeta\sass\vect{A}\smwrap{sA}\Gamma\yields C
\justifies
\zeta\sass\Gamma\yields A\ninfix C
\using \ninfix R
\endprooftree
\\\\
47. & 
\prooftree
\Xi\langle\vect{A}\smwrap{1}\vect{B}\rangle\yields D\quad\cdots\quad
\Xi\langle\vect{A}\smwrap{sA}\vect{B}\rangle\yields D
\justifies
\Xi\langle\vect{A\ndprod B}\rangle\yields D
\using \ndprod L
\endprooftree \tb
\prooftree
\zeta_1\sass\Delta\yields A\tb\zeta_2\sass\Gamma\yields B
\justifies
\zeta_1\mun\zeta_2\sass\Delta\smwrap{k}\Gamma\yields A\ndprod B
\using \ndprod R
\endprooftree
\end{array}
$
\end{center}
\caption{Binary synthetic multiplicative rules}
\label{slunbinders}
\end{figure}

\commentout{
\section{Rules for semantically inactive non-deterministic defined multiplicatives}

\begin{figure}[ht]
\begin{center}
$
\begin{array}{lc}
61. &
\prooftree
\Gamma\yields A\tb
\Delta\langle\vect{C}\rangle\yields D
\justifies
\Delta\langle\Gamma, \vect{A\nlif C}\rangle\yields D
\using \nlif L_1
\endprooftree \tb
\prooftree
\Gamma\yields A \tb
\Delta\langle\vect{C}\rangle\yields D
\justifies
\Delta\langle\vect{A\nlif C}, \Gamma\rangle\yields D
\using \nlif L_2
\endprooftree
\\\\
&\prooftree
\vect{A}, \Gamma\yields C\tb
\Gamma, \vect{A}\yields C
\justifies
\Gamma\yields A\nlif C
\using \nlif R
\endprooftree
\\\\
62. &
\prooftree
\Delta\langle\vect{A}, \vect{B}\rangle\yields D\tb
\Delta\langle\vect{B}, \vect{A}\rangle\yields D
\justifies
\Delta\langle\vect{A\nlip B}\rangle\yields D\ass
\using \nlip L
\endprooftree
\\\\
&\prooftree
\Gamma_1\yields A\tb\Gamma_2\yields B
\justifies
\Gamma_1, \Gamma_2\yields A\nlip B
\using \nlip R_1
\endprooftree
\tb
\prooftree
\Gamma_1\yields B\tb\Gamma_2\yields A
\justifies
\Gamma_1, \Gamma_2\yields A\nlip B
\using \nlip R_2
\endprooftree
\end{array}
$
\end{center}
\caption{Semantically inactive non-deterministic derived multiplicative rules}
\label{inslunbinders}
\end{figure}
}

\section{Limited contraction and limited expansion connectives of {\bf GDL}}

The limited contraction $|$,
of Figure~\ref{sllca}, J\"{a}ger (2005\cite{jaeger:2005}), has application to
anaphora, and the limited expansion $+$, to words as types.

\begin{figure}[ht]
\begin{center}
$
\begin{array}{ll}
48.\tb &
 \prooftree
\zeta_1\sass\Gamma\yields A\tb\Xi(\zeta_2\sass\Delta_1, \langle \vect{A}\rangle, \Delta_2\langle\vect{B}\rangle)\yields D
\justifies
\Xi(\zeta_1\mun\zeta_2\sass\Delta_1\langle\Gamma\rangle, \Delta_2\langle\vect{B|A}\rangle)\yields D
\using |L
\endprooftree
\tab
\prooftree
\Xi\langle \vect{B_0}; \ldots; \vect{B_n}\rangle\yields D
\justifies
\Xi\langle \vect{B_0|A}; \ldots; \vect{B_n|A}\rangle\yields D|A
\using |R
\endprooftree
\\\\
49. & \prooftree
\Xi(\Lambda)\yields A
\justifies
\Xi(\vect{0}\yields A
\using +L
\endprooftree\tb
\prooftree
\justifies
\nil\sass\Lambda\yields 0
\using +R
\endprooftree
\\\\
&\prooftree
\Xi\langle\vect{v+w}\rangle\yields D
\justifies
\Xi\langle\vect{v}, \vect{w}\rangle\yields D
\using +L
\endprooftree \tb
\prooftree
\zeta_1\sass\Delta\yields v\tb\zeta_2\sass \Gamma\yields w
\justifies
\zeta_1\mun\zeta_2\sass\Delta, \Gamma\yields v+w
\using +R
\endprooftree
\end{array}$
\end{center}
\caption{Limited contraction and limited expansion}
\label{sllca}
\end{figure}
\noindent
The limited contraction can be used for anaphora in an assignment like
$\syncnst{it}\ass(S\scircum{}N)\sinfix{}(S|N)$
for,
e.g.,
\pcnst{the} \syncnst{company$_i$ said it$_i$} $\syncnst{flourished}\ass S$,
and it can be used for $\syncnst{such\ that}$ relativisation
in an assignment
$\syncnst{such\ that}\ass$ $(\CN\bsl\CN)/(S|N)$ for,
say,
$\syncnst{man\ such\ that\mbox{$_i$}\ he\mbox{$_i$}\ thinks\ Mary\ loves\ him\mbox{$_i$}}$\ass \CN.  
The interesting thing about this example is that it has a positive (succedent) occurrence
of the J\"ager bar $S!N$. This necessitates a pronoun in the
corresponding antcecedent, and allows more such pronouns, all
cobound as required to the head noun resticted by the \lingform{such that}
relativisation.

The limited wexpansion
$+$ of Morrill and Valent\'{\i}n (2014\cite{mv:words}) has
application to words as types, for example in relation to the semantically inactive
multiplicatives, $\syncnst{is}\ass (\syncnst{there}\fununder S)/N$ for
\lingform{There is John}. The intended reading of this example is that John exists:
\syncnst{there} makes no semantic contribution and is argument to
a semantically inactive under.

\section{The difference operator of {\bf GDL$'$}}

The difference operator is a connective for which the Cut-rule does
not make sense \commentout{ZZZexample/explain}. Therefore it is considered a metalogical
connective added to the system {\bf GDL} without Cut,
yielding the system {\bf GDL'}.
The rules are as follows:

\begin{figure}[ht]
\begin{center}
$
\begin{array}{lc}
50. &
\prooftree
\Xi\langle A\rangle\yields C
\justifies
\Xi\langle A-B\rangle\yields C
\using -L
\endprooftree
\tab
\prooftree
\Xi\yields A
\justifies
\Xi\yields A-B
\using -R, \not\vdash \Xi\yields B
\endprooftree
\end{array}
$
\end{center}
\caption{Difference rules}
\end{figure}

\noindent
The difference operator is a means to define exceptions. For example, to avoid generation
of \unacc \pcnst{the} \pcnst{extremely} \pcnst{man} with an intensifier type $(\CN/\CN)/(\CN/\CN)$ applying
to the empty string of type $\CN/\CN$, we can instead assign an intensifier type
$(\CN/\CN)/((\CN/\CN)-I)$ excluding the empty string.

\chapter{Completeness of {\bf hDL} for syntactical models, and Cut-elimination}

\commentout{ZZZDependent on logico-algebraic foundations.}

\label{semCut}

\part{SEMANTICS}

\noindent
In this part we present the Curry-Howard type-logical semantics of Generalised Displacement
Logic {\bf GDL}.
In Chapter~\ref{typeschap} we recall the definition of syntactic types of Generalised Displacement Logic,
adding the semantic type map to intuitionistic types.
In Chapter~\ref{semchap} we define the semantic representation language.
In Chapter~\ref{slsc} we present the semantically labelled sequent calculus for {\bf GDL}.

\chapter{Generalised Displacement
Logic {\bf GDL} Types}

\label{typeschap}

In this chapter we see the semantic type map for Generalised Displacement
Logic, {\bf GDL}.

\section{Semantic types}

Recall the following operations on sets:
\disp{
\begin{itemize}
\item[a.]
Functional exponentiation: $X^Y = \mbox{the set of all total functions from $Y$ to $X$}$
\item[b.]
Cartesian product: $X\times Y = \{\langle x, y\rangle|\ x\in X\ \&\ y\in Y\}$
\item[c.]
Disjoint union: $X\uplus Y = (\{1\}\times X)\cup(\{2\}\times Y)$
\item[d.]
$n$-th Cross product, $n\in{\cal N}$: $\begin{array}[t]{rcl}
X^0 & = & \{\nil\}\\
X^{n+1} & = & X^n\times X
\end{array}$
\end{itemize}}
We shall use these to build semantic domains.
The set $\cal T$ of {\em semantic types\/} of the semantic
representation language is defined on the
basis of a set $\delta$ of {\em basic semantic types\/} as follows:
\disp{$
{\cal T}  ::=  \delta\ |\ F\ |\ \top\ |\ 
{\cal T}{+}{\cal T}\ |\ 
{\cal T}\&{\cal T}\ |\ 
{\cal T}{\rightarrow}{\cal T}\ |\ 
{\bf M}{\cal T}\ |\ 
{\bf L}{\cal T}\ |\ 
{\cal T}^+
$}
A \techterm{semantic frame\/} comprises a family $\{D_\tau\}_{\tau\in\delta}$
of non-empty \techterm{basic type domains\/} and nonempty sets
$V$ of feature values and $+$ of possible worlds.
This induces a nonempty \techterm{type domain} $D_\tau$ for each type $\tau$ as follows:
\disp{$
\begin{array}[t]{rclll}
D_F & = & V\\
D_\top & = & \{\emptyset\}
\\
D_{\tau_1{+}\tau_2} & = & D_{\tau_1}\uplus{D_{\tau_2}}
\\
D_{\tau_1\&\tau_2} & = & D_{\tau_1}\times{D_{\tau_2}}
\\
D_{\tau_1{\rightarrow}\tau_2} & = & D_{\tau_2}^{D_{\tau_1}}
\\
D_{{\bf M}\tau} & = & W\times D_\tau
\\
D_{{\bf L}\tau} & = &  D_\tau^W
\\
D_{\tau^+} & = & \bigcup_{n>0}(D_{\tau})^n
\end{array}
$}

\section{Syntactic types}

The syntactic types of our categorial logic are sorted
according to the number of points of discontinuity their expressions contain.
Each \techterm{type predicate letter\/} will have a sort,
the sort of it's expressions, and an arity,
the number of feature term arguments it takes, which
are naturals, and there is primitive semantic type
map $t$. Assuming to be already given ordinary feature terms 
interpreted in the domain $F$,
where $P$ is a type predicate letter of sort $i$ and arity $n$
and  $t_1, \ldots, t_n$ are terms,
$Pt_1\ldots t_n$ is an (atomic) type of sort $i$ of the corresponding semantic type.
Compound types are formed by connectives as shown in Figure~\ref{GCLtypes},
and the homomorphic semantic type map $T$ associates these with semantic
types and hence, through $D$, semantic domains.

\begin{figure}[ht]
\small$$
\begin{array}{lrclrclll}
\begin{array}{lrclrclll}
1. & \Tp{}_i & ::= & \Tp{}_{i{+}j}/\Tp{}_j & T(C/B) & = & T(B){\rightarrow}T(C)
 & \mbox{over \cite{lambek:mathematics}}\\
2. & \Tp{}_j & ::= & \Tp{}_i\bsl\Tp{}_{i{+}j} & T(A\bsl C) & = & T(A){\rightarrow}T(C)
 & \mbox{under \cite{lambek:mathematics}}\\
3. & \Tp{}_{i{+}j} & ::= & \Tp{}_i\product\Tp{}_j & T(A\product B) & = & T(A)\&T(B) &
\mbox{times \cite{lambek:mathematics}}\\
4. & \Tp{}_0 & ::= & I & T(I) & = & \top & \mbox{continuous unit \cite{lambek:88}}\\
5, k. & \Tp{}_{i'} & ::= & \Tp{}_{i{+}j}\scircum{k}\Tp{}_j, 1\le k\le i' & T(C\scircum{k} B) & = & T(B){\rightarrow}T(C) & \mbox{circumfix \cite{mvf:tdc}}\\
6, k. & \Tp{}_j & ::= & \Tp{}_{i'}\sinfix{k}\Tp{}_{i{+}j}, 1\le k\le i' & T(A\sinfix{k} C) & = & T(A){\rightarrow}T(C) & \mbox{infix \cite{mvf:tdc}}\\
7, k. & \Tp{}_{i{+}j} & ::= & \Tp{}_{i'}\swprod{k}\Tp{}_j, 1\le k\le i' & T(A\swprod{k} B) & = & T(A)\&T(B)& \mbox{wrap \cite{mvf:tdc}}\\
8. & \Tp{}_1 & ::= & J & T(J) & = & \top & \mbox{discontinuous unit \cite{mvf:tdc}}\\
9. & \Tp{}_i & ::= & \Tp{}_i\aconj\Tp{}_i & T(A\aconj B) & = & T(A)\&T(B) & \mbox{with \cite{lambek:61, morrill:galt}}\\
10. & \Tp{}_i & ::= & \Tp{}_i\adisj\Tp{}_i & T(A\adisj B) & = & T(A){+}T(B) & \mbox{plus \cite{lambek:61, morrill:galt}}\\
11. & \Tp{}_i & ::= & \bigwedge V\Tp{}_i & T(\bigwedge vA) & = & F{\rightarrow}T(A) & \mbox{1st order univ.\ qu.\ 
\cite{morrill:tlg}}\\
12. & \Tp{}_i & ::= & \bigvee V\Tp{}_i & T(\bigvee vA) & = & F\&T(A) & \mbox{1st order exist.\ qu.\ 
\cite{morrill:tlg}}\\
13. & \Tp{}_i & ::= & \mymod\Tp{}_i & T(\mymod A) & = & {\bf L}T(A) & \mbox{univ.~norm.~modality \cite{morrill:ib90}}\\
14. & \Tp{}_i & ::= & \emod\Tp{}_i & T(\emod A) & = & {\bf M}T(A) & \mbox{exist.~norm.~modality \cite{moortgat:handbook}}\\
15. & \Tp{}_i & ::= & \abrack{}\Tp{}_i & T(\abrack A) & = & T(A) & \mbox{univ.\ brack.~modality 
\cite{morrill:92, moortgat:95}}\\
16. & \Tp{}_i & ::= & \mybrack{}\Tp{}_i & T(\mybrack A) & = & T(A) & \mbox{exist.\ brack.~modality
\cite{morrill:92, moortgat:95}}\\
17. & \Tp{}_0 & ::= & \univexp\Tp{}_0 & T(\univexp A) & = & T(A) & \mbox{univ.~subexponential
\cite{bhlm:91}}\\
18. & \Tp{}_0 & ::= & \exstexp\Tp{}_0 & T(\exstexp A) & = & T(A)^+ & \mbox{exist.~subexponential
\cite{morrill:tlg}}\\
19. & \Tp{}_i & := & \Tp{}_{i{+}j}\argover \Tp{}_j & T(C\argover B) & = & \top\mathrm{\ if\ } T(C)=\top
& \mbox{left sem.\ inactive over \cite{mv:words}} \\ 
20. & \Tp{}_j & := & \Tp{}_i\fununder\Tp{}_{i{+}j} & T(A\fununder C) & = & T(C)\mathrm{\ if\ } T(A)=\top 
& \mbox{left sem.\ inactive under \cite{mv:words}}\\ 
21. & \Tp{}_{i{+}j} & ::= & \Tp{}_i\leftiprod\Tp{}_j & T(A\leftiprod B) & = & T(B)\mathrm{\ if\ } T(A)=\top &
\mbox{left sem.\ inactive cont.\ product \cite{mv:words}}\\
22. & \Tp{}_i & := & \Tp{}_{i{+}j}\funover \Tp{}_j & T(C\funover B) & = & T(C)\mathrm{\ if\ } T(B)=\top 
& \mbox{right sem.\ inactive over \cite{mv:words}}\\ 
23. & \Tp{}_j & := & \Tp{}_i\argunder\Tp{}_{i{+}j} & T(A\argunder C) & = & \top\mathrm{\ if\ } T(C)=\top 
& \mbox{right sem.\ inactive under \cite{mv:words}}\\ 
24. & \Tp{}_{i{+}j} & ::= & \Tp{}_i\rightiprod\Tp{}_j & T(A\rightiprod B) & = & T(A)\mathrm{\ if\ } T(B)=\top &
\mbox{right sem.\ inactive cont.\ product \cite{mv:words}}\\
25, k. & \Tp{}_{i'} & := & \Tp{}_{i{+}j}\funcircum{k} \Tp{}_j, 1\le k\le i'& T(C\funcircum{k} B) & = & \top\mathrm{\ if\ } T(C)=\top 
& \mbox{upper sem.\ inactive extract \cite{mv:words}}\\ 
26, k. & \Tp{}_j & := & \Tp{}_{i'}\arginfix{k}\Tp{}_{i{+}j}, 1\le k\le i' & T(A\arginfix{k} C) & = & T(C) \mathrm{\ if\ } T(A)=\top 
& \mbox{upper sem.\ inactive infix \cite{mv:words}}\\ 
27, k. & \Tp{}_{i{+}j} & ::= & \Tp{}_{i'}\upperiwprod{k}\Tp{}_j, 1\le k\le i' & T(A\upperiwprod{k} B) & = & T(B)\mathrm{\ if\ } T(A)=\top &
\mbox{upper sem.\ inactive disc.\ product \cite{mv:words}}\\
28, k. & \Tp{}_{i'} & := & \Tp{}_{i{+}j}\argcircum{k} \Tp{}_j, 1\le k\le i' & T(C\argcircum{k} B) & = & T(C)\mathrm{\ if\ } T(B)=\top
& \mbox{lower sem.\ inactive extract \cite{mv:words}} \\ 
29, k. & \Tp{}_j & := & \Tp{}_{i'}\funinfix{k}\Tp{}_{i{+}j}, 1\le k\le i' & T(A\funinfix{k} C) & = & \top\mathrm{\ if\ } T(C)=\top 
& \mbox{lower sem.\ inactive infix \cite{mv:words}}\\ 
30, k. & \Tp{}_{i{+}j} & ::= & \Tp{}_{i'}\loweriwprod{k}\Tp{}_j, 1\le k\le i' & T(A\loweriwprod{k} B) & = & T(A)\mathrm{\ if\ } T(B)=\top &
\mbox{lower sem.\ inactive disc.\ product \cite{mv:words}}\\
31. & \Tp{}_i & ::= & \Tp{}_i\iaconj\Tp{}_i & T(A\iaconj B) & = & T(A)=T(B) & \mbox{sem.\ 
inactive with \cite{morrill:tlg}}\\
32. & \Tp{}_i & ::= & \Tp{}_i\iadisj\Tp{}_i & T(A\iadisj B) & = & T(A)=T(B) & \mbox{sem.\
inactive plus \cite{morrill:tlg}}\\
33. & \Tp{}_i & ::= & \forall V\Tp{}_i & T(\forall vA) & = & T(A) & \mbox{sem.\ inactive 1st order univ.\ qu.\ \cite{morrill:tlg}}\\
34. & \Tp{}_i & ::= & \exists V\Tp{}_i & T(\exists vA) & = & T(A) & \mbox{sem.\ inactive 1st order exist.\ qu.\ \cite{morrill:tlg}}\\
35. & \Tp{}_i & ::= & \imod\Tp{}_i & T(\imod A) & = & T(A) & \mbox{sem.\ inactive universal modality \cite{morrill:ib90}}\\
36. & \Tp{}_i & ::= & \iemod\Tp{}_i & T(\iemod A) & = & T(A) & \mbox{sem.\ inactive existential modality \cite{morrill:tlg}}\\
37. & \Tp{}_i & ::= & \leftproj\Tp{}_{i'} &
T(\leftproj A) & = & T(A) &
\mbox{left projection
\cite{mvf:tbilisi}}\\
38. & \Tp{}_i & ::= & \rightproj\Tp{}_{i'} &
T(\rightproj A) & = & T(A) &
\mbox{right projection
\cite{mvf:tbilisi}}\\
39. & \Tp{}_{i'} & ::= & \leftinj\Tp{}_i & 
T(\leftinj A) & = & T(A) &
\mbox{left injection
\cite{mvf:tbilisi}}\\
40. & \Tp{}_{i'} & ::= & \rightinj\Tp{}_i & 
T(\rightinj A) & = & T(A) &
\mbox{right injection
\cite{mvf:tbilisi}}\\
41, k. & \Tp{}_{i'} & ::= & \ssplit{k}\Tp{}_i & 
T(\ssplit{k}A) & = & T(A) &
\mbox{split
\cite{morrill:merenciano}}\\
42, k. & \Tp{}_i & ::= & \sbridge{k}\Tp{}_{i'} &
T(\sbridge{k}A) & = & T(A) &
\mbox{bridge
\cite{morrill:merenciano}}\\
43. & \Tp{}_i & ::= & \Tp{}_{i{+}j}\ndiv\Tp{}_j^{\overline p} &  
T(B\ndiv A) & = & 
T(A){\rightarrow}T(B) 
 &  \mbox{non-det.\ division \cite{mvf:tdc}}\\
44. & \Tp{}_{i{+}j} & ::= & \Tp{}_i\nprod\Tp{}_j & 
T(A\nprod B) & = & T(A)\&T(B)
 & \mbox{non-det.\ continuous product  \cite{mvf:tdc}}\\
45. & \Tp{}_{i'} & ::= &  \Tp{}_{i{+}j}\nextract\Tp{}_j^{\overline p} &
T(C\nextract B) & = & T(B){\rightarrow}T(C)
 & \mbox{non-det.\ extract  \cite{mvf:tdc}}\\
46. & \Tp{}_j & ::= & \Tp{}_{i'}^{\overline p}\ninfix\Tp{}_{i{+}j} &
T(A\ninfix C) & = & T(A){\rightarrow}T(C)
 & \mbox{non-det.\ infix  \cite{mvf:tdc}}\\
47. & \Tp{}_{i{+}j} & ::= & \Tp{}_{i'}\ndprod\Tp{}_j & 
T(A\ndprod B) & = & T(A)\&T(B)
 & \mbox{non-det.\ discontinuous product  \cite{mvf:tdc}}\\
48. & \Tp{}_{i{+}j} & ::= & \Tp{}_{i{+}j}|\Tp{}_j & T(B|A) & = & T(A){\rightarrow}T(B) &
\mbox{limited contraction \cite{jaeger:2005}}\\ 
49. & \Tp{}_0 & ::= & w & T(w) & = & \top & \mbox{limited expansion \cite{mv:words}}\\
50. & \Tp{}_i & ::= & \Tp{}_i-\Tp{}_i & T(A-B) & = & T(A) & \mbox{difference \cite{mv:jcss13}}\\
\end{array}
\end{array}
$$
\caption{Types of {\bf GDL}}
\label{GCLtypes}
\end{figure}

\chapter{Semantic Representation Language}

\label{semchap}

In this chapter we define terms of the semantic representation
language, interpretation of the semantic representation language,
and conversion laws of the semantic representation language.

\section{Terms}

The sets $\Phi_\tau$ of {\em terms\/} of type
$\tau$ for each semantic type $\tau$ are defined on the basis
of sets $C_\tau$ of constants of type $\tau$
and denumerably infinite sets $V_\tau$ of variables
of type $\tau$ for each type $\tau$ as follows:
\disp{$
\begin{array}[t]{rcll}
\Phi_\tau & ::= & C_\tau & \mbox{constants}\\
\Phi_\tau & ::= & V_\tau & \mbox{variables}\\
\Phi_\top & ::= & \zero & \mbox{null element}\\
\Phi_\tau & ::= & \Phi_{\tau_1+\tau_2}\casearrow V_{\tau_1}.\Phi_\tau;\ V_{\tau_2}.\Phi_\tau
& \mbox{case statement}\\
\Phi_{\tau+\tau'} & ::= & \iota_1\Phi_\tau & \mbox{first injection}\\
\Phi_{\tau'+\tau} & ::= & \iota_2\Phi_\tau & \mbox{second injection}\\
\Phi_\tau & ::= & \pi_1\Phi_{\tau\&\tau'} & \mbox{first projection}\\
\Phi_\tau & ::= & \pi_2\Phi_{\tau'\&\tau} & \mbox{second projection}\\
\Phi_{\tau\&\tau'} & ::= & (\Phi_\tau, \Phi_{\tau'}) & \mbox{ordered pair formation}\\
\Phi_\tau & ::= & (\Phi_{\tau'\rightarrow\tau}\,\Phi_{\tau'}) & \mbox{functional application}\\
\Phi_{\tau\rightarrow\tau'} & ::= & \lambda V_\tau\Phi_{\tau'} & \mbox{functional abstraction}\\
\Phi_\tau & ::= & \dn\Phi_{{\bf L}\tau} & \mbox{extensionalization}\\
\Phi_{{\bf L}\tau} & ::= & \up\Phi_\tau & \mbox{intensionalization}\\
\Phi_\tau & ::= & \dnc\Phi_{{\bf M}\tau} & \mbox{projection}\\
\Phi_{{\bf M}\tau} & ::= & \upc\Phi_\tau & \mbox{injection}\\
\Phi_{\tau^+} & ::= & [\Phi_{\tau}]\ |\ [\Phi_{\tau}|\Phi_{\tau^+}] & \mbox{non-empty list construction}
\end{array}
$}

\section{Interpretation}

Given a semantic frame,
a \techterm{valuation} $f$ mapping each
constant of type $\tau$ into an element of $D_\tau$,
an assignment $g$ mapping each
variable of type $\tau$ into an element of $D_\tau$,
and a world $i\in W$,
each term $\phi$ of type $\tau$ receives an interpretation $[\phi]^{g, i}\in D_\tau$
as shown in Figure~\ref{semint}, where the \techterm{update} $g[x:=m]$ is
$(g-\{(x, g(x)\})\cup\{(x, m)\}$, i.e.\ the function which sends $x$ to $m$ and agrees
with $g$ elsewhere.
\begin{figure}[ht]
$$
\begin{array}[t]{rcll}
[a]^{g, i} & = & f(a)\ \mbox{for constant}\ a\in C_\tau\\
{}[x]^{g, i} & = & g(x)\ \mbox{for variable}\ x\in V_\tau\\
{}[0]^{g, i} & = & \emptyset\\
{}[\phi\casearrow x.\psi; y.\chi]^{g, i} & = & \left \{
\begin{array}{ll}
[\psi]^{g[x:={\bf snd}([\phi]^{g, i})], i} & \mbox{if}\ {\bf fst}([\phi]^{g, i}) = 1\\
{}[\chi]^{g[y:={\bf snd}([\phi]^{g, i})], i} & \mbox{if}\ {\bf fst}([\phi]^{g, i}) = 2
\end{array} 
\right . \\
{}[\iota_1\phi]^{g, i} & = & \langle 1, [\phi]^{g, i}\rangle \\
{}[\iota_2\phi]^{g, i} & = & \langle 2, [\phi]^{g, i}\rangle \\
{}[\pi_1\phi]^{g, i} & = & {\bf fst}([\phi]^{g, i}) \\
{} [\pi_2\phi]^{g, i} & = & {\bf snd}([\phi]^{g, i}) \\
{}[(\phi, \psi)]^{g, i} & = & \langle [\phi]^{g, i}, [\psi]^{g, i}\rangle\\
{}[(\phi\,\psi)]^{g, i} & = & [\phi]^{g, i}([\psi]^{g, i})\\
{}[\lambda x\phi]^{g, i} & = & d \mapsto [\phi]^{g[x:=d], i}\\
{}[\dn\phi]^{g, i} & = & [\phi]^{g, i}(i)\\
{}[\up\phi]^{g, i} & = & j\mapsto [\phi]^{g, j}\\
{}[\dnc\phi]^{g, i} & = & {\bf snd}([\phi]^{g, i})\\
{}[\upc\phi]^{g, i} & = & \langle i, [\phi]^{g, i}\rangle\\
{}[[\phi]]^ {g, i} & = & \langle[\phi]^{g, i}, \nil\rangle\\
{}[[\phi|\psi]]^{g, i} & = & \langle[\phi]^{g, i}, [\psi]^{g, i}\rangle
\end{array}
$$
\caption{Interpretation of the semantic representation language}
\label{semint}
\end{figure}

In $x.\phi$, $\lambda x\phi$ or $\up\phi$, $\phi$ is the \techterm{scope\/}
of $x.$, $\lambda x$ or $\up$.
An occurrence of a variable $x$ in a term
is called {\em free\/} if and only if it does not fall within the scope
of any $x.$ or $\lambda x$; 
otherwise it is {\em bound\/} (by the closest
$x.$ or $\lambda x$ within the scope of which it falls).
The result $\phi\{\psi_1/x_1, \ldots, \psi_n/x_n\}$ of substituting terms 
$\psi_1, \ldots, \psi_n$
for variables 
$x_1, \ldots, x_n$ 
of the same types respectively
in a term $\phi$ is the result of simultaneously replacing by $\psi_i$
every free
occurrence of $x_i$ in $\phi$.
We say that $\psi$ is {\em free for $x$ in $\phi$} if and only if
no variable in $\psi$ becomes bound in $\phi\{\psi/x\}$.
We say that a term is \techterm{modally closed\/} if and only if
every occurrence of $\dn$ occurs within the scope of an $\up$.
A modally closed term is denotationally invariant across worlds.
We say that a term $\psi$ is \techterm{modally free for $x$ in $\phi$} if
and only if either $\psi$ is modally closed,
or no free occurrence of $x$ in $\phi$ is within the scope of an $\up$.

\section{Conversion}

The laws of conversion in Figure~\ref{convlaws} obtain.
\begin{figure}[ht]
\begin{center}
$
\begin{array}[t]{rcl}
\phi\casearrow y.\psi; z.\chi & \equiv & \phi\casearrow x.(\psi\{x/y\});
z.\chi\\
& \mbox{if $x$ is not free in $\psi$}\\
& \mbox{and is free for $y$ in $\psi$}\\
\phi\casearrow y.\psi; z.\chi & \equiv & \phi\casearrow y.\psi;
x.(\chi\{x/z\})\\
& \mbox{if $x$ is not free in $\chi$}\\
& \mbox{and is free for $z$ in $\chi$}\\
\lambda y\phi & \equiv & \lambda x(\phi\{x/y\})\\
& \mbox{if $x$ is not free in $\phi$}\\
&\mbox{and is free for $y$ in $\phi$}\\
& \mbox{\em $\alpha$-conversion}\\\\
\iota_1\phi\casearrow y.\psi; z.\chi & \equiv &
\psi\{\phi/y\}\\
& \mbox{if $\phi$ is free for $y$ in $\psi$}\\
& \mbox{and modally free for $y$ in $\psi$}\\
\iota_2\phi\casearrow y.\psi; z.\chi & \equiv &
\chi\{\phi/z\}\\
& \mbox{if $\phi$ is free for $z$ in $\chi$}\\
& \mbox{and modally free for $z$ in $\chi$}\\
\pi_1(\phi, \psi) & \equiv & \phi\\
\pi_2(\phi, \psi) & \equiv & \psi\\
(\lambda x\phi\,\psi) & \equiv & \phi\{\psi/x\}\\
& \mbox{if $\psi$ is free for $x$ in $\phi$,}\\
& \mbox{and modally free for $x$ in $\phi$}\\
\dn\up\phi & \equiv & \phi\\
\dnc\upc\phi & \equiv & \phi\\
& \mbox{\em $\beta$-conversion}\\\\
(\pi_1\phi, \pi_2\phi) & \equiv & \phi\\
\lambda x(\phi\,x) & \equiv & \phi\\
& \mbox{if $x$ is not free in $\phi$}\\
\up\dn\phi & \equiv & \phi\\
& \mbox{if $\phi$ is modally closed}\\
\upc\dnc\phi & \equiv & \phi\\
& \mbox{\em $\eta$-conversion}
\end{array}
$
\end{center}
\caption{Semantic conversion laws}
\label{convlaws}
\end{figure}
For completeness,
the so-called commuting conversions for the case statement
are given in Figure~\ref{commconvlaws}.

\begin{figure}[ht]
\begin{center}
$
\begin{array}[t]{rcll}
\phi\casearrow x.\iota_1\psi; y.\iota_1\chi & \equiv & \iota_1(\phi\casearrow x.\psi; y.\chi)\\
\phi\casearrow x.\iota_2\psi; y.\iota_2\chi & \equiv & \iota_2(\phi\casearrow x.\psi; y.\chi)\\
\phi\casearrow x.\pi_1\psi; y.\iota_1\chi & \equiv & \pi_1(\phi\casearrow x.\psi; y.\chi)\\
\phi\casearrow x.\pi_2\psi; y.\iota_2\chi & \equiv & \pi_2(\phi\casearrow x.\psi; y.\chi)\\
\phi\casearrow x.(\delta, \psi); y.(\delta, \chi) & \equiv & (\delta, \phi\casearrow x.\psi; y.\chi)\\
\phi\casearrow x.(\psi, \delta); y.(\chi, \delta) & \equiv & (\phi\casearrow x.\psi; y.\chi, \delta)\\
\phi\casearrow x.(\delta\,\psi); y.(\delta\,\chi) & \equiv & (\delta\,\phi\casearrow x.\psi; y.\chi)\\
\phi\casearrow x.(\psi\,\delta); y.(\chi\,\delta) & \equiv & (\phi\casearrow x.\psi; y.\chi\,\delta)\\
\\
\phi\casearrow x.\lambda z\psi; y.\lambda z\chi & \equiv & \lambda z(\phi\casearrow x.\psi; y.\chi)\\
& \mbox{if $z$ is not free in $\phi$}\\
\\
\phi\casearrow x.\dn\psi; y.\dn\chi & \equiv & \dn(\phi\casearrow x.\psi; y.\chi)\\
\\
\phi\casearrow x.\up\psi; y.\up\chi & \equiv & \up(\phi\casearrow x.\psi; y.\chi)\\
\multicolumn{3}{l}{\mbox{if $\phi$ is modally closed}}\\
\\
\phi\casearrow x.\dnc\psi; y.\dnc\chi & \equiv & \dnc(\phi\casearrow x.\psi; y.\chi)\\
\phi\casearrow x.\upc\psi; y.\upc\chi & \equiv & \upc(\phi\casearrow x.\psi; y.\chi)\\
\phi\casearrow x.[\delta|\psi]; y.[\delta|\chi] & \equiv & [\delta|\phi\casearrow x.\psi; y.\chi]\\
\phi\casearrow x.[\psi|\delta]; y.[\chi|\delta] & \equiv & [\phi\casearrow x.\psi; y.\chi|\delta]\\
\end{array}
$
\end{center}
\caption{Semantic commuting conversion laws}
\label{commconvlaws}
\end{figure}

\chapter{Semantically Labelled Sequent Calculus}

\label{slsc}

In this chapter we give semantically labelled Gentzen sequent calculus rules
according to Curry-Howard formulas-as-types/proofs-as-programs
(Girard, Tayler and Lafont 1989\cite{girardtaylaf}),
but we shall do this in relation to the enriched notion of hedge sequent calculus.

\section{Hedge sequent calculus}

We  recall the notions of hedge sequent calculus from Chapter~\ref{seqcalcchap}.
In Gentzen sequent calculus for 
displacement calculus with bracket modalities (structural inhibition)
the left hand, antecedent, sides of sequents contain bracket constructors,
but subexponentials
(structural facilitation) have no special sequent punctuation.
There is motivation to provide a special antecedent notation of \scare{stoups}
for structural facilitation as well as that of brackets for structural inhibition.
For example, the permutation rules for $\univexp$ mean that the calculus
without stoups does not have the finite proof property: the permutation rules
can be applied in a senseless cyclic fashion to generate an infinite number of
proofs of the same theorem.
The use of stoups restores the finite proof property and it also facilitates the
proof-theoretic property of focusing (Chapter~\ref{focchap}).

\techterm{Stoups\/} (cf.~the linear logic of Girard 2011\cite{girard:blind})  are stores read as multisets for re-usable (non-linear)
resources which appear at the left of a configuration marked off by
a semicolon
(when the stoup is empty the semicolon may be omitted). The stoup of linear
logic is for resources which can be contracted (copied) or weakened (deleted).
By contrast, our stoup is for a linguistically motivated variant of contraction, and
does not allow weakening. 
Furthermore, whereas linear logic is commutative,
our logic is in general noncommutative and the stoup is also used for resources
which are commutative.\footnote{To anticipate linguistically a little, a hypothetical subtype emitted by a relative pronoun
corresponding to a long-distance
dependency will enter a stoup, percolate in stoups, maybe contracting
to create (parasitic) gaps, and finally permute into a (host)
extraction site.}

A configuration together with a stoup is a \techterm{zone\/}.
The bracket constructor applies not to a configuration alone but to a configuration
with a stoup, i.e\ a zone: a reusable resource is specific to its 
bracketed domain.

Zones $\Zone$ (notated $\Xi$, possibly with subindices), stoups $\Stoup$ (notated $\zeta$, possibly
with subindices) and
configurations $\Config$ (notated $\Gamma, \Delta$, possibly with
subindices) are defined by
($\emptyset$ is the empty stoup;
$\Lambda$ is the empty configuration;
the \techterm{separator} $\sep$ marks points of discontinuity.); note that only types of sort $0$
can go into the stoup:
reusable types of other sorts would not preserve the sequent
antecedent-succedent sort equality under contraction:
\disp{$\begin{array}[t]{rcl}
\Zone{} & ::= & {\Stoup}; {\Config}\\
{\Stoup} & ::= & \emptyset\ |\ {\Tp}_0, {\Stoup}\\
{\Config} & ::= & \Lambda\ |\ {\Tterm}, {\Config}\\
{\Tterm} & ::= & \sep\ |\ {\Tp}_0\ |\ {\Tp}_{i{>}0}\{\underbrace{{\Config}:
\ldots : {\Config}}_{i\ \mbox{\scriptsize{\bf Config}'s}}\}\ |\ [\Zone{}]
\end{array}$}

For a type $A$, its sort $s(A)$ is the $i \mathrm{\ such\ that\ }A\in {\Tp}_i$;
for example:
\disp{$s((S\scircum{1}N)\scircum{1} N)=s((S\scircum{1}N)\scircum{2} N)=2$}
For a configuration $\Gamma$, its sort 
$s(\Gamma)$ is $|\Gamma|_{\mathrm{1}}$,
i.e.\ the number of points of discontinuity $\sep$ which $\Gamma$ contains.
For a zone $\Xi$, its sort $s(\Xi)$ is the sort of its configuration
since stoup types are of sort is $0$; for example:
\disp{$\begin{array}[t]{l}s(N; \sep, \sep, (S\scircum{1} N)\scircum{2} N\{N/\CN, \CN: \sep\}) =
s(\sep, \sep, (S\scircum{1} N)\scircum{2} N\{N/\CN, \CN: \sep\}) = 3
\end{array}$}
Sequents are of the form:
\disp{$\Zone{}\yields {\Tp}\mathrm{\ such\ that\ }s(\Zone{})= s({\Tp})$}
A sequent $\Xi\yields A$ is \techterm{valid}, $\models
\Xi\yields A$, iff $[\Xi] \subseteq [A]$ in every interpretation.

The \techterm{figure\/} $\vect{A}$ of a type $A$ is defined by:
\disp{$
\vect{A} = \left\{
\begin{array}{ll}
A & \mbox{if\ } s(A)=0\\
A\{\underbrace{\sep: \ldots: \sep}_{s(A)\ 1\mathit{'s}}\} & \mbox{if\ } s(A)>0
\end{array}\right.$}

Where $\Gamma$ is a configuration of sort $i$ and $\Delta_1, \ldots, \Delta_i$
are configurations,
the \techterm{fold\/} $\Gamma\otimes\langle\Delta_1: \ldots:  \Delta_i\rangle$
is the result of replacing the successive \sep{}'s in $\Gamma$
by $\Delta_1, \ldots, \Delta_i$ respectively;
similarly, 
where $\Xi$ is a zone of sort $i$ and $\Delta_1, \ldots, \Delta_i$
are configurations,
the \techterm{fold\/} $\Xi\otimes\langle\Delta_1: \ldots:  \Delta_i\rangle$
is the result of replacing the successive \sep{}'s in the configuration of $\Xi$
by $\Delta_1, \ldots, \Delta_i$ respectively.

Where $\Gamma$ is a configuration of sort $i$,
the hyperoccurrence notation
$\Delta\langle\Gamma\rangle$ abbreviates
$\Delta_0(\Gamma\otimes\langle\Delta_1: \ldots:  \Delta_i\rangle)$,
i.e.\ a context configuration $\Delta$ (which is externally $\Delta_0$
and internally $\Delta_1, \ldots, \Delta_i$) with
a potentially discontinuous distinguished subconfiguration $\Gamma$;
similarly, where $\Xi$ is a zone of sort $i$,
the hyperoccurrence notation
$\Xi\langle\Gamma\rangle$ abbreviates
$\zeta; \Delta_0(\Gamma\otimes\langle\Delta_1: \ldots:  \Delta_i\rangle)$
where $\Xi=\zeta; \Delta$,
i.e.\ a context zone $\zeta; \Delta$ (which is externally $\zeta; \Delta_0$
and internally $\Delta_1, \ldots, \Delta_i$) with
a potentially discontinuous distinguished subconfiguration $\Gamma$.

Where $\Delta$ is a configuration of sort $i>0$ and $\Gamma$ is a configuration, the
$k$\/th \emph{metalinguistic intercalation} $\Delta\smwrap{k}\Gamma$, $1\le k\le i$,  is given by:
\disp{$
\Delta\smwrap{k}\Gamma =_{df} \Delta\otimes\langle\underbrace{\sep: \ldots: \sep}_{k{-}1\ 1\mbox{\footnotesize 's}}: \Gamma: \underbrace{\sep: \ldots: \sep}_{i{-}k\ 1\mbox{\footnotesize 's}}\rangle$
}
\noindent
i.e.\ the $k$th metalinguistic intercalation
$\Delta\smwrap{k}\Gamma$ is the configuration
resulting from replacing by $\Gamma$ the $k$\/th separator
in $\Delta$.

A semantically labelled sequent is a sequent in which the antecedent syntactic type occurrences
$A_1$, \ldots, $A_n$ are labelled by distinct variables $x_1, \ldots, x_n$ of semantic types 
$T(A_1), \ldots, T(A_n)$ respectively,
and the succedent type $A$ is labelled by a term of type $T(A)$ with free variables drawn
from $x_1, \ldots, x_n$.

\section{The identity rules}

The identity axiom is (\ref{idex}a);
since we adopt the convention that empty stoups can be omitted,
we write (\ref{idex}b):
\
\disp{ a.\tb 
\prooftree
\justifies
\emptyset; \vect{P}\ass x\yields P\ass x
\using \mathit{id}
\endprooftree\label{idex}
\tb b.\tb 
\prooftree
\justifies
\vect{P}\ass x\yields P\ass x
\using \mathit{id}
\endprooftree}
This states that a type is derivable from itself,
asserting the reflexivity of set inclusion and the derivability relation.
The Cut rule is:
\disp{$
\prooftree
\zeta_1; \Gamma\yields A\tb \zeta_2; \Delta\langle\vect{A}\rangle\yields B
\justifies
\zeta_1\mun\zeta_2; \Delta\langle\Gamma\rangle\yields B
\using {\it Cut}
\endprooftree$}
This states the contextual generalisation of the transitivity
of set inclusion and the derivability relation.

\section{Continuous multiplicatives}

The continuous multiplicatives  \{$/$, $\bsl$, $\product$, $I$\}
of Figure~\ref{slcmult},
Lambek (1958\cite{lambek:mathematics};
1988\cite{lambek:88}), are defined in relation to concatenation; they
are the basic means of categorial (sub)categorization.

\begin{figure}[ht]
\begin{center}
$
\begin{array}{lc}
1. &\prooftree
\zeta_1\sass\Gamma\yields B\ass\psi \tb
\zeta_2\sass\Delta\langle\vect{C}\ass z\rangle\yields D\ass\omega
\justifies
\zeta_1\mun\zeta_2\sass\Delta\langle\vect{C/B}\ass x, \Gamma\rangle\yields D\ass\omega\subst{(x\ \psi)/z}
\using / L
\endprooftree \tb
\prooftree
\zeta\sass\Gamma, \vect{B}\ass y\yields C\ass\chi
\justifies
\zeta\sass\Gamma\yields C/B\ass\lambda y\chi
\using / R
\endprooftree
\\\\
2. &
\prooftree
\zeta_1\sass\Gamma\yields A\ass\phi \tb
\zeta_2\sass\Delta\langle\vect{C}\ass z\rangle\yields D\ass\omega
\justifies
\zeta_1\mun\zeta_2\sass\Delta\langle\Gamma, \vect{A\bsl C}\ass y\rangle\yields D\ass\omega\subst{(y\ \phi)/z}
\using \bsl L
\endprooftree \tb
\prooftree
\zeta\sass\vect{A}\ass x, \Gamma\yields C\ass\chi
\justifies
\zeta\sass\Gamma\yields A\bsl C\ass\lambda x\chi
\using \bsl R
\endprooftree
\\\\
3. \tb & \prooftree
\Xi\langle\vect{A}\ass x, \vect{B}\ass y\rangle\yields D\ass\omega
\justifies
\Xi\langle\vect{A\product B}\ass z\rangle\yields D\ass\omega\subst{\pi_1 z/x, \pi_2 z/y}
\using \product L
\endprooftree \tb
\prooftree
\zeta_1\sass\Gamma_1\yields A\ass\phi\tb\zeta_2\sass\Gamma_2\yields B\ass\psi
\justifies
\zeta_1\mun\zeta_2\sass\Gamma_1, \Gamma_2\yields A\product B\ass(\phi, \psi)
\using \product R
\endprooftree
\\\\
4. &
\prooftree
\Xi\langle\Lambda\rangle\yields A\ass\phi
\justifies
\Xi\langle\vect{I}\ass x\rangle\yields A\ass\phi
\using IL
\endprooftree\tb
\prooftree
\justifies
\Lambda\yields I\ass\zero
\using IR
\endprooftree
\end{array}
$
\end{center}
\caption{Continuous multiplicative rules}
\label{slcmult}
\end{figure}

\clearpage

The directional divisions over, $/$, and under, $\bsl$, are exemplified
by assignments such as $\syncnst{the}\ass N/\CN\ass\iota$ for
$\syncnst{the man}\ass N\ass(\iota\ \semcnst{man})$ and $\syncnst{sings}\ass N\bsl S\ass\semcnst{sing}$ for
$\syncnst{John sings}\ass S\ass(\semcnst{sing}\ \semcnst{j})$,
and $\syncnst{loves}\ass$ $(N\bsl S)/N\ass\semcnst{love}$ for 
$\syncnst{John loves Mary}\ass S\ass((\semcnst{love}\ \semcnst{m})\ \semcnst{j})$.
Hence, for \syncnst{the man}:
\disp{$
\prooftree
\CN\yields\CN\tb N\yields N
\justifies
N/\CN, \CN\yields N
\using /L
\endprooftree$}
And for \syncnst{John sings} and \syncnst{John loves Mary}:
\disp{$
\prooftree
N\yields N\tb S\yields S
\justifies
N, N\bsl S\yields S
\using \bsl L
\endprooftree
\tb
\prooftree
N\yields N
\prooftree
N\yields N\tb S\yields S
\justifies
N, N\bsl S\yields S
\using \bsl L
\endprooftree
\justifies
N, (N\bsl S)/N, N\yields S
\using /L
\endprooftree
$}
The continuous product $\product$ is exemplified by a \scare{small clause}
assignment
$$\pcnst{considers}\ass (N\bsl S)/(N\product(\CN/\CN))\ass
\lambda x(\semcnst{consider}\ ((\pi_2 x\ \lambda y[y=\pi_1 x])\ \pi_1 x))$$
for $\syncnst{John considers Mary socialist}\ass S\ass$ $((\semcnst{consider}$ $((\semcnst{socialist}\ \lambda y[y=\semcnst{m}])\ \semcnst{m}))\ \semcnst{j})$.
\disp{$
\prooftree
\prooftree
N\yields N\tb 
\prooftree
\prooftree
\CN\yields\CN\tb\CN\yields\CN
\justifies
\CN/\CN, \CN\yields \CN
\using /L
\endprooftree
\justifies
\CN/\CN\yields \CN/\CN
\using /R
\endprooftree
\justifies
N, \CN/\CN\yields N\product(\CN/\CN)
\using \product R
\endprooftree
\prooftree
N\yields N\tb S\yields S
\justifies
N, N\bsl S\yields S
\using \bsl L
\endprooftree
\justifies
N, (N\bsl S)/(N\product(\CN/\CN)), N, \CN/\CN\yields S
\using /L
\endprooftree$}
Of course this use of product is not essential: we could just as well
have used  $((N\bsl S)/(\CN/\CN))/N$ since in general we have both
$A/(C\product B)\yields (A/B)/C$ (currying) and
$(A/B)/C\yields A/(C\product B)$ (uncurrying). For essential use of
product, in the antecedent, see the treatment of past participles
in Morrill (2011\cite{morrill:oxford}), of which the assignment to
\lingform{loved} in Chapter~\ref{lexchap} here is a discontinuous generalisation.\footnote{The continuous
unit can be used together with additive disjunction to express
the optionality of a complement as in $\syncnst{eats}\ass (N\bsl S)/(N\adisj I)$ for
$\syncnst{John eats fish}\ass S$ and $\syncnst{John eats}\ass S$.
It can also we used in conjunction with difference to prevent the null string
being supplied as argument to an intensifier as in
$\syncnst{very}\ass(\CN/\CN)/((\CN/\CN)-I)$ for
$\syncnst{very tall man}\ass \CN$ but $\unacc \syncnst{very man}\ass\CN$.}

\section{Discontinuous multiplicatives}

\label{dismultsect}

The discontinuous multiplicatives \{$\scircum{}$, $\sinfix{}$, $\swprod{}$, $J$\}
of Figure~\ref{sldmult},
the displacement connectives,
Morrill and Valent\'{\i}n (2010\cite{mv:disp}),
Morrill, Valent\'{\i}n and Fadda  (2011\cite{mvf:tdc}),
are defined in relation to intercalation; they are the basic means
of obtaining displacement effects.

\newpage

\begin{figure}[ht]
\begin{center}
$
\begin{array}{lc}
5. &
\prooftree
\zeta_1\sass\Gamma\yields B\ass\psi \tb
\zeta_2\sass\Delta\langle\vect{C}\ass z\rangle\yields D\ass\omega
\justifies
\zeta_1\mun\zeta_2\sass\Delta\langle\vect{C\scircum{k} B}\ass x\smwrap{k}\Gamma\rangle\yields D\ass\omega\subst{(x\ \psi)/z}
\using \scircum{k} L
\endprooftree \tb
\prooftree
\zeta\sass\Gamma\smwrap{k}\vect{B}\ass y\yields C\ass\chi
\justifies
\zeta\sass\Gamma\yields C\scircum{k} B\ass\lambda y\chi
\using \scircum{k} R
\endprooftree
\\\\
6. & 
\prooftree
\zeta_1\sass\Gamma\yields A\ass\phi \tb
\zeta_2\sass\Delta\langle\vect{C}\ass z\rangle\yields D\ass\omega
\justifies
\zeta_1\mun\zeta_2\sass\Delta\langle\Gamma\smwrap{k}\vect{A\sinfix{k} C}\ass y\rangle\yields D\ass\omega\subst{(y\ \phi)/z}
\using \sinfix{k} L
\endprooftree \tb
\prooftree
\zeta\sass\vect{A}\ass x\smwrap{k}\Gamma\yields C\ass\chi
\justifies
\zeta\sass\Gamma\yields A\sinfix{k} C\ass\lambda x\chi
\using \sinfix{k} R
\endprooftree
\\\\
7.\tb &
\prooftree
\Xi\langle\vect{A}\ass x\smwrap{k}\vect{B}\ass y\rangle\yields D\ass\omega
\justifies
\Xi\langle\vect{A\swprod{k} B}\ass z\rangle\yields D\ass\omega\subst{\pi_1 z/x, \pi_2 z/y}
\using \swprod{k} L
\endprooftree \tb
\prooftree
\zeta_1\sass\Gamma_1\yields A\ass\phi\tb\zeta_2\sass\Gamma_2\yields B\ass\psi
\justifies
\zeta_1\mun\zeta_2\sass\Gamma_1\smwrap{k}\Gamma_2\yields A\swprod{k} B
\using \swprod{k} R
\endprooftree
\\\\
8. &
\prooftree
\Xi\langle\sep\rangle\yields A\ass\phi
\justifies
\Xi\langle\vect{J}\ass x\rangle\yields A\ass\phi
\using JL
\endprooftree\tb
\prooftree
\justifies
\sep\yields J\ass\zero
\using JR
\endprooftree
\end{array}
$
\end{center}
\caption{Discontinuous multiplicative rules}
\label{sldmult}
\end{figure}

When the value of the $k$ subscript is one it may be omitted,
i.e.\ it defaults to one.
Circumfixation, 
$\scircum{}$,
is exemplified by a discontinuous idiom assignment
$\syncnst{gives}\add\sep\add\syncnst{the}\add\syncnst{cold}\add\syncnst{shoulder}\ass$ 
$(N\bsl S)\scircum{}N\ass \semcnst{shun}$ for
$\syncnst{Mary gives the man the cold shoulder}\ass S\ass((\semcnst{shun}\ (\iota\ \semcnst{man}))\ 
\semcnst{m})$:
\disp{$
\prooftree
\prooftree
\CN\yields\CN\tb N\yields N
\justifies
N/\CN, \CN\yields N
\using /L
\endprooftree
\prooftree
N\yields N\tb S\yields S
\justifies
N, N\bsl S\yields S
\using \bsl L
\endprooftree
\justifies
N, (N\bsl S)\circum{}N\{N/\CN, \CN\}\yields S
\using \circum{}L
\endprooftree$}
Infixation, 
$\sinfix{}$,
and extractation together are exemplified by a quantifier assignment
$$\syncnst{everyone}\ass (S\scircum{} N)\sinfix{}S\ass\lambda x\forall y[(\semcnst{person}\ y)\rightarrow
(x\ y)]$$
simulating Montague's S14 quantifying in:
\disp{$
\prooftree
\prooftree
\ldots, N, \ldots\yields S
\justifies
\ldots, \sep, \ldots\yields S\circum{}N
\using \circum{}R
\endprooftree
\prooftree
\justifies
S\yields S
\using {\it id}
\endprooftree
\justifies
\ldots, (S\circum{} N)\infix{} S, \ldots\yields S
\using \infix{}L
\endprooftree$} 
Circumfixation and discontinuous product,
$\swprod{}$,
are illustrated in an assignment to a relative pronoun
$$\syncnst{that}\ass(\CN\bsl\CN)/((S\circum{}N)\swprod{}I)\ass\lambda x\lambda y\lambda z[(y\ z)\wedge(\pi_1 x\ z)]$$
allowing both peripheral and medial extraction,
$\syncnst{that John likes}\ass\CN\bsl\CN\ass\lambda y\lambda z[(y\ z)\wedge((\semcnst{like}\ z)\ \semcnst{j})]$ and
$\syncnst{that John saw today}\ass\CN\bsl\CN\ass\lambda y\lambda z[(y\ z)\wedge(\semcnst{today}\ ((\semcnst{see}\ z)\ \semcnst{j}))]$:\footnote{Use of the discontinuous product unit, 
$J$,
in conjunction with difference is illustrated in a pronoun assignment
$\syncnst{him}\ass(((S\scircum{}N)\scircum{2}N)-(J\product((N\bsl S)\scircum{}N))\sinfix{2}(S\scircum{}N)$
blocking a subject antecedent (a Principle B effect).}
\disp{$
\prooftree
\prooftree
\prooftree
N, (N\bsl S)/N, N\yields S
\justifies
N, (N\bsl S)/N, \sep\yields S\circum{}N
\using \circum{}R
\endprooftree
\prooftree
\justifies
\yields I
\using IL
\endprooftree
\justifies
N, (N\bsl S)/N\yields (S\circum{}N)\wprod I
\using \wprod R
\endprooftree
\CN\bsl\CN\yields \CN\bsl\CN
\justifies
(\CN\bsl\CN)/((S\circum N)\wprod I), N, (N\bsl S)/N\yields \CN\bsl\CN
\using /L
\endprooftree
$}
\disp{$
\prooftree
\prooftree
\prooftree
N, (N\bsl S)/N, N, S\bsl S\yields S
\justifies
N, (N\bsl S)/N, \sep, S\bsl S\yields S\circum{}N
\using \circum{}R
\endprooftree
\prooftree
\justifies
\yields I
\using IL
\endprooftree
\justifies
N, (N\bsl S)/N, S\bsl S\yields (S\circum{}N)\wprod I
\using \wprod R
\endprooftree
\CN\bsl\CN\yields \CN\bsl\CN
\justifies
(\CN\bsl\CN)/((S\circum N)\wprod I), N, (N\bsl S)/N, S\bsl S\yields \CN\bsl\CN
\using /L
\endprooftree
$}

\section{Additives}

The additive conjunction and disjunction \{$\aconj$, $\adisj$\}
of Figure~\ref{sladd},
Lambek (1961\cite{lambek:61}), 
Morrill (1990\cite{morrill:galt}), and Kanazawa (1992\cite{kanazawa:additives}),
capture polymorphism.

\begin{figure}[ht]
\begin{center}
$
\begin{array}{lc}
9.\tb & 
\prooftree
\Xi\langle\vect{A}\ass x\rangle\yields C\ass\chi
\justifies
\Xi\langle\vect{A\aconj B}\ass z\rangle\yields C\ass\chi\subst{\pi_1 z/x}
\using \aconj L_1
\endprooftree\tb
\prooftree
\Xi\langle\vect{B}\ass y\rangle\yields C\ass\chi
\justifies
\Xi\langle\vect{A\aconj B}\ass z\rangle\yields C\ass\chi\subst{\pi_2 z/y}
\using \aconj L_2
\endprooftree
\\\\
&\prooftree
\Xi\yields A\ass\phi\tb\Xi\yields B\ass\psi
\justifies
\Xi\yields A\aconj B\ass(\phi, \psi)
\using \aconj R
\endprooftree
\\\\
10. &
\prooftree
\Xi\langle\vect{A}\ass x\rangle\yields C\ass\chi_1\tb\Xi\langle\vect{B}\ass y\rangle\yields C\ass\chi_2
\justifies
\Xi\langle\vect{A\adisj B}\ass z\rangle\yields C\ass z\casearrow x.\chi_1; y.\chi_2
\using \adisj L
\endprooftree
\\\\
& \prooftree
\Xi\yields A\ass\phi
\justifies
\Xi\yields A\adisj B\ass\iota_1\phi
\using \adisj R_1
\endprooftree\tb
\prooftree
\Xi\yields B\ass\psi
\justifies
\Xi \yields A\adisj B\ass\iota_2\psi
\using \adisj R_2
\endprooftree
\end{array}
$
\end{center}
\caption{Additive rules}
\label{sladd}
\end{figure}


For example the additive conjunction $\aconj$ can be used for
$\syncnst{rice}\ass N\aconj\CN\ass((\semcnst{gen}\ \semcnst{rice}), \semcnst{rice})$ as in $\syncnst{rice grows}\ass S\ass(\semcnst{grow}\ (\semcnst{gen}\ \semcnst{rice}))$
and $\syncnst{the rice grows}\ass S\ass(\semcnst{grow}\ (\iota\ \semcnst{rice}))$:
\disp{$
\prooftree
\prooftree
N\yields N
\justifies
N\aconj\CN\yields N
\using \aconj L_1
\endprooftree
S\yields S
\justifies
N\aconj\CN, N\bsl S\yields S
\using \bsl L
\endprooftree
\tb
\prooftree
N/\CN, \CN, N\bsl S\yields S
\justifies
N/\CN, N\aconj\CN, N\bsl S\yields S
\using \aconj L_2
\endprooftree
$}
The additive disjunction $\adisj$ can be used for $\syncnst{is}\ass(N\bsl S)/(N\adisj(\CN/\CN))\ass
\lambda x\lambda yx\rightarrow z.[z=y]; w.((w\ \lambda u[y=u])\ y)$
as in $\syncnst{Tully is Cicero}\ass S\ass[t=c]$ and $\syncnst{Tully is humanist}\ass S\ass
((\semcnst{humanist}\ \lambda u[t=u])\ \semcnst{t})$:
\disp{$
\prooftree
\prooftree
N\yields N
\justifies
N\yields N\adisj(\CN/\CN)
\using \adisj R_1
\endprooftree
N\bsl S\yields N\bsl S
\justifies
(N\bsl S)/(N\adisj(\CN/\CN)), N\yields N\bsl S
\using /L
\endprooftree
\tb
\prooftree
\prooftree
\CN/\CN\yields \CN/\CN
\justifies
\CN/\CN\yields N\adisj(\CN/\CN)
\using \adisj R_2
\endprooftree
N\bsl S\yields N\bsl S
\justifies
(N\bsl S)/(N\adisj(\CN/\CN)), \CN/\CN\yields N\bsl S
\using /L
\endprooftree
$}

\section{Quantifiers}

The 1st order quantifiers \{$\bigwedge$, $\bigvee$\}
of Figure~\ref{slquant},
Morrill (1994\cite{morrill:tlg}),
have application to features.

\newpage

\begin{figure}[ht]
\begin{center} 
 $
 \begin{array}{lc}
11.\tb & \prooftree
 \Xi\langle \vect{A[t/v]}\ass x\rangle\yields B\ass\psi
 \justifies
\Xi\langle\vect{\bigwedge vA}\ass z\rangle\yields B\ass\psi\subst{(z\ t)/x}
 \using \bigwedge L
 \endprooftree\tb
 \prooftree
 \Xi\yields A[a/v]\ass\phi
 \justifies
 \Xi\yields \bigwedge vA\ass\lambda v\phi
 \using \bigwedge R^\dagger
 \endprooftree
\\\\
12. &
\prooftree
\Xi\langle\vect{A[a/v]}\ass x\rangle\yields B\ass\psi
\justifies
\Xi\langle\vect{\bigvee vA}\ass z\rangle\yields B\ass\psi\subst{\pi_2 z/x}
\using \bigvee L^\dagger
\endprooftree\tb
\prooftree
\Xi\yields A[t/v]\ass\phi
\justifies
\Xi\yields\bigvee vA\ass(t, \phi)
\using \bigvee R
\endprooftree
\end{array}
$
\end{center}
\caption{Quantifier rules; $^\dagger$ indicates that there is no $a$ in the conclusion}
\label{slquant}
\end{figure}

For example, we can generalise over singular and plural number in
$\syncnst{sheep}\ass$ $\bigwedge{}n\CN n\ass\lambda n(n\ \semcnst{sheep})$ for
$\syncnst{the sheep grazes}\ass S\ass(\semcnst{graze}\ (\iota\ (\semcnst{sg}\ \semcnst{sheep})))$ and $\syncnst{the sheep graze}\ass S\ass(\semcnst{graze}\ (\iota\ (\semcnst{pl}\ \semcnst{sheep})))$:
\disp{$\small
\prooftree
\CN{\it sg}\yields \CN{\it sg}
\justifies
\bigwedge n\CN n\yields \CN{\it sg}
\using \bigwedge L
\endprooftree
\tab
\prooftree
\CN{\it pl}\yields \CN{\it pl}
\justifies
\bigwedge n\CN n\yields \CN{\it pl}
\using \bigwedge L
\endprooftree
$}
And we can express a past, present or future tense finite sentence complement:
$$\syncnst{said}\ass (N\bsl S)/\bigvee{}tSf(t)\ass\lambda x(\semcnst{say}\ (\pi_1 x\ \pi_2 x))$$
in
$$\syncnst{John said Mary walked}\ass S\ass((\semcnst{say}\ (\semcnst{past}\ (\semcnst{walk}\ \semcnst{m})))\ \semcnst{j}),$$
$$\syncnst{John said Mary walks}\ass S\ass((\semcnst{say}\ (\semcnst{pres}\ (\semcnst{walk}\ \semcnst{m})))\ \semcnst{j})$$ and
$$\syncnst{John said Mary will walk}\ass S\ass((\semcnst{say}\ (\semcnst{fut}\ (\semcnst{walk}\ \semcnst{m})))\ \semcnst{j}):$$
\disp{$\small
\prooftree
Sf({\it past})\yields Sf({\it past})
\justifies
Sf({\it past})\yields \bigvee t Sf(t)
\using \bigvee R
\endprooftree
\tb
\prooftree
Sf({\it pres})\yields Sf({\it pres})
\justifies
Sf({\it pres})\yields \bigvee t Sf(t)
\using \bigvee R
\endprooftree
\tb
\prooftree
Sf({\it fut})\yields Sf({\it fut})
\justifies
Sf({\it fut})\yields \bigvee t Sf(t)
\using \bigvee R
\endprooftree
$}

\section{Normal modalities}

With respect to the normal modalities \{$\mymod$, $\emod$\} of Figure~\ref{slmod},
Morrill (1990\cite{morrill:ib90}) and
Moortgat (1997\cite{moortgat:handbook}),
the universal has application to intensionality.
\begin{figure}[ht]
\begin{center}
$
\begin{array}{lcc}
13. & \prooftree
\Xi\langle\vect{A}\ass x\rangle\yields B\ass\psi
 \justifies
\Xi\langle\vect{\mymod A}\ass z\rangle\yields B\ass\psi\subst{\dn z/x}
 \using \mymod L
 \endprooftree 
 &
\prooftree
 \diagmod\Xi\yields A\ass\phi
 \justifies
 \diagmod\Xi\yields \mymod A\ass\up\phi
 \using \mymod R
 \endprooftree
 \\\\
 14.\tb &
 \prooftree
\diagmod\Xi\langle\vect{A}\ass x\rangle\yields \diagemod B\ass\psi
 \justifies
\diagmod\Xi\langle\vect{\emod A}\ass z\rangle\yields \diagemod B\ass\psi\subst{\dnc z/x}
 \using \emod L
 \endprooftree 
 &
\prooftree
 \Xi\yields A\ass\phi
 \justifies
 \Xi\yields \emod A\ass\upc\phi
 \using \emod R
 \endprooftree
 \end{array}
 $
 \end{center}
\caption{Normal modality rules; $\diagmod/\diagemod$ marks a structure all the types of
which have principal connective a box/diamond}
\label{slmod}
\end{figure}
For example,
for a propositional attitude verb we can have an assignment
$\syncnst{believes}\ass \mymod((N\bsl S)/\mymod S)\ass\semcnst{believe}$
with a modality outermost since the word,
like all words,
has a sense,
and its sentential complement is an intensional domain,
but its subject is not.\footnote{The existential normal modality has no application known to us,
but we think a candidate may be dynamic semantics,
when the indices are interpreted like discourse markers.}

\section{Bracket modalities}

The bracket modalities \{$\abrack$, $\mybrack$\} of Figure~\ref{slbrmod},
Morrill (1992\cite{morrill:92}) and
Moortgat (1995\cite{moortgat:95}),
have application to syntactical domains
such as
 islands. They are semantically transparent.
\begin{figure}[ht]
\begin{center}
$
\begin{array}{lcc}
15.\tb & \prooftree
\Xi\langle \vect{A}\ass x\rangle\yields B\ass\psi
\justifies
\Xi\langle[\vect{\abrack A}\ass x]\rangle\yields B\ass\psi
\using \abrack L
\endprooftree
 &
\prooftree
[\Xi]\yields A\ass\phi
\justifies
\Xi\yields \abrack A\ass\phi
\using \abrack R
\endprooftree
\\\\
16. & 
\prooftree
\Xi\langle[\vect{A}\ass x]\rangle\yields B\ass\psi
\justifies
\Xi\langle\vect{\mybrack A}\ass x\rangle\yields B\ass\psi
\using \mybrack L
\endprooftree
&
\prooftree
\Xi\yields A\ass\phi
\justifies
[\Xi]\yields \mybrack A\ass\phi
\using \mybrack R
\endprooftree
\end{array}$
 \end{center}
\caption{Bracket modality rules}
\label{slbrmod}
\end{figure}
For example,
$\syncnst{walks}\ass\mybrack{}N\bsl S\ass$ $\semcnst{walk}$ for the sentential subject condition,
and
$$\syncnst{before}\ass\abrack{}(\VP\bsl\VP)/\VP\ass\lambda x\lambda y\lambda z((\semcnst{before}\ (x\ z))\ (y\ z))$$
for the adverbial island constraint,
which are weak islands, 
and can contain parasitic gaps, see the next section;
for a strong island such as a coordinate structure, which cannot contain a parasitic gap,
we define doubly bracketed strong islands --- $\syncnst{and}\ass(S\bsl \abrack\abrack S)/S\ass
\lambda x\lambda y[y\wedge x]$.
\disp{
a.\tb
\prooftree
\prooftree
N\yields N
\justifies
[N]\yields\mybrack N
\using \mybrack R
\endprooftree
S\yields S
\justifies
[N], \mybrack N\bsl S
\using \bsl L
\endprooftree
\tb b.\tb
\prooftree
S\yields S\tb
\prooftree
S\yields S\tb
\prooftree
\prooftree
S\yields S
\justifies
[\abrack S]\yields S
\using \abrack L
\endprooftree
\justifies
[[\abrack\abrack S]]\yields S
\using \abrack L
\endprooftree
\justifies
[[S, S\bsl\abrack\abrack S]]\yields S
\using \bsl S
\endprooftree
\justifies
[[S, (S\bsl\abrack\abrack S)/S]]\yields S
\using / S
\endprooftree}
Without bracket modalities we do not capture the Coordinate Structure
Constraint (CSC) that coordinate structures are islands to left extraction
such as relativisation,
e.g.
$$\unacc \lingform{man that John laughs and \lingform{Mary likes}}$$
would be derived from the theorem
$$\CN, (\CN\bsl\CN)/(S/N), N, N\bsl S, (S\bsl S)/S, N, (N\bsl S)/N\yields \CN.$$

\section{Subexponentials}

The exponentials \{$\univexp$, $\exstexp$\} of Figure~\ref{exp2},
Girard (1987\cite{girard:87}),
Barry et al.\ (1991\cite{bhlm:91}),
Morrill (1994\cite{morrill:tlg}, 2002\cite{morrill:para02}),
2011\cite{morrill:oxford}) and
Morrill and Valent\'{\i}n (2015\cite{cctlg:nl},
have application to sharing.
\begin{figure}[ht]
\begin{center}
$
\begin{array}{lc}
17. & 
 \prooftree
\Xi(\zeta\mun\{A\ass x\}\sass\Gamma_1, \Gamma_2)\yields B\ass\psi
 \justifies
\Xi(\zeta\sass\Gamma_1, \univexp A\ass x, \Gamma_2)\yields B\ass\psi
 \using \univexp L
 \endprooftree \tb
\prooftree
\zeta\sass \yields A\ass\phi
 \justifies
\zeta; \yields \univexp A\ass\phi
 \using \univexp R, \zeta \neq \emptyset
 \endprooftree \\\\
&  \prooftree
\Xi(\zeta\sass\Gamma_1, A\ass x, \Gamma_2)\yields B\ass\psi
 \justifies
\Xi(\zeta\mun\{A\ass x\}\sass\Gamma_1, \Gamma_2)\yields B\ass\psi
 \using \univexp P
 \endprooftree
 \\\\
 & \commentout{ \prooftree
 \Xi(\zeta\mun\{\mbox{\fbox{A}}\ass x\}; \Gamma_1, [\{A\ass y\}; \Gamma_2], \Gamma_3)\yields B\ass\psi
 \justifies
 \Xi(\zeta\mun\{\mbox{\fbox{A}}\ass x\}; \Gamma_1, [\nil; [\nil; \Gamma_2]], \Gamma_3)\yields B\ass\psi\subst{x/y}
 \using C
 \endprooftree}
 \prooftree
 \Xi(\zeta\mun\{A_0\ass x_0, \ldots, A_n\ass x_n\}; \Gamma_1, [\{A_0\ass y_0, \ldots, A_n\ass y_n\}; \Gamma_2], \Gamma_3)\yields B\ass\psi
 \justifies
 \Xi(\zeta\mun\{A_0\ass x_0, \ldots A_n\ass x_n\}; \Gamma_1, \Gamma_2, \Gamma_3)\yields B\ass\psi\subst{x_0/y_0, \ldots, x_n/y_n}
 \using \univexp C
 \endprooftree
 \\\\
 18. \tb &
 \prooftree
 \Xi(A\ass x_1)\yields D\ass\omega([x_1])\tb
 \Xi(A\ass x_1, A\ass x_2)\yields D\ass\omega([x_1, x_2])\tb
 \cdots
 \justifies
 \Xi(\exstexp A\ass x)\yields D\ass\omega(x)
 \using \exstexp L
 \endprooftree
 \\\\
&\prooftree
\Xi\yields A\ass\phi
\justifies
\Xi\yields \exstexp A\ass[\phi]
\using \exstexp R
\endprooftree
\tb
\prooftree
\zeta\sass\Gamma\yields A\ass\phi\tb\zeta'\sass\Delta\yields \exstexp A\ass\psi
\justifies
\zeta\mun\zeta'\sass\Gamma, \Delta\yields \exstexp A\ass[\phi|\psi]
\using \exstexp M
\endprooftree
\end{array}
$
 \end{center}
\caption{Exponential rules}
\label{exp2}
\end{figure}
Using the universal exponential, 
\univexp,
we can assign to a relative pronoun thus:
$$\syncnst{that}\ass(\CN\bsl\CN)/(S/\univexp N)\ass\lambda x\lambda y\lambda z[(y\ z)\wedge(x\ z)]$$ allowing parasitic extraction,
Morrill (2011\cite{morrill:oxford}),
Morrill and Valent\'{\i}n (2015\cite{cctlg:nl}),
such as
$$\syncnst{paper that John filed without reading}\ass\CN\ass
\lambda z[(\semcnst{paper}\ z)\wedge[((\semcnst{file}\ z)\ \semcnst{j})\wedge\neg((\semcnst{read}\ z)\ \semcnst{j})]],$$
where parasitic gaps can appear only in (weak) islands,
but can be iterated in (weak) subislands (see Chapter~\ref{relchap}).\footnote{The universal exponential
enjoys the laws $\univexp A\yields A$ and $\univexp A\yields\univexp\univexp A$
of {\bf S4} modal logic, which are derived as follows:

(i)\tb a.\tb
$
\prooftree
\prooftree
A\yields A
\justifies
A; \yields A
\using \univexp P
\endprooftree
\justifies
\univexp A\yields A
\using \univexp L
\endprooftree
\tb b.\tb
\prooftree
\prooftree
\prooftree
\prooftree
A\yields A
\justifies
A; \yields A
\using \univexp P
\endprooftree
\justifies
A; \yields \univexp A
\using \univexp R
\endprooftree
\justifies
A; \yields \univexp\univexp A
\using \univexp R
\endprooftree
\justifies
A \yields \univexp\univexp A
\using \univexp L
\endprooftree
$
}
Using the existential exponential, 
\exstexp,
we can assign a coordinator type
$\syncnst{and}\ass(\exstexp S\bsl S)/S\ass(\alpha^+\ \semcnst{and})$ allowing iterated coordination
as in $\syncnst{I came, I saw and I conquered}\ass N\ass[(\semcnst{come}\ 
\semcnst{i})\wedge[(\semcnst{see}\ \semcnst{i})\wedge(\semcnst{conquer}\ 
\semcnst{i})]]$ where $\alpha^+$ is a map apply function; see Chapter~\ref{coordchap}.

\section{Semantically inactive connectives}

\begin{figure}
\begin{center}
$
\begin{array}{lcc}
19.\tb & \prooftree
\zeta\sass  \Gamma\yields B\ass\psi \tb
\Xi(\zeta'\sass  \Delta_1\langle\vect{C}\ass z\rangle, \Delta_2)\yields D\ass\omega
\justifies
\Xi(\zeta\mun\zeta'\sass  \Delta_1, \langle\vect{C\argover B}\ass x, \Gamma\rangle, \Delta_2)\yields D\ass\omega\{\zero/z\}
\using \argover L
\endprooftree
&
\prooftree
\zeta\sass \Gamma, \vect{B}\ass y\yields C\ass\chi
\justifies
\zeta\sass \Gamma\yields C\argover B\ass\zero
\using \argover R
\endprooftree
\\\\
20. & \prooftree
\zeta\sass  \Gamma\yields A\ass\phi \tb
\Xi(\zeta'\sass  \Delta_1, \langle\vect{C}\ass z\rangle, \Delta_2)\yields D\ass\omega
\justifies
\Xi(\zeta\mun\zeta'\sass  \Delta_1, \langle\Gamma, \vect{A\fununder C}\ass y\rangle, \Delta_2)\yields D\ass\omega\{y/z\}
\using \fununder L
\endprooftree
&
\prooftree
\zeta\sass  \vect{A}\ass x, \Gamma\yields C\ass\chi
\justifies
\zeta\sass \Gamma\yields A\fununder C\ass\chi
\using \fununder R
\endprooftree\\\\
21. &
\prooftree
\Xi\langle\vect{A}\ass x, \vect{B}\ass y\rangle\yields D\ass\omega
\justifies
\Xi\langle\vect{A\leftiprod B}\ass z\rangle\yields D\ass\omega\{z/y\}
\using \leftiprod L
\endprooftree &
\prooftree
\zeta\sass  \Gamma_1\yields A\ass\phi\tb\zeta'\sass  \Gamma_2\yields B\ass\psi
\justifies
\zeta\mun\zeta'\sass  \Gamma_1, \Gamma_2\yields A\leftiprod B\ass\psi
\using \leftiprod R
\endprooftree\\\\
22. &
\prooftree
\zeta\sass  \Gamma\yields B\ass\psi \tb
\Xi(\zeta'\sass  \Delta_1\langle\vect{C}\ass z\rangle, \Delta_2)\yields D\ass\omega
\justifies
\Xi(\zeta\mun\zeta'\sass  \Delta_1, \langle\vect{C\funover B}\ass x, \Gamma\rangle, \Delta_2)\yields D\ass\omega\{x/z\}
\using \funover L
\endprooftree
&
\prooftree
\zeta\sass \Gamma, \vect{B}\ass y\yields C\ass\chi
\justifies
\zeta\sass \Gamma\yields C\funover B\ass\chi
\using \funover R
\endprooftree
\\\\
23. &\prooftree
\zeta\sass  \Gamma\yields A\ass\phi \tb
\Xi(\zeta'\sass  \Delta_1, \langle\vect{C}\ass z\rangle, \Delta_2)\yields D\ass\omega
\justifies
\Xi(\zeta\mun\zeta'\sass  \Delta_1, \langle\Gamma, \vect{A\argunder C}\ass y\rangle, \Delta_2)\yields D\ass\omega\{\zero/z\}
\using \argunder L
\endprooftree
&
\prooftree
\zeta\sass  \vect{A}\ass x, \Gamma\yields C\ass\chi
\justifies
\zeta\sass \Gamma\yields A\argunder C\ass\zero
\using \argunder R
\endprooftree\\\\
24. &
\prooftree
\Xi\langle\vect{A}\ass x, \vect{B}\ass y\rangle\yields D\ass\omega
\justifies
\Xi\langle\vect{A\rightiprod B}\ass z\rangle\yields D\ass\omega\{z/x\}
\using \rightiprod L
\endprooftree 
&
\prooftree
\zeta\sass  \Gamma_1\yields A\ass\phi\tb\zeta'\sass  \Gamma_2\yields B\ass\psi
\justifies
\zeta\mun\zeta'\sass  \Gamma_1, \Gamma_2\yields A\rightiprod B\ass\phi
\using \rightiprod R
\endprooftree\
\end{array}$
\end{center}
\caption{Semantically inactive continuous multiplicative rules}
\label{csim}
\end{figure}

\begin{figure}[ht]
\begin{center}
$
\begin{array}{lcc}
25.\tb &\prooftree
\zeta\sass  \Gamma\yields B\ass\psi \tb
\Xi(\zeta'\sass  \Delta_1\langle\vect{C}\ass z\rangle, \Delta_2)\yields D\ass\omega
\justifies
\Xi(\zeta\mun\zeta'\sass  \Delta_1, \langle\vect{C\funcircum{k} B}\ass x\smwrap{k} \Gamma\rangle, \Delta_2)\yields D\ass\omega\{x/z\}
\using \funcircum{k} L
\endprooftree
&
\prooftree
\zeta\sass\Gamma\smwrap{k}\vect{B}\ass y\yields C\ass\chi
\justifies
\zeta; \Gamma\yields C\funcircum{k} B\ass\chi
\using \funcircum{k} R
\endprooftree
\\\\
26. & \prooftree
\zeta\sass  \Gamma\yields A\ass\phi \tb
\Omega(\zeta'\sass  \Delta_1, \langle\vect{C}\ass z\rangle, \Delta_2)\yields D\ass\omega
\justifies
\Omega(\zeta\mun\zeta'\sass  \Delta_1, \langle\Gamma\smwrap{k} \vect{A\arginfix{k} C}\ass y\rangle, \Delta_2)\yields D\ass\omega\{\zero/z\}
\using \arginfix{k} L
\endprooftree 
&
\prooftree
\zeta\sass\vect{A}\ass x\smwrap{k}\Gamma\yields C\ass\chi
\justifies
\zeta\sass \Gamma\yields A\arginfix{k} C\ass\zero
\using \arginfix{k} R
\endprooftree
\\\\
27. &
\prooftree
\Omega\langle\vect{A}\ass x\smwrap{k}\vect{B}\ass y\rangle\yields D\ass\omega
\justifies
\Omega\langle\vect{A\upperiwprod{k} B}\ass z\rangle\yields D\ass\omega\subst{z/y}
\using \upperiwprod{k} L
\endprooftree
&
\prooftree
\zeta\sass  \Gamma_1\yields A\ass\phi\tb\zeta'\sass  \Gamma_2\yields B\ass\psi
\justifies
\zeta\mun\zeta'\sass  \Gamma_1\smwrap{k} \Gamma_2\yields A\upperiwprod{k} B\ass\psi
\using \upperiwprod{k} R
\endprooftree
\\\\
28. & \prooftree
\zeta\sass  \Gamma\yields B\ass\psi \tb
\Omega(\zeta'\sass  \Delta_1\langle\vect{C}\ass z\rangle, \Delta_2)\yields D\ass\omega
\justifies
\Omega(\zeta\mun\zeta'\sass  \Delta_1, \langle\vect{C\argcircum{k} B}\ass x\smwrap{k} \Gamma\rangle, \Delta_2)\yields D\ass\omega\{\zero/z\}
\using \argcircum{k} L
\endprooftree 
&
\prooftree
\zeta\sass \Gamma\smwrap{k}\vect{B}\ass y\yields C\ass\chi
\justifies
\zeta\sass \Gamma\yields C\argcircum{k} B\ass\zero
\using \argcircum{k} R
\endprooftree
\\\\
29. &\prooftree
\zeta\sass  \Gamma\yields A\ass\phi \tb
\Omega(\zeta'\sass  \Delta_1, \langle\vect{C}\ass z\rangle, \Delta_2)\yields D\ass\omega
\justifies
\Omega(\zeta\mun\zeta'\sass  \Delta_1, \langle\Gamma\smwrap{k} \vect{A\funinfix{k} C}\ass y\rangle, \Delta_2)\yields D\ass\omega\{y/z\}
\using \funinfix{k} L
\endprooftree
&
\prooftree
\zeta\sass \vect{A}\ass x\smwrap{k}\Gamma\yields C\ass\chi
\justifies
\zeta\sass \Gamma\yields A\funinfix{k} C\ass\chi
\using \funinfix{k} R
\endprooftree
\\\\
30. &
\prooftree
\Omega\langle\vect{A}\ass x\smwrap{k}\vect{B}\ass y\rangle\yields D\ass\omega
\justifies
\Omega\langle\vect{A\loweriwprod{k} B}\ass z\rangle\yields D\ass\omega\subst{z/x}
\using \loweriwprod{k} L
\endprooftree &
\prooftree
\zeta\sass  \Gamma_1\yields A\ass\phi\tb\zeta'\sass  \Gamma_2\yields B\ass\psi
\justifies
\zeta\mun\zeta'\sass  \Gamma_1\smwrap{k} \Gamma_2\yields A\loweriwprod{k} B\ass\phi
\using \loweriwprod R\commentout{ZZZwhy small?}
\endprooftree
\end{array}$
\end{center}
\caption{Semantically inactive discontinuous multiplicative rules}
\label{dsim}
\end{figure}

\noindent
The semantically inactive multiplicatives \{$\argover$, $\fununder$, $\funover$, $\argunder$,
$\leftiprod$, $\rightiprod$, 
$\funcircum{}$, $\arginfix{}$, $\argcircum{}$, $\funinfix{}$,
$\upperiwprod{}$, $\loweriwprod{}$\}
of Figures~\ref{csim} and~\ref{dsim}, Morrill and Valent\'{\i}n (2014\cite{mv:words}),
can be used for the subcategorization without vacuous lambda abstraction
of semantically void elements.
For example:
\disp{
\begin{tabular}[t]{ll}
a. &  $\syncnst{rains}\ass \syncnst{it}\fununder S\ass\semcnst{rain}$ for $\syncnst{it rains}\ass S\ass
\semcnst{rain}$\\
b. & $\syncnst{give}\ass (N\bsl S)/(N{\RIGHTcircle}\syncnst{the}+\syncnst{cold}\add\syncnst{shoulder})\ass\semcnst{shun}$
\end{tabular}\label{Wex}}
This is a type logical formulation of the old idea that heads can co-occur with
fixed strings to yield non-compositional meanings, and it runs up against
the standard problem that some such strings can be compositionally modified,
for example:
\disp{John gave Mary the same offensively cold shoulder that she gave him.\label{offcold}}
(Gazdar et al. 1985\cite{gkps:gpsg}), i.e.\ there is \emph{semi}-compositionality,
or semi-idiomaticity,
depending on whether we want to see the bottle as half empty or as half full.
In the face of this dilemma it seems we can only let the grammar decide and take consolation in
the fact that all grammars leak.
We shall in fact assume a treatment of the form (\ref{Wex}b),
and we suggest that (\ref{offcold}) is interpreted by \scare{poetic license}.

\begin{figure}[ht]
\begin{center}
$
\begin{array}{lcc}
31.\tb &
\prooftree
\Xi\langle\vect{A}\ass x\rangle\yields C\ass\chi
\justifies
\Xi\langle\vect{A\iaconj B}\ass x\rangle\yields C\ass\chi
\using \iaconj L_1
\endprooftree
&
\prooftree
\Xi\langle\vect{B}\ass y\rangle\yields C\ass\chi
\justifies
\Xi\langle\vect{A\iaconj B}\ass y\rangle\yields C\ass\chi
\using \iaconj L_2
\endprooftree
\\\\
&\multicolumn{2}{c}{\prooftree
\Xi\yields A\ass\chi\tb\Xi\yields B\ass\chi
\justifies
\Xi\yields A\iaconj B\ass\chi
\using \iaconj R
\endprooftree}
\\\\
32. &\multicolumn{2}{c}{\prooftree
\Xi\langle\vect{A}\ass z\rangle\yields C\ass\chi\tb\Xi\langle\vect{B}\ass z\rangle\yields C\ass\chi
\justifies
\Xi\langle\vect{A\iadisj B}\ass z\rangle\yields C\ass\chi
\using \iadisj L
\endprooftree}
\\\\
& \prooftree
\Xi\yields A\ass\phi
\justifies
\Xi\yields A\iadisj B\ass\phi
\using \iadisj R_1
\endprooftree
&
\prooftree
\Xi\yields B\ass\psi
\justifies
\Xi \yields A\iadisj B\ass\psi
\using \iadisj R_2
\endprooftree
\end{array}$
\end{center}
\caption{Semantically inactive additive rules}
\label{sia}
\end{figure}

The semantically inactive additives \{$\iaconj$, $\iadisj$\}
of Figure~\ref{sia}, Morrill (1994\cite{morrill:tlg}),
can be used for polymorphism which is syntactic but not semantically active.
For example (assuming as semantics an identity function across types),
$\syncnst{to}\ass(\PP/N)\iaconj(\VP/\VP)\ass\lambda xx$ for
$$\syncnst{to Paris}\ass\PP\mbox{ and }\syncnst{to run}\ass\VP
\mbox{ and } \syncnst{thinks}\ass(N\bsl S)/(S\iadisj\CP)\ass\semcnst{think}$$
for 
$$\syncnst{John thinks Suzy sings}\ass S\ass((\semcnst{think}\ (\semcnst{sing}\ \semcnst{s}))\ \semcnst{j})$$ and 
$$\syncnst{John thinks that Suzy sings}\ass S\ass((\semcnst{think}\ (\semcnst{sing}\ \semcnst{s}))\ \semcnst{j}).$$

\begin{figure}[ht]
\begin{center} 
 $
 \begin{array}{lcc}
33.\tb &
 \prooftree
\Xi\langle \vect{A[t/v]}\ass x\rangle\yields B\ass\psi
 \justifies
\Xi\langle\vect{\forall vA}\ass x\rangle\yields B\ass\psi
 \using \forall L
 \endprooftree
 &
 \prooftree
\Xi\yields A[a/v]\ass\phi
 \justifies
\Xi\yields \forall vA\ass\phi
 \using \forall R^\dagger
 \endprooftree
 \\\\
 34. &
 \prooftree
\Xi\langle\vect{A[a/v]}\ass x\rangle\yields B\ass\psi
\justifies
\Xi\langle\vect{\exists vA}\ass x\rangle\yields B\ass\psi
\using \exists L^\dagger
\endprooftree
&
\prooftree
\Xi\yields A[t/v]\ass\phi
\justifies
\Xi\yields\exists vA\ass\phi
\using \exists R
\endprooftree
\end{array}$
\end{center}
\caption{Semantically inactive quantifier rules;
$^\dagger$ indicates that there is no $a$ in the conclusion}
\label{siq}
\end{figure}

The semantically inactive  first-order quantifiers of Figure~\ref{siq}, see
Morrill (1994\cite{morrill:tlg}),
have application to syntactic features.
For example, to transmit number agreement features in a definite noun phrase
there can be an assignment
$\syncnst{the}\ass\forall n(Nt(n)/\CN n)\ass\iota$ indicating that the number and gender
on a definite noun phrase come from its head.
And in a sentence,
$\syncnst{likes}\ass((\exists g Nt(s(g))\bsl S)/\exists a Na)\ass$ $\semcnst{like}$ indicating that the object
can have any agreement features but the subject must be third person singular,
of any gender.

\begin{figure}[ht]
\begin{center}
$
\begin{array}{lcc}
35. &
 \prooftree
\Xi\langle\vect{A}\ass x\rangle\yields B\ass\psi
 \justifies
\Xi\langle\vect{\imod A}\ass x\rangle\yields B\ass\psi
 \using \imod L
 \endprooftree 
 &
\prooftree
 \diagmod\Xi\yields A\ass\phi
 \justifies
\diagmod\Xi\yields \imod A\ass\phi
 \using \imod R
 \endprooftree
 \\\\
 36. \tb &
 \prooftree
\diagmod\Xi\langle\vect{A}\ass x\rangle\yields \diagemod B\ass\psi
 \justifies
\diagmod\Xi\langle\vect{\iemod A}\ass x\rangle\yields \diagemod B\ass\psi
 \using \iemod L
 \endprooftree
 &
 \prooftree
\Xi\yields A\ass\phi
 \justifies
\Xi\yields \iemod A\ass\phi
 \using \iemod R
 \endprooftree \end{array}
$
 \end{center}
\caption{Semantically inactive normal modality rules; $\diagmod/\diagemod$ marks a structure all the types of
which have principal connective a box/diamond}
\label{sinm}
\end{figure}

With respect to the semantically inactive modalities \{$\imod$, $\iemod$\} of Figure~\ref{sinm},
Hepple (1990\cite{hepple:phd}) and
Morrill (1994\cite{morrill:tlg}),
the universal can be applied to classify rigid designators such as proper names
as in e.g.~$\syncnst{John}\ass\imod N\ass \semcnst{j}.$

\begin{figure}[ht]
\begin{center}
$
\begin{array}{lcc}
37. & 
\prooftree
\Xi\langle\vect{A}\ass x\rangle\yields B\ass\psi
\justifies
\Xi\langle\vect{A}\ass x, \sep\rangle\yields B\ass\psi
\using \leftproj L
\endprooftree
&
\prooftree
\zeta\sass\Gamma, \sep\yields A\ass\phi
\justifies
\zeta\sass\Gamma\yields \leftproj A\ass\phi
\using \leftproj R
\endprooftree\\\\
38. & \prooftree
\Xi\langle\vect{A}\ass x\rangle\yields B\ass\psi
\justifies
\Xi\langle\sep, \vect{A}\ass x\rangle\yields B\ass\psi
\using \rightproj L
\endprooftree
&
\prooftree
\zeta\sass\sep, \Gamma\yields A\ass\phi
\justifies
\zeta\sass\Gamma\yields\rightproj A\ass\phi
\using \rightproj R
\endprooftree
\\\\
39. &
\prooftree
\Xi\langle \vect{A}\ass x, \sep\rangle\yields B\ass\psi
\justifies
\Xi\langle\vect{\leftinj A}\ass x\rangle\yields B\ass\psi
\using \leftinj L
\endprooftree 
&
\prooftree
\zeta\sass\Gamma\yields A\ass\phi
\justifies
\zeta\sass\Gamma, \sep\yields\leftinj A\ass\phi
\using \leftinj R
\endprooftree
\\\\
40. &
\prooftree
\Xi\langle\sep, \vect{A}\ass x\rangle\yields B\ass\psi
\justifies
\Xi\langle\vect{\rightinj A}\ass x\rangle\yields B\ass\psi
\using \rightinj L
\endprooftree
&
\prooftree
\zeta\sass\Gamma\yields A\ass\phi
\justifies
\zeta\sass\sep, \Gamma\yields\rightinj A\ass\phi
\using \leftinj R
\endprooftree
\\\\
41. & \prooftree
\Xi\langle \vect{B}\ass y\rangle\yields C\ass\chi
\justifies
\Xi\langle\vect{\ssplit{k} B}\ass y\smwrap{k}\Lambda\rangle\yields C\ass\chi
\using \ssplit{k} L
\endprooftree
&
\prooftree
\zeta\sass\Delta\smwrap{k}\Lambda\yields A\ass\psi
\justifies
\zeta\sass\Delta\yields\ssplit{k} A\ass\psi
\using \ssplit{k} R
\endprooftree
\\\\
42.\tb &
\prooftree
\Xi\langle\vect{B}\ass y\smwrap{k}\Lambda\rangle\yields C\ass\chi
\justifies
\Xi\langle\vect{\sbridge{k} B}\ass y\rangle\yields C\ass\chi
\using \sbridge{k} L
\endprooftree
&
\prooftree
\zeta\sass\Delta\yields A\ass\psi
\justifies
\zeta\sass\Delta\smwrap{k}\Lambda\yields\sbridge{k} A\ass\psi
\using \sbridge{k} R
\endprooftree
\end{array}
$
\end{center}
\caption{Deterministic synthetic multiplicative rules}
\label{dsm}
\end{figure}

The deterministic synthetic multiplicatives of Figure~\ref{dsm},
left and right projection and injection \{$\leftproj$, $\rightproj$, 
$\leftinj$, $\rightinj$\}, Morrill et al.\ (2009\cite{mvf:tbilisi}),
and split and bridge \{$\ssplit{}$, $\sbridge{}$\},
Morrill and Merenciano (1996\cite{morrill:merenciano}),
are abbreviatory unary operators. By way of example,
a relative pronoun assignment allowing both peripheral and medial
extraction may now be expressed
$\syncnst{that}\ass (\CN\bsl\CN)/\sbridge{}(S\scircum{}N)\ass\lambda x\lambda y\lambda z[(y\ z)\wedge(x\ z)]$.

\begin{figure}[ht]
\begin{center}
\rotatebox{-0}{\scriptsize
$
\begin{array}{lcc}
43. & \prooftree
\zeta\sass\Gamma\yields A\ass\phi \tb
\Xi(\zeta'\sass\Delta_1, \langle\vect{C}\ass z\rangle, \Delta_2)\yields D\ass\omega
\justifies
\Xi(\zeta\mun\zeta'\sass\Delta_1, \langle\Gamma, \vect{B\ndiv A}\ass y\rangle, \Delta_2)\yields D\ass\omega\subst{(y\ \phi)/z}
\using \ndiv L_1
\endprooftree 
&
\prooftree
\zeta\sass\Gamma\yields A\ass\phi \tb
\Xi(\zeta'\sass\Delta_1, \langle\vect{C}\ass z\rangle, \Delta_2)\yields D\ass\omega
\justifies
\Xi(\zeta\mun\zeta'\sass\Delta_1, \langle\vect{B\ndiv A}\ass y, \Gamma\rangle, \Delta_2)\yields D\ass\omega\subst{(y\ \phi)/z}
\using \ndiv L_2
\endprooftree\\\\
& \multicolumn{2}{c}{\prooftree
\zeta\sass\vect{A}\ass x, \Gamma\yields C\ass\chi\tb
\zeta\sass\Gamma, \vect{A}\ass x\yields C\ass\chi
\justifies
\zeta\sass\Gamma\yields C\ndiv A\ass\lambda x\chi
\using \ndiv R
\endprooftree}
\\\\
44. & \multicolumn{2}{c}{\prooftree
\Xi\langle\vect{A}\ass x, \vect{B}\ass y\rangle\yields D\ass\omega\tb
\Xi\langle\vect{B}\ass y, \vect{A}\ass x\rangle\yields D\ass\omega
\justifies
\Xi\langle\vect{A\nprod B}\ass z\rangle\yields D\ass\omega\subst{\pi_1 z/x, \pi_2 z/y}
\using \nprod L
\endprooftree}
\\\\
& \prooftree
\zeta\sass\Gamma_1\yields A\ass\phi\tb\zeta'\sass\Gamma_2\yields B\ass\psi
\justifies
\zeta\mun\zeta'\sass\Gamma_1, \Gamma_2\yields A\nprod B\ass(\phi, \psi)
\using \nprod R_1
\endprooftree &
\prooftree
\zeta\sass\Gamma_1\yields B\ass\psi\tb\zeta'\sass\Gamma_2\yields A\ass\phi
\justifies
\zeta\mun\zeta'\sass\Gamma_1, \Gamma_2\yields A\nprod B\ass(\phi, \psi)
\using \nprod R_2
\endprooftree
\\\\
45. & \prooftree
\zeta\sass\Gamma\yields A\ass\psi \tb
\Xi(\zeta'\sass\Delta_1, \langle\vect{B}\ass z\rangle, \Delta_2)\yields D\ass\omega
\justifies
\Xi(\zeta\mun\zeta'\sass\Delta_1, \langle\vect{B\nextract A}\ass x\smwrap{k}\Gamma\rangle, \Delta_2)\yields D\ass\omega\subst{(x\ \psi)/z}
\using \nextract L
\endprooftree 
&
\prooftree
\zeta\sass\Gamma\smwrap{1}\vect{B}\ass y\yields C\ass\chi\tb\cdots\tb
\zeta\sass\Gamma\smwrap{sC'-sB}\vect{B}\ass y\yields C\ass\chi
\justifies
\zeta\sass\Gamma\yields C\nextract B\ass\lambda y\chi
\using \nextract R
\endprooftree
\\\\
46. & \prooftree
\zeta\sass\Gamma\yields A\ass\phi \tb
\Xi(\zeta'\sass\Delta_1, \langle\vect{B}\ass z\rangle, \Delta_2)\yields D\ass\omega
\justifies
\Xi(\zeta\mun\zeta'\sass\Delta_1, \langle\Gamma\smwrap{k}\vect{A\ninfix B}\ass y\rangle, \Delta_2)\yields D\ass\omega\subst{(y\ \phi)/z}
\using \ninfix L
\endprooftree
&
\prooftree
\zeta\sass\vect{A}\ass x\smwrap{1}\Gamma\yields C\ass\chi\tb\cdots\tb
\zeta\sass\vect{A}\ass x\smwrap{sA}\Gamma\yields C\ass\chi
\justifies
\zeta\sass\Gamma\yields A\ninfix C\ass\lambda x\chi
\using \ninfix R
\endprooftree
\\\\
47.\tb & \prooftree
\Xi\langle\vect{A}\ass x\smwrap{1}\vect{B}\ass y\rangle\yields D\ass\omega\tb\cdots\tb
\Xi\langle\vect{A}\ass x\smwrap{sA}\vect{B}\ass y\rangle\yields D\ass\omega
\justifies
\Xi\langle\vect{A\ndprod B}\ass z\rangle\yields D\ass\omega\subst{\pi_1 z/x, \pi_2 z/y}
\using \ndprod L
\endprooftree
&
\prooftree
\zeta\sass\Gamma_1\yields A\ass\phi\tb\zeta'\sass\Gamma_2\yields B\ass\psi
\justifies
\zeta\mun\zeta'\sass\Gamma_1\smwrap{k}\Gamma_2\yields A\ndprod B\ass(\phi, \psi)
\using \ndprod R
\endprooftree
\end{array}
$}
\end{center}
\caption{Non-deterministic synthetic multiplicative rules}
\label{nsm}
\end{figure}

The continuous and discontinuous
non-deterministic synthetic multiplicatives \{$\ndiv$, $\nprod$, $\nextract$, $\ninfix$, $\ndprod$\}
of Figure~\ref{nsm}, Morrill,
Valent\'{\i}n and Fadda (2011\cite{mvf:tdc}),
are abbreviatory binary operators. By way of example of the continuous operators,
an adsentential preposition can be allowed to appear both sentence initially
and sentence finally by an assignment such as
$\syncnst{in}\ass (S\ndiv S)/N\ass\semcnst{in}$ for e.g.
$$\syncnst{in summer John swims}\ass S\ass((\semcnst{in}\ \semcnst{summer})\ (\semcnst{swim}\ \semcnst{j}))$$ and
$$\syncnst{John swims in summer}\ass S\ass((\semcnst{in}\ \semcnst{summer})\ (\semcnst{swim}\ \semcnst{j})),$$
and complement alternation can be expressed in e.g.~$\syncnst{talks}\ass(N\bsl S)/(\mathit{PPto}\nprod\mathit{PPabout})\ass\semcnst{talk}$
for both
$$\syncnst{John talks to Mary about Bill}\ass S\ass((\semcnst{talk}\ (\semcnst{m}, \semcnst{b}))\ \semcnst{j})$$
and
$$\syncnst{John talks about Bill to Mary}\ass S\ass((\semcnst{talk}\ (\semcnst{m}, \semcnst{b}))\ \semcnst{j}).$$
By way of example of the discontinuous operators,
the particle shift of a particle verb can be captured by e.g.\
$\syncnst{calls}\add\sep\add\syncnst{up}\add\sep\ass \ssplit{}(N\bsl S)\nextract N\ass\semcnst{phone}$
which produces both
$\syncnst{John calls}$ $\syncnst{Mary up}\ass S\ass((\semcnst{phone}\ \semcnst{m})\ \semcnst{j})$ and $\syncnst{John calls up Mary}\ass S\ass((\semcnst{phone}\ \semcnst{m})\ \semcnst{j})$.

\clearpage

\section{Limited contraction and limited expansion}

The limited contraction and limited expansion of Figure~\ref{sllca2}
can be used for anaphora and for words as types respectively.

\begin{figure}[ht]
\begin{center}
$
\begin{array}{lc}
48. &
 \prooftree
\zeta\sass\Gamma\yields A\ass\phi\tb\zeta'\sass\Delta\langle \vect{A}\ass x; \vect{B}\ass y\rangle\yields D\ass\omega
\justifies
\zeta\mun\zeta'\sass\Delta\langle\Gamma; \vect{B|A}\ass z\rangle\yields D\ass\omega\subst{\phi/x, (z\ \phi)/y}
\using |L
\endprooftree
\\\\
\tb&\prooftree
\zeta\sass\Gamma\langle \vect{B_0}\ass y_0: \ldots: \vect{B_n}\ass y_n\rangle\yields D\ass\omega 
\justifies
\zeta\sass\Gamma\langle \vect{B_0|A}\ass z_0: \ldots: \vect{B_n|A}\ass z_n\rangle\yields D|A\ass\lambda x\omega\subst{(z_0\ x)/y_0, \ldots, (z_n\ x)/y_n}
\using |R
\endprooftree
\\\\
49.\tb & \prooftree
\Xi\langle\Lambda\rangle\yields A\ass\phi
\justifies
\Xi\langle\vect{0}\ass x\rangle\yields A\ass\phi
\using +L
\endprooftree
\tb\prooftree
\Xi\langle\vect{v\ass x, \vect{w}}\ass y\rangle\yields D\ass\omega
\justifies
\Xi\langle\vect{v+w}\ass z\rangle\yields D\ass\omega\subst{z/x, z/y}
\using +L
\endprooftree\\\\
 & 
\prooftree
\justifies
\Lambda\yields 0\ass\zero
\using +R
\endprooftree
\tb
\prooftree
\zeta_1; \Gamma_1\yields v\ass\phi\tb
\zeta_2; \Gamma_2\yields w\ass\phi
\justifies
\zeta_1\mun\zeta_2; \Gamma_1, \Gamma_2\yields v+w\ass\nil
\using +R
\endprooftree
\end{array}$
\end{center}
\caption{Limited contraction and limited expansion rules}
\label{sllca2}
\end{figure}

The limited contraction, $|$,
of  J\"{a}ger (2005\cite{jaeger:2005}),
can be used for anaphora in an assignment like
$\syncnst{it}\ass$ $(S\scircum{}N)\sinfix{}(S|N)\ass\lambda xx$
for,
e.g.,
$\syncnst{the company$_i$ said it$_i$}$ $\syncnst{flourished}\ass S\ass
((\semcnst{say}\ (\semcnst{flourish}\ (\iota\ \semcnst{company})))$ $(\iota\ \semcnst{company}))$,
and it can be used for $\syncnst{such that}$ relativisation
in an assignment
$\syncnst{such that}\ass$ $(\CN\bsl\CN)/$ $(S|N)\ass\lambda x\lambda y\lambda z[(y\ z)\wedge(x\ z)]$ for,
say,
$\syncnst{man such that$_i$ he$_i$ thinks}$ $\syncnst{Mary loves him$_i$}\ass \CN\ass
\lambda z[(\semcnst{man}\ z)\wedge((\semcnst{think}\ ((\semcnst{love}\ z)\ \semcnst{m}))\ z)]$;
see Chapter~\ref{monfrag}. 
The limited expansion can be used for words as types
as in Morrill and Valent\'{\i}n (2014\cite{mv:words}) for semantically void words.
For example $\syncnst{rains}\ass \mathit{it}\bsl S\ass\lambda x\semcnst{rain}$ for
\lingform{it rains}.

\section{Difference}

\commentout{ZZZAdmissibility of Cut?}

\disp{$
\prooftree
\zeta_1; \Gamma\yields A\ass\phi \tb
\zeta_2; \Delta\langle \vect{A}\ass x\rangle\yields B\ass\psi
\justifies
\zeta_1\mun\zeta_2; \Delta\langle\Gamma\rangle\yields B\ass\psi\subst{\phi/x}
\using \mathit{Cut}
\endprooftree
$
}

Negation as failure:\footnote{For difference
$-$ (connective number 50), however, Cut is not appropriate.
We consider this a \techterm{metalogical\/} connective which is added
to the system for which Cut-elimination has been proved.}
\disp{$
50.\ \prooftree
\justifies
\Delta\yields \neg A
\using \neg R, \not\vdash \Delta\yields A
\endprooftree
$}
Difference: $A-B = A\&\neg B$:
\disp{$
\prooftree
\Delta\yields A
\justifies
\Delta\yields A-B
\using - R, \not\vdash \Delta\yields B
\endprooftree
$}

\part{PARSING}

Even in Cut-free sequent proof search there are in general many 
sequent proofs for each theorem, differing in inessential rule
reorderings.
And in particular, in Cut-free hedge sequent proof search there are in general
many semantically equivalent hedge sequent proofs for each theorem,
giving rise to great redundancy in hedge sequent calculus
parsing/theorem proving. In this part we present Andreoli's
method of \scare{focusing}
for greatly reducing the redundancy of hedge sequent calculus.
In addition, theorems are subject to van Benthem's \scare{count} invariance properties
necessary (though not sufficient) for theoremhood.
In this part we also present infinitary count invariance for categorial
logic including subexponentials, which can be used in hedge
sequent calculus proof search to filter
hedge sequents and subsequents by rapidly establishing
non-count invariance and hence non-provability.

\chapter{Focalised Sequent Calculus}

\label{focchap}

Categorial grammar operates under the slogans
`grammar as logic' and `parsing as deduction'.
The grammar is a substructural, indeed sublinear,
logic such as displacement logic,
and the parsing paradigm is typically backward-chaining sequent proof-search.
Spurious ambiguity is the phenomenon whereby distinct derivations in grammar
may assign the same structural reading,
resulting in redundancy in the parse search space and inefficiency in parsing.
Understanding the problem depends on identifying the essential mathematical structure
of derivations.
This is trivial in the case of CFG, where the parse structures
are ordered trees;
in the case of categorial grammar, the parse structures
are proof nets. However, with respect to multiplicatives intrinsic
proof nets have not yet been given for displacement calculus {\bf D},
and proof nets for additives, for example,
which have applications to polymorphism, are complex.
In this chapter we approach parsing as deduction for 
Full Displacement Logic $\FDL$
by means of the proof-theoretic technique of focalisation,
and we prove completeness for the case of displacement calculus with additives
{\bf DA}. 

\section{Spurious ambiguity}

\subsection{Introduction}

In CFG sequential rewriting derivations exhibit spurious ambiguity:
distinct rewriting derivations may correspond to the same parse structure (tree) and the
same structural reading.
In this case it is transparent to develop parsing algorithms avoiding spurious ambiguity
by reference to parse trees. 
In categorial grammar (CG) the problem is more subtle. The Cut-free Lambek
sequent proof search space is finite, but involves a combinatorial explosion
of spuriously ambiguous sequential proofs: sequent proofs of the same
endsequent with the same Curry-Howard term reading.
This can be understood, analogously to CFG, as inessential
rule reorderings, which we can parallelise in underlying geometric parse structures
which are (planar) proof nets.
The planarity of Lambek proof nets reflects that the formalism is continuous or
concatenative. But the challenge of natural grammar is discontinuity or
apparent movement, whereby there is syntactic/semantic mismatch,
or elements appearing out of place. Hence the subsumption of Lambek
calculus, as a logic of strings with appending,
by displacement calculus \D{}, as a logic of strings with holes
including plugging as well as
appending (Morrill et al.\ 2011\cite{mvf:tdc}).

Proof nets for \D{} must be partially nonplanar; steps
towards intrinsic correctness criteria for displacement proof nets are
made in Fadda (2010\cite{fadda:phd}) 
and 
Moot (2014\cite{moot:lamfes}). 
However, even in the case
of Lambek calculus, in our experience parsing by reference to intrinsic criteria 
(Morrill 2011\cite{morrill:oxford} appendix B,
Moot and Retor\'{e} 2012\cite{mootret}) 
is not more efficient than parsing by reference to
extrinsic criteria of normalised sequent calculus 
(Hendriks 1993\cite{hendriksH:phd},
Morrill 2011\cite{morrill:logprodisp}, 2012\cite{morrill:catlogdem}).
In its turn, on the other hand,
normalisation does not extend to product left rules and product unit
left rules, nor to additives.
The focalisation of Andreoli (1992\cite{andreoli:92})
represents a methodology midway between proof nets and normalisation.
Here we apply the focusing discipline 
to the parsing as deduction of $\FDL$.

\subsection{Spurious ambiguity in CFG and CG}

\label{spambcfglc}

Consider the following production rules:
\disp{$
\begin{array}[t]{l}
S \rightarrow Q\ \VP\\
Q \rightarrow \Det\ \CN\\
\VP \rightarrow \TV\ N
\end{array}
$}
These generate the following sequential rewriting derivations:
\disp{$
\begin{array}[t]{l}
S \rightarrow Q\ \VP \rightarrow \Det\ \CN\ \VP \rightarrow \Det\ \CN\ \TV\ N\\
S \rightarrow Q\ \VP \rightarrow Q\ \TV\ N \rightarrow \Det\ \CN\ \TV\ N
\end{array}
$}
These sequential rewriting derivations correspond to the
same parellelised parse structure:
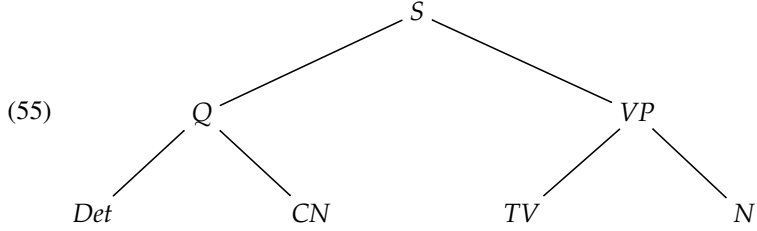
\disp{$
\diagram
 &&& S\\
 &Q\urrline &&&&\VP\ullline\\
\Det\urline && \CN\ulline && \TV\urline && N\ulline
\enddiagram
$}
And they correspond to the same structural reading;
sequential rewriting has \techterm{spurious ambiguity}.

Recall the definitions of types, configurations and sequents in the Lambek
calculus \AL{} (Lambek 1958\cite{lambek:mathematics}), in terms of a set $\cal P$ of primitive types,
where $\Lambda$ is the metalinguistic empty string:
\disp{
Types $\cal F ::= {\cal P}\ |\ {\cal F}\bsl{\cal F}\ |\ {\cal F}/{\cal F}\ |\ {\cal F}\product{\cal F}$\\
Configurations ${\cal O} ::= \Lambda\ |\ {\cal F}\ |\ {\cal O}, {\cal O}$\\
Sequents $\Sigma ::= {\cal O}\yields{\cal F}$}
$\Delta(\Gamma)$ indicates a configuration $\Delta$ and
a distinguished subconfiguration $\Gamma$; the logical rules of \AL{} are:

\disp{
$
\prooftree
\Gamma\yields A \tb
\Delta(C)\yields D
\justifies
\Delta(A\bsl C)\yields D
\using \bsl L
\endprooftree \tb
\prooftree
A, \Gamma\yields C
\justifies
\Gamma\yields A\bsl C
\using \bsl R
\endprooftree$
\vtab
$\prooftree
\Gamma\yields B\tb
\Delta(C)\yields D
\justifies
\Delta(C/B,  \Gamma)\yields D
\using / L
\endprooftree \tb
\prooftree
\Gamma, B\yields C
\justifies
\Gamma\yields C/B
\using /R
\endprooftree$
\vtab
$\prooftree
\Delta(A, B)\yields D
\justifies
\Delta(A\product B)\yields D
\using \product L
\endprooftree$
\tb
$\prooftree
\Gamma_1\yields A\tb\Gamma_2\yields B
\justifies
\Gamma_1, \Gamma_2\yields A\product B
\using \product R
\endprooftree
$
}

\begin{figure*}
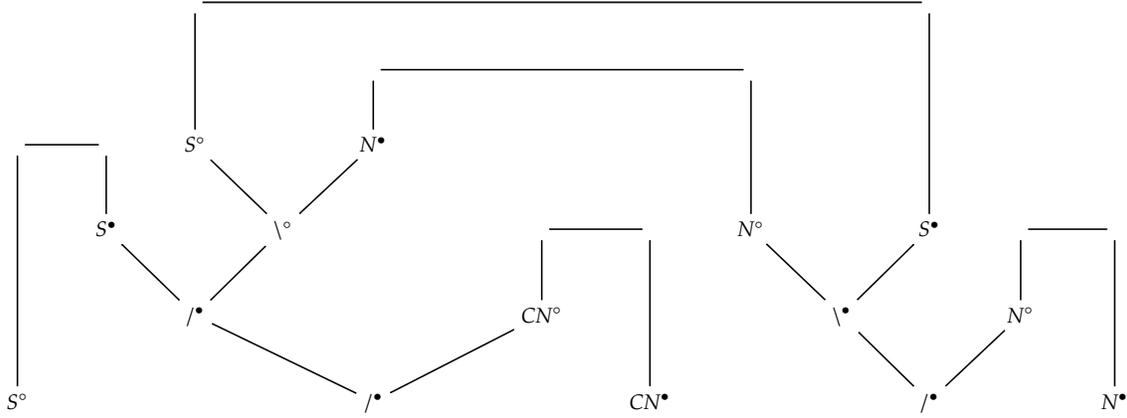

\spreaddiagramcolumns{-1ex}
\spreaddiagramrows{-1ex}
\begin{center}
$\footnotesize
\diagram
&&&&&&&&&&&&\\
&&&&&&&&&&&&\\
&& S\out && N\inp\\
& S\inp && \bsl\out\ulline\urline &&&&& N\out\xline'[-2,0]'[-2,-4]'[-1,-4] && S\inp\xline'[-3,0]'[-3,-8]'[-1,-8]&&\\
&& /\inp\ulline\urline&&&& \CN\out&&&\bsl\inp\ulline\urline&& N\out\\
S\out\xline'[-3,0]'[-3,1]'[-2,1]&&&&/\inp\ullline\urrline&&&\CN\inp\xline'[-2,0]'[-2,-1]'[-1,-1]&&&/\inp\ulline\urline && N\inp\xline'[-2,0]'[-2,-1]'[-1,-1]
\enddiagram
$
\end{center}
\caption{Proof net}
\label{pn}
\end{figure*}

\noindent
Even amongst Cut-free proofs there is spurious ambiguity; consider for example
the sequential derivations of Figure~\ref{seqder}.
\begin{figure*}
\begin{center}\footnotesize
\prooftree
\CN\yields\CN
\prooftree
\prooftree
\prooftree
N\yields N
\prooftree
N\yields N\tb S\yields S
\justifies
N, N\bsl S\yields S
\using \bsl L
\endprooftree
\justifies
N, (N\bsl S)/N, N\yields S
\using /L
\endprooftree
\justifies
(N\bsl S)/N, N\yields N\bsl S
\using \bsl R
\endprooftree
S\yields S
\justifies
S/(N\bsl S), (N\bsl S)/N, N\yields S
\using /L
\endprooftree
\justifies
(S/(N\bsl S))/\CN, \CN, (N\bsl S)/N, N\yields S
\using  /L
\endprooftree
\tab
\prooftree
N\yields N
\prooftree
\CN\yields\CN
\prooftree
\prooftree
\prooftree
N\yields N\tb S\yields S
\justifies
N, N\bsl S\yields S
\using \bsl L
\endprooftree
\justifies
N\bsl S\yields N\bsl S
\using \bsl R
\endprooftree
S\yields S
\justifies
S/(N\bsl S), N\bsl S\yields S
\using /L
\endprooftree
\justifies
(S/(N\bsl S))/\CN, \CN, N\bsl S\yields S
\using /L
\endprooftree
\justifies
(S/(N\bsl S))/\CN, \CN, (N\bsl S)/N, N\yields S
\using /L
\endprooftree
\end{center}
\caption{Spurious ambiguity}
\label{seqder}
\end{figure*}
These have the same parallelised parse structure (proof net),
given in Figure~\ref{pn}.

Lambek proof structures are planar
graphs which must satisfy certain global and local properties to be correct as proofs
(proof nets). Proof nets provide a geometric perspective
on derivational equivalence. Alternatively we may identify the same
algebraic parse structure (Curry-Howard term):
$((x_{\mathit{Det}}\,x_{\mathit{CN}})$ $\lambda x((x_{\mathit{TV}}\,x_{N})\,x))$
for the categorial examples here.

\section{Focalisation for \FDL}
\label{foc}

The discipline of focalisation depends fundamentally on the distinction
between invertible and noninvertible rules. A rule is invertible (or reversable)
if its premises are derivable from its conclusion --- for example $\bsl R$
and $\adisj L$ --- otherwise it is noninvertible.
In focalisation situated (antecedent, input, \inp{} / succedent, output, \out) connectives are classified as of negative (asynchronous)
or positive (synchronous) \emph{polarity\/} according as their rule is invertible
or not respectively.
There are alternating phases of don't-care non-deterministic negative rule application,
and positive rule application locking on to \techterm{focalised\/} formulas.
Given a sequent, invertible rules are applied in a don't care non-determinitic fashion until
no longer possible. Once given a subgoal
with no occurrences of negative formulas, one chooses a positive formula
as principal formula; 
we say it is focalised --- in our presentation this formula is boxed ---
and we apply proof search to its subformulas while these remain positive. 
When one finds a negative formula or
a literal, invertible rules are applied in a don't care non-determinitic fashion again until
no longer possible, when another positive formula is chosen, and so on.

We define input and output synchronous and asynchronous types 
of ${\bf FDL}$, $P, Q, N, M$,
as follows:
\disp{$
$}
Sequents now include \techterm{stoups} $\cal Z$ which are lists of sort 0 types
to which structural rules can apply. Sequents with such stoups can be defined:
\disp{$
\begin{array}[t]{rcl}
\Stoup & ::= & \emptyset\ |\ \Tp{}_0^\bullet, {\Stoup}\\
\Config{} & ::= & \Lambda\ |\ \sep\ |\ \Tp{}^\bullet_0\ |\ \Tp{}^\bullet_{i{>}0}\{\underbrace{\Config{}: \ldots: \Config{}}_{i\ \mbox{\scriptsize{\bf Config}'s}}\}\ |\ \Config{}, \Config{}\ |\ [\Stoup{}; \Config{}]\\
\end{array}
$}
The definition of sequents $\Sigma$ is:
\disp{
$\Sigma ::= \Stoup{}; \Config{}\yields \Tp{}^\circ$ such that $s(\Config) = s(\Tp{}^\circ)$}
The focused sequent calculus for this categorial logic
is defined in the following sections.

\section{Asynchronous rules}

\disp{$
\prooftree
\justifies
\vect{P}\ass x\yields P\ass x
\using \mathit{id}, \mbox{$P$ atomic}
\endprooftree
$}

\subsection{Primary connectives}

\begin{figure}[ht]
\begin{center}
$
\begin{array}{lclc}
1.
&\prooftree
\zeta\sass \Gamma, \vect{B}\ass y\yields C\ass\chi
\justifies
\zeta\sass \Gamma\yields C/B\ass\lambda y\chi
\using / R
\endprooftree
&
2. & \prooftree
\zeta\sass  \vect{A}\ass x, \Gamma\yields C\ass\chi
\justifies
\zeta\sass \Gamma\yields A\bsl C\ass\lambda x\chi
\using \bsl R
\endprooftree
\\\\
3. & \prooftree
\Xi\langle\vect{A}\ass x, \vect{B}\ass y\rangle\yields D\ass\omega
\justifies
\Xi\langle\vect{A\product B}\ass z\rangle\yields D\ass\omega\{\pi_1 z/x, \pi_2 z/y\}
\using \product L
\endprooftree
&
4.
&\prooftree
\Xi\langle\Lambda\rangle\yields A\ass\phi
\justifies
\Xi\langle\vect{I}\ass x\rangle\yields A\ass\phi
\using IL
\endprooftree
\\\\
5. &
\prooftree
\zeta\sass \Gamma\smwrap{k}\vect{B}\ass y\yields C\ass\chi
\justifies
\zeta\sass \Gamma\yields C\scircum{k} B\ass\lambda y\chi
\using \scircum{k} R
\endprooftree
&
6. &\prooftree
\zeta\sass \vect{A}\ass x\smwrap{k}\Gamma\yields C\ass\chi
\justifies
\zeta\sass \Gamma\yields A\sinfix{k} C\ass\lambda x\chi
\using \sinfix{k} R
\endprooftree
\\\\
7. & \prooftree
\Xi\langle\vect{A}\ass x\smwrap{k}\vect{B}\ass y\rangle\yields D\ass\omega
\justifies
\Xi\langle\vect{A\swprod{k} B}\ass z\rangle\yields D\ass\omega\subst{\pi_1 z/x, \pi_2 z/y}
\using \swprod{k} L
\endprooftree
&
8.
&
\prooftree
\Xi\langle\sep\rangle\yields A\ass\phi
\justifies
\Xi\langle\vect{J}\ass x\rangle\yields A\ass\phi
\using JL
\endprooftree
\end{array}
$
\end{center}
\caption{Asynchronous multiplicative rules}
\label{apmult}
\end{figure}

\begin{figure}[ht]
\begin{center}
$
\begin{array}{lc}
9. &
\prooftree
\Xi\yields A\ass\phi\tb\Xi\yields B\ass\psi
\justifies
\Xi\yields A\aconj B\ass(\phi, \psi)
\using \aconj R
\endprooftree
\\\\
10. &
\prooftree
\Xi\langle\vect{A}\ass x\rangle\yields C\ass\chi_1\tb\Xi\langle\vect{B}\ass y\rangle\yields C\ass\chi_2
\justifies
\Xi\langle\vect{A\adisj B}\ass z\rangle\yields C\ass z\casearrow x.\chi_1; y.\chi_2
\using \adisj L
\endprooftree
\end{array}
$
\end{center}
\caption{Asynchronous additive rules}
\label{apadd}
\end{figure}

\begin{figure}[ht]
\begin{center} 
 $
 \begin{array}{lclc}
 11. &
 \prooftree
\Xi\yields A[a/v]\ass\phi
 \justifies
\Xi\yields \bigwedge vA\ass\lambda v\phi
 \using \bigwedge R^\dagger
 \endprooftree
&
12.
&
\prooftree
\Xi\langle\vect{A[a/v]}\ass x\rangle\yields B\ass\psi
\justifies
\Xi\langle\vect{\bigvee vA}\ass z\rangle\yields B\ass\psi\subst{\pi_2 z/x}
\using \bigvee L^\dagger
\endprooftree
\end{array}
$
\end{center}
\caption{Asynchronous quantifier rules, 
where $^\dagger$ indicates that there is no $a$ in the conclusion}
\label{aslquants}
\end{figure}

\begin{figure}[ht]
\begin{center}
$
\begin{array}{lclc}
13. &
\prooftree
\diagmod\Xi\yields A\ass\phi
 \justifies
\diagmod\Xi\yields \mymod A\ass\up\phi
 \using \mymod R
 \endprooftree
 &
 14.
 &
 \prooftree
\diagmod\Xi\langle\vect{A}\ass x\rangle\yields \diagemod B\ass\psi
 \justifies
\diagmod\Xi\langle\vect{\emod A}\ass z\rangle\yields \diagemod B\ass\psi\subst{\dnc z/x}
 \using \emod L
 \endprooftree 
 \end{array}
 $
 \end{center}
\caption{Asynchronous normal modality rules; $\diagmod/\diagemod$ marks a structure all the types of
which have principal connective a box/diamond}
\label{aslmods}
\end{figure}

\begin{figure}[ht]
\begin{center}
$
\begin{array}{lclc}
15. &
\prooftree
{}[\Xi]\yields A\ass\phi
\justifies
\Xi\yields \abrack A\ass\phi
\using \abrack R
\endprooftree
&
16.
&
\prooftree
\Xi\langle[\vect{A}\ass x]\rangle\yields B\ass\psi
\justifies
\Xi\langle\vect{\mybrack A}\ass x\rangle\yields B\ass\psi
\using \mybrack L
\endprooftree
\end{array}
$
 \end{center}
\caption{Asynchronous bracket modality rules}
\label{aslbrmods}
\end{figure}

\begin{figure}[ht]
\begin{center}
$
\begin{array}{lc}
17. & \prooftree
\zeta\sass\Lambda \yields A\ass\phi
 \justifies
\zeta; \Lambda \yields \univexp A\ass\phi
 \using \univexp R
 \endprooftree
\\\\
&
 \prooftree
 \Xi(\zeta\mun\{A\ass x\}\sass\Gamma_1, \Gamma_2)\yields B\ass\psi
 \justifies
 \Xi(\zeta\sass\Gamma_1, \univexp A\ass x, \Gamma_2)\yields B\ass\psi
 \using \univexp P
 \endprooftree 
\end{array}
$
 \end{center}
\caption{Asynchronous subexponential rules}
\label{aslexps}
\end{figure}

\commentout{

\subsection{Limited expansion}

\begin{figure}[ht]
\begin{center}
$
\begin{array}{lc}
20. & \prooftree
\Xi\langle\Lambda\rangle\yields A\ass\phi
\justifies
\Xi\langle\vect{0}\ass x\rangle\yields A\ass\phi
\using +L
\endprooftree
\tb\prooftree
\Xi\langle\vect{v}\ass x, \vect{w}\ass y\rangle\yields D\ass\omega
\justifies
\Xi\langle\vect{W(v+w)}\ass z\rangle\yields D\ass\omega\subst{z/x, z/y}
\using WL
\endprooftree
\end{array}
$
\end{center}
\caption{Asynchronous limited expansion rules}
\label{aswat}
\end{figure}

}

\clearpage

\subsection{Semantically inactive variants}

\begin{figure}[ht]
\begin{center}
$
\begin{array}{lclc}
19. & \prooftree
\zeta\sass \Gamma, \vect{B}\ass y\yields C\ass\chi
\justifies
\zeta\sass \Gamma\yields C\argover B\ass\zero
\using \argover R
\endprooftree
&
20. & \prooftree
\zeta\sass  \vect{A}\ass x, \Gamma\yields C\ass\chi
\justifies
\zeta\sass \Gamma\yields A\fununder C\ass\chi
\using \fununder R
\endprooftree
\\\\
21. & \prooftree
\zeta\sass \Gamma, \vect{B}\ass y\yields C\ass\chi
\justifies
\zeta\sass \Gamma\yields C\funover B\ass\chi
\using \funover R
\endprooftree &
22. & \prooftree
\zeta\sass  \vect{A}\ass x, \Gamma\yields C\ass\chi
\justifies
\zeta\sass \Gamma\yields A\argunder C\ass\zero
\using \argunder R
\endprooftree
\\\\
23. & \multicolumn{3}{c}{\prooftree
\Xi\langle\vect{A}\ass x, \vect{B}\ass y\rangle\yields D\ass\omega
\justifies
\Xi\langle\vect{A\leftiprod B}\ass z\rangle\yields D\ass\omega\{z/y\}
\using \leftiprod L
\endprooftree}
\\\\
24. & \multicolumn{3}{c}{\prooftree
\Xi\langle\vect{A}\ass x, \vect{B}\ass y\rangle\yields D\ass\omega
\justifies
\Xi\langle\vect{A\rightiprod B}\ass z\rangle\yields D\ass\omega\{z/x\}
\using \rightiprod L
\endprooftree}
\end{array}$
\end{center}
\caption{Asynchronous semantically inactive continuous multiplicative rules}
\label{aunmult}
\end{figure}

\begin{figure}[ht]
\begin{center}
$
\begin{array}{lclc}
25. &
\prooftree
\zeta\sass\Gamma\smwrap{k}\vect{B}\ass y\yields C\ass\chi
\justifies
\zeta; \Gamma\yields C\funcircum{k} B\ass\chi
\using \funcircum{k} R
\endprooftree
&
26. &
\prooftree
\zeta\sass\vect{A}\ass x\smwrap{k}\Gamma\yields C\ass\chi
\justifies
\zeta\sass \Gamma\yields A\arginfix{k} C\ass\zero
\using \arginfix{k} R
\endprooftree
\\\\
27. &
\prooftree
\zeta\sass \Gamma\smwrap{k}\vect{B}\ass y\yields C\ass\chi
\justifies
\zeta\sass \Gamma\yields C\argcircum{k} B\ass\zero
\using \argcircum{k} R
\endprooftree
&
28. &
\prooftree
\zeta\sass \vect{A}\ass x\smwrap{k}\Gamma\yields C\ass\chi
\justifies
\zeta\sass \Gamma\yields A\funinfix{k} C\ass\chi
\using \funinfix{k} R
\endprooftree
\\\\
29. & \multicolumn{3}{c}{\prooftree
\Xi\langle\vect{A}\ass x\smwrap{k}\vect{B}\ass y\rangle\yields D\ass\omega
\justifies
\Xi\langle\vect{A\upperiwprod{k} B}\ass z\rangle\yields D\ass\omega\subst{z/y}
\using \upperiwprod{k} L
\endprooftree}\\\\
30. & \multicolumn{3}{c}{\prooftree
\Xi\langle\vect{A}\ass x\smwrap{k}\vect{B}\ass y\rangle\yields D\ass\omega
\justifies
\Xi\langle\vect{A\loweriwprod{k} B}\ass z\rangle\yields D\ass\omega\subst{z/x}
\using \loweriwprod{k} L
\endprooftree}
\end{array}$
\end{center}
\caption{Asynchronous semantically inactive discontinuous multiplicative rules}
\label{aslunimult}
\end{figure}

\begin{figure}[ht]
\begin{center}
$
\begin{array}{lc}
31. &
\prooftree
\Xi\yields A\ass\chi\tb\Xi\yields B\ass\chi
\justifies
\Xi\yields A\iaconj B\ass\chi
\using \iaconj R
\endprooftree
\\\\
32. &
\prooftree
\Xi\langle\vect{A}\ass z\rangle\yields C\ass\chi\tb\Xi\langle\vect{B}\ass z\rangle\yields C\ass\chi
\justifies
\Xi\langle\vect{A\iadisj B}\ass z\rangle\yields C\ass\chi
\using \iadisj L
\endprooftree
\end{array}
$
\end{center}
\caption{Asynchronous semantically inactive additive rules}
\label{aslunadds}
\end{figure}

\begin{figure}[ht]
\begin{center} 
 $
 \begin{array}{lclc}
 33. &
 \prooftree
\Xi\yields A[a/v]\ass\phi
 \justifies
\Xi\yields \forall vA\ass\phi
 \using \forall R^\dagger
 \endprooftree
&
34. &
\prooftree
\Xi\langle\vect{A[a/v]}\ass x\rangle\yields B\ass\psi
\justifies
\Xi\langle\vect{\exists vA}\ass x\rangle\yields B\ass\psi
\using \exists L^\dagger
\endprooftree
\end{array}
$
\end{center}
\caption{Asynchronous semantically inactive quantifier rules, 
where $^\dagger$ indicates that there is no $a$ in the conclusion}
\label{aslunquants}
\end{figure}

\begin{figure}[ht]
\begin{center}
$
\begin{array}{lclc}
35. &
\prooftree
 \diagmod\Xi\yields A\ass\phi
 \justifies
\diagmod\Xi\yields \imod A\ass\phi
 \using \imod R
 \endprooftree
 &
 36. &
 \prooftree
\diagmod\Xi\langle\vect{A}\ass x\rangle\yields \diagemod B\ass\psi
 \justifies
\diagmod\Xi\langle\vect{\iemod A}\ass x\rangle\yields \diagemod B\ass\psi
 \using \iemod L
 \endprooftree
 \end{array} 
 $
 \end{center}
\caption{Asynchronous semantically inactive normal modality rules; $\diagmod/\diagemod$ marks a structure all the types of
which have principal connective a box/diamond}
\label{aslunmods}
\end{figure}

\clearpage

\subsection{Unary synthetic multiplicatives}

\begin{figure}[ht]
\begin{center}
$
\begin{array}{lclc}
37. & \prooftree
\zeta\sass\Gamma, \sep\yields A\ass\phi
\justifies
\zeta\sass\Gamma\yields \leftproj A\ass\phi
\using \leftproj R
\endprooftree
&
38. &
\prooftree
\Xi\langle \vect{A}\ass x, \sep\rangle\yields B\ass\psi
\justifies
\Xi\langle\vect{\leftinj A}\ass x\rangle\yields B\ass\psi
\using \leftinj L
\endprooftree 
\\\\
39. & \prooftree
\zeta\sass\sep, \Gamma\yields A\ass\phi
\justifies
\zeta\sass\Gamma\yields\rightproj A\ass\phi
\using \rightproj R
\endprooftree
\\\\
40. &
\prooftree
\Xi\langle\sep, \vect{A}\ass x\rangle\yields B\ass\psi
\justifies
\Xi\langle\vect{\rightinj A}\ass x\rangle\yields B\ass\psi
\using \rightinj L
\endprooftree
\\\\
41. &
\prooftree
\zeta\sass\Delta\smwrap{k}\Lambda\yields B\ass\psi
\justifies
\zeta\sass\Delta\yields\ssplit{k} B\ass\psi
\using \ssplit{k} R
\endprooftree
&
42. &
\prooftree
\Xi\langle\vect{B}\ass y\smwrap{k}\Lambda\rangle\yields C\ass\chi
\justifies
\Xi\langle\vect{\sbridge{k} B}\ass y\rangle\yields C\ass\chi
\using \sbridge{k} L
\endprooftree
\end{array}
$
\end{center}
\caption{Asynchronous deterministic synthetic multiplicative rules}
\label{aslununders}
\end{figure}

\subsection{Binary synthetic multiplicatives}

\begin{figure}[ht]
\begin{center}
$
\begin{array}{lc}
43. &
\prooftree
\zeta\sass\vect{A}\ass x, \Gamma\yields C\ass\chi\tb
\zeta\sass\Gamma, \vect{A}\ass x\yields C\ass\chi
\justifies
\zeta\sass\Gamma\yields C\ndiv A\ass\lambda x\chi
\using \ndiv R
\endprooftree
\\\\
44. & \prooftree
\Xi\langle\vect{A}\ass x, \vect{B}\ass y\rangle\yields D\ass\omega\tb
\Xi\langle\vect{B}\ass y, \vect{A}\ass x\rangle\yields D\ass\omega
\justifies
\Xi\langle\vect{A\nprod B}\ass z\rangle\yields D\ass\omega\subst{\pi_1 z/x, \pi_2 z/y}
\using \nprod L
\endprooftree
\\\\
45. &
\prooftree
\zeta\sass\Gamma\smwrap{1}\vect{B}\ass y\yields C\ass\chi\tb\cdots\tb
\zeta\sass\Gamma\smwrap{sC'-sB}\vect{B}\ass y\yields C\ass\chi
\justifies
\zeta\sass\Gamma\yields C\nextract B\ass\lambda y\chi
\using \nextract R
\endprooftree
\\\\
46. &
\prooftree
\zeta\sass\vect{A}\ass x\smwrap{1}\Gamma\yields C\ass\chi\tb\cdots\tb
\zeta\sass\vect{A}\ass x\smwrap{sA}\Gamma\yields C\ass\chi
\justifies
\zeta\sass\Gamma\yields A\ninfix C\ass\lambda x\chi
\using \ninfix R
\endprooftree
\\\\
47. & \prooftree
\Xi\langle\vect{A}\ass x\smwrap{1}\vect{B}\ass y\rangle\yields D\ass\omega\tb\cdots\tb
\Xi\langle\vect{A}\ass x\smwrap{sA}\vect{B}\ass y\rangle\yields D\ass\omega
\justifies
\Xi\langle\vect{A\ndprod B}\ass z\rangle\yields D\ass\omega\subst{\pi_1 z/x, \pi_2 z/y}
\using \ndprod L
\endprooftree
\end{array}$
\end{center}
\caption{Asynchronous non-deterministic synthetic multiplicative rules}
\label{sslunbinders}
\end{figure}

\clearpage

\section{Synchronous left rules}

\disp{$
\prooftree
\justifies
\vect{\mbox{\fbox{$P$}}}\ass x\yields P\ass x
\using \mathit{id}\mbox{,\ if\ atomic\_focus(inp, $P$)}
\endprooftree
$}

\subsection{Primary connectives}

\begin{figure}[ht]
\begin{center}
$
\begin{array}{lcc}
1. & \prooftree
\zeta_1\sass  \Gamma\yields \mbox{\fbox{$P$}}\ass\psi \tb
\zeta_2\sass  \Delta\langle\vect{\mbox{\fbox{$Q$}}}\ass z\rangle\yields D\ass\omega
\justifies
\zeta_1\mun\zeta_2\sass  \Delta\langle\vect{\mbox{\fbox{$Q/P$}}}\ass x, \Gamma\rangle\yields D\ass\omega\{(x\ \psi)/z\}
\using / L
\endprooftree &
\prooftree
\zeta_1\sass  \Gamma\yields N\ass\psi \tb
\zeta_2\sass  \Delta\langle\vect{\mbox{\fbox{$Q$}}}\ass z\rangle\yields D\ass\omega
\justifies
\zeta_1\mun\zeta_2\sass  \Delta\langle\vect{\mbox{\fbox{$Q/N$}}}\ass x, \Gamma\rangle\yields D\ass\omega\{(x\ \psi)/z\}
\using / L
\endprooftree
\\\\
&\prooftree
\zeta_1\sass  \Gamma\yields \mbox{\fbox{$P$}}\ass\psi \tb
\zeta_2\sass  \Delta\langle\vect{M}\ass z\rangle\yields D\ass\omega
\justifies
\zeta_1\mun\zeta_2\sass  \Delta\langle\vect{\mbox{\fbox{$M/P$}}}\ass x, \Gamma\rangle\yields D\ass\omega\{(x\ \psi)/z\}
\using / L
\endprooftree
&
\prooftree
\zeta_1\sass  \Gamma\yields M\ass\psi \tb
\zeta_2\sass  \Delta\langle\vect{N}\ass z\rangle, \Delta\yields D\ass\omega
\justifies
\zeta_1\mun\zeta_2\sass  \Delta\langle\vect{\mbox{\fbox{$N/M$}}}\ass x, \Gamma\rangle\yields D\ass\omega\{(x\ \psi)/z\}
\using / L
\endprooftree
\\\\
2. & \prooftree
\zeta_1\sass  \Gamma\yields \mbox{\fbox{$P$}}\ass\phi \tb
\zeta_2\sass  \Delta\langle\vect{\mbox{\fbox{$Q$}}}\ass z\rangle\yields D\ass\omega
\justifies
\zeta_1\mun\zeta_2\sass  \Delta_1, \langle\Gamma, \vect{\mbox{\fbox{$P\bsl Q$}}}\ass y\rangle\yields D\ass\omega\{(y\ \phi)/z\}
\using \bsl L
\endprooftree 
&
\prooftree
\zeta_1\sass  \Gamma\yields \mbox{\fbox{$P$}}\ass\phi \tb
\zeta_2\sass  \Delta\langle\vect{M}\ass z\rangle\yields D\ass\omega
\justifies
\zeta_1\mun \zeta_2\sass  \Delta_1, \langle\Gamma, \vect{\mbox{\fbox{$P\bsl M$}}}\ass y\rangle\yields D\ass\omega\{(y\ \phi)/z\}
\using \bsl L
\endprooftree
\\\\
\end{array}$
\end{center}
\caption{Left synchronous continuous multiplicative rules}
\label{spcmult}
\end{figure}

\begin{figure}[ht]
\begin{center}
$
\begin{array}{lcc}
5. & \prooftree
\zeta_1\sass \Gamma\yields \mbox{\fbox{$P$}}\ass\psi \tb
\zeta_2\sass \Delta\langle\vect{\mbox{\fbox{$Q$}}}\ass z\rangle\yields D\ass\omega
\justifies
\zeta_1\mun\zeta_2\sass\Delta\langle\vect{\mbox{\fbox{$Q\scircum{k} P$}}}\ass x\smwrap{k}\Gamma\rangle\yields D\ass\omega\{(x\ \psi)/z\}
\using \scircum{k} L
\endprooftree 
&
\prooftree
\zeta\sass\Gamma\yields N\ass\psi \tb
\zeta_1\sass \Delta\langle\vect{\mbox{\fbox{$Q$}}}\ass z\rangle\yields D\ass\omega
\justifies
\zeta_1\mun\zeta_2\sass \Delta\langle\vect{\mbox{\fbox{$Q\scircum{k} N$}}}\ass x\smwrap{k}\Gamma\rangle\yields D\ass\omega\{(x\ \psi)/z\}
\using \scircum{k} L
\endprooftree
\\\\
&\prooftree
\zeta\sass\Gamma\yields \mbox{\fbox{$P$}}\ass\psi \tb
\zeta_1\sass \Delta\langle\vect{M}\ass z\rangle, \Delta\yields D\ass\omega
\justifies
\zeta_1\mun\zeta_2\sass \Delta\langle\vect{\mbox{\fbox{$M\scircum{k} P$}}}\ass x\smwrap{k}\Gamma\rangle\yields D\ass\omega\{(x\ \psi)/z\}
\using \scircum{k} L
\endprooftree 
&
\prooftree
\zeta_1\sass \Gamma\yields N\ass\psi \tb
\zeta_2\sass \Delta\langle\vect{M}\ass z\rangle, \Delta_2)\yields D\ass\omega
\justifies
\zeta_1\mun\zeta_2\sass \Delta\langle\vect{\mbox{\fbox{$M\scircum{k} N$}}}\ass x\smwrap{k}\Gamma\rangle\yields D\ass\omega\{(x\ \psi)/z\}
\using \scircum{k} L
\endprooftree
\\\\
6. & \prooftree
\zeta_1\sass \Gamma\yields \mbox{\fbox{$P$}}\ass\phi \tb
\zeta_2\sass \Delta\langle\vect{\mbox{\fbox{$Q$}}}\ass z\rangle, \Delta\yields D\ass\omega
\justifies
\zeta_1\mun\zeta_2\sass\Delta\langle\Gamma\smwrap{k}\vect{\mbox{\fbox{$P\sinfix{k} Q$}}}\ass y\rangle\yields D\ass\omega\{(y\ \phi)/z\}
\using \sinfix{k} L
\endprooftree 
&
\prooftree
\zeta_1\sass \Gamma\yields \mbox{\fbox{$P$}}\ass\phi \tb
\zeta_2\sass\Delta\langle\vect{M}\ass z\rangle\yields D\ass\omega
\justifies
\zeta_1\mun\zeta_2\sass\Delta\langle\Gamma\smwrap{k}\vect{\mbox{\fbox{$P\sinfix{k} M$}}}\ass y\rangle\yields D\ass\omega\{(y\ \phi)/z\}
\using \sinfix{k} L
\endprooftree
\\\\
\end{array}$
\end{center}
\caption{Left synchronous discontinuous multiplicative rules}
\label{spimult}
\end{figure}

\begin{figure}[ht]
\begin{center}
$
\begin{array}{lcc}
9. &
\prooftree
\Xi\langle\vect{\mbox{\fbox{$Q$}}}\ass x\rangle\yields C\ass\chi
\justifies
\Xi\langle\vect{\mbox{\fbox{$Q\aconj B$}}}\ass z\rangle\yields C\ass\chi\subst{\pi_1 z/x}
\using \aconj L_1
\endprooftree
&
\prooftree
\Xi\langle\vect{M}\ass x\rangle\yields C\ass\chi
\justifies
\Xi\langle\vect{\mbox{\fbox{$M\aconj B$}}}\ass z\rangle\yields C\ass\chi\subst{\pi_1 z/x}
\using \aconj L_1
\endprooftree
\\\\
&
\prooftree
\Xi\langle\vect{\mbox{\fbox{$Q$}}}\ass y\rangle\yields C\ass\chi
\justifies
\Xi\langle\vect{\mbox{\fbox{$A\aconj Q$}}}\ass z\rangle\yields C\ass\chi\subst{\pi_2 z/y}
\using \aconj L_2
\endprooftree
&
\prooftree
\Xi\langle\vect{M}\ass y\rangle\yields C\ass\chi
\justifies
\Xi\langle\vect{\mbox{\fbox{$A\aconj M$}}}\ass z\rangle\yields C\ass\chi\subst{\pi_2 z/y}
\using \aconj L_2
\endprooftree
\end{array}$
\end{center}
\caption{Left synchronous additive rules}
\label{spadd}
\end{figure}

\begin{figure}[ht]
\begin{center} 
 $
 \begin{array}{lcc}
11. & \prooftree
\Xi\langle \vect{\mbox{\fbox{$Q[t/v]$}}}\ass x\rangle\yields B\ass\psi
 \justifies
\Xi\langle\vect{\mbox{\fbox{$\bigwedge vQ$}}}\ass z\rangle\yields B\ass\psi\subst{(z\ t)/x}
 \using \bigwedge L
 \endprooftree
 &
 \prooftree
\Xi\langle \vect{M[t/v]}\ass x\rangle\yields B\ass\psi
 \justifies
\Xi\langle\vect{\mbox{\fbox{$\bigwedge vM$}}}\ass z\rangle\yields B\ass\psi\subst{(z\ t)/x}
 \using \bigwedge L
 \endprooftree 
\end{array}$
\end{center}
\caption{Left synchronous quantifier rules}
\label{slquants}
\end{figure}

\begin{figure}[ht]
\begin{center}
$
\begin{array}{lcc}
13. & \prooftree
\Xi\langle\vect{\mbox{\fbox{$Q$}}}\ass x\rangle\yields B\ass\psi
 \justifies
\Xi\langle\vect{\mbox{\fbox{$\mymod Q$}}}\ass z\rangle\yields B\ass\psi\subst{\dn z/x}
 \using \mymod L
 \endprooftree 
 &
 \prooftree
\Xi\langle\vect{M}\ass x\rangle\yields B\ass\psi
 \justifies
\Xi\langle\vect{\mbox{\fbox{$\mymod M$}}}\ass z\rangle\yields B\ass\psi\subst{\dn z/x}
 \using \mymod L
 \endprooftree 
\end{array}
$
 \end{center}
\caption{Left synchronous normal modality rules}
\label{sslmods}
\end{figure}

\begin{figure}[ht]
\begin{center}
$
\begin{array}{lcc}
15. & \prooftree
\Xi\langle \vect{\mbox{\fbox{$Q$}}}\ass x\rangle\yields B\ass\psi
\justifies
\Xi\langle[\vect{\mbox{\fbox{$\abrack Q$}}}\ass x]\rangle\yields B\ass\psi
\using \abrack L
\endprooftree &
\prooftree
\Xi\langle \vect{M}\ass x\rangle\yields B\ass\psi
\justifies
\Xi\langle[\vect{\mbox{\fbox{$\abrack M$}}}\ass x]\rangle\yields B\ass\psi
\using \abrack L
\endprooftree 
\end{array}$
 \end{center}
\caption{Left synchronous bracket modality rules}
\label{sslbrmods}
\end{figure}

\commentout{

\begin{figure}[ht]
\begin{center}
$
\begin{array}{lcc}
19. &
\prooftree
\zeta\sass\Gamma\yields \fbox{$P$}\ass\phi\tb\zeta'\sass\Delta\langle \vect{P}\ass x; \vect{\fbox{$Q$}}\ass y\rangle\yields D\ass\omega
\justifies
\zeta\mun\zeta'\sass\Delta\langle\Gamma; \vect{\fbox{$Q|P$}}\ass z\rangle\yields D\ass\omega\subst{\phi/x, (z\ \phi)/y}
\using |L
\endprooftree
&
\prooftree
\zeta\sass\Gamma\yields N\ass\phi\tb\zeta'\sass\Delta\langle \vect{N}\ass x; \vect{\fbox{$Q$}}\ass y\rangle\yields D\ass\omega
\justifies
\zeta\mun\zeta'\sass\Delta\langle\Gamma; \vect{\fbox{$Q|N$}}\ass z\rangle\yields D\ass\omega\subst{\phi/x, (z\ \phi)/y}
\using |L
\endprooftree
\\\\
&\prooftree
\zeta\sass\Gamma\yields \fbox{$P$}\ass\phi\tb\zeta'\sass\Delta\langle \vect{P}\ass x; \vect{M}\ass y\rangle\yields D\ass\omega
\justifies
\zeta\mun\zeta'\sass\Delta\langle\Gamma; \vect{\fbox{$M|P$}}\ass z\rangle\yields D\ass\omega\subst{\phi/x, (z\ \phi)/y}
\using |L
\endprooftree
&
\prooftree
\zeta\sass\Gamma\yields N\ass\phi\tb\zeta'\sass\Delta\langle \vect{N}\ass x; \vect{M}\ass y\rangle\yields D\ass\omega
\justifies
\zeta\mun\zeta'\sass\Delta\langle\Gamma; \vect{\fbox{$M|N$}}\ass z\rangle\yields D\ass\omega\subst{\phi/x, (z\ \phi)/y}
\using |L
\endprooftree
\end{array}
$
\commentout{ZZZexpansion?}
\end{center}
\caption{Left synchronous limited contraction for anaphora rules}
\label{sllcas}
\end{figure}

}

\clearpage

\subsection{Semantically inactive variants}

\begin{figure}[ht]
\begin{center}
$
\begin{array}{lcc}
19. & \prooftree
\zeta\sass  \Gamma\yields \mbox{\fbox{$P$}}\ass\psi \tb
\Xi(\zeta'\sass  \Delta_1\langle\vect{\mbox{\fbox{$Q$}}}\ass z\rangle, \Delta_2)\yields D\ass\omega
\justifies
\Xi(\zeta\mun\zeta'\sass  \Delta_1, \langle\vect{\mbox{\fbox{$Q\argover P$}}}\ass x, \Gamma\rangle, \Delta_2)\yields D\ass\omega\{\zero/z\}
\using \argover L
\endprooftree
&
\prooftree
\zeta\sass  \Gamma\yields N\ass\psi \tb
\Xi(\zeta'\sass  \Delta_1, \langle\vect{\mbox{\fbox{$Q$}}}\ass z\rangle, \Delta_2)\yields D\ass\omega
\justifies
\Xi(\zeta\mun\zeta'\sass  \Delta_1, \langle\vect{\mbox{\fbox{$Q\argover N$}}}\ass x, \Gamma\rangle, \Delta_2)\yields D\ass\omega\{\zero/z\}
\using \argover L
\endprooftree
\\\\
&\prooftree
\zeta\sass  \Gamma\yields \mbox{\fbox{$P$}}\ass\psi \tb
\Xi(\zeta'\sass  \Delta_1, \langle\vect{M}\ass z\rangle, \Delta_2)\yields D\ass\omega
\justifies
\Xi(\zeta\mun\zeta'\sass  \Delta_1, \langle\vect{\mbox{\fbox{$M\argover P$}}}\ass x, \Gamma\rangle, \Delta_2)\yields D\ass\omega\{\zero/z\}
\using \argover L
\endprooftree
&
\prooftree
\zeta\sass  \Gamma\yields M\ass\psi \tb
\Xi(\zeta'\sass  \Delta_1, \langle\vect{N}\ass z\rangle, \Delta_2)\yields D\ass\omega
\justifies
\Xi(\zeta\mun\zeta\sass  \Delta_1, \langle\vect{\mbox{\fbox{$N\argover M$}}}\ass x, \Gamma\rangle, \Delta_2)\yields D\ass\omega\{\zero/z\}
\using \argover L
\endprooftree
\\\\
\end{array}$
\end{center}
\caption{Left synchronous semantically inactive continuous multiplicative rules}
\label{spcunmult}
\end{figure}

\begin{figure}[ht]
\begin{center}
$
\begin{array}{lcc}
25. &\prooftree
\zeta\sass  \Gamma\yields \mbox{\fbox{$P$}}\ass\psi \tb
\Xi(\zeta'\sass  \Delta_1\langle\vect{\mbox{\fbox{$Q$}}}\ass z\rangle, \Delta_2)\yields D\ass\omega
\justifies
\Xi(\zeta\mun\zeta'\sass  \Delta_1, \langle\vect{\mbox{\fbox{$Q\funcircum{k} P$}}}\ass x\smwrap{k} \Gamma\rangle, \Delta_2)\yields D\ass\omega\{x/z\}
\using \funcircum{k} L
\endprooftree
&
\prooftree
\zeta\sass  \Gamma\yields N\ass\psi \tb
\Xi(\zeta'\sass  \Delta_1, \langle\vect{\mbox{\fbox{$Q$}}}\ass z\rangle, \Delta_2)\yields D\ass\omega
\justifies
\Xi(\zeta\mun\zeta'\sass  \Delta_1, \langle\vect{\mbox{\fbox{$Q\funcircum{k} N$}}}\ass x\smwrap{k} \Gamma\rangle, \Delta_2)\yields D\ass\omega\{x/z\}
\using \funcircum{k} L
\endprooftree
\\\\
&\prooftree
\zeta\sass  \Gamma\yields \mbox{\fbox{$P$}}\ass\psi \tb
\Xi(\zeta'\sass  \Delta_1, \langle\vect{M}\ass z\rangle, \Delta_2)\yields D\ass\omega
\justifies
\Xi(\zeta\mun\zeta'\sass  \Delta_1, \langle\vect{\mbox{\fbox{$M\funcircum{k} P$}}}\ass x\smwrap{k} \Gamma\rangle, \Delta_2)\yields D\ass\omega\{x/z\}
\using \funcircum{k} L
\endprooftree 
&
\prooftree
\zeta\sass  \Gamma\yields M\ass\psi \tb
\Xi(\zeta'\sass  \Delta_1, \langle\vect{N}\ass z\rangle, \Delta_2)\yields D\ass\omega
\justifies
\Xi(\zeta\mun\zeta\sass  \Delta_1, \langle\vect{\mbox{\fbox{$N\funcircum{k} M$}}}\ass x\smwrap{k} \Gamma\rangle, \Delta_2)\yields D\ass\omega\{x/z\}
\using \funcircum{k} L
\endprooftree
\\\\
\end{array}$
\end{center}
\caption{Left synchronous semantically inactive discontinuous multiplicative rules}
\label{spcunimult}
\end{figure}

\begin{figure}[ht]
\begin{center}
$
\begin{array}{lcc}
31. &
\prooftree
\Xi\langle\vect{\mbox{\fbox{$Q$}}}\ass x\rangle\yields C\ass\chi
\justifies
\Xi\langle\vect{\mbox{\fbox{$Q\iaconj B$}}}\ass x\rangle\yields C\ass\chi
\using \iaconj L_1
\endprooftree
&
\prooftree
\Xi\langle\vect{M}\ass x\rangle\yields C\ass\chi
\justifies
\Xi\langle\vect{\mbox{\fbox{$M\iaconj B$}}}\ass x\rangle\yields C\ass\chi
\using \iaconj L_1
\endprooftree
\\\\
&
\prooftree
\Xi\langle\vect{\mbox{\fbox{$Q$}}}\ass y\rangle\yields C\ass\chi
\justifies
\Xi\langle\vect{\mbox{\fbox{$A\iaconj Q$}}}\ass y\rangle\yields C\ass\chi
\using \iaconj L_2
\endprooftree
&
\prooftree
\Xi\langle\vect{M}\ass y\rangle\yields C\ass\chi
\justifies
\Xi\langle\vect{\mbox{\fbox{$A\iaconj M$}}}\ass y\rangle\yields C\ass\chi
\using \iaconj L_2
\endprooftree
\end{array}$
\end{center}
\caption{Semantically inactive additives}
\label{lsslunadds}
\end{figure}

\begin{figure}[ht]
\begin{center} 
 $
 \begin{array}{lcc}
33. &
 \prooftree
\Xi\langle \vect{\mbox{\fbox{$Q[t/v]$}}}\ass x\rangle\yields B\ass\psi
 \justifies
\Xi\langle\vect{\mbox{\fbox{$\forall vQ$}}}\ass x\rangle\yields B\ass\psi
 \using \forall L
 \endprooftree
 &
  \prooftree
\Xi\langle \vect{M[t/v]}\ass x\rangle\yields B\ass\psi
 \justifies
\Xi\langle\vect{\mbox{\fbox{$\forall vM$}}}\ass x\rangle\yields B\ass\psi
 \using \forall L
 \endprooftree
\end{array}$
\end{center}
\caption{Left synchronous semantically inactive quantifier rules}
\label{lsslunquants}
\end{figure}

\begin{figure}[ht]
\begin{center}
$
\begin{array}{lcc}
35. &
 \prooftree
\Xi\langle\vect{\mbox{\fbox{$Q$}}}\ass x\rangle\yields B\ass\psi
 \justifies
\Xi\langle\vect{\mbox{\fbox{$\imod Q$}}}\ass x\rangle\yields B\ass\psi
 \using \imod L
 \endprooftree 
 &
\prooftree
\Xi\langle\vect{M}\ass x\rangle\yields B\ass\psi
 \justifies
\Xi\langle\vect{\mbox{\fbox{$\imod M$}}}\ass x\rangle\yields B\ass\psi
 \using \imod L
 \endprooftree 
 \end{array}
$
 \end{center}
\caption{Left synchronous semantically inactive normal modality rules}
\label{lsslunmods}
\end{figure}

\clearpage

\subsection{Unary synthetic multiplicatives}

\begin{figure}[ht]
\begin{center}
$
\begin{array}{lcc}
37. & 
\prooftree
\Xi\langle\vect{\mbox{\fbox{$Q$}}}\ass x\rangle\yields B\ass\psi
\justifies
\Xi\langle\vect{\mbox{\fbox{$\leftproj Q$}}}\ass x, \sep\rangle\yields B\ass\psi
\using \leftproj L
\endprooftree
&
\prooftree
\Xi\langle\vect{M}\ass x\rangle\yields B\ass\psi
\justifies
\Xi\langle\vect{\mbox{\fbox{$\leftproj M$}}}\ass x, \sep\rangle\yields B\ass\psi
\using \leftproj L
\endprooftree
\\\\
38. & \prooftree
\Xi\langle\vect{\mbox{\fbox{$Q$}}}\ass x\rangle\yields B\ass\psi
\justifies
\Xi\langle\sep, \vect{\mbox{\fbox{$\rightproj Q$}}}\ass x\rangle\yields B\ass\psi
\using \rightproj L
\endprooftree
&
\prooftree
\Xi\langle\vect{M}\ass x\rangle\yields B\ass\psi
\justifies
\Xi\langle\sep, \vect{\mbox{\fbox{$\rightproj M$}}}\ass x\rangle\yields B\ass\psi
\using \rightproj L
\endprooftree
\\\\
41. & \prooftree
\Xi\langle \vect{\mbox{\fbox{$Q$}}}\ass y\rangle\yields C\ass\chi
\justifies
\Xi\langle\vect{\mbox{\fbox{$\ssplit{k} Q$}}}\ass y\smwrap{k}\Lambda\rangle\yields C\ass\chi
\using \ssplit{k} L
\endprooftree
&
\prooftree
\Xi\langle \vect{M}\ass y\rangle\yields C\ass\chi
\justifies
\Xi\langle\vect{\mbox{\fbox{$\ssplit{k} M$}}}\ass y\smwrap{k}\Lambda\rangle\yields C\ass\chi
\using \ssplit{k} L
\endprooftree
\end{array}
$
\end{center}
\caption{Left synchronous deterministic synthetic multiplicative rules}
\label{lsslununders}
\end{figure}

\subsection{Binary synthetic multiplicatives}

\begin{figure}[ht]
\begin{center}
$
\begin{array}{lcc}
43. & \prooftree
\zeta\sass\Gamma\yields \mbox{\fbox{$P$}}\ass\phi \tb
\Xi(\zeta'\sass\Delta_1, \langle\vect{\mbox{\fbox{$Q$}}}\ass z\rangle, \Delta_2)\yields D\ass\omega
\justifies
\Xi(\zeta\mun\zeta'\sass\Delta_1, \langle\Gamma, \vect{\mbox{\fbox{$Q\ndiv P$}}}\ass y\rangle, \Delta_2)\yields D\ass\omega\subst{(y\ \phi)/z}
\using \ndiv L_1
\endprooftree &
\prooftree
\zeta\sass\Gamma\yields N\ass\phi \tb
\Xi(\zeta'\sass\Delta_1, \vect{\mbox{\fbox{$Q$}}}\ass z, \Delta_2)\yields D\ass\omega
\justifies
\Delta\langle\zeta\mun\zeta'\sass\Delta_1, \Gamma, \vect{\mbox{\fbox{$Q\ndiv N$}}}\ass y, \Delta_2\rangle\yields D\ass\omega\subst{(y\ \phi)/z}
\using \ndiv L_1
\endprooftree
\\\\
&
\prooftree
\zeta\sass\Gamma\yields \mbox{\fbox{$P$}}\ass\phi \tb
\Xi(\zeta'\sass\Delta_1, \langle\vect{M}\ass z\rangle, \Delta_2)\yields D\ass\omega
\justifies
\Xi(\zeta\mun\zeta'\sass\Delta_1, \langle\Gamma, \vect{\mbox{\fbox{$M\ndiv P$}}}\ass y\rangle, \Delta_2)\yields D\ass\omega\subst{(y\ \phi)/z}
\using \ndiv L_1
\endprooftree
&
\prooftree
\zeta\sass\Gamma\yields N\ass\phi \tb
\Xi(\zeta'\sass\Delta_1, \langle\vect{M}\ass z\rangle, \Delta_2)\yields D\ass\omega
\justifies
\Xi(\zeta\mun\zeta'\sass\Delta_1, \langle\Gamma, \vect{\mbox{\fbox{$M\ndiv N$}}}\ass y\rangle, \Delta_2)\yields D\ass\omega\subst{(y\ \phi)/z}
\using \ndiv L_1
\endprooftree
\\\\
\end{array}
$
\end{center}
\caption{Left synchronous continuous non-deterministic synthetic multiplicative rules}
\label{laslunbindersI}
\end{figure}

\begin{figure}[ht]
\begin{center}
$
\begin{array}{lcc}
45. & \prooftree
\zeta\sass\Gamma\yields \mbox{\fbox{$P$}}\ass\psi \tb
\Xi(\zeta'\sass\Delta_1, \langle\vect{\mbox{\fbox{$Q$}}}\ass z\rangle, \Delta_2)\yields D\ass\omega
\justifies
\Xi(\zeta\mun\zeta'\sass\Delta_1, \langle\vect{\mbox{\fbox{$Q\nextract P$}}}\ass x\smwrap{k}\Gamma\rangle, \Delta_2)\yields D\ass\omega\subst{(x\ \psi)/z}
\using \nextract L
\endprooftree 
&
\prooftree
\zeta\sass\Gamma\yields N\ass\psi \tb
\Xi(\zeta'\sass\Delta_1, \langle\vect{\mbox{\fbox{$Q$}}}\ass z\rangle, \Delta_2)\yields D\ass\omega
\justifies
\Xi(\zeta\mun\zeta'\sass\Delta_1, \langle\vect{\mbox{\fbox{$Q\nextract N$}}}\ass x\smwrap{k}\Gamma\rangle, \Delta_2)\yields D\ass\omega\subst{(x\ \psi)/z}
\using \nextract L
\endprooftree
\\\\
&
\prooftree
\zeta\sass\Gamma\yields \mbox{\fbox{$P$}}\ass\psi \tb
\Xi(\zeta'\sass\Delta_1, \langle\vect{M}\ass z\rangle, \Delta_2)\yields D\ass\omega
\justifies
\Xi\langle\zeta\mun\zeta'\sass\Delta_1, \langle\vect{\mbox{\fbox{$M\nextract P$}}}\ass x\smwrap{k}\Gamma\rangle, \Delta_2)\yields D\ass\omega\subst{(x\ \psi)/z}
\using \nextract L
\endprooftree
&
\prooftree
\zeta\sass\Gamma\yields N\ass\psi \tb
\Xi(\zeta'\sass\Delta_1, \langle\vect{M}\ass z\rangle, \Delta_2)\yields D\ass\omega
\justifies
\Xi(\zeta\mun\zeta'\sass\Delta_1, \langle\vect{\mbox{\fbox{$M\nextract N$}}}\ass x\smwrap{k}\Gamma\rangle, \Delta_2)\yields D\ass\omega\subst{(x\ \psi)/z}
\using \nextract L
\endprooftree
\\\\
\end{array}
$
\end{center}
\caption{Left synchronous discontinuous non-deterministic synthetic multiplicative rules}
\label{laslunbindersII}
\end{figure}

\clearpage

\section{Synchronous stoup rules}

\begin{figure}[ht]
\begin{center}
$
 \begin{array}{lcc}
 17. & \multicolumn{2}{c}{\prooftree
\Xi(\zeta\sass\Gamma_1, A\ass x, \Gamma_2)\yields B\ass\psi
 \justifies
 \Xi(\zeta\mun\{\mbox{\fbox{$A$}}\ass x\}\sass\Gamma_1, \Gamma_2)\yields B\ass\psi
 \using \univexp L
 \endprooftree}
 \\\\
 & \multicolumn{2}{c}{\prooftree
 \Xi(\zeta\mun\{\mbox{\fbox{A}}\ass x\}; \Gamma_1, [\{A\ass y\}; \Gamma_2], \Gamma_3)\yields B\ass\psi
 \justifies
 \Xi(\zeta\mun\{\mbox{\fbox{A}}\ass x\}; \Gamma_1, \Gamma_2, \Gamma_3)\yields B\ass\psi\subst{x/y}
 \using C
 \endprooftree}
 \end{array}$
 \end{center}
\caption{Synchronous exponential stoup rules}
\label{sslexps}
\end{figure}

\section{Synchronous right rules}

\disp{$
\prooftree
\justifies
\vect{P}\ass x\yields \mbox{\fbox{$P$}}\ass x
\using \mathit{id}\mbox{,\ if\ atomic\_focus(out, $P$)}
\endprooftree
$}

\subsection{Primary connectives}

\begin{figure}[ht]
\begin{center}
$
\begin{array}{lc}
3. &
\prooftree
\zeta\sass  \Gamma_1\yields \mbox{\fbox{$P_1$}}\ass\phi\tb\zeta'\sass  \Gamma_2\yields \mbox{\fbox{$P_2$}}\ass\psi
\justifies
\zeta\mun\zeta'\sass  \Gamma_1, \Gamma_2\yields \mbox{\fbox{$P_1\product P_2$}}\ass(\phi, \psi)
\using \product R
\endprooftree\tb
\prooftree
\zeta\sass  \Gamma_1\yields \mbox{\fbox{$P$}}\ass\phi\tb\zeta'\sass  \Gamma_2\yields N\ass\psi
\justifies
\zeta\mun\zeta'\sass  \Gamma_1, \Gamma_2\yields \mbox{\fbox{$P\product N$}}\ass(\phi, \psi)
\using \product R
\endprooftree
\\\\
&
\prooftree
\zeta\sass  \Gamma_1\yields N\ass\phi\tb\zeta'\sass  \Gamma_2\yields \mbox{\fbox{$P$}}\ass\psi
\justifies
\zeta\mun\zeta'\sass  \Gamma_1, \Gamma_2\yields \mbox{\fbox{$N\product P$}}\ass(\phi, \psi)
\using \product R
\endprooftree\tb
\prooftree
\zeta\sass  \Gamma_1\yields N_1\ass\phi\tb\zeta'\sass  \Gamma_2\yields N_2\ass\psi
\justifies
\zeta\mun\zeta'\sass  \Gamma_1, \Gamma_2\yields \mbox{\fbox{$N_1\product N_2$}}\ass(\phi, \psi)
\using \product R
\endprooftree
\\\\
4. & 
\prooftree
\justifies
\Lambda\yields \mbox{\fbox{$I$}}\ass\zero
\using IR
\endprooftree
\end{array}
$
\end{center}
\caption{Right synchronous continuous multiplicative rules}
\label{rspcmult}
\end{figure}

\begin{figure}[ht]
\begin{center}
$
\begin{array}{lc}
7. &
\prooftree
\zeta\sass \Gamma_1\yields \mbox{\fbox{$P_1$}}\ass\phi\tb\zeta'\sass\Gamma_2\yields \mbox{\fbox{$P_2$}}\ass\psi
\justifies
\zeta\mun\zeta'\sass\Gamma_1\smwrap{k}\Gamma_2\yields \mbox{\fbox{$P_1\swprod{k}P_2$}}\ass(\phi, \psi)
\using \swprod{k} R
\endprooftree\tb
\prooftree
\zeta\sass\Gamma_1\yields \mbox{\fbox{$P$}}\ass\phi\tb\zeta'\sass\Gamma_2\yields N\ass\psi
\justifies
\zeta\mun\zeta'\sass\Gamma_1\smwrap{k}\Gamma_2\yields \mbox{\fbox{$P\swprod{k}N$}}\ass(\phi, \psi)
\using \swprod{k} R
\endprooftree
\\\\
&
\prooftree
\zeta\sass\Gamma_1\yields N\ass\phi\tb\zeta'\sass\Gamma_2\yields \mbox{\fbox{$P$}}\ass\psi
\justifies
\zeta\mun\zeta'\sass \Gamma_1\smwrap{k}\Gamma_2\yields \mbox{\fbox{$N\swprod{k}P$}}\ass(\phi, \psi)
\using \swprod{k} R
\endprooftree\tb
\prooftree
\zeta\sass\Gamma_1\yields N_1\ass\phi\tb\zeta'\sass\Gamma_2\yields N_2\ass\psi
\justifies
\zeta\mun\zeta'\sass\Gamma_1\smwrap{k}\Gamma_2\yields \mbox{\fbox{$N_1\swprod{k}N_2$}}\ass(\phi, \psi)
\using \swprod{k} R
\endprooftree
\\\\
8. &
\prooftree
\justifies
\sep\yields \mbox{\fbox{$J$}}\ass\zero
\using JR
\endprooftree
\end{array}
$
\end{center}
\caption{Right synchronous discontinuous multiplicative rules}
\label{rspimult}
\end{figure}

\begin{figure}[ht]
\begin{center}
$
\begin{array}{lc}
10. & \prooftree
\Xi\yields \mbox{\fbox{$P$}}\ass\phi
\justifies
\Xi\yields \mbox{\fbox{$P\adisj B$}}\ass\iota_1\phi
\using \adisj R_1
\endprooftree
\tb
\prooftree
\Xi\yields N\ass\phi
\justifies
\Xi\yields \mbox{\fbox{$N\adisj B$}}\ass\iota_1\phi
\using \adisj R_1
\endprooftree
\\\\
&
\prooftree
\Xi\yields \mbox{\fbox{$P$}}\ass\psi
\justifies
\Xi \yields \mbox{\fbox{$A\adisj P$}}\ass\iota_2\psi
\using \adisj R_2
\endprooftree
\tb
\prooftree
\Xi\yields N\ass\psi
\justifies
\Xi \yields \mbox{\fbox{$A\adisj N$}}\ass\iota_2\psi
\using \adisj R_2
\endprooftree
\end{array}
$
\end{center}
\caption{Right synchronous additive rules}
\label{rspadd}
\end{figure}

\begin{figure}[ht]
\begin{center} 
$
\begin{array}{lc}
12. & 
\prooftree
\Xi\yields \mbox{\fbox{$P[t/v]$}}\ass\phi
\justifies
\Xi\yields\mbox{\fbox{$\bigvee vP$}}\ass(t, \phi)
\using \bigvee R
\endprooftree
\tb
\prooftree
\Xi\yields N[t/v]\ass\phi
\justifies
\Xi\yields\mbox{\fbox{$\bigvee vN$}}\ass(t, \phi)
\using \bigvee R
\endprooftree
\end{array}$
\end{center}
\caption{Right synchronous quantifier rules}
\label{rslquants}
\end{figure}

\begin{figure}[ht]
\begin{center}
$
\begin{array}{lc}
14. &
\prooftree
\Xi\yields \mbox{\fbox{$P$}}\ass\phi
 \justifies
\Xi\yields \mbox{\fbox{$\emod P$}}\ass\upc\phi
 \using \emod R
 \endprooftree
\tb
\prooftree
\Xi\yields N\ass\phi
 \justifies
\Xi\yields \mbox{\fbox{$\emod N$}}\ass\upc\phi
 \using \emod R
 \endprooftree
 \end{array}
 $
 \end{center}
\caption{Right synchronous normal modality rules}
\label{rsslmods}
\end{figure}

\begin{figure}[ht]
\begin{center}
$
\begin{array}{lc}
16. &
\prooftree
\Xi\yields \mbox{\fbox{$P$}}\ass\phi
\justifies
[\Xi]\yields\mbox{\fbox{$\mybrack P$}}\ass\phi
\using \mybrack R
\endprooftree
 \tb
\prooftree
\Xi\yields N\ass\phi
\justifies
[\Xi]\yields\mbox{\fbox{$\mybrack N$}}\ass\phi
\using \mybrack R
\endprooftree
\end{array}$
 \end{center}
\caption{Right synchronous bracket modality rules}
\label{rsslbrmods}
\end{figure}

\begin{figure}[ht]
\begin{center}
 $
 \begin{array}{lcc}
 18. & \prooftree
\Xi\yields \mbox{\fbox{$P$}}\ass\phi
\justifies
\Xi\yields \mbox{\fbox{$\exstexp P$}}\ass[\phi]
\using \exstexp R
\endprooftree
&
 \prooftree
\Xi\yields N\ass\phi
\justifies
\Xi\yields  \mbox{\fbox{$\exstexp N$}}\ass[\phi]
\using \exstexp R
\endprooftree
\\\\
&
\prooftree
\zeta\sass\Gamma\yields \mbox{\fbox{$P$}}\ass\phi\tb\zeta'\sass \Delta\yields \mbox{\fbox{$\exstexp P$}}\ass\psi
\justifies
\zeta\mun\zeta'\sass\Gamma, \Delta\yields \mbox{\fbox{$\exstexp P$}}\ass[\phi|\psi]
\using \exstexp M
\endprooftree
&
\prooftree
\zeta\sass\Gamma\yields N\ass\phi\tb\zeta'\sass \Delta\yields \exstexp N\ass\psi
\justifies
\zeta\mun\zeta'\sass\Gamma, \Delta\yields \mbox{\fbox{$\exstexp N$}}\ass[\phi|\psi]
\using \exstexp M
\endprooftree
\end{array}
$
 \end{center}
\caption{Right synchronous subexponential rules}
\label{rsslexps}
\end{figure}

\clearpage

\subsection{Semantically inactive variants}

\begin{figure}[ht]
\begin{center}
$
\begin{array}{lcc}
23. &
\prooftree
\zeta\sass  \Gamma_1\yields \mbox{\fbox{$P_1$}}\ass\phi\tb\zeta'\sass  \Gamma_2\yields \mbox{\fbox{$P_2$}}\ass\psi
\justifies
\zeta\mun\zeta'\sass  \Gamma_1, \Gamma_2\yields \mbox{\fbox{$P_1\leftiprod P_2$}}\ass\psi
\using \leftiprod R
\endprooftree\tb
\prooftree
\zeta\sass  \Gamma_1\yields \mbox{\fbox{$P$}}\ass\phi\tb\zeta'\sass  \Gamma_2\yields N\ass\psi
\justifies
\zeta\mun\zeta'\sass  \Gamma_1, \Gamma_2\yields \mbox{\fbox{$P\leftiprod N$}}\ass\psi
\using \leftiprod R
\endprooftree
\\\\
&
\prooftree
\zeta\sass  \Gamma_1\yields N\ass\phi\tb\zeta'\sass  \Gamma_2\yields \mbox{\fbox{$P$}}\ass\psi
\justifies
\zeta\mun\zeta'\sass  \Gamma_1, \Gamma_2\yields \mbox{\fbox{$N\leftiprod P$}}\ass\psi
\using \leftiprod R
\endprooftree\tb
\prooftree
\zeta\sass  \Gamma_1\yields N_1\ass\phi\tb\zeta'\sass  \Gamma_2\yields N_2\ass\psi
\justifies
\zeta\mun\zeta'\sass  \Gamma_1, \Gamma_2\yields \mbox{\fbox{$N_1\leftiprod N_2$}}\ass\psi
\using \leftiprod R
\endprooftree
\\\\
24. &
\prooftree
\zeta\sass  \Gamma_1\yields \mbox{\fbox{$P_1$}}\ass\phi\tb\zeta'\sass  \Gamma_2\yields \mbox{\fbox{$P_2$}}\ass\psi
\justifies
\zeta\mun\zeta'\sass  \Gamma_1, \Gamma_2\yields \mbox{\fbox{$P_1\rightiprod P_2$}}\ass\phi
\using \rightiprod R
\endprooftree\tb
\prooftree
\zeta\sass  \Gamma_1\yields \mbox{\fbox{$P$}}\ass\phi\tb\zeta'\sass  \Gamma_2\yields N\ass\psi
\justifies
\zeta\mun\zeta'\sass  \Gamma_1, \Gamma_2\yields \mbox{\fbox{$P\rightiprod N$}}\ass\phi
\using \rightiprod R
\endprooftree
\\\\
\end{array}
$
\end{center}
\caption{Right synchronous rules for semantically inactive continuous multiplicatives}
\label{rsuncconc}
\end{figure}

\begin{figure}[ht]
\begin{center}
$
\begin{array}{lcc}
29. &
\prooftree
\zeta\sass  \Gamma_1\yields \mbox{\fbox{$P_1$}}\ass\phi\tb\zeta'\sass  \Gamma_2\yields \mbox{\fbox{$P_2$}}\ass\psi
\justifies
\zeta\mun\zeta'\sass  \Gamma_1\smwrap{k} \Gamma_2\yields \mbox{\fbox{$P_1\upperiwprod{k} P_2$}}\ass\psi
\using \upperiwprod{k} R
\endprooftree\tb
\prooftree
\zeta\sass  \Gamma_1\yields \mbox{\fbox{$P$}}\ass\phi\tb\zeta'\sass  \Gamma_2\yields N\ass\psi
\justifies
\zeta\mun\zeta'\sass  \Gamma_1\smwrap{k} \Gamma_2\yields \mbox{\fbox{$P\upperiwprod{k} N$}}\ass\psi
\using \upperiwprod{k} R
\endprooftree
\\\\
&
\prooftree
\zeta\sass  \Gamma_1\yields N\ass\phi\tb\zeta'\sass  \Gamma_2\yields \mbox{\fbox{$P$}}\ass\psi
\justifies
\zeta\mun\zeta'\sass  \Gamma_1\smwrap{k} \Gamma_2\yields \mbox{\fbox{$N\upperiwprod{k} P$}}\ass\psi
\using \upperiwprod{k} R
\endprooftree\tb
\prooftree
\zeta\sass  \Gamma_1\yields N_1\ass\phi\tb\zeta'\sass  \Gamma_2\yields N_2\ass\psi
\justifies
\zeta\mun\zeta'\sass  \Gamma_1\smwrap{k} \Gamma_2\yields \mbox{\fbox{$N_1\upperiwprod{k} N_2$}}\ass\psi
\using \upperiwprod{k} R
\endprooftree\\\\
30. &
\prooftree
\zeta\sass  \Gamma_1\yields \mbox{\fbox{$P_1$}}\ass\phi\tb\zeta'\sass  \Gamma_2\yields \mbox{\fbox{$P_2$}}\ass\psi
\justifies
\zeta\mun\zeta'\sass  \Gamma_1\smwrap{k} \Gamma_2\yields \mbox{\fbox{$P_1\loweriwprod{k} P_2$}}\ass\phi
\using \loweriwprod R
\endprooftree\tb
\prooftree
\zeta\sass  \Gamma_1\yields \mbox{\fbox{$P$}}\ass\phi\tb\zeta'\sass  \Gamma_2\yields N\ass\psi
\justifies
\zeta\mun\zeta'\sass  \Gamma_1\smwrap{k} \Gamma_2\yields \mbox{\fbox{$P\loweriwprod{k} N$}}\ass\phi
\using \loweriwprod{k} R
\endprooftree
\\\\
\end{array}
$
\end{center}
\caption{Right synchronous rules for semantically inactive discontinuous multiplicatives}
\label{rsuncinterc}
\end{figure}

\begin{figure}[ht]
\begin{center}
$
\begin{array}{lcc}
32. &
\prooftree
\Xi\yields \mbox{\fbox{$P$}}\ass\phi
\justifies
\Xi\yields \mbox{\fbox{$P\iadisj B$}}\ass\phi
\using \iadisj R_1
\endprooftree\tb
\prooftree
\Xi\yields N\ass\phi
\justifies
\Xi\yields \mbox{\fbox{$N\iadisj B$}}\ass\phi
\using \iadisj R_1
\endprooftree
\\\\
&
\prooftree
\Xi\yields P\ass\psi
\justifies
\Xi \yields \mbox{\fbox{$A\iadisj P$}}\ass\psi
\using \iadisj R_2
\endprooftree
\tb
\prooftree
\Xi\yields N\ass\psi
\justifies
\Xi \yields \mbox{\fbox{$A\iadisj N$}}\ass\psi
\using \iadisj R_2
\endprooftree
\end{array}$
\end{center}
\caption{Right synchronous semantically inactive additive rules}
\label{rsslunadds}
\end{figure}

\begin{figure}[ht]
\begin{center} 
$
\begin{array}{lcc}
34. &
\prooftree
\Xi\yields \mbox{\fbox{$P[t/v]$}}\ass\phi
\justifies
\Xi\yields\mbox{\fbox{$\exists vP$}}\ass\phi
\using \exists R
\endprooftree
\tb
\prooftree
\Xi\yields N[t/v]\ass\phi
\justifies
\Xi\yields\mbox{\fbox{$\exists vN$}}\ass\phi
\using \exists R
\endprooftree
\end{array}
$
\end{center}
\caption{Right synchronous semantically inactive quantifier rules}
\label{rsslunquants}
\end{figure}

\begin{figure}[ht]
\begin{center}
$
\begin{array}{lcc}
36. &
\prooftree
\Xi\yields \mbox{\fbox{$P$}}\ass\phi
 \justifies
\Xi\yields \mbox{\fbox{$\iemod P$}}\ass\phi
 \using \iemod R
 \endprooftree
\tb
\prooftree
\Xi\yields N\ass\phi
 \justifies
\Xi\yields \mbox{\fbox{$\iemod N$}}\ass\phi
 \using \iemod R
 \endprooftree
 \end{array}
 $
 \end{center}
\caption{Right synchronous semantically inactive normal modality rules}
\label{rsslunmods}
\end{figure}

\clearpage

\subsection{Unary synthetic multiplicatives}

\begin{figure}[ht]
\begin{center}
$
\begin{array}{rcc}
39. & \prooftree
\zeta\sass\Gamma\yields \mbox{\fbox{$P$}}\ass\phi
\justifies
\zeta\sass\Gamma, \sep\yields\mbox{\fbox{$\leftinj P$}}\ass\phi
\using \leftinj R
\endprooftree\tb
\prooftree
\zeta\sass\Gamma\yields N\ass\phi
\justifies
\zeta\sass\Gamma, \sep\yields \mbox{\fbox{$\leftinj N$}}\ass\phi
\using \leftinj R
\endprooftree
\\\\
40. &
\prooftree
\zeta\sass\Gamma\yields \mbox{\fbox{$P$}}\ass\phi
\justifies
\zeta\sass\sep, \Gamma\yields\mbox{\fbox{$\rightinj P$}}\ass\phi
\using \rightinj R
\endprooftree\tb
\prooftree
\zeta\sass\Gamma\yields N\ass\phi
\justifies
\zeta\sass\sep, \Gamma\yields \mbox{\fbox{$\rightinj N$}}\ass\phi
\using \leftinj R
\endprooftree
\\\\
42. &
\prooftree
\zeta\sass\Delta\yields \mbox{\fbox{$P$}}\ass\psi
\justifies
\zeta\sass\Delta\smwrap{k}\Lambda\yields\mbox{\fbox{$\sbridge{k} P$}}\ass\psi
\using \sbridge{k} R
\endprooftree
\tb
\prooftree
\zeta\sass\Delta\yields N\ass\psi
\justifies
\zeta\sass\Delta\smwrap{k}\Lambda\yields\mbox{\fbox{$\sbridge{k} N$}}\ass\psi
\using \sbridge{k} R
\endprooftree
\end{array}
$
\end{center}
\caption{Right synchronous deterministic synthetic multiplicative rules}
\label{rsslununders}
\end{figure}

\subsection{Binary synthetic multiplicatives}

\begin{figure}[ht]
\begin{center}
$
\begin{array}{lcc}
44. &
\prooftree
\zeta\sass\Gamma_1\yields \mbox{\fbox{$P_1$}}\ass\phi\tb\zeta'\sass\Gamma_2\yields \mbox{\fbox{$P_2$}}\ass\psi
\justifies
\zeta\mun\zeta'\sass\Gamma_1, \Gamma_2\yields \mbox{\fbox{$P_1\nprod P_2$}}\ass(\phi, \psi)
\using \nprod R_1
\endprooftree
\tb
\prooftree
\zeta\sass\Gamma_1\yields \mbox{\fbox{$P$}}\ass\phi\tb\zeta'\sass\Gamma_2\yields N\ass\psi
\justifies
\zeta\mun\zeta'\sass\Gamma_1, \Gamma_2\yields \mbox{\fbox{$P\nprod N$}}\ass(\phi, \psi)
\using \nprod R_1
\endprooftree
\\\\
&
\prooftree
\zeta\sass\Gamma_1\yields \mbox{\fbox{$N$}}\ass\phi\tb\zeta'\sass\Gamma_2\yields P\ass\psi
\justifies
\zeta\mun\zeta'\sass\Gamma_1, \Gamma_2\yields \mbox{\fbox{$N\nprod P$}}\ass(\phi, \psi)
\using \nprod R_1
\endprooftree
\tb
\prooftree
\zeta\sass\Gamma_1\yields N_1\ass\phi\tb\zeta'\sass\Gamma_2\yields N_2\ass\psi
\justifies
\zeta\mun\zeta'\sass\Gamma_1, \Gamma_2\yields \mbox{\fbox{$N_1\nprod N_2$}}\ass(\phi, \psi)
\using \nprod R_1
\endprooftree
\\\\
&
\prooftree
\zeta\sass\Gamma_1\yields \mbox{\fbox{$P_2$}}\ass\psi\tb\zeta'\sass\Gamma_2\yields \mbox{\fbox{$P_1$}}\ass\phi
\justifies
\zeta\mun\zeta'\sass\Gamma_1, \Gamma_2\yields \mbox{\fbox{$P_1\nprod P_2$}}\ass(\phi, \psi)
\using \nprod R_2
\endprooftree
\tb
\prooftree
\zeta\sass\Gamma_1\yields N\ass\psi\tb\zeta'\sass\Gamma_2\yields \mbox{\fbox{$P$}}\ass\phi
\justifies
\zeta\mun\zeta'\sass\Gamma_1, \Gamma_2\yields \mbox{\fbox{$P\nprod N$}}\ass(\phi, \psi)
\using \nprod R_2
\endprooftree
\\\\
&
\prooftree
\zeta\sass\Gamma_1\yields \mbox{\fbox{$P$}}\ass\psi\tb\zeta'\sass\Gamma_2\yields N\ass\phi
\justifies
\zeta\mun\zeta'\sass\Gamma_1, \Gamma_2\yields \mbox{\fbox{$N\nprod P$}}\ass(\phi, \psi)
\using \nprod R_2
\endprooftree
\tb
\prooftree
\zeta\sass\Gamma_1\yields N_2\ass\psi\tb\zeta'\sass\Gamma_2\yields N_1\ass\phi
\justifies
\zeta\mun\zeta'\sass\Gamma_1, \Gamma_2\yields \mbox{\fbox{$N_1\nprod N_2$}}\ass(\phi, \psi)
\using \nprod R_2
\endprooftree
\end{array}$
\end{center}
\caption{Right synchronous continuous non-deterministic derived multiplicative rules}
\label{raslunbindersI}
\end{figure}

\begin{figure}[ht]
\begin{center}
$
\begin{array}{lcc}
47. &
\prooftree
\zeta\sass\Gamma_1\yields \mbox{\fbox{$P_1$}}\ass\phi\tb\zeta'\sass\Gamma_2\yields \mbox{\fbox{$P_2$}}\ass\psi
\justifies
\zeta\mun\zeta'\sass\Gamma_1\smwrap{k}\Gamma_2\yields \mbox{\fbox{$P_1\ndprod P_2$}}\ass(\phi, \psi)
\using \ndprod R
\endprooftree
\tb
\prooftree
\zeta\sass\Gamma_1\yields \mbox{\fbox{$P$}}\ass\phi\tb\zeta'\sass\Gamma_2\yields B\ass\psi
\justifies
\zeta\mun\zeta'\sass\Gamma_1\smwrap{k}\Gamma_2\yields \mbox{\fbox{$P\ndprod N$}}\ass(\phi, \psi)
\using \ndprod R
\endprooftree
\\\\
&\prooftree
\zeta\sass\Gamma_1\yields N\ass\phi\tb\zeta'\sass\Gamma_2\yields\mbox{\fbox{$P$}}\ass\psi
\justifies
\zeta\mun\zeta'\sass\Gamma_1\smwrap{k}\Gamma_2\yields \mbox{\fbox{$N\ndprod P$}}\ass(\phi, \psi)
\using \ndprod R
\endprooftree
\tb
\prooftree
\zeta\sass\Gamma_1\yields N_1\ass\phi\tb\zeta'\sass\Gamma_2\yields N_2\ass\psi
\justifies
\zeta\mun\zeta'\sass\Gamma_1\smwrap{k}\Gamma_2\yields \mbox{\fbox{$N_1\ndprod N_2$}}\ass(\phi, \psi)
\using \ndprod R
\endprooftree
\end{array}$
\end{center}
\caption{Right synchronous discontinuous non-deterministic synthetic multiplicative rules}
\label{raslunbindersII}
\end{figure}

\begin{figure}[ht]
\begin{center}
$
\begin{array}{lc}
48. &
\prooftree
\zeta\sass\Gamma\langle \vect{B_0}\ass y_0; \ldots; \vect{\fbox{$Q$}}; \ldots; \vect{B_n}\ass y_n\rangle\yields D\ass\omega 
\justifies
\zeta\sass\Gamma\langle \vect{B_0|A}\ass z_0; \ldots; \vect{\fbox{$Q|A$}}; \ldots; \vect{B_n|A}\ass z_n\rangle\yields D|A\ass\lambda x\omega\subst{(z_0\ x)/y_0, \ldots, (z_n\ x)/y_n}
\using |R
\endprooftree\\\\
 &
\prooftree
\zeta\sass\Gamma\langle \vect{B_0}\ass y_0; \ldots; \vect{M}; \ldots; \vect{B_n}\ass y_n\rangle\yields D\ass\omega 
\justifies
\zeta\sass\Gamma\langle \vect{B_0|A}\ass z_0; \ldots; \vect{\fbox{$M|A$}}; \ldots; \vect{B_n|A}\ass z_n\rangle\yields D|A\ass\lambda x\omega\subst{(z_0\ x)/y_0, \ldots, (z_n\ x)/y_n}
\using |R
\endprooftree
\end{array}
$
\end{center}
\caption{Right synchronous rules for limited contraction}
\label{sllcas}
\end{figure}

\begin{figure}[ht]
\begin{center}
$
\begin{array}{lcc}
49. & 
\prooftree
\justifies
\emptyset; \Lambda\yields \mbox{\fbox{$W(0)$}}\ass\zero
\using WR
\endprooftree
&
\prooftree
\zeta_1; \Gamma_1\yields \mbox{\fbox{$W(v)$}}\ass\phi\tb
\zeta_2; \Gamma_2\yields \mbox{\fbox{$W(w)$}}\ass\phi
\justifies
\zeta_1\mun\zeta_2; \Gamma_1, \Gamma_2\yields\mbox{\fbox{$W(v+w)$}}\ass\nil
\using WR
\endprooftree
\end{array}
$
\end{center}
\caption{Right synchronous rules for limited expansion}
\label{rsuncinterc2}
\end{figure}

\clearpage

\begin{figure}[ht]
\begin{center}
$
\begin{array}{lcc}
50. & \prooftree
\Xi\yields \fbox{$P_1$}\ass\psi\tb\not\vdash\Xi\yields \fbox{$P_2$}\ass\_
\justifies
\Xi\yields \fbox{$P_1\diff P_2$}\ass\psi
\using \diff R
\endprooftree
&
\prooftree
\Xi\yields \fbox{$P$}\ass\psi\tb\not\vdash\Xi\yields N\ass\_
\justifies
\Xi\yields \fbox{$P\diff N$}\ass\psi
\using \diff R
\endprooftree\\\\
&
\prooftree
\Xi\yields N\ass\psi\tb\not\vdash\Xi\yields \fbox{$P$}\ass\_
\justifies
\Xi\yields \fbox{$N\diff P$}\ass\psi
\using \diff R
\endprooftree
&
\prooftree
\Xi\yields N_1\ass\psi\tb\not\vdash\Xi\yields N_2\ass\_
\justifies
\Xi\yields \fbox{$N_1\diff N_2$}\ass\psi
\using \diff R
\endprooftree
\end{array}
$
\end{center}
\caption{Right synchronous rules for difference}
\label{sldiffs}
\end{figure}

\section{Completeness of focalisation for \DA}

\label{compl}

We shall be dealing with three systems: the displacement calculus \DA{} with sequents notated $\Delta\yields A$, the \techterm{weakly focalised\/} displacement
calculus with additives \DAf{} with sequents notated $\Delta\lyieldsw A$, and the \techterm{strongly focalised\/} displacement calculus with additives \DAF{} with sequents notated
$\Delta\lyields A$. Sequents of both \DAf{} and \DAF{} may contain at most one focalised formula,
possibly $A$.
When a \DAf{} sequent is notated $\Delta\lyieldsw A\pf$, it means that the sequent possibly contains a (unique)  focalised formula. Otherwise, $\Delta\lyieldsw A$ means
that the sequent does not contain a focus. 

In this section we prove the strong focalisation property for the displacement
calculus with additives \DA{}.
The focalisation property for Linear Logic was discovered 
by Andreoli (1992\cite{andreoli:92}). 
In this paper we follow the proof idea from \cite{Laurent04bnote}, which we
adapt to the intuitionistic non-commutative case \DA{} with twin multiplicative modes of combination, the continuous (concatenation) and the discontinuous
(intercalation) products. The proof relies heavily on the Cut-elimination property for
weakly focalised \DA{} which is proved
in the appendix. In our presentation of focalisation
we have avoided the \techterm{react} rules of \cite{andreoli:92} and \cite{Chaudhuri:2006:FIM:1292706}, and use instead a simpler, box, notation 
suitable for non-commutativity.

\DAF{} is a subsystem of \DAf{}. \DAf{}
has the focusing rules \foc{} and Cut rules $\pcut_1$, $\pcut_2$, $\ncut_1$ and $\ncut_2$\footnote{If it is convenient,  we may drop 
the subscripts.} shown in (\ref{complfocD:defs}), and the synchronous and asynchronous rules displayed before, which are
read as allowing in synchronous rules the occurrence of asynchronous formulas, and in asynchronous rules as allowing arbitrary sequents with possibly
one focalised formula. 
\DAF{} has the focusing rules but not the Cut rules, and the synchronous and asynchronous rules displayed before, which are such that focalised
sequents cannot contain any complex asynchronous formulas, whereas sequents with at least one complex asynchronous formula cannot contain a
focalised formula. Hence, strongly focalised proof search operates in alternating asynchronous and synchronous phases. 
The weakly focalised calculus \DAf{} is an intermediate logic
which we use to prove the completeness of \DAF{} for \DA{}.
\disp{\mini$
\begin{array}[t]{l}
\prooftree
\Delta\langle \vect{\fbox{$Q$}}\rangle\lyieldsw A
\justifies
\Delta\langle \vect{Q}\rangle\lyieldsw A
\using \foc
\endprooftree
\tb
\prooftree
\Delta\lyieldsw \fbox{$P$}
\justifies
\Delta\lyieldsw P
\using \foc
\endprooftree\\\\
\prooftree
\Gamma\lyieldsw \fbox{$P$}
\tb 
\Delta\langle\vect P\rangle\lyieldsw C\pf 
\justifies
\Delta\langle\Gamma\rangle\lyieldsw C\pf
\using\pcut_1
\endprooftree
\tb
\prooftree
\Gamma\lyieldsw N\pf 
\tb 
\Delta\langle\vect{\fbox{$N$}}\rangle\lyieldsw C
\justifies
\Delta\langle\Gamma\rangle\lyieldsw C\pf
\using\pcut_2
\endprooftree\\\\
\prooftree
\Gamma\lyieldsw P\pf
\tb 
\Delta\langle\vect P\rangle\lyieldsw C
\justifies
\Delta\langle\Gamma\rangle\lyieldsw C\pf
\using\ncut_1
\endprooftree
\tb
\prooftree
\Gamma\lyieldsw N\
\tb \Delta\langle\vect{N}\rangle\lyieldsw C\pf
\justifies
\Delta\langle\Gamma\rangle\lyieldsw C\pf
\using \ncut_2
\endprooftree
\end{array}$}
\label{complfocD:defs}

As before, $Q$ and $P$ denote synchronous formulas in input and output position respectively, whereas $M$ and $N$ 
denote asynchronous formulas in input and output position respectively; we will abbreviate thus: \emph{left sync., right synch., left async., and right async.}.
Occurrences of $P,Q,M$ and $N$ are supposed not to be focalised, which means that their
focalised occurrence \emph{must\/} be signalled with a box. By contrast, occurrences of $A,B,C$ may be focalised (if the sequent they occur in is judged\pf{}) or not.
The table below summarizes the notational convention on formulas $P,Q,M$ and $N$:
\begin{center}$
\begin{tabular}{| l | c |  c|}
\hline
& \mbox{input} & \mbox{output} \\ \hline
mbox{sync.} & $\mathbf{Q}$ & $\mathbf{P}$  \\ \hline
\mbox{async.} & $\mathbf{M}$ & $\mathbf{N}$  \\
hline
\end{tabular}$\end{center}
2

\subsection{Embedding of \DA{} into \DA$_{\bf foc}$}

\label{complfocD:emb}

The identity axiom we consider for \DA{} and for both \DAf{} and \DAF{} is restricted to atomic types;
recalling that atomic types are classified into positive bias {\it At}$^+$ and negative bias {\it At}$^-$:
\disp{\begin{tabular}[t]{l}
If $P\in\mbox{\it At}^+$, P\lyieldsw \fbox{$P$} and P\lyields \fbox{$P$}\\
If $Q\in\mbox{\it At}^-$, \fbox{$Q$}\lyieldsw Q and \fbox{$Q$}\lyields Q\\
\end{tabular}}
 In fact, the Identity rule holds of any type A. 
It has the following formulation in the sequent calculi considered here:

\disp{\mini
$
\left\{
\begin{array}{lll}
\vect{A}\yields A&&\mbox{ in }\DA{}\\
\vect{P}\lyieldsw \fbox{$P$}&\tb \vect{\fbox{$N$}}\lyieldsw N&\mbox{ in }\DAf{}\\
\vect{P}\lyields P&\tb \vect{N}\lyields  N&\mbox{ in }\DAF{}\\
\end{array}\right.
$
}

The Identity axiom for arbitrary types is also known as \techterm{Eta-expansion}. Eta-expansion is easy to prove in both \DA{} and \DAf{}, 
but the same is not the case for \DAF{}. This is the reason to consider what we have called weak focalisation, which helps us to 
prove smoothly this crucial property for the proof of strong focalisation.

\begin{theorem}
\emph{(Embedding of \DA{} into \DAf{})}
For any configuration $\Delta$ and type $A$, we have that if $\Delta\yields A$ then $\Delta\lyieldsw A$.
\label{complfocD:embthm1}
\end{theorem}

\prf{
We proceed by induction on the length of the derivation of \DA{} proofs. In the following lines, we apply the induction hypothesis (i.h.) for each premise of \DA{} rules (with 
the exception of the Identity rule and the right rules of units):\\\\
- Identity axiom: 
\disp{\mini$
\prooftree
\vect{P}\lyieldsw \fbox{P}
\justifies
\vect{P}\lyieldsw P
\using \foc
\endprooftree
\tb
\prooftree
\vect{\fbox{N}}\lyieldsw N
\justifies
\vect{N}\lyieldsw N
\using \foc
\endprooftree
$}
- Cut rule: just apply \ncut.

\noindent
- Units
\disp{\mini$
\prooftree
\justifies
\Lambda\yields I
\using I R
\endprooftree
\tb\leadsto\tb
\prooftree
\prooftree
\justifies
\Lambda\lyieldsw \fbox{$I$}
\using IR
\endprooftree
\justifies
\Lambda\lyieldsw I
\using \foc
\endprooftree
$}
\disp{\mini$
\prooftree
\justifies
\sep\yields J
\using J R
\endprooftree
\tb\leadsto\tb
\prooftree
\prooftree
\justifies
\sep\lyieldsw \fbox{$J$}
\using JR
\endprooftree
\justifies
\sep\lyieldsw J
\using \foc
\endprooftree
$}
Left unit rules apply as in the case of \DA{}.\\\\

\noindent
- Left discontinuous product: directly translates.
%

\noindent
- Right discontinuous product. There are cases $P_1\swprod{k}P_2$, $N_1\swprod{k}N_2$, $N\swprod{k}P$ and $P\swprod{k}N$.
 We show one representative example:\\

\noindent{\mini$
\prooftree
\Delta\yields P
\tb 
\Gamma\yields N
\justifies
\Delta\smwrap{k}\Gamma\yields P\swprod{k}N
\using \swprod{k}R
\endprooftree
\tb\leadsto\tb$
$
\prooftree
\prooftree
\Gamma\lyieldsw N
\tb
\prooftree
\Delta\lyieldsw P
\tb
\prooftree
\vect{P}\lyieldsw \fbox{$P$}
\tb
\prooftree
\fbox{$\vect{N}$}\lyieldsw N
\justifies
\vect{N}\lyieldsw N
\using foc
\endprooftree
\justifies 
\vect{P}\smwrap{k}\vect{N}\yields \fbox{$P\swprod{k}N$}
\using\swprod{k}R
\endprooftree
\justifies
\Delta\smwrap{k}\vect{N}\lyieldsw \fbox{$P\swprod{k}N$}
\using \ncut
\endprooftree
\justifies
\Delta\smwrap{k}\Gamma\lyieldsw \fbox{$P\swprod{k}N$}
\using\ncut
\endprooftree
\justifies
\Delta\smwrap{k}\Gamma\lyieldsw P\swprod{k}N
\using foc
\endprooftree
$}\\

\noindent
- Left discontinuous \scircum{k} rule (the left rule for \sinfix{k} is entirely similar). Like in the case for the right discontinuous
product \swprod{k} rule, we only show one representative example:\\

\noindent{\mini$
\prooftree
\Gamma\yields P
\tb
\Delta\langle\vect{N}\rangle\yields A
\justifies 
\Delta\langle\vect{N\scircum{k}P}\smwrap{k}\Gamma\rangle\yields A
\using\scircum{k}L
\endprooftree
\tb\leadsto\tb$
$
\prooftree
\prooftree
\Gamma\lyieldsw P
\prooftree
\endprooftree
\tb
\prooftree
\prooftree
\vect{P}\lyieldsw\fbox{P}
\tb
\vect{\fbox{$N$}}\lyieldsw N
\justifies
\vect{\fbox{$N\scircum{k}P$}}\smwrap{k}\vect{P}\lyieldsw N
\using\scircum{k}L
\endprooftree
\tb
\Delta\langle\vect{N}\rangle\lyieldsw A
\justifies
\Delta\langle\vect{\fbox{$N\scircum{k}P$}}\smwrap{k}\vect{P}\rangle\lyieldsw A
\using \ncut
\endprooftree
\justifies
\Delta\langle\vect{\fbox{$N\scircum{k}P$}}\smwrap{k}\Gamma\rangle\lyieldsw A
\using \ncut
\endprooftree
\justifies 
\Delta\langle\vect{N\scircum{k}P}\smwrap{k}\Gamma\rangle\lyieldsw A
\using\foc
\endprooftree
$}\\

\noindent
- Right discontinuous \scircum{k} rule (the right discontinuous rule for \sinfix{k}{} is entirely similar):
\disp{\mini$
\prooftree
\Delta\smwrap{k}\vect{A}\yields B
\justifies
\Delta\yields B\scircum{k} A
\using \scircum{k} R
\endprooftree
\ \ \ \ \leadsto\ \ \ \ 
\prooftree
\Delta\smwrap{k}\vect{A}\lyieldsw B
\justifies
\Delta\lyieldsw B\scircum{k} A
\using \scircum{k} R
\endprooftree
$}
- Product and implicative continuous rules. These follow the same pattern as the discontinuous case.
We interchange the metalinguistic $k$-th intercalation $\smwrap{k}$  with the metalinguistic concatenation ',', and we
interchange \swprod{k}{}, \scircum{k}{} and \sinfix{k}{} with \product{}, $/$, and $\bsl$ respectively.
Concerning additives, conjunction Right translates directly and
we consider then conjunction Left (disjunction is symmetric):\\

\noindent{\mini
\prooftree
\Delta\langle \vect{P}\rangle\yields C
\justifies
\Delta\langle \vect{P\aconj M}\rangle\yields C
\using\aconj L
\endprooftree
$\ \ \ \ \ \ \ \leadsto\ \ \ \ \ \ \ $
\prooftree
\vect{P\aconj M}\lyieldsw P\tb
\Delta\langle \vect{P}\rangle\lyieldsw C
\justifies
\Delta\langle \vect{P\aconj M}\rangle\lyieldsw C
\using\ncut
\endprooftree}\\

\noindent
where by Eta expansion and application of the \foc{} rule, we have $\vect{P\aconj M}\lyieldsw P$.
}

\subsection{Embedding of \DAf{} into \DAF{}}

\label{complfocD:emb2}

\begin{theorem}
\emph{(Embedding of \DAf{} into \DAF{})}
For any configuration $\Delta$ and type $A$, we have that if $\Delta\lyieldsw A$ with one focalised formula
and no asynchronous formula occurrence, then $\Delta\lyields A$ with the same formula focalised.
If $\Delta\lyieldsw A$ with no focalised formula and with at least one asynchronous formula, then 
$\Delta\lyields A$.
\label{complfocD:embthm2}
\end{theorem}

\prf{
We proceed by induction on the size of \DAf{} sequents.\footnote{For a given type $A$,
the \emph{size\/} of $A$, $|A|$, is the number of connectives in $A$. By recursion on configurations we have:
$
\begin{array}{rcl}
|\Lambda| & ::= & 0 \\
|\vect{A},\Delta| & ::= & |A| + |\Delta|, \mbox{for\ } sA=0\\
|\sep,\Delta| & ::= & |\Delta|\\
|A\{\Delta_1:\cdots:\Delta_{sA}\}|& ::= &|A|+\sum\limits_{i=1}^{sA}|\Delta_i|
\end{array}$

\noindent
Moreover, we have:

\noindent
$
|\Delta\langle\vect{\fbox{$Q$}}\rangle\lyieldsw A|=|\Delta\langle\vect{Q}\rangle\lyieldsw A|\\
|\Delta\lyieldsw \fbox{$P$}|=|\Delta\lyieldsw P|
$
} 
We consider Cut-free \DAf{} proofs which match the sequents of this theorem. If the last rule is logical (i.e., it is not an instance of the \foc{} rule) the 
i.h.\ applies
directly  and we get \DAF{} proofs of the same end-sequent. Now, let us suppose that the last rule is not logical, i.e.~it is an instance of the \foc{} rule.
Let us suppose that the end sequent $\Delta\lyieldsw A$ is a synchronous sequent. Suppose for example that the focalised formula
is in the succedent:
\disp{\mini
$
\prooftree
\Delta\lyieldsw \fbox{$P$}
\justifies
\Delta\lyieldsw P
\using \foc
\endprooftree
$
}
The sequent $\Delta\lyieldsw \fbox{$P$}$ arises from a synchronous rule to which we can apply i.h..
Let us suppose now that the end-sequent contains at least one asynchronous formula. We see three cases which are illustrative:
\disp{\mini
\begin{tabular}[t]{ll}
a. & $\Delta\langle \vect{A\swprod{k}B}\rangle\lyieldsw \fbox{$P$}$\\
b. & $\Delta\langle \vect{\fbox{$Q$}}\rangle\lyieldsw B\scircum{k}A$\\
c. & $\Delta\langle \vect{\fbox{$Q$}}\rangle\lyieldsw A\aconj B$
\end{tabular}\label{complfocD:embthm2cases}}
We have by Eta expansion that $\vect{A\swprod{k}B}\lyieldsw \vect{\fbox{$A\swprod{k}B$}}$. We apply to this sequent the invertible
\swprod{k}{} left rule, whence $\vect{A}|_k\vect{B}\lyieldsw \vect{\fbox{$A\swprod{k}B$}}$. In case (\ref{complfocD:embthm2cases}a), 
we have the following proof in \DAf{}:
\disp{\mini
$
\prooftree
\prooftree
\vect{A}\smwrap{k}\vect{B}\lyieldsw \vect{\fbox{$A\swprod{k}B$}}
\tb
\Delta\langle \vect{A\swprod{k}B}\rangle\lyieldsw \fbox{$P$}
\justifies
\Delta\langle \vect{A}\smwrap{k}\vect{B}\rangle\lyieldsw \fbox{$P$}
\using\pcut_1
\endprooftree
\justifies
\Delta\langle \vect{A}\smwrap{k}\vect{B}\rangle\lyieldsw P
\using\foc
\endprooftree
$
}
To the above \DAf{} proof we apply Cut-elimination and we get the Cut-free \DAf{} end-sequent $\Delta\langle \vect{A}$ $|_k\vect{B}\rangle\lyieldsw P$.
We have
$|\Delta\langle \vect{A}|_k\vect{B}\rangle\lyieldsw P| <  | \Delta\langle \vect{A\swprod{k}B}\rangle\lyieldsw P|$.
We can apply then i.h.~and we derive the provable
\DAF{} sequent $\Delta\langle \vect{A}\smwrap{k}\vect{B}\rangle\lyields P$ to which we can apply the left \swprod{k}{} rule. We have obtained 
$\Delta\langle \vect{A\swprod{k}B}\rangle\lyields P$. In the same way, we have that 
$\vect{\fbox{$B\scircum{k} A$}}|_k\vect{A}\lyieldsw B$.
Thus, in case (\ref{complfocD:embthm2cases}b), we have the following proof in \DAf{}:
\disp{\mini
$
\prooftree
\prooftree
\Delta\langle \vect{\fbox{$Q$}}\rangle\lyieldsw B\scircum{k}A
\tb
\vect{\fbox{$B\scircum{k} A$}}\smwrap{k}\vect{A}\lyieldsw B
\justifies
\Delta\langle \vect{\fbox{$Q$}}\rangle\smwrap{k}\vect{A}\lyieldsw B
\using\pcut_2
\endprooftree
\justifies
\Delta\langle \vect{Q}\rangle\smwrap{k}\vect{A}\lyieldsw B
\using\foc
\endprooftree
$
}
As before, we apply Cut-elimination to the above proof. We get the Cut-free \DAf{} end-sequent $\Delta\langle \vect{Q}\rangle|_k\vect{A}$ \lyieldsw $B$. It has
size less than $|\Delta\langle \vect{Q}\rangle\lyieldsw B\scircum{k}A|$. We can apply i.h.\ and we get the \DAF{} provable sequent 
$\Delta\langle \vect{Q}\rangle|_k\vect{A}\lyields B$ to which we apply the \scircum{k}{} right rule. 
In case (\ref{complfocD:embthm2cases}c):

\disp{
\prooftree
\Delta\langle \vect{\fbox{Q}}\rangle \lyieldsw A\aconj B
\justifies
\Delta\langle \vect{Q}\rangle \lyieldsw A\aconj B
\using\foc
\endprooftree}

\noindent
by applying the \foc{} rule we get the provable \DAf{} sequents $\Delta\langle \vect{Q}\rangle \lyieldsw A$
and $\Delta\langle \vect{Q}\rangle \lyieldsw B$. These sequents have smaller size than $\Delta\langle \vect{Q}\rangle \lyieldsw A\aconj B$. The aforementioned sequents have a Cut-free proof in \DAf{}. We apply i.h.\ and we get $\Delta\langle \vect{Q}\rangle \lyields A$ and  $\Delta\langle \vect{Q}\rangle \lyields B$.
We apply the \aconj{} right rule in \DAF{}, and we get $\Delta\langle \vect{Q}\rangle \lyields A\aconj B$.
}

\section{Cut-elimination for displacement calculus with additives}

We prove this by induction on the complexity $(d,h)$ of top-most instances of $Cut$, where $d$ is the size\footnote{The size of $|A|$ is the number of connectives appearing
in $A$.} of the cut formula and $h$ is the length of the derivation the last rule of which is the Cut rule.
There are four cases to consider: Cut with axiom in the minor premise, Cut with axiom in the major premise, principal Cuts, and
permutation conversions. In each case, the complexity of the Cut is reduced. In order to save space, we will not be exhaustive showing all the cases
because many follow the same pattern. In particular, for any synchronous logical rule there are always four cases to consider corresponding to 
the polarity of the subformulas. Here, and in the following, we will show only one representative example. 
Concerning continuous and discontinuous formulas, we will show only the discontinuous cases
(discontinuous connectives are less known than the continuous ones of the plain Lambek Calculus). For the continuous instances, the reader has only to interchange
the meta-linguistic wrap $|_k$ with the meta-linguistic concatenation $','$, \swprod{k}{} with \product{}, \scircum{k} with $/$ and 
\sinfix{k}{} with \bsl{}. The units cases (principal case and permutation conversion cases) are completely trivial.\\\\
\prf{
- $Id$ cases: 
\disp{
\prooftree
\vect{P}\lyieldsw\fbox{$P$}\tb 
\Delta\langle\vect{P}\rangle\lyieldsw B\pf
\justifies
\Delta\langle\vect{P}\rangle\lyieldsw B\pf
\using\pcut_1
\endprooftree
\tb$\leadsto$\tb
$\Delta\langle\vect{P}\rangle\lyieldsw B$\pf
\tb \\\\\\
\prooftree
\Delta\lyieldsw N\pf
\tb
\vect{\fbox{$N$}}\lyieldsw N
\justifies
\Delta\langle\vect{N}\rangle\lyieldsw B\pf
\using\pcut_2
\endprooftree
\tb$\leadsto$\tb
$\Delta\langle\vect{N}\rangle\lyieldsw B\pf$
}
The attentive reader may have wondered whether the following $Id$ case could arise:\\
\disp{
\prooftree
\fbox{$\vect{Q}$}\yields Q
\tb
\Gamma\langle\vect{Q}\rangle\yields A
\justifies
\Gamma\langle\fbox{Q}\rangle\yields A
\using \ncut_i
\endprooftree
\label{patholCut1}
}
If $Q$ were a primitive type $q$, and $\Gamma$ were not the empty context, we would have then a Cut-free 
underivable sequent. For example, if the right premise of the Cut rule in~(\ref{patholCut1})
were the derivable sequent $q,q\bsl s\yields s$, we would have then as conclusion:
\disp{
$
\fbox{q},q\bsl s\yields s
$\label{patholCutEx}}
Since the primitive type $q$ in the antecedent is focalised, there is no possibility of applying the
\bsl{} left rule, which is a synchronous rule that needs that its active formula to be focalised. 
Principal cases:\\\\
\product{} \foc{} cases:
\disp{ 
\prooftree
\prooftree
\Delta\lyieldsw \fbox{$P$}
\justifies
\Delta\lyieldsw P
\using\foc
\endprooftree
\tb
\Gamma\langle\vect{P}\rangle\lyieldsw A\pf
\justifies
\Gamma\langle\Delta\rangle\lyieldsw A\pf
\using\ncut_1
\endprooftree
\tb$\leadsto$\tb
\prooftree
\Delta\lyieldsw \fbox{$P$}
\tb
\Gamma\langle\vect{P}\rangle\lyieldsw A\pf
\justifies
\Gamma\langle\Delta\rangle\lyieldsw A\pf
\using\pcut_1
\endprooftree
}
\disp{ 
\prooftree
\Delta\lyieldsw N
\tb
\prooftree
\Delta\langle\fbox{$N$}\rangle\lyieldsw A
\justifies
\Gamma\langle\vect{N}\rangle\lyieldsw A
\using\foc
\endprooftree
\justifies
\Gamma\langle\Delta\rangle\lyieldsw A
\using\ncut_2
\endprooftree
\tb$\leadsto$\tb
\prooftree
\Delta\lyieldsw N
\tb
\Gamma\langle\vect{\fbox{$N$}}\rangle\lyieldsw A
\justifies
\Gamma\langle\Delta\rangle\lyieldsw A
\using\pcut_2
\endprooftree
}
\product{} logical connectives:
\disp{
\prooftree
\prooftree
\Delta|_k\vect{P_1}\lyieldsw P_2\pf
\justifies
\Delta\lyieldsw P_2\scircum{k}P_1\pf
\using\scircum{k}R
\endprooftree
\tb
\prooftree
\Gamma_1\lyieldsw \fbox{$P_1$}
\tb
\Gamma_2\langle\vect{P_2}\rangle\lyieldsw A
\justifies
\Gamma_2\langle\vect{\fbox{$P_2\scircum{k}P_1$}}|_k\Gamma_1\rangle\lyieldsw A
\using\scircum{k}L
\endprooftree
\justifies
\Gamma_2\langle\Delta|_k\Gamma_1\rangle\lyieldsw A\pf
\using\pcut_2
\endprooftree
\tb$\leadsto$\tb\\\\\\
\prooftree
\Gamma_1\lyieldsw\fbox{$P_1$}
\tb
\prooftree
\Delta|_k\vect{P_1}\lyields P_2\pf
\tb
\Gamma_2\langle\vect{P_2}\rangle\lyieldsw A
\justifies
\Gamma_2\langle\Delta|_k\vect{P_1}\rangle\lyieldsw A\pf
\using\ncut_1
\endprooftree
\justifies
\Gamma_2\langle\Delta|_k\Gamma_1\rangle\lyieldsw A\pf
\using\pcut_1
\endprooftree
}
The case of \sinfix{k}{} is entirely similar to the \scircum{k}{} case
and the case of \sinfix{k}{} is entirely similar to the \scircum{k}{} case.
\disp{ 
\prooftree
\prooftree
\Delta_1\lyieldsw\fbox{$P$}
\tb
\Delta_2\lyieldsw N
\justifies
\Delta_1|_k\Delta_2\lyieldsw\fbox{$P\swprod{k}N$}
\using\swprod{k}R
\endprooftree
\tb
\prooftree
\Gamma\langle\vect{P}|_k\vect{N}\rangle\lyieldsw A\pf
\justifies
\Gamma\langle\vect{P\swprod{k}N}\rangle\lyieldsw A\pf
\using\swprod{k}L
\endprooftree
\justifies
\Gamma\langle\Delta_1|_k\Delta_2\rangle\lyieldsw A\pf
\using\pcut_1
\endprooftree
\tb$\leadsto\tb$\\\\\\
\prooftree
\Delta_2\lyieldsw N
\tb
\prooftree
\Delta_1\lyieldsws \fbox{$P$}
\tb
\Gamma\langle\vect{P}|_k\vect{N}\rangle\lyieldsw A\pf
\justifies
\Gamma\langle\Delta_1|_k\vect{N}\rangle\lyieldsw A\pf
\using\pcut_1
\endprooftree
\justifies
\Gamma\langle\Delta_1|_k\Delta_2\rangle\lyieldsw A\pf
\using\ncut_2
\endprooftree
}\ \\
\disp{ 
$
\prooftree
\prooftree
\Delta\lyieldsw Q\pf
\tb
\Delta\lyieldsw A\pf
\justifies
\Delta\lyieldsw Q\aconj A\pf
\using\aconj R
\endprooftree
\tb
\prooftree
\Gamma\langle\vect{\fbox{$Q$}}\rangle\lyieldsw B
\justifies
\Gamma\langle\vect{\fbox{$Q\aconj A$}}\lyieldsw B
\using\aconj L
\endprooftree
\justifies
\Gamma\langle\Delta\rangle\lyieldsw B\pf
\using\pcut_2
\endprooftree
\tb\leadsto\tb\\\\\\
\prooftree
\Delta\lyields Q\pf\tb
\Gamma\langle\vect{\fbox{$Q$}}\rangle\lyieldsw B
\justifies
\Gamma\langle\Delta\rangle\lyieldsw B\pf
\using\pcut_2
\endprooftree
$
}
\disp{ 
$
\prooftree
\prooftree
\Delta\lyieldsw M\pf
\tb
\Delta\lyieldsw A\pf
\justifies
\Delta\lyieldsw M\aconj A\pf
\using\aconj R
\endprooftree
\tb
\prooftree
\Gamma\langle\vect M\rangle\lyieldsw B
\justifies
\Gamma\langle\vect{\fbox{$M\aconj A$}}\rangle\lyieldsw B
\using\aconj L
\endprooftree
\justifies
\Gamma\langle\Delta\rangle\lyieldsw B\pf
\using\pcut_2
\endprooftree
\tb\leadsto\tb\\\\\\
\prooftree
\Delta\lyields M\pf\tb
\Gamma\langle\vect M\rangle\lyieldsw B
\justifies
\Gamma\langle\Delta\rangle\lyieldsw B\pf
\using\ncut_1
\endprooftree
$
}
%
- Left commutative \pcut{} conversions:
\disp{
$
\prooftree
\prooftree
\Delta\langle\vect{\fbox{$Q$}}\rangle\lyieldsw N
\justifies
\Delta\langle\vect{Q}\rangle\lyieldsw N
\using\foc
\endprooftree
\tb 
\Gamma\langle\vect{\fbox{$N$}}\rangle\lyieldsw C
\justifies
\Gamma\langle\Delta\langle\vect{Q}\rangle\rangle\lyieldsw C
\using\pcut_2
\endprooftree
\tb\leadsto\tb
\prooftree
\prooftree
\Delta\langle\vect{\fbox{$Q$}}\rangle\lyieldsw N
\tb
\Gamma\langle\vect{\fbox{$N$}}\rangle\lyieldsw C
\justifies
\Gamma\langle\Delta\langle\vect{\fbox{$Q$}}\rangle\rangle\lyieldsw C
\using\pcut_2
\endprooftree
\justifies
\Gamma\langle\Delta\langle\vect{Q}\rangle\rangle\lyieldsw C
\using\foc
\endprooftree$
}
\disp{ 
$
\prooftree
\prooftree
\Delta\langle \vect{A}|_k\vect{B}\rangle\lyieldsw\fbox{$P$}
\justifies
\Delta\langle \vect{A\swprod{k}B}\rangle\lyieldsw\fbox{$P$}
\using\swprod{k}L
\endprooftree
\tb 
\Gamma\langle\vect{P}\rangle\lyieldsw C\pf
\justifies
\Gamma\langle\Delta\langle \vect{A\swprod{k}B}\rangle\rangle\lyieldsw C\pf
\using\pcut_1
\endprooftree
\tb\leadsto\tb\\\\\\
\prooftree
\prooftree
\Delta\langle\vect{A}|_k\vect{B}\rangle\lyieldsw\fbox{$P$}
\tb
\Gamma\langle\vect{P}\rangle\lyieldsw C\pf
\justifies
\Gamma\langle\Delta\langle \vect{A}|_k\vect{B}\rangle\rangle\lyieldsw C\pf
\using\pcut_1
\endprooftree
\justifies
\Gamma\langle\Delta\langle \vect{A\swprod{k}B}\rangle\rangle\lyieldsw C\pf
\using\swprod{k}L
\endprooftree
$
}
\disp{ 
$
\prooftree
\prooftree
\Delta\langle \vect{A}|_k\vect{B}\rangle\lyieldsw N\pf
\justifies
\Delta\langle \vect{A\swprod{k}B}\rangle\lyieldsw N\pf
\using\swprod{k}L
\endprooftree
\tb 
\Gamma\langle\fbox{$\vect{N}$}\rangle\lyieldsw C
\justifies
\Gamma\langle\Delta\langle \vect{A\swprod{k}B}\rangle\rangle\lyieldsw C\pf
\using\pcut_2
\endprooftree
\tb\leadsto\tb\\\\\\
\prooftree
\prooftree
\Delta\langle\vect{A}|_k\vect{B}\rangle\lyieldsw N\pf
\tb
\Gamma\langle\fbox{$\vect{N}$}\rangle\lyieldsw C
\justifies
\Gamma\langle\Delta\langle \vect{A}|_k\vect{B}\rangle\rangle\lyieldsw C\pf
\using\pcut_2
\endprooftree
\justifies
\Gamma\langle\Delta\langle \vect{A\swprod{k}B}\rangle\rangle\lyieldsw C\pf
\using\swprod{k}L
\endprooftree
$
}
\disp{ 
$
\prooftree
\prooftree
\Gamma_1\lyieldsw \fbox{$P_1$}
\tb
\Gamma_2\langle\vect{\fbox{$N_1$}}\rangle\lyieldsw N
\justifies
\Gamma_2\langle\vect{\fbox{$N_1\scircum{k}P_1$}}|_k\Gamma_1\rangle\lyieldsw N
\using\scircum{k}L
\endprooftree
\tb
\Theta\langle\vect{\fbox{$N$}}\rangle\lyieldsw C
\justifies
\Theta\langle\Gamma_2\langle\vect{\fbox{$N_1\scircum{k}P_1$}}|_k\Gamma_1\rangle\rangle\lyieldsw C
\using\pcut_2
\endprooftree
\tb\leadsto\tb\\\\\\
\prooftree
\Gamma_1\lyieldsw\fbox{$P_1$}
\tb
\prooftree
\Gamma_1\langle\vect{\fbox{$N_1$}}\rangle\lyieldsw N
\tb
\Theta\langle\vect{\fbox{$N$}}\rangle\lyieldsw C
\justifies
\Theta\langle\Gamma_2\langle\vect{\fbox{$N_1$}}\rangle\rangle\lyieldsw C
\using\pcut_2
\endprooftree
\justifies
\Theta\langle\Gamma_2\langle\vect{\fbox{$N_1\scircum{k}P_1$}}|_k\Gamma_1\rangle\rangle\lyieldsw C
\using\scircum{k}L
\endprooftree
$

}
\disp{ 
$
\prooftree
\prooftree
\Gamma\langle \vect A\rangle\lyieldsw \fbox{$P$}
\tb
\Gamma\langle \vect B\rangle\lyieldsw \fbox{$P$}
\justifies
\Gamma\langle \vect{A\adisj B}\rangle\lyieldsw \fbox{$P$}
\using\adisj L
\endprooftree
\tb
\Delta\langle\vect P\rangle\lyieldsw C\pf
\justifies
\Delta\langle\Gamma\langle \vect{A\adisj B}\rangle\rangle\lyieldsw C\pf
\using\pcut_1
\endprooftree
\tb\leadsto\tb\\\\\\
\prooftree
\prooftree
\Gamma\langle\vect A\rangle\lyieldsw\fbox{$P$}
\tb
\Delta\langle\vect P\rangle\lyieldsw C\pf
\justifies
\Delta\langle\Gamma\langle\vect A\rangle\rangle\lyieldsw C\pf
\using\pcut_1
\endprooftree
\tb
\prooftree
\Gamma\langle\vect B\rangle\lyieldsw\fbox{$P$}
\tb
\Delta\langle\vect P\rangle\lyieldsw C\pf
\justifies
\Delta\langle\Gamma\langle\vect B\rangle\rangle\lyieldsw C\pf
\using\pcut_1
\endprooftree
\justifies
\Delta\langle\Gamma\langle \vect{A\adisj B}\rangle\rangle\lyieldsw C\pf
\using\adisj L
\endprooftree
$
}

%
\disp{ 
$
\prooftree
\prooftree
\Gamma\langle \vect A\rangle\lyieldsw N\pf
\tb
\Gamma\langle \vect B\rangle\lyieldsw N\pf
\justifies
\Gamma\langle \vect{A\adisj B}\rangle\lyieldsw N\pf
\using\adisj L
\endprooftree
\tb
\Delta\langle\fbox{$\vect N$}\rangle\lyieldsw C
\justifies
\Delta\langle\Gamma\langle \vect{A\adisj B}\rangle\rangle\lyieldsw C\pf
\using\pcut_2
\endprooftree
\tb\leadsto\tb\\\\\\
\prooftree
\prooftree
\Gamma\langle\vect A\rangle\lyieldsw N\pf
\tb
\Delta\langle\fbox{$\vect N$}\rangle\lyieldsw C
\justifies
\Delta\langle\Gamma\langle\vect A\rangle\rangle\lyieldsw C\pf
\using\pcut_2
\endprooftree
\tb
\prooftree
\Gamma\langle\vect B\rangle\lyieldsw N\pf
\tb
\Delta\langle\fbox{$\vect N$}\rangle\lyieldsw C
\justifies
\Delta\langle\Gamma\langle\vect B\rangle\rangle\lyieldsw C\pf
\using\pcut_2
\endprooftree
\justifies
\Delta\langle\Gamma\langle \vect{A\adisj B}\rangle\rangle\lyieldsw C\pf
\using\adisj L
\endprooftree
$
}

- Right commutative \pcut{} conversions (unordered multiple distinguished occurrences are separated by semicolons):
\disp{ 
$
\prooftree
\Delta\lyieldsw\fbox{$P$}
\tb
\prooftree
\Gamma\langle\vect{P};\vect{\fbox{$Q$}}\rangle\lyieldsw C
\justifies
\Gamma\langle\vect{P};\vect{Q}\rangle\lyieldsw C
\using\foc
\endprooftree
\justifies
\Gamma\langle\Delta;\vect{Q}\rangle\lyieldsw C
\using\pcut_1
\endprooftree
\tb\leadsto\tb
\prooftree
\prooftree
\Delta\lyieldsw\fbox{$P$}
\tb
\Gamma\langle\vect{P};\vect{\fbox{$Q$}}\rangle\lyieldsw C
\justifies
\Gamma\langle\Delta;\vect{\fbox{$Q$}}\rangle\lyieldsw C
\using\pcut_1
\endprooftree
\justifies
\Gamma\langle\Delta;\vect{Q}\rangle\lyieldsw C
\using\foc
\endprooftree
$
}
\disp{ 
$
\prooftree
\Delta\lyieldsw\fbox{$P_1$}
\tb
\prooftree
\Gamma\langle\vect{P_1}\rangle\lyieldsw\fbox{$P_2$}
\justifies
\Gamma\langle\vect{P_1}\rangle\lyieldsw P_2
\using\foc
\endprooftree
\justifies
\Gamma\langle \Delta\rangle\lyieldsw P_2
\using\pcut_1
\endprooftree
\tb\leadsto\tb
\prooftree
\prooftree
\Delta\lyieldsw\fbox{$P_1$}
\tb
\Gamma\langle\vect{P_1}\rangle\lyieldsw \fbox{$P_2$}
\justifies
\Gamma\langle\Delta\rangle\lyieldsw \fbox{$P_2$}
\using\pcut_1
\endprooftree
\justifies
\Gamma\langle\Delta\rangle\lyieldsw P_2
\using\foc
\endprooftree
$
}
\disp{ 
$
\prooftree
\Delta\lyieldsw\fbox{$P$}
\tb
\prooftree
\Gamma\langle\vect{P}\rangle|_k\vect{A}\lyieldsw B\pf
\justifies
\Gamma\langle\vect{P}\rangle\lyieldsw B\scircum{k}A\pf
\using\scircum{k}R
\endprooftree
\justifies
\Gamma\langle\Delta\rangle\lyieldsw B\scircum{k}A\pf
\using\pcut_1
\endprooftree
\tb\leadsto\tb
\prooftree
\prooftree
\Delta\lyieldsw\fbox{$P$}
\tb
\Gamma\langle\vect{P}\rangle|_k \vect{A}\lyieldsw B\pf
\justifies
\Gamma\langle\Delta\rangle|_k \vect{A}\lyieldsw B\pf
\using\pcut_1
\endprooftree
\justifies
\Gamma\langle\Delta\rangle\lyieldsw B\scircum{k}A\pf
\using\scircum{k}R
\endprooftree
$
}
\disp{ 
$
\prooftree
\Delta\lyieldsw N\pf
\tb
\prooftree
\Gamma\langle\fbox{$\vect{N}$}\rangle|_k\vect{A}\lyieldsw B
\justifies
\Gamma\langle\fbox{$\vect{N}$}\rangle\lyieldsw B\scircum{k}A
\using\scircum{k}R
\endprooftree
\justifies
\Gamma\langle\Delta\rangle\lyieldsw B\scircum{k}A\pf
\using\pcut_2
\endprooftree
\tb\leadsto\tb
\prooftree
\prooftree
\Delta\lyieldsw N\pf
\tb
\Gamma\langle\fbox{$\vect{N}$}\rangle|_k \vect{A}\lyieldsw B
\justifies
\Gamma\langle\Delta\rangle|_k \vect{A}\lyieldsw B\pf
\using\pcut_2
\endprooftree
\justifies
\Gamma\langle\Delta\rangle\lyieldsw B\scircum{k}A\pf
\using\scircum{k}R
\endprooftree
$
}
\disp{ 
$
\prooftree
\Delta\lyieldsw\fbox{$P$}
\tb
\prooftree
\Gamma\langle\vect{P};\vect{A}|_k\vect{B}\rangle\lyieldsw C\pf
\justifies
\Gamma\langle\vect{P};\vect{A\swprod{k}B}\rangle\lyieldsw C\pf
\using\swprod{k}L
\endprooftree
\justifies
\Gamma\langle\Delta;\vect{A\swprod{k}B}\rangle\lyieldsw C\pf
\using\pcut_1
\endprooftree
\tb\leadsto\tb
\prooftree
\prooftree
\Delta\lyieldsw\fbox{$P$}
\tb
\Gamma\langle\vect{P};\vect{A}|_k\vect{B}\rangle\lyieldsw C\pf
\justifies
\Gamma\langle\Delta;\vect{A}|_k\vect{B}\rangle\lyieldsw C\pf
\using\pcut_1
\endprooftree
\justifies
\Gamma\langle\Delta;\vect{A\swprod{k}B}\rangle\lyieldsw C\pf
\using\swprod{k}L
\endprooftree
$
}
\disp{
$
\prooftree
\Delta\lyieldsw N\pf
\tb
\prooftree
\Gamma\langle\vect{\fbox{$N$}};\vect{A}|_k\vect{B}\rangle\lyieldsw C
\justifies
\Gamma\langle\vect{\fbox{$N$}};\vect{A\swprod{k}B}\rangle\lyieldsw C
\using\swprod{k}L
\endprooftree
\justifies
\Gamma\langle\Delta;\vect{A\swprod{k}B}\rangle\lyieldsw C\pf
\using\pcut_2
\endprooftree
\tb\leadsto\tb\\\\\\
\prooftree
\prooftree
\Delta\lyieldsw N\pf\tb\Gamma\langle\vect{\fbox{$N$}};\vect{A}|_k\vect{B}\rangle\lyieldsw C
\justifies
\Gamma\langle\Delta;\vect{A}|_k\vect{B}\rangle\lyieldsw C\pf
\using\pcut_2
\endprooftree
\justifies
\Gamma\langle\Delta;\vect{A\swprod{k}B}\rangle\lyieldsw C\pf
\using\swprod{k}L
\endprooftree
$
}
\disp{
$
\prooftree
\Delta\lyieldsw\fbox{$P$}
\tb
\prooftree
\Gamma\lyieldsw\fbox{$P_1$}
\tb
\Theta\langle\vect{P_2};\vect{P}\rangle\lyieldsw C
\justifies
\Theta\langle\vect{\fbox{$P_2\scircum{k}P_1$}}|_k\Gamma;\vect{P}\rangle\lyieldsw C
\using\scircum{k}L
\endprooftree
\justifies
\Theta\langle\vect{\fbox{$P_2\scircum{k}P_1$}}|_k\Gamma;\Delta\rangle\lyieldsw C
\using\pcut_1
\endprooftree
\tb\leadsto\tb\\\\\\
\prooftree
\Gamma\lyieldsw\fbox{$P_1$}
\tb
\prooftree
\Delta\lyieldsw\fbox{$P$}
\tb
\Theta\langle\vect{P_2};\vect{P}\rangle\lyieldsw C
\justifies
\Theta\langle\vect{P_2};\Delta\rangle\lyieldsw C
\using\pcut_1
\endprooftree
\justifies
\Theta\langle\vect{\fbox{$P_2\scircum{k}P_1$}}|_k\Gamma;\Delta\rangle\lyieldsw C
\using\scircum{k}L
\endprooftree
$
}
\disp{ 
$
\prooftree
\Delta\lyieldsw\fbox{$P$}
\tb
\prooftree
\Gamma\langle\vect P\rangle\lyieldsw A\pf
\tb
\Gamma\langle\vect P\rangle\lyieldsw B\pf
\justifies
\Gamma\langle\vect P\rangle\lyieldsw A\aconj B\pf
\using\aconj R
\endprooftree
\justifies
\Gamma\langle\Delta\rangle\lyieldsw A\aconj B\pf
\using\pcut_1
\endprooftree
\tb\leadsto\tb\\\\\\
\prooftree
\prooftree
\Delta\lyieldsw \fbox{$P$}
\tb
\Gamma\langle\vect P\rangle\lyieldsw A\pf
\justifies
\Gamma\langle\Delta\rangle\lyieldsw A\pf
\using\pcut_1
\endprooftree
\tb
\prooftree
\Delta\lyieldsw \fbox{$P$}
\tb
\Gamma\langle\vect P\rangle\lyieldsw B\pf
\justifies
\Gamma\langle\Delta\rangle\lyieldsw B\pf
\using\pcut_1
\endprooftree
\justifies
\Gamma\langle\Delta\rangle\lyieldsw A\aconj B\pf
\using\aconj R
\endprooftree
$
}
\disp{ 
$
\prooftree
\Delta\lyieldsw N\pf
\tb
\prooftree
\Gamma\langle\fbox{$\vect N$}\rangle\lyieldsw A
\tb
\Gamma\langle\fbox{$\vect N$}\rangle\lyieldsw B
\justifies
\Gamma\langle\fbox{$\vect N$}\rangle\lyieldsw A\aconj B
\using\aconj R
\endprooftree
\justifies
\Gamma\langle\Delta\rangle\lyieldsw A\aconj B\pf
\using\pcut_2
\endprooftree
\tb\leadsto\tb\\\\\\
\prooftree
\prooftree
\Delta\lyieldsw N\pf
\tb
\Gamma\langle\fbox{$\vect N$}\rangle\lyieldsw A
\justifies
\Gamma\langle\Delta\rangle\lyieldsw A\pf
\using\pcut_2
\endprooftree
\tb
\prooftree
\Delta\lyieldsw N\pf
\tb
\Gamma\langle\fbox{$\vect N$}\rangle\lyieldsw B
\justifies
\Gamma\langle\Delta\rangle\lyieldsw B\pf
\using\pcut_2
\endprooftree
\justifies
\Gamma\langle\Delta\rangle\lyieldsw A\aconj B\pf
\using\aconj R
\endprooftree
$
}
- Left commutative \ncut{} conversions:\\
\disp{
$
\prooftree
\prooftree
\Delta\langle\vect{\fbox{$Q$}}\rangle\lyieldsw P
\justifies
\Delta\langle\vect{Q}\rangle\lyieldsw P
\using\foc
\endprooftree
\tb
\Gamma\langle\vect{P}\rangle\lyieldsw C
\justifies
\Gamma\langle\Delta\langle \vect{Q}\rangle\rangle\lyieldsw C
\using\ncut_1
\endprooftree
\tb\leadsto\tb
\prooftree
\prooftree
\Delta\vect{\fbox{$Q$}}\lyieldsw P
\tb
\Gamma\langle\vect{P}\rangle\lyieldsw C
\justifies
\Gamma\langle\Delta\langle \vect{\fbox{$Q$}}\rangle\rangle\lyieldsw C
\using\ncut_1
\endprooftree
\justifies
\Gamma\langle\Delta\langle \vect{Q}\rangle\rangle\lyieldsw C
\using\foc
\endprooftree
$
}
\disp{ 
$
\prooftree
\prooftree
\Delta\langle \vect{A}|_k\vect{B}\rangle\lyieldsw P\pf
\justifies
\Delta\langle \vect{A\swprod{k}B}\rangle\lyieldsw P\pf
\using\swprod{k}L
\endprooftree
\tb 
\Gamma\langle\vect{P}\rangle\lyieldsw C
\justifies
\Gamma\langle\Delta\langle \vect{A\swprod{k}B}\rangle\rangle\lyieldsw C\pf
\using\ncut_1
\endprooftree
\tb\leadsto\tb\\\\\\
\prooftree
\prooftree
\Delta\langle\vect{A}|_k\vect{B}\rangle\lyieldsw P\pf
\tb
\Gamma\langle\vect{P}\rangle\lyieldsw C
\justifies
\Gamma\langle\Delta\langle \vect{A}|_k\vect{B}\rangle\rangle\lyieldsw C\pf
\using\ncut_1
\endprooftree
\justifies
\Gamma\langle\Delta\langle \vect{A\swprod{k}B}\rangle\rangle\lyieldsw C\pf
\using\swprod{k}L
\endprooftree
$
}
\disp{ 
$
\prooftree
\prooftree
\Delta\langle \vect{A}|_k\vect{B}\rangle\lyieldsw N
\justifies
\Delta\langle \vect{A\swprod{k}B}\rangle\lyieldsw N
\using\swprod{k}L
\endprooftree
\tb 
\Gamma\langle\vect{N}\rangle\lyieldsw C\pf
\justifies
\Gamma\langle\Delta\langle \vect{A\swprod{k}B}\rangle\rangle\lyieldsw C\pf
\using\ncut_2
\endprooftree
\tb\leadsto\tb\\\\\\
\prooftree
\prooftree
\Delta\langle\vect{A}|_k\vect{B}\rangle\lyieldsw N
\tb
\Gamma\langle\vect{N}\rangle\lyieldsw C\pf
\justifies
\Gamma\langle\Delta\langle \vect{A}|_k\vect{B}\rangle\rangle\lyieldsw C\pf
\using\ncut_2
\endprooftree
\justifies
\Gamma\langle\Delta\langle \vect{A\swprod{k}B}\rangle\rangle\lyieldsw C\pf
\using\swprod{k}L
\endprooftree
$
}
\disp{ 
$
\prooftree
\prooftree
\Gamma_1\lyieldsw \fbox{$P_1$}
\tb
\Gamma_2\langle\vect{\fbox{$N_1$}}\rangle\lyieldsw P
\justifies
\Gamma_2\langle\vect{\fbox{$N_1\scircum{k}P_1$}}|_k\Gamma_1\rangle\lyieldsw P
\using\scircum{k}L
\endprooftree
\tb
\Theta\langle\vect{P}\rangle\lyieldsw C
\justifies
\Theta\langle\Gamma_2\langle\vect{\fbox{$N_1\scircum{k}P_1$}}|_k\Gamma_1\rangle\rangle\lyieldsw C
\using\ncut_1
\endprooftree
\tb\leadsto\tb\\\\\\
\prooftree
\Gamma_1\lyieldsw\fbox{$P_1$}
\tb
\prooftree
\Gamma_1\langle\vect{\fbox{$N_1$}}\rangle\lyieldsw P
\tb
\Theta\langle\vect{P}\rangle\lyieldsw C
\justifies
\Theta\langle\Gamma_2\langle\vect{\fbox{$N_1$}}\rangle\rangle\lyieldsw C
\using\ncut_1
\endprooftree
\justifies
\Theta\langle\Gamma_2\langle\vect{\fbox{$N_1\scircum{k}P_1$}}|_k\Gamma_1\rangle\rangle\lyieldsw C
\using\scircum{k}L
\endprooftree
$
}
%
\disp{ 
$
\prooftree
\prooftree
\Gamma\langle \vect A\rangle\lyieldsw P\pf
\tb
\Gamma\langle \vect B\rangle\lyieldsw P\pf
\justifies
\Gamma\langle \vect{A\adisj B}\rangle\lyieldsw P\pf
\using\adisj L
\endprooftree
\tb
\Delta\langle\vect P\rangle\lyieldsw C\pf
\justifies
\Delta\langle\Gamma\langle \vect{A\adisj B}\rangle\rangle\lyieldsw C\pf
\using\ncut_1
\endprooftree
\tb\leadsto\tb\\\\\\
\prooftree
\prooftree
\Gamma\langle\vect A\rangle\lyieldsw P\pf
\tb
\Delta\langle\vect P\rangle\lyieldsw C
\justifies
\Delta\langle\Gamma\langle\vect A\rangle\rangle\lyieldsw C\pf
\using\ncut_1
\endprooftree
\tb
\prooftree
\Gamma\langle\vect B\rangle\lyieldsw P\pf
\tb
\Delta\langle\vect P\rangle\lyieldsw C
\justifies
\Delta\langle\Gamma\langle\vect B\rangle\rangle\lyieldsw C\pf
\using\ncut_1
\endprooftree
\justifies
\Delta\langle\Gamma\langle \vect{A\adisj B}\rangle\rangle\lyieldsw C\pf
\using\adisj L
\endprooftree
$
}

%
\disp{ 
$
\prooftree
\prooftree
\Gamma\langle \vect A\rangle\lyieldsw N
\tb
\Gamma\langle \vect B\rangle\lyieldsw N
\justifies
\Gamma\langle \vect{A\adisj B}\rangle\lyieldsw N
\using\adisj L
\endprooftree
\tb
\Delta\langle \vect N\rangle\lyieldsw C\pf
\justifies
\Delta\langle\Gamma\langle \vect{A\adisj B}\rangle\rangle\lyieldsw C\pf
\using\ncut_2
\endprooftree
\tb\leadsto\tb\\\\\\
\prooftree
\prooftree
\Gamma\langle\vect A\rangle\lyieldsw N
\tb
\Delta\langle\vect N\rangle\lyieldsw C\pf
\justifies
\Delta\langle\Gamma\langle\vect A\rangle\rangle\lyieldsw C\pf
\using\ncut_2
\endprooftree
\tb
\prooftree
\Gamma\langle\vect B\rangle\lyieldsw N
\tb
\Delta\langle\vect N\rangle\lyieldsw C\pf
\justifies
\Delta\langle\Gamma\langle\vect B\rangle\rangle\lyieldsw C\pf
\using\ncut_2
\endprooftree
\justifies
\Delta\langle\Gamma\langle \vect{A\adisj B}\rangle\rangle\lyieldsw C\pf
\using\adisj L
\endprooftree
$
}

- Right commutative \ncut{} conversions:\\
\disp{ 
$
\prooftree
\Delta\lyieldsw N
\tb
\prooftree
\Gamma\langle\vect{N};\vect{\fbox{$Q$}}\rangle\lyieldsw C
\justifies
\Gamma\langle\vect{N};\vect{Q}\rangle\lyieldsw C
\using\foc
\endprooftree
\justifies
\Gamma\langle\Delta;\vect{Q}\rangle\lyieldsw C
\using\ncut_2
\endprooftree
\tb\leadsto\tb
\prooftree
\prooftree
\Delta\lyieldsw N
\tb
\Gamma\langle\vect{N};\vect{\fbox{$Q$}}\rangle\lyieldsw C
\justifies
\Gamma\langle\Delta;\vect{\fbox{$Q$}}\rangle\lyieldsw C
\using\ncut_2
\endprooftree
\justifies
\Gamma\langle\Delta;\vect{Q}\rangle\lyieldsw C
\using\foc
\endprooftree
$
}
\disp{ 
$
\prooftree
\Delta\lyieldsw N
\tb
\prooftree
\Gamma\langle\vect{N}\rangle\lyieldsw\fbox{$P$}
\justifies
\Gamma\langle\vect{N}\rangle\lyieldsw P
\using\foc
\endprooftree
\justifies
\Gamma\langle\Delta\rangle\lyieldsw P
\using\ncut_2
\endprooftree
\tb\leadsto\tb
\prooftree
\prooftree
\Delta\lyieldsw N
\tb
\Gamma\langle\vect{N}\rangle\lyieldsw \fbox{$P$}
\justifies
\Gamma\langle\Delta\rangle\lyieldsw \fbox{$P$}
\using\ncut_2
\endprooftree
\justifies
\Gamma\langle\Delta\rangle\lyieldsw P
\using\foc
\endprooftree
$
}
\disp{ 
$
\prooftree
\Delta\lyieldsw P\pf
\tb
\prooftree
\Gamma\langle\vect{P}\rangle|_k\vect{A}\lyieldsw B
\justifies
\Gamma\langle\vect{P}\rangle\lyieldsw B\scircum{k}A
\using\scircum{k}R
\endprooftree
\justifies
\Gamma\langle\Delta\rangle\lyieldsw B\scircum{k}A\pf
\using\ncut_1
\endprooftree
\tb\leadsto\tb
\prooftree
\prooftree
\Delta\lyieldsw P\pf
\tb
\Gamma\langle\vect{P}\rangle|_k \vect{A}\lyieldsw B
\justifies
\Gamma\langle\Delta\rangle|_k \vect{A}\lyieldsw B\pf
\using\ncut_1
\endprooftree
\justifies
\Gamma\langle\Delta\rangle\lyieldsw B\scircum{k}A\pf
\using\scircum{k}R
\endprooftree
$
}
\disp{ 
$
\prooftree
\Delta\lyieldsw N
\tb
\prooftree
\Gamma\langle\vect{P}\rangle|_k\vect{A}\lyieldsw B\pf
\justifies
\Gamma\langle\vect{P}\rangle\lyieldsw B\scircum{k}A\pf
\using\scircum{k}R
\endprooftree
\justifies
\Gamma\langle\Delta\rangle\lyieldsw B\scircum{k}A\pf
\using\ncut_2
\endprooftree
\tb\leadsto\tb
\prooftree
\prooftree
\Delta\lyieldsw N
\tb
\Gamma\langle\vect{P}\rangle|_k \vect{A}\lyieldsw B\pf
\justifies
\Gamma\langle\Delta\rangle|_k \vect{A}\lyieldsw B\pf
\using\ncut_2
\endprooftree
\justifies
\Gamma\langle\Delta\rangle\lyieldsw B\scircum{k}A\pf
\using\scircum{k}R
\endprooftree
$
}
\disp{ 
$
\prooftree
\Delta\lyieldsw P\pf
\tb
\prooftree
\Gamma\langle\vect{P};\vect{A}|_k\vect{B}\rangle\lyieldsw C
\justifies
\Gamma\langle\vect{P};\vect{A\swprod{k}B}\rangle\lyieldsw C
\using\swprod{k}L
\endprooftree
\justifies
\Gamma\langle\Delta;\vect{A\swprod{k}B}\rangle\lyieldsw C\pf
\using\ncut_1
\endprooftree
\tb\leadsto\tb
\prooftree
\prooftree
\Delta\lyieldsw P\pf
\tb
\Gamma\langle\vect{P};\vect{A}|_k\vect{B}\rangle\lyieldsw C
\justifies
\Gamma\langle\Delta;\vect{A}|_k\vect{B}\rangle\lyieldsw C\pf
\using\ncut_1
\endprooftree
\justifies
\Gamma\langle\Delta;\vect{A\swprod{k}B}\rangle\lyieldsw C\pf
\using\swprod{k}L
\endprooftree
$
}
\disp{ 
$
\prooftree
\Delta\lyieldsw N
\tb
\prooftree
\Gamma\langle\vect{N};\vect{A}|_k\vect{B}\rangle\lyieldsw C\pf
\justifies
\Gamma\langle\vect{N};\vect{A\swprod{k}B}\rangle\lyieldsw C\pf
\using\swprod{k}L
\endprooftree
\justifies
\Gamma\langle\Delta;\vect{A\swprod{k}B}\rangle\lyieldsw C\pf
\using\ncut_2
\endprooftree
\tb\leadsto\tb
\prooftree
\prooftree
\Delta\lyieldsw N\tb\Gamma\langle\vect{N};\vect{A}|_k\vect{B}\rangle\lyieldsw C\pf
\justifies
\Gamma\langle\Delta;\vect{A}|_k\vect{B}\rangle\lyieldsw C\pf
\using\ncut_2
\endprooftree
\justifies
\Gamma\langle\Delta;\vect{A\swprod{k}B}\rangle\lyieldsw C\pf
\using\swprod{k}L
\endprooftree
$
}
\disp{
$
\prooftree
\Delta\lyieldsw N
\tb
\prooftree
\Gamma\lyieldsw\fbox{$P_1$}
\tb
\Theta\langle\vect{P_2};\vect{N}\rangle\lyieldsw C
\justifies
\Theta\langle\vect{\fbox{$P_2\scircum{k}P_1$}}|_k\Gamma;\vect{N}\rangle\lyieldsw C
\using\scircum{k}L
\endprooftree
\justifies
\Theta\langle\vect{\fbox{$P_2\scircum{k}P_1$}}|_k\Gamma;\Delta\rangle\lyieldsw C
\using\ncut_2
\endprooftree
\tb\leadsto\tb\\\\\\
\prooftree
\Gamma\lyieldsw\fbox{$P_1$}
\tb
\prooftree
\Delta\lyieldsw N
\tb
\Theta\langle\vect{P_2};\vect{N}\rangle\lyieldsw C
\justifies
\Theta\langle\vect{P_2};\Delta\rangle\lyieldsw C
\using\ncut_2
\endprooftree
\justifies
\Theta\langle\vect{\fbox{$P_2\scircum{k}P_1$}}|_k\Gamma;\Delta\rangle\lyieldsw C
\using\scircum{k}L
\endprooftree
$}
\disp{ 
$
\prooftree
\Delta\lyieldsw P\pf
\tb
\prooftree
\Gamma\langle\vect P\rangle\lyieldsw A
\tb
\Gamma\langle\vect P\rangle\lyieldsw B
\justifies
\Gamma\langle\vect P\rangle\lyieldsw A\aconj B
\using\aconj R
\endprooftree
\justifies
\Gamma\langle\Delta\rangle\lyieldsw A\aconj B\pf
\using\ncut_1
\endprooftree
\tb\leadsto\tb\\\\\\
\prooftree
\prooftree
\Delta\lyieldsw P
\tb
\Gamma\langle\vect P\rangle\lyieldsw A
\justifies
\Gamma\langle\Delta\rangle\lyieldsw A\pf
\using\ncut_1
\endprooftree
\tb
\prooftree
\Delta\lyieldsw P\pf
\tb
\Gamma\langle\vect P\rangle\lyieldsw B
\justifies
\Gamma\langle\Delta\rangle\lyieldsw B\pf
\using\ncut_1
\endprooftree
\justifies
\Gamma\langle\Delta\rangle\lyieldsw A\aconj B\pf
\using\aconj R
\endprooftree
$
}
\disp{ 
$
\prooftree
\Delta\lyieldsw N
\tb
\prooftree
\Gamma\langle\vect N\rangle\lyieldsw A\pf
\tb
\Gamma\langle\vect N\rangle\lyieldsw B\pf
\justifies
\Gamma\langle\vect N\rangle\lyieldsw A\aconj B\pf
\using\aconj R
\endprooftree
\justifies
\Gamma\langle\Delta\rangle\lyieldsw A\aconj B\pf
\using\ncut_2
\endprooftree
\tb\leadsto\tb\\\\\\
\prooftree
\prooftree
\Delta\lyieldsw N
\tb
\Gamma\langle\vect N\rangle\lyieldsw A\pf
\justifies
\Gamma\langle\Delta\rangle\lyieldsw A\pf
\using\ncut_2
\endprooftree
\tb
\prooftree
\Delta\lyieldsw N
\tb
\Gamma\langle\vect N\rangle\lyieldsw B\pf
\justifies
\Gamma\langle\Delta\rangle\lyieldsw B\pf
\using\ncut_2
\endprooftree
\justifies
\Gamma\langle\Delta\rangle\lyieldsw A\aconj B\pf
\using\aconj R
\endprooftree
$
}\ \\
}

\section{Non-admissibility of Cut}

\commentout{
ZZZ}

\chapter{Count-invariance}

\label{countchap}

We define infinitary count invariance for categorial
logic extending count invariance for multiplicatives
(van Benthem 1991\cite{benthem:action})
and additives and bracket modalities (Valent\'{\i}n et al. 2013\cite{vsm:add})
to include subexponentials (Kuznetsov et al.~2017\cite{mol16count}). The count invariance provides a powerful tool
for pruning proof search in categorial parsing/theorem-proving.

\section{Introduction}

We make some introductory remarks on non-linearity
since the main feature here is infinitary count invariance
for subexponentials.

\subsection{Sharing}

In standard logic information does not have multiplicity. Thus where $\add$
is the notion of addition of information and $\le$ is the notion of inclusion
of information we have $x\add x\le x$ and $x\le x\add x$; together these
two properties amount to {idempotency}: $x\add x=x$.
These properties are expressed by the rules of inference of Contraction and
Expansion:
\disp{$
\prooftree
\Delta(A, A)\yields B
\justifies
\Delta(A)\yields B
\using \mathrm{Contraction}
\endprooftree
$
\vtab
$\prooftree
\Delta(A)\yields B
\justifies
\Delta(A, A)\yields B
\using \mathrm{Expansion}
\endprooftree
$}
Linguistic resources do not have these properties: 
grammaticality is not generally preserved under addition or deletion
of copies of words or expressions.
However, there are some constructions manifesting something
similiar.
Parasitic gaps involve a kind of Contraction. Parasitic gaps cannot occur anywhere, for example
\disp{\unacc the slave that$_i$ John sold $e_i$ to $e_i$}Rather, we assume here that as the term \scare{parasitic} suggests,
a parasitic gap must fall within an island. Extraction from weak islands can become fully acceptable when accompanied
by a cobound non-island extraction:
\disp{\begin{tabular}[t]{ll}
a. & the man that$_i$ [the friends of $e_i$] admire $e_i$ \\
b. & the paper that$_i$ John filed $e_i$ [without reading $e_i$] \\
c. & the paper that$_i$ [the editor of $e_i$] filed $e_i$ [without reading $e_i$]
\end{tabular}}
And iterated coordination allows a kind of Expansion: 
\disp{
John likes, Mary dislikes and Bill loves London.}
That is, in logical grammar a \techterm{controlled\/} use of idempotency, or sharing, 
is motivated. Girard (1987\cite{girard:87}) introduced exponentials for such control.
Morrill and Valent\'{\i}n (2015\cite{cctlg:nl}) uses versions of the exponentials,
dubbed \scare{subexponentials} by 
Kanovich et al.~(2018\cite{2018subexp})
to treat (parasitic) gaps and
iterated coordination in categorial grammar.

\subsection{Count invariance}

Van Benthem (1991\cite{benthem:action}) 
introduces count invariance for multiplicatives in (sub)linear
logic, which involves simply checking the equality of the number of positive
and negative occurrences of each atom in a sequent. Thus where $\#_P(\Sigma)$
is the count of the atom $P$ in the sequent $\Sigma$ we have:
\disp{$\vdash \Sigma \Longrightarrow \forall P, \#_P(\Sigma)=0$\label{multeq}}
I.e.~there must be an exact  balance between the number of positive
and the number of negative occurrences of each atom.
This provides a necessary,
although of course not sufficient, criterion for theoremhood, and can be checked
very quickly. It can thus be used as a filter in proof search: if backward chaining
proof search generates an initial endsequent goal, or a subgoal, 
which does not satisfy the count invariant,
then the (sub)goal can be safely made to fail immediately.
This notion of count for multiplicatives was included in the categorial parser/theorem-prover
CatLog (Morrill 2012\cite{morrill:catlogdem}).

In Valent\'{\i}n et al.\ (2013\cite{vsm:add}) the idea is extended to additives (and bracket modalities).
To treat additives, instead of a single count for each atom $P$ of a sequent $\Sigma$ we have
a minimum count $\#_{\mymin, P}(\Sigma)$ and a maximum count $\#_{\mymax, P}(\Sigma)$
and for a sequent to be a theorem it must satisfy two inequations:
\disp{$\vdash \Sigma \Longrightarrow \forall P, \#_{\mymin, P}(\Sigma)\le 0\le \#_{\mymax, P}(\Sigma)$
\label{addineq}}
I.e.~the count functions $\#_{\mymin, P}$ and $\#_{\mymax, P}$ 
define an interval which must include the point
of equilibrium $0$; in the case of the multiplicatives,
$\#_{\mymin, P} = \#_{\mymax, P} = \#_P$ and (\ref{addineq}) reduces to the special case
(\ref{multeq}).
This generalised notion of count for additives and bracket modalities is included in the categorial
parser/theorem-prover CatLog2.\footnote{{\tt http://www.cs.upc.edu/\~{}droman/index.php}}

The structure of the rest of the chapter is as follows.
In Section~\ref{infcntalg} we present the infinitary count algebra which we employ,
define the fragment of categorial logic for which we illustrate count invariance,
and define the (infinitary) count functions for this fragment.
In Section~\ref{proofsect} we 
state and prove our count invariance theorem. 

\section{Infinitary count algebra}

\label{infcntalg}

We consider terms built over constants
$0$, $1$, $\mybot$, $\mytop$ and $\topbot$ 
by binary operations
of plus ($+$), minus ($-$), minimum (\mymin)
and maximum (\mymax), and unary operations of positive extrapolation (\posex)
and negative extrapolation (\negex) as follows where $i$ and $j$ are integers
and $n$ is a positive integer:\footnote{Positive and negative extrapolation can be defined
by $\posex(x) = \frac{2x^2}{x-|x|}$ and $\negex(x) = \frac{-2x^2}{-x-|x|}$ where division by zero is infinite
and obeys the sign rule.}\\

$
\begin{array}{c|cccc}
+ & j & \mybot & \mytop & \topbot\\
\hline
i & i{+}j & \mybot & \mytop & \topbot\\
\mybot & \mybot & \mybot & \topbot & \topbot\\
\mytop & \mytop & \topbot & \mytop & \topbot\\
\topbot & \topbot & \topbot & \topbot & \topbot
\end{array}
$
\tab
$
\begin{array}{c|cccc}
- & j & \mybot & \mytop & \topbot\\
\hline
i & i{-}j & \mytop & \mybot & \topbot\\
\mybot & \mybot & \topbot & \mybot & \topbot\\
\mytop & \mytop & \mytop & \topbot & \topbot\\
\topbot & \topbot & \topbot & \topbot & \topbot
\end{array}
$

\ \\

$
\begin{array}{c|cccc}
\mymin & j & \mybot & \mytop & \topbot\\
\hline
i & \frac{|i+j|-|i-j|}{2} & \mybot & i & \mybot\\
\mybot & \mybot & \mybot & \mybot & \mybot\\
\mytop & j & \mybot & \mytop & \mybot\\
\topbot & \mybot & \mybot & \mybot & \mybot
\end{array}
$
\tab
$
\begin{array}{c|cccc}
\mymax & j & \mybot & \mytop & \topbot\\
\hline
i & \frac{|i+j|+|i-j|}{2} & i & \mytop & \mytop\\
\mybot & j & \mybot & \mytop & \mytop\\
\mytop & \mytop & \mytop & \mytop & \mytop\\
\topbot & \mytop & \mytop & \mytop & \mytop 
\end{array}
$

\ \\

$
\begin{array}{c|cccccc}
 & \posex & \negex \\
\hline
-n & -n & \mybot\\
0 & 0 & 0 \\
+n & \mytop & +n \\
\mybot & \mybot & \mybot \\
\mytop & \mytop & \mytop \\
\topbot & \topbot & \topbot 
\end{array}
$

\ \\

Where for primitive types \PrimTypes, $Q\in \PrimTypes{\cup}\{[]\}$,
$m\in\{\mymin, \mymax\}$ and $\overline{\mymin}=\mymax$ and
$\overline{\mymax}=\mymin$.

$$
\begin{array}{l}
\#_{m, Q}(\Gamma\yields A) =  \#^\out_{m, Q}(A)-\#^\inp_{\overline{m}, Q}(\Gamma)
\end{array}
$$

\noindent
where for types $\Tp$:

$$
\begin{array}{lcl}
\Tp  ::=  \PrimTypes\ |\ \\
\Tp\bsl\Tp\ |\ \Tp/\Tp\ |\ \Tp\product\Tp\ |\ \\
\Tp\aconj\Tp\ |\ \Tp\adisj\Tp\ |\ \\
\abrack\Tp\ |\ \mybrack\Tp\ |\ \\
\univexp\Tp\ |\ \exstexp\Tp
\end{array}
$$

\noindent
for $P\in\PrimTypes$, $p\in\{\inp, \out\}$, and $\overline{\inp}=\out$ and $\overline{\out}=\inp$\commentout{ZZZ we have P o Q:?}

\ \\

$
\begin{array}{rcl}
\#^p_{m, Q}(P) & = &
\begin{array}{ll}
1 & \mbox{if $Q=P$}\\
0 & \mbox{if $Q\neq P$}
\end{array}\\
\#^p_{m, Q}(A\bsl C) & = &
\#^p_{m, Q}(C)-\#^{\overline{p}}_{\overline{m}, Q}(A)\\
\#^p_{m, Q}(C/B) & = &
\#^p_{m, Q}(C)-\#^{\overline{p}}_{\overline{m}, Q}(B)\\
\#^p_{m, Q}(A\product B) & = &
\#^p_{m, Q}(A)+\#^p_{m, Q}(B)\\
\#^\out_{m, Q}(A\aconj B) & = & \overline{m}(\#^\out_{m, Q}(A), \#^\out_{m, Q}(B))\\
\#^\inp_{m, Q}(A\aconj B) & = & m(\#^\inp_{m, Q}(A), \#^\inp_{m, Q}(B))\\
\#^\out_{m, Q}(A\adisj B) & = & m(\#^\out_{m, Q}(A), \#^\out_{m, Q}(B))\\
\#^\inp_{m, Q}(A\adisj B) & = & \overline{m}(\#^\inp_{m, Q}(A), \#^\inp_{m, Q}(B))\\
\#^p_{m, P}(\abrack A) & = & \#^p_{m, P}(A)\\
\#^p_{m, []}(\abrack A) & = & \#^p_{m, []}(A) - 1\\
\#^p_{m, P}(\mybrack A) & = & \#^p_{m, P}(A)\\
\#^p_{m, []}(\mybrack A) & = & \#^p_{m, []}(A) + 1\\
\#^p_{\mymin, Q}(\univexp A) & = & \negex(\#^p_{\mymin, Q}(A))\\
\#^\out_{\mymax, Q}(\univexp A) & = & \posex(\#^\out_{\mymax, Q}(A))\\
\#^\inp_{\mymax, P}(\univexp A) & = & \posex(\#^\inp_{\mymax, P}(A))\\
\#^\inp_{\mymax, []}(\univexp A) & = & \mytop\\
\#^p_{\mymin, Q}(\exstexp A) & = & \negex(\#^p_{\mymin, Q}(A))\\
\#^p_{\mymax, Q}(\exstexp A) & = & \posex(\#^p_{\mymax, Q}(A))
\end{array}
$

\ \\

\noindent
To present sequents we define 
\techterm{configurations\/} $\Config$
and \techterm{tree terms} $\Tterm$ by mutual recursion
in terms of types $\Tp$ as follows, where $\Lambda$ is the metalinguistic empty string:

\disp{$\begin{array}[t]{rcl}
\Config & ::= & \Lambda\ |\ \Tterm, \Config\\
\Tterm & ::= & \Tp\ |\ [\Config]
\end{array}$}

\noindent
The rules for the fragment of categorial logic are as shown in Figure~\ref{rulefig}.
\begin{figure}
\begin{center}
 \prooftree
 \Gamma\yields A\tb \Delta(C)\yields D
 \justifies
 \Delta(\Gamma, A\bsl C)\yields D
 \using \bsl L
 \endprooftree
\tb
 \prooftree
 \Gamma\yields B\tb \Delta(C)\yields D
 \justifies
 \Delta(C/B, \Gamma)\yields D
 \using \bsl L
 \endprooftree
 \vtab
  \prooftree
 A, \Gamma\yields C
 \justifies
 \Gamma\yields A\bsl C
 \using \bsl R
 \endprooftree
\tb
 \prooftree
 \Gamma, B\yields C
 \justifies
 \Gamma\yields C/B
 \using / R
 \endprooftree
\vtab
\prooftree
 \Delta(A, B)\yields C
 \justifies
 \Delta(A\product B)\yields C
 \using \product L
 \endprooftree
\tb
 \prooftree
 \Gamma_1\yields A\tb \Gamma_2\yields B
 \justifies
 \Gamma_1, \Gamma_2\yields A\product B
 \using \product R
 \endprooftree
\vtab
 \prooftree
 \Delta(A)\yields D
 \justifies
 \Delta(A\aconj B)\yields D
 \using \aconj L_1
 \endprooftree
\tb
 \prooftree
 \Delta(B)\yields D
 \justifies
 \Delta(A\aconj B)\yields D
 \using \aconj L_2
 \endprooftree
\tb
 \prooftree
 \Gamma\yields A\tb\Gamma\yields B
 \justifies
 \Gamma\yields A\aconj B
 \using \aconj R
 \endprooftree
\vtab
\prooftree
 \Gamma\yields A
 \justifies
 \Gamma\yields A\adisj B
 \using \adisj R_1
 \endprooftree
\tb
 \prooftree
 \Gamma\yields B
 \justifies
 \Gamma\yields A\adisj B
 \using \adisj R_2
 \endprooftree
\tb
 \prooftree
 \Delta(A)\yields D\tb\Delta(B)\yields D
 \justifies
 \Delta(A\adisj B)\yields D
 \using \adisj L
 \endprooftree
\vtab
 \prooftree
 \Gamma(A)\yields B
 \justifies
 \Gamma([\abrack A])\yields B
 \using \abrack L
 \endprooftree
 \tb
\prooftree
{}[\Gamma]\yields A
\justifies
\Gamma\yields \abrack A
\using \abrack R
\endprooftree
\vtab
\prooftree
\Gamma([A])\yields B
\justifies
\Gamma(\mybrack A)\yields B
\using \mybrack L
\endprooftree
\tb
\prooftree
\Gamma\yields A
\justifies
{}[\Gamma]\yields \mybrack A
\using \mybrack R
\endprooftree
\vtab
\prooftree
\Delta(A)\yields D
\justifies
\Delta(\univexp A)\yields D
\using \univexp L
\endprooftree
\tb
\prooftree
A_1, \ldots, A_n \yields A
\justifies
A_1, \ldots, A_n \yields \univexp A
\using \univexp R\commentout{ZZZsoft?}
\endprooftree
\vtab
\prooftree
\Delta(\Gamma, \univexp A)\yields D
\justifies
\Delta(\univexp A, \Gamma)\yields D
\using \univexp P_1
\endprooftree
\tb
\prooftree
\Delta(\univexp A, \Gamma)\yields D
\justifies
\Delta(\Gamma, \univexp A)\yields D
\using \univexp P_2
\endprooftree
\tb
\prooftree
\Delta(\univexp A_0, \ldots, \univexp A_n, [\univexp A_0, \ldots, \univexp A_n, \Gamma])\yields D
\justifies
\Delta(\univexp A_0, \ldots, \univexp A_n, \Gamma)\yields D
\using \univexp C
\endprooftree
\vtab
\prooftree
\Delta(A)\yields D\tb \Delta(A, A)\yields D\tb \ldots
\justifies
\Delta(\exstexp A)\yields D
\using \exstexp L
\endprooftree
\tb
\prooftree
\Gamma_1\yields C\tb\Gamma_2\yields\exstexp C
\justifies
\Gamma_1, \Gamma_2\yields\exstexp C
\using \exstexp M
\endprooftree
\end{center}
\caption{Rules for the categorial logic fragment}
\label{rulefig}
\end{figure}
For tree terms and configurations, counts are as follows:

\ \\

$
\begin{array}{rcl}
\#^!_{m, Q}(\zeta_1, \zeta_2) & = & \#^!_{m, Q}(\zeta_1)+ \#^!_{m, Q}(\zeta_2)\\
\#^!_{\mymin, Q}(A) & = & \negex(\#^\inp_{\mymin, Q}(A))\\
\#^!_{\mymax, P}(A) & = & \posex(\#^\inp_{\mymax, P}(A))\\
\#^!_{\mymax, []}(A) & = & \mytop\\
\#^!_{m, Q}(\nil) & = & 0\\
\#^\inp_{m, Q}(\Gamma, \Delta) & = & \#^\inp_{m, Q}(\Gamma)+\#^\inp_{m, Q}(\Delta)\\
\#^\inp_{m, []}([\Gamma]) & = & \#^\inp_{m, []}(\Gamma)'\\
\#\inp_{m, P}([\Gamma]) & = & \#\inp_{m, P}(\Gamma)\\
\#^\inp_{m, Q}(A) & = & \#^\inp_{m, Q}(A) \mbox{ for $A\in\Tp$}\\
\#^\inp_{m, Q}(\Lambda) & = & 0
\end{array}
$

\section{Theorem and proof}

\label{proofsect}

\ \\

{\bf Definition}

\ \\

We say that $t\mle 0$ (and that $0\mge t$)  if and only if $t=\topbot$ or $t=\mybot$ or $t\le 0$,
and we say that $0\mle t$ (and that $t\mge 0$) if and only if $t=\topbot$ or $t=\mytop$ or $0\le t$.

\ \\

{\bf Theorem} 

\ \\

$\vdash \Sigma \Longrightarrow \forall Q\in \PrimTypes\cup\{[]\}, \#_{\mymin, Q}(\Sigma) \mle 0 \mle
 \#_{\mymax, Q}(\Sigma)$
 
 \ \\
 
 {\bf Proof} By induction on the length of derivations.
 
 \ \\
 
 \subsection{Multiplicatives}
 
\commentout{ZZZ \ih}

 \begin{itemize}
 
 \item
 
 $
 \prooftree
 \Gamma\yields A\tb \Delta(C)\yields D
 \justifies
 \Delta(\Gamma, A\bsl C)\yields D
 \using \bsl L
 \endprooftree
 $\\
 
 For every atom or bracket,\\
  
 $
 \begin{array}{l}
 \#_m(\Delta(\Gamma, A\bsl C)\yields D) =\\
 \#^\out_m(D)-\#^\inp_{\overline{m}}(\Delta)-\#^\inp_{\overline{m}}(\Gamma)-\#^\inp_{\overline{m}}(A\bsl C) = \\
\#^\out_m(D)-\#^\inp_{\overline{m}}(\Delta)-\#^\inp_{\overline{m}}(\Gamma)-\#^\inp_{\overline{m}}(C)+\#^\out_m(A) = \\
\#^\out_m(A)-\#^\inp_{\overline{m}}(\Gamma) + \#^\out_m(D)-\#^\inp_{\overline{m}}(\Delta)-\#^\inp_{\overline{m}}(C) =\\
\#_m(\Gamma\yields A) + \#_m(\Delta(C)\yields D)
\end{array}
 $\\
 
By induction hypothesis (\ih),
$\#_\mymin(\Gamma\yields A)\mle 0$ and $ \#_\mymin(\Delta(C)\yields D)\mle 0$.
Therefore $ \#_\mymin(\Delta(\Gamma, A\bsl C)\yields D) =
\#_\mymin(\Gamma\yields A) + \#_\mymin(\Delta(C)\yields D)\mle 0$.
Similarly, $0\mle \#_\mymax(\Delta(\Gamma, A\bsl C)\yields D) =
\#_\mymax(\Gamma\yields A) + \#_\mymax(\Delta(C)\yields D)$.
Therefore:\\ 
 
 $ \#_\mymin(\Delta(\Gamma, A\bsl C)\yields D)\mle0\mle \#_\mymax(\Delta(\Gamma, A\bsl C)\yields D)$ \\
 
 \item
 
 $
 \prooftree
 \Gamma\yields B\tb \Delta(C)\yields D
 \justifies
 \Delta(C/B, \Gamma)\yields D
 \using \bsl L
 \endprooftree
 $\\
 
 Like $\bsl L$.\\
 
 \item
 
 $
 \prooftree
 A, \Gamma\yields C
 \justifies
 \Gamma\yields A\bsl C
 \using \bsl R
 \endprooftree
 $\\
  
For every atom or bracket,\\

 $
 \begin{array}{l}
\#_m(\Gamma\yields A\bsl C) =\\
\#^\out_m(A\bsl C)-\#^\inp_{\overline{m}}(\Gamma) = \\
\#^\out_m(C)-\#^\inp_{\overline{m}}(A)-\#^\inp_{\overline{m}}(\Gamma) = \\
\#_m(A, \Gamma\yields C)
\end{array}
$\\

Therefore by \ih,\\
 
$\#_\mymin(\Gamma\yields A\bsl C)\mle0 \mle \#_\mymax(\Gamma\yields A\bsl C)$\\

\item
 
 $
 \prooftree
 \Gamma, B\yields C
 \justifies
 \Gamma\yields C/B
 \using / R
 \endprooftree
 $\\
 
 Like $\bsl R$.\\
 
 \item
 
 $
 \prooftree
 \Delta(A, B)\yields C
 \justifies
 \Delta(A\product B)\yields C
 \using \product L
 \endprooftree
 $\\
 
 For every atom or bracket,\\
 
 $
 \begin{array}{l}
 \#_m(\Delta(A\product B)\yields C) =\\
 \#^\out_m(C)-\#^\inp_{\overline{m}}(\Delta)-\#^\inp_{\overline{m}}(A\product B) =\\
 \#^\out_m(C)-\#^\inp_{\overline{m}}(\Delta)-\#^\inp_{\overline{m}}(A)-\#^\inp_{\overline{m}}(B) =\\
 \#_m(\Delta(A, B)\yields C)
 \end{array}
 $\\
 
Therefore by \ih,\\ 
 
 $\#_\mymin(\Delta(A\product B)\yields C)\mle0\mle  \#_\mymax(\Delta(A\product B)\yields C)$\\
 
 \item
 
 $
 \prooftree
 \Gamma_1\yields A\tb \Gamma_2\yields B
 \justifies
 \Gamma_1, \Gamma_2\yields A\product B
 \using \product R
 \endprooftree
 $\\

For every atom or bracket,\\
 
 $
 \begin{array}{l}
 \#_m(\Gamma_1, \Gamma_2\yields A\product B) =\\
 \#^\out_m(A\product B)-\#^\inp_{\overline{m}}(\Gamma_1, \Gamma_2) =\\
 \#^\out_m(A)-\#^\inp_{\overline{m}}(\Gamma_1) +
  \#^\out_m(B)-\#^\inp_{\overline{m}}(\Gamma_2) = \\
 \#_m(\Gamma_1\yields A) + \#_m(\Gamma_2\yields B)
  \end{array}
 $\\
 
 Therefore by \ih,\\
 
 $ \#_\mymin(\Gamma_1, \Gamma_2\yields A\product B)\mle0\mle  \#_\mymax(\Gamma_1, \Gamma_2\yields A\product B)$\\
 
 \end{itemize}
 
 \subsection{Additives} 
 
 \ 
 
 \begin{itemize}
 
 \item
 
 $
 \prooftree
 \Delta(A)\yields D
 \justifies
 \Delta(A\aconj B)\yields D
 \using \aconj L_1
 \endprooftree
 $\\
 
 For every atom or bracket,\\
 
 $
 \begin{array}{l}
 \#_\mymin( \Delta(A\aconj B)\yields D) =\\
 \#^\out_\mymin(D)-\#^\inp_\mymax(\Delta)-\#^\inp_\mymax(A\aconj B) =\\
 \#^\out_\mymin(D)-\#^\inp_\mymax(\Delta)-\mymax(\#^\inp_\mymax(A), \#^\inp_\mymax(B)) \mle\\
 \#^\out_\mymin(D)-\#^\inp_\mymax(\Delta)-\#^\inp_\mymax(A) =\\
 \#_\mymin( \Delta(A)\yields D) \mle 0\ i.h.
 \end{array}
 $\\

And\\

$
 \begin{array}{l}
 \#_\mymax(\Delta(A\aconj B)\yields D) =\\
 \#^\out_\mymax(D)-\#^\inp_\mymin(\Delta)-\#^\inp_\mymin(A\aconj B) =\\
 \#^\out_\mymax(D)-\#^\inp_\mymin(\Delta)-\mymin(\#^\inp_\mymin(A), \#^\inp_\mymin(B)) \mge\\
 \#^\out_\mymax(D)-\#^\inp_\mymin(\Delta)-\#^\inp_\mymin(A) =\\
 \#_\mymax( \Delta(A)\yields D) \mge 0\ i.h.
 \end{array}
 $\\
 
 Therefore:\\
 
 $\#_\mymin( \Delta(A\aconj B)\yields D) \mle 0\mle  \#_\mymax(\Delta(A\aconj B)\yields D)$\\

\item

 $
 \prooftree
 \Delta(B)\yields D
 \justifies
 \Delta(A\aconj B)\yields D
 \using \aconj L_2
 \endprooftree
 $\\
 
 Like $\aconj L_1$.\\
 
 \item
 
 $
 \prooftree
 \Gamma\yields A\tb\Gamma\yields B
 \justifies
 \Gamma\yields A\aconj B
 \using \aconj R
 \endprooftree
 $\\
 
 $
 \begin{array}{l}
 \#_\mymin( \Gamma\yields A\aconj B) =\\
 \#^\out_\mymin(A\aconj B)-\#^\inp_\mymax(\Gamma) =\\
 \mymax(\#^\out_\mymin(A), \#^\out_\mymin(B))-\#^\inp_\mymax(\Gamma) =\\
 \mymax(\#^\out_\mymin(A)-\#^\inp_\mymax(\Gamma), \#^\out_\mymin(B)-\#^\inp_\mymax(\Gamma)) =\\
 \underbrace{\mymax(\underbrace{\#_\mymin(\Gamma\yields A)}_{\mle 0\ i.h.}, \underbrace{\#_\mymin(\Gamma\yields B)}_{\mle 0\ i.h.})}_{\mle 0}
 \end{array}
 $\\
 
 And\\
 
 $
 \begin{array}{l}
 \#_\mymax( \Gamma\yields A\aconj B) =\\
 \#^\out_\mymax(A\aconj B)-\#^\inp_\mymin(\Gamma) =\\
 \mymin(\#^\out_\mymax(A), \#^\out_\mymax(B))-\#^\inp_\mymin(\Gamma) =\\
 \mymin(\#^\out_\mymax(A)-\#^\inp_\mymin(\Gamma), \#^\out_\mymax(B)-\#^\inp_\mymin(\Gamma)) =\\
 \underbrace{\mymin(\underbrace{\#_\mymax(\Gamma\yields A)}_{0 \mle\ i.h.}, \underbrace{\#_\mymax(\Gamma\yields B)}_{0 \mle\ i.h.})}_{0 \mle}
 \end{array}
 $\\
 
 Therefore:\\
 
 $ \#_\mymin( \Gamma\yields A\aconj B) \mle 0\mle \#_\mymax( \Gamma\yields A\aconj B)$.\\

\item

 $
 \prooftree
 \Gamma\yields A
 \justifies
 \Gamma\yields A\adisj B
 \using \adisj R_1
 \endprooftree
 $\\
 
 $
 \begin{array}{l}
 \#_\mymin(\Gamma\yields A\adisj B) =\\
 \#^\out_\mymin(A\adisj B)-\#^\inp_\mymax(\Gamma) =\\
 \mymin(\#^\out_\mymin(A), \#^\inp_\mymin(B))-\#^\inp_\mymax(\Gamma) \mle\\
\#^\out_\mymin(A)-\#^\inp_\mymax(\Gamma) =\\
 \#_\mymin(\Gamma\yields A) \mle 0\ \ih
 \end{array}
 $\\

And\\

$
 \begin{array}{l}
 \#_\mymax(\Gamma\yields A\adisj B) =\\
\#^\out_\mymax(A\adisj B)-\#^\inp_\mymin(\Gamma) =\\
\mymax(\#^\inp_\mymax(A), \#^\inp_\mymax(B))-\#^\inp_\mymin(\Gamma) \mge\\
 \#^\inp_\mymax(A)-\#^\inp_\mymin(\Gamma) =\\
 \#_\mymax(\Gamma\yields A) \mge 0\ \ih
 \end{array}
 $\\

\item

 $
 \prooftree
 \Gamma\yields B
 \justifies
 \Gamma\yields A\adisj B
 \using \adisj R_2
 \endprooftree
 $\\
 
 Like $\adisj R_1$.\\
 
 \item
 
 $
 \prooftree
 \Delta(A)\yields D\tb\Delta(B)\yields D
 \justifies
 \Delta(A\adisj B)\yields D
 \using \adisj L
 \endprooftree
 $\\
 
 For every atom or bracket,\\
 
 $
 \begin{array}{l}
 \#_\mymin(\Delta(A\adisj B)\yields D) =\\
  \#^\out_\mymin(D)-\#^\inp_\mymax(\Delta)-\#^\inp_\mymax(A\adisj B) =\\
 \#^\out_\mymin(D)-\#^\inp_\mymax(\Delta)- \mymin(\#^\inp_\mymax(A), \#^\inp_\mymax(B)) =\\
 \mymax(\#^\out_\mymin(D)-\#^\inp_\mymax(\Delta)-\#^\inp_\mymax(A), \#^\out_\mymin(D)-\#^\inp_\mymax(\Delta)-\#^\inp_\mymax(B)) =\\
 \underbrace{\mymax(\underbrace{\#_\mymin(\Delta(A)\yields D)}_{\mle 0\ \ih}, \underbrace{\#_\mymin(\Delta(B)\yields D)}_{\mle 0\ \ih})}_{\mle 0}
 \end{array}
 $
 
 \ 
 
 $0\mle  \#_\mymax(\Xi(A\adisj B)\yields D)$ similarly
 
 \end{itemize}
 
 \subsection{Bracket modalities}
 
 \ 
 
 \begin{itemize}
 
 \item
 
 $
 \prooftree
 \Gamma(A)\yields B
 \justifies
 \Gamma([\abrack A])\yields B
 \using \abrack L
 \endprooftree
 $\\
 
 For atoms:\\
 
 $
 \begin{array}{l}
 \#_{m, P}(\Gamma([\abrack A])\yields B) =\\
 \#^\out_{m, P}(B) - \#^\inp_{\overline{m}, P}(\Gamma([\abrack A])) =\\
 \#^\out_{m, P}(B) - \#^\inp_{\overline{m}, P}(\Gamma) - \#^\inp_{\overline{m}, P}([\abrack A])) =\\
 \#^\out_{m, P}(B) - \#^\inp_{\overline{m}, P}(\Gamma) - \#^\inp_{\overline{m}, P}(\abrack A)) =\\
 \#^\out_{m, P}(B) - \#^\inp_{\overline{m}, P}(\Gamma) - \#^\inp_{\overline{m}, P}(A)) =\\
  \#^\out_{m, P}(B) - \#^\inp_{\overline{m}, P}(\Gamma(A)) =\\
 \#_{m, P}(\Gamma(A)\yields B)\\
 \end{array}
 $\\
 
 I.e.\ the property for the conclusion follows from the induccion hypothesis for the premise
 since brackets and bracket modalities
 are transparent to atom count.\\
 
 For brackets:\\

$
 \begin{array}{l}
 \#_{m, []}(\Gamma([\abrack A])\yields B) =\\
 \#^\out_{m, []}(B) - \#^\inp_{\overline{m}, []}(\Gamma([\abrack A])) =\\
 \#^\out_{m, []}(B) - \#^\inp_{\overline{m}, []}(\Gamma) - \#^\inp_{\overline{m}, []}([\abrack A])) =\\
 \#^\out_{m, []}(B) - \#^\inp_{\overline{m}, []}(\Gamma) - \#^\inp_{\overline{m}, []}(\abrack A)-1 =\\
 \#^\out_{m, []}(B) - \#^\inp_{\overline{m}, []}(\Gamma) - \#^\inp_{\overline{m}, []}(A)+1-1 =\\
 \#^\out_{m, []}(B) - \#^\inp_{\overline{m}, []}(\Gamma) - \#^\inp_{\overline{m}, []}(A))=\\
\#^\out_{m, []}(B) - \#^\inp_{\overline{m}, []}(\Gamma(A)) =\\
 \#_{m, []}(\Gamma(A)\yields B) =\\
 \end{array}
$\\

Therefore by \ih,\\ 

$\#_{\mymin}(\Gamma([\abrack A])\yields B)\mle 0\mle \#_{\mymax}(\Gamma([\abrack A])\yields B)$\\

\item

$
\prooftree
{}[\Gamma]\yields A
\justifies
\Gamma\yields \abrack A
\using \abrack R
\endprooftree
$\\

For atoms:\\

$\#_{m, P}(\Gamma\yields \abrack A) = \#_{m, P}(\Gamma\yields A) = \#_{m, P}([\Gamma]\yields A)$\\

Since brackets and bracket modalities are transparant to atom count.\\

For brackets:\\

$
\begin{array}{l}
\#_{m, []}(\Gamma\yields \abrack A) =\\
\#^\out_{m, []}(\abrack A) - \#^\inp_{\overline{m}, []}(\Gamma) =\\
\#^\out_{m, []}(A) -1 - \#^\inp_{\overline{m}, []}(\Gamma) =\\
\#^\out_{m, []}(A) - (\#^\inp_{\overline{m}, []}(\Gamma)+1) =\\
\#^\out_{m, []}(A) - \#^\inp_{\overline{m}, []}([\Gamma]) =\\
\#_{m, []}([\Gamma]\yields A)
\end{array}
$

\ \\

Therefore by \ih

\ 

$\#_{\mymin}(\Gamma\yields \abrack A)\mle 0\mle \#_{\mymax}(\Gamma\yields \abrack A)$\\

\item

$
\prooftree
\Gamma([A])\yields B
\justifies
\Gamma(\mybrack A)\yields B
\using \mybrack L
\endprooftree
$\\

For atoms,\\

$\#_{m, P}(\Gamma(\mybrack A)\yields B) = \#_{m, P}(\Gamma([A])\yields B)$\\

since brackets and bracket modalities are transparent to atom count.\\

For brackets,\\

$
\begin{array}{l}
\#_{m, []}(\Gamma(\mybrack A)\yields B) =\\
\#^\out_{m, []}(B) - \#^\inp_{\overline{m}, []}(\Gamma)- \#^\inp_{\overline{m}, []}(\mybrack A) =\\
\#^\out_{m, []}(B) - \#^\inp_{\overline{m}, []}(\Gamma) -(\#^\inp_{\overline{m}, []}(A)+1) =\\
\#^\out_{m, []}(B) - \#^\inp_{\overline{m}, []}(\Gamma) -\#^\inp_{\overline{m}, []}([A]) =\\
 \#_{m, []}(\Gamma([A])\yields B)
 \end{array}
$\\

Therefore by \ih\\

$\#_{\mymin}(\Gamma(\mybrack A)\yields B)\mle 0\mle \#_{\mymax}(\Gamma(\mybrack A)\yields B)$\\

\item

$
\prooftree
\Gamma\yields A
\justifies
{}[\Gamma]\yields \mybrack A
\using \mybrack R
\endprooftree
$\\

For atoms,\\

$\#_{m, P}([\Gamma]\yields \mybrack A) = \#_{m, P}(\Gamma\yields A)$\\

since brackets and bracket modalities are transparent to atom count.\\

For brackets,\\

$
\begin{array}{l}
\#_{m, []}([\Gamma]\yields \mybrack A) =\\
\#^\out_{m, []}(\mybrack A) - \#^\inp_{\overline{m}, []}([\Gamma]) =\\
\#^\out_{m, []}(A) + 1 - \#^\inp_{\overline{m}, []}(\Gamma) -1 =\\
\#^\out_{m, []}(A)- \#^\inp_{\overline{m}, []}(\Gamma) =\\
\#_{m, []}(\Gamma\yields A)\\
\end{array}
$\\

Therefore by \ih:\\ 

$\#_{\mymin}([\Gamma]\yields \mybrack A)\mle 0\mle \#_{\mymax}([\Gamma]\yields \mybrack A)$

\end{itemize}

\subsection{Subexponentials}

\ 

\begin{itemize}

\item

$
\prooftree
\Delta(A)\yields D
\justifies
\Delta(\univexp A)\yields D
\using \univexp L
\endprooftree
$\\

For atoms,\\

$
\begin{array}{l}
\#_{\mymin, P}(\Delta(\univexp A)\yields D) =\\
\#^\out_{\mymin, P}(D)-\#^\inp_{\mymax, P}(\Delta)-\#^\inp_{\mymax, P}(\univexp A) =\\
\#^\out_{\mymin, P}(D)-\#^\inp_{\mymax, P}(\Delta)-X^+(\#^\inp_{\mymax, P}(A)) \mle\\
\#^\out_{\mymin, P}(D)-\#^\inp_{\mymax, P}(\Delta)-\#^\inp_{\mymax, P}(A) =\\
\#_{\mymin, P}(\Delta(A)\yields D) \mle 0\ \ih
\end{array}
$\\

And\\

$
\begin{array}{l}
\#_{\mymax, P}(\Delta(\univexp A)\yields D) =\\
\#^\out_{\mymax, P}(D)-\#^\inp_{\mymin, P}(\Delta)-\#^\inp_{\mymin, P}(\univexp A) =\\
\#^\out_{\mymax, P}(D)-\#^\inp_{\mymin, P}(\Delta)-X^-(\#^\inp_{\mymin, P}(A)) \mge\\
\#^\out_{\mymax, P}(D)-\#^\inp_{\mymin, P}(\Delta)-\#^\inp_{\mymin, P}(A) =\\
\#_{\mymax, P}(\Delta(A)\yields D) \mge 0\ i.h.
\end{array}
$\\

For brackets,\\

$
\begin{array}{l}
\#_{\mymin, []}(\Delta(\univexp A)\yields D) =\\
\#^\out_{\mymin, []}(D)-\#^\inp_{\mymax, []}(\Delta)-\#^\inp_{\mymax, []}(\univexp A) =\\
\#^\out_{\mymin, []}(D)-\#^\inp_{\mymax, []}(\Delta)-\mytop \mle\\
\#^\out_{\mymin, []}(D)-\#^\inp_{\mymax, []}(\Delta)-\#^\inp_{\mymax, []}(A) =\\
\#_{\mymin, []}(\Delta(A)\yields D) \mle 0\ i.h.
\end{array}
$\\

And\\

$
\begin{array}{l}
\#_{\mymax, []}(\Delta(\univexp A)\yields D) =\\
\#^\out_{\mymax, []}(D)-\#^\inp_{\mymin, []}(\Delta)-\#^\inp_{\mymin, []}(\univexp A) =\\
\#^\out_{\mymax, []}(D)-\#^\inp_{\mymin, []}(\Delta)-X^-(\#^\inp_{\mymin, []}(A)) \mge\\
\#^\out_{\mymax, []}(D)-\#^\inp_{\mymin, []}(\Delta)-\#^\inp_{\mymin, []}(A) =\\
\#_{\mymax, []}(\Delta(A)\yields D) \mge 0\ \ih
\end{array}
$

\ 

\item

\prooftree
\univexp A_1, \ldots, \univexp A_n\yields A
\justifies
\univexp A_1, \ldots, \univexp A_n\yields \univexp A
\using \univexp R
\endprooftree\\

For atoms and brackets,\\

$
\begin{array}{l}
\#_{\mymin, Q}(\univexp A_1, \ldots, \univexp A_n\yields \univexp A) =\\
\#^\out_{\mymin, Q}(\univexp A)-\#^\inp_{\mymax, Q}(\univexp A_1, \ldots, \univexp A_n) =\\
X^-(\#^\out_{\mymin, Q}(A))-\#^\inp_{\mymax, Q}(\univexp A_1, \ldots, \univexp A_n) \mle\\
\#^\out_{\mymin, Q}(A)-\#^\inp_{\mymax, Q}(\univexp A_1, \ldots, \univexp A_n) =\\
\#_{\mymin, Q}(\univexp A_1, \ldots, \univexp A_n\yields A) \mle 0\ \ih\\
 \end{array}
$\\

And,\\

$
\begin{array}{l}
\#_{\mymax, Q}(\univexp A_1, \ldots, \univexp A_n\yields \univexp A) =\\
\#^\out_{\mymax, Q}(\univexp A)-\#^\inp_{\mymin, Q}(\univexp A_1, \ldots, \univexp A_n) =\\
X^+(\#^\out_{\mymax, Q}(A))-\#^\inp_{\mymin, Q}(\univexp A_1, \ldots, \univexp A_n) \mge\\
\#^\out_{\mymax, Q}(A)-\#^\inp_{\mymin, Q}(\univexp A_1, \ldots, \univexp A_n) =\\
\#_{\mymax, Q}(\univexp A_1, \ldots, \univexp A_n\yields A) \mge 0\ \ih\\
 \end{array}
$\\

\ \\

\item

\prooftree
\Delta(\univexp A_0, \ldots, \univexp A_n, [\univexp A_0, \ldots, \univexp A_n, \Gamma])\yields D
\justifies
\Delta(\univexp A_0, \ldots, \univexp A_n, \Gamma)\yields D
\using \univexp C
\endprooftree\\

For atoms,\\

$
\begin{array}{l}
\#_\mymin(\Delta(\univexp A_0, \ldots, \univexp A_n, \Gamma)\yields D) =\\
\#^\out_\mymin(D)-\#^\inp_\mymax(\Delta, \Gamma)-\#^\inp_\mymax(\univexp A_0)-\cdots-
\#^\inp_\mymax(\univexp A_n) =\\
\#^\out_\mymin(D)-\#^\inp_\mymax(\Delta, \Gamma)-X^+(\#^\inp_\mymax(A_0))-\cdots-
X^+(\#^\inp_\mymax(A_n)) \mle\\
\#^\out_\mymin(D)-\#^\inp_\mymax(\Delta, [\Gamma])-\\X^+(\#^\inp_\mymax(A_0))-\cdots-
X^+(\#^\inp_\mymax(A_n))-X^+(\#^\inp_\mymax(A_0))-\cdots-
X^+(\#^\inp_\mymax(A_n)) =\\
\#_\mymin(\Delta(\univexp A_0, \ldots, \univexp A_n, [\univexp A_0, \ldots, \univexp A_n, \Gamma])\yields D)
\end{array}
$\\

For brackets,\\

$
\begin{array}{l}
\#_\mymin(\Delta(\univexp A_0, \ldots, \univexp A_n, \Gamma)\yields D) =\\
\#^\out_\mymin(D)-\#^\inp_\mymax(\Delta, \Gamma)-\#^\inp_\mymax(\univexp A_0)-\cdots-
\#^\inp_\mymax(\univexp A_n) =\\
\#^\out_\mymin(D)-\#^\inp_\mymax(\Delta, \Gamma)-\mytop-\cdots-
\mytop \mle\\
\#^\out_\mymin(D)-\#^\inp_\mymax(\Delta, [\Gamma])-\mytop-\cdots-
\mytop-\mytop-\cdots-
\mytop =\\
\#_\mymin(\Delta(\univexp A_0, \ldots, \univexp A_n, [\univexp A_0, \ldots, \univexp A_n, \Gamma])\yields D)
\end{array}
$\\

And for atoms and brackets,\\

$
\begin{array}{l}
\#_\mymax(\Delta(\univexp A_0, \ldots, \univexp A_n, \Gamma)\yields D) =\\
\#^\out_\mymax(D)-\#^\inp_\mymin(\Delta, \Gamma)-\#^\inp_\mymin(\univexp A_0)-\cdots-
\#^\inp_\mymin(\univexp A_n) =\\
\#^\out_\mymax(D)-\#^\inp_\mymin(\Delta, \Gamma)-X^-(\#^\inp_\mymin(A_0))-\cdots-
X^-(\#^\inp_\mymin(A_n)) \mle\\
\#^\out_\mymax(D)-\#^\inp_\mymin(\Delta, [\Gamma])-\\X^-(\#^\inp_\mymin(A_0))-\cdots-
X^-(\#^\inp_\mymin(A_n))-X^+(\#^\inp_\mymin(A_0))-\cdots-
X^-(\#^\inp_\mymin(A_n)) =\\
\#_\mymax(\Delta(\univexp A_0, \ldots, \univexp A_n, [\univexp A_0, \ldots, \univexp A_n, \Gamma])\yields D)
\end{array}
$\\

\item

$
\prooftree
\Delta(A)\yields D\tb \Delta(A, A)\yields D \ldots
\justifies
\Delta(\exstexp A)\yields D
\using \exstexp L
\endprooftree
$\\

For atoms and brackets,\\

$
\begin{array}{l}
\#_\mymin(\Delta(\exstexp A)\yields D) =\\
\#^\out_\mymin(D) - \#^\inp_\mymax(\Delta)-\#^\inp_\mymax(\exstexp A) =\\
\#^\out_\mymin(D) - \#^\inp_\mymax(\Delta)-X^+(\#^\inp_\mymax(A)) \mle\\
\#^\out_\mymin(D) - \#^\inp_\mymax(\Delta)-\#^\inp_\mymax(A) =\\
\#_\mymin(\Delta(A)\yields D) \mle 0\ \ih
\end{array}
$\\

And\\

$
\begin{array}{l}
\#_\mymax(\Delta(\exstexp A)\yields D) =\\
\#^\out_\mymax(D) - \#^\inp_\mymin(\Delta)-\#^\inp_\mymin(\exstexp A) =\\
\#^\out_\mymax(D) - \#^\inp_\mymin(\Delta)-X^-(\#^\inp_\mymin(A)) \mge\\
\#^\out_\mymax(D) - \#^\inp_\mymin(\Delta)-\#^\inp_\mymin(A) =\\
\#_\mymax(\Delta(A)\yields D) \mge 0\ \ih
\end{array}
$\\

\item

\prooftree
\Gamma_0\yields C\tb\cdots\tb\Gamma_n\yields C
\justifies
\Gamma_1, \ldots, \Gamma_n\yields\exstexp C
\using \exstexp R
\endprooftree\\

For atoms and brackets,\\

$
\begin{array}{l}
\#_\mymin(\Gamma_0, \ldots, \Gamma_n\yields \exstexp A) =\\
\#^\out_\mymin(\exstexp A)-\#^\inp_\mymax(\Gamma_0)-\cdots -\#^\inp_\mymax(\Gamma_n) =\\
X^-(\#^\out_\mymin(A))-\#^\inp_\mymax(\Gamma_0)-\cdots -\#^\inp_\mymax(\Gamma_n) \mle\\
n\cdot(\#^\out_\mymin(A))-\#^\inp_\mymax(\Gamma_0)-\cdots -\#^\inp_\mymax(\Gamma_n) \mle\\
\underbrace{\underbrace{\#^\out_\mymin(A)-\#^\inp_\mymax(\Gamma_0)}_{\mle 0\ \ih}+\cdots+\underbrace{\#^\out_\mymin(A)-\#^\inp_\mymax(\Gamma_n)}_{\mle 0\ \ih}}_{\mle 0}
\end{array}
$\\

And,\\

$
\begin{array}{l}
\#_\mymax(\Gamma_0, \ldots, \Gamma_n\yields \exstexp A) =\\
\#^\out_\mymax(\exstexp A)-\#^\inp_\mymin(\Gamma_0)-\cdots -\#^\inp_\mymin(\Gamma_n) =\\
X^-(\#^\out_\mymax(A))-\#^\inp_\mymin(\Gamma_0)-\cdots -\#^\inp_\mymin(\Gamma_n) \mge\\
n\cdot(\#^\out_\mymax(A))-\#^\inp_\mymin(\Gamma_0)-\cdots -\#^\inp_\mymin(\Gamma_n) =\\
\underbrace{\underbrace{\#^\out_\mymax(A)-\#^\inp_\mymin(\Gamma_0)}_{\mge 0\ \ih}+\cdots+\underbrace{\#^\out_\mymax(A)-\#^\inp_\mymin(\Gamma_n)}_{\mge 0\ \ih}}_{\mge 0}
\end{array}
$

\end{itemize}

\part{GRAMMAR}

\noindent
In this part we exemplify the applications to grammar of categorial logic.
Chapter~\ref{lexchap} contains the lexicon.
The subsequent chapters cover linguistic applications including
Montague grammar, coordination, discontinuity and relativisation.

\chapter{Lexicon}

\label{lexchap}


\ 

\noindent
$
{\bf 007}: {\blacksquare}{\forall}gNt(s(g)): {\it 007}\\
{\bf a}: {\blacksquare}{\forall}g({\forall}f((Sf{{\uparrow}{}}{\blacksquare}Nt(s(g))){{\downarrow}{}}Sf)/{\it CN}{\it s(g)}): \lambda A\lambda B\exists C[({\it A}\ {\it C})\wedge ({\it B}\ {\it C})]\\
{\bf admire}: {\square}(({\langle\rangle}({\exists}aNa{-}{\exists}gNt(s(g)))\backslash Sf)/{\exists}aNa): \mbox{\^{}}\lambda A\lambda B({\it Pres}\ ((\mbox{\v{}}{\it admire}\ {\it A})\ {\it B}))\\
{\bf And}: {\blacksquare}{\forall}f(Sf/Sf): \lambda A{\it A}\\
{\bf and}: {\blacksquare}{\forall}f((?{\blacksquare}Sf\backslash {[]^{-1}}{[]^{-1}}Sf)/{\blacksquare}Sf): (\Phinplus{}\ {\it 0}\ {\it and})\\
{\bf and}: {\blacksquare}{\forall}a{\forall}f((?{\blacksquare}({\langle\rangle}Na\backslash Sf)\backslash {[]^{-1}}{[]^{-1}}({\langle\rangle}Na\backslash Sf))/{\blacksquare}({\langle\rangle}Na\backslash Sf)): (\Phinplus{}\ ({\it s}\ {\it 0})\ {\it and})\\
{\bf and}: {\blacksquare}{\forall}a{\forall}f((?{\blacksquare}(Sf/!Na)\backslash {[]^{-1}}{[]^{-1}}(Sf/!Na))/{\blacksquare}(Sf/!Na)): (\Phinplus{}\ ({\it s}\ {\it 0})\ {\it and})\\
{\bf and}: {\blacksquare}{\forall}f((?{\blacksquare}(Sf/{\exists}aNa)\backslash {[]^{-1}}{[]^{-1}}(Sf/{\exists}aNa))/{\blacksquare}(Sf/{\exists}aNa)): (\Phinplus{}\ ({\it s}\ {\it 0})\ {\it and})\\
{\bf and}: {\blacksquare}{\forall}w{\forall}a{\forall}b{\forall}f(({\blacksquare}((Sf{{\uparrow}{}}((({\langle\rangle}Na\backslash Sf){\multimapinv}Ww)/Nb)){{\begin{picture}(7,7)(0,-2)\put(1.5,-3){\rotatebox{90}{$\multimap$}}\end{picture}}{_{_{2}}}}Ww)\backslash {[]^{-1}}{[]^{-1}}((Sf{{\uparrow}{}}((({\langle\rangle}Na\backslash Sf){\multimapinv}Ww)/Nb)){{\begin{picture}(7,7)(0,-2)\put(1.5,-3){\rotatebox{90}{$\multimap$}}\end{picture}}{_{_{2}}}}Ww))/\\
\tab\mbox{$\hat{\ }$}\mbox{$\hat{\ }$}{\blacksquare}((Sf{{\uparrow}{}}((({\langle\rangle}Na\backslash Sf){\multimapinv}Ww)/Nb)){{\begin{picture}(7,7)(0,-2)\put(1.5,-3){\rotatebox{90}{$\multimap$}}\end{picture}}{_{_{2}}}}Ww)): \lambda A\lambda B\lambda C[({\it B}\ {\it C})\wedge ({\it A}\ {\it C})]\\
{\bf and}: {\blacksquare}{\forall}f{\forall}a((?{\blacksquare}(({\langle\rangle}Na\backslash Sf)/{\exists}bNb)\backslash {[]^{-1}}{[]^{-1}}(({\langle\rangle}Na\backslash Sf)/{\exists}bNb))/{\blacksquare}(({\langle\rangle}Na\backslash Sf)/{\exists}bNb)): (\Phinplus{}\ ({\it s}\ ({\it s}\ {\it 0}))\ {\it and})\\
{\bf and}: {\blacksquare}{\forall}f{\forall}a((?{\blacksquare}((({\langle\rangle}Na\backslash Sf)/{\exists}bNb)\backslash ({\langle\rangle}Na\backslash Sf))\backslash {[]^{-1}}{[]^{-1}}((({\langle\rangle}Na\backslash Sf)/{\exists}bNb)\backslash ({\langle\rangle}Na\backslash Sf)))/{\blacksquare}((({\langle\rangle}Na\backslash Sf)/\\
\tab{\exists}bNb)\backslash ({\langle\rangle}Na\backslash Sf))): (\Phinplus{}\ ({\it s}\ ({\it s}\ {\it 0}))\ {\it and})\\
{\bf and}: {\blacksquare}{\forall}f{\forall}a(({\blacksquare}((({\langle\rangle}Na\backslash Sf)/({\exists}bNb{\oplus}{\exists}g(({\it CN}{\it g}/{\it CN}{\it g}){\sqcup}({\it CN}{\it g}\backslash {\it CN}{\it g}))))\backslash ({\langle\rangle}Na\backslash Sf))\backslash {[]^{-1}}{[]^{-1}}((({\langle\rangle}Na\backslash Sf)/\\
\tab({\exists}bNb{\oplus}{\exists}g(({\it CN}{\it g}/{\it CN}{\it g}){\sqcup}({\it CN}{\it g}\backslash {\it CN}{\it g}))))\backslash ({\langle\rangle}Na\backslash Sf)))/{\blacksquare}((({\langle\rangle}Na\backslash Sf)/({\exists}bNb{\oplus}{\exists}g(({\it CN}{\it g}/{\it CN}{\it g}){\sqcup}\\
\tab({\it CN}{\it g}\backslash {\it CN}{\it g}))))\backslash ({\langle\rangle}Na\backslash Sf))): \lambda A\lambda B\lambda C\lambda D[(({\it B}\ {\it C})\ {\it D})\wedge (({\it A}\ {\it C})\ {\it D})]\\
{\bf and}: {\blacksquare}{\forall}a{\forall}b{\forall}f((?{\blacksquare}((({\langle\rangle}Na\backslash Sf)/({\exists}cNc{\oplus}{\it CP}b))\backslash ({\langle\rangle}Na\backslash Sf))\backslash {[]^{-1}}{[]^{-1}}((({\langle\rangle}Na\backslash Sf)/({\exists}cNc{\oplus}{\it CP}b))\backslash ({\langle\rangle}Na\backslash Sf)))/\\
\tab{\blacksquare}((({\langle\rangle}Na\backslash Sf)/({\exists}cNc{\oplus}{\it CP}b))\backslash ({\langle\rangle}Na\backslash Sf))): (\Phinplus{}\ ({\it s}\ ({\it s}\ {\it 0}))\ {\it and})\\
{\bf and}: {\blacksquare}{\forall}a{\forall}b{\forall}f((?{\blacksquare}(({\langle\rangle}Na\backslash Sf)/{\it PP}{\it b})\backslash {[]^{-1}}{[]^{-1}}(({\langle\rangle}Na\backslash Sf)/{\it PP}{\it b}))/{\blacksquare}(({\langle\rangle}Na\backslash Sf)/{\it PP}{\it b})): (\Phinplus{}\ ({\it s}\ ({\it s}\ {\it 0}))\ {\it and})\\
{\bf and}: {\blacksquare}{\forall}a{\forall}b{\forall}f((?{\blacksquare}((({\langle\rangle}Na\backslash Sf)/({\exists}cNc{\bullet}{\it PP}{\it b}))\backslash ({\langle\rangle}Na\backslash Sf))\backslash {[]^{-1}}{[]^{-1}}((({\langle\rangle}Na\backslash Sf)/({\exists}cNc{\bullet}{\it PP}{\it b}))\backslash ({\langle\rangle}Na\backslash Sf)))/\\
\tab{\blacksquare}((({\langle\rangle}Na\backslash Sf)/({\exists}cNc{\bullet}{\it PP}{\it b}))\backslash ({\langle\rangle}Na\backslash Sf))): (\Phinplus{}\ ({\it s}\ ({\it s}\ {\it 0}))\ {\it and})\\
{\bf and}: {\blacksquare}{\forall}a{\forall}b{\forall}f((?{\blacksquare}(({\langle\rangle}Na\backslash Sf)/!Nb)\backslash {[]^{-1}}{[]^{-1}}(({\langle\rangle}Na\backslash Sf)/!Nb))/{\blacksquare}(({\langle\rangle}Na\backslash Sf)/!Nb)): (\Phinplus{}\ ({\it s}\ ({\it s}\ {\it 0}))\ {\it and})\\
{\bf ate}: {\square}(({\langle\rangle}{\exists}aNa\backslash Sf)/{\exists}aNa): \mbox{\^{}}\lambda A\lambda B({\it Past}\ ((\mbox{\v{}}{\it eat}\ {\it A})\ {\it B}))\\
{\bf bagels}: {\square}(Nt(p(n)){\&}{\it CN}{\it p(n)}): \mbox{\^{}}(({\it gen}\ \mbox{\v{}}{\it bagels}), \mbox{\v{}}{\it bagels})\\
{\bf barn}: {\square}{\it CN}{\it s(n)}: {\it barn}\\
{\bf be}: {\square}(({\langle\rangle}W[there]{\multimap}Sb)/{\exists}aNa): \mbox{\^{}}\lambda A(\mbox{\v{}}{\it be}\ {\it A})\\
{\bf before}: {\blacksquare}({\forall}a{\forall}f(({\langle\rangle}Na\backslash Sf)\backslash ({\langle\rangle}Na\backslash Sf))/Sf): \lambda A\lambda B\lambda C(({\it before}\ {\it A})\ ({\it B}\ {\it C}))\\
{\bf beginning}: {\square}{\it CN}{\it s(n)}: {\it beginning}\\
{\bf believes}: {\square}(({\langle\rangle}{\exists}gNt(s(g))\backslash Sf)/({\it CP}that{\sqcup}{\square}Sf)): \mbox{\^{}}\lambda A\lambda B({\it Pres}\ ((\mbox{\v{}}{\it believe}\ {\it A})\ {\it B}))\\
{\bf bill}: {\blacksquare}Nt(s(m)): {\it b}\\
{\bf bond}: {\blacksquare}Nt(s(m)): {\it b}\\
{\bf book}: {\square}{\it CN}{\it s(n)}: {\it book}\\
{\bf bought}: {\square}(({\langle\rangle}{\exists}aNa\backslash Sf)/({\exists}aNa{\bullet}{\exists}aNa)): \mbox{\^{}}\lambda A\lambda B({\it Past}\ (((\mbox{\v{}}{\it buy}\ \pi_1{\it A})\ \pi_2{\it A})\ {\it B}))\\
{\bf bought}: {\square}(({\langle\rangle}{\exists}aNa\backslash Sf)/{\exists}aNa): \mbox{\^{}}\lambda A\lambda B({\it Past}\ ((\mbox{\v{}}{\it buy}\ {\it A})\ {\it B}))\\
{\bf by}: {\blacksquare}{\forall}a((({\langle\rangle}Na\backslash S-)\backslash ({\langle\rangle}Na\backslash S-))/Na): \lambda A\lambda B\lambda C[[{\it C}={\it A}]\wedge ({\it B}\ {\it C})]\\
{\bf by}: {\square}({\forall}n({\it CN}{\it n}\backslash {\it CN}{\it n})/{\exists}aNa): \mbox{\^{}}\lambda A\lambda B((\mbox{\v{}}{\it by}\ {\it A})\ {\it B})\\
{\bf buys}: {\square}(({\langle\rangle}{\exists}gNt(s(g))\backslash Sf)/({\exists}aNa{\bullet}{\exists}aNa)): \mbox{\^{}}\lambda A\lambda B({\it Pres}\ (((\mbox{\v{}}{\it buy}\ \pi_1{\it A})\ \pi_2{\it A})\ {\it B}))\\
{\bf calls}: {\square}(({\langle\rangle}{\exists}gNt(s(g))\backslash Sf)/{\exists}a((W[up]{\LEFTcircle}Na){\sqcup}(Na{\RIGHTcircle}W[up]))): \mbox{\^{}}\lambda A\lambda B((\mbox{\v{}}{\it phone}\ {\it A})\ {\it B})\\
{\bf catch}: {\square}(({\langle\rangle}{\exists}aNa\backslash Sb)/{\exists}aNa): \mbox{\^{}}\lambda A\lambda B((\mbox{\v{}}{\it catch}\ {\it A})\ {\it B})\\
{\bf cezanne}: {\blacksquare}Nt(s(m)): {\it c}\\
{\bf cd}: {\square}{\it CN}{\it s(n)}: {\it cd}\\
{\bf charles}: {\blacksquare}Nt(s(m)): {\it c}\\
{\bf clark}: {\blacksquare}{\forall}gNt(s(g)): {\it c}\\
{\bf coffee}: {\square}(Nt(s(n)){\&}{\it CN}{\it s(n)}): \mbox{\^{}}(({\it gen}\ \mbox{\v{}}{\it coffee}), \mbox{\v{}}{\it coffee})\\
{\bf created}: {\square}(({\langle\rangle}{\exists}aNa\backslash Sf)/{\exists}aNa): \mbox{\^{}}\lambda A\lambda B({\it Past}\ ((\mbox{\v{}}{\it create}\ {\it A})\ {\it B}))\\
{\bf darkness}: {\square}({\it CN}{\it s(n)}{\&}Nt(s(n))): \mbox{\^{}}(\mbox{\v{}}{\it darkness}, ({\it gen}\ \mbox{\v{}}{\it darkness}))\\
{\bf deep}: {\square}{\it CN}{\it s(n)}: {\it deep}\\
{\bf did}: {\blacksquare}{\forall}a{\forall}g{\forall}b{\forall}h(((({\langle\rangle}Na\backslash Sg){{\uparrow}{}}({\langle\rangle}Nb\backslash Sh))/({\exists}c{\langle\rangle}Nc\backslash Sf))\backslash (({\langle\rangle}Na\backslash Sg){{\uparrow}{}}({\langle\rangle}Nb\backslash Sh))): \lambda A\lambda B(({\it A}\ {\it B})\ {\it B})\\
{\bf did}{+}{\bf too}: ((({\langle\rangle}NA\backslash SB){{\uparrow}{}}({\langle\rangle}NC\backslash SD))/({\langle\rangle}NE\backslash SF))\backslash (({\langle\rangle}NG\backslash SH){{\uparrow}{}}(NI\backslash SJ)): \lambda K\lambda L(({\it K}\ {\it L})\ {\it L})\\
{\bf doesnt}: {\blacksquare}{\forall}g{\forall}a((Sg{{\uparrow}{}}(({\langle\rangle}Na\backslash Sf)/({\langle\rangle}Na\backslash Sb))){{\downarrow}{}}Sg): \lambda A\neg ({\it A}\ \lambda B\lambda C({\it B}\ {\it C}))\\
{\bf dog}: {\square}{\it CN}{\it s(n)}: {\it dog}\\
{\bf donuts}: {\square}(Nt(p(n)){\&}{\it CN}{\it p(n)}): \mbox{\^{}}(({\it gen}\ \mbox{\v{}}{\it donuts}), \mbox{\v{}}{\it donuts})\\
{\bf earth}: {\square}{\it CN}{\it s(n)}: {\it earth}\\
{\bf eat}: {\square}(({\langle\rangle}{\exists}aNa\backslash Sb)/{\exists}aNa): \mbox{\^{}}\lambda A\lambda B((\mbox{\v{}}{\it eat}\ {\it A})\ {\it B})\\
{\bf edinburgh}: {\blacksquare}Nt(s(n)): {\it e}\\
{\bf editor}: {\square}({\forall}g{\it CN}{\it s(g)}/{\it PP}{\it of}): {\it editor}\\
{\bf every}: {\blacksquare}{\forall}g({\forall}f((Sf{{\uparrow}{}}Nt(s(g))){{\downarrow}{}}Sf)/{\it CN}{\it s(g)}): \lambda A\lambda B\forall C[({\it A}\ {\it C})\rightarrow ({\it B}\ {\it C})]\\
{\bf everyone}: {\square}{\forall}f((Sf{{\uparrow}{}}{\forall}gNt(g)){{\downarrow}{}}Sf): \mbox{\^{}}\lambda A\forall B[(\mbox{\v{}}{\it person}\ {\it B})\rightarrow ({\it A}\ {\it B})]\\
{\bf face}: {\square}{\it CN}{\it s(n)}: {\it face}\\
{\bf fell}: {\square}({\exists}a{\langle\rangle}Na\backslash Sf): \mbox{\^{}}\lambda A({\it Past}\ (\mbox{\v{}}{\it fall}\ {\it A}))\\
{\bf filed}: {\square}(({\langle\rangle}{\exists}gNt(s(g))\backslash Sf)/{\exists}aNa): \mbox{\^{}}\lambda A\lambda B({\it Past}\ ((\mbox{\v{}}{\it file}\ {\it A})\ {\it B}))\\
{\bf finds}: {\square}(({\langle\rangle}{\exists}gNt(s(g))\backslash Sf)/{\exists}aNa): \mbox{\^{}}\lambda A\lambda B({\it Pres}\ ((\mbox{\v{}}{\it find}\ {\it A})\ {\it B}))\\
{\bf fish}: {\square}{\it CN}{\it s(n)}: {\it fish}\\
{\bf for}: {\blacksquare}({\it PP}{\it for}/{\exists}aNa): \lambda A{\it A}\\
{\bf form}: {\square}({\it CN}{\it s(n)}{\&}Nt(s(n))): \mbox{\^{}}(\mbox{\v{}}{\it form}, ({\it gen}\ \mbox{\v{}}{\it form}))\\
{\bf fortunately}: {\square}{\forall}f(\mbox{$\check{\ }$}Sf{{}{\downarrow}{}}Sf): {\it fortunately}\\
{\bf friends}: {\square}({\it CN}{\it p}/{\it PP}{\it of}): {\it friends}\\
{\bf from}: {\square}(({\forall}a{\forall}f(({\langle\rangle}Na\backslash Sf)\backslash ({\langle\rangle}Na\backslash Sf)){\&}{\forall}n({\it CN}{\it n}\backslash {\it CN}{\it n}))/{\exists}bNb): \mbox{\^{}}\lambda A((\mbox{\v{}}{\it fromadv}\ {\it A}), (\mbox{\v{}}{\it fromadn}\ {\it A}))\\
{\bf gave}: {\square}(({\langle\rangle}{\exists}aNa\backslash Sf)/({\exists}bNb{\bullet}{\it PP}{\it to})): \mbox{\^{}}\lambda A\lambda B({\it Past}\ (((\mbox{\v{}}{\it give}\ \pi_2{\it A})\ \pi_1{\it A})\ {\it B}))\\
{\bf gave}: {\square}(({\langle\rangle}{\exists}gNt(s(g))\backslash Sf)/({\exists}aNa{\RIGHTcircle}W[the,cold,shoulder])): \mbox{\^{}}\lambda A\lambda B({\it Past}\ ((\mbox{\v{}}{\it shun}\ {\it A})\ {\it B}))\\
{\bf gave}: {\square}((({\langle\rangle}{\exists}aNa\backslash Sf)/{\exists}aNa)/{\exists}aNa): \mbox{\^{}}\lambda A\lambda B\lambda C({\it Past}\ (((\mbox{\v{}}{\it give}\ {\it A})\ {\it B})\ {\it C}))\\
{\bf girl}: {\square}{\it CN}{\it s(f)}: {\it girl}\\
{\bf gives}: {\square}(({\langle\rangle}{\exists}gNt(s(g))\backslash Sf)/({\exists}aNa{\RIGHTcircle}W[the,cold,shoulder])): \mbox{\^{}}\lambda A\lambda B({\it Pres}\ ((\mbox{\v{}}{\it shun}\ {\it A})\ {\it B}))\\
{\bf God}: {\blacksquare}Nt(s(m)): {\it God}\\
{\bf good}: {\square}{\forall}n({\it CN}{\it n}/{\it CN}{\it n}): {\it good}\\
{\bf has}: {\square}(({\langle\rangle}{\exists}gNt(s(g))\backslash Sf)/{\exists}aNa): \mbox{\^{}}\lambda A\lambda B({\it Pres}\ ((\mbox{\v{}}{\it have}\ {\it A})\ {\it B}))\\
{\bf he}: {\blacksquare}{[]^{-1}}{\forall}g(({\blacksquare}Sg{|}{\blacksquare}Nt(s(m)))/({\langle\rangle}Nt(s(m))\backslash Sg)): \lambda A{\it A}\\
{\bf heaven}: {\square}{\it CN}{\it s(n)}: {\it heaven}\\
{\bf her}: {\blacksquare}{\forall}g{\forall}a((({\langle\rangle}Na\backslash Sg){{}{\uparrow}{}}{\blacksquare}Nt(s(f))){{}{\downarrow}{}}({\blacksquare}({\langle\rangle}Na\backslash Sg){|}{\blacksquare}Nt(s(f)))): \lambda A{\it A}\\
{\bf himself}: {\blacksquare}{\forall}f((({\langle\rangle}Nt(s(m))\backslash Sf){{}{\uparrow}{}}Nt(s(m))){{}{\downarrow}{}}({\langle\rangle}Nt(s(m))\backslash Sf)): \lambda A\lambda B(({\it A}\ {\it B})\ {\it B})\\
{\bf horse}: {\square}{\it CN}{\it s(n)}: {\it horse}\\
{\bf in}: {\square}({\forall}a{\forall}f(({\langle\rangle}Na\backslash Sf)\backslash ({\langle\rangle}Na\backslash Sf))/{\exists}aNa): \mbox{\^{}}\lambda A\lambda B\lambda C((\mbox{\v{}}{\it in}\ {\it A})\ ({\it B}\ {\it C}))\\
{\bf in}: {\square}({\forall}f(Sf{\div}Sf)/{\exists}aNa): {\it in}\\
{\bf is}: {\blacksquare}(({\langle\rangle}{\exists}gNt(s(g))\backslash Sf)/({\exists}aNa{\oplus}({\exists}g(({\it CN}{\it g}/{\it CN}{\it g}){\sqcup}({\it CN}{\it g}\backslash {\it CN}{\it g})){-}I))): \lambda A\lambda B({\it Pres}\ ({\it A}\casearrow C.[{\it B}={\it C}]; D.(({\it D}\ \lambda E[{\it E}={\it B}])\ {\it B})))\\
{\bf it}: {\blacksquare}W[it]: {\it 0}\\
{\bf it}: {\blacksquare}{\forall}f{\forall}a((({\langle\rangle}Na\backslash Sf){{}{\uparrow}{}}{\blacksquare}Nt(s(n))){{}{\downarrow}{}}({\blacksquare}({\langle\rangle}Na\backslash Sf){|}{\blacksquare}Nt(s(n)))): \lambda A{\it A}\\
{\bf it}: {\blacksquare}{[]^{-1}}{\forall}f(({\blacksquare}Sf{|}{\blacksquare}Nt(s(n)))/({\langle\rangle}Nt(s(n))\backslash Sf)): \lambda A{\it A}\\
{\bf jogs}: {\square}({\langle\rangle}{\exists}gNt(s(g))\backslash Sf): \mbox{\^{}}\lambda A({\it Pres}\ (\mbox{\v{}}{\it jog}\ {\it A}))\\
{\bf john}: {\blacksquare}Nt(s(m)): {\it j}\\
{\bf laughs}: {\square}({\langle\rangle}{\exists}gNt(s(g))\backslash Sf): \mbox{\^{}}\lambda A({\it Pres}\ (\mbox{\v{}}{\it laugh}\ {\it A}))\\
{\bf left}: {\square}({\langle\rangle}{\exists}gNt(s(g))\backslash Sf): \mbox{\^{}}\lambda A({\it Pres}\ (\mbox{\v{}}{\it leave}\ {\it A}))\\
{\bf let}: {\square}(Sim/Sb): {\it let}\\
{\bf light}: {\square}({\it CN}{\it s(n)}{\&}Nt(s(n))): \mbox{\^{}}(\mbox{\v{}}{\it light}, ({\it gen}\ \mbox{\v{}}{\it light}))\\
{\bf likes}: {\square}(({\langle\rangle}{\exists}gNt(s(g))\backslash Sf)/{\exists}aNa): \mbox{\^{}}\lambda A\lambda B({\it Pres}\ ((\mbox{\v{}}{\it like}\ {\it A})\ {\it B}))\\
{\bf logic}: {\square}(Nt(s(n)){\&}{\it CN}{\it s(n)}): \mbox{\^{}}(({\it gen}\ \mbox{\v{}}{\it logic}), \mbox{\v{}}{\it logic})\\
{\bf london}: {\blacksquare}Nt(s(n)): {\it l}\\
{\bf loses}: {\square}(({\langle\rangle}{\exists}gNt(s(g))\backslash Sf)/{\exists}aNa): \mbox{\^{}}\lambda A\lambda B({\it Pres}\ ((\mbox{\v{}}{\it lose}\ {\it A})\ {\it B}))\\
{\bf love}: {\square}(({\langle\rangle}{\exists}aNa\backslash Sb)/{\exists}aNa): \mbox{\^{}}\lambda A\lambda B((\mbox{\v{}}{\it love}\ {\it A})\ {\it B})\\
{\bf loved}: {\square}{\forall}a{\forall}b((({\langle\rangle}Na\backslash S-){{}{\uparrow}{}}Nb){{}{\odot}{}}((({\langle\rangle}Na\backslash S-){{}{\uparrow}{}}Nb){{}{\downarrow}{}}{\forall}g({\it CN}{\it g}\backslash {\it CN}{\it g}))): \mbox{\^{}}(\mbox{\v{}}{\it love}, \lambda A\lambda B\lambda C[({\it B}\ {\it C})\wedge \exists D(({\it A}\ {\it C})\ {\it D})])\\
{\bf loves}: {\square}(({\langle\rangle}{\exists}gNt(s(g))\backslash Sf)/{\exists}aNa): \mbox{\^{}}\lambda A\lambda B({\it Pres}\ ((\mbox{\v{}}{\it love}\ {\it A})\ {\it B}))\\
{\bf man}: {\square}{\it CN}{\it s(m)}: {\it man}\\
{\bf mary}: {\blacksquare}Nt(s(f)): {\it m}\\
{\bf met}: {\square}(({\langle\rangle}{\exists}aNa\backslash Sf)/{\exists}aNa): \mbox{\^{}}\lambda A\lambda B({\it Past}\ ((\mbox{\v{}}{\it meet}\ {\it A})\ {\it B}))\\
{\bf more}: {\blacksquare}{\forall}h{\forall}g{\forall}f((Sf{{}{\uparrow}{}}(((Sh{{}{\uparrow}{}}Nt(p(g))){{}{\downarrow}{}}Sh)/{\it CN}{\it p(g)})){{}{\downarrow}{}}(Sf/\mbox{$\hat{\ }$}({\it CP}than{{}{\uparrow}{}}{\blacksquare}(((Sh{{}{\uparrow}{}}Nt(p(g))){{}{\downarrow}{}}Sh)/{\it CN}{\it p(g)})))): \lambda A\lambda B[|\lambda C({\it A}\ \lambda D\lambda E[({\it D}\ {\it C})\wedge ({\it E}\ {\it C})])|>|\lambda F\mbox{\v{}}({\it B}\ \lambda G\lambda H[({\it G}\ {\it F})\wedge ({\it H}\ {\it F})])|]\\
{\bf mountain}: {\square}{\it CN}{\it s(n)}: {\it mountain}\\
{\bf moved}: {\square}({\langle\rangle}{\exists}aNa\backslash Sf): \mbox{\^{}}\lambda A({\it Past}\ (\mbox{\v{}}{\it move}\ {\it A}))\\
{\bf necessarily}: {\blacksquare}(SA/{\square}SA): {\it Nec}\\
{\bf of}: {\square}(({\forall}n({\it CN}{\it n}\backslash {\it CN}{\it n})/{\blacksquare}{\exists}bNb){\&}({\it PP}{\it of}/{\exists}aNa)): \mbox{\^{}}(\mbox{\v{}}{\it of}, \lambda A{\it A})\\
{\bf or}: {\blacksquare}{\forall}f((?{\blacksquare}Sf\backslash {[]^{-1}}{[]^{-1}}Sf)/{\blacksquare}Sf): (\Phinplus{}\ {\it 0}\ {\it or})\\
{\bf or}: {\blacksquare}{\forall}a{\forall}f((?{\blacksquare}({\langle\rangle}Na\backslash Sf)\backslash {[]^{-1}}{[]^{-1}}({\langle\rangle}Na\backslash Sf))/{\blacksquare}({\langle\rangle}Na\backslash Sf)): (\Phinplus{}\ ({\it s}\ {\it 0})\ {\it or})\\
{\bf or}: {\blacksquare}{\forall}f((?{\blacksquare}(Sf/({\langle\rangle}{\exists}gNt(s(g))\backslash Sf))\backslash {[]^{-1}}{[]^{-1}}(Sf/({\langle\rangle}{\exists}gNt(s(g))\backslash Sf)))/{\blacksquare}(Sf/({\langle\rangle}{\exists}gNt(s(g))\backslash Sf))): (\Phinplus{}\ ({\it s}\ {\it 0})\ {\it or})\\
{\bf or}: {\blacksquare}{\forall}a{\forall}f((?{\blacksquare}((({\langle\rangle}Na\backslash Sf)/{\exists}bNb)/{\exists}bNb)\backslash {[]^{-1}}{[]^{-1}}((({\langle\rangle}Na\backslash Sf)/{\exists}bNb)/{\exists}bNb))/{\blacksquare}((({\langle\rangle}Na\backslash Sf)/{\exists}bNb)/{\exists}bNb)): (\Phinplus{}\ ({\it s}\ ({\it s}\ ({\it s}\ {\it 0})))\ {\it or})\\
{\bf painting}: {\square}({\it CN}{\it s(n)}/{\it PP}{\it of}): \mbox{\^{}}\lambda A((\mbox{\v{}}{\it of}\ {\it A})\ \mbox{\v{}}{\it painting})\\
{\bf paper}: {\square}{\it CN}{\it s(n)}: {\it paper}\\
{\bf park}: {\square}{\it CN}{\it s(n)}: {\it park}\\
{\bf past}: {\square}{\forall}a{\forall}f((({\langle\rangle}Na\backslash Sf)\backslash ({\langle\rangle}Na\backslash Sf))/{\exists}bNb): \mbox{\^{}}\lambda A\lambda B\lambda C((\mbox{\v{}}{\it past}\ {\it A})\ ({\it B}\ {\it C}))\\
{\bf perseverance}: {\square}(Nt(s(n)){\&}{\it CN}{\it s(n)}): \mbox{\^{}}(({\it gen}\ \mbox{\v{}}{\it perseverance}), \mbox{\v{}}{\it perseverance})\\
{\bf peter}: {\blacksquare}Nt(s(m)): {\it p}\\
{\bf phonetics}: {\square}(Nt(s(n)){\&}{\it CN}{\it s(n)}): \mbox{\^{}}(({\it gen}\ \mbox{\v{}}{\it phonetics}), \mbox{\v{}}{\it phonetics})\\
{\bf praises}: {\square}(({\langle\rangle}{\exists}gNt(s(g))\backslash Sf)/{\exists}aNa): \mbox{\^{}}\lambda A\lambda B({\it Pres}\ ((\mbox{\v{}}{\it praise}\ {\it A})\ {\it B}))\\
{\bf raced}: {\square}({\langle\rangle}{\exists}aNa\backslash Sf): \mbox{\^{}}\lambda A({\it Past}\ (\mbox{\v{}}{\it race}\ {\it A}))\\
{\bf raced}: {\square}{\forall}a{\forall}b((({\langle\rangle}Na\backslash S-){{}{\uparrow}{}}Nb){{}{\odot}{}}((({\langle\rangle}Na\backslash S-){{}{\uparrow}{}}Nb){{}{\downarrow}{}}{\forall}g({\it CN}{\it g}\backslash {\it CN}{\it g}))): \mbox{\^{}}(\mbox{\v{}}{\it race2}, \lambda A\lambda B\lambda C[({\it B}\ {\it C})\wedge \exists D(({\it A}\ {\it C})\ {\it D})])\\
{\bf rains}: {\square}({\langle\rangle}W[it]{\multimap}Sf): \mbox{\^{}}({\it Pres}\ \mbox{\v{}}{\it itrains})\\
{\bf reading}: {\square}(({\langle\rangle}{\exists}aNa\backslash Spsp)/{\exists}aNa): \mbox{\^{}}\lambda A\lambda B((\mbox{\v{}}{\it read}\ {\it A})\ {\it B})\\
{\bf robin}: {\blacksquare}{\forall}gNt(s(g)): {\it r}\\
{\bf said}: {\square}(({\langle\rangle}{\exists}aNa\backslash Sf)/Sim): \mbox{\^{}}\lambda A\lambda B({\it Past}\ ((\mbox{\v{}}{\it say}\ {\it A})\ {\it B}))\\
{\bf saw}: {\square}(({\langle\rangle}{\exists}aNa\backslash Sf)/({\exists}aNa{\oplus}{\it CP}that)): \mbox{\^{}}\lambda A\lambda B({\it Past}\ (({\it A}\casearrow C.(\mbox{\v{}}{\it seee}\ {\it C}); D.(\mbox{\v{}}{\it seet}\ {\it D}))\ {\it B}))\\
{\bf seeks}: {\square}(({\langle\rangle}{\exists}gNt(s(g))\backslash Sf)/{\square}{\forall}a{\forall}f(((Na\backslash Sf)/{\exists}bNb)\backslash (Na\backslash Sf))): \mbox{\^{}}\lambda A\lambda B((\mbox{\v{}}{\it tries}\ \mbox{\^{}}((\mbox{\v{}}{\it A}\ \mbox{\v{}}{\it find})\ {\it B}))\ {\it B})\\
{\bf sees}: {\square}(({\langle\rangle}{\exists}gNt(s(g))\backslash Sf)/{\exists}aNa): \mbox{\^{}}\lambda A\lambda B({\it Pres}\ ((\mbox{\v{}}{\it see}\ {\it A})\ {\it B}))\\
{\bf sent}: {\square}(({\langle\rangle}{\exists}aNa\backslash Sf)/({\exists}bNb{\bullet}{\it PP}{\it to})): \mbox{\^{}}\lambda A\lambda B({\it Past}\ (((\mbox{\v{}}{\it sent}\ \pi_2{\it A})\ \pi_1{\it A})\ {\it B}))\\
{\bf sent}: {\square}((({\langle\rangle}{\exists}aNa\backslash Sf)/{\exists}aNa)/{\exists}aNa): \mbox{\^{}}\lambda A\lambda B\lambda C({\it Past}\ (((\mbox{\v{}}{\it send}\ {\it A})\ {\it B})\ {\it C}))\\
{\bf she}: {\blacksquare}{[]^{-1}}{\forall}g(({\blacksquare}Sg{|}{\blacksquare}Nt(s(f)))/({\langle\rangle}Nt(s(f))\backslash Sg)): \lambda A{\it A}\\
{\bf sings}: {\square}({\langle\rangle}{\exists}gNt(s(g))\backslash Sf): \mbox{\^{}}\lambda A({\it Pres}\ (\mbox{\v{}}{\it sing}\ {\it A}))\\
{\bf slept}: {\square}({\langle\rangle}{\exists}gNt(s(g))\backslash Sf): \mbox{\^{}}\lambda A({\it Past}\ (\mbox{\v{}}{\it sleep}\ {\it A}))\\
{\bf slowly}: {\square}{\forall}a{\forall}f({\square}({\langle\rangle}Na\backslash Sf)\backslash ({\langle\rangle}{\square}Na\backslash Sf)): \mbox{\^{}}\lambda A\lambda B(\mbox{\v{}}{\it slowly}\ \mbox{\^{}}(\mbox{\v{}}{\it A}\ \mbox{\v{}}{\it B}))\\
{\bf sneezed}: {\square}({\langle\rangle}{\exists}gNt(s(g))\backslash Sf): \mbox{\^{}}\lambda A({\it Past}\ (\mbox{\v{}}{\it sneeze}\ {\it A}))\\
{\bf sold}: {\square}(({\langle\rangle}{\exists}aNa\backslash Sf)/({\exists}bNb{\bullet}{\it PP}{\it for})): \mbox{\^{}}\lambda A\lambda B({\it Past}\ (((\mbox{\v{}}{\it sell}\ \pi_2{\it A})\ \pi_1{\it A})\ {\it B}))\\
{\bf someone}: {\square}{\forall}f((Sf{{}{\uparrow}{}}{\blacksquare}{\forall}gNt(g)){{}{\downarrow}{}}Sf): \mbox{\^{}}\lambda A\exists B[(\mbox{\v{}}{\it person}\ {\it B})\wedge ({\it A}\ {\it B})]\\
{\bf Spirit}: {\square}{\it CN}{\it s(m)}: {\it Spirit}\\
{\bf studies}: {\square}(({\langle\rangle}{\exists}gNt(s(g))\backslash Sf)/{\exists}aNa): \mbox{\^{}}\lambda A\lambda B({\it Pres}\ ((\mbox{\v{}}{\it study}\ {\it A})\ {\it B}))\\
{\bf such}{+}{\bf that}: {\blacksquare}{\forall}n(({\it CN}{\it n}\backslash {\it CN}{\it n})/(Sf{|}{\blacksquare}Nt(n))): \lambda A\lambda B\lambda C[({\it B}\ {\it C})\wedge ({\it A}\ {\it C})]\\
{\bf suzy}: {\blacksquare}Nt(s(f)): {\it s}\\
{\bf talks}: {\square}({\langle\rangle}{\exists}gNt(s(g))\backslash Sf): \mbox{\^{}}\lambda A({\it Pres}\ (\mbox{\v{}}{\it talk}\ {\it A}))\\
{\bf tall}: {\square}{\forall}g({\it CN}{\it g}/{\it CN}{\it g}): {\it tall}\\
{\bf teetotal}: {\square}{\forall}n({\it CN}{\it n}/{\it CN}{\it n}): \mbox{\^{}}\lambda A\lambda B[({\it A}\ {\it B})\wedge (\mbox{\v{}}{\it teetotal}\ {\it B})]\\
{\bf tenmilliondollars}: {\square}Nt(s(n)): {\it tenmilliondollars}\\
{\bf than}: {\blacksquare}({\it CP}than/{\square}Sf): \lambda A{\it A}\\
{\bf that}: {\blacksquare}({\it CP}that/{\square}Sf): \lambda A{\it A}\\
{\bf that}: {\blacksquare}{\forall}n({[]^{-1}}{[]^{-1}}({\it CN}{\it n}\backslash {\it CN}{\it n})/{\blacksquare}(({\langle\rangle}Nt(n){\sqcap}!{\blacksquare}Nt(n))\backslash Sf)): \lambda A\lambda B\lambda C[({\it B}\ {\it C})\wedge ({\it A}\ {\it C})]\\
{\bf the}: {\blacksquare}{\forall}n(Nt(n)/{\it CN}{\it n}): \iota \\
{\bf the}{+}{\bf cold}{+}{\bf shoulder}: {\blacksquare}W[the,cold,shoulder]: {\it 0}\\
{\bf there}: {\blacksquare}W[there]: {\it 0}\\
{\bf thinks}: {\square}(({\langle\rangle}{\exists}gNt(s(g))\backslash Sf)/({\it CP}that{\sqcup}{\square}Sf)): \mbox{\^{}}\lambda A\lambda B({\it Pres}\ ((\mbox{\v{}}{\it think}\ {\it A})\ {\it B}))\\
{\bf to}: {\blacksquare}(({\it PP}{\it to}/{\exists}aNa){\sqcap}{\forall}n(({\langle\rangle}Nn\backslash Si)/({\langle\rangle}Nn\backslash Sb))): \lambda A{\it A}\\
{\bf today}: {\square}{\forall}a{\forall}f(({\langle\rangle}Na\backslash Sf)\backslash ({\langle\rangle}Na\backslash Sf)): \mbox{\^{}}\lambda A\lambda B(\mbox{\v{}}{\it today}\ ({\it A}\ {\it B}))\\
{\bf tries}: {\square}(({\langle\rangle}{\exists}gNt(s(g))\backslash Sf)/{\square}({\langle\rangle}{\exists}gNt(s(g))\backslash Si)): \mbox{\^{}}\lambda A\lambda B((\mbox{\v{}}{\it tries}\ \mbox{\^{}}(\mbox{\v{}}{\it A}\ {\it B}))\ {\it B})\\
{\bf unicorn}: {\square}{\it CN}{\it s(n)}: {\it unicorn}\\
{\bf up}: {\blacksquare}W[up]: {\it 0}\\
{\bf upon}: {\square}(({\forall}b{\forall}f(({\langle\rangle}Nb\backslash Sf)\backslash ({\langle\rangle}Nb\backslash Sf)){\&}{\forall}g({\it CN}{\it g}\backslash {\it CN}{\it g}))/{\exists}aNa): \mbox{\^{}}\lambda A((\mbox{\v{}}{\it uponadv}\ {\it A}), (\mbox{\v{}}{\it uponadn}\ {\it A}))\\
{\bf void}: {\square}{\forall}g({\it CN}{\it g}/{\it CN}{\it g}): {\it void}\\
{\bf walk}: {\square}({\langle\rangle}({\exists}aNa{-}{\exists}gNt(s(g)))\backslash Sf): \mbox{\^{}}\lambda A({\it Pres}\ (\mbox{\v{}}{\it walk}\ {\it A}))\\
{\bf walk}: {\square}({\langle\rangle}{\exists}aNa\backslash Sb): \mbox{\^{}}\lambda A(\mbox{\v{}}{\it walk}\ {\it A})\\
{\bf walks}: {\square}({\langle\rangle}{\exists}gNt(s(g))\backslash Sf): \mbox{\^{}}\lambda A({\it Pres}\ (\mbox{\v{}}{\it walk}\ {\it A}))\\
{\bf was}: {\blacksquare}(({\langle\rangle}{\exists}gNt(s(g))\backslash Sf)/({\exists}aNa{\oplus}({\exists}g(({\it CN}{\it g}/{\it CN}{\it g}){\sqcup}({\it CN}{\it g}\backslash {\it CN}{\it g})){-}I))): \lambda A\lambda B({\it Past}\ ({\it A}\casearrow C.[{\it B}={\it C}]; D.(({\it D}\ \lambda E[{\it E}={\it B}])\ {\it B})))\\
{\bf was}: {\square}(({\langle\rangle}W[there]{\multimap}Sf)/{\exists}aNa): \mbox{\^{}}\lambda A({\it Past}\ (\mbox{\v{}}{\it be}\ {\it A}))\\
{\bf waters}: {\square}{\it CN}{\it p(n)}: {\it waters}\\
{\bf which}: {\blacksquare}{\forall}n{\forall}m((Nt(n){{}{\uparrow}{}}Nt(m)){{}{\downarrow}{}}({[]^{-1}}{[]^{-1}}({\it CN}{\it m}\backslash {\it CN}{\it m})/{\blacksquare}(({\langle\rangle}Nt(n){\sqcap}!{\blacksquare}Nt(n))\backslash Sf))): \lambda A\lambda B\lambda C\lambda D[({\it C}\ {\it D})\wedge ({\it B}\ ({\it A}\ {\it D}))]\\
{\bf who}: {\blacksquare}{\forall}h{\forall}n({[]^{-1}}{[]^{-1}}(Nt(n)\backslash ((Sh{{}{\uparrow}{}}Nt(n)){{}{\downarrow}{}}Sh))/{\blacksquare}(({\langle\rangle}Nt(n){\sqcap}!{\blacksquare}Nt(n))\backslash Sf)): \lambda A\lambda B\lambda C[({\it A}\ {\it B})\wedge ({\it C}\ {\it B})]\\
{\bf will}: {\blacksquare}{\forall}a(({\langle\rangle}Na\backslash Sf)/({\langle\rangle}Na\backslash Sb)): \lambda A\lambda B({\it Fut}\ ({\it A}\ {\it B}))\\
{\bf without}: {\square}({\forall}g({\it CN}{\it g}\backslash {\it CN}{\it g})/{\exists}aNa): \mbox{\^{}}\lambda A\lambda B\lambda C[({\it B}\ {\it C})\wedge \neg ((\mbox{\v{}}{\it with}\ {\it A})\ {\it C})]\\
{\bf without}: {\blacksquare}{\forall}a{\forall}f({[]^{-1}}(({\langle\rangle}Na\backslash Sf)\backslash ({\langle\rangle}Na\backslash Sf))/({\langle\rangle}Na\backslash Spsp)): \lambda A\lambda B\lambda C[({\it B}\ {\it C})\wedge \neg ({\it A}\ {\it C})]\\
{\bf woman}: {\square}{\it CN}{\it s(f)}: {\it woman}\\
{\bf yesterday}: {\square}{\forall}a{\forall}f(({\langle\rangle}Na\backslash Sf)\backslash ({\langle\rangle}Na\backslash Sf)): \mbox{\^{}}\lambda A\lambda B(\mbox{\v{}}{\it yesterday}\ ({\it A}\ {\it B}))\\$

\chapter{Initial Examples} 

\label{initexchap}

The derivations we give have been computer-generated from
the lexicon given in Chapter~\ref{lexchap} and a parser for the categorial logic.
The implementation is a categorial parser/theorem-prover CatLog2 comprising
6000 lines of Prolog using backward chaining proof-search in 
the Gentzen sequent calculus (Morrill 2011\cite{morrill:logprodisp}), and
the focusing of Andreoli (1992\cite{andreoli:92}), see Chapter~\ref{focchap};
in addition to focusing, the implementation exploits count-invariance 
(van Benthem 1991\cite{benthem:action}; Valent\'{\i}n, Serret and Morrill (2013\cite{vsm:add}),
see Chapter~\ref{countchap}.
In focusing, proofs are built in alternating phases of don't care non-deterministic
invertible/asynchronous rule application and focused noninvertible/synchronous rule application.
The boxes in our derivations mark the focused types, which are the
active types of synchronous rule application.
All the reader needs to have in mind is that a boxed type in the conclusion
of an inference step is always the active type of that inference step.
The first example is as follows:\footnote{Note how in the input
to CatLog brackets mark islands: single brackets for weak islands such
as subjects and double brackets for
strong islands such as relative clauses and coordinate structures,
Morrill (2011\citep{morrill:oxford}, Chapter~5).
Morrill, Kuznetsov, Kanovich and Scedrov (2018\cite{brackind:2018}) considers induction of such brackets.}
\disp{
${}[{\bf john}]{+}{\bf walks}: Sf$
\label{ex12}}
Note that in our syntactical form the subject is a bracketed domain,
and this will always be the case --- implementing that subjects are weak
islands.
Lookup in our lexicon yields the following semantically labelled sequent:
\disp{
$[{\blacksquare}Nt(s(m)): {\it j}], {\square}({\langle\rangle}{\exists}gNt(s(g))\backslash Sf): \mbox{\^{}}\lambda A({\it Pres}\ (\mbox{\v{}}{\it walk}\ {\it A}))\ \Rightarrow\ Sf$}
The lexical types are semantically modalised outermost, and this will always be
the case --- implementing that word meanings are intensions/senses;
the modality of the proper name subject is semantically inactive
(we take proper names to be rigid designators), while the modality of the tensed verb is
semantically active (the interpretation of tensed verbs depends on the temporal
reference points).
The verb projects a finite sentence (feature $f$) when it combines with a third person
singular (bracketed) subject of any gender (the existential quantification);
the actual subject is masculine (feature $m$).

The derivation is as follows:

\disp{
\noindent{\footnotesize
\prooftree
\prooftree
\prooftree
\prooftree
\prooftree
\prooftree
\justifies
\mbox{\fbox{$Nt(s(m))$}}\ \Rightarrow\ Nt(s(m))
\endprooftree
\justifies
\mbox{\fbox{${\blacksquare}Nt(s(m))$}}\ \Rightarrow\ Nt(s(m))
\using {\blacksquare}L
\endprooftree
\justifies
{\blacksquare}Nt(s(m))\ \Rightarrow\ \fbox{${\exists}gNt(s(g))$}
\using {\exists}R
\endprooftree
\justifies
[{\blacksquare}Nt(s(m))]\ \Rightarrow\ \fbox{${\langle\rangle}{\exists}gNt(s(g))$}
\using {\langle\rangle}R
\endprooftree
\prooftree
\justifies
\mbox{\fbox{$Sf$}}\ \Rightarrow\ Sf
\endprooftree
\justifies
[{\blacksquare}Nt(s(m))], \mbox{\fbox{${\langle\rangle}{\exists}gNt(s(g))\backslash Sf$}}\ \Rightarrow\ Sf
\using {\backslash}L
\endprooftree
\justifies
[{\blacksquare}Nt(s(m))], \mbox{\fbox{${\square}({\langle\rangle}{\exists}gNt(s(g))\backslash Sf)$}}\ \Rightarrow\ Sf
\using {\Box}L
\endprooftree}
}

\noindent
The flow of information in the semantic reading of derivations can be illustrated for
the case in hand as follows; note that in practice the steps of this information
flow are implemented by unification stepwise with derivation.
First, variables for the antecedent semantics are added in the endsequent:
\disp{
$
[{\blacksquare}Nt(s(m)): x], {\square}({\langle\rangle}{\exists}gNt(s(g))\backslash Sf): y\ \Rightarrow\ Sf
$}
Reading bottom-up,
at the lowest inference step (\mymod{}L)
the verb semantics is replaced by the extension $z$ and the subject semantics $x$ is
carried over:
\disp{
\prooftree
[{\blacksquare}Nt(s(m)): x], \mbox{\fbox{${\langle\rangle}{\exists}gNt(s(g))\backslash Sf$}}: 
z\ \Rightarrow\ Sf
\justifies
[{\blacksquare}Nt(s(m)): x], {\square}({\langle\rangle}{\exists}gNt(s(g))\backslash Sf): y\ \Rightarrow\ Sf
\using {\Box}L
\endprooftree}
At the second inference we propagate the subject semantics on the argument branch:
\disp{
\prooftree
\prooftree
[{\blacksquare}Nt(s(m)): x]\ \Rightarrow\ \fbox{${\langle\rangle}{\exists}gNt(s(g))$}\tb
\mbox{\fbox{$Sf$}}\ \Rightarrow\ Sf
\justifies
[{\blacksquare}Nt(s(m)): x], \mbox{\fbox{${\langle\rangle}{\exists}gNt(s(g))\backslash Sf$}}: z\ \Rightarrow\ Sf
\using {\backslash}L
\endprooftree
\justifies
[{\blacksquare}Nt(s(m)): x], {\square}({\langle\rangle}{\exists}gNt(s(g))\backslash Sf): y\ \Rightarrow\ Sf
\using {\Box}L
\endprooftree}
The next three inferences involve semantically transparent copying of the antecedent semantics:
\commentout{

\vspace{0.15in}

\noindent{\footnotesize
\prooftree
\prooftree
\prooftree
\prooftree
\prooftree
\prooftree
\justifies
\mbox{\fbox{$Nt(s(m))$}}\ \Rightarrow\ Nt(s(m))
\endprooftree
\justifies
\mbox{\fbox{${\blacksquare}Nt(s(m))$}}\ \Rightarrow\ Nt(s(m))
\using {\blacksquare}L
\endprooftree
\justifies
{\blacksquare}Nt(s(m)): x\ \Rightarrow\ \fbox{${\exists}gNt(s(g))$}
\using {\exists}R
\endprooftree
\justifies
[{\blacksquare}Nt(s(m)): x]\ \Rightarrow\ \fbox{${\langle\rangle}{\exists}gNt(s(g))$}
\using {\langle\rangle}R
\endprooftree
\prooftree
\justifies
\mbox{\fbox{$Sf$}}\ \Rightarrow\ Sf
\endprooftree
\justifies
[{\blacksquare}Nt(s(m)): x], \mbox{\fbox{${\langle\rangle}{\exists}gNt(s(g))\backslash Sf$}}: 
z\ \Rightarrow\ Sf
\using {\backslash}L
\endprooftree
\justifies
[{\blacksquare}Nt(s(m)): x], {\square}({\langle\rangle}{\exists}gNt(s(g))\backslash Sf): y\ \Rightarrow\ Sf
\using {\Box}L
\endprooftree}

\vspace{0.15in}

\noindent{\footnotesize
\prooftree
\prooftree
\prooftree
\prooftree
\prooftree
\prooftree
\justifies
\mbox{\fbox{$Nt(s(m))$}}\ \Rightarrow\ Nt(s(m))
\endprooftree
\justifies
\mbox{\fbox{${\blacksquare}Nt(s(m)): x$}}\ \Rightarrow\ Nt(s(m))
\using {\blacksquare}L
\endprooftree
\justifies
{\blacksquare}Nt(s(m)): x\ \Rightarrow\ \fbox{${\exists}gNt(s(g))$}
\using {\exists}R
\endprooftree
\justifies
[{\blacksquare}Nt(s(m)): x]\ \Rightarrow\ \fbox{${\langle\rangle}{\exists}gNt(s(g))$}
\using {\langle\rangle}R
\endprooftree
\prooftree
\justifies
\mbox{\fbox{$Sf$}}\ \Rightarrow\ Sf
\endprooftree
\justifies
[{\blacksquare}Nt(s(m)): x], \mbox{\fbox{${\langle\rangle}{\exists}gNt(s(g))\backslash Sf$}}: 
z\ \Rightarrow\ Sf
\using {\backslash}L
\endprooftree
\justifies
[{\blacksquare}Nt(s(m)): x], {\square}({\langle\rangle}{\exists}gNt(s(g))\backslash Sf): y\ \Rightarrow\ Sf
\using {\Box}L
\endprooftree}

\vspace{0.15in}

}
\disp{
\prooftree
\prooftree
\prooftree
\prooftree
\prooftree
\prooftree
\justifies
\mbox{\fbox{$Nt(s(m))$}: x}\ \Rightarrow\ Nt(s(m))
\endprooftree
\justifies
\mbox{\fbox{${\blacksquare}Nt(s(m)): x$}}\ \Rightarrow\ Nt(s(m))
\using {\blacksquare}L
\endprooftree
\justifies
{\blacksquare}Nt(s(m)): x\ \Rightarrow\ \fbox{${\exists}gNt(s(g))$}
\using {\exists}R
\endprooftree
\justifies
[{\blacksquare}Nt(s(m)): x]\ \Rightarrow\ \fbox{${\langle\rangle}{\exists}gNt(s(g))$}
\using {\langle\rangle}R
\endprooftree
\prooftree
\justifies
\mbox{\fbox{$Sf$}}\ \Rightarrow\ Sf
\endprooftree
\justifies
[{\blacksquare}Nt(s(m)): x], \mbox{\fbox{${\langle\rangle}{\exists}gNt(s(g))\backslash Sf$}}: 
z\ \Rightarrow\ Sf
\using {\backslash}L
\endprooftree
\justifies
[{\blacksquare}Nt(s(m)): x], {\square}({\langle\rangle}{\exists}gNt(s(g))\backslash Sf): y\ \Rightarrow\ Sf
\using {\Box}L
\endprooftree}
At the identity axiom the antecedent semantics is copied to the succedent:
\disp{
\prooftree
\prooftree
\prooftree
\prooftree
\prooftree
\prooftree
\justifies
\mbox{\fbox{$Nt(s(m)): x$}}\ \Rightarrow\ Nt(s(m)): x
\endprooftree
\justifies
\mbox{\fbox{${\blacksquare}Nt(s(m)): x$}}\ \Rightarrow\ Nt(s(m))
\using {\blacksquare}L
\endprooftree
\justifies
{\blacksquare}Nt(s(m)): x\ \Rightarrow\ \fbox{${\exists}gNt(s(g))$}
\using {\exists}R
\endprooftree
\justifies
[{\blacksquare}Nt(s(m)): x]\ \Rightarrow\ \fbox{${\langle\rangle}{\exists}gNt(s(g))$}
\using {\langle\rangle}R
\endprooftree
\prooftree
\justifies
\mbox{\fbox{$Sf$}}\ \Rightarrow\ Sf
\endprooftree
\justifies
[{\blacksquare}Nt(s(m)): x], \mbox{\fbox{${\langle\rangle}{\exists}gNt(s(g))\backslash Sf$}}: 
z\ \Rightarrow\ Sf
\using {\backslash}L
\endprooftree
\justifies
[{\blacksquare}Nt(s(m)): x], {\square}({\langle\rangle}{\exists}gNt(s(g))\backslash Sf): y\ \Rightarrow\ Sf
\using {\Box}L
\endprooftree}
In a following phase the succedent semantics is copied from premises to
conclusions as far as the root of the argument branch:
\commentout{

\vspace{0.15in}

\noindent{\footnotesize
\prooftree
\prooftree
\prooftree
\prooftree
\prooftree
\prooftree
\justifies
\mbox{\fbox{$Nt(s(m)): x$}}\ \Rightarrow\ Nt(s(m)): x
\endprooftree
\justifies
\mbox{\fbox{${\blacksquare}Nt(s(m)): x$}}\ \Rightarrow\ Nt(s(m)): x
\using {\blacksquare}L
\endprooftree
\justifies
{\blacksquare}Nt(s(m)): x\ \Rightarrow\ \fbox{${\exists}gNt(s(g))$}
\using {\exists}R
\endprooftree
\justifies
[{\blacksquare}Nt(s(m)): x]\ \Rightarrow\ \fbox{${\langle\rangle}{\exists}gNt(s(g))$}
\using {\langle\rangle}R
\endprooftree
\prooftree
\justifies
\mbox{\fbox{$Sf$}}\ \Rightarrow\ Sf
\endprooftree
\justifies
[{\blacksquare}Nt(s(m)): x], \mbox{\fbox{${\langle\rangle}{\exists}gNt(s(g))\backslash Sf$}}: 
z\ \Rightarrow\ Sf
\using {\backslash}L
\endprooftree
\justifies
[{\blacksquare}Nt(s(m)): x], {\square}({\langle\rangle}{\exists}gNt(s(g))\backslash Sf): y\ \Rightarrow\ Sf
\using {\Box}L
\endprooftree}

\vspace{0.15in}

\noindent{\footnotesize
\prooftree
\prooftree
\prooftree
\prooftree
\prooftree
\prooftree
\justifies
\mbox{\fbox{$Nt(s(m)): x$}}\ \Rightarrow\ Nt(s(m)): x
\endprooftree
\justifies
\mbox{\fbox{${\blacksquare}Nt(s(m)): x$}}\ \Rightarrow\ Nt(s(m)): x
\using {\blacksquare}L
\endprooftree
\justifies
{\blacksquare}Nt(s(m)): x\ \Rightarrow\ \fbox{${\exists}gNt(s(g))$}: x
\using {\exists}R
\endprooftree
\justifies
[{\blacksquare}Nt(s(m)): x]\ \Rightarrow\ \fbox{${\langle\rangle}{\exists}gNt(s(g))$}
\using {\langle\rangle}R
\endprooftree
\prooftree
\justifies
\mbox{\fbox{$Sf$}}\ \Rightarrow\ Sf
\endprooftree
\justifies
[{\blacksquare}Nt(s(m)): x], \mbox{\fbox{${\langle\rangle}{\exists}gNt(s(g))\backslash Sf$}}: 
z\ \Rightarrow\ Sf
\using {\backslash}L
\endprooftree
\justifies
[{\blacksquare}Nt(s(m)): x], {\square}({\langle\rangle}{\exists}gNt(s(g))\backslash Sf): y\ \Rightarrow\ Sf
\using {\Box}L
\endprooftree}

}
\disp{
\prooftree
\prooftree
\prooftree
\prooftree
\prooftree
\prooftree
\justifies
\mbox{\fbox{$Nt(s(m)): x$}}\ \Rightarrow\ Nt(s(m)): x
\endprooftree
\justifies
\mbox{\fbox{${\blacksquare}Nt(s(m)): x$}}\ \Rightarrow\ Nt(s(m)): x
\using {\blacksquare}L
\endprooftree
\justifies
{\blacksquare}Nt(s(m)): x\ \Rightarrow\ \fbox{${\exists}gNt(s(g))$}: x
\using {\exists}R
\endprooftree
\justifies
[{\blacksquare}Nt(s(m)): x]\ \Rightarrow\ \fbox{${\langle\rangle}{\exists}gNt(s(g)): x$}
\using {\langle\rangle}R
\endprooftree
\prooftree
\justifies
\mbox{\fbox{$Sf$}}\ \Rightarrow\ Sf
\endprooftree
\justifies
[{\blacksquare}Nt(s(m)): x], \mbox{\fbox{${\langle\rangle}{\exists}gNt(s(g))\backslash Sf$}}: 
z\ \Rightarrow\ Sf
\using {\backslash}L
\endprooftree
\justifies
[{\blacksquare}Nt(s(m)): x], {\square}({\langle\rangle}{\exists}gNt(s(g))\backslash Sf): y\ \Rightarrow\ Sf
\using {\Box}L
\endprooftree}
Now the functor value semantics in the antecedent of the value branch is labelled
with a new variable $w$:
\disp{
\prooftree
\prooftree
\prooftree
\prooftree
\prooftree
\prooftree
\justifies
\mbox{\fbox{$Nt(s(m)): x$}}\ \Rightarrow\ Nt(s(m)): x
\endprooftree
\justifies
\mbox{\fbox{${\blacksquare}Nt(s(m)): x$}}\ \Rightarrow\ Nt(s(m)): x
\using {\blacksquare}L
\endprooftree
\justifies
{\blacksquare}Nt(s(m)): x\ \Rightarrow\ \fbox{${\exists}gNt(s(g))$}: x
\using {\exists}R
\endprooftree
\justifies
[{\blacksquare}Nt(s(m)): x]\ \Rightarrow\ \fbox{${\langle\rangle}{\exists}gNt(s(g)): x$}
\using {\langle\rangle}R
\endprooftree
\prooftree
\justifies
\mbox{\fbox{$Sf: w$}}\ \Rightarrow\ Sf
\endprooftree
\justifies
[{\blacksquare}Nt(s(m)): x], \mbox{\fbox{${\langle\rangle}{\exists}gNt(s(g))\backslash Sf$}}: 
z\ \Rightarrow\ Sf
\using {\backslash}L
\endprooftree
\justifies
[{\blacksquare}Nt(s(m)): x], {\square}({\langle\rangle}{\exists}gNt(s(g))\backslash Sf): y\ \Rightarrow\ Sf
\using {\Box}L
\endprooftree}
At the $\id$ axiom this semantics is copied from antecedent to succedent:
\disp{
\prooftree
\prooftree
\prooftree
\prooftree
\prooftree
\prooftree
\justifies
\mbox{\fbox{$Nt(s(m)): x$}}\ \Rightarrow\ Nt(s(m)): x
\endprooftree
\justifies
\mbox{\fbox{${\blacksquare}Nt(s(m)): {\it j}$}}\ \Rightarrow\ Nt(s(m)): {\it j}
\using {\blacksquare}L
\endprooftree
\justifies
{\blacksquare}Nt(s(m)): x\ \Rightarrow\ \fbox{${\exists}gNt(s(g))$}: x
\using {\exists}R
\endprooftree
\justifies
[{\blacksquare}Nt(s(m)): x]\ \Rightarrow\ \fbox{${\langle\rangle}{\exists}gNt(s(g)): x$}
\using {\langle\rangle}R
\endprooftree
\prooftree
\justifies
\mbox{\fbox{$Sf: w$}}\ \Rightarrow\ Sf: w
\endprooftree
\justifies
[{\blacksquare}Nt(s(m)): x], \mbox{\fbox{${\langle\rangle}{\exists}gNt(s(g))\backslash Sf$}}: z\ \Rightarrow\ Sf
\using {\backslash}L
\endprooftree
\justifies
[{\blacksquare}Nt(s(m)): x], {\square}({\langle\rangle}{\exists}gNt(s(g))\backslash Sf): y\ \Rightarrow\ Sf
\using {\Box}L
\endprooftree}
In the $\bsl L$ conclusion succedent the semantics of the major premise is subject to the
substitution of $w$ by the functional application of the functor $z$ to the argument $x$:
\disp{
\prooftree
\prooftree
\prooftree
\prooftree
\prooftree
\prooftree
\justifies
\mbox{\fbox{$Nt(s(m)): x$}}\ \Rightarrow\ Nt(s(m)): x
\endprooftree
\justifies
\mbox{\fbox{${\blacksquare}Nt(s(m)): x$}}\ \Rightarrow\ Nt(s(m)): x
\using {\blacksquare}L
\endprooftree
\justifies
{\blacksquare}Nt(s(m)): x\ \Rightarrow\ \fbox{${\exists}gNt(s(g))$}: x
\using {\exists}R
\endprooftree
\justifies
[{\blacksquare}Nt(s(m)): x]\ \Rightarrow\ \fbox{${\langle\rangle}{\exists}gNt(s(g)): x$}
\using {\langle\rangle}R
\endprooftree
\prooftree
\justifies
\mbox{\fbox{$Sf: w$}}\ \Rightarrow\ Sf: w
\endprooftree
\justifies
[{\blacksquare}Nt(s(m)): x], \mbox{\fbox{${\langle\rangle}{\exists}gNt(s(g))\backslash Sf$}}: 
z\ \Rightarrow\ Sf: w\{(z\ x)/w\} = (z\ x)
\using {\backslash}L
\endprooftree
\justifies
[{\blacksquare}Nt(s(m)): x], {\square}({\langle\rangle}{\exists}gNt(s(g))\backslash Sf): y\ \Rightarrow\ Sf
\using {\Box}L
\endprooftree}
And thence to the conclusion of the endsequent:
\disp{
\prooftree
\prooftree
\prooftree
\prooftree
\prooftree
\prooftree
\justifies
\mbox{\fbox{$Nt(s(m)): x$}}\ \Rightarrow\ Nt(s(m)): x
\endprooftree
\justifies
\mbox{\fbox{${\blacksquare}Nt(s(m)): x$}}\ \Rightarrow\ Nt(s(m)): x
\using {\blacksquare}L
\endprooftree
\justifies
{\blacksquare}Nt(s(m)): x\ \Rightarrow\ \fbox{${\exists}gNt(s(g))$}: x
\using {\exists}R
\endprooftree
\justifies
[{\blacksquare}Nt(s(m)): x]\ \Rightarrow\ \fbox{${\langle\rangle}{\exists}gNt(s(g)): x$}
\using {\langle\rangle}R
\endprooftree
\prooftree
\justifies
\mbox{\fbox{$Sf: w$}}\ \Rightarrow\ Sf: w
\endprooftree
\justifies
[{\blacksquare}Nt(s(m)): x], \mbox{\fbox{${\langle\rangle}{\exists}gNt(s(g))\backslash Sf$}}: 
z\ \Rightarrow\ Sf: (z\ x)
\using {\backslash}L
\endprooftree
\justifies
[{\blacksquare}Nt(s(m)): x], {\square}({\langle\rangle}{\exists}gNt(s(g))\backslash Sf): y\ \Rightarrow\ Sf: (z\ x)\{\mbox{\v{}}y/z\} = (\mbox{\v{}}y\ x)
\using {\Box}L
\endprooftree}
Now we can substitute in the lexical semantics ${\it j}$ for \lingform{John\/} ($x$) and
$\mbox{\^{}}\lambda A({\it Pres}\ (\mbox{\v{}}{\it walk}\ {\it A}))$ for \lingform{walks\/}
($y$) and evaluate:\footnote{Montague's Intensional Logic assigned nonlogical constants of type $\tau$
a denotation in the intension of $\tau$ and then interpreted a constant
with respect to a world as its extension in that world.
By contrast  our semantic representation language assigns constants denotations in their own type,
so our semantic representations have explicit extensionalisations of
intensional constants.}
\disp{$\begin{array}[t]{l}
(\mbox{\v{}}\mbox{\^{}}\lambda A({\it Pres}\ (\mbox{\v{}}{\it walk}\ {\it A}))\ {\it j}) =\\
(\lambda A({\it Pres}\ (\mbox{\v{}}{\it walk}\ {\it A}))\ {\it j}) =\\
({\it Pres}\ (\mbox{\v{}}{\it walk}\ {\it j}))
\end{array}$}
\noindent
(As we have said, this elucidation 
is not exactly how CatLog2 extracts semantics; CatLog2 uses unification and instantiation
of metavariables to deliver in a single pass the unevaluated semantics of the upwards
and downward phases, and then
normalises.)

\begin{figure}
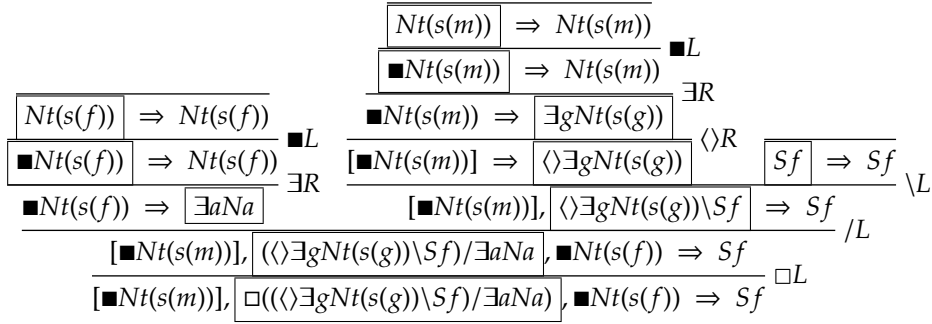

\begin{center}
\prooftree
\prooftree
\prooftree
\prooftree
\prooftree
\justifies
\mbox{\fbox{$Nt(s(f))$}}\ \Rightarrow\ Nt(s(f))
\endprooftree
\justifies
\mbox{\fbox{${\blacksquare}Nt(s(f))$}}\ \Rightarrow\ Nt(s(f))
\using {\blacksquare}L
\endprooftree
\justifies
{\blacksquare}Nt(s(f))\ \Rightarrow\ \fbox{${\exists}aNa$}
\using {\exists}R
\endprooftree
\prooftree
\prooftree
\prooftree
\prooftree
\prooftree
\justifies
\mbox{\fbox{$Nt(s(m))$}}\ \Rightarrow\ Nt(s(m))
\endprooftree
\justifies
\mbox{\fbox{${\blacksquare}Nt(s(m))$}}\ \Rightarrow\ Nt(s(m))
\using {\blacksquare}L
\endprooftree
\justifies
{\blacksquare}Nt(s(m))\ \Rightarrow\ \fbox{${\exists}gNt(s(g))$}
\using {\exists}R
\endprooftree
\justifies
[{\blacksquare}Nt(s(m))]\ \Rightarrow\ \fbox{${\langle\rangle}{\exists}gNt(s(g))$}
\using {\langle\rangle}R
\endprooftree
\prooftree
\justifies
\mbox{\fbox{$Sf$}}\ \Rightarrow\ Sf
\endprooftree
\justifies
[{\blacksquare}Nt(s(m))], \mbox{\fbox{${\langle\rangle}{\exists}gNt(s(g))\backslash Sf$}}\ \Rightarrow\ Sf
\using {\backslash}L
\endprooftree
\justifies
[{\blacksquare}Nt(s(m))], \mbox{\fbox{$({\langle\rangle}{\exists}gNt(s(g))\backslash Sf)/{\exists}aNa$}}, {\blacksquare}Nt(s(f))\ \Rightarrow\ Sf
\using {/}L
\endprooftree
\justifies
[{\blacksquare}Nt(s(m))], \mbox{\fbox{${\square}(({\langle\rangle}{\exists}gNt(s(g))\backslash Sf)/{\exists}aNa)$}}, {\blacksquare}Nt(s(f))\ \Rightarrow\ Sf
\using {\Box}L
\endprooftree
\end{center}
\caption{Derivation for \lingform{John loves Mary}}
\label{jlm}
\end{figure}

By way of a second example, the following is a simple transitive sentence:
\disp{
$[{\bf john}]{+}{\bf loves}{+}{\bf mary}: Sf$}
Lexical lookup yields:
\disp{
$[{\blacksquare}Nt(s(m)): {\it j}], {\square}(({\langle\rangle}{\exists}gNt(s(g))\backslash Sf)/{\exists}aNa): \mbox{\^{}}\lambda A\lambda B({\it Pres}\ ((\mbox{\v{}}{\it love}\ {\it A})\ {\it B})), {\blacksquare}Nt(s(f)): {\it m}\ \Rightarrow\ Sf$}
There is the derivation given in Figure~\ref{jlm}.
Reading upwards from the endsequent, the first inference removes the intensionality modality
from the transitive verb, and then over left selects the object to analyse as the
argument of the transitive verb; this is done by existential right instantiating the agreement 
feature to
third person singular feminine,
followed by (semantically inactive) intensionality modality left.
The right hand branch is the same as for example~(\ref{ex12}) after the first inference.
All this delivers semantics:
\disp{
$({\it Pres}\ ((\mbox{\v{}}{\it love}\ {\it m})\ {\it j}))$}

The next example has a subordinate clause:
\disp{
$[{\bf john}]{+}{\bf thinks}{+}[{\bf mary}]{+}{\bf walks}: Sf$
\label{ex3}}
Lexical lookup yields the following;
note that the propositional attitude verb is polymorphic with respect to
a complementised or uncomplementised sentential argument,
expressed with a semantically inactive additive disjunction:
\disp{
$[{\blacksquare}Nt(s(m)): {\it j}], {\square}(({\langle\rangle}{\exists}gNt(s(g))\backslash Sf)/({\it CP}that{\sqcup}{\square}Sf)): \mbox{\^{}}\lambda A\lambda B({\it Pres}\ ((\mbox{\v{}}{\it think}\ {\it A})\ {\it B})), \\{}[{\blacksquare}Nt(s(f)): {\it m}], {\square}({\langle\rangle}{\exists}gNt(s(g))\backslash Sf): \mbox{\^{}}\lambda C({\it Pres}\ (\mbox{\v{}}{\it walk}\ {\it C}))\ \Rightarrow\ Sf$}
This has the derivation given in Figure~\ref{jtmw}.
\begin{figure}
\begin{center}
\resizebox{\textwidth}{!}{\prooftree
\prooftree
\prooftree
\prooftree
\prooftree
\prooftree
\prooftree
\prooftree
\prooftree
\prooftree
\justifies
\mbox{\fbox{$Nt(s(f))$}}\ \Rightarrow\ Nt(s(f))
\endprooftree
\justifies
\mbox{\fbox{${\blacksquare}Nt(s(f))$}}\ \Rightarrow\ Nt(s(f))
\using {\blacksquare}L
\endprooftree
\justifies
{\blacksquare}Nt(s(f))\ \Rightarrow\ \fbox{${\exists}gNt(s(g))$}
\using {\exists}R
\endprooftree
\justifies
[{\blacksquare}Nt(s(f))]\ \Rightarrow\ \fbox{${\langle\rangle}{\exists}gNt(s(g))$}
\using {\langle\rangle}R
\endprooftree
\prooftree
\justifies
\mbox{\fbox{$Sf$}}\ \Rightarrow\ Sf
\endprooftree
\justifies
[{\blacksquare}Nt(s(f))], \mbox{\fbox{${\langle\rangle}{\exists}gNt(s(g))\backslash Sf$}}\ \Rightarrow\ Sf
\using {\backslash}L
\endprooftree
\justifies
[{\blacksquare}Nt(s(f))], \mbox{\fbox{${\square}({\langle\rangle}{\exists}gNt(s(g))\backslash Sf)$}}\ \Rightarrow\ Sf
\using {\Box}L
\endprooftree
\justifies
[{\blacksquare}Nt(s(f))], {\square}({\langle\rangle}{\exists}gNt(s(g))\backslash Sf)\ \Rightarrow\ {\square}Sf
\using {\Box}R
\endprooftree
\justifies
[{\blacksquare}Nt(s(f))], {\square}({\langle\rangle}{\exists}gNt(s(g))\backslash Sf)\ \Rightarrow\ \fbox{${\it CP}that{\sqcup}{\square}Sf$}
\using {\sqcup}R
\endprooftree
\prooftree
\prooftree
\prooftree
\prooftree
\prooftree
\justifies
\mbox{\fbox{$Nt(s(m))$}}\ \Rightarrow\ Nt(s(m))
\endprooftree
\justifies
\mbox{\fbox{${\blacksquare}Nt(s(m))$}}\ \Rightarrow\ Nt(s(m))
\using {\blacksquare}L
\endprooftree
\justifies
{\blacksquare}Nt(s(m))\ \Rightarrow\ \fbox{${\exists}gNt(s(g))$}
\using {\exists}R
\endprooftree
\justifies
[{\blacksquare}Nt(s(m))]\ \Rightarrow\ \fbox{${\langle\rangle}{\exists}gNt(s(g))$}
\using {\langle\rangle}R
\endprooftree
\prooftree
\justifies
\mbox{\fbox{$Sf$}}\ \Rightarrow\ Sf
\endprooftree
\justifies
[{\blacksquare}Nt(s(m))], \mbox{\fbox{${\langle\rangle}{\exists}gNt(s(g))\backslash Sf$}}\ \Rightarrow\ Sf
\using {\backslash}L
\endprooftree
\justifies
[{\blacksquare}Nt(s(m))], \mbox{\fbox{$({\langle\rangle}{\exists}gNt(s(g))\backslash Sf)/({\it CP}that{\sqcup}{\square}Sf)$}}, [{\blacksquare}Nt(s(f))], {\square}({\langle\rangle}{\exists}gNt(s(g))\backslash Sf)\ \Rightarrow\ Sf
\using {/}L
\endprooftree
\justifies
[{\blacksquare}Nt(s(m))], \mbox{\fbox{${\square}(({\langle\rangle}{\exists}gNt(s(g))\backslash Sf)/({\it CP}that{\sqcup}{\square}Sf))$}}, [{\blacksquare}Nt(s(f))], {\square}({\langle\rangle}{\exists}gNt(s(g))\backslash Sf)\ \Rightarrow\ Sf
\using {\Box}L
\endprooftree}
\end{center}
\caption{Derivation for \lingform{John thinks Mary walks}}
\label{jtmw}
\end{figure}
Reading bottom-up,
following elimination of the intensionality modality on the propositional
attitude verb, over left partitions in such a way as to supply the subordinate clause
as the propositional argument. Again, the righthand subtree 
is the same as for example~(\ref{jlm}) after the first inference. In the lefthand
subtree semantically inactive additive conjunction right selects the
modalised uncomplementized sentence type. The succedent modality is removed, this being licensed
by the fact that all the antecedent types are modalised, and the remaining derivation
is also like that for example~(\ref{jlm}).
The derivation delivers semantics:
\disp{
$({\it Pres}\ ((\mbox{\v{}}{\it think}\ \mbox{\^{}}({\it Pres}\ (\mbox{\v{}}{\it walk}\ {\it m})))\ {\it j}))$}

\commentout{

The next example has a subsubordinate clause:
\disp{
$[{\bf mary}]{+}{\bf believes}{+}[{\bf john}]{+}{\bf thinks}{+}[{\bf mary}]{+}{\bf walks}: Sf$}
Lexical lookup yields:
\disp{
$[{\blacksquare}Nt(s(f)): {\it m}], {\square}(({\langle\rangle}{\exists}gNt(s(g))\backslash Sf)/({\it CP}that{\sqcup}{\square}Sf)): \mbox{\^{}}\lambda A\lambda B({\it Pres}\ ((\mbox{\v{}}{\it believe}\ {\it A})\ {\it B})), \\{}[{\blacksquare}Nt(s(m)): {\it j}], {\square}(({\langle\rangle}{\exists}gNt(s(g))\backslash Sf)/({\it CP}that{\sqcup}{\square}Sf)): \mbox{\^{}}\lambda C\lambda D({\it Pres}\ ((\mbox{\v{}}{\it think}\ {\it C})\ {\it D})), \\{}[{\blacksquare}Nt(s(f)): {\it m}], {\square}({\langle\rangle}{\exists}gNt(s(g))\backslash Sf): \mbox{\^{}}\lambda E({\it Pres}\ (\mbox{\v{}}{\it walk}\ {\it E}))\ \Rightarrow\ Sf$}
There is the derivation given in Figure~\ref{mbjtmw}.
\begin{figure}\tiny
\begin{center}
\prooftree
\prooftree
\prooftree
\prooftree
\prooftree
\prooftree
\prooftree
\prooftree
\prooftree
\prooftree
\prooftree
\prooftree
\prooftree
\prooftree
\justifies
\mbox{\fbox{$Nt(s(f))$}}\ \Rightarrow\ Nt(s(f))
\endprooftree
\justifies
\mbox{\fbox{${\blacksquare}Nt(s(f))$}}\ \Rightarrow\ Nt(s(f))
\using {\blacksquare}L
\endprooftree
\justifies
{\blacksquare}Nt(s(f))\ \Rightarrow\ \fbox{${\exists}gNt(s(g))$}
\using {\exists}R
\endprooftree
\justifies
[{\blacksquare}Nt(s(f))]\ \Rightarrow\ \fbox{${\langle\rangle}{\exists}gNt(s(g))$}
\using {\langle\rangle}R
\endprooftree
\prooftree
\justifies
\mbox{\fbox{$Sf$}}\ \Rightarrow\ Sf
\endprooftree
\justifies
[{\blacksquare}Nt(s(f))], \mbox{\fbox{${\langle\rangle}{\exists}gNt(s(g))\backslash Sf$}}\ \Rightarrow\ Sf
\using {\backslash}L
\endprooftree
\justifies
[{\blacksquare}Nt(s(f))], \mbox{\fbox{${\square}({\langle\rangle}{\exists}gNt(s(g))\backslash Sf)$}}\ \Rightarrow\ Sf
\using {\Box}L
\endprooftree
\justifies
[{\blacksquare}Nt(s(f))], {\square}({\langle\rangle}{\exists}gNt(s(g))\backslash Sf)\ \Rightarrow\ {\square}Sf
\using {\Box}R
\endprooftree
\justifies
[{\blacksquare}Nt(s(f))], {\square}({\langle\rangle}{\exists}gNt(s(g))\backslash Sf)\ \Rightarrow\ \fbox{${\it CP}that{\sqcup}{\square}Sf$}
\using {\sqcup}R
\endprooftree
\prooftree
\prooftree
\prooftree
\prooftree
\prooftree
\justifies
\mbox{\fbox{$Nt(s(m))$}}\ \Rightarrow\ Nt(s(m))
\endprooftree
\justifies
\mbox{\fbox{${\blacksquare}Nt(s(m))$}}\ \Rightarrow\ Nt(s(m))
\using {\blacksquare}L
\endprooftree
\justifies
{\blacksquare}Nt(s(m))\ \Rightarrow\ \fbox{${\exists}gNt(s(g))$}
\using {\exists}R
\endprooftree
\justifies
[{\blacksquare}Nt(s(m))]\ \Rightarrow\ \fbox{${\langle\rangle}{\exists}gNt(s(g))$}
\using {\langle\rangle}R
\endprooftree
\prooftree
\justifies
\mbox{\fbox{$Sf$}}\ \Rightarrow\ Sf
\endprooftree
\justifies
[{\blacksquare}Nt(s(m))], \mbox{\fbox{${\langle\rangle}{\exists}gNt(s(g))\backslash Sf$}}\ \Rightarrow\ Sf
\using {\backslash}L
\endprooftree
\justifies
[{\blacksquare}Nt(s(m))], \mbox{\fbox{$({\langle\rangle}{\exists}gNt(s(g))\backslash Sf)/({\it CP}that{\sqcup}{\square}Sf)$}}, [{\blacksquare}Nt(s(f))], {\square}({\langle\rangle}{\exists}gNt(s(g))\backslash Sf)\ \Rightarrow\ Sf
\using {/}L
\endprooftree
\justifies
[{\blacksquare}Nt(s(m))], \mbox{\fbox{${\square}(({\langle\rangle}{\exists}gNt(s(g))\backslash Sf)/({\it CP}that{\sqcup}{\square}Sf))$}}, [{\blacksquare}Nt(s(f))], {\square}({\langle\rangle}{\exists}gNt(s(g))\backslash Sf)\ \Rightarrow\ Sf
\using {\Box}L
\endprooftree
\justifies
[{\blacksquare}Nt(s(m))], {\square}(({\langle\rangle}{\exists}gNt(s(g))\backslash Sf)/({\it CP}that{\sqcup}{\square}Sf)), [{\blacksquare}Nt(s(f))], {\square}({\langle\rangle}{\exists}gNt(s(g))\backslash Sf)\ \Rightarrow\ {\square}Sf
\using {\Box}R
\endprooftree
\justifies
[{\blacksquare}Nt(s(m))], {\square}(({\langle\rangle}{\exists}gNt(s(g))\backslash Sf)/({\it CP}that{\sqcup}{\square}Sf)), [{\blacksquare}Nt(s(f))], {\square}({\langle\rangle}{\exists}gNt(s(g))\backslash Sf)\ \Rightarrow\ \fbox{${\it CP}that{\sqcup}{\square}Sf$}
\using {\sqcup}R
\endprooftree
\prooftree
\prooftree
\prooftree
\prooftree
\prooftree
\justifies
\mbox{\fbox{$Nt(s(f))$}}\ \Rightarrow\ Nt(s(f))
\endprooftree
\justifies
\mbox{\fbox{${\blacksquare}Nt(s(f))$}}\ \Rightarrow\ Nt(s(f))
\using {\blacksquare}L
\endprooftree
\justifies
{\blacksquare}Nt(s(f))\ \Rightarrow\ \fbox{${\exists}gNt(s(g))$}
\using {\exists}R
\endprooftree
\justifies
[{\blacksquare}Nt(s(f))]\ \Rightarrow\ \fbox{${\langle\rangle}{\exists}gNt(s(g))$}
\using {\langle\rangle}R
\endprooftree
\prooftree
\justifies
\mbox{\fbox{$Sf$}}\ \Rightarrow\ Sf
\endprooftree
\justifies
[{\blacksquare}Nt(s(f))], \mbox{\fbox{${\langle\rangle}{\exists}gNt(s(g))\backslash Sf$}}\ \Rightarrow\ Sf
\using {\backslash}L
\endprooftree
\justifies
[{\blacksquare}Nt(s(f))], \mbox{\fbox{$({\langle\rangle}{\exists}gNt(s(g))\backslash Sf)/({\it CP}that{\sqcup}{\square}Sf)$}}, [{\blacksquare}Nt(s(m))], {\square}(({\langle\rangle}{\exists}gNt(s(g))\backslash Sf)/({\it CP}that{\sqcup}{\square}Sf)), [{\blacksquare}Nt(s(f))], {\square}({\langle\rangle}{\exists}gNt(s(g))\backslash Sf)\ \Rightarrow\ Sf
\using {/}L
\endprooftree
\justifies
[{\blacksquare}Nt(s(f))], \mbox{\fbox{${\square}(({\langle\rangle}{\exists}gNt(s(g))\backslash Sf)/({\it CP}that{\sqcup}{\square}Sf))$}}, [{\blacksquare}Nt(s(m))], {\square}(({\langle\rangle}{\exists}gNt(s(g))\backslash Sf)/({\it CP}that{\sqcup}{\square}Sf)), [{\blacksquare}Nt(s(f))], {\square}({\langle\rangle}{\exists}gNt(s(g))\backslash Sf)\ \Rightarrow\ Sf
\using {\Box}L
\endprooftree
\end{center}
\caption{Derivation for \lingform{Mary believes John thinks Mary walks}}
\label{mbjtmw}
\end{figure}
The derivation illustrates repetition of the same patterns of inference as in the 
analysis of example~(\ref{ex3}).
It delivers semantics:
\disp{
$({\it Pres}\ ((\mbox{\v{}}{\it believe}\ \mbox{\^{}}({\it Pres}\ ((\mbox{\v{}}{\it think}\ \mbox{\^{}}({\it Pres}\ (\mbox{\v{}}{\it walk}\ {\it m})))\ {\it j})))\ {\it m}))$}

}

The following example involves a ditransitive verb:
\disp{
$[{\bf mary}]{+}{\bf buys}{+}{\bf john}{+}{\bf coffee}: Sf$}
Lexical lookup is as follows;
note the use of product (multiplicative conjunction) for the ditransitive verb, 
and the use of additive conjunction for the polymorphism of the mass noun {\it coffee\/}
which can appear either as a bare nominal or with an article: 
\disp{
$[{\blacksquare}Nt(s(f)): {\it m}], {\square}(({\langle\rangle}{\exists}gNt(s(g))\backslash Sf)/({\exists}aNa{\bullet}{\exists}aNa)): \mbox{\^{}}\lambda A\lambda B({\it Pres}\ (((\mbox{\v{}}{\it buy}\ \pi_1{\it A})\ \pi_2{\it A})\ {\it B})),\\ {\blacksquare}Nt(s(m)): {\it j}, {\square}(Nt(s(n)){\&}{\it CN}{\it s(n)}): \mbox{\^{}}(({\it gen}\ \mbox{\v{}}{\it coffee}), \mbox{\v{}}{\it coffee})\ \Rightarrow\ Sf$}
There is the derivation given in Figure~\ref{mbjc}.
\begin{figure}
\begin{center}
\scriptsize
\prooftree
\prooftree
\prooftree
\prooftree
\prooftree
\prooftree
\justifies
\mbox{\fbox{$Nt(s(m))$}}\ \Rightarrow\ Nt(s(m))
\endprooftree
\justifies
\mbox{\fbox{${\blacksquare}Nt(s(m))$}}\ \Rightarrow\ Nt(s(m))
\using {\blacksquare}L
\endprooftree
\justifies
{\blacksquare}Nt(s(m))\ \Rightarrow\ \fbox{${\exists}aNa$}
\using {\exists}R
\endprooftree
\prooftree
\prooftree
\prooftree
\prooftree
\justifies
\mbox{\fbox{$Nt(s(n))$}}\ \Rightarrow\ Nt(s(n))
\endprooftree
\justifies
\mbox{\fbox{$Nt(s(n)){\&}{\it CN}{\it s(n)}$}}\ \Rightarrow\ Nt(s(n))
\using {\&}L
\endprooftree
\justifies
\mbox{\fbox{${\square}(Nt(s(n)){\&}{\it CN}{\it s(n)})$}}\ \Rightarrow\ Nt(s(n))
\using {\Box}L
\endprooftree
\justifies
{\square}(Nt(s(n)){\&}{\it CN}{\it s(n)})\ \Rightarrow\ \fbox{${\exists}aNa$}
\using {\exists}R
\endprooftree
\justifies
{\blacksquare}Nt(s(m)), {\square}(Nt(s(n)){\&}{\it CN}{\it s(n)})\ \Rightarrow\ \fbox{${\exists}aNa{\bullet}{\exists}aNa$}
\using {\bullet}R
\endprooftree
\prooftree
\prooftree
\prooftree
\prooftree
\prooftree
\justifies
\mbox{\fbox{$Nt(s(f))$}}\ \Rightarrow\ Nt(s(f))
\endprooftree
\justifies
\mbox{\fbox{${\blacksquare}Nt(s(f))$}}\ \Rightarrow\ Nt(s(f))
\using {\blacksquare}L
\endprooftree
\justifies
{\blacksquare}Nt(s(f))\ \Rightarrow\ \fbox{${\exists}gNt(s(g))$}
\using {\exists}R
\endprooftree
\justifies
[{\blacksquare}Nt(s(f))]\ \Rightarrow\ \fbox{${\langle\rangle}{\exists}gNt(s(g))$}
\using {\langle\rangle}R
\endprooftree
\prooftree
\justifies
\mbox{\fbox{$Sf$}}\ \Rightarrow\ Sf
\endprooftree
\justifies
[{\blacksquare}Nt(s(f))], \mbox{\fbox{${\langle\rangle}{\exists}gNt(s(g))\backslash Sf$}}\ \Rightarrow\ Sf
\using {\backslash}L
\endprooftree
\justifies
[{\blacksquare}Nt(s(f))], \mbox{\fbox{$({\langle\rangle}{\exists}gNt(s(g))\backslash Sf)/({\exists}aNa{\bullet}{\exists}aNa)$}}, {\blacksquare}Nt(s(m)), {\square}(Nt(s(n)){\&}{\it CN}{\it s(n)})\ \Rightarrow\ Sf
\using {/}L
\endprooftree
\justifies
[{\blacksquare}Nt(s(f))], \mbox{\fbox{${\square}(({\langle\rangle}{\exists}gNt(s(g))\backslash Sf)/({\exists}aNa{\bullet}{\exists}aNa))$}}, {\blacksquare}Nt(s(m)), {\square}(Nt(s(n)){\&}{\it CN}{\it s(n)})\ \Rightarrow\ Sf
\using {\Box}L
\endprooftree
\end{center}
\caption{Derivation for \lingform{Mary buys John coffee}}
\label{mbjc}
\end{figure}
After removal of the outer modality of the ditransitive verb, the partitioning of over left
selects the two objects as the verb's product argument,
partitioned in turn by continuous product right. The indirect object \lingform{John}
is analysed by existential right and inactive modality left inferences;
the direct object \lingform{coffee} is analysed by 
existential right and (active) modality left inferences followed by selection of the
bare noun type by additive conjunction left.
The rightmost subtree is as usual for an intransitive sentence.
This delivers semantics as follows in which a `generic' operator applies to {\it coffee}:
\disp{
$({\it Pres}\ (((\mbox{\v{}}{\it buy}\ {\it j})\ ({\it gen}\ \mbox{\v{}}{\it coffee}))\ {\it m}))$}

The next example includes a definite article:
\disp{
$[{\bf the}{+}{\bf man}]{+}{\bf walks}: Sf$}
We treat the definite article simply as an iota operator which returns the unique
individual in the context of discourse satisfying its common noun argument (Carpenter 
1997\cite{carpenter:96});
this unicity is presupposed by the use of the definite.
Lexical lookup yields the semantically labelled sequent:
\disp{
$[{\blacksquare}{\forall}n(Nt(n)/{\it CN}{\it n}): \iota , {\square}{\it CN}{\it s(m)}: {\it man}], {\square}({\langle\rangle}{\exists}gNt(s(g))\backslash Sf): \mbox{\^{}}\lambda A({\it Pres}\ (\mbox{\v{}}{\it walk}\ {\it A}))\ \Rightarrow\ Sf$}
There is the derivation given in Figure~\ref{tmw}.
\begin{figure}
\begin{center}
\prooftree
\prooftree
\prooftree
\prooftree
\prooftree
\prooftree
\prooftree
\prooftree
\prooftree
\justifies
\mbox{\fbox{${\it CN}{\it s(m)}$}}\ \Rightarrow\ {\it CN}{\it s(m)}
\endprooftree
\justifies
\mbox{\fbox{${\square}{\it CN}{\it s(m)}$}}\ \Rightarrow\ {\it CN}{\it s(m)}
\using {\Box}L
\endprooftree
\prooftree
\justifies
\mbox{\fbox{$Nt(s(m))$}}\ \Rightarrow\ Nt(s(m))
\endprooftree
\justifies
\mbox{\fbox{$Nt(s(m))/{\it CN}{\it s(m)}$}}, {\square}{\it CN}{\it s(m)}\ \Rightarrow\ Nt(s(m))
\using {/}L
\endprooftree
\justifies
\mbox{\fbox{${\forall}n(Nt(n)/{\it CN}{\it n})$}}, {\square}{\it CN}{\it s(m)}\ \Rightarrow\ Nt(s(m))
\using {\forall}L
\endprooftree
\justifies
\mbox{\fbox{${\blacksquare}{\forall}n(Nt(n)/{\it CN}{\it n})$}}, {\square}{\it CN}{\it s(m)}\ \Rightarrow\ Nt(s(m))
\using {\blacksquare}L
\endprooftree
\justifies
{\blacksquare}{\forall}n(Nt(n)/{\it CN}{\it n}), {\square}{\it CN}{\it s(m)}\ \Rightarrow\ \fbox{${\exists}gNt(s(g))$}
\using {\exists}R
\endprooftree
\justifies
[{\blacksquare}{\forall}n(Nt(n)/{\it CN}{\it n}), {\square}{\it CN}{\it s(m)}]\ \Rightarrow\ \fbox{${\langle\rangle}{\exists}gNt(s(g))$}
\using {\langle\rangle}R
\endprooftree
\prooftree
\justifies
\mbox{\fbox{$Sf$}}\ \Rightarrow\ Sf
\endprooftree
\justifies
[{\blacksquare}{\forall}n(Nt(n)/{\it CN}{\it n}), {\square}{\it CN}{\it s(m)}], \mbox{\fbox{${\langle\rangle}{\exists}gNt(s(g))\backslash Sf$}}\ \Rightarrow\ Sf
\using {\backslash}L
\endprooftree
\justifies
[{\blacksquare}{\forall}n(Nt(n)/{\it CN}{\it n}), {\square}{\it CN}{\it s(m)}], \mbox{\fbox{${\square}({\langle\rangle}{\exists}gNt(s(g))\backslash Sf)$}}\ \Rightarrow\ Sf
\using {\Box}L
\endprooftree
\end{center}
\caption{Derivation for \lingform{The man walks}}
\label{tmw}
\end{figure}
This is like the derivation of an intransitive sentence before,
but with the analysis of the definite noun phrase subject at the top left.
The derivation delivers semantics:
\disp{
$({\it Pres}\ (\mbox{\v{}}{\it walk}\ (\iota \ \mbox{\v{}}{\it man})))$}

\begin{figure}
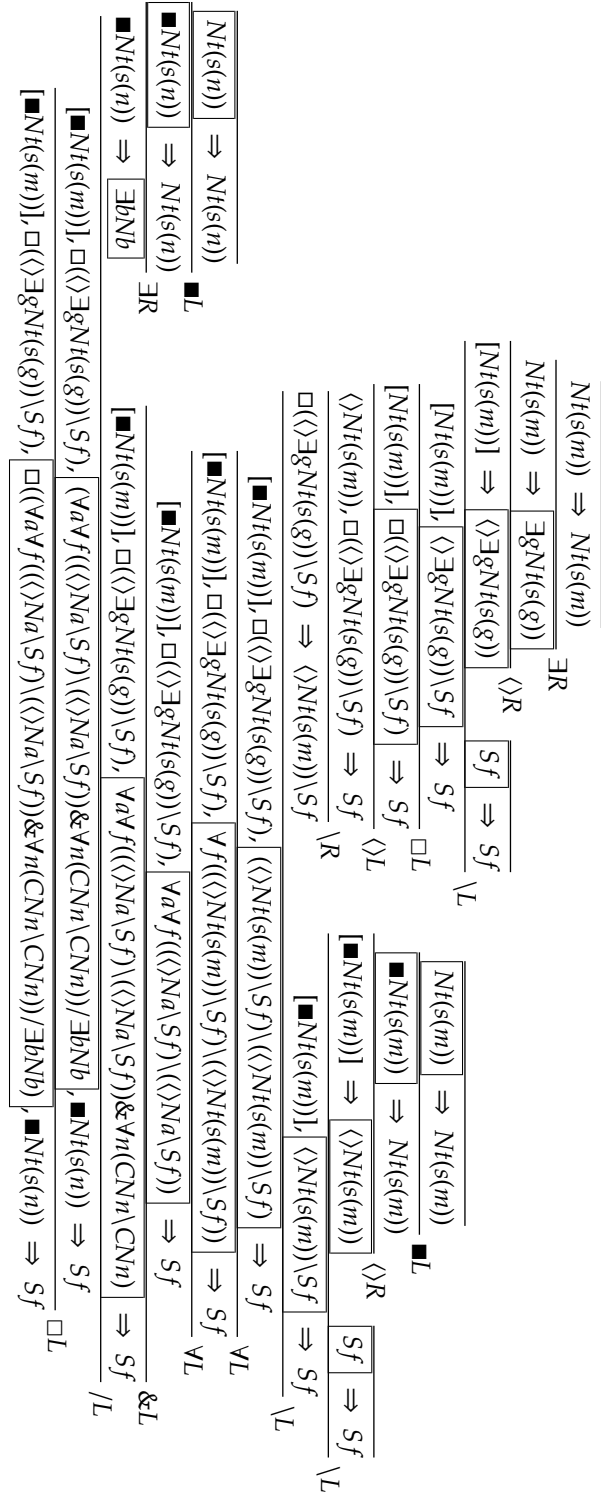

\begin{center}
\rotatebox{-90}{
\prooftree
\prooftree
\prooftree
\prooftree
\prooftree
\justifies
\mbox{\fbox{$Nt(s(n))$}}\ \Rightarrow\ Nt(s(n))
\endprooftree
\justifies
\mbox{\fbox{${\blacksquare}Nt(s(n))$}}\ \Rightarrow\ Nt(s(n))
\using {\blacksquare}L
\endprooftree
\justifies
{\blacksquare}Nt(s(n))\ \Rightarrow\ \fbox{${\exists}bNb$}
\using {\exists}R
\endprooftree
\prooftree
\prooftree
\prooftree
\prooftree
\prooftree
\prooftree
\prooftree
\prooftree
\prooftree
\prooftree
\prooftree
\justifies
Nt(s(m))\ \Rightarrow\ Nt(s(m))
\endprooftree
\justifies
Nt(s(m))\ \Rightarrow\ \fbox{${\exists}gNt(s(g))$}
\using {\exists}R
\endprooftree
\justifies
[Nt(s(m))]\ \Rightarrow\ \fbox{${\langle\rangle}{\exists}gNt(s(g))$}
\using {\langle\rangle}R
\endprooftree
\prooftree
\justifies
\mbox{\fbox{$Sf$}}\ \Rightarrow\ Sf
\endprooftree
\justifies
[Nt(s(m))], \mbox{\fbox{${\langle\rangle}{\exists}gNt(s(g))\backslash Sf$}}\ \Rightarrow\ Sf
\using {\backslash}L
\endprooftree
\justifies
[Nt(s(m))], \mbox{\fbox{${\square}({\langle\rangle}{\exists}gNt(s(g))\backslash Sf)$}}\ \Rightarrow\ Sf
\using {\Box}L
\endprooftree
\justifies
{\langle\rangle}Nt(s(m)), {\square}({\langle\rangle}{\exists}gNt(s(g))\backslash Sf)\ \Rightarrow\ Sf
\using {\langle\rangle}L
\endprooftree
\justifies
{\square}({\langle\rangle}{\exists}gNt(s(g))\backslash Sf)\ \Rightarrow\ {\langle\rangle}Nt(s(m))\backslash Sf
\using {\backslash}R
\endprooftree
\prooftree
\prooftree
\prooftree
\prooftree
\justifies
\mbox{\fbox{$Nt(s(m))$}}\ \Rightarrow\ Nt(s(m))
\endprooftree
\justifies
\mbox{\fbox{${\blacksquare}Nt(s(m))$}}\ \Rightarrow\ Nt(s(m))
\using {\blacksquare}L
\endprooftree
\justifies
[{\blacksquare}Nt(s(m))]\ \Rightarrow\ \fbox{${\langle\rangle}Nt(s(m))$}
\using {\langle\rangle}R
\endprooftree
\prooftree
\justifies
\mbox{\fbox{$Sf$}}\ \Rightarrow\ Sf
\endprooftree
\justifies
[{\blacksquare}Nt(s(m))], \mbox{\fbox{${\langle\rangle}Nt(s(m))\backslash Sf$}}\ \Rightarrow\ Sf
\using {\backslash}L
\endprooftree
\justifies
[{\blacksquare}Nt(s(m))], {\square}({\langle\rangle}{\exists}gNt(s(g))\backslash Sf), \mbox{\fbox{$({\langle\rangle}Nt(s(m))\backslash Sf)\backslash ({\langle\rangle}Nt(s(m))\backslash Sf)$}}\ \Rightarrow\ Sf
\using {\backslash}L
\endprooftree
\justifies
[{\blacksquare}Nt(s(m))], {\square}({\langle\rangle}{\exists}gNt(s(g))\backslash Sf), \mbox{\fbox{${\forall}f(({\langle\rangle}Nt(s(m))\backslash Sf)\backslash ({\langle\rangle}Nt(s(m))\backslash Sf))$}}\ \Rightarrow\ Sf
\using {\forall}L
\endprooftree
\justifies
[{\blacksquare}Nt(s(m))], {\square}({\langle\rangle}{\exists}gNt(s(g))\backslash Sf), \mbox{\fbox{${\forall}a{\forall}f(({\langle\rangle}Na\backslash Sf)\backslash ({\langle\rangle}Na\backslash Sf))$}}\ \Rightarrow\ Sf
\using {\forall}L
\endprooftree
\justifies
[{\blacksquare}Nt(s(m))], {\square}({\langle\rangle}{\exists}gNt(s(g))\backslash Sf), \mbox{\fbox{${\forall}a{\forall}f(({\langle\rangle}Na\backslash Sf)\backslash ({\langle\rangle}Na\backslash Sf)){\&}{\forall}n({\it CN}{\it n}\backslash {\it CN}{\it n})$}}\ \Rightarrow\ Sf
\using {\&}L
\endprooftree
\justifies
[{\blacksquare}Nt(s(m))], {\square}({\langle\rangle}{\exists}gNt(s(g))\backslash Sf), \mbox{\fbox{$({\forall}a{\forall}f(({\langle\rangle}Na\backslash Sf)\backslash ({\langle\rangle}Na\backslash Sf)){\&}{\forall}n({\it CN}{\it n}\backslash {\it CN}{\it n}))/{\exists}bNb$}}, {\blacksquare}Nt(s(n))\ \Rightarrow\ Sf
\using {/}L
\endprooftree
\justifies
[{\blacksquare}Nt(s(m))], {\square}({\langle\rangle}{\exists}gNt(s(g))\backslash Sf), \mbox{\fbox{${\square}(({\forall}a{\forall}f(({\langle\rangle}Na\backslash Sf)\backslash ({\langle\rangle}Na\backslash Sf)){\&}{\forall}n({\it CN}{\it n}\backslash {\it CN}{\it n}))/{\exists}bNb)$}}, {\blacksquare}Nt(s(n))\ \Rightarrow\ Sf
\using {\Box}L
\endprooftree
}
\end{center}
\caption{Derivation for \lingform{John walks from Edinburgh}}
\label{jwfe}
\end{figure}

The next two examples have adverbial and adnominal prepositional modification
respectively. We consider the adverbial case first:
\disp{
$[{\bf john}]{+}{\bf walks}{+}{\bf from}{+}{\bf edinburgh}: Sf$}
Lexical lookup inserts a single value-polymorphic prepositional type,
which uses semantically active additive conjunction:
\disp{
$[{\blacksquare}Nt(s(m)): {\it j}], {\square}({\langle\rangle}{\exists}gNt(s(g))\backslash Sf): \mbox{\^{}}\lambda A({\it Pres}\ (\mbox{\v{}}{\it walk}\ {\it A})), {\square}(({\forall}a{\forall}f(({\langle\rangle}Na\backslash Sf)\backslash\\ ({\langle\rangle}Na\backslash Sf)){\&}{\forall}n({\it CN}{\it n}\backslash {\it CN}{\it n}))/{\exists}bNb): \mbox{\^{}}\lambda B((\mbox{\v{}}{\it fromadv}\ {\it B}), (\mbox{\v{}}{\it fromadn}\ {\it B})), {\blacksquare}Nt(s(n)): {\it e}\ \Rightarrow\ Sf$}
There is the derivation given in Figure~\ref{jwfe}.
After elimination of the outer modality of the preposition, over left
selects as the prepositional 
argument the prepositional object, which is analysed in the leftmost subtree.
In the sister subtree additive conjunction left selects the adverbial type for the prepositional phrase
and for all left instantiates the subject agreement and verb form features
to third person singular masculine, and finite. Following under left,
in the middle subtree \lingform{walks} is analysed as the intransitive verb second argument
of the adverbial preposition; 
note the analysis of the higher-order type by the under right rule,
which lowers the conclusion succedent hypothetical subtype into the premise antecedent.
The rightmost subtree is an intransitive sentence case again.
All this delivers the semantics:
\disp{
$(((\mbox{\v{}}{\it fromadv}\ {\it e})\ \lambda B({\it Pres}\ (\mbox{\v{}}{\it walk}\ {\it B})))\ {\it j})$}

The adnominal case is:
\disp{
$[{\bf the}{+}{\bf man}{+}{\bf from}{+}{\bf edinburgh}]{+}{\bf walks}: Sf$}
Lexical lookup yields:
\disp{
$[{\blacksquare}{\forall}n(Nt(n)/{\it CN}{\it n}): \iota , {\square}{\it CN}{\it s(m)}: {\it man}, {\square}(({\forall}a{\forall}f(({\langle\rangle}Na\backslash Sf)\backslash ({\langle\rangle}Na\backslash Sf)){\&}\\{\forall}n({\it CN}{\it n}\backslash {\it CN}{\it n}))/{\exists}bNb): \mbox{\^{}}\lambda A((\mbox{\v{}}{\it fromadv}\ {\it A}), (\mbox{\v{}}{\it fromadn}\ {\it A})), {\blacksquare}Nt(s(n)): {\it e}],\\ {\square}({\langle\rangle}{\exists}gNt(s(g))\backslash Sf): \mbox{\^{}}\lambda B({\it Pres}\ (\mbox{\v{}}{\it walk}\ {\it B}))\ \Rightarrow\ Sf$}
There is the derivation given in Figure~\ref{tmfew}.
\begin{figure}
\begin{center}
\rotatebox{-90}{
\prooftree
\prooftree
\prooftree
\prooftree
\prooftree
\prooftree
\prooftree
\prooftree
\prooftree
\prooftree
\prooftree
\prooftree
\justifies
\mbox{\fbox{$Nt(s(n))$}}\ \Rightarrow\ Nt(s(n))
\endprooftree
\justifies
\mbox{\fbox{${\blacksquare}Nt(s(n))$}}\ \Rightarrow\ Nt(s(n))
\using {\blacksquare}L
\endprooftree
\justifies
{\blacksquare}Nt(s(n))\ \Rightarrow\ \fbox{${\exists}bNb$}
\using {\exists}R
\endprooftree
\prooftree
\prooftree
\prooftree
\prooftree
\prooftree
\justifies
\mbox{\fbox{${\it CN}{\it s(m)}$}}\ \Rightarrow\ {\it CN}{\it s(m)}
\endprooftree
\justifies
\mbox{\fbox{${\square}{\it CN}{\it s(m)}$}}\ \Rightarrow\ {\it CN}{\it s(m)}
\using {\Box}L
\endprooftree
\prooftree
\justifies
\mbox{\fbox{${\it CN}{\it s(m)}$}}\ \Rightarrow\ {\it CN}{\it s(m)}
\endprooftree
\justifies
{\square}{\it CN}{\it s(m)}, \mbox{\fbox{${\it CN}{\it s(m)}\backslash {\it CN}{\it s(m)}$}}\ \Rightarrow\ {\it CN}{\it s(m)}
\using {\backslash}L
\endprooftree
\justifies
{\square}{\it CN}{\it s(m)}, \mbox{\fbox{${\forall}n({\it CN}{\it n}\backslash {\it CN}{\it n})$}}\ \Rightarrow\ {\it CN}{\it s(m)}
\using {\forall}L
\endprooftree
\justifies
{\square}{\it CN}{\it s(m)}, \mbox{\fbox{${\forall}a{\forall}f(({\langle\rangle}Na\backslash Sf)\backslash ({\langle\rangle}Na\backslash Sf)){\&}{\forall}n({\it CN}{\it n}\backslash {\it CN}{\it n})$}}\ \Rightarrow\ {\it CN}{\it s(m)}
\using {\&}L
\endprooftree
\justifies
{\square}{\it CN}{\it s(m)}, \mbox{\fbox{$({\forall}a{\forall}f(({\langle\rangle}Na\backslash Sf)\backslash ({\langle\rangle}Na\backslash Sf)){\&}{\forall}n({\it CN}{\it n}\backslash {\it CN}{\it n}))/{\exists}bNb$}}, {\blacksquare}Nt(s(n))\ \Rightarrow\ {\it CN}{\it s(m)}
\using {/}L
\endprooftree
\justifies
{\square}{\it CN}{\it s(m)}, \mbox{\fbox{${\square}(({\forall}a{\forall}f(({\langle\rangle}Na\backslash Sf)\backslash ({\langle\rangle}Na\backslash Sf)){\&}{\forall}n({\it CN}{\it n}\backslash {\it CN}{\it n}))/{\exists}bNb)$}}, {\blacksquare}Nt(s(n))\ \Rightarrow\ {\it CN}{\it s(m)}
\using {\Box}L
\endprooftree
\prooftree
\justifies
\mbox{\fbox{$Nt(s(m))$}}\ \Rightarrow\ Nt(s(m))
\endprooftree
\justifies
\mbox{\fbox{$Nt(s(m))/{\it CN}{\it s(m)}$}}, {\square}{\it CN}{\it s(m)}, {\square}(({\forall}a{\forall}f(({\langle\rangle}Na\backslash Sf)\backslash ({\langle\rangle}Na\backslash Sf)){\&}{\forall}n({\it CN}{\it n}\backslash {\it CN}{\it n}))/{\exists}bNb), {\blacksquare}Nt(s(n))\ \Rightarrow\ Nt(s(m))
\using {/}L
\endprooftree
\justifies
\mbox{\fbox{${\forall}n(Nt(n)/{\it CN}{\it n})$}}, {\square}{\it CN}{\it s(m)}, {\square}(({\forall}a{\forall}f(({\langle\rangle}Na\backslash Sf)\backslash ({\langle\rangle}Na\backslash Sf)){\&}{\forall}n({\it CN}{\it n}\backslash {\it CN}{\it n}))/{\exists}bNb), {\blacksquare}Nt(s(n))\ \Rightarrow\ Nt(s(m))
\using {\forall}L
\endprooftree
\justifies
\mbox{\fbox{${\blacksquare}{\forall}n(Nt(n)/{\it CN}{\it n})$}}, {\square}{\it CN}{\it s(m)}, {\square}(({\forall}a{\forall}f(({\langle\rangle}Na\backslash Sf)\backslash ({\langle\rangle}Na\backslash Sf)){\&}{\forall}n({\it CN}{\it n}\backslash {\it CN}{\it n}))/{\exists}bNb), {\blacksquare}Nt(s(n))\ \Rightarrow\ Nt(s(m))
\using {\blacksquare}L
\endprooftree
\justifies
{\blacksquare}{\forall}n(Nt(n)/{\it CN}{\it n}), {\square}{\it CN}{\it s(m)}, {\square}(({\forall}a{\forall}f(({\langle\rangle}Na\backslash Sf)\backslash ({\langle\rangle}Na\backslash Sf)){\&}{\forall}n({\it CN}{\it n}\backslash {\it CN}{\it n}))/{\exists}bNb), {\blacksquare}Nt(s(n))\ \Rightarrow\ \fbox{${\exists}gNt(s(g))$}
\using {\exists}R
\endprooftree
\justifies
[{\blacksquare}{\forall}n(Nt(n)/{\it CN}{\it n}), {\square}{\it CN}{\it s(m)}, {\square}(({\forall}a{\forall}f(({\langle\rangle}Na\backslash Sf)\backslash ({\langle\rangle}Na\backslash Sf)){\&}{\forall}n({\it CN}{\it n}\backslash {\it CN}{\it n}))/{\exists}bNb), {\blacksquare}Nt(s(n))]\ \Rightarrow\ \fbox{${\langle\rangle}{\exists}gNt(s(g))$}
\using {\langle\rangle}R
\endprooftree
\prooftree
\justifies
\mbox{\fbox{$Sf$}}\ \Rightarrow\ Sf
\endprooftree
\justifies
[{\blacksquare}{\forall}n(Nt(n)/{\it CN}{\it n}), {\square}{\it CN}{\it s(m)}, {\square}(({\forall}a{\forall}f(({\langle\rangle}Na\backslash Sf)\backslash ({\langle\rangle}Na\backslash Sf)){\&}{\forall}n({\it CN}{\it n}\backslash {\it CN}{\it n}))/{\exists}bNb), {\blacksquare}Nt(s(n))], \mbox{\fbox{${\langle\rangle}{\exists}gNt(s(g))\backslash Sf$}}\ \Rightarrow\ Sf
\using {\backslash}L
\endprooftree
\justifies
[{\blacksquare}{\forall}n(Nt(n)/{\it CN}{\it n}), {\square}{\it CN}{\it s(m)}, {\square}(({\forall}a{\forall}f(({\langle\rangle}Na\backslash Sf)\backslash ({\langle\rangle}Na\backslash Sf)){\&}{\forall}n({\it CN}{\it n}\backslash {\it CN}{\it n}))/{\exists}bNb), {\blacksquare}Nt(s(n))], \mbox{\fbox{${\square}({\langle\rangle}{\exists}gNt(s(g))\backslash Sf)$}}\ \Rightarrow\ Sf
\using {\Box}L
\endprooftree
}
\end{center}
\caption{Derivation for \lingform{The man from Edinburgh walks}}
\label{tmfew}
\end{figure}
In the first two steps the intransitive verb \lingform{walks\/} is prepared to apply
to the complex subject. Bracket right and exists right follow,
then (inactive) modality left and for all left on the determiner,
which then applies to the complex common noun. The result of modality
left on the preposition applies to the prepositional object and in the major premise
additive conjunction left selects the adnominal prepositional type.
The semantics delivered is:
\disp{
$({\it Pres}\ (\mbox{\v{}}{\it walk}\ (\iota \ ((\mbox{\v{}}{\it fromadn}\ {\it e})\ \mbox{\v{}}{\it man}))))$}

The last two initial examples involve the copula with nominal and (intersective) adjectival
complementation respectively. We consider first the nominal case:
\disp{
$[{\bf bond}]{+}{\bf is}{+}{\bf 007}: Sf$}
Lexical lookup inserts a single argument-polymorphic copula type,
which uses both semantically active and semantically inactive additive disjunction:\footnote{The difference operator (Morrill and Valent\'{\i}n 2014\cite{mv:jcss13}) for linguistic
exceptions is also used.
It involves negation as failure, 
which cannot easily be displayed. We do not dwell on this operator here.}
\disp{
$[{\blacksquare}Nt(s(m)): {\it b}], {\blacksquare}(({\langle\rangle}{\exists}gNt(s(g))\backslash Sf)/({\exists}aNa{\oplus}({\exists}g(({\it CN}{\it g}/{\it CN}{\it g}){\sqcup}({\it CN}{\it g}\backslash {\it CN}{\it g})){-}I))):$ $\lambda A\lambda B({\it Pres}$ $\ ({\it A}\casearrow C.[{\it B}={\it C}]; D.(({\it D}\ \lambda E[{\it E}={\it B}])\ {\it B}))), {\blacksquare}{\forall}gNt(s(g)): {\it 007}\ \Rightarrow\ Sf$}
There is the derivation given in Figure~\ref{bi7}.
\begin{figure}
\begin{center}
\resizebox{\textwidth}{!}{
\prooftree
\prooftree
\prooftree
\prooftree
\prooftree
\prooftree
\prooftree
\justifies
\mbox{\fbox{$Nt(s(A))$}}\ \Rightarrow\ Nt(s(A))
\endprooftree
\justifies
\mbox{\fbox{${\forall}gNt(s(g))$}}\ \Rightarrow\ Nt(s(A))
\using {\forall}L
\endprooftree
\justifies
\mbox{\fbox{${\blacksquare}{\forall}gNt(s(g))$}}\ \Rightarrow\ Nt(s(A))
\using {\blacksquare}L
\endprooftree
\justifies
{\blacksquare}{\forall}gNt(s(g))\ \Rightarrow\ \fbox{${\exists}aNa$}
\using {\exists}R
\endprooftree
\justifies
{\blacksquare}{\forall}gNt(s(g))\ \Rightarrow\ \fbox{${\exists}aNa{\oplus}({\exists}g(({\it CN}{\it g}/{\it CN}{\it g}){\sqcup}({\it CN}{\it g}\backslash {\it CN}{\it g})){-}I)$}
\using {\oplus}R
\endprooftree
\prooftree
\prooftree
\prooftree
\prooftree
\prooftree
\justifies
\mbox{\fbox{$Nt(s(m))$}}\ \Rightarrow\ Nt(s(m))
\endprooftree
\justifies
\mbox{\fbox{${\blacksquare}Nt(s(m))$}}\ \Rightarrow\ Nt(s(m))
\using {\blacksquare}L
\endprooftree
\justifies
{\blacksquare}Nt(s(m))\ \Rightarrow\ \fbox{${\exists}gNt(s(g))$}
\using {\exists}R
\endprooftree
\justifies
[{\blacksquare}Nt(s(m))]\ \Rightarrow\ \fbox{${\langle\rangle}{\exists}gNt(s(g))$}
\using {\langle\rangle}R
\endprooftree
\prooftree
\justifies
\mbox{\fbox{$Sf$}}\ \Rightarrow\ Sf
\endprooftree
\justifies
[{\blacksquare}Nt(s(m))], \mbox{\fbox{${\langle\rangle}{\exists}gNt(s(g))\backslash Sf$}}\ \Rightarrow\ Sf
\using {\backslash}L
\endprooftree
\justifies
[{\blacksquare}Nt(s(m))], \mbox{\fbox{$({\langle\rangle}{\exists}gNt(s(g))\backslash Sf)/({\exists}aNa{\oplus}({\exists}g(({\it CN}{\it g}/{\it CN}{\it g}){\sqcup}({\it CN}{\it g}\backslash {\it CN}{\it g})){-}I))$}}, {\blacksquare}{\forall}gNt(s(g))\ \Rightarrow\ Sf
\using {/}L
\endprooftree
\justifies
[{\blacksquare}Nt(s(m))], \mbox{\fbox{${\blacksquare}(({\langle\rangle}{\exists}gNt(s(g))\backslash Sf)/({\exists}aNa{\oplus}({\exists}g(({\it CN}{\it g}/{\it CN}{\it g}){\sqcup}({\it CN}{\it g}\backslash {\it CN}{\it g})){-}I)))$}}, {\blacksquare}{\forall}gNt(s(g))\ \Rightarrow\ Sf
\using {\blacksquare}L
\endprooftree}
\end{center}
\caption{Derivation for \lingform{Bond is 007}}
\label{bi7}
\end{figure}
After elimination of the outer copula modality the copula is applied to its nominal complement.
Additive disjunction right selects the first, nominal, disjunct.
The derivation delivers semantics:
\disp{
$({\it Pres}\ [{\it b}={\it 007}])$}

The (intersective) adjectival case is:
\disp{
$[{\bf bond}]{+}{\bf is}{+}{\bf teetotal}: Sf$}
Lexical lookup yields:
\disp{
$[{\blacksquare}Nt(s(m)): {\it b}], {\blacksquare}(({\langle\rangle}{\exists}gNt(s(g))\backslash Sf)/({\exists}aNa{\oplus}({\exists}g(({\it CN}{\it g}/{\it CN}{\it g}){\sqcup}({\it CN}{\it g}\backslash {\it CN}{\it g})){-}I))): \lambda A\lambda B({\it Pres}$ $\ ({\it A}\casearrow C.[{\it B}={\it C}]; D.(({\it D}\ \lambda E[{\it E}={\it B}])\ {\it B}))), {\square}{\forall}n({\it CN}{\it n}/{\it CN}{\it n}): \mbox{\^{}}\lambda F\lambda G[({\it F}\ {\it G})\wedge (\mbox{\v{}}{\it teetotal}\ {\it G})]\ \Rightarrow\ Sf$}
There is the derivation given in Figure~\ref{bit}.
\begin{figure}
\begin{center}
\resizebox{\textwidth}{!}{
\prooftree
\prooftree
\prooftree
\prooftree
\prooftree
\prooftree
\prooftree
\prooftree
\prooftree
\prooftree
\prooftree
\justifies
{\it CN}{\it A}\ \Rightarrow\ {\it CN}{\it A}
\endprooftree
\prooftree
\justifies
\mbox{\fbox{${\it CN}{\it A}$}}\ \Rightarrow\ {\it CN}{\it A}
\endprooftree
\justifies
\mbox{\fbox{${\it CN}{\it A}/{\it CN}{\it A}$}}, {\it CN}{\it A}\ \Rightarrow\ {\it CN}{\it A}
\using {/}L
\endprooftree
\justifies
\mbox{\fbox{${\forall}n({\it CN}{\it n}/{\it CN}{\it n})$}}, {\it CN}{\it A}\ \Rightarrow\ {\it CN}{\it A}
\using {\forall}L
\endprooftree
\justifies
\mbox{\fbox{${\square}{\forall}n({\it CN}{\it n}/{\it CN}{\it n})$}}, {\it CN}{\it A}\ \Rightarrow\ {\it CN}{\it A}
\using {\Box}L
\endprooftree
\justifies
{\square}{\forall}n({\it CN}{\it n}/{\it CN}{\it n})\ \Rightarrow\ {\it CN}{\it A}/{\it CN}{\it A}
\using {/}R
\endprooftree
\justifies
{\square}{\forall}n({\it CN}{\it n}/{\it CN}{\it n})\ \Rightarrow\ \fbox{$({\it CN}{\it A}/{\it CN}{\it A}){\sqcup}({\it CN}{\it A}\backslash {\it CN}{\it A})$}
\using {\sqcup}R
\endprooftree
\justifies
{\square}{\forall}n({\it CN}{\it n}/{\it CN}{\it n})\ \Rightarrow\ \fbox{${\exists}g(({\it CN}{\it g}/{\it CN}{\it g}){\sqcup}({\it CN}{\it g}\backslash {\it CN}{\it g}))$}
\using {\exists}R
\endprooftree
\justifies
{\square}{\forall}n({\it CN}{\it n}/{\it CN}{\it n})\ \Rightarrow\ \fbox{${\exists}g(({\it CN}{\it g}/{\it CN}{\it g}){\sqcup}({\it CN}{\it g}\backslash {\it CN}{\it g})){-}I$}
\using {-}R
\endprooftree
\justifies
{\square}{\forall}n({\it CN}{\it n}/{\it CN}{\it n})\ \Rightarrow\ \fbox{${\exists}aNa{\oplus}({\exists}g(({\it CN}{\it g}/{\it CN}{\it g}){\sqcup}({\it CN}{\it g}\backslash {\it CN}{\it g})){-}I)$}
\using {\oplus}R
\endprooftree
\prooftree
\prooftree
\prooftree
\prooftree
\prooftree
\justifies
\mbox{\fbox{$Nt(s(m))$}}\ \Rightarrow\ Nt(s(m))
\endprooftree
\justifies
\mbox{\fbox{${\blacksquare}Nt(s(m))$}}\ \Rightarrow\ Nt(s(m))
\using {\blacksquare}L
\endprooftree
\justifies
{\blacksquare}Nt(s(m))\ \Rightarrow\ \fbox{${\exists}gNt(s(g))$}
\using {\exists}R
\endprooftree
\justifies
[{\blacksquare}Nt(s(m))]\ \Rightarrow\ \fbox{${\langle\rangle}{\exists}gNt(s(g))$}
\using {\langle\rangle}R
\endprooftree
\prooftree
\justifies
\mbox{\fbox{$Sf$}}\ \Rightarrow\ Sf
\endprooftree
\justifies
[{\blacksquare}Nt(s(m))], \mbox{\fbox{${\langle\rangle}{\exists}gNt(s(g))\backslash Sf$}}\ \Rightarrow\ Sf
\using {\backslash}L
\endprooftree
\justifies
[{\blacksquare}Nt(s(m))], \mbox{\fbox{$({\langle\rangle}{\exists}gNt(s(g))\backslash Sf)/({\exists}aNa{\oplus}({\exists}g(({\it CN}{\it g}/{\it CN}{\it g}){\sqcup}({\it CN}{\it g}\backslash {\it CN}{\it g})){-}I))$}}, {\square}{\forall}n({\it CN}{\it n}/{\it CN}{\it n})\ \Rightarrow\ Sf
\using {/}L
\endprooftree
\justifies
[{\blacksquare}Nt(s(m))], \mbox{\fbox{${\blacksquare}(({\langle\rangle}{\exists}gNt(s(g))\backslash Sf)/({\exists}aNa{\oplus}({\exists}g(({\it CN}{\it g}/{\it CN}{\it g}){\sqcup}({\it CN}{\it g}\backslash {\it CN}{\it g})){-}I)))$}}, {\square}{\forall}n({\it CN}{\it n}/{\it CN}{\it n})\ \Rightarrow\ Sf
\using {\blacksquare}L
\endprooftree}
\end{center}
\caption{Derivation for \lingform{Bond is teetotal}},
\label{bit}
\end{figure}
After elimination of its outer modality, the copula is applied to its adjectival complement.
Semantically active additive disjunction right selects the second disjunct.
The difference right rule checks that the antecedent is not empty, but this is not displayed.
Exists right substitutes the existentially quantified variable for a metavariable
$A$ and semantically inactive additive disjunction right then selects
the adjectival disjunct.
The following semantics is delivered:
\disp{
$({\it Pres}\ (\mbox{\v{}}{\it teetotal}\ {\it b}))$}

\chapter{The Montague Fragment}

\label{monfrag}

In this chapter we give derivations of the Montague grammar
fragment examples analysed in Chapter~7
of Dowty, Wall and Peters (1981\cite{dwp:81}), DWP.
This \scare{Montague test} was performed
on the 9th of October 2015 at the Colloque de Sintaxe et S\'emantique \`a Paris.
The example sentences are shown in Figure~\ref{dwpexfig}.
\begin{figure}
{
str(dwp('(7-7)'), [b([john]), walks], s(f)).

str(dwp('(7-16)'), [b([every, man]), talks], s(f)).

str(dwp('(7-19)'), [b([the, fish]), walks], s(f)).

str(dwp('(7-32)'), [b([every, man]), b([b([walks, or, talks])])], s(f)).

str(dwp('(7-34)'), [b([b([b([every, man]), walks, or, b([every, man]), talks])])], s(f)).

str(dwp('(7-39)'), [b([b([b([a, woman]), walks, and, b([she]), talks])])], s(f)).

str(dwp('(7-43, 45)'), [b([john]), believes, that, b([a, fish]), walks], s(f)). 

str(dwp('(7-48, 49, 52)'), [b([every, man]), believes, that, b([a, fish]), walks], s(f)).

str(dwp('(7-57)'), [b([every, fish, such, that, b([it]), walks]), talks], s(f)).

str(dwp('(7-60, 62)'), [b([john]), seeks, a, unicorn], s(f)).

str(dwp('(7-73)'), [b([john]), is, bill], s(f)).

str(dwp('(7-76)'), [b([john]), is, a, man], s(f)).

str(dwp('(7-83)'), [necessarily, b([john]), walks], s(f)).

str(dwp('(7-86)'), [b([john]), walks, slowly], s(f)).

str(dwp('(7-91)'), [b([john]), tries, to, walk], s(f)).

str(dwp('(7-94)'), [b([john]), tries, to, b([b([catch, a, fish, and, eat, it])])], s(f)).

str(dwp('(7-98)'), [b([john]), finds, a, unicorn], s(f)).

str(dwp('(7-105)'), [b([every, man, such, that, b([he]), loves, a, woman]), loses, her], s(f)).

str(dwp('(7-110)'), [b([john]), walks, in, a, park], s(f)).

str(dwp('(7-116, 118)'), [b([every, man]), doesnt, walk], s(f)).
}
\caption{The mini-corpus of the Montague test}
\label{dwpexfig}
\end{figure}
The lexicon is shown in Figure~\ref{dwplexfig}.
\begin{figure}
{\footnotesize
{\bf a}: ${\blacksquare}{\forall}g({\forall}f((Sf{{\uparrow}{}}{\blacksquare}Nt(s(g))){{\downarrow}{}}Sf)/{\CN}{s(g)}): \lambda A\lambda B\exists C[({A}\ {C})\wedge ({B}\ {C})]$\\
{\bf and}: ${\blacksquare}{\forall}f((?{\blacksquare}Sf\backslash {[]^{-1}}{[]^{-1}}Sf)/{\blacksquare}Sf): (\Phinplus{}\ {0}\ {and})$\\
{\bf and}: ${\blacksquare}{\forall}a{\forall}f((?{\blacksquare}({\langle\rangle}Na\backslash Sf)\backslash {[]^{-1}}{[]^{-1}}({\langle\rangle}Na\backslash Sf))/{\blacksquare}({\langle\rangle}Na\backslash Sf)): (\Phinplus{}\ ({s}\ {0})\ {and})$\\
{\bf believes}: ${\square}(({\langle\rangle}{\exists}gNt(s(g))\backslash Sf)/({CP}that{\sqcup}{\square}Sf)): \mbox{\^{}}\lambda A\lambda B({Pres}\ ((\mbox{\v{}}{believe}\ {A})\ {B}))$\\
{\bf bill}: ${\blacksquare}Nt(s(m)):  {b}$\\
{\bf catch}: ${\square}(({\langle\rangle}{\exists}aNa\backslash Sb)/{\exists}aNa): \mbox{\^{}}\lambda A\lambda B((\mbox{\v{}}{catch}\ {A})\ {B})$\\
{\bf doesnt}: ${\blacksquare}{\forall}g{\forall}a((Sg{{\uparrow}{}}(({\langle\rangle}Na\backslash Sf)/({\langle\rangle}Na\backslash Sb))){{\downarrow}{}}Sg): \lambda A\neg ({A}\ \lambda B\lambda C({B}\ {C}))$\\
{\bf eat}: ${\square}(({\langle\rangle}{\exists}aNa\backslash Sb)/{\exists}aNa): \mbox{\^{}}\lambda A\lambda B((\mbox{\v{}}{eat}\ {A})\ {B})$\\
{\bf every}: ${\blacksquare}{\forall}g({\forall}f((Sf{{\uparrow}{}}Nt(s(g))){{\downarrow}{}}Sf)/{\CN}{s(g)}): \lambda A\lambda B\forall C[({A}\ {C})\rightarrow ({B}\ {C})]$\\
{\bf finds}: ${\square}(({\langle\rangle}{\exists}gNt(s(g))\backslash Sf)/{\exists}aNa): \mbox{\^{}}\lambda A\lambda B({Pres}\ ((\mbox{\v{}}{find}\ {A})\ {B}))$\\
{\bf fish}: ${\square}{\CN}{s(n)}: {fish}$\\
{\bf he}: ${\blacksquare}{[]^{-1}}{\forall}g(({\blacksquare}Sg{|}{\blacksquare}Nt(s(m)))/({\langle\rangle}Nt(s(m))\backslash Sg)): \lambda A{A}$\\
{\bf her}: ${\blacksquare}{\forall}g{\forall}a((({\langle\rangle}Na\backslash Sg){{\uparrow}{}}{\blacksquare}Nt(s(f))){{\downarrow}{}}({\blacksquare}({\langle\rangle}Na\backslash Sg){|}{\blacksquare}Nt(s(f)))): \lambda A{A}$\\
{\bf in}: ${\square}({\forall}a{\forall}f(({\langle\rangle}Na\backslash Sf)\backslash ({\langle\rangle}Na\backslash Sf))/{\exists}aNa): \mbox{\^{}}\lambda A\lambda B\lambda C((\mbox{\v{}}{in}\ {A})\ ({B}\ {C}))$\\
{\bf is}: ${\blacksquare}(({\langle\rangle}{\exists}gNt(s(g))\backslash Sf)/({\exists}aNa{\oplus}({\exists}g(({\CN}{g}/{\CN}{g}){\sqcup}({\CN}{g}\backslash {\CN}{g})){-}I))): \lambda A\lambda B({Pres}\ ({A}\rightarrow C.[{B}={C}]; D.(({D}\ \lambda E[{E}={B}])\ {B})))$\\
{\bf it}: ${\blacksquare}{\forall}f{\forall}a((({\langle\rangle}Na\backslash Sf){{\uparrow}{}}{\blacksquare}Nt(s(n))){{\downarrow}{}}({\blacksquare}({\langle\rangle}Na\backslash Sf){|}{\blacksquare}Nt(s(n)))): \lambda A{A}$\\
{\bf it}: ${\blacksquare}{[]^{-1}}{\forall}f(({\blacksquare}Sf{|}{\blacksquare}Nt(s(n)))/({\langle\rangle}Nt(s(n))\backslash Sf)): \lambda A{A}$\\
{\bf john}: ${\blacksquare}Nt(s(m)): {j}$\\
{\bf loses}: ${\square}(({\langle\rangle}{\exists}gNt(s(g))\backslash Sf)/{\exists}aNa): \mbox{\^{}}\lambda A\lambda B({Pres}\ ((\mbox{\v{}}{lose}\ {A})\ {B}))$\\
{\bf loves}: ${\square}(({\langle\rangle}{\exists}gNt(s(g))\backslash Sf)/{\exists}aNa): \mbox{\^{}}\lambda A\lambda B({Pres}\ ((\mbox{\v{}}{love}\ {A})\ {B}))$\\
{\bf man}: ${\square}{\CN}{s(m)}: {man}$\\
{\bf necessarily}: ${\blacksquare}(SA/{\square}SA): {Nec}$\\
{\bf or}: ${\blacksquare}{\forall}f((?{\blacksquare}Sf\backslash {[]^{-1}}{[]^{-1}}Sf)/{\blacksquare}Sf): (\Phinplus{}\ {0}\ {or})$\\
{\bf or}: ${\blacksquare}{\forall}a{\forall}f((?{\blacksquare}({\langle\rangle}Na\backslash Sf)\backslash {[]^{-1}}{[]^{-1}}({\langle\rangle}Na\backslash Sf))/{\blacksquare}({\langle\rangle}Na\backslash Sf)): (\Phinplus{}\ ({s}\ {0})\ {or})$\\
{\bf or}: ${\blacksquare}{\forall}f((?{\blacksquare}(Sf/({\langle\rangle}{\exists}gNt(s(g))\backslash Sf))\backslash {[]^{-1}}{[]^{-1}}(Sf/({\langle\rangle}{\exists}gNt(s(g))\backslash Sf)))/{\blacksquare}(Sf/({\langle\rangle}{\exists}gNt(s(g))\backslash Sf))): (\Phinplus{}\ ({s}\ {0})\ {or})$\\
{\bf park}: ${\square}{\CN}{s(n)}: {park}$\\
{\bf seeks}: ${\square}(({\langle\rangle}{\exists}gNt(s(g))\backslash Sf)/{\square}{\forall}a{\forall}f(((Na\backslash Sf)/{\exists}bNb)\backslash (Na\backslash Sf))): \mbox{\^{}}\lambda A\lambda B((\mbox{\v{}}{tries}\ \mbox{\^{}}((\mbox{\v{}}{A}\ \mbox{\v{}}{find})\ {B}))\ {B})$\\
{\bf she}: ${\blacksquare}{[]^{-1}}{\forall}g(({\blacksquare}Sg{|}{\blacksquare}Nt(s(f)))/({\langle\rangle}Nt(s(f))\backslash Sg)): \lambda A{A}$\\
{\bf slowly}: ${\square}{\forall}a{\forall}f({\square}({\langle\rangle}Na\backslash Sf)\backslash ({\langle\rangle}{\square}Na\backslash Sf)): \mbox{\^{}}\lambda A\lambda B(\mbox{\v{}}{slowly}\ \mbox{\^{}}(\mbox{\v{}}{A}\ \mbox{\v{}}{B}))$\\
{\bf such}{+}{\bf that}: ${\blacksquare}{\forall}n(({\CN}{n}\backslash {\CN}{n})/(Sf{|}{\blacksquare}Nt(n))): \lambda A\lambda B\lambda C[({B}\ {C})\wedge ({A}\ {C})]$\\
{\bf talks}: ${\square}({\langle\rangle}{\exists}gNt(s(g))\backslash Sf): \mbox{\^{}}\lambda A({Pres}\ (\mbox{\v{}}{talk}\ {A}))$\\
{\bf that}: ${\blacksquare}({\CP}that/{\square}Sf): \lambda A{A}$\\
{\bf the}: ${\blacksquare}{\forall}n(Nt(n)/{\CN}{n}): \iota$ \\
{\bf to}: ${\blacksquare}(({\PP}{to}/{\exists}aNa){\sqcap}{\forall}n(({\langle\rangle}Nn\backslash Si)/({\langle\rangle}Nn\backslash Sb))): \lambda A{A}$\\
{\bf unicorn}: ${\square}{\CN}{s(n)}: {unicorn}$\\
{\bf walks}: ${\square}({\langle\rangle}{\exists}gNt(s(g))\backslash Sf): \mbox{\^{}}\lambda A({Pres}\ (\mbox{\v{}}{walk}\ {A}))$\\
{\bf woman}: ${\square}{\CN}{s(f)}: {woman}$\\
}
\caption{Lexicon for the Montague test}
\label{dwplexfig}
\end{figure}
The first example is as follows, the same as example~(\ref{ex12}) of Chapter~\ref{initexchap}.
(We continue to include the indexation of CatLog2,
which contains the numeration of the source,
within the display.)
\disp{
(dwp((7-7))) $[{\bf john}]{+}{\bf walks}: Sf$}
Recall that in our syntactical forms the subjects are bracketed domains
--- implementing that subjects are weak
islands.
Lookup in our lexicon yields the following semantically labelled sequent:
\disp{
$[{\blacksquare}Nt(s(m)): {\it j}], {\square}({\langle\rangle}{\exists}gNt(s(g))\backslash Sf): \mbox{\^{}}\lambda A({\it Pres}\ (\mbox{\v{}}{\it walk}\ {\it A}))\ \Rightarrow\ Sf$}
As always the lexical types are semantically modalized outermost
--- implementing that word meanings are intensions/senses;
the modality of the proper name subject is semantically inactive
(proper names are rigid designators), while the modality of the tensed verb is
semantically active (the interpretation of tensed verbs depends on the temporal
reference points).
The verb projects a finite sentence (feature $f$) when it combines with a third person
singular (bracketed) subject of any gender;
the actual subject is masculine (feature $m$).

The derivation is as follows:
\vspace{0.15in}
$$
\prooftree
\prooftree
\prooftree
\prooftree
\prooftree
\prooftree
\justifies
\mbox{\fbox{$Nt(s(m))$}}\ \Rightarrow\ Nt(s(m))
\endprooftree
\justifies
\mbox{\fbox{${\blacksquare}Nt(s(m))$}}\ \Rightarrow\ Nt(s(m))
\using {\blacksquare}L
\endprooftree
\justifies
{\blacksquare}Nt(s(m))\ \Rightarrow\ \fbox{${\exists}gNt(s(g))$}
\using {\exists}R
\endprooftree
\justifies
[{\blacksquare}Nt(s(m))]\ \Rightarrow\ \fbox{${\langle\rangle}{\exists}gNt(s(g))$}
\using {\langle\rangle}R
\endprooftree
\prooftree
\justifies
\mbox{\fbox{$Sf$}}\ \Rightarrow\ Sf
\endprooftree
\justifies
[{\blacksquare}Nt(s(m))], \mbox{\fbox{${\langle\rangle}{\exists}gNt(s(g))\backslash Sf$}}\ \Rightarrow\ Sf
\using {\backslash}L
\endprooftree
\justifies
[{\blacksquare}Nt(s(m))], \mbox{\fbox{${\square}({\langle\rangle}{\exists}gNt(s(g))\backslash Sf)$}}\ \Rightarrow\ Sf
\using {\Box}L
\endprooftree
$$
\vspace{0.15in}
\noindent
The semantics delivered by the derivation of this example is:
\disp{
$({\it Pres}\ (\mbox{\v{}}{\it walk}\ {\it j}))$}

The next example involves a quantifier phrase in subject position:
\disp{
(dwp((7-16))) $[{\bf every}{+}{\bf man}]{+}{\bf talks}: Sf$}
Lookup yields the following semantically labelled sequent:
\disp{
$[{\blacksquare}{\forall}g({\forall}f((Sf{{}{\uparrow}{}}Nt(s(g))){{}{\downarrow}{}}Sf)/{\it CN}{\it s(g)}): \lambda A\lambda B\forall C[({\it A}\ {\it C})\rightarrow ({\it B}\ {\it C})], {\square}{\it CN}{\it s(m)}: {\it man}],\\{\square}({\langle\rangle}{\exists}gNt(s(g))\backslash$ $Sf): \mbox{\^{}}\lambda D({\it Pres}\ (\mbox{\v{}}{\it talk}\ {\it D}))\ \Rightarrow\ Sf$}
The semantic modality of the quantifier is inactive since its semantics is purely logical.
Within its modality the type for the quantifier is a functor seeking a count noun to its right;
the feature variable $g$ transmits gender from the count noun argument to the value of the functor.
The functor yields a generalised quantifier type which will infix at the result of extracting a
nominal in a sentence,
simulating Montague's rule of term insertion, or quantifying in, S14.
The derivation is given below:
\vspace{0.15in}
$$
\prooftree
\prooftree
\prooftree
\prooftree
\prooftree
\justifies
\mbox{\fbox{${\it CN}{\it s(m)}$}}\ \Rightarrow\ {\it CN}{\it s(m)}
\endprooftree
\justifies
\mbox{\fbox{${\square}{\it CN}{\it s(m)}$}}\ \Rightarrow\ {\it CN}{\it s(m)}
\using {\Box}L
\endprooftree
\prooftree
\prooftree
\prooftree
\prooftree
\prooftree
\prooftree
\prooftree
\prooftree
\justifies
Nt(s(m))\ \Rightarrow\ Nt(s(m))
\endprooftree
\justifies
Nt(s(m))\ \Rightarrow\ \fbox{${\exists}gNt(s(g))$}
\using {\exists}R
\endprooftree
\justifies
[Nt(s(m))]\ \Rightarrow\ \fbox{${\langle\rangle}{\exists}gNt(s(g))$}
\using {\langle\rangle}R
\endprooftree
\prooftree
\justifies
\mbox{\fbox{$Sf$}}\ \Rightarrow\ Sf
\endprooftree
\justifies
[Nt(s(m))], \mbox{\fbox{${\langle\rangle}{\exists}gNt(s(g))\backslash Sf$}}\ \Rightarrow\ Sf
\using {\backslash}L
\endprooftree
\justifies
[Nt(s(m))], \mbox{\fbox{${\square}({\langle\rangle}{\exists}gNt(s(g))\backslash Sf)$}}\ \Rightarrow\ Sf
\using {\Box}L
\endprooftree
\justifies
[{\tt 1}], {\square}({\langle\rangle}{\exists}gNt(s(g))\backslash Sf)\ \Rightarrow\ Sf{{}{\uparrow}{}}Nt(s(m))
\using {\uparrow}R
\endprooftree
\prooftree
\justifies
\mbox{\fbox{$Sf$}}\ \Rightarrow\ Sf
\endprooftree
\justifies
[\mbox{\fbox{$(Sf{{}{\uparrow}{}}Nt(s(m))){{}{\downarrow}{}}Sf$}}], {\square}({\langle\rangle}{\exists}gNt(s(g))\backslash Sf)\ \Rightarrow\ Sf
\using {\downarrow}L
\endprooftree
\justifies
[\mbox{\fbox{${\forall}f((Sf{{}{\uparrow}{}}Nt(s(m))){{}{\downarrow}{}}Sf)$}}], {\square}({\langle\rangle}{\exists}gNt(s(g))\backslash Sf)\ \Rightarrow\ Sf
\using {\forall}L
\endprooftree
\justifies
[\mbox{\fbox{${\forall}f((Sf{{}{\uparrow}{}}Nt(s(m))){{}{\downarrow}{}}Sf)/{\it CN}{\it s(m)}$}}, {\square}{\it CN}{\it s(m)}], {\square}({\langle\rangle}{\exists}gNt(s(g))\backslash Sf)\ \Rightarrow\ Sf
\using {/}L
\endprooftree
\justifies
[\mbox{\fbox{${\forall}g({\forall}f((Sf{{}{\uparrow}{}}Nt(s(g))){{}{\downarrow}{}}Sf)/{\it CN}{\it s(g)})$}}, {\square}{\it CN}{\it s(m)}], {\square}({\langle\rangle}{\exists}gNt(s(g))\backslash Sf)\ \Rightarrow\ Sf
\using {\forall}L
\endprooftree
\justifies
[\mbox{\fbox{${\blacksquare}{\forall}g({\forall}f((Sf{{}{\uparrow}{}}Nt(s(g))){{}{\downarrow}{}}Sf)/{\it CN}{\it s(g)})$}}, {\square}{\it CN}{\it s(m)}], {\square}({\langle\rangle}{\exists}gNt(s(g))\backslash Sf)\ \Rightarrow\ Sf
\using {\blacksquare}L
\endprooftree
$$
\vspace{0.15in}
\noindent
This delivers the required semantics:
\disp{
$\forall C[(\mbox{\v{}}{\it man}\ {\it C})\rightarrow ({\it Pres}\ (\mbox{\v{}}{\it talk}\ {\it C}))]$}

The next example of DWP is:
\disp{
(dwp((7-19))) $[{\bf the}{+}{\bf fish}]{+}{\bf walks}: Sf$}
Montague analysed the definite article Russelian style as universal quantification with
unicity, 
but this does not reflect its presuppositional character.
In our grammar we assume that the presupposition of unicity is somehow otherwise given,
and the article maps Hilbert style to a simple nominal with a higher-order iota logical constant.
Lexical lookup yields the following semantically labelled sequent: 
\disp{
$[{\blacksquare}{\forall}n(Nt(n)/{\it CN}{\it n}): \iota , {\square}{\it CN}{\it s(n)}: {\it fish}], {\square}({\langle\rangle}{\exists}gNt(s(g))\backslash Sf): \mbox{\^{}}\lambda A({\it Pres}\ (\mbox{\v{}}{\it walk}\ {\it A}))\ \Rightarrow\ Sf$}
This has the derivation:
\vspace{0.15in}
$$
\prooftree
\prooftree
\prooftree
\prooftree
\prooftree
\prooftree
\prooftree
\prooftree
\prooftree
\justifies
\mbox{\fbox{${\it CN}{\it s(n)}$}}\ \Rightarrow\ {\it CN}{\it s(n)}
\endprooftree
\justifies
\mbox{\fbox{${\square}{\it CN}{\it s(n)}$}}\ \Rightarrow\ {\it CN}{\it s(n)}
\using {\Box}L
\endprooftree
\prooftree
\justifies
\mbox{\fbox{$Nt(s(n))$}}\ \Rightarrow\ Nt(s(n))
\endprooftree
\justifies
\mbox{\fbox{$Nt(s(n))/{\it CN}{\it s(n)}$}}, {\square}{\it CN}{\it s(n)}\ \Rightarrow\ Nt(s(n))
\using {/}L
\endprooftree
\justifies
\mbox{\fbox{${\forall}n(Nt(n)/{\it CN}{\it n})$}}, {\square}{\it CN}{\it s(n)}\ \Rightarrow\ Nt(s(n))
\using {\forall}L
\endprooftree
\justifies
\mbox{\fbox{${\blacksquare}{\forall}n(Nt(n)/{\it CN}{\it n})$}}, {\square}{\it CN}{\it s(n)}\ \Rightarrow\ Nt(s(n))
\using {\blacksquare}L
\endprooftree
\justifies
{\blacksquare}{\forall}n(Nt(n)/{\it CN}{\it n}), {\square}{\it CN}{\it s(n)}\ \Rightarrow\ \fbox{${\exists}gNt(s(g))$}
\using {\exists}R
\endprooftree
\justifies
[{\blacksquare}{\forall}n(Nt(n)/{\it CN}{\it n}), {\square}{\it CN}{\it s(n)}]\ \Rightarrow\ \fbox{${\langle\rangle}{\exists}gNt(s(g))$}
\using {\langle\rangle}R
\endprooftree
\prooftree
\justifies
\mbox{\fbox{$Sf$}}\ \Rightarrow\ Sf
\endprooftree
\justifies
[{\blacksquare}{\forall}n(Nt(n)/{\it CN}{\it n}), {\square}{\it CN}{\it s(n)}], \mbox{\fbox{${\langle\rangle}{\exists}gNt(s(g))\backslash Sf$}}\ \Rightarrow\ Sf
\using {\backslash}L
\endprooftree
\justifies
[{\blacksquare}{\forall}n(Nt(n)/{\it CN}{\it n}), {\square}{\it CN}{\it s(n)}], \mbox{\fbox{${\square}({\langle\rangle}{\exists}gNt(s(g))\backslash Sf)$}}\ \Rightarrow\ Sf
\using {\Box}L
\endprooftree
$$
\vspace{0.15in}
\noindent
The derivation delivers semantics:
\disp{
$({\it Pres}\ (\mbox{\v{}}{\it walk}\ (\iota \ \mbox{\v{}}{\it fish})))$}

The next example involves subject quantification and verb phrase coordination:
\disp{
(dwp((7-32))) $[{\bf every}{+}{\bf man}]{+}[[{\bf walks}{+}{\bf or}{+}{\bf talks}]]: Sf$}
Lexical lookup inserting the disjunctive verb phrase coordinator yields the semantically
labelled sequent:
\disp{
$[{\blacksquare}{\forall}g({\forall}f((Sf{{}{\uparrow}{}}Nt(s(g))){{}{\downarrow}{}}Sf)/{\it CN}{\it s(g)}): \lambda A\lambda B\forall C[({\it A}\ {\it C})\rightarrow ({\it B}\ {\it C})], {\square}{\it CN}{\it s(m)}: {\it man}],\\{}[[{\square}({\langle\rangle}{\exists}gNt(s(g))\backslash$ $Sf): \mbox{\^{}}\lambda D({\it Pres}\ (\mbox{\v{}}{\it walk}\ {\it D})), {\blacksquare}{\forall}a{\forall}f((?{\blacksquare}({\langle\rangle}Na\backslash Sf)\backslash {[]^{-1}}{[]^{-1}}({\langle\rangle}Na\backslash Sf))/\\{}{\blacksquare}({\langle\rangle}Na\backslash Sf)): (\Phinplus{}\ ({\it s}\ {\it 0})\ {\it or}),$ ${\square}({\langle\rangle}{\exists}gNt(s(g))\backslash Sf): \mbox{\^{}}\lambda E({\it Pres}\ (\mbox{\v{}}{\it talk}\ {\it E}))]]\ \Rightarrow\ Sf$}
The combinator lexical semantics
$(\Phinplus\ (s\ 0)\ {\it or})$
of the disjunctive coordinator is such that:
$$((((\Phi^{n+}\ (s\ 0)\ {\it or})\ x)\ [y])\ z) =
(((\Phinplus\ 0\ {\it or})\ (x\ z))\ (\alphaplus{}\ [y]\ z)) =
(((\Phinplus\ 0\ {\it or})\ (x\ z))\ [(y\ z)]) =
{}[(y\ z)\vee(x\ z)]$$ 
See Chapter~\ref{coordchap}.
Syntactically,
the value of the coordinator type projects double brackets,
making the coordinate structure a string island;
the disjuncts are semantically inactive modal domains,
making the coordinate structure a scope island.
The derivation is as follows:
{
\vspace{0.15in}
$$
\resizebox{\textwidth}{!}{
\prooftree
\prooftree
\prooftree
\prooftree
\prooftree
\justifies
\mbox{\fbox{${\it CN}{\it s(m)}$}}\ \Rightarrow\ {\it CN}{\it s(m)}
\endprooftree
\justifies
\mbox{\fbox{${\square}{\it CN}{\it s(m)}$}}\ \Rightarrow\ {\it CN}{\it s(m)}
\using {\Box}L
\endprooftree
\prooftree
\prooftree
\prooftree
\prooftree
\prooftree
\prooftree
\prooftree
\prooftree
\prooftree
\prooftree
\prooftree
\prooftree
\prooftree
\prooftree
\prooftree
\justifies
Nt(s(m))\ \Rightarrow\ Nt(s(m))
\endprooftree
\justifies
Nt(s(m))\ \Rightarrow\ \fbox{${\exists}gNt(s(g))$}
\using {\exists}R
\endprooftree
\justifies
[Nt(s(m))]\ \Rightarrow\ \fbox{${\langle\rangle}{\exists}gNt(s(g))$}
\using {\langle\rangle}R
\endprooftree
\prooftree
\justifies
\mbox{\fbox{$Sf$}}\ \Rightarrow\ Sf
\endprooftree
\justifies
[Nt(s(m))], \mbox{\fbox{${\langle\rangle}{\exists}gNt(s(g))\backslash Sf$}}\ \Rightarrow\ Sf
\using {\backslash}L
\endprooftree
\justifies
[Nt(s(m))], \mbox{\fbox{${\square}({\langle\rangle}{\exists}gNt(s(g))\backslash Sf)$}}\ \Rightarrow\ Sf
\using {\Box}L
\endprooftree
\justifies
{\langle\rangle}Nt(s(m)), {\square}({\langle\rangle}{\exists}gNt(s(g))\backslash Sf)\ \Rightarrow\ Sf
\using {\langle\rangle}L
\endprooftree
\justifies
{\square}({\langle\rangle}{\exists}gNt(s(g))\backslash Sf)\ \Rightarrow\ {\langle\rangle}Nt(s(m))\backslash Sf
\using {\backslash}R
\endprooftree
\justifies
{\square}({\langle\rangle}{\exists}gNt(s(g))\backslash Sf)\ \Rightarrow\ {\blacksquare}({\langle\rangle}Nt(s(m))\backslash Sf)
\using {\blacksquare}R
\endprooftree
\prooftree
\prooftree
\prooftree
\prooftree
\prooftree
\prooftree
\prooftree
\prooftree
\prooftree
\prooftree
\justifies
Nt(s(m))\ \Rightarrow\ Nt(s(m))
\endprooftree
\justifies
Nt(s(m))\ \Rightarrow\ \fbox{${\exists}gNt(s(g))$}
\using {\exists}R
\endprooftree
\justifies
[Nt(s(m))]\ \Rightarrow\ \fbox{${\langle\rangle}{\exists}gNt(s(g))$}
\using {\langle\rangle}R
\endprooftree
\prooftree
\justifies
\mbox{\fbox{$Sf$}}\ \Rightarrow\ Sf
\endprooftree
\justifies
[Nt(s(m))], \mbox{\fbox{${\langle\rangle}{\exists}gNt(s(g))\backslash Sf$}}\ \Rightarrow\ Sf
\using {\backslash}L
\endprooftree
\justifies
[Nt(s(m))], \mbox{\fbox{${\square}({\langle\rangle}{\exists}gNt(s(g))\backslash Sf)$}}\ \Rightarrow\ Sf
\using {\Box}L
\endprooftree
\justifies
{\langle\rangle}Nt(s(m)), {\square}({\langle\rangle}{\exists}gNt(s(g))\backslash Sf)\ \Rightarrow\ Sf
\using {\langle\rangle}L
\endprooftree
\justifies
{\square}({\langle\rangle}{\exists}gNt(s(g))\backslash Sf)\ \Rightarrow\ {\langle\rangle}Nt(s(m))\backslash Sf
\using {\backslash}R
\endprooftree
\justifies
{\square}({\langle\rangle}{\exists}gNt(s(g))\backslash Sf)\ \Rightarrow\ {\blacksquare}({\langle\rangle}Nt(s(m))\backslash Sf)
\using {\blacksquare}R
\endprooftree
\justifies
{\square}({\langle\rangle}{\exists}gNt(s(g))\backslash Sf)\ \Rightarrow\ \fbox{$?{\blacksquare}({\langle\rangle}Nt(s(m))\backslash Sf)$}
\using {?}R
\endprooftree
\prooftree
\prooftree
\prooftree
\prooftree
\prooftree
\justifies
Nt(s(m))\ \Rightarrow\ Nt(s(m))
\endprooftree
\justifies
[Nt(s(m))]\ \Rightarrow\ \fbox{${\langle\rangle}Nt(s(m))$}
\using {\langle\rangle}R
\endprooftree
\prooftree
\justifies
\mbox{\fbox{$Sf$}}\ \Rightarrow\ Sf
\endprooftree
\justifies
[Nt(s(m))], \mbox{\fbox{${\langle\rangle}Nt(s(m))\backslash Sf$}}\ \Rightarrow\ Sf
\using {\backslash}L
\endprooftree
\justifies
[Nt(s(m))], [\mbox{\fbox{${[]^{-1}}({\langle\rangle}Nt(s(m))\backslash Sf)$}}]\ \Rightarrow\ Sf
\using {[]^{-1}}L
\endprooftree
\justifies
[Nt(s(m))], [[\mbox{\fbox{${[]^{-1}}{[]^{-1}}({\langle\rangle}Nt(s(m))\backslash Sf)$}}]]\ \Rightarrow\ Sf
\using {[]^{-1}}L
\endprooftree
\justifies
[Nt(s(m))], [[{\square}({\langle\rangle}{\exists}gNt(s(g))\backslash Sf), \mbox{\fbox{$?{\blacksquare}({\langle\rangle}Nt(s(m))\backslash Sf)\backslash {[]^{-1}}{[]^{-1}}({\langle\rangle}Nt(s(m))\backslash Sf)$}}]]\ \Rightarrow\ Sf
\using {\backslash}L
\endprooftree
\justifies
[Nt(s(m))], [[{\square}({\langle\rangle}{\exists}gNt(s(g))\backslash Sf), \mbox{\fbox{$(?{\blacksquare}({\langle\rangle}Nt(s(m))\backslash Sf)\backslash {[]^{-1}}{[]^{-1}}({\langle\rangle}Nt(s(m))\backslash Sf))/{\blacksquare}({\langle\rangle}Nt(s(m))\backslash Sf)$}}, {\square}({\langle\rangle}{\exists}gNt(s(g))\backslash Sf)]]\ \Rightarrow\ Sf
\using {/}L
\endprooftree
\justifies
[Nt(s(m))], [[{\square}({\langle\rangle}{\exists}gNt(s(g))\backslash Sf), \mbox{\fbox{${\forall}f((?{\blacksquare}({\langle\rangle}Nt(s(m))\backslash Sf)\backslash {[]^{-1}}{[]^{-1}}({\langle\rangle}Nt(s(m))\backslash Sf))/{\blacksquare}({\langle\rangle}Nt(s(m))\backslash Sf))$}}, {\square}({\langle\rangle}{\exists}gNt(s(g))\backslash Sf)]]\ \Rightarrow\ Sf
\using {\forall}L
\endprooftree
\justifies
[Nt(s(m))], [[{\square}({\langle\rangle}{\exists}gNt(s(g))\backslash Sf), \mbox{\fbox{${\forall}a{\forall}f((?{\blacksquare}({\langle\rangle}Na\backslash Sf)\backslash {[]^{-1}}{[]^{-1}}({\langle\rangle}Na\backslash Sf))/{\blacksquare}({\langle\rangle}Na\backslash Sf))$}}, {\square}({\langle\rangle}{\exists}gNt(s(g))\backslash Sf)]]\ \Rightarrow\ Sf
\using {\forall}L
\endprooftree
\justifies
[Nt(s(m))], [[{\square}({\langle\rangle}{\exists}gNt(s(g))\backslash Sf), \mbox{\fbox{${\blacksquare}{\forall}a{\forall}f((?{\blacksquare}({\langle\rangle}Na\backslash Sf)\backslash {[]^{-1}}{[]^{-1}}({\langle\rangle}Na\backslash Sf))/{\blacksquare}({\langle\rangle}Na\backslash Sf))$}}, {\square}({\langle\rangle}{\exists}gNt(s(g))\backslash Sf)]]\ \Rightarrow\ Sf
\using {\blacksquare}L
\endprooftree
\justifies
[{\tt 1}], [[{\square}({\langle\rangle}{\exists}gNt(s(g))\backslash Sf), {\blacksquare}{\forall}a{\forall}f((?{\blacksquare}({\langle\rangle}Na\backslash Sf)\backslash {[]^{-1}}{[]^{-1}}({\langle\rangle}Na\backslash Sf))/{\blacksquare}({\langle\rangle}Na\backslash Sf)), {\square}({\langle\rangle}{\exists}gNt(s(g))\backslash Sf)]]\ \Rightarrow\ Sf{{}{\uparrow}{}}Nt(s(m))
\using {\uparrow}R
\endprooftree
\prooftree
\justifies
\mbox{\fbox{$Sf$}}\ \Rightarrow\ Sf
\endprooftree
\justifies
[\mbox{\fbox{$(Sf{{}{\uparrow}{}}Nt(s(m))){{}{\downarrow}{}}Sf$}}], [[{\square}({\langle\rangle}{\exists}gNt(s(g))\backslash Sf), {\blacksquare}{\forall}a{\forall}f((?{\blacksquare}({\langle\rangle}Na\backslash Sf)\backslash {[]^{-1}}{[]^{-1}}({\langle\rangle}Na\backslash Sf))/{\blacksquare}({\langle\rangle}Na\backslash Sf)), {\square}({\langle\rangle}{\exists}gNt(s(g))\backslash Sf)]]\ \Rightarrow\ Sf
\using {\downarrow}L
\endprooftree
\justifies
[\mbox{\fbox{${\forall}f((Sf{{}{\uparrow}{}}Nt(s(m))){{}{\downarrow}{}}Sf)$}}], [[{\square}({\langle\rangle}{\exists}gNt(s(g))\backslash Sf), {\blacksquare}{\forall}a{\forall}f((?{\blacksquare}({\langle\rangle}Na\backslash Sf)\backslash {[]^{-1}}{[]^{-1}}({\langle\rangle}Na\backslash Sf))/{\blacksquare}({\langle\rangle}Na\backslash Sf)), {\square}({\langle\rangle}{\exists}gNt(s(g))\backslash Sf)]]\ \Rightarrow\ Sf
\using {\forall}L
\endprooftree
\justifies
[\mbox{\fbox{${\forall}f((Sf{{}{\uparrow}{}}Nt(s(m))){{}{\downarrow}{}}Sf)/{\it CN}{\it s(m)}$}}, {\square}{\it CN}{\it s(m)}], [[{\square}({\langle\rangle}{\exists}gNt(s(g))\backslash Sf), {\blacksquare}{\forall}a{\forall}f((?{\blacksquare}({\langle\rangle}Na\backslash Sf)\backslash {[]^{-1}}{[]^{-1}}({\langle\rangle}Na\backslash Sf))/{\blacksquare}({\langle\rangle}Na\backslash Sf)), {\square}({\langle\rangle}{\exists}gNt(s(g))\backslash Sf)]]\ \Rightarrow\ Sf
\using {/}L
\endprooftree
\justifies
[\mbox{\fbox{${\forall}g({\forall}f((Sf{{}{\uparrow}{}}Nt(s(g))){{}{\downarrow}{}}Sf)/{\it CN}{\it s(g)})$}}, {\square}{\it CN}{\it s(m)}], [[{\square}({\langle\rangle}{\exists}gNt(s(g))\backslash Sf), {\blacksquare}{\forall}a{\forall}f((?{\blacksquare}({\langle\rangle}Na\backslash Sf)\backslash {[]^{-1}}{[]^{-1}}({\langle\rangle}Na\backslash Sf))/{\blacksquare}({\langle\rangle}Na\backslash Sf)), {\square}({\langle\rangle}{\exists}gNt(s(g))\backslash Sf)]]\ \Rightarrow\ Sf
\using {\forall}L
\endprooftree
\justifies
[\mbox{\fbox{${\blacksquare}{\forall}g({\forall}f((Sf{{}{\uparrow}{}}Nt(s(g))){{}{\downarrow}{}}Sf)/{\it CN}{\it s(g)})$}}, {\square}{\it CN}{\it s(m)}], [[{\square}({\langle\rangle}{\exists}gNt(s(g))\backslash Sf), {\blacksquare}{\forall}a{\forall}f((?{\blacksquare}({\langle\rangle}Na\backslash Sf)\backslash {[]^{-1}}{[]^{-1}}({\langle\rangle}Na\backslash Sf))/{\blacksquare}({\langle\rangle}Na\backslash Sf)), {\square}({\langle\rangle}{\exists}gNt(s(g))\backslash Sf)]]\ \Rightarrow\ Sf
\using {\blacksquare}L
\endprooftree}
$$
\vspace{0.15in}}
\noindent
This delivers semantics:
\disp{
$\forall C[(\mbox{\v{}}{\it man}\ {\it C})\rightarrow [({\it Pres}\ (\mbox{\v{}}{\it walk}\ {\it C}))\vee ({\it Pres}\ (\mbox{\v{}}{\it talk}\ {\it C}))]]$}

The next example is:
\disp{
(dwp((7-34))) $[[[{\bf every}{+}{\bf man}]{+}{\bf walks}{+}{\bf or}{+}[{\bf every}{+}{\bf man}]{+}{\bf talks}]]: Sf$}
This involves sentential coordination:
\disp{
$[[[{\blacksquare}{\forall}g({\forall}f((Sf{{}{\uparrow}{}}Nt(s(g))){{}{\downarrow}{}}Sf)/{\it CN}{\it s(g)}): \lambda A\lambda B\forall C[({\it A}\ {\it C})\rightarrow ({\it B}\ {\it C})], {\square}{\it CN}{\it s(m)}: {\it man}],\\{}{\square}({\langle\rangle}{\exists}gNt(s(g))\backslash Sf): \mbox{\^{}}\lambda D({\it Pres}\ (\mbox{\v{}}{\it walk}\ {\it D})), {\blacksquare}{\forall}f((?{\blacksquare}Sf\backslash {[]^{-1}}{[]^{-1}}Sf)/{\blacksquare}Sf): (\Phinplus{}\ {\it 0}\ {\it or}),\\{}[{\blacksquare}{\forall}g({\forall}f((Sf{{}{\uparrow}{}}Nt(s(g))){{}{\downarrow}{}}Sf)/{\it CN}{\it s(g)}): \lambda E\lambda F\forall G[({\it E}\ {\it G})\rightarrow ({\it F}\ {\it G})], {\square}{\it CN}{\it s(m)}: {\it man}],\\{}{\square}({\langle\rangle}{\exists}gNt(s(g))\backslash Sf): \mbox{\^{}}\lambda H({\it Pres}\ (\mbox{\v{}}{\it talk}\ {\it H}))]]\ \Rightarrow\ Sf$}
Because the disjuncts are (semantically inactive) modal domains the quantifier phrases
can only take scope within their disjuncts.
The derivation is as follows:
\vspace{0.15in}
$$
\rotatebox{-90}{\tiny
\prooftree
\prooftree
\prooftree
\prooftree
\prooftree
\prooftree
\prooftree
\prooftree
\prooftree
\justifies
\mbox{\fbox{${\it CN}{\it s(m)}$}}\ \Rightarrow\ {\it CN}{\it s(m)}
\endprooftree
\justifies
\mbox{\fbox{${\square}{\it CN}{\it s(m)}$}}\ \Rightarrow\ {\it CN}{\it s(m)}
\using {\Box}L
\endprooftree
\prooftree
\prooftree
\prooftree
\prooftree
\prooftree
\prooftree
\prooftree
\prooftree
\justifies
Nt(s(m))\ \Rightarrow\ Nt(s(m))
\endprooftree
\justifies
Nt(s(m))\ \Rightarrow\ \fbox{${\exists}gNt(s(g))$}
\using {\exists}R
\endprooftree
\justifies
[Nt(s(m))]\ \Rightarrow\ \fbox{${\langle\rangle}{\exists}gNt(s(g))$}
\using {\langle\rangle}R
\endprooftree
\prooftree
\justifies
\mbox{\fbox{$Sf$}}\ \Rightarrow\ Sf
\endprooftree
\justifies
[Nt(s(m))], \mbox{\fbox{${\langle\rangle}{\exists}gNt(s(g))\backslash Sf$}}\ \Rightarrow\ Sf
\using {\backslash}L
\endprooftree
\justifies
[Nt(s(m))], \mbox{\fbox{${\square}({\langle\rangle}{\exists}gNt(s(g))\backslash Sf)$}}\ \Rightarrow\ Sf
\using {\Box}L
\endprooftree
\justifies
[{\tt 1}], {\square}({\langle\rangle}{\exists}gNt(s(g))\backslash Sf)\ \Rightarrow\ Sf{{}{\uparrow}{}}Nt(s(m))
\using {\uparrow}R
\endprooftree
\prooftree
\justifies
\mbox{\fbox{$Sf$}}\ \Rightarrow\ Sf
\endprooftree
\justifies
[\mbox{\fbox{$(Sf{{}{\uparrow}{}}Nt(s(m))){{}{\downarrow}{}}Sf$}}], {\square}({\langle\rangle}{\exists}gNt(s(g))\backslash Sf)\ \Rightarrow\ Sf
\using {\downarrow}L
\endprooftree
\justifies
[\mbox{\fbox{${\forall}f((Sf{{}{\uparrow}{}}Nt(s(m))){{}{\downarrow}{}}Sf)$}}], {\square}({\langle\rangle}{\exists}gNt(s(g))\backslash Sf)\ \Rightarrow\ Sf
\using {\forall}L
\endprooftree
\justifies
[\mbox{\fbox{${\forall}f((Sf{{}{\uparrow}{}}Nt(s(m))){{}{\downarrow}{}}Sf)/{\it CN}{\it s(m)}$}}, {\square}{\it CN}{\it s(m)}], {\square}({\langle\rangle}{\exists}gNt(s(g))\backslash Sf)\ \Rightarrow\ Sf
\using {/}L
\endprooftree
\justifies
[\mbox{\fbox{${\forall}g({\forall}f((Sf{{}{\uparrow}{}}Nt(s(g))){{}{\downarrow}{}}Sf)/{\it CN}{\it s(g)})$}}, {\square}{\it CN}{\it s(m)}], {\square}({\langle\rangle}{\exists}gNt(s(g))\backslash Sf)\ \Rightarrow\ Sf
\using {\forall}L
\endprooftree
\justifies
[\mbox{\fbox{${\blacksquare}{\forall}g({\forall}f((Sf{{}{\uparrow}{}}Nt(s(g))){{}{\downarrow}{}}Sf)/{\it CN}{\it s(g)})$}}, {\square}{\it CN}{\it s(m)}], {\square}({\langle\rangle}{\exists}gNt(s(g))\backslash Sf)\ \Rightarrow\ Sf
\using {\blacksquare}L
\endprooftree
\justifies
[{\blacksquare}{\forall}g({\forall}f((Sf{{}{\uparrow}{}}Nt(s(g))){{}{\downarrow}{}}Sf)/{\it CN}{\it s(g)}), {\square}{\it CN}{\it s(m)}], {\square}({\langle\rangle}{\exists}gNt(s(g))\backslash Sf)\ \Rightarrow\ {\blacksquare}Sf
\using {\blacksquare}R
\endprooftree
\prooftree
\prooftree
\prooftree
\prooftree
\prooftree
\prooftree
\prooftree
\prooftree
\justifies
\mbox{\fbox{${\it CN}{\it s(m)}$}}\ \Rightarrow\ {\it CN}{\it s(m)}
\endprooftree
\justifies
\mbox{\fbox{${\square}{\it CN}{\it s(m)}$}}\ \Rightarrow\ {\it CN}{\it s(m)}
\using {\Box}L
\endprooftree
\prooftree
\prooftree
\prooftree
\prooftree
\prooftree
\prooftree
\prooftree
\prooftree
\justifies
Nt(s(m))\ \Rightarrow\ Nt(s(m))
\endprooftree
\justifies
Nt(s(m))\ \Rightarrow\ \fbox{${\exists}gNt(s(g))$}
\using {\exists}R
\endprooftree
\justifies
[Nt(s(m))]\ \Rightarrow\ \fbox{${\langle\rangle}{\exists}gNt(s(g))$}
\using {\langle\rangle}R
\endprooftree
\prooftree
\justifies
\mbox{\fbox{$Sf$}}\ \Rightarrow\ Sf
\endprooftree
\justifies
[Nt(s(m))], \mbox{\fbox{${\langle\rangle}{\exists}gNt(s(g))\backslash Sf$}}\ \Rightarrow\ Sf
\using {\backslash}L
\endprooftree
\justifies
[Nt(s(m))], \mbox{\fbox{${\square}({\langle\rangle}{\exists}gNt(s(g))\backslash Sf)$}}\ \Rightarrow\ Sf
\using {\Box}L
\endprooftree
\justifies
[{\tt 1}], {\square}({\langle\rangle}{\exists}gNt(s(g))\backslash Sf)\ \Rightarrow\ Sf{{}{\uparrow}{}}Nt(s(m))
\using {\uparrow}R
\endprooftree
\prooftree
\justifies
\mbox{\fbox{$Sf$}}\ \Rightarrow\ Sf
\endprooftree
\justifies
[\mbox{\fbox{$(Sf{{}{\uparrow}{}}Nt(s(m))){{}{\downarrow}{}}Sf$}}], {\square}({\langle\rangle}{\exists}gNt(s(g))\backslash Sf)\ \Rightarrow\ Sf
\using {\downarrow}L
\endprooftree
\justifies
[\mbox{\fbox{${\forall}f((Sf{{}{\uparrow}{}}Nt(s(m))){{}{\downarrow}{}}Sf)$}}], {\square}({\langle\rangle}{\exists}gNt(s(g))\backslash Sf)\ \Rightarrow\ Sf
\using {\forall}L
\endprooftree
\justifies
[\mbox{\fbox{${\forall}f((Sf{{}{\uparrow}{}}Nt(s(m))){{}{\downarrow}{}}Sf)/{\it CN}{\it s(m)}$}}, {\square}{\it CN}{\it s(m)}], {\square}({\langle\rangle}{\exists}gNt(s(g))\backslash Sf)\ \Rightarrow\ Sf
\using {/}L
\endprooftree
\justifies
[\mbox{\fbox{${\forall}g({\forall}f((Sf{{}{\uparrow}{}}Nt(s(g))){{}{\downarrow}{}}Sf)/{\it CN}{\it s(g)})$}}, {\square}{\it CN}{\it s(m)}], {\square}({\langle\rangle}{\exists}gNt(s(g))\backslash Sf)\ \Rightarrow\ Sf
\using {\forall}L
\endprooftree
\justifies
[\mbox{\fbox{${\blacksquare}{\forall}g({\forall}f((Sf{{}{\uparrow}{}}Nt(s(g))){{}{\downarrow}{}}Sf)/{\it CN}{\it s(g)})$}}, {\square}{\it CN}{\it s(m)}], {\square}({\langle\rangle}{\exists}gNt(s(g))\backslash Sf)\ \Rightarrow\ Sf
\using {\blacksquare}L
\endprooftree
\justifies
[{\blacksquare}{\forall}g({\forall}f((Sf{{}{\uparrow}{}}Nt(s(g))){{}{\downarrow}{}}Sf)/{\it CN}{\it s(g)}), {\square}{\it CN}{\it s(m)}], {\square}({\langle\rangle}{\exists}gNt(s(g))\backslash Sf)\ \Rightarrow\ {\blacksquare}Sf
\using {\blacksquare}R
\endprooftree
\justifies
[{\blacksquare}{\forall}g({\forall}f((Sf{{}{\uparrow}{}}Nt(s(g))){{}{\downarrow}{}}Sf)/{\it CN}{\it s(g)}), {\square}{\it CN}{\it s(m)}], {\square}({\langle\rangle}{\exists}gNt(s(g))\backslash Sf)\ \Rightarrow\ \fbox{$?{\blacksquare}Sf$}
\using {?}R
\endprooftree
\prooftree
\prooftree
\prooftree
\justifies
\mbox{\fbox{$Sf$}}\ \Rightarrow\ Sf
\endprooftree
\justifies
[\mbox{\fbox{${[]^{-1}}Sf$}}]\ \Rightarrow\ Sf
\using {[]^{-1}}L
\endprooftree
\justifies
[[\mbox{\fbox{${[]^{-1}}{[]^{-1}}Sf$}}]]\ \Rightarrow\ Sf
\using {[]^{-1}}L
\endprooftree
\justifies
[[[{\blacksquare}{\forall}g({\forall}f((Sf{{}{\uparrow}{}}Nt(s(g))){{}{\downarrow}{}}Sf)/{\it CN}{\it s(g)}), {\square}{\it CN}{\it s(m)}], {\square}({\langle\rangle}{\exists}gNt(s(g))\backslash Sf), \mbox{\fbox{$?{\blacksquare}Sf\backslash {[]^{-1}}{[]^{-1}}Sf$}}]]\ \Rightarrow\ Sf
\using {\backslash}L
\endprooftree
\justifies
[[[{\blacksquare}{\forall}g({\forall}f((Sf{{}{\uparrow}{}}Nt(s(g))){{}{\downarrow}{}}Sf)/{\it CN}{\it s(g)}), {\square}{\it CN}{\it s(m)}], {\square}({\langle\rangle}{\exists}gNt(s(g))\backslash Sf), \mbox{\fbox{$(?{\blacksquare}Sf\backslash {[]^{-1}}{[]^{-1}}Sf)/{\blacksquare}Sf$}}, [{\blacksquare}{\forall}g({\forall}f((Sf{{}{\uparrow}{}}Nt(s(g))){{}{\downarrow}{}}Sf)/{\it CN}{\it s(g)}), {\square}{\it CN}{\it s(m)}], {\square}({\langle\rangle}{\exists}gNt(s(g))\backslash Sf)]]\ \Rightarrow\ Sf
\using {/}L
\endprooftree
\justifies
[[[{\blacksquare}{\forall}g({\forall}f((Sf{{}{\uparrow}{}}Nt(s(g))){{}{\downarrow}{}}Sf)/{\it CN}{\it s(g)}), {\square}{\it CN}{\it s(m)}], {\square}({\langle\rangle}{\exists}gNt(s(g))\backslash Sf), \mbox{\fbox{${\forall}f((?{\blacksquare}Sf\backslash {[]^{-1}}{[]^{-1}}Sf)/{\blacksquare}Sf)$}}, [{\blacksquare}{\forall}g({\forall}f((Sf{{}{\uparrow}{}}Nt(s(g))){{}{\downarrow}{}}Sf)/{\it CN}{\it s(g)}), {\square}{\it CN}{\it s(m)}], {\square}({\langle\rangle}{\exists}gNt(s(g))\backslash Sf)]]\ \Rightarrow\ Sf
\using {\forall}L
\endprooftree
\justifies
[[[{\blacksquare}{\forall}g({\forall}f((Sf{{}{\uparrow}{}}Nt(s(g))){{}{\downarrow}{}}Sf)/{\it CN}{\it s(g)}), {\square}{\it CN}{\it s(m)}], {\square}({\langle\rangle}{\exists}gNt(s(g))\backslash Sf), \mbox{\fbox{${\blacksquare}{\forall}f((?{\blacksquare}Sf\backslash {[]^{-1}}{[]^{-1}}Sf)/{\blacksquare}Sf)$}}, [{\blacksquare}{\forall}g({\forall}f((Sf{{}{\uparrow}{}}Nt(s(g))){{}{\downarrow}{}}Sf)/{\it CN}{\it s(g)}), {\square}{\it CN}{\it s(m)}], {\square}({\langle\rangle}{\exists}gNt(s(g))\backslash Sf)]]\ \Rightarrow\ Sf
\using {\blacksquare}L
\endprooftree}
$$
\vspace{0.15in}

The semantics assigned is:
\disp{
$[\forall H[(\mbox{\v{}}{\it man}\ {\it H})\rightarrow ({\it Pres}\ (\mbox{\v{}}{\it walk}\ {\it H}))]\vee \forall C[(\mbox{\v{}}{\it man}\ {\it C})\rightarrow ({\it Pres}\ (\mbox{\v{}}{\it talk}\ {\it C}))]]$}

The next example involves anaphora and sentential conjunction:
\disp{
(dwp((7-39))) $[[[{\bf a}{+}{\bf woman}]{+}{\bf walks}{+}{\bf and}{+}[{\bf she}]{+}{\bf talks}]]: Sf$}
Our lexical type assignment to the nominative pronoun ensures that it combines to
the right with a verb phrase,
i.e.\ that it occupies subject position:
\disp{
$[[[{\blacksquare}{\forall}g({\forall}f((Sf{{}{\uparrow}{}}{\blacksquare}Nt(s(g))){{}{\downarrow}{}}Sf)/{\it CN}{\it s(g)}): \lambda A\lambda B\exists C[({\it A}\ {\it C})\wedge ({\it B}\ {\it C})], {\square}{\it CN}{\it s(f)}: {\it woman}],\\{\square}({\langle\rangle}{\exists}gNt(s(g))\backslash$ $Sf): \mbox{\^{}}\lambda D({\it Pres}\ (\mbox{\v{}}{\it walk}\ {\it D})), {\blacksquare}{\forall}f((?{\blacksquare}Sf\backslash {[]^{-1}}{[]^{-1}}Sf)/{\blacksquare}Sf): (\Phinplus{}\ {\it 0}\ {\it and}),\\{}[{\blacksquare}{[]^{-1}}{\forall}g(({\blacksquare}Sg{|}{\blacksquare}Nt(s(f)))/$ $({\langle\rangle}Nt(s(f))\backslash Sg)): \lambda E{\it E}],
{\square}({\langle\rangle}{\exists}gNt(s(g))\backslash Sf): \mbox{\^{}}\lambda F({\it Pres}\ (\mbox{\v{}}{\it talk}\ {\it F}))]]\\\Rightarrow\ Sf$}
The limited contraction for anaphora type constructor is used to create anaphoric
dependency.
There is the following derivation where the treatment of quantification and anaphora
(and coordination) interact in such a way that the pronoun is bound by the quantificational
noun phrase antecedent.
\vspace{0.15in}
$$
\rotatebox{-90}{\tiny
\prooftree
\prooftree
\prooftree
\prooftree
\prooftree
\justifies
\mbox{\fbox{${\it CN}{\it s(f)}$}}\ \Rightarrow\ {\it CN}{\it s(f)}
\endprooftree
\justifies
\mbox{\fbox{${\square}{\it CN}{\it s(f)}$}}\ \Rightarrow\ {\it CN}{\it s(f)}
\using {\Box}L
\endprooftree
\prooftree
\prooftree
\prooftree
\prooftree
\prooftree
\prooftree
\prooftree
\prooftree
\prooftree
\prooftree
\prooftree
\prooftree
\prooftree
\prooftree
\justifies
Nt(s(f))\ \Rightarrow\ Nt(s(f))
\endprooftree
\justifies
Nt(s(f))\ \Rightarrow\ \fbox{${\exists}gNt(s(g))$}
\using {\exists}R
\endprooftree
\justifies
[Nt(s(f))]\ \Rightarrow\ \fbox{${\langle\rangle}{\exists}gNt(s(g))$}
\using {\langle\rangle}R
\endprooftree
\prooftree
\justifies
\mbox{\fbox{$Sf$}}\ \Rightarrow\ Sf
\endprooftree
\justifies
[Nt(s(f))], \mbox{\fbox{${\langle\rangle}{\exists}gNt(s(g))\backslash Sf$}}\ \Rightarrow\ Sf
\using {\backslash}L
\endprooftree
\justifies
[Nt(s(f))], \mbox{\fbox{${\square}({\langle\rangle}{\exists}gNt(s(g))\backslash Sf)$}}\ \Rightarrow\ Sf
\using {\Box}L
\endprooftree
\justifies
{\langle\rangle}Nt(s(f)), {\square}({\langle\rangle}{\exists}gNt(s(g))\backslash Sf)\ \Rightarrow\ Sf
\using {\langle\rangle}L
\endprooftree
\justifies
{\square}({\langle\rangle}{\exists}gNt(s(g))\backslash Sf)\ \Rightarrow\ {\langle\rangle}Nt(s(f))\backslash Sf
\using {\backslash}R
\endprooftree
\prooftree
\prooftree
\prooftree
\prooftree
\justifies
\mbox{\fbox{$Nt(s(f))$}}\ \Rightarrow\ Nt(s(f))
\endprooftree
\justifies
\mbox{\fbox{${\blacksquare}Nt(s(f))$}}\ \Rightarrow\ Nt(s(f))
\using {\blacksquare}L
\endprooftree
\justifies
{\blacksquare}Nt(s(f))\ \Rightarrow\ {\blacksquare}Nt(s(f))
\using {\blacksquare}R
\endprooftree
\prooftree
\prooftree
\prooftree
\prooftree
\prooftree
\prooftree
\justifies
\mbox{\fbox{$Sf$}}\ \Rightarrow\ Sf
\endprooftree
\justifies
\mbox{\fbox{${\blacksquare}Sf$}}\ \Rightarrow\ Sf
\using {\blacksquare}L
\endprooftree
\justifies
{\blacksquare}Sf\ \Rightarrow\ {\blacksquare}Sf
\using {\blacksquare}R
\endprooftree
\prooftree
\prooftree
\prooftree
\prooftree
\prooftree
\prooftree
\prooftree
\prooftree
\prooftree
\justifies
\mbox{\fbox{$Nt(s(f))$}}\ \Rightarrow\ Nt(s(f))
\endprooftree
\justifies
\mbox{\fbox{${\blacksquare}Nt(s(f))$}}\ \Rightarrow\ Nt(s(f))
\using {\blacksquare}L
\endprooftree
\justifies
{\blacksquare}Nt(s(f))\ \Rightarrow\ \fbox{${\exists}gNt(s(g))$}
\using {\exists}R
\endprooftree
\justifies
[{\blacksquare}Nt(s(f))]\ \Rightarrow\ \fbox{${\langle\rangle}{\exists}gNt(s(g))$}
\using {\langle\rangle}R
\endprooftree
\prooftree
\justifies
\mbox{\fbox{$Sf$}}\ \Rightarrow\ Sf
\endprooftree
\justifies
[{\blacksquare}Nt(s(f))], \mbox{\fbox{${\langle\rangle}{\exists}gNt(s(g))\backslash Sf$}}\ \Rightarrow\ Sf
\using {\backslash}L
\endprooftree
\justifies
[{\blacksquare}Nt(s(f))], \mbox{\fbox{${\square}({\langle\rangle}{\exists}gNt(s(g))\backslash Sf)$}}\ \Rightarrow\ Sf
\using {\Box}L
\endprooftree
\justifies
[{\blacksquare}Nt(s(f))], {\square}({\langle\rangle}{\exists}gNt(s(g))\backslash Sf)\ \Rightarrow\ {\blacksquare}Sf
\using {\blacksquare}R
\endprooftree
\justifies
[{\blacksquare}Nt(s(f))], {\square}({\langle\rangle}{\exists}gNt(s(g))\backslash Sf)\ \Rightarrow\ \fbox{$?{\blacksquare}Sf$}
\using {?}R
\endprooftree
\prooftree
\prooftree
\prooftree
\justifies
\mbox{\fbox{$Sf$}}\ \Rightarrow\ Sf
\endprooftree
\justifies
[\mbox{\fbox{${[]^{-1}}Sf$}}]\ \Rightarrow\ Sf
\using {[]^{-1}}L
\endprooftree
\justifies
[[\mbox{\fbox{${[]^{-1}}{[]^{-1}}Sf$}}]]\ \Rightarrow\ Sf
\using {[]^{-1}}L
\endprooftree
\justifies
[[[{\blacksquare}Nt(s(f))], {\square}({\langle\rangle}{\exists}gNt(s(g))\backslash Sf), \mbox{\fbox{$?{\blacksquare}Sf\backslash {[]^{-1}}{[]^{-1}}Sf$}}]]\ \Rightarrow\ Sf
\using {\backslash}L
\endprooftree
\justifies
[[[{\blacksquare}Nt(s(f))], {\square}({\langle\rangle}{\exists}gNt(s(g))\backslash Sf), \mbox{\fbox{$(?{\blacksquare}Sf\backslash {[]^{-1}}{[]^{-1}}Sf)/{\blacksquare}Sf$}}, {\blacksquare}Sf]]\ \Rightarrow\ Sf
\using {/}L
\endprooftree
\justifies
[[[{\blacksquare}Nt(s(f))], {\square}({\langle\rangle}{\exists}gNt(s(g))\backslash Sf), \mbox{\fbox{${\forall}f((?{\blacksquare}Sf\backslash {[]^{-1}}{[]^{-1}}Sf)/{\blacksquare}Sf)$}}, {\blacksquare}Sf]]\ \Rightarrow\ Sf
\using {\forall}L
\endprooftree
\justifies
[[[{\blacksquare}Nt(s(f))], {\square}({\langle\rangle}{\exists}gNt(s(g))\backslash Sf), \mbox{\fbox{${\blacksquare}{\forall}f((?{\blacksquare}Sf\backslash {[]^{-1}}{[]^{-1}}Sf)/{\blacksquare}Sf)$}}, {\blacksquare}Sf]]\ \Rightarrow\ Sf
\using {\blacksquare}L
\endprooftree
\justifies
[[[{\blacksquare}Nt(s(f))], {\square}({\langle\rangle}{\exists}gNt(s(g))\backslash Sf), {\blacksquare}{\forall}f((?{\blacksquare}Sf\backslash {[]^{-1}}{[]^{-1}}Sf)/{\blacksquare}Sf), \mbox{\fbox{${\blacksquare}Sf{|}{\blacksquare}Nt(s(f))$}}]]\ \Rightarrow\ Sf
\using {|}L
\endprooftree
\justifies
[[[{\blacksquare}Nt(s(f))], {\square}({\langle\rangle}{\exists}gNt(s(g))\backslash Sf), {\blacksquare}{\forall}f((?{\blacksquare}Sf\backslash {[]^{-1}}{[]^{-1}}Sf)/{\blacksquare}Sf), \mbox{\fbox{$({\blacksquare}Sf{|}{\blacksquare}Nt(s(f)))/({\langle\rangle}Nt(s(f))\backslash Sf)$}}, {\square}({\langle\rangle}{\exists}gNt(s(g))\backslash Sf)]]\ \Rightarrow\ Sf
\using {/}L
\endprooftree
\justifies
[[[{\blacksquare}Nt(s(f))], {\square}({\langle\rangle}{\exists}gNt(s(g))\backslash Sf), {\blacksquare}{\forall}f((?{\blacksquare}Sf\backslash {[]^{-1}}{[]^{-1}}Sf)/{\blacksquare}Sf), \mbox{\fbox{${\forall}g(({\blacksquare}Sg{|}{\blacksquare}Nt(s(f)))/({\langle\rangle}Nt(s(f))\backslash Sg))$}}, {\square}({\langle\rangle}{\exists}gNt(s(g))\backslash Sf)]]\ \Rightarrow\ Sf
\using {\forall}L
\endprooftree
\justifies
[[[{\blacksquare}Nt(s(f))], {\square}({\langle\rangle}{\exists}gNt(s(g))\backslash Sf), {\blacksquare}{\forall}f((?{\blacksquare}Sf\backslash {[]^{-1}}{[]^{-1}}Sf)/{\blacksquare}Sf), [\mbox{\fbox{${[]^{-1}}{\forall}g(({\blacksquare}Sg{|}{\blacksquare}Nt(s(f)))/({\langle\rangle}Nt(s(f))\backslash Sg))$}}], {\square}({\langle\rangle}{\exists}gNt(s(g))\backslash Sf)]]\ \Rightarrow\ Sf
\using {[]^{-1}}L
\endprooftree
\justifies
[[[{\blacksquare}Nt(s(f))], {\square}({\langle\rangle}{\exists}gNt(s(g))\backslash Sf), {\blacksquare}{\forall}f((?{\blacksquare}Sf\backslash {[]^{-1}}{[]^{-1}}Sf)/{\blacksquare}Sf), [\mbox{\fbox{${\blacksquare}{[]^{-1}}{\forall}g(({\blacksquare}Sg{|}{\blacksquare}Nt(s(f)))/({\langle\rangle}Nt(s(f))\backslash Sg))$}}], {\square}({\langle\rangle}{\exists}gNt(s(g))\backslash Sf)]]\ \Rightarrow\ Sf
\using {\blacksquare}L
\endprooftree
\justifies
[[[{\tt 1}], {\square}({\langle\rangle}{\exists}gNt(s(g))\backslash Sf), {\blacksquare}{\forall}f((?{\blacksquare}Sf\backslash {[]^{-1}}{[]^{-1}}Sf)/{\blacksquare}Sf), [{\blacksquare}{[]^{-1}}{\forall}g(({\blacksquare}Sg{|}{\blacksquare}Nt(s(f)))/({\langle\rangle}Nt(s(f))\backslash Sg))], {\square}({\langle\rangle}{\exists}gNt(s(g))\backslash Sf)]]\ \Rightarrow\ Sf{{}{\uparrow}{}}{\blacksquare}Nt(s(f))
\using {\uparrow}R
\endprooftree
\prooftree
\justifies
\mbox{\fbox{$Sf$}}\ \Rightarrow\ Sf
\endprooftree
\justifies
[[[\mbox{\fbox{$(Sf{{}{\uparrow}{}}{\blacksquare}Nt(s(f))){{}{\downarrow}{}}Sf$}}], {\square}({\langle\rangle}{\exists}gNt(s(g))\backslash Sf), {\blacksquare}{\forall}f((?{\blacksquare}Sf\backslash {[]^{-1}}{[]^{-1}}Sf)/{\blacksquare}Sf), [{\blacksquare}{[]^{-1}}{\forall}g(({\blacksquare}Sg{|}{\blacksquare}Nt(s(f)))/({\langle\rangle}Nt(s(f))\backslash Sg))], {\square}({\langle\rangle}{\exists}gNt(s(g))\backslash Sf)]]\ \Rightarrow\ Sf
\using {\downarrow}L
\endprooftree
\justifies
[[[\mbox{\fbox{${\forall}f((Sf{{}{\uparrow}{}}{\blacksquare}Nt(s(f))){{}{\downarrow}{}}Sf)$}}], {\square}({\langle\rangle}{\exists}gNt(s(g))\backslash Sf), {\blacksquare}{\forall}f((?{\blacksquare}Sf\backslash {[]^{-1}}{[]^{-1}}Sf)/{\blacksquare}Sf), [{\blacksquare}{[]^{-1}}{\forall}g(({\blacksquare}Sg{|}{\blacksquare}Nt(s(f)))/({\langle\rangle}Nt(s(f))\backslash Sg))], {\square}({\langle\rangle}{\exists}gNt(s(g))\backslash Sf)]]\ \Rightarrow\ Sf
\using {\forall}L
\endprooftree
\justifies
[[[\mbox{\fbox{${\forall}f((Sf{{}{\uparrow}{}}{\blacksquare}Nt(s(f))){{}{\downarrow}{}}Sf)/{\it CN}{\it s(f)}$}}, {\square}{\it CN}{\it s(f)}], {\square}({\langle\rangle}{\exists}gNt(s(g))\backslash Sf), {\blacksquare}{\forall}f((?{\blacksquare}Sf\backslash {[]^{-1}}{[]^{-1}}Sf)/{\blacksquare}Sf), [{\blacksquare}{[]^{-1}}{\forall}g(({\blacksquare}Sg{|}{\blacksquare}Nt(s(f)))/({\langle\rangle}Nt(s(f))\backslash Sg))], {\square}({\langle\rangle}{\exists}gNt(s(g))\backslash Sf)]]\ \Rightarrow\ Sf
\using {/}L
\endprooftree
\justifies
[[[\mbox{\fbox{${\forall}g({\forall}f((Sf{{}{\uparrow}{}}{\blacksquare}Nt(s(g))){{}{\downarrow}{}}Sf)/{\it CN}{\it s(g)})$}}, {\square}{\it CN}{\it s(f)}], {\square}({\langle\rangle}{\exists}gNt(s(g))\backslash Sf), {\blacksquare}{\forall}f((?{\blacksquare}Sf\backslash {[]^{-1}}{[]^{-1}}Sf)/{\blacksquare}Sf), [{\blacksquare}{[]^{-1}}{\forall}g(({\blacksquare}Sg{|}{\blacksquare}Nt(s(f)))/({\langle\rangle}Nt(s(f))\backslash Sg))], {\square}({\langle\rangle}{\exists}gNt(s(g))\backslash Sf)]]\ \Rightarrow\ Sf
\using {\forall}L
\endprooftree
\justifies
[[[\mbox{\fbox{${\blacksquare}{\forall}g({\forall}f((Sf{{}{\uparrow}{}}{\blacksquare}Nt(s(g))){{}{\downarrow}{}}Sf)/{\it CN}{\it s(g)})$}}, {\square}{\it CN}{\it s(f)}], {\square}({\langle\rangle}{\exists}gNt(s(g))\backslash Sf), {\blacksquare}{\forall}f((?{\blacksquare}Sf\backslash {[]^{-1}}{[]^{-1}}Sf)/{\blacksquare}Sf), [{\blacksquare}{[]^{-1}}{\forall}g(({\blacksquare}Sg{|}{\blacksquare}Nt(s(f)))/({\langle\rangle}Nt(s(f))\backslash Sg))], {\square}({\langle\rangle}{\exists}gNt(s(g))\backslash Sf)]]\ \Rightarrow\ Sf
\using {\blacksquare}L
\endprooftree}
$$
\vspace{0.15in}
\noindent
This assigns the correct semantics:
\disp{
$\exists C[(\mbox{\v{}}{\it woman}\ {\it C})\wedge [({\it Pres}\ (\mbox{\v{}}{\it walk}\ {\it C}))\wedge ({\it Pres}\ (\mbox{\v{}}{\it talk}\ {\it C}))]]$}

The next analyses of DWP are for the de re or specific versus de dicto or non-specific
ambiguity of:
\disp{
\vspace{0.15in}
(dwp((7-43, 45))) $[{\bf john}]{+}{\bf believes}{+}{\bf that}{+}[{\bf a}{+}{\bf fish}]{+}{\bf walks}: Sf$}
On Montague's account the indefinite quantifier phrase can quantity in at the matrix level
yielding the de re reading in which the propositional attitude verb is within the scope of
the existential quantification (John's belief is directed towards a specific fish) or it can quantify
in at the level of the subordinate clause in which case the existential quantification is within the
scope of the propositional attitude verb (John has no specific fish in mind).
Our grammar conserves this account.
Lexical lookup yields the semantically labelled sequent:
\disp{
$[{\blacksquare}Nt(s(m)): {\it j}], {\square}(({\langle\rangle}{\exists}gNt(s(g))\backslash Sf)/({\it CP}that{\sqcup}{\square}Sf)): \mbox{\^{}}\lambda A\lambda B({\it Pres}\ ((\mbox{\v{}}{\it believe}\ {\it A})\ {\it B})), {\blacksquare}({\it CP}that/$ ${\square}Sf): \lambda C{\it C}, [{\blacksquare}{\forall}g({\forall}f((Sf{{}{\uparrow}{}}{\blacksquare}Nt(s(g))){{}{\downarrow}{}}Sf)/{\it CN}{\it s(g)}): \lambda D\lambda E\exists F[({\it D}\ {\it F})\wedge ({\it E}\ {\it F})], {\square}{\it CN}{\it s(n)}: {\it fish}],$ ${\square}({\langle\rangle}{\exists}gNt(s(g))\backslash Sf): \mbox{\^{}}\lambda G({\it Pres}\ (\mbox{\v{}}{\it walk}\ {\it G}))\ \Rightarrow\ Sf$}
In Montague grammar all term phrases have the same semantic type,
and this requirement necessitates the type raising of proper names so that they have the same type 
as quantifier phrases;
a meaning postulate is then required to make them rigid designators (i.e.\ denote
the same individual in all worlds).
Note how in our type logical grammar proper names are assigned their lower type,
and our semantically inactive modality lexical semantics encodes that they are intensional,
but rigid designators. 
\vspace{0.15in}
$$
{\tiny
\prooftree
\prooftree
\prooftree
\prooftree
\prooftree
\justifies
\mbox{\fbox{${\it CN}{\it s(n)}$}}\ \Rightarrow\ {\it CN}{\it s(n)}
\endprooftree
\justifies
\mbox{\fbox{${\square}{\it CN}{\it s(n)}$}}\ \Rightarrow\ {\it CN}{\it s(n)}
\using {\Box}L
\endprooftree
\prooftree
\prooftree
\prooftree
\prooftree
\prooftree
\prooftree
\prooftree
\prooftree
\prooftree
\prooftree
\prooftree
\prooftree
\prooftree
\prooftree
\prooftree
\justifies
\mbox{\fbox{$Nt(s(n))$}}\ \Rightarrow\ Nt(s(n))
\endprooftree
\justifies
\mbox{\fbox{${\blacksquare}Nt(s(n))$}}\ \Rightarrow\ Nt(s(n))
\using {\blacksquare}L
\endprooftree
\justifies
{\blacksquare}Nt(s(n))\ \Rightarrow\ \fbox{${\exists}gNt(s(g))$}
\using {\exists}R
\endprooftree
\justifies
[{\blacksquare}Nt(s(n))]\ \Rightarrow\ \fbox{${\langle\rangle}{\exists}gNt(s(g))$}
\using {\langle\rangle}R
\endprooftree
\prooftree
\justifies
\mbox{\fbox{$Sf$}}\ \Rightarrow\ Sf
\endprooftree
\justifies
[{\blacksquare}Nt(s(n))], \mbox{\fbox{${\langle\rangle}{\exists}gNt(s(g))\backslash Sf$}}\ \Rightarrow\ Sf
\using {\backslash}L
\endprooftree
\justifies
[{\blacksquare}Nt(s(n))], \mbox{\fbox{${\square}({\langle\rangle}{\exists}gNt(s(g))\backslash Sf)$}}\ \Rightarrow\ Sf
\using {\Box}L
\endprooftree
\justifies
[{\blacksquare}Nt(s(n))], {\square}({\langle\rangle}{\exists}gNt(s(g))\backslash Sf)\ \Rightarrow\ {\square}Sf
\using {\Box}R
\endprooftree
\prooftree
\justifies
\mbox{\fbox{${\it CP}that$}}\ \Rightarrow\ {\it CP}that
\endprooftree
\justifies
\mbox{\fbox{${\it CP}that/{\square}Sf$}}, [{\blacksquare}Nt(s(n))], {\square}({\langle\rangle}{\exists}gNt(s(g))\backslash Sf)\ \Rightarrow\ {\it CP}that
\using {/}L
\endprooftree
\justifies
\mbox{\fbox{${\blacksquare}({\it CP}that/{\square}Sf)$}}, [{\blacksquare}Nt(s(n))], {\square}({\langle\rangle}{\exists}gNt(s(g))\backslash Sf)\ \Rightarrow\ {\it CP}that
\using {\blacksquare}L
\endprooftree
\justifies
{\blacksquare}({\it CP}that/{\square}Sf), [{\blacksquare}Nt(s(n))], {\square}({\langle\rangle}{\exists}gNt(s(g))\backslash Sf)\ \Rightarrow\ \fbox{${\it CP}that{\sqcup}{\square}Sf$}
\using {\sqcup}R
\endprooftree
\prooftree
\prooftree
\prooftree
\prooftree
\prooftree
\justifies
\mbox{\fbox{$Nt(s(m))$}}\ \Rightarrow\ Nt(s(m))
\endprooftree
\justifies
\mbox{\fbox{${\blacksquare}Nt(s(m))$}}\ \Rightarrow\ Nt(s(m))
\using {\blacksquare}L
\endprooftree
\justifies
{\blacksquare}Nt(s(m))\ \Rightarrow\ \fbox{${\exists}gNt(s(g))$}
\using {\exists}R
\endprooftree
\justifies
[{\blacksquare}Nt(s(m))]\ \Rightarrow\ \fbox{${\langle\rangle}{\exists}gNt(s(g))$}
\using {\langle\rangle}R
\endprooftree
\prooftree
\justifies
\mbox{\fbox{$Sf$}}\ \Rightarrow\ Sf
\endprooftree
\justifies
[{\blacksquare}Nt(s(m))], \mbox{\fbox{${\langle\rangle}{\exists}gNt(s(g))\backslash Sf$}}\ \Rightarrow\ Sf
\using {\backslash}L
\endprooftree
\justifies
[{\blacksquare}Nt(s(m))], \mbox{\fbox{$({\langle\rangle}{\exists}gNt(s(g))\backslash Sf)/({\it CP}that{\sqcup}{\square}Sf)$}}, {\blacksquare}({\it CP}that/{\square}Sf), [{\blacksquare}Nt(s(n))], {\square}({\langle\rangle}{\exists}gNt(s(g))\backslash Sf)\ \Rightarrow\ Sf
\using {/}L
\endprooftree
\justifies
[{\blacksquare}Nt(s(m))], \mbox{\fbox{${\square}(({\langle\rangle}{\exists}gNt(s(g))\backslash Sf)/({\it CP}that{\sqcup}{\square}Sf))$}}, {\blacksquare}({\it CP}that/{\square}Sf), [{\blacksquare}Nt(s(n))], {\square}({\langle\rangle}{\exists}gNt(s(g))\backslash Sf)\ \Rightarrow\ Sf
\using {\Box}L
\endprooftree
\justifies
[{\blacksquare}Nt(s(m))], {\square}(({\langle\rangle}{\exists}gNt(s(g))\backslash Sf)/({\it CP}that{\sqcup}{\square}Sf)), {\blacksquare}({\it CP}that/{\square}Sf), [{\tt 1}], {\square}({\langle\rangle}{\exists}gNt(s(g))\backslash Sf)\ \Rightarrow\ Sf{{}{\uparrow}{}}{\blacksquare}Nt(s(n))
\using {\uparrow}R
\endprooftree
\prooftree
\justifies
\mbox{\fbox{$Sf$}}\ \Rightarrow\ Sf
\endprooftree
\justifies
[{\blacksquare}Nt(s(m))], {\square}(({\langle\rangle}{\exists}gNt(s(g))\backslash Sf)/({\it CP}that{\sqcup}{\square}Sf)), {\blacksquare}({\it CP}that/{\square}Sf), [\mbox{\fbox{$(Sf{{}{\uparrow}{}}{\blacksquare}Nt(s(n))){{}{\downarrow}{}}Sf$}}], {\square}({\langle\rangle}{\exists}gNt(s(g))\backslash Sf)\ \Rightarrow\ Sf
\using {\downarrow}L
\endprooftree
\justifies
[{\blacksquare}Nt(s(m))], {\square}(({\langle\rangle}{\exists}gNt(s(g))\backslash Sf)/({\it CP}that{\sqcup}{\square}Sf)), {\blacksquare}({\it CP}that/{\square}Sf), [\mbox{\fbox{${\forall}f((Sf{{}{\uparrow}{}}{\blacksquare}Nt(s(n))){{}{\downarrow}{}}Sf)$}}], {\square}({\langle\rangle}{\exists}gNt(s(g))\backslash Sf)\ \Rightarrow\ Sf
\using {\forall}L
\endprooftree
\justifies
[{\blacksquare}Nt(s(m))], {\square}(({\langle\rangle}{\exists}gNt(s(g))\backslash Sf)/({\it CP}that{\sqcup}{\square}Sf)), {\blacksquare}({\it CP}that/{\square}Sf), [\mbox{\fbox{${\forall}f((Sf{{}{\uparrow}{}}{\blacksquare}Nt(s(n))){{}{\downarrow}{}}Sf)/{\it CN}{\it s(n)}$}}, {\square}{\it CN}{\it s(n)}], {\square}({\langle\rangle}{\exists}gNt(s(g))\backslash Sf)\ \Rightarrow\ Sf
\using {/}L
\endprooftree
\justifies
[{\blacksquare}Nt(s(m))], {\square}(({\langle\rangle}{\exists}gNt(s(g))\backslash Sf)/({\it CP}that{\sqcup}{\square}Sf)), {\blacksquare}({\it CP}that/{\square}Sf), [\mbox{\fbox{${\forall}g({\forall}f((Sf{{}{\uparrow}{}}{\blacksquare}Nt(s(g))){{}{\downarrow}{}}Sf)/{\it CN}{\it s(g)})$}}, {\square}{\it CN}{\it s(n)}], {\square}({\langle\rangle}{\exists}gNt(s(g))\backslash Sf)\ \Rightarrow\ Sf
\using {\forall}L
\endprooftree
\justifies
[{\blacksquare}Nt(s(m))], {\square}(({\langle\rangle}{\exists}gNt(s(g))\backslash Sf)/({\it CP}that{\sqcup}{\square}Sf)), {\blacksquare}({\it CP}that/{\square}Sf), [\mbox{\fbox{${\blacksquare}{\forall}g({\forall}f((Sf{{}{\uparrow}{}}{\blacksquare}Nt(s(g))){{}{\downarrow}{}}Sf)/{\it CN}{\it s(g)})$}}, {\square}{\it CN}{\it s(n)}], {\square}({\langle\rangle}{\exists}gNt(s(g))\backslash Sf)\ \Rightarrow\ Sf
\using {\blacksquare}L
\endprooftree}
$$
\vspace{0.15in}
\noindent
This de re derivation delivers semantics:
\disp{
$\exists C[(\mbox{\v{}}{\it fish}\ {\it C})\wedge ({\it Pres}\ ((\mbox{\v{}}{\it believe}\ \mbox{\^{}}({\it Pres}\ (\mbox{\v{}}{\it walk}\ {\it C})))\ {\it j}))]$}
This has existential commitment with respect to fish:
it cannot be true without some fish existing in the actual world.

The modalization of the complement sentence corresponds to translation into an
intensional formula, 
i.e.\ a proposition,
as argument to the propositional attitude verb.
The $\square R$ inference in the de re derivation depends on the hypothetical subtype
introduced by the indefinite article being modal.
Observe that by contrast in our type logical grammar the corresponding
subtype of {\it every\/} is not modal, 
capturing that its quantification is limited to local scope (something Montague did not
capture).

The de dicto derivation is:
\vspace{0.15in}
$$
{\tiny
\prooftree
\prooftree
\prooftree
\prooftree
\prooftree
\prooftree
\prooftree
\prooftree
\prooftree
\prooftree
\prooftree
\justifies
\mbox{\fbox{${\it CN}{\it s(n)}$}}\ \Rightarrow\ {\it CN}{\it s(n)}
\endprooftree
\justifies
\mbox{\fbox{${\square}{\it CN}{\it s(n)}$}}\ \Rightarrow\ {\it CN}{\it s(n)}
\using {\Box}L
\endprooftree
\prooftree
\prooftree
\prooftree
\prooftree
\prooftree
\prooftree
\prooftree
\prooftree
\prooftree
\justifies
\mbox{\fbox{$Nt(s(n))$}}\ \Rightarrow\ Nt(s(n))
\endprooftree
\justifies
\mbox{\fbox{${\blacksquare}Nt(s(n))$}}\ \Rightarrow\ Nt(s(n))
\using {\blacksquare}L
\endprooftree
\justifies
{\blacksquare}Nt(s(n))\ \Rightarrow\ \fbox{${\exists}gNt(s(g))$}
\using {\exists}R
\endprooftree
\justifies
[{\blacksquare}Nt(s(n))]\ \Rightarrow\ \fbox{${\langle\rangle}{\exists}gNt(s(g))$}
\using {\langle\rangle}R
\endprooftree
\prooftree
\justifies
\mbox{\fbox{$Sf$}}\ \Rightarrow\ Sf
\endprooftree
\justifies
[{\blacksquare}Nt(s(n))], \mbox{\fbox{${\langle\rangle}{\exists}gNt(s(g))\backslash Sf$}}\ \Rightarrow\ Sf
\using {\backslash}L
\endprooftree
\justifies
[{\blacksquare}Nt(s(n))], \mbox{\fbox{${\square}({\langle\rangle}{\exists}gNt(s(g))\backslash Sf)$}}\ \Rightarrow\ Sf
\using {\Box}L
\endprooftree
\justifies
[{\tt 1}], {\square}({\langle\rangle}{\exists}gNt(s(g))\backslash Sf)\ \Rightarrow\ Sf{{}{\uparrow}{}}{\blacksquare}Nt(s(n))
\using {\uparrow}R
\endprooftree
\prooftree
\justifies
\mbox{\fbox{$Sf$}}\ \Rightarrow\ Sf
\endprooftree
\justifies
[\mbox{\fbox{$(Sf{{}{\uparrow}{}}{\blacksquare}Nt(s(n))){{}{\downarrow}{}}Sf$}}], {\square}({\langle\rangle}{\exists}gNt(s(g))\backslash Sf)\ \Rightarrow\ Sf
\using {\downarrow}L
\endprooftree
\justifies
[\mbox{\fbox{${\forall}f((Sf{{}{\uparrow}{}}{\blacksquare}Nt(s(n))){{}{\downarrow}{}}Sf)$}}], {\square}({\langle\rangle}{\exists}gNt(s(g))\backslash Sf)\ \Rightarrow\ Sf
\using {\forall}L
\endprooftree
\justifies
[\mbox{\fbox{${\forall}f((Sf{{}{\uparrow}{}}{\blacksquare}Nt(s(n))){{}{\downarrow}{}}Sf)/{\it CN}{\it s(n)}$}}, {\square}{\it CN}{\it s(n)}], {\square}({\langle\rangle}{\exists}gNt(s(g))\backslash Sf)\ \Rightarrow\ Sf
\using {/}L
\endprooftree
\justifies
[\mbox{\fbox{${\forall}g({\forall}f((Sf{{}{\uparrow}{}}{\blacksquare}Nt(s(g))){{}{\downarrow}{}}Sf)/{\it CN}{\it s(g)})$}}, {\square}{\it CN}{\it s(n)}], {\square}({\langle\rangle}{\exists}gNt(s(g))\backslash Sf)\ \Rightarrow\ Sf
\using {\forall}L
\endprooftree
\justifies
[\mbox{\fbox{${\blacksquare}{\forall}g({\forall}f((Sf{{}{\uparrow}{}}{\blacksquare}Nt(s(g))){{}{\downarrow}{}}Sf)/{\it CN}{\it s(g)})$}}, {\square}{\it CN}{\it s(n)}], {\square}({\langle\rangle}{\exists}gNt(s(g))\backslash Sf)\ \Rightarrow\ Sf
\using {\blacksquare}L
\endprooftree
\justifies
[{\blacksquare}{\forall}g({\forall}f((Sf{{}{\uparrow}{}}{\blacksquare}Nt(s(g))){{}{\downarrow}{}}Sf)/{\it CN}{\it s(g)}), {\square}{\it CN}{\it s(n)}], {\square}({\langle\rangle}{\exists}gNt(s(g))\backslash Sf)\ \Rightarrow\ {\square}Sf
\using {\Box}R
\endprooftree
\prooftree
\justifies
\mbox{\fbox{${\it CP}that$}}\ \Rightarrow\ {\it CP}that
\endprooftree
\justifies
\mbox{\fbox{${\it CP}that/{\square}Sf$}}, [{\blacksquare}{\forall}g({\forall}f((Sf{{}{\uparrow}{}}{\blacksquare}Nt(s(g))){{}{\downarrow}{}}Sf)/{\it CN}{\it s(g)}), {\square}{\it CN}{\it s(n)}], {\square}({\langle\rangle}{\exists}gNt(s(g))\backslash Sf)\ \Rightarrow\ {\it CP}that
\using {/}L
\endprooftree
\justifies
\mbox{\fbox{${\blacksquare}({\it CP}that/{\square}Sf)$}}, [{\blacksquare}{\forall}g({\forall}f((Sf{{}{\uparrow}{}}{\blacksquare}Nt(s(g))){{}{\downarrow}{}}Sf)/{\it CN}{\it s(g)}), {\square}{\it CN}{\it s(n)}], {\square}({\langle\rangle}{\exists}gNt(s(g))\backslash Sf)\ \Rightarrow\ {\it CP}that
\using {\blacksquare}L
\endprooftree
\justifies
{\blacksquare}({\it CP}that/{\square}Sf), [{\blacksquare}{\forall}g({\forall}f((Sf{{}{\uparrow}{}}{\blacksquare}Nt(s(g))){{}{\downarrow}{}}Sf)/{\it CN}{\it s(g)}), {\square}{\it CN}{\it s(n)}], {\square}({\langle\rangle}{\exists}gNt(s(g))\backslash Sf)\ \Rightarrow\ \fbox{${\it CP}that{\sqcup}{\square}Sf$}
\using {\sqcup}R
\endprooftree
\prooftree
\prooftree
\prooftree
\prooftree
\prooftree
\justifies
\mbox{\fbox{$Nt(s(m))$}}\ \Rightarrow\ Nt(s(m))
\endprooftree
\justifies
\mbox{\fbox{${\blacksquare}Nt(s(m))$}}\ \Rightarrow\ Nt(s(m))
\using {\blacksquare}L
\endprooftree
\justifies
{\blacksquare}Nt(s(m))\ \Rightarrow\ \fbox{${\exists}gNt(s(g))$}
\using {\exists}R
\endprooftree
\justifies
[{\blacksquare}Nt(s(m))]\ \Rightarrow\ \fbox{${\langle\rangle}{\exists}gNt(s(g))$}
\using {\langle\rangle}R
\endprooftree
\prooftree
\justifies
\mbox{\fbox{$Sf$}}\ \Rightarrow\ Sf
\endprooftree
\justifies
[{\blacksquare}Nt(s(m))], \mbox{\fbox{${\langle\rangle}{\exists}gNt(s(g))\backslash Sf$}}\ \Rightarrow\ Sf
\using {\backslash}L
\endprooftree
\justifies
[{\blacksquare}Nt(s(m))], \mbox{\fbox{$({\langle\rangle}{\exists}gNt(s(g))\backslash Sf)/({\it CP}that{\sqcup}{\square}Sf)$}}, {\blacksquare}({\it CP}that/{\square}Sf), [{\blacksquare}{\forall}g({\forall}f((Sf{{}{\uparrow}{}}{\blacksquare}Nt(s(g))){{}{\downarrow}{}}Sf)/{\it CN}{\it s(g)}), {\square}{\it CN}{\it s(n)}], {\square}({\langle\rangle}{\exists}gNt(s(g))\backslash Sf)\ \Rightarrow\ Sf
\using {/}L
\endprooftree
\justifies
[{\blacksquare}Nt(s(m))], \mbox{\fbox{${\square}(({\langle\rangle}{\exists}gNt(s(g))\backslash Sf)/({\it CP}that{\sqcup}{\square}Sf))$}}, {\blacksquare}({\it CP}that/{\square}Sf), [{\blacksquare}{\forall}g({\forall}f((Sf{{}{\uparrow}{}}{\blacksquare}Nt(s(g))){{}{\downarrow}{}}Sf)/{\it CN}{\it s(g)}), {\square}{\it CN}{\it s(n)}], {\square}({\langle\rangle}{\exists}gNt(s(g))\backslash Sf)\ \Rightarrow\ Sf
\using {\Box}L
\endprooftree}
$$
\vspace{0.15in}
\noindent
This delivers semantics:
\disp{
$({\it Pres}\ ((\mbox{\v{}}{\it believe}\ \mbox{\^{}}\exists F[(\mbox{\v{}}{\it fish}\ {\it F})\wedge ({\it Pres}\ (\mbox{\v{}}{\it walk}\ {\it F}))])\ {\it j}))$}
This does not have existential commitment with respect to fish:
it can be true without any fish existing in the actual world.
The next example combines the previous ambiguity with alternative quantifier
scopings at the matrix level resulting in a total of three readings:
\disp{
(dwp((7-48, 49, 52))) $[{\bf every}{+}{\bf man}]{+}{\bf believes}{+}{\bf that}{+}[{\bf a}{+}{\bf fish}]{+}{\bf walks}: Sf$}
Lexical lookup yields the semantically labelled sequent:
\disp{
$[{\blacksquare}{\forall}g({\forall}f((Sf{{}{\uparrow}{}}Nt(s(g))){{}{\downarrow}{}}Sf)/{\it CN}{\it s(g)}): \lambda A\lambda B\forall C[({\it A}\ {\it C})\rightarrow ({\it B}\ {\it C})], {\square}{\it CN}{\it s(m)}: {\it man}],\\{\square}(({\langle\rangle}{\exists}gNt(s(g))\backslash$ $Sf)/({\it CP}that{\sqcup}{\square}Sf)): \mbox{\^{}}\lambda D\lambda E({\it Pres}\ ((\mbox{\v{}}{\it believe}\ {\it D})\ {\it E})), {\blacksquare}({\it CP}that/{\square}Sf): \lambda F{\it F},$ $[{\blacksquare}{\forall}g({\forall}f((Sf{{}{\uparrow}{}}{\blacksquare}Nt(s(g))){{}{\downarrow}{}}$ $Sf)/{\it CN}{\it s(g)}): \lambda G\lambda H\exists I[({\it G}\ {\it I})\wedge ({\it H}\ {\it I})], {\square}{\it CN}{\it s(n)}: {\it fish}],\\{\square}({\langle\rangle}{\exists}gNt(s(g))\backslash Sf): \mbox{\^{}}\lambda J({\it Pres}\ (\mbox{\v{}}{\it walk}\ {\it J}))\ \Rightarrow\ Sf$}
A first derivation, 
in which the existential quantifies in at the widest level,
is thus:
\vspace{0.15in}
$$
\resizebox{\textwidth}{!}{
\prooftree
\prooftree
\prooftree
\prooftree
\prooftree
\justifies
\mbox{\fbox{${\it CN}{\it s(n)}$}}\ \Rightarrow\ {\it CN}{\it s(n)}
\endprooftree
\justifies
\mbox{\fbox{${\square}{\it CN}{\it s(n)}$}}\ \Rightarrow\ {\it CN}{\it s(n)}
\using {\Box}L
\endprooftree
\prooftree
\prooftree
\prooftree
\prooftree
\prooftree
\prooftree
\prooftree
\prooftree
\justifies
\mbox{\fbox{${\it CN}{\it s(m)}$}}\ \Rightarrow\ {\it CN}{\it s(m)}
\endprooftree
\justifies
\mbox{\fbox{${\square}{\it CN}{\it s(m)}$}}\ \Rightarrow\ {\it CN}{\it s(m)}
\using {\Box}L
\endprooftree
\prooftree
\prooftree
\prooftree
\prooftree
\prooftree
\prooftree
\prooftree
\prooftree
\prooftree
\prooftree
\prooftree
\prooftree
\prooftree
\prooftree
\prooftree
\justifies
\mbox{\fbox{$Nt(s(n))$}}\ \Rightarrow\ Nt(s(n))
\endprooftree
\justifies
\mbox{\fbox{${\blacksquare}Nt(s(n))$}}\ \Rightarrow\ Nt(s(n))
\using {\blacksquare}L
\endprooftree
\justifies
{\blacksquare}Nt(s(n))\ \Rightarrow\ \fbox{${\exists}gNt(s(g))$}
\using {\exists}R
\endprooftree
\justifies
[{\blacksquare}Nt(s(n))]\ \Rightarrow\ \fbox{${\langle\rangle}{\exists}gNt(s(g))$}
\using {\langle\rangle}R
\endprooftree
\prooftree
\justifies
\mbox{\fbox{$Sf$}}\ \Rightarrow\ Sf
\endprooftree
\justifies
[{\blacksquare}Nt(s(n))], \mbox{\fbox{${\langle\rangle}{\exists}gNt(s(g))\backslash Sf$}}\ \Rightarrow\ Sf
\using {\backslash}L
\endprooftree
\justifies
[{\blacksquare}Nt(s(n))], \mbox{\fbox{${\square}({\langle\rangle}{\exists}gNt(s(g))\backslash Sf)$}}\ \Rightarrow\ Sf
\using {\Box}L
\endprooftree
\justifies
[{\blacksquare}Nt(s(n))], {\square}({\langle\rangle}{\exists}gNt(s(g))\backslash Sf)\ \Rightarrow\ {\square}Sf
\using {\Box}R
\endprooftree
\prooftree
\justifies
\mbox{\fbox{${\it CP}that$}}\ \Rightarrow\ {\it CP}that
\endprooftree
\justifies
\mbox{\fbox{${\it CP}that/{\square}Sf$}}, [{\blacksquare}Nt(s(n))], {\square}({\langle\rangle}{\exists}gNt(s(g))\backslash Sf)\ \Rightarrow\ {\it CP}that
\using {/}L
\endprooftree
\justifies
\mbox{\fbox{${\blacksquare}({\it CP}that/{\square}Sf)$}}, [{\blacksquare}Nt(s(n))], {\square}({\langle\rangle}{\exists}gNt(s(g))\backslash Sf)\ \Rightarrow\ {\it CP}that
\using {\blacksquare}L
\endprooftree
\justifies
{\blacksquare}({\it CP}that/{\square}Sf), [{\blacksquare}Nt(s(n))], {\square}({\langle\rangle}{\exists}gNt(s(g))\backslash Sf)\ \Rightarrow\ \fbox{${\it CP}that{\sqcup}{\square}Sf$}
\using {\sqcup}R
\endprooftree
\prooftree
\prooftree
\prooftree
\prooftree
\justifies
Nt(s(m))\ \Rightarrow\ Nt(s(m))
\endprooftree
\justifies
Nt(s(m))\ \Rightarrow\ \fbox{${\exists}gNt(s(g))$}
\using {\exists}R
\endprooftree
\justifies
[Nt(s(m))]\ \Rightarrow\ \fbox{${\langle\rangle}{\exists}gNt(s(g))$}
\using {\langle\rangle}R
\endprooftree
\prooftree
\justifies
\mbox{\fbox{$Sf$}}\ \Rightarrow\ Sf
\endprooftree
\justifies
[Nt(s(m))], \mbox{\fbox{${\langle\rangle}{\exists}gNt(s(g))\backslash Sf$}}\ \Rightarrow\ Sf
\using {\backslash}L
\endprooftree
\justifies
[Nt(s(m))], \mbox{\fbox{$({\langle\rangle}{\exists}gNt(s(g))\backslash Sf)/({\it CP}that{\sqcup}{\square}Sf)$}}, {\blacksquare}({\it CP}that/{\square}Sf), [{\blacksquare}Nt(s(n))], {\square}({\langle\rangle}{\exists}gNt(s(g))\backslash Sf)\ \Rightarrow\ Sf
\using {/}L
\endprooftree
\justifies
[Nt(s(m))], \mbox{\fbox{${\square}(({\langle\rangle}{\exists}gNt(s(g))\backslash Sf)/({\it CP}that{\sqcup}{\square}Sf))$}}, {\blacksquare}({\it CP}that/{\square}Sf), [{\blacksquare}Nt(s(n))], {\square}({\langle\rangle}{\exists}gNt(s(g))\backslash Sf)\ \Rightarrow\ Sf
\using {\Box}L
\endprooftree
\justifies
[{\tt 1}], {\square}(({\langle\rangle}{\exists}gNt(s(g))\backslash Sf)/({\it CP}that{\sqcup}{\square}Sf)), {\blacksquare}({\it CP}that/{\square}Sf), [{\blacksquare}Nt(s(n))], {\square}({\langle\rangle}{\exists}gNt(s(g))\backslash Sf)\ \Rightarrow\ Sf{{}{\uparrow}{}}Nt(s(m))
\using {\uparrow}R
\endprooftree
\prooftree
\justifies
\mbox{\fbox{$Sf$}}\ \Rightarrow\ Sf
\endprooftree
\justifies
[\mbox{\fbox{$(Sf{{}{\uparrow}{}}Nt(s(m))){{}{\downarrow}{}}Sf$}}], {\square}(({\langle\rangle}{\exists}gNt(s(g))\backslash Sf)/({\it CP}that{\sqcup}{\square}Sf)), {\blacksquare}({\it CP}that/{\square}Sf), [{\blacksquare}Nt(s(n))], {\square}({\langle\rangle}{\exists}gNt(s(g))\backslash Sf)\ \Rightarrow\ Sf
\using {\downarrow}L
\endprooftree
\justifies
[\mbox{\fbox{${\forall}f((Sf{{}{\uparrow}{}}Nt(s(m))){{}{\downarrow}{}}Sf)$}}], {\square}(({\langle\rangle}{\exists}gNt(s(g))\backslash Sf)/({\it CP}that{\sqcup}{\square}Sf)), {\blacksquare}({\it CP}that/{\square}Sf), [{\blacksquare}Nt(s(n))], {\square}({\langle\rangle}{\exists}gNt(s(g))\backslash Sf)\ \Rightarrow\ Sf
\using {\forall}L
\endprooftree
\justifies
[\mbox{\fbox{${\forall}f((Sf{{}{\uparrow}{}}Nt(s(m))){{}{\downarrow}{}}Sf)/{\it CN}{\it s(m)}$}}, {\square}{\it CN}{\it s(m)}], {\square}(({\langle\rangle}{\exists}gNt(s(g))\backslash Sf)/({\it CP}that{\sqcup}{\square}Sf)), {\blacksquare}({\it CP}that/{\square}Sf), [{\blacksquare}Nt(s(n))], {\square}({\langle\rangle}{\exists}gNt(s(g))\backslash Sf)\ \Rightarrow\ Sf
\using {/}L
\endprooftree
\justifies
[\mbox{\fbox{${\forall}g({\forall}f((Sf{{}{\uparrow}{}}Nt(s(g))){{}{\downarrow}{}}Sf)/{\it CN}{\it s(g)})$}}, {\square}{\it CN}{\it s(m)}], {\square}(({\langle\rangle}{\exists}gNt(s(g))\backslash Sf)/({\it CP}that{\sqcup}{\square}Sf)), {\blacksquare}({\it CP}that/{\square}Sf), [{\blacksquare}Nt(s(n))], {\square}({\langle\rangle}{\exists}gNt(s(g))\backslash Sf)\ \Rightarrow\ Sf
\using {\forall}L
\endprooftree
\justifies
[\mbox{\fbox{${\blacksquare}{\forall}g({\forall}f((Sf{{}{\uparrow}{}}Nt(s(g))){{}{\downarrow}{}}Sf)/{\it CN}{\it s(g)})$}}, {\square}{\it CN}{\it s(m)}], {\square}(({\langle\rangle}{\exists}gNt(s(g))\backslash Sf)/({\it CP}that{\sqcup}{\square}Sf)), {\blacksquare}({\it CP}that/{\square}Sf), [{\blacksquare}Nt(s(n))], {\square}({\langle\rangle}{\exists}gNt(s(g))\backslash Sf)\ \Rightarrow\ Sf
\using {\blacksquare}L
\endprooftree
\justifies
[{\blacksquare}{\forall}g({\forall}f((Sf{{}{\uparrow}{}}Nt(s(g))){{}{\downarrow}{}}Sf)/{\it CN}{\it s(g)}), {\square}{\it CN}{\it s(m)}], {\square}(({\langle\rangle}{\exists}gNt(s(g))\backslash Sf)/({\it CP}that{\sqcup}{\square}Sf)), {\blacksquare}({\it CP}that/{\square}Sf), [{\tt 1}], {\square}({\langle\rangle}{\exists}gNt(s(g))\backslash Sf)\ \Rightarrow\ Sf{{}{\uparrow}{}}{\blacksquare}Nt(s(n))
\using {\uparrow}R
\endprooftree
\prooftree
\justifies
\mbox{\fbox{$Sf$}}\ \Rightarrow\ Sf
\endprooftree
\justifies
[{\blacksquare}{\forall}g({\forall}f((Sf{{}{\uparrow}{}}Nt(s(g))){{}{\downarrow}{}}Sf)/{\it CN}{\it s(g)}), {\square}{\it CN}{\it s(m)}], {\square}(({\langle\rangle}{\exists}gNt(s(g))\backslash Sf)/({\it CP}that{\sqcup}{\square}Sf)), {\blacksquare}({\it CP}that/{\square}Sf), [\mbox{\fbox{$(Sf{{}{\uparrow}{}}{\blacksquare}Nt(s(n))){{}{\downarrow}{}}Sf$}}], {\square}({\langle\rangle}{\exists}gNt(s(g))\backslash Sf)\ \Rightarrow\ Sf
\using {\downarrow}L
\endprooftree
\justifies
[{\blacksquare}{\forall}g({\forall}f((Sf{{}{\uparrow}{}}Nt(s(g))){{}{\downarrow}{}}Sf)/{\it CN}{\it s(g)}), {\square}{\it CN}{\it s(m)}], {\square}(({\langle\rangle}{\exists}gNt(s(g))\backslash Sf)/({\it CP}that{\sqcup}{\square}Sf)), {\blacksquare}({\it CP}that/{\square}Sf), [\mbox{\fbox{${\forall}f((Sf{{}{\uparrow}{}}{\blacksquare}Nt(s(n))){{}{\downarrow}{}}Sf)$}}], {\square}({\langle\rangle}{\exists}gNt(s(g))\backslash Sf)\ \Rightarrow\ Sf
\using {\forall}L
\endprooftree
\justifies
[{\blacksquare}{\forall}g({\forall}f((Sf{{}{\uparrow}{}}Nt(s(g))){{}{\downarrow}{}}Sf)/{\it CN}{\it s(g)}), {\square}{\it CN}{\it s(m)}], {\square}(({\langle\rangle}{\exists}gNt(s(g))\backslash Sf)/({\it CP}that{\sqcup}{\square}Sf)), {\blacksquare}({\it CP}that/{\square}Sf), [\mbox{\fbox{${\forall}f((Sf{{}{\uparrow}{}}{\blacksquare}Nt(s(n))){{}{\downarrow}{}}Sf)/{\it CN}{\it s(n)}$}}, {\square}{\it CN}{\it s(n)}], {\square}({\langle\rangle}{\exists}gNt(s(g))\backslash Sf)\ \Rightarrow\ Sf
\using {/}L
\endprooftree
\justifies
[{\blacksquare}{\forall}g({\forall}f((Sf{{}{\uparrow}{}}Nt(s(g))){{}{\downarrow}{}}Sf)/{\it CN}{\it s(g)}), {\square}{\it CN}{\it s(m)}], {\square}(({\langle\rangle}{\exists}gNt(s(g))\backslash Sf)/({\it CP}that{\sqcup}{\square}Sf)), {\blacksquare}({\it CP}that/{\square}Sf), [\mbox{\fbox{${\forall}g({\forall}f((Sf{{}{\uparrow}{}}{\blacksquare}Nt(s(g))){{}{\downarrow}{}}Sf)/{\it CN}{\it s(g)})$}}, {\square}{\it CN}{\it s(n)}], {\square}({\langle\rangle}{\exists}gNt(s(g))\backslash Sf)\ \Rightarrow\ Sf
\using {\forall}L
\endprooftree
\justifies
[{\blacksquare}{\forall}g({\forall}f((Sf{{}{\uparrow}{}}Nt(s(g))){{}{\downarrow}{}}Sf)/{\it CN}{\it s(g)}), {\square}{\it CN}{\it s(m)}], {\square}(({\langle\rangle}{\exists}gNt(s(g))\backslash Sf)/({\it CP}that{\sqcup}{\square}Sf)), {\blacksquare}({\it CP}that/{\square}Sf), [\mbox{\fbox{${\blacksquare}{\forall}g({\forall}f((Sf{{}{\uparrow}{}}{\blacksquare}Nt(s(g))){{}{\downarrow}{}}Sf)/{\it CN}{\it s(g)})$}}, {\square}{\it CN}{\it s(n)}], {\square}({\langle\rangle}{\exists}gNt(s(g))\backslash Sf)\ \Rightarrow\ Sf
\using {\blacksquare}L
\endprooftree}
$$
\vspace{0.15in}
\noindent
This delivers semantics:
\disp{
$\exists C[(\mbox{\v{}}{\it fish}\ {\it C})\wedge \forall G[(\mbox{\v{}}{\it man}\ {\it G})\rightarrow ({\it Pres}\ ((\mbox{\v{}}{\it believe}\ \mbox{\^{}}({\it Pres}\ (\mbox{\v{}}{\it walk}\ {\it C})))\ {\it G}))]]$}
A second derivation,
in which the existential quantifies in outside of the propositional attitude verb
but within the universal, is thus:
\vspace{0.15in}
$$
\resizebox{\textwidth}{!}{
\prooftree
\prooftree
\prooftree
\prooftree
\prooftree
\justifies
\mbox{\fbox{${\it CN}{\it s(m)}$}}\ \Rightarrow\ {\it CN}{\it s(m)}
\endprooftree
\justifies
\mbox{\fbox{${\square}{\it CN}{\it s(m)}$}}\ \Rightarrow\ {\it CN}{\it s(m)}
\using {\Box}L
\endprooftree
\prooftree
\prooftree
\prooftree
\prooftree
\prooftree
\prooftree
\prooftree
\prooftree
\justifies
\mbox{\fbox{${\it CN}{\it s(n)}$}}\ \Rightarrow\ {\it CN}{\it s(n)}
\endprooftree
\justifies
\mbox{\fbox{${\square}{\it CN}{\it s(n)}$}}\ \Rightarrow\ {\it CN}{\it s(n)}
\using {\Box}L
\endprooftree
\prooftree
\prooftree
\prooftree
\prooftree
\prooftree
\prooftree
\prooftree
\prooftree
\prooftree
\prooftree
\prooftree
\prooftree
\prooftree
\prooftree
\prooftree
\justifies
\mbox{\fbox{$Nt(s(n))$}}\ \Rightarrow\ Nt(s(n))
\endprooftree
\justifies
\mbox{\fbox{${\blacksquare}Nt(s(n))$}}\ \Rightarrow\ Nt(s(n))
\using {\blacksquare}L
\endprooftree
\justifies
{\blacksquare}Nt(s(n))\ \Rightarrow\ \fbox{${\exists}gNt(s(g))$}
\using {\exists}R
\endprooftree
\justifies
[{\blacksquare}Nt(s(n))]\ \Rightarrow\ \fbox{${\langle\rangle}{\exists}gNt(s(g))$}
\using {\langle\rangle}R
\endprooftree
\prooftree
\justifies
\mbox{\fbox{$Sf$}}\ \Rightarrow\ Sf
\endprooftree
\justifies
[{\blacksquare}Nt(s(n))], \mbox{\fbox{${\langle\rangle}{\exists}gNt(s(g))\backslash Sf$}}\ \Rightarrow\ Sf
\using {\backslash}L
\endprooftree
\justifies
[{\blacksquare}Nt(s(n))], \mbox{\fbox{${\square}({\langle\rangle}{\exists}gNt(s(g))\backslash Sf)$}}\ \Rightarrow\ Sf
\using {\Box}L
\endprooftree
\justifies
[{\blacksquare}Nt(s(n))], {\square}({\langle\rangle}{\exists}gNt(s(g))\backslash Sf)\ \Rightarrow\ {\square}Sf
\using {\Box}R
\endprooftree
\prooftree
\justifies
\mbox{\fbox{${\it CP}that$}}\ \Rightarrow\ {\it CP}that
\endprooftree
\justifies
\mbox{\fbox{${\it CP}that/{\square}Sf$}}, [{\blacksquare}Nt(s(n))], {\square}({\langle\rangle}{\exists}gNt(s(g))\backslash Sf)\ \Rightarrow\ {\it CP}that
\using {/}L
\endprooftree
\justifies
\mbox{\fbox{${\blacksquare}({\it CP}that/{\square}Sf)$}}, [{\blacksquare}Nt(s(n))], {\square}({\langle\rangle}{\exists}gNt(s(g))\backslash Sf)\ \Rightarrow\ {\it CP}that
\using {\blacksquare}L
\endprooftree
\justifies
{\blacksquare}({\it CP}that/{\square}Sf), [{\blacksquare}Nt(s(n))], {\square}({\langle\rangle}{\exists}gNt(s(g))\backslash Sf)\ \Rightarrow\ \fbox{${\it CP}that{\sqcup}{\square}Sf$}
\using {\sqcup}R
\endprooftree
\prooftree
\prooftree
\prooftree
\prooftree
\justifies
Nt(s(m))\ \Rightarrow\ Nt(s(m))
\endprooftree
\justifies
Nt(s(m))\ \Rightarrow\ \fbox{${\exists}gNt(s(g))$}
\using {\exists}R
\endprooftree
\justifies
[Nt(s(m))]\ \Rightarrow\ \fbox{${\langle\rangle}{\exists}gNt(s(g))$}
\using {\langle\rangle}R
\endprooftree
\prooftree
\justifies
\mbox{\fbox{$Sf$}}\ \Rightarrow\ Sf
\endprooftree
\justifies
[Nt(s(m))], \mbox{\fbox{${\langle\rangle}{\exists}gNt(s(g))\backslash Sf$}}\ \Rightarrow\ Sf
\using {\backslash}L
\endprooftree
\justifies
[Nt(s(m))], \mbox{\fbox{$({\langle\rangle}{\exists}gNt(s(g))\backslash Sf)/({\it CP}that{\sqcup}{\square}Sf)$}}, {\blacksquare}({\it CP}that/{\square}Sf), [{\blacksquare}Nt(s(n))], {\square}({\langle\rangle}{\exists}gNt(s(g))\backslash Sf)\ \Rightarrow\ Sf
\using {/}L
\endprooftree
\justifies
[Nt(s(m))], \mbox{\fbox{${\square}(({\langle\rangle}{\exists}gNt(s(g))\backslash Sf)/({\it CP}that{\sqcup}{\square}Sf))$}}, {\blacksquare}({\it CP}that/{\square}Sf), [{\blacksquare}Nt(s(n))], {\square}({\langle\rangle}{\exists}gNt(s(g))\backslash Sf)\ \Rightarrow\ Sf
\using {\Box}L
\endprooftree
\justifies
[Nt(s(m))], {\square}(({\langle\rangle}{\exists}gNt(s(g))\backslash Sf)/({\it CP}that{\sqcup}{\square}Sf)), {\blacksquare}({\it CP}that/{\square}Sf), [{\tt 1}], {\square}({\langle\rangle}{\exists}gNt(s(g))\backslash Sf)\ \Rightarrow\ Sf{{}{\uparrow}{}}{\blacksquare}Nt(s(n))
\using {\uparrow}R
\endprooftree
\prooftree
\justifies
\mbox{\fbox{$Sf$}}\ \Rightarrow\ Sf
\endprooftree
\justifies
[Nt(s(m))], {\square}(({\langle\rangle}{\exists}gNt(s(g))\backslash Sf)/({\it CP}that{\sqcup}{\square}Sf)), {\blacksquare}({\it CP}that/{\square}Sf), [\mbox{\fbox{$(Sf{{}{\uparrow}{}}{\blacksquare}Nt(s(n))){{}{\downarrow}{}}Sf$}}], {\square}({\langle\rangle}{\exists}gNt(s(g))\backslash Sf)\ \Rightarrow\ Sf
\using {\downarrow}L
\endprooftree
\justifies
[Nt(s(m))], {\square}(({\langle\rangle}{\exists}gNt(s(g))\backslash Sf)/({\it CP}that{\sqcup}{\square}Sf)), {\blacksquare}({\it CP}that/{\square}Sf), [\mbox{\fbox{${\forall}f((Sf{{}{\uparrow}{}}{\blacksquare}Nt(s(n))){{}{\downarrow}{}}Sf)$}}], {\square}({\langle\rangle}{\exists}gNt(s(g))\backslash Sf)\ \Rightarrow\ Sf
\using {\forall}L
\endprooftree
\justifies
[Nt(s(m))], {\square}(({\langle\rangle}{\exists}gNt(s(g))\backslash Sf)/({\it CP}that{\sqcup}{\square}Sf)), {\blacksquare}({\it CP}that/{\square}Sf), [\mbox{\fbox{${\forall}f((Sf{{}{\uparrow}{}}{\blacksquare}Nt(s(n))){{}{\downarrow}{}}Sf)/{\it CN}{\it s(n)}$}}, {\square}{\it CN}{\it s(n)}], {\square}({\langle\rangle}{\exists}gNt(s(g))\backslash Sf)\ \Rightarrow\ Sf
\using {/}L
\endprooftree
\justifies
[Nt(s(m))], {\square}(({\langle\rangle}{\exists}gNt(s(g))\backslash Sf)/({\it CP}that{\sqcup}{\square}Sf)), {\blacksquare}({\it CP}that/{\square}Sf), [\mbox{\fbox{${\forall}g({\forall}f((Sf{{}{\uparrow}{}}{\blacksquare}Nt(s(g))){{}{\downarrow}{}}Sf)/{\it CN}{\it s(g)})$}}, {\square}{\it CN}{\it s(n)}], {\square}({\langle\rangle}{\exists}gNt(s(g))\backslash Sf)\ \Rightarrow\ Sf
\using {\forall}L
\endprooftree
\justifies
[Nt(s(m))], {\square}(({\langle\rangle}{\exists}gNt(s(g))\backslash Sf)/({\it CP}that{\sqcup}{\square}Sf)), {\blacksquare}({\it CP}that/{\square}Sf), [\mbox{\fbox{${\blacksquare}{\forall}g({\forall}f((Sf{{}{\uparrow}{}}{\blacksquare}Nt(s(g))){{}{\downarrow}{}}Sf)/{\it CN}{\it s(g)})$}}, {\square}{\it CN}{\it s(n)}], {\square}({\langle\rangle}{\exists}gNt(s(g))\backslash Sf)\ \Rightarrow\ Sf
\using {\blacksquare}L
\endprooftree
\justifies
[{\tt 1}], {\square}(({\langle\rangle}{\exists}gNt(s(g))\backslash Sf)/({\it CP}that{\sqcup}{\square}Sf)), {\blacksquare}({\it CP}that/{\square}Sf), [{\blacksquare}{\forall}g({\forall}f((Sf{{}{\uparrow}{}}{\blacksquare}Nt(s(g))){{}{\downarrow}{}}Sf)/{\it CN}{\it s(g)}), {\square}{\it CN}{\it s(n)}], {\square}({\langle\rangle}{\exists}gNt(s(g))\backslash Sf)\ \Rightarrow\ Sf{{}{\uparrow}{}}Nt(s(m))
\using {\uparrow}R
\endprooftree
\prooftree
\justifies
\mbox{\fbox{$Sf$}}\ \Rightarrow\ Sf
\endprooftree
\justifies
[\mbox{\fbox{$(Sf{{}{\uparrow}{}}Nt(s(m))){{}{\downarrow}{}}Sf$}}], {\square}(({\langle\rangle}{\exists}gNt(s(g))\backslash Sf)/({\it CP}that{\sqcup}{\square}Sf)), {\blacksquare}({\it CP}that/{\square}Sf), [{\blacksquare}{\forall}g({\forall}f((Sf{{}{\uparrow}{}}{\blacksquare}Nt(s(g))){{}{\downarrow}{}}Sf)/{\it CN}{\it s(g)}), {\square}{\it CN}{\it s(n)}], {\square}({\langle\rangle}{\exists}gNt(s(g))\backslash Sf)\ \Rightarrow\ Sf
\using {\downarrow}L
\endprooftree
\justifies
[\mbox{\fbox{${\forall}f((Sf{{}{\uparrow}{}}Nt(s(m))){{}{\downarrow}{}}Sf)$}}], {\square}(({\langle\rangle}{\exists}gNt(s(g))\backslash Sf)/({\it CP}that{\sqcup}{\square}Sf)), {\blacksquare}({\it CP}that/{\square}Sf), [{\blacksquare}{\forall}g({\forall}f((Sf{{}{\uparrow}{}}{\blacksquare}Nt(s(g))){{}{\downarrow}{}}Sf)/{\it CN}{\it s(g)}), {\square}{\it CN}{\it s(n)}], {\square}({\langle\rangle}{\exists}gNt(s(g))\backslash Sf)\ \Rightarrow\ Sf
\using {\forall}L
\endprooftree
\justifies
[\mbox{\fbox{${\forall}f((Sf{{}{\uparrow}{}}Nt(s(m))){{}{\downarrow}{}}Sf)/{\it CN}{\it s(m)}$}}, {\square}{\it CN}{\it s(m)}], {\square}(({\langle\rangle}{\exists}gNt(s(g))\backslash Sf)/({\it CP}that{\sqcup}{\square}Sf)), {\blacksquare}({\it CP}that/{\square}Sf), [{\blacksquare}{\forall}g({\forall}f((Sf{{}{\uparrow}{}}{\blacksquare}Nt(s(g))){{}{\downarrow}{}}Sf)/{\it CN}{\it s(g)}), {\square}{\it CN}{\it s(n)}], {\square}({\langle\rangle}{\exists}gNt(s(g))\backslash Sf)\ \Rightarrow\ Sf
\using {/}L
\endprooftree
\justifies
[\mbox{\fbox{${\forall}g({\forall}f((Sf{{}{\uparrow}{}}Nt(s(g))){{}{\downarrow}{}}Sf)/{\it CN}{\it s(g)})$}}, {\square}{\it CN}{\it s(m)}], {\square}(({\langle\rangle}{\exists}gNt(s(g))\backslash Sf)/({\it CP}that{\sqcup}{\square}Sf)), {\blacksquare}({\it CP}that/{\square}Sf), [{\blacksquare}{\forall}g({\forall}f((Sf{{}{\uparrow}{}}{\blacksquare}Nt(s(g))){{}{\downarrow}{}}Sf)/{\it CN}{\it s(g)}), {\square}{\it CN}{\it s(n)}], {\square}({\langle\rangle}{\exists}gNt(s(g))\backslash Sf)\ \Rightarrow\ Sf
\using {\forall}L
\endprooftree
\justifies
[\mbox{\fbox{${\blacksquare}{\forall}g({\forall}f((Sf{{}{\uparrow}{}}Nt(s(g))){{}{\downarrow}{}}Sf)/{\it CN}{\it s(g)})$}}, {\square}{\it CN}{\it s(m)}], {\square}(({\langle\rangle}{\exists}gNt(s(g))\backslash Sf)/({\it CP}that{\sqcup}{\square}Sf)), {\blacksquare}({\it CP}that/{\square}Sf), [{\blacksquare}{\forall}g({\forall}f((Sf{{}{\uparrow}{}}{\blacksquare}Nt(s(g))){{}{\downarrow}{}}Sf)/{\it CN}{\it s(g)}), {\square}{\it CN}{\it s(n)}], {\square}({\langle\rangle}{\exists}gNt(s(g))\backslash Sf)\ \Rightarrow\ Sf
\using {\blacksquare}L
\endprooftree}
$$
\vspace{0.15in}
\noindent
This delivers semantics:
\disp{
$\forall C[(\mbox{\v{}}{\it man}\ {\it C})\rightarrow \exists G[(\mbox{\v{}}{\it fish}\ {\it G})\wedge ({\it Pres}\ ((\mbox{\v{}}{\it believe}\ \mbox{\^{}}({\it Pres}\ (\mbox{\v{}}{\it walk}\ {\it G})))\ {\it C}))]]$}
The third derivation, in which the existential takes narrowest scope,
is thus:
\vspace{0.15in}
$$
\resizebox{\textwidth}{!}{
\prooftree
\prooftree
\prooftree
\prooftree
\prooftree
\justifies
\mbox{\fbox{${\it CN}{\it s(m)}$}}\ \Rightarrow\ {\it CN}{\it s(m)}
\endprooftree
\justifies
\mbox{\fbox{${\square}{\it CN}{\it s(m)}$}}\ \Rightarrow\ {\it CN}{\it s(m)}
\using {\Box}L
\endprooftree
\prooftree
\prooftree
\prooftree
\prooftree
\prooftree
\prooftree
\prooftree
\prooftree
\prooftree
\prooftree
\prooftree
\prooftree
\prooftree
\prooftree
\justifies
\mbox{\fbox{${\it CN}{\it s(n)}$}}\ \Rightarrow\ {\it CN}{\it s(n)}
\endprooftree
\justifies
\mbox{\fbox{${\square}{\it CN}{\it s(n)}$}}\ \Rightarrow\ {\it CN}{\it s(n)}
\using {\Box}L
\endprooftree
\prooftree
\prooftree
\prooftree
\prooftree
\prooftree
\prooftree
\prooftree
\prooftree
\prooftree
\justifies
\mbox{\fbox{$Nt(s(n))$}}\ \Rightarrow\ Nt(s(n))
\endprooftree
\justifies
\mbox{\fbox{${\blacksquare}Nt(s(n))$}}\ \Rightarrow\ Nt(s(n))
\using {\blacksquare}L
\endprooftree
\justifies
{\blacksquare}Nt(s(n))\ \Rightarrow\ \fbox{${\exists}gNt(s(g))$}
\using {\exists}R
\endprooftree
\justifies
[{\blacksquare}Nt(s(n))]\ \Rightarrow\ \fbox{${\langle\rangle}{\exists}gNt(s(g))$}
\using {\langle\rangle}R
\endprooftree
\prooftree
\justifies
\mbox{\fbox{$Sf$}}\ \Rightarrow\ Sf
\endprooftree
\justifies
[{\blacksquare}Nt(s(n))], \mbox{\fbox{${\langle\rangle}{\exists}gNt(s(g))\backslash Sf$}}\ \Rightarrow\ Sf
\using {\backslash}L
\endprooftree
\justifies
[{\blacksquare}Nt(s(n))], \mbox{\fbox{${\square}({\langle\rangle}{\exists}gNt(s(g))\backslash Sf)$}}\ \Rightarrow\ Sf
\using {\Box}L
\endprooftree
\justifies
[{\tt 1}], {\square}({\langle\rangle}{\exists}gNt(s(g))\backslash Sf)\ \Rightarrow\ Sf{{}{\uparrow}{}}{\blacksquare}Nt(s(n))
\using {\uparrow}R
\endprooftree
\prooftree
\justifies
\mbox{\fbox{$Sf$}}\ \Rightarrow\ Sf
\endprooftree
\justifies
[\mbox{\fbox{$(Sf{{}{\uparrow}{}}{\blacksquare}Nt(s(n))){{}{\downarrow}{}}Sf$}}], {\square}({\langle\rangle}{\exists}gNt(s(g))\backslash Sf)\ \Rightarrow\ Sf
\using {\downarrow}L
\endprooftree
\justifies
[\mbox{\fbox{${\forall}f((Sf{{}{\uparrow}{}}{\blacksquare}Nt(s(n))){{}{\downarrow}{}}Sf)$}}], {\square}({\langle\rangle}{\exists}gNt(s(g))\backslash Sf)\ \Rightarrow\ Sf
\using {\forall}L
\endprooftree
\justifies
[\mbox{\fbox{${\forall}f((Sf{{}{\uparrow}{}}{\blacksquare}Nt(s(n))){{}{\downarrow}{}}Sf)/{\it CN}{\it s(n)}$}}, {\square}{\it CN}{\it s(n)}], {\square}({\langle\rangle}{\exists}gNt(s(g))\backslash Sf)\ \Rightarrow\ Sf
\using {/}L
\endprooftree
\justifies
[\mbox{\fbox{${\forall}g({\forall}f((Sf{{}{\uparrow}{}}{\blacksquare}Nt(s(g))){{}{\downarrow}{}}Sf)/{\it CN}{\it s(g)})$}}, {\square}{\it CN}{\it s(n)}], {\square}({\langle\rangle}{\exists}gNt(s(g))\backslash Sf)\ \Rightarrow\ Sf
\using {\forall}L
\endprooftree
\justifies
[\mbox{\fbox{${\blacksquare}{\forall}g({\forall}f((Sf{{}{\uparrow}{}}{\blacksquare}Nt(s(g))){{}{\downarrow}{}}Sf)/{\it CN}{\it s(g)})$}}, {\square}{\it CN}{\it s(n)}], {\square}({\langle\rangle}{\exists}gNt(s(g))\backslash Sf)\ \Rightarrow\ Sf
\using {\blacksquare}L
\endprooftree
\justifies
[{\blacksquare}{\forall}g({\forall}f((Sf{{}{\uparrow}{}}{\blacksquare}Nt(s(g))){{}{\downarrow}{}}Sf)/{\it CN}{\it s(g)}), {\square}{\it CN}{\it s(n)}], {\square}({\langle\rangle}{\exists}gNt(s(g))\backslash Sf)\ \Rightarrow\ {\square}Sf
\using {\Box}R
\endprooftree
\prooftree
\justifies
\mbox{\fbox{${\it CP}that$}}\ \Rightarrow\ {\it CP}that
\endprooftree
\justifies
\mbox{\fbox{${\it CP}that/{\square}Sf$}}, [{\blacksquare}{\forall}g({\forall}f((Sf{{}{\uparrow}{}}{\blacksquare}Nt(s(g))){{}{\downarrow}{}}Sf)/{\it CN}{\it s(g)}), {\square}{\it CN}{\it s(n)}], {\square}({\langle\rangle}{\exists}gNt(s(g))\backslash Sf)\ \Rightarrow\ {\it CP}that
\using {/}L
\endprooftree
\justifies
\mbox{\fbox{${\blacksquare}({\it CP}that/{\square}Sf)$}}, [{\blacksquare}{\forall}g({\forall}f((Sf{{}{\uparrow}{}}{\blacksquare}Nt(s(g))){{}{\downarrow}{}}Sf)/{\it CN}{\it s(g)}), {\square}{\it CN}{\it s(n)}], {\square}({\langle\rangle}{\exists}gNt(s(g))\backslash Sf)\ \Rightarrow\ {\it CP}that
\using {\blacksquare}L
\endprooftree
\justifies
{\blacksquare}({\it CP}that/{\square}Sf), [{\blacksquare}{\forall}g({\forall}f((Sf{{}{\uparrow}{}}{\blacksquare}Nt(s(g))){{}{\downarrow}{}}Sf)/{\it CN}{\it s(g)}), {\square}{\it CN}{\it s(n)}], {\square}({\langle\rangle}{\exists}gNt(s(g))\backslash Sf)\ \Rightarrow\ \fbox{${\it CP}that{\sqcup}{\square}Sf$}
\using {\sqcup}R
\endprooftree
\prooftree
\prooftree
\prooftree
\prooftree
\justifies
Nt(s(m))\ \Rightarrow\ Nt(s(m))
\endprooftree
\justifies
Nt(s(m))\ \Rightarrow\ \fbox{${\exists}gNt(s(g))$}
\using {\exists}R
\endprooftree
\justifies
[Nt(s(m))]\ \Rightarrow\ \fbox{${\langle\rangle}{\exists}gNt(s(g))$}
\using {\langle\rangle}R
\endprooftree
\prooftree
\justifies
\mbox{\fbox{$Sf$}}\ \Rightarrow\ Sf
\endprooftree
\justifies
[Nt(s(m))], \mbox{\fbox{${\langle\rangle}{\exists}gNt(s(g))\backslash Sf$}}\ \Rightarrow\ Sf
\using {\backslash}L
\endprooftree
\justifies
[Nt(s(m))], \mbox{\fbox{$({\langle\rangle}{\exists}gNt(s(g))\backslash Sf)/({\it CP}that{\sqcup}{\square}Sf)$}}, {\blacksquare}({\it CP}that/{\square}Sf), [{\blacksquare}{\forall}g({\forall}f((Sf{{}{\uparrow}{}}{\blacksquare}Nt(s(g))){{}{\downarrow}{}}Sf)/{\it CN}{\it s(g)}), {\square}{\it CN}{\it s(n)}], {\square}({\langle\rangle}{\exists}gNt(s(g))\backslash Sf)\ \Rightarrow\ Sf
\using {/}L
\endprooftree
\justifies
[Nt(s(m))], \mbox{\fbox{${\square}(({\langle\rangle}{\exists}gNt(s(g))\backslash Sf)/({\it CP}that{\sqcup}{\square}Sf))$}}, {\blacksquare}({\it CP}that/{\square}Sf), [{\blacksquare}{\forall}g({\forall}f((Sf{{}{\uparrow}{}}{\blacksquare}Nt(s(g))){{}{\downarrow}{}}Sf)/{\it CN}{\it s(g)}), {\square}{\it CN}{\it s(n)}], {\square}({\langle\rangle}{\exists}gNt(s(g))\backslash Sf)\ \Rightarrow\ Sf
\using {\Box}L
\endprooftree
\justifies
[{\tt 1}], {\square}(({\langle\rangle}{\exists}gNt(s(g))\backslash Sf)/({\it CP}that{\sqcup}{\square}Sf)), {\blacksquare}({\it CP}that/{\square}Sf), [{\blacksquare}{\forall}g({\forall}f((Sf{{}{\uparrow}{}}{\blacksquare}Nt(s(g))){{}{\downarrow}{}}Sf)/{\it CN}{\it s(g)}), {\square}{\it CN}{\it s(n)}], {\square}({\langle\rangle}{\exists}gNt(s(g))\backslash Sf)\ \Rightarrow\ Sf{{}{\uparrow}{}}Nt(s(m))
\using {\uparrow}R
\endprooftree
\prooftree
\justifies
\mbox{\fbox{$Sf$}}\ \Rightarrow\ Sf
\endprooftree
\justifies
[\mbox{\fbox{$(Sf{{}{\uparrow}{}}Nt(s(m))){{}{\downarrow}{}}Sf$}}], {\square}(({\langle\rangle}{\exists}gNt(s(g))\backslash Sf)/({\it CP}that{\sqcup}{\square}Sf)), {\blacksquare}({\it CP}that/{\square}Sf), [{\blacksquare}{\forall}g({\forall}f((Sf{{}{\uparrow}{}}{\blacksquare}Nt(s(g))){{}{\downarrow}{}}Sf)/{\it CN}{\it s(g)}), {\square}{\it CN}{\it s(n)}], {\square}({\langle\rangle}{\exists}gNt(s(g))\backslash Sf)\ \Rightarrow\ Sf
\using {\downarrow}L
\endprooftree
\justifies
[\mbox{\fbox{${\forall}f((Sf{{}{\uparrow}{}}Nt(s(m))){{}{\downarrow}{}}Sf)$}}], {\square}(({\langle\rangle}{\exists}gNt(s(g))\backslash Sf)/({\it CP}that{\sqcup}{\square}Sf)), {\blacksquare}({\it CP}that/{\square}Sf), [{\blacksquare}{\forall}g({\forall}f((Sf{{}{\uparrow}{}}{\blacksquare}Nt(s(g))){{}{\downarrow}{}}Sf)/{\it CN}{\it s(g)}), {\square}{\it CN}{\it s(n)}], {\square}({\langle\rangle}{\exists}gNt(s(g))\backslash Sf)\ \Rightarrow\ Sf
\using {\forall}L
\endprooftree
\justifies
[\mbox{\fbox{${\forall}f((Sf{{}{\uparrow}{}}Nt(s(m))){{}{\downarrow}{}}Sf)/{\it CN}{\it s(m)}$}}, {\square}{\it CN}{\it s(m)}], {\square}(({\langle\rangle}{\exists}gNt(s(g))\backslash Sf)/({\it CP}that{\sqcup}{\square}Sf)), {\blacksquare}({\it CP}that/{\square}Sf), [{\blacksquare}{\forall}g({\forall}f((Sf{{}{\uparrow}{}}{\blacksquare}Nt(s(g))){{}{\downarrow}{}}Sf)/{\it CN}{\it s(g)}), {\square}{\it CN}{\it s(n)}], {\square}({\langle\rangle}{\exists}gNt(s(g))\backslash Sf)\ \Rightarrow\ Sf
\using {/}L
\endprooftree
\justifies
[\mbox{\fbox{${\forall}g({\forall}f((Sf{{}{\uparrow}{}}Nt(s(g))){{}{\downarrow}{}}Sf)/{\it CN}{\it s(g)})$}}, {\square}{\it CN}{\it s(m)}], {\square}(({\langle\rangle}{\exists}gNt(s(g))\backslash Sf)/({\it CP}that{\sqcup}{\square}Sf)), {\blacksquare}({\it CP}that/{\square}Sf), [{\blacksquare}{\forall}g({\forall}f((Sf{{}{\uparrow}{}}{\blacksquare}Nt(s(g))){{}{\downarrow}{}}Sf)/{\it CN}{\it s(g)}), {\square}{\it CN}{\it s(n)}], {\square}({\langle\rangle}{\exists}gNt(s(g))\backslash Sf)\ \Rightarrow\ Sf
\using {\forall}L
\endprooftree
\justifies
[\mbox{\fbox{${\blacksquare}{\forall}g({\forall}f((Sf{{}{\uparrow}{}}Nt(s(g))){{}{\downarrow}{}}Sf)/{\it CN}{\it s(g)})$}}, {\square}{\it CN}{\it s(m)}], {\square}(({\langle\rangle}{\exists}gNt(s(g))\backslash Sf)/({\it CP}that{\sqcup}{\square}Sf)), {\blacksquare}({\it CP}that/{\square}Sf), [{\blacksquare}{\forall}g({\forall}f((Sf{{}{\uparrow}{}}{\blacksquare}Nt(s(g))){{}{\downarrow}{}}Sf)/{\it CN}{\it s(g)}), {\square}{\it CN}{\it s(n)}], {\square}({\langle\rangle}{\exists}gNt(s(g))\backslash Sf)\ \Rightarrow\ Sf
\using {\blacksquare}L
\endprooftree}
$$
\vspace{0.15in}
\noindent
This delivers semantics:
\disp{
$\forall C[(\mbox{\v{}}{\it man}\ {\it C})\rightarrow ({\it Pres}\ ((\mbox{\v{}}{\it believe}\ \mbox{\^{}}\exists J[(\mbox{\v{}}{\it fish}\ {\it J})\wedge ({\it Pres}\ (\mbox{\v{}}{\it walk}\ {\it J}))])\ {\it C}))]$}

The next example involves the {\it such that\/} relativization of Montague's fragment:
\disp{
(dwp((7-57))) $[{\bf every}{+}{\bf fish}{+}{\bf such}{+}{\bf that}{+}[{\bf it}]{+}{\bf walks}]{+}{\bf talks}: Sf$}
This is relativization in which the clause subordinate to {\it such that\/} must contain a
pronoun bound by the head which is modified by the relative clause.
Semantically,
the meaning of the body of the relative clause abstracted over the pronoun meaning
restricts the head modified.
The relative pronoun {\it such that\/} is categorized accordingly using the limited
contraction for anaphora type constructor.
Lexical lookup produces the following semantically labelled sequent:
\disp{
$[{\blacksquare}{\forall}g({\forall}f((Sf{{}{\uparrow}{}}Nt(s(g))){{}{\downarrow}{}}Sf)/{\it CN}{\it s(g)}): \lambda A\lambda B\forall C[({\it A}\ {\it C})\rightarrow ({\it B}\ {\it C})], {\square}{\it CN}{\it s(n)}: {\it fish},\\{\blacksquare}{\forall}n(({\it CN}{\it n}\backslash {\it CN}{\it n})/$ $(Sf{|}{\blacksquare}Nt(n))): \lambda D\lambda E\lambda F[({\it E}\ {\it F})\wedge ({\it D}\ {\it F})], [{\blacksquare}{[]^{-1}}{\forall}f(({\blacksquare}Sf{|}{\blacksquare}Nt(s(n)))/\\{}({\langle\rangle}Nt(s(n))\backslash Sf)): \lambda G{\it G}], $ ${\square}({\langle\rangle}{\exists}gNt(s(g))\backslash Sf): \mbox{\^{}}\lambda H({\it Pres}\ (\mbox{\v{}}{\it walk}\ {\it H}))], {\square}({\langle\rangle}{\exists}gNt(s(g))\backslash Sf):\\\mbox{\^{}}\lambda I({\it Pres}\ (\mbox{\v{}}{\it talk}\ {\it I}))\ \Rightarrow\ Sf$}
This sequent has the derivation:
\vspace{0.15in}
$$
{\tiny
\prooftree
\prooftree
\prooftree
\prooftree
\prooftree
\prooftree
\prooftree
\prooftree
\prooftree
\prooftree
\prooftree
\prooftree
\prooftree
\prooftree
\prooftree
\prooftree
\prooftree
\justifies
Nt(s(n))\ \Rightarrow\ Nt(s(n))
\endprooftree
\justifies
Nt(s(n))\ \Rightarrow\ \fbox{${\exists}gNt(s(g))$}
\using {\exists}R
\endprooftree
\justifies
[Nt(s(n))]\ \Rightarrow\ \fbox{${\langle\rangle}{\exists}gNt(s(g))$}
\using {\langle\rangle}R
\endprooftree
\prooftree
\justifies
\mbox{\fbox{$Sf$}}\ \Rightarrow\ Sf
\endprooftree
\justifies
[Nt(s(n))], \mbox{\fbox{${\langle\rangle}{\exists}gNt(s(g))\backslash Sf$}}\ \Rightarrow\ Sf
\using {\backslash}L
\endprooftree
\justifies
[Nt(s(n))], \mbox{\fbox{${\square}({\langle\rangle}{\exists}gNt(s(g))\backslash Sf)$}}\ \Rightarrow\ Sf
\using {\Box}L
\endprooftree
\justifies
{\langle\rangle}Nt(s(n)), {\square}({\langle\rangle}{\exists}gNt(s(g))\backslash Sf)\ \Rightarrow\ Sf
\using {\langle\rangle}L
\endprooftree
\justifies
{\square}({\langle\rangle}{\exists}gNt(s(g))\backslash Sf)\ \Rightarrow\ {\langle\rangle}Nt(s(n))\backslash Sf
\using {\backslash}R
\endprooftree
\prooftree
\prooftree
\prooftree
\justifies
\mbox{\fbox{$Sf$}}\ \Rightarrow\ Sf
\endprooftree
\justifies
\mbox{\fbox{${\blacksquare}Sf$}}\ \Rightarrow\ Sf
\using {\blacksquare}L
\endprooftree
\justifies
\mbox{\fbox{${\blacksquare}Sf{|}{\blacksquare}Nt(s(n))$}}\ \Rightarrow\ {\blacksquare}Sf{|}{\blacksquare}Nt(s(n))
\using {|}R
\endprooftree
\justifies
\mbox{\fbox{$({\blacksquare}Sf{|}{\blacksquare}Nt(s(n)))/({\langle\rangle}Nt(s(n))\backslash Sf)$}}, {\square}({\langle\rangle}{\exists}gNt(s(g))\backslash Sf)\ \Rightarrow\ Sf{|}{\blacksquare}Nt(s(n))
\using {/}L
\endprooftree
\justifies
\mbox{\fbox{${\forall}f(({\blacksquare}Sf{|}{\blacksquare}Nt(s(n)))/({\langle\rangle}Nt(s(n))\backslash Sf))$}}, {\square}({\langle\rangle}{\exists}gNt(s(g))\backslash Sf)\ \Rightarrow\ Sf{|}{\blacksquare}Nt(s(n))
\using {\forall}L
\endprooftree
\justifies
[\mbox{\fbox{${[]^{-1}}{\forall}f(({\blacksquare}Sf{|}{\blacksquare}Nt(s(n)))/({\langle\rangle}Nt(s(n))\backslash Sf))$}}], {\square}({\langle\rangle}{\exists}gNt(s(g))\backslash Sf)\ \Rightarrow\ Sf{|}{\blacksquare}Nt(s(n))
\using {[]^{-1}}L
\endprooftree
\justifies
[\mbox{\fbox{${\blacksquare}{[]^{-1}}{\forall}f(({\blacksquare}Sf{|}{\blacksquare}Nt(s(n)))/({\langle\rangle}Nt(s(n))\backslash Sf))$}}], {\square}({\langle\rangle}{\exists}gNt(s(g))\backslash Sf)\ \Rightarrow\ Sf{|}{\blacksquare}Nt(s(n))
\using {\blacksquare}L
\endprooftree
\prooftree
\prooftree
\prooftree
\justifies
\mbox{\fbox{${\it CN}{\it s(n)}$}}\ \Rightarrow\ {\it CN}{\it s(n)}
\endprooftree
\justifies
\mbox{\fbox{${\square}{\it CN}{\it s(n)}$}}\ \Rightarrow\ {\it CN}{\it s(n)}
\using {\Box}L
\endprooftree
\prooftree
\justifies
\mbox{\fbox{${\it CN}{\it s(n)}$}}\ \Rightarrow\ {\it CN}{\it s(n)}
\endprooftree
\justifies
{\square}{\it CN}{\it s(n)}, \mbox{\fbox{${\it CN}{\it s(n)}\backslash {\it CN}{\it s(n)}$}}\ \Rightarrow\ {\it CN}{\it s(n)}
\using {\backslash}L
\endprooftree
\justifies
{\square}{\it CN}{\it s(n)}, \mbox{\fbox{$({\it CN}{\it s(n)}\backslash {\it CN}{\it s(n)})/(Sf{|}{\blacksquare}Nt(s(n)))$}}, [{\blacksquare}{[]^{-1}}{\forall}f(({\blacksquare}Sf{|}{\blacksquare}Nt(s(n)))/({\langle\rangle}Nt(s(n))\backslash Sf))], {\square}({\langle\rangle}{\exists}gNt(s(g))\backslash Sf)\ \Rightarrow\ {\it CN}{\it s(n)}
\using {/}L
\endprooftree
\justifies
{\square}{\it CN}{\it s(n)}, \mbox{\fbox{${\forall}n(({\it CN}{\it n}\backslash {\it CN}{\it n})/(Sf{|}{\blacksquare}Nt(n)))$}}, [{\blacksquare}{[]^{-1}}{\forall}f(({\blacksquare}Sf{|}{\blacksquare}Nt(s(n)))/({\langle\rangle}Nt(s(n))\backslash Sf))], {\square}({\langle\rangle}{\exists}gNt(s(g))\backslash Sf)\ \Rightarrow\ {\it CN}{\it s(n)}
\using {\forall}L
\endprooftree
\justifies
{\square}{\it CN}{\it s(n)}, \mbox{\fbox{${\blacksquare}{\forall}n(({\it CN}{\it n}\backslash {\it CN}{\it n})/(Sf{|}{\blacksquare}Nt(n)))$}}, [{\blacksquare}{[]^{-1}}{\forall}f(({\blacksquare}Sf{|}{\blacksquare}Nt(s(n)))/({\langle\rangle}Nt(s(n))\backslash Sf))], {\square}({\langle\rangle}{\exists}gNt(s(g))\backslash Sf)\ \Rightarrow\ {\it CN}{\it s(n)}
\using {\blacksquare}L
\endprooftree
\prooftree
\prooftree
\prooftree
\prooftree
\prooftree
\prooftree
\prooftree
\prooftree
\justifies
Nt(s(n))\ \Rightarrow\ Nt(s(n))
\endprooftree
\justifies
Nt(s(n))\ \Rightarrow\ \fbox{${\exists}gNt(s(g))$}
\using {\exists}R
\endprooftree
\justifies
[Nt(s(n))]\ \Rightarrow\ \fbox{${\langle\rangle}{\exists}gNt(s(g))$}
\using {\langle\rangle}R
\endprooftree
\prooftree
\justifies
\mbox{\fbox{$Sf$}}\ \Rightarrow\ Sf
\endprooftree
\justifies
[Nt(s(n))], \mbox{\fbox{${\langle\rangle}{\exists}gNt(s(g))\backslash Sf$}}\ \Rightarrow\ Sf
\using {\backslash}L
\endprooftree
\justifies
[Nt(s(n))], \mbox{\fbox{${\square}({\langle\rangle}{\exists}gNt(s(g))\backslash Sf)$}}\ \Rightarrow\ Sf
\using {\Box}L
\endprooftree
\justifies
[{\tt 1}], {\square}({\langle\rangle}{\exists}gNt(s(g))\backslash Sf)\ \Rightarrow\ Sf{{}{\uparrow}{}}Nt(s(n))
\using {\uparrow}R
\endprooftree
\prooftree
\justifies
\mbox{\fbox{$Sf$}}\ \Rightarrow\ Sf
\endprooftree
\justifies
[\mbox{\fbox{$(Sf{{}{\uparrow}{}}Nt(s(n))){{}{\downarrow}{}}Sf$}}], {\square}({\langle\rangle}{\exists}gNt(s(g))\backslash Sf)\ \Rightarrow\ Sf
\using {\downarrow}L
\endprooftree
\justifies
[\mbox{\fbox{${\forall}f((Sf{{}{\uparrow}{}}Nt(s(n))){{}{\downarrow}{}}Sf)$}}], {\square}({\langle\rangle}{\exists}gNt(s(g))\backslash Sf)\ \Rightarrow\ Sf
\using {\forall}L
\endprooftree
\justifies
[\mbox{\fbox{${\forall}f((Sf{{}{\uparrow}{}}Nt(s(n))){{}{\downarrow}{}}Sf)/{\it CN}{\it s(n)}$}}, {\square}{\it CN}{\it s(n)}, {\blacksquare}{\forall}n(({\it CN}{\it n}\backslash {\it CN}{\it n})/(Sf{|}{\blacksquare}Nt(n))), [{\blacksquare}{[]^{-1}}{\forall}f(({\blacksquare}Sf{|}{\blacksquare}Nt(s(n)))/({\langle\rangle}Nt(s(n))\backslash Sf))], {\square}({\langle\rangle}{\exists}gNt(s(g))\backslash Sf)], {\square}({\langle\rangle}{\exists}gNt(s(g))\backslash Sf)\ \Rightarrow\ Sf
\using {/}L
\endprooftree
\justifies
[\mbox{\fbox{${\forall}g({\forall}f((Sf{{}{\uparrow}{}}Nt(s(g))){{}{\downarrow}{}}Sf)/{\it CN}{\it s(g)})$}}, {\square}{\it CN}{\it s(n)}, {\blacksquare}{\forall}n(({\it CN}{\it n}\backslash {\it CN}{\it n})/(Sf{|}{\blacksquare}Nt(n))), [{\blacksquare}{[]^{-1}}{\forall}f(({\blacksquare}Sf{|}{\blacksquare}Nt(s(n)))/({\langle\rangle}Nt(s(n))\backslash Sf))], {\square}({\langle\rangle}{\exists}gNt(s(g))\backslash Sf)], {\square}({\langle\rangle}{\exists}gNt(s(g))\backslash Sf)\ \Rightarrow\ Sf
\using {\forall}L
\endprooftree
\justifies
[\mbox{\fbox{${\blacksquare}{\forall}g({\forall}f((Sf{{}{\uparrow}{}}Nt(s(g))){{}{\downarrow}{}}Sf)/{\it CN}{\it s(g)})$}}, {\square}{\it CN}{\it s(n)}, {\blacksquare}{\forall}n(({\it CN}{\it n}\backslash {\it CN}{\it n})/(Sf{|}{\blacksquare}Nt(n))), [{\blacksquare}{[]^{-1}}{\forall}f(({\blacksquare}Sf{|}{\blacksquare}Nt(s(n)))/({\langle\rangle}Nt(s(n))\backslash Sf))], {\square}({\langle\rangle}{\exists}gNt(s(g))\backslash Sf)], {\square}({\langle\rangle}{\exists}gNt(s(g))\backslash Sf)\ \Rightarrow\ Sf
\using {\blacksquare}L
\endprooftree}
$$
\vspace{0.15in}
\noindent
This delivers semantics:
\disp{
$\forall C[[(\mbox{\v{}}{\it fish}\ {\it C})\wedge ({\it Pres}\ (\mbox{\v{}}{\it walk}\ {\it C}))]\rightarrow ({\it Pres}\ (\mbox{\v{}}{\it talk}\ {\it C}))]$}

The next example of DWP involves {\it seek\/} which is an intensional object verb
synonymous with {\it try to find}:
\disp{
(dwp((7-60, 62))) $[{\bf john}]{+}{\bf seeks}{+}{\bf a}{+}{\bf unicorn}: Sf$}
The sentence has a specific reading in which there is a particular unicorn
which John is trying to find,
and a non-specific reading in which John is just trying to bring it about that he finds some,
any,
unicorn.
The former reading has existential commitment with respect to unicorns,
the latter does not.
Lexical lookup gives the following semantically labelled sequent:
\disp{
$[{\blacksquare}Nt(s(m)): {\it j}], {\square}(({\langle\rangle}{\exists}gNt(s(g))\backslash Sf)/{\square}{\forall}a{\forall}f(((Na\backslash Sf)/{\exists}bNb)\backslash (Na\backslash Sf))):\\
 \mbox{\^{}}\lambda A\lambda B((\mbox{\v{}}{\it tries}\ \mbox{\^{}}((\mbox{\v{}}{\it A}\ \mbox{\v{}}{\it find})\ {\it B}))\ {\it B}), {\blacksquare}{\forall}g({\forall}f((Sf{{}{\uparrow}{}}{\blacksquare}Nt(s(g))){{}{\downarrow}{}}Sf)/{\it CN}{\it s(g)}): \lambda C\lambda D\exists E[({\it C}\ {\it E})\wedge ({\it D}\ {\it E})],$ ${\square}{\it CN}{\it s(n)}: {\it unicorn}\ \Rightarrow\ Sf$}
For the specific reading there is the derivation:
\vspace{0.15in}
$$
{\tiny
\prooftree
\prooftree
\prooftree
\prooftree
\prooftree
\justifies
\mbox{\fbox{${\it CN}{\it s(n)}$}}\ \Rightarrow\ {\it CN}{\it s(n)}
\endprooftree
\justifies
\mbox{\fbox{${\square}{\it CN}{\it s(n)}$}}\ \Rightarrow\ {\it CN}{\it s(n)}
\using {\Box}L
\endprooftree
\prooftree
\prooftree
\prooftree
\prooftree
\prooftree
\prooftree
\prooftree
\prooftree
\prooftree
\prooftree
\prooftree
\prooftree
\prooftree
\prooftree
\justifies
\mbox{\fbox{$Nt(s(n))$}}\ \Rightarrow\ Nt(s(n))
\endprooftree
\justifies
\mbox{\fbox{${\blacksquare}Nt(s(n))$}}\ \Rightarrow\ Nt(s(n))
\using {\blacksquare}L
\endprooftree
\justifies
{\blacksquare}Nt(s(n))\ \Rightarrow\ \fbox{${\exists}bNb$}
\using {\exists}R
\endprooftree
\prooftree
\prooftree
\justifies
N21\ \Rightarrow\ N21
\endprooftree
\prooftree
\justifies
\mbox{\fbox{$S22$}}\ \Rightarrow\ S22
\endprooftree
\justifies
N21, \mbox{\fbox{$N21\backslash S22$}}\ \Rightarrow\ S22
\using {\backslash}L
\endprooftree
\justifies
N21, \mbox{\fbox{$(N21\backslash S22)/{\exists}bNb$}}, {\blacksquare}Nt(s(n))\ \Rightarrow\ S22
\using {/}L
\endprooftree
\justifies
(N21\backslash S22)/{\exists}bNb, {\blacksquare}Nt(s(n))\ \Rightarrow\ N21\backslash S22
\using {\backslash}R
\endprooftree
\justifies
{\blacksquare}Nt(s(n))\ \Rightarrow\ ((N21\backslash S22)/{\exists}bNb)\backslash (N21\backslash S22)
\using {\backslash}R
\endprooftree
\justifies
{\blacksquare}Nt(s(n))\ \Rightarrow\ {\forall}f(((N21\backslash Sf)/{\exists}bNb)\backslash (N21\backslash Sf))
\using {\forall}R
\endprooftree
\justifies
{\blacksquare}Nt(s(n))\ \Rightarrow\ {\forall}a{\forall}f(((Na\backslash Sf)/{\exists}bNb)\backslash (Na\backslash Sf))
\using {\forall}R
\endprooftree
\justifies
{\blacksquare}Nt(s(n))\ \Rightarrow\ {\square}{\forall}a{\forall}f(((Na\backslash Sf)/{\exists}bNb)\backslash (Na\backslash Sf))
\using {\Box}R
\endprooftree
\prooftree
\prooftree
\prooftree
\prooftree
\prooftree
\justifies
\mbox{\fbox{$Nt(s(m))$}}\ \Rightarrow\ Nt(s(m))
\endprooftree
\justifies
\mbox{\fbox{${\blacksquare}Nt(s(m))$}}\ \Rightarrow\ Nt(s(m))
\using {\blacksquare}L
\endprooftree
\justifies
{\blacksquare}Nt(s(m))\ \Rightarrow\ \fbox{${\exists}gNt(s(g))$}
\using {\exists}R
\endprooftree
\justifies
[{\blacksquare}Nt(s(m))]\ \Rightarrow\ \fbox{${\langle\rangle}{\exists}gNt(s(g))$}
\using {\langle\rangle}R
\endprooftree
\prooftree
\justifies
\mbox{\fbox{$Sf$}}\ \Rightarrow\ Sf
\endprooftree
\justifies
[{\blacksquare}Nt(s(m))], \mbox{\fbox{${\langle\rangle}{\exists}gNt(s(g))\backslash Sf$}}\ \Rightarrow\ Sf
\using {\backslash}L
\endprooftree
\justifies
[{\blacksquare}Nt(s(m))], \mbox{\fbox{$({\langle\rangle}{\exists}gNt(s(g))\backslash Sf)/{\square}{\forall}a{\forall}f(((Na\backslash Sf)/{\exists}bNb)\backslash (Na\backslash Sf))$}}, {\blacksquare}Nt(s(n))\ \Rightarrow\ Sf
\using {/}L
\endprooftree
\justifies
[{\blacksquare}Nt(s(m))], \mbox{\fbox{${\square}(({\langle\rangle}{\exists}gNt(s(g))\backslash Sf)/{\square}{\forall}a{\forall}f(((Na\backslash Sf)/{\exists}bNb)\backslash (Na\backslash Sf)))$}}, {\blacksquare}Nt(s(n))\ \Rightarrow\ Sf
\using {\Box}L
\endprooftree
\justifies
[{\blacksquare}Nt(s(m))], {\square}(({\langle\rangle}{\exists}gNt(s(g))\backslash Sf)/{\square}{\forall}a{\forall}f(((Na\backslash Sf)/{\exists}bNb)\backslash (Na\backslash Sf))), {\tt 1}\ \Rightarrow\ Sf{{}{\uparrow}{}}{\blacksquare}Nt(s(n))
\using {\uparrow}R
\endprooftree
\prooftree
\justifies
\mbox{\fbox{$Sf$}}\ \Rightarrow\ Sf
\endprooftree
\justifies
[{\blacksquare}Nt(s(m))], {\square}(({\langle\rangle}{\exists}gNt(s(g))\backslash Sf)/{\square}{\forall}a{\forall}f(((Na\backslash Sf)/{\exists}bNb)\backslash (Na\backslash Sf))), \mbox{\fbox{$(Sf{{}{\uparrow}{}}{\blacksquare}Nt(s(n))){{}{\downarrow}{}}Sf$}}\ \Rightarrow\ Sf
\using {\downarrow}L
\endprooftree
\justifies
[{\blacksquare}Nt(s(m))], {\square}(({\langle\rangle}{\exists}gNt(s(g))\backslash Sf)/{\square}{\forall}a{\forall}f(((Na\backslash Sf)/{\exists}bNb)\backslash (Na\backslash Sf))), \mbox{\fbox{${\forall}f((Sf{{}{\uparrow}{}}{\blacksquare}Nt(s(n))){{}{\downarrow}{}}Sf)$}}\ \Rightarrow\ Sf
\using {\forall}L
\endprooftree
\justifies
[{\blacksquare}Nt(s(m))], {\square}(({\langle\rangle}{\exists}gNt(s(g))\backslash Sf)/{\square}{\forall}a{\forall}f(((Na\backslash Sf)/{\exists}bNb)\backslash (Na\backslash Sf))), \mbox{\fbox{${\forall}f((Sf{{}{\uparrow}{}}{\blacksquare}Nt(s(n))){{}{\downarrow}{}}Sf)/{\it CN}{\it s(n)}$}}, {\square}{\it CN}{\it s(n)}\ \Rightarrow\ Sf
\using {/}L
\endprooftree
\justifies
[{\blacksquare}Nt(s(m))], {\square}(({\langle\rangle}{\exists}gNt(s(g))\backslash Sf)/{\square}{\forall}a{\forall}f(((Na\backslash Sf)/{\exists}bNb)\backslash (Na\backslash Sf))), \mbox{\fbox{${\forall}g({\forall}f((Sf{{}{\uparrow}{}}{\blacksquare}Nt(s(g))){{}{\downarrow}{}}Sf)/{\it CN}{\it s(g)})$}}, {\square}{\it CN}{\it s(n)}\ \Rightarrow\ Sf
\using {\forall}L
\endprooftree
\justifies
[{\blacksquare}Nt(s(m))], {\square}(({\langle\rangle}{\exists}gNt(s(g))\backslash Sf)/{\square}{\forall}a{\forall}f(((Na\backslash Sf)/{\exists}bNb)\backslash (Na\backslash Sf))), \mbox{\fbox{${\blacksquare}{\forall}g({\forall}f((Sf{{}{\uparrow}{}}{\blacksquare}Nt(s(g))){{}{\downarrow}{}}Sf)/{\it CN}{\it s(g)})$}}, {\square}{\it CN}{\it s(n)}\ \Rightarrow\ Sf
\using {\blacksquare}L
\endprooftree}
$$
\vspace{0.15in}
\noindent
This delivers the semantics with existential commitment:
\disp{ 
$\exists C[(\mbox{\v{}}{\it unicorn}\ {\it C})\wedge ((\mbox{\v{}}{\it tries}\ \mbox{\^{}}((\mbox{\v{}}{\it find}\ {\it C})\ {\it j}))\ {\it j})]$}

For the non-specific reading there is the derivation:
\vspace{0.15in}
$$
{\tiny
\prooftree
\prooftree
\prooftree
\prooftree
\prooftree
\prooftree
\prooftree
\prooftree
\prooftree
\prooftree
\prooftree
\prooftree
\justifies
\mbox{\fbox{${\it CN}{\it s(n)}$}}\ \Rightarrow\ {\it CN}{\it s(n)}
\endprooftree
\justifies
\mbox{\fbox{${\square}{\it CN}{\it s(n)}$}}\ \Rightarrow\ {\it CN}{\it s(n)}
\using {\Box}L
\endprooftree
\prooftree
\prooftree
\prooftree
\prooftree
\prooftree
\prooftree
\prooftree
\justifies
\mbox{\fbox{$Nt(s(n))$}}\ \Rightarrow\ Nt(s(n))
\endprooftree
\justifies
\mbox{\fbox{${\blacksquare}Nt(s(n))$}}\ \Rightarrow\ Nt(s(n))
\using {\blacksquare}L
\endprooftree
\justifies
{\blacksquare}Nt(s(n))\ \Rightarrow\ \fbox{${\exists}bNb$}
\using {\exists}R
\endprooftree
\prooftree
\prooftree
\justifies
N19\ \Rightarrow\ N19
\endprooftree
\prooftree
\justifies
\mbox{\fbox{$S20$}}\ \Rightarrow\ S20
\endprooftree
\justifies
N19, \mbox{\fbox{$N19\backslash S20$}}\ \Rightarrow\ S20
\using {\backslash}L
\endprooftree
\justifies
N19, \mbox{\fbox{$(N19\backslash S20)/{\exists}bNb$}}, {\blacksquare}Nt(s(n))\ \Rightarrow\ S20
\using {/}L
\endprooftree
\justifies
N19, (N19\backslash S20)/{\exists}bNb, {\tt 1}\ \Rightarrow\ S20{{}{\uparrow}{}}{\blacksquare}Nt(s(n))
\using {\uparrow}R
\endprooftree
\prooftree
\justifies
\mbox{\fbox{$S20$}}\ \Rightarrow\ S20
\endprooftree
\justifies
N19, (N19\backslash S20)/{\exists}bNb, \mbox{\fbox{$(S20{{}{\uparrow}{}}{\blacksquare}Nt(s(n))){{}{\downarrow}{}}S20$}}\ \Rightarrow\ S20
\using {\downarrow}L
\endprooftree
\justifies
N19, (N19\backslash S20)/{\exists}bNb, \mbox{\fbox{${\forall}f((Sf{{}{\uparrow}{}}{\blacksquare}Nt(s(n))){{}{\downarrow}{}}Sf)$}}\ \Rightarrow\ S20
\using {\forall}L
\endprooftree
\justifies
N19, (N19\backslash S20)/{\exists}bNb, \mbox{\fbox{${\forall}f((Sf{{}{\uparrow}{}}{\blacksquare}Nt(s(n))){{}{\downarrow}{}}Sf)/{\it CN}{\it s(n)}$}}, {\square}{\it CN}{\it s(n)}\ \Rightarrow\ S20
\using {/}L
\endprooftree
\justifies
N19, (N19\backslash S20)/{\exists}bNb, \mbox{\fbox{${\forall}g({\forall}f((Sf{{}{\uparrow}{}}{\blacksquare}Nt(s(g))){{}{\downarrow}{}}Sf)/{\it CN}{\it s(g)})$}}, {\square}{\it CN}{\it s(n)}\ \Rightarrow\ S20
\using {\forall}L
\endprooftree
\justifies
N19, (N19\backslash S20)/{\exists}bNb, \mbox{\fbox{${\blacksquare}{\forall}g({\forall}f((Sf{{}{\uparrow}{}}{\blacksquare}Nt(s(g))){{}{\downarrow}{}}Sf)/{\it CN}{\it s(g)})$}}, {\square}{\it CN}{\it s(n)}\ \Rightarrow\ S20
\using {\blacksquare}L
\endprooftree
\justifies
(N19\backslash S20)/{\exists}bNb, {\blacksquare}{\forall}g({\forall}f((Sf{{}{\uparrow}{}}{\blacksquare}Nt(s(g))){{}{\downarrow}{}}Sf)/{\it CN}{\it s(g)}), {\square}{\it CN}{\it s(n)}\ \Rightarrow\ N19\backslash S20
\using {\backslash}R
\endprooftree
\justifies
{\blacksquare}{\forall}g({\forall}f((Sf{{}{\uparrow}{}}{\blacksquare}Nt(s(g))){{}{\downarrow}{}}Sf)/{\it CN}{\it s(g)}), {\square}{\it CN}{\it s(n)}\ \Rightarrow\ ((N19\backslash S20)/{\exists}bNb)\backslash (N19\backslash S20)
\using {\backslash}R
\endprooftree
\justifies
{\blacksquare}{\forall}g({\forall}f((Sf{{}{\uparrow}{}}{\blacksquare}Nt(s(g))){{}{\downarrow}{}}Sf)/{\it CN}{\it s(g)}), {\square}{\it CN}{\it s(n)}\ \Rightarrow\ {\forall}f(((N19\backslash Sf)/{\exists}bNb)\backslash (N19\backslash Sf))
\using {\forall}R
\endprooftree
\justifies
{\blacksquare}{\forall}g({\forall}f((Sf{{}{\uparrow}{}}{\blacksquare}Nt(s(g))){{}{\downarrow}{}}Sf)/{\it CN}{\it s(g)}), {\square}{\it CN}{\it s(n)}\ \Rightarrow\ {\forall}a{\forall}f(((Na\backslash Sf)/{\exists}bNb)\backslash (Na\backslash Sf))
\using {\forall}R
\endprooftree
\justifies
{\blacksquare}{\forall}g({\forall}f((Sf{{}{\uparrow}{}}{\blacksquare}Nt(s(g))){{}{\downarrow}{}}Sf)/{\it CN}{\it s(g)}), {\square}{\it CN}{\it s(n)}\ \Rightarrow\ {\square}{\forall}a{\forall}f(((Na\backslash Sf)/{\exists}bNb)\backslash (Na\backslash Sf))
\using {\Box}R
\endprooftree
\prooftree
\prooftree
\prooftree
\prooftree
\prooftree
\justifies
\mbox{\fbox{$Nt(s(m))$}}\ \Rightarrow\ Nt(s(m))
\endprooftree
\justifies
\mbox{\fbox{${\blacksquare}Nt(s(m))$}}\ \Rightarrow\ Nt(s(m))
\using {\blacksquare}L
\endprooftree
\justifies
{\blacksquare}Nt(s(m))\ \Rightarrow\ \fbox{${\exists}gNt(s(g))$}
\using {\exists}R
\endprooftree
\justifies
[{\blacksquare}Nt(s(m))]\ \Rightarrow\ \fbox{${\langle\rangle}{\exists}gNt(s(g))$}
\using {\langle\rangle}R
\endprooftree
\prooftree
\justifies
\mbox{\fbox{$Sf$}}\ \Rightarrow\ Sf
\endprooftree
\justifies
[{\blacksquare}Nt(s(m))], \mbox{\fbox{${\langle\rangle}{\exists}gNt(s(g))\backslash Sf$}}\ \Rightarrow\ Sf
\using {\backslash}L
\endprooftree
\justifies
[{\blacksquare}Nt(s(m))], \mbox{\fbox{$({\langle\rangle}{\exists}gNt(s(g))\backslash Sf)/{\square}{\forall}a{\forall}f(((Na\backslash Sf)/{\exists}bNb)\backslash (Na\backslash Sf))$}}, {\blacksquare}{\forall}g({\forall}f((Sf{{}{\uparrow}{}}{\blacksquare}Nt(s(g))){{}{\downarrow}{}}Sf)/{\it CN}{\it s(g)}), {\square}{\it CN}{\it s(n)}\ \Rightarrow\ Sf
\using {/}L
\endprooftree
\justifies
[{\blacksquare}Nt(s(m))], \mbox{\fbox{${\square}(({\langle\rangle}{\exists}gNt(s(g))\backslash Sf)/{\square}{\forall}a{\forall}f(((Na\backslash Sf)/{\exists}bNb)\backslash (Na\backslash Sf)))$}}, {\blacksquare}{\forall}g({\forall}f((Sf{{}{\uparrow}{}}{\blacksquare}Nt(s(g))){{}{\downarrow}{}}Sf)/{\it CN}{\it s(g)}), {\square}{\it CN}{\it s(n)}\ \Rightarrow\ Sf
\using {\Box}L
\endprooftree}
$$
\vspace{0.15in}
\noindent
This delivers the semantics without existential commitment:
\disp{
$((\mbox{\v{}}{\it tries}\ \mbox{\^{}}\exists G[(\mbox{\v{}}{\it unicorn}\ {\it G})\wedge ((\mbox{\v{}}{\it find}\ {\it G})\ {\it j})])\ {\it j})$}

The next examples involve the copula of identity. First, and minimally:
\disp{
(dwp((7-73))) $[{\bf john}]{+}{\bf is}{+}{\bf bill}: Sf$}
For this there is the semantically labelled sequent:
\disp{
$[{\blacksquare}Nt(s(m)): {\it j}], {\blacksquare}(({\langle\rangle}{\exists}gNt(s(g))\backslash Sf)/({\exists}aNa{\oplus}({\exists}g(({\it CN}{\it g}/{\it CN}{\it g}){\sqcup}({\it CN}{\it g}\backslash {\it CN}{\it g})){-}I))):\\\lambda A\lambda B({\it Pres}$\ $({\it A}\casearrow C.[{\it B}={\it C}]; D.(({\it D}\ \lambda E[{\it E}={\it B}])\ {\it B}))), {\blacksquare}Nt(s(m)): {\it b}\ \Rightarrow\ Sf$}
This has the derivation:
\vspace{0.15in}
$$
\resizebox{\textwidth}{!}{
\prooftree
\prooftree
\prooftree
\prooftree
\prooftree
\prooftree
\justifies
\mbox{\fbox{$Nt(s(m))$}}\ \Rightarrow\ Nt(s(m))
\endprooftree
\justifies
\mbox{\fbox{${\blacksquare}Nt(s(m))$}}\ \Rightarrow\ Nt(s(m))
\using {\blacksquare}L
\endprooftree
\justifies
{\blacksquare}Nt(s(m))\ \Rightarrow\ \fbox{${\exists}aNa$}
\using {\exists}R
\endprooftree
\justifies
{\blacksquare}Nt(s(m))\ \Rightarrow\ \fbox{${\exists}aNa{\oplus}({\exists}g(({\it CN}{\it g}/{\it CN}{\it g}){\sqcup}({\it CN}{\it g}\backslash {\it CN}{\it g})){-}I)$}
\using {\oplus}R
\endprooftree
\prooftree
\prooftree
\prooftree
\prooftree
\prooftree
\justifies
\mbox{\fbox{$Nt(s(m))$}}\ \Rightarrow\ Nt(s(m))
\endprooftree
\justifies
\mbox{\fbox{${\blacksquare}Nt(s(m))$}}\ \Rightarrow\ Nt(s(m))
\using {\blacksquare}L
\endprooftree
\justifies
{\blacksquare}Nt(s(m))\ \Rightarrow\ \fbox{${\exists}gNt(s(g))$}
\using {\exists}R
\endprooftree
\justifies
[{\blacksquare}Nt(s(m))]\ \Rightarrow\ \fbox{${\langle\rangle}{\exists}gNt(s(g))$}
\using {\langle\rangle}R
\endprooftree
\prooftree
\justifies
\mbox{\fbox{$Sf$}}\ \Rightarrow\ Sf
\endprooftree
\justifies
[{\blacksquare}Nt(s(m))], \mbox{\fbox{${\langle\rangle}{\exists}gNt(s(g))\backslash Sf$}}\ \Rightarrow\ Sf
\using {\backslash}L
\endprooftree
\justifies
[{\blacksquare}Nt(s(m))], \mbox{\fbox{$({\langle\rangle}{\exists}gNt(s(g))\backslash Sf)/({\exists}aNa{\oplus}({\exists}g(({\it CN}{\it g}/{\it CN}{\it g}){\sqcup}({\it CN}{\it g}\backslash {\it CN}{\it g})){-}I))$}}, {\blacksquare}Nt(s(m))\ \Rightarrow\ Sf
\using {/}L
\endprooftree
\justifies
[{\blacksquare}Nt(s(m))], \mbox{\fbox{${\blacksquare}(({\langle\rangle}{\exists}gNt(s(g))\backslash Sf)/({\exists}aNa{\oplus}({\exists}g(({\it CN}{\it g}/{\it CN}{\it g}){\sqcup}({\it CN}{\it g}\backslash {\it CN}{\it g})){-}I)))$}}, {\blacksquare}Nt(s(m))\ \Rightarrow\ Sf
\using {\blacksquare}L
\endprooftree}
$$
\vspace{0.15in}
\noindent
It delivers semantics:
\disp{
$({\it Pres}\ [{\it j}={\it b}])$}

Second, and more subtly:
\disp{
(dwp((7-76))) $[{\bf john}]{+}{\bf is}{+}{\bf a}{+}{\bf man}: Sf$}
Lexical lookup (with the same lexical entry for the copula)
yields the semantically annotated sequent:
\disp{
$[{\blacksquare}Nt(s(m)): {\it j}], {\blacksquare}(({\langle\rangle}{\exists}gNt(s(g))\backslash Sf)/({\exists}aNa{\oplus}({\exists}g(({\it CN}{\it g}/{\it CN}{\it g}){\sqcup}({\it CN}{\it g}\backslash {\it CN}{\it g})){-}I))): \lambda A\lambda B({\it Pres}$\ $({\it A}\casearrow C.[{\it B}={\it C}]; D.(({\it D}\ \lambda E[{\it E}={\it B}])\ {\it B}))), {\blacksquare}{\forall}g({\forall}f((Sf{{}{\uparrow}{}}{\blacksquare}Nt(s(g))){{}{\downarrow}{}}Sf)/{\it CN}{\it s(g)}): \lambda F\lambda G\exists H[({\it F}\ {\it H})\wedge ({\it G}\ {\it H})], {\square}{\it CN}{\it s(m)}: {\it man}\ \Rightarrow\ Sf$}
This has the derivation:
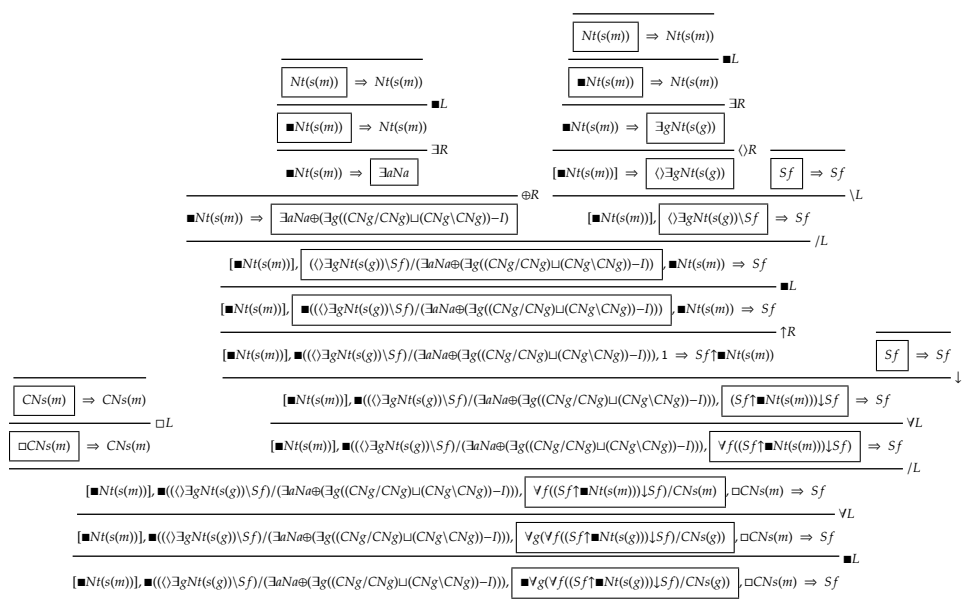
\vspace{0.15in}
$$
{\tiny
\prooftree
\prooftree
\prooftree
\prooftree
\prooftree
\justifies
\mbox{\fbox{${\it CN}{\it s(m)}$}}\ \Rightarrow\ {\it CN}{\it s(m)}
\endprooftree
\justifies
\mbox{\fbox{${\square}{\it CN}{\it s(m)}$}}\ \Rightarrow\ {\it CN}{\it s(m)}
\using {\Box}L
\endprooftree
\prooftree
\prooftree
\prooftree
\prooftree
\prooftree
\prooftree
\prooftree
\prooftree
\prooftree
\justifies
\mbox{\fbox{$Nt(s(m))$}}\ \Rightarrow\ Nt(s(m))
\endprooftree
\justifies
\mbox{\fbox{${\blacksquare}Nt(s(m))$}}\ \Rightarrow\ Nt(s(m))
\using {\blacksquare}L
\endprooftree
\justifies
{\blacksquare}Nt(s(m))\ \Rightarrow\ \fbox{${\exists}aNa$}
\using {\exists}R
\endprooftree
\justifies
{\blacksquare}Nt(s(m))\ \Rightarrow\ \fbox{${\exists}aNa{\oplus}({\exists}g(({\it CN}{\it g}/{\it CN}{\it g}){\sqcup}({\it CN}{\it g}\backslash {\it CN}{\it g})){-}I)$}
\using {\oplus}R
\endprooftree
\prooftree
\prooftree
\prooftree
\prooftree
\prooftree
\justifies
\mbox{\fbox{$Nt(s(m))$}}\ \Rightarrow\ Nt(s(m))
\endprooftree
\justifies
\mbox{\fbox{${\blacksquare}Nt(s(m))$}}\ \Rightarrow\ Nt(s(m))
\using {\blacksquare}L
\endprooftree
\justifies
{\blacksquare}Nt(s(m))\ \Rightarrow\ \fbox{${\exists}gNt(s(g))$}
\using {\exists}R
\endprooftree
\justifies
[{\blacksquare}Nt(s(m))]\ \Rightarrow\ \fbox{${\langle\rangle}{\exists}gNt(s(g))$}
\using {\langle\rangle}R
\endprooftree
\prooftree
\justifies
\mbox{\fbox{$Sf$}}\ \Rightarrow\ Sf
\endprooftree
\justifies
[{\blacksquare}Nt(s(m))], \mbox{\fbox{${\langle\rangle}{\exists}gNt(s(g))\backslash Sf$}}\ \Rightarrow\ Sf
\using {\backslash}L
\endprooftree
\justifies
[{\blacksquare}Nt(s(m))], \mbox{\fbox{$({\langle\rangle}{\exists}gNt(s(g))\backslash Sf)/({\exists}aNa{\oplus}({\exists}g(({\it CN}{\it g}/{\it CN}{\it g}){\sqcup}({\it CN}{\it g}\backslash {\it CN}{\it g})){-}I))$}}, {\blacksquare}Nt(s(m))\ \Rightarrow\ Sf
\using {/}L
\endprooftree
\justifies
[{\blacksquare}Nt(s(m))], \mbox{\fbox{${\blacksquare}(({\langle\rangle}{\exists}gNt(s(g))\backslash Sf)/({\exists}aNa{\oplus}({\exists}g(({\it CN}{\it g}/{\it CN}{\it g}){\sqcup}({\it CN}{\it g}\backslash {\it CN}{\it g})){-}I)))$}}, {\blacksquare}Nt(s(m))\ \Rightarrow\ Sf
\using {\blacksquare}L
\endprooftree
\justifies
[{\blacksquare}Nt(s(m))], {\blacksquare}(({\langle\rangle}{\exists}gNt(s(g))\backslash Sf)/({\exists}aNa{\oplus}({\exists}g(({\it CN}{\it g}/{\it CN}{\it g}){\sqcup}({\it CN}{\it g}\backslash {\it CN}{\it g})){-}I))), {\tt 1}\ \Rightarrow\ Sf{{}{\uparrow}{}}{\blacksquare}Nt(s(m))
\using {\uparrow}R
\endprooftree
\prooftree
\justifies
\mbox{\fbox{$Sf$}}\ \Rightarrow\ Sf
\endprooftree
\justifies
[{\blacksquare}Nt(s(m))], {\blacksquare}(({\langle\rangle}{\exists}gNt(s(g))\backslash Sf)/({\exists}aNa{\oplus}({\exists}g(({\it CN}{\it g}/{\it CN}{\it g}){\sqcup}({\it CN}{\it g}\backslash {\it CN}{\it g})){-}I))), \mbox{\fbox{$(Sf{{}{\uparrow}{}}{\blacksquare}Nt(s(m))){{}{\downarrow}{}}Sf$}}\ \Rightarrow\ Sf
\using {\downarrow}L
\endprooftree
\justifies
[{\blacksquare}Nt(s(m))], {\blacksquare}(({\langle\rangle}{\exists}gNt(s(g))\backslash Sf)/({\exists}aNa{\oplus}({\exists}g(({\it CN}{\it g}/{\it CN}{\it g}){\sqcup}({\it CN}{\it g}\backslash {\it CN}{\it g})){-}I))), \mbox{\fbox{${\forall}f((Sf{{}{\uparrow}{}}{\blacksquare}Nt(s(m))){{}{\downarrow}{}}Sf)$}}\ \Rightarrow\ Sf
\using {\forall}L
\endprooftree
\justifies
[{\blacksquare}Nt(s(m))], {\blacksquare}(({\langle\rangle}{\exists}gNt(s(g))\backslash Sf)/({\exists}aNa{\oplus}({\exists}g(({\it CN}{\it g}/{\it CN}{\it g}){\sqcup}({\it CN}{\it g}\backslash {\it CN}{\it g})){-}I))), \mbox{\fbox{${\forall}f((Sf{{}{\uparrow}{}}{\blacksquare}Nt(s(m))){{}{\downarrow}{}}Sf)/{\it CN}{\it s(m)}$}}, {\square}{\it CN}{\it s(m)}\ \Rightarrow\ Sf
\using {/}L
\endprooftree
\justifies
[{\blacksquare}Nt(s(m))], {\blacksquare}(({\langle\rangle}{\exists}gNt(s(g))\backslash Sf)/({\exists}aNa{\oplus}({\exists}g(({\it CN}{\it g}/{\it CN}{\it g}){\sqcup}({\it CN}{\it g}\backslash {\it CN}{\it g})){-}I))), \mbox{\fbox{${\forall}g({\forall}f((Sf{{}{\uparrow}{}}{\blacksquare}Nt(s(g))){{}{\downarrow}{}}Sf)/{\it CN}{\it s(g)})$}}, {\square}{\it CN}{\it s(m)}\ \Rightarrow\ Sf
\using {\forall}L
\endprooftree
\justifies
[{\blacksquare}Nt(s(m))], {\blacksquare}(({\langle\rangle}{\exists}gNt(s(g))\backslash Sf)/({\exists}aNa{\oplus}({\exists}g(({\it CN}{\it g}/{\it CN}{\it g}){\sqcup}({\it CN}{\it g}\backslash {\it CN}{\it g})){-}I))), \mbox{\fbox{${\blacksquare}{\forall}g({\forall}f((Sf{{}{\uparrow}{}}{\blacksquare}Nt(s(g))){{}{\downarrow}{}}Sf)/{\it CN}{\it s(g)})$}}, {\square}{\it CN}{\it s(m)}\ \Rightarrow\ Sf
\using {\blacksquare}L
\endprooftree}
$$
\vspace{0.15in}
\noindent
The derivation delivers the semantics:
\disp{
$\exists C[(\mbox{\v{}}{\it man}\ {\it C})\wedge ({\it Pres}\ [{\it j}={\it C}])]$}
Leaving aside tense,
this is logically equivalent to $(\mbox{\v{}}{\it man}\ {\it j})$,
as required.
This correct interaction of the copula of identity with an indefinitely quantified complement
is a nice prediction of Montague grammar. 
It is preserved in type logical grammar,
and simplified by the lower type of the copula nominal complement.

The next example involves an intensional adsentential modifier:
\disp{
(dwp((7-83))) ${\bf necessarily}{+}[{\bf john}]{+}{\bf walks}: Sf$}
Lexical lookup yields the following semantically labelled sequent:
\disp{
${\blacksquare}(SA/{\square}SA): {\it Nec}, [{\blacksquare}Nt(s(m)): {\it j}], {\square}({\langle\rangle}{\exists}gNt(s(g))\backslash Sf): \mbox{\^{}}\lambda B({\it Pres}\ (\mbox{\v{}}{\it walk}\ {\it B}))\ \Rightarrow\ Sf$}
This has the derivation:
\vspace{0.15in}
$$
\prooftree
\prooftree
\prooftree
\prooftree
\prooftree
\prooftree
\prooftree
\prooftree
\prooftree
\justifies
\mbox{\fbox{$Nt(s(m))$}}\ \Rightarrow\ Nt(s(m))
\endprooftree
\justifies
\mbox{\fbox{${\blacksquare}Nt(s(m))$}}\ \Rightarrow\ Nt(s(m))
\using {\blacksquare}L
\endprooftree
\justifies
{\blacksquare}Nt(s(m))\ \Rightarrow\ \fbox{${\exists}gNt(s(g))$}
\using {\exists}R
\endprooftree
\justifies
[{\blacksquare}Nt(s(m))]\ \Rightarrow\ \fbox{${\langle\rangle}{\exists}gNt(s(g))$}
\using {\langle\rangle}R
\endprooftree
\prooftree
\justifies
\mbox{\fbox{$Sf$}}\ \Rightarrow\ Sf
\endprooftree
\justifies
[{\blacksquare}Nt(s(m))], \mbox{\fbox{${\langle\rangle}{\exists}gNt(s(g))\backslash Sf$}}\ \Rightarrow\ Sf
\using {\backslash}L
\endprooftree
\justifies
[{\blacksquare}Nt(s(m))], \mbox{\fbox{${\square}({\langle\rangle}{\exists}gNt(s(g))\backslash Sf)$}}\ \Rightarrow\ Sf
\using {\Box}L
\endprooftree
\justifies
[{\blacksquare}Nt(s(m))], {\square}({\langle\rangle}{\exists}gNt(s(g))\backslash Sf)\ \Rightarrow\ {\square}Sf
\using {\Box}R
\endprooftree
\prooftree
\justifies
\mbox{\fbox{$Sf$}}\ \Rightarrow\ Sf
\endprooftree
\justifies
\mbox{\fbox{$Sf/{\square}Sf$}}, [{\blacksquare}Nt(s(m))], {\square}({\langle\rangle}{\exists}gNt(s(g))\backslash Sf)\ \Rightarrow\ Sf
\using {/}L
\endprooftree
\justifies
\mbox{\fbox{${\blacksquare}(Sf/{\square}Sf)$}}, [{\blacksquare}Nt(s(m))], {\square}({\langle\rangle}{\exists}gNt(s(g))\backslash Sf)\ \Rightarrow\ Sf
\using {\blacksquare}L
\endprooftree
$$
\vspace{0.15in}
\noindent
The derivation delivers semantics:
\disp{
$({\it Nec}\ \mbox{\^{}}({\it Pres}\ (\mbox{\v{}}{\it walk}\ {\it j})))$}

The following example involves an adverb:
\disp{
(dwp((7-86))) $[{\bf john}]{+}{\bf walks}{+}{\bf slowly}: Sf$}
This is also assumed to create an intensional context.\footnote{The subject argument
of `slowly' is intensionalized in order for it to be modally closed to undergo
$\beta$-conversion in the semantics. But then this does not give the usual
verb phrase type for `walks slowly'; the ramifications of this remain to be explored.}
Lexical lookup yields:
\disp{
$[{\blacksquare}Nt(s(m)): {\it j}], {\square}({\langle\rangle}{\exists}gNt(s(g))\backslash Sf): \mbox{\^{}}\lambda A({\it Pres}\ (\mbox{\v{}}{\it walk}\ {\it A})), {\square}{\forall}a{\forall}f({\square}({\langle\rangle}Na\backslash Sf)\backslash ({\langle\rangle}{\square}Na\backslash Sf)): \mbox{\^{}}\lambda B\lambda C(\mbox{\v{}}{\it slowly}\ \mbox{\^{}}(\mbox{\v{}}{\it B}\ \mbox{\v{}}{\it C}))\ \Rightarrow\ Sf$}
This has the derivation:
\vspace{0.15in}
$$\small
\prooftree
\prooftree
\prooftree
\prooftree
\prooftree
\prooftree
\prooftree
\prooftree
\prooftree
\prooftree
\prooftree
\prooftree
\justifies
Nt(s(m))\ \Rightarrow\ Nt(s(m))
\endprooftree
\justifies
Nt(s(m))\ \Rightarrow\ \fbox{${\exists}gNt(s(g))$}
\using {\exists}R
\endprooftree
\justifies
[Nt(s(m))]\ \Rightarrow\ \fbox{${\langle\rangle}{\exists}gNt(s(g))$}
\using {\langle\rangle}R
\endprooftree
\prooftree
\justifies
\mbox{\fbox{$Sf$}}\ \Rightarrow\ Sf
\endprooftree
\justifies
[Nt(s(m))], \mbox{\fbox{${\langle\rangle}{\exists}gNt(s(g))\backslash Sf$}}\ \Rightarrow\ Sf
\using {\backslash}L
\endprooftree
\justifies
[Nt(s(m))], \mbox{\fbox{${\square}({\langle\rangle}{\exists}gNt(s(g))\backslash Sf)$}}\ \Rightarrow\ Sf
\using {\Box}L
\endprooftree
\justifies
{\langle\rangle}Nt(s(m)), {\square}({\langle\rangle}{\exists}gNt(s(g))\backslash Sf)\ \Rightarrow\ Sf
\using {\langle\rangle}L
\endprooftree
\justifies
{\square}({\langle\rangle}{\exists}gNt(s(g))\backslash Sf)\ \Rightarrow\ {\langle\rangle}Nt(s(m))\backslash Sf
\using {\backslash}R
\endprooftree
\justifies
{\square}({\langle\rangle}{\exists}gNt(s(g))\backslash Sf)\ \Rightarrow\ {\square}({\langle\rangle}Nt(s(m))\backslash Sf)
\using {\Box}R
\endprooftree
\prooftree
\prooftree
\prooftree
\prooftree
\prooftree
\justifies
\mbox{\fbox{$Nt(s(m))$}}\ \Rightarrow\ Nt(s(m))
\endprooftree
\justifies
\mbox{\fbox{${\blacksquare}Nt(s(m))$}}\ \Rightarrow\ Nt(s(m))
\using {\blacksquare}L
\endprooftree
\justifies
{\blacksquare}Nt(s(m))\ \Rightarrow\ {\square}Nt(s(m))
\using {\Box}R
\endprooftree
\justifies
[{\blacksquare}Nt(s(m))]\ \Rightarrow\ \fbox{${\langle\rangle}{\square}Nt(s(m))$}
\using {\langle\rangle}R
\endprooftree
\prooftree
\justifies
\mbox{\fbox{$Sf$}}\ \Rightarrow\ Sf
\endprooftree
\justifies
[{\blacksquare}Nt(s(m))], \mbox{\fbox{${\langle\rangle}{\square}Nt(s(m))\backslash Sf$}}\ \Rightarrow\ Sf
\using {\backslash}L
\endprooftree
\justifies
[{\blacksquare}Nt(s(m))], {\square}({\langle\rangle}{\exists}gNt(s(g))\backslash Sf), \mbox{\fbox{${\square}({\langle\rangle}Nt(s(m))\backslash Sf)\backslash ({\langle\rangle}{\square}Nt(s(m))\backslash Sf)$}}\ \Rightarrow\ Sf
\using {\backslash}L
\endprooftree
\justifies
[{\blacksquare}Nt(s(m))], {\square}({\langle\rangle}{\exists}gNt(s(g))\backslash Sf), \mbox{\fbox{${\forall}f({\square}({\langle\rangle}Nt(s(m))\backslash Sf)\backslash ({\langle\rangle}{\square}Nt(s(m))\backslash Sf))$}}\ \Rightarrow\ Sf
\using {\forall}L
\endprooftree
\justifies
[{\blacksquare}Nt(s(m))], {\square}({\langle\rangle}{\exists}gNt(s(g))\backslash Sf), \mbox{\fbox{${\forall}a{\forall}f({\square}({\langle\rangle}Na\backslash Sf)\backslash ({\langle\rangle}{\square}Na\backslash Sf))$}}\ \Rightarrow\ Sf
\using {\forall}L
\endprooftree
\justifies
[{\blacksquare}Nt(s(m))], {\square}({\langle\rangle}{\exists}gNt(s(g))\backslash Sf), \mbox{\fbox{${\square}{\forall}a{\forall}f({\square}({\langle\rangle}Na\backslash Sf)\backslash ({\langle\rangle}{\square}Na\backslash Sf))$}}\ \Rightarrow\ Sf
\using {\Box}L
\endprooftree
$$
\vspace{0.15in}
\noindent
It delivers semantics:
\disp{
$(\mbox{\v{}}{\it slowly}\ \mbox{\^{}}({\it Pres}\ (\mbox{\v{}}{\it walk}\ {\it j})))$}

The next example involves an equi control verb:
\disp{
(dwp((7-91))) $[{\bf john}]{+}{\bf tries}{+}{\bf to}{+}{\bf walk}: Sf$}
We lexically analyse the equi semantics a a relation of trying between the subject
and a proposition of which the subject is agent (something Montague did not do).
Lexical lookup yields:
\disp{
$[{\blacksquare}Nt(s(m)): {\it j}], {\square}(({\langle\rangle}{\exists}gNt(s(g))\backslash Sf)/{\square}({\langle\rangle}{\exists}gNt(s(g))\backslash Si)): \mbox{\^{}}\lambda A\lambda B((\mbox{\v{}}{\it tries}\ \mbox{\^{}}(\mbox{\v{}}{\it A}\ {\it B}))\ {\it B}),\\
 {\blacksquare}(({\it PP}{\it to}/{\exists}aNa){\sqcap}{\forall}n(({\langle\rangle}Nn\backslash Si)/({\langle\rangle}Nn\backslash Sb))): \lambda C{\it C}, {\square}({\langle\rangle}{\exists}aNa\backslash Sb): \mbox{\^{}}\lambda D(\mbox{\v{}}{\it walk}\ {\it D})\ \Rightarrow\ Sf$}
This has the derivation:
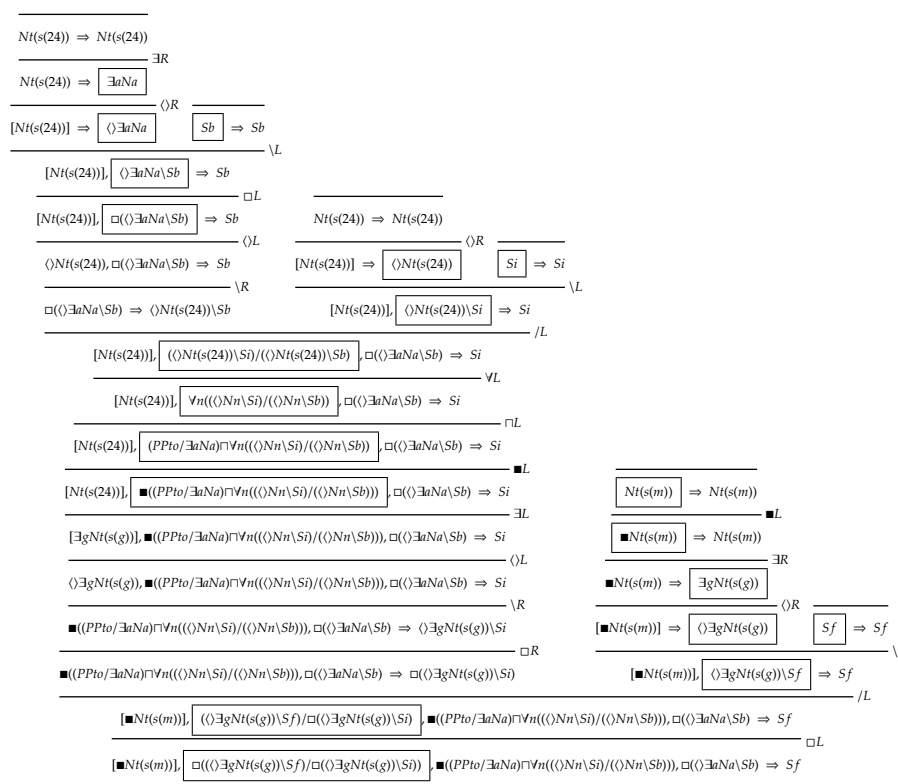
\vspace{0.15in}
$$
{\tiny
\prooftree
\prooftree
\prooftree
\prooftree
\prooftree
\prooftree
\prooftree
\prooftree
\prooftree
\prooftree
\prooftree
\prooftree
\prooftree
\prooftree
\prooftree
\prooftree
\prooftree
\justifies
Nt(s(24))\ \Rightarrow\ Nt(s(24))
\endprooftree
\justifies
Nt(s(24))\ \Rightarrow\ \fbox{${\exists}aNa$}
\using {\exists}R
\endprooftree
\justifies
[Nt(s(24))]\ \Rightarrow\ \fbox{${\langle\rangle}{\exists}aNa$}
\using {\langle\rangle}R
\endprooftree
\prooftree
\justifies
\mbox{\fbox{$Sb$}}\ \Rightarrow\ Sb
\endprooftree
\justifies
[Nt(s(24))], \mbox{\fbox{${\langle\rangle}{\exists}aNa\backslash Sb$}}\ \Rightarrow\ Sb
\using {\backslash}L
\endprooftree
\justifies
[Nt(s(24))], \mbox{\fbox{${\square}({\langle\rangle}{\exists}aNa\backslash Sb)$}}\ \Rightarrow\ Sb
\using {\Box}L
\endprooftree
\justifies
{\langle\rangle}Nt(s(24)), {\square}({\langle\rangle}{\exists}aNa\backslash Sb)\ \Rightarrow\ Sb
\using {\langle\rangle}L
\endprooftree
\justifies
{\square}({\langle\rangle}{\exists}aNa\backslash Sb)\ \Rightarrow\ {\langle\rangle}Nt(s(24))\backslash Sb
\using {\backslash}R
\endprooftree
\prooftree
\prooftree
\prooftree
\justifies
Nt(s(24))\ \Rightarrow\ Nt(s(24))
\endprooftree
\justifies
[Nt(s(24))]\ \Rightarrow\ \fbox{${\langle\rangle}Nt(s(24))$}
\using {\langle\rangle}R
\endprooftree
\prooftree
\justifies
\mbox{\fbox{$Si$}}\ \Rightarrow\ Si
\endprooftree
\justifies
[Nt(s(24))], \mbox{\fbox{${\langle\rangle}Nt(s(24))\backslash Si$}}\ \Rightarrow\ Si
\using {\backslash}L
\endprooftree
\justifies
[Nt(s(24))], \mbox{\fbox{$({\langle\rangle}Nt(s(24))\backslash Si)/({\langle\rangle}Nt(s(24))\backslash Sb)$}}, {\square}({\langle\rangle}{\exists}aNa\backslash Sb)\ \Rightarrow\ Si
\using {/}L
\endprooftree
\justifies
[Nt(s(24))], \mbox{\fbox{${\forall}n(({\langle\rangle}Nn\backslash Si)/({\langle\rangle}Nn\backslash Sb))$}}, {\square}({\langle\rangle}{\exists}aNa\backslash Sb)\ \Rightarrow\ Si
\using {\forall}L
\endprooftree
\justifies
[Nt(s(24))], \mbox{\fbox{$({\it PP}{\it to}/{\exists}aNa){\sqcap}{\forall}n(({\langle\rangle}Nn\backslash Si)/({\langle\rangle}Nn\backslash Sb))$}}, {\square}({\langle\rangle}{\exists}aNa\backslash Sb)\ \Rightarrow\ Si
\using {\sqcap}L
\endprooftree
\justifies
[Nt(s(24))], \mbox{\fbox{${\blacksquare}(({\it PP}{\it to}/{\exists}aNa){\sqcap}{\forall}n(({\langle\rangle}Nn\backslash Si)/({\langle\rangle}Nn\backslash Sb)))$}}, {\square}({\langle\rangle}{\exists}aNa\backslash Sb)\ \Rightarrow\ Si
\using {\blacksquare}L
\endprooftree
\justifies
[{\exists}gNt(s(g))], {\blacksquare}(({\it PP}{\it to}/{\exists}aNa){\sqcap}{\forall}n(({\langle\rangle}Nn\backslash Si)/({\langle\rangle}Nn\backslash Sb))), {\square}({\langle\rangle}{\exists}aNa\backslash Sb)\ \Rightarrow\ Si
\using {\exists}L
\endprooftree
\justifies
{\langle\rangle}{\exists}gNt(s(g)), {\blacksquare}(({\it PP}{\it to}/{\exists}aNa){\sqcap}{\forall}n(({\langle\rangle}Nn\backslash Si)/({\langle\rangle}Nn\backslash Sb))), {\square}({\langle\rangle}{\exists}aNa\backslash Sb)\ \Rightarrow\ Si
\using {\langle\rangle}L
\endprooftree
\justifies
{\blacksquare}(({\it PP}{\it to}/{\exists}aNa){\sqcap}{\forall}n(({\langle\rangle}Nn\backslash Si)/({\langle\rangle}Nn\backslash Sb))), {\square}({\langle\rangle}{\exists}aNa\backslash Sb)\ \Rightarrow\ {\langle\rangle}{\exists}gNt(s(g))\backslash Si
\using {\backslash}R
\endprooftree
\justifies
{\blacksquare}(({\it PP}{\it to}/{\exists}aNa){\sqcap}{\forall}n(({\langle\rangle}Nn\backslash Si)/({\langle\rangle}Nn\backslash Sb))), {\square}({\langle\rangle}{\exists}aNa\backslash Sb)\ \Rightarrow\ {\square}({\langle\rangle}{\exists}gNt(s(g))\backslash Si)
\using {\Box}R
\endprooftree
\prooftree
\prooftree
\prooftree
\prooftree
\prooftree
\justifies
\mbox{\fbox{$Nt(s(m))$}}\ \Rightarrow\ Nt(s(m))
\endprooftree
\justifies
\mbox{\fbox{${\blacksquare}Nt(s(m))$}}\ \Rightarrow\ Nt(s(m))
\using {\blacksquare}L
\endprooftree
\justifies
{\blacksquare}Nt(s(m))\ \Rightarrow\ \fbox{${\exists}gNt(s(g))$}
\using {\exists}R
\endprooftree
\justifies
[{\blacksquare}Nt(s(m))]\ \Rightarrow\ \fbox{${\langle\rangle}{\exists}gNt(s(g))$}
\using {\langle\rangle}R
\endprooftree
\prooftree
\justifies
\mbox{\fbox{$Sf$}}\ \Rightarrow\ Sf
\endprooftree
\justifies
[{\blacksquare}Nt(s(m))], \mbox{\fbox{${\langle\rangle}{\exists}gNt(s(g))\backslash Sf$}}\ \Rightarrow\ Sf
\using {\backslash}L
\endprooftree
\justifies
[{\blacksquare}Nt(s(m))], \mbox{\fbox{$({\langle\rangle}{\exists}gNt(s(g))\backslash Sf)/{\square}({\langle\rangle}{\exists}gNt(s(g))\backslash Si)$}}, {\blacksquare}(({\it PP}{\it to}/{\exists}aNa){\sqcap}{\forall}n(({\langle\rangle}Nn\backslash Si)/({\langle\rangle}Nn\backslash Sb))), {\square}({\langle\rangle}{\exists}aNa\backslash Sb)\ \Rightarrow\ Sf
\using {/}L
\endprooftree
\justifies
[{\blacksquare}Nt(s(m))], \mbox{\fbox{${\square}(({\langle\rangle}{\exists}gNt(s(g))\backslash Sf)/{\square}({\langle\rangle}{\exists}gNt(s(g))\backslash Si))$}}, {\blacksquare}(({\it PP}{\it to}/{\exists}aNa){\sqcap}{\forall}n(({\langle\rangle}Nn\backslash Si)/({\langle\rangle}Nn\backslash Sb))), {\square}({\langle\rangle}{\exists}aNa\backslash Sb)\ \Rightarrow\ Sf
\using {\Box}L
\endprooftree}
$$
\vspace{0.15in}
\noindent
It delivers the semantics:
\disp{
$((\mbox{\v{}}{\it tries}\ \mbox{\^{}}(\mbox{\v{}}{\it walk}\ {\it j}))\ {\it j})$}
I.e.\ John tries to bring it about that he (John) walks.

The next example involves control, 
quantification,
coordination and anaphora:
\disp{
(dwp((7-94))) $[{\bf john}]{+}{\bf tries}{+}{\bf to}{+}[[{\bf catch}{+}{\bf a}{+}{\bf fish}{+}{\bf and}{+}{\bf eat}{+}{\bf it}]]: Sf$}
The sentence is ambiguous as to whether {\it a fish\/} is wide scope
(with existential commitment) or narrow scope (without existential commitment)
with respect to {\it tries\/},
but in both cases it must be the antecedent of {\it it}.
Lexical lookup inserting the verb phrase coordinator and the 
pronoun assignment yields the semantically labelled sequent:
\disp{
$[{\blacksquare}Nt(s(m)): {\it j}], {\square}(({\langle\rangle}{\exists}gNt(s(g))\backslash Sf)/{\square}({\langle\rangle}{\exists}gNt(s(g))\backslash Si)): \mbox{\^{}}\lambda A\lambda B((\mbox{\v{}}{\it tries}\ \mbox{\^{}}(\mbox{\v{}}{\it A}\ {\it B}))\ {\it B}),\\
 {\blacksquare}(({\it PP}{\it to}/{\exists}aNa){\sqcap}{\forall}n(({\langle\rangle}Nn\backslash Si)/({\langle\rangle}Nn\backslash Sb))): \lambda C{\it C}, [[{\square}(({\langle\rangle}{\exists}aNa\backslash Sb)/{\exists}aNa):\\\mbox{\^{}}\lambda D\lambda E((\mbox{\v{}}{\it catch}\ {\it D})\ {\it E}),
  {\blacksquare}{\forall}g({\forall}f((Sf{{}{\uparrow}{}}{\blacksquare}Nt(s(g))){{}{\downarrow}{}}Sf)/{\it CN}{\it s(g)}): \lambda F\lambda G\exists H[({\it F}\ {\it H})\wedge ({\it G}\ {\it H})],\\{\square}{\it CN}{\it s(n)}: {\it fish},
   {\blacksquare}{\forall}a{\forall}f((?{\blacksquare}({\langle\rangle}Na\backslash Sf)\backslash {[]^{-1}}{[]^{-1}}({\langle\rangle}Na\backslash Sf))/{\blacksquare}({\langle\rangle}Na\backslash Sf)): (\Phinplus{}\ ({\it s}\ {\it 0})\ {\it and}),\\{\square}(({\langle\rangle}{\exists}aNa\backslash Sb)/{\exists}aNa): \mbox{\^{}}\lambda I\lambda J((\mbox{\v{}}{\it eat}\ {\it I})\ {\it J}), {\blacksquare}{\forall}f{\forall}a((({\langle\rangle}Na\backslash Sf){{}{\uparrow}{}}{\blacksquare}Nt(s(n))){{}{\downarrow}{}}\\({\blacksquare}({\langle\rangle}Na\backslash Sf){|}{\blacksquare}Nt(s(n)))): \lambda K{\it K}]]\ \Rightarrow\ Sf$}
The wide scope existential derivation is:
\vspace{0.15in}
$$
\tiny
\prooftree
\prooftree
\prooftree
\prooftree
\prooftree
\prooftree
\prooftree
\prooftree
\justifies
\mbox{\fbox{$Nt(s(n))$}}\ \Rightarrow\ Nt(s(n))
\endprooftree
\justifies
\mbox{\fbox{${\blacksquare}Nt(s(n))$}}\ \Rightarrow\ Nt(s(n))
\using {\blacksquare}L
\endprooftree
\justifies
{\blacksquare}Nt(s(n))\ \Rightarrow\ \fbox{${\exists}aNa$}
\using {\exists}R
\endprooftree
\prooftree
\prooftree
\prooftree
\prooftree
\justifies
Nt(s(50))\ \Rightarrow\ Nt(s(50))
\endprooftree
\justifies
Nt(s(50))\ \Rightarrow\ \fbox{${\exists}aNa$}
\using {\exists}R
\endprooftree
\justifies
[Nt(s(50))]\ \Rightarrow\ \fbox{${\langle\rangle}{\exists}aNa$}
\using {\langle\rangle}R
\endprooftree
\prooftree
\justifies
\mbox{\fbox{$Sb$}}\ \Rightarrow\ Sb
\endprooftree
\justifies
[Nt(s(50))], \mbox{\fbox{${\langle\rangle}{\exists}aNa\backslash Sb$}}\ \Rightarrow\ Sb
\using {\backslash}L
\endprooftree
\justifies
[Nt(s(50))], \mbox{\fbox{$({\langle\rangle}{\exists}aNa\backslash Sb)/{\exists}aNa$}}, {\blacksquare}Nt(s(n))\ \Rightarrow\ Sb
\using {/}L
\endprooftree
\justifies
[Nt(s(50))], \mbox{\fbox{${\square}(({\langle\rangle}{\exists}aNa\backslash Sb)/{\exists}aNa)$}}, {\blacksquare}Nt(s(n))\ \Rightarrow\ Sb
\using {\Box}L
\endprooftree
\justifies
{\langle\rangle}Nt(s(50)), {\square}(({\langle\rangle}{\exists}aNa\backslash Sb)/{\exists}aNa), {\blacksquare}Nt(s(n))\ \Rightarrow\ Sb
\using {\langle\rangle}L
\endprooftree
\justifies
{\square}(({\langle\rangle}{\exists}aNa\backslash Sb)/{\exists}aNa), {\blacksquare}Nt(s(n))\ \Rightarrow\ {\langle\rangle}Nt(s(50))\backslash Sb
\using {\backslash}R
\endprooftree
\justifies
\begin{array}{c}
{\square}(({\langle\rangle}{\exists}aNa\backslash Sb)/{\exists}aNa), {\tt 1}\ \Rightarrow\ ({\langle\rangle}Nt(s(50))\backslash Sb){{}{\uparrow}{}}{\blacksquare}Nt(s(n))\\
\mbox{\footnotesize\textcircled{1}}
\end{array}
\using {\uparrow}R
\endprooftree
$$
$$
\resizebox{\textwidth}{!}{
\prooftree
\prooftree
\prooftree
\prooftree
\justifies
\mbox{\fbox{$Nt(s(n))$}}\ \Rightarrow\ Nt(s(n))
\endprooftree
\justifies
\mbox{\fbox{${\blacksquare}Nt(s(n))$}}\ \Rightarrow\ Nt(s(n))
\using {\blacksquare}L
\endprooftree
\justifies
{\blacksquare}Nt(s(n))\ \Rightarrow\ {\blacksquare}Nt(s(n))
\using {\blacksquare}R
\endprooftree
\prooftree
\prooftree
\prooftree
\prooftree
\prooftree
\prooftree
\prooftree
\prooftree
\prooftree
\prooftree
\prooftree
\justifies
Nt(s(50))\ \Rightarrow\ Nt(s(50))
\endprooftree
\justifies
[Nt(s(50))]\ \Rightarrow\ \fbox{${\langle\rangle}Nt(s(50))$}
\using {\langle\rangle}R
\endprooftree
\prooftree
\justifies
\mbox{\fbox{$Sb$}}\ \Rightarrow\ Sb
\endprooftree
\justifies
[Nt(s(50))], \mbox{\fbox{${\langle\rangle}Nt(s(50))\backslash Sb$}}\ \Rightarrow\ Sb
\using {\backslash}L
\endprooftree
\justifies
[Nt(s(50))], \mbox{\fbox{${\blacksquare}({\langle\rangle}Nt(s(50))\backslash Sb)$}}\ \Rightarrow\ Sb
\using {\blacksquare}L
\endprooftree
\justifies
{\langle\rangle}Nt(s(50)), {\blacksquare}({\langle\rangle}Nt(s(50))\backslash Sb)\ \Rightarrow\ Sb
\using {\langle\rangle}L
\endprooftree
\justifies
{\blacksquare}({\langle\rangle}Nt(s(50))\backslash Sb)\ \Rightarrow\ {\langle\rangle}Nt(s(50))\backslash Sb
\using {\backslash}R
\endprooftree
\justifies
{\blacksquare}({\langle\rangle}Nt(s(50))\backslash Sb)\ \Rightarrow\ {\blacksquare}({\langle\rangle}Nt(s(50))\backslash Sb)
\using {\blacksquare}R
\endprooftree
\prooftree
\prooftree
\prooftree
\prooftree
\prooftree
\prooftree
\prooftree
\prooftree
\prooftree
\prooftree
\justifies
\mbox{\fbox{$Nt(s(n))$}}\ \Rightarrow\ Nt(s(n))
\endprooftree
\justifies
\mbox{\fbox{${\blacksquare}Nt(s(n))$}}\ \Rightarrow\ Nt(s(n))
\using {\blacksquare}L
\endprooftree
\justifies
{\blacksquare}Nt(s(n))\ \Rightarrow\ \fbox{${\exists}aNa$}
\using {\exists}R
\endprooftree
\prooftree
\prooftree
\prooftree
\prooftree
\justifies
Nt(s(50))\ \Rightarrow\ Nt(s(50))
\endprooftree
\justifies
Nt(s(50))\ \Rightarrow\ \fbox{${\exists}aNa$}
\using {\exists}R
\endprooftree
\justifies
[Nt(s(50))]\ \Rightarrow\ \fbox{${\langle\rangle}{\exists}aNa$}
\using {\langle\rangle}R
\endprooftree
\prooftree
\justifies
\mbox{\fbox{$Sb$}}\ \Rightarrow\ Sb
\endprooftree
\justifies
[Nt(s(50))], \mbox{\fbox{${\langle\rangle}{\exists}aNa\backslash Sb$}}\ \Rightarrow\ Sb
\using {\backslash}L
\endprooftree
\justifies
[Nt(s(50))], \mbox{\fbox{$({\langle\rangle}{\exists}aNa\backslash Sb)/{\exists}aNa$}}, {\blacksquare}Nt(s(n))\ \Rightarrow\ Sb
\using {/}L
\endprooftree
\justifies
[Nt(s(50))], \mbox{\fbox{${\square}(({\langle\rangle}{\exists}aNa\backslash Sb)/{\exists}aNa)$}}, {\blacksquare}Nt(s(n))\ \Rightarrow\ Sb
\using {\Box}L
\endprooftree
\justifies
{\langle\rangle}Nt(s(50)), {\square}(({\langle\rangle}{\exists}aNa\backslash Sb)/{\exists}aNa), {\blacksquare}Nt(s(n))\ \Rightarrow\ Sb
\using {\langle\rangle}L
\endprooftree
\justifies
{\square}(({\langle\rangle}{\exists}aNa\backslash Sb)/{\exists}aNa), {\blacksquare}Nt(s(n))\ \Rightarrow\ {\langle\rangle}Nt(s(50))\backslash Sb
\using {\backslash}R
\endprooftree
\justifies
{\square}(({\langle\rangle}{\exists}aNa\backslash Sb)/{\exists}aNa), {\blacksquare}Nt(s(n))\ \Rightarrow\ {\blacksquare}({\langle\rangle}Nt(s(50))\backslash Sb)
\using {\blacksquare}R
\endprooftree
\justifies
{\square}(({\langle\rangle}{\exists}aNa\backslash Sb)/{\exists}aNa), {\blacksquare}Nt(s(n))\ \Rightarrow\ \fbox{$?{\blacksquare}({\langle\rangle}Nt(s(50))\backslash Sb)$}
\using {?}R
\endprooftree
\prooftree
\prooftree
\prooftree
\prooftree
\prooftree
\justifies
Nt(s(50))\ \Rightarrow\ Nt(s(50))
\endprooftree
\justifies
[Nt(s(50))]\ \Rightarrow\ \fbox{${\langle\rangle}Nt(s(50))$}
\using {\langle\rangle}R
\endprooftree
\prooftree
\justifies
\mbox{\fbox{$Sb$}}\ \Rightarrow\ Sb
\endprooftree
\justifies
[Nt(s(50))], \mbox{\fbox{${\langle\rangle}Nt(s(50))\backslash Sb$}}\ \Rightarrow\ Sb
\using {\backslash}L
\endprooftree
\justifies
[Nt(s(50))], [\mbox{\fbox{${[]^{-1}}({\langle\rangle}Nt(s(50))\backslash Sb)$}}]\ \Rightarrow\ Sb
\using {[]^{-1}}L
\endprooftree
\justifies
[Nt(s(50))], [[\mbox{\fbox{${[]^{-1}}{[]^{-1}}({\langle\rangle}Nt(s(50))\backslash Sb)$}}]]\ \Rightarrow\ Sb
\using {[]^{-1}}L
\endprooftree
\justifies
[Nt(s(50))], [[{\square}(({\langle\rangle}{\exists}aNa\backslash Sb)/{\exists}aNa), {\blacksquare}Nt(s(n)), \mbox{\fbox{$?{\blacksquare}({\langle\rangle}Nt(s(50))\backslash Sb)\backslash {[]^{-1}}{[]^{-1}}({\langle\rangle}Nt(s(50))\backslash Sb)$}}]]\ \Rightarrow\ Sb
\using {\backslash}L
\endprooftree
\justifies
[Nt(s(50))], [[{\square}(({\langle\rangle}{\exists}aNa\backslash Sb)/{\exists}aNa), {\blacksquare}Nt(s(n)), \mbox{\fbox{$(?{\blacksquare}({\langle\rangle}Nt(s(50))\backslash Sb)\backslash {[]^{-1}}{[]^{-1}}({\langle\rangle}Nt(s(50))\backslash Sb))/{\blacksquare}({\langle\rangle}Nt(s(50))\backslash Sb)$}}, {\blacksquare}({\langle\rangle}Nt(s(50))\backslash Sb)]]\ \Rightarrow\ Sb
\using {/}L
\endprooftree
\justifies
[Nt(s(50))], [[{\square}(({\langle\rangle}{\exists}aNa\backslash Sb)/{\exists}aNa), {\blacksquare}Nt(s(n)), \mbox{\fbox{${\forall}f((?{\blacksquare}({\langle\rangle}Nt(s(50))\backslash Sf)\backslash {[]^{-1}}{[]^{-1}}({\langle\rangle}Nt(s(50))\backslash Sf))/{\blacksquare}({\langle\rangle}Nt(s(50))\backslash Sf))$}}, {\blacksquare}({\langle\rangle}Nt(s(50))\backslash Sb)]]\ \Rightarrow\ Sb
\using {\forall}L
\endprooftree
\justifies
[Nt(s(50))], [[{\square}(({\langle\rangle}{\exists}aNa\backslash Sb)/{\exists}aNa), {\blacksquare}Nt(s(n)), \mbox{\fbox{${\forall}a{\forall}f((?{\blacksquare}({\langle\rangle}Na\backslash Sf)\backslash {[]^{-1}}{[]^{-1}}({\langle\rangle}Na\backslash Sf))/{\blacksquare}({\langle\rangle}Na\backslash Sf))$}}, {\blacksquare}({\langle\rangle}Nt(s(50))\backslash Sb)]]\ \Rightarrow\ Sb
\using {\forall}L
\endprooftree
\justifies
[Nt(s(50))], [[{\square}(({\langle\rangle}{\exists}aNa\backslash Sb)/{\exists}aNa), {\blacksquare}Nt(s(n)), \mbox{\fbox{${\blacksquare}{\forall}a{\forall}f((?{\blacksquare}({\langle\rangle}Na\backslash Sf)\backslash {[]^{-1}}{[]^{-1}}({\langle\rangle}Na\backslash Sf))/{\blacksquare}({\langle\rangle}Na\backslash Sf))$}}, {\blacksquare}({\langle\rangle}Nt(s(50))\backslash Sb)]]\ \Rightarrow\ Sb
\using {\blacksquare}L
\endprooftree
\justifies
\begin{array}{c}
[Nt(s(50))], [[{\square}(({\langle\rangle}{\exists}aNa\backslash Sb)/{\exists}aNa), {\blacksquare}Nt(s(n)), {\blacksquare}{\forall}a{\forall}f((?{\blacksquare}({\langle\rangle}Na\backslash Sf)\backslash {[]^{-1}}{[]^{-1}}({\langle\rangle}Na\backslash Sf))/{\blacksquare}({\langle\rangle}Na\backslash Sf)), \mbox{\fbox{${\blacksquare}({\langle\rangle}Nt(s(50))\backslash Sb){|}{\blacksquare}Nt(s(n))$}}]]\ \Rightarrow\ Sb\\
\mbox{\footnotesize\textcircled{2}}
\end{array}
\using {|}L
\endprooftree}
$$
$$\tiny
\prooftree
\prooftree
\prooftree
\prooftree
\prooftree
\prooftree
\mbox{\footnotesize\textcircled{1}}\tab\tab\tab\tab\tab\tab\tab\tab
\mbox{\footnotesize\textcircled{2}}
\justifies
[Nt(s(50))], [[{\square}(({\langle\rangle}{\exists}aNa\backslash Sb)/{\exists}aNa), {\blacksquare}Nt(s(n)), {\blacksquare}{\forall}a{\forall}f((?{\blacksquare}({\langle\rangle}Na\backslash Sf)\backslash {[]^{-1}}{[]^{-1}}({\langle\rangle}Na\backslash Sf))/{\blacksquare}({\langle\rangle}Na\backslash Sf)), {\square}(({\langle\rangle}{\exists}aNa\backslash Sb)/{\exists}aNa), \mbox{\fbox{$(({\langle\rangle}Nt(s(50))\backslash Sb){{}{\uparrow}{}}{\blacksquare}Nt(s(n))){{}{\downarrow}{}}({\blacksquare}({\langle\rangle}Nt(s(50))\backslash Sb){|}{\blacksquare}Nt(s(n)))$}}]]\ \Rightarrow\ Sb
\using {\downarrow}L
\endprooftree
\justifies
[Nt(s(50))], [[{\square}(({\langle\rangle}{\exists}aNa\backslash Sb)/{\exists}aNa), {\blacksquare}Nt(s(n)), {\blacksquare}{\forall}a{\forall}f((?{\blacksquare}({\langle\rangle}Na\backslash Sf)\backslash {[]^{-1}}{[]^{-1}}({\langle\rangle}Na\backslash Sf))/{\blacksquare}({\langle\rangle}Na\backslash Sf)), {\square}(({\langle\rangle}{\exists}aNa\backslash Sb)/{\exists}aNa), \mbox{\fbox{${\forall}a((({\langle\rangle}Na\backslash Sb){{}{\uparrow}{}}{\blacksquare}Nt(s(n))){{}{\downarrow}{}}({\blacksquare}({\langle\rangle}Na\backslash Sb){|}{\blacksquare}Nt(s(n))))$}}]]\ \Rightarrow\ Sb
\using {\forall}L
\endprooftree
\justifies
[Nt(s(50))], [[{\square}(({\langle\rangle}{\exists}aNa\backslash Sb)/{\exists}aNa), {\blacksquare}Nt(s(n)), {\blacksquare}{\forall}a{\forall}f((?{\blacksquare}({\langle\rangle}Na\backslash Sf)\backslash {[]^{-1}}{[]^{-1}}({\langle\rangle}Na\backslash Sf))/{\blacksquare}({\langle\rangle}Na\backslash Sf)), {\square}(({\langle\rangle}{\exists}aNa\backslash Sb)/{\exists}aNa), \mbox{\fbox{${\forall}f{\forall}a((({\langle\rangle}Na\backslash Sf){{}{\uparrow}{}}{\blacksquare}Nt(s(n))){{}{\downarrow}{}}({\blacksquare}({\langle\rangle}Na\backslash Sf){|}{\blacksquare}Nt(s(n))))$}}]]\ \Rightarrow\ Sb
\using {\forall}L
\endprooftree
\justifies
[Nt(s(50))], [[{\square}(({\langle\rangle}{\exists}aNa\backslash Sb)/{\exists}aNa), {\blacksquare}Nt(s(n)), {\blacksquare}{\forall}a{\forall}f((?{\blacksquare}({\langle\rangle}Na\backslash Sf)\backslash {[]^{-1}}{[]^{-1}}({\langle\rangle}Na\backslash Sf))/{\blacksquare}({\langle\rangle}Na\backslash Sf)), {\square}(({\langle\rangle}{\exists}aNa\backslash Sb)/{\exists}aNa), \mbox{\fbox{${\blacksquare}{\forall}f{\forall}a((({\langle\rangle}Na\backslash Sf){{}{\uparrow}{}}{\blacksquare}Nt(s(n))){{}{\downarrow}{}}({\blacksquare}({\langle\rangle}Na\backslash Sf){|}{\blacksquare}Nt(s(n))))$}}]]\ \Rightarrow\ Sb
\using {\blacksquare}L
\endprooftree
\justifies
{\langle\rangle}Nt(s(50)), [[{\square}(({\langle\rangle}{\exists}aNa\backslash Sb)/{\exists}aNa), {\blacksquare}Nt(s(n)), {\blacksquare}{\forall}a{\forall}f((?{\blacksquare}({\langle\rangle}Na\backslash Sf)\backslash {[]^{-1}}{[]^{-1}}({\langle\rangle}Na\backslash Sf))/{\blacksquare}({\langle\rangle}Na\backslash Sf)), {\square}(({\langle\rangle}{\exists}aNa\backslash Sb)/{\exists}aNa), {\blacksquare}{\forall}f{\forall}a((({\langle\rangle}Na\backslash Sf){{}{\uparrow}{}}{\blacksquare}Nt(s(n))){{}{\downarrow}{}}({\blacksquare}({\langle\rangle}Na\backslash Sf){|}{\blacksquare}Nt(s(n))))]]\ \Rightarrow\ Sb
\using {\langle\rangle}L
\endprooftree
\justifies
\begin{array}{c}
{}[[{\square}(({\langle\rangle}{\exists}aNa\backslash Sb)/{\exists}aNa), {\blacksquare}Nt(s(n)), {\blacksquare}{\forall}a{\forall}f((?{\blacksquare}({\langle\rangle}Na\backslash Sf)\backslash {[]^{-1}}{[]^{-1}}({\langle\rangle}Na\backslash Sf))/{\blacksquare}({\langle\rangle}Na\backslash Sf)), {\square}(({\langle\rangle}{\exists}aNa\backslash Sb)/{\exists}aNa), {\blacksquare}{\forall}f{\forall}a((({\langle\rangle}Na\backslash Sf){{}{\uparrow}{}}{\blacksquare}Nt(s(n))){{}{\downarrow}{}}({\blacksquare}({\langle\rangle}Na\backslash Sf){|}{\blacksquare}Nt(s(n))))]]\ \Rightarrow\ {\langle\rangle}Nt(s(50))\backslash Sb\\
\mbox{\footnotesize\textcircled{3}}
\end{array}
\using {\backslash}R
\endprooftree
$$
$$\tiny
\rotatebox{-90}{
\prooftree
\prooftree
\prooftree
\prooftree
\prooftree
\prooftree
\prooftree
\prooftree
\mbox{\footnotesize\textcircled{3}\tab\tab\tab\tab\tab\tab\tab\tab}
\prooftree
\prooftree
\prooftree
\justifies
Nt(s(50))\ \Rightarrow\ Nt(s(50))
\endprooftree
\justifies
[Nt(s(50))]\ \Rightarrow\ \fbox{${\langle\rangle}Nt(s(50))$}
\using {\langle\rangle}R
\endprooftree
\prooftree
\justifies
\mbox{\fbox{$Si$}}\ \Rightarrow\ Si
\endprooftree
\justifies
[Nt(s(50))], \mbox{\fbox{${\langle\rangle}Nt(s(50))\backslash Si$}}\ \Rightarrow\ Si
\using {\backslash}L
\endprooftree
\justifies
[Nt(s(50))], \mbox{\fbox{$({\langle\rangle}Nt(s(50))\backslash Si)/({\langle\rangle}Nt(s(50))\backslash Sb)$}}, [[{\square}(({\langle\rangle}{\exists}aNa\backslash Sb)/{\exists}aNa), {\blacksquare}Nt(s(n)), {\blacksquare}{\forall}a{\forall}f((?{\blacksquare}({\langle\rangle}Na\backslash Sf)\backslash {[]^{-1}}{[]^{-1}}({\langle\rangle}Na\backslash Sf))/{\blacksquare}({\langle\rangle}Na\backslash Sf)), {\square}(({\langle\rangle}{\exists}aNa\backslash Sb)/{\exists}aNa), {\blacksquare}{\forall}f{\forall}a((({\langle\rangle}Na\backslash Sf){{}{\uparrow}{}}{\blacksquare}Nt(s(n))){{}{\downarrow}{}}({\blacksquare}({\langle\rangle}Na\backslash Sf){|}{\blacksquare}Nt(s(n))))]]\ \Rightarrow\ Si
\using {/}L
\endprooftree
\justifies
[Nt(s(50))], \mbox{\fbox{${\forall}n(({\langle\rangle}Nn\backslash Si)/({\langle\rangle}Nn\backslash Sb))$}}, [[{\square}(({\langle\rangle}{\exists}aNa\backslash Sb)/{\exists}aNa), {\blacksquare}Nt(s(n)), {\blacksquare}{\forall}a{\forall}f((?{\blacksquare}({\langle\rangle}Na\backslash Sf)\backslash {[]^{-1}}{[]^{-1}}({\langle\rangle}Na\backslash Sf))/{\blacksquare}({\langle\rangle}Na\backslash Sf)), {\square}(({\langle\rangle}{\exists}aNa\backslash Sb)/{\exists}aNa), {\blacksquare}{\forall}f{\forall}a((({\langle\rangle}Na\backslash Sf){{}{\uparrow}{}}{\blacksquare}Nt(s(n))){{}{\downarrow}{}}({\blacksquare}({\langle\rangle}Na\backslash Sf){|}{\blacksquare}Nt(s(n))))]]\ \Rightarrow\ Si
\using {\forall}L
\endprooftree
\justifies
[Nt(s(50))], \mbox{\fbox{$({\it PP}{\it to}/{\exists}aNa){\sqcap}{\forall}n(({\langle\rangle}Nn\backslash Si)/({\langle\rangle}Nn\backslash Sb))$}}, [[{\square}(({\langle\rangle}{\exists}aNa\backslash Sb)/{\exists}aNa), {\blacksquare}Nt(s(n)), {\blacksquare}{\forall}a{\forall}f((?{\blacksquare}({\langle\rangle}Na\backslash Sf)\backslash {[]^{-1}}{[]^{-1}}({\langle\rangle}Na\backslash Sf))/{\blacksquare}({\langle\rangle}Na\backslash Sf)), {\square}(({\langle\rangle}{\exists}aNa\backslash Sb)/{\exists}aNa), {\blacksquare}{\forall}f{\forall}a((({\langle\rangle}Na\backslash Sf){{}{\uparrow}{}}{\blacksquare}Nt(s(n))){{}{\downarrow}{}}({\blacksquare}({\langle\rangle}Na\backslash Sf){|}{\blacksquare}Nt(s(n))))]]\ \Rightarrow\ Si
\using {\sqcap}L
\endprooftree
\justifies
[Nt(s(50))], \mbox{\fbox{${\blacksquare}(({\it PP}{\it to}/{\exists}aNa){\sqcap}{\forall}n(({\langle\rangle}Nn\backslash Si)/({\langle\rangle}Nn\backslash Sb)))$}}, [[{\square}(({\langle\rangle}{\exists}aNa\backslash Sb)/{\exists}aNa), {\blacksquare}Nt(s(n)), {\blacksquare}{\forall}a{\forall}f((?{\blacksquare}({\langle\rangle}Na\backslash Sf)\backslash {[]^{-1}}{[]^{-1}}({\langle\rangle}Na\backslash Sf))/{\blacksquare}({\langle\rangle}Na\backslash Sf)), {\square}(({\langle\rangle}{\exists}aNa\backslash Sb)/{\exists}aNa), {\blacksquare}{\forall}f{\forall}a((({\langle\rangle}Na\backslash Sf){{}{\uparrow}{}}{\blacksquare}Nt(s(n))){{}{\downarrow}{}}({\blacksquare}({\langle\rangle}Na\backslash Sf){|}{\blacksquare}Nt(s(n))))]]\ \Rightarrow\ Si
\using {\blacksquare}L
\endprooftree
\justifies
[{\exists}gNt(s(g))], {\blacksquare}(({\it PP}{\it to}/{\exists}aNa){\sqcap}{\forall}n(({\langle\rangle}Nn\backslash Si)/({\langle\rangle}Nn\backslash Sb))), [[{\square}(({\langle\rangle}{\exists}aNa\backslash Sb)/{\exists}aNa), {\blacksquare}Nt(s(n)), {\blacksquare}{\forall}a{\forall}f((?{\blacksquare}({\langle\rangle}Na\backslash Sf)\backslash {[]^{-1}}{[]^{-1}}({\langle\rangle}Na\backslash Sf))/{\blacksquare}({\langle\rangle}Na\backslash Sf)), {\square}(({\langle\rangle}{\exists}aNa\backslash Sb)/{\exists}aNa), {\blacksquare}{\forall}f{\forall}a((({\langle\rangle}Na\backslash Sf){{}{\uparrow}{}}{\blacksquare}Nt(s(n))){{}{\downarrow}{}}({\blacksquare}({\langle\rangle}Na\backslash Sf){|}{\blacksquare}Nt(s(n))))]]\ \Rightarrow\ Si
\using {\exists}L
\endprooftree
\justifies
{\langle\rangle}{\exists}gNt(s(g)), {\blacksquare}(({\it PP}{\it to}/{\exists}aNa){\sqcap}{\forall}n(({\langle\rangle}Nn\backslash Si)/({\langle\rangle}Nn\backslash Sb))), [[{\square}(({\langle\rangle}{\exists}aNa\backslash Sb)/{\exists}aNa), {\blacksquare}Nt(s(n)), {\blacksquare}{\forall}a{\forall}f((?{\blacksquare}({\langle\rangle}Na\backslash Sf)\backslash {[]^{-1}}{[]^{-1}}({\langle\rangle}Na\backslash Sf))/{\blacksquare}({\langle\rangle}Na\backslash Sf)), {\square}(({\langle\rangle}{\exists}aNa\backslash Sb)/{\exists}aNa), {\blacksquare}{\forall}f{\forall}a((({\langle\rangle}Na\backslash Sf){{}{\uparrow}{}}{\blacksquare}Nt(s(n))){{}{\downarrow}{}}({\blacksquare}({\langle\rangle}Na\backslash Sf){|}{\blacksquare}Nt(s(n))))]]\ \Rightarrow\ Si
\using {\langle\rangle}L
\endprooftree
\justifies
{\blacksquare}(({\it PP}{\it to}/{\exists}aNa){\sqcap}{\forall}n(({\langle\rangle}Nn\backslash Si)/({\langle\rangle}Nn\backslash Sb))), [[{\square}(({\langle\rangle}{\exists}aNa\backslash Sb)/{\exists}aNa), {\blacksquare}Nt(s(n)), {\blacksquare}{\forall}a{\forall}f((?{\blacksquare}({\langle\rangle}Na\backslash Sf)\backslash {[]^{-1}}{[]^{-1}}({\langle\rangle}Na\backslash Sf))/{\blacksquare}({\langle\rangle}Na\backslash Sf)), {\square}(({\langle\rangle}{\exists}aNa\backslash Sb)/{\exists}aNa), {\blacksquare}{\forall}f{\forall}a((({\langle\rangle}Na\backslash Sf){{}{\uparrow}{}}{\blacksquare}Nt(s(n))){{}{\downarrow}{}}({\blacksquare}({\langle\rangle}Na\backslash Sf){|}{\blacksquare}Nt(s(n))))]]\ \Rightarrow\ {\langle\rangle}{\exists}gNt(s(g))\backslash Si
\using {\backslash}R
\endprooftree
\justifies
\begin{array}{c}
{\blacksquare}(({\it PP}{\it to}/{\exists}aNa){\sqcap}{\forall}n(({\langle\rangle}Nn\backslash Si)/({\langle\rangle}Nn\backslash Sb))), [[{\square}(({\langle\rangle}{\exists}aNa\backslash Sb)/{\exists}aNa), {\blacksquare}Nt(s(n)), {\blacksquare}{\forall}a{\forall}f((?{\blacksquare}({\langle\rangle}Na\backslash Sf)\backslash {[]^{-1}}{[]^{-1}}({\langle\rangle}Na\backslash Sf))/{\blacksquare}({\langle\rangle}Na\backslash Sf)), {\square}(({\langle\rangle}{\exists}aNa\backslash Sb)/{\exists}aNa), {\blacksquare}{\forall}f{\forall}a((({\langle\rangle}Na\backslash Sf){{}{\uparrow}{}}{\blacksquare}Nt(s(n))){{}{\downarrow}{}}({\blacksquare}({\langle\rangle}Na\backslash Sf){|}{\blacksquare}Nt(s(n))))]]\ \Rightarrow\ {\square}({\langle\rangle}{\exists}gNt(s(g))\backslash Si)\\
{\footnotesize\textcircled{4}}
\end{array}
\using {\Box}R
\endprooftree}
$$
$$\tiny
\rotatebox{-90}{
\prooftree
\prooftree
\prooftree
\prooftree
\prooftree
\justifies
\mbox{\fbox{${\it CN}{\it s(n)}$}}\ \Rightarrow\ {\it CN}{\it s(n)}
\endprooftree
\justifies
\mbox{\fbox{${\square}{\it CN}{\it s(n)}$}}\ \Rightarrow\ {\it CN}{\it s(n)}
\using {\Box}L
\endprooftree
\prooftree
\prooftree
\prooftree
\prooftree
\prooftree
\mbox{\footnotesize\textcircled{4}\tab\tab\tab\tab\tab\tab\tab\tab}\tab
\prooftree
\prooftree
\prooftree
\prooftree
\prooftree
\justifies
\mbox{\fbox{$Nt(s(m))$}}\ \Rightarrow\ Nt(s(m))
\endprooftree
\justifies
\mbox{\fbox{${\blacksquare}Nt(s(m))$}}\ \Rightarrow\ Nt(s(m))
\using {\blacksquare}L
\endprooftree
\justifies
{\blacksquare}Nt(s(m))\ \Rightarrow\ \fbox{${\exists}gNt(s(g))$}
\using {\exists}R
\endprooftree
\justifies
[{\blacksquare}Nt(s(m))]\ \Rightarrow\ \fbox{${\langle\rangle}{\exists}gNt(s(g))$}
\using {\langle\rangle}R
\endprooftree
\prooftree
\justifies
\mbox{\fbox{$Sf$}}\ \Rightarrow\ Sf
\endprooftree
\justifies
[{\blacksquare}Nt(s(m))], \mbox{\fbox{${\langle\rangle}{\exists}gNt(s(g))\backslash Sf$}}\ \Rightarrow\ Sf
\using {\backslash}L
\endprooftree
\justifies
\begin{array}{c}
{}[{\blacksquare}Nt(s(m))], \mbox{\fbox{$({\langle\rangle}{\exists}gNt(s(g))\backslash Sf)/{\square}({\langle\rangle}{\exists}gNt(s(g))\backslash Si)$}}, {\blacksquare}(({\it PP}{\it to}/{\exists}aNa){\sqcap}{\forall}n(({\langle\rangle}Nn\backslash Si)/({\langle\rangle}Nn\backslash Sb))), [[{\square}(({\langle\rangle}{\exists}aNa\backslash Sb)/{\exists}aNa), {\blacksquare}Nt(s(n)), {\blacksquare}{\forall}a{\forall}f((?{\blacksquare}({\langle\rangle}Na\backslash Sf)\backslash {[]^{-1}}{[]^{-1}}({\langle\rangle}Na\backslash Sf))/{\blacksquare}({\langle\rangle}Na\backslash Sf)), {\square}(({\langle\rangle}{\exists}aNa\backslash Sb)/{\exists}aNa),\\
{\blacksquare}{\forall}f{\forall}a((({\langle\rangle}Na\backslash Sf){{}{\uparrow}{}}{\blacksquare}Nt(s(n))){{}{\downarrow}{}}({\blacksquare}({\langle\rangle}Na\backslash Sf){|}{\blacksquare}Nt(s(n))))]]\ \Rightarrow\ Sf
\end{array}
\using {/}L
\endprooftree
\justifies
\begin{array}{c}
{}[{\blacksquare}Nt(s(m))], \mbox{\fbox{${\square}(({\langle\rangle}{\exists}gNt(s(g))\backslash Sf)/{\square}({\langle\rangle}{\exists}gNt(s(g))\backslash Si))$}}, {\blacksquare}(({\it PP}{\it to}/{\exists}aNa){\sqcap}{\forall}n(({\langle\rangle}Nn\backslash Si)/({\langle\rangle}Nn\backslash Sb))), [[{\square}(({\langle\rangle}{\exists}aNa\backslash Sb)/{\exists}aNa), {\blacksquare}Nt(s(n)), {\blacksquare}{\forall}a{\forall}f((?{\blacksquare}({\langle\rangle}Na\backslash Sf)\backslash {[]^{-1}}{[]^{-1}}({\langle\rangle}Na\backslash Sf))/{\blacksquare}({\langle\rangle}Na\backslash Sf)), {\square}(({\langle\rangle}{\exists}aNa\backslash Sb)/{\exists}aNa),\\
{\blacksquare}{\forall}f{\forall}a((({\langle\rangle}Na\backslash Sf){{}{\uparrow}{}}{\blacksquare}Nt(s(n))){{}{\downarrow}{}}({\blacksquare}({\langle\rangle}Na\backslash Sf){|}{\blacksquare}Nt(s(n))))]]\ \Rightarrow\ Sf
\end{array}
\using {\Box}L
\endprooftree
\justifies
\begin{array}{c}
{}[{\blacksquare}Nt(s(m))], {\square}(({\langle\rangle}{\exists}gNt(s(g))\backslash Sf)/{\square}({\langle\rangle}{\exists}gNt(s(g))\backslash Si)), {\blacksquare}(({\it PP}{\it to}/{\exists}aNa){\sqcap}{\forall}n(({\langle\rangle}Nn\backslash Si)/({\langle\rangle}Nn\backslash Sb))), [[{\square}(({\langle\rangle}{\exists}aNa\backslash Sb)/{\exists}aNa), {\tt 1}, {\blacksquare}{\forall}a{\forall}f((?{\blacksquare}({\langle\rangle}Na\backslash Sf)\backslash {[]^{-1}}{[]^{-1}}({\langle\rangle}Na\backslash Sf))/{\blacksquare}({\langle\rangle}Na\backslash Sf)), {\square}(({\langle\rangle}{\exists}aNa\backslash Sb)/{\exists}aNa),\\
{\blacksquare}{\forall}f{\forall}a((({\langle\rangle}Na\backslash Sf){{}{\uparrow}{}}{\blacksquare}Nt(s(n))){{}{\downarrow}{}}({\blacksquare}({\langle\rangle}Na\backslash Sf){|}{\blacksquare}Nt(s(n))))]]\ \Rightarrow\ Sf{{}{\uparrow}{}}{\blacksquare}Nt(s(n))
\end{array}
\using {\uparrow}R
\endprooftree
\prooftree
\justifies
\mbox{\fbox{$Sf$}}\ \Rightarrow\ Sf
\endprooftree
\justifies
\begin{array}{c}
{}[{\blacksquare}Nt(s(m))], {\square}(({\langle\rangle}{\exists}gNt(s(g))\backslash Sf)/{\square}({\langle\rangle}{\exists}gNt(s(g))\backslash Si)), {\blacksquare}(({\it PP}{\it to}/{\exists}aNa){\sqcap}{\forall}n(({\langle\rangle}Nn\backslash Si)/({\langle\rangle}Nn\backslash Sb))), [[{\square}(({\langle\rangle}{\exists}aNa\backslash Sb)/{\exists}aNa), \mbox{\fbox{$(Sf{{}{\uparrow}{}}{\blacksquare}Nt(s(n))){{}{\downarrow}{}}Sf$}},\\
{\blacksquare}{\forall}a{\forall}f((?{\blacksquare}({\langle\rangle}Na\backslash Sf)\backslash {[]^{-1}}{[]^{-1}}({\langle\rangle}Na\backslash Sf))/{\blacksquare}({\langle\rangle}Na\backslash Sf)), {\square}(({\langle\rangle}{\exists}aNa\backslash Sb)/{\exists}aNa), {\blacksquare}{\forall}f{\forall}a((({\langle\rangle}Na\backslash Sf){{}{\uparrow}{}}{\blacksquare}Nt(s(n))){{}{\downarrow}{}}({\blacksquare}({\langle\rangle}Na\backslash Sf){|}{\blacksquare}Nt(s(n))))]]\ \Rightarrow\ Sf
\end{array}
\using {\downarrow}L
\endprooftree
\justifies
\begin{array}{c}
{}[{\blacksquare}Nt(s(m))], {\square}(({\langle\rangle}{\exists}gNt(s(g))\backslash Sf)/{\square}({\langle\rangle}{\exists}gNt(s(g))\backslash Si)), {\blacksquare}(({\it PP}{\it to}/{\exists}aNa){\sqcap}{\forall}n(({\langle\rangle}Nn\backslash Si)/({\langle\rangle}Nn\backslash Sb))), [[{\square}(({\langle\rangle}{\exists}aNa\backslash Sb)/{\exists}aNa), \mbox{\fbox{${\forall}f((Sf{{}{\uparrow}{}}{\blacksquare}Nt(s(n))){{}{\downarrow}{}}Sf)$}},\\
{\blacksquare}{\forall}a{\forall}f((?{\blacksquare}({\langle\rangle}Na\backslash Sf)\backslash {[]^{-1}}{[]^{-1}}({\langle\rangle}Na\backslash Sf))/{\blacksquare}({\langle\rangle}Na\backslash Sf)), {\square}(({\langle\rangle}{\exists}aNa\backslash Sb)/{\exists}aNa), {\blacksquare}{\forall}f{\forall}a((({\langle\rangle}Na\backslash Sf){{}{\uparrow}{}}{\blacksquare}Nt(s(n))){{}{\downarrow}{}}({\blacksquare}({\langle\rangle}Na\backslash Sf){|}{\blacksquare}Nt(s(n))))]]\ \Rightarrow\ Sf
\end{array}
\using {\forall}L
\endprooftree
\justifies
\begin{array}{c}
{}[{\blacksquare}Nt(s(m))], {\square}(({\langle\rangle}{\exists}gNt(s(g))\backslash Sf)/{\square}({\langle\rangle}{\exists}gNt(s(g))\backslash Si)), {\blacksquare}(({\it PP}{\it to}/{\exists}aNa){\sqcap}{\forall}n(({\langle\rangle}Nn\backslash Si)/({\langle\rangle}Nn\backslash Sb))), [[{\square}(({\langle\rangle}{\exists}aNa\backslash Sb)/{\exists}aNa), \mbox{\fbox{${\forall}f((Sf{{}{\uparrow}{}}{\blacksquare}Nt(s(n))){{}{\downarrow}{}}Sf)/{\it CN}{\it s(n)}$}},\\
{\square}{\it CN}{\it s(n)}, {\blacksquare}{\forall}a{\forall}f((?{\blacksquare}({\langle\rangle}Na\backslash Sf)\backslash {[]^{-1}}{[]^{-1}}({\langle\rangle}Na\backslash Sf))/{\blacksquare}({\langle\rangle}Na\backslash Sf)), {\square}(({\langle\rangle}{\exists}aNa\backslash Sb)/{\exists}aNa), {\blacksquare}{\forall}f{\forall}a((({\langle\rangle}Na\backslash Sf){{}{\uparrow}{}}{\blacksquare}Nt(s(n))){{}{\downarrow}{}}({\blacksquare}({\langle\rangle}Na\backslash Sf){|}{\blacksquare}Nt(s(n))))]]\ \Rightarrow\ Sf
\end{array}
\using {/}L
\endprooftree
\justifies
\begin{array}{c}
{}[{\blacksquare}Nt(s(m))], {\square}(({\langle\rangle}{\exists}gNt(s(g))\backslash Sf)/{\square}({\langle\rangle}{\exists}gNt(s(g))\backslash Si)), {\blacksquare}(({\it PP}{\it to}/{\exists}aNa){\sqcap}{\forall}n(({\langle\rangle}Nn\backslash Si)/({\langle\rangle}Nn\backslash Sb))), [[{\square}(({\langle\rangle}{\exists}aNa\backslash Sb)/{\exists}aNa), \mbox{\fbox{${\forall}g({\forall}f((Sf{{}{\uparrow}{}}{\blacksquare}Nt(s(g))){{}{\downarrow}{}}Sf)/{\it CN}{\it s(g)})$}},\\
{\square}{\it CN}{\it s(n)}, {\blacksquare}{\forall}a{\forall}f((?{\blacksquare}({\langle\rangle}Na\backslash Sf)\backslash {[]^{-1}}{[]^{-1}}({\langle\rangle}Na\backslash Sf))/{\blacksquare}({\langle\rangle}Na\backslash Sf)), {\square}(({\langle\rangle}{\exists}aNa\backslash Sb)/{\exists}aNa), {\blacksquare}{\forall}f{\forall}a((({\langle\rangle}Na\backslash Sf){{}{\uparrow}{}}{\blacksquare}Nt(s(n))){{}{\downarrow}{}}({\blacksquare}({\langle\rangle}Na\backslash Sf){|}{\blacksquare}Nt(s(n))))]]\ \Rightarrow\ Sf
\end{array}
\using {\forall}L
\endprooftree
\justifies
\begin{array}{c}
{}[{\blacksquare}Nt(s(m))], {\square}(({\langle\rangle}{\exists}gNt(s(g))\backslash Sf)/{\square}({\langle\rangle}{\exists}gNt(s(g))\backslash Si)), {\blacksquare}(({\it PP}{\it to}/{\exists}aNa){\sqcap}{\forall}n(({\langle\rangle}Nn\backslash Si)/({\langle\rangle}Nn\backslash Sb))), [[{\square}(({\langle\rangle}{\exists}aNa\backslash Sb)/{\exists}aNa), \mbox{\fbox{${\blacksquare}{\forall}g({\forall}f((Sf{{}{\uparrow}{}}{\blacksquare}Nt(s(g))){{}{\downarrow}{}}Sf)/{\it CN}{\it s(g)})$}},\\
 {\square}{\it CN}{\it s(n)}, {\blacksquare}{\forall}a{\forall}f((?{\blacksquare}({\langle\rangle}Na\backslash Sf)\backslash {[]^{-1}}{[]^{-1}}({\langle\rangle}Na\backslash Sf))/{\blacksquare}({\langle\rangle}Na\backslash Sf)), {\square}(({\langle\rangle}{\exists}aNa\backslash Sb)/{\exists}aNa), {\blacksquare}{\forall}f{\forall}a((({\langle\rangle}Na\backslash Sf){{}{\uparrow}{}}{\blacksquare}Nt(s(n))){{}{\downarrow}{}}({\blacksquare}({\langle\rangle}Na\backslash Sf){|}{\blacksquare}Nt(s(n))))]]\ \Rightarrow\ Sf
 \end{array}
\using {\blacksquare}L
\endprooftree}
$$
\vspace{0.15in}
This delivers semantics with existential commitment:
\disp{
$\exists C[(\mbox{\v{}}{\it fish}\ {\it C})\wedge ((\mbox{\v{}}{\it tries}\ \mbox{\^{}}[((\mbox{\v{}}{\it catch}\ {\it C})\ {\it j})\wedge ((\mbox{\v{}}{\it eat}\ {\it C})\ {\it j})])\ {\it j})]$}
The existential narrow scope derivation is:
\vspace{0.15in}
$$
\tiny
\prooftree
\prooftree
\prooftree
\prooftree
\prooftree
\prooftree
\prooftree
\prooftree
\justifies
\mbox{\fbox{$Nt(s(n))$}}\ \Rightarrow\ Nt(s(n))
\endprooftree
\justifies
\mbox{\fbox{${\blacksquare}Nt(s(n))$}}\ \Rightarrow\ Nt(s(n))
\using {\blacksquare}L
\endprooftree
\justifies
{\blacksquare}Nt(s(n))\ \Rightarrow\ \fbox{${\exists}aNa$}
\using {\exists}R
\endprooftree
\prooftree
\prooftree
\prooftree
\prooftree
\justifies
Nt(s(50))\ \Rightarrow\ Nt(s(50))
\endprooftree
\justifies
Nt(s(50))\ \Rightarrow\ \fbox{${\exists}aNa$}
\using {\exists}R
\endprooftree
\justifies
[Nt(s(50))]\ \Rightarrow\ \fbox{${\langle\rangle}{\exists}aNa$}
\using {\langle\rangle}R
\endprooftree
\prooftree
\justifies
\mbox{\fbox{$Sb$}}\ \Rightarrow\ Sb
\endprooftree
\justifies
[Nt(s(50))], \mbox{\fbox{${\langle\rangle}{\exists}aNa\backslash Sb$}}\ \Rightarrow\ Sb
\using {\backslash}L
\endprooftree
\justifies
[Nt(s(50))], \mbox{\fbox{$({\langle\rangle}{\exists}aNa\backslash Sb)/{\exists}aNa$}}, {\blacksquare}Nt(s(n))\ \Rightarrow\ Sb
\using {/}L
\endprooftree
\justifies
[Nt(s(50))], \mbox{\fbox{${\square}(({\langle\rangle}{\exists}aNa\backslash Sb)/{\exists}aNa)$}}, {\blacksquare}Nt(s(n))\ \Rightarrow\ Sb
\using {\Box}L
\endprooftree
\justifies
{\langle\rangle}Nt(s(50)), {\square}(({\langle\rangle}{\exists}aNa\backslash Sb)/{\exists}aNa), {\blacksquare}Nt(s(n))\ \Rightarrow\ Sb
\using {\langle\rangle}L
\endprooftree
\justifies
{\square}(({\langle\rangle}{\exists}aNa\backslash Sb)/{\exists}aNa), {\blacksquare}Nt(s(n))\ \Rightarrow\ {\langle\rangle}Nt(s(50))\backslash Sb
\using {\backslash}R
\endprooftree
\justifies
\begin{array}{c}
{\square}(({\langle\rangle}{\exists}aNa\backslash Sb)/{\exists}aNa), {\tt 1}\ \Rightarrow\ ({\langle\rangle}Nt(s(50))\backslash Sb){{}{\uparrow}{}}{\blacksquare}Nt(s(n))\\
\mbox{\footnotesize\textcircled{1}}
\end{array}
\using {\uparrow}R
\endprooftree
\tb
\prooftree
\prooftree
\prooftree
\prooftree
\prooftree
\prooftree
\prooftree
\prooftree
\prooftree
\justifies
\mbox{\fbox{$Nt(s(n))$}}\ \Rightarrow\ Nt(s(n))
\endprooftree
\justifies
\mbox{\fbox{${\blacksquare}Nt(s(n))$}}\ \Rightarrow\ Nt(s(n))
\using {\blacksquare}L
\endprooftree
\justifies
{\blacksquare}Nt(s(n))\ \Rightarrow\ \fbox{${\exists}aNa$}
\using {\exists}R
\endprooftree
\prooftree
\prooftree
\prooftree
\prooftree
\justifies
Nt(s(50))\ \Rightarrow\ Nt(s(50))
\endprooftree
\justifies
Nt(s(50))\ \Rightarrow\ \fbox{${\exists}aNa$}
\using {\exists}R
\endprooftree
\justifies
[Nt(s(50))]\ \Rightarrow\ \fbox{${\langle\rangle}{\exists}aNa$}
\using {\langle\rangle}R
\endprooftree
\prooftree
\justifies
\mbox{\fbox{$Sb$}}\ \Rightarrow\ Sb
\endprooftree
\justifies
[Nt(s(50))], \mbox{\fbox{${\langle\rangle}{\exists}aNa\backslash Sb$}}\ \Rightarrow\ Sb
\using {\backslash}L
\endprooftree
\justifies
[Nt(s(50))], \mbox{\fbox{$({\langle\rangle}{\exists}aNa\backslash Sb)/{\exists}aNa$}}, {\blacksquare}Nt(s(n))\ \Rightarrow\ Sb
\using {/}L
\endprooftree
\justifies
[Nt(s(50))], \mbox{\fbox{${\square}(({\langle\rangle}{\exists}aNa\backslash Sb)/{\exists}aNa)$}}, {\blacksquare}Nt(s(n))\ \Rightarrow\ Sb
\using {\Box}L
\endprooftree
\justifies
{\langle\rangle}Nt(s(50)), {\square}(({\langle\rangle}{\exists}aNa\backslash Sb)/{\exists}aNa), {\blacksquare}Nt(s(n))\ \Rightarrow\ Sb
\using {\langle\rangle}L
\endprooftree
\justifies
{\square}(({\langle\rangle}{\exists}aNa\backslash Sb)/{\exists}aNa), {\blacksquare}Nt(s(n))\ \Rightarrow\ {\langle\rangle}Nt(s(50))\backslash Sb
\using {\backslash}R
\endprooftree
\justifies
{\square}(({\langle\rangle}{\exists}aNa\backslash Sb)/{\exists}aNa), {\blacksquare}Nt(s(n))\ \Rightarrow\ {\blacksquare}({\langle\rangle}Nt(s(50))\backslash Sb)
\using {\blacksquare}R
\endprooftree
\justifies
\begin{array}{c}
{\square}(({\langle\rangle}{\exists}aNa\backslash Sb)/{\exists}aNa), {\blacksquare}Nt(s(n))\ \Rightarrow\ \fbox{$?{\blacksquare}({\langle\rangle}Nt(s(50))\backslash Sb)$}\\
\mbox{\footnotesize\textcircled{2}}
\end{array}
\using {?}R
\endprooftree
$$
$$\tiny
\rotatebox{-90}{
\prooftree
\prooftree
\prooftree
\prooftree
\prooftree
\prooftree
\mbox{\footnotesize\textcircled{1}}\tab
\prooftree
\prooftree
\prooftree
\prooftree
\justifies
\mbox{\fbox{$Nt(s(n))$}}\ \Rightarrow\ Nt(s(n))
\endprooftree
\justifies
\mbox{\fbox{${\blacksquare}Nt(s(n))$}}\ \Rightarrow\ Nt(s(n))
\using {\blacksquare}L
\endprooftree
\justifies
{\blacksquare}Nt(s(n))\ \Rightarrow\ {\blacksquare}Nt(s(n))
\using {\blacksquare}R
\endprooftree
\prooftree
\prooftree
\prooftree
\prooftree
\prooftree
\prooftree
\prooftree
\prooftree
\prooftree
\prooftree
\prooftree
\justifies
Nt(s(50))\ \Rightarrow\ Nt(s(50))
\endprooftree
\justifies
[Nt(s(50))]\ \Rightarrow\ \fbox{${\langle\rangle}Nt(s(50))$}
\using {\langle\rangle}R
\endprooftree
\prooftree
\justifies
\mbox{\fbox{$Sb$}}\ \Rightarrow\ Sb
\endprooftree
\justifies
[Nt(s(50))], \mbox{\fbox{${\langle\rangle}Nt(s(50))\backslash Sb$}}\ \Rightarrow\ Sb
\using {\backslash}L
\endprooftree
\justifies
[Nt(s(50))], \mbox{\fbox{${\blacksquare}({\langle\rangle}Nt(s(50))\backslash Sb)$}}\ \Rightarrow\ Sb
\using {\blacksquare}L
\endprooftree
\justifies
{\langle\rangle}Nt(s(50)), {\blacksquare}({\langle\rangle}Nt(s(50))\backslash Sb)\ \Rightarrow\ Sb
\using {\langle\rangle}L
\endprooftree
\justifies
{\blacksquare}({\langle\rangle}Nt(s(50))\backslash Sb)\ \Rightarrow\ {\langle\rangle}Nt(s(50))\backslash Sb
\using {\backslash}R
\endprooftree
\justifies
{\blacksquare}({\langle\rangle}Nt(s(50))\backslash Sb)\ \Rightarrow\ {\blacksquare}({\langle\rangle}Nt(s(50))\backslash Sb)
\using {\blacksquare}R
\endprooftree
\prooftree
\mbox{\footnotesize\textcircled{2}}\tab
\prooftree
\prooftree
\prooftree
\prooftree
\prooftree
\justifies
Nt(s(50))\ \Rightarrow\ Nt(s(50))
\endprooftree
\justifies
[Nt(s(50))]\ \Rightarrow\ \fbox{${\langle\rangle}Nt(s(50))$}
\using {\langle\rangle}R
\endprooftree
\prooftree
\justifies
\mbox{\fbox{$Sb$}}\ \Rightarrow\ Sb
\endprooftree
\justifies
[Nt(s(50))], \mbox{\fbox{${\langle\rangle}Nt(s(50))\backslash Sb$}}\ \Rightarrow\ Sb
\using {\backslash}L
\endprooftree
\justifies
[Nt(s(50))], [\mbox{\fbox{${[]^{-1}}({\langle\rangle}Nt(s(50))\backslash Sb)$}}]\ \Rightarrow\ Sb
\using {[]^{-1}}L
\endprooftree
\justifies
[Nt(s(50))], [[\mbox{\fbox{${[]^{-1}}{[]^{-1}}({\langle\rangle}Nt(s(50))\backslash Sb)$}}]]\ \Rightarrow\ Sb
\using {[]^{-1}}L
\endprooftree
\justifies
[Nt(s(50))], [[{\square}(({\langle\rangle}{\exists}aNa\backslash Sb)/{\exists}aNa), {\blacksquare}Nt(s(n)), \mbox{\fbox{$?{\blacksquare}({\langle\rangle}Nt(s(50))\backslash Sb)\backslash {[]^{-1}}{[]^{-1}}({\langle\rangle}Nt(s(50))\backslash Sb)$}}]]\ \Rightarrow\ Sb
\using {\backslash}L
\endprooftree
\justifies
{}[Nt(s(50))], [[{\square}(({\langle\rangle}{\exists}aNa\backslash Sb)/{\exists}aNa), {\blacksquare}Nt(s(n)), \mbox{\fbox{$(?{\blacksquare}({\langle\rangle}Nt(s(50))\backslash Sb)\backslash {[]^{-1}}{[]^{-1}}({\langle\rangle}Nt(s(50))\backslash Sb))/{\blacksquare}({\langle\rangle}Nt(s(50))\backslash Sb)$}}, {\blacksquare}({\langle\rangle}Nt(s(50))\backslash Sb)]]\ \Rightarrow\ Sb
\using {/}L
\endprooftree
\justifies
[Nt(s(50))], [[{\square}(({\langle\rangle}{\exists}aNa\backslash Sb)/{\exists}aNa), {\blacksquare}Nt(s(n)), \mbox{\fbox{${\forall}f((?{\blacksquare}({\langle\rangle}Nt(s(50))\backslash Sf)\backslash {[]^{-1}}{[]^{-1}}({\langle\rangle}Nt(s(50))\backslash Sf))/{\blacksquare}({\langle\rangle}Nt(s(50))\backslash Sf))$}}, {\blacksquare}({\langle\rangle}Nt(s(50))\backslash Sb)]]\ \Rightarrow\ Sb
\using {\forall}L
\endprooftree
\justifies
[Nt(s(50))], [[{\square}(({\langle\rangle}{\exists}aNa\backslash Sb)/{\exists}aNa), {\blacksquare}Nt(s(n)), \mbox{\fbox{${\forall}a{\forall}f((?{\blacksquare}({\langle\rangle}Na\backslash Sf)\backslash {[]^{-1}}{[]^{-1}}({\langle\rangle}Na\backslash Sf))/{\blacksquare}({\langle\rangle}Na\backslash Sf))$}}, {\blacksquare}({\langle\rangle}Nt(s(50))\backslash Sb)]]\ \Rightarrow\ Sb
\using {\forall}L
\endprooftree
\justifies
[Nt(s(50))], [[{\square}(({\langle\rangle}{\exists}aNa\backslash Sb)/{\exists}aNa), {\blacksquare}Nt(s(n)), \mbox{\fbox{${\blacksquare}{\forall}a{\forall}f((?{\blacksquare}({\langle\rangle}Na\backslash Sf)\backslash {[]^{-1}}{[]^{-1}}({\langle\rangle}Na\backslash Sf))/{\blacksquare}({\langle\rangle}Na\backslash Sf))$}}, {\blacksquare}({\langle\rangle}Nt(s(50))\backslash Sb)]]\ \Rightarrow\ Sb
\using {\blacksquare}L
\endprooftree
\justifies
[Nt(s(50))], [[{\square}(({\langle\rangle}{\exists}aNa\backslash Sb)/{\exists}aNa), {\blacksquare}Nt(s(n)), {\blacksquare}{\forall}a{\forall}f((?{\blacksquare}({\langle\rangle}Na\backslash Sf)\backslash {[]^{-1}}{[]^{-1}}({\langle\rangle}Na\backslash Sf))/{\blacksquare}({\langle\rangle}Na\backslash Sf)), \mbox{\fbox{${\blacksquare}({\langle\rangle}Nt(s(50))\backslash Sb){|}{\blacksquare}Nt(s(n))$}}]]\ \Rightarrow\ Sb
\using {|}L
\endprooftree
\justifies
[Nt(s(50))], [[{\square}(({\langle\rangle}{\exists}aNa\backslash Sb)/{\exists}aNa), {\blacksquare}Nt(s(n)), {\blacksquare}{\forall}a{\forall}f((?{\blacksquare}({\langle\rangle}Na\backslash Sf)\backslash {[]^{-1}}{[]^{-1}}({\langle\rangle}Na\backslash Sf))/{\blacksquare}({\langle\rangle}Na\backslash Sf)), {\square}(({\langle\rangle}{\exists}aNa\backslash Sb)/{\exists}aNa), \mbox{\fbox{$(({\langle\rangle}Nt(s(50))\backslash Sb){{}{\uparrow}{}}{\blacksquare}Nt(s(n))){{}{\downarrow}{}}({\blacksquare}({\langle\rangle}Nt(s(50))\backslash Sb){|}{\blacksquare}Nt(s(n)))$}}]]\ \Rightarrow\ Sb
\using {\downarrow}L
\endprooftree
\justifies
[Nt(s(50))], [[{\square}(({\langle\rangle}{\exists}aNa\backslash Sb)/{\exists}aNa), {\blacksquare}Nt(s(n)), {\blacksquare}{\forall}a{\forall}f((?{\blacksquare}({\langle\rangle}Na\backslash Sf)\backslash {[]^{-1}}{[]^{-1}}({\langle\rangle}Na\backslash Sf))/{\blacksquare}({\langle\rangle}Na\backslash Sf)), {\square}(({\langle\rangle}{\exists}aNa\backslash Sb)/{\exists}aNa), \mbox{\fbox{${\forall}a((({\langle\rangle}Na\backslash Sb){{}{\uparrow}{}}{\blacksquare}Nt(s(n))){{}{\downarrow}{}}({\blacksquare}({\langle\rangle}Na\backslash Sb){|}{\blacksquare}Nt(s(n))))$}}]]\ \Rightarrow\ Sb
\using {\forall}L
\endprooftree
\justifies
[Nt(s(50))], [[{\square}(({\langle\rangle}{\exists}aNa\backslash Sb)/{\exists}aNa), {\blacksquare}Nt(s(n)), {\blacksquare}{\forall}a{\forall}f((?{\blacksquare}({\langle\rangle}Na\backslash Sf)\backslash {[]^{-1}}{[]^{-1}}({\langle\rangle}Na\backslash Sf))/{\blacksquare}({\langle\rangle}Na\backslash Sf)), {\square}(({\langle\rangle}{\exists}aNa\backslash Sb)/{\exists}aNa), \mbox{\fbox{${\forall}f{\forall}a((({\langle\rangle}Na\backslash Sf){{}{\uparrow}{}}{\blacksquare}Nt(s(n))){{}{\downarrow}{}}({\blacksquare}({\langle\rangle}Na\backslash Sf){|}{\blacksquare}Nt(s(n))))$}}]]\ \Rightarrow\ Sb
\using {\forall}L
\endprooftree
\justifies
[Nt(s(50))], [[{\square}(({\langle\rangle}{\exists}aNa\backslash Sb)/{\exists}aNa), {\blacksquare}Nt(s(n)), {\blacksquare}{\forall}a{\forall}f((?{\blacksquare}({\langle\rangle}Na\backslash Sf)\backslash {[]^{-1}}{[]^{-1}}({\langle\rangle}Na\backslash Sf))/{\blacksquare}({\langle\rangle}Na\backslash Sf)), {\square}(({\langle\rangle}{\exists}aNa\backslash Sb)/{\exists}aNa), \mbox{\fbox{${\blacksquare}{\forall}f{\forall}a((({\langle\rangle}Na\backslash Sf){{}{\uparrow}{}}{\blacksquare}Nt(s(n))){{}{\downarrow}{}}({\blacksquare}({\langle\rangle}Na\backslash Sf){|}{\blacksquare}Nt(s(n))))$}}]]\ \Rightarrow\ Sb
\using {\blacksquare}L
\endprooftree
\justifies
{\langle\rangle}Nt(s(50)), [[{\square}(({\langle\rangle}{\exists}aNa\backslash Sb)/{\exists}aNa), {\blacksquare}Nt(s(n)), {\blacksquare}{\forall}a{\forall}f((?{\blacksquare}({\langle\rangle}Na\backslash Sf)\backslash {[]^{-1}}{[]^{-1}}({\langle\rangle}Na\backslash Sf))/{\blacksquare}({\langle\rangle}Na\backslash Sf)), {\square}(({\langle\rangle}{\exists}aNa\backslash Sb)/{\exists}aNa), {\blacksquare}{\forall}f{\forall}a((({\langle\rangle}Na\backslash Sf){{}{\uparrow}{}}{\blacksquare}Nt(s(n))){{}{\downarrow}{}}({\blacksquare}({\langle\rangle}Na\backslash Sf){|}{\blacksquare}Nt(s(n))))]]\ \Rightarrow\ Sb
\using {\langle\rangle}L
\endprooftree
\justifies
\begin{array}{c}
{}[[{\square}(({\langle\rangle}{\exists}aNa\backslash Sb)/{\exists}aNa), {\blacksquare}Nt(s(n)), {\blacksquare}{\forall}a{\forall}f((?{\blacksquare}({\langle\rangle}Na\backslash Sf)\backslash {[]^{-1}}{[]^{-1}}({\langle\rangle}Na\backslash Sf))/{\blacksquare}({\langle\rangle}Na\backslash Sf)), {\square}(({\langle\rangle}{\exists}aNa\backslash Sb)/{\exists}aNa), {\blacksquare}{\forall}f{\forall}a((({\langle\rangle}Na\backslash Sf){{}{\uparrow}{}}{\blacksquare}Nt(s(n))){{}{\downarrow}{}}({\blacksquare}({\langle\rangle}Na\backslash Sf){|}{\blacksquare}Nt(s(n))))]]\ \Rightarrow\ {\langle\rangle}Nt(s(50))\backslash Sb\\
\mbox{\footnotesize\textcircled{3}}
\end{array}
\using {\backslash}R
\endprooftree}
$$
$$\tiny
\rotatebox{-90}{
\prooftree
\prooftree
\prooftree
\prooftree
\prooftree
\justifies
\mbox{\fbox{${\it CN}{\it s(n)}$}}\ \Rightarrow\ {\it CN}{\it s(n)}
\endprooftree
\justifies
\mbox{\fbox{${\square}{\it CN}{\it s(n)}$}}\ \Rightarrow\ {\it CN}{\it s(n)}
\using {\Box}L
\endprooftree
\prooftree
\prooftree
\prooftree
\prooftree
\prooftree
\prooftree
\prooftree
\prooftree
\prooftree
\prooftree
\prooftree
\prooftree
\prooftree
\mbox{\footnotesize\textcircled{3}\tab\tab\tab\tab}
\prooftree
\prooftree
\prooftree
\justifies
Nt(s(50))\ \Rightarrow\ Nt(s(50))
\endprooftree
\justifies
[Nt(s(50))]\ \Rightarrow\ \fbox{${\langle\rangle}Nt(s(50))$}
\using {\langle\rangle}R
\endprooftree
\prooftree
\justifies
\mbox{\fbox{$Si$}}\ \Rightarrow\ Si
\endprooftree
\justifies
[Nt(s(50))], \mbox{\fbox{${\langle\rangle}Nt(s(50))\backslash Si$}}\ \Rightarrow\ Si
\using {\backslash}L
\endprooftree
\justifies
\begin{array}{c}
{}[Nt(s(50))], \mbox{\fbox{$({\langle\rangle}Nt(s(50))\backslash Si)/({\langle\rangle}Nt(s(50))\backslash Sb)$}}, [[{\square}(({\langle\rangle}{\exists}aNa\backslash Sb)/{\exists}aNa), {\blacksquare}Nt(s(n)), {\blacksquare}{\forall}a{\forall}f((?{\blacksquare}({\langle\rangle}Na\backslash Sf)\backslash {[]^{-1}}{[]^{-1}}({\langle\rangle}Na\backslash Sf))/{\blacksquare}({\langle\rangle}Na\backslash Sf)),\\
{\square}(({\langle\rangle}{\exists}aNa\backslash Sb)/{\exists}aNa), {\blacksquare}{\forall}f{\forall}a((({\langle\rangle}Na\backslash Sf){{}{\uparrow}{}}{\blacksquare}Nt(s(n))){{}{\downarrow}{}}({\blacksquare}({\langle\rangle}Na\backslash Sf){|}{\blacksquare}Nt(s(n))))]]\ \Rightarrow\ Si
\end{array}
\using {/}L
\endprooftree
\justifies
\begin{array}{c}
{}[Nt(s(50))], \mbox{\fbox{${\forall}n(({\langle\rangle}Nn\backslash Si)/({\langle\rangle}Nn\backslash Sb))$}}, [[{\square}(({\langle\rangle}{\exists}aNa\backslash Sb)/{\exists}aNa), {\blacksquare}Nt(s(n)), {\blacksquare}{\forall}a{\forall}f((?{\blacksquare}({\langle\rangle}Na\backslash Sf)\backslash {[]^{-1}}{[]^{-1}}({\langle\rangle}Na\backslash Sf))/{\blacksquare}({\langle\rangle}Na\backslash Sf)),\\
{\square}(({\langle\rangle}{\exists}aNa\backslash Sb)/{\exists}aNa), {\blacksquare}{\forall}f{\forall}a((({\langle\rangle}Na\backslash Sf){{}{\uparrow}{}}{\blacksquare}Nt(s(n))){{}{\downarrow}{}}({\blacksquare}({\langle\rangle}Na\backslash Sf){|}{\blacksquare}Nt(s(n))))]]\ \Rightarrow\ Si
\end{array}
\using {\forall}L
\endprooftree
\justifies
\begin{array}{c}
{}[Nt(s(50))], \mbox{\fbox{$({\it PP}{\it to}/{\exists}aNa){\sqcap}{\forall}n(({\langle\rangle}Nn\backslash Si)/({\langle\rangle}Nn\backslash Sb))$}}, [[{\square}(({\langle\rangle}{\exists}aNa\backslash Sb)/{\exists}aNa), {\blacksquare}Nt(s(n)), {\blacksquare}{\forall}a{\forall}f((?{\blacksquare}({\langle\rangle}Na\backslash Sf)\backslash {[]^{-1}}{[]^{-1}}({\langle\rangle}Na\backslash Sf))/{\blacksquare}({\langle\rangle}Na\backslash Sf)),\\
{\square}(({\langle\rangle}{\exists}aNa\backslash Sb)/{\exists}aNa), {\blacksquare}{\forall}f{\forall}a((({\langle\rangle}Na\backslash Sf){{}{\uparrow}{}}{\blacksquare}Nt(s(n))){{}{\downarrow}{}}({\blacksquare}({\langle\rangle}Na\backslash Sf){|}{\blacksquare}Nt(s(n))))]]\ \Rightarrow\ Si
\end{array}
\using {\sqcap}L
\endprooftree
\justifies
\begin{array}{c}
{}[Nt(s(50))], \mbox{\fbox{${\blacksquare}(({\it PP}{\it to}/{\exists}aNa){\sqcap}{\forall}n(({\langle\rangle}Nn\backslash Si)/({\langle\rangle}Nn\backslash Sb)))$}}, [[{\square}(({\langle\rangle}{\exists}aNa\backslash Sb)/{\exists}aNa), {\blacksquare}Nt(s(n)), {\blacksquare}{\forall}a{\forall}f((?{\blacksquare}({\langle\rangle}Na\backslash Sf)\backslash {[]^{-1}}{[]^{-1}}({\langle\rangle}Na\backslash Sf))/{\blacksquare}({\langle\rangle}Na\backslash Sf)),\\
{\square}(({\langle\rangle}{\exists}aNa\backslash Sb)/{\exists}aNa), {\blacksquare}{\forall}f{\forall}a((({\langle\rangle}Na\backslash Sf){{}{\uparrow}{}}{\blacksquare}Nt(s(n))){{}{\downarrow}{}}({\blacksquare}({\langle\rangle}Na\backslash Sf){|}{\blacksquare}Nt(s(n))))]]\ \Rightarrow\ Si
\end{array}
\using {\blacksquare}L
\endprooftree
\justifies
\begin{array}{c}
{}[{\exists}gNt(s(g))], {\blacksquare}(({\it PP}{\it to}/{\exists}aNa){\sqcap}{\forall}n(({\langle\rangle}Nn\backslash Si)/({\langle\rangle}Nn\backslash Sb))), [[{\square}(({\langle\rangle}{\exists}aNa\backslash Sb)/{\exists}aNa), {\blacksquare}Nt(s(n)), {\blacksquare}{\forall}a{\forall}f((?{\blacksquare}({\langle\rangle}Na\backslash Sf)\backslash {[]^{-1}}{[]^{-1}}({\langle\rangle}Na\backslash Sf))/{\blacksquare}({\langle\rangle}Na\backslash Sf)),\\
{\square}(({\langle\rangle}{\exists}aNa\backslash Sb)/{\exists}aNa), {\blacksquare}{\forall}f{\forall}a((({\langle\rangle}Na\backslash Sf){{}{\uparrow}{}}{\blacksquare}Nt(s(n))){{}{\downarrow}{}}({\blacksquare}({\langle\rangle}Na\backslash Sf){|}{\blacksquare}Nt(s(n))))]]\ \Rightarrow\ Si
\end{array}
\using {\exists}L
\endprooftree
\justifies
\begin{array}{c}
{\langle\rangle}{\exists}gNt(s(g)), {\blacksquare}(({\it PP}{\it to}/{\exists}aNa){\sqcap}{\forall}n(({\langle\rangle}Nn\backslash Si)/({\langle\rangle}Nn\backslash Sb))), [[{\square}(({\langle\rangle}{\exists}aNa\backslash Sb)/{\exists}aNa), {\blacksquare}Nt(s(n)), {\blacksquare}{\forall}a{\forall}f((?{\blacksquare}({\langle\rangle}Na\backslash Sf)\backslash {[]^{-1}}{[]^{-1}}({\langle\rangle}Na\backslash Sf))/{\blacksquare}({\langle\rangle}Na\backslash Sf)),\\
{\square}(({\langle\rangle}{\exists}aNa\backslash Sb)/{\exists}aNa), {\blacksquare}{\forall}f{\forall}a((({\langle\rangle}Na\backslash Sf){{}{\uparrow}{}}{\blacksquare}Nt(s(n))){{}{\downarrow}{}}({\blacksquare}({\langle\rangle}Na\backslash Sf){|}{\blacksquare}Nt(s(n))))]]\ \Rightarrow\ Si
\end{array}
\using {\langle\rangle}L
\endprooftree
\justifies
\begin{array}{c}
{\blacksquare}(({\it PP}{\it to}/{\exists}aNa){\sqcap}{\forall}n(({\langle\rangle}Nn\backslash Si)/({\langle\rangle}Nn\backslash Sb))), [[{\square}(({\langle\rangle}{\exists}aNa\backslash Sb)/{\exists}aNa), {\blacksquare}Nt(s(n)), {\blacksquare}{\forall}a{\forall}f((?{\blacksquare}({\langle\rangle}Na\backslash Sf)\backslash {[]^{-1}}{[]^{-1}}({\langle\rangle}Na\backslash Sf))/{\blacksquare}({\langle\rangle}Na\backslash Sf)),\\ {\square}(({\langle\rangle}{\exists}aNa\backslash Sb)/{\exists}aNa), {\blacksquare}{\forall}f{\forall}a((({\langle\rangle}Na\backslash Sf){{}{\uparrow}{}}{\blacksquare}Nt(s(n))){{}{\downarrow}{}}({\blacksquare}({\langle\rangle}Na\backslash Sf){|}{\blacksquare}Nt(s(n))))]]\ \Rightarrow\ {\langle\rangle}{\exists}gNt(s(g))\backslash Si
\end{array}
\using {\backslash}R
\endprooftree
\justifies
\begin{array}{c}
{\blacksquare}(({\it PP}{\it to}/{\exists}aNa){\sqcap}{\forall}n(({\langle\rangle}Nn\backslash Si)/({\langle\rangle}Nn\backslash Sb))), [[{\square}(({\langle\rangle}{\exists}aNa\backslash Sb)/{\exists}aNa), {\blacksquare}Nt(s(n)), {\blacksquare}{\forall}a{\forall}f((?{\blacksquare}({\langle\rangle}Na\backslash Sf)\backslash {[]^{-1}}{[]^{-1}}({\langle\rangle}Na\backslash Sf))/{\blacksquare}({\langle\rangle}Na\backslash Sf)),\\
{\square}(({\langle\rangle}{\exists}aNa\backslash Sb)/{\exists}aNa), {\blacksquare}{\forall}f{\forall}a((({\langle\rangle}Na\backslash Sf){{}{\uparrow}{}}{\blacksquare}Nt(s(n))){{}{\downarrow}{}}({\blacksquare}({\langle\rangle}Na\backslash Sf){|}{\blacksquare}Nt(s(n))))]]\ \Rightarrow\ {\square}({\langle\rangle}{\exists}gNt(s(g))\backslash Si)
\end{array}
\using {\Box}R
\endprooftree
\prooftree
\prooftree
\prooftree
\prooftree
\prooftree
\justifies
\mbox{\fbox{$Nt(s(m))$}}\ \Rightarrow\ Nt(s(m))
\endprooftree
\justifies
\mbox{\fbox{${\blacksquare}Nt(s(m))$}}\ \Rightarrow\ Nt(s(m))
\using {\blacksquare}L
\endprooftree
\justifies
{\blacksquare}Nt(s(m))\ \Rightarrow\ \fbox{${\exists}gNt(s(g))$}
\using {\exists}R
\endprooftree
\justifies
[{\blacksquare}Nt(s(m))]\ \Rightarrow\ \fbox{${\langle\rangle}{\exists}gNt(s(g))$}
\using {\langle\rangle}R
\endprooftree
\prooftree
\justifies
\mbox{\fbox{$Sf$}}\ \Rightarrow\ Sf
\endprooftree
\justifies
[{\blacksquare}Nt(s(m))], \mbox{\fbox{${\langle\rangle}{\exists}gNt(s(g))\backslash Sf$}}\ \Rightarrow\ Sf
\using {\backslash}L
\endprooftree
\justifies
\begin{array}{c}
{}[{\blacksquare}Nt(s(m))], \mbox{\fbox{$({\langle\rangle}{\exists}gNt(s(g))\backslash Sf)/{\square}({\langle\rangle}{\exists}gNt(s(g))\backslash Si)$}}, {\blacksquare}(({\it PP}{\it to}/{\exists}aNa){\sqcap}{\forall}n(({\langle\rangle}Nn\backslash Si)/({\langle\rangle}Nn\backslash Sb))), [[{\square}(({\langle\rangle}{\exists}aNa\backslash Sb)/{\exists}aNa), {\blacksquare}Nt(s(n)), {\blacksquare}{\forall}a{\forall}f((?{\blacksquare}({\langle\rangle}Na\backslash Sf)\backslash {[]^{-1}}{[]^{-1}}({\langle\rangle}Na\backslash Sf))/{\blacksquare}({\langle\rangle}Na\backslash Sf)),\\
{\square}(({\langle\rangle}{\exists}aNa\backslash Sb)/{\exists}aNa), {\blacksquare}{\forall}f{\forall}a((({\langle\rangle}Na\backslash Sf){{}{\uparrow}{}}{\blacksquare}Nt(s(n))){{}{\downarrow}{}}({\blacksquare}({\langle\rangle}Na\backslash Sf){|}{\blacksquare}Nt(s(n))))]]\ \Rightarrow\ Sf
\end{array}
\using {/}L
\endprooftree
\justifies
\begin{array}{c}
[{\blacksquare}Nt(s(m))], \mbox{\fbox{${\square}(({\langle\rangle}{\exists}gNt(s(g))\backslash Sf)/{\square}({\langle\rangle}{\exists}gNt(s(g))\backslash Si))$}}, {\blacksquare}(({\it PP}{\it to}/{\exists}aNa){\sqcap}{\forall}n(({\langle\rangle}Nn\backslash Si)/({\langle\rangle}Nn\backslash Sb))), [[{\square}(({\langle\rangle}{\exists}aNa\backslash Sb)/{\exists}aNa), {\blacksquare}Nt(s(n)), {\blacksquare}{\forall}a{\forall}f((?{\blacksquare}({\langle\rangle}Na\backslash Sf)\backslash {[]^{-1}}{[]^{-1}}({\langle\rangle}Na\backslash Sf))/{\blacksquare}({\langle\rangle}Na\backslash Sf)),\\
{\square}(({\langle\rangle}{\exists}aNa\backslash Sb)/{\exists}aNa), {\blacksquare}{\forall}f{\forall}a((({\langle\rangle}Na\backslash Sf){{}{\uparrow}{}}{\blacksquare}Nt(s(n))){{}{\downarrow}{}}({\blacksquare}({\langle\rangle}Na\backslash Sf){|}{\blacksquare}Nt(s(n))))]]\ \Rightarrow\ Sf
\end{array}
\using {\Box}L
\endprooftree
\justifies
\begin{array}{c}
{}[{\blacksquare}Nt(s(m))], {\square}(({\langle\rangle}{\exists}gNt(s(g))\backslash Sf)/{\square}({\langle\rangle}{\exists}gNt(s(g))\backslash Si)), {\blacksquare}(({\it PP}{\it to}/{\exists}aNa){\sqcap}{\forall}n(({\langle\rangle}Nn\backslash Si)/({\langle\rangle}Nn\backslash Sb))), [[{\square}(({\langle\rangle}{\exists}aNa\backslash Sb)/{\exists}aNa), {\tt 1}, {\blacksquare}{\forall}a{\forall}f((?{\blacksquare}({\langle\rangle}Na\backslash Sf)\backslash {[]^{-1}}{[]^{-1}}({\langle\rangle}Na\backslash Sf))/{\blacksquare}({\langle\rangle}Na\backslash Sf)),\\
{\square}(({\langle\rangle}{\exists}aNa\backslash Sb)/{\exists}aNa), {\blacksquare}{\forall}f{\forall}a((({\langle\rangle}Na\backslash Sf){{}{\uparrow}{}}{\blacksquare}Nt(s(n))){{}{\downarrow}{}}({\blacksquare}({\langle\rangle}Na\backslash Sf){|}{\blacksquare}Nt(s(n))))]]\ \Rightarrow\ Sf{{}{\uparrow}{}}{\blacksquare}Nt(s(n))
\end{array}
\using {\uparrow}R
\endprooftree
\prooftree
\justifies
\mbox{\fbox{$Sf$}}\ \Rightarrow\ Sf
\endprooftree
\justifies
\begin{array}{c}
{}[{\blacksquare}Nt(s(m))], {\square}(({\langle\rangle}{\exists}gNt(s(g))\backslash Sf)/{\square}({\langle\rangle}{\exists}gNt(s(g))\backslash Si)), {\blacksquare}(({\it PP}{\it to}/{\exists}aNa){\sqcap}{\forall}n(({\langle\rangle}Nn\backslash Si)/({\langle\rangle}Nn\backslash Sb))), [[{\square}(({\langle\rangle}{\exists}aNa\backslash Sb)/{\exists}aNa), \mbox{\fbox{$(Sf{{}{\uparrow}{}}{\blacksquare}Nt(s(n))){{}{\downarrow}{}}Sf$}},\\ {\blacksquare}{\forall}a{\forall}f((?{\blacksquare}({\langle\rangle}Na\backslash Sf)\backslash {[]^{-1}}{[]^{-1}}({\langle\rangle}Na\backslash Sf))/{\blacksquare}({\langle\rangle}Na\backslash Sf)), {\square}(({\langle\rangle}{\exists}aNa\backslash Sb)/{\exists}aNa), {\blacksquare}{\forall}f{\forall}a((({\langle\rangle}Na\backslash Sf){{}{\uparrow}{}}{\blacksquare}Nt(s(n))){{}{\downarrow}{}}({\blacksquare}({\langle\rangle}Na\backslash Sf){|}{\blacksquare}Nt(s(n))))]]\ \Rightarrow\ Sf
\end{array}
\using {\downarrow}L
\endprooftree
\justifies
\begin{array}{c}
{}[{\blacksquare}Nt(s(m))], {\square}(({\langle\rangle}{\exists}gNt(s(g))\backslash Sf)/{\square}({\langle\rangle}{\exists}gNt(s(g))\backslash Si)), {\blacksquare}(({\it PP}{\it to}/{\exists}aNa){\sqcap}{\forall}n(({\langle\rangle}Nn\backslash Si)/({\langle\rangle}Nn\backslash Sb))), [[{\square}(({\langle\rangle}{\exists}aNa\backslash Sb)/{\exists}aNa), \mbox{\fbox{${\forall}f((Sf{{}{\uparrow}{}}{\blacksquare}Nt(s(n))){{}{\downarrow}{}}Sf)$}},\\
{\blacksquare}{\forall}a{\forall}f((?{\blacksquare}({\langle\rangle}Na\backslash Sf)\backslash {[]^{-1}}{[]^{-1}}({\langle\rangle}Na\backslash Sf))/{\blacksquare}({\langle\rangle}Na\backslash Sf)), {\square}(({\langle\rangle}{\exists}aNa\backslash Sb)/{\exists}aNa), {\blacksquare}{\forall}f{\forall}a((({\langle\rangle}Na\backslash Sf){{}{\uparrow}{}}{\blacksquare}Nt(s(n))){{}{\downarrow}{}}({\blacksquare}({\langle\rangle}Na\backslash Sf){|}{\blacksquare}Nt(s(n))))]]\ \Rightarrow\ Sf
\end{array}
\using {\forall}L
\endprooftree
\justifies
\begin{array}{c}
{}[{\blacksquare}Nt(s(m))], {\square}(({\langle\rangle}{\exists}gNt(s(g))\backslash Sf)/{\square}({\langle\rangle}{\exists}gNt(s(g))\backslash Si)), {\blacksquare}(({\it PP}{\it to}/{\exists}aNa){\sqcap}{\forall}n(({\langle\rangle}Nn\backslash Si)/({\langle\rangle}Nn\backslash Sb))), [[{\square}(({\langle\rangle}{\exists}aNa\backslash Sb)/{\exists}aNa), \mbox{\fbox{${\forall}f((Sf{{}{\uparrow}{}}{\blacksquare}Nt(s(n))){{}{\downarrow}{}}Sf)/{\it CN}{\it s(n)}$}},\\
{\square}{\it CN}{\it s(n)}, {\blacksquare}{\forall}a{\forall}f((?{\blacksquare}({\langle\rangle}Na\backslash Sf)\backslash {[]^{-1}}{[]^{-1}}({\langle\rangle}Na\backslash Sf))/{\blacksquare}({\langle\rangle}Na\backslash Sf)), {\square}(({\langle\rangle}{\exists}aNa\backslash Sb)/{\exists}aNa), {\blacksquare}{\forall}f{\forall}a((({\langle\rangle}Na\backslash Sf){{}{\uparrow}{}}{\blacksquare}Nt(s(n))){{}{\downarrow}{}}({\blacksquare}({\langle\rangle}Na\backslash Sf){|}{\blacksquare}Nt(s(n))))]]\ \Rightarrow\ Sf
\end{array}
\using {/}L
\endprooftree
\justifies
\begin{array}{c}
{}[{\blacksquare}Nt(s(m))], {\square}(({\langle\rangle}{\exists}gNt(s(g))\backslash Sf)/{\square}({\langle\rangle}{\exists}gNt(s(g))\backslash Si)), {\blacksquare}(({\it PP}{\it to}/{\exists}aNa){\sqcap}{\forall}n(({\langle\rangle}Nn\backslash Si)/({\langle\rangle}Nn\backslash Sb))), [[{\square}(({\langle\rangle}{\exists}aNa\backslash Sb)/{\exists}aNa), \mbox{\fbox{${\forall}g({\forall}f((Sf{{}{\uparrow}{}}{\blacksquare}Nt(s(g))){{}{\downarrow}{}}Sf)/{\it CN}{\it s(g)})$}},\\ 
{\square}{\it CN}{\it s(n)}, {\blacksquare}{\forall}a{\forall}f((?{\blacksquare}({\langle\rangle}Na\backslash Sf)\backslash {[]^{-1}}{[]^{-1}}({\langle\rangle}Na\backslash Sf))/{\blacksquare}({\langle\rangle}Na\backslash Sf)), {\square}(({\langle\rangle}{\exists}aNa\backslash Sb)/{\exists}aNa), {\blacksquare}{\forall}f{\forall}a((({\langle\rangle}Na\backslash Sf){{}{\uparrow}{}}{\blacksquare}Nt(s(n))){{}{\downarrow}{}}({\blacksquare}({\langle\rangle}Na\backslash Sf){|}{\blacksquare}Nt(s(n))))]]\ \Rightarrow\ Sf
\end{array}
\using {\forall}L
\endprooftree
\justifies
\begin{array}{c}
{}[{\blacksquare}Nt(s(m))], {\square}(({\langle\rangle}{\exists}gNt(s(g))\backslash Sf)/{\square}({\langle\rangle}{\exists}gNt(s(g))\backslash Si)), {\blacksquare}(({\it PP}{\it to}/{\exists}aNa){\sqcap}{\forall}n(({\langle\rangle}Nn\backslash Si)/({\langle\rangle}Nn\backslash Sb))), [[{\square}(({\langle\rangle}{\exists}aNa\backslash Sb)/{\exists}aNa), \mbox{\fbox{${\blacksquare}{\forall}g({\forall}f((Sf{{}{\uparrow}{}}{\blacksquare}Nt(s(g))){{}{\downarrow}{}}Sf)/{\it CN}{\it s(g)})$}},\\
{\square}{\it CN}{\it s(n)}, {\blacksquare}{\forall}a{\forall}f((?{\blacksquare}({\langle\rangle}Na\backslash Sf)\backslash {[]^{-1}}{[]^{-1}}({\langle\rangle}Na\backslash Sf))/{\blacksquare}({\langle\rangle}Na\backslash Sf)), {\square}(({\langle\rangle}{\exists}aNa\backslash Sb)/{\exists}aNa), {\blacksquare}{\forall}f{\forall}a((({\langle\rangle}Na\backslash Sf){{}{\uparrow}{}}{\blacksquare}Nt(s(n))){{}{\downarrow}{}}({\blacksquare}({\langle\rangle}Na\backslash Sf){|}{\blacksquare}Nt(s(n))))]]\ \Rightarrow\ Sf
\end{array}
\using {\blacksquare}L
\endprooftree}
$$
\vspace{0.15in}
\noindent
This delivers semantics without existential commitment:
\disp{
$((\mbox{\v{}}{\it tries}\ \mbox{\^{}}\exists F[(\mbox{\v{}}{\it fish}\ {\it F})\wedge [((\mbox{\v{}}{\it catch}\ {\it F})\ {\it j})\wedge ((\mbox{\v{}}{\it eat}\ {\it F})\ {\it j})]])\ {\it j})$}

The next example involves an extensional transitive verb:
\disp{
(dwp((7-98))) $[{\bf john}]{+}{\bf finds}{+}{\bf a}{+}{\bf unicorn}: Sf$}
The sentence cannot be true unless a unicorn exists.
Montague treated extensional and intensional verbs
uniformly syntactically by raising the type of extensional verbs to
accommodate intensional verbs (\scare{raising to the worst case'}),
and then using meaning postulates to capture the existential commitment of the former.
Our type logical treatment allows assignment of the lower type to the existential which,
as well as being simpler, captures the existential commitment automatically. 
Lexical lookup yields:
\disp{
$[{\blacksquare}Nt(s(m)): {\it j}], {\square}(({\langle\rangle}{\exists}gNt(s(g))\backslash Sf)/{\exists}aNa): \mbox{\^{}}\lambda A\lambda B({\it Pres}\ ((\mbox{\v{}}{\it find}\ {\it A})\ {\it B})), {\blacksquare}{\forall}g({\forall}f((Sf{{}{\uparrow}{}}{\blacksquare}Nt(s(g))){{}{\downarrow}{}}$ $Sf)/{\it CN}{\it s(g)}): \lambda C\lambda D\exists E[({\it C}\ {\it E})\wedge ({\it D}\ {\it E})], {\square}{\it CN}{\it s(n)}: {\it unicorn}\ \Rightarrow\ Sf$}
This has the derivation:
$$
\vspace{0.15in}
{\tiny
\prooftree
\prooftree
\prooftree
\prooftree
\prooftree
\justifies
\mbox{\fbox{${\it CN}{\it s(n)}$}}\ \Rightarrow\ {\it CN}{\it s(n)}
\endprooftree
\justifies
\mbox{\fbox{${\square}{\it CN}{\it s(n)}$}}\ \Rightarrow\ {\it CN}{\it s(n)}
\using {\Box}L
\endprooftree
\prooftree
\prooftree
\prooftree
\prooftree
\prooftree
\prooftree
\prooftree
\prooftree
\justifies
\mbox{\fbox{$Nt(s(n))$}}\ \Rightarrow\ Nt(s(n))
\endprooftree
\justifies
\mbox{\fbox{${\blacksquare}Nt(s(n))$}}\ \Rightarrow\ Nt(s(n))
\using {\blacksquare}L
\endprooftree
\justifies
{\blacksquare}Nt(s(n))\ \Rightarrow\ \fbox{${\exists}aNa$}
\using {\exists}R
\endprooftree
\prooftree
\prooftree
\prooftree
\prooftree
\prooftree
\justifies
\mbox{\fbox{$Nt(s(m))$}}\ \Rightarrow\ Nt(s(m))
\endprooftree
\justifies
\mbox{\fbox{${\blacksquare}Nt(s(m))$}}\ \Rightarrow\ Nt(s(m))
\using {\blacksquare}L
\endprooftree
\justifies
{\blacksquare}Nt(s(m))\ \Rightarrow\ \fbox{${\exists}gNt(s(g))$}
\using {\exists}R
\endprooftree
\justifies
[{\blacksquare}Nt(s(m))]\ \Rightarrow\ \fbox{${\langle\rangle}{\exists}gNt(s(g))$}
\using {\langle\rangle}R
\endprooftree
\prooftree
\justifies
\mbox{\fbox{$Sf$}}\ \Rightarrow\ Sf
\endprooftree
\justifies
[{\blacksquare}Nt(s(m))], \mbox{\fbox{${\langle\rangle}{\exists}gNt(s(g))\backslash Sf$}}\ \Rightarrow\ Sf
\using {\backslash}L
\endprooftree
\justifies
[{\blacksquare}Nt(s(m))], \mbox{\fbox{$({\langle\rangle}{\exists}gNt(s(g))\backslash Sf)/{\exists}aNa$}}, {\blacksquare}Nt(s(n))\ \Rightarrow\ Sf
\using {/}L
\endprooftree
\justifies
[{\blacksquare}Nt(s(m))], \mbox{\fbox{${\square}(({\langle\rangle}{\exists}gNt(s(g))\backslash Sf)/{\exists}aNa)$}}, {\blacksquare}Nt(s(n))\ \Rightarrow\ Sf
\using {\Box}L
\endprooftree
\justifies
[{\blacksquare}Nt(s(m))], {\square}(({\langle\rangle}{\exists}gNt(s(g))\backslash Sf)/{\exists}aNa), {\tt 1}\ \Rightarrow\ Sf{{}{\uparrow}{}}{\blacksquare}Nt(s(n))
\using {\uparrow}R
\endprooftree
\prooftree
\justifies
\mbox{\fbox{$Sf$}}\ \Rightarrow\ Sf
\endprooftree
\justifies
[{\blacksquare}Nt(s(m))], {\square}(({\langle\rangle}{\exists}gNt(s(g))\backslash Sf)/{\exists}aNa), \mbox{\fbox{$(Sf{{}{\uparrow}{}}{\blacksquare}Nt(s(n))){{}{\downarrow}{}}Sf$}}\ \Rightarrow\ Sf
\using {\downarrow}L
\endprooftree
\justifies
[{\blacksquare}Nt(s(m))], {\square}(({\langle\rangle}{\exists}gNt(s(g))\backslash Sf)/{\exists}aNa), \mbox{\fbox{${\forall}f((Sf{{}{\uparrow}{}}{\blacksquare}Nt(s(n))){{}{\downarrow}{}}Sf)$}}\ \Rightarrow\ Sf
\using {\forall}L
\endprooftree
\justifies
[{\blacksquare}Nt(s(m))], {\square}(({\langle\rangle}{\exists}gNt(s(g))\backslash Sf)/{\exists}aNa), \mbox{\fbox{${\forall}f((Sf{{}{\uparrow}{}}{\blacksquare}Nt(s(n))){{}{\downarrow}{}}Sf)/{\it CN}{\it s(n)}$}}, {\square}{\it CN}{\it s(n)}\ \Rightarrow\ Sf
\using {/}L
\endprooftree
\justifies
[{\blacksquare}Nt(s(m))], {\square}(({\langle\rangle}{\exists}gNt(s(g))\backslash Sf)/{\exists}aNa), \mbox{\fbox{${\forall}g({\forall}f((Sf{{}{\uparrow}{}}{\blacksquare}Nt(s(g))){{}{\downarrow}{}}Sf)/{\it CN}{\it s(g)})$}}, {\square}{\it CN}{\it s(n)}\ \Rightarrow\ Sf
\using {\forall}L
\endprooftree
\justifies
[{\blacksquare}Nt(s(m))], {\square}(({\langle\rangle}{\exists}gNt(s(g))\backslash Sf)/{\exists}aNa), \mbox{\fbox{${\blacksquare}{\forall}g({\forall}f((Sf{{}{\uparrow}{}}{\blacksquare}Nt(s(g))){{}{\downarrow}{}}Sf)/{\it CN}{\it s(g)})$}}, {\square}{\it CN}{\it s(n)}\ \Rightarrow\ Sf
\using {\blacksquare}L
\endprooftree}
$$
\vspace{0.15in}
\noindent
It yields a semantics with existential commitment:
\disp{
$\exists C[(\mbox{\v{}}{\it unicorn}\ {\it C})\wedge ({\it Pres}\ ((\mbox{\v{}}{\it find}\ {\it C})\ {\it j}))]$}

DWP continue with a donkey sentence,
for which Montague grammar does not make the right prediction,
and nor our cover grammar:
\disp{
(dwp((7-105))) $[{\bf every}{+}{\bf man}{+}{\bf such}{+}{\bf that}{+}[{\bf he}]{+}{\bf loves}{+}{\bf a}{+}{\bf woman}]{+}{\bf loses}{+}{\bf her}: Sf$}
Lexical lookup yields:
\disp{
$[{\blacksquare}{\forall}g({\forall}f((Sf{{}{\uparrow}{}}Nt(s(g))){{}{\downarrow}{}}Sf)/{\it CN}{\it s(g)}): \lambda A\lambda B\forall C[({\it A}\ {\it C})\rightarrow ({\it B}\ {\it C})], {\square}{\it CN}{\it s(m)}: {\it man},\\{\blacksquare}{\forall}n(({\it CN}{\it n}\backslash {\it CN}{\it n})/$ $(Sf{|}{\blacksquare}Nt(n))): \lambda D\lambda E\lambda F[({\it E}\ {\it F})\wedge ({\it D}\ {\it F})], [{\blacksquare}{[]^{-1}}{\forall}g(({\blacksquare}Sg{|}{\blacksquare}Nt(s(m)))/\\({\langle\rangle}Nt(s(m))\backslash Sg)): \lambda G{\it G}],{\square}(({\langle\rangle}{\exists}gNt(s(g))\backslash Sf)/{\exists}aNa): \mbox{\^{}}\lambda H\lambda I({\it Pres}\ ((\mbox{\v{}}{\it love}\ {\it H})\ {\it I})),\\{\blacksquare}{\forall}g({\forall}f((Sf{{}{\uparrow}{}}{\blacksquare}Nt(s(g))){{}{\downarrow}{}}Sf)/{\it CN}{\it s(g)}): \lambda J\lambda K\exists L[({\it J}\ {\it L})\wedge ({\it K}\ {\it L})], {\square}{\it CN}{\it s(f)}: {\it woman}],\\{\square}(({\langle\rangle}{\exists}gNt(s(g))\backslash Sf)/{\exists}aNa): \mbox{\^{}}\lambda M\lambda N({\it Pres}\ ((\mbox{\v{}}{\it lose}\ {\it M})\ {\it N})),\\
 {\blacksquare}{\forall}g{\forall}a((({\langle\rangle}Na\backslash Sg){{}{\uparrow}{}}{\blacksquare}Nt(s(f))){{}{\downarrow}{}}({\blacksquare}({\langle\rangle}Na\backslash Sg){|}{\blacksquare}Nt(s(f)))): \lambda O{\it O}\ \Rightarrow\ Sf$}
There is a dominant reading in which {\it a woman\/},
which is the donkey anaphora antecedent,
is understood as quantified universally,
but Montague grammar obtains only a subordinate reading in which
{\it a woman\/} is quantified existentially at the matrix level.
There is the derivation:
\vspace{0.15in}
$$
\tiny
\prooftree
\prooftree
\prooftree
\prooftree
\prooftree
\prooftree
\prooftree
\prooftree
\justifies
\mbox{\fbox{$Nt(s(f))$}}\ \Rightarrow\ Nt(s(f))
\endprooftree
\justifies
\mbox{\fbox{${\blacksquare}Nt(s(f))$}}\ \Rightarrow\ Nt(s(f))
\using {\blacksquare}L
\endprooftree
\justifies
{\blacksquare}Nt(s(f))\ \Rightarrow\ \fbox{${\exists}aNa$}
\using {\exists}R
\endprooftree
\prooftree
\prooftree
\prooftree
\prooftree
\justifies
Nt(s(m))\ \Rightarrow\ Nt(s(m))
\endprooftree
\justifies
Nt(s(m))\ \Rightarrow\ \fbox{${\exists}gNt(s(g))$}
\using {\exists}R
\endprooftree
\justifies
[Nt(s(m))]\ \Rightarrow\ \fbox{${\langle\rangle}{\exists}gNt(s(g))$}
\using {\langle\rangle}R
\endprooftree
\prooftree
\justifies
\mbox{\fbox{$Sf$}}\ \Rightarrow\ Sf
\endprooftree
\justifies
[Nt(s(m))], \mbox{\fbox{${\langle\rangle}{\exists}gNt(s(g))\backslash Sf$}}\ \Rightarrow\ Sf
\using {\backslash}L
\endprooftree
\justifies
[Nt(s(m))], \mbox{\fbox{$({\langle\rangle}{\exists}gNt(s(g))\backslash Sf)/{\exists}aNa$}}, {\blacksquare}Nt(s(f))\ \Rightarrow\ Sf
\using {/}L
\endprooftree
\justifies
[Nt(s(m))], \mbox{\fbox{${\square}(({\langle\rangle}{\exists}gNt(s(g))\backslash Sf)/{\exists}aNa)$}}, {\blacksquare}Nt(s(f))\ \Rightarrow\ Sf
\using {\Box}L
\endprooftree
\justifies
{\langle\rangle}Nt(s(m)), {\square}(({\langle\rangle}{\exists}gNt(s(g))\backslash Sf)/{\exists}aNa), {\blacksquare}Nt(s(f))\ \Rightarrow\ Sf
\using {\langle\rangle}L
\endprooftree
\justifies
{\square}(({\langle\rangle}{\exists}gNt(s(g))\backslash Sf)/{\exists}aNa), {\blacksquare}Nt(s(f))\ \Rightarrow\ {\langle\rangle}Nt(s(m))\backslash Sf
\using {\backslash}R
\endprooftree
\justifies
\begin{array}{c}
{\square}(({\langle\rangle}{\exists}gNt(s(g))\backslash Sf)/{\exists}aNa), {\tt 1}\ \Rightarrow\ ({\langle\rangle}Nt(s(m))\backslash Sf){{}{\uparrow}{}}{\blacksquare}Nt(s(f))\\
\mbox{\footnotesize\textcircled{1}}
\end{array}
\using {\uparrow}R
\endprooftree
$$

$$
\tiny
\prooftree
\prooftree
\prooftree
\prooftree
\prooftree
\prooftree
\prooftree
\prooftree
\prooftree
\prooftree
\prooftree
\justifies
\mbox{\fbox{$Nt(s(f))$}}\ \Rightarrow\ Nt(s(f))
\endprooftree
\justifies
\mbox{\fbox{${\blacksquare}Nt(s(f))$}}\ \Rightarrow\ Nt(s(f))
\using {\blacksquare}L
\endprooftree
\justifies
{\blacksquare}Nt(s(f))\ \Rightarrow\ \fbox{${\exists}aNa$}
\using {\exists}R
\endprooftree
\prooftree
\prooftree
\prooftree
\prooftree
\justifies
Nt(s(m))\ \Rightarrow\ Nt(s(m))
\endprooftree
\justifies
Nt(s(m))\ \Rightarrow\ \fbox{${\exists}gNt(s(g))$}
\using {\exists}R
\endprooftree
\justifies
[Nt(s(m))]\ \Rightarrow\ \fbox{${\langle\rangle}{\exists}gNt(s(g))$}
\using {\langle\rangle}R
\endprooftree
\prooftree
\justifies
\mbox{\fbox{$Sf$}}\ \Rightarrow\ Sf
\endprooftree
\justifies
[Nt(s(m))], \mbox{\fbox{${\langle\rangle}{\exists}gNt(s(g))\backslash Sf$}}\ \Rightarrow\ Sf
\using {\backslash}L
\endprooftree
\justifies
[Nt(s(m))], \mbox{\fbox{$({\langle\rangle}{\exists}gNt(s(g))\backslash Sf)/{\exists}aNa$}}, {\blacksquare}Nt(s(f))\ \Rightarrow\ Sf
\using {/}L
\endprooftree
\justifies
[Nt(s(m))], \mbox{\fbox{${\square}(({\langle\rangle}{\exists}gNt(s(g))\backslash Sf)/{\exists}aNa)$}}, {\blacksquare}Nt(s(f))\ \Rightarrow\ Sf
\using {\Box}L
\endprooftree
\justifies
{\langle\rangle}Nt(s(m)), {\square}(({\langle\rangle}{\exists}gNt(s(g))\backslash Sf)/{\exists}aNa), {\blacksquare}Nt(s(f))\ \Rightarrow\ Sf
\using {\langle\rangle}L
\endprooftree
\justifies
{\square}(({\langle\rangle}{\exists}gNt(s(g))\backslash Sf)/{\exists}aNa), {\blacksquare}Nt(s(f))\ \Rightarrow\ {\langle\rangle}Nt(s(m))\backslash Sf
\using {\backslash}R
\endprooftree
\prooftree
\prooftree
\prooftree
\justifies
\mbox{\fbox{$Sf$}}\ \Rightarrow\ Sf
\endprooftree
\justifies
\mbox{\fbox{${\blacksquare}Sf$}}\ \Rightarrow\ Sf
\using {\blacksquare}L
\endprooftree
\justifies
\mbox{\fbox{${\blacksquare}Sf{|}{\blacksquare}Nt(s(m))$}}\ \Rightarrow\ {\blacksquare}Sf{|}{\blacksquare}Nt(s(m))
\using {|}R
\endprooftree
\justifies
\mbox{\fbox{$({\blacksquare}Sf{|}{\blacksquare}Nt(s(m)))/({\langle\rangle}Nt(s(m))\backslash Sf)$}}, {\square}(({\langle\rangle}{\exists}gNt(s(g))\backslash Sf)/{\exists}aNa), {\blacksquare}Nt(s(f))\ \Rightarrow\ Sf{|}{\blacksquare}Nt(s(m))
\using {/}L
\endprooftree
\justifies
\mbox{\fbox{${\forall}g(({\blacksquare}Sg{|}{\blacksquare}Nt(s(m)))/({\langle\rangle}Nt(s(m))\backslash Sg))$}}, {\square}(({\langle\rangle}{\exists}gNt(s(g))\backslash Sf)/{\exists}aNa), {\blacksquare}Nt(s(f))\ \Rightarrow\ Sf{|}{\blacksquare}Nt(s(m))
\using {\forall}L
\endprooftree
\justifies
[\mbox{\fbox{${[]^{-1}}{\forall}g(({\blacksquare}Sg{|}{\blacksquare}Nt(s(m)))/({\langle\rangle}Nt(s(m))\backslash Sg))$}}], {\square}(({\langle\rangle}{\exists}gNt(s(g))\backslash Sf)/{\exists}aNa), {\blacksquare}Nt(s(f))\ \Rightarrow\ Sf{|}{\blacksquare}Nt(s(m))
\using {[]^{-1}}L
\endprooftree
\justifies
\begin{array}{c}
[\mbox{\fbox{${\blacksquare}{[]^{-1}}{\forall}g(({\blacksquare}Sg{|}{\blacksquare}Nt(s(m)))/({\langle\rangle}Nt(s(m))\backslash Sg))$}}], {\square}(({\langle\rangle}{\exists}gNt(s(g))\backslash Sf)/{\exists}aNa), {\blacksquare}Nt(s(f))\ \Rightarrow\ Sf{|}{\blacksquare}Nt(s(m))\\
\mbox{\footnotesize\textcircled{2}}
\end{array}
\using {\blacksquare}L
\endprooftree
$$
$$\tiny
\rotatebox{-90}{
\prooftree
\prooftree
\prooftree
\prooftree
\mbox{\footnotesize\textcircled{1}}\tab
\prooftree
\prooftree
\prooftree
\prooftree
\justifies
\mbox{\fbox{$Nt(s(f))$}}\ \Rightarrow\ Nt(s(f))
\endprooftree
\justifies
\mbox{\fbox{${\blacksquare}Nt(s(f))$}}\ \Rightarrow\ Nt(s(f))
\using {\blacksquare}L
\endprooftree
\justifies
{\blacksquare}Nt(s(f))\ \Rightarrow\ {\blacksquare}Nt(s(f))
\using {\blacksquare}R
\endprooftree
\prooftree
\prooftree
\prooftree
\prooftree
\prooftree
\prooftree
\mbox{\footnotesize\textcircled{2}\tab\tab\tab\tab}
\prooftree
\prooftree
\prooftree
\justifies
\mbox{\fbox{${\it CN}{\it s(m)}$}}\ \Rightarrow\ {\it CN}{\it s(m)}
\endprooftree
\justifies
\mbox{\fbox{${\square}{\it CN}{\it s(m)}$}}\ \Rightarrow\ {\it CN}{\it s(m)}
\using {\Box}L
\endprooftree
\prooftree
\justifies
\mbox{\fbox{${\it CN}{\it s(m)}$}}\ \Rightarrow\ {\it CN}{\it s(m)}
\endprooftree
\justifies
{\square}{\it CN}{\it s(m)}, \mbox{\fbox{${\it CN}{\it s(m)}\backslash {\it CN}{\it s(m)}$}}\ \Rightarrow\ {\it CN}{\it s(m)}
\using {\backslash}L
\endprooftree
\justifies
{\square}{\it CN}{\it s(m)}, \mbox{\fbox{$({\it CN}{\it s(m)}\backslash {\it CN}{\it s(m)})/(Sf{|}{\blacksquare}Nt(s(m)))$}}, [{\blacksquare}{[]^{-1}}{\forall}g(({\blacksquare}Sg{|}{\blacksquare}Nt(s(m)))/({\langle\rangle}Nt(s(m))\backslash Sg))], {\square}(({\langle\rangle}{\exists}gNt(s(g))\backslash Sf)/{\exists}aNa), {\blacksquare}Nt(s(f))\ \Rightarrow\ {\it CN}{\it s(m)}
\using {/}L
\endprooftree
\justifies
{\square}{\it CN}{\it s(m)}, \mbox{\fbox{${\forall}n(({\it CN}{\it n}\backslash {\it CN}{\it n})/(Sf{|}{\blacksquare}Nt(n)))$}}, [{\blacksquare}{[]^{-1}}{\forall}g(({\blacksquare}Sg{|}{\blacksquare}Nt(s(m)))/({\langle\rangle}Nt(s(m))\backslash Sg))], {\square}(({\langle\rangle}{\exists}gNt(s(g))\backslash Sf)/{\exists}aNa), {\blacksquare}Nt(s(f))\ \Rightarrow\ {\it CN}{\it s(m)}
\using {\forall}L
\endprooftree
\justifies
{\square}{\it CN}{\it s(m)}, \mbox{\fbox{${\blacksquare}{\forall}n(({\it CN}{\it n}\backslash {\it CN}{\it n})/(Sf{|}{\blacksquare}Nt(n)))$}}, [{\blacksquare}{[]^{-1}}{\forall}g(({\blacksquare}Sg{|}{\blacksquare}Nt(s(m)))/({\langle\rangle}Nt(s(m))\backslash Sg))], {\square}(({\langle\rangle}{\exists}gNt(s(g))\backslash Sf)/{\exists}aNa), {\blacksquare}Nt(s(f))\ \Rightarrow\ {\it CN}{\it s(m)}
\using {\blacksquare}L
\endprooftree
\prooftree
\prooftree
\prooftree
\prooftree
\prooftree
\prooftree
\prooftree
\justifies
Nt(s(m))\ \Rightarrow\ Nt(s(m))
\endprooftree
\justifies
[Nt(s(m))]\ \Rightarrow\ \fbox{${\langle\rangle}Nt(s(m))$}
\using {\langle\rangle}R
\endprooftree
\prooftree
\justifies
\mbox{\fbox{$Sf$}}\ \Rightarrow\ Sf
\endprooftree
\justifies
[Nt(s(m))], \mbox{\fbox{${\langle\rangle}Nt(s(m))\backslash Sf$}}\ \Rightarrow\ Sf
\using {\backslash}L
\endprooftree
\justifies
[Nt(s(m))], \mbox{\fbox{${\blacksquare}({\langle\rangle}Nt(s(m))\backslash Sf)$}}\ \Rightarrow\ Sf
\using {\blacksquare}L
\endprooftree
\justifies
[{\tt 1}], {\blacksquare}({\langle\rangle}Nt(s(m))\backslash Sf)\ \Rightarrow\ Sf{{}{\uparrow}{}}Nt(s(m))
\using {\uparrow}R
\endprooftree
\prooftree
\justifies
\mbox{\fbox{$Sf$}}\ \Rightarrow\ Sf
\endprooftree
\justifies
[\mbox{\fbox{$(Sf{{}{\uparrow}{}}Nt(s(m))){{}{\downarrow}{}}Sf$}}], {\blacksquare}({\langle\rangle}Nt(s(m))\backslash Sf)\ \Rightarrow\ Sf
\using {\downarrow}L
\endprooftree
\justifies
[\mbox{\fbox{${\forall}f((Sf{{}{\uparrow}{}}Nt(s(m))){{}{\downarrow}{}}Sf)$}}], {\blacksquare}({\langle\rangle}Nt(s(m))\backslash Sf)\ \Rightarrow\ Sf
\using {\forall}L
\endprooftree
\justifies
[\mbox{\fbox{${\forall}f((Sf{{}{\uparrow}{}}Nt(s(m))){{}{\downarrow}{}}Sf)/{\it CN}{\it s(m)}$}}, {\square}{\it CN}{\it s(m)}, {\blacksquare}{\forall}n(({\it CN}{\it n}\backslash {\it CN}{\it n})/(Sf{|}{\blacksquare}Nt(n))), [{\blacksquare}{[]^{-1}}{\forall}g(({\blacksquare}Sg{|}{\blacksquare}Nt(s(m)))/({\langle\rangle}Nt(s(m))\backslash Sg))], {\square}(({\langle\rangle}{\exists}gNt(s(g))\backslash Sf)/{\exists}aNa), {\blacksquare}Nt(s(f))], {\blacksquare}({\langle\rangle}Nt(s(m))\backslash Sf)\ \Rightarrow\ Sf
\using {/}L
\endprooftree
\justifies
[\mbox{\fbox{${\forall}g({\forall}f((Sf{{}{\uparrow}{}}Nt(s(g))){{}{\downarrow}{}}Sf)/{\it CN}{\it s(g)})$}}, {\square}{\it CN}{\it s(m)}, {\blacksquare}{\forall}n(({\it CN}{\it n}\backslash {\it CN}{\it n})/(Sf{|}{\blacksquare}Nt(n))), [{\blacksquare}{[]^{-1}}{\forall}g(({\blacksquare}Sg{|}{\blacksquare}Nt(s(m)))/({\langle\rangle}Nt(s(m))\backslash Sg))], {\square}(({\langle\rangle}{\exists}gNt(s(g))\backslash Sf)/{\exists}aNa), {\blacksquare}Nt(s(f))], {\blacksquare}({\langle\rangle}Nt(s(m))\backslash Sf)\ \Rightarrow\ Sf
\using {\forall}L
\endprooftree
\justifies
[\mbox{\fbox{${\blacksquare}{\forall}g({\forall}f((Sf{{}{\uparrow}{}}Nt(s(g))){{}{\downarrow}{}}Sf)/{\it CN}{\it s(g)})$}}, {\square}{\it CN}{\it s(m)}, {\blacksquare}{\forall}n(({\it CN}{\it n}\backslash {\it CN}{\it n})/(Sf{|}{\blacksquare}Nt(n))), [{\blacksquare}{[]^{-1}}{\forall}g(({\blacksquare}Sg{|}{\blacksquare}Nt(s(m)))/({\langle\rangle}Nt(s(m))\backslash Sg))], {\square}(({\langle\rangle}{\exists}gNt(s(g))\backslash Sf)/{\exists}aNa), {\blacksquare}Nt(s(f))], {\blacksquare}({\langle\rangle}Nt(s(m))\backslash Sf)\ \Rightarrow\ Sf
\using {\blacksquare}L
\endprooftree
\justifies
[{\blacksquare}{\forall}g({\forall}f((Sf{{}{\uparrow}{}}Nt(s(g))){{}{\downarrow}{}}Sf)/{\it CN}{\it s(g)}), {\square}{\it CN}{\it s(m)}, {\blacksquare}{\forall}n(({\it CN}{\it n}\backslash {\it CN}{\it n})/(Sf{|}{\blacksquare}Nt(n))), [{\blacksquare}{[]^{-1}}{\forall}g(({\blacksquare}Sg{|}{\blacksquare}Nt(s(m)))/({\langle\rangle}Nt(s(m))\backslash Sg))], {\square}(({\langle\rangle}{\exists}gNt(s(g))\backslash Sf)/{\exists}aNa), {\blacksquare}Nt(s(f))], \mbox{\fbox{${\blacksquare}({\langle\rangle}Nt(s(m))\backslash Sf){|}{\blacksquare}Nt(s(f))$}}\ \Rightarrow\ Sf
\using {|}L
\endprooftree
\justifies
[{\blacksquare}{\forall}g({\forall}f((Sf{{}{\uparrow}{}}Nt(s(g))){{}{\downarrow}{}}Sf)/{\it CN}{\it s(g)}), {\square}{\it CN}{\it s(m)}, {\blacksquare}{\forall}n(({\it CN}{\it n}\backslash {\it CN}{\it n})/(Sf{|}{\blacksquare}Nt(n))), [{\blacksquare}{[]^{-1}}{\forall}g(({\blacksquare}Sg{|}{\blacksquare}Nt(s(m)))/({\langle\rangle}Nt(s(m))\backslash Sg))], {\square}(({\langle\rangle}{\exists}gNt(s(g))\backslash Sf)/{\exists}aNa), {\blacksquare}Nt(s(f))], {\square}(({\langle\rangle}{\exists}gNt(s(g))\backslash Sf)/{\exists}aNa), \mbox{\fbox{$(({\langle\rangle}Nt(s(m))\backslash Sf){{}{\uparrow}{}}{\blacksquare}Nt(s(f))){{}{\downarrow}{}}({\blacksquare}({\langle\rangle}Nt(s(m))\backslash Sf){|}{\blacksquare}Nt(s(f)))$}}\ \Rightarrow\ Sf
\using {\downarrow}L
\endprooftree
\justifies
[{\blacksquare}{\forall}g({\forall}f((Sf{{}{\uparrow}{}}Nt(s(g))){{}{\downarrow}{}}Sf)/{\it CN}{\it s(g)}), {\square}{\it CN}{\it s(m)}, {\blacksquare}{\forall}n(({\it CN}{\it n}\backslash {\it CN}{\it n})/(Sf{|}{\blacksquare}Nt(n))), [{\blacksquare}{[]^{-1}}{\forall}g(({\blacksquare}Sg{|}{\blacksquare}Nt(s(m)))/({\langle\rangle}Nt(s(m))\backslash Sg))], {\square}(({\langle\rangle}{\exists}gNt(s(g))\backslash Sf)/{\exists}aNa), {\blacksquare}Nt(s(f))], {\square}(({\langle\rangle}{\exists}gNt(s(g))\backslash Sf)/{\exists}aNa), \mbox{\fbox{${\forall}a((({\langle\rangle}Na\backslash Sf){{}{\uparrow}{}}{\blacksquare}Nt(s(f))){{}{\downarrow}{}}({\blacksquare}({\langle\rangle}Na\backslash Sf){|}{\blacksquare}Nt(s(f))))$}}\ \Rightarrow\ Sf
\using {\forall}L
\endprooftree
\justifies
[{\blacksquare}{\forall}g({\forall}f((Sf{{}{\uparrow}{}}Nt(s(g))){{}{\downarrow}{}}Sf)/{\it CN}{\it s(g)}), {\square}{\it CN}{\it s(m)}, {\blacksquare}{\forall}n(({\it CN}{\it n}\backslash {\it CN}{\it n})/(Sf{|}{\blacksquare}Nt(n))), [{\blacksquare}{[]^{-1}}{\forall}g(({\blacksquare}Sg{|}{\blacksquare}Nt(s(m)))/({\langle\rangle}Nt(s(m))\backslash Sg))], {\square}(({\langle\rangle}{\exists}gNt(s(g))\backslash Sf)/{\exists}aNa), {\blacksquare}Nt(s(f))], {\square}(({\langle\rangle}{\exists}gNt(s(g))\backslash Sf)/{\exists}aNa), \mbox{\fbox{${\forall}g{\forall}a((({\langle\rangle}Na\backslash Sg){{}{\uparrow}{}}{\blacksquare}Nt(s(f))){{}{\downarrow}{}}({\blacksquare}({\langle\rangle}Na\backslash Sg){|}{\blacksquare}Nt(s(f))))$}}\ \Rightarrow\ Sf
\using {\forall}L
\endprooftree
\justifies
\begin{array}{c}
{}[{\blacksquare}{\forall}g({\forall}f((Sf{{}{\uparrow}{}}Nt(s(g))){{}{\downarrow}{}}Sf)/{\it CN}{\it s(g)}), {\square}{\it CN}{\it s(m)}, {\blacksquare}{\forall}n(({\it CN}{\it n}\backslash {\it CN}{\it n})/(Sf{|}{\blacksquare}Nt(n))), [{\blacksquare}{[]^{-1}}{\forall}g(({\blacksquare}Sg{|}{\blacksquare}Nt(s(m)))/({\langle\rangle}Nt(s(m))\backslash Sg))], {\square}(({\langle\rangle}{\exists}gNt(s(g))\backslash Sf)/{\exists}aNa), {\blacksquare}Nt(s(f))], {\square}(({\langle\rangle}{\exists}gNt(s(g))\backslash Sf)/{\exists}aNa), \mbox{\fbox{${\blacksquare}{\forall}g{\forall}a((({\langle\rangle}Na\backslash Sg){{}{\uparrow}{}}{\blacksquare}Nt(s(f))){{}{\downarrow}{}}({\blacksquare}({\langle\rangle}Na\backslash Sg){|}{\blacksquare}Nt(s(f))))$}}\ \Rightarrow\ Sf\\
\mbox{\footnotesize\textcircled{3}}
\end{array}
\using {\blacksquare}L
\endprooftree}
$$
$$\tiny
\rotatebox{-90}{
\prooftree
\prooftree
\prooftree
\prooftree
\prooftree
\justifies
\mbox{\fbox{${\it CN}{\it s(f)}$}}\ \Rightarrow\ {\it CN}{\it s(f)}
\endprooftree
\justifies
\mbox{\fbox{${\square}{\it CN}{\it s(f)}$}}\ \Rightarrow\ {\it CN}{\it s(f)}
\using {\Box}L
\endprooftree
\prooftree
\prooftree
\prooftree
\mbox{\footnotesize\textcircled{3}}
\justifies
\begin{array}{c}
{}[{\blacksquare}{\forall}g({\forall}f((Sf{{}{\uparrow}{}}Nt(s(g))){{}{\downarrow}{}}Sf)/{\it CN}{\it s(g)}), {\square}{\it CN}{\it s(m)}, {\blacksquare}{\forall}n(({\it CN}{\it n}\backslash {\it CN}{\it n})/(Sf{|}{\blacksquare}Nt(n))), [{\blacksquare}{[]^{-1}}{\forall}g(({\blacksquare}Sg{|}{\blacksquare}Nt(s(m)))/({\langle\rangle}Nt(s(m))\backslash Sg))], {\square}(({\langle\rangle}{\exists}gNt(s(g))\backslash Sf)/{\exists}aNa), {\tt 1}],\\
{\square}(({\langle\rangle}{\exists}gNt(s(g))\backslash Sf)/{\exists}aNa), {\blacksquare}{\forall}g{\forall}a((({\langle\rangle}Na\backslash Sg){{}{\uparrow}{}}{\blacksquare}Nt(s(f))){{}{\downarrow}{}}({\blacksquare}({\langle\rangle}Na\backslash Sg){|}{\blacksquare}Nt(s(f))))\ \Rightarrow\ Sf{{}{\uparrow}{}}{\blacksquare}Nt(s(f))
\end{array}
\using {\uparrow}R
\endprooftree
\prooftree
\justifies
\mbox{\fbox{$Sf$}}\ \Rightarrow\ Sf
\endprooftree
\justifies
\begin{array}{c}
[{\blacksquare}{\forall}g({\forall}f((Sf{{}{\uparrow}{}}Nt(s(g))){{}{\downarrow}{}}Sf)/{\it CN}{\it s(g)}), {\square}{\it CN}{\it s(m)}, {\blacksquare}{\forall}n(({\it CN}{\it n}\backslash {\it CN}{\it n})/(Sf{|}{\blacksquare}Nt(n))), [{\blacksquare}{[]^{-1}}{\forall}g(({\blacksquare}Sg{|}{\blacksquare}Nt(s(m)))/({\langle\rangle}Nt(s(m))\backslash Sg))], {\square}(({\langle\rangle}{\exists}gNt(s(g))\backslash Sf)/{\exists}aNa),\\
\mbox{\fbox{$(Sf{{}{\uparrow}{}}{\blacksquare}Nt(s(f))){{}{\downarrow}{}}Sf$}}], {\square}(({\langle\rangle}{\exists}gNt(s(g))\backslash Sf)/{\exists}aNa), {\blacksquare}{\forall}g{\forall}a((({\langle\rangle}Na\backslash Sg){{}{\uparrow}{}}{\blacksquare}Nt(s(f))){{}{\downarrow}{}}({\blacksquare}({\langle\rangle}Na\backslash Sg){|}{\blacksquare}Nt(s(f))))\ \Rightarrow\ Sf
\end{array}
\using {\downarrow}L
\endprooftree
\justifies
\begin{array}{c}
[{\blacksquare}{\forall}g({\forall}f((Sf{{}{\uparrow}{}}Nt(s(g))){{}{\downarrow}{}}Sf)/{\it CN}{\it s(g)}), {\square}{\it CN}{\it s(m)}, {\blacksquare}{\forall}n(({\it CN}{\it n}\backslash {\it CN}{\it n})/(Sf{|}{\blacksquare}Nt(n))), [{\blacksquare}{[]^{-1}}{\forall}g(({\blacksquare}Sg{|}{\blacksquare}Nt(s(m)))/({\langle\rangle}Nt(s(m))\backslash Sg))], {\square}(({\langle\rangle}{\exists}gNt(s(g))\backslash Sf)/{\exists}aNa),\\
\mbox{\fbox{${\forall}f((Sf{{}{\uparrow}{}}{\blacksquare}Nt(s(f))){{}{\downarrow}{}}Sf)$}}], {\square}(({\langle\rangle}{\exists}gNt(s(g))\backslash Sf)/{\exists}aNa), {\blacksquare}{\forall}g{\forall}a((({\langle\rangle}Na\backslash Sg){{}{\uparrow}{}}{\blacksquare}Nt(s(f))){{}{\downarrow}{}}({\blacksquare}({\langle\rangle}Na\backslash Sg){|}{\blacksquare}Nt(s(f))))\ \Rightarrow\ Sf
\end{array}
\using {\forall}L
\endprooftree
\justifies
\begin{array}{c}
{}[{\blacksquare}{\forall}g({\forall}f((Sf{{}{\uparrow}{}}Nt(s(g))){{}{\downarrow}{}}Sf)/{\it CN}{\it s(g)}), {\square}{\it CN}{\it s(m)}, {\blacksquare}{\forall}n(({\it CN}{\it n}\backslash {\it CN}{\it n})/(Sf{|}{\blacksquare}Nt(n))), [{\blacksquare}{[]^{-1}}{\forall}g(({\blacksquare}Sg{|}{\blacksquare}Nt(s(m)))/({\langle\rangle}Nt(s(m))\backslash Sg))], {\square}(({\langle\rangle}{\exists}gNt(s(g))\backslash Sf)/{\exists}aNa),\\
\mbox{\fbox{${\forall}f((Sf{{}{\uparrow}{}}{\blacksquare}Nt(s(f))){{}{\downarrow}{}}Sf)/{\it CN}{\it s(f)}$}}, {\square}{\it CN}{\it s(f)}], {\square}(({\langle\rangle}{\exists}gNt(s(g))\backslash Sf)/{\exists}aNa), {\blacksquare}{\forall}g{\forall}a((({\langle\rangle}Na\backslash Sg){{}{\uparrow}{}}{\blacksquare}Nt(s(f))){{}{\downarrow}{}}({\blacksquare}({\langle\rangle}Na\backslash Sg){|}{\blacksquare}Nt(s(f))))\ \Rightarrow\ Sf
\end{array}
\using {/}L
\endprooftree
\justifies
\begin{array}{c}
{}[{\blacksquare}{\forall}g({\forall}f((Sf{{}{\uparrow}{}}Nt(s(g))){{}{\downarrow}{}}Sf)/{\it CN}{\it s(g)}), {\square}{\it CN}{\it s(m)}, {\blacksquare}{\forall}n(({\it CN}{\it n}\backslash {\it CN}{\it n})/(Sf{|}{\blacksquare}Nt(n))), [{\blacksquare}{[]^{-1}}{\forall}g(({\blacksquare}Sg{|}{\blacksquare}Nt(s(m)))/({\langle\rangle}Nt(s(m))\backslash Sg))], {\square}(({\langle\rangle}{\exists}gNt(s(g))\backslash Sf)/{\exists}aNa),\\
\mbox{\fbox{${\forall}g({\forall}f((Sf{{}{\uparrow}{}}{\blacksquare}Nt(s(g))){{}{\downarrow}{}}Sf)/{\it CN}{\it s(g)})$}}, {\square}{\it CN}{\it s(f)}], {\square}(({\langle\rangle}{\exists}gNt(s(g))\backslash Sf)/{\exists}aNa), {\blacksquare}{\forall}g{\forall}a((({\langle\rangle}Na\backslash Sg){{}{\uparrow}{}}{\blacksquare}Nt(s(f))){{}{\downarrow}{}}({\blacksquare}({\langle\rangle}Na\backslash Sg){|}{\blacksquare}Nt(s(f))))\ \Rightarrow\ Sf
\end{array}
\using {\forall}L
\endprooftree
\justifies
\begin{array}{c}
{}[{\blacksquare}{\forall}g({\forall}f((Sf{{}{\uparrow}{}}Nt(s(g))){{}{\downarrow}{}}Sf)/{\it CN}{\it s(g)}), {\square}{\it CN}{\it s(m)}, {\blacksquare}{\forall}n(({\it CN}{\it n}\backslash {\it CN}{\it n})/(Sf{|}{\blacksquare}Nt(n))), [{\blacksquare}{[]^{-1}}{\forall}g(({\blacksquare}Sg{|}{\blacksquare}Nt(s(m)))/({\langle\rangle}Nt(s(m))\backslash Sg))], {\square}(({\langle\rangle}{\exists}gNt(s(g))\backslash Sf)/{\exists}aNa),\\
\mbox{\fbox{${\blacksquare}{\forall}g({\forall}f((Sf{{}{\uparrow}{}}{\blacksquare}Nt(s(g))){{}{\downarrow}{}}Sf)/{\it CN}{\it s(g)})$}}, {\square}{\it CN}{\it s(f)}], {\square}(({\langle\rangle}{\exists}gNt(s(g))\backslash Sf)/{\exists}aNa), {\blacksquare}{\forall}g{\forall}a((({\langle\rangle}Na\backslash Sg){{}{\uparrow}{}}{\blacksquare}Nt(s(f))){{}{\downarrow}{}}({\blacksquare}({\langle\rangle}Na\backslash Sg){|}{\blacksquare}Nt(s(f))))\ \Rightarrow\ Sf
\end{array}
\using {\blacksquare}L
\endprooftree}
$$
\vspace{0.15in}
\noindent
This delivers semantics:
\disp{
$\exists C[(\mbox{\v{}}{\it woman}\ {\it C})\wedge \forall G[[(\mbox{\v{}}{\it man}\ {\it G})\wedge ({\it Pres}\ ((\mbox{\v{}}{\it love}\ {\it C})\ {\it G}))]\rightarrow ({\it Pres}\ ((\mbox{\v{}}{\it lose}\ {\it C})\ {\it G}))]]$}

The assignment of lowest type in type logical grammar also means that existential
commitment of a preposition comes without the need for devices such as the meaning
postulates of Montague grammar:
\disp{
(dwp((7-110))) $[{\bf john}]{+}{\bf walks}{+}{\bf in}{+}{\bf a}{+}{\bf park}: Sf$}
Lexical lookup for this example yields the semantically annotated sequent:
\disp{
$[{\blacksquare}Nt(s(m)): {\it j}], {\square}({\langle\rangle}{\exists}gNt(s(g))\backslash Sf): \mbox{\^{}}\lambda A({\it Pres}\ (\mbox{\v{}}{\it walk}\ {\it A})), {\square}({\forall}a{\forall}f(({\langle\rangle}Na\backslash Sf)\backslash ({\langle\rangle}Na\backslash Sf))/\\{\exists}aNa): \mbox{\^{}}\lambda B\lambda C\lambda D((\mbox{\v{}}{\it in}\ {\it B})\ ({\it C}\ {\it D})), {\blacksquare}{\forall}g({\forall}f((Sf{{}{\uparrow}{}}{\blacksquare}Nt(s(g))){{}{\downarrow}{}}Sf)/{\it CN}{\it s(g)}):\\\lambda E\lambda F\exists G[({\it E}\ {\it G})\wedge ({\it F}\ {\it G})], {\square}{\it CN}{\it s(n)}: {\it park}\ \Rightarrow\ Sf$}
This sequent has the proof:
\vspace{0.15in}
$$
\tiny
\prooftree
\prooftree
\prooftree
\prooftree
\prooftree
\justifies
\mbox{\fbox{${\it CN}{\it s(n)}$}}\ \Rightarrow\ {\it CN}{\it s(n)}
\endprooftree
\justifies
\mbox{\fbox{${\square}{\it CN}{\it s(n)}$}}\ \Rightarrow\ {\it CN}{\it s(n)}
\using {\Box}L
\endprooftree
\prooftree
\prooftree
\prooftree
\prooftree
\prooftree
\prooftree
\prooftree
\prooftree
\justifies
\mbox{\fbox{$Nt(s(n))$}}\ \Rightarrow\ Nt(s(n))
\endprooftree
\justifies
\mbox{\fbox{${\blacksquare}Nt(s(n))$}}\ \Rightarrow\ Nt(s(n))
\using {\blacksquare}L
\endprooftree
\justifies
{\blacksquare}Nt(s(n))\ \Rightarrow\ \fbox{${\exists}aNa$}
\using {\exists}R
\endprooftree
\prooftree
\prooftree
\prooftree
\prooftree
\prooftree
\prooftree
\prooftree
\prooftree
\prooftree
\prooftree
\justifies
Nt(s(m))\ \Rightarrow\ Nt(s(m))
\endprooftree
\justifies
Nt(s(m))\ \Rightarrow\ \fbox{${\exists}gNt(s(g))$}
\using {\exists}R
\endprooftree
\justifies
[Nt(s(m))]\ \Rightarrow\ \fbox{${\langle\rangle}{\exists}gNt(s(g))$}
\using {\langle\rangle}R
\endprooftree
\prooftree
\justifies
\mbox{\fbox{$Sf$}}\ \Rightarrow\ Sf
\endprooftree
\justifies
[Nt(s(m))], \mbox{\fbox{${\langle\rangle}{\exists}gNt(s(g))\backslash Sf$}}\ \Rightarrow\ Sf
\using {\backslash}L
\endprooftree
\justifies
[Nt(s(m))], \mbox{\fbox{${\square}({\langle\rangle}{\exists}gNt(s(g))\backslash Sf)$}}\ \Rightarrow\ Sf
\using {\Box}L
\endprooftree
\justifies
{\langle\rangle}Nt(s(m)), {\square}({\langle\rangle}{\exists}gNt(s(g))\backslash Sf)\ \Rightarrow\ Sf
\using {\langle\rangle}L
\endprooftree
\justifies
{\square}({\langle\rangle}{\exists}gNt(s(g))\backslash Sf)\ \Rightarrow\ {\langle\rangle}Nt(s(m))\backslash Sf
\using {\backslash}R
\endprooftree
\prooftree
\prooftree
\prooftree
\prooftree
\justifies
\mbox{\fbox{$Nt(s(m))$}}\ \Rightarrow\ Nt(s(m))
\endprooftree
\justifies
\mbox{\fbox{${\blacksquare}Nt(s(m))$}}\ \Rightarrow\ Nt(s(m))
\using {\blacksquare}L
\endprooftree
\justifies
[{\blacksquare}Nt(s(m))]\ \Rightarrow\ \fbox{${\langle\rangle}Nt(s(m))$}
\using {\langle\rangle}R
\endprooftree
\prooftree
\justifies
\mbox{\fbox{$Sf$}}\ \Rightarrow\ Sf
\endprooftree
\justifies
[{\blacksquare}Nt(s(m))], \mbox{\fbox{${\langle\rangle}Nt(s(m))\backslash Sf$}}\ \Rightarrow\ Sf
\using {\backslash}L
\endprooftree
\justifies
[{\blacksquare}Nt(s(m))], {\square}({\langle\rangle}{\exists}gNt(s(g))\backslash Sf), \mbox{\fbox{$({\langle\rangle}Nt(s(m))\backslash Sf)\backslash ({\langle\rangle}Nt(s(m))\backslash Sf)$}}\ \Rightarrow\ Sf
\using {\backslash}L
\endprooftree
\justifies
[{\blacksquare}Nt(s(m))], {\square}({\langle\rangle}{\exists}gNt(s(g))\backslash Sf), \mbox{\fbox{${\forall}f(({\langle\rangle}Nt(s(m))\backslash Sf)\backslash ({\langle\rangle}Nt(s(m))\backslash Sf))$}}\ \Rightarrow\ Sf
\using {\forall}L
\endprooftree
\justifies
[{\blacksquare}Nt(s(m))], {\square}({\langle\rangle}{\exists}gNt(s(g))\backslash Sf), \mbox{\fbox{${\forall}a{\forall}f(({\langle\rangle}Na\backslash Sf)\backslash ({\langle\rangle}Na\backslash Sf))$}}\ \Rightarrow\ Sf
\using {\forall}L
\endprooftree
\justifies
[{\blacksquare}Nt(s(m))], {\square}({\langle\rangle}{\exists}gNt(s(g))\backslash Sf), \mbox{\fbox{${\forall}a{\forall}f(({\langle\rangle}Na\backslash Sf)\backslash ({\langle\rangle}Na\backslash Sf))/{\exists}aNa$}}, {\blacksquare}Nt(s(n))\ \Rightarrow\ Sf
\using {/}L
\endprooftree
\justifies
[{\blacksquare}Nt(s(m))], {\square}({\langle\rangle}{\exists}gNt(s(g))\backslash Sf), \mbox{\fbox{${\square}({\forall}a{\forall}f(({\langle\rangle}Na\backslash Sf)\backslash ({\langle\rangle}Na\backslash Sf))/{\exists}aNa)$}}, {\blacksquare}Nt(s(n))\ \Rightarrow\ Sf
\using {\Box}L
\endprooftree
\justifies
[{\blacksquare}Nt(s(m))], {\square}({\langle\rangle}{\exists}gNt(s(g))\backslash Sf), {\square}({\forall}a{\forall}f(({\langle\rangle}Na\backslash Sf)\backslash ({\langle\rangle}Na\backslash Sf))/{\exists}aNa), {\tt 1}\ \Rightarrow\ Sf{{}{\uparrow}{}}{\blacksquare}Nt(s(n))
\using {\uparrow}R
\endprooftree
\prooftree
\justifies
\mbox{\fbox{$Sf$}}\ \Rightarrow\ Sf
\endprooftree
\justifies
[{\blacksquare}Nt(s(m))], {\square}({\langle\rangle}{\exists}gNt(s(g))\backslash Sf), {\square}({\forall}a{\forall}f(({\langle\rangle}Na\backslash Sf)\backslash ({\langle\rangle}Na\backslash Sf))/{\exists}aNa), \mbox{\fbox{$(Sf{{}{\uparrow}{}}{\blacksquare}Nt(s(n))){{}{\downarrow}{}}Sf$}}\ \Rightarrow\ Sf
\using {\downarrow}L
\endprooftree
\justifies
[{\blacksquare}Nt(s(m))], {\square}({\langle\rangle}{\exists}gNt(s(g))\backslash Sf), {\square}({\forall}a{\forall}f(({\langle\rangle}Na\backslash Sf)\backslash ({\langle\rangle}Na\backslash Sf))/{\exists}aNa), \mbox{\fbox{${\forall}f((Sf{{}{\uparrow}{}}{\blacksquare}Nt(s(n))){{}{\downarrow}{}}Sf)$}}\ \Rightarrow\ Sf
\using {\forall}L
\endprooftree
\justifies
[{\blacksquare}Nt(s(m))], {\square}({\langle\rangle}{\exists}gNt(s(g))\backslash Sf), {\square}({\forall}a{\forall}f(({\langle\rangle}Na\backslash Sf)\backslash ({\langle\rangle}Na\backslash Sf))/{\exists}aNa), \mbox{\fbox{${\forall}f((Sf{{}{\uparrow}{}}{\blacksquare}Nt(s(n))){{}{\downarrow}{}}Sf)/{\it CN}{\it s(n)}$}}, {\square}{\it CN}{\it s(n)}\ \Rightarrow\ Sf
\using {/}L
\endprooftree
\justifies
[{\blacksquare}Nt(s(m))], {\square}({\langle\rangle}{\exists}gNt(s(g))\backslash Sf), {\square}({\forall}a{\forall}f(({\langle\rangle}Na\backslash Sf)\backslash ({\langle\rangle}Na\backslash Sf))/{\exists}aNa), \mbox{\fbox{${\forall}g({\forall}f((Sf{{}{\uparrow}{}}{\blacksquare}Nt(s(g))){{}{\downarrow}{}}Sf)/{\it CN}{\it s(g)})$}}, {\square}{\it CN}{\it s(n)}\ \Rightarrow\ Sf
\using {\forall}L
\endprooftree
\justifies
[{\blacksquare}Nt(s(m))], {\square}({\langle\rangle}{\exists}gNt(s(g))\backslash Sf), {\square}({\forall}a{\forall}f(({\langle\rangle}Na\backslash Sf)\backslash ({\langle\rangle}Na\backslash Sf))/{\exists}aNa), \mbox{\fbox{${\blacksquare}{\forall}g({\forall}f((Sf{{}{\uparrow}{}}{\blacksquare}Nt(s(g))){{}{\downarrow}{}}Sf)/{\it CN}{\it s(g)})$}}, {\square}{\it CN}{\it s(n)}\ \Rightarrow\ Sf
\using {\blacksquare}L
\endprooftree
$$
\vspace{0.15in}
\noindent
It delivers the semantics (with existential comittment):
\disp{
$\exists C[(\mbox{\v{}}{\it park}\ {\it C})\wedge ((\mbox{\v{}}{\it in}\ {\it C})\ ({\it Pres}\ (\mbox{\v{}}{\it walk}\ {\it j})))]$}

Finally DWP analyse the ambiguous example:
\disp{
(dwp((7-116, 118))) $[{\bf every}{+}{\bf man}]{+}{\bf doesnt}{+}{\bf walk}: Sf$}
This has a dominant reading in which the universal has narrow scope
with respect to the negation, and a subordinate reading in which
the universal has wide scope with respect to the negation.
Lexical lookup yields:
\disp{
$[{\blacksquare}{\forall}g({\forall}f((Sf{{}{\uparrow}{}}Nt(s(g))){{}{\downarrow}{}}Sf)/{\it CN}{\it s(g)}): \lambda A\lambda B\forall C[({\it A}\ {\it C})\rightarrow ({\it B}\ {\it C})], {\square}{\it CN}{\it s(m)}: {\it man}],\\
 {\blacksquare}{\forall}g{\forall}a((Sg{{}{\uparrow}{}}(({\langle\rangle}Na\backslash Sf)/({\langle\rangle}Na\backslash Sb))){{}{\downarrow}{}}Sg): \lambda D\neg ({\it D}\ \lambda E\lambda F({\it E}\ {\it F})), {\square}({\langle\rangle}{\exists}aNa\backslash Sb): \mbox{\^{}}\lambda G(\mbox{\v{}}{\it walk}\ {\it G})\\\Rightarrow\ Sf$}
This has a first derivation:
\vspace{0.15in}
$$
\tiny
\prooftree
\prooftree
\prooftree
\prooftree
\prooftree
\justifies
\mbox{\fbox{${\it CN}{\it s(m)}$}}\ \Rightarrow\ {\it CN}{\it s(m)}
\endprooftree
\justifies
\mbox{\fbox{${\square}{\it CN}{\it s(m)}$}}\ \Rightarrow\ {\it CN}{\it s(m)}
\using {\Box}L
\endprooftree
\prooftree
\prooftree
\prooftree
\prooftree
\prooftree
\prooftree
\prooftree
\prooftree
\prooftree
\prooftree
\prooftree
\prooftree
\prooftree
\prooftree
\prooftree
\prooftree
\justifies
Nt(s(m))\ \Rightarrow\ Nt(s(m))
\endprooftree
\justifies
Nt(s(m))\ \Rightarrow\ \fbox{${\exists}aNa$}
\using {\exists}R
\endprooftree
\justifies
[Nt(s(m))]\ \Rightarrow\ \fbox{${\langle\rangle}{\exists}aNa$}
\using {\langle\rangle}R
\endprooftree
\prooftree
\justifies
\mbox{\fbox{$Sb$}}\ \Rightarrow\ Sb
\endprooftree
\justifies
[Nt(s(m))], \mbox{\fbox{${\langle\rangle}{\exists}aNa\backslash Sb$}}\ \Rightarrow\ Sb
\using {\backslash}L
\endprooftree
\justifies
[Nt(s(m))], \mbox{\fbox{${\square}({\langle\rangle}{\exists}aNa\backslash Sb)$}}\ \Rightarrow\ Sb
\using {\Box}L
\endprooftree
\justifies
{\langle\rangle}Nt(s(m)), {\square}({\langle\rangle}{\exists}aNa\backslash Sb)\ \Rightarrow\ Sb
\using {\langle\rangle}L
\endprooftree
\justifies
{\square}({\langle\rangle}{\exists}aNa\backslash Sb)\ \Rightarrow\ {\langle\rangle}Nt(s(m))\backslash Sb
\using {\backslash}R
\endprooftree
\prooftree
\prooftree
\prooftree
\justifies
Nt(s(m))\ \Rightarrow\ Nt(s(m))
\endprooftree
\justifies
[Nt(s(m))]\ \Rightarrow\ \fbox{${\langle\rangle}Nt(s(m))$}
\using {\langle\rangle}R
\endprooftree
\prooftree
\justifies
\mbox{\fbox{$Sf$}}\ \Rightarrow\ Sf
\endprooftree
\justifies
[Nt(s(m))], \mbox{\fbox{${\langle\rangle}Nt(s(m))\backslash Sf$}}\ \Rightarrow\ Sf
\using {\backslash}L
\endprooftree
\justifies
[Nt(s(m))], \mbox{\fbox{$({\langle\rangle}Nt(s(m))\backslash Sf)/({\langle\rangle}Nt(s(m))\backslash Sb)$}}, {\square}({\langle\rangle}{\exists}aNa\backslash Sb)\ \Rightarrow\ Sf
\using {/}L
\endprooftree
\justifies
[Nt(s(m))], {\tt 1}, {\square}({\langle\rangle}{\exists}aNa\backslash Sb)\ \Rightarrow\ Sf{{}{\uparrow}{}}(({\langle\rangle}Nt(s(m))\backslash Sf)/({\langle\rangle}Nt(s(m))\backslash Sb))
\using {\uparrow}R
\endprooftree
\prooftree
\justifies
\mbox{\fbox{$Sf$}}\ \Rightarrow\ Sf
\endprooftree
\justifies
[Nt(s(m))], \mbox{\fbox{$(Sf{{}{\uparrow}{}}(({\langle\rangle}Nt(s(m))\backslash Sf)/({\langle\rangle}Nt(s(m))\backslash Sb))){{}{\downarrow}{}}Sf$}}, {\square}({\langle\rangle}{\exists}aNa\backslash Sb)\ \Rightarrow\ Sf
\using {\downarrow}L
\endprooftree
\justifies
[Nt(s(m))], \mbox{\fbox{${\forall}a((Sf{{}{\uparrow}{}}(({\langle\rangle}Na\backslash Sf)/({\langle\rangle}Na\backslash Sb))){{}{\downarrow}{}}Sf)$}}, {\square}({\langle\rangle}{\exists}aNa\backslash Sb)\ \Rightarrow\ Sf
\using {\forall}L
\endprooftree
\justifies
[Nt(s(m))], \mbox{\fbox{${\forall}g{\forall}a((Sg{{}{\uparrow}{}}(({\langle\rangle}Na\backslash Sf)/({\langle\rangle}Na\backslash Sb))){{}{\downarrow}{}}Sg)$}}, {\square}({\langle\rangle}{\exists}aNa\backslash Sb)\ \Rightarrow\ Sf
\using {\forall}L
\endprooftree
\justifies
[Nt(s(m))], \mbox{\fbox{${\blacksquare}{\forall}g{\forall}a((Sg{{}{\uparrow}{}}(({\langle\rangle}Na\backslash Sf)/({\langle\rangle}Na\backslash Sb))){{}{\downarrow}{}}Sg)$}}, {\square}({\langle\rangle}{\exists}aNa\backslash Sb)\ \Rightarrow\ Sf
\using {\blacksquare}L
\endprooftree
\justifies
[{\tt 1}], {\blacksquare}{\forall}g{\forall}a((Sg{{}{\uparrow}{}}(({\langle\rangle}Na\backslash Sf)/({\langle\rangle}Na\backslash Sb))){{}{\downarrow}{}}Sg), {\square}({\langle\rangle}{\exists}aNa\backslash Sb)\ \Rightarrow\ Sf{{}{\uparrow}{}}Nt(s(m))
\using {\uparrow}R
\endprooftree
\prooftree
\justifies
\mbox{\fbox{$Sf$}}\ \Rightarrow\ Sf
\endprooftree
\justifies
[\mbox{\fbox{$(Sf{{}{\uparrow}{}}Nt(s(m))){{}{\downarrow}{}}Sf$}}], {\blacksquare}{\forall}g{\forall}a((Sg{{}{\uparrow}{}}(({\langle\rangle}Na\backslash Sf)/({\langle\rangle}Na\backslash Sb))){{}{\downarrow}{}}Sg), {\square}({\langle\rangle}{\exists}aNa\backslash Sb)\ \Rightarrow\ Sf
\using {\downarrow}L
\endprooftree
\justifies
[\mbox{\fbox{${\forall}f((Sf{{}{\uparrow}{}}Nt(s(m))){{}{\downarrow}{}}Sf)$}}], {\blacksquare}{\forall}g{\forall}a((Sg{{}{\uparrow}{}}(({\langle\rangle}Na\backslash Sf)/({\langle\rangle}Na\backslash Sb))){{}{\downarrow}{}}Sg), {\square}({\langle\rangle}{\exists}aNa\backslash Sb)\ \Rightarrow\ Sf
\using {\forall}L
\endprooftree
\justifies
[\mbox{\fbox{${\forall}f((Sf{{}{\uparrow}{}}Nt(s(m))){{}{\downarrow}{}}Sf)/{\it CN}{\it s(m)}$}}, {\square}{\it CN}{\it s(m)}], {\blacksquare}{\forall}g{\forall}a((Sg{{}{\uparrow}{}}(({\langle\rangle}Na\backslash Sf)/({\langle\rangle}Na\backslash Sb))){{}{\downarrow}{}}Sg), {\square}({\langle\rangle}{\exists}aNa\backslash Sb)\ \Rightarrow\ Sf
\using {/}L
\endprooftree
\justifies
[\mbox{\fbox{${\forall}g({\forall}f((Sf{{}{\uparrow}{}}Nt(s(g))){{}{\downarrow}{}}Sf)/{\it CN}{\it s(g)})$}}, {\square}{\it CN}{\it s(m)}], {\blacksquare}{\forall}g{\forall}a((Sg{{}{\uparrow}{}}(({\langle\rangle}Na\backslash Sf)/({\langle\rangle}Na\backslash Sb))){{}{\downarrow}{}}Sg), {\square}({\langle\rangle}{\exists}aNa\backslash Sb)\ \Rightarrow\ Sf
\using {\forall}L
\endprooftree
\justifies
[\mbox{\fbox{${\blacksquare}{\forall}g({\forall}f((Sf{{}{\uparrow}{}}Nt(s(g))){{}{\downarrow}{}}Sf)/{\it CN}{\it s(g)})$}}, {\square}{\it CN}{\it s(m)}], {\blacksquare}{\forall}g{\forall}a((Sg{{}{\uparrow}{}}(({\langle\rangle}Na\backslash Sf)/({\langle\rangle}Na\backslash Sb))){{}{\downarrow}{}}Sg), {\square}({\langle\rangle}{\exists}aNa\backslash Sb)\ \Rightarrow\ Sf
\using {\blacksquare}L
\endprooftree
$$
\vspace{0.15in}
\noindent
This delivers the subordinate reading:
\disp{
$\forall C[(\mbox{\v{}}{\it man}\ {\it C})\rightarrow \neg (\mbox{\v{}}{\it walk}\ {\it C})]$}
There is also the derivation:
\vspace{0.15in}
$$
\tiny
\prooftree
\prooftree
\prooftree
\prooftree
\prooftree
\prooftree
\prooftree
\prooftree
\prooftree
\prooftree
\justifies
\mbox{\fbox{${\it CN}{\it s(m)}$}}\ \Rightarrow\ {\it CN}{\it s(m)}
\endprooftree
\justifies
\mbox{\fbox{${\square}{\it CN}{\it s(m)}$}}\ \Rightarrow\ {\it CN}{\it s(m)}
\using {\Box}L
\endprooftree
\prooftree
\prooftree
\prooftree
\prooftree
\prooftree
\prooftree
\prooftree
\prooftree
\prooftree
\prooftree
\prooftree
\justifies
Nt(s(m))\ \Rightarrow\ Nt(s(m))
\endprooftree
\justifies
Nt(s(m))\ \Rightarrow\ \fbox{${\exists}aNa$}
\using {\exists}R
\endprooftree
\justifies
[Nt(s(m))]\ \Rightarrow\ \fbox{${\langle\rangle}{\exists}aNa$}
\using {\langle\rangle}R
\endprooftree
\prooftree
\justifies
\mbox{\fbox{$Sb$}}\ \Rightarrow\ Sb
\endprooftree
\justifies
[Nt(s(m))], \mbox{\fbox{${\langle\rangle}{\exists}aNa\backslash Sb$}}\ \Rightarrow\ Sb
\using {\backslash}L
\endprooftree
\justifies
[Nt(s(m))], \mbox{\fbox{${\square}({\langle\rangle}{\exists}aNa\backslash Sb)$}}\ \Rightarrow\ Sb
\using {\Box}L
\endprooftree
\justifies
{\langle\rangle}Nt(s(m)), {\square}({\langle\rangle}{\exists}aNa\backslash Sb)\ \Rightarrow\ Sb
\using {\langle\rangle}L
\endprooftree
\justifies
{\square}({\langle\rangle}{\exists}aNa\backslash Sb)\ \Rightarrow\ {\langle\rangle}Nt(s(m))\backslash Sb
\using {\backslash}R
\endprooftree
\prooftree
\prooftree
\prooftree
\justifies
Nt(s(m))\ \Rightarrow\ Nt(s(m))
\endprooftree
\justifies
[Nt(s(m))]\ \Rightarrow\ \fbox{${\langle\rangle}Nt(s(m))$}
\using {\langle\rangle}R
\endprooftree
\prooftree
\justifies
\mbox{\fbox{$Sf$}}\ \Rightarrow\ Sf
\endprooftree
\justifies
[Nt(s(m))], \mbox{\fbox{${\langle\rangle}Nt(s(m))\backslash Sf$}}\ \Rightarrow\ Sf
\using {\backslash}L
\endprooftree
\justifies
[Nt(s(m))], \mbox{\fbox{$({\langle\rangle}Nt(s(m))\backslash Sf)/({\langle\rangle}Nt(s(m))\backslash Sb)$}}, {\square}({\langle\rangle}{\exists}aNa\backslash Sb)\ \Rightarrow\ Sf
\using {/}L
\endprooftree
\justifies
[{\tt 1}], ({\langle\rangle}Nt(s(m))\backslash Sf)/({\langle\rangle}Nt(s(m))\backslash Sb), {\square}({\langle\rangle}{\exists}aNa\backslash Sb)\ \Rightarrow\ Sf{{}{\uparrow}{}}Nt(s(m))
\using {\uparrow}R
\endprooftree
\prooftree
\justifies
\mbox{\fbox{$Sf$}}\ \Rightarrow\ Sf
\endprooftree
\justifies
[\mbox{\fbox{$(Sf{{}{\uparrow}{}}Nt(s(m))){{}{\downarrow}{}}Sf$}}], ({\langle\rangle}Nt(s(m))\backslash Sf)/({\langle\rangle}Nt(s(m))\backslash Sb), {\square}({\langle\rangle}{\exists}aNa\backslash Sb)\ \Rightarrow\ Sf
\using {\downarrow}L
\endprooftree
\justifies
[\mbox{\fbox{${\forall}f((Sf{{}{\uparrow}{}}Nt(s(m))){{}{\downarrow}{}}Sf)$}}], ({\langle\rangle}Nt(s(m))\backslash Sf)/({\langle\rangle}Nt(s(m))\backslash Sb), {\square}({\langle\rangle}{\exists}aNa\backslash Sb)\ \Rightarrow\ Sf
\using {\forall}L
\endprooftree
\justifies
[\mbox{\fbox{${\forall}f((Sf{{}{\uparrow}{}}Nt(s(m))){{}{\downarrow}{}}Sf)/{\it CN}{\it s(m)}$}}, {\square}{\it CN}{\it s(m)}], ({\langle\rangle}Nt(s(m))\backslash Sf)/({\langle\rangle}Nt(s(m))\backslash Sb), {\square}({\langle\rangle}{\exists}aNa\backslash Sb)\ \Rightarrow\ Sf
\using {/}L
\endprooftree
\justifies
[\mbox{\fbox{${\forall}g({\forall}f((Sf{{}{\uparrow}{}}Nt(s(g))){{}{\downarrow}{}}Sf)/{\it CN}{\it s(g)})$}}, {\square}{\it CN}{\it s(m)}], ({\langle\rangle}Nt(s(m))\backslash Sf)/({\langle\rangle}Nt(s(m))\backslash Sb), {\square}({\langle\rangle}{\exists}aNa\backslash Sb)\ \Rightarrow\ Sf
\using {\forall}L
\endprooftree
\justifies
[\mbox{\fbox{${\blacksquare}{\forall}g({\forall}f((Sf{{}{\uparrow}{}}Nt(s(g))){{}{\downarrow}{}}Sf)/{\it CN}{\it s(g)})$}}, {\square}{\it CN}{\it s(m)}], ({\langle\rangle}Nt(s(m))\backslash Sf)/({\langle\rangle}Nt(s(m))\backslash Sb), {\square}({\langle\rangle}{\exists}aNa\backslash Sb)\ \Rightarrow\ Sf
\using {\blacksquare}L
\endprooftree
\justifies
[{\blacksquare}{\forall}g({\forall}f((Sf{{}{\uparrow}{}}Nt(s(g))){{}{\downarrow}{}}Sf)/{\it CN}{\it s(g)}), {\square}{\it CN}{\it s(m)}], {\tt 1}, {\square}({\langle\rangle}{\exists}aNa\backslash Sb)\ \Rightarrow\ Sf{{}{\uparrow}{}}(({\langle\rangle}Nt(s(m))\backslash Sf)/({\langle\rangle}Nt(s(m))\backslash Sb))
\using {\uparrow}R
\endprooftree
\prooftree
\justifies
\mbox{\fbox{$Sf$}}\ \Rightarrow\ Sf
\endprooftree
\justifies
[{\blacksquare}{\forall}g({\forall}f((Sf{{}{\uparrow}{}}Nt(s(g))){{}{\downarrow}{}}Sf)/{\it CN}{\it s(g)}), {\square}{\it CN}{\it s(m)}], \mbox{\fbox{$(Sf{{}{\uparrow}{}}(({\langle\rangle}Nt(s(m))\backslash Sf)/({\langle\rangle}Nt(s(m))\backslash Sb))){{}{\downarrow}{}}Sf$}}, {\square}({\langle\rangle}{\exists}aNa\backslash Sb)\ \Rightarrow\ Sf
\using {\downarrow}L
\endprooftree
\justifies
[{\blacksquare}{\forall}g({\forall}f((Sf{{}{\uparrow}{}}Nt(s(g))){{}{\downarrow}{}}Sf)/{\it CN}{\it s(g)}), {\square}{\it CN}{\it s(m)}], \mbox{\fbox{${\forall}a((Sf{{}{\uparrow}{}}(({\langle\rangle}Na\backslash Sf)/({\langle\rangle}Na\backslash Sb))){{}{\downarrow}{}}Sf)$}}, {\square}({\langle\rangle}{\exists}aNa\backslash Sb)\ \Rightarrow\ Sf
\using {\forall}L
\endprooftree
\justifies
[{\blacksquare}{\forall}g({\forall}f((Sf{{}{\uparrow}{}}Nt(s(g))){{}{\downarrow}{}}Sf)/{\it CN}{\it s(g)}), {\square}{\it CN}{\it s(m)}], \mbox{\fbox{${\forall}g{\forall}a((Sg{{}{\uparrow}{}}(({\langle\rangle}Na\backslash Sf)/({\langle\rangle}Na\backslash Sb))){{}{\downarrow}{}}Sg)$}}, {\square}({\langle\rangle}{\exists}aNa\backslash Sb)\ \Rightarrow\ Sf
\using {\forall}L
\endprooftree
\justifies
[{\blacksquare}{\forall}g({\forall}f((Sf{{}{\uparrow}{}}Nt(s(g))){{}{\downarrow}{}}Sf)/{\it CN}{\it s(g)}), {\square}{\it CN}{\it s(m)}], \mbox{\fbox{${\blacksquare}{\forall}g{\forall}a((Sg{{}{\uparrow}{}}(({\langle\rangle}Na\backslash Sf)/({\langle\rangle}Na\backslash Sb))){{}{\downarrow}{}}Sg)$}}, {\square}({\langle\rangle}{\exists}aNa\backslash Sb)\ \Rightarrow\ Sf
\using {\blacksquare}L
\endprooftree
$$
\vspace{0.15in}
\noindent
This delivers the dominant reading:
\disp{
$\neg \forall G[(\mbox{\v{}}{\it man}\ {\it G})\rightarrow (\mbox{\v{}}{\it walk}\ {\it G})]$}

\chapter{Coordination}

\label{coordchap}

In this chapter we analyse examples of coordination,
cf.~Morrill~(2011\cite{morrill:oxford}, Chapter~3, Section~10).

\section{Constituent and `non-constituent'  coordination}

\label{crd:1}

To express the lexical semantics of coordination, including iterated coordination
(e.g.~\lingform{Bill, Mary,} \lingform{Suzy and} \lingform{Fred})
and various arities (zeroary e.g.~sentence,
unary e.g.~verb phrase, binary e.g.~transitive verb, \ldots), we use combinators: a non-empty list
map apply $\alphaplus$ and a non-empty list map $\mathbf{\Phi^n}$ combinator $\Phinplus$. 

The
combinator $\mathbf{\Phi}$ is such that $\mathbf{\Phi}\ x\ y\ z\ w= x\ (y\ w)\ (z\ w)$ (Curry and Feys
1958\cite{curryfeys}).
The non-empty list
map apply combinator $\alphaplus$ is as follows:
\disp{
$
\begin{array}[t]{rcl}
(\alphaplus\ [x]\ y) & = & [(x\ y)]\\
(\alphaplus\ [x, y| z]\ w) & = & [(x\ w)|(\alphaplus\ [y|z]\ w)]
\end{array}
$}
The non-empty list map $\mathbf{\Phi^n}$ combinator $\Phinplus$ is thus:
\disp{
$
\begin{array}[t]{rcl}
(((\Phinplus\ 0\ {\it and})\ x)\ [y]) & = & [y\wedge x]\\
(((\Phinplus\ 0\ {\it or})\ x)\ [y]) & = & [y\vee x]\\
(((\Phinplus\ 0\ {\it and})\ x)\ [y, z|w]) & = & [y\wedge(((\Phinplus\ 0\ {\it and})\ x)\ [z|w])]\\
(((\Phinplus\ 0\ {\it or})\ x)\ [y, z|w]) & = & [y\vee(((\Phinplus\ 0\ {\it or})\ x)\ [z|w])]\\
((((\Phinplus\ (s\ n)\ c)\ x)\ y)\ z) & = & (((\Phinplus\ n\ c)\ (x\ z))\ (\alphaplus\ y\ z))
\end{array}
$}
These equations mean that in semantic evaluation any subterm of the form
on the left is to be replaced by that on the right, successively.

\subsection{Sentence coordination}

The first example is simple sentential conjunction; as we have said subjects
are bracketed since they are (weak) islands and
coordinate structures are doubly bracketed corresponding to the fact that they are
strong islands; these brackets are given in the input:
\disp{
$[[[{\bf john}]{+}{\bf praises}{+}{\bf mary}{+}{\bf and}{+}[{\bf john}]{+}{\bf laughs}]]: Sf$}
Lexical lookup yields  the following where the coordinator type is
essentially $(\exstexp X\bsl\abrack\abrack X)/X$ with $X=S$.
\disp{
$\begin{array}[t]{l}
[[[{\blacksquare}Nt(s(m)): {\it j}], {\square}(({\langle\rangle}{\exists}gNt(s(g))\backslash Sf)/{\exists}aNa): \mbox{\^{}}\lambda A\lambda B({\it Pres}\ ((\mbox{\v{}}{\it praise}\ {\it A})\ {\it B})),\\ {\blacksquare}Nt(s(f)): {\it m},
 {\blacksquare}{\forall}f((?{\blacksquare}Sf\backslash {[]^{-1}}{[]^{-1}}Sf)/{\blacksquare}Sf): (\Phinplus\ {\it 0}\ {\it and}), [{\blacksquare}Nt(s(m)): {\it j}],\\ {\square}({\langle\rangle}{\exists}gNt(s(g))\backslash Sf): \mbox{\^{}}\lambda C({\it Pres}\ (\mbox{\v{}}{\it laugh}\ {\it C}))]]\ \Rightarrow\ Sf
 \end{array}$}
The left conjunt is marked with the existential exponential to allow iterated coordination,
which we will illustrate later;
the conjunts are marked with semantically inactive normal modalities to make coordinate
structures scope islands to quantifiers other than indefinites.
There is the derivation:

\vspace{0.15in}

\begin{center}
\noindent\rotatebox{-90}{\scriptsize
\prooftree
\prooftree
\prooftree
\prooftree
\prooftree
\prooftree
\prooftree
\prooftree
\prooftree
\prooftree
\justifies
\mbox{\fbox{$Nt(s(m))$}}\ \Rightarrow\ Nt(s(m))
\endprooftree
\justifies
\mbox{\fbox{${\blacksquare}Nt(s(m))$}}\ \Rightarrow\ Nt(s(m))
\using {\blacksquare}L
\endprooftree
\justifies
{\blacksquare}Nt(s(m))\ \Rightarrow\ \fbox{${\exists}gNt(s(g))$}
\using {\exists}R
\endprooftree
\justifies
[{\blacksquare}Nt(s(m))]\ \Rightarrow\ \fbox{${\langle\rangle}{\exists}gNt(s(g))$}
\using {\langle\rangle}R
\endprooftree
\prooftree
\justifies
\mbox{\fbox{$Sf$}}\ \Rightarrow\ Sf
\endprooftree
\justifies
[{\blacksquare}Nt(s(m))], \mbox{\fbox{${\langle\rangle}{\exists}gNt(s(g))\backslash Sf$}}\ \Rightarrow\ Sf
\using {\backslash}L
\endprooftree
\justifies
[{\blacksquare}Nt(s(m))], \mbox{\fbox{${\square}({\langle\rangle}{\exists}gNt(s(g))\backslash Sf)$}}\ \Rightarrow\ Sf
\using {\Box}L
\endprooftree
\justifies
\begin{array}{c}
[{\blacksquare}Nt(s(m))], {\square}({\langle\rangle}{\exists}gNt(s(g))\backslash Sf)\ \Rightarrow\ {\blacksquare}Sf\\
\mbox{\footnotesize\textcircled{1}}
\end{array}
\using {\blacksquare}R
\endprooftree
\prooftree
\prooftree
\prooftree
\prooftree
\prooftree
\prooftree
\prooftree
\prooftree
\justifies
\mbox{\fbox{$Nt(s(f))$}}\ \Rightarrow\ Nt(s(f))
\endprooftree
\justifies
\mbox{\fbox{${\blacksquare}Nt(s(f))$}}\ \Rightarrow\ Nt(s(f))
\using {\blacksquare}L
\endprooftree
\justifies
{\blacksquare}Nt(s(f))\ \Rightarrow\ \fbox{${\exists}aNa$}
\using {\exists}R
\endprooftree
\prooftree
\prooftree
\prooftree
\prooftree
\prooftree
\justifies
\mbox{\fbox{$Nt(s(m))$}}\ \Rightarrow\ Nt(s(m))
\endprooftree
\justifies
\mbox{\fbox{${\blacksquare}Nt(s(m))$}}\ \Rightarrow\ Nt(s(m))
\using {\blacksquare}L
\endprooftree
\justifies
{\blacksquare}Nt(s(m))\ \Rightarrow\ \fbox{${\exists}gNt(s(g))$}
\using {\exists}R
\endprooftree
\justifies
[{\blacksquare}Nt(s(m))]\ \Rightarrow\ \fbox{${\langle\rangle}{\exists}gNt(s(g))$}
\using {\langle\rangle}R
\endprooftree
\prooftree
\justifies
\mbox{\fbox{$Sf$}}\ \Rightarrow\ Sf
\endprooftree
\justifies
[{\blacksquare}Nt(s(m))], \mbox{\fbox{${\langle\rangle}{\exists}gNt(s(g))\backslash Sf$}}\ \Rightarrow\ Sf
\using {\backslash}L
\endprooftree
\justifies
[{\blacksquare}Nt(s(m))], \mbox{\fbox{$({\langle\rangle}{\exists}gNt(s(g))\backslash Sf)/{\exists}aNa$}}, {\blacksquare}Nt(s(f))\ \Rightarrow\ Sf
\using {/}L
\endprooftree
\justifies
[{\blacksquare}Nt(s(m))], \mbox{\fbox{${\square}(({\langle\rangle}{\exists}gNt(s(g))\backslash Sf)/{\exists}aNa)$}}, {\blacksquare}Nt(s(f))\ \Rightarrow\ Sf
\using {\Box}L
\endprooftree
\justifies
[{\blacksquare}Nt(s(m))], {\square}(({\langle\rangle}{\exists}gNt(s(g))\backslash Sf)/{\exists}aNa), {\blacksquare}Nt(s(f))\ \Rightarrow\ {\blacksquare}Sf
\using {\blacksquare}R
\endprooftree
\justifies
\begin{array}{c}
[{\blacksquare}Nt(s(m))], {\square}(({\langle\rangle}{\exists}gNt(s(g))\backslash Sf)/{\exists}aNa), {\blacksquare}Nt(s(f))\ \Rightarrow\ \fbox{$?{\blacksquare}Sf$}\\
\mbox{\footnotesize\textcircled{2}}
\end{array}
\using {?}R
\endprooftree
\prooftree
\prooftree
\prooftree
\justifies
\mbox{\fbox{$Sf$}}\ \Rightarrow\ Sf
\endprooftree
\justifies
[\mbox{\fbox{${[]^{-1}}Sf$}}]\ \Rightarrow\ Sf
\using {[]^{-1}}L
\endprooftree
\justifies
\begin{array}{c}
{}[[\mbox{\fbox{${[]^{-1}}{[]^{-1}}Sf$}}]]\ \Rightarrow\ Sf\\
\mbox{\footnotesize\textcircled{3}}
\end{array}
\using {[]^{-1}}L
\endprooftree
\justifies
[[[{\blacksquare}Nt(s(m))], {\square}(({\langle\rangle}{\exists}gNt(s(g))\backslash Sf)/{\exists}aNa), {\blacksquare}Nt(s(f)), \mbox{\fbox{$?{\blacksquare}Sf\backslash {[]^{-1}}{[]^{-1}}Sf$}}]]\ \Rightarrow\ Sf
\using {\backslash}L
\endprooftree
\justifies
[[[{\blacksquare}Nt(s(m))], {\square}(({\langle\rangle}{\exists}gNt(s(g))\backslash Sf)/{\exists}aNa), {\blacksquare}Nt(s(f)), \mbox{\fbox{$(?{\blacksquare}Sf\backslash {[]^{-1}}{[]^{-1}}Sf)/{\blacksquare}Sf$}}, [{\blacksquare}Nt(s(m))], {\square}({\langle\rangle}{\exists}gNt(s(g))\backslash Sf)]]\ \Rightarrow\ Sf
\using {/}L
\endprooftree
\justifies
[[[{\blacksquare}Nt(s(m))], {\square}(({\langle\rangle}{\exists}gNt(s(g))\backslash Sf)/{\exists}aNa), {\blacksquare}Nt(s(f)), \mbox{\fbox{${\forall}f((?{\blacksquare}Sf\backslash {[]^{-1}}{[]^{-1}}Sf)/{\blacksquare}Sf)$}}, [{\blacksquare}Nt(s(m))], {\square}({\langle\rangle}{\exists}gNt(s(g))\backslash Sf)]]\ \Rightarrow\ Sf
\using {\forall}L
\endprooftree
\justifies
[[[{\blacksquare}Nt(s(m))], {\square}(({\langle\rangle}{\exists}gNt(s(g))\backslash Sf)/{\exists}aNa), {\blacksquare}Nt(s(f)), \mbox{\fbox{${\blacksquare}{\forall}f((?{\blacksquare}Sf\backslash {[]^{-1}}{[]^{-1}}Sf)/{\blacksquare}Sf)$}}, [{\blacksquare}Nt(s(m))], {\square}({\langle\rangle}{\exists}gNt(s(g))\backslash Sf)]]\ \Rightarrow\ Sf
\using {\blacksquare}L
\endprooftree}
\end{center}

\vspace{0.15in}

\noindent
In the conclusions of rules boxes mark the active type of the rule application, i.e.~the
type which is decomposed in the premises; as we have seen
the boxes mark \techterm{focused\/} types (Andreoli 1992\cite{andreoli:92}), 
which are the active types in
the conclusions of noninvertible rules, subject to focusing rule application;
the active types in conclusions of invertible rules are not marked; 
whenever a type is boxed it is the active type
--- this improves readability.
Reading from the root, the type of the coordinator projecting the construction
is decomposed, eliminating the semantically inactive outermost modality
and instantiating the tense feature to finite ($f$). 
The result then applies forwards to the righthand conjunct,
the analysis of which is shown in the subtree marked \textcircled{1}:
after removal of the semantically inactive modality in the succedent
(which is licensed by the fact that the antecedents are modalised),
left box elimination is applied to the intransitive verb;
this then applies to its subject and in the minor premise the subject
bracket modality is removed, then the agreement instantiated to
third person singular masculine ($t(s(m))$), and then the semantically inactive subject
modality is removed and the axiom is matched.
The coordinator then applies to the lefthand conjunct,
in the subtree marked \textcircled{2}:
the existential exponential is removed directly from the succedent, since
the coordination is not iterative, and then the semantically inactive succedent
modality is eliminated (as required the antecedents are modalised), and the outer
modality of the transitive verb is removed;
the transitive verb then applies to its object, where the agreement is instantiated to
third person singular feminine ($t(s(f))$), and the analysis of the resulting
verb phrase is the same as in \textcircled{1}.
Finally, in the subderivation \textcircled{3} the coordinator checks off
the doubly bracketed context which it requires/projects.
All this delivers the correct reading under:
\disp{
$[({\it Pres}\ ((\mbox{\v{}}{\it praise}\ {\it m})\ {\it j}))\wedge ({\it Pres}\ (\mbox{\v{}}{\it laugh}\ {\it j}))]$}

\subsection{Verb phrase coordination}

The next example is one of verb phrase conjunction:
\disp{
$[{\bf john}]{+}[[{\bf praises}{+}{\bf mary}{+}{\bf and}{+}{\bf laughs}]]: Sf$}
Lexical lookup yields the coordinator type
basically $(\exstexp X\bsl\abrack\abrack X)/X$ with $X=N\bsl S$.
\disp{
$
\begin{array}[t]{l}
[{\blacksquare}Nt(s(m)): {\it j}], [[{\square}(({\langle\rangle}{\exists}gNt(s(g))\backslash Sf)/{\exists}aNa): \mbox{\^{}}\lambda A\lambda B({\it Pres}\ ((\mbox{\v{}}{\it praise}\ {\it A})\ {\it B})),\\ {\blacksquare}Nt(s(f)): {\it m},
 {\blacksquare}{\forall}a{\forall}f((?{\blacksquare}({\langle\rangle}Na\backslash Sf)\backslash {[]^{-1}}{[]^{-1}}({\langle\rangle}Na\backslash Sf))/{\blacksquare}({\langle\rangle}Na\backslash Sf)):\\ (\Phinplus\ ({\it s}\ {\it 0})\ {\it and}), {\square}({\langle\rangle}{\exists}gNt(s(g))\backslash Sf): \mbox{\^{}}\lambda C({\it Pres}\ (\mbox{\v{}}{\it laugh}\ {\it C}))]]\ \Rightarrow\ Sf
 \end{array}$}
The coordination combinator semantics
$(\Phinplus\ (s\ 0)\ {\it and})$ is such that:
\disp{$
\begin{array}[t]{l}
((((\Phinplus\ (s\ 0)\ {\it and})\ x)\ [y])\ z) =\\
(((\Phinplus\ 0\ {\it and})\ (x\ z))\ (\alphaplus\ [y]\ z)) =\\
(((\Phinplus\ 0\ {\it and})\ (x\ z))\ [(y\ z)]) =\\
{}[(y\ z)\wedge(x\ z)]
\end{array}$}
There is the derivation:

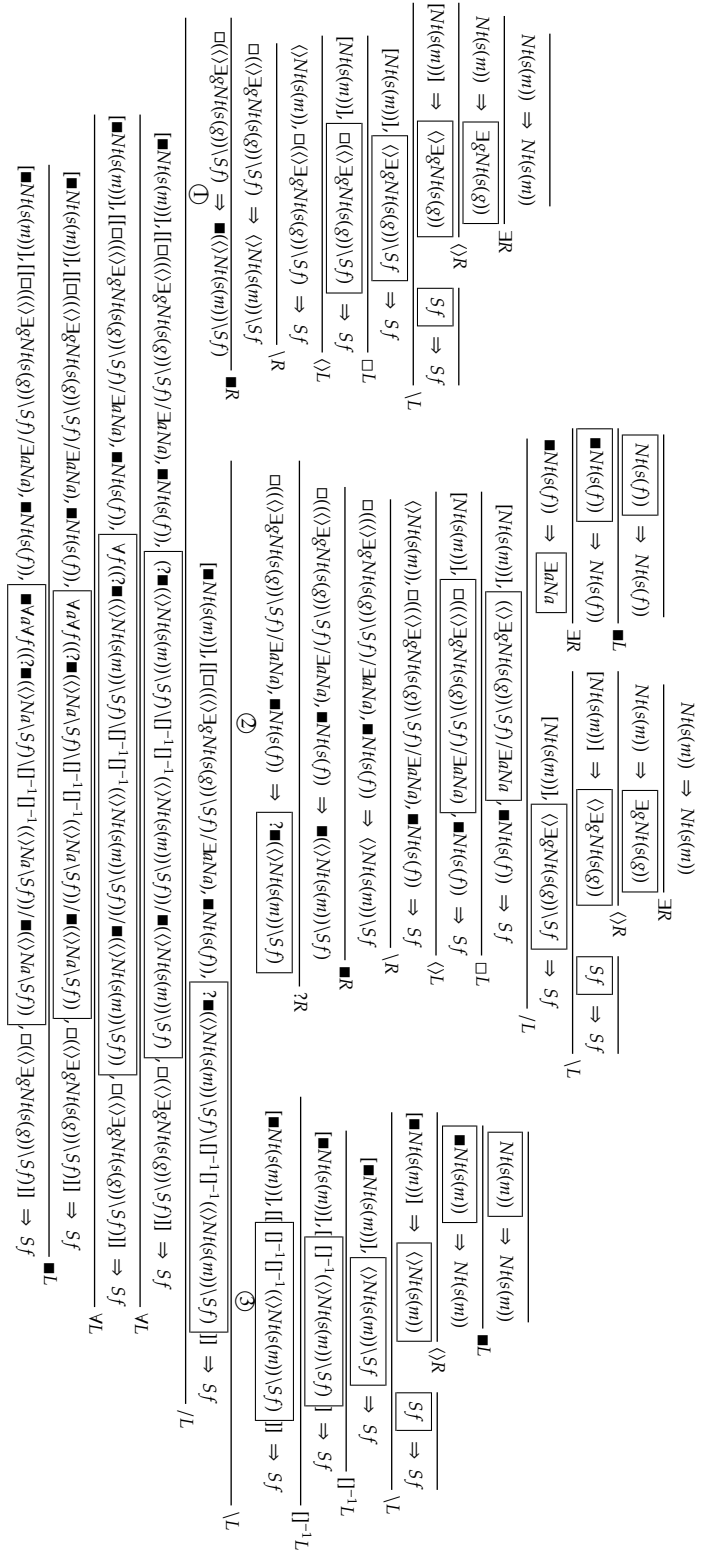
\vspace{0.15in}

\begin{center}
\rotatebox{-90}{\scriptsize
\prooftree
\prooftree
\prooftree
\prooftree
\prooftree
\prooftree
\prooftree
\prooftree
\prooftree
\prooftree
\prooftree
\prooftree
\justifies
Nt(s(m))\ \Rightarrow\ Nt(s(m))
\endprooftree
\justifies
Nt(s(m))\ \Rightarrow\ \fbox{${\exists}gNt(s(g))$}
\using {\exists}R
\endprooftree
\justifies
[Nt(s(m))]\ \Rightarrow\ \fbox{${\langle\rangle}{\exists}gNt(s(g))$}
\using {\langle\rangle}R
\endprooftree
\prooftree
\justifies
\mbox{\fbox{$Sf$}}\ \Rightarrow\ Sf
\endprooftree
\justifies
[Nt(s(m))], \mbox{\fbox{${\langle\rangle}{\exists}gNt(s(g))\backslash Sf$}}\ \Rightarrow\ Sf
\using {\backslash}L
\endprooftree
\justifies
[Nt(s(m))], \mbox{\fbox{${\square}({\langle\rangle}{\exists}gNt(s(g))\backslash Sf)$}}\ \Rightarrow\ Sf
\using {\Box}L
\endprooftree
\justifies
{\langle\rangle}Nt(s(m)), {\square}({\langle\rangle}{\exists}gNt(s(g))\backslash Sf)\ \Rightarrow\ Sf
\using {\langle\rangle}L
\endprooftree
\justifies
{\square}({\langle\rangle}{\exists}gNt(s(g))\backslash Sf)\ \Rightarrow\ {\langle\rangle}Nt(s(m))\backslash Sf
\using {\backslash}R
\endprooftree
\justifies
\begin{array}{c}
{\square}({\langle\rangle}{\exists}gNt(s(g))\backslash Sf)\ \Rightarrow\ {\blacksquare}({\langle\rangle}Nt(s(m))\backslash Sf)\\
\mbox{\footnotesize\textcircled{1}}
\end{array}
\using {\blacksquare}R
\endprooftree
\prooftree
\prooftree
\prooftree
\prooftree
\prooftree
\prooftree
\prooftree
\prooftree
\prooftree
\prooftree
\justifies
\mbox{\fbox{$Nt(s(f))$}}\ \Rightarrow\ Nt(s(f))
\endprooftree
\justifies
\mbox{\fbox{${\blacksquare}Nt(s(f))$}}\ \Rightarrow\ Nt(s(f))
\using {\blacksquare}L
\endprooftree
\justifies
{\blacksquare}Nt(s(f))\ \Rightarrow\ \fbox{${\exists}aNa$}
\using {\exists}R
\endprooftree
\prooftree
\prooftree
\prooftree
\prooftree
\justifies
Nt(s(m))\ \Rightarrow\ Nt(s(m))
\endprooftree
\justifies
Nt(s(m))\ \Rightarrow\ \fbox{${\exists}gNt(s(g))$}
\using {\exists}R
\endprooftree
\justifies
[Nt(s(m))]\ \Rightarrow\ \fbox{${\langle\rangle}{\exists}gNt(s(g))$}
\using {\langle\rangle}R
\endprooftree
\prooftree
\justifies
\mbox{\fbox{$Sf$}}\ \Rightarrow\ Sf
\endprooftree
\justifies
[Nt(s(m))], \mbox{\fbox{${\langle\rangle}{\exists}gNt(s(g))\backslash Sf$}}\ \Rightarrow\ Sf
\using {\backslash}L
\endprooftree
\justifies
[Nt(s(m))], \mbox{\fbox{$({\langle\rangle}{\exists}gNt(s(g))\backslash Sf)/{\exists}aNa$}}, {\blacksquare}Nt(s(f))\ \Rightarrow\ Sf
\using {/}L
\endprooftree
\justifies
[Nt(s(m))], \mbox{\fbox{${\square}(({\langle\rangle}{\exists}gNt(s(g))\backslash Sf)/{\exists}aNa)$}}, {\blacksquare}Nt(s(f))\ \Rightarrow\ Sf
\using {\Box}L
\endprooftree
\justifies
{\langle\rangle}Nt(s(m)), {\square}(({\langle\rangle}{\exists}gNt(s(g))\backslash Sf)/{\exists}aNa), {\blacksquare}Nt(s(f))\ \Rightarrow\ Sf
\using {\langle\rangle}L
\endprooftree
\justifies
{\square}(({\langle\rangle}{\exists}gNt(s(g))\backslash Sf)/{\exists}aNa), {\blacksquare}Nt(s(f))\ \Rightarrow\ {\langle\rangle}Nt(s(m))\backslash Sf
\using {\backslash}R
\endprooftree
\justifies
{\square}(({\langle\rangle}{\exists}gNt(s(g))\backslash Sf)/{\exists}aNa), {\blacksquare}Nt(s(f))\ \Rightarrow\ {\blacksquare}({\langle\rangle}Nt(s(m))\backslash Sf)
\using {\blacksquare}R
\endprooftree
\justifies
\begin{array}{c}
{\square}(({\langle\rangle}{\exists}gNt(s(g))\backslash Sf)/{\exists}aNa), {\blacksquare}Nt(s(f))\ \Rightarrow\ \fbox{$?{\blacksquare}({\langle\rangle}Nt(s(m))\backslash Sf)$}\\
\mbox{\footnotesize\textcircled{2}}
\end{array}
\using {?}R
\endprooftree
\prooftree
\prooftree
\prooftree
\prooftree
\prooftree
\prooftree
\justifies
\mbox{\fbox{$Nt(s(m))$}}\ \Rightarrow\ Nt(s(m))
\endprooftree
\justifies
\mbox{\fbox{${\blacksquare}Nt(s(m))$}}\ \Rightarrow\ Nt(s(m))
\using {\blacksquare}L
\endprooftree
\justifies
[{\blacksquare}Nt(s(m))]\ \Rightarrow\ \fbox{${\langle\rangle}Nt(s(m))$}
\using {\langle\rangle}R
\endprooftree
\prooftree
\justifies
\mbox{\fbox{$Sf$}}\ \Rightarrow\ Sf
\endprooftree
\justifies
[{\blacksquare}Nt(s(m))], \mbox{\fbox{${\langle\rangle}Nt(s(m))\backslash Sf$}}\ \Rightarrow\ Sf
\using {\backslash}L
\endprooftree
\justifies
[{\blacksquare}Nt(s(m))], [\mbox{\fbox{${[]^{-1}}({\langle\rangle}Nt(s(m))\backslash Sf)$}}]\ \Rightarrow\ Sf
\using {[]^{-1}}L
\endprooftree
\justifies
\begin{array}{c}
{}[{\blacksquare}Nt(s(m))], [[\mbox{\fbox{${[]^{-1}}{[]^{-1}}({\langle\rangle}Nt(s(m))\backslash Sf)$}}]]\ \Rightarrow\ Sf\\
\mbox{\footnotesize\textcircled{3}}
\end{array}
\using {[]^{-1}}L
\endprooftree
\justifies
[{\blacksquare}Nt(s(m))], [[{\square}(({\langle\rangle}{\exists}gNt(s(g))\backslash Sf)/{\exists}aNa), {\blacksquare}Nt(s(f)), \mbox{\fbox{$?{\blacksquare}({\langle\rangle}Nt(s(m))\backslash Sf)\backslash {[]^{-1}}{[]^{-1}}({\langle\rangle}Nt(s(m))\backslash Sf)$}}]]\ \Rightarrow\ Sf
\using {\backslash}L
\endprooftree
\justifies
[{\blacksquare}Nt(s(m))], [[{\square}(({\langle\rangle}{\exists}gNt(s(g))\backslash Sf)/{\exists}aNa), {\blacksquare}Nt(s(f)), \mbox{\fbox{$(?{\blacksquare}({\langle\rangle}Nt(s(m))\backslash Sf)\backslash {[]^{-1}}{[]^{-1}}({\langle\rangle}Nt(s(m))\backslash Sf))/{\blacksquare}({\langle\rangle}Nt(s(m))\backslash Sf)$}}, {\square}({\langle\rangle}{\exists}gNt(s(g))\backslash Sf)]]\ \Rightarrow\ Sf
\using {/}L
\endprooftree
\justifies
[{\blacksquare}Nt(s(m))], [[{\square}(({\langle\rangle}{\exists}gNt(s(g))\backslash Sf)/{\exists}aNa), {\blacksquare}Nt(s(f)), \mbox{\fbox{${\forall}f((?{\blacksquare}({\langle\rangle}Nt(s(m))\backslash Sf)\backslash {[]^{-1}}{[]^{-1}}({\langle\rangle}Nt(s(m))\backslash Sf))/{\blacksquare}({\langle\rangle}Nt(s(m))\backslash Sf))$}}, {\square}({\langle\rangle}{\exists}gNt(s(g))\backslash Sf)]]\ \Rightarrow\ Sf
\using {\forall}L
\endprooftree
\justifies
[{\blacksquare}Nt(s(m))], [[{\square}(({\langle\rangle}{\exists}gNt(s(g))\backslash Sf)/{\exists}aNa), {\blacksquare}Nt(s(f)), \mbox{\fbox{${\forall}a{\forall}f((?{\blacksquare}({\langle\rangle}Na\backslash Sf)\backslash {[]^{-1}}{[]^{-1}}({\langle\rangle}Na\backslash Sf))/{\blacksquare}({\langle\rangle}Na\backslash Sf))$}}, {\square}({\langle\rangle}{\exists}gNt(s(g))\backslash Sf)]]\ \Rightarrow\ Sf
\using {\forall}L
\endprooftree
\justifies
[{\blacksquare}Nt(s(m))], [[{\square}(({\langle\rangle}{\exists}gNt(s(g))\backslash Sf)/{\exists}aNa), {\blacksquare}Nt(s(f)), \mbox{\fbox{${\blacksquare}{\forall}a{\forall}f((?{\blacksquare}({\langle\rangle}Na\backslash Sf)\backslash {[]^{-1}}{[]^{-1}}({\langle\rangle}Na\backslash Sf))/{\blacksquare}({\langle\rangle}Na\backslash Sf))$}}, {\square}({\langle\rangle}{\exists}gNt(s(g))\backslash Sf)]]\ \Rightarrow\ Sf
\using {\blacksquare}L
\endprooftree}
\end{center}

\vspace{0.15in}

\noindent
Reading from the root, first the outermost semantically inactive modality of the coordinator
is removed and the agreement features of the subject, $t(s(m))$, and the value of the tense
feature, $f$, are instantiated. 
The coordinator then applies to the righthand conjunct, for which the subderivation
is \textcircled{1}: the succedent semantically inactive modality is removed (the antecedent
is modalised) and the argument modally bracketed nominal is lowered into the
antecedent where its bracket modality is unfolded; the modality of the intransitive verb is removed,
and the subderivation finishes applying to the subject with bracket modality right
and existential quantifier right rules.
The coordinator then applies to the transitive verb plus object lefthand conjunct
in the subderivation rooted at \textcircled{2}: after existential exponential right,
semantically inactive modality right (the antecedents are modalised), and under right
lowering the subject subtype into the antecedent, the bracket modality of the subject is
unfolded; the transitive verb type is then selected, its modality removed,
and applied with modality rules and instantiation of existentially quantified
features to the object (left subsubderivation) and subject (middle subsubderivation).
In the subderivation rooted at \textcircled{3} the context of the coordinate structure
is recognised: double bracketing, and a (bracketed) proper name subject.
All this delivers semantics:
\disp{
$[({\it Pres}\ ((\mbox{\v{}}{\it praise}\ {\it m})\ {\it j}))\wedge ({\it Pres}\ (\mbox{\v{}}{\it laugh}\ {\it j}))]$}

\subsection{Transitive verb coordination}

The next example is of transitive verb coordination, with a complex non-standard
constituent transitive verb in the right-hand conjunt.
\disp{
$[{\bf john}]{+}[[{\bf likes}{+}{\bf and}{+}{\bf will}{+}{\bf love}]]{+}{\bf london}: Sf$}
Appropriate lexical lookup yields the following where the coordinator type is
essentially $(\exstexp X\bsl$ $\abrack\abrack X)/X$ with $X=(N\bsl S)/N$.
\disp{
$\begin{array}[t]{l}
[{\blacksquare}Nt(s(m)): {\it j}], [[{\square}(({\langle\rangle}{\exists}gNt(s(g))\backslash Sf)/{\exists}aNa): \mbox{\^{}}\lambda A\lambda B({\it Pres}\ ((\mbox{\v{}}{\it like}\ {\it A})\ {\it B})),\\
 {\blacksquare}{\forall}f{\forall}a((?{\blacksquare}(({\langle\rangle}Na\backslash Sf)/{\exists}bNb)\backslash {[]^{-1}}{[]^{-1}}(({\langle\rangle}Na\backslash Sf)/{\exists}bNb))/{\blacksquare}(({\langle\rangle}Na\backslash Sf)/{\exists}bNb)):\\ (\Phinplus\ ({\it s}\ ({\it s}\ {\it 0}))\ {\it and}),
  {\blacksquare}{\forall}a(({\langle\rangle}Na\backslash Sf)/({\langle\rangle}Na\backslash Sb)): \lambda C\lambda D({\it Fut}\ ({\it C}\ {\it D})),\\ {\square}(({\langle\rangle}{\exists}aNa\backslash Sb)/{\exists}aNa): \mbox{\^{}}\lambda E\lambda F((\mbox{\v{}}{\it love}\ {\it E})\ {\it F})]], {\blacksquare}Nt(s(n)): {\it l}\ \Rightarrow\ Sf
\end{array}$}
There is the following derivation.
Again, 
after eliminating the modality of the coordinator,
and instantiating the left universal quantifiers for tense and subject agreement,
at the root,
there are three main subderivations marked \textcircled{1},
\textcircled{2}
and \textcircled{3} deriving the right conjunct,
the left conjunct and the coordinate structure context respectively.
In \textcircled{1},
after removing the modality on the right and lowering the object and subject
arguments into the antecedent we have to analyse a sequence which is
essentially S-Aux-TV-O. This involves analysing TV-O as VP
(left subsubderivation),
and S-VP as S (right subsubderivation).
Subtree \textcircled{2} involves essentially derivation of the identity
TV yields TV.
Subderivation \textcircled{3} checks the double bracketing of the coordinate
structure and recognises the right object context and left subject context.
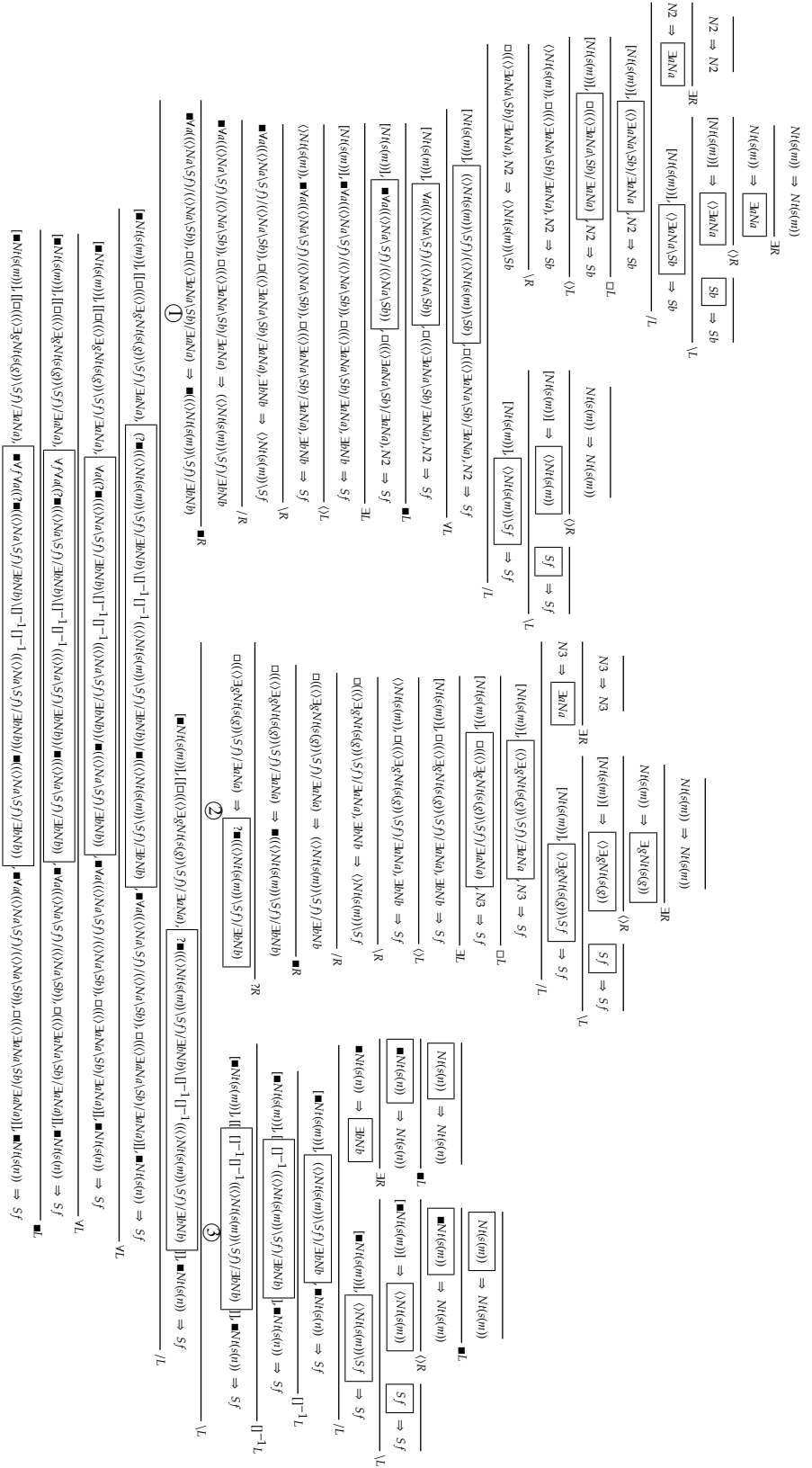
\vspace{0.15in}
$$
\rotatebox{-90}{\tiny
\prooftree
\prooftree
\prooftree
\prooftree
\prooftree
\prooftree
\prooftree
\prooftree
\prooftree
\prooftree
\prooftree
\prooftree
\prooftree
\prooftree
\prooftree
\prooftree
\prooftree
\prooftree
\justifies
N2\ \Rightarrow\ N2
\endprooftree
\justifies
N2\ \Rightarrow\ \fbox{${\exists}aNa$}
\using {\exists}R
\endprooftree
\prooftree
\prooftree
\prooftree
\prooftree
\justifies
Nt(s(m))\ \Rightarrow\ Nt(s(m))
\endprooftree
\justifies
Nt(s(m))\ \Rightarrow\ \fbox{${\exists}aNa$}
\using {\exists}R
\endprooftree
\justifies
[Nt(s(m))]\ \Rightarrow\ \fbox{${\langle\rangle}{\exists}aNa$}
\using {\langle\rangle}R
\endprooftree
\prooftree
\justifies
\mbox{\fbox{$Sb$}}\ \Rightarrow\ Sb
\endprooftree
\justifies
[Nt(s(m))], \mbox{\fbox{${\langle\rangle}{\exists}aNa\backslash Sb$}}\ \Rightarrow\ Sb
\using {\backslash}L
\endprooftree
\justifies
[Nt(s(m))], \mbox{\fbox{$({\langle\rangle}{\exists}aNa\backslash Sb)/{\exists}aNa$}}, N2\ \Rightarrow\ Sb
\using {/}L
\endprooftree
\justifies
[Nt(s(m))], \mbox{\fbox{${\square}(({\langle\rangle}{\exists}aNa\backslash Sb)/{\exists}aNa)$}}, N2\ \Rightarrow\ Sb
\using {\Box}L
\endprooftree
\justifies
{\langle\rangle}Nt(s(m)), {\square}(({\langle\rangle}{\exists}aNa\backslash Sb)/{\exists}aNa), N2\ \Rightarrow\ Sb
\using {\langle\rangle}L
\endprooftree
\justifies
{\square}(({\langle\rangle}{\exists}aNa\backslash Sb)/{\exists}aNa), N2\ \Rightarrow\ {\langle\rangle}Nt(s(m))\backslash Sb
\using {\backslash}R
\endprooftree
\prooftree
\prooftree
\prooftree
\justifies
Nt(s(m))\ \Rightarrow\ Nt(s(m))
\endprooftree
\justifies
[Nt(s(m))]\ \Rightarrow\ \fbox{${\langle\rangle}Nt(s(m))$}
\using {\langle\rangle}R
\endprooftree
\prooftree
\justifies
\mbox{\fbox{$Sf$}}\ \Rightarrow\ Sf
\endprooftree
\justifies
[Nt(s(m))], \mbox{\fbox{${\langle\rangle}Nt(s(m))\backslash Sf$}}\ \Rightarrow\ Sf
\using {\backslash}L
\endprooftree
\justifies
[Nt(s(m))], \mbox{\fbox{$({\langle\rangle}Nt(s(m))\backslash Sf)/({\langle\rangle}Nt(s(m))\backslash Sb)$}}, {\square}(({\langle\rangle}{\exists}aNa\backslash Sb)/{\exists}aNa), N2\ \Rightarrow\ Sf
\using {/}L
\endprooftree
\justifies
[Nt(s(m))], \mbox{\fbox{${\forall}a(({\langle\rangle}Na\backslash Sf)/({\langle\rangle}Na\backslash Sb))$}}, {\square}(({\langle\rangle}{\exists}aNa\backslash Sb)/{\exists}aNa), N2\ \Rightarrow\ Sf
\using {\forall}L
\endprooftree
\justifies
[Nt(s(m))], \mbox{\fbox{${\blacksquare}{\forall}a(({\langle\rangle}Na\backslash Sf)/({\langle\rangle}Na\backslash Sb))$}}, {\square}(({\langle\rangle}{\exists}aNa\backslash Sb)/{\exists}aNa), N2\ \Rightarrow\ Sf
\using {\blacksquare}L
\endprooftree
\justifies
[Nt(s(m))], {\blacksquare}{\forall}a(({\langle\rangle}Na\backslash Sf)/({\langle\rangle}Na\backslash Sb)), {\square}(({\langle\rangle}{\exists}aNa\backslash Sb)/{\exists}aNa), {\exists}bNb\ \Rightarrow\ Sf
\using {\exists}L
\endprooftree
\justifies
{\langle\rangle}Nt(s(m)), {\blacksquare}{\forall}a(({\langle\rangle}Na\backslash Sf)/({\langle\rangle}Na\backslash Sb)), {\square}(({\langle\rangle}{\exists}aNa\backslash Sb)/{\exists}aNa), {\exists}bNb\ \Rightarrow\ Sf
\using {\langle\rangle}L
\endprooftree
\justifies
{\blacksquare}{\forall}a(({\langle\rangle}Na\backslash Sf)/({\langle\rangle}Na\backslash Sb)), {\square}(({\langle\rangle}{\exists}aNa\backslash Sb)/{\exists}aNa), {\exists}bNb\ \Rightarrow\ {\langle\rangle}Nt(s(m))\backslash Sf
\using {\backslash}R
\endprooftree
\justifies
{\blacksquare}{\forall}a(({\langle\rangle}Na\backslash Sf)/({\langle\rangle}Na\backslash Sb)), {\square}(({\langle\rangle}{\exists}aNa\backslash Sb)/{\exists}aNa)\ \Rightarrow\ ({\langle\rangle}Nt(s(m))\backslash Sf)/{\exists}bNb
\using {/}R
\endprooftree
\justifies
\begin{array}{c}
{\blacksquare}{\forall}a(({\langle\rangle}Na\backslash Sf)/({\langle\rangle}Na\backslash Sb)), {\square}(({\langle\rangle}{\exists}aNa\backslash Sb)/{\exists}aNa)\ \Rightarrow\ {\blacksquare}(({\langle\rangle}Nt(s(m))\backslash Sf)/{\exists}bNb)\\
\mbox{\footnotesize\textcircled{1}}
\end{array}
\using {\blacksquare}R
\endprooftree
\prooftree
\prooftree
\prooftree
\prooftree
\prooftree
\prooftree
\prooftree
\prooftree
\prooftree
\prooftree
\prooftree
\justifies
N3\ \Rightarrow\ N3
\endprooftree
\justifies
N3\ \Rightarrow\ \fbox{${\exists}aNa$}
\using {\exists}R
\endprooftree
\prooftree
\prooftree
\prooftree
\prooftree
\justifies
Nt(s(m))\ \Rightarrow\ Nt(s(m))
\endprooftree
\justifies
Nt(s(m))\ \Rightarrow\ \fbox{${\exists}gNt(s(g))$}
\using {\exists}R
\endprooftree
\justifies
[Nt(s(m))]\ \Rightarrow\ \fbox{${\langle\rangle}{\exists}gNt(s(g))$}
\using {\langle\rangle}R
\endprooftree
\prooftree
\justifies
\mbox{\fbox{$Sf$}}\ \Rightarrow\ Sf
\endprooftree
\justifies
[Nt(s(m))], \mbox{\fbox{${\langle\rangle}{\exists}gNt(s(g))\backslash Sf$}}\ \Rightarrow\ Sf
\using {\backslash}L
\endprooftree
\justifies
[Nt(s(m))], \mbox{\fbox{$({\langle\rangle}{\exists}gNt(s(g))\backslash Sf)/{\exists}aNa$}}, N3\ \Rightarrow\ Sf
\using {/}L
\endprooftree
\justifies
[Nt(s(m))], \mbox{\fbox{${\square}(({\langle\rangle}{\exists}gNt(s(g))\backslash Sf)/{\exists}aNa)$}}, N3\ \Rightarrow\ Sf
\using {\Box}L
\endprooftree
\justifies
[Nt(s(m))], {\square}(({\langle\rangle}{\exists}gNt(s(g))\backslash Sf)/{\exists}aNa), {\exists}bNb\ \Rightarrow\ Sf
\using {\exists}L
\endprooftree
\justifies
{\langle\rangle}Nt(s(m)), {\square}(({\langle\rangle}{\exists}gNt(s(g))\backslash Sf)/{\exists}aNa), {\exists}bNb\ \Rightarrow\ Sf
\using {\langle\rangle}L
\endprooftree
\justifies
{\square}(({\langle\rangle}{\exists}gNt(s(g))\backslash Sf)/{\exists}aNa), {\exists}bNb\ \Rightarrow\ {\langle\rangle}Nt(s(m))\backslash Sf
\using {\backslash}R
\endprooftree
\justifies
{\square}(({\langle\rangle}{\exists}gNt(s(g))\backslash Sf)/{\exists}aNa)\ \Rightarrow\ ({\langle\rangle}Nt(s(m))\backslash Sf)/{\exists}bNb
\using {/}R
\endprooftree
\justifies
{\square}(({\langle\rangle}{\exists}gNt(s(g))\backslash Sf)/{\exists}aNa)\ \Rightarrow\ {\blacksquare}(({\langle\rangle}Nt(s(m))\backslash Sf)/{\exists}bNb)
\using {\blacksquare}R
\endprooftree
\justifies
\begin{array}{c}
{\square}(({\langle\rangle}{\exists}gNt(s(g))\backslash Sf)/{\exists}aNa)\ \Rightarrow\ \fbox{$?{\blacksquare}(({\langle\rangle}Nt(s(m))\backslash Sf)/{\exists}bNb)$}\\
\mbox{\footnotesize\textcircled{2}}
\end{array}
\using {?}R
\endprooftree
\prooftree
\prooftree
\prooftree
\prooftree
\prooftree
\prooftree
\justifies
\mbox{\fbox{$Nt(s(n))$}}\ \Rightarrow\ Nt(s(n))
\endprooftree
\justifies
\mbox{\fbox{${\blacksquare}Nt(s(n))$}}\ \Rightarrow\ Nt(s(n))
\using {\blacksquare}L
\endprooftree
\justifies
{\blacksquare}Nt(s(n))\ \Rightarrow\ \fbox{${\exists}bNb$}
\using {\exists}R
\endprooftree
\prooftree
\prooftree
\prooftree
\prooftree
\justifies
\mbox{\fbox{$Nt(s(m))$}}\ \Rightarrow\ Nt(s(m))
\endprooftree
\justifies
\mbox{\fbox{${\blacksquare}Nt(s(m))$}}\ \Rightarrow\ Nt(s(m))
\using {\blacksquare}L
\endprooftree
\justifies
[{\blacksquare}Nt(s(m))]\ \Rightarrow\ \fbox{${\langle\rangle}Nt(s(m))$}
\using {\langle\rangle}R
\endprooftree
\prooftree
\justifies
\mbox{\fbox{$Sf$}}\ \Rightarrow\ Sf
\endprooftree
\justifies
[{\blacksquare}Nt(s(m))], \mbox{\fbox{${\langle\rangle}Nt(s(m))\backslash Sf$}}\ \Rightarrow\ Sf
\using {\backslash}L
\endprooftree
\justifies
[{\blacksquare}Nt(s(m))], \mbox{\fbox{$({\langle\rangle}Nt(s(m))\backslash Sf)/{\exists}bNb$}}, {\blacksquare}Nt(s(n))\ \Rightarrow\ Sf
\using {/}L
\endprooftree
\justifies
[{\blacksquare}Nt(s(m))], [\mbox{\fbox{${[]^{-1}}(({\langle\rangle}Nt(s(m))\backslash Sf)/{\exists}bNb)$}}], {\blacksquare}Nt(s(n))\ \Rightarrow\ Sf
\using {[]^{-1}}L
\endprooftree
\justifies
\begin{array}{c}
[{\blacksquare}Nt(s(m))], [[\mbox{\fbox{${[]^{-1}}{[]^{-1}}(({\langle\rangle}Nt(s(m))\backslash Sf)/{\exists}bNb)$}}]], {\blacksquare}Nt(s(n))\ \Rightarrow\ Sf\\
\mbox{\footnotesize\textcircled{3}}
\end{array}
\using {[]^{-1}}L
\endprooftree
\justifies
[{\blacksquare}Nt(s(m))], [[{\square}(({\langle\rangle}{\exists}gNt(s(g))\backslash Sf)/{\exists}aNa), \mbox{\fbox{$?{\blacksquare}(({\langle\rangle}Nt(s(m))\backslash Sf)/{\exists}bNb)\backslash {[]^{-1}}{[]^{-1}}(({\langle\rangle}Nt(s(m))\backslash Sf)/{\exists}bNb)$}}]], {\blacksquare}Nt(s(n))\ \Rightarrow\ Sf
\using {\backslash}L
\endprooftree
\justifies
[{\blacksquare}Nt(s(m))], [[{\square}(({\langle\rangle}{\exists}gNt(s(g))\backslash Sf)/{\exists}aNa), \mbox{\fbox{$(?{\blacksquare}(({\langle\rangle}Nt(s(m))\backslash Sf)/{\exists}bNb)\backslash {[]^{-1}}{[]^{-1}}(({\langle\rangle}Nt(s(m))\backslash Sf)/{\exists}bNb))/{\blacksquare}(({\langle\rangle}Nt(s(m))\backslash Sf)/{\exists}bNb)$}}, {\blacksquare}{\forall}a(({\langle\rangle}Na\backslash Sf)/({\langle\rangle}Na\backslash Sb)), {\square}(({\langle\rangle}{\exists}aNa\backslash Sb)/{\exists}aNa)]], {\blacksquare}Nt(s(n))\ \Rightarrow\ Sf
\using {/}L
\endprooftree
\justifies
[{\blacksquare}Nt(s(m))], [[{\square}(({\langle\rangle}{\exists}gNt(s(g))\backslash Sf)/{\exists}aNa), \mbox{\fbox{${\forall}a((?{\blacksquare}(({\langle\rangle}Na\backslash Sf)/{\exists}bNb)\backslash {[]^{-1}}{[]^{-1}}(({\langle\rangle}Na\backslash Sf)/{\exists}bNb))/{\blacksquare}(({\langle\rangle}Na\backslash Sf)/{\exists}bNb))$}}, {\blacksquare}{\forall}a(({\langle\rangle}Na\backslash Sf)/({\langle\rangle}Na\backslash Sb)), {\square}(({\langle\rangle}{\exists}aNa\backslash Sb)/{\exists}aNa)]], {\blacksquare}Nt(s(n))\ \Rightarrow\ Sf
\using {\forall}L
\endprooftree
\justifies
[{\blacksquare}Nt(s(m))], [[{\square}(({\langle\rangle}{\exists}gNt(s(g))\backslash Sf)/{\exists}aNa), \mbox{\fbox{${\forall}f{\forall}a((?{\blacksquare}(({\langle\rangle}Na\backslash Sf)/{\exists}bNb)\backslash {[]^{-1}}{[]^{-1}}(({\langle\rangle}Na\backslash Sf)/{\exists}bNb))/{\blacksquare}(({\langle\rangle}Na\backslash Sf)/{\exists}bNb))$}}, {\blacksquare}{\forall}a(({\langle\rangle}Na\backslash Sf)/({\langle\rangle}Na\backslash Sb)), {\square}(({\langle\rangle}{\exists}aNa\backslash Sb)/{\exists}aNa)]], {\blacksquare}Nt(s(n))\ \Rightarrow\ Sf
\using {\forall}L
\endprooftree
\justifies
[{\blacksquare}Nt(s(m))], [[{\square}(({\langle\rangle}{\exists}gNt(s(g))\backslash Sf)/{\exists}aNa), \mbox{\fbox{${\blacksquare}{\forall}f{\forall}a((?{\blacksquare}(({\langle\rangle}Na\backslash Sf)/{\exists}bNb)\backslash {[]^{-1}}{[]^{-1}}(({\langle\rangle}Na\backslash Sf)/{\exists}bNb))/{\blacksquare}(({\langle\rangle}Na\backslash Sf)/{\exists}bNb))$}}, {\blacksquare}{\forall}a(({\langle\rangle}Na\backslash Sf)/({\langle\rangle}Na\backslash Sb)), {\square}(({\langle\rangle}{\exists}aNa\backslash Sb)/{\exists}aNa)]], {\blacksquare}Nt(s(n))\ \Rightarrow\ Sf
\using {\blacksquare}L
\endprooftree}
$$
\vspace{0.15in}
\noindent
The coordination combinator semantics is such that
\disp{
$\begin{array}[t]{l}
(((((\Phinplus\ (s\ (s\ 0))\ {\it and})\ x)\ [y])\ z)\ w) =\\
((((\Phinplus\ (s\ 0)\ {\it and})\ (x\ z))\ (\alphaplus\ [y]\ z))\ w) =\\
((((\Phinplus\ (s\ 0)\ {\it and})\ (x\ z))\ [(y\ z)])\ w) =\\
(((\Phinplus\ 0\ {\it and})\ ((x\ z)\ w))\ (\alphaplus\ [(y\ z)]\ w)) =\\
(((\Phinplus\ 0\ {\it and})\ ((x\ z)\ w))\ [((y\ z)\ w)]) =\\
{}[((y\ z)\ w)\wedge((x\ z)\ w)]
\end{array}$}
All this delivers semantics:
\disp{
$[({\it Pres}\ ((\mbox{\v{}}{\it like}\ {\it l})\ {\it j}))\wedge ({\it Fut}\ ((\mbox{\v{}}{\it love}\ {\it l})\ {\it j}))]$}

\subsection{Ditransitive verb coordination}

The next example is of ditransitive verb disjunction:
\disp{
$[{\bf john}]{+}[[{\bf gave}{+}{\bf or}{+}{\bf sent}]]{+}{\bf mary}{+}{\bf the}{+}{\bf book}: Sf$}
Appropriate lexical lookup yields the semantically labelled sequent where the coordinator type is
essentially $(\exstexp X\bsl\abrack\abrack X)/X$ with $X=((N\bsl S)/N)/N$
(using the curried ditransitive verb type):
\disp{
$\begin{array}[t]{l}
[{\blacksquare}Nt(s(m)): {\it j}], [[{\square}((({\langle\rangle}{\exists}aNa\backslash Sf)/{\exists}aNa)/{\exists}aNa): \mbox{\^{}}\lambda A\lambda B\lambda C({\it Past}\ (((\mbox{\v{}}{\it give}\ {\it A})\ {\it B})\ {\it C})),\\
 {\blacksquare}{\forall}a{\forall}f((?{\blacksquare}((({\langle\rangle}Na\backslash Sf)/{\exists}bNb)/{\exists}bNb)\backslash {[]^{-1}}{[]^{-1}}((({\langle\rangle}Na\backslash Sf)/{\exists}bNb)/{\exists}bNb))/\\{\blacksquare}((({\langle\rangle}Na\backslash Sf)/{\exists}bNb)/{\exists}bNb)): (\Phinplus\ ({\it s}\ ({\it s}\ ({\it s}\ {\it 0})))\ {\it or}), {\square}((({\langle\rangle}{\exists}aNa\backslash Sf)/{\exists}aNa)/{\exists}aNa): \\\mbox{\^{}}\lambda D\lambda E\lambda F({\it Past}\ (((\mbox{\v{}}{\it send}\ {\it D})\ {\it E})\ {\it F}))]],{\blacksquare}Nt(s(f)): {\it m}, {\blacksquare}{\forall}n(Nt(n)/{\it CN}{\it n}): \iota,\\ {\square}{\it CN}{\it s(n)}: {\it book}\ \Rightarrow\ Sf
 \end{array}$}
There is the derivation:
\vspace{0.15in}
$${\tiny
\prooftree
\prooftree
\prooftree
\prooftree
\prooftree
\prooftree
\prooftree
\prooftree
\prooftree
\prooftree
\prooftree
\justifies
N1\ \Rightarrow\ N1
\endprooftree
\justifies
N1\ \Rightarrow\ \fbox{${\exists}aNa$}
\using {\exists}R
\endprooftree
\prooftree
\prooftree
\prooftree
\justifies
N2\ \Rightarrow\ N2
\endprooftree
\justifies
N2\ \Rightarrow\ \fbox{${\exists}aNa$}
\using {\exists}R
\endprooftree
\prooftree
\prooftree
\prooftree
\prooftree
\justifies
Nt(s(m))\ \Rightarrow\ Nt(s(m))
\endprooftree
\justifies
Nt(s(m))\ \Rightarrow\ \fbox{${\exists}aNa$}
\using {\exists}R
\endprooftree
\justifies
[Nt(s(m))]\ \Rightarrow\ \fbox{${\langle\rangle}{\exists}aNa$}
\using {\langle\rangle}R
\endprooftree
\prooftree
\justifies
\mbox{\fbox{$Sf$}}\ \Rightarrow\ Sf
\endprooftree
\justifies
[Nt(s(m))], \mbox{\fbox{${\langle\rangle}{\exists}aNa\backslash Sf$}}\ \Rightarrow\ Sf
\using {\backslash}L
\endprooftree
\justifies
[Nt(s(m))], \mbox{\fbox{$({\langle\rangle}{\exists}aNa\backslash Sf)/{\exists}aNa$}}, N2\ \Rightarrow\ Sf
\using {/}L
\endprooftree
\justifies
[Nt(s(m))], \mbox{\fbox{$(({\langle\rangle}{\exists}aNa\backslash Sf)/{\exists}aNa)/{\exists}aNa$}}, N1, N2\ \Rightarrow\ Sf
\using {/}L
\endprooftree
\justifies
[Nt(s(m))], \mbox{\fbox{${\square}((({\langle\rangle}{\exists}aNa\backslash Sf)/{\exists}aNa)/{\exists}aNa)$}}, N1, N2\ \Rightarrow\ Sf
\using {\Box}L
\endprooftree
\justifies
[Nt(s(m))], {\square}((({\langle\rangle}{\exists}aNa\backslash Sf)/{\exists}aNa)/{\exists}aNa), N1, {\exists}bNb\ \Rightarrow\ Sf
\using {\exists}L
\endprooftree
\justifies
[Nt(s(m))], {\square}((({\langle\rangle}{\exists}aNa\backslash Sf)/{\exists}aNa)/{\exists}aNa), {\exists}bNb, {\exists}bNb\ \Rightarrow\ Sf
\using {\exists}L
\endprooftree
\justifies
{\langle\rangle}Nt(s(m)), {\square}((({\langle\rangle}{\exists}aNa\backslash Sf)/{\exists}aNa)/{\exists}aNa), {\exists}bNb, {\exists}bNb\ \Rightarrow\ Sf
\using {\langle\rangle}L
\endprooftree
\justifies
{\square}((({\langle\rangle}{\exists}aNa\backslash Sf)/{\exists}aNa)/{\exists}aNa), {\exists}bNb, {\exists}bNb\ \Rightarrow\ {\langle\rangle}Nt(s(m))\backslash Sf
\using {\backslash}R
\endprooftree
\justifies
{\square}((({\langle\rangle}{\exists}aNa\backslash Sf)/{\exists}aNa)/{\exists}aNa), {\exists}bNb\ \Rightarrow\ ({\langle\rangle}Nt(s(m))\backslash Sf)/{\exists}bNb
\using {/}R
\endprooftree
\justifies
{\square}((({\langle\rangle}{\exists}aNa\backslash Sf)/{\exists}aNa)/{\exists}aNa)\ \Rightarrow\ (({\langle\rangle}Nt(s(m))\backslash Sf)/{\exists}bNb)/{\exists}bNb
\using {/}R
\endprooftree
\justifies
\begin{array}{c}
{\square}((({\langle\rangle}{\exists}aNa\backslash Sf)/{\exists}aNa)/{\exists}aNa)\ \Rightarrow\ {\blacksquare}((({\langle\rangle}Nt(s(m))\backslash Sf)/{\exists}bNb)/{\exists}bNb)\\
\mbox{\footnotesize\textcircled{1}}
\end{array}
\using {\blacksquare}R
\endprooftree
\prooftree
\prooftree
\vdots
\justifies
{\square}((({\langle\rangle}{\exists}aNa\backslash Sf)/{\exists}aNa)/{\exists}aNa)\ \Rightarrow\ {\blacksquare}((({\langle\rangle}Nt(s(m))\backslash Sf)/{\exists}bNb)/{\exists}bNb)
\using {\blacksquare}R
\endprooftree
\justifies
\begin{array}{c}
{\square}((({\langle\rangle}{\exists}aNa\backslash Sf)/{\exists}aNa)/{\exists}aNa)\ \Rightarrow\ \fbox{$?{\blacksquare}((({\langle\rangle}Nt(s(m))\backslash Sf)/{\exists}bNb)/{\exists}bNb)$}\\
\mbox{\footnotesize\textcircled{2}}
\end{array}
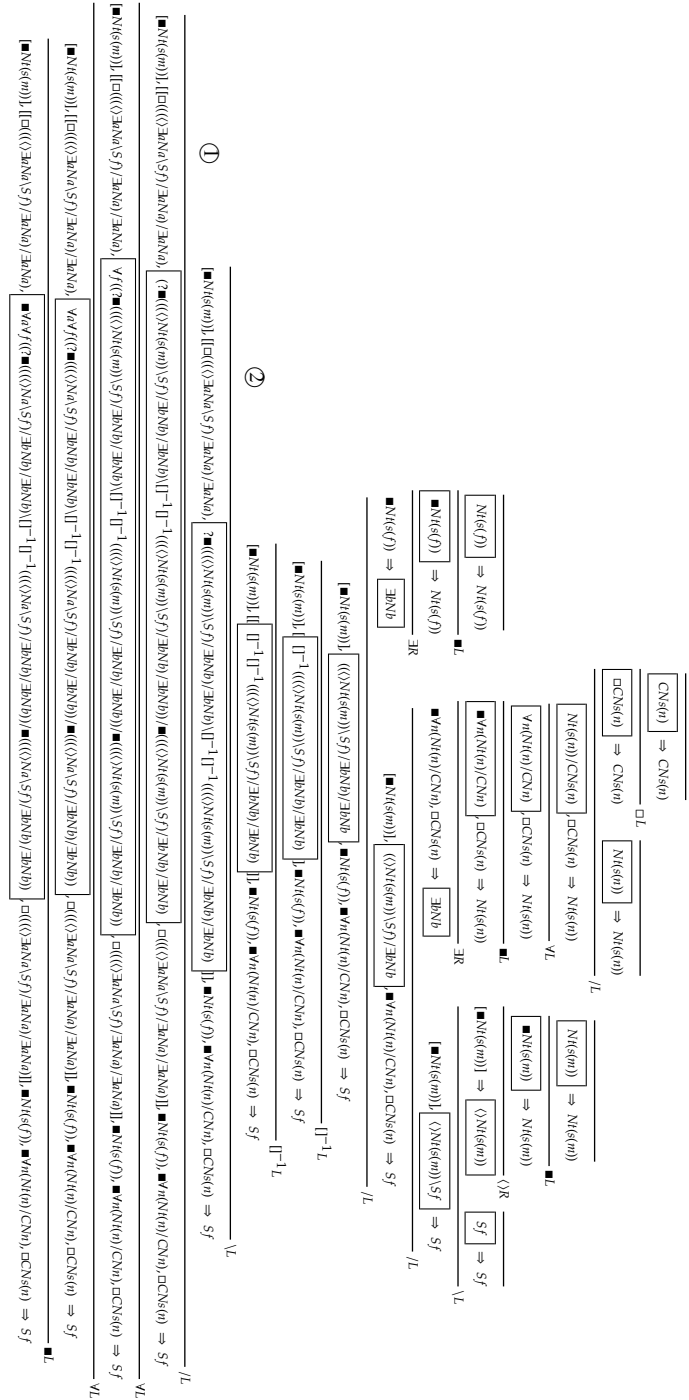
\using {?}R
\endprooftree}
$$

\begin{center}
\rotatebox{-90}{\tiny
\prooftree
\prooftree
\prooftree
\prooftree
\mbox{\footnotesize\textcircled{1}}\tab
\prooftree
\mbox{\footnotesize\textcircled{2}}\tab
\prooftree
\prooftree
\prooftree
\prooftree
\prooftree
\prooftree
\justifies
\mbox{\fbox{$Nt(s(f))$}}\ \Rightarrow\ Nt(s(f))
\endprooftree
\justifies
\mbox{\fbox{${\blacksquare}Nt(s(f))$}}\ \Rightarrow\ Nt(s(f))
\using {\blacksquare}L
\endprooftree
\justifies
{\blacksquare}Nt(s(f))\ \Rightarrow\ \fbox{${\exists}bNb$}
\using {\exists}R
\endprooftree
\prooftree
\prooftree
\prooftree
\prooftree
\prooftree
\prooftree
\prooftree
\justifies
\mbox{\fbox{${\it CN}{\it s(n)}$}}\ \Rightarrow\ {\it CN}{\it s(n)}
\endprooftree
\justifies
\mbox{\fbox{${\square}{\it CN}{\it s(n)}$}}\ \Rightarrow\ {\it CN}{\it s(n)}
\using {\Box}L
\endprooftree
\prooftree
\justifies
\mbox{\fbox{$Nt(s(n))$}}\ \Rightarrow\ Nt(s(n))
\endprooftree
\justifies
\mbox{\fbox{$Nt(s(n))/{\it CN}{\it s(n)}$}}, {\square}{\it CN}{\it s(n)}\ \Rightarrow\ Nt(s(n))
\using {/}L
\endprooftree
\justifies
\mbox{\fbox{${\forall}n(Nt(n)/{\it CN}{\it n})$}}, {\square}{\it CN}{\it s(n)}\ \Rightarrow\ Nt(s(n))
\using {\forall}L
\endprooftree
\justifies
\mbox{\fbox{${\blacksquare}{\forall}n(Nt(n)/{\it CN}{\it n})$}}, {\square}{\it CN}{\it s(n)}\ \Rightarrow\ Nt(s(n))
\using {\blacksquare}L
\endprooftree
\justifies
{\blacksquare}{\forall}n(Nt(n)/{\it CN}{\it n}), {\square}{\it CN}{\it s(n)}\ \Rightarrow\ \fbox{${\exists}bNb$}
\using {\exists}R
\endprooftree
\prooftree
\prooftree
\prooftree
\prooftree
\justifies
\mbox{\fbox{$Nt(s(m))$}}\ \Rightarrow\ Nt(s(m))
\endprooftree
\justifies
\mbox{\fbox{${\blacksquare}Nt(s(m))$}}\ \Rightarrow\ Nt(s(m))
\using {\blacksquare}L
\endprooftree
\justifies
[{\blacksquare}Nt(s(m))]\ \Rightarrow\ \fbox{${\langle\rangle}Nt(s(m))$}
\using {\langle\rangle}R
\endprooftree
\prooftree
\justifies
\mbox{\fbox{$Sf$}}\ \Rightarrow\ Sf
\endprooftree
\justifies
[{\blacksquare}Nt(s(m))], \mbox{\fbox{${\langle\rangle}Nt(s(m))\backslash Sf$}}\ \Rightarrow\ Sf
\using {\backslash}L
\endprooftree
\justifies
[{\blacksquare}Nt(s(m))], \mbox{\fbox{$({\langle\rangle}Nt(s(m))\backslash Sf)/{\exists}bNb$}}, {\blacksquare}{\forall}n(Nt(n)/{\it CN}{\it n}), {\square}{\it CN}{\it s(n)}\ \Rightarrow\ Sf
\using {/}L
\endprooftree
\justifies
[{\blacksquare}Nt(s(m))], \mbox{\fbox{$(({\langle\rangle}Nt(s(m))\backslash Sf)/{\exists}bNb)/{\exists}bNb$}}, {\blacksquare}Nt(s(f)), {\blacksquare}{\forall}n(Nt(n)/{\it CN}{\it n}), {\square}{\it CN}{\it s(n)}\ \Rightarrow\ Sf
\using {/}L
\endprooftree
\justifies
[{\blacksquare}Nt(s(m))], [\mbox{\fbox{${[]^{-1}}((({\langle\rangle}Nt(s(m))\backslash Sf)/{\exists}bNb)/{\exists}bNb)$}}], {\blacksquare}Nt(s(f)), {\blacksquare}{\forall}n(Nt(n)/{\it CN}{\it n}), {\square}{\it CN}{\it s(n)}\ \Rightarrow\ Sf
\using {[]^{-1}}L
\endprooftree
\justifies
[{\blacksquare}Nt(s(m))], [[\mbox{\fbox{${[]^{-1}}{[]^{-1}}((({\langle\rangle}Nt(s(m))\backslash Sf)/{\exists}bNb)/{\exists}bNb)$}}]], {\blacksquare}Nt(s(f)), {\blacksquare}{\forall}n(Nt(n)/{\it CN}{\it n}), {\square}{\it CN}{\it s(n)}\ \Rightarrow\ Sf
\using {[]^{-1}}L
\endprooftree\justifies
[{\blacksquare}Nt(s(m))], [[{\square}((({\langle\rangle}{\exists}aNa\backslash Sf)/{\exists}aNa)/{\exists}aNa), \mbox{\fbox{$?{\blacksquare}((({\langle\rangle}Nt(s(m))\backslash Sf)/{\exists}bNb)/{\exists}bNb)\backslash {[]^{-1}}{[]^{-1}}((({\langle\rangle}Nt(s(m))\backslash Sf)/{\exists}bNb)/{\exists}bNb)$}}]], {\blacksquare}Nt(s(f)), {\blacksquare}{\forall}n(Nt(n)/{\it CN}{\it n}), {\square}{\it CN}{\it s(n)}\ \Rightarrow\ Sf
\using {\backslash}L
\endprooftree
\justifies
[{\blacksquare}Nt(s(m))], [[{\square}((({\langle\rangle}{\exists}aNa\backslash Sf)/{\exists}aNa)/{\exists}aNa), \mbox{\fbox{$(?{\blacksquare}((({\langle\rangle}Nt(s(m))\backslash Sf)/{\exists}bNb)/{\exists}bNb)\backslash {[]^{-1}}{[]^{-1}}((({\langle\rangle}Nt(s(m))\backslash Sf)/{\exists}bNb)/{\exists}bNb))/{\blacksquare}((({\langle\rangle}Nt(s(m))\backslash Sf)/{\exists}bNb)/{\exists}bNb)$}}, {\square}((({\langle\rangle}{\exists}aNa\backslash Sf)/{\exists}aNa)/{\exists}aNa)]], {\blacksquare}Nt(s(f)), {\blacksquare}{\forall}n(Nt(n)/{\it CN}{\it n}), {\square}{\it CN}{\it s(n)}\ \Rightarrow\ Sf
\using {/}L
\endprooftree
\justifies
[{\blacksquare}Nt(s(m))], [[{\square}((({\langle\rangle}{\exists}aNa\backslash Sf)/{\exists}aNa)/{\exists}aNa), \mbox{\fbox{${\forall}f((?{\blacksquare}((({\langle\rangle}Nt(s(m))\backslash Sf)/{\exists}bNb)/{\exists}bNb)\backslash {[]^{-1}}{[]^{-1}}((({\langle\rangle}Nt(s(m))\backslash Sf)/{\exists}bNb)/{\exists}bNb))/{\blacksquare}((({\langle\rangle}Nt(s(m))\backslash Sf)/{\exists}bNb)/{\exists}bNb))$}}, {\square}((({\langle\rangle}{\exists}aNa\backslash Sf)/{\exists}aNa)/{\exists}aNa)]], {\blacksquare}Nt(s(f)), {\blacksquare}{\forall}n(Nt(n)/{\it CN}{\it n}), {\square}{\it CN}{\it s(n)}\ \Rightarrow\ Sf
\using {\forall}L
\endprooftree
\justifies
[{\blacksquare}Nt(s(m))], [[{\square}((({\langle\rangle}{\exists}aNa\backslash Sf)/{\exists}aNa)/{\exists}aNa), \mbox{\fbox{${\forall}a{\forall}f((?{\blacksquare}((({\langle\rangle}Na\backslash Sf)/{\exists}bNb)/{\exists}bNb)\backslash {[]^{-1}}{[]^{-1}}((({\langle\rangle}Na\backslash Sf)/{\exists}bNb)/{\exists}bNb))/{\blacksquare}((({\langle\rangle}Na\backslash Sf)/{\exists}bNb)/{\exists}bNb))$}}, {\square}((({\langle\rangle}{\exists}aNa\backslash Sf)/{\exists}aNa)/{\exists}aNa)]], {\blacksquare}Nt(s(f)), {\blacksquare}{\forall}n(Nt(n)/{\it CN}{\it n}), {\square}{\it CN}{\it s(n)}\ \Rightarrow\ Sf
\using {\forall}L
\endprooftree
\justifies
[{\blacksquare}Nt(s(m))], [[{\square}((({\langle\rangle}{\exists}aNa\backslash Sf)/{\exists}aNa)/{\exists}aNa), \mbox{\fbox{${\blacksquare}{\forall}a{\forall}f((?{\blacksquare}((({\langle\rangle}Na\backslash Sf)/{\exists}bNb)/{\exists}bNb)\backslash {[]^{-1}}{[]^{-1}}((({\langle\rangle}Na\backslash Sf)/{\exists}bNb)/{\exists}bNb))/{\blacksquare}((({\langle\rangle}Na\backslash Sf)/{\exists}bNb)/{\exists}bNb))$}}, {\square}((({\langle\rangle}{\exists}aNa\backslash Sf)/{\exists}aNa)/{\exists}aNa)]], {\blacksquare}Nt(s(f)), {\blacksquare}{\forall}n(Nt(n)/{\it CN}{\it n}), {\square}{\it CN}{\it s(n)}\ \Rightarrow\ Sf
\using {\blacksquare}L
\endprooftree}
\end{center}

\vspace{0.15in}

\noindent
As usual the coordinator modality is removed and the subject agreement and
tense feature are instantiated at the root.
Then the subderivations \textcircled{1} and \textcircled{2} derive the
right and left conjuncts respectively by essentially
deriving the identity TTV yields TTV on the ditransitive verb type TTV:
they are the same except for the initial $?R$ in \textcircled{2},
hence the latter is elided.
The remaining context derivation checks the double bracketing as usual
and for the rest amounts to analysis of a sequence S-TTV-O1-O2
where S and O1 are proper names and O2 is a definite noun phrase.

The example is interesting in that it illustrates coordination of arity three.
The coordination combinator semantics is such that:
\disp{
$\begin{array}[t]{l}
((((((\Phinplus\ (s\ (s\ (s\ 0)))\ {\it or})\ x)\ [y])\ z)\ w)\ u) =\\
(((((\Phinplus\ (s\ (s\ 0))\ {\it or})\ (x\ z))\ (\alphaplus\ [y]\ z))\ w)\ u) =\\
(((((\Phinplus\ (s\ (s\ 0))\ {\it or})\ (x\ z))\ [(y\ z)])\ w)\ u) =\\
((((\Phinplus\ (s\ 0)\ {\it or})\ ((x\ z)\ w))\ (\alphaplus\ [(y\ z)]\ w))\ u) =\\
((((\Phinplus\ (s\ 0)\ {\it or})\ ((x\ z)\ w))\ [((y\ z)\ w])\ u) =\\
(((\Phinplus\ 0\ {\it or})\ (((x\ z)\ w)\ u))\ (\alphaplus\ [((y\ z)\ w)]\ u)) =\\
(((\Phinplus\ 0\ {\it or})\ (((x\ z)\ w)\ u))\ [(((y\ z)\ w)\ u)]) =\\
{}[(((y\ z)\ w)\ u)\wedge(((x\ z)\ w)\ u)]
\end{array}$}
All this delivers semantics:
\disp{
$[({\it Past}\ (((\mbox{\v{}}{\it give}\ {\it m})\ (\iota \ \mbox{\v{}}{\it book}))\ {\it j}))\vee ({\it Past}\ (((\mbox{\v{}}{\it send}\ {\it m})\ (\iota \ \mbox{\v{}}{\it book}))\ {\it j}))]$}

\subsection{Subject coordination}

We continue with subject disjunction:
\disp{
$[[[{\bf john}]{+}{\bf or}{+}[{\bf mary}]]]{+}{\bf sings}: Sf$}
Appropriate lexical lookup inserts a coordinator over lifted subject noun phrases
(cf. Montague 1973\cite{montague:ptq}), 
essentially $(\exstexp X\bsl\abrack\abrack X)/X$ with $X=$ $S/(N\bsl S)$.
\disp{
$\begin{array}[t]{l}
[[[{\blacksquare}Nt(s(m)): {\it j}],
{\blacksquare}{\forall}f((?{\blacksquare}(Sf/({\langle\rangle}{\exists}gNt(s(g))\backslash Sf))\backslash {[]^{-1}}{[]^{-1}}(Sf/({\langle\rangle}{\exists}gNt(s(g))\backslash Sf)))/\\{\blacksquare}(Sf/({\langle\rangle}{\exists}gNt(s(g))\backslash Sf))): (\Phinplus\ ({\it s}\ {\it 0})\ {\it or}), [{\blacksquare}Nt(s(f)): {\it m}]]],\\ {\square}({\langle\rangle}{\exists}gNt(s(g))\backslash Sf): \mbox{\^{}}\lambda A({\it Pres}\ (\mbox{\v{}}{\it sing}\ {\it A}))\ \Rightarrow\ Sf
\end{array}$}
There is the derivation:

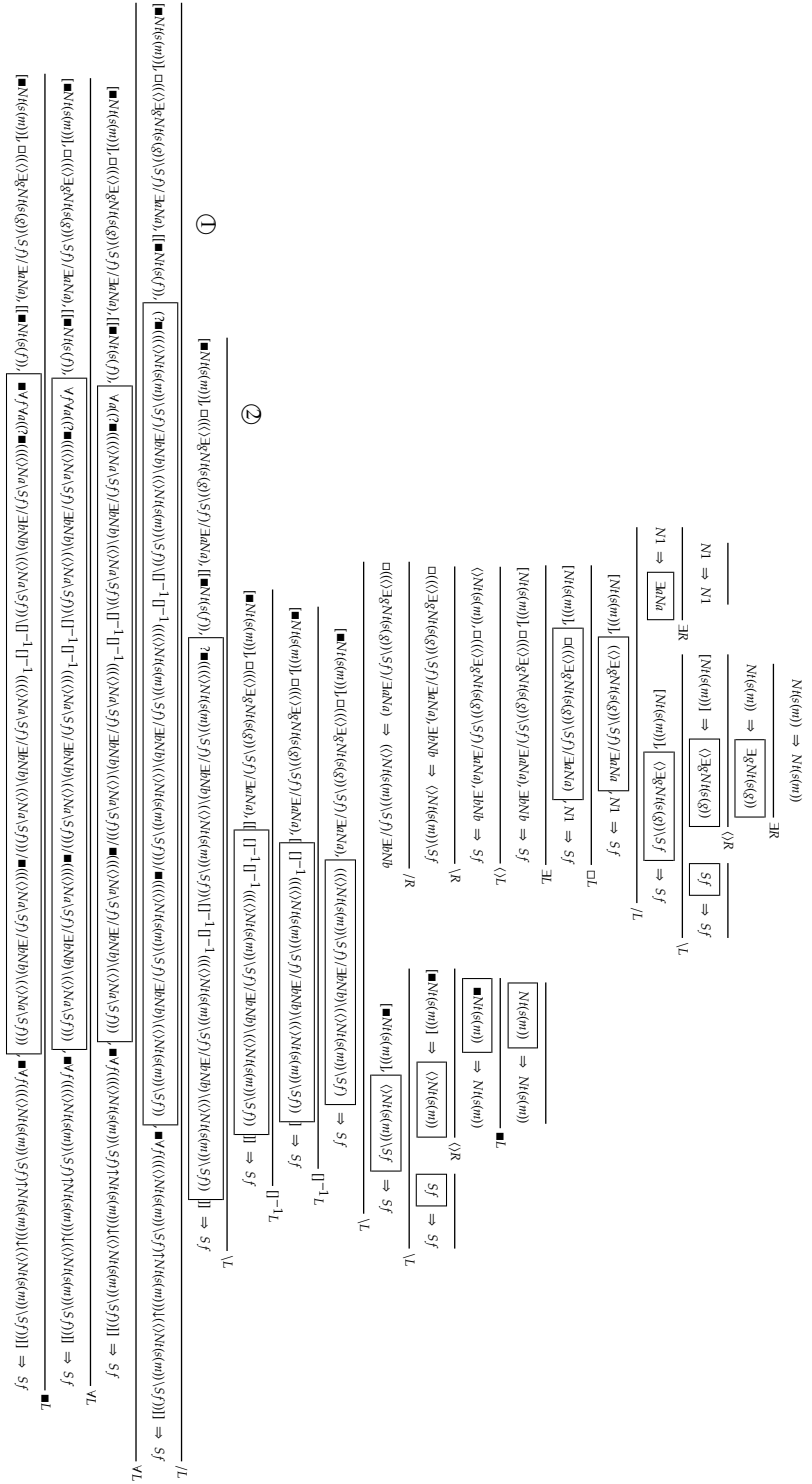
\vspace{0.15in}

\begin{center}
\rotatebox{-90}{\tiny
\prooftree
\prooftree
\prooftree
\prooftree
\prooftree
\prooftree
\prooftree
\prooftree
\prooftree
\prooftree
\justifies
\mbox{\fbox{$Nt(s(f))$}}\ \Rightarrow\ Nt(s(f))
\endprooftree
\justifies
\mbox{\fbox{${\blacksquare}Nt(s(f))$}}\ \Rightarrow\ Nt(s(f))
\using {\blacksquare}L
\endprooftree
\justifies
{\blacksquare}Nt(s(f))\ \Rightarrow\ \fbox{${\exists}gNt(s(g))$}
\using {\exists}R
\endprooftree
\justifies
[{\blacksquare}Nt(s(f))]\ \Rightarrow\ \fbox{${\langle\rangle}{\exists}gNt(s(g))$}
\using {\langle\rangle}R
\endprooftree
\prooftree
\justifies
\mbox{\fbox{$Sf$}}\ \Rightarrow\ Sf
\endprooftree
\justifies
[{\blacksquare}Nt(s(f))], \mbox{\fbox{${\langle\rangle}{\exists}gNt(s(g))\backslash Sf$}}\ \Rightarrow\ Sf
\using {\backslash}L
\endprooftree
\justifies
[{\blacksquare}Nt(s(f))]\ \Rightarrow\ Sf/({\langle\rangle}{\exists}gNt(s(g))\backslash Sf)
\using {/}R
\endprooftree
\justifies
[{\blacksquare}Nt(s(f))]\ \Rightarrow\ {\blacksquare}(Sf/({\langle\rangle}{\exists}gNt(s(g))\backslash Sf))
\using {\blacksquare}R
\endprooftree
\prooftree
\prooftree
\prooftree
\prooftree
\prooftree
\prooftree
\prooftree
\prooftree
\prooftree
\justifies
\mbox{\fbox{$Nt(s(m))$}}\ \Rightarrow\ Nt(s(m))
\endprooftree
\justifies
\mbox{\fbox{${\blacksquare}Nt(s(m))$}}\ \Rightarrow\ Nt(s(m))
\using {\blacksquare}L
\endprooftree
\justifies
{\blacksquare}Nt(s(m))\ \Rightarrow\ \fbox{${\exists}gNt(s(g))$}
\using {\exists}R
\endprooftree
\justifies
[{\blacksquare}Nt(s(m))]\ \Rightarrow\ \fbox{${\langle\rangle}{\exists}gNt(s(g))$}
\using {\langle\rangle}R
\endprooftree
\prooftree
\justifies
\mbox{\fbox{$Sf$}}\ \Rightarrow\ Sf
\endprooftree
\justifies
[{\blacksquare}Nt(s(m))], \mbox{\fbox{${\langle\rangle}{\exists}gNt(s(g))\backslash Sf$}}\ \Rightarrow\ Sf
\using {\backslash}L
\endprooftree
\justifies
[{\blacksquare}Nt(s(m))]\ \Rightarrow\ Sf/({\langle\rangle}{\exists}gNt(s(g))\backslash Sf)
\using {/}R
\endprooftree
\justifies
[{\blacksquare}Nt(s(m))]\ \Rightarrow\ {\blacksquare}(Sf/({\langle\rangle}{\exists}gNt(s(g))\backslash Sf))
\using {\blacksquare}R
\endprooftree
\justifies
[{\blacksquare}Nt(s(m))]\ \Rightarrow\ \fbox{$?{\blacksquare}(Sf/({\langle\rangle}{\exists}gNt(s(g))\backslash Sf))$}
\using {?}R
\endprooftree
\prooftree
\prooftree
\prooftree
\prooftree
\prooftree
\prooftree
\prooftree
\prooftree
\prooftree
\prooftree
\prooftree
\justifies
Nt(s(1))\ \Rightarrow\ Nt(s(1))
\endprooftree
\justifies
Nt(s(1))\ \Rightarrow\ \fbox{${\exists}gNt(s(g))$}
\using {\exists}R
\endprooftree
\justifies
[Nt(s(1))]\ \Rightarrow\ \fbox{${\langle\rangle}{\exists}gNt(s(g))$}
\using {\langle\rangle}R
\endprooftree
\prooftree
\justifies
\mbox{\fbox{$Sf$}}\ \Rightarrow\ Sf
\endprooftree
\justifies
[Nt(s(1))], \mbox{\fbox{${\langle\rangle}{\exists}gNt(s(g))\backslash Sf$}}\ \Rightarrow\ Sf
\using {\backslash}L
\endprooftree
\justifies
[Nt(s(1))], \mbox{\fbox{${\square}({\langle\rangle}{\exists}gNt(s(g))\backslash Sf)$}}\ \Rightarrow\ Sf
\using {\Box}L
\endprooftree
\justifies
[{\exists}gNt(s(g))], {\square}({\langle\rangle}{\exists}gNt(s(g))\backslash Sf)\ \Rightarrow\ Sf
\using {\exists}L
\endprooftree
\justifies
{\langle\rangle}{\exists}gNt(s(g)), {\square}({\langle\rangle}{\exists}gNt(s(g))\backslash Sf)\ \Rightarrow\ Sf
\using {\langle\rangle}L
\endprooftree
\justifies
{\square}({\langle\rangle}{\exists}gNt(s(g))\backslash Sf)\ \Rightarrow\ {\langle\rangle}{\exists}gNt(s(g))\backslash Sf
\using {\backslash}R
\endprooftree
\prooftree
\justifies
\mbox{\fbox{$Sf$}}\ \Rightarrow\ Sf
\endprooftree
\justifies
\mbox{\fbox{$Sf/({\langle\rangle}{\exists}gNt(s(g))\backslash Sf)$}}, {\square}({\langle\rangle}{\exists}gNt(s(g))\backslash Sf)\ \Rightarrow\ Sf
\using {/}L
\endprooftree
\justifies
[\mbox{\fbox{${[]^{-1}}(Sf/({\langle\rangle}{\exists}gNt(s(g))\backslash Sf))$}}], {\square}({\langle\rangle}{\exists}gNt(s(g))\backslash Sf)\ \Rightarrow\ Sf
\using {[]^{-1}}L
\endprooftree
\justifies
[[\mbox{\fbox{${[]^{-1}}{[]^{-1}}(Sf/({\langle\rangle}{\exists}gNt(s(g))\backslash Sf))$}}]], {\square}({\langle\rangle}{\exists}gNt(s(g))\backslash Sf)\ \Rightarrow\ Sf
\using {[]^{-1}}L
\endprooftree
\justifies
[[[{\blacksquare}Nt(s(m))], \mbox{\fbox{$?{\blacksquare}(Sf/({\langle\rangle}{\exists}gNt(s(g))\backslash Sf))\backslash {[]^{-1}}{[]^{-1}}(Sf/({\langle\rangle}{\exists}gNt(s(g))\backslash Sf))$}}]], {\square}({\langle\rangle}{\exists}gNt(s(g))\backslash Sf)\ \Rightarrow\ Sf
\using {\backslash}L
\endprooftree
\justifies
[[[{\blacksquare}Nt(s(m))], \mbox{\fbox{$(?{\blacksquare}(Sf/({\langle\rangle}{\exists}gNt(s(g))\backslash Sf))\backslash {[]^{-1}}{[]^{-1}}(Sf/({\langle\rangle}{\exists}gNt(s(g))\backslash Sf)))/{\blacksquare}(Sf/({\langle\rangle}{\exists}gNt(s(g))\backslash Sf))$}}, [{\blacksquare}Nt(s(f))]]], {\square}({\langle\rangle}{\exists}gNt(s(g))\backslash Sf)\ \Rightarrow\ Sf
\using {/}L
\endprooftree
\justifies
[[[{\blacksquare}Nt(s(m))], \mbox{\fbox{${\forall}f((?{\blacksquare}(Sf/({\langle\rangle}{\exists}gNt(s(g))\backslash Sf))\backslash {[]^{-1}}{[]^{-1}}(Sf/({\langle\rangle}{\exists}gNt(s(g))\backslash Sf)))/{\blacksquare}(Sf/({\langle\rangle}{\exists}gNt(s(g))\backslash Sf)))$}}, [{\blacksquare}Nt(s(f))]]], {\square}({\langle\rangle}{\exists}gNt(s(g))\backslash Sf)\ \Rightarrow\ Sf
\using {\forall}L
\endprooftree
\justifies
[[[{\blacksquare}Nt(s(m))], \mbox{\fbox{${\blacksquare}{\forall}f((?{\blacksquare}(Sf/({\langle\rangle}{\exists}gNt(s(g))\backslash Sf))\backslash {[]^{-1}}{[]^{-1}}(Sf/({\langle\rangle}{\exists}gNt(s(g))\backslash Sf)))/{\blacksquare}(Sf/({\langle\rangle}{\exists}gNt(s(g))\backslash Sf)))$}}, [{\blacksquare}Nt(s(f))]]], {\square}({\langle\rangle}{\exists}gNt(s(g))\backslash Sf)\ \Rightarrow\ Sf
\using {\blacksquare}L
\endprooftree}
\end{center}

\vspace{0.15in}

\noindent
In the same manner that we have seen before,
in the left subderivation the righthand conjunct is analysed
as of the lifted type, and in the middle subderivation the lefthand
conjunct is analysed as of the lifted type;
these are the same except for the eventual $?R$ of the latter,
and they centre on the over right lowering
of the higher order verb phrase argument into the antecedent
where it subsequently applies as a functor.
In the righthand derivation the brackets are checked and then application
of the coordinate structure to its verb phrase context involves essentially
derivation of the identity VP yields VP.
This delivers the semantics:
\disp{
$[({\it Pres}\ (\mbox{\v{}}{\it sing}\ {\it j}))\vee ({\it Pres}\ (\mbox{\v{}}{\it sing}\ {\it m}))]$}

\subsection{Object coordination}

Object conjunction, including an object reflexive, is illustrated by:
\disp{
$[{\bf john}]{+}{\bf loves}{+}[[{\bf mary}{+}{\bf and}{+}{\bf himself}]]: Sf$}
The coordination is
basically $(\exstexp X\bsl\abrack\abrack X)/X$ with $X=((N\bsl S)/N)\bsl(N\bsl S)$.
\disp{
\vspace{0.15in}
$\begin{array}[t]{l}
[{\blacksquare}Nt(s(m)): {\it j}], {\square}(({\langle\rangle}{\exists}gNt(s(g))\backslash Sf)/{\exists}aNa): \mbox{\^{}}\lambda A\lambda B({\it Pres}\ ((\mbox{\v{}}{\it love}\ {\it A})\ {\it B})),
{}\mbox{$[[{\blacksquare}Nt(s(f)):{\it m},$}\\
 {\blacksquare}{\forall}f{\forall}a((?{\blacksquare}((({\langle\rangle}Na\backslash Sf)/{\exists}bNb)\backslash ({\langle\rangle}Na\backslash Sf))\backslash {[]^{-1}}{[]^{-1}}((({\langle\rangle}Na\backslash Sf)/{\exists}bNb)\backslash ({\langle\rangle}Na\backslash Sf)))/\\
 {\blacksquare}((({\langle\rangle}Na\backslash Sf)/{\exists}bNb)\backslash ({\langle\rangle}Na\backslash Sf))): (\Phinplus\ ({\it s}\ ({\it s}\ {\it 0}))\ {\it and}),\\ {\blacksquare}{\forall}f((({\langle\rangle}Nt(s(m))\backslash Sf){{\uparrow}{}}Nt(s(m))){{\downarrow}{}}({\langle\rangle}Nt(s(m))\backslash Sf)): \lambda C\lambda D(({\it C}\ {\it D})\ {\it D})]]\ \Rightarrow\ Sf
 \end{array}$}
There is the derivation:


{\tiny
\begin{center}
\prooftree
\prooftree
\prooftree
\prooftree
\prooftree
\prooftree
\prooftree
\prooftree
\prooftree
\prooftree
\prooftree
\prooftree
\prooftree
\justifies
Nt(s(m))\ \Rightarrow\ Nt(s(m))
\endprooftree
\justifies
Nt(s(m))\ \Rightarrow\ \fbox{${\exists}bNb$}
\using {\exists}R
\endprooftree
\prooftree
\prooftree
\prooftree
\justifies
Nt(s(m))\ \Rightarrow\ Nt(s(m))
\endprooftree
\justifies
[Nt(s(m))]\ \Rightarrow\ \fbox{${\langle\rangle}Nt(s(m))$}
\using {\langle\rangle}R
\endprooftree
\prooftree
\justifies
\mbox{\fbox{$Sf$}}\ \Rightarrow\ Sf
\endprooftree
\justifies
[Nt(s(m))], \mbox{\fbox{${\langle\rangle}Nt(s(m))\backslash Sf$}}\ \Rightarrow\ Sf
\using {\backslash}L
\endprooftree
\justifies
[Nt(s(m))], \mbox{\fbox{$({\langle\rangle}Nt(s(m))\backslash Sf)/{\exists}bNb$}}, Nt(s(m))\ \Rightarrow\ Sf
\using {/}L
\endprooftree
\justifies
{\langle\rangle}Nt(s(m)), ({\langle\rangle}Nt(s(m))\backslash Sf)/{\exists}bNb, Nt(s(m))\ \Rightarrow\ Sf
\using {\langle\rangle}L
\endprooftree
\justifies
({\langle\rangle}Nt(s(m))\backslash Sf)/{\exists}bNb, Nt(s(m))\ \Rightarrow\ {\langle\rangle}Nt(s(m))\backslash Sf
\using {\backslash}R
\endprooftree
\justifies
({\langle\rangle}Nt(s(m))\backslash Sf)/{\exists}bNb, {\tt 1}\ \Rightarrow\ ({\langle\rangle}Nt(s(m))\backslash Sf){{\uparrow}{}}Nt(s(m))
\using {\uparrow}R
\endprooftree
\prooftree
\prooftree
\prooftree
\justifies
Nt(s(m))\ \Rightarrow\ Nt(s(m))
\endprooftree
\justifies
[Nt(s(m))]\ \Rightarrow\ \fbox{${\langle\rangle}Nt(s(m))$}
\using {\langle\rangle}R
\endprooftree
\prooftree
\justifies
\mbox{\fbox{$Sf$}}\ \Rightarrow\ Sf
\endprooftree
\justifies
[Nt(s(m))], \mbox{\fbox{${\langle\rangle}Nt(s(m))\backslash Sf$}}\ \Rightarrow\ Sf
\using {\backslash}L
\endprooftree
\justifies
[Nt(s(m))], ({\langle\rangle}Nt(s(m))\backslash Sf)/{\exists}bNb, \mbox{\fbox{$(({\langle\rangle}Nt(s(m))\backslash Sf){{\uparrow}{}}Nt(s(m))){{\downarrow}{}}({\langle\rangle}Nt(s(m))\backslash Sf)$}}\ \Rightarrow\ Sf
\using {\downarrow}L
\endprooftree
\justifies
[Nt(s(m))], ({\langle\rangle}Nt(s(m))\backslash Sf)/{\exists}bNb, \mbox{\fbox{${\forall}f((({\langle\rangle}Nt(s(m))\backslash Sf){{\uparrow}{}}Nt(s(m))){{\downarrow}{}}({\langle\rangle}Nt(s(m))\backslash Sf))$}}\ \Rightarrow\ Sf
\using {\forall}L
\endprooftree
\justifies
[Nt(s(m))], ({\langle\rangle}Nt(s(m))\backslash Sf)/{\exists}bNb, \mbox{\fbox{${\blacksquare}{\forall}f((({\langle\rangle}Nt(s(m))\backslash Sf){{\uparrow}{}}Nt(s(m))){{\downarrow}{}}({\langle\rangle}Nt(s(m))\backslash Sf))$}}\ \Rightarrow\ Sf
\using {\blacksquare}L
\endprooftree
\justifies
{\langle\rangle}Nt(s(m)), ({\langle\rangle}Nt(s(m))\backslash Sf)/{\exists}bNb, {\blacksquare}{\forall}f((({\langle\rangle}Nt(s(m))\backslash Sf){{\uparrow}{}}Nt(s(m))){{\downarrow}{}}({\langle\rangle}Nt(s(m))\backslash Sf))\ \Rightarrow\ Sf
\using {\langle\rangle}L
\endprooftree
\justifies
({\langle\rangle}Nt(s(m))\backslash Sf)/{\exists}bNb, {\blacksquare}{\forall}f((({\langle\rangle}Nt(s(m))\backslash Sf){{\uparrow}{}}Nt(s(m))){{\downarrow}{}}({\langle\rangle}Nt(s(m))\backslash Sf))\ \Rightarrow\ {\langle\rangle}Nt(s(m))\backslash Sf
\using {\backslash}R
\endprooftree
\justifies
{\blacksquare}{\forall}f((({\langle\rangle}Nt(s(m))\backslash Sf){{\uparrow}{}}Nt(s(m))){{\downarrow}{}}({\langle\rangle}Nt(s(m))\backslash Sf))\ \Rightarrow\ (({\langle\rangle}Nt(s(m))\backslash Sf)/{\exists}bNb)\backslash ({\langle\rangle}Nt(s(m))\backslash Sf)
\using {\backslash}R
\endprooftree
\justifies
\begin{array}{c}
{\blacksquare}{\forall}f((({\langle\rangle}Nt(s(m))\backslash Sf){{\uparrow}{}}Nt(s(m))){{\downarrow}{}}({\langle\rangle}Nt(s(m))\backslash Sf))\ \Rightarrow\ {\blacksquare}((({\langle\rangle}Nt(s(m))\backslash Sf)/{\exists}bNb)\backslash ({\langle\rangle}Nt(s(m))\backslash Sf))\\
\mbox{\footnotesize\textcircled{1}}
\end{array}
\using {\blacksquare}R
\endprooftree

\prooftree
\prooftree
\prooftree
\prooftree
\prooftree
\prooftree
\prooftree
\prooftree
\prooftree
\justifies
\mbox{\fbox{$Nt(s(f))$}}\ \Rightarrow\ Nt(s(f))
\endprooftree
\justifies
\mbox{\fbox{${\blacksquare}Nt(s(f))$}}\ \Rightarrow\ Nt(s(f))
\using {\blacksquare}L
\endprooftree
\justifies
{\blacksquare}Nt(s(f))\ \Rightarrow\ \fbox{${\exists}bNb$}
\using {\exists}R
\endprooftree
\prooftree
\prooftree
\prooftree
\justifies
Nt(s(m))\ \Rightarrow\ Nt(s(m))
\endprooftree
\justifies
[Nt(s(m))]\ \Rightarrow\ \fbox{${\langle\rangle}Nt(s(m))$}
\using {\langle\rangle}R
\endprooftree
\prooftree
\justifies
\mbox{\fbox{$Sf$}}\ \Rightarrow\ Sf
\endprooftree
\justifies
[Nt(s(m))], \mbox{\fbox{${\langle\rangle}Nt(s(m))\backslash Sf$}}\ \Rightarrow\ Sf
\using {\backslash}L
\endprooftree
\justifies
[Nt(s(m))], \mbox{\fbox{$({\langle\rangle}Nt(s(m))\backslash Sf)/{\exists}bNb$}}, {\blacksquare}Nt(s(f))\ \Rightarrow\ Sf
\using {/}L
\endprooftree
\justifies
{\langle\rangle}Nt(s(m)), ({\langle\rangle}Nt(s(m))\backslash Sf)/{\exists}bNb, {\blacksquare}Nt(s(f))\ \Rightarrow\ Sf
\using {\langle\rangle}L
\endprooftree
\justifies
({\langle\rangle}Nt(s(m))\backslash Sf)/{\exists}bNb, {\blacksquare}Nt(s(f))\ \Rightarrow\ {\langle\rangle}Nt(s(m))\backslash Sf
\using {\backslash}R
\endprooftree
\justifies
{\blacksquare}Nt(s(f))\ \Rightarrow\ (({\langle\rangle}Nt(s(m))\backslash Sf)/{\exists}bNb)\backslash ({\langle\rangle}Nt(s(m))\backslash Sf)
\using {\backslash}R
\endprooftree
\justifies
{\blacksquare}Nt(s(f))\ \Rightarrow\ {\blacksquare}((({\langle\rangle}Nt(s(m))\backslash Sf)/{\exists}bNb)\backslash ({\langle\rangle}Nt(s(m))\backslash Sf))
\using {\blacksquare}R
\endprooftree
\justifies
\begin{array}{c}
{\blacksquare}Nt(s(f))\ \Rightarrow\ \fbox{$?{\blacksquare}((({\langle\rangle}Nt(s(m))\backslash Sf)/{\exists}bNb)\backslash ({\langle\rangle}Nt(s(m))\backslash Sf))$}\\
\mbox{\footnotesize\textcircled{2}}
\end{array}
\using {?}R
\endprooftree
\end{center}}

\begin{center}
\rotatebox{-90}{\tiny
\prooftree
\prooftree
\prooftree
\prooftree
\mbox{\footnotesize\textcircled{1}}\tab
\prooftree
\mbox{\footnotesize\textcircled{2}}\tab
\prooftree
\prooftree
\prooftree
\prooftree
\prooftree
\prooftree
\prooftree
\prooftree
\prooftree
\prooftree
\prooftree
\justifies
N1\ \Rightarrow\ N1
\endprooftree
\justifies
N1\ \Rightarrow\ \fbox{${\exists}aNa$}
\using {\exists}R
\endprooftree
\prooftree
\prooftree
\prooftree
\prooftree
\justifies
Nt(s(m))\ \Rightarrow\ Nt(s(m))
\endprooftree
\justifies
Nt(s(m))\ \Rightarrow\ \fbox{${\exists}gNt(s(g))$}
\using {\exists}R
\endprooftree
\justifies
[Nt(s(m))]\ \Rightarrow\ \fbox{${\langle\rangle}{\exists}gNt(s(g))$}
\using {\langle\rangle}R
\endprooftree
\prooftree
\justifies
\mbox{\fbox{$Sf$}}\ \Rightarrow\ Sf
\endprooftree
\justifies
[Nt(s(m))], \mbox{\fbox{${\langle\rangle}{\exists}gNt(s(g))\backslash Sf$}}\ \Rightarrow\ Sf
\using {\backslash}L
\endprooftree
\justifies
[Nt(s(m))], \mbox{\fbox{$({\langle\rangle}{\exists}gNt(s(g))\backslash Sf)/{\exists}aNa$}}, N1\ \Rightarrow\ Sf
\using {/}L
\endprooftree
\justifies
[Nt(s(m))], \mbox{\fbox{${\square}(({\langle\rangle}{\exists}gNt(s(g))\backslash Sf)/{\exists}aNa)$}}, N1\ \Rightarrow\ Sf
\using {\Box}L
\endprooftree
\justifies
[Nt(s(m))], {\square}(({\langle\rangle}{\exists}gNt(s(g))\backslash Sf)/{\exists}aNa), {\exists}bNb\ \Rightarrow\ Sf
\using {\exists}L
\endprooftree
\justifies
{\langle\rangle}Nt(s(m)), {\square}(({\langle\rangle}{\exists}gNt(s(g))\backslash Sf)/{\exists}aNa), {\exists}bNb\ \Rightarrow\ Sf
\using {\langle\rangle}L
\endprooftree
\justifies
{\square}(({\langle\rangle}{\exists}gNt(s(g))\backslash Sf)/{\exists}aNa), {\exists}bNb\ \Rightarrow\ {\langle\rangle}Nt(s(m))\backslash Sf
\using {\backslash}R
\endprooftree
\justifies
{\square}(({\langle\rangle}{\exists}gNt(s(g))\backslash Sf)/{\exists}aNa)\ \Rightarrow\ ({\langle\rangle}Nt(s(m))\backslash Sf)/{\exists}bNb
\using {/}R
\endprooftree
\prooftree
\prooftree
\prooftree
\prooftree
\justifies
\mbox{\fbox{$Nt(s(m))$}}\ \Rightarrow\ Nt(s(m))
\endprooftree
\justifies
\mbox{\fbox{${\blacksquare}Nt(s(m))$}}\ \Rightarrow\ Nt(s(m))
\using {\blacksquare}L
\endprooftree
\justifies
[{\blacksquare}Nt(s(m))]\ \Rightarrow\ \fbox{${\langle\rangle}Nt(s(m))$}
\using {\langle\rangle}R
\endprooftree
\prooftree
\justifies
\mbox{\fbox{$Sf$}}\ \Rightarrow\ Sf
\endprooftree
\justifies
[{\blacksquare}Nt(s(m))], \mbox{\fbox{${\langle\rangle}Nt(s(m))\backslash Sf$}}\ \Rightarrow\ Sf
\using {\backslash}L
\endprooftree
\justifies
[{\blacksquare}Nt(s(m))], {\square}(({\langle\rangle}{\exists}gNt(s(g))\backslash Sf)/{\exists}aNa), \mbox{\fbox{$(({\langle\rangle}Nt(s(m))\backslash Sf)/{\exists}bNb)\backslash ({\langle\rangle}Nt(s(m))\backslash Sf)$}}\ \Rightarrow\ Sf
\using {\backslash}L
\endprooftree
\justifies
[{\blacksquare}Nt(s(m))], {\square}(({\langle\rangle}{\exists}gNt(s(g))\backslash Sf)/{\exists}aNa), [\mbox{\fbox{${[]^{-1}}((({\langle\rangle}Nt(s(m))\backslash Sf)/{\exists}bNb)\backslash ({\langle\rangle}Nt(s(m))\backslash Sf))$}}]\ \Rightarrow\ Sf
\using {[]^{-1}}L
\endprooftree
\justifies
[{\blacksquare}Nt(s(m))], {\square}(({\langle\rangle}{\exists}gNt(s(g))\backslash Sf)/{\exists}aNa), [[\mbox{\fbox{${[]^{-1}}{[]^{-1}}((({\langle\rangle}Nt(s(m))\backslash Sf)/{\exists}bNb)\backslash ({\langle\rangle}Nt(s(m))\backslash Sf))$}}]]\ \Rightarrow\ Sf
\using {[]^{-1}}L
\endprooftree\justifies
[{\blacksquare}Nt(s(m))], {\square}(({\langle\rangle}{\exists}gNt(s(g))\backslash Sf)/{\exists}aNa), [[{\blacksquare}Nt(s(f)), \mbox{\fbox{$?{\blacksquare}((({\langle\rangle}Nt(s(m))\backslash Sf)/{\exists}bNb)\backslash ({\langle\rangle}Nt(s(m))\backslash Sf))\backslash {[]^{-1}}{[]^{-1}}((({\langle\rangle}Nt(s(m))\backslash Sf)/{\exists}bNb)\backslash ({\langle\rangle}Nt(s(m))\backslash Sf))$}}]]\ \Rightarrow\ Sf
\using {\backslash}L
\endprooftree
\justifies
[{\blacksquare}Nt(s(m))], {\square}(({\langle\rangle}{\exists}gNt(s(g))\backslash Sf)/{\exists}aNa), [[{\blacksquare}Nt(s(f)), \mbox{\fbox{$(?{\blacksquare}((({\langle\rangle}Nt(s(m))\backslash Sf)/{\exists}bNb)\backslash ({\langle\rangle}Nt(s(m))\backslash Sf))\backslash {[]^{-1}}{[]^{-1}}((({\langle\rangle}Nt(s(m))\backslash Sf)/{\exists}bNb)\backslash ({\langle\rangle}Nt(s(m))\backslash Sf)))/{\blacksquare}((({\langle\rangle}Nt(s(m))\backslash Sf)/{\exists}bNb)\backslash ({\langle\rangle}Nt(s(m))\backslash Sf))$}}, {\blacksquare}{\forall}f((({\langle\rangle}Nt(s(m))\backslash Sf){{\uparrow}{}}Nt(s(m))){{\downarrow}{}}({\langle\rangle}Nt(s(m))\backslash Sf))]]\ \Rightarrow\ Sf
\using {/}L
\endprooftree
\justifies
[{\blacksquare}Nt(s(m))], {\square}(({\langle\rangle}{\exists}gNt(s(g))\backslash Sf)/{\exists}aNa), [[{\blacksquare}Nt(s(f)), \mbox{\fbox{${\forall}a((?{\blacksquare}((({\langle\rangle}Na\backslash Sf)/{\exists}bNb)\backslash ({\langle\rangle}Na\backslash Sf))\backslash {[]^{-1}}{[]^{-1}}((({\langle\rangle}Na\backslash Sf)/{\exists}bNb)\backslash ({\langle\rangle}Na\backslash Sf)))/{\blacksquare}((({\langle\rangle}Na\backslash Sf)/{\exists}bNb)\backslash ({\langle\rangle}Na\backslash Sf)))$}}, {\blacksquare}{\forall}f((({\langle\rangle}Nt(s(m))\backslash Sf){{\uparrow}{}}Nt(s(m))){{\downarrow}{}}({\langle\rangle}Nt(s(m))\backslash Sf))]]\ \Rightarrow\ Sf
\using {\forall}L
\endprooftree
\justifies
[{\blacksquare}Nt(s(m))], {\square}(({\langle\rangle}{\exists}gNt(s(g))\backslash Sf)/{\exists}aNa), [[{\blacksquare}Nt(s(f)), \mbox{\fbox{${\forall}f{\forall}a((?{\blacksquare}((({\langle\rangle}Na\backslash Sf)/{\exists}bNb)\backslash ({\langle\rangle}Na\backslash Sf))\backslash {[]^{-1}}{[]^{-1}}((({\langle\rangle}Na\backslash Sf)/{\exists}bNb)\backslash ({\langle\rangle}Na\backslash Sf)))/{\blacksquare}((({\langle\rangle}Na\backslash Sf)/{\exists}bNb)\backslash ({\langle\rangle}Na\backslash Sf)))$}}, {\blacksquare}{\forall}f((({\langle\rangle}Nt(s(m))\backslash Sf){{\uparrow}{}}Nt(s(m))){{\downarrow}{}}({\langle\rangle}Nt(s(m))\backslash Sf))]]\ \Rightarrow\ Sf
\using {\forall}L
\endprooftree
\justifies
[{\blacksquare}Nt(s(m))], {\square}(({\langle\rangle}{\exists}gNt(s(g))\backslash Sf)/{\exists}aNa), [[{\blacksquare}Nt(s(f)), \mbox{\fbox{${\blacksquare}{\forall}f{\forall}a((?{\blacksquare}((({\langle\rangle}Na\backslash Sf)/{\exists}bNb)\backslash ({\langle\rangle}Na\backslash Sf))\backslash {[]^{-1}}{[]^{-1}}((({\langle\rangle}Na\backslash Sf)/{\exists}bNb)\backslash ({\langle\rangle}Na\backslash Sf)))/{\blacksquare}((({\langle\rangle}Na\backslash Sf)/{\exists}bNb)\backslash ({\langle\rangle}Na\backslash Sf)))$}}, {\blacksquare}{\forall}f((({\langle\rangle}Nt(s(m))\backslash Sf){{\uparrow}{}}Nt(s(m))){{\downarrow}{}}({\langle\rangle}Nt(s(m))\backslash Sf))]]\ \Rightarrow\ Sf
\using {\blacksquare}L
\endprooftree}
\end{center}

\vspace{0.15in}

\noindent
Again the main left (\textcircled{1}), main middle (\textcircled{2}) and main right
subderivations are for the right conjunt, left conjunct, and context respectively.
The subtree \textcircled{1} essentially derives that a subject-oriented reflexive
$((N\bsl S)\circum N)\infix(N\bsl S)$ yields a lifted object type $((N\bsl S)/N)\bsl(N\bsl S)$:
\disp{\mini
\prooftree
\prooftree
\prooftree
\prooftree
(N\bsl S)/N, N\yields N\bsl S
\justifies
(N\bsl S)/N, \sep\yields(N\bsl S)\circum N
\using \circum R
\endprooftree\tb
N, N\bsl S\yields S
\justifies
N, (N\bsl S)/N, ((N\bsl S)\circum N)\infix(N\bsl S)\yields S
\using \infix L
\endprooftree
\justifies
(N\bsl S)/N, ((N\bsl S)\circum N)\infix(N\bsl S)\yields N\bsl S
\using \bsl R
\endprooftree
\justifies
((N\bsl S)\circum N)\infix(N\bsl S)\yields ((N\bsl S)/N)\bsl(N\bsl S)
\using \bsl R
\endprooftree}
The derivation centres on the successive $\infix L$ and $\circum R$ half way up.
The subtree \textcircled{2} essentially derives that a nominal $N$ yields a lifted object type
$((N\bsl S)/N)\bsl(N\bsl S)$:
\disp{$
N\yields ((N\bsl S)/N)\bsl(N\bsl S)$}
The remaining main subtree checks the brackets and applies the coordinate structure
to the transitive verb (deriving TV\yields TV) and subject (deriving N\yields N)
contexts.
All this delivers the correct semantics:
\disp{
$[({\it Pres}\ ((\mbox{\v{}}{\it love}\ {\it m})\ {\it j}))\wedge ({\it Pres}\ ((\mbox{\v{}}{\it love}\ {\it j})\ {\it j}))]$}

\subsection{Right node raising coordination}

The next example is an instance of right node raising:
\disp{
$[[[{\bf john}]{+}{\bf likes}{+}{\bf and}{+}[{\bf mary}]{+}{\bf loves}]]{+}{\bf london}: Sf$}
Appropriate lexical lookup yields  the following where the coordinator type is
essentially $(\exstexp X\bsl$ $\abrack\abrack X)/X$ with $X=S/N$.
\disp{
$\begin{array}[t]{l}
[[[{\blacksquare}Nt(s(m)): {\it j}], {\square}(({\langle\rangle}{\exists}gNt(s(g))\backslash Sf)/{\exists}aNa): \mbox{\^{}}\lambda A\lambda B({\it Pres}\ ((\mbox{\v{}}{\it like}\ {\it A})\ {\it B})),\\
 {\blacksquare}{\forall}f((?{\blacksquare}(Sf/{\exists}aNa)\backslash {[]^{-1}}{[]^{-1}}(Sf/{\exists}aNa))/{\blacksquare}(Sf/{\exists}aNa)): (\Phinplus\ ({\it s}\ {\it 0})\ {\it and}),\\{} [{\blacksquare}Nt(s(f)): {\it m}], {\square}(({\langle\rangle}{\exists}gNt(s(g))\backslash Sf)/{\exists}aNa): \mbox{\^{}}\lambda C\lambda D({\it Pres}\ ((\mbox{\v{}}{\it love}\ {\it C})\ {\it D}))]],\\ {\blacksquare}Nt(s(n)): {\it l}\ \Rightarrow\ Sf
 \end{array}$}
There is the derivation below.

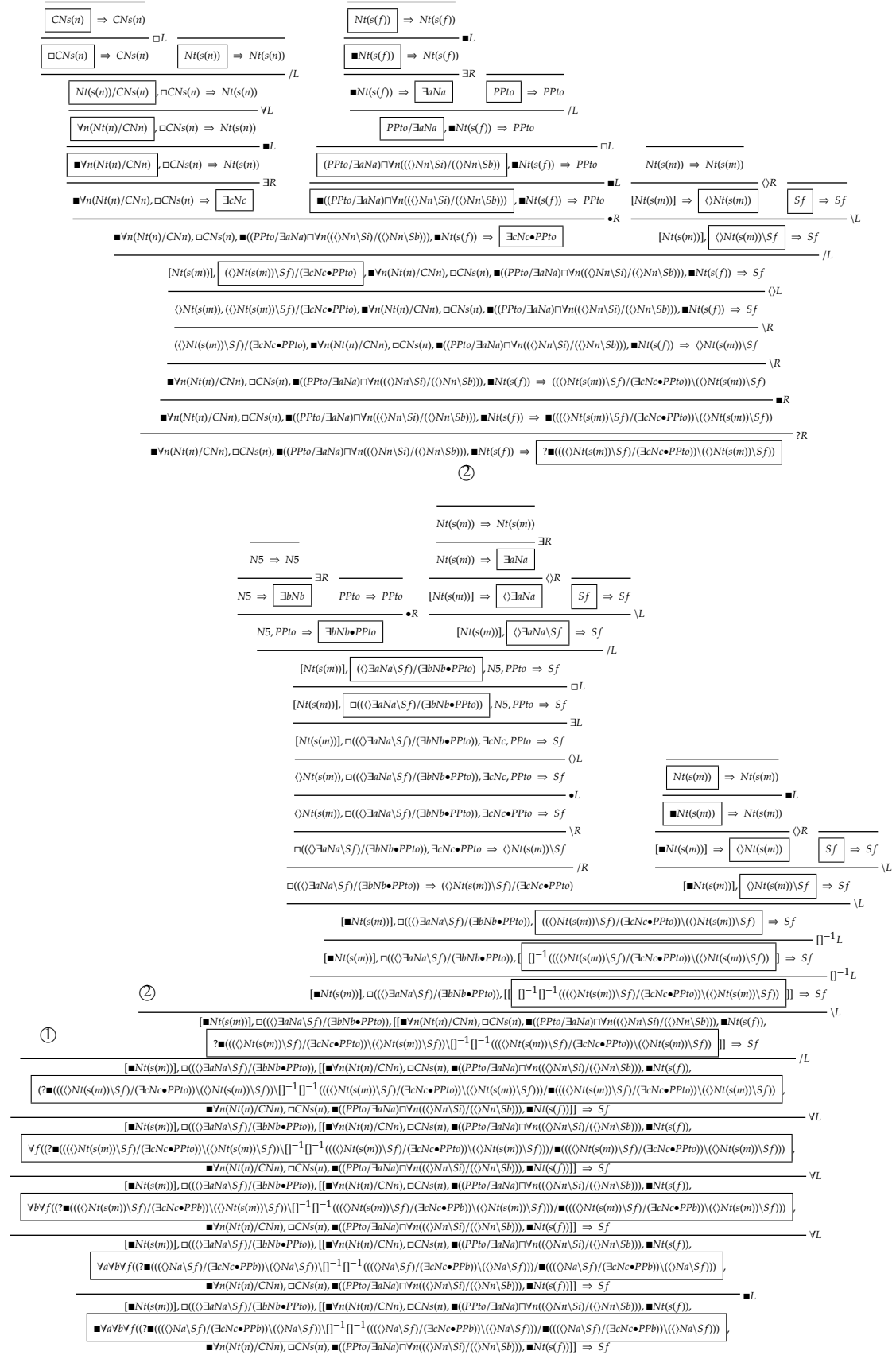
\vspace*{0.15in}

\begin{center}
\rotatebox{-90}{\tiny
\prooftree
\prooftree
\prooftree
\prooftree
\prooftree
\prooftree
\prooftree
\prooftree
\prooftree
\prooftree
\justifies
N2\ \Rightarrow\ N2
\endprooftree
\justifies
N2\ \Rightarrow\ \fbox{${\exists}aNa$}
\using {\exists}R
\endprooftree
\prooftree
\prooftree
\prooftree
\prooftree
\prooftree
\justifies
\mbox{\fbox{$Nt(s(f))$}}\ \Rightarrow\ Nt(s(f))
\endprooftree
\justifies
\mbox{\fbox{${\blacksquare}Nt(s(f))$}}\ \Rightarrow\ Nt(s(f))
\using {\blacksquare}L
\endprooftree
\justifies
{\blacksquare}Nt(s(f))\ \Rightarrow\ \fbox{${\exists}gNt(s(g))$}
\using {\exists}R
\endprooftree
\justifies
[{\blacksquare}Nt(s(f))]\ \Rightarrow\ \fbox{${\langle\rangle}{\exists}gNt(s(g))$}
\using {\langle\rangle}R
\endprooftree
\prooftree
\justifies
\mbox{\fbox{$Sf$}}\ \Rightarrow\ Sf
\endprooftree
\justifies
[{\blacksquare}Nt(s(f))], \mbox{\fbox{${\langle\rangle}{\exists}gNt(s(g))\backslash Sf$}}\ \Rightarrow\ Sf
\using {\backslash}L
\endprooftree
\justifies
[{\blacksquare}Nt(s(f))], \mbox{\fbox{$({\langle\rangle}{\exists}gNt(s(g))\backslash Sf)/{\exists}aNa$}}, N2\ \Rightarrow\ Sf
\using {/}L
\endprooftree
\justifies
[{\blacksquare}Nt(s(f))], \mbox{\fbox{${\square}(({\langle\rangle}{\exists}gNt(s(g))\backslash Sf)/{\exists}aNa)$}}, N2\ \Rightarrow\ Sf
\using {\Box}L
\endprooftree
\justifies
[{\blacksquare}Nt(s(f))], {\square}(({\langle\rangle}{\exists}gNt(s(g))\backslash Sf)/{\exists}aNa), {\exists}aNa\ \Rightarrow\ Sf
\using {\exists}L
\endprooftree
\justifies
[{\blacksquare}Nt(s(f))], {\square}(({\langle\rangle}{\exists}gNt(s(g))\backslash Sf)/{\exists}aNa)\ \Rightarrow\ Sf/{\exists}aNa
\using {/}R
\endprooftree
\justifies
[{\blacksquare}Nt(s(f))], {\square}(({\langle\rangle}{\exists}gNt(s(g))\backslash Sf)/{\exists}aNa)\ \Rightarrow\ {\blacksquare}(Sf/{\exists}aNa)
\using {\blacksquare}R
\endprooftree
\prooftree
\prooftree
\prooftree
\prooftree
\prooftree
\prooftree
\prooftree
\prooftree
\prooftree
\justifies
N3\ \Rightarrow\ N3
\endprooftree
\justifies
N3\ \Rightarrow\ \fbox{${\exists}aNa$}
\using {\exists}R
\endprooftree
\prooftree
\prooftree
\prooftree
\prooftree
\prooftree
\justifies
\mbox{\fbox{$Nt(s(m))$}}\ \Rightarrow\ Nt(s(m))
\endprooftree
\justifies
\mbox{\fbox{${\blacksquare}Nt(s(m))$}}\ \Rightarrow\ Nt(s(m))
\using {\blacksquare}L
\endprooftree
\justifies
{\blacksquare}Nt(s(m))\ \Rightarrow\ \fbox{${\exists}gNt(s(g))$}
\using {\exists}R
\endprooftree
\justifies
[{\blacksquare}Nt(s(m))]\ \Rightarrow\ \fbox{${\langle\rangle}{\exists}gNt(s(g))$}
\using {\langle\rangle}R
\endprooftree
\prooftree
\justifies
\mbox{\fbox{$Sf$}}\ \Rightarrow\ Sf
\endprooftree
\justifies
[{\blacksquare}Nt(s(m))], \mbox{\fbox{${\langle\rangle}{\exists}gNt(s(g))\backslash Sf$}}\ \Rightarrow\ Sf
\using {\backslash}L
\endprooftree
\justifies
[{\blacksquare}Nt(s(m))], \mbox{\fbox{$({\langle\rangle}{\exists}gNt(s(g))\backslash Sf)/{\exists}aNa$}}, N3\ \Rightarrow\ Sf
\using {/}L
\endprooftree
\justifies
[{\blacksquare}Nt(s(m))], \mbox{\fbox{${\square}(({\langle\rangle}{\exists}gNt(s(g))\backslash Sf)/{\exists}aNa)$}}, N3\ \Rightarrow\ Sf
\using {\Box}L
\endprooftree
\justifies
[{\blacksquare}Nt(s(m))], {\square}(({\langle\rangle}{\exists}gNt(s(g))\backslash Sf)/{\exists}aNa), {\exists}aNa\ \Rightarrow\ Sf
\using {\exists}L
\endprooftree
\justifies
[{\blacksquare}Nt(s(m))], {\square}(({\langle\rangle}{\exists}gNt(s(g))\backslash Sf)/{\exists}aNa)\ \Rightarrow\ Sf/{\exists}aNa
\using {/}R
\endprooftree
\justifies
[{\blacksquare}Nt(s(m))], {\square}(({\langle\rangle}{\exists}gNt(s(g))\backslash Sf)/{\exists}aNa)\ \Rightarrow\ {\blacksquare}(Sf/{\exists}aNa)
\using {\blacksquare}R
\endprooftree
\justifies
[{\blacksquare}Nt(s(m))], {\square}(({\langle\rangle}{\exists}gNt(s(g))\backslash Sf)/{\exists}aNa)\ \Rightarrow\ \fbox{$?{\blacksquare}(Sf/{\exists}aNa)$}
\using {?}R
\endprooftree
\prooftree
\prooftree
\prooftree
\prooftree
\prooftree
\prooftree
\justifies
\mbox{\fbox{$Nt(s(n))$}}\ \Rightarrow\ Nt(s(n))
\endprooftree
\justifies
\mbox{\fbox{${\blacksquare}Nt(s(n))$}}\ \Rightarrow\ Nt(s(n))
\using {\blacksquare}L
\endprooftree
\justifies
{\blacksquare}Nt(s(n))\ \Rightarrow\ \fbox{${\exists}aNa$}
\using {\exists}R
\endprooftree
\prooftree
\justifies
\mbox{\fbox{$Sf$}}\ \Rightarrow\ Sf
\endprooftree
\justifies
\mbox{\fbox{$Sf/{\exists}aNa$}}, {\blacksquare}Nt(s(n))\ \Rightarrow\ Sf
\using {/}L
\endprooftree
\justifies
[\mbox{\fbox{${[]^{-1}}(Sf/{\exists}aNa)$}}], {\blacksquare}Nt(s(n))\ \Rightarrow\ Sf
\using {[]^{-1}}L
\endprooftree
\justifies
[[\mbox{\fbox{${[]^{-1}}{[]^{-1}}(Sf/{\exists}aNa)$}}]], {\blacksquare}Nt(s(n))\ \Rightarrow\ Sf
\using {[]^{-1}}L
\endprooftree
\justifies
[[[{\blacksquare}Nt(s(m))], {\square}(({\langle\rangle}{\exists}gNt(s(g))\backslash Sf)/{\exists}aNa), \mbox{\fbox{$?{\blacksquare}(Sf/{\exists}aNa)\backslash {[]^{-1}}{[]^{-1}}(Sf/{\exists}aNa)$}}]], {\blacksquare}Nt(s(n))\ \Rightarrow\ Sf
\using {\backslash}L
\endprooftree
\justifies
[[[{\blacksquare}Nt(s(m))], {\square}(({\langle\rangle}{\exists}gNt(s(g))\backslash Sf)/{\exists}aNa), \mbox{\fbox{$(?{\blacksquare}(Sf/{\exists}aNa)\backslash {[]^{-1}}{[]^{-1}}(Sf/{\exists}aNa))/{\blacksquare}(Sf/{\exists}aNa)$}}, [{\blacksquare}Nt(s(f))], {\square}(({\langle\rangle}{\exists}gNt(s(g))\backslash Sf)/{\exists}aNa)]], {\blacksquare}Nt(s(n))\ \Rightarrow\ Sf
\using {/}L
\endprooftree
\justifies
[[[{\blacksquare}Nt(s(m))], {\square}(({\langle\rangle}{\exists}gNt(s(g))\backslash Sf)/{\exists}aNa), \mbox{\fbox{${\forall}f((?{\blacksquare}(Sf/{\exists}aNa)\backslash {[]^{-1}}{[]^{-1}}(Sf/{\exists}aNa))/{\blacksquare}(Sf/{\exists}aNa))$}}, [{\blacksquare}Nt(s(f))], {\square}(({\langle\rangle}{\exists}gNt(s(g))\backslash Sf)/{\exists}aNa)]], {\blacksquare}Nt(s(n))\ \Rightarrow\ Sf
\using {\forall}L
\endprooftree
\justifies
[[[{\blacksquare}Nt(s(m))], {\square}(({\langle\rangle}{\exists}gNt(s(g))\backslash Sf)/{\exists}aNa), \mbox{\fbox{${\blacksquare}{\forall}f((?{\blacksquare}(Sf/{\exists}aNa)\backslash {[]^{-1}}{[]^{-1}}(Sf/{\exists}aNa))/{\blacksquare}(Sf/{\exists}aNa))$}}, [{\blacksquare}Nt(s(f))], {\square}(({\langle\rangle}{\exists}gNt(s(g))\backslash Sf)/{\exists}aNa)]], {\blacksquare}Nt(s(n))\ \Rightarrow\ Sf
\using {\blacksquare}L
\endprooftree}
\end{center}

\vspace*{0.15in}

As ever the left and middle subderivations are for the right and left conjuncts
respectively, and involve in this case essentially the derivation of
$N, \TV \yields S/N$ which proceeds:
\disp{$
\prooftree
N, \TV, N\yields S
\justifies
N, \TV\yields S/N
\using /R
\endprooftree$}
The right, context, subderivation, applies the coordinate structure of type, basically,
$S/N$ to its right node raised $N$.
All this assigns semantics:
\disp{
$[({\it Pres}\ ((\mbox{\v{}}{\it like}\ {\it l})\ {\it j}))\wedge ({\it Pres}\ ((\mbox{\v{}}{\it love}\ {\it l})\ {\it m}))]$}

\commentout{

There follows conjunction of complex prepositional transitive verb phrases:
\disp{
(crd(21)) $[{\bf john}]{+}[[{\bf gave}{+}{\bf the}{+}{\bf book}{+}{\bf and}{+}{\bf sent}{+}{\bf the}{+}{\bf cd}]]{+}{\bf to}{+}{\bf mary}: Sf$}
Appropriate lexical lookup yields the following where the
coordinator is essentially of the form $(\exstexp X\bsl$ $\abrack\abrack X)/X$ where
$X=$.
\disp{
$[{\blacksquare}Nt(s(m)): {\it j}], [[{\square}(({\langle\rangle}{\exists}aNa\backslash Sf)/({\exists}bNb{\bullet}{\it PP}{\it to})): \mbox{\^{}}\lambda A\lambda B({\it Past}\ (((\mbox{\v{}}{\it give}\ \pi_2{\it A})\ \pi_1{\it A})\ {\it B})), {\blacksquare}{\forall}n(Nt(n)/{\it CN}{\it n}): \iota , {\square}{\it CN}{\it s(n)}: {\it book}, {\blacksquare}{\forall}a{\forall}b{\forall}f((?{\blacksquare}(({\langle\rangle}Na\backslash Sf)/{\it PP}{\it b})\backslash {[]^{-1}}{[]^{-1}}(({\langle\rangle}Na\backslash Sf)/{\it PP}{\it b}))/{\blacksquare}(({\langle\rangle}Na\backslash Sf)/{\it PP}{\it b})): (\Phinplus\ ({\it s}\ ({\it s}\ {\it 0}))\ {\it and}), {\square}(({\langle\rangle}{\exists}aNa\backslash Sf)/({\exists}bNb{\bullet}{\it PP}{\it to})): \mbox{\^{}}\lambda C\lambda D({\it Past}\ (((\mbox{\v{}}{\it sent}\ \pi_2{\it C})\ \pi_1{\it C})\ {\it D})), {\blacksquare}{\forall}n(Nt(n)/{\it CN}{\it n}): \iota , {\square}{\it CN}{\it s(n)}: {\it cd}]], {\blacksquare}(({\it PP}{\it to}/{\exists}aNa){\sqcap}{\forall}n(({\langle\rangle}Nn\backslash Si)/({\langle\rangle}Nn\backslash Sb))): \lambda E{\it E}, {\blacksquare}Nt(s(f)): {\it m}\ \Rightarrow\ Sf$}
There is the derivation:

\clearpage

\vspace{0.15in}

{\tiny
\prooftree
\prooftree
\prooftree
\prooftree
\prooftree
\prooftree
\prooftree
\prooftree
\prooftree
\prooftree
\prooftree
\prooftree
\prooftree
\justifies
\mbox{\fbox{${\it CN}{\it s(n)}$}}\ \Rightarrow\ {\it CN}{\it s(n)}
\endprooftree
\justifies
\mbox{\fbox{${\square}{\it CN}{\it s(n)}$}}\ \Rightarrow\ {\it CN}{\it s(n)}
\using {\Box}L
\endprooftree
\prooftree
\justifies
\mbox{\fbox{$Nt(s(n))$}}\ \Rightarrow\ Nt(s(n))
\endprooftree
\justifies
\mbox{\fbox{$Nt(s(n))/{\it CN}{\it s(n)}$}}, {\square}{\it CN}{\it s(n)}\ \Rightarrow\ Nt(s(n))
\using {/}L
\endprooftree
\justifies
\mbox{\fbox{${\forall}n(Nt(n)/{\it CN}{\it n})$}}, {\square}{\it CN}{\it s(n)}\ \Rightarrow\ Nt(s(n))
\using {\forall}L
\endprooftree
\justifies
\mbox{\fbox{${\blacksquare}{\forall}n(Nt(n)/{\it CN}{\it n})$}}, {\square}{\it CN}{\it s(n)}\ \Rightarrow\ Nt(s(n))
\using {\blacksquare}L
\endprooftree
\justifies
{\blacksquare}{\forall}n(Nt(n)/{\it CN}{\it n}), {\square}{\it CN}{\it s(n)}\ \Rightarrow\ \fbox{${\exists}bNb$}
\using {\exists}R
\endprooftree
\prooftree
\justifies
{\it PP}{\it to}\ \Rightarrow\ {\it PP}{\it to}
\endprooftree
\justifies
{\blacksquare}{\forall}n(Nt(n)/{\it CN}{\it n}), {\square}{\it CN}{\it s(n)}, {\it PP}{\it to}\ \Rightarrow\ \fbox{${\exists}bNb{\bullet}{\it PP}{\it to}$}
\using {\bullet}R
\endprooftree
\prooftree
\prooftree
\prooftree
\prooftree
\justifies
Nt(s(m))\ \Rightarrow\ Nt(s(m))
\endprooftree
\justifies
Nt(s(m))\ \Rightarrow\ \fbox{${\exists}aNa$}
\using {\exists}R
\endprooftree
\justifies
[Nt(s(m))]\ \Rightarrow\ \fbox{${\langle\rangle}{\exists}aNa$}
\using {\langle\rangle}R
\endprooftree
\prooftree
\justifies
\mbox{\fbox{$Sf$}}\ \Rightarrow\ Sf
\endprooftree
\justifies
[Nt(s(m))], \mbox{\fbox{${\langle\rangle}{\exists}aNa\backslash Sf$}}\ \Rightarrow\ Sf
\using {\backslash}L
\endprooftree
\justifies
[Nt(s(m))], \mbox{\fbox{$({\langle\rangle}{\exists}aNa\backslash Sf)/({\exists}bNb{\bullet}{\it PP}{\it to})$}}, {\blacksquare}{\forall}n(Nt(n)/{\it CN}{\it n}), {\square}{\it CN}{\it s(n)}, {\it PP}{\it to}\ \Rightarrow\ Sf
\using {/}L
\endprooftree
\justifies
[Nt(s(m))], \mbox{\fbox{${\square}(({\langle\rangle}{\exists}aNa\backslash Sf)/({\exists}bNb{\bullet}{\it PP}{\it to}))$}}, {\blacksquare}{\forall}n(Nt(n)/{\it CN}{\it n}), {\square}{\it CN}{\it s(n)}, {\it PP}{\it to}\ \Rightarrow\ Sf
\using {\Box}L
\endprooftree
\justifies
{\langle\rangle}Nt(s(m)), {\square}(({\langle\rangle}{\exists}aNa\backslash Sf)/({\exists}bNb{\bullet}{\it PP}{\it to})), {\blacksquare}{\forall}n(Nt(n)/{\it CN}{\it n}), {\square}{\it CN}{\it s(n)}, {\it PP}{\it to}\ \Rightarrow\ Sf
\using {\langle\rangle}L
\endprooftree
\justifies
{\square}(({\langle\rangle}{\exists}aNa\backslash Sf)/({\exists}bNb{\bullet}{\it PP}{\it to})), {\blacksquare}{\forall}n(Nt(n)/{\it CN}{\it n}), {\square}{\it CN}{\it s(n)}, {\it PP}{\it to}\ \Rightarrow\ {\langle\rangle}Nt(s(m))\backslash Sf
\using {\backslash}R
\endprooftree
\justifies
{\square}(({\langle\rangle}{\exists}aNa\backslash Sf)/({\exists}bNb{\bullet}{\it PP}{\it to})), {\blacksquare}{\forall}n(Nt(n)/{\it CN}{\it n}), {\square}{\it CN}{\it s(n)}\ \Rightarrow\ ({\langle\rangle}Nt(s(m))\backslash Sf)/{\it PP}{\it to}
\using {/}R
\endprooftree
\justifies
\begin{array}{c}
{\square}(({\langle\rangle}{\exists}aNa\backslash Sf)/({\exists}bNb{\bullet}{\it PP}{\it to})), {\blacksquare}{\forall}n(Nt(n)/{\it CN}{\it n}), {\square}{\it CN}{\it s(n)}\ \Rightarrow\ {\blacksquare}(({\langle\rangle}Nt(s(m))\backslash Sf)/{\it PP}{\it to})\\
\mbox{\footnotesize\textcircled{1}}
\end{array}
\using {\blacksquare}R
\endprooftree

\prooftree
\prooftree
\prooftree
\prooftree
\prooftree
\prooftree
\prooftree
\prooftree
\prooftree
\prooftree
\prooftree
\prooftree
\prooftree
\prooftree
\justifies
\mbox{\fbox{${\it CN}{\it s(n)}$}}\ \Rightarrow\ {\it CN}{\it s(n)}
\endprooftree
\justifies
\mbox{\fbox{${\square}{\it CN}{\it s(n)}$}}\ \Rightarrow\ {\it CN}{\it s(n)}
\using {\Box}L
\endprooftree
\prooftree
\justifies
\mbox{\fbox{$Nt(s(n))$}}\ \Rightarrow\ Nt(s(n))
\endprooftree
\justifies
\mbox{\fbox{$Nt(s(n))/{\it CN}{\it s(n)}$}}, {\square}{\it CN}{\it s(n)}\ \Rightarrow\ Nt(s(n))
\using {/}L
\endprooftree
\justifies
\mbox{\fbox{${\forall}n(Nt(n)/{\it CN}{\it n})$}}, {\square}{\it CN}{\it s(n)}\ \Rightarrow\ Nt(s(n))
\using {\forall}L
\endprooftree
\justifies
\mbox{\fbox{${\blacksquare}{\forall}n(Nt(n)/{\it CN}{\it n})$}}, {\square}{\it CN}{\it s(n)}\ \Rightarrow\ Nt(s(n))
\using {\blacksquare}L
\endprooftree
\justifies
{\blacksquare}{\forall}n(Nt(n)/{\it CN}{\it n}), {\square}{\it CN}{\it s(n)}\ \Rightarrow\ \fbox{${\exists}bNb$}
\using {\exists}R
\endprooftree
\prooftree
\justifies
{\it PP}{\it to}\ \Rightarrow\ {\it PP}{\it to}
\endprooftree
\justifies
{\blacksquare}{\forall}n(Nt(n)/{\it CN}{\it n}), {\square}{\it CN}{\it s(n)}, {\it PP}{\it to}\ \Rightarrow\ \fbox{${\exists}bNb{\bullet}{\it PP}{\it to}$}
\using {\bullet}R
\endprooftree
\prooftree
\prooftree
\prooftree
\prooftree
\justifies
Nt(s(m))\ \Rightarrow\ Nt(s(m))
\endprooftree
\justifies
Nt(s(m))\ \Rightarrow\ \fbox{${\exists}aNa$}
\using {\exists}R
\endprooftree
\justifies
[Nt(s(m))]\ \Rightarrow\ \fbox{${\langle\rangle}{\exists}aNa$}
\using {\langle\rangle}R
\endprooftree
\prooftree
\justifies
\mbox{\fbox{$Sf$}}\ \Rightarrow\ Sf
\endprooftree
\justifies
[Nt(s(m))], \mbox{\fbox{${\langle\rangle}{\exists}aNa\backslash Sf$}}\ \Rightarrow\ Sf
\using {\backslash}L
\endprooftree
\justifies
[Nt(s(m))], \mbox{\fbox{$({\langle\rangle}{\exists}aNa\backslash Sf)/({\exists}bNb{\bullet}{\it PP}{\it to})$}}, {\blacksquare}{\forall}n(Nt(n)/{\it CN}{\it n}), {\square}{\it CN}{\it s(n)}, {\it PP}{\it to}\ \Rightarrow\ Sf
\using {/}L
\endprooftree
\justifies
[Nt(s(m))], \mbox{\fbox{${\square}(({\langle\rangle}{\exists}aNa\backslash Sf)/({\exists}bNb{\bullet}{\it PP}{\it to}))$}}, {\blacksquare}{\forall}n(Nt(n)/{\it CN}{\it n}), {\square}{\it CN}{\it s(n)}, {\it PP}{\it to}\ \Rightarrow\ Sf
\using {\Box}L
\endprooftree
\justifies
{\langle\rangle}Nt(s(m)), {\square}(({\langle\rangle}{\exists}aNa\backslash Sf)/({\exists}bNb{\bullet}{\it PP}{\it to})), {\blacksquare}{\forall}n(Nt(n)/{\it CN}{\it n}), {\square}{\it CN}{\it s(n)}, {\it PP}{\it to}\ \Rightarrow\ Sf
\using {\langle\rangle}L
\endprooftree
\justifies
{\square}(({\langle\rangle}{\exists}aNa\backslash Sf)/({\exists}bNb{\bullet}{\it PP}{\it to})), {\blacksquare}{\forall}n(Nt(n)/{\it CN}{\it n}), {\square}{\it CN}{\it s(n)}, {\it PP}{\it to}\ \Rightarrow\ {\langle\rangle}Nt(s(m))\backslash Sf
\using {\backslash}R
\endprooftree
\justifies
{\square}(({\langle\rangle}{\exists}aNa\backslash Sf)/({\exists}bNb{\bullet}{\it PP}{\it to})), {\blacksquare}{\forall}n(Nt(n)/{\it CN}{\it n}), {\square}{\it CN}{\it s(n)}\ \Rightarrow\ ({\langle\rangle}Nt(s(m))\backslash Sf)/{\it PP}{\it to}
\using {/}R
\endprooftree
\justifies
{\square}(({\langle\rangle}{\exists}aNa\backslash Sf)/({\exists}bNb{\bullet}{\it PP}{\it to})), {\blacksquare}{\forall}n(Nt(n)/{\it CN}{\it n}), {\square}{\it CN}{\it s(n)}\ \Rightarrow\ {\blacksquare}(({\langle\rangle}Nt(s(m))\backslash Sf)/{\it PP}{\it to})
\using {\blacksquare}R
\endprooftree
\justifies
\begin{array}{c}
{\square}(({\langle\rangle}{\exists}aNa\backslash Sf)/({\exists}bNb{\bullet}{\it PP}{\it to})), {\blacksquare}{\forall}n(Nt(n)/{\it CN}{\it n}), {\square}{\it CN}{\it s(n)}\ \Rightarrow\ \fbox{$?{\blacksquare}(({\langle\rangle}Nt(s(m))\backslash Sf)/{\it PP}{\it to})$}\\
\mbox{\footnotesize\textcircled{2}}
\end{array}
\using {?}R
\endprooftree

\rotatebox{-90}{\scriptsize
\prooftree
\prooftree
\prooftree
\prooftree
\prooftree
\mbox{\footnotesize\textcircled{1}}\tab
\prooftree
\mbox{\footnotesize\textcircled{2}}\tab
\prooftree
\prooftree
\prooftree
\prooftree
\prooftree
\prooftree
\prooftree
\prooftree
\prooftree
\justifies
\mbox{\fbox{$Nt(s(f))$}}\ \Rightarrow\ Nt(s(f))
\endprooftree
\justifies
\mbox{\fbox{${\blacksquare}Nt(s(f))$}}\ \Rightarrow\ Nt(s(f))
\using {\blacksquare}L
\endprooftree
\justifies
{\blacksquare}Nt(s(f))\ \Rightarrow\ \fbox{${\exists}aNa$}
\using {\exists}R
\endprooftree
\prooftree
\justifies
\mbox{\fbox{${\it PP}{\it to}$}}\ \Rightarrow\ {\it PP}{\it to}
\endprooftree
\justifies
\mbox{\fbox{${\it PP}{\it to}/{\exists}aNa$}}, {\blacksquare}Nt(s(f))\ \Rightarrow\ {\it PP}{\it to}
\using {/}L
\endprooftree
\justifies
\mbox{\fbox{$({\it PP}{\it to}/{\exists}aNa){\sqcap}{\forall}n(({\langle\rangle}Nn\backslash Si)/({\langle\rangle}Nn\backslash Sb))$}}, {\blacksquare}Nt(s(f))\ \Rightarrow\ {\it PP}{\it to}
\using {\sqcap}L
\endprooftree
\justifies
\mbox{\fbox{${\blacksquare}(({\it PP}{\it to}/{\exists}aNa){\sqcap}{\forall}n(({\langle\rangle}Nn\backslash Si)/({\langle\rangle}Nn\backslash Sb)))$}}, {\blacksquare}Nt(s(f))\ \Rightarrow\ {\it PP}{\it to}
\using {\blacksquare}L
\endprooftree
\prooftree
\prooftree
\prooftree
\prooftree
\justifies
\mbox{\fbox{$Nt(s(m))$}}\ \Rightarrow\ Nt(s(m))
\endprooftree
\justifies
\mbox{\fbox{${\blacksquare}Nt(s(m))$}}\ \Rightarrow\ Nt(s(m))
\using {\blacksquare}L
\endprooftree
\justifies
[{\blacksquare}Nt(s(m))]\ \Rightarrow\ \fbox{${\langle\rangle}Nt(s(m))$}
\using {\langle\rangle}R
\endprooftree
\prooftree
\justifies
\mbox{\fbox{$Sf$}}\ \Rightarrow\ Sf
\endprooftree
\justifies
[{\blacksquare}Nt(s(m))], \mbox{\fbox{${\langle\rangle}Nt(s(m))\backslash Sf$}}\ \Rightarrow\ Sf
\using {\backslash}L
\endprooftree
\justifies
[{\blacksquare}Nt(s(m))], \mbox{\fbox{$({\langle\rangle}Nt(s(m))\backslash Sf)/{\it PP}{\it to}$}}, {\blacksquare}(({\it PP}{\it to}/{\exists}aNa){\sqcap}{\forall}n(({\langle\rangle}Nn\backslash Si)/({\langle\rangle}Nn\backslash Sb))), {\blacksquare}Nt(s(f))\ \Rightarrow\ Sf
\using {/}L
\endprooftree
\justifies
[{\blacksquare}Nt(s(m))], [\mbox{\fbox{${[]^{-1}}(({\langle\rangle}Nt(s(m))\backslash Sf)/{\it PP}{\it to})$}}], {\blacksquare}(({\it PP}{\it to}/{\exists}aNa){\sqcap}{\forall}n(({\langle\rangle}Nn\backslash Si)/({\langle\rangle}Nn\backslash Sb))), {\blacksquare}Nt(s(f))\ \Rightarrow\ Sf
\using {[]^{-1}}L
\endprooftree
\justifies
[{\blacksquare}Nt(s(m))], [[\mbox{\fbox{${[]^{-1}}{[]^{-1}}(({\langle\rangle}Nt(s(m))\backslash Sf)/{\it PP}{\it to})$}}]], {\blacksquare}(({\it PP}{\it to}/{\exists}aNa){\sqcap}{\forall}n(({\langle\rangle}Nn\backslash Si)/({\langle\rangle}Nn\backslash Sb))), {\blacksquare}Nt(s(f))\ \Rightarrow\ Sf
\using {[]^{-1}}L
\endprooftree
\justifies
[{\blacksquare}Nt(s(m))], [[{\square}(({\langle\rangle}{\exists}aNa\backslash Sf)/({\exists}bNb{\bullet}{\it PP}{\it to})), {\blacksquare}{\forall}n(Nt(n)/{\it CN}{\it n}), {\square}{\it CN}{\it s(n)}, \mbox{\fbox{$?{\blacksquare}(({\langle\rangle}Nt(s(m))\backslash Sf)/{\it PP}{\it to})\backslash {[]^{-1}}{[]^{-1}}(({\langle\rangle}Nt(s(m))\backslash Sf)/{\it PP}{\it to})$}}]], {\blacksquare}(({\it PP}{\it to}/{\exists}aNa){\sqcap}{\forall}n(({\langle\rangle}Nn\backslash Si)/({\langle\rangle}Nn\backslash Sb))), {\blacksquare}Nt(s(f))\ \Rightarrow\ Sf
\using {\backslash}L
\endprooftree
\justifies
[{\blacksquare}Nt(s(m))], [[{\square}(({\langle\rangle}{\exists}aNa\backslash Sf)/({\exists}bNb{\bullet}{\it PP}{\it to})), {\blacksquare}{\forall}n(Nt(n)/{\it CN}{\it n}), {\square}{\it CN}{\it s(n)}, \mbox{\fbox{$(?{\blacksquare}(({\langle\rangle}Nt(s(m))\backslash Sf)/{\it PP}{\it to})\backslash {[]^{-1}}{[]^{-1}}(({\langle\rangle}Nt(s(m))\backslash Sf)/{\it PP}{\it to}))/{\blacksquare}(({\langle\rangle}Nt(s(m))\backslash Sf)/{\it PP}{\it to})$}}, {\square}(({\langle\rangle}{\exists}aNa\backslash Sf)/({\exists}bNb{\bullet}{\it PP}{\it to})), {\blacksquare}{\forall}n(Nt(n)/{\it CN}{\it n}), {\square}{\it CN}{\it s(n)}]], {\blacksquare}(({\it PP}{\it to}/{\exists}aNa){\sqcap}{\forall}n(({\langle\rangle}Nn\backslash Si)/({\langle\rangle}Nn\backslash Sb))), {\blacksquare}Nt(s(f))\ \Rightarrow\ Sf
\using {/}L
\endprooftree
\justifies
[{\blacksquare}Nt(s(m))], [[{\square}(({\langle\rangle}{\exists}aNa\backslash Sf)/({\exists}bNb{\bullet}{\it PP}{\it to})), {\blacksquare}{\forall}n(Nt(n)/{\it CN}{\it n}), {\square}{\it CN}{\it s(n)}, \mbox{\fbox{${\forall}f((?{\blacksquare}(({\langle\rangle}Nt(s(m))\backslash Sf)/{\it PP}{\it to})\backslash {[]^{-1}}{[]^{-1}}(({\langle\rangle}Nt(s(m))\backslash Sf)/{\it PP}{\it to}))/{\blacksquare}(({\langle\rangle}Nt(s(m))\backslash Sf)/{\it PP}{\it to}))$}}, {\square}(({\langle\rangle}{\exists}aNa\backslash Sf)/({\exists}bNb{\bullet}{\it PP}{\it to})), {\blacksquare}{\forall}n(Nt(n)/{\it CN}{\it n}), {\square}{\it CN}{\it s(n)}]], {\blacksquare}(({\it PP}{\it to}/{\exists}aNa){\sqcap}{\forall}n(({\langle\rangle}Nn\backslash Si)/({\langle\rangle}Nn\backslash Sb))), {\blacksquare}Nt(s(f))\ \Rightarrow\ Sf
\using {\forall}L
\endprooftree
\justifies
[{\blacksquare}Nt(s(m))], [[{\square}(({\langle\rangle}{\exists}aNa\backslash Sf)/({\exists}bNb{\bullet}{\it PP}{\it to})), {\blacksquare}{\forall}n(Nt(n)/{\it CN}{\it n}), {\square}{\it CN}{\it s(n)}, \mbox{\fbox{${\forall}b{\forall}f((?{\blacksquare}(({\langle\rangle}Nt(s(m))\backslash Sf)/{\it PP}{\it b})\backslash {[]^{-1}}{[]^{-1}}(({\langle\rangle}Nt(s(m))\backslash Sf)/{\it PP}{\it b}))/{\blacksquare}(({\langle\rangle}Nt(s(m))\backslash Sf)/{\it PP}{\it b}))$}}, {\square}(({\langle\rangle}{\exists}aNa\backslash Sf)/({\exists}bNb{\bullet}{\it PP}{\it to})), {\blacksquare}{\forall}n(Nt(n)/{\it CN}{\it n}), {\square}{\it CN}{\it s(n)}]], {\blacksquare}(({\it PP}{\it to}/{\exists}aNa){\sqcap}{\forall}n(({\langle\rangle}Nn\backslash Si)/({\langle\rangle}Nn\backslash Sb))), {\blacksquare}Nt(s(f))\ \Rightarrow\ Sf
\using {\forall}L
\endprooftree
\justifies
[{\blacksquare}Nt(s(m))], [[{\square}(({\langle\rangle}{\exists}aNa\backslash Sf)/({\exists}bNb{\bullet}{\it PP}{\it to})), {\blacksquare}{\forall}n(Nt(n)/{\it CN}{\it n}), {\square}{\it CN}{\it s(n)}, \mbox{\fbox{${\forall}a{\forall}b{\forall}f((?{\blacksquare}(({\langle\rangle}Na\backslash Sf)/{\it PP}{\it b})\backslash {[]^{-1}}{[]^{-1}}(({\langle\rangle}Na\backslash Sf)/{\it PP}{\it b}))/{\blacksquare}(({\langle\rangle}Na\backslash Sf)/{\it PP}{\it b}))$}}, {\square}(({\langle\rangle}{\exists}aNa\backslash Sf)/({\exists}bNb{\bullet}{\it PP}{\it to})), {\blacksquare}{\forall}n(Nt(n)/{\it CN}{\it n}), {\square}{\it CN}{\it s(n)}]], {\blacksquare}(({\it PP}{\it to}/{\exists}aNa){\sqcap}{\forall}n(({\langle\rangle}Nn\backslash Si)/({\langle\rangle}Nn\backslash Sb))), {\blacksquare}Nt(s(f))\ \Rightarrow\ Sf
\using {\forall}L
\endprooftree
\justifies
[{\blacksquare}Nt(s(m))], [[{\square}(({\langle\rangle}{\exists}aNa\backslash Sf)/({\exists}bNb{\bullet}{\it PP}{\it to})), {\blacksquare}{\forall}n(Nt(n)/{\it CN}{\it n}), {\square}{\it CN}{\it s(n)}, \mbox{\fbox{${\blacksquare}{\forall}a{\forall}b{\forall}f((?{\blacksquare}(({\langle\rangle}Na\backslash Sf)/{\it PP}{\it b})\backslash {[]^{-1}}{[]^{-1}}(({\langle\rangle}Na\backslash Sf)/{\it PP}{\it b}))/{\blacksquare}(({\langle\rangle}Na\backslash Sf)/{\it PP}{\it b}))$}}, {\square}(({\langle\rangle}{\exists}aNa\backslash Sf)/({\exists}bNb{\bullet}{\it PP}{\it to})), {\blacksquare}{\forall}n(Nt(n)/{\it CN}{\it n}), {\square}{\it CN}{\it s(n)}]], {\blacksquare}(({\it PP}{\it to}/{\exists}aNa){\sqcap}{\forall}n(({\langle\rangle}Nn\backslash Si)/({\langle\rangle}Nn\backslash Sb))), {\blacksquare}Nt(s(f))\ \Rightarrow\ Sf
\using {\blacksquare}L
\endprooftree}}

\vspace{0.15in}

\noindent
This delivers semantics:
\disp{
$[({\it Past}\ (((\mbox{\v{}}{\it give}\ {\it m})\ (\iota \ \mbox{\v{}}{\it book}))\ {\it j}))\wedge ({\it Past}\ (((\mbox{\v{}}{\it sent}\ {\it m})\ (\iota \ \mbox{\v{}}{\it cd}))\ {\it j}))]$}

}

\subsection{Argument-cluster left node raising coordination}

The following example is of non-standard constituent argument-cluster coordination,
or, left node raising: 
\disp{
$[{\bf john}]{+}{\bf gave}{+}[[{\bf the}{+}{\bf book}{+}{\bf to}{+}{\bf mary}{+}{\bf and}{+}{\bf the}{+}{\bf cd}{+}{\bf to}{+}{\bf suzy}]]: Sf$}
Appropriate lexical lookup yields the following where the coordinator type is
essentially $(\exstexp X\bsl$ $\abrack\abrack X)/X$ with $X=((N\bsl S)/(N\product\PP))\bsl(N\bsl S)$
(using the uncurried prepositional ditransitive verb type):
\disp{
$\begin{array}[t]{l}
[{\blacksquare}Nt(s(m)): {\it j}], {\square}(({\langle\rangle}{\exists}aNa\backslash Sf)/({\exists}bNb{\bullet}{\it PP}{\it to})): \mbox{\^{}}\lambda A\lambda B({\it Past}\ (((\mbox{\v{}}{\it give}\ \pi_2{\it A})\ \pi_1{\it A})\ {\it B})), \\
{}[[{\blacksquare}{\forall}n(Nt(n)/{\it CN}{\it n}): \iota , {\square}{\it CN}{\it s(n)}: {\it book}, {\blacksquare}(({\it PP}{\it to}/{\exists}aNa){\sqcap}{\forall}n(({\langle\rangle}Nn\backslash Si)/({\langle\rangle}Nn\backslash Sb))):\\ \lambda C{\it C}, {\blacksquare}Nt(s(f)): {\it m},
 {\blacksquare}{\forall}a{\forall}b{\forall}f((?{\blacksquare}((({\langle\rangle}Na\backslash Sf)/({\exists}cNc{\bullet}{\it PP}{\it b}))\backslash ({\langle\rangle}Na\backslash Sf))\backslash\\ {[]^{-1}}{[]^{-1}}((({\langle\rangle}Na\backslash Sf)/({\exists}cNc{\bullet}{\it PP}{\it b}))\backslash ({\langle\rangle}Na\backslash Sf)))/
 {\blacksquare}((({\langle\rangle}Na\backslash Sf)/({\exists}cNc{\bullet}{\it PP}{\it b}))\backslash ({\langle\rangle}Na\backslash Sf))):\\ (\Phinplus\ ({\it s}\ ({\it s}\ {\it 0}))\ {\it and}), {\blacksquare}{\forall}n(Nt(n)/{\it CN}{\it n}): \iota , {\square}{\it CN}{\it s(n)}: {\it cd},\\
 {\blacksquare}(({\it PP}{\it to}/{\exists}aNa){\sqcap}{\forall}n(({\langle\rangle}Nn\backslash Si)/({\langle\rangle}Nn\backslash Sb))):\lambda D{\it D},{\blacksquare}Nt(s(f)): {\it s}]]\ \Rightarrow\ Sf
 \end{array}$}
This has the following derivation.
In \textcircled{1} the righthand conjunct is analysed as essentially of the shape
$((N\bsl S)/(N\product \PP))\bsl(N\bsl S)$.
The main action is in the initial unfolding of this succedent to yield a `canonical' sequent:
\disp{$\mini
\prooftree
\prooftree
N, (N\bsl S)/(N\product \PP), N/\CN, \CN, \PP, N\yields S
\justifies
(N\bsl S)/(N\product \PP), N/\CN, \CN, \PP, N\yields N\bsl S
\using \bsl R
\endprooftree
\justifies
N/\CN, \CN, \PP, N\yields ((N\bsl S)/(N\product \PP))\bsl(N\bsl S)
\using \bsl R
\endprooftree
$}
Subderivation \textcircled{2} is exactly the same --- except for the additional
bottommost existential
exponential right rule.
Reading upwards in the main derivation,
after modality elimination and instantiation of features on the coordinator type
and application to the two conjuncts, the brackets are checked and there is
application of the whole coordinate structure to the left node raised verb
$(N\bsl S)/(N\product \PP)$ (left subsubderivation) and to the subject $N$
(right subsubderivation).
\vspace{0.15in}
$${\tiny
\prooftree
\prooftree
\prooftree
\prooftree
\prooftree
\prooftree
\prooftree
\prooftree
\prooftree
\prooftree
\prooftree
\prooftree
\justifies
\mbox{\fbox{${\it CN}{\it s(n)}$}}\ \Rightarrow\ {\it CN}{\it s(n)}
\endprooftree
\justifies
\mbox{\fbox{${\square}{\it CN}{\it s(n)}$}}\ \Rightarrow\ {\it CN}{\it s(n)}
\using {\Box}L
\endprooftree
\prooftree
\justifies
\mbox{\fbox{$Nt(s(n))$}}\ \Rightarrow\ Nt(s(n))
\endprooftree
\justifies
\mbox{\fbox{$Nt(s(n))/{\it CN}{\it s(n)}$}}, {\square}{\it CN}{\it s(n)}\ \Rightarrow\ Nt(s(n))
\using {/}L
\endprooftree
\justifies
\mbox{\fbox{${\forall}n(Nt(n)/{\it CN}{\it n})$}}, {\square}{\it CN}{\it s(n)}\ \Rightarrow\ Nt(s(n))
\using {\forall}L
\endprooftree
\justifies
\mbox{\fbox{${\blacksquare}{\forall}n(Nt(n)/{\it CN}{\it n})$}}, {\square}{\it CN}{\it s(n)}\ \Rightarrow\ Nt(s(n))
\using {\blacksquare}L
\endprooftree
\justifies
{\blacksquare}{\forall}n(Nt(n)/{\it CN}{\it n}), {\square}{\it CN}{\it s(n)}\ \Rightarrow\ \fbox{${\exists}cNc$}
\using {\exists}R
\endprooftree
\prooftree
\prooftree
\prooftree
\prooftree
\prooftree
\prooftree
\justifies
\mbox{\fbox{$Nt(s(f))$}}\ \Rightarrow\ Nt(s(f))
\endprooftree
\justifies
\mbox{\fbox{${\blacksquare}Nt(s(f))$}}\ \Rightarrow\ Nt(s(f))
\using {\blacksquare}L
\endprooftree
\justifies
{\blacksquare}Nt(s(f))\ \Rightarrow\ \fbox{${\exists}aNa$}
\using {\exists}R
\endprooftree
\prooftree
\justifies
\mbox{\fbox{${\it PP}{\it to}$}}\ \Rightarrow\ {\it PP}{\it to}
\endprooftree
\justifies
\mbox{\fbox{${\it PP}{\it to}/{\exists}aNa$}}, {\blacksquare}Nt(s(f))\ \Rightarrow\ {\it PP}{\it to}
\using {/}L
\endprooftree
\justifies
\mbox{\fbox{$({\it PP}{\it to}/{\exists}aNa){\sqcap}{\forall}n(({\langle\rangle}Nn\backslash Si)/({\langle\rangle}Nn\backslash Sb))$}}, {\blacksquare}Nt(s(f))\ \Rightarrow\ {\it PP}{\it to}
\using {\sqcap}L
\endprooftree
\justifies
\mbox{\fbox{${\blacksquare}(({\it PP}{\it to}/{\exists}aNa){\sqcap}{\forall}n(({\langle\rangle}Nn\backslash Si)/({\langle\rangle}Nn\backslash Sb)))$}}, {\blacksquare}Nt(s(f))\ \Rightarrow\ {\it PP}{\it to}
\using {\blacksquare}L
\endprooftree
\justifies
{\blacksquare}{\forall}n(Nt(n)/{\it CN}{\it n}), {\square}{\it CN}{\it s(n)}, {\blacksquare}(({\it PP}{\it to}/{\exists}aNa){\sqcap}{\forall}n(({\langle\rangle}Nn\backslash Si)/({\langle\rangle}Nn\backslash Sb))), {\blacksquare}Nt(s(f))\ \Rightarrow\ \fbox{${\exists}cNc{\bullet}{\it PP}{\it to}$}
\using {\bullet}R
\endprooftree
\prooftree
\prooftree
\prooftree
\justifies
Nt(s(m))\ \Rightarrow\ Nt(s(m))
\endprooftree
\justifies
[Nt(s(m))]\ \Rightarrow\ \fbox{${\langle\rangle}Nt(s(m))$}
\using {\langle\rangle}R
\endprooftree
\prooftree
\justifies
\mbox{\fbox{$Sf$}}\ \Rightarrow\ Sf
\endprooftree
\justifies
[Nt(s(m))], \mbox{\fbox{${\langle\rangle}Nt(s(m))\backslash Sf$}}\ \Rightarrow\ Sf
\using {\backslash}L
\endprooftree
\justifies
[Nt(s(m))], \mbox{\fbox{$({\langle\rangle}Nt(s(m))\backslash Sf)/({\exists}cNc{\bullet}{\it PP}{\it to})$}}, {\blacksquare}{\forall}n(Nt(n)/{\it CN}{\it n}), {\square}{\it CN}{\it s(n)}, {\blacksquare}(({\it PP}{\it to}/{\exists}aNa){\sqcap}{\forall}n(({\langle\rangle}Nn\backslash Si)/({\langle\rangle}Nn\backslash Sb))), {\blacksquare}Nt(s(f))\ \Rightarrow\ Sf
\using {/}L
\endprooftree
\justifies
{\langle\rangle}Nt(s(m)), ({\langle\rangle}Nt(s(m))\backslash Sf)/({\exists}cNc{\bullet}{\it PP}{\it to}), {\blacksquare}{\forall}n(Nt(n)/{\it CN}{\it n}), {\square}{\it CN}{\it s(n)}, {\blacksquare}(({\it PP}{\it to}/{\exists}aNa){\sqcap}{\forall}n(({\langle\rangle}Nn\backslash Si)/({\langle\rangle}Nn\backslash Sb))), {\blacksquare}Nt(s(f))\ \Rightarrow\ Sf
\using {\langle\rangle}L
\endprooftree
\justifies
({\langle\rangle}Nt(s(m))\backslash Sf)/({\exists}cNc{\bullet}{\it PP}{\it to}), {\blacksquare}{\forall}n(Nt(n)/{\it CN}{\it n}), {\square}{\it CN}{\it s(n)}, {\blacksquare}(({\it PP}{\it to}/{\exists}aNa){\sqcap}{\forall}n(({\langle\rangle}Nn\backslash Si)/({\langle\rangle}Nn\backslash Sb))), {\blacksquare}Nt(s(f))\ \Rightarrow\ {\langle\rangle}Nt(s(m))\backslash Sf
\using {\backslash}R
\endprooftree
\justifies
{\blacksquare}{\forall}n(Nt(n)/{\it CN}{\it n}), {\square}{\it CN}{\it s(n)}, {\blacksquare}(({\it PP}{\it to}/{\exists}aNa){\sqcap}{\forall}n(({\langle\rangle}Nn\backslash Si)/({\langle\rangle}Nn\backslash Sb))), {\blacksquare}Nt(s(f))\ \Rightarrow\ (({\langle\rangle}Nt(s(m))\backslash Sf)/({\exists}cNc{\bullet}{\it PP}{\it to}))\backslash ({\langle\rangle}Nt(s(m))\backslash Sf)
\using {\backslash}R
\endprooftree
\justifies
\begin{array}{c}
{\blacksquare}{\forall}n(Nt(n)/{\it CN}{\it n}), {\square}{\it CN}{\it s(n)}, {\blacksquare}(({\it PP}{\it to}/{\exists}aNa){\sqcap}{\forall}n(({\langle\rangle}Nn\backslash Si)/({\langle\rangle}Nn\backslash Sb))), {\blacksquare}Nt(s(f))\ \Rightarrow\ {\blacksquare}((({\langle\rangle}Nt(s(m))\backslash Sf)/({\exists}cNc{\bullet}{\it PP}{\it to}))\backslash ({\langle\rangle}Nt(s(m))\backslash Sf))\\
\mbox{\footnotesize\textcircled{1}}
\end{array}
\using {\blacksquare}R
\endprooftree}
$$
$$
{\tiny
\prooftree
\prooftree
\prooftree
\prooftree
\prooftree
\prooftree
\prooftree
\prooftree
\prooftree
\prooftree
\prooftree
\prooftree
\prooftree
\justifies
\mbox{\fbox{${\it CN}{\it s(n)}$}}\ \Rightarrow\ {\it CN}{\it s(n)}
\endprooftree
\justifies
\mbox{\fbox{${\square}{\it CN}{\it s(n)}$}}\ \Rightarrow\ {\it CN}{\it s(n)}
\using {\Box}L
\endprooftree
\prooftree
\justifies
\mbox{\fbox{$Nt(s(n))$}}\ \Rightarrow\ Nt(s(n))
\endprooftree
\justifies
\mbox{\fbox{$Nt(s(n))/{\it CN}{\it s(n)}$}}, {\square}{\it CN}{\it s(n)}\ \Rightarrow\ Nt(s(n))
\using {/}L
\endprooftree
\justifies
\mbox{\fbox{${\forall}n(Nt(n)/{\it CN}{\it n})$}}, {\square}{\it CN}{\it s(n)}\ \Rightarrow\ Nt(s(n))
\using {\forall}L
\endprooftree
\justifies
\mbox{\fbox{${\blacksquare}{\forall}n(Nt(n)/{\it CN}{\it n})$}}, {\square}{\it CN}{\it s(n)}\ \Rightarrow\ Nt(s(n))
\using {\blacksquare}L
\endprooftree
\justifies
{\blacksquare}{\forall}n(Nt(n)/{\it CN}{\it n}), {\square}{\it CN}{\it s(n)}\ \Rightarrow\ \fbox{${\exists}cNc$}
\using {\exists}R
\endprooftree
\prooftree
\prooftree
\prooftree
\prooftree
\prooftree
\prooftree
\justifies
\mbox{\fbox{$Nt(s(f))$}}\ \Rightarrow\ Nt(s(f))
\endprooftree
\justifies
\mbox{\fbox{${\blacksquare}Nt(s(f))$}}\ \Rightarrow\ Nt(s(f))
\using {\blacksquare}L
\endprooftree
\justifies
{\blacksquare}Nt(s(f))\ \Rightarrow\ \fbox{${\exists}aNa$}
\using {\exists}R
\endprooftree
\prooftree
\justifies
\mbox{\fbox{${\it PP}{\it to}$}}\ \Rightarrow\ {\it PP}{\it to}
\endprooftree
\justifies
\mbox{\fbox{${\it PP}{\it to}/{\exists}aNa$}}, {\blacksquare}Nt(s(f))\ \Rightarrow\ {\it PP}{\it to}
\using {/}L
\endprooftree
\justifies
\mbox{\fbox{$({\it PP}{\it to}/{\exists}aNa){\sqcap}{\forall}n(({\langle\rangle}Nn\backslash Si)/({\langle\rangle}Nn\backslash Sb))$}}, {\blacksquare}Nt(s(f))\ \Rightarrow\ {\it PP}{\it to}
\using {\sqcap}L
\endprooftree
\justifies
\mbox{\fbox{${\blacksquare}(({\it PP}{\it to}/{\exists}aNa){\sqcap}{\forall}n(({\langle\rangle}Nn\backslash Si)/({\langle\rangle}Nn\backslash Sb)))$}}, {\blacksquare}Nt(s(f))\ \Rightarrow\ {\it PP}{\it to}
\using {\blacksquare}L
\endprooftree
\justifies
{\blacksquare}{\forall}n(Nt(n)/{\it CN}{\it n}), {\square}{\it CN}{\it s(n)}, {\blacksquare}(({\it PP}{\it to}/{\exists}aNa){\sqcap}{\forall}n(({\langle\rangle}Nn\backslash Si)/({\langle\rangle}Nn\backslash Sb))), {\blacksquare}Nt(s(f))\ \Rightarrow\ \fbox{${\exists}cNc{\bullet}{\it PP}{\it to}$}
\using {\bullet}R
\endprooftree
\prooftree
\prooftree
\prooftree
\justifies
Nt(s(m))\ \Rightarrow\ Nt(s(m))
\endprooftree
\justifies
[Nt(s(m))]\ \Rightarrow\ \fbox{${\langle\rangle}Nt(s(m))$}
\using {\langle\rangle}R
\endprooftree
\prooftree
\justifies
\mbox{\fbox{$Sf$}}\ \Rightarrow\ Sf
\endprooftree
\justifies
[Nt(s(m))], \mbox{\fbox{${\langle\rangle}Nt(s(m))\backslash Sf$}}\ \Rightarrow\ Sf
\using {\backslash}L
\endprooftree
\justifies
[Nt(s(m))], \mbox{\fbox{$({\langle\rangle}Nt(s(m))\backslash Sf)/({\exists}cNc{\bullet}{\it PP}{\it to})$}}, {\blacksquare}{\forall}n(Nt(n)/{\it CN}{\it n}), {\square}{\it CN}{\it s(n)}, {\blacksquare}(({\it PP}{\it to}/{\exists}aNa){\sqcap}{\forall}n(({\langle\rangle}Nn\backslash Si)/({\langle\rangle}Nn\backslash Sb))), {\blacksquare}Nt(s(f))\ \Rightarrow\ Sf
\using {/}L
\endprooftree
\justifies
{\langle\rangle}Nt(s(m)), ({\langle\rangle}Nt(s(m))\backslash Sf)/({\exists}cNc{\bullet}{\it PP}{\it to}), {\blacksquare}{\forall}n(Nt(n)/{\it CN}{\it n}), {\square}{\it CN}{\it s(n)}, {\blacksquare}(({\it PP}{\it to}/{\exists}aNa){\sqcap}{\forall}n(({\langle\rangle}Nn\backslash Si)/({\langle\rangle}Nn\backslash Sb))), {\blacksquare}Nt(s(f))\ \Rightarrow\ Sf
\using {\langle\rangle}L
\endprooftree
\justifies
({\langle\rangle}Nt(s(m))\backslash Sf)/({\exists}cNc{\bullet}{\it PP}{\it to}), {\blacksquare}{\forall}n(Nt(n)/{\it CN}{\it n}), {\square}{\it CN}{\it s(n)}, {\blacksquare}(({\it PP}{\it to}/{\exists}aNa){\sqcap}{\forall}n(({\langle\rangle}Nn\backslash Si)/({\langle\rangle}Nn\backslash Sb))), {\blacksquare}Nt(s(f))\ \Rightarrow\ {\langle\rangle}Nt(s(m))\backslash Sf
\using {\backslash}R
\endprooftree
\justifies
{\blacksquare}{\forall}n(Nt(n)/{\it CN}{\it n}), {\square}{\it CN}{\it s(n)}, {\blacksquare}(({\it PP}{\it to}/{\exists}aNa){\sqcap}{\forall}n(({\langle\rangle}Nn\backslash Si)/({\langle\rangle}Nn\backslash Sb))), {\blacksquare}Nt(s(f))\ \Rightarrow\ (({\langle\rangle}Nt(s(m))\backslash Sf)/({\exists}cNc{\bullet}{\it PP}{\it to}))\backslash ({\langle\rangle}Nt(s(m))\backslash Sf)
\using {\backslash}R
\endprooftree
\justifies
{\blacksquare}{\forall}n(Nt(n)/{\it CN}{\it n}), {\square}{\it CN}{\it s(n)}, {\blacksquare}(({\it PP}{\it to}/{\exists}aNa){\sqcap}{\forall}n(({\langle\rangle}Nn\backslash Si)/({\langle\rangle}Nn\backslash Sb))), {\blacksquare}Nt(s(f))\ \Rightarrow\ {\blacksquare}((({\langle\rangle}Nt(s(m))\backslash Sf)/({\exists}cNc{\bullet}{\it PP}{\it to}))\backslash ({\langle\rangle}Nt(s(m))\backslash Sf))
\using {\blacksquare}R
\endprooftree
\justifies
\begin{array}{c}
{\blacksquare}{\forall}n(Nt(n)/{\it CN}{\it n}), {\square}{\it CN}{\it s(n)}, {\blacksquare}(({\it PP}{\it to}/{\exists}aNa){\sqcap}{\forall}n(({\langle\rangle}Nn\backslash Si)/({\langle\rangle}Nn\backslash Sb))), {\blacksquare}Nt(s(f))\ \Rightarrow\ \fbox{$?{\blacksquare}((({\langle\rangle}Nt(s(m))\backslash Sf)/({\exists}cNc{\bullet}{\it PP}{\it to}))\backslash ({\langle\rangle}Nt(s(m))\backslash Sf))$}\\
\mbox{\footnotesize\textcircled{2}}
\end{array}
\using {?}R
\endprooftree}
$$
$$
\rotatebox{0}{\tiny
\prooftree
\prooftree
\prooftree
\prooftree
\prooftree
\mbox{\footnotesize\textcircled{1}}\tab
\prooftree
\mbox{\footnotesize\textcircled{2}}\tab
\prooftree
\prooftree
\prooftree
\prooftree
\prooftree
\prooftree
\prooftree
\prooftree
\prooftree
\prooftree
\prooftree
\prooftree
\prooftree
\justifies
N5\ \Rightarrow\ N5
\endprooftree
\justifies
N5\ \Rightarrow\ \fbox{${\exists}bNb$}
\using {\exists}R
\endprooftree
\prooftree
\justifies
{\it PP}{\it to}\ \Rightarrow\ {\it PP}{\it to}
\endprooftree
\justifies
N5, {\it PP}{\it to}\ \Rightarrow\ \fbox{${\exists}bNb{\bullet}{\it PP}{\it to}$}
\using {\bullet}R
\endprooftree
\prooftree
\prooftree
\prooftree
\prooftree
\justifies
Nt(s(m))\ \Rightarrow\ Nt(s(m))
\endprooftree
\justifies
Nt(s(m))\ \Rightarrow\ \fbox{${\exists}aNa$}
\using {\exists}R
\endprooftree
\justifies
[Nt(s(m))]\ \Rightarrow\ \fbox{${\langle\rangle}{\exists}aNa$}
\using {\langle\rangle}R
\endprooftree
\prooftree
\justifies
\mbox{\fbox{$Sf$}}\ \Rightarrow\ Sf
\endprooftree
\justifies
[Nt(s(m))], \mbox{\fbox{${\langle\rangle}{\exists}aNa\backslash Sf$}}\ \Rightarrow\ Sf
\using {\backslash}L
\endprooftree
\justifies
[Nt(s(m))], \mbox{\fbox{$({\langle\rangle}{\exists}aNa\backslash Sf)/({\exists}bNb{\bullet}{\it PP}{\it to})$}}, N5, {\it PP}{\it to}\ \Rightarrow\ Sf
\using {/}L
\endprooftree
\justifies
[Nt(s(m))], \mbox{\fbox{${\square}(({\langle\rangle}{\exists}aNa\backslash Sf)/({\exists}bNb{\bullet}{\it PP}{\it to}))$}}, N5, {\it PP}{\it to}\ \Rightarrow\ Sf
\using {\Box}L
\endprooftree
\justifies
[Nt(s(m))], {\square}(({\langle\rangle}{\exists}aNa\backslash Sf)/({\exists}bNb{\bullet}{\it PP}{\it to})), {\exists}cNc, {\it PP}{\it to}\ \Rightarrow\ Sf
\using {\exists}L
\endprooftree
\justifies
{\langle\rangle}Nt(s(m)), {\square}(({\langle\rangle}{\exists}aNa\backslash Sf)/({\exists}bNb{\bullet}{\it PP}{\it to})), {\exists}cNc, {\it PP}{\it to}\ \Rightarrow\ Sf
\using {\langle\rangle}L
\endprooftree
\justifies
{\langle\rangle}Nt(s(m)), {\square}(({\langle\rangle}{\exists}aNa\backslash Sf)/({\exists}bNb{\bullet}{\it PP}{\it to})), {\exists}cNc{\bullet}{\it PP}{\it to}\ \Rightarrow\ Sf
\using {\bullet}L
\endprooftree
\justifies
{\square}(({\langle\rangle}{\exists}aNa\backslash Sf)/({\exists}bNb{\bullet}{\it PP}{\it to})), {\exists}cNc{\bullet}{\it PP}{\it to}\ \Rightarrow\ {\langle\rangle}Nt(s(m))\backslash Sf
\using {\backslash}R
\endprooftree
\justifies
{\square}(({\langle\rangle}{\exists}aNa\backslash Sf)/({\exists}bNb{\bullet}{\it PP}{\it to}))\ \Rightarrow\ ({\langle\rangle}Nt(s(m))\backslash Sf)/({\exists}cNc{\bullet}{\it PP}{\it to})
\using {/}R
\endprooftree
\prooftree
\prooftree
\prooftree
\prooftree
\justifies
\mbox{\fbox{$Nt(s(m))$}}\ \Rightarrow\ Nt(s(m))
\endprooftree
\justifies
\mbox{\fbox{${\blacksquare}Nt(s(m))$}}\ \Rightarrow\ Nt(s(m))
\using {\blacksquare}L
\endprooftree
\justifies
[{\blacksquare}Nt(s(m))]\ \Rightarrow\ \fbox{${\langle\rangle}Nt(s(m))$}
\using {\langle\rangle}R
\endprooftree
\prooftree
\justifies
\mbox{\fbox{$Sf$}}\ \Rightarrow\ Sf
\endprooftree
\justifies
[{\blacksquare}Nt(s(m))], \mbox{\fbox{${\langle\rangle}Nt(s(m))\backslash Sf$}}\ \Rightarrow\ Sf
\using {\backslash}L
\endprooftree
\justifies
[{\blacksquare}Nt(s(m))], {\square}(({\langle\rangle}{\exists}aNa\backslash Sf)/({\exists}bNb{\bullet}{\it PP}{\it to})), \mbox{\fbox{$(({\langle\rangle}Nt(s(m))\backslash Sf)/({\exists}cNc{\bullet}{\it PP}{\it to}))\backslash ({\langle\rangle}Nt(s(m))\backslash Sf)$}}\ \Rightarrow\ Sf
\using {\backslash}L
\endprooftree
\justifies
[{\blacksquare}Nt(s(m))], {\square}(({\langle\rangle}{\exists}aNa\backslash Sf)/({\exists}bNb{\bullet}{\it PP}{\it to})), [\mbox{\fbox{${[]^{-1}}((({\langle\rangle}Nt(s(m))\backslash Sf)/({\exists}cNc{\bullet}{\it PP}{\it to}))\backslash ({\langle\rangle}Nt(s(m))\backslash Sf))$}}]\ \Rightarrow\ Sf
\using {[]^{-1}}L
\endprooftree
\justifies
[{\blacksquare}Nt(s(m))], {\square}(({\langle\rangle}{\exists}aNa\backslash Sf)/({\exists}bNb{\bullet}{\it PP}{\it to})), [[\mbox{\fbox{${[]^{-1}}{[]^{-1}}((({\langle\rangle}Nt(s(m))\backslash Sf)/({\exists}cNc{\bullet}{\it PP}{\it to}))\backslash ({\langle\rangle}Nt(s(m))\backslash Sf))$}}]]\ \Rightarrow\ Sf
\using {[]^{-1}}L
\endprooftree
\justifies
\begin{array}{c}
{}[{\blacksquare}Nt(s(m))], {\square}(({\langle\rangle}{\exists}aNa\backslash Sf)/({\exists}bNb{\bullet}{\it PP}{\it to})), [[{\blacksquare}{\forall}n(Nt(n)/{\it CN}{\it n}), {\square}{\it CN}{\it s(n)}, {\blacksquare}(({\it PP}{\it to}/{\exists}aNa){\sqcap}{\forall}n(({\langle\rangle}Nn\backslash Si)/({\langle\rangle}Nn\backslash Sb))), {\blacksquare}Nt(s(f)),\\
 \mbox{\fbox{$?{\blacksquare}((({\langle\rangle}Nt(s(m))\backslash Sf)/({\exists}cNc{\bullet}{\it PP}{\it to}))\backslash ({\langle\rangle}Nt(s(m))\backslash Sf))\backslash {[]^{-1}}{[]^{-1}}((({\langle\rangle}Nt(s(m))\backslash Sf)/({\exists}cNc{\bullet}{\it PP}{\it to}))\backslash ({\langle\rangle}Nt(s(m))\backslash Sf))$}}]]\ \Rightarrow\ Sf
\end{array}
\using {\backslash}L
\endprooftree
\justifies
\begin{array}{c}
{}[{\blacksquare}Nt(s(m))], {\square}(({\langle\rangle}{\exists}aNa\backslash Sf)/({\exists}bNb{\bullet}{\it PP}{\it to})), [[{\blacksquare}{\forall}n(Nt(n)/{\it CN}{\it n}), {\square}{\it CN}{\it s(n)}, {\blacksquare}(({\it PP}{\it to}/{\exists}aNa){\sqcap}{\forall}n(({\langle\rangle}Nn\backslash Si)/({\langle\rangle}Nn\backslash Sb))), {\blacksquare}Nt(s(f)),\\
\mbox{\fbox{$(?{\blacksquare}((({\langle\rangle}Nt(s(m))\backslash Sf)/({\exists}cNc{\bullet}{\it PP}{\it to}))\backslash ({\langle\rangle}Nt(s(m))\backslash Sf))\backslash {[]^{-1}}{[]^{-1}}((({\langle\rangle}Nt(s(m))\backslash Sf)/({\exists}cNc{\bullet}{\it PP}{\it to}))\backslash ({\langle\rangle}Nt(s(m))\backslash Sf)))/{\blacksquare}((({\langle\rangle}Nt(s(m))\backslash Sf)/({\exists}cNc{\bullet}{\it PP}{\it to}))\backslash ({\langle\rangle}Nt(s(m))\backslash Sf))$}},\\
{\blacksquare}{\forall}n(Nt(n)/{\it CN}{\it n}), {\square}{\it CN}{\it s(n)}, {\blacksquare}(({\it PP}{\it to}/{\exists}aNa){\sqcap}{\forall}n(({\langle\rangle}Nn\backslash Si)/({\langle\rangle}Nn\backslash Sb))), {\blacksquare}Nt(s(f))]]\ \Rightarrow\ Sf
\end{array}
\using {/}L
\endprooftree
\justifies
\begin{array}{c}
{}[{\blacksquare}Nt(s(m))], {\square}(({\langle\rangle}{\exists}aNa\backslash Sf)/({\exists}bNb{\bullet}{\it PP}{\it to})), [[{\blacksquare}{\forall}n(Nt(n)/{\it CN}{\it n}), {\square}{\it CN}{\it s(n)}, {\blacksquare}(({\it PP}{\it to}/{\exists}aNa){\sqcap}{\forall}n(({\langle\rangle}Nn\backslash Si)/({\langle\rangle}Nn\backslash Sb))), {\blacksquare}Nt(s(f)),\\
\mbox{\fbox{${\forall}f((?{\blacksquare}((({\langle\rangle}Nt(s(m))\backslash Sf)/({\exists}cNc{\bullet}{\it PP}{\it to}))\backslash ({\langle\rangle}Nt(s(m))\backslash Sf))\backslash {[]^{-1}}{[]^{-1}}((({\langle\rangle}Nt(s(m))\backslash Sf)/({\exists}cNc{\bullet}{\it PP}{\it to}))\backslash ({\langle\rangle}Nt(s(m))\backslash Sf)))/{\blacksquare}((({\langle\rangle}Nt(s(m))\backslash Sf)/({\exists}cNc{\bullet}{\it PP}{\it to}))\backslash ({\langle\rangle}Nt(s(m))\backslash Sf)))$}},\\
{\blacksquare}{\forall}n(Nt(n)/{\it CN}{\it n}), {\square}{\it CN}{\it s(n)}, {\blacksquare}(({\it PP}{\it to}/{\exists}aNa){\sqcap}{\forall}n(({\langle\rangle}Nn\backslash Si)/({\langle\rangle}Nn\backslash Sb))), {\blacksquare}Nt(s(f))]]\ \Rightarrow\ Sf
\end{array}
\using {\forall}L
\endprooftree
\justifies
\begin{array}{c}
{}[{\blacksquare}Nt(s(m))], {\square}(({\langle\rangle}{\exists}aNa\backslash Sf)/({\exists}bNb{\bullet}{\it PP}{\it to})), [[{\blacksquare}{\forall}n(Nt(n)/{\it CN}{\it n}), {\square}{\it CN}{\it s(n)}, {\blacksquare}(({\it PP}{\it to}/{\exists}aNa){\sqcap}{\forall}n(({\langle\rangle}Nn\backslash Si)/({\langle\rangle}Nn\backslash Sb))), {\blacksquare}Nt(s(f)),\\
\mbox{\fbox{${\forall}b{\forall}f((?{\blacksquare}((({\langle\rangle}Nt(s(m))\backslash Sf)/({\exists}cNc{\bullet}{\it PP}{\it b}))\backslash ({\langle\rangle}Nt(s(m))\backslash Sf))\backslash {[]^{-1}}{[]^{-1}}((({\langle\rangle}Nt(s(m))\backslash Sf)/({\exists}cNc{\bullet}{\it PP}{\it b}))\backslash ({\langle\rangle}Nt(s(m))\backslash Sf)))/{\blacksquare}((({\langle\rangle}Nt(s(m))\backslash Sf)/({\exists}cNc{\bullet}{\it PP}{\it b}))\backslash ({\langle\rangle}Nt(s(m))\backslash Sf)))$}},\\
{\blacksquare}{\forall}n(Nt(n)/{\it CN}{\it n}), {\square}{\it CN}{\it s(n)}, {\blacksquare}(({\it PP}{\it to}/{\exists}aNa){\sqcap}{\forall}n(({\langle\rangle}Nn\backslash Si)/({\langle\rangle}Nn\backslash Sb))), {\blacksquare}Nt(s(f))]]\ \Rightarrow\ Sf
\end{array}
\using {\forall}L
\endprooftree
\justifies
\begin{array}{c}
{}[{\blacksquare}Nt(s(m))], {\square}(({\langle\rangle}{\exists}aNa\backslash Sf)/({\exists}bNb{\bullet}{\it PP}{\it to})), [[{\blacksquare}{\forall}n(Nt(n)/{\it CN}{\it n}), {\square}{\it CN}{\it s(n)}, {\blacksquare}(({\it PP}{\it to}/{\exists}aNa){\sqcap}{\forall}n(({\langle\rangle}Nn\backslash Si)/({\langle\rangle}Nn\backslash Sb))), {\blacksquare}Nt(s(f)),\\
\mbox{\fbox{${\forall}a{\forall}b{\forall}f((?{\blacksquare}((({\langle\rangle}Na\backslash Sf)/({\exists}cNc{\bullet}{\it PP}{\it b}))\backslash ({\langle\rangle}Na\backslash Sf))\backslash {[]^{-1}}{[]^{-1}}((({\langle\rangle}Na\backslash Sf)/({\exists}cNc{\bullet}{\it PP}{\it b}))\backslash ({\langle\rangle}Na\backslash Sf)))/{\blacksquare}((({\langle\rangle}Na\backslash Sf)/({\exists}cNc{\bullet}{\it PP}{\it b}))\backslash ({\langle\rangle}Na\backslash Sf)))$}},\\
{\blacksquare}{\forall}n(Nt(n)/{\it CN}{\it n}), {\square}{\it CN}{\it s(n)}, {\blacksquare}(({\it PP}{\it to}/{\exists}aNa){\sqcap}{\forall}n(({\langle\rangle}Nn\backslash Si)/({\langle\rangle}Nn\backslash Sb))), {\blacksquare}Nt(s(f))]]\ \Rightarrow\ Sf
\end{array}
\using {\forall}L
\endprooftree
\justifies
\begin{array}{c}
{}[{\blacksquare}Nt(s(m))], {\square}(({\langle\rangle}{\exists}aNa\backslash Sf)/({\exists}bNb{\bullet}{\it PP}{\it to})), [[{\blacksquare}{\forall}n(Nt(n)/{\it CN}{\it n}), {\square}{\it CN}{\it s(n)}, {\blacksquare}(({\it PP}{\it to}/{\exists}aNa){\sqcap}{\forall}n(({\langle\rangle}Nn\backslash Si)/({\langle\rangle}Nn\backslash Sb))), {\blacksquare}Nt(s(f)),\\
\mbox{\fbox{${\blacksquare}{\forall}a{\forall}b{\forall}f((?{\blacksquare}((({\langle\rangle}Na\backslash Sf)/({\exists}cNc{\bullet}{\it PP}{\it b}))\backslash ({\langle\rangle}Na\backslash Sf))\backslash {[]^{-1}}{[]^{-1}}((({\langle\rangle}Na\backslash Sf)/({\exists}cNc{\bullet}{\it PP}{\it b}))\backslash ({\langle\rangle}Na\backslash Sf)))/{\blacksquare}((({\langle\rangle}Na\backslash Sf)/({\exists}cNc{\bullet}{\it PP}{\it b}))\backslash ({\langle\rangle}Na\backslash Sf)))$}},\\
{\blacksquare}{\forall}n(Nt(n)/{\it CN}{\it n}), {\square}{\it CN}{\it s(n)}, {\blacksquare}(({\it PP}{\it to}/{\exists}aNa){\sqcap}{\forall}n(({\langle\rangle}Nn\backslash Si)/({\langle\rangle}Nn\backslash Sb))), {\blacksquare}Nt(s(f))]]\ \Rightarrow\ Sf
\end{array}
\using {\blacksquare}L
\endprooftree}
$$
\vspace{0.15in}
\noindent
All this correctly assigns semantics:
\disp{
$[({\it Past}\ (((\mbox{\v{}}{\it give}\ {\it m})\ (\iota \ \mbox{\v{}}{\it book}))\ {\it j}))\wedge ({\it Past}\ (((\mbox{\v{}}{\it give}\ {\it s})\ (\iota \ \mbox{\v{}}{\it cd}))\ {\it j}))]$}

\subsection{Argument plus modifier left node raising coordination}

The next example has LNR with arguments and adverbs in the conjuncts:
\disp{
$[{\bf john}]{+}{\bf saw}{+}[[{\bf mary}{+}{\bf today}{+}{\bf and}{+}{\bf bill}{+}{\bf yesterday}]]: Sf$}
Appropriate lexical lookup yields the following where the
coordinator is essentially of the form $(\exstexp X\bsl \abrack\abrack X)/X$ where
$X=((N\bsl S)/N)\bsl(N\bsl S)$.\footnote{Here `saw` is polymorphic between seeing an entity ({\it seee})
and seeing a proposition ({\it seet}).
We shall see in Section~\ref{unlikesect} how this allows coordination over `unlike' types.
The coordination here is over regular transitive verbs,
but we wish to show now the integration of node raising with the propensity
for such other features.}
\disp{
$\begin{array}[t]{l}
[{\blacksquare}Nt(s(m)): {\it j}], {\square}(({\langle\rangle}{\exists}aNa\backslash Sf)/({\exists}aNa{\oplus}{\it CP}that)):\\
\mbox{\^{}}\lambda A\lambda B({\it Past}\ (({\it A}\casearrow C.(\mbox{\v{}}{\it seee}\ {\it C}); D.(\mbox{\v{}}{\it seet}\ {\it D}))\ {\it B})),\\
{} [[{\blacksquare}Nt(s(f)): {\it m}, {\square}{\forall}a{\forall}f(({\langle\rangle}Na\backslash Sf)\backslash ({\langle\rangle}Na\backslash Sf)): \mbox{\^{}}\lambda E\lambda F(\mbox{\v{}}{\it today}\ ({\it E}\ {\it F})), \\{\blacksquare}{\forall}f{\forall}a((?{\blacksquare}((({\langle\rangle}Na\backslash Sf)/{\exists}bNb)\backslash ({\langle\rangle}Na\backslash Sf))\backslash
 {[]^{-1}}{[]^{-1}}((({\langle\rangle}Na\backslash Sf)/{\exists}bNb)\backslash ({\langle\rangle}Na\backslash Sf)))/\\{\blacksquare}((({\langle\rangle}Na\backslash Sf)/{\exists}bNb)\backslash ({\langle\rangle}Na\backslash Sf))): (\Phinplus\ ({\it s}\ ({\it s}\ {\it 0}))\ {\it and}), {\blacksquare}Nt(s(m)): {\it b}, \\{\square}{\forall}a{\forall}f(({\langle\rangle}Na\backslash Sf)\backslash ({\langle\rangle}Na\backslash Sf)): \mbox{\^{}}\lambda G\lambda H(\mbox{\v{}}{\it yesterday}\ ({\it G}\ {\it H}))]]\ \Rightarrow\ Sf
 \end{array}$}
The example has the following derivation.
In \textcircled{1} the righthand conjunct is analysed as essentially of the shape
$((N\bsl S)/N)\bsl(N\bsl S)$. The main action is in the initial unfolding of this succedent
to yield a `canonical' sequent:
\disp{$\mini
\prooftree
\prooftree
N, (N\bsl S)/N, N, (N\bsl S)\bsl(N\bsl S)\yields S
\justifies
(N\bsl S)/N, N, (N\bsl S)\bsl(N\bsl S)\yields N\bsl S
\using \bsl R
\endprooftree
\justifies
N, (N\bsl S)\bsl(N\bsl S)\yields ((N\bsl S)/N)\bsl(N\bsl S)
\using \bsl R
\endprooftree$}
The subtree \textcircled{2} for the lefthand conjunct is exactly the same --- except
for the bottommost existential exponential right rule and the gender of the object ---
hence it has been elided.
The trunk of the main derivation and the checking of the bracket context
are fairly standard by now. In the left and right subsubderivations
the (polymorphic) verb type is shown to yield the left node raised
transitive verb $(N\bsl S)/N$ and subject $N$ coordinate structure arguments.


{\tiny
\begin{center}
\prooftree
\prooftree
\prooftree
\prooftree
\prooftree
\prooftree
\prooftree
\prooftree
\prooftree
\prooftree
\prooftree
\prooftree
\prooftree
\prooftree
\justifies
\mbox{\fbox{$Nt(s(m))$}}\ \Rightarrow\ Nt(s(m))
\endprooftree
\justifies
\mbox{\fbox{${\blacksquare}Nt(s(m))$}}\ \Rightarrow\ Nt(s(m))
\using {\blacksquare}L
\endprooftree
\justifies
{\blacksquare}Nt(s(m))\ \Rightarrow\ \fbox{${\exists}bNb$}
\using {\exists}R
\endprooftree
\prooftree
\prooftree
\prooftree
\justifies
Nt(s(m))\ \Rightarrow\ Nt(s(m))
\endprooftree
\justifies
[Nt(s(m))]\ \Rightarrow\ \fbox{${\langle\rangle}Nt(s(m))$}
\using {\langle\rangle}R
\endprooftree
\prooftree
\justifies
\mbox{\fbox{$Sf$}}\ \Rightarrow\ Sf
\endprooftree
\justifies
[Nt(s(m))], \mbox{\fbox{${\langle\rangle}Nt(s(m))\backslash Sf$}}\ \Rightarrow\ Sf
\using {\backslash}L
\endprooftree
\justifies
[Nt(s(m))], \mbox{\fbox{$({\langle\rangle}Nt(s(m))\backslash Sf)/{\exists}bNb$}}, {\blacksquare}Nt(s(m))\ \Rightarrow\ Sf
\using {/}L
\endprooftree
\justifies
{\langle\rangle}Nt(s(m)), ({\langle\rangle}Nt(s(m))\backslash Sf)/{\exists}bNb, {\blacksquare}Nt(s(m))\ \Rightarrow\ Sf
\using {\langle\rangle}L
\endprooftree
\justifies
({\langle\rangle}Nt(s(m))\backslash Sf)/{\exists}bNb, {\blacksquare}Nt(s(m))\ \Rightarrow\ {\langle\rangle}Nt(s(m))\backslash Sf
\using {\backslash}R
\endprooftree
\prooftree
\prooftree
\prooftree
\justifies
Nt(s(m))\ \Rightarrow\ Nt(s(m))
\endprooftree
\justifies
[Nt(s(m))]\ \Rightarrow\ \fbox{${\langle\rangle}Nt(s(m))$}
\using {\langle\rangle}R
\endprooftree
\prooftree
\justifies
\mbox{\fbox{$Sf$}}\ \Rightarrow\ Sf
\endprooftree
\justifies
[Nt(s(m))], \mbox{\fbox{${\langle\rangle}Nt(s(m))\backslash Sf$}}\ \Rightarrow\ Sf
\using {\backslash}L
\endprooftree
\justifies
[Nt(s(m))], ({\langle\rangle}Nt(s(m))\backslash Sf)/{\exists}bNb, {\blacksquare}Nt(s(m)), \mbox{\fbox{$({\langle\rangle}Nt(s(m))\backslash Sf)\backslash ({\langle\rangle}Nt(s(m))\backslash Sf)$}}\ \Rightarrow\ Sf
\using {\backslash}L
\endprooftree
\justifies
[Nt(s(m))], ({\langle\rangle}Nt(s(m))\backslash Sf)/{\exists}bNb, {\blacksquare}Nt(s(m)), \mbox{\fbox{${\forall}f(({\langle\rangle}Nt(s(m))\backslash Sf)\backslash ({\langle\rangle}Nt(s(m))\backslash Sf))$}}\ \Rightarrow\ Sf
\using {\forall}L
\endprooftree
\justifies
[Nt(s(m))], ({\langle\rangle}Nt(s(m))\backslash Sf)/{\exists}bNb, {\blacksquare}Nt(s(m)), \mbox{\fbox{${\forall}a{\forall}f(({\langle\rangle}Na\backslash Sf)\backslash ({\langle\rangle}Na\backslash Sf))$}}\ \Rightarrow\ Sf
\using {\forall}L
\endprooftree
\justifies
[Nt(s(m))], ({\langle\rangle}Nt(s(m))\backslash Sf)/{\exists}bNb, {\blacksquare}Nt(s(m)), \mbox{\fbox{${\square}{\forall}a{\forall}f(({\langle\rangle}Na\backslash Sf)\backslash ({\langle\rangle}Na\backslash Sf))$}}\ \Rightarrow\ Sf
\using {\Box}L
\endprooftree
\justifies
{\langle\rangle}Nt(s(m)), ({\langle\rangle}Nt(s(m))\backslash Sf)/{\exists}bNb, {\blacksquare}Nt(s(m)), {\square}{\forall}a{\forall}f(({\langle\rangle}Na\backslash Sf)\backslash ({\langle\rangle}Na\backslash Sf))\ \Rightarrow\ Sf
\using {\langle\rangle}L
\endprooftree
\justifies
({\langle\rangle}Nt(s(m))\backslash Sf)/{\exists}bNb, {\blacksquare}Nt(s(m)), {\square}{\forall}a{\forall}f(({\langle\rangle}Na\backslash Sf)\backslash ({\langle\rangle}Na\backslash Sf))\ \Rightarrow\ {\langle\rangle}Nt(s(m))\backslash Sf
\using {\backslash}R
\endprooftree
\justifies
{\blacksquare}Nt(s(m)), {\square}{\forall}a{\forall}f(({\langle\rangle}Na\backslash Sf)\backslash ({\langle\rangle}Na\backslash Sf))\ \Rightarrow\ (({\langle\rangle}Nt(s(m))\backslash Sf)/{\exists}bNb)\backslash ({\langle\rangle}Nt(s(m))\backslash Sf)
\using {\backslash}R
\endprooftree
\justifies
\begin{array}{c}
{\blacksquare}Nt(s(m)), {\square}{\forall}a{\forall}f(({\langle\rangle}Na\backslash Sf)\backslash ({\langle\rangle}Na\backslash Sf))\ \Rightarrow\ {\blacksquare}((({\langle\rangle}Nt(s(m))\backslash Sf)/{\exists}bNb)\backslash ({\langle\rangle}Nt(s(m))\backslash Sf))\\
\mbox{\footnotesize\textcircled{1}}
\end{array}
\using {\blacksquare}R
\endprooftree

\rotatebox{-90}{
\resizebox{\textheight}{!}{
\prooftree
\prooftree
\prooftree
\prooftree
\mbox{\footnotesize\textcircled{1}}\tab
\prooftree
\prooftree
\vdots
\justifies
\begin{array}{c}
{\blacksquare}Nt(s(f)), {\square}{\forall}a{\forall}f(({\langle\rangle}Na\backslash Sf)\backslash ({\langle\rangle}Na\backslash Sf))\ \Rightarrow\ \fbox{$?{\blacksquare}((({\langle\rangle}Nt(s(m))\backslash Sf)/{\exists}bNb)\backslash ({\langle\rangle}Nt(s(m))\backslash Sf))$}\\
\end{array}
\using {?}R
\endprooftree
\prooftree
\prooftree
\prooftree
\prooftree
\prooftree
\prooftree
\prooftree
\prooftree
\prooftree
\prooftree
\prooftree
\prooftree
\justifies
N1\ \Rightarrow\ N1
\endprooftree
\justifies
N1\ \Rightarrow\ \fbox{${\exists}aNa$}
\using {\exists}R
\endprooftree
\justifies
N1\ \Rightarrow\ \fbox{${\exists}aNa{\oplus}{\it CP}that$}
\using {\oplus}R
\endprooftree
\prooftree
\prooftree
\prooftree
\prooftree
\justifies
Nt(s(m))\ \Rightarrow\ Nt(s(m))
\endprooftree
\justifies
Nt(s(m))\ \Rightarrow\ \fbox{${\exists}aNa$}
\using {\exists}R
\endprooftree
\justifies
[Nt(s(m))]\ \Rightarrow\ \fbox{${\langle\rangle}{\exists}aNa$}
\using {\langle\rangle}R
\endprooftree
\prooftree
\justifies
\mbox{\fbox{$Sf$}}\ \Rightarrow\ Sf
\endprooftree
\justifies
[Nt(s(m))], \mbox{\fbox{${\langle\rangle}{\exists}aNa\backslash Sf$}}\ \Rightarrow\ Sf
\using {\backslash}L
\endprooftree
\justifies
[Nt(s(m))], \mbox{\fbox{$({\langle\rangle}{\exists}aNa\backslash Sf)/({\exists}aNa{\oplus}{\it CP}that)$}}, N1\ \Rightarrow\ Sf
\using {/}L
\endprooftree
\justifies
[Nt(s(m))], \mbox{\fbox{${\square}(({\langle\rangle}{\exists}aNa\backslash Sf)/({\exists}aNa{\oplus}{\it CP}that))$}}, N1\ \Rightarrow\ Sf
\using {\Box}L
\endprooftree
\justifies
[Nt(s(m))], {\square}(({\langle\rangle}{\exists}aNa\backslash Sf)/({\exists}aNa{\oplus}{\it CP}that)), {\exists}bNb\ \Rightarrow\ Sf
\using {\exists}L
\endprooftree
\justifies
{\langle\rangle}Nt(s(m)), {\square}(({\langle\rangle}{\exists}aNa\backslash Sf)/({\exists}aNa{\oplus}{\it CP}that)), {\exists}bNb\ \Rightarrow\ Sf
\using {\langle\rangle}L
\endprooftree
\justifies
{\square}(({\langle\rangle}{\exists}aNa\backslash Sf)/({\exists}aNa{\oplus}{\it CP}that)), {\exists}bNb\ \Rightarrow\ {\langle\rangle}Nt(s(m))\backslash Sf
\using {\backslash}R
\endprooftree
\justifies
{\square}(({\langle\rangle}{\exists}aNa\backslash Sf)/({\exists}aNa{\oplus}{\it CP}that))\ \Rightarrow\ ({\langle\rangle}Nt(s(m))\backslash Sf)/{\exists}bNb
\using {/}R
\endprooftree
\prooftree
\prooftree
\prooftree
\prooftree
\justifies
\mbox{\fbox{$Nt(s(m))$}}\ \Rightarrow\ Nt(s(m))
\endprooftree
\justifies
\mbox{\fbox{${\blacksquare}Nt(s(m))$}}\ \Rightarrow\ Nt(s(m))
\using {\blacksquare}L
\endprooftree
\justifies
[{\blacksquare}Nt(s(m))]\ \Rightarrow\ \fbox{${\langle\rangle}Nt(s(m))$}
\using {\langle\rangle}R
\endprooftree
\prooftree
\justifies
\mbox{\fbox{$Sf$}}\ \Rightarrow\ Sf
\endprooftree
\justifies
[{\blacksquare}Nt(s(m))], \mbox{\fbox{${\langle\rangle}Nt(s(m))\backslash Sf$}}\ \Rightarrow\ Sf
\using {\backslash}L
\endprooftree
\justifies
[{\blacksquare}Nt(s(m))], {\square}(({\langle\rangle}{\exists}aNa\backslash Sf)/({\exists}aNa{\oplus}{\it CP}that)), \mbox{\fbox{$(({\langle\rangle}Nt(s(m))\backslash Sf)/{\exists}bNb)\backslash ({\langle\rangle}Nt(s(m))\backslash Sf)$}}\ \Rightarrow\ Sf
\using {\backslash}L
\endprooftree
\justifies
[{\blacksquare}Nt(s(m))], {\square}(({\langle\rangle}{\exists}aNa\backslash Sf)/({\exists}aNa{\oplus}{\it CP}that)), [\mbox{\fbox{${[]^{-1}}((({\langle\rangle}Nt(s(m))\backslash Sf)/{\exists}bNb)\backslash ({\langle\rangle}Nt(s(m))\backslash Sf))$}}]\ \Rightarrow\ Sf
\using {[]^{-1}}L
\endprooftree
\justifies
[{\blacksquare}Nt(s(m))], {\square}(({\langle\rangle}{\exists}aNa\backslash Sf)/({\exists}aNa{\oplus}{\it CP}that)), [[\mbox{\fbox{${[]^{-1}}{[]^{-1}}((({\langle\rangle}Nt(s(m))\backslash Sf)/{\exists}bNb)\backslash ({\langle\rangle}Nt(s(m))\backslash Sf))$}}]]\ \Rightarrow\ Sf
\using {[]^{-1}}L
\endprooftree
\justifies
[{\blacksquare}Nt(s(m))], {\square}(({\langle\rangle}{\exists}aNa\backslash Sf)/({\exists}aNa{\oplus}{\it CP}that)), [[{\blacksquare}Nt(s(f)), {\square}{\forall}a{\forall}f(({\langle\rangle}Na\backslash Sf)\backslash ({\langle\rangle}Na\backslash Sf)), \mbox{\fbox{$?{\blacksquare}((({\langle\rangle}Nt(s(m))\backslash Sf)/{\exists}bNb)\backslash ({\langle\rangle}Nt(s(m))\backslash Sf))\backslash {[]^{-1}}{[]^{-1}}((({\langle\rangle}Nt(s(m))\backslash Sf)/{\exists}bNb)\backslash ({\langle\rangle}Nt(s(m))\backslash Sf))$}}]]\ \Rightarrow\ Sf
\using {\backslash}L
\endprooftree
\justifies
[{\blacksquare}Nt(s(m))], {\square}(({\langle\rangle}{\exists}aNa\backslash Sf)/({\exists}aNa{\oplus}{\it CP}that)), [[{\blacksquare}Nt(s(f)), {\square}{\forall}a{\forall}f(({\langle\rangle}Na\backslash Sf)\backslash ({\langle\rangle}Na\backslash Sf)), \mbox{\fbox{$(?{\blacksquare}((({\langle\rangle}Nt(s(m))\backslash Sf)/{\exists}bNb)\backslash ({\langle\rangle}Nt(s(m))\backslash Sf))\backslash {[]^{-1}}{[]^{-1}}((({\langle\rangle}Nt(s(m))\backslash Sf)/{\exists}bNb)\backslash ({\langle\rangle}Nt(s(m))\backslash Sf)))/{\blacksquare}((({\langle\rangle}Nt(s(m))\backslash Sf)/{\exists}bNb)\backslash ({\langle\rangle}Nt(s(m))\backslash Sf))$}}, {\blacksquare}Nt(s(m)), {\square}{\forall}a{\forall}f(({\langle\rangle}Na\backslash Sf)\backslash ({\langle\rangle}Na\backslash Sf))]]\ \Rightarrow\ Sf
\using {/}L
\endprooftree
\justifies
[{\blacksquare}Nt(s(m))], {\square}(({\langle\rangle}{\exists}aNa\backslash Sf)/({\exists}aNa{\oplus}{\it CP}that)), [[{\blacksquare}Nt(s(f)), {\square}{\forall}a{\forall}f(({\langle\rangle}Na\backslash Sf)\backslash ({\langle\rangle}Na\backslash Sf)), \mbox{\fbox{${\forall}a((?{\blacksquare}((({\langle\rangle}Na\backslash Sf)/{\exists}bNb)\backslash ({\langle\rangle}Na\backslash Sf))\backslash {[]^{-1}}{[]^{-1}}((({\langle\rangle}Na\backslash Sf)/{\exists}bNb)\backslash ({\langle\rangle}Na\backslash Sf)))/{\blacksquare}((({\langle\rangle}Na\backslash Sf)/{\exists}bNb)\backslash ({\langle\rangle}Na\backslash Sf)))$}}, {\blacksquare}Nt(s(m)), {\square}{\forall}a{\forall}f(({\langle\rangle}Na\backslash Sf)\backslash ({\langle\rangle}Na\backslash Sf))]]\ \Rightarrow\ Sf
\using {\forall}L
\endprooftree
\justifies
[{\blacksquare}Nt(s(m))], {\square}(({\langle\rangle}{\exists}aNa\backslash Sf)/({\exists}aNa{\oplus}{\it CP}that)), [[{\blacksquare}Nt(s(f)), {\square}{\forall}a{\forall}f(({\langle\rangle}Na\backslash Sf)\backslash ({\langle\rangle}Na\backslash Sf)), \mbox{\fbox{${\forall}f{\forall}a((?{\blacksquare}((({\langle\rangle}Na\backslash Sf)/{\exists}bNb)\backslash ({\langle\rangle}Na\backslash Sf))\backslash {[]^{-1}}{[]^{-1}}((({\langle\rangle}Na\backslash Sf)/{\exists}bNb)\backslash ({\langle\rangle}Na\backslash Sf)))/{\blacksquare}((({\langle\rangle}Na\backslash Sf)/{\exists}bNb)\backslash ({\langle\rangle}Na\backslash Sf)))$}}, {\blacksquare}Nt(s(m)), {\square}{\forall}a{\forall}f(({\langle\rangle}Na\backslash Sf)\backslash ({\langle\rangle}Na\backslash Sf))]]\ \Rightarrow\ Sf
\using {\forall}L
\endprooftree
\justifies
[{\blacksquare}Nt(s(m))], {\square}(({\langle\rangle}{\exists}aNa\backslash Sf)/({\exists}aNa{\oplus}{\it CP}that)), [[{\blacksquare}Nt(s(f)), {\square}{\forall}a{\forall}f(({\langle\rangle}Na\backslash Sf)\backslash ({\langle\rangle}Na\backslash Sf)), \mbox{\fbox{${\blacksquare}{\forall}f{\forall}a((?{\blacksquare}((({\langle\rangle}Na\backslash Sf)/{\exists}bNb)\backslash ({\langle\rangle}Na\backslash Sf))\backslash {[]^{-1}}{[]^{-1}}((({\langle\rangle}Na\backslash Sf)/{\exists}bNb)\backslash ({\langle\rangle}Na\backslash Sf)))/{\blacksquare}((({\langle\rangle}Na\backslash Sf)/{\exists}bNb)\backslash ({\langle\rangle}Na\backslash Sf)))$}}, {\blacksquare}Nt(s(m)), {\square}{\forall}a{\forall}f(({\langle\rangle}Na\backslash Sf)\backslash ({\langle\rangle}Na\backslash Sf))]]\ \Rightarrow\ Sf
\using {\blacksquare}L
\endprooftree}}
\end{center}}

\vspace{0.15in}

\noindent
This delivers the semantics:
\disp{
$[(\mbox{\v{}}{\it today}\ ({\it Past}\ ((\mbox{\v{}}{\it seee}\ {\it m})\ {\it j})))\wedge (\mbox{\v{}}{\it yesterday}\ ({\it Past}\ ((\mbox{\v{}}{\it seee}\ {\it b})\ {\it j})))]$}

\commentout{

Again, another result of lexical consultation is:
\disp{
$[{\blacksquare}Nt(s(m)): {\it j}], {\square}(({\langle\rangle}{\exists}aNa\backslash Sf)/({\exists}aNa{\oplus}{\it CP}that)): \mbox{\^{}}\lambda A\lambda B({\it Past}\ (({\it A}\casearrow C.(\mbox{\v{}}{\it seee}\ {\it C}); D.(\mbox{\v{}}{\it seet}\ {\it D}))\ {\it B})), [[{\blacksquare}Nt(s(f)): {\it m}, {\square}{\forall}a{\forall}f(({\langle\rangle}Na\backslash Sf)\backslash ({\langle\rangle}Na\backslash Sf)): \mbox{\^{}}\lambda E\lambda F(\mbox{\v{}}{\it today}\ ({\it E}\ {\it F})), {\blacksquare}{\forall}a{\forall}b{\forall}f((?{\blacksquare}((({\langle\rangle}Na\backslash Sf)/({\exists}cNc{\oplus}{\it CP}b))\backslash ({\langle\rangle}Na\backslash Sf))\backslash\\
 {[]^{-1}}{[]^{-1}}((({\langle\rangle}Na\backslash Sf)/({\exists}cNc{\oplus}{\it CP}b))\backslash ({\langle\rangle}Na\backslash Sf)))/{\blacksquare}((({\langle\rangle}Na\backslash Sf)/({\exists}cNc{\oplus}{\it CP}b))\backslash ({\langle\rangle}Na\backslash Sf))): (\Phinplus\ ({\it s}\ ({\it s}\ {\it 0}))\ {\it and}), \\{\blacksquare}Nt(s(m)): {\it b}, {\square}{\forall}a{\forall}f(({\langle\rangle}Na\backslash Sf)\backslash ({\langle\rangle}Na\backslash Sf)): \mbox{\^{}}\lambda G\lambda H(\mbox{\v{}}{\it yesterday}\ ({\it G}\ {\it H}))]]\ \Rightarrow\ Sf$}
Here the coordination is over the polymorphic verb type.
This has the derivation:

\vspace{0.15in}

{\tiny

\prooftree
\prooftree
\prooftree
\prooftree
\prooftree
\prooftree
\prooftree
\prooftree
\prooftree
\prooftree
\prooftree
\prooftree
\prooftree
\prooftree
\prooftree
\justifies
\mbox{\fbox{$Nt(s(m))$}}\ \Rightarrow\ Nt(s(m))
\endprooftree
\justifies
\mbox{\fbox{${\blacksquare}Nt(s(m))$}}\ \Rightarrow\ Nt(s(m))
\using {\blacksquare}L
\endprooftree
\justifies
{\blacksquare}Nt(s(m))\ \Rightarrow\ \fbox{${\exists}cNc$}
\using {\exists}R
\endprooftree
\justifies
{\blacksquare}Nt(s(m))\ \Rightarrow\ \fbox{${\exists}cNc{\oplus}{\it CP}that$}
\using {\oplus}R
\endprooftree
\prooftree
\prooftree
\prooftree
\justifies
Nt(s(m))\ \Rightarrow\ Nt(s(m))
\endprooftree
\justifies
[Nt(s(m))]\ \Rightarrow\ \fbox{${\langle\rangle}Nt(s(m))$}
\using {\langle\rangle}R
\endprooftree
\prooftree
\justifies
\mbox{\fbox{$Sf$}}\ \Rightarrow\ Sf
\endprooftree
\justifies
[Nt(s(m))], \mbox{\fbox{${\langle\rangle}Nt(s(m))\backslash Sf$}}\ \Rightarrow\ Sf
\using {\backslash}L
\endprooftree
\justifies
[Nt(s(m))], \mbox{\fbox{$({\langle\rangle}Nt(s(m))\backslash Sf)/({\exists}cNc{\oplus}{\it CP}that)$}}, {\blacksquare}Nt(s(m))\ \Rightarrow\ Sf
\using {/}L
\endprooftree
\justifies
{\langle\rangle}Nt(s(m)), ({\langle\rangle}Nt(s(m))\backslash Sf)/({\exists}cNc{\oplus}{\it CP}that), {\blacksquare}Nt(s(m))\ \Rightarrow\ Sf
\using {\langle\rangle}L
\endprooftree
\justifies
({\langle\rangle}Nt(s(m))\backslash Sf)/({\exists}cNc{\oplus}{\it CP}that), {\blacksquare}Nt(s(m))\ \Rightarrow\ {\langle\rangle}Nt(s(m))\backslash Sf
\using {\backslash}R
\endprooftree
\prooftree
\prooftree
\prooftree
\justifies
Nt(s(m))\ \Rightarrow\ Nt(s(m))
\endprooftree
\justifies
[Nt(s(m))]\ \Rightarrow\ \fbox{${\langle\rangle}Nt(s(m))$}
\using {\langle\rangle}R
\endprooftree
\prooftree
\justifies
\mbox{\fbox{$Sf$}}\ \Rightarrow\ Sf
\endprooftree
\justifies
[Nt(s(m))], \mbox{\fbox{${\langle\rangle}Nt(s(m))\backslash Sf$}}\ \Rightarrow\ Sf
\using {\backslash}L
\endprooftree
\justifies
[Nt(s(m))], ({\langle\rangle}Nt(s(m))\backslash Sf)/({\exists}cNc{\oplus}{\it CP}that), {\blacksquare}Nt(s(m)), \mbox{\fbox{$({\langle\rangle}Nt(s(m))\backslash Sf)\backslash ({\langle\rangle}Nt(s(m))\backslash Sf)$}}\ \Rightarrow\ Sf
\using {\backslash}L
\endprooftree
\justifies
[Nt(s(m))], ({\langle\rangle}Nt(s(m))\backslash Sf)/({\exists}cNc{\oplus}{\it CP}that), {\blacksquare}Nt(s(m)), \mbox{\fbox{${\forall}f(({\langle\rangle}Nt(s(m))\backslash Sf)\backslash ({\langle\rangle}Nt(s(m))\backslash Sf))$}}\ \Rightarrow\ Sf
\using {\forall}L
\endprooftree
\justifies
[Nt(s(m))], ({\langle\rangle}Nt(s(m))\backslash Sf)/({\exists}cNc{\oplus}{\it CP}that), {\blacksquare}Nt(s(m)), \mbox{\fbox{${\forall}a{\forall}f(({\langle\rangle}Na\backslash Sf)\backslash ({\langle\rangle}Na\backslash Sf))$}}\ \Rightarrow\ Sf
\using {\forall}L
\endprooftree
\justifies
[Nt(s(m))], ({\langle\rangle}Nt(s(m))\backslash Sf)/({\exists}cNc{\oplus}{\it CP}that), {\blacksquare}Nt(s(m)), \mbox{\fbox{${\square}{\forall}a{\forall}f(({\langle\rangle}Na\backslash Sf)\backslash ({\langle\rangle}Na\backslash Sf))$}}\ \Rightarrow\ Sf
\using {\Box}L
\endprooftree
\justifies
{\langle\rangle}Nt(s(m)), ({\langle\rangle}Nt(s(m))\backslash Sf)/({\exists}cNc{\oplus}{\it CP}that), {\blacksquare}Nt(s(m)), {\square}{\forall}a{\forall}f(({\langle\rangle}Na\backslash Sf)\backslash ({\langle\rangle}Na\backslash Sf))\ \Rightarrow\ Sf
\using {\langle\rangle}L
\endprooftree
\justifies
({\langle\rangle}Nt(s(m))\backslash Sf)/({\exists}cNc{\oplus}{\it CP}that), {\blacksquare}Nt(s(m)), {\square}{\forall}a{\forall}f(({\langle\rangle}Na\backslash Sf)\backslash ({\langle\rangle}Na\backslash Sf))\ \Rightarrow\ {\langle\rangle}Nt(s(m))\backslash Sf
\using {\backslash}R
\endprooftree
\justifies
{\blacksquare}Nt(s(m)), {\square}{\forall}a{\forall}f(({\langle\rangle}Na\backslash Sf)\backslash ({\langle\rangle}Na\backslash Sf))\ \Rightarrow\ (({\langle\rangle}Nt(s(m))\backslash Sf)/({\exists}cNc{\oplus}{\it CP}that))\backslash ({\langle\rangle}Nt(s(m))\backslash Sf)
\using {\backslash}R
\endprooftree
\justifies
\begin{array}{c}
{\blacksquare}Nt(s(m)), {\square}{\forall}a{\forall}f(({\langle\rangle}Na\backslash Sf)\backslash ({\langle\rangle}Na\backslash Sf))\ \Rightarrow\ {\blacksquare}((({\langle\rangle}Nt(s(m))\backslash Sf)/({\exists}cNc{\oplus}{\it CP}that))\backslash ({\langle\rangle}Nt(s(m))\backslash Sf))\\
\mbox{\footnotesize\textcircled{1}}
\end{array}
\using {\blacksquare}R
\endprooftree

\prooftree
\prooftree
\prooftree
\prooftree
\prooftree
\prooftree
\prooftree
\prooftree
\prooftree
\prooftree
\prooftree
\prooftree
\prooftree
\prooftree
\prooftree
\prooftree
\justifies
\mbox{\fbox{$Nt(s(f))$}}\ \Rightarrow\ Nt(s(f))
\endprooftree
\justifies
\mbox{\fbox{${\blacksquare}Nt(s(f))$}}\ \Rightarrow\ Nt(s(f))
\using {\blacksquare}L
\endprooftree
\justifies
{\blacksquare}Nt(s(f))\ \Rightarrow\ \fbox{${\exists}cNc$}
\using {\exists}R
\endprooftree
\justifies
{\blacksquare}Nt(s(f))\ \Rightarrow\ \fbox{${\exists}cNc{\oplus}{\it CP}that$}
\using {\oplus}R
\endprooftree
\prooftree
\prooftree
\prooftree
\justifies
Nt(s(m))\ \Rightarrow\ Nt(s(m))
\endprooftree
\justifies
[Nt(s(m))]\ \Rightarrow\ \fbox{${\langle\rangle}Nt(s(m))$}
\using {\langle\rangle}R
\endprooftree
\prooftree
\justifies
\mbox{\fbox{$Sf$}}\ \Rightarrow\ Sf
\endprooftree
\justifies
[Nt(s(m))], \mbox{\fbox{${\langle\rangle}Nt(s(m))\backslash Sf$}}\ \Rightarrow\ Sf
\using {\backslash}L
\endprooftree
\justifies
[Nt(s(m))], \mbox{\fbox{$({\langle\rangle}Nt(s(m))\backslash Sf)/({\exists}cNc{\oplus}{\it CP}that)$}}, {\blacksquare}Nt(s(f))\ \Rightarrow\ Sf
\using {/}L
\endprooftree
\justifies
{\langle\rangle}Nt(s(m)), ({\langle\rangle}Nt(s(m))\backslash Sf)/({\exists}cNc{\oplus}{\it CP}that), {\blacksquare}Nt(s(f))\ \Rightarrow\ Sf
\using {\langle\rangle}L
\endprooftree
\justifies
({\langle\rangle}Nt(s(m))\backslash Sf)/({\exists}cNc{\oplus}{\it CP}that), {\blacksquare}Nt(s(f))\ \Rightarrow\ {\langle\rangle}Nt(s(m))\backslash Sf
\using {\backslash}R
\endprooftree
\prooftree
\prooftree
\prooftree
\justifies
Nt(s(m))\ \Rightarrow\ Nt(s(m))
\endprooftree
\justifies
[Nt(s(m))]\ \Rightarrow\ \fbox{${\langle\rangle}Nt(s(m))$}
\using {\langle\rangle}R
\endprooftree
\prooftree
\justifies
\mbox{\fbox{$Sf$}}\ \Rightarrow\ Sf
\endprooftree
\justifies
[Nt(s(m))], \mbox{\fbox{${\langle\rangle}Nt(s(m))\backslash Sf$}}\ \Rightarrow\ Sf
\using {\backslash}L
\endprooftree
\justifies
[Nt(s(m))], ({\langle\rangle}Nt(s(m))\backslash Sf)/({\exists}cNc{\oplus}{\it CP}that), {\blacksquare}Nt(s(f)), \mbox{\fbox{$({\langle\rangle}Nt(s(m))\backslash Sf)\backslash ({\langle\rangle}Nt(s(m))\backslash Sf)$}}\ \Rightarrow\ Sf
\using {\backslash}L
\endprooftree
\justifies
[Nt(s(m))], ({\langle\rangle}Nt(s(m))\backslash Sf)/({\exists}cNc{\oplus}{\it CP}that), {\blacksquare}Nt(s(f)), \mbox{\fbox{${\forall}f(({\langle\rangle}Nt(s(m))\backslash Sf)\backslash ({\langle\rangle}Nt(s(m))\backslash Sf))$}}\ \Rightarrow\ Sf
\using {\forall}L
\endprooftree
\justifies
[Nt(s(m))], ({\langle\rangle}Nt(s(m))\backslash Sf)/({\exists}cNc{\oplus}{\it CP}that), {\blacksquare}Nt(s(f)), \mbox{\fbox{${\forall}a{\forall}f(({\langle\rangle}Na\backslash Sf)\backslash ({\langle\rangle}Na\backslash Sf))$}}\ \Rightarrow\ Sf
\using {\forall}L
\endprooftree
\justifies
[Nt(s(m))], ({\langle\rangle}Nt(s(m))\backslash Sf)/({\exists}cNc{\oplus}{\it CP}that), {\blacksquare}Nt(s(f)), \mbox{\fbox{${\square}{\forall}a{\forall}f(({\langle\rangle}Na\backslash Sf)\backslash ({\langle\rangle}Na\backslash Sf))$}}\ \Rightarrow\ Sf
\using {\Box}L
\endprooftree
\justifies
{\langle\rangle}Nt(s(m)), ({\langle\rangle}Nt(s(m))\backslash Sf)/({\exists}cNc{\oplus}{\it CP}that), {\blacksquare}Nt(s(f)), {\square}{\forall}a{\forall}f(({\langle\rangle}Na\backslash Sf)\backslash ({\langle\rangle}Na\backslash Sf))\ \Rightarrow\ Sf
\using {\langle\rangle}L
\endprooftree
\justifies
({\langle\rangle}Nt(s(m))\backslash Sf)/({\exists}cNc{\oplus}{\it CP}that), {\blacksquare}Nt(s(f)), {\square}{\forall}a{\forall}f(({\langle\rangle}Na\backslash Sf)\backslash ({\langle\rangle}Na\backslash Sf))\ \Rightarrow\ {\langle\rangle}Nt(s(m))\backslash Sf
\using {\backslash}R
\endprooftree
\justifies
{\blacksquare}Nt(s(f)), {\square}{\forall}a{\forall}f(({\langle\rangle}Na\backslash Sf)\backslash ({\langle\rangle}Na\backslash Sf))\ \Rightarrow\ (({\langle\rangle}Nt(s(m))\backslash Sf)/({\exists}cNc{\oplus}{\it CP}that))\backslash ({\langle\rangle}Nt(s(m))\backslash Sf)
\using {\backslash}R
\endprooftree
\justifies
{\blacksquare}Nt(s(f)), {\square}{\forall}a{\forall}f(({\langle\rangle}Na\backslash Sf)\backslash ({\langle\rangle}Na\backslash Sf))\ \Rightarrow\ {\blacksquare}((({\langle\rangle}Nt(s(m))\backslash Sf)/({\exists}cNc{\oplus}{\it CP}that))\backslash ({\langle\rangle}Nt(s(m))\backslash Sf))
\using {\blacksquare}R
\endprooftree
\justifies
\begin{array}{c}
{\blacksquare}Nt(s(f)), {\square}{\forall}a{\forall}f(({\langle\rangle}Na\backslash Sf)\backslash ({\langle\rangle}Na\backslash Sf))\ \Rightarrow\ \fbox{$?{\blacksquare}((({\langle\rangle}Nt(s(m))\backslash Sf)/({\exists}cNc{\oplus}{\it CP}that))\backslash ({\langle\rangle}Nt(s(m))\backslash Sf))$}\\
\mbox{\footnotesize\textcircled{2}}
\end{array}
\using {?}R
\endprooftree

\rotatebox{-90}{
\prooftree
\prooftree
\prooftree
\prooftree
\prooftree
\mbox{\footnotesize\textcircled{1}}\tab
\prooftree
\mbox{\footnotesize\textcircled{2}}\tab
\prooftree
\prooftree
\prooftree
\prooftree
\prooftree
\prooftree
\prooftree
\prooftree
\prooftree
\prooftree
\prooftree
\prooftree
\prooftree
\justifies
N6\ \Rightarrow\ N6
\endprooftree
\justifies
N6\ \Rightarrow\ \fbox{${\exists}aNa$}
\using {\exists}R
\endprooftree
\justifies
N6\ \Rightarrow\ \fbox{${\exists}aNa{\oplus}{\it CP}that$}
\using {\oplus}R
\endprooftree
\prooftree
\prooftree
\prooftree
\prooftree
\justifies
Nt(s(m))\ \Rightarrow\ Nt(s(m))
\endprooftree
\justifies
Nt(s(m))\ \Rightarrow\ \fbox{${\exists}aNa$}
\using {\exists}R
\endprooftree
\justifies
[Nt(s(m))]\ \Rightarrow\ \fbox{${\langle\rangle}{\exists}aNa$}
\using {\langle\rangle}R
\endprooftree
\prooftree
\justifies
\mbox{\fbox{$Sf$}}\ \Rightarrow\ Sf
\endprooftree
\justifies
[Nt(s(m))], \mbox{\fbox{${\langle\rangle}{\exists}aNa\backslash Sf$}}\ \Rightarrow\ Sf
\using {\backslash}L
\endprooftree
\justifies
[Nt(s(m))], \mbox{\fbox{$({\langle\rangle}{\exists}aNa\backslash Sf)/({\exists}aNa{\oplus}{\it CP}that)$}}, N6\ \Rightarrow\ Sf
\using {/}L
\endprooftree
\justifies
[Nt(s(m))], \mbox{\fbox{${\square}(({\langle\rangle}{\exists}aNa\backslash Sf)/({\exists}aNa{\oplus}{\it CP}that))$}}, N6\ \Rightarrow\ Sf
\using {\Box}L
\endprooftree
\justifies
[Nt(s(m))], {\square}(({\langle\rangle}{\exists}aNa\backslash Sf)/({\exists}aNa{\oplus}{\it CP}that)), {\exists}cNc\ \Rightarrow\ Sf
\using {\exists}L
\endprooftree
\justifies
{\langle\rangle}Nt(s(m)), {\square}(({\langle\rangle}{\exists}aNa\backslash Sf)/({\exists}aNa{\oplus}{\it CP}that)), {\exists}cNc\ \Rightarrow\ Sf
\using {\langle\rangle}L
\endprooftree
\prooftree
\prooftree
\prooftree
\prooftree
\prooftree
\justifies
{\it CP}that\ \Rightarrow\ {\it CP}that
\endprooftree
\justifies
{\it CP}that\ \Rightarrow\ \fbox{${\exists}aNa{\oplus}{\it CP}that$}
\using {\oplus}R
\endprooftree
\prooftree
\prooftree
\prooftree
\prooftree
\justifies
Nt(s(m))\ \Rightarrow\ Nt(s(m))
\endprooftree
\justifies
Nt(s(m))\ \Rightarrow\ \fbox{${\exists}aNa$}
\using {\exists}R
\endprooftree
\justifies
[Nt(s(m))]\ \Rightarrow\ \fbox{${\langle\rangle}{\exists}aNa$}
\using {\langle\rangle}R
\endprooftree
\prooftree
\justifies
\mbox{\fbox{$Sf$}}\ \Rightarrow\ Sf
\endprooftree
\justifies
[Nt(s(m))], \mbox{\fbox{${\langle\rangle}{\exists}aNa\backslash Sf$}}\ \Rightarrow\ Sf
\using {\backslash}L
\endprooftree
\justifies
[Nt(s(m))], \mbox{\fbox{$({\langle\rangle}{\exists}aNa\backslash Sf)/({\exists}aNa{\oplus}{\it CP}that)$}}, {\it CP}that\ \Rightarrow\ Sf
\using {/}L
\endprooftree
\justifies
[Nt(s(m))], \mbox{\fbox{${\square}(({\langle\rangle}{\exists}aNa\backslash Sf)/({\exists}aNa{\oplus}{\it CP}that))$}}, {\it CP}that\ \Rightarrow\ Sf
\using {\Box}L
\endprooftree
\justifies
{\langle\rangle}Nt(s(m)), {\square}(({\langle\rangle}{\exists}aNa\backslash Sf)/({\exists}aNa{\oplus}{\it CP}that)), {\it CP}that\ \Rightarrow\ Sf
\using {\langle\rangle}L
\endprooftree
\justifies
{\langle\rangle}Nt(s(m)), {\square}(({\langle\rangle}{\exists}aNa\backslash Sf)/({\exists}aNa{\oplus}{\it CP}that)), {\exists}cNc{\oplus}{\it CP}that\ \Rightarrow\ Sf
\using {\oplus}L
\endprooftree
\justifies
{\square}(({\langle\rangle}{\exists}aNa\backslash Sf)/({\exists}aNa{\oplus}{\it CP}that)), {\exists}cNc{\oplus}{\it CP}that\ \Rightarrow\ {\langle\rangle}Nt(s(m))\backslash Sf
\using {\backslash}R
\endprooftree
\justifies
{\square}(({\langle\rangle}{\exists}aNa\backslash Sf)/({\exists}aNa{\oplus}{\it CP}that))\ \Rightarrow\ ({\langle\rangle}Nt(s(m))\backslash Sf)/({\exists}cNc{\oplus}{\it CP}that)
\using {/}R
\endprooftree
\prooftree
\prooftree
\prooftree
\prooftree
\justifies
\mbox{\fbox{$Nt(s(m))$}}\ \Rightarrow\ Nt(s(m))
\endprooftree
\justifies
\mbox{\fbox{${\blacksquare}Nt(s(m))$}}\ \Rightarrow\ Nt(s(m))
\using {\blacksquare}L
\endprooftree
\justifies
[{\blacksquare}Nt(s(m))]\ \Rightarrow\ \fbox{${\langle\rangle}Nt(s(m))$}
\using {\langle\rangle}R
\endprooftree
\prooftree
\justifies
\mbox{\fbox{$Sf$}}\ \Rightarrow\ Sf
\endprooftree
\justifies
[{\blacksquare}Nt(s(m))], \mbox{\fbox{${\langle\rangle}Nt(s(m))\backslash Sf$}}\ \Rightarrow\ Sf
\using {\backslash}L
\endprooftree
\justifies
[{\blacksquare}Nt(s(m))], {\square}(({\langle\rangle}{\exists}aNa\backslash Sf)/({\exists}aNa{\oplus}{\it CP}that)), \mbox{\fbox{$(({\langle\rangle}Nt(s(m))\backslash Sf)/({\exists}cNc{\oplus}{\it CP}that))\backslash ({\langle\rangle}Nt(s(m))\backslash Sf)$}}\ \Rightarrow\ Sf
\using {\backslash}L
\endprooftree
\justifies
[{\blacksquare}Nt(s(m))], {\square}(({\langle\rangle}{\exists}aNa\backslash Sf)/({\exists}aNa{\oplus}{\it CP}that)), [\mbox{\fbox{${[]^{-1}}((({\langle\rangle}Nt(s(m))\backslash Sf)/({\exists}cNc{\oplus}{\it CP}that))\backslash ({\langle\rangle}Nt(s(m))\backslash Sf))$}}]\ \Rightarrow\ Sf
\using {[]^{-1}}L
\endprooftree
\justifies
[{\blacksquare}Nt(s(m))], {\square}(({\langle\rangle}{\exists}aNa\backslash Sf)/({\exists}aNa{\oplus}{\it CP}that)), [[\mbox{\fbox{${[]^{-1}}{[]^{-1}}((({\langle\rangle}Nt(s(m))\backslash Sf)/({\exists}cNc{\oplus}{\it CP}that))\backslash ({\langle\rangle}Nt(s(m))\backslash Sf))$}}]]\ \Rightarrow\ Sf
\using {[]^{-1}}L
\endprooftree
\justifies
[{\blacksquare}Nt(s(m))], {\square}(({\langle\rangle}{\exists}aNa\backslash Sf)/({\exists}aNa{\oplus}{\it CP}that)), [[{\blacksquare}Nt(s(f)), {\square}{\forall}a{\forall}f(({\langle\rangle}Na\backslash Sf)\backslash ({\langle\rangle}Na\backslash Sf)), \mbox{\fbox{$?{\blacksquare}((({\langle\rangle}Nt(s(m))\backslash Sf)/({\exists}cNc{\oplus}{\it CP}that))\backslash ({\langle\rangle}Nt(s(m))\backslash Sf))\backslash {[]^{-1}}{[]^{-1}}((({\langle\rangle}Nt(s(m))\backslash Sf)/({\exists}cNc{\oplus}{\it CP}that))\backslash ({\langle\rangle}Nt(s(m))\backslash Sf))$}}]]\ \Rightarrow\ Sf
\using {\backslash}L
\endprooftree
\justifies
[{\blacksquare}Nt(s(m))], {\square}(({\langle\rangle}{\exists}aNa\backslash Sf)/({\exists}aNa{\oplus}{\it CP}that)), [[{\blacksquare}Nt(s(f)), {\square}{\forall}a{\forall}f(({\langle\rangle}Na\backslash Sf)\backslash ({\langle\rangle}Na\backslash Sf)), \mbox{\fbox{$(?{\blacksquare}((({\langle\rangle}Nt(s(m))\backslash Sf)/({\exists}cNc{\oplus}{\it CP}that))\backslash ({\langle\rangle}Nt(s(m))\backslash Sf))\backslash {[]^{-1}}{[]^{-1}}((({\langle\rangle}Nt(s(m))\backslash Sf)/({\exists}cNc{\oplus}{\it CP}that))\backslash ({\langle\rangle}Nt(s(m))\backslash Sf)))/{\blacksquare}((({\langle\rangle}Nt(s(m))\backslash Sf)/({\exists}cNc{\oplus}{\it CP}that))\backslash ({\langle\rangle}Nt(s(m))\backslash Sf))$}}, {\blacksquare}Nt(s(m)), {\square}{\forall}a{\forall}f(({\langle\rangle}Na\backslash Sf)\backslash ({\langle\rangle}Na\backslash Sf))]]\ \Rightarrow\ Sf
\using {/}L
\endprooftree
\justifies
[{\blacksquare}Nt(s(m))], {\square}(({\langle\rangle}{\exists}aNa\backslash Sf)/({\exists}aNa{\oplus}{\it CP}that)), [[{\blacksquare}Nt(s(f)), {\square}{\forall}a{\forall}f(({\langle\rangle}Na\backslash Sf)\backslash ({\langle\rangle}Na\backslash Sf)), \mbox{\fbox{${\forall}f((?{\blacksquare}((({\langle\rangle}Nt(s(m))\backslash Sf)/({\exists}cNc{\oplus}{\it CP}that))\backslash ({\langle\rangle}Nt(s(m))\backslash Sf))\backslash {[]^{-1}}{[]^{-1}}((({\langle\rangle}Nt(s(m))\backslash Sf)/({\exists}cNc{\oplus}{\it CP}that))\backslash ({\langle\rangle}Nt(s(m))\backslash Sf)))/{\blacksquare}((({\langle\rangle}Nt(s(m))\backslash Sf)/({\exists}cNc{\oplus}{\it CP}that))\backslash ({\langle\rangle}Nt(s(m))\backslash Sf)))$}}, {\blacksquare}Nt(s(m)), {\square}{\forall}a{\forall}f(({\langle\rangle}Na\backslash Sf)\backslash ({\langle\rangle}Na\backslash Sf))]]\ \Rightarrow\ Sf
\using {\forall}L
\endprooftree
\justifies
[{\blacksquare}Nt(s(m))], {\square}(({\langle\rangle}{\exists}aNa\backslash Sf)/({\exists}aNa{\oplus}{\it CP}that)), [[{\blacksquare}Nt(s(f)), {\square}{\forall}a{\forall}f(({\langle\rangle}Na\backslash Sf)\backslash ({\langle\rangle}Na\backslash Sf)), \mbox{\fbox{${\forall}b{\forall}f((?{\blacksquare}((({\langle\rangle}Nt(s(m))\backslash Sf)/({\exists}cNc{\oplus}{\it CP}b))\backslash ({\langle\rangle}Nt(s(m))\backslash Sf))\backslash {[]^{-1}}{[]^{-1}}((({\langle\rangle}Nt(s(m))\backslash Sf)/({\exists}cNc{\oplus}{\it CP}b))\backslash ({\langle\rangle}Nt(s(m))\backslash Sf)))/{\blacksquare}((({\langle\rangle}Nt(s(m))\backslash Sf)/({\exists}cNc{\oplus}{\it CP}b))\backslash ({\langle\rangle}Nt(s(m))\backslash Sf)))$}}, {\blacksquare}Nt(s(m)), {\square}{\forall}a{\forall}f(({\langle\rangle}Na\backslash Sf)\backslash ({\langle\rangle}Na\backslash Sf))]]\ \Rightarrow\ Sf
\using {\forall}L
\endprooftree
\justifies
[{\blacksquare}Nt(s(m))], {\square}(({\langle\rangle}{\exists}aNa\backslash Sf)/({\exists}aNa{\oplus}{\it CP}that)), [[{\blacksquare}Nt(s(f)), {\square}{\forall}a{\forall}f(({\langle\rangle}Na\backslash Sf)\backslash ({\langle\rangle}Na\backslash Sf)), \mbox{\fbox{${\forall}a{\forall}b{\forall}f((?{\blacksquare}((({\langle\rangle}Na\backslash Sf)/({\exists}cNc{\oplus}{\it CP}b))\backslash ({\langle\rangle}Na\backslash Sf))\backslash {[]^{-1}}{[]^{-1}}((({\langle\rangle}Na\backslash Sf)/({\exists}cNc{\oplus}{\it CP}b))\backslash ({\langle\rangle}Na\backslash Sf)))/{\blacksquare}((({\langle\rangle}Na\backslash Sf)/({\exists}cNc{\oplus}{\it CP}b))\backslash ({\langle\rangle}Na\backslash Sf)))$}}, {\blacksquare}Nt(s(m)), {\square}{\forall}a{\forall}f(({\langle\rangle}Na\backslash Sf)\backslash ({\langle\rangle}Na\backslash Sf))]]\ \Rightarrow\ Sf
\using {\forall}L
\endprooftree
\justifies
[{\blacksquare}Nt(s(m))], {\square}(({\langle\rangle}{\exists}aNa\backslash Sf)/({\exists}aNa{\oplus}{\it CP}that)), [[{\blacksquare}Nt(s(f)), {\square}{\forall}a{\forall}f(({\langle\rangle}Na\backslash Sf)\backslash ({\langle\rangle}Na\backslash Sf)), \mbox{\fbox{${\blacksquare}{\forall}a{\forall}b{\forall}f((?{\blacksquare}((({\langle\rangle}Na\backslash Sf)/({\exists}cNc{\oplus}{\it CP}b))\backslash ({\langle\rangle}Na\backslash Sf))\backslash {[]^{-1}}{[]^{-1}}((({\langle\rangle}Na\backslash Sf)/({\exists}cNc{\oplus}{\it CP}b))\backslash ({\langle\rangle}Na\backslash Sf)))/{\blacksquare}((({\langle\rangle}Na\backslash Sf)/({\exists}cNc{\oplus}{\it CP}b))\backslash ({\langle\rangle}Na\backslash Sf)))$}}, {\blacksquare}Nt(s(m)), {\square}{\forall}a{\forall}f(({\langle\rangle}Na\backslash Sf)\backslash ({\langle\rangle}Na\backslash Sf))]]\ \Rightarrow\ Sf
\using {\blacksquare}L
\endprooftree}}

\vspace{0.15in}

\noindent
This delivers the same semantics:
\disp{
$[(\mbox{\v{}}{\it today}\ ({\it Past}\ ((\mbox{\v{}}{\it seee}\ {\it m})\ {\it j})))\wedge (\mbox{\v{}}{\it yesterday}\ ({\it Past}\ ((\mbox{\v{}}{\it seee}\ {\it b})\ {\it j})))]$
}}

\subsection{Across-the-board extraction}

This example is (medial) sentential across-the-board extraction:
\disp{
${\bf man}{+}[[{\bf that}{+}[[[{\bf john}]{+}{\bf saw}{+}{\bf yesterday}{+}{\bf and}{+}[{\bf bill}]{+}{\bf saw}{+}{\bf today}]]]]: {\it CN}{\it s(m)}$}
Appropriate lexical lookup yields the semantically annotated sequent where the
coordinator is essentially of the form $(\exstexp X\bsl \abrack\abrack X)/X$ where
$X=S/\univexp N$.
\disp{
$\begin{array}[t]{l}
{\square}{\it CN}{\it s(m)}: {\it man}, [[{\blacksquare}{\forall}n({[]^{-1}}{[]^{-1}}({\it CN}{\it n}\backslash {\it CN}{\it n})/{\blacksquare}(({\langle\rangle}Nt(n){\sqcap}!{\blacksquare}Nt(n))\backslash Sf)):\\ \lambda A\lambda B\lambda C[({\it B}\ {\it C})\wedge ({\it A}\ {\it C})], [[[{\blacksquare}Nt(s(m)): {\it j}], {\square}(({\langle\rangle}{\exists}aNa\backslash Sf)/({\exists}aNa{\oplus}{\it CP}that)):\\ \mbox{\^{}}\lambda D\lambda E({\it Past}\ (({\it D}\casearrow F.(\mbox{\v{}}{\it seee}\ {\it F}); G.(\mbox{\v{}}{\it seet}\ {\it G}))\ {\it E})), {\square}{\forall}a{\forall}f(({\langle\rangle}Na\backslash Sf)\backslash ({\langle\rangle}Na\backslash Sf)):\\ \mbox{\^{}}\lambda H\lambda I(\mbox{\v{}}{\it yesterday}\ ({\it H}\ {\it I})),
{\blacksquare}{\forall}a{\forall}f((?{\blacksquare}(Sf/!Na)\backslash {[]^{-1}}{[]^{-1}}(Sf/!Na))/{\blacksquare}(Sf/!Na)):\\ (\Phinplus\ ({\it s}\ {\it 0})\ {\it and}), [{\blacksquare}Nt(s(m)): {\it b}], {\square}(({\langle\rangle}{\exists}aNa\backslash Sf)/({\exists}aNa{\oplus}{\it CP}that)):\\ \mbox{\^{}}\lambda J\lambda K({\it Past}\ (({\it J}\casearrow L.(\mbox{\v{}}{\it seee}\ {\it L}); M.(\mbox{\v{}}{\it seet}\ {\it M}))\ {\it K})),\\ {\square}{\forall}a{\forall}f(({\langle\rangle}Na\backslash Sf)\backslash ({\langle\rangle}Na\backslash Sf)): \mbox{\^{}}\lambda N\lambda O(\mbox{\v{}}{\it today}\ ({\it N}\ {\it O}))]]]]\ \Rightarrow\ {\it CN}{\it s(m)}
\end{array}$}
The relative pronoun is essentially of the form $(\CN\bsl\CN)/((\mybrack N\iaconj\univexp N)\bsl S)$
where the semantically inactively conjoined nominals $\mybrack N$
and $\univexp N$
are for subject relativisation
and object relativisation respectively.
There is the following derivation.
In \textcircled{1} the righthand conjunct is derived as a type of the shape $S/\univexp N$.
The universal exponential argument is lowered into the antecedent and into the stoup:
\disp{$\mini
\prooftree
\prooftree
N; \ldots\yields S
\justifies
\ldots, \univexp N\yields S
\using \univexp L
\endprooftree
\justifies
\ldots\yields S/\univexp N
\using /R
\endprooftree$} 
The nominal percolates leftwards in the stoup into the minor premise
at the application of the adverb to the verb phrase and leftwards again in the stoup 
into the minor premise at the application of the (polymorphic) transitive verb to its
object: at the top leftmost subderivation $\univexp P$ brings the nominal
out of the stoup to fulfil the role of this object.
Subtree \textcircled{2} which analyses the lefthand conjunct
is exactly the same as \textcircled{1} --- except for the bottommost existential
exponential right rule --- hence it has been elided.
The principal new action in the main derivation occurs above \textcircled{3}
where the coordinate structure is supplied as the higher order
relative pronoun argument.
The succession has the form:
\disp{$\mini
\prooftree
\prooftree
\prooftree
N; \ldots\yields S
\justifies
\univexp N, \ldots\yields S
\using \univexp L
\endprooftree
\justifies
\mybrack N\iaconj\univexp N, \ldots\yields S
\using \iaconj L
\endprooftree
\justifies
\ldots\yields (\mybrack N\iaconj\univexp N)\bsl S
\using \bsl R
\endprooftree$}
Once the coordinate structure (here `\ldots') is analysed as $S/\univexp N$
the hypothetical nominal subtype which has entered the stoup yields the $\univexp N$
in the subderivation at the top, which has the form:
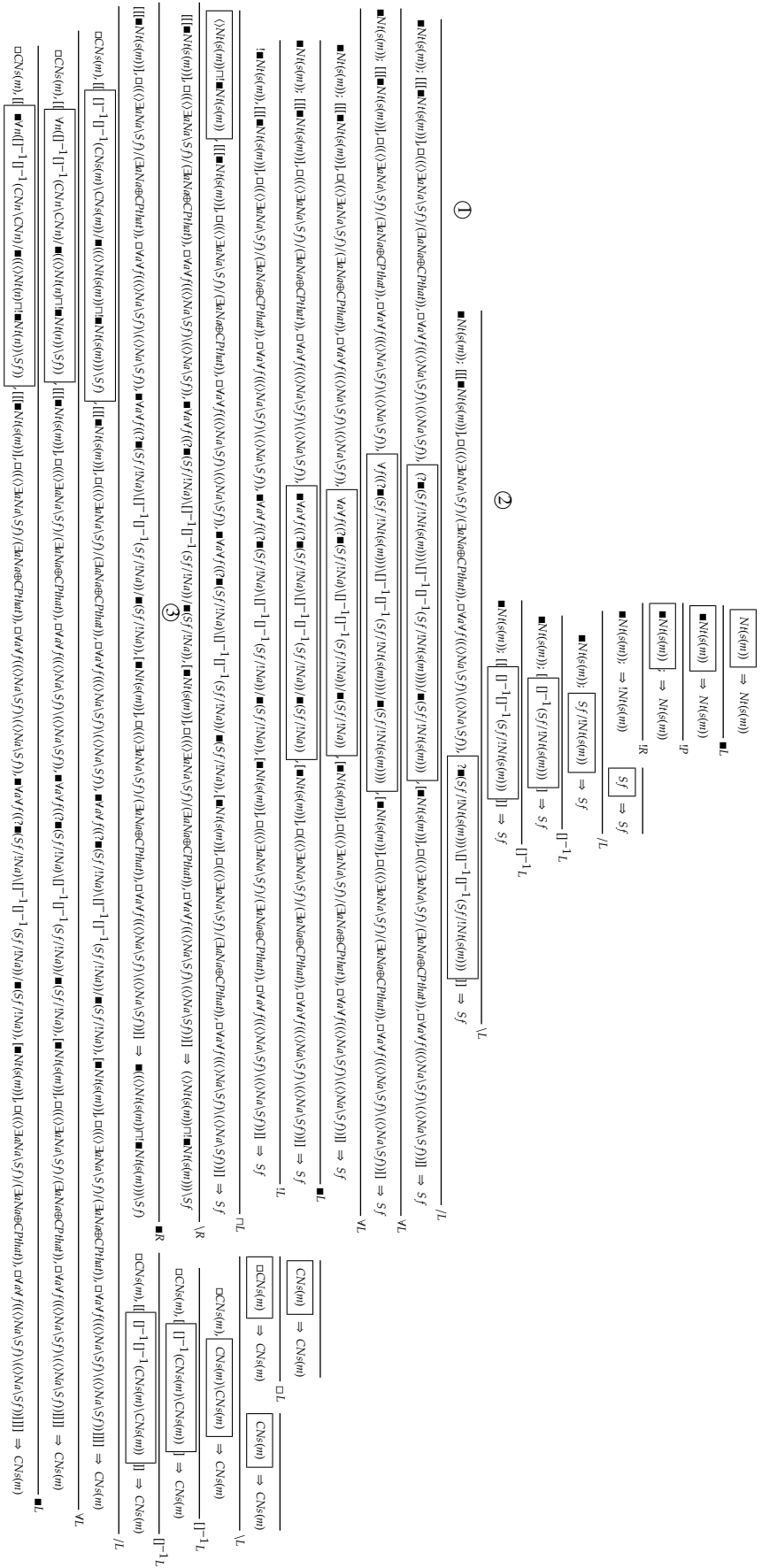
\disp{$\mini
\prooftree
\prooftree
\prooftree
N\yields N
\justifies
N; \yields N
\using \univexp P
\endprooftree
\justifies
N; \yields \univexp N
\using \univexp R
\endprooftree
S\yields S
\justifies
N; S/\univexp N\yields S
\using /L
\endprooftree$}

\vspace{0.15in}

\begin{center}{\tiny
\prooftree
\prooftree
\prooftree
\prooftree
\prooftree
\prooftree
\prooftree
\prooftree
\prooftree
\prooftree
\prooftree
\prooftree
\prooftree
\prooftree
\prooftree
\justifies
Nt(s(m))\ \Rightarrow\ Nt(s(m))
\endprooftree
\justifies
\mbox{\fbox{$Nt(s(m))$}};\ \ \Rightarrow\ Nt(s(m))
\using {!}P
\endprooftree
\justifies
Nt(s(m));\ \ \Rightarrow\ \fbox{${\exists}aNa$}
\using {\exists}R
\endprooftree
\justifies
Nt(s(m));\ \ \Rightarrow\ \fbox{${\exists}aNa{\oplus}{\it CP}that$}
\using {\oplus}R
\endprooftree
\prooftree
\prooftree
\prooftree
\prooftree
\justifies
Nt(s(m))\ \Rightarrow\ Nt(s(m))
\endprooftree
\justifies
Nt(s(m))\ \Rightarrow\ \fbox{${\exists}aNa$}
\using {\exists}R
\endprooftree
\justifies
[Nt(s(m))]\ \Rightarrow\ \fbox{${\langle\rangle}{\exists}aNa$}
\using {\langle\rangle}R
\endprooftree
\prooftree
\justifies
\mbox{\fbox{$Sf$}}\ \Rightarrow\ Sf
\endprooftree
\justifies
[Nt(s(m))], \mbox{\fbox{${\langle\rangle}{\exists}aNa\backslash Sf$}}\ \Rightarrow\ Sf
\using {\backslash}L
\endprooftree
\justifies
Nt(s(m));\ [Nt(s(m))], \mbox{\fbox{$({\langle\rangle}{\exists}aNa\backslash Sf)/({\exists}aNa{\oplus}{\it CP}that)$}}\ \Rightarrow\ Sf
\using {/}L
\endprooftree
\justifies
Nt(s(m));\ [Nt(s(m))], \mbox{\fbox{${\square}(({\langle\rangle}{\exists}aNa\backslash Sf)/({\exists}aNa{\oplus}{\it CP}that))$}}\ \Rightarrow\ Sf
\using {\Box}L
\endprooftree
\justifies
Nt(s(m));\ {\langle\rangle}Nt(s(m)), {\square}(({\langle\rangle}{\exists}aNa\backslash Sf)/({\exists}aNa{\oplus}{\it CP}that))\ \Rightarrow\ Sf
\using {\langle\rangle}L
\endprooftree
\justifies
Nt(s(m));\ {\square}(({\langle\rangle}{\exists}aNa\backslash Sf)/({\exists}aNa{\oplus}{\it CP}that))\ \Rightarrow\ {\langle\rangle}Nt(s(m))\backslash Sf
\using {\backslash}R
\endprooftree
\prooftree
\prooftree
\prooftree
\prooftree
\justifies
\mbox{\fbox{$Nt(s(m))$}}\ \Rightarrow\ Nt(s(m))
\endprooftree
\justifies
\mbox{\fbox{${\blacksquare}Nt(s(m))$}}\ \Rightarrow\ Nt(s(m))
\using {\blacksquare}L
\endprooftree
\justifies
[{\blacksquare}Nt(s(m))]\ \Rightarrow\ \fbox{${\langle\rangle}Nt(s(m))$}
\using {\langle\rangle}R
\endprooftree
\prooftree
\justifies
\mbox{\fbox{$Sf$}}\ \Rightarrow\ Sf
\endprooftree
\justifies
[{\blacksquare}Nt(s(m))], \mbox{\fbox{${\langle\rangle}Nt(s(m))\backslash Sf$}}\ \Rightarrow\ Sf
\using {\backslash}L
\endprooftree
\justifies
Nt(s(m));\ [{\blacksquare}Nt(s(m))], {\square}(({\langle\rangle}{\exists}aNa\backslash Sf)/({\exists}aNa{\oplus}{\it CP}that)), \mbox{\fbox{$({\langle\rangle}Nt(s(m))\backslash Sf)\backslash ({\langle\rangle}Nt(s(m))\backslash Sf)$}}\ \Rightarrow\ Sf
\using {\backslash}L
\endprooftree
\justifies
Nt(s(m));\ [{\blacksquare}Nt(s(m))], {\square}(({\langle\rangle}{\exists}aNa\backslash Sf)/({\exists}aNa{\oplus}{\it CP}that)), \mbox{\fbox{${\forall}f(({\langle\rangle}Nt(s(m))\backslash Sf)\backslash ({\langle\rangle}Nt(s(m))\backslash Sf))$}}\ \Rightarrow\ Sf
\using {\forall}L
\endprooftree
\justifies
Nt(s(m));\ [{\blacksquare}Nt(s(m))], {\square}(({\langle\rangle}{\exists}aNa\backslash Sf)/({\exists}aNa{\oplus}{\it CP}that)), \mbox{\fbox{${\forall}a{\forall}f(({\langle\rangle}Na\backslash Sf)\backslash ({\langle\rangle}Na\backslash Sf))$}}\ \Rightarrow\ Sf
\using {\forall}L
\endprooftree
\justifies
Nt(s(m));\ [{\blacksquare}Nt(s(m))], {\square}(({\langle\rangle}{\exists}aNa\backslash Sf)/({\exists}aNa{\oplus}{\it CP}that)), \mbox{\fbox{${\square}{\forall}a{\forall}f(({\langle\rangle}Na\backslash Sf)\backslash ({\langle\rangle}Na\backslash Sf))$}}\ \Rightarrow\ Sf
\using {\Box}L
\endprooftree
\justifies
[{\blacksquare}Nt(s(m))], {\square}(({\langle\rangle}{\exists}aNa\backslash Sf)/({\exists}aNa{\oplus}{\it CP}that)), {\square}{\forall}a{\forall}f(({\langle\rangle}Na\backslash Sf)\backslash ({\langle\rangle}Na\backslash Sf)), !Nt(s(m))\ \Rightarrow\ Sf
\using {!}L
\endprooftree
\justifies
[{\blacksquare}Nt(s(m))], {\square}(({\langle\rangle}{\exists}aNa\backslash Sf)/({\exists}aNa{\oplus}{\it CP}that)), {\square}{\forall}a{\forall}f(({\langle\rangle}Na\backslash Sf)\backslash ({\langle\rangle}Na\backslash Sf))\ \Rightarrow\ Sf/!Nt(s(m))
\using {/}R
\endprooftree
\justifies
\begin{array}{c}
[{\blacksquare}Nt(s(m))], {\square}(({\langle\rangle}{\exists}aNa\backslash Sf)/({\exists}aNa{\oplus}{\it CP}that)), {\square}{\forall}a{\forall}f(({\langle\rangle}Na\backslash Sf)\backslash ({\langle\rangle}Na\backslash Sf))\ \Rightarrow\ {\blacksquare}(Sf/!Nt(s(m)))\\
\mbox{\footnotesize\textcircled{1}}
\end{array}
\using {\blacksquare}R
\endprooftree

\prooftree
\vdots
\justifies
\begin{array}{c}
[{\blacksquare}Nt(s(m))], {\square}(({\langle\rangle}{\exists}aNa\backslash Sf)/({\exists}aNa{\oplus}{\it CP}that)), {\square}{\forall}a{\forall}f(({\langle\rangle}Na\backslash Sf)\backslash ({\langle\rangle}Na\backslash Sf))\ \Rightarrow\ \fbox{$?{\blacksquare}(Sf/!Nt(s(m)))$}\\
\mbox{\footnotesize\textcircled{2}}
\end{array}
\using {?}R
\endprooftree}
\end{center}

\begin{center}
\rotatebox{-90}{\tiny
\prooftree
\prooftree
\prooftree
\prooftree
\prooftree
\prooftree
\prooftree
\prooftree
\prooftree
\prooftree
\prooftree
\mbox{\footnotesize\textcircled{1}}\tab
\prooftree
\mbox{\footnotesize\textcircled{2}}\tab
\prooftree
\prooftree
\prooftree
\prooftree
\prooftree
\prooftree
\prooftree
\justifies
\mbox{\fbox{$Nt(s(m))$}}\ \Rightarrow\ Nt(s(m))
\endprooftree
\justifies
\mbox{\fbox{${\blacksquare}Nt(s(m))$}}\ \Rightarrow\ Nt(s(m))
\using {\blacksquare}L
\endprooftree
\justifies
\mbox{\fbox{${\blacksquare}Nt(s(m))$}};\ \ \Rightarrow\ Nt(s(m))
\using {!}P
\endprooftree
\justifies
{\blacksquare}Nt(s(m));\ \ \Rightarrow\ !Nt(s(m))
\using {!}R
\endprooftree
\prooftree
\justifies
\mbox{\fbox{$Sf$}}\ \Rightarrow\ Sf
\endprooftree
\justifies
{\blacksquare}Nt(s(m));\ \mbox{\fbox{$Sf/!Nt(s(m))$}}\ \Rightarrow\ Sf
\using {/}L
\endprooftree
\justifies
{\blacksquare}Nt(s(m));\ [\mbox{\fbox{${[]^{-1}}(Sf/!Nt(s(m)))$}}]\ \Rightarrow\ Sf
\using {[]^{-1}}L
\endprooftree
\justifies
{\blacksquare}Nt(s(m));\ [[\mbox{\fbox{${[]^{-1}}{[]^{-1}}(Sf/!Nt(s(m)))$}}]]\ \Rightarrow\ Sf
\using {[]^{-1}}L
\endprooftree
\justifies
{\blacksquare}Nt(s(m));\ [[[{\blacksquare}Nt(s(m))], {\square}(({\langle\rangle}{\exists}aNa\backslash Sf)/({\exists}aNa{\oplus}{\it CP}that)), {\square}{\forall}a{\forall}f(({\langle\rangle}Na\backslash Sf)\backslash ({\langle\rangle}Na\backslash Sf)), \mbox{\fbox{$?{\blacksquare}(Sf/!Nt(s(m)))\backslash {[]^{-1}}{[]^{-1}}(Sf/!Nt(s(m)))$}}]]\ \Rightarrow\ Sf
\using {\backslash}L
\endprooftree
\justifies
{\blacksquare}Nt(s(m));\ [[[{\blacksquare}Nt(s(m))], {\square}(({\langle\rangle}{\exists}aNa\backslash Sf)/({\exists}aNa{\oplus}{\it CP}that)), {\square}{\forall}a{\forall}f(({\langle\rangle}Na\backslash Sf)\backslash ({\langle\rangle}Na\backslash Sf)), \mbox{\fbox{$(?{\blacksquare}(Sf/!Nt(s(m)))\backslash {[]^{-1}}{[]^{-1}}(Sf/!Nt(s(m))))/{\blacksquare}(Sf/!Nt(s(m)))$}}, [{\blacksquare}Nt(s(m))], {\square}(({\langle\rangle}{\exists}aNa\backslash Sf)/({\exists}aNa{\oplus}{\it CP}that)), {\square}{\forall}a{\forall}f(({\langle\rangle}Na\backslash Sf)\backslash ({\langle\rangle}Na\backslash Sf))]]\ \Rightarrow\ Sf
\using {/}L
\endprooftree
\justifies
{\blacksquare}Nt(s(m));\ [[[{\blacksquare}Nt(s(m))], {\square}(({\langle\rangle}{\exists}aNa\backslash Sf)/({\exists}aNa{\oplus}{\it CP}that)), {\square}{\forall}a{\forall}f(({\langle\rangle}Na\backslash Sf)\backslash ({\langle\rangle}Na\backslash Sf)), \mbox{\fbox{${\forall}f((?{\blacksquare}(Sf/!Nt(s(m)))\backslash {[]^{-1}}{[]^{-1}}(Sf/!Nt(s(m))))/{\blacksquare}(Sf/!Nt(s(m))))$}}, [{\blacksquare}Nt(s(m))], {\square}(({\langle\rangle}{\exists}aNa\backslash Sf)/({\exists}aNa{\oplus}{\it CP}that)), {\square}{\forall}a{\forall}f(({\langle\rangle}Na\backslash Sf)\backslash ({\langle\rangle}Na\backslash Sf))]]\ \Rightarrow\ Sf
\using {\forall}L
\endprooftree
\justifies
{\blacksquare}Nt(s(m));\ [[[{\blacksquare}Nt(s(m))], {\square}(({\langle\rangle}{\exists}aNa\backslash Sf)/({\exists}aNa{\oplus}{\it CP}that)), {\square}{\forall}a{\forall}f(({\langle\rangle}Na\backslash Sf)\backslash ({\langle\rangle}Na\backslash Sf)), \mbox{\fbox{${\forall}a{\forall}f((?{\blacksquare}(Sf/!Na)\backslash {[]^{-1}}{[]^{-1}}(Sf/!Na))/{\blacksquare}(Sf/!Na))$}}, [{\blacksquare}Nt(s(m))], {\square}(({\langle\rangle}{\exists}aNa\backslash Sf)/({\exists}aNa{\oplus}{\it CP}that)), {\square}{\forall}a{\forall}f(({\langle\rangle}Na\backslash Sf)\backslash ({\langle\rangle}Na\backslash Sf))]]\ \Rightarrow\ Sf
\using {\forall}L
\endprooftree
\justifies
{\blacksquare}Nt(s(m));\ [[[{\blacksquare}Nt(s(m))], {\square}(({\langle\rangle}{\exists}aNa\backslash Sf)/({\exists}aNa{\oplus}{\it CP}that)), {\square}{\forall}a{\forall}f(({\langle\rangle}Na\backslash Sf)\backslash ({\langle\rangle}Na\backslash Sf)), \mbox{\fbox{${\blacksquare}{\forall}a{\forall}f((?{\blacksquare}(Sf/!Na)\backslash {[]^{-1}}{[]^{-1}}(Sf/!Na))/{\blacksquare}(Sf/!Na))$}}, [{\blacksquare}Nt(s(m))], {\square}(({\langle\rangle}{\exists}aNa\backslash Sf)/({\exists}aNa{\oplus}{\it CP}that)), {\square}{\forall}a{\forall}f(({\langle\rangle}Na\backslash Sf)\backslash ({\langle\rangle}Na\backslash Sf))]]\ \Rightarrow\ Sf
\using {\blacksquare}L
\endprooftree
\justifies
!{\blacksquare}Nt(s(m)), [[[{\blacksquare}Nt(s(m))], {\square}(({\langle\rangle}{\exists}aNa\backslash Sf)/({\exists}aNa{\oplus}{\it CP}that)), {\square}{\forall}a{\forall}f(({\langle\rangle}Na\backslash Sf)\backslash ({\langle\rangle}Na\backslash Sf)), {\blacksquare}{\forall}a{\forall}f((?{\blacksquare}(Sf/!Na)\backslash {[]^{-1}}{[]^{-1}}(Sf/!Na))/{\blacksquare}(Sf/!Na)), [{\blacksquare}Nt(s(m))], {\square}(({\langle\rangle}{\exists}aNa\backslash Sf)/({\exists}aNa{\oplus}{\it CP}that)), {\square}{\forall}a{\forall}f(({\langle\rangle}Na\backslash Sf)\backslash ({\langle\rangle}Na\backslash Sf))]]\ \Rightarrow\ Sf
\using {!}L
\endprooftree
\justifies
\mbox{\fbox{${\langle\rangle}Nt(s(m)){\sqcap}!{\blacksquare}Nt(s(m))$}}, [[[{\blacksquare}Nt(s(m))], {\square}(({\langle\rangle}{\exists}aNa\backslash Sf)/({\exists}aNa{\oplus}{\it CP}that)), {\square}{\forall}a{\forall}f(({\langle\rangle}Na\backslash Sf)\backslash ({\langle\rangle}Na\backslash Sf)), {\blacksquare}{\forall}a{\forall}f((?{\blacksquare}(Sf/!Na)\backslash {[]^{-1}}{[]^{-1}}(Sf/!Na))/{\blacksquare}(Sf/!Na)), [{\blacksquare}Nt(s(m))], {\square}(({\langle\rangle}{\exists}aNa\backslash Sf)/({\exists}aNa{\oplus}{\it CP}that)), {\square}{\forall}a{\forall}f(({\langle\rangle}Na\backslash Sf)\backslash ({\langle\rangle}Na\backslash Sf))]]\ \Rightarrow\ Sf
\using {\sqcap}L
\endprooftree
\justifies
\begin{array}{c}
{}[[[{\blacksquare}Nt(s(m))], {\square}(({\langle\rangle}{\exists}aNa\backslash Sf)/({\exists}aNa{\oplus}{\it CP}that)), {\square}{\forall}a{\forall}f(({\langle\rangle}Na\backslash Sf)\backslash ({\langle\rangle}Na\backslash Sf)), {\blacksquare}{\forall}a{\forall}f((?{\blacksquare}(Sf/!Na)\backslash {[]^{-1}}{[]^{-1}}(Sf/!Na))/{\blacksquare}(Sf/!Na)), [{\blacksquare}Nt(s(m))], {\square}(({\langle\rangle}{\exists}aNa\backslash Sf)/({\exists}aNa{\oplus}{\it CP}that)), {\square}{\forall}a{\forall}f(({\langle\rangle}Na\backslash Sf)\backslash ({\langle\rangle}Na\backslash Sf))]]\ \Rightarrow\ ({\langle\rangle}Nt(s(m)){\sqcap}!{\blacksquare}Nt(s(m)))\backslash Sf\\
\mbox{\footnotesize\textcircled{3}}
\end{array}
\using {\backslash}R
\endprooftree
\justifies
[[[{\blacksquare}Nt(s(m))], {\square}(({\langle\rangle}{\exists}aNa\backslash Sf)/({\exists}aNa{\oplus}{\it CP}that)), {\square}{\forall}a{\forall}f(({\langle\rangle}Na\backslash Sf)\backslash ({\langle\rangle}Na\backslash Sf)), {\blacksquare}{\forall}a{\forall}f((?{\blacksquare}(Sf/!Na)\backslash {[]^{-1}}{[]^{-1}}(Sf/!Na))/{\blacksquare}(Sf/!Na)), [{\blacksquare}Nt(s(m))], {\square}(({\langle\rangle}{\exists}aNa\backslash Sf)/({\exists}aNa{\oplus}{\it CP}that)), {\square}{\forall}a{\forall}f(({\langle\rangle}Na\backslash Sf)\backslash ({\langle\rangle}Na\backslash Sf))]]\ \Rightarrow\ {\blacksquare}(({\langle\rangle}Nt(s(m)){\sqcap}!{\blacksquare}Nt(s(m)))\backslash Sf)
\using {\blacksquare}R
\endprooftree
\prooftree
\prooftree
\prooftree
\prooftree
\prooftree
\justifies
\mbox{\fbox{${\it CN}{\it s(m)}$}}\ \Rightarrow\ {\it CN}{\it s(m)}
\endprooftree
\justifies
\mbox{\fbox{${\square}{\it CN}{\it s(m)}$}}\ \Rightarrow\ {\it CN}{\it s(m)}
\using {\Box}L
\endprooftree
\prooftree
\justifies
\mbox{\fbox{${\it CN}{\it s(m)}$}}\ \Rightarrow\ {\it CN}{\it s(m)}
\endprooftree
\justifies
{\square}{\it CN}{\it s(m)}, \mbox{\fbox{${\it CN}{\it s(m)}\backslash {\it CN}{\it s(m)}$}}\ \Rightarrow\ {\it CN}{\it s(m)}
\using {\backslash}L
\endprooftree
\justifies
{\square}{\it CN}{\it s(m)}, [\mbox{\fbox{${[]^{-1}}({\it CN}{\it s(m)}\backslash {\it CN}{\it s(m)})$}}]\ \Rightarrow\ {\it CN}{\it s(m)}
\using {[]^{-1}}L
\endprooftree
\justifies
{\square}{\it CN}{\it s(m)}, [[\mbox{\fbox{${[]^{-1}}{[]^{-1}}({\it CN}{\it s(m)}\backslash {\it CN}{\it s(m)})$}}]]\ \Rightarrow\ {\it CN}{\it s(m)}
\using {[]^{-1}}L
\endprooftree
\justifies
{\square}{\it CN}{\it s(m)}, [[\mbox{\fbox{${[]^{-1}}{[]^{-1}}({\it CN}{\it s(m)}\backslash {\it CN}{\it s(m)})/{\blacksquare}(({\langle\rangle}Nt(s(m)){\sqcap}!{\blacksquare}Nt(s(m)))\backslash Sf)$}}, [[[{\blacksquare}Nt(s(m))], {\square}(({\langle\rangle}{\exists}aNa\backslash Sf)/({\exists}aNa{\oplus}{\it CP}that)), {\square}{\forall}a{\forall}f(({\langle\rangle}Na\backslash Sf)\backslash ({\langle\rangle}Na\backslash Sf)), {\blacksquare}{\forall}a{\forall}f((?{\blacksquare}(Sf/!Na)\backslash {[]^{-1}}{[]^{-1}}(Sf/!Na))/{\blacksquare}(Sf/!Na)), [{\blacksquare}Nt(s(m))], {\square}(({\langle\rangle}{\exists}aNa\backslash Sf)/({\exists}aNa{\oplus}{\it CP}that)), {\square}{\forall}a{\forall}f(({\langle\rangle}Na\backslash Sf)\backslash ({\langle\rangle}Na\backslash Sf))]]]]\ \Rightarrow\ {\it CN}{\it s(m)}
\using {/}L
\endprooftree
\justifies
{\square}{\it CN}{\it s(m)}, [[\mbox{\fbox{${\forall}n({[]^{-1}}{[]^{-1}}({\it CN}{\it n}\backslash {\it CN}{\it n})/{\blacksquare}(({\langle\rangle}Nt(n){\sqcap}!{\blacksquare}Nt(n))\backslash Sf))$}}, [[[{\blacksquare}Nt(s(m))], {\square}(({\langle\rangle}{\exists}aNa\backslash Sf)/({\exists}aNa{\oplus}{\it CP}that)), {\square}{\forall}a{\forall}f(({\langle\rangle}Na\backslash Sf)\backslash ({\langle\rangle}Na\backslash Sf)), {\blacksquare}{\forall}a{\forall}f((?{\blacksquare}(Sf/!Na)\backslash {[]^{-1}}{[]^{-1}}(Sf/!Na))/{\blacksquare}(Sf/!Na)), [{\blacksquare}Nt(s(m))], {\square}(({\langle\rangle}{\exists}aNa\backslash Sf)/({\exists}aNa{\oplus}{\it CP}that)), {\square}{\forall}a{\forall}f(({\langle\rangle}Na\backslash Sf)\backslash ({\langle\rangle}Na\backslash Sf))]]]]\ \Rightarrow\ {\it CN}{\it s(m)}
\using {\forall}L
\endprooftree
\justifies
{\square}{\it CN}{\it s(m)}, [[\mbox{\fbox{${\blacksquare}{\forall}n({[]^{-1}}{[]^{-1}}({\it CN}{\it n}\backslash {\it CN}{\it n})/{\blacksquare}(({\langle\rangle}Nt(n){\sqcap}!{\blacksquare}Nt(n))\backslash Sf))$}}, [[[{\blacksquare}Nt(s(m))], {\square}(({\langle\rangle}{\exists}aNa\backslash Sf)/({\exists}aNa{\oplus}{\it CP}that)), {\square}{\forall}a{\forall}f(({\langle\rangle}Na\backslash Sf)\backslash ({\langle\rangle}Na\backslash Sf)), {\blacksquare}{\forall}a{\forall}f((?{\blacksquare}(Sf/!Na)\backslash {[]^{-1}}{[]^{-1}}(Sf/!Na))/{\blacksquare}(Sf/!Na)), [{\blacksquare}Nt(s(m))], {\square}(({\langle\rangle}{\exists}aNa\backslash Sf)/({\exists}aNa{\oplus}{\it CP}that)), {\square}{\forall}a{\forall}f(({\langle\rangle}Na\backslash Sf)\backslash ({\langle\rangle}Na\backslash Sf))]]]]\ \Rightarrow\ {\it CN}{\it s(m)}
\using {\blacksquare}L
\endprooftree}
\end{center}

\vspace{0.15in}

\noindent
This delivers semantics:
\disp{
$\lambda C[(\mbox{\v{}}{\it man}\ {\it C})\wedge [(\mbox{\v{}}{\it yesterday}\ ({\it Past}\ ((\mbox{\v{}}{\it seee}\ {\it C})\ {\it j})))\wedge (\mbox{\v{}}{\it today}\ ({\it Past}\ ((\mbox{\v{}}{\it seee}\ {\it C})\ {\it b})))]]$}

Note that (with bracket modalities) the CSC is respected in TLG as in 
phrase structure grammar and categorial grammar 
formalisms assuming like type coordination schemata,
because there is no coordinator type instance conjoining
sentences only one of which contains a gap.

\commentout{

And~crd(25) is verb phrase (medial) Across-The-Board extraction:
\disp{
(crd(25)) ${\bf man}{+}[[{\bf that}{+}[{\bf john}]{+}[[{\bf saw}{+}{\bf yesterday}{+}{\bf and}{+}{\bf met}{+}{\bf today}]]]]: {\it CN}{\it s(m)}$}
Appropriate lexical insertion yields the sequent:
\disp{
${\square}{\it CN}{\it s(m)}: {\it man}, [[{\blacksquare}{\forall}n({[]^{-1}}{[]^{-1}}({\it CN}{\it n}\backslash {\it CN}{\it n})/{\blacksquare}(({\langle\rangle}Nt(n){\sqcap}!{\blacksquare}Nt(n))\backslash Sf)): \lambda A\lambda B\lambda C[({\it B}\ {\it C})\wedge ({\it A}\ {\it C})], [{\blacksquare}Nt(s(m)): {\it j}], [[{\square}(({\langle\rangle}{\exists}aNa\backslash Sf)/({\exists}aNa{\oplus}{\it CP}that)): \mbox{\^{}}\lambda D\lambda E({\it Past}\ (({\it D}\casearrow F.(\mbox{\v{}}{\it seee}\ {\it F}); G.(\mbox{\v{}}{\it seet}\ {\it G}))\ {\it E})), {\square}{\forall}a{\forall}f(({\langle\rangle}Na\backslash Sf)\backslash ({\langle\rangle}Na\backslash Sf)): \mbox{\^{}}\lambda H\lambda I(\mbox{\v{}}{\it yesterday}\ ({\it H}\ {\it I})),\\
{\blacksquare}{\forall}a{\forall}b{\forall}f((?{\blacksquare}(({\langle\rangle}Na\backslash Sf)/!Nb)\backslash {[]^{-1}}{[]^{-1}}(({\langle\rangle}Na\backslash Sf)/!Nb))/{\blacksquare}(({\langle\rangle}Na\backslash Sf)/!Nb)): (\Phinplus\ ({\it s}\ ({\it s}\ {\it 0}))\ {\it and}), {\square}(({\langle\rangle}{\exists}aNa\backslash Sf)/{\exists}aNa): \mbox{\^{}}\lambda J\lambda K({\it Past}\ ((\mbox{\v{}}{\it meet}\ {\it J})\ {\it K})), {\square}{\forall}a{\forall}f(({\langle\rangle}Na\backslash Sf)\backslash ({\langle\rangle}Na\backslash Sf)): \mbox{\^{}}\lambda L\lambda M(\mbox{\v{}}{\it today}\ ({\it L}\ {\it M}))]]]]\ \Rightarrow\ {\it CN}{\it s(m)}$}
There is the derivation:

\vspace{0.15in}

{\tiny
\prooftree
\prooftree
\prooftree
\prooftree
\prooftree
\prooftree
\prooftree
\prooftree
\prooftree
\prooftree
\prooftree
\prooftree
\prooftree
\prooftree
\justifies
\mbox{\fbox{$Nt(s(m))$}}\ \Rightarrow\ Nt(s(m))
\endprooftree
\justifies
\mbox{\fbox{${\blacksquare}Nt(s(m))$}}\ \Rightarrow\ Nt(s(m))
\using {\blacksquare}L
\endprooftree
\justifies
{\blacksquare}Nt(s(m))\ \Rightarrow\ \fbox{${\exists}bNb$}
\using {\exists}R
\endprooftree
\prooftree
\prooftree
\prooftree
\justifies
Nt(s(m))\ \Rightarrow\ Nt(s(m))
\endprooftree
\justifies
[Nt(s(m))]\ \Rightarrow\ \fbox{${\langle\rangle}Nt(s(m))$}
\using {\langle\rangle}R
\endprooftree
\prooftree
\justifies
\mbox{\fbox{$Sf$}}\ \Rightarrow\ Sf
\endprooftree
\justifies
[Nt(s(m))], \mbox{\fbox{${\langle\rangle}Nt(s(m))\backslash Sf$}}\ \Rightarrow\ Sf
\using {\backslash}L
\endprooftree
\justifies
[Nt(s(m))], \mbox{\fbox{$({\langle\rangle}Nt(s(m))\backslash Sf)/{\exists}bNb$}}, {\blacksquare}Nt(s(m))\ \Rightarrow\ Sf
\using {/}L
\endprooftree
\justifies
{\langle\rangle}Nt(s(m)), ({\langle\rangle}Nt(s(m))\backslash Sf)/{\exists}bNb, {\blacksquare}Nt(s(m))\ \Rightarrow\ Sf
\using {\langle\rangle}L
\endprooftree
\justifies
({\langle\rangle}Nt(s(m))\backslash Sf)/{\exists}bNb, {\blacksquare}Nt(s(m))\ \Rightarrow\ {\langle\rangle}Nt(s(m))\backslash Sf
\using {\backslash}R
\endprooftree
\prooftree
\prooftree
\prooftree
\justifies
Nt(s(m))\ \Rightarrow\ Nt(s(m))
\endprooftree
\justifies
[Nt(s(m))]\ \Rightarrow\ \fbox{${\langle\rangle}Nt(s(m))$}
\using {\langle\rangle}R
\endprooftree
\prooftree
\justifies
\mbox{\fbox{$Sf$}}\ \Rightarrow\ Sf
\endprooftree
\justifies
[Nt(s(m))], \mbox{\fbox{${\langle\rangle}Nt(s(m))\backslash Sf$}}\ \Rightarrow\ Sf
\using {\backslash}L
\endprooftree
\justifies
[Nt(s(m))], ({\langle\rangle}Nt(s(m))\backslash Sf)/{\exists}bNb, {\blacksquare}Nt(s(m)), \mbox{\fbox{$({\langle\rangle}Nt(s(m))\backslash Sf)\backslash ({\langle\rangle}Nt(s(m))\backslash Sf)$}}\ \Rightarrow\ Sf
\using {\backslash}L
\endprooftree
\justifies
[Nt(s(m))], ({\langle\rangle}Nt(s(m))\backslash Sf)/{\exists}bNb, {\blacksquare}Nt(s(m)), \mbox{\fbox{${\forall}f(({\langle\rangle}Nt(s(m))\backslash Sf)\backslash ({\langle\rangle}Nt(s(m))\backslash Sf))$}}\ \Rightarrow\ Sf
\using {\forall}L
\endprooftree
\justifies
[Nt(s(m))], ({\langle\rangle}Nt(s(m))\backslash Sf)/{\exists}bNb, {\blacksquare}Nt(s(m)), \mbox{\fbox{${\forall}a{\forall}f(({\langle\rangle}Na\backslash Sf)\backslash ({\langle\rangle}Na\backslash Sf))$}}\ \Rightarrow\ Sf
\using {\forall}L
\endprooftree
\justifies
[Nt(s(m))], ({\langle\rangle}Nt(s(m))\backslash Sf)/{\exists}bNb, {\blacksquare}Nt(s(m)), \mbox{\fbox{${\square}{\forall}a{\forall}f(({\langle\rangle}Na\backslash Sf)\backslash ({\langle\rangle}Na\backslash Sf))$}}\ \Rightarrow\ Sf
\using {\Box}L
\endprooftree
\justifies
{\langle\rangle}Nt(s(m)), ({\langle\rangle}Nt(s(m))\backslash Sf)/{\exists}bNb, {\blacksquare}Nt(s(m)), {\square}{\forall}a{\forall}f(({\langle\rangle}Na\backslash Sf)\backslash ({\langle\rangle}Na\backslash Sf))\ \Rightarrow\ Sf
\using {\langle\rangle}L
\endprooftree
\justifies
({\langle\rangle}Nt(s(m))\backslash Sf)/{\exists}bNb, {\blacksquare}Nt(s(m)), {\square}{\forall}a{\forall}f(({\langle\rangle}Na\backslash Sf)\backslash ({\langle\rangle}Na\backslash Sf))\ \Rightarrow\ {\langle\rangle}Nt(s(m))\backslash Sf
\using {\backslash}R
\endprooftree
\justifies
{\blacksquare}Nt(s(m)), {\square}{\forall}a{\forall}f(({\langle\rangle}Na\backslash Sf)\backslash ({\langle\rangle}Na\backslash Sf))\ \Rightarrow\ (({\langle\rangle}Nt(s(m))\backslash Sf)/{\exists}bNb)\backslash ({\langle\rangle}Nt(s(m))\backslash Sf)
\using {\backslash}R
\endprooftree
\justifies
\begin{array}{c}
{\blacksquare}Nt(s(m)), {\square}{\forall}a{\forall}f(({\langle\rangle}Na\backslash Sf)\backslash ({\langle\rangle}Na\backslash Sf))\ \Rightarrow\ {\blacksquare}((({\langle\rangle}Nt(s(m))\backslash Sf)/{\exists}bNb)\backslash ({\langle\rangle}Nt(s(m))\backslash Sf))\\
\mbox{\footnotesize\textcircled{1}}
\end{array}
\using {\blacksquare}R
\endprooftree

\prooftree
\prooftree
\prooftree
\prooftree
\prooftree
\prooftree
\prooftree
\prooftree
\prooftree
\prooftree
\prooftree
\prooftree
\prooftree
\prooftree
\prooftree
\justifies
\mbox{\fbox{$Nt(s(f))$}}\ \Rightarrow\ Nt(s(f))
\endprooftree
\justifies
\mbox{\fbox{${\blacksquare}Nt(s(f))$}}\ \Rightarrow\ Nt(s(f))
\using {\blacksquare}L
\endprooftree
\justifies
{\blacksquare}Nt(s(f))\ \Rightarrow\ \fbox{${\exists}bNb$}
\using {\exists}R
\endprooftree
\prooftree
\prooftree
\prooftree
\justifies
Nt(s(m))\ \Rightarrow\ Nt(s(m))
\endprooftree
\justifies
[Nt(s(m))]\ \Rightarrow\ \fbox{${\langle\rangle}Nt(s(m))$}
\using {\langle\rangle}R
\endprooftree
\prooftree
\justifies
\mbox{\fbox{$Sf$}}\ \Rightarrow\ Sf
\endprooftree
\justifies
[Nt(s(m))], \mbox{\fbox{${\langle\rangle}Nt(s(m))\backslash Sf$}}\ \Rightarrow\ Sf
\using {\backslash}L
\endprooftree
\justifies
[Nt(s(m))], \mbox{\fbox{$({\langle\rangle}Nt(s(m))\backslash Sf)/{\exists}bNb$}}, {\blacksquare}Nt(s(f))\ \Rightarrow\ Sf
\using {/}L
\endprooftree
\justifies
{\langle\rangle}Nt(s(m)), ({\langle\rangle}Nt(s(m))\backslash Sf)/{\exists}bNb, {\blacksquare}Nt(s(f))\ \Rightarrow\ Sf
\using {\langle\rangle}L
\endprooftree
\justifies
({\langle\rangle}Nt(s(m))\backslash Sf)/{\exists}bNb, {\blacksquare}Nt(s(f))\ \Rightarrow\ {\langle\rangle}Nt(s(m))\backslash Sf
\using {\backslash}R
\endprooftree
\prooftree
\prooftree
\prooftree
\justifies
Nt(s(m))\ \Rightarrow\ Nt(s(m))
\endprooftree
\justifies
[Nt(s(m))]\ \Rightarrow\ \fbox{${\langle\rangle}Nt(s(m))$}
\using {\langle\rangle}R
\endprooftree
\prooftree
\justifies
\mbox{\fbox{$Sf$}}\ \Rightarrow\ Sf
\endprooftree
\justifies
[Nt(s(m))], \mbox{\fbox{${\langle\rangle}Nt(s(m))\backslash Sf$}}\ \Rightarrow\ Sf
\using {\backslash}L
\endprooftree
\justifies
[Nt(s(m))], ({\langle\rangle}Nt(s(m))\backslash Sf)/{\exists}bNb, {\blacksquare}Nt(s(f)), \mbox{\fbox{$({\langle\rangle}Nt(s(m))\backslash Sf)\backslash ({\langle\rangle}Nt(s(m))\backslash Sf)$}}\ \Rightarrow\ Sf
\using {\backslash}L
\endprooftree
\justifies
[Nt(s(m))], ({\langle\rangle}Nt(s(m))\backslash Sf)/{\exists}bNb, {\blacksquare}Nt(s(f)), \mbox{\fbox{${\forall}f(({\langle\rangle}Nt(s(m))\backslash Sf)\backslash ({\langle\rangle}Nt(s(m))\backslash Sf))$}}\ \Rightarrow\ Sf
\using {\forall}L
\endprooftree
\justifies
[Nt(s(m))], ({\langle\rangle}Nt(s(m))\backslash Sf)/{\exists}bNb, {\blacksquare}Nt(s(f)), \mbox{\fbox{${\forall}a{\forall}f(({\langle\rangle}Na\backslash Sf)\backslash ({\langle\rangle}Na\backslash Sf))$}}\ \Rightarrow\ Sf
\using {\forall}L
\endprooftree
\justifies
[Nt(s(m))], ({\langle\rangle}Nt(s(m))\backslash Sf)/{\exists}bNb, {\blacksquare}Nt(s(f)), \mbox{\fbox{${\square}{\forall}a{\forall}f(({\langle\rangle}Na\backslash Sf)\backslash ({\langle\rangle}Na\backslash Sf))$}}\ \Rightarrow\ Sf
\using {\Box}L
\endprooftree
\justifies
{\langle\rangle}Nt(s(m)), ({\langle\rangle}Nt(s(m))\backslash Sf)/{\exists}bNb, {\blacksquare}Nt(s(f)), {\square}{\forall}a{\forall}f(({\langle\rangle}Na\backslash Sf)\backslash ({\langle\rangle}Na\backslash Sf))\ \Rightarrow\ Sf
\using {\langle\rangle}L
\endprooftree
\justifies
({\langle\rangle}Nt(s(m))\backslash Sf)/{\exists}bNb, {\blacksquare}Nt(s(f)), {\square}{\forall}a{\forall}f(({\langle\rangle}Na\backslash Sf)\backslash ({\langle\rangle}Na\backslash Sf))\ \Rightarrow\ {\langle\rangle}Nt(s(m))\backslash Sf
\using {\backslash}R
\endprooftree
\justifies
{\blacksquare}Nt(s(f)), {\square}{\forall}a{\forall}f(({\langle\rangle}Na\backslash Sf)\backslash ({\langle\rangle}Na\backslash Sf))\ \Rightarrow\ (({\langle\rangle}Nt(s(m))\backslash Sf)/{\exists}bNb)\backslash ({\langle\rangle}Nt(s(m))\backslash Sf)
\using {\backslash}R
\endprooftree
\justifies
{\blacksquare}Nt(s(f)), {\square}{\forall}a{\forall}f(({\langle\rangle}Na\backslash Sf)\backslash ({\langle\rangle}Na\backslash Sf))\ \Rightarrow\ {\blacksquare}((({\langle\rangle}Nt(s(m))\backslash Sf)/{\exists}bNb)\backslash ({\langle\rangle}Nt(s(m))\backslash Sf))
\using {\blacksquare}R
\endprooftree
\justifies
\begin{array}{c}
{\blacksquare}Nt(s(f)), {\square}{\forall}a{\forall}f(({\langle\rangle}Na\backslash Sf)\backslash ({\langle\rangle}Na\backslash Sf))\ \Rightarrow\ \fbox{$?{\blacksquare}((({\langle\rangle}Nt(s(m))\backslash Sf)/{\exists}bNb)\backslash ({\langle\rangle}Nt(s(m))\backslash Sf))$}\\
\mbox{\footnotesize\textcircled{2}}
\end{array}
\using {?}R
\endprooftree

\rotatebox{-90}{
\prooftree
\prooftree
\prooftree
\prooftree
\mbox{\footnotesize\textcircled{1}}\tab
\prooftree
\mbox{\footnotesize\textcircled{2}}\tab
\prooftree
\prooftree
\prooftree
\prooftree
\prooftree
\prooftree
\prooftree
\prooftree
\prooftree
\prooftree
\prooftree
\prooftree
\justifies
N1\ \Rightarrow\ N1
\endprooftree
\justifies
N1\ \Rightarrow\ \fbox{${\exists}aNa$}
\using {\exists}R
\endprooftree
\justifies
N1\ \Rightarrow\ \fbox{${\exists}aNa{\oplus}{\it CP}that$}
\using {\oplus}R
\endprooftree
\prooftree
\prooftree
\prooftree
\prooftree
\justifies
Nt(s(m))\ \Rightarrow\ Nt(s(m))
\endprooftree
\justifies
Nt(s(m))\ \Rightarrow\ \fbox{${\exists}aNa$}
\using {\exists}R
\endprooftree
\justifies
[Nt(s(m))]\ \Rightarrow\ \fbox{${\langle\rangle}{\exists}aNa$}
\using {\langle\rangle}R
\endprooftree
\prooftree
\justifies
\mbox{\fbox{$Sf$}}\ \Rightarrow\ Sf
\endprooftree
\justifies
[Nt(s(m))], \mbox{\fbox{${\langle\rangle}{\exists}aNa\backslash Sf$}}\ \Rightarrow\ Sf
\using {\backslash}L
\endprooftree
\justifies
[Nt(s(m))], \mbox{\fbox{$({\langle\rangle}{\exists}aNa\backslash Sf)/({\exists}aNa{\oplus}{\it CP}that)$}}, N1\ \Rightarrow\ Sf
\using {/}L
\endprooftree
\justifies
[Nt(s(m))], \mbox{\fbox{${\square}(({\langle\rangle}{\exists}aNa\backslash Sf)/({\exists}aNa{\oplus}{\it CP}that))$}}, N1\ \Rightarrow\ Sf
\using {\Box}L
\endprooftree
\justifies
[Nt(s(m))], {\square}(({\langle\rangle}{\exists}aNa\backslash Sf)/({\exists}aNa{\oplus}{\it CP}that)), {\exists}bNb\ \Rightarrow\ Sf
\using {\exists}L
\endprooftree
\justifies
{\langle\rangle}Nt(s(m)), {\square}(({\langle\rangle}{\exists}aNa\backslash Sf)/({\exists}aNa{\oplus}{\it CP}that)), {\exists}bNb\ \Rightarrow\ Sf
\using {\langle\rangle}L
\endprooftree
\justifies
{\square}(({\langle\rangle}{\exists}aNa\backslash Sf)/({\exists}aNa{\oplus}{\it CP}that)), {\exists}bNb\ \Rightarrow\ {\langle\rangle}Nt(s(m))\backslash Sf
\using {\backslash}R
\endprooftree
\justifies
{\square}(({\langle\rangle}{\exists}aNa\backslash Sf)/({\exists}aNa{\oplus}{\it CP}that))\ \Rightarrow\ ({\langle\rangle}Nt(s(m))\backslash Sf)/{\exists}bNb
\using {/}R
\endprooftree
\prooftree
\prooftree
\prooftree
\prooftree
\justifies
\mbox{\fbox{$Nt(s(m))$}}\ \Rightarrow\ Nt(s(m))
\endprooftree
\justifies
\mbox{\fbox{${\blacksquare}Nt(s(m))$}}\ \Rightarrow\ Nt(s(m))
\using {\blacksquare}L
\endprooftree
\justifies
[{\blacksquare}Nt(s(m))]\ \Rightarrow\ \fbox{${\langle\rangle}Nt(s(m))$}
\using {\langle\rangle}R
\endprooftree
\prooftree
\justifies
\mbox{\fbox{$Sf$}}\ \Rightarrow\ Sf
\endprooftree
\justifies
[{\blacksquare}Nt(s(m))], \mbox{\fbox{${\langle\rangle}Nt(s(m))\backslash Sf$}}\ \Rightarrow\ Sf
\using {\backslash}L
\endprooftree
\justifies
[{\blacksquare}Nt(s(m))], {\square}(({\langle\rangle}{\exists}aNa\backslash Sf)/({\exists}aNa{\oplus}{\it CP}that)), \mbox{\fbox{$(({\langle\rangle}Nt(s(m))\backslash Sf)/{\exists}bNb)\backslash ({\langle\rangle}Nt(s(m))\backslash Sf)$}}\ \Rightarrow\ Sf
\using {\backslash}L
\endprooftree
\justifies
[{\blacksquare}Nt(s(m))], {\square}(({\langle\rangle}{\exists}aNa\backslash Sf)/({\exists}aNa{\oplus}{\it CP}that)), [\mbox{\fbox{${[]^{-1}}((({\langle\rangle}Nt(s(m))\backslash Sf)/{\exists}bNb)\backslash ({\langle\rangle}Nt(s(m))\backslash Sf))$}}]\ \Rightarrow\ Sf
\using {[]^{-1}}L
\endprooftree
\justifies
[{\blacksquare}Nt(s(m))], {\square}(({\langle\rangle}{\exists}aNa\backslash Sf)/({\exists}aNa{\oplus}{\it CP}that)), [[\mbox{\fbox{${[]^{-1}}{[]^{-1}}((({\langle\rangle}Nt(s(m))\backslash Sf)/{\exists}bNb)\backslash ({\langle\rangle}Nt(s(m))\backslash Sf))$}}]]\ \Rightarrow\ Sf
\using {[]^{-1}}L
\endprooftree
\justifies
[{\blacksquare}Nt(s(m))], {\square}(({\langle\rangle}{\exists}aNa\backslash Sf)/({\exists}aNa{\oplus}{\it CP}that)), [[{\blacksquare}Nt(s(f)), {\square}{\forall}a{\forall}f(({\langle\rangle}Na\backslash Sf)\backslash ({\langle\rangle}Na\backslash Sf)), \mbox{\fbox{$?{\blacksquare}((({\langle\rangle}Nt(s(m))\backslash Sf)/{\exists}bNb)\backslash ({\langle\rangle}Nt(s(m))\backslash Sf))\backslash {[]^{-1}}{[]^{-1}}((({\langle\rangle}Nt(s(m))\backslash Sf)/{\exists}bNb)\backslash ({\langle\rangle}Nt(s(m))\backslash Sf))$}}]]\ \Rightarrow\ Sf
\using {\backslash}L
\endprooftree
\justifies
[{\blacksquare}Nt(s(m))], {\square}(({\langle\rangle}{\exists}aNa\backslash Sf)/({\exists}aNa{\oplus}{\it CP}that)), [[{\blacksquare}Nt(s(f)), {\square}{\forall}a{\forall}f(({\langle\rangle}Na\backslash Sf)\backslash ({\langle\rangle}Na\backslash Sf)), \mbox{\fbox{$(?{\blacksquare}((({\langle\rangle}Nt(s(m))\backslash Sf)/{\exists}bNb)\backslash ({\langle\rangle}Nt(s(m))\backslash Sf))\backslash {[]^{-1}}{[]^{-1}}((({\langle\rangle}Nt(s(m))\backslash Sf)/{\exists}bNb)\backslash ({\langle\rangle}Nt(s(m))\backslash Sf)))/{\blacksquare}((({\langle\rangle}Nt(s(m))\backslash Sf)/{\exists}bNb)\backslash ({\langle\rangle}Nt(s(m))\backslash Sf))$}}, {\blacksquare}Nt(s(m)), {\square}{\forall}a{\forall}f(({\langle\rangle}Na\backslash Sf)\backslash ({\langle\rangle}Na\backslash Sf))]]\ \Rightarrow\ Sf
\using {/}L
\endprooftree
\justifies
[{\blacksquare}Nt(s(m))], {\square}(({\langle\rangle}{\exists}aNa\backslash Sf)/({\exists}aNa{\oplus}{\it CP}that)), [[{\blacksquare}Nt(s(f)), {\square}{\forall}a{\forall}f(({\langle\rangle}Na\backslash Sf)\backslash ({\langle\rangle}Na\backslash Sf)), \mbox{\fbox{${\forall}a((?{\blacksquare}((({\langle\rangle}Na\backslash Sf)/{\exists}bNb)\backslash ({\langle\rangle}Na\backslash Sf))\backslash {[]^{-1}}{[]^{-1}}((({\langle\rangle}Na\backslash Sf)/{\exists}bNb)\backslash ({\langle\rangle}Na\backslash Sf)))/{\blacksquare}((({\langle\rangle}Na\backslash Sf)/{\exists}bNb)\backslash ({\langle\rangle}Na\backslash Sf)))$}}, {\blacksquare}Nt(s(m)), {\square}{\forall}a{\forall}f(({\langle\rangle}Na\backslash Sf)\backslash ({\langle\rangle}Na\backslash Sf))]]\ \Rightarrow\ Sf
\using {\forall}L
\endprooftree
\justifies
[{\blacksquare}Nt(s(m))], {\square}(({\langle\rangle}{\exists}aNa\backslash Sf)/({\exists}aNa{\oplus}{\it CP}that)), [[{\blacksquare}Nt(s(f)), {\square}{\forall}a{\forall}f(({\langle\rangle}Na\backslash Sf)\backslash ({\langle\rangle}Na\backslash Sf)), \mbox{\fbox{${\forall}f{\forall}a((?{\blacksquare}((({\langle\rangle}Na\backslash Sf)/{\exists}bNb)\backslash ({\langle\rangle}Na\backslash Sf))\backslash {[]^{-1}}{[]^{-1}}((({\langle\rangle}Na\backslash Sf)/{\exists}bNb)\backslash ({\langle\rangle}Na\backslash Sf)))/{\blacksquare}((({\langle\rangle}Na\backslash Sf)/{\exists}bNb)\backslash ({\langle\rangle}Na\backslash Sf)))$}}, {\blacksquare}Nt(s(m)), {\square}{\forall}a{\forall}f(({\langle\rangle}Na\backslash Sf)\backslash ({\langle\rangle}Na\backslash Sf))]]\ \Rightarrow\ Sf
\using {\forall}L
\endprooftree
\justifies
[{\blacksquare}Nt(s(m))], {\square}(({\langle\rangle}{\exists}aNa\backslash Sf)/({\exists}aNa{\oplus}{\it CP}that)), [[{\blacksquare}Nt(s(f)), {\square}{\forall}a{\forall}f(({\langle\rangle}Na\backslash Sf)\backslash ({\langle\rangle}Na\backslash Sf)), \mbox{\fbox{${\blacksquare}{\forall}f{\forall}a((?{\blacksquare}((({\langle\rangle}Na\backslash Sf)/{\exists}bNb)\backslash ({\langle\rangle}Na\backslash Sf))\backslash {[]^{-1}}{[]^{-1}}((({\langle\rangle}Na\backslash Sf)/{\exists}bNb)\backslash ({\langle\rangle}Na\backslash Sf)))/{\blacksquare}((({\langle\rangle}Na\backslash Sf)/{\exists}bNb)\backslash ({\langle\rangle}Na\backslash Sf)))$}}, {\blacksquare}Nt(s(m)), {\square}{\forall}a{\forall}f(({\langle\rangle}Na\backslash Sf)\backslash ({\langle\rangle}Na\backslash Sf))]]\ \Rightarrow\ Sf
\using {\blacksquare}L
\endprooftree}}

\vspace{0.15in}

\noindent
This delivers semantics:
\disp{
$\lambda C[(\mbox{\v{}}{\it man}\ {\it C})\wedge [(\mbox{\v{}}{\it yesterday}\ ({\it Past}\ ((\mbox{\v{}}{\it seee}\ {\it C})\ {\it j})))\wedge (\mbox{\v{}}{\it today}\ ({\it Past}\ ((\mbox{\v{}}{\it meet}\ {\it C})\ {\it j})))]]$}

}

\section{Iterated coordination}

We consider examples of iterated coordination.

\subsection{Iterated coordination of addicity zero}

Minimally we have the example:
\disp{
$[[[{\bf john}]{+}{\bf walks}{+}[{\bf mary}]{+}{\bf talks}{+}{\bf and}{+}[{\bf bill}]{+}{\bf sings}]]: Sf$}
Appropriate lexical lookup yields:
\disp{
$\begin{array}[t]{l}
[[[{\blacksquare}Nt(s(m)): {\it j}], {\square}({\langle\rangle}{\exists}gNt(s(g))\backslash Sf): \mbox{\^{}}\lambda A({\it Pres}\ (\mbox{\v{}}{\it walk}\ {\it A})),\\{} [{\blacksquare}Nt(s(f)): {\it m}], {\square}({\langle\rangle}{\exists}gNt(s(g))\backslash Sf): \mbox{\^{}}\lambda B({\it Pres}\ (\mbox{\v{}}{\it talk}\ {\it B})), \\{\blacksquare}{\forall}f((?{\blacksquare}Sf\backslash {[]^{-1}}{[]^{-1}}Sf)/{\blacksquare}Sf): (\Phinplus\ {\it 0}\ {\it and}), [{\blacksquare}Nt(s(m)): {\it b}],\\ {\square}({\langle\rangle}{\exists}gNt(s(g))\backslash Sf): \mbox{\^{}}\lambda C({\it Pres}\ (\mbox{\v{}}{\it sing}\ {\it C}))]]\ \Rightarrow\ Sf
\end{array}$}
The coordination combinator $(\Phinplus\ 0\ {\it and})$ is such that:
\disp{
$\begin{array}[t]{l}
(((\Phinplus\ 0\ {\it and})\ x)\ [y, z]) =\\
{}[y\wedge(((\Phinplus\ 0\ {\it and})\ x)\ [z])] =\\
{}[y\wedge[z\wedge x]]
\end{array}$}
There is the derivation:

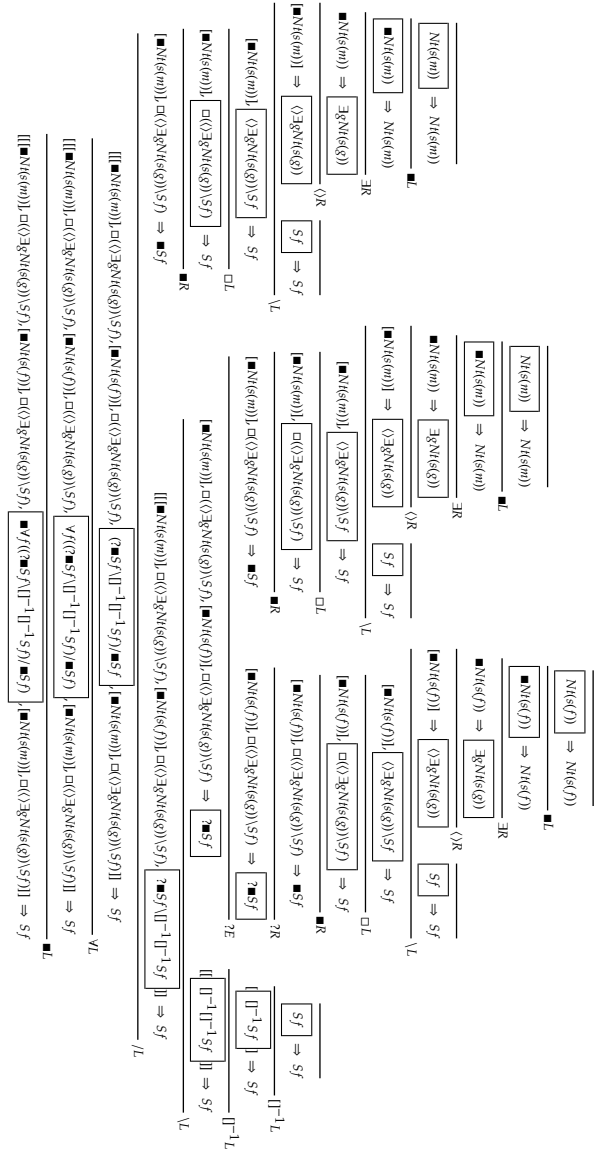
\vspace{0.15in}

\begin{center}
\rotatebox{-90}{\tiny
\prooftree
\prooftree
\prooftree
\prooftree
\prooftree
\prooftree
\prooftree
\prooftree
\prooftree
\prooftree
\justifies
\mbox{\fbox{$Nt(s(m))$}}\ \Rightarrow\ Nt(s(m))
\endprooftree
\justifies
\mbox{\fbox{${\blacksquare}Nt(s(m))$}}\ \Rightarrow\ Nt(s(m))
\using {\blacksquare}L
\endprooftree
\justifies
{\blacksquare}Nt(s(m))\ \Rightarrow\ \fbox{${\exists}gNt(s(g))$}
\using {\exists}R
\endprooftree
\justifies
[{\blacksquare}Nt(s(m))]\ \Rightarrow\ \fbox{${\langle\rangle}{\exists}gNt(s(g))$}
\using {\langle\rangle}R
\endprooftree
\prooftree
\justifies
\mbox{\fbox{$Sf$}}\ \Rightarrow\ Sf
\endprooftree
\justifies
[{\blacksquare}Nt(s(m))], \mbox{\fbox{${\langle\rangle}{\exists}gNt(s(g))\backslash Sf$}}\ \Rightarrow\ Sf
\using {\backslash}L
\endprooftree
\justifies
[{\blacksquare}Nt(s(m))], \mbox{\fbox{${\square}({\langle\rangle}{\exists}gNt(s(g))\backslash Sf)$}}\ \Rightarrow\ Sf
\using {\Box}L
\endprooftree
\justifies
[{\blacksquare}Nt(s(m))], {\square}({\langle\rangle}{\exists}gNt(s(g))\backslash Sf)\ \Rightarrow\ {\blacksquare}Sf
\using {\blacksquare}R
\endprooftree
\prooftree
\prooftree
\prooftree
\prooftree
\prooftree
\prooftree
\prooftree
\prooftree
\prooftree
\justifies
\mbox{\fbox{$Nt(s(m))$}}\ \Rightarrow\ Nt(s(m))
\endprooftree
\justifies
\mbox{\fbox{${\blacksquare}Nt(s(m))$}}\ \Rightarrow\ Nt(s(m))
\using {\blacksquare}L
\endprooftree
\justifies
{\blacksquare}Nt(s(m))\ \Rightarrow\ \fbox{${\exists}gNt(s(g))$}
\using {\exists}R
\endprooftree
\justifies
[{\blacksquare}Nt(s(m))]\ \Rightarrow\ \fbox{${\langle\rangle}{\exists}gNt(s(g))$}
\using {\langle\rangle}R
\endprooftree
\prooftree
\justifies
\mbox{\fbox{$Sf$}}\ \Rightarrow\ Sf
\endprooftree
\justifies
[{\blacksquare}Nt(s(m))], \mbox{\fbox{${\langle\rangle}{\exists}gNt(s(g))\backslash Sf$}}\ \Rightarrow\ Sf
\using {\backslash}L
\endprooftree
\justifies
[{\blacksquare}Nt(s(m))], \mbox{\fbox{${\square}({\langle\rangle}{\exists}gNt(s(g))\backslash Sf)$}}\ \Rightarrow\ Sf
\using {\Box}L
\endprooftree
\justifies
[{\blacksquare}Nt(s(m))], {\square}({\langle\rangle}{\exists}gNt(s(g))\backslash Sf)\ \Rightarrow\ {\blacksquare}Sf
\using {\blacksquare}R
\endprooftree
\prooftree
\prooftree
\prooftree
\prooftree
\prooftree
\prooftree
\prooftree
\prooftree
\justifies
\mbox{\fbox{$Nt(s(f))$}}\ \Rightarrow\ Nt(s(f))
\endprooftree
\justifies
\mbox{\fbox{${\blacksquare}Nt(s(f))$}}\ \Rightarrow\ Nt(s(f))
\using {\blacksquare}L
\endprooftree
\justifies
{\blacksquare}Nt(s(f))\ \Rightarrow\ \fbox{${\exists}gNt(s(g))$}
\using {\exists}R
\endprooftree
\justifies
[{\blacksquare}Nt(s(f))]\ \Rightarrow\ \fbox{${\langle\rangle}{\exists}gNt(s(g))$}
\using {\langle\rangle}R
\endprooftree
\prooftree
\justifies
\mbox{\fbox{$Sf$}}\ \Rightarrow\ Sf
\endprooftree
\justifies
[{\blacksquare}Nt(s(f))], \mbox{\fbox{${\langle\rangle}{\exists}gNt(s(g))\backslash Sf$}}\ \Rightarrow\ Sf
\using {\backslash}L
\endprooftree
\justifies
[{\blacksquare}Nt(s(f))], \mbox{\fbox{${\square}({\langle\rangle}{\exists}gNt(s(g))\backslash Sf)$}}\ \Rightarrow\ Sf
\using {\Box}L
\endprooftree
\justifies
[{\blacksquare}Nt(s(f))], {\square}({\langle\rangle}{\exists}gNt(s(g))\backslash Sf)\ \Rightarrow\ {\blacksquare}Sf
\using {\blacksquare}R
\endprooftree
\justifies
[{\blacksquare}Nt(s(f))], {\square}({\langle\rangle}{\exists}gNt(s(g))\backslash Sf)\ \Rightarrow\ \fbox{$?{\blacksquare}Sf$}
\using {?}R
\endprooftree
\justifies
[{\blacksquare}Nt(s(m))], {\square}({\langle\rangle}{\exists}gNt(s(g))\backslash Sf), [{\blacksquare}Nt(s(f))], {\square}({\langle\rangle}{\exists}gNt(s(g))\backslash Sf)\ \Rightarrow\ \fbox{$?{\blacksquare}Sf$}
\using {?}E
\endprooftree
\prooftree
\prooftree
\prooftree
\justifies
\mbox{\fbox{$Sf$}}\ \Rightarrow\ Sf
\endprooftree
\justifies
[\mbox{\fbox{${[]^{-1}}Sf$}}]\ \Rightarrow\ Sf
\using {[]^{-1}}L
\endprooftree
\justifies
[[\mbox{\fbox{${[]^{-1}}{[]^{-1}}Sf$}}]]\ \Rightarrow\ Sf
\using {[]^{-1}}L
\endprooftree
\justifies
[[[{\blacksquare}Nt(s(m))], {\square}({\langle\rangle}{\exists}gNt(s(g))\backslash Sf), [{\blacksquare}Nt(s(f))], {\square}({\langle\rangle}{\exists}gNt(s(g))\backslash Sf), \mbox{\fbox{$?{\blacksquare}Sf\backslash {[]^{-1}}{[]^{-1}}Sf$}}]]\ \Rightarrow\ Sf
\using {\backslash}L
\endprooftree
\justifies
[[[{\blacksquare}Nt(s(m))], {\square}({\langle\rangle}{\exists}gNt(s(g))\backslash Sf), [{\blacksquare}Nt(s(f))], {\square}({\langle\rangle}{\exists}gNt(s(g))\backslash Sf), \mbox{\fbox{$(?{\blacksquare}Sf\backslash {[]^{-1}}{[]^{-1}}Sf)/{\blacksquare}Sf$}}, [{\blacksquare}Nt(s(m))], {\square}({\langle\rangle}{\exists}gNt(s(g))\backslash Sf)]]\ \Rightarrow\ Sf
\using {/}L
\endprooftree
\justifies
[[[{\blacksquare}Nt(s(m))], {\square}({\langle\rangle}{\exists}gNt(s(g))\backslash Sf), [{\blacksquare}Nt(s(f))], {\square}({\langle\rangle}{\exists}gNt(s(g))\backslash Sf), \mbox{\fbox{${\forall}f((?{\blacksquare}Sf\backslash {[]^{-1}}{[]^{-1}}Sf)/{\blacksquare}Sf)$}}, [{\blacksquare}Nt(s(m))], {\square}({\langle\rangle}{\exists}gNt(s(g))\backslash Sf)]]\ \Rightarrow\ Sf
\using {\forall}L
\endprooftree
\justifies
[[[{\blacksquare}Nt(s(m))], {\square}({\langle\rangle}{\exists}gNt(s(g))\backslash Sf), [{\blacksquare}Nt(s(f))], {\square}({\langle\rangle}{\exists}gNt(s(g))\backslash Sf), \mbox{\fbox{${\blacksquare}{\forall}f((?{\blacksquare}Sf\backslash {[]^{-1}}{[]^{-1}}Sf)/{\blacksquare}Sf)$}}, [{\blacksquare}Nt(s(m))], {\square}({\langle\rangle}{\exists}gNt(s(g))\backslash Sf)]]\ \Rightarrow\ Sf
\using {\blacksquare}L
\endprooftree}
\end{center}

\vspace{0.15in}

\noindent
This delivers semantics:
\disp{
$[({\it Pres}\ (\mbox{\v{}}{\it walk}\ {\it j}))\wedge [({\it Pres}\ (\mbox{\v{}}{\it talk}\ {\it m}))\wedge ({\it Pres}\ (\mbox{\v{}}{\it sing}\ {\it b}))]]$}

\subsection{Iterated coordination of addicity one}

There is the example of verb phrase iterated coordination:
\disp{
$[{\bf john}]{+}[[{\bf walks}{+}{\bf talks}{+}{\bf and}{+}{\bf sings}]]: Sf$}
Appropriate lexical insertion yields:
\disp{
$\begin{array}[t]{l}
[{\blacksquare}Nt(s(m)): {\it j}], [[{\square}({\langle\rangle}{\exists}gNt(s(g))\backslash Sf): \mbox{\^{}}\lambda A({\it Pres}\ (\mbox{\v{}}{\it walk}\ {\it A})),\\ {\square}({\langle\rangle}{\exists}gNt(s(g))\backslash Sf): \mbox{\^{}}\lambda B({\it Pres}\ (\mbox{\v{}}{\it talk}\ {\it B})),\\
 {\blacksquare}{\forall}a{\forall}f((?{\blacksquare}({\langle\rangle}Na\backslash Sf)\backslash {[]^{-1}}{[]^{-1}}({\langle\rangle}Na\backslash Sf))/{\blacksquare}({\langle\rangle}Na\backslash Sf)): (\Phinplus\ ({\it s}\ {\it 0})\ {\it and}),\\ {\square}({\langle\rangle}{\exists}gNt(s(g))\backslash Sf): \mbox{\^{}}\lambda C({\it Pres}\ (\mbox{\v{}}{\it sing}\ {\it C}))]]\ \Rightarrow\ Sf
 \end{array}$}
The coordinator lexical semantics $(\Phinplus\ (s\ 0)\ {\it and})$ is such that:
\disp{
$\begin{array}[t]{l}
((((\Phinplus\ (s\ 0)\ {\it and})\ x)\ [y, z])\ w) =\\
(((\Phinplus\ 0\ {\it and})\ (x\ w))\ (\alphaplus\ [y, z]\ w)) =\\
(((\Phinplus\ 0\ {\it and})\ (x\ w))\ [(y\ w)|(\alphaplus\ [z]\ w)]) =\\
(((\Phinplus\ 0\ {\it and})\ (x\ w))\ [(y\ w), (z\ w)]) =\\
{}[(y\ w)\wedge(((\Phinplus\ 0\ {\it and})\ (x\ w))\ [(z\ w)])] =\\
{}[(y\ w)\wedge[(z\ w)\wedge(x\ w)]]
\end{array}$}
There is the derivation:

\vspace{0.15in}

\begin{center}
\rotatebox{-90}{\tiny
\prooftree
\prooftree
\prooftree
\prooftree
\prooftree
\prooftree
\prooftree
\prooftree
\prooftree
\prooftree
\prooftree
\prooftree
\justifies
Nt(s(m))\ \Rightarrow\ Nt(s(m))
\endprooftree
\justifies
Nt(s(m))\ \Rightarrow\ \fbox{${\exists}gNt(s(g))$}
\using {\exists}R
\endprooftree
\justifies
[Nt(s(m))]\ \Rightarrow\ \fbox{${\langle\rangle}{\exists}gNt(s(g))$}
\using {\langle\rangle}R
\endprooftree
\prooftree
\justifies
\mbox{\fbox{$Sf$}}\ \Rightarrow\ Sf
\endprooftree
\justifies
[Nt(s(m))], \mbox{\fbox{${\langle\rangle}{\exists}gNt(s(g))\backslash Sf$}}\ \Rightarrow\ Sf
\using {\backslash}L
\endprooftree
\justifies
[Nt(s(m))], \mbox{\fbox{${\square}({\langle\rangle}{\exists}gNt(s(g))\backslash Sf)$}}\ \Rightarrow\ Sf
\using {\Box}L
\endprooftree
\justifies
{\langle\rangle}Nt(s(m)), {\square}({\langle\rangle}{\exists}gNt(s(g))\backslash Sf)\ \Rightarrow\ Sf
\using {\langle\rangle}L
\endprooftree
\justifies
{\square}({\langle\rangle}{\exists}gNt(s(g))\backslash Sf)\ \Rightarrow\ {\langle\rangle}Nt(s(m))\backslash Sf
\using {\backslash}R
\endprooftree
\justifies
{\square}({\langle\rangle}{\exists}gNt(s(g))\backslash Sf)\ \Rightarrow\ {\blacksquare}({\langle\rangle}Nt(s(m))\backslash Sf)
\using {\blacksquare}R
\endprooftree
\prooftree
\prooftree
\prooftree
\prooftree
\prooftree
\prooftree
\prooftree
\prooftree
\prooftree
\prooftree
\justifies
Nt(s(m))\ \Rightarrow\ Nt(s(m))
\endprooftree
\justifies
Nt(s(m))\ \Rightarrow\ \fbox{${\exists}gNt(s(g))$}
\using {\exists}R
\endprooftree
\justifies
[Nt(s(m))]\ \Rightarrow\ \fbox{${\langle\rangle}{\exists}gNt(s(g))$}
\using {\langle\rangle}R
\endprooftree
\prooftree
\justifies
\mbox{\fbox{$Sf$}}\ \Rightarrow\ Sf
\endprooftree
\justifies
[Nt(s(m))], \mbox{\fbox{${\langle\rangle}{\exists}gNt(s(g))\backslash Sf$}}\ \Rightarrow\ Sf
\using {\backslash}L
\endprooftree
\justifies
[Nt(s(m))], \mbox{\fbox{${\square}({\langle\rangle}{\exists}gNt(s(g))\backslash Sf)$}}\ \Rightarrow\ Sf
\using {\Box}L
\endprooftree
\justifies
{\langle\rangle}Nt(s(m)), {\square}({\langle\rangle}{\exists}gNt(s(g))\backslash Sf)\ \Rightarrow\ Sf
\using {\langle\rangle}L
\endprooftree
\justifies
{\square}({\langle\rangle}{\exists}gNt(s(g))\backslash Sf)\ \Rightarrow\ {\langle\rangle}Nt(s(m))\backslash Sf
\using {\backslash}R
\endprooftree
\justifies
{\square}({\langle\rangle}{\exists}gNt(s(g))\backslash Sf)\ \Rightarrow\ {\blacksquare}({\langle\rangle}Nt(s(m))\backslash Sf)
\using {\blacksquare}R
\endprooftree
\prooftree
\prooftree
\prooftree
\prooftree
\prooftree
\prooftree
\prooftree
\prooftree
\prooftree
\justifies
Nt(s(m))\ \Rightarrow\ Nt(s(m))
\endprooftree
\justifies
Nt(s(m))\ \Rightarrow\ \fbox{${\exists}gNt(s(g))$}
\using {\exists}R
\endprooftree
\justifies
[Nt(s(m))]\ \Rightarrow\ \fbox{${\langle\rangle}{\exists}gNt(s(g))$}
\using {\langle\rangle}R
\endprooftree
\prooftree
\justifies
\mbox{\fbox{$Sf$}}\ \Rightarrow\ Sf
\endprooftree
\justifies
[Nt(s(m))], \mbox{\fbox{${\langle\rangle}{\exists}gNt(s(g))\backslash Sf$}}\ \Rightarrow\ Sf
\using {\backslash}L
\endprooftree
\justifies
[Nt(s(m))], \mbox{\fbox{${\square}({\langle\rangle}{\exists}gNt(s(g))\backslash Sf)$}}\ \Rightarrow\ Sf
\using {\Box}L
\endprooftree
\justifies
{\langle\rangle}Nt(s(m)), {\square}({\langle\rangle}{\exists}gNt(s(g))\backslash Sf)\ \Rightarrow\ Sf
\using {\langle\rangle}L
\endprooftree
\justifies
{\square}({\langle\rangle}{\exists}gNt(s(g))\backslash Sf)\ \Rightarrow\ {\langle\rangle}Nt(s(m))\backslash Sf
\using {\backslash}R
\endprooftree
\justifies
{\square}({\langle\rangle}{\exists}gNt(s(g))\backslash Sf)\ \Rightarrow\ {\blacksquare}({\langle\rangle}Nt(s(m))\backslash Sf)
\using {\blacksquare}R
\endprooftree
\justifies
{\square}({\langle\rangle}{\exists}gNt(s(g))\backslash Sf)\ \Rightarrow\ \fbox{$?{\blacksquare}({\langle\rangle}Nt(s(m))\backslash Sf)$}
\using {?}R
\endprooftree
\justifies
{\square}({\langle\rangle}{\exists}gNt(s(g))\backslash Sf), {\square}({\langle\rangle}{\exists}gNt(s(g))\backslash Sf)\ \Rightarrow\ \fbox{$?{\blacksquare}({\langle\rangle}Nt(s(m))\backslash Sf)$}
\using {?}E
\endprooftree
\prooftree
\prooftree
\prooftree
\prooftree
\prooftree
\prooftree
\justifies
\mbox{\fbox{$Nt(s(m))$}}\ \Rightarrow\ Nt(s(m))
\endprooftree
\justifies
\mbox{\fbox{${\blacksquare}Nt(s(m))$}}\ \Rightarrow\ Nt(s(m))
\using {\blacksquare}L
\endprooftree
\justifies
[{\blacksquare}Nt(s(m))]\ \Rightarrow\ \fbox{${\langle\rangle}Nt(s(m))$}
\using {\langle\rangle}R
\endprooftree
\prooftree
\justifies
\mbox{\fbox{$Sf$}}\ \Rightarrow\ Sf
\endprooftree
\justifies
[{\blacksquare}Nt(s(m))], \mbox{\fbox{${\langle\rangle}Nt(s(m))\backslash Sf$}}\ \Rightarrow\ Sf
\using {\backslash}L
\endprooftree
\justifies
[{\blacksquare}Nt(s(m))], [\mbox{\fbox{${[]^{-1}}({\langle\rangle}Nt(s(m))\backslash Sf)$}}]\ \Rightarrow\ Sf
\using {[]^{-1}}L
\endprooftree
\justifies
[{\blacksquare}Nt(s(m))], [[\mbox{\fbox{${[]^{-1}}{[]^{-1}}({\langle\rangle}Nt(s(m))\backslash Sf)$}}]]\ \Rightarrow\ Sf
\using {[]^{-1}}L
\endprooftree
\justifies
[{\blacksquare}Nt(s(m))], [[{\square}({\langle\rangle}{\exists}gNt(s(g))\backslash Sf), {\square}({\langle\rangle}{\exists}gNt(s(g))\backslash Sf), \mbox{\fbox{$?{\blacksquare}({\langle\rangle}Nt(s(m))\backslash Sf)\backslash {[]^{-1}}{[]^{-1}}({\langle\rangle}Nt(s(m))\backslash Sf)$}}]]\ \Rightarrow\ Sf
\using {\backslash}L
\endprooftree
\justifies
[{\blacksquare}Nt(s(m))], [[{\square}({\langle\rangle}{\exists}gNt(s(g))\backslash Sf), {\square}({\langle\rangle}{\exists}gNt(s(g))\backslash Sf), \mbox{\fbox{$(?{\blacksquare}({\langle\rangle}Nt(s(m))\backslash Sf)\backslash {[]^{-1}}{[]^{-1}}({\langle\rangle}Nt(s(m))\backslash Sf))/{\blacksquare}({\langle\rangle}Nt(s(m))\backslash Sf)$}}, {\square}({\langle\rangle}{\exists}gNt(s(g))\backslash Sf)]]\ \Rightarrow\ Sf
\using {/}L
\endprooftree
\justifies
[{\blacksquare}Nt(s(m))], [[{\square}({\langle\rangle}{\exists}gNt(s(g))\backslash Sf), {\square}({\langle\rangle}{\exists}gNt(s(g))\backslash Sf), \mbox{\fbox{${\forall}f((?{\blacksquare}({\langle\rangle}Nt(s(m))\backslash Sf)\backslash {[]^{-1}}{[]^{-1}}({\langle\rangle}Nt(s(m))\backslash Sf))/{\blacksquare}({\langle\rangle}Nt(s(m))\backslash Sf))$}}, {\square}({\langle\rangle}{\exists}gNt(s(g))\backslash Sf)]]\ \Rightarrow\ Sf
\using {\forall}L
\endprooftree
\justifies
[{\blacksquare}Nt(s(m))], [[{\square}({\langle\rangle}{\exists}gNt(s(g))\backslash Sf), {\square}({\langle\rangle}{\exists}gNt(s(g))\backslash Sf), \mbox{\fbox{${\forall}a{\forall}f((?{\blacksquare}({\langle\rangle}Na\backslash Sf)\backslash {[]^{-1}}{[]^{-1}}({\langle\rangle}Na\backslash Sf))/{\blacksquare}({\langle\rangle}Na\backslash Sf))$}}, {\square}({\langle\rangle}{\exists}gNt(s(g))\backslash Sf)]]\ \Rightarrow\ Sf
\using {\forall}L
\endprooftree
\justifies
[{\blacksquare}Nt(s(m))], [[{\square}({\langle\rangle}{\exists}gNt(s(g))\backslash Sf), {\square}({\langle\rangle}{\exists}gNt(s(g))\backslash Sf), \mbox{\fbox{${\blacksquare}{\forall}a{\forall}f((?{\blacksquare}({\langle\rangle}Na\backslash Sf)\backslash {[]^{-1}}{[]^{-1}}({\langle\rangle}Na\backslash Sf))/{\blacksquare}({\langle\rangle}Na\backslash Sf))$}}, {\square}({\langle\rangle}{\exists}gNt(s(g))\backslash Sf)]]\ \Rightarrow\ Sf
\using {\blacksquare}L
\endprooftree}
\end{center}

\vspace{0.15in}

\noindent
This delivers semantics:
\disp{
$[({\it Pres}\ (\mbox{\v{}}{\it walk}\ {\it j}))\wedge [({\it Pres}\ (\mbox{\v{}}{\it talk}\ {\it j}))\wedge ({\it Pres}\ (\mbox{\v{}}{\it sing}\ {\it j}))]]$}

\subsection{Iterated coordination of addicity two}

There is the example of transitive verb phrase iterated coordination:
\disp{
$[{\bf john}]{+}[[{\bf praises}{+}{\bf likes}{+}{\bf and}{+}{\bf will}{+}{\bf love}]]{+}{\bf london}: Sf$}
Lexical insertion yields:
\disp{
$\begin{array}[t]{l}
[{\blacksquare}Nt(s(m)): {\it j}], [[{\square}(({\langle\rangle}{\exists}gNt(s(g))\backslash Sf)/{\exists}aNa): \mbox{\^{}}\lambda A\lambda B({\it Pres}\ ((\mbox{\v{}}{\it praise}\ {\it A})\ {\it B})),\\
 {\square}(({\langle\rangle}{\exists}gNt(s(g))\backslash Sf)/{\exists}aNa): \mbox{\^{}}\lambda C\lambda D({\it Pres}\ ((\mbox{\v{}}{\it like}\ {\it C})\ {\it D})), \\{\blacksquare}{\forall}f{\forall}a((?{\blacksquare}(({\langle\rangle}Na\backslash Sf)/{\exists}bNb)\backslash {[]^{-1}}{[]^{-1}}(({\langle\rangle}Na\backslash Sf)/{\exists}bNb))/{\blacksquare}(({\langle\rangle}Na\backslash Sf)/{\exists}bNb)):\\ (\Phinplus\ ({\it s}\ ({\it s}\ {\it 0}))\ {\it and}), {\blacksquare}{\forall}a(({\langle\rangle}Na\backslash Sf)/({\langle\rangle}Na\backslash Sb)): \lambda E\lambda F({\it Fut}\ ({\it E}\ {\it F})),\\ {\square}(({\langle\rangle}{\exists}aNa\backslash Sb)/{\exists}aNa): \mbox{\^{}}\lambda G\lambda H((\mbox{\v{}}{\it love}\ {\it G})\ {\it H})]],
 {\blacksquare}Nt(s(n)): {\it l}\ \Rightarrow\ Sf
 \end{array}$}
The coordination combinator semantics is such that:
\disp{
$\begin{array}[t]{l}
(((((\Phinplus\ (s\ (s\ 0))\ {\it and})\ x)\ [y, z])\ w)\ u) =\\
((((\Phinplus\ (s\ 0)\ {\it and})\ (x\ w))\ (\alphaplus\ [y, z]\ w))\ u) =\\
((((\Phinplus\ (s\ 0)\ {\it and})\ (x\ w))\ [(y\ w), (z\ w)])\ u) =\\
(((\Phinplus\ 0\ {\it and})\ ((x\ w)\ u))\ (\alphaplus\ [(y\ w), (z\ w)]\ u)) =\\
(((\Phinplus\ 0\ {\it and})\ ((x\ w)\ u))\ [((y\ w)\ u), ((z\ w)\ u)]) =\\
{}[((y\ w)\ u)\wedge[((z\ w)\ u)\wedge((x\ w)\ u)]]
\end{array}$}
There is the derivation:

\vspace{0.15in}

\begin{center}
{\scriptsize
\prooftree
\prooftree
\prooftree
\prooftree
\prooftree
\prooftree
\prooftree
\prooftree
\prooftree
\prooftree
\prooftree
\prooftree
\prooftree
\prooftree
\justifies
N1\ \Rightarrow\ N1
\endprooftree
\justifies
N1\ \Rightarrow\ \fbox{${\exists}aNa$}
\using {\exists}R
\endprooftree
\prooftree
\prooftree
\prooftree
\prooftree
\justifies
Nt(s(m))\ \Rightarrow\ Nt(s(m))
\endprooftree
\justifies
Nt(s(m))\ \Rightarrow\ \fbox{${\exists}aNa$}
\using {\exists}R
\endprooftree
\justifies
[Nt(s(m))]\ \Rightarrow\ \fbox{${\langle\rangle}{\exists}aNa$}
\using {\langle\rangle}R
\endprooftree
\prooftree
\justifies
\mbox{\fbox{$Sb$}}\ \Rightarrow\ Sb
\endprooftree
\justifies
[Nt(s(m))], \mbox{\fbox{${\langle\rangle}{\exists}aNa\backslash Sb$}}\ \Rightarrow\ Sb
\using {\backslash}L
\endprooftree
\justifies
[Nt(s(m))], \mbox{\fbox{$({\langle\rangle}{\exists}aNa\backslash Sb)/{\exists}aNa$}}, N1\ \Rightarrow\ Sb
\using {/}L
\endprooftree
\justifies
[Nt(s(m))], \mbox{\fbox{${\square}(({\langle\rangle}{\exists}aNa\backslash Sb)/{\exists}aNa)$}}, N1\ \Rightarrow\ Sb
\using {\Box}L
\endprooftree
\justifies
{\langle\rangle}Nt(s(m)), {\square}(({\langle\rangle}{\exists}aNa\backslash Sb)/{\exists}aNa), N1\ \Rightarrow\ Sb
\using {\langle\rangle}L
\endprooftree
\justifies
{\square}(({\langle\rangle}{\exists}aNa\backslash Sb)/{\exists}aNa), N1\ \Rightarrow\ {\langle\rangle}Nt(s(m))\backslash Sb
\using {\backslash}R
\endprooftree
\prooftree
\prooftree
\prooftree
\justifies
Nt(s(m))\ \Rightarrow\ Nt(s(m))
\endprooftree
\justifies
[Nt(s(m))]\ \Rightarrow\ \fbox{${\langle\rangle}Nt(s(m))$}
\using {\langle\rangle}R
\endprooftree
\prooftree
\justifies
\mbox{\fbox{$Sf$}}\ \Rightarrow\ Sf
\endprooftree
\justifies
[Nt(s(m))], \mbox{\fbox{${\langle\rangle}Nt(s(m))\backslash Sf$}}\ \Rightarrow\ Sf
\using {\backslash}L
\endprooftree
\justifies
[Nt(s(m))], \mbox{\fbox{$({\langle\rangle}Nt(s(m))\backslash Sf)/({\langle\rangle}Nt(s(m))\backslash Sb)$}}, {\square}(({\langle\rangle}{\exists}aNa\backslash Sb)/{\exists}aNa), N1\ \Rightarrow\ Sf
\using {/}L
\endprooftree
\justifies
[Nt(s(m))], \mbox{\fbox{${\forall}a(({\langle\rangle}Na\backslash Sf)/({\langle\rangle}Na\backslash Sb))$}}, {\square}(({\langle\rangle}{\exists}aNa\backslash Sb)/{\exists}aNa), N1\ \Rightarrow\ Sf
\using {\forall}L
\endprooftree
\justifies
[Nt(s(m))], \mbox{\fbox{${\blacksquare}{\forall}a(({\langle\rangle}Na\backslash Sf)/({\langle\rangle}Na\backslash Sb))$}}, {\square}(({\langle\rangle}{\exists}aNa\backslash Sb)/{\exists}aNa), N1\ \Rightarrow\ Sf
\using {\blacksquare}L
\endprooftree
\justifies
[Nt(s(m))], {\blacksquare}{\forall}a(({\langle\rangle}Na\backslash Sf)/({\langle\rangle}Na\backslash Sb)), {\square}(({\langle\rangle}{\exists}aNa\backslash Sb)/{\exists}aNa), {\exists}bNb\ \Rightarrow\ Sf
\using {\exists}L
\endprooftree
\justifies
{\langle\rangle}Nt(s(m)), {\blacksquare}{\forall}a(({\langle\rangle}Na\backslash Sf)/({\langle\rangle}Na\backslash Sb)), {\square}(({\langle\rangle}{\exists}aNa\backslash Sb)/{\exists}aNa), {\exists}bNb\ \Rightarrow\ Sf
\using {\langle\rangle}L
\endprooftree
\justifies
{\blacksquare}{\forall}a(({\langle\rangle}Na\backslash Sf)/({\langle\rangle}Na\backslash Sb)), {\square}(({\langle\rangle}{\exists}aNa\backslash Sb)/{\exists}aNa), {\exists}bNb\ \Rightarrow\ {\langle\rangle}Nt(s(m))\backslash Sf
\using {\backslash}R
\endprooftree
\justifies
{\blacksquare}{\forall}a(({\langle\rangle}Na\backslash Sf)/({\langle\rangle}Na\backslash Sb)), {\square}(({\langle\rangle}{\exists}aNa\backslash Sb)/{\exists}aNa)\ \Rightarrow\ ({\langle\rangle}Nt(s(m))\backslash Sf)/{\exists}bNb
\using {/}R
\endprooftree
\justifies
\begin{array}{c}
{\blacksquare}{\forall}a(({\langle\rangle}Na\backslash Sf)/({\langle\rangle}Na\backslash Sb)), {\square}(({\langle\rangle}{\exists}aNa\backslash Sb)/{\exists}aNa)\ \Rightarrow\ {\blacksquare}(({\langle\rangle}Nt(s(m))\backslash Sf)/{\exists}bNb)\\
\mbox{\footnotesize\textcircled{1}}
\end{array}
\using {\blacksquare}R
\endprooftree}
\end{center}

\begin{center}
\rotatebox{-90}{\scriptsize
\prooftree
\prooftree
\prooftree
\prooftree
\prooftree
\prooftree
\prooftree
\prooftree
\prooftree
\prooftree
\justifies
N1\ \Rightarrow\ N1
\endprooftree
\justifies
N1\ \Rightarrow\ \fbox{${\exists}aNa$}
\using {\exists}R
\endprooftree
\prooftree
\prooftree
\prooftree
\prooftree
\justifies
Nt(s(m))\ \Rightarrow\ Nt(s(m))
\endprooftree
\justifies
Nt(s(m))\ \Rightarrow\ \fbox{${\exists}gNt(s(g))$}
\using {\exists}R
\endprooftree
\justifies
[Nt(s(m))]\ \Rightarrow\ \fbox{${\langle\rangle}{\exists}gNt(s(g))$}
\using {\langle\rangle}R
\endprooftree
\prooftree
\justifies
\mbox{\fbox{$Sf$}}\ \Rightarrow\ Sf
\endprooftree
\justifies
[Nt(s(m))], \mbox{\fbox{${\langle\rangle}{\exists}gNt(s(g))\backslash Sf$}}\ \Rightarrow\ Sf
\using {\backslash}L
\endprooftree
\justifies
[Nt(s(m))], \mbox{\fbox{$({\langle\rangle}{\exists}gNt(s(g))\backslash Sf)/{\exists}aNa$}}, N1\ \Rightarrow\ Sf
\using {/}L
\endprooftree
\justifies
[Nt(s(m))], \mbox{\fbox{${\square}(({\langle\rangle}{\exists}gNt(s(g))\backslash Sf)/{\exists}aNa)$}}, N1\ \Rightarrow\ Sf
\using {\Box}L
\endprooftree
\justifies
[Nt(s(m))], {\square}(({\langle\rangle}{\exists}gNt(s(g))\backslash Sf)/{\exists}aNa), {\exists}bNb\ \Rightarrow\ Sf
\using {\exists}L
\endprooftree
\justifies
{\langle\rangle}Nt(s(m)), {\square}(({\langle\rangle}{\exists}gNt(s(g))\backslash Sf)/{\exists}aNa), {\exists}bNb\ \Rightarrow\ Sf
\using {\langle\rangle}L
\endprooftree
\justifies
{\square}(({\langle\rangle}{\exists}gNt(s(g))\backslash Sf)/{\exists}aNa), {\exists}bNb\ \Rightarrow\ {\langle\rangle}Nt(s(m))\backslash Sf
\using {\backslash}R
\endprooftree
\justifies
{\square}(({\langle\rangle}{\exists}gNt(s(g))\backslash Sf)/{\exists}aNa)\ \Rightarrow\ ({\langle\rangle}Nt(s(m))\backslash Sf)/{\exists}bNb
\using {/}R
\endprooftree
\justifies
{\square}(({\langle\rangle}{\exists}gNt(s(g))\backslash Sf)/{\exists}aNa)\ \Rightarrow\ {\blacksquare}(({\langle\rangle}Nt(s(m))\backslash Sf)/{\exists}bNb)
\using {\blacksquare}R
\endprooftree
\prooftree
\prooftree
\prooftree
\prooftree
\prooftree
\prooftree
\prooftree
\prooftree
\prooftree
\prooftree
\justifies
N2\ \Rightarrow\ N2
\endprooftree
\justifies
N2\ \Rightarrow\ \fbox{${\exists}aNa$}
\using {\exists}R
\endprooftree
\prooftree
\prooftree
\prooftree
\prooftree
\justifies
Nt(s(m))\ \Rightarrow\ Nt(s(m))
\endprooftree
\justifies
Nt(s(m))\ \Rightarrow\ \fbox{${\exists}gNt(s(g))$}
\using {\exists}R
\endprooftree
\justifies
[Nt(s(m))]\ \Rightarrow\ \fbox{${\langle\rangle}{\exists}gNt(s(g))$}
\using {\langle\rangle}R
\endprooftree
\prooftree
\justifies
\mbox{\fbox{$Sf$}}\ \Rightarrow\ Sf
\endprooftree
\justifies
[Nt(s(m))], \mbox{\fbox{${\langle\rangle}{\exists}gNt(s(g))\backslash Sf$}}\ \Rightarrow\ Sf
\using {\backslash}L
\endprooftree
\justifies
[Nt(s(m))], \mbox{\fbox{$({\langle\rangle}{\exists}gNt(s(g))\backslash Sf)/{\exists}aNa$}}, N2\ \Rightarrow\ Sf
\using {/}L
\endprooftree
\justifies
[Nt(s(m))], \mbox{\fbox{${\square}(({\langle\rangle}{\exists}gNt(s(g))\backslash Sf)/{\exists}aNa)$}}, N2\ \Rightarrow\ Sf
\using {\Box}L
\endprooftree
\justifies
[Nt(s(m))], {\square}(({\langle\rangle}{\exists}gNt(s(g))\backslash Sf)/{\exists}aNa), {\exists}bNb\ \Rightarrow\ Sf
\using {\exists}L
\endprooftree
\justifies
{\langle\rangle}Nt(s(m)), {\square}(({\langle\rangle}{\exists}gNt(s(g))\backslash Sf)/{\exists}aNa), {\exists}bNb\ \Rightarrow\ Sf
\using {\langle\rangle}L
\endprooftree
\justifies
{\square}(({\langle\rangle}{\exists}gNt(s(g))\backslash Sf)/{\exists}aNa), {\exists}bNb\ \Rightarrow\ {\langle\rangle}Nt(s(m))\backslash Sf
\using {\backslash}R
\endprooftree
\justifies
{\square}(({\langle\rangle}{\exists}gNt(s(g))\backslash Sf)/{\exists}aNa)\ \Rightarrow\ ({\langle\rangle}Nt(s(m))\backslash Sf)/{\exists}bNb
\using {/}R
\endprooftree
\justifies
{\square}(({\langle\rangle}{\exists}gNt(s(g))\backslash Sf)/{\exists}aNa)\ \Rightarrow\ {\blacksquare}(({\langle\rangle}Nt(s(m))\backslash Sf)/{\exists}bNb)
\using {\blacksquare}R
\endprooftree
\justifies
{\square}(({\langle\rangle}{\exists}gNt(s(g))\backslash Sf)/{\exists}aNa)\ \Rightarrow\ \fbox{$?{\blacksquare}(({\langle\rangle}Nt(s(m))\backslash Sf)/{\exists}bNb)$}
\using {?}R
\endprooftree
\justifies
\begin{array}{c}
{\square}(({\langle\rangle}{\exists}gNt(s(g))\backslash Sf)/{\exists}aNa), {\square}(({\langle\rangle}{\exists}gNt(s(g))\backslash Sf)/{\exists}aNa)\ \Rightarrow\ \fbox{$?{\blacksquare}(({\langle\rangle}Nt(s(m))\backslash Sf)/{\exists}bNb)$}\\
\mbox{\footnotesize\textcircled{2}}
\end{array}
\using {?}E
\endprooftree}
\end{center}

\begin{center}
\rotatebox{-90}{\tiny
\prooftree
\prooftree
\prooftree
\prooftree
\mbox{\footnotesize\textcircled{1}}\tab
\prooftree
\mbox{\footnotesize\textcircled{2}}\tab
\prooftree
\prooftree
\prooftree
\prooftree
\prooftree
\prooftree
\justifies
\mbox{\fbox{$Nt(s(n))$}}\ \Rightarrow\ Nt(s(n))
\endprooftree
\justifies
\mbox{\fbox{${\blacksquare}Nt(s(n))$}}\ \Rightarrow\ Nt(s(n))
\using {\blacksquare}L
\endprooftree
\justifies
{\blacksquare}Nt(s(n))\ \Rightarrow\ \fbox{${\exists}bNb$}
\using {\exists}R
\endprooftree
\prooftree
\prooftree
\prooftree
\prooftree
\justifies
\mbox{\fbox{$Nt(s(m))$}}\ \Rightarrow\ Nt(s(m))
\endprooftree
\justifies
\mbox{\fbox{${\blacksquare}Nt(s(m))$}}\ \Rightarrow\ Nt(s(m))
\using {\blacksquare}L
\endprooftree
\justifies
[{\blacksquare}Nt(s(m))]\ \Rightarrow\ \fbox{${\langle\rangle}Nt(s(m))$}
\using {\langle\rangle}R
\endprooftree
\prooftree
\justifies
\mbox{\fbox{$Sf$}}\ \Rightarrow\ Sf
\endprooftree
\justifies
[{\blacksquare}Nt(s(m))], \mbox{\fbox{${\langle\rangle}Nt(s(m))\backslash Sf$}}\ \Rightarrow\ Sf
\using {\backslash}L
\endprooftree
\justifies
[{\blacksquare}Nt(s(m))], \mbox{\fbox{$({\langle\rangle}Nt(s(m))\backslash Sf)/{\exists}bNb$}}, {\blacksquare}Nt(s(n))\ \Rightarrow\ Sf
\using {/}L
\endprooftree
\justifies
[{\blacksquare}Nt(s(m))], [\mbox{\fbox{${[]^{-1}}(({\langle\rangle}Nt(s(m))\backslash Sf)/{\exists}bNb)$}}], {\blacksquare}Nt(s(n))\ \Rightarrow\ Sf
\using {[]^{-1}}L
\endprooftree
\justifies
[{\blacksquare}Nt(s(m))], [[\mbox{\fbox{${[]^{-1}}{[]^{-1}}(({\langle\rangle}Nt(s(m))\backslash Sf)/{\exists}bNb)$}}]], {\blacksquare}Nt(s(n))\ \Rightarrow\ Sf
\using {[]^{-1}}L
\endprooftree
\justifies
[{\blacksquare}Nt(s(m))], [[{\square}(({\langle\rangle}{\exists}gNt(s(g))\backslash Sf)/{\exists}aNa), {\square}(({\langle\rangle}{\exists}gNt(s(g))\backslash Sf)/{\exists}aNa), \mbox{\fbox{$?{\blacksquare}(({\langle\rangle}Nt(s(m))\backslash Sf)/{\exists}bNb)\backslash {[]^{-1}}{[]^{-1}}(({\langle\rangle}Nt(s(m))\backslash Sf)/{\exists}bNb)$}}]], {\blacksquare}Nt(s(n))\ \Rightarrow\ Sf
\using {\backslash}L
\endprooftree
\justifies
[{\blacksquare}Nt(s(m))], [[{\square}(({\langle\rangle}{\exists}gNt(s(g))\backslash Sf)/{\exists}aNa), {\square}(({\langle\rangle}{\exists}gNt(s(g))\backslash Sf)/{\exists}aNa), \mbox{\fbox{$(?{\blacksquare}(({\langle\rangle}Nt(s(m))\backslash Sf)/{\exists}bNb)\backslash {[]^{-1}}{[]^{-1}}(({\langle\rangle}Nt(s(m))\backslash Sf)/{\exists}bNb))/{\blacksquare}(({\langle\rangle}Nt(s(m))\backslash Sf)/{\exists}bNb)$}}, {\blacksquare}{\forall}a(({\langle\rangle}Na\backslash Sf)/({\langle\rangle}Na\backslash Sb)), {\square}(({\langle\rangle}{\exists}aNa\backslash Sb)/{\exists}aNa)]], {\blacksquare}Nt(s(n))\ \Rightarrow\ Sf
\using {/}L
\endprooftree
\justifies
[{\blacksquare}Nt(s(m))], [[{\square}(({\langle\rangle}{\exists}gNt(s(g))\backslash Sf)/{\exists}aNa), {\square}(({\langle\rangle}{\exists}gNt(s(g))\backslash Sf)/{\exists}aNa), \mbox{\fbox{${\forall}a((?{\blacksquare}(({\langle\rangle}Na\backslash Sf)/{\exists}bNb)\backslash {[]^{-1}}{[]^{-1}}(({\langle\rangle}Na\backslash Sf)/{\exists}bNb))/{\blacksquare}(({\langle\rangle}Na\backslash Sf)/{\exists}bNb))$}}, {\blacksquare}{\forall}a(({\langle\rangle}Na\backslash Sf)/({\langle\rangle}Na\backslash Sb)), {\square}(({\langle\rangle}{\exists}aNa\backslash Sb)/{\exists}aNa)]], {\blacksquare}Nt(s(n))\ \Rightarrow\ Sf
\using {\forall}L
\endprooftree
\justifies
[{\blacksquare}Nt(s(m))], [[{\square}(({\langle\rangle}{\exists}gNt(s(g))\backslash Sf)/{\exists}aNa), {\square}(({\langle\rangle}{\exists}gNt(s(g))\backslash Sf)/{\exists}aNa), \mbox{\fbox{${\forall}f{\forall}a((?{\blacksquare}(({\langle\rangle}Na\backslash Sf)/{\exists}bNb)\backslash {[]^{-1}}{[]^{-1}}(({\langle\rangle}Na\backslash Sf)/{\exists}bNb))/{\blacksquare}(({\langle\rangle}Na\backslash Sf)/{\exists}bNb))$}}, {\blacksquare}{\forall}a(({\langle\rangle}Na\backslash Sf)/({\langle\rangle}Na\backslash Sb)), {\square}(({\langle\rangle}{\exists}aNa\backslash Sb)/{\exists}aNa)]], {\blacksquare}Nt(s(n))\ \Rightarrow\ Sf
\using {\forall}L
\endprooftree
\justifies
[{\blacksquare}Nt(s(m))], [[{\square}(({\langle\rangle}{\exists}gNt(s(g))\backslash Sf)/{\exists}aNa), {\square}(({\langle\rangle}{\exists}gNt(s(g))\backslash Sf)/{\exists}aNa), \mbox{\fbox{${\blacksquare}{\forall}f{\forall}a((?{\blacksquare}(({\langle\rangle}Na\backslash Sf)/{\exists}bNb)\backslash {[]^{-1}}{[]^{-1}}(({\langle\rangle}Na\backslash Sf)/{\exists}bNb))/{\blacksquare}(({\langle\rangle}Na\backslash Sf)/{\exists}bNb))$}}, {\blacksquare}{\forall}a(({\langle\rangle}Na\backslash Sf)/({\langle\rangle}Na\backslash Sb)), {\square}(({\langle\rangle}{\exists}aNa\backslash Sb)/{\exists}aNa)]], {\blacksquare}Nt(s(n))\ \Rightarrow\ Sf
\using {\blacksquare}L
\endprooftree}
\end{center}

\vspace{0.15in}

\noindent
All this assigns the correct semantics:
\disp{
$[({\it Pres}\ ((\mbox{\v{}}{\it praise}\ {\it l})\ {\it j}))\wedge [({\it Pres}\ ((\mbox{\v{}}{\it like}\ {\it l})\ {\it j}))\wedge ({\it Fut}\ ((\mbox{\v{}}{\it love}\ {\it l})\ {\it j}))]]$}

\section{Coordination of unlike types}

\label{unlikesect}

In the following we have coordinate unlike types with nominal and
adjectival complementation of \emph{is}.
\disp{
$[{\bf bond}]{+}{\bf is}{+}[[{\bf 007}{+}{\bf and}{+}{\bf teetotal}]]: Sf$}
Together with a suitable coordinator type, a polymorphic assignment to the copula
of the form $(N\bsl S)/(N\adisj$ $(\CN/\CN))$ predicts such coordination under a \emph{like\/}
type scheme (Morrill 1990\cite{morrill:galt}; Johnson and Bayer 1995\cite{johnsonbayer:95};
Bayer 1996\cite{bayer:96}).
Lexical lookup of types yields the following, where the coordinator is of the form
$(X\bsl \abrack\abrack X)/X$ where $X=(N\bsl S)/(N\adisj((\CN/\CN)\iadisj(\CN\bsl\CN)))$.
\disp{
$\begin{array}[t]{l}
[{\blacksquare}Nt(s(m)): {\it b}], {\blacksquare}(({\langle\rangle}{\exists}gNt(s(g))\backslash Sf)/({\exists}aNa{\oplus}({\exists}g(({\it CN}{\it g}/{\it CN}{\it g}){\sqcup}({\it CN}{\it g}\backslash {\it CN}{\it g})){-}I))):\\ \lambda A\lambda B({\it Pres}\ ({\it A}\casearrow C.[{\it B}={\it C}]; D.(({\it D}\ \lambda E[{\it E}={\it B}])\ {\it B}))), [[{\blacksquare}{\forall}gNt(s(g)): {\it 007},\\ {\blacksquare}{\forall}f{\forall}a(({\blacksquare}((({\langle\rangle}Na\backslash Sf)/({\exists}bNb{\oplus}{\exists}g(({\it CN}{\it g}/{\it CN}{\it g}){\sqcup}({\it CN}{\it g}\backslash {\it CN}{\it g}))))\backslash ({\langle\rangle}Na\backslash Sf))\backslash\\ {[]^{-1}}{[]^{-1}}((({\langle\rangle}Na\backslash Sf)/({\exists}bNb{\oplus}{\exists}g(({\it CN}{\it g}/{\it CN}{\it g}){\sqcup}({\it CN}{\it g}\backslash {\it CN}{\it g}))))\backslash ({\langle\rangle}Na\backslash Sf)))/\\{\blacksquare}((({\langle\rangle}Na\backslash Sf)/({\exists}bNb{\oplus}{\exists}g(({\it CN}{\it g}/{\it CN}{\it g}){\sqcup}({\it CN}{\it g}\backslash {\it CN}{\it g}))))\backslash ({\langle\rangle}Na\backslash Sf))): \\\lambda F\lambda G\lambda H\lambda I[(({\it G}\ {\it H})\ {\it I})\wedge (({\it F}\ {\it H})\ {\it I})], {\square}{\forall}n({\it CN}{\it n}/{\it CN}{\it n}): \mbox{\^{}}\lambda J\lambda K[({\it J}\ {\it K})\wedge (\mbox{\v{}}{\it teetotal}\ {\it K})]]]\\ \Rightarrow~Sf
\end{array}$}
The derivation is as shown below.

\begin{center}\scriptsize
\prooftree
\prooftree
\prooftree
\prooftree
\prooftree
\prooftree
\prooftree
\prooftree
\prooftree
\prooftree
\prooftree
\prooftree
\justifies
{\it CN}{\it A}\ \Rightarrow\ {\it CN}{\it A}
\endprooftree
\prooftree
\justifies
\mbox{\fbox{${\it CN}{\it A}$}}\ \Rightarrow\ {\it CN}{\it A}
\endprooftree
\justifies
\mbox{\fbox{${\it CN}{\it A}/{\it CN}{\it A}$}}, {\it CN}{\it A}\ \Rightarrow\ {\it CN}{\it A}
\using {/}L
\endprooftree
\justifies
\mbox{\fbox{${\forall}n({\it CN}{\it n}/{\it CN}{\it n})$}}, {\it CN}{\it A}\ \Rightarrow\ {\it CN}{\it A}
\using {\forall}L
\endprooftree
\justifies
\mbox{\fbox{${\square}{\forall}n({\it CN}{\it n}/{\it CN}{\it n})$}}, {\it CN}{\it A}\ \Rightarrow\ {\it CN}{\it A}
\using {\Box}L
\endprooftree
\justifies
{\square}{\forall}n({\it CN}{\it n}/{\it CN}{\it n})\ \Rightarrow\ \fbox{${\it CN}{\it A}/{\it CN}{\it A}$}
\using {\sqcup}R
\endprooftree
\justifies
{\square}{\forall}n({\it CN}{\it n}/{\it CN}{\it n})\ \Rightarrow\ \fbox{${\exists}g({\it CN}{\it g}/{\it CN}{\it g})$}
\using {\exists}R
\endprooftree
\justifies
{\square}{\forall}n({\it CN}{\it n}/{\it CN}{\it n})\ \Rightarrow\ \fbox{${\exists}bNb{\oplus}{\exists}g({\it CN}{\it g}/{\it CN}{\it g})$}
\using {\oplus}R
\endprooftree
\prooftree
\prooftree
\prooftree
\justifies
Nt(s(m))\ \Rightarrow\ Nt(s(m))
\endprooftree
\justifies
[Nt(s(m))]\ \Rightarrow\ \fbox{${\langle\rangle}Nt(s(m))$}
\using {\langle\rangle}R
\endprooftree
\prooftree
\justifies
\mbox{\fbox{$Sf$}}\ \Rightarrow\ Sf
\endprooftree
\justifies
[Nt(s(m))], \mbox{\fbox{${\langle\rangle}Nt(s(m))\backslash Sf$}}\ \Rightarrow\ Sf
\using {\backslash}L
\endprooftree
\justifies
[Nt(s(m))], \mbox{\fbox{$({\langle\rangle}Nt(s(m))\backslash Sf)/({\exists}bNb{\oplus}{\exists}g({\it CN}{\it g}/{\it CN}{\it g}))$}}, {\square}{\forall}n({\it CN}{\it n}/{\it CN}{\it n})\ \Rightarrow\ Sf
\using {/}L
\endprooftree
\justifies
{\langle\rangle}Nt(s(m)), ({\langle\rangle}Nt(s(m))\backslash Sf)/({\exists}bNb{\oplus}{\exists}g({\it CN}{\it g}/{\it CN}{\it g})), {\square}{\forall}n({\it CN}{\it n}/{\it CN}{\it n})\ \Rightarrow\ Sf
\using {\langle\rangle}L
\endprooftree
\justifies
({\langle\rangle}Nt(s(m))\backslash Sf)/({\exists}bNb{\oplus}{\exists}g({\it CN}{\it g}/{\it CN}{\it g})), {\square}{\forall}n({\it CN}{\it n}/{\it CN}{\it n})\ \Rightarrow\ {\langle\rangle}Nt(s(m))\backslash Sf
\using {\backslash}R
\endprooftree
\justifies
{\square}{\forall}n({\it CN}{\it n}/{\it CN}{\it n})\ \Rightarrow\ (({\langle\rangle}Nt(s(m))\backslash Sf)/({\exists}bNb{\oplus}{\exists}g({\it CN}{\it g}/{\it CN}{\it g})))\backslash ({\langle\rangle}Nt(s(m))\backslash Sf)
\using {\backslash}R
\endprooftree
\justifies
\begin{array}{c}
{\square}{\forall}n({\it CN}{\it n}/{\it CN}{\it n})\ \Rightarrow\ {\blacksquare}((({\langle\rangle}Nt(s(m))\backslash Sf)/({\exists}bNb{\oplus}{\exists}g({\it CN}{\it g}/{\it CN}{\it g})))\backslash ({\langle\rangle}Nt(s(m))\backslash Sf))\\
\mbox{\footnotesize\textcircled{1}}
\end{array}
\using {\blacksquare}R
\endprooftree

\prooftree
\prooftree
\prooftree
\prooftree
\prooftree
\prooftree
\prooftree
\prooftree
\prooftree
\prooftree
\justifies
\mbox{\fbox{$Nt(s(A))$}}\ \Rightarrow\ Nt(s(A))
\endprooftree
\justifies
\mbox{\fbox{${\forall}gNt(s(g))$}}\ \Rightarrow\ Nt(s(A))
\using {\forall}L
\endprooftree
\justifies
\mbox{\fbox{${\blacksquare}{\forall}gNt(s(g))$}}\ \Rightarrow\ Nt(s(A))
\using {\blacksquare}L
\endprooftree
\justifies
{\blacksquare}{\forall}gNt(s(g))\ \Rightarrow\ \fbox{${\exists}bNb$}
\using {\exists}R
\endprooftree
\justifies
{\blacksquare}{\forall}gNt(s(g))\ \Rightarrow\ \fbox{${\exists}bNb{\oplus}{\exists}g({\it CN}{\it g}/{\it CN}{\it g})$}
\using {\oplus}R
\endprooftree
\prooftree
\prooftree
\prooftree
\justifies
Nt(s(m))\ \Rightarrow\ Nt(s(m))
\endprooftree
\justifies
[Nt(s(m))]\ \Rightarrow\ \fbox{${\langle\rangle}Nt(s(m))$}
\using {\langle\rangle}R
\endprooftree
\prooftree
\justifies
\mbox{\fbox{$Sf$}}\ \Rightarrow\ Sf
\endprooftree
\justifies
[Nt(s(m))], \mbox{\fbox{${\langle\rangle}Nt(s(m))\backslash Sf$}}\ \Rightarrow\ Sf
\using {\backslash}L
\endprooftree
\justifies
[Nt(s(m))], \mbox{\fbox{$({\langle\rangle}Nt(s(m))\backslash Sf)/({\exists}bNb{\oplus}{\exists}g({\it CN}{\it g}/{\it CN}{\it g}))$}}, {\blacksquare}{\forall}gNt(s(g))\ \Rightarrow\ Sf
\using {/}L
\endprooftree
\justifies
{\langle\rangle}Nt(s(m)), ({\langle\rangle}Nt(s(m))\backslash Sf)/({\exists}bNb{\oplus}{\exists}g({\it CN}{\it g}/{\it CN}{\it g})), {\blacksquare}{\forall}gNt(s(g))\ \Rightarrow\ Sf
\using {\langle\rangle}L
\endprooftree
\justifies
({\langle\rangle}Nt(s(m))\backslash Sf)/({\exists}bNb{\oplus}{\exists}g({\it CN}{\it g}/{\it CN}{\it g})), {\blacksquare}{\forall}gNt(s(g))\ \Rightarrow\ {\langle\rangle}Nt(s(m))\backslash Sf
\using {\backslash}R
\endprooftree
\justifies
{\blacksquare}{\forall}gNt(s(g))\ \Rightarrow\ (({\langle\rangle}Nt(s(m))\backslash Sf)/({\exists}bNb{\oplus}{\exists}g({\it CN}{\it g}/{\it CN}{\it g})))\backslash ({\langle\rangle}Nt(s(m))\backslash Sf)
\using {\backslash}R
\endprooftree
\justifies
\begin{array}{c}
{\blacksquare}{\forall}gNt(s(g))\ \Rightarrow\ {\blacksquare}((({\langle\rangle}Nt(s(m))\backslash Sf)/({\exists}bNb{\oplus}{\exists}g({\it CN}{\it g}/{\it CN}{\it g})))\backslash ({\langle\rangle}Nt(s(m))\backslash Sf))\\
\mbox{\footnotesize\textcircled{2}}
\end{array}
\using {\blacksquare}R
\endprooftree

\prooftree
\prooftree
\prooftree
\prooftree
\prooftree
\prooftree
\prooftree
\justifies
N1\ \Rightarrow\ N1
\endprooftree
\justifies
N1\ \Rightarrow\ \fbox{${\exists}aNa$}
\using {\exists}R
\endprooftree
\justifies
N1\ \Rightarrow\ \fbox{${\exists}aNa{\oplus}{\exists}g({\it CN}{\it g}/{\it CN}{\it g})$}
\using {\oplus}R
\endprooftree
\prooftree
\prooftree
\prooftree
\prooftree
\justifies
Nt(s(m))\ \Rightarrow\ Nt(s(m))
\endprooftree
\justifies
Nt(s(m))\ \Rightarrow\ \fbox{${\exists}gNt(s(g))$}
\using {\exists}R
\endprooftree
\justifies
[Nt(s(m))]\ \Rightarrow\ \fbox{${\langle\rangle}{\exists}gNt(s(g))$}
\using {\langle\rangle}R
\endprooftree
\prooftree
\justifies
\mbox{\fbox{$Sf$}}\ \Rightarrow\ Sf
\endprooftree
\justifies
[Nt(s(m))], \mbox{\fbox{${\langle\rangle}{\exists}gNt(s(g))\backslash Sf$}}\ \Rightarrow\ Sf
\using {\backslash}L
\endprooftree
\justifies
[Nt(s(m))], \mbox{\fbox{$({\langle\rangle}{\exists}gNt(s(g))\backslash Sf)/({\exists}aNa{\oplus}{\exists}g({\it CN}{\it g}/{\it CN}{\it g})$}}, N1\ \Rightarrow\ Sf
\using {/}L
\endprooftree
\justifies
[Nt(s(m))], \mbox{\fbox{${\blacksquare}(({\langle\rangle}{\exists}gNt(s(g))\backslash Sf)/({\exists}aNa{\oplus}{\exists}g({\it CN}{\it g}/{\it CN}{\it g})))$}}, N1\ \Rightarrow\ Sf
\using {\blacksquare}L
\endprooftree
\justifies
[Nt(s(m))], {\blacksquare}(({\langle\rangle}{\exists}gNt(s(g))\backslash Sf)/({\exists}aNa{\oplus}{\exists}g({\it CN}{\it g}/{\it CN}{\it g}))), {\exists}bNb\ \Rightarrow\ Sf
\using {\exists}L
\endprooftree
\justifies
\begin{array}{c}
{\langle\rangle}Nt(s(m)), {\blacksquare}(({\langle\rangle}{\exists}gNt(s(g))\backslash Sf)/({\exists}aNa{\oplus}{\exists}g({\it CN}{\it g}/{\it CN}{\it g}))), {\exists}bNb\ \Rightarrow\ Sf\\
\mbox{\footnotesize\textcircled{3}}
\end{array}
\using {\langle\rangle}L
\endprooftree

\rotatebox{-0}{\tiny
\prooftree
\prooftree
\prooftree
\prooftree
\prooftree
\prooftree
\prooftree
\vdots
\justifies
{\it CN}{\it 5581}/{\it CN}{\it 5581}\ \Rightarrow\ \fbox{${\exists}g(({\it CN}{\it g}/{\it CN}{\it g}){\sqcup}({\it CN}{\it g}\backslash {\it CN}{\it g}))$}
\using {\exists}R
\endprooftree
\justifies
{\it CN}{\it 5581}/{\it CN}{\it 5581}\ \Rightarrow\ \fbox{${\exists}g(({\it CN}{\it g}/{\it CN}{\it g}){\sqcup}({\it CN}{\it g}\backslash {\it CN}{\it g}))-I$}
\using {-}R
\endprooftree
\justifies
{\it CN}{\it 5581}/{\it CN}{\it 5581}\ \Rightarrow\ \fbox{${\exists}aNa{\oplus}({\exists}g(({\it CN}{\it g}/{\it CN}{\it g}){\sqcup}({\it CN}{\it g}\backslash {\it CN}{\it g})){-}I)$}
\using {\oplus}R
\endprooftree
\prooftree
\prooftree
\prooftree
\prooftree
\justifies
Nt(s(m))\ \Rightarrow\ Nt(s(m))
\endprooftree
\justifies
Nt(s(m))\ \Rightarrow\ \fbox{${\exists}gNt(s(g))$}
\using {\exists}R
\endprooftree
\justifies
[Nt(s(m))]\ \Rightarrow\ \fbox{${\langle\rangle}{\exists}gNt(s(g))$}
\using {\langle\rangle}R
\endprooftree
\prooftree
\justifies
\mbox{\fbox{$Sf$}}\ \Rightarrow\ Sf
\endprooftree
\justifies
[Nt(s(m))], \mbox{\fbox{${\langle\rangle}{\exists}gNt(s(g))\backslash Sf$}}\ \Rightarrow\ Sf
\using {\backslash}L
\endprooftree
\justifies
[Nt(s(m))], \mbox{\fbox{$({\langle\rangle}{\exists}gNt(s(g))\backslash Sf)/({\exists}aNa{\oplus}({\exists}g(({\it CN}{\it g}/{\it CN}{\it g}){\sqcup}({\it CN}{\it g}\backslash {\it CN}{\it g})){-}I))$}}, {\it CN}{\it 5581}/{\it CN}{\it 5581}\ \Rightarrow\ Sf
\using {/}L
\endprooftree
\justifies
[Nt(s(m))], \mbox{\fbox{${\blacksquare}(({\langle\rangle}{\exists}gNt(s(g))\backslash Sf)/({\exists}aNa{\oplus}{\exists}g(({\it CN}{\it g}/{\it CN}{\it g})))$}}, {\it CN}{\it 5581}/{\it CN}{\it 5581}\ \Rightarrow\ Sf
\using {\blacksquare}L
\endprooftree
\justifies
[Nt(s(m))], {\blacksquare}(({\langle\rangle}{\exists}gNt(s(g))\backslash Sf)/({\exists}aNa{\oplus}{\exists}g({\it CN}{\it g}/{\it CN}{\it g}))), {\exists}g({\it CN}{\it g}/{\it CN}{\it g})\ \Rightarrow\ Sf
\using {\exists}L
\endprooftree
\justifies
\begin{array}{c}
{\langle\rangle}Nt(s(m)), {\blacksquare}(({\langle\rangle}{\exists}gNt(s(g))\backslash Sf)/({\exists}aNa{\oplus}{\exists}g({\it CN}{\it g}/{\it CN}{\it g}))), {\exists}g({\it CN}{\it g}/{\it CN}{\it g})\ \Rightarrow\ Sf\\
\mbox{\footnotesize\textcircled{4}}
\end{array}
\using {\langle\rangle}L
\endprooftree}
\end{center}
\begin{center}
\rotatebox{-90}{\tiny
\prooftree
\prooftree
\prooftree
\prooftree
\mbox{\footnotesize\textcircled{1}}\tab\tab
\prooftree
\mbox{\footnotesize\textcircled{2}}\tab
\prooftree
\prooftree
\prooftree
\prooftree
\prooftree
\prooftree
\mbox{\footnotesize\textcircled{3}}\tab\tab\tab
\mbox{\footnotesize\textcircled{4}}
\justifies
{\langle\rangle}Nt(s(m)), {\blacksquare}(({\langle\rangle}{\exists}gNt(s(g))\backslash Sf)/({\exists}aNa{\oplus}{\exists}g({\it CN}{\it g}/{\it CN}{\it g}))), {\exists}bNb{\oplus}{\exists}g({\it CN}{\it g}/{\it CN}{\it g})\ \Rightarrow\ Sf
\using {\oplus}L
\endprooftree
\justifies
{\blacksquare}(({\langle\rangle}{\exists}gNt(s(g))\backslash Sf)/({\exists}aNa{\oplus}{\exists}g({\it CN}{\it g}/{\it CN}{\it g}))), {\exists}bNb{\oplus}{\exists}g({\it CN}{\it g}/{\it CN}{\it g})\ \Rightarrow\ {\langle\rangle}Nt(s(m))\backslash Sf
\using {\backslash}R
\endprooftree
\justifies
{\blacksquare}(({\langle\rangle}{\exists}gNt(s(g))\backslash Sf)/({\exists}aNa{\oplus}{\exists}g({\it CN}{\it g}/{\it CN}{\it g}))\ \Rightarrow\ ({\langle\rangle}Nt(s(m))\backslash Sf)/({\exists}bNb{\oplus}{\exists}g({\it CN}{\it g}/{\it CN}{\it g}))
\using {/}R
\endprooftree
\prooftree
\prooftree
\prooftree
\prooftree
\justifies
\mbox{\fbox{$Nt(s(m))$}}\ \Rightarrow\ Nt(s(m))
\endprooftree
\justifies
\mbox{\fbox{${\blacksquare}Nt(s(m))$}}\ \Rightarrow\ Nt(s(m))
\using {\blacksquare}L
\endprooftree
\justifies
[{\blacksquare}Nt(s(m))]\ \Rightarrow\ \fbox{${\langle\rangle}Nt(s(m))$}
\using {\langle\rangle}R
\endprooftree
\prooftree
\justifies
\mbox{\fbox{$Sf$}}\ \Rightarrow\ Sf
\endprooftree
\justifies
[{\blacksquare}Nt(s(m))], \mbox{\fbox{${\langle\rangle}Nt(s(m))\backslash Sf$}}\ \Rightarrow\ Sf
\using {\backslash}L
\endprooftree
\justifies
[{\blacksquare}Nt(s(m))], {\blacksquare}(({\langle\rangle}{\exists}gNt(s(g))\backslash Sf)/({\exists}aNa{\oplus}{\exists}g({\it CN}{\it g}/{\it CN}{\it g}))), \mbox{\fbox{$(({\langle\rangle}Nt(s(m))\backslash Sf)/({\exists}bNb{\oplus}{\exists}g({\it CN}{\it g}/{\it CN}{\it g})))\backslash ({\langle\rangle}Nt(s(m))\backslash Sf)$}}\ \Rightarrow\ Sf
\using {\backslash}L
\endprooftree
\justifies
[{\blacksquare}Nt(s(m))], {\blacksquare}(({\langle\rangle}{\exists}gNt(s(g))\backslash Sf)/({\exists}aNa{\oplus}{\exists}g({\it CN}{\it g}/{\it CN}{\it g}))), [\mbox{\fbox{${[]^{-1}}((({\langle\rangle}Nt(s(m))\backslash Sf)/({\exists}bNb{\oplus}{\exists}g({\it CN}{\it g}/{\it CN}{\it g}))))\backslash ({\langle\rangle}Nt(s(m))\backslash Sf))$}}]\ \Rightarrow\ Sf
\using {[]^{-1}}L
\endprooftree
\justifies
[{\blacksquare}Nt(s(m))], {\blacksquare}(({\langle\rangle}{\exists}gNt(s(g))\backslash Sf)/({\exists}aNa{\oplus}{\exists}g({\it CN}{\it g}/{\it CN}{\it g}))), [[\mbox{\fbox{${[]^{-1}}{[]^{-1}}((({\langle\rangle}Nt(s(m))\backslash Sf)/({\exists}bNb{\oplus}{\exists}g({\it CN}{\it g}/{\it CN}{\it g}))))\backslash ({\langle\rangle}Nt(s(m))\backslash Sf))$}}]]\ \Rightarrow\ Sf
\using {[]^{-1}}L
\endprooftree
\justifies
\begin{array}{c}
{}[{\blacksquare}Nt(s(m))], {\blacksquare}(({\langle\rangle}{\exists}gNt(s(g))\backslash Sf)/({\exists}aNa{\oplus}{\exists}g({\it CN}{\it g}/{\it CN}{\it g}))), [[{\blacksquare}{\forall}gNt(s(g)),\\
\mbox{\fbox{${\blacksquare}((({\langle\rangle}Nt(s(m))\backslash Sf)/({\exists}bNb{\oplus}{\exists}g({\it CN}{\it g}/{\it CN}{\it g})))\backslash ({\langle\rangle}Nt(s(m))\backslash Sf))\backslash {[]^{-1}}{[]^{-1}}((({\langle\rangle}Nt(s(m))\backslash Sf)/({\exists}bNb{\oplus}{\exists}g({\it CN}{\it g}/{\it CN}{\it g})))\backslash ({\langle\rangle}Nt(s(m))\backslash Sf))$}}]]\ \Rightarrow\ Sf
\end{array}
\using {\backslash}L
\endprooftree
\justifies
\begin{array}{c}
{}[{\blacksquare}Nt(s(m))], {\blacksquare}(({\langle\rangle}{\exists}gNt(s(g))\backslash Sf)/({\exists}aNa{\oplus}{\exists}g({\it CN}{\it g}/{\it CN}{\it g}))), [[{\blacksquare}{\forall}gNt(s(g)),\\
\mbox{\fbox{$({\blacksquare}((({\langle\rangle}Nt(s(m))\backslash Sf)/({\exists}bNb{\oplus}{\exists}g({\it CN}{\it g}/{\it CN}{\it g})))\backslash ({\langle\rangle}Nt(s(m))\backslash Sf))\backslash {[]^{-1}}{[]^{-1}}((({\langle\rangle}Nt(s(m))\backslash Sf)/({\exists}bNb{\oplus}{\exists}g({\it CN}{\it g}/{\it CN}{\it g})))\backslash ({\langle\rangle}Nt(s(m))\backslash Sf)))/{\blacksquare}((({\langle\rangle}Nt(s(m))\backslash Sf)/({\exists}bNb{\oplus}{\exists}g({\it CN}{\it g}/{\it CN}{\it g})))\backslash ({\langle\rangle}Nt(s(m))\backslash Sf))$}},\\
{\square}{\forall}n({\it CN}{\it n}/{\it CN}{\it n})]]\ \Rightarrow\ Sf
\end{array}
\using {/}L
\endprooftree
\justifies
\begin{array}{c}
{}[{\blacksquare}Nt(s(m))], {\blacksquare}(({\langle\rangle}{\exists}gNt(s(g))\backslash Sf)/({\exists}aNa{\oplus}{\exists}g({\it CN}{\it g}/{\it CN}{\it g}))), [[{\blacksquare}{\forall}gNt(s(g)),\\
\mbox{\fbox{${\forall}a(({\blacksquare}((({\langle\rangle}Na\backslash Sf)/({\exists}bNb{\oplus}{\exists}g({\it CN}{\it g}/{\it CN}{\it g})))\backslash ({\langle\rangle}Na\backslash Sf))\backslash {[]^{-1}}{[]^{-1}}((({\langle\rangle}Na\backslash Sf)/({\exists}bNb{\oplus}{\exists}g({\it CN}{\it g}/{\it CN}{\it g})))\backslash ({\langle\rangle}Na\backslash Sf)))/{\blacksquare}((({\langle\rangle}Na\backslash Sf)/({\exists}bNb{\oplus}{\exists}g({\it CN}{\it g}/{\it CN}{\it g})))\backslash ({\langle\rangle}Na\backslash Sf)))$}},\\
{\square}{\forall}n({\it CN}{\it n}/{\it CN}{\it n})]]\ \Rightarrow\ Sf
\end{array}
\using {\forall}L
\endprooftree
\justifies
\begin{array}{c}
{}[{\blacksquare}Nt(s(m))], {\blacksquare}(({\langle\rangle}{\exists}gNt(s(g))\backslash Sf)/({\exists}aNa{\oplus}{\exists}g({\it CN}{\it g}/{\it CN}{\it g}))), [[{\blacksquare}{\forall}gNt(s(g)),\\
\mbox{\fbox{${\forall}f{\forall}a(({\blacksquare}((({\langle\rangle}Na\backslash Sf)/({\exists}bNb{\oplus}{\exists}g({\it CN}{\it g}/{\it CN}{\it g})))\backslash ({\langle\rangle}Na\backslash Sf))\backslash {[]^{-1}}{[]^{-1}}((({\langle\rangle}Na\backslash Sf)/({\exists}bNb{\oplus}{\exists}g({\it CN}{\it g}/{\it CN}{\it g}))))\backslash ({\langle\rangle}Na\backslash Sf)))/{\blacksquare}((({\langle\rangle}Na\backslash Sf)/({\exists}bNb{\oplus}{\exists}g({\it CN}{\it g}/{\it CN}{\it g})))\backslash ({\langle\rangle}Na\backslash Sf)))$}},\\
{\square}{\forall}n({\it CN}{\it n}/{\it CN}{\it n})]]\ \Rightarrow\ Sf
\end{array}
\using {\forall}L
\endprooftree
\justifies
\begin{array}{c}
{}[{\blacksquare}Nt(s(m))], {\blacksquare}(({\langle\rangle}{\exists}gNt(s(g))\backslash Sf)/({\exists}aNa{\oplus}{\exists}g({\it CN}{\it g}/{\it CN}{\it g}))), [[{\blacksquare}{\forall}gNt(s(g)),\\
\mbox{\fbox{${\blacksquare}{\forall}f{\forall}a(({\blacksquare}((({\langle\rangle}Na\backslash Sf)/({\exists}bNb{\oplus}{\exists}g({\it CN}{\it g}/{\it CN}{\it g})))\backslash ({\langle\rangle}Na\backslash Sf))\backslash {[]^{-1}}{[]^{-1}}((({\langle\rangle}Na\backslash Sf)/({\exists}bNb{\oplus}{\exists}g({\it CN}{\it g}/{\it CN}{\it g})))\backslash ({\langle\rangle}Na\backslash Sf)))/{\blacksquare}((({\langle\rangle}Na\backslash Sf)/({\exists}bNb{\oplus}{\exists}g({\it CN}{\it g}/{\it CN}{\it g})))\backslash ({\langle\rangle}Na\backslash Sf)))$}},\\
{\square}{\forall}n({\it CN}{\it n}/{\it CN}{\it n})]]\ \Rightarrow\ Sf
\end{array}
\using {\blacksquare}L
\endprooftree}
\end{center}


\noindent
This yields semantics:
\disp{
$[({\it Pres}\ [{\it b}={\it 007}])\wedge ({\it Pres}\ (\mbox{\v{}}{\it teetotal}\ {\it b}))]$.}
The same account can be given for other verbs, for example our polymorphic
type for \lingform{saw} will allow under a suitable coordinator type
\scare{unlike} coordination such as \lingform{John saw the facts and that}
\lingform{Mary had been right}.

\chapter{The 
examples of Morrill, Valent\'{\i}n \& Fadda (2011)}

\label{mvfchap}

In this chapter we analyse the displacement examples of the article
Morrill, Valent\'{\i}n and Fadda (2011\cite{mvf:tdc}) 
presenting the displacement calculus.

\section{English examples}

The first example, 
however,
is modified in view of Morrill and Valent\'{\i}n (2014\cite{mv:words}).
It is a discontinuous idiom.
(We include the indexation of CatLog,
which contains the numeration of the source,
within the example displays.)
\
\disp{
(tdc(43)) $[{\bf mary}]{+}{\bf gave}{+}{\bf the}{+}{\bf man}{+}{\bf the}{+}{\bf cold}{+}{\bf shoulder}: Sf$}
Lexical lookup yields:
\disp{
$[{\blacksquare}Nt(s(f)): {\it m}], {\square}(({\langle\rangle}{\exists}gNt(s(g))\backslash Sf)/({\exists}aNa{\RIGHTcircle}W[the,cold,shoulder])):\\\mbox{\^{}}\lambda A\lambda B({\it Past}\ ((\mbox{\v{}}{\it shun}\ {\it A})\ {\it B})),$ $ {\blacksquare}{\forall}n(Nt(n)/{\it CN}{\it n}): \iota , {\square}{\it CN}{\it s(m)}: {\it man}, {\blacksquare}W[the,cold,shoulder]: {\it 0}\\\Rightarrow\ Sf$}
There is the derivation:
\vspace{0.15in}
$$
\tiny
\prooftree
\prooftree
\prooftree
\prooftree
\prooftree
\prooftree
\prooftree
\prooftree
\prooftree
\justifies
\mbox{\fbox{${\it CN}{\it s(m)}$}}\ \Rightarrow\ {\it CN}{\it s(m)}
\endprooftree
\justifies
\mbox{\fbox{${\square}{\it CN}{\it s(m)}$}}\ \Rightarrow\ {\it CN}{\it s(m)}
\using {\Box}L
\endprooftree
\prooftree
\justifies
\mbox{\fbox{$Nt(s(m))$}}\ \Rightarrow\ Nt(s(m))
\endprooftree
\justifies
\mbox{\fbox{$Nt(s(m))/{\it CN}{\it s(m)}$}}, {\square}{\it CN}{\it s(m)}\ \Rightarrow\ Nt(s(m))
\using {/}L
\endprooftree
\justifies
\mbox{\fbox{${\forall}n(Nt(n)/{\it CN}{\it n})$}}, {\square}{\it CN}{\it s(m)}\ \Rightarrow\ Nt(s(m))
\using {\forall}L
\endprooftree
\justifies
\mbox{\fbox{${\blacksquare}{\forall}n(Nt(n)/{\it CN}{\it n})$}}, {\square}{\it CN}{\it s(m)}\ \Rightarrow\ Nt(s(m))
\using {\blacksquare}L
\endprooftree
\justifies
{\blacksquare}{\forall}n(Nt(n)/{\it CN}{\it n}), {\square}{\it CN}{\it s(m)}\ \Rightarrow\ \fbox{${\exists}aNa$}
\using {\exists}R
\endprooftree
\prooftree
\prooftree
\justifies
\mbox{\fbox{$W[the,cold,shoulder]$}}\ \Rightarrow\ W[the,cold,shoulder]
\endprooftree
\justifies
\mbox{\fbox{${\blacksquare}W[the,cold,shoulder]$}}\ \Rightarrow\ W[the,cold,shoulder]
\using {\blacksquare}L
\endprooftree
\justifies
{\blacksquare}{\forall}n(Nt(n)/{\it CN}{\it n}), {\square}{\it CN}{\it s(m)}, {\blacksquare}W[the,cold,shoulder]\ \Rightarrow\ \fbox{${\exists}aNa{\RIGHTcircle}W[the,cold,shoulder]$}
\using {\RIGHTcircle}R
\endprooftree
\prooftree
\prooftree
\prooftree
\prooftree
\prooftree
\justifies
\mbox{\fbox{$Nt(s(f))$}}\ \Rightarrow\ Nt(s(f))
\endprooftree
\justifies
\mbox{\fbox{${\blacksquare}Nt(s(f))$}}\ \Rightarrow\ Nt(s(f))
\using {\blacksquare}L
\endprooftree
\justifies
{\blacksquare}Nt(s(f))\ \Rightarrow\ \fbox{${\exists}gNt(s(g))$}
\using {\exists}R
\endprooftree
\justifies
[{\blacksquare}Nt(s(f))]\ \Rightarrow\ \fbox{${\langle\rangle}{\exists}gNt(s(g))$}
\using {\langle\rangle}R
\endprooftree
\prooftree
\justifies
\mbox{\fbox{$Sf$}}\ \Rightarrow\ Sf
\endprooftree
\justifies
[{\blacksquare}Nt(s(f))], \mbox{\fbox{${\langle\rangle}{\exists}gNt(s(g))\backslash Sf$}}\ \Rightarrow\ Sf
\using {\backslash}L
\endprooftree
\justifies
[{\blacksquare}Nt(s(f))], \mbox{\fbox{$({\langle\rangle}{\exists}gNt(s(g))\backslash Sf)/({\exists}aNa{\RIGHTcircle}W[the,cold,shoulder])$}}, {\blacksquare}{\forall}n(Nt(n)/{\it CN}{\it n}), {\square}{\it CN}{\it s(m)}, {\blacksquare}W[the,cold,shoulder]\ \Rightarrow\ Sf
\using {/}L
\endprooftree
\justifies
[{\blacksquare}Nt(s(f))], \mbox{\fbox{${\square}(({\langle\rangle}{\exists}gNt(s(g))\backslash Sf)/({\exists}aNa{\RIGHTcircle}W[the,cold,shoulder]))$}}, {\blacksquare}{\forall}n(Nt(n)/{\it CN}{\it n}), {\square}{\it CN}{\it s(m)}, {\blacksquare}W[the,cold,shoulder]\ \Rightarrow\ Sf
\using {\Box}L
\endprooftree
$$
\vspace{0.15in}
\noindent
This delivers semantics:
\disp{
$({\it Past}\ ((\mbox{\v{}}{\it shun}\ (\iota \ \mbox{\v{}}{\it man}))\ {\it m}))$}

Similarly:
\disp{
(tdc(4343)) $[{\bf mary}]{+}{\bf gave}{+}{\bf john}{+}{\bf the}{+}{\bf cold}{+}{\bf shoulder}: Sf$}
Lexical lookup yields:
\disp{
$[{\blacksquare}Nt(s(f)): {\it m}], {\square}(({\langle\rangle}{\exists}gNt(s(g))\backslash Sf)/({\exists}aNa{\RIGHTcircle}W[the,cold,shoulder])):\\ \mbox{\^{}}\lambda A\lambda B({\it Past}\ ((\mbox{\v{}}{\it shun}\ {\it A})\ {\it B})), {\blacksquare}Nt(s(m)): {\it j}, {\blacksquare}W[the,cold,shoulder]: {\it 0}\ \Rightarrow\ Sf$}
There is the derivation:
\vspace{0.15in}
$$
\rotatebox{-0}{\tiny
\prooftree
\prooftree
\prooftree
\prooftree
\prooftree
\prooftree
\justifies
\mbox{\fbox{$Nt(s(m))$}}\ \Rightarrow\ Nt(s(m))
\endprooftree
\justifies
\mbox{\fbox{${\blacksquare}Nt(s(m))$}}\ \Rightarrow\ Nt(s(m))
\using {\blacksquare}L
\endprooftree
\justifies
{\blacksquare}Nt(s(m))\ \Rightarrow\ \fbox{${\exists}aNa$}
\using {\exists}R
\endprooftree
\prooftree
\prooftree
\justifies
\mbox{\fbox{$W[the,cold,shoulder]$}}\ \Rightarrow\ W[the,cold,shoulder]
\endprooftree
\justifies
\mbox{\fbox{${\blacksquare}W[the,cold,shoulder]$}}\ \Rightarrow\ W[the,cold,shoulder]
\using {\blacksquare}L
\endprooftree
\justifies
{\blacksquare}Nt(s(m)), {\blacksquare}W[the,cold,shoulder]\ \Rightarrow\ \fbox{${\exists}aNa{\RIGHTcircle}W[the,cold,shoulder]$}
\using {\RIGHTcircle}R
\endprooftree
\prooftree
\prooftree
\prooftree
\prooftree
\prooftree
\justifies
\mbox{\fbox{$Nt(s(f))$}}\ \Rightarrow\ Nt(s(f))
\endprooftree
\justifies
\mbox{\fbox{${\blacksquare}Nt(s(f))$}}\ \Rightarrow\ Nt(s(f))
\using {\blacksquare}L
\endprooftree
\justifies
{\blacksquare}Nt(s(f))\ \Rightarrow\ \fbox{${\exists}gNt(s(g))$}
\using {\exists}R
\endprooftree
\justifies
[{\blacksquare}Nt(s(f))]\ \Rightarrow\ \fbox{${\langle\rangle}{\exists}gNt(s(g))$}
\using {\langle\rangle}R
\endprooftree
\prooftree
\justifies
\mbox{\fbox{$Sf$}}\ \Rightarrow\ Sf
\endprooftree
\justifies
[{\blacksquare}Nt(s(f))], \mbox{\fbox{${\langle\rangle}{\exists}gNt(s(g))\backslash Sf$}}\ \Rightarrow\ Sf
\using {\backslash}L
\endprooftree
\justifies
[{\blacksquare}Nt(s(f))], \mbox{\fbox{$({\langle\rangle}{\exists}gNt(s(g))\backslash Sf)/({\exists}aNa{\RIGHTcircle}W[the,cold,shoulder])$}}, {\blacksquare}Nt(s(m)), {\blacksquare}W[the,cold,shoulder]\ \Rightarrow\ Sf
\using {/}L
\endprooftree
\justifies
[{\blacksquare}Nt(s(f))], \mbox{\fbox{${\square}(({\langle\rangle}{\exists}gNt(s(g))\backslash Sf)/({\exists}aNa{\RIGHTcircle}W[the,cold,shoulder]))$}}, {\blacksquare}Nt(s(m)), {\blacksquare}W[the,cold,shoulder]\ \Rightarrow\ Sf
\using {\Box}L
\endprooftree}
$$
\vspace{0.15in}

\noindent
This delivers semantics:
\disp{
$({\it Past}\ ((\mbox{\v{}}{\it shun}\ {\it j})\ {\it m}))$}

The next example has medial quantification:
\disp{
(tdc(47)) $[{\bf john}]{+}{\bf gave}{+}{\bf every}{+}{\bf book}{+}{\bf to}{+}{\bf mary}: Sf$}
Lexical lookup yields:
\disp{
$[{\blacksquare}Nt(s(m)): {\it j}], {\square}(({\langle\rangle}{\exists}aNa\backslash Sf)/({\exists}bNb{\bullet}{\it PP}{\it to})): \mbox{\^{}}\lambda A\lambda B({\it Past}\ (((\mbox{\v{}}{\it give}\ \pi_2{\it A})\ \pi_1{\it A})\ {\it B})),\\
 {\blacksquare}{\forall}g({\forall}f((Sf{{}{\uparrow}{}}Nt(s(g))){{}{\downarrow}{}}Sf)/{\it CN}{\it s(g)}): \lambda C\lambda D\forall E[({\it C}\ {\it E})\rightarrow ({\it D}\ {\it E})], {\square}{\it CN}{\it s(n)}: {\it book},\\
  {\blacksquare}(({\it PP}{\it to}/{\exists}aNa){\sqcap}{\forall}n(({\langle\rangle}Nn\backslash Si)/({\langle\rangle}Nn\backslash Sb))): \lambda F{\it F}, {\blacksquare}Nt(s(f)): {\it m}\ \Rightarrow\ Sf$}
There is the derivation:

\vspace{0.15in}

\begin{center}
\rotatebox{-90}
{\tiny
\prooftree
\prooftree
\prooftree
\prooftree
\prooftree
\justifies
\mbox{\fbox{${\it CN}{\it s(n)}$}}\ \Rightarrow\ {\it CN}{\it s(n)}
\endprooftree
\justifies
\mbox{\fbox{${\square}{\it CN}{\it s(n)}$}}\ \Rightarrow\ {\it CN}{\it s(n)}
\using {\Box}L
\endprooftree
\prooftree
\prooftree
\prooftree
\prooftree
\prooftree
\prooftree
\prooftree
\prooftree
\justifies
Nt(s(n))\ \Rightarrow\ Nt(s(n))
\endprooftree
\justifies
Nt(s(n))\ \Rightarrow\ \fbox{${\exists}bNb$}
\using {\exists}R
\endprooftree
\prooftree
\prooftree
\prooftree
\prooftree
\prooftree
\prooftree
\justifies
\mbox{\fbox{$Nt(s(f))$}}\ \Rightarrow\ Nt(s(f))
\endprooftree
\justifies
\mbox{\fbox{${\blacksquare}Nt(s(f))$}}\ \Rightarrow\ Nt(s(f))
\using {\blacksquare}L
\endprooftree
\justifies
{\blacksquare}Nt(s(f))\ \Rightarrow\ \fbox{${\exists}aNa$}
\using {\exists}R
\endprooftree
\prooftree
\justifies
\mbox{\fbox{${\it PP}{\it to}$}}\ \Rightarrow\ {\it PP}{\it to}
\endprooftree
\justifies
\mbox{\fbox{${\it PP}{\it to}/{\exists}aNa$}}, {\blacksquare}Nt(s(f))\ \Rightarrow\ {\it PP}{\it to}
\using {/}L
\endprooftree
\justifies
\mbox{\fbox{$({\it PP}{\it to}/{\exists}aNa){\sqcap}{\forall}n(({\langle\rangle}Nn\backslash Si)/({\langle\rangle}Nn\backslash Sb))$}}, {\blacksquare}Nt(s(f))\ \Rightarrow\ {\it PP}{\it to}
\using {\sqcap}L
\endprooftree
\justifies
\mbox{\fbox{${\blacksquare}(({\it PP}{\it to}/{\exists}aNa){\sqcap}{\forall}n(({\langle\rangle}Nn\backslash Si)/({\langle\rangle}Nn\backslash Sb)))$}}, {\blacksquare}Nt(s(f))\ \Rightarrow\ {\it PP}{\it to}
\using {\blacksquare}L
\endprooftree
\justifies
Nt(s(n)), {\blacksquare}(({\it PP}{\it to}/{\exists}aNa){\sqcap}{\forall}n(({\langle\rangle}Nn\backslash Si)/({\langle\rangle}Nn\backslash Sb))), {\blacksquare}Nt(s(f))\ \Rightarrow\ \fbox{${\exists}bNb{\bullet}{\it PP}{\it to}$}
\using {\bullet}R
\endprooftree
\prooftree
\prooftree
\prooftree
\prooftree
\prooftree
\justifies
\mbox{\fbox{$Nt(s(m))$}}\ \Rightarrow\ Nt(s(m))
\endprooftree
\justifies
\mbox{\fbox{${\blacksquare}Nt(s(m))$}}\ \Rightarrow\ Nt(s(m))
\using {\blacksquare}L
\endprooftree
\justifies
{\blacksquare}Nt(s(m))\ \Rightarrow\ \fbox{${\exists}aNa$}
\using {\exists}R
\endprooftree
\justifies
[{\blacksquare}Nt(s(m))]\ \Rightarrow\ \fbox{${\langle\rangle}{\exists}aNa$}
\using {\langle\rangle}R
\endprooftree
\prooftree
\justifies
\mbox{\fbox{$Sf$}}\ \Rightarrow\ Sf
\endprooftree
\justifies
[{\blacksquare}Nt(s(m))], \mbox{\fbox{${\langle\rangle}{\exists}aNa\backslash Sf$}}\ \Rightarrow\ Sf
\using {\backslash}L
\endprooftree
\justifies
[{\blacksquare}Nt(s(m))], \mbox{\fbox{$({\langle\rangle}{\exists}aNa\backslash Sf)/({\exists}bNb{\bullet}{\it PP}{\it to})$}}, Nt(s(n)), {\blacksquare}(({\it PP}{\it to}/{\exists}aNa){\sqcap}{\forall}n(({\langle\rangle}Nn\backslash Si)/({\langle\rangle}Nn\backslash Sb))), {\blacksquare}Nt(s(f))\ \Rightarrow\ Sf
\using {/}L
\endprooftree
\justifies
[{\blacksquare}Nt(s(m))], \mbox{\fbox{${\square}(({\langle\rangle}{\exists}aNa\backslash Sf)/({\exists}bNb{\bullet}{\it PP}{\it to}))$}}, Nt(s(n)), {\blacksquare}(({\it PP}{\it to}/{\exists}aNa){\sqcap}{\forall}n(({\langle\rangle}Nn\backslash Si)/({\langle\rangle}Nn\backslash Sb))), {\blacksquare}Nt(s(f))\ \Rightarrow\ Sf
\using {\Box}L
\endprooftree
\justifies
[{\blacksquare}Nt(s(m))], {\square}(({\langle\rangle}{\exists}aNa\backslash Sf)/({\exists}bNb{\bullet}{\it PP}{\it to})), {\tt 1}, {\blacksquare}(({\it PP}{\it to}/{\exists}aNa){\sqcap}{\forall}n(({\langle\rangle}Nn\backslash Si)/({\langle\rangle}Nn\backslash Sb))), {\blacksquare}Nt(s(f))\ \Rightarrow\ Sf{{}{\uparrow}{}}Nt(s(n))
\using {\uparrow}R
\endprooftree
\prooftree
\justifies
\mbox{\fbox{$Sf$}}\ \Rightarrow\ Sf
\endprooftree
\justifies
[{\blacksquare}Nt(s(m))], {\square}(({\langle\rangle}{\exists}aNa\backslash Sf)/({\exists}bNb{\bullet}{\it PP}{\it to})), \mbox{\fbox{$(Sf{{}{\uparrow}{}}Nt(s(n))){{}{\downarrow}{}}Sf$}}, {\blacksquare}(({\it PP}{\it to}/{\exists}aNa){\sqcap}{\forall}n(({\langle\rangle}Nn\backslash Si)/({\langle\rangle}Nn\backslash Sb))), {\blacksquare}Nt(s(f))\ \Rightarrow\ Sf
\using {\downarrow}L
\endprooftree
\justifies
[{\blacksquare}Nt(s(m))], {\square}(({\langle\rangle}{\exists}aNa\backslash Sf)/({\exists}bNb{\bullet}{\it PP}{\it to})), \mbox{\fbox{${\forall}f((Sf{{}{\uparrow}{}}Nt(s(n))){{}{\downarrow}{}}Sf)$}}, {\blacksquare}(({\it PP}{\it to}/{\exists}aNa){\sqcap}{\forall}n(({\langle\rangle}Nn\backslash Si)/({\langle\rangle}Nn\backslash Sb))), {\blacksquare}Nt(s(f))\ \Rightarrow\ Sf
\using {\forall}L
\endprooftree
\justifies
[{\blacksquare}Nt(s(m))], {\square}(({\langle\rangle}{\exists}aNa\backslash Sf)/({\exists}bNb{\bullet}{\it PP}{\it to})), \mbox{\fbox{${\forall}f((Sf{{}{\uparrow}{}}Nt(s(n))){{}{\downarrow}{}}Sf)/{\it CN}{\it s(n)}$}}, {\square}{\it CN}{\it s(n)}, {\blacksquare}(({\it PP}{\it to}/{\exists}aNa){\sqcap}{\forall}n(({\langle\rangle}Nn\backslash Si)/({\langle\rangle}Nn\backslash Sb))), {\blacksquare}Nt(s(f))\ \Rightarrow\ Sf
\using {/}L
\endprooftree
\justifies
[{\blacksquare}Nt(s(m))], {\square}(({\langle\rangle}{\exists}aNa\backslash Sf)/({\exists}bNb{\bullet}{\it PP}{\it to})), \mbox{\fbox{${\forall}g({\forall}f((Sf{{}{\uparrow}{}}Nt(s(g))){{}{\downarrow}{}}Sf)/{\it CN}{\it s(g)})$}}, {\square}{\it CN}{\it s(n)}, {\blacksquare}(({\it PP}{\it to}/{\exists}aNa){\sqcap}{\forall}n(({\langle\rangle}Nn\backslash Si)/({\langle\rangle}Nn\backslash Sb))), {\blacksquare}Nt(s(f))\ \Rightarrow\ Sf
\using {\forall}L
\endprooftree
\justifies
[{\blacksquare}Nt(s(m))], {\square}(({\langle\rangle}{\exists}aNa\backslash Sf)/({\exists}bNb{\bullet}{\it PP}{\it to})), \mbox{\fbox{${\blacksquare}{\forall}g({\forall}f((Sf{{}{\uparrow}{}}Nt(s(g))){{}{\downarrow}{}}Sf)/{\it CN}{\it s(g)})$}}, {\square}{\it CN}{\it s(n)}, {\blacksquare}(({\it PP}{\it to}/{\exists}aNa){\sqcap}{\forall}n(({\langle\rangle}Nn\backslash Si)/({\langle\rangle}Nn\backslash Sb))), {\blacksquare}Nt(s(f))\ \Rightarrow\ Sf
\using {\blacksquare}L
\endprooftree}
\end{center}

\vspace{0.15in}

\noindent
This delivers semantics:
\disp{
$\forall C[(\mbox{\v{}}{\it book}\ {\it C})\rightarrow ({\it Past}\ (((\mbox{\v{}}{\it give}\ {\it m})\ {\it C})\ {\it j}))]$}

The following example has subordinate clause existential quantification,
exhibiting de re/de dicto ambiguity:
\disp{
(tdc(50)) $[{\bf mary}]{+}{\bf thinks}{+}[{\bf someone}]{+}{\bf left}: Sf$}
Lexical lookup yields:
\disp{
$[{\blacksquare}Nt(s(f)): {\it m}], {\square}(({\langle\rangle}{\exists}gNt(s(g))\backslash Sf)/({\it CP}that{\sqcup}{\square}Sf)): \mbox{\^{}}\lambda A\lambda B({\it Pres}\ ((\mbox{\v{}}{\it think}\ {\it A})\ {\it B})), \\{}[{\square}{\forall}f((Sf{{}{\uparrow}{}}{\blacksquare}{\forall}gNt(g)){{}{\downarrow}{}}Sf): \mbox{\^{}}\lambda C\exists D[(\mbox{\v{}}{\it person}\ {\it D})\wedge ({\it C}\ {\it D})]], {\square}({\langle\rangle}{\exists}gNt(s(g))\backslash Sf):\\ \mbox{\^{}}\lambda E({\it Pres}\ (\mbox{\v{}}{\it leave}\ {\it E}))\ \Rightarrow\ Sf$}
There is the de re derivation:
\vspace{0.15in}
$$
\tiny
\prooftree
\prooftree
\prooftree
\prooftree
\prooftree
\prooftree
\prooftree
\prooftree
\prooftree
\prooftree
\prooftree
\prooftree
\prooftree
\prooftree
\prooftree
\justifies
\mbox{\fbox{$Nt(s(A))$}}\ \Rightarrow\ Nt(s(A))
\endprooftree
\justifies
\mbox{\fbox{${\forall}gNt(g)$}}\ \Rightarrow\ Nt(s(A))
\using {\forall}L
\endprooftree
\justifies
\mbox{\fbox{${\blacksquare}{\forall}gNt(g)$}}\ \Rightarrow\ Nt(s(A))
\using {\blacksquare}L
\endprooftree
\justifies
{\blacksquare}{\forall}gNt(g)\ \Rightarrow\ \fbox{${\exists}gNt(s(g))$}
\using {\exists}R
\endprooftree
\justifies
[{\blacksquare}{\forall}gNt(g)]\ \Rightarrow\ \fbox{${\langle\rangle}{\exists}gNt(s(g))$}
\using {\langle\rangle}R
\endprooftree
\prooftree
\justifies
\mbox{\fbox{$Sf$}}\ \Rightarrow\ Sf
\endprooftree
\justifies
[{\blacksquare}{\forall}gNt(g)], \mbox{\fbox{${\langle\rangle}{\exists}gNt(s(g))\backslash Sf$}}\ \Rightarrow\ Sf
\using {\backslash}L
\endprooftree
\justifies
[{\blacksquare}{\forall}gNt(g)], \mbox{\fbox{${\square}({\langle\rangle}{\exists}gNt(s(g))\backslash Sf)$}}\ \Rightarrow\ Sf
\using {\Box}L
\endprooftree
\justifies
[{\blacksquare}{\forall}gNt(g)], {\square}({\langle\rangle}{\exists}gNt(s(g))\backslash Sf)\ \Rightarrow\ {\square}Sf
\using {\Box}R
\endprooftree
\justifies
[{\blacksquare}{\forall}gNt(g)], {\square}({\langle\rangle}{\exists}gNt(s(g))\backslash Sf)\ \Rightarrow\ \fbox{${\it CP}that{\sqcup}{\square}Sf$}
\using {\sqcup}R
\endprooftree
\prooftree
\prooftree
\prooftree
\prooftree
\prooftree
\justifies
\mbox{\fbox{$Nt(s(f))$}}\ \Rightarrow\ Nt(s(f))
\endprooftree
\justifies
\mbox{\fbox{${\blacksquare}Nt(s(f))$}}\ \Rightarrow\ Nt(s(f))
\using {\blacksquare}L
\endprooftree
\justifies
{\blacksquare}Nt(s(f))\ \Rightarrow\ \fbox{${\exists}gNt(s(g))$}
\using {\exists}R
\endprooftree
\justifies
[{\blacksquare}Nt(s(f))]\ \Rightarrow\ \fbox{${\langle\rangle}{\exists}gNt(s(g))$}
\using {\langle\rangle}R
\endprooftree
\prooftree
\justifies
\mbox{\fbox{$Sf$}}\ \Rightarrow\ Sf
\endprooftree
\justifies
[{\blacksquare}Nt(s(f))], \mbox{\fbox{${\langle\rangle}{\exists}gNt(s(g))\backslash Sf$}}\ \Rightarrow\ Sf
\using {\backslash}L
\endprooftree
\justifies
[{\blacksquare}Nt(s(f))], \mbox{\fbox{$({\langle\rangle}{\exists}gNt(s(g))\backslash Sf)/({\it CP}that{\sqcup}{\square}Sf)$}}, [{\blacksquare}{\forall}gNt(g)], {\square}({\langle\rangle}{\exists}gNt(s(g))\backslash Sf)\ \Rightarrow\ Sf
\using {/}L
\endprooftree
\justifies
[{\blacksquare}Nt(s(f))], \mbox{\fbox{${\square}(({\langle\rangle}{\exists}gNt(s(g))\backslash Sf)/({\it CP}that{\sqcup}{\square}Sf))$}}, [{\blacksquare}{\forall}gNt(g)], {\square}({\langle\rangle}{\exists}gNt(s(g))\backslash Sf)\ \Rightarrow\ Sf
\using {\Box}L
\endprooftree
\justifies
[{\blacksquare}Nt(s(f))], {\square}(({\langle\rangle}{\exists}gNt(s(g))\backslash Sf)/({\it CP}that{\sqcup}{\square}Sf)), [{\tt 1}], {\square}({\langle\rangle}{\exists}gNt(s(g))\backslash Sf)\ \Rightarrow\ Sf{{}{\uparrow}{}}{\blacksquare}{\forall}gNt(g)
\using {\uparrow}R
\endprooftree
\prooftree
\justifies
\mbox{\fbox{$Sf$}}\ \Rightarrow\ Sf
\endprooftree
\justifies
[{\blacksquare}Nt(s(f))], {\square}(({\langle\rangle}{\exists}gNt(s(g))\backslash Sf)/({\it CP}that{\sqcup}{\square}Sf)), [\mbox{\fbox{$(Sf{{}{\uparrow}{}}{\blacksquare}{\forall}gNt(g)){{}{\downarrow}{}}Sf$}}], {\square}({\langle\rangle}{\exists}gNt(s(g))\backslash Sf)\ \Rightarrow\ Sf
\using {\downarrow}L
\endprooftree
\justifies
[{\blacksquare}Nt(s(f))], {\square}(({\langle\rangle}{\exists}gNt(s(g))\backslash Sf)/({\it CP}that{\sqcup}{\square}Sf)), [\mbox{\fbox{${\forall}f((Sf{{}{\uparrow}{}}{\blacksquare}{\forall}gNt(g)){{}{\downarrow}{}}Sf)$}}], {\square}({\langle\rangle}{\exists}gNt(s(g))\backslash Sf)\ \Rightarrow\ Sf
\using {\forall}L
\endprooftree
\justifies
[{\blacksquare}Nt(s(f))], {\square}(({\langle\rangle}{\exists}gNt(s(g))\backslash Sf)/({\it CP}that{\sqcup}{\square}Sf)), [\mbox{\fbox{${\square}{\forall}f((Sf{{}{\uparrow}{}}{\blacksquare}{\forall}gNt(g)){{}{\downarrow}{}}Sf)$}}], {\square}({\langle\rangle}{\exists}gNt(s(g))\backslash Sf)\ \Rightarrow\ Sf
\using {\Box}L
\endprooftree
$$
\vspace{0.15in}
\noindent
This delivers semantics:
\disp{
$\exists B[(\mbox{\v{}}{\it person}\ {\it B})\wedge ({\it Pres}\ ((\mbox{\v{}}{\it think}\ \mbox{\^{}}({\it Pres}\ (\mbox{\v{}}{\it leave}\ {\it B})))\ {\it m}))]$}
And the de dicto derivation:
\vspace{0.15in}
$$
{\tiny
\prooftree
\prooftree
\prooftree
\prooftree
\prooftree
\prooftree
\prooftree
\prooftree
\prooftree
\prooftree
\prooftree
\prooftree
\prooftree
\prooftree
\prooftree
\justifies
\mbox{\fbox{$Nt(s(A))$}}\ \Rightarrow\ Nt(s(A))
\endprooftree
\justifies
\mbox{\fbox{${\forall}gNt(g)$}}\ \Rightarrow\ Nt(s(A))
\using {\forall}L
\endprooftree
\justifies
\mbox{\fbox{${\blacksquare}{\forall}gNt(g)$}}\ \Rightarrow\ Nt(s(A))
\using {\blacksquare}L
\endprooftree
\justifies
{\blacksquare}{\forall}gNt(g)\ \Rightarrow\ \fbox{${\exists}gNt(s(g))$}
\using {\exists}R
\endprooftree
\justifies
[{\blacksquare}{\forall}gNt(g)]\ \Rightarrow\ \fbox{${\langle\rangle}{\exists}gNt(s(g))$}
\using {\langle\rangle}R
\endprooftree
\prooftree
\justifies
\mbox{\fbox{$Sf$}}\ \Rightarrow\ Sf
\endprooftree
\justifies
[{\blacksquare}{\forall}gNt(g)], \mbox{\fbox{${\langle\rangle}{\exists}gNt(s(g))\backslash Sf$}}\ \Rightarrow\ Sf
\using {\backslash}L
\endprooftree
\justifies
[{\blacksquare}{\forall}gNt(g)], \mbox{\fbox{${\square}({\langle\rangle}{\exists}gNt(s(g))\backslash Sf)$}}\ \Rightarrow\ Sf
\using {\Box}L
\endprooftree
\justifies
[{\tt 1}], {\square}({\langle\rangle}{\exists}gNt(s(g))\backslash Sf)\ \Rightarrow\ Sf{{}{\uparrow}{}}{\blacksquare}{\forall}gNt(g)
\using {\uparrow}R
\endprooftree
\prooftree
\justifies
\mbox{\fbox{$Sf$}}\ \Rightarrow\ Sf
\endprooftree
\justifies
[\mbox{\fbox{$(Sf{{}{\uparrow}{}}{\blacksquare}{\forall}gNt(g)){{}{\downarrow}{}}Sf$}}], {\square}({\langle\rangle}{\exists}gNt(s(g))\backslash Sf)\ \Rightarrow\ Sf
\using {\downarrow}L
\endprooftree
\justifies
[\mbox{\fbox{${\forall}f((Sf{{}{\uparrow}{}}{\blacksquare}{\forall}gNt(g)){{}{\downarrow}{}}Sf)$}}], {\square}({\langle\rangle}{\exists}gNt(s(g))\backslash Sf)\ \Rightarrow\ Sf
\using {\forall}L
\endprooftree
\justifies
[\mbox{\fbox{${\square}{\forall}f((Sf{{}{\uparrow}{}}{\blacksquare}{\forall}gNt(g)){{}{\downarrow}{}}Sf)$}}], {\square}({\langle\rangle}{\exists}gNt(s(g))\backslash Sf)\ \Rightarrow\ Sf
\using {\Box}L
\endprooftree
\justifies
[{\square}{\forall}f((Sf{{}{\uparrow}{}}{\blacksquare}{\forall}gNt(g)){{}{\downarrow}{}}Sf)], {\square}({\langle\rangle}{\exists}gNt(s(g))\backslash Sf)\ \Rightarrow\ {\square}Sf
\using {\Box}R
\endprooftree
\justifies
[{\square}{\forall}f((Sf{{}{\uparrow}{}}{\blacksquare}{\forall}gNt(g)){{}{\downarrow}{}}Sf)], {\square}({\langle\rangle}{\exists}gNt(s(g))\backslash Sf)\ \Rightarrow\ \fbox{${\it CP}that{\sqcup}{\square}Sf$}
\using {\sqcup}R
\endprooftree
\prooftree
\prooftree
\prooftree
\prooftree
\prooftree
\justifies
\mbox{\fbox{$Nt(s(f))$}}\ \Rightarrow\ Nt(s(f))
\endprooftree
\justifies
\mbox{\fbox{${\blacksquare}Nt(s(f))$}}\ \Rightarrow\ Nt(s(f))
\using {\blacksquare}L
\endprooftree
\justifies
{\blacksquare}Nt(s(f))\ \Rightarrow\ \fbox{${\exists}gNt(s(g))$}
\using {\exists}R
\endprooftree
\justifies
[{\blacksquare}Nt(s(f))]\ \Rightarrow\ \fbox{${\langle\rangle}{\exists}gNt(s(g))$}
\using {\langle\rangle}R
\endprooftree
\prooftree
\justifies
\mbox{\fbox{$Sf$}}\ \Rightarrow\ Sf
\endprooftree
\justifies
[{\blacksquare}Nt(s(f))], \mbox{\fbox{${\langle\rangle}{\exists}gNt(s(g))\backslash Sf$}}\ \Rightarrow\ Sf
\using {\backslash}L
\endprooftree
\justifies
[{\blacksquare}Nt(s(f))], \mbox{\fbox{$({\langle\rangle}{\exists}gNt(s(g))\backslash Sf)/({\it CP}that{\sqcup}{\square}Sf)$}}, [{\square}{\forall}f((Sf{{}{\uparrow}{}}{\blacksquare}{\forall}gNt(g)){{}{\downarrow}{}}Sf)], {\square}({\langle\rangle}{\exists}gNt(s(g))\backslash Sf)\ \Rightarrow\ Sf
\using {/}L
\endprooftree
\justifies
[{\blacksquare}Nt(s(f))], \mbox{\fbox{${\square}(({\langle\rangle}{\exists}gNt(s(g))\backslash Sf)/({\it CP}that{\sqcup}{\square}Sf))$}}, [{\square}{\forall}f((Sf{{}{\uparrow}{}}{\blacksquare}{\forall}gNt(g)){{}{\downarrow}{}}Sf)], {\square}({\langle\rangle}{\exists}gNt(s(g))\backslash Sf)\ \Rightarrow\ Sf
\using {\Box}L
\endprooftree}
$$
\vspace{0.15in}

\noindent
This delivers semantics:
\disp{
$({\it Pres}\ ((\mbox{\v{}}{\it think}\ \mbox{\^{}}\exists D[(\mbox{\v{}}{\it person}\ {\it D})\wedge ({\it Pres}\ (\mbox{\v{}}{\it leave}\ {\it D}))])\ {\it m}))$}

The next example exhibits classical quantifier scope ambiguity:
\disp{
(tdc(53)) $[{\bf everyone}]{+}{\bf loves}{+}{\bf someone}: Sf$}
There is the subject wide scope reading (cf.\ everyone loves their (respective) mother)
and the object wide scope reading (cf.\ everyone loves (one and) the (same) queen).
Lexical lookup yields:
\disp{
$[{\square}{\forall}f((Sf{{}{\uparrow}{}}{\forall}gNt(g)){{}{\downarrow}{}}Sf): \mbox{\^{}}\lambda A\forall B[(\mbox{\v{}}{\it person}\ {\it B})\rightarrow ({\it A}\ {\it B})]], {\square}(({\langle\rangle}{\exists}gNt(s(g))\backslash Sf)/{\exists}aNa):\\ \mbox{\^{}}\lambda C\lambda D({\it Pres}\ ((\mbox{\v{}}{\it love}\ {\it C})\ {\it D})),
 {\square}{\forall}f((Sf{{}{\uparrow}{}}{\blacksquare}{\forall}gNt(g)){{}{\downarrow}{}}Sf): \mbox{\^{}}\lambda E\exists F[(\mbox{\v{}}{\it person}\ {\it F})\wedge ({\it E}\ {\it F})]\ \Rightarrow\ Sf$}
There is the subject wide scope derivation as follows in which the subject quantifier
is processed closest to the root:
\vspace{0.15in}
$$
{\tiny
\prooftree
\prooftree
\prooftree
\prooftree
\prooftree
\prooftree
\prooftree
\prooftree
\prooftree
\prooftree
\prooftree
\prooftree
\prooftree
\prooftree
\justifies
\mbox{\fbox{$Nt(A)$}}\ \Rightarrow\ Nt(A)
\endprooftree
\justifies
\mbox{\fbox{${\forall}gNt(g)$}}\ \Rightarrow\ Nt(A)
\using {\forall}L
\endprooftree
\justifies
\mbox{\fbox{${\blacksquare}{\forall}gNt(g)$}}\ \Rightarrow\ Nt(A)
\using {\blacksquare}L
\endprooftree
\justifies
{\blacksquare}{\forall}gNt(g)\ \Rightarrow\ \fbox{${\exists}aNa$}
\using {\exists}R
\endprooftree
\prooftree
\prooftree
\prooftree
\prooftree
\prooftree
\justifies
\mbox{\fbox{$Nt(s(A))$}}\ \Rightarrow\ Nt(s(A))
\endprooftree
\justifies
\mbox{\fbox{${\forall}gNt(g)$}}\ \Rightarrow\ Nt(s(A))
\using {\forall}L
\endprooftree
\justifies
{\forall}gNt(g)\ \Rightarrow\ \fbox{${\exists}gNt(s(g))$}
\using {\exists}R
\endprooftree
\justifies
[{\forall}gNt(g)]\ \Rightarrow\ \fbox{${\langle\rangle}{\exists}gNt(s(g))$}
\using {\langle\rangle}R
\endprooftree
\prooftree
\justifies
\mbox{\fbox{$Sf$}}\ \Rightarrow\ Sf
\endprooftree
\justifies
[{\forall}gNt(g)], \mbox{\fbox{${\langle\rangle}{\exists}gNt(s(g))\backslash Sf$}}\ \Rightarrow\ Sf
\using {\backslash}L
\endprooftree
\justifies
[{\forall}gNt(g)], \mbox{\fbox{$({\langle\rangle}{\exists}gNt(s(g))\backslash Sf)/{\exists}aNa$}}, {\blacksquare}{\forall}gNt(g)\ \Rightarrow\ Sf
\using {/}L
\endprooftree
\justifies
[{\forall}gNt(g)], \mbox{\fbox{${\square}(({\langle\rangle}{\exists}gNt(s(g))\backslash Sf)/{\exists}aNa)$}}, {\blacksquare}{\forall}gNt(g)\ \Rightarrow\ Sf
\using {\Box}L
\endprooftree
\justifies
[{\forall}gNt(g)], {\square}(({\langle\rangle}{\exists}gNt(s(g))\backslash Sf)/{\exists}aNa), {\tt 1}\ \Rightarrow\ Sf{{}{\uparrow}{}}{\blacksquare}{\forall}gNt(g)
\using {\uparrow}R
\endprooftree
\prooftree
\justifies
\mbox{\fbox{$Sf$}}\ \Rightarrow\ Sf
\endprooftree
\justifies
[{\forall}gNt(g)], {\square}(({\langle\rangle}{\exists}gNt(s(g))\backslash Sf)/{\exists}aNa), \mbox{\fbox{$(Sf{{}{\uparrow}{}}{\blacksquare}{\forall}gNt(g)){{}{\downarrow}{}}Sf$}}\ \Rightarrow\ Sf
\using {\downarrow}L
\endprooftree
\justifies
[{\forall}gNt(g)], {\square}(({\langle\rangle}{\exists}gNt(s(g))\backslash Sf)/{\exists}aNa), \mbox{\fbox{${\forall}f((Sf{{}{\uparrow}{}}{\blacksquare}{\forall}gNt(g)){{}{\downarrow}{}}Sf)$}}\ \Rightarrow\ Sf
\using {\forall}L
\endprooftree
\justifies
[{\forall}gNt(g)], {\square}(({\langle\rangle}{\exists}gNt(s(g))\backslash Sf)/{\exists}aNa), \mbox{\fbox{${\square}{\forall}f((Sf{{}{\uparrow}{}}{\blacksquare}{\forall}gNt(g)){{}{\downarrow}{}}Sf)$}}\ \Rightarrow\ Sf
\using {\Box}L
\endprooftree
\justifies
[{\tt 1}], {\square}(({\langle\rangle}{\exists}gNt(s(g))\backslash Sf)/{\exists}aNa), {\square}{\forall}f((Sf{{}{\uparrow}{}}{\blacksquare}{\forall}gNt(g)){{}{\downarrow}{}}Sf)\ \Rightarrow\ Sf{{}{\uparrow}{}}{\forall}gNt(g)
\using {\uparrow}R
\endprooftree
\prooftree
\justifies
\mbox{\fbox{$Sf$}}\ \Rightarrow\ Sf
\endprooftree
\justifies
[\mbox{\fbox{$(Sf{{}{\uparrow}{}}{\forall}gNt(g)){{}{\downarrow}{}}Sf$}}], {\square}(({\langle\rangle}{\exists}gNt(s(g))\backslash Sf)/{\exists}aNa), {\square}{\forall}f((Sf{{}{\uparrow}{}}{\blacksquare}{\forall}gNt(g)){{}{\downarrow}{}}Sf)\ \Rightarrow\ Sf
\using {\downarrow}L
\endprooftree
\justifies
[\mbox{\fbox{${\forall}f((Sf{{}{\uparrow}{}}{\forall}gNt(g)){{}{\downarrow}{}}Sf)$}}], {\square}(({\langle\rangle}{\exists}gNt(s(g))\backslash Sf)/{\exists}aNa), {\square}{\forall}f((Sf{{}{\uparrow}{}}{\blacksquare}{\forall}gNt(g)){{}{\downarrow}{}}Sf)\ \Rightarrow\ Sf
\using {\forall}L
\endprooftree
\justifies
[\mbox{\fbox{${\square}{\forall}f((Sf{{}{\uparrow}{}}{\forall}gNt(g)){{}{\downarrow}{}}Sf)$}}], {\square}(({\langle\rangle}{\exists}gNt(s(g))\backslash Sf)/{\exists}aNa), {\square}{\forall}f((Sf{{}{\uparrow}{}}{\blacksquare}{\forall}gNt(g)){{}{\downarrow}{}}Sf)\ \Rightarrow\ Sf
\using {\Box}L
\endprooftree}
$$
\vspace{0.15in}
This delivers semantics:
\disp{
$\forall B[(\mbox{\v{}}{\it person}\ {\it B})\rightarrow \exists E[(\mbox{\v{}}{\it person}\ {\it E})\wedge ({\it Pres}\ ((\mbox{\v{}}{\it love}\ {\it E})\ {\it B}))]]$}
And there is the object wide scope derivation as follows in which the object quantifier
is processed closest to the root:
\vspace{0.15in}
$$
{\tiny
\prooftree
\prooftree
\prooftree
\prooftree
\prooftree
\prooftree
\prooftree
\prooftree
\prooftree
\prooftree
\prooftree
\prooftree
\prooftree
\prooftree
\justifies
\mbox{\fbox{$Nt(A)$}}\ \Rightarrow\ Nt(A)
\endprooftree
\justifies
\mbox{\fbox{${\forall}gNt(g)$}}\ \Rightarrow\ Nt(A)
\using {\forall}L
\endprooftree
\justifies
\mbox{\fbox{${\blacksquare}{\forall}gNt(g)$}}\ \Rightarrow\ Nt(A)
\using {\blacksquare}L
\endprooftree
\justifies
{\blacksquare}{\forall}gNt(g)\ \Rightarrow\ \fbox{${\exists}aNa$}
\using {\exists}R
\endprooftree
\prooftree
\prooftree
\prooftree
\prooftree
\prooftree
\justifies
\mbox{\fbox{$Nt(s(A))$}}\ \Rightarrow\ Nt(s(A))
\endprooftree
\justifies
\mbox{\fbox{${\forall}gNt(g)$}}\ \Rightarrow\ Nt(s(A))
\using {\forall}L
\endprooftree
\justifies
{\forall}gNt(g)\ \Rightarrow\ \fbox{${\exists}gNt(s(g))$}
\using {\exists}R
\endprooftree
\justifies
[{\forall}gNt(g)]\ \Rightarrow\ \fbox{${\langle\rangle}{\exists}gNt(s(g))$}
\using {\langle\rangle}R
\endprooftree
\prooftree
\justifies
\mbox{\fbox{$Sf$}}\ \Rightarrow\ Sf
\endprooftree
\justifies
[{\forall}gNt(g)], \mbox{\fbox{${\langle\rangle}{\exists}gNt(s(g))\backslash Sf$}}\ \Rightarrow\ Sf
\using {\backslash}L
\endprooftree
\justifies
[{\forall}gNt(g)], \mbox{\fbox{$({\langle\rangle}{\exists}gNt(s(g))\backslash Sf)/{\exists}aNa$}}, {\blacksquare}{\forall}gNt(g)\ \Rightarrow\ Sf
\using {/}L
\endprooftree
\justifies
[{\forall}gNt(g)], \mbox{\fbox{${\square}(({\langle\rangle}{\exists}gNt(s(g))\backslash Sf)/{\exists}aNa)$}}, {\blacksquare}{\forall}gNt(g)\ \Rightarrow\ Sf
\using {\Box}L
\endprooftree
\justifies
[{\tt 1}], {\square}(({\langle\rangle}{\exists}gNt(s(g))\backslash Sf)/{\exists}aNa), {\blacksquare}{\forall}gNt(g)\ \Rightarrow\ Sf{{}{\uparrow}{}}{\forall}gNt(g)
\using {\uparrow}R
\endprooftree
\prooftree
\justifies
\mbox{\fbox{$Sf$}}\ \Rightarrow\ Sf
\endprooftree
\justifies
[\mbox{\fbox{$(Sf{{}{\uparrow}{}}{\forall}gNt(g)){{}{\downarrow}{}}Sf$}}], {\square}(({\langle\rangle}{\exists}gNt(s(g))\backslash Sf)/{\exists}aNa), {\blacksquare}{\forall}gNt(g)\ \Rightarrow\ Sf
\using {\downarrow}L
\endprooftree
\justifies
[\mbox{\fbox{${\forall}f((Sf{{}{\uparrow}{}}{\forall}gNt(g)){{}{\downarrow}{}}Sf)$}}], {\square}(({\langle\rangle}{\exists}gNt(s(g))\backslash Sf)/{\exists}aNa), {\blacksquare}{\forall}gNt(g)\ \Rightarrow\ Sf
\using {\forall}L
\endprooftree
\justifies
[\mbox{\fbox{${\square}{\forall}f((Sf{{}{\uparrow}{}}{\forall}gNt(g)){{}{\downarrow}{}}Sf)$}}], {\square}(({\langle\rangle}{\exists}gNt(s(g))\backslash Sf)/{\exists}aNa), {\blacksquare}{\forall}gNt(g)\ \Rightarrow\ Sf
\using {\Box}L
\endprooftree
\justifies
[{\square}{\forall}f((Sf{{}{\uparrow}{}}{\forall}gNt(g)){{}{\downarrow}{}}Sf)], {\square}(({\langle\rangle}{\exists}gNt(s(g))\backslash Sf)/{\exists}aNa), {\tt 1}\ \Rightarrow\ Sf{{}{\uparrow}{}}{\blacksquare}{\forall}gNt(g)
\using {\uparrow}R
\endprooftree
\prooftree
\justifies
\mbox{\fbox{$Sf$}}\ \Rightarrow\ Sf
\endprooftree
\justifies
[{\square}{\forall}f((Sf{{}{\uparrow}{}}{\forall}gNt(g)){{}{\downarrow}{}}Sf)], {\square}(({\langle\rangle}{\exists}gNt(s(g))\backslash Sf)/{\exists}aNa), \mbox{\fbox{$(Sf{{}{\uparrow}{}}{\blacksquare}{\forall}gNt(g)){{}{\downarrow}{}}Sf$}}\ \Rightarrow\ Sf
\using {\downarrow}L
\endprooftree
\justifies
[{\square}{\forall}f((Sf{{}{\uparrow}{}}{\forall}gNt(g)){{}{\downarrow}{}}Sf)], {\square}(({\langle\rangle}{\exists}gNt(s(g))\backslash Sf)/{\exists}aNa), \mbox{\fbox{${\forall}f((Sf{{}{\uparrow}{}}{\blacksquare}{\forall}gNt(g)){{}{\downarrow}{}}Sf)$}}\ \Rightarrow\ Sf
\using {\forall}L
\endprooftree
\justifies
[{\square}{\forall}f((Sf{{}{\uparrow}{}}{\forall}gNt(g)){{}{\downarrow}{}}Sf)], {\square}(({\langle\rangle}{\exists}gNt(s(g))\backslash Sf)/{\exists}aNa), \mbox{\fbox{${\square}{\forall}f((Sf{{}{\uparrow}{}}{\blacksquare}{\forall}gNt(g)){{}{\downarrow}{}}Sf)$}}\ \Rightarrow\ Sf
\using {\Box}L
\endprooftree}
$$
\vspace{0.15in}
\noindent
This delivers the semantics:
\disp{
$\exists B[(\mbox{\v{}}{\it person}\ {\it B})\wedge \forall E[(\mbox{\v{}}{\it person}\ {\it E})\rightarrow ({\it Pres}\ ((\mbox{\v{}}{\it love}\ {\it B})\ {\it E}))]]$}

The next example is of medial relativisation:
\disp{
(tdc(54)) ${\bf dog}{+}[[{\bf that}{+}[{\bf mary}]{+}{\bf saw}{+}{\bf today}]]: {\it CN}{\it s(n)}$}
But it is not analysed as in Morrill, Valent\'{\i}n and Fadda (2011\cite{mvf:tdc})
but as in Morrill (2011\cite{morrill:oxford}), Chapter~5.
Note the double brackets for the strong island relative clause.
The lexical lookup yields:
\disp{
\vspace{0.15in}
${\square}{\it CN}{\it s(n)}: {\it dog}, [[{\blacksquare}{\forall}n({[]^{-1}}{[]^{-1}}({\it CN}{\it n}\backslash {\it CN}{\it n})/{\blacksquare}(({\langle\rangle}Nt(n){\sqcap}!{\blacksquare}Nt(n))\backslash Sf)):\\\lambda A\lambda B\lambda C[({\it B}\ {\it C})\wedge ({\it A}\ {\it C})], [{\blacksquare}Nt(s(f)): {\it m}], {\square}(({\langle\rangle}{\exists}aNa\backslash Sf)/({\exists}aNa{\oplus}{\it CP}that)):\\ \mbox{\^{}}\lambda D\lambda E({\it Past}\ (({\it D}\casearrow F.(\mbox{\v{}}{\it seee}\ {\it F}); G.(\mbox{\v{}}{\it seet}\ {\it G}))\ {\it E})), {\square}{\forall}a{\forall}f(({\langle\rangle}Na\backslash Sf)\backslash ({\langle\rangle}Na\backslash Sf)):\\ \mbox{\^{}}\lambda H\lambda I(\mbox{\v{}}{\it today}\ ({\it H}\ {\it I}))]]\ \Rightarrow\ {\it CN}{\it s(n)}$}
There is the derivation:

\vspace{0.15in}

\begin{center}
\rotatebox{-90}{
\resizebox{\textheight}{!}{
\prooftree
\prooftree
\prooftree
\prooftree
\prooftree
\prooftree
\prooftree
\prooftree
\prooftree
\prooftree
\prooftree
\prooftree
\prooftree
\prooftree
\prooftree
\prooftree
\prooftree
\prooftree
\prooftree
\prooftree
\justifies
\mbox{\fbox{$Nt(s(n))$}}\ \Rightarrow\ Nt(s(n))
\endprooftree
\justifies
\mbox{\fbox{${\blacksquare}Nt(s(n))$}}\ \Rightarrow\ Nt(s(n))
\using {\blacksquare}L
\endprooftree
\justifies
\mbox{\fbox{${\blacksquare}Nt(s(n))$}};\ \ \Rightarrow\ Nt(s(n))
\using {!}P
\endprooftree
\justifies
{\blacksquare}Nt(s(n));\ \ \Rightarrow\ \fbox{${\exists}aNa$}
\using {\exists}R
\endprooftree
\justifies
{\blacksquare}Nt(s(n));\ \ \Rightarrow\ \fbox{${\exists}aNa{\oplus}{\it CP}that$}
\using {\oplus}R
\endprooftree
\prooftree
\prooftree
\prooftree
\prooftree
\justifies
Nt(s(f))\ \Rightarrow\ Nt(s(f))
\endprooftree
\justifies
Nt(s(f))\ \Rightarrow\ \fbox{${\exists}aNa$}
\using {\exists}R
\endprooftree
\justifies
[Nt(s(f))]\ \Rightarrow\ \fbox{${\langle\rangle}{\exists}aNa$}
\using {\langle\rangle}R
\endprooftree
\prooftree
\justifies
\mbox{\fbox{$Sf$}}\ \Rightarrow\ Sf
\endprooftree
\justifies
[Nt(s(f))], \mbox{\fbox{${\langle\rangle}{\exists}aNa\backslash Sf$}}\ \Rightarrow\ Sf
\using {\backslash}L
\endprooftree
\justifies
{\blacksquare}Nt(s(n));\ [Nt(s(f))], \mbox{\fbox{$({\langle\rangle}{\exists}aNa\backslash Sf)/({\exists}aNa{\oplus}{\it CP}that)$}}\ \Rightarrow\ Sf
\using {/}L
\endprooftree
\justifies
{\blacksquare}Nt(s(n));\ [Nt(s(f))], \mbox{\fbox{${\square}(({\langle\rangle}{\exists}aNa\backslash Sf)/({\exists}aNa{\oplus}{\it CP}that))$}}\ \Rightarrow\ Sf
\using {\Box}L
\endprooftree
\justifies
{\blacksquare}Nt(s(n));\ {\langle\rangle}Nt(s(f)), {\square}(({\langle\rangle}{\exists}aNa\backslash Sf)/({\exists}aNa{\oplus}{\it CP}that))\ \Rightarrow\ Sf
\using {\langle\rangle}L
\endprooftree
\justifies
{\blacksquare}Nt(s(n));\ {\square}(({\langle\rangle}{\exists}aNa\backslash Sf)/({\exists}aNa{\oplus}{\it CP}that))\ \Rightarrow\ {\langle\rangle}Nt(s(f))\backslash Sf
\using {\backslash}R
\endprooftree
\prooftree
\prooftree
\prooftree
\prooftree
\justifies
\mbox{\fbox{$Nt(s(f))$}}\ \Rightarrow\ Nt(s(f))
\endprooftree
\justifies
\mbox{\fbox{${\blacksquare}Nt(s(f))$}}\ \Rightarrow\ Nt(s(f))
\using {\blacksquare}L
\endprooftree
\justifies
[{\blacksquare}Nt(s(f))]\ \Rightarrow\ \fbox{${\langle\rangle}Nt(s(f))$}
\using {\langle\rangle}R
\endprooftree
\prooftree
\justifies
\mbox{\fbox{$Sf$}}\ \Rightarrow\ Sf
\endprooftree
\justifies
[{\blacksquare}Nt(s(f))], \mbox{\fbox{${\langle\rangle}Nt(s(f))\backslash Sf$}}\ \Rightarrow\ Sf
\using {\backslash}L
\endprooftree
\justifies
{\blacksquare}Nt(s(n));\ [{\blacksquare}Nt(s(f))], {\square}(({\langle\rangle}{\exists}aNa\backslash Sf)/({\exists}aNa{\oplus}{\it CP}that)), \mbox{\fbox{$({\langle\rangle}Nt(s(f))\backslash Sf)\backslash ({\langle\rangle}Nt(s(f))\backslash Sf)$}}\ \Rightarrow\ Sf
\using {\backslash}L
\endprooftree
\justifies
{\blacksquare}Nt(s(n));\ [{\blacksquare}Nt(s(f))], {\square}(({\langle\rangle}{\exists}aNa\backslash Sf)/({\exists}aNa{\oplus}{\it CP}that)), \mbox{\fbox{${\forall}f(({\langle\rangle}Nt(s(f))\backslash Sf)\backslash ({\langle\rangle}Nt(s(f))\backslash Sf))$}}\ \Rightarrow\ Sf
\using {\forall}L
\endprooftree
\justifies
{\blacksquare}Nt(s(n));\ [{\blacksquare}Nt(s(f))], {\square}(({\langle\rangle}{\exists}aNa\backslash Sf)/({\exists}aNa{\oplus}{\it CP}that)), \mbox{\fbox{${\forall}a{\forall}f(({\langle\rangle}Na\backslash Sf)\backslash ({\langle\rangle}Na\backslash Sf))$}}\ \Rightarrow\ Sf
\using {\forall}L
\endprooftree
\justifies
{\blacksquare}Nt(s(n));\ [{\blacksquare}Nt(s(f))], {\square}(({\langle\rangle}{\exists}aNa\backslash Sf)/({\exists}aNa{\oplus}{\it CP}that)), \mbox{\fbox{${\square}{\forall}a{\forall}f(({\langle\rangle}Na\backslash Sf)\backslash ({\langle\rangle}Na\backslash Sf))$}}\ \Rightarrow\ Sf
\using {\Box}L
\endprooftree
\justifies
!{\blacksquare}Nt(s(n)), [{\blacksquare}Nt(s(f))], {\square}(({\langle\rangle}{\exists}aNa\backslash Sf)/({\exists}aNa{\oplus}{\it CP}that)), {\square}{\forall}a{\forall}f(({\langle\rangle}Na\backslash Sf)\backslash ({\langle\rangle}Na\backslash Sf))\ \Rightarrow\ Sf
\using {!}L
\endprooftree
\justifies
\mbox{\fbox{${\langle\rangle}Nt(s(n)){\sqcap}!{\blacksquare}Nt(s(n))$}}, [{\blacksquare}Nt(s(f))], {\square}(({\langle\rangle}{\exists}aNa\backslash Sf)/({\exists}aNa{\oplus}{\it CP}that)), {\square}{\forall}a{\forall}f(({\langle\rangle}Na\backslash Sf)\backslash ({\langle\rangle}Na\backslash Sf))\ \Rightarrow\ Sf
\using {\sqcap}L
\endprooftree
\justifies
[{\blacksquare}Nt(s(f))], {\square}(({\langle\rangle}{\exists}aNa\backslash Sf)/({\exists}aNa{\oplus}{\it CP}that)), {\square}{\forall}a{\forall}f(({\langle\rangle}Na\backslash Sf)\backslash ({\langle\rangle}Na\backslash Sf))\ \Rightarrow\ ({\langle\rangle}Nt(s(n)){\sqcap}!{\blacksquare}Nt(s(n)))\backslash Sf
\using {\backslash}R
\endprooftree
\justifies
[{\blacksquare}Nt(s(f))], {\square}(({\langle\rangle}{\exists}aNa\backslash Sf)/({\exists}aNa{\oplus}{\it CP}that)), {\square}{\forall}a{\forall}f(({\langle\rangle}Na\backslash Sf)\backslash ({\langle\rangle}Na\backslash Sf))\ \Rightarrow\ {\blacksquare}(({\langle\rangle}Nt(s(n)){\sqcap}!{\blacksquare}Nt(s(n)))\backslash Sf)
\using {\blacksquare}R
\endprooftree
\prooftree
\prooftree
\prooftree
\prooftree
\prooftree
\justifies
\mbox{\fbox{${\it CN}{\it s(n)}$}}\ \Rightarrow\ {\it CN}{\it s(n)}
\endprooftree
\justifies
\mbox{\fbox{${\square}{\it CN}{\it s(n)}$}}\ \Rightarrow\ {\it CN}{\it s(n)}
\using {\Box}L
\endprooftree
\prooftree
\justifies
\mbox{\fbox{${\it CN}{\it s(n)}$}}\ \Rightarrow\ {\it CN}{\it s(n)}
\endprooftree
\justifies
{\square}{\it CN}{\it s(n)}, \mbox{\fbox{${\it CN}{\it s(n)}\backslash {\it CN}{\it s(n)}$}}\ \Rightarrow\ {\it CN}{\it s(n)}
\using {\backslash}L
\endprooftree
\justifies
{\square}{\it CN}{\it s(n)}, [\mbox{\fbox{${[]^{-1}}({\it CN}{\it s(n)}\backslash {\it CN}{\it s(n)})$}}]\ \Rightarrow\ {\it CN}{\it s(n)}
\using {[]^{-1}}L
\endprooftree
\justifies
{\square}{\it CN}{\it s(n)}, [[\mbox{\fbox{${[]^{-1}}{[]^{-1}}({\it CN}{\it s(n)}\backslash {\it CN}{\it s(n)})$}}]]\ \Rightarrow\ {\it CN}{\it s(n)}
\using {[]^{-1}}L
\endprooftree
\justifies
{\square}{\it CN}{\it s(n)}, [[\mbox{\fbox{${[]^{-1}}{[]^{-1}}({\it CN}{\it s(n)}\backslash {\it CN}{\it s(n)})/{\blacksquare}(({\langle\rangle}Nt(s(n)){\sqcap}!{\blacksquare}Nt(s(n)))\backslash Sf)$}}, [{\blacksquare}Nt(s(f))], {\square}(({\langle\rangle}{\exists}aNa\backslash Sf)/({\exists}aNa{\oplus}{\it CP}that)), {\square}{\forall}a{\forall}f(({\langle\rangle}Na\backslash Sf)\backslash ({\langle\rangle}Na\backslash Sf))]]\ \Rightarrow\ {\it CN}{\it s(n)}
\using {/}L
\endprooftree
\justifies
{\square}{\it CN}{\it s(n)}, [[\mbox{\fbox{${\forall}n({[]^{-1}}{[]^{-1}}({\it CN}{\it n}\backslash {\it CN}{\it n})/{\blacksquare}(({\langle\rangle}Nt(n){\sqcap}!{\blacksquare}Nt(n))\backslash Sf))$}}, [{\blacksquare}Nt(s(f))], {\square}(({\langle\rangle}{\exists}aNa\backslash Sf)/({\exists}aNa{\oplus}{\it CP}that)), {\square}{\forall}a{\forall}f(({\langle\rangle}Na\backslash Sf)\backslash ({\langle\rangle}Na\backslash Sf))]]\ \Rightarrow\ {\it CN}{\it s(n)}
\using {\forall}L
\endprooftree
\justifies
{\square}{\it CN}{\it s(n)}, [[\mbox{\fbox{${\blacksquare}{\forall}n({[]^{-1}}{[]^{-1}}({\it CN}{\it n}\backslash {\it CN}{\it n})/{\blacksquare}(({\langle\rangle}Nt(n){\sqcap}!{\blacksquare}Nt(n))\backslash Sf))$}}, [{\blacksquare}Nt(s(f))], {\square}(({\langle\rangle}{\exists}aNa\backslash Sf)/({\exists}aNa{\oplus}{\it CP}that)), {\square}{\forall}a{\forall}f(({\langle\rangle}Na\backslash Sf)\backslash ({\langle\rangle}Na\backslash Sf))]]\ \Rightarrow\ {\it CN}{\it s(n)}
\using {\blacksquare}L
\endprooftree}}
\end{center}

\vspace{0.15in}

\noindent
This delivers semantics:
\disp{
$\lambda C[(\mbox{\v{}}{\it dog}\ {\it C})\wedge (\mbox{\v{}}{\it today}\ ({\it Past}\ ((\mbox{\v{}}{\it seee}\ {\it C})\ {\it m})))]$}

The next example is of VP~ellipsis:
\disp{
(tdc(58a)) $[{\bf john}]{+}{\bf slept}{+}{\bf before}{+}[{\bf mary}]{+}{\bf did}: Sf$}
\disp{
$[{\blacksquare}Nt(s(m)): {\it j}], {\square}({\langle\rangle}{\exists}gNt(s(g))\backslash Sf): \mbox{\^{}}\lambda A({\it Past}\ (\mbox{\v{}}{\it sleep}\ {\it A})), {\blacksquare}({\forall}a{\forall}f(({\langle\rangle}Na\backslash Sf)\backslash ({\langle\rangle}Na\backslash Sf))/Sf): \lambda B\lambda C\lambda D(({\it before}\ {\it B})\ ({\it C}\ {\it D})), [{\blacksquare}Nt(s(f)): {\it m}], {\blacksquare}{\forall}a{\forall}g{\forall}b{\forall}h(((({\langle\rangle}Na\backslash Sg){{}{\uparrow}{}}({\langle\rangle}Nb\backslash Sh))/({\exists}c{\langle\rangle}Nc\backslash Sf))\backslash\\
 (({\langle\rangle}Na\backslash Sg){{}{\uparrow}{}}({\langle\rangle}Nb\backslash Sh))): \lambda E\lambda F(({\it E}\ {\it F})\ {\it F})\ \Rightarrow\ Sf$}

\vspace{0.15in}

\begin{center}
\rotatebox{-90}{\tiny
\prooftree
\prooftree
\prooftree
\prooftree
\prooftree
\prooftree
\prooftree
\prooftree
\prooftree
\prooftree
\prooftree
\prooftree
\prooftree
\prooftree
\prooftree
\prooftree
\prooftree
\justifies
\mbox{\fbox{$Nt(s(f))$}}\ \Rightarrow\ Nt(s(f))
\endprooftree
\justifies
\mbox{\fbox{${\blacksquare}Nt(s(f))$}}\ \Rightarrow\ Nt(s(f))
\using {\blacksquare}L
\endprooftree
\justifies
[{\blacksquare}Nt(s(f))]\ \Rightarrow\ \fbox{${\langle\rangle}Nt(s(f))$}
\using {\langle\rangle}R
\endprooftree
\justifies
[{\blacksquare}Nt(s(f))]\ \Rightarrow\ \fbox{${\exists}c{\langle\rangle}Nc$}
\using {\exists}R
\endprooftree
\prooftree
\justifies
\mbox{\fbox{$Sf$}}\ \Rightarrow\ Sf
\endprooftree
\justifies
[{\blacksquare}Nt(s(f))], \mbox{\fbox{${\exists}c{\langle\rangle}Nc\backslash Sf$}}\ \Rightarrow\ Sf
\using {\backslash}L
\endprooftree
\prooftree
\prooftree
\prooftree
\prooftree
\prooftree
\prooftree
\prooftree
\prooftree
\justifies
Nt(s(m))\ \Rightarrow\ Nt(s(m))
\endprooftree
\justifies
[Nt(s(m))]\ \Rightarrow\ \fbox{${\langle\rangle}Nt(s(m))$}
\using {\langle\rangle}R
\endprooftree
\prooftree
\justifies
\mbox{\fbox{$Sf$}}\ \Rightarrow\ Sf
\endprooftree
\justifies
[Nt(s(m))], \mbox{\fbox{${\langle\rangle}Nt(s(m))\backslash Sf$}}\ \Rightarrow\ Sf
\using {\backslash}L
\endprooftree
\justifies
{\langle\rangle}Nt(s(m)), {\langle\rangle}Nt(s(m))\backslash Sf\ \Rightarrow\ Sf
\using {\langle\rangle}L
\endprooftree
\justifies
{\langle\rangle}Nt(s(m))\backslash Sf\ \Rightarrow\ {\langle\rangle}Nt(s(m))\backslash Sf
\using {\backslash}R
\endprooftree
\prooftree
\prooftree
\prooftree
\justifies
Nt(s(m))\ \Rightarrow\ Nt(s(m))
\endprooftree
\justifies
[Nt(s(m))]\ \Rightarrow\ \fbox{${\langle\rangle}Nt(s(m))$}
\using {\langle\rangle}R
\endprooftree
\prooftree
\justifies
\mbox{\fbox{$Sf$}}\ \Rightarrow\ Sf
\endprooftree
\justifies
[Nt(s(m))], \mbox{\fbox{${\langle\rangle}Nt(s(m))\backslash Sf$}}\ \Rightarrow\ Sf
\using {\backslash}L
\endprooftree
\justifies
[Nt(s(m))], {\langle\rangle}Nt(s(m))\backslash Sf, \mbox{\fbox{$({\langle\rangle}Nt(s(m))\backslash Sf)\backslash ({\langle\rangle}Nt(s(m))\backslash Sf)$}}\ \Rightarrow\ Sf
\using {\backslash}L
\endprooftree
\justifies
[Nt(s(m))], {\langle\rangle}Nt(s(m))\backslash Sf, \mbox{\fbox{${\forall}f(({\langle\rangle}Nt(s(m))\backslash Sf)\backslash ({\langle\rangle}Nt(s(m))\backslash Sf))$}}\ \Rightarrow\ Sf
\using {\forall}L
\endprooftree
\justifies
[Nt(s(m))], {\langle\rangle}Nt(s(m))\backslash Sf, \mbox{\fbox{${\forall}a{\forall}f(({\langle\rangle}Na\backslash Sf)\backslash ({\langle\rangle}Na\backslash Sf))$}}\ \Rightarrow\ Sf
\using {\forall}L
\endprooftree
\justifies
[Nt(s(m))], {\langle\rangle}Nt(s(m))\backslash Sf, \mbox{\fbox{${\forall}a{\forall}f(({\langle\rangle}Na\backslash Sf)\backslash ({\langle\rangle}Na\backslash Sf))/Sf$}}, [{\blacksquare}Nt(s(f))], {\exists}c{\langle\rangle}Nc\backslash Sf\ \Rightarrow\ Sf
\using {/}L
\endprooftree
\justifies
[Nt(s(m))], {\langle\rangle}Nt(s(m))\backslash Sf, \mbox{\fbox{${\blacksquare}({\forall}a{\forall}f(({\langle\rangle}Na\backslash Sf)\backslash ({\langle\rangle}Na\backslash Sf))/Sf)$}}, [{\blacksquare}Nt(s(f))], {\exists}c{\langle\rangle}Nc\backslash Sf\ \Rightarrow\ Sf
\using {\blacksquare}L
\endprooftree
\justifies
{\langle\rangle}Nt(s(m)), {\langle\rangle}Nt(s(m))\backslash Sf, {\blacksquare}({\forall}a{\forall}f(({\langle\rangle}Na\backslash Sf)\backslash ({\langle\rangle}Na\backslash Sf))/Sf), [{\blacksquare}Nt(s(f))], {\exists}c{\langle\rangle}Nc\backslash Sf\ \Rightarrow\ Sf
\using {\langle\rangle}L
\endprooftree
\justifies
{\langle\rangle}Nt(s(m))\backslash Sf, {\blacksquare}({\forall}a{\forall}f(({\langle\rangle}Na\backslash Sf)\backslash ({\langle\rangle}Na\backslash Sf))/Sf), [{\blacksquare}Nt(s(f))], {\exists}c{\langle\rangle}Nc\backslash Sf\ \Rightarrow\ {\langle\rangle}Nt(s(m))\backslash Sf
\using {\backslash}R
\endprooftree
\justifies
{\tt 1}, {\blacksquare}({\forall}a{\forall}f(({\langle\rangle}Na\backslash Sf)\backslash ({\langle\rangle}Na\backslash Sf))/Sf), [{\blacksquare}Nt(s(f))], {\exists}c{\langle\rangle}Nc\backslash Sf\ \Rightarrow\ ({\langle\rangle}Nt(s(m))\backslash Sf){{}{\uparrow}{}}({\langle\rangle}Nt(s(m))\backslash Sf)
\using {\uparrow}R
\endprooftree
\justifies
{\tt 1}, {\blacksquare}({\forall}a{\forall}f(({\langle\rangle}Na\backslash Sf)\backslash ({\langle\rangle}Na\backslash Sf))/Sf), [{\blacksquare}Nt(s(f))]\ \Rightarrow\ (({\langle\rangle}Nt(s(m))\backslash Sf){{}{\uparrow}{}}({\langle\rangle}Nt(s(m))\backslash Sf))/({\exists}c{\langle\rangle}Nc\backslash Sf)
\using {/}R
\endprooftree
\prooftree
\prooftree
\prooftree
\prooftree
\prooftree
\prooftree
\prooftree
\prooftree
\justifies
Nt(s(m))\ \Rightarrow\ Nt(s(m))
\endprooftree
\justifies
Nt(s(m))\ \Rightarrow\ \fbox{${\exists}gNt(s(g))$}
\using {\exists}R
\endprooftree
\justifies
[Nt(s(m))]\ \Rightarrow\ \fbox{${\langle\rangle}{\exists}gNt(s(g))$}
\using {\langle\rangle}R
\endprooftree
\prooftree
\justifies
\mbox{\fbox{$Sf$}}\ \Rightarrow\ Sf
\endprooftree
\justifies
[Nt(s(m))], \mbox{\fbox{${\langle\rangle}{\exists}gNt(s(g))\backslash Sf$}}\ \Rightarrow\ Sf
\using {\backslash}L
\endprooftree
\justifies
[Nt(s(m))], \mbox{\fbox{${\square}({\langle\rangle}{\exists}gNt(s(g))\backslash Sf)$}}\ \Rightarrow\ Sf
\using {\Box}L
\endprooftree
\justifies
{\langle\rangle}Nt(s(m)), {\square}({\langle\rangle}{\exists}gNt(s(g))\backslash Sf)\ \Rightarrow\ Sf
\using {\langle\rangle}L
\endprooftree
\justifies
{\square}({\langle\rangle}{\exists}gNt(s(g))\backslash Sf)\ \Rightarrow\ {\langle\rangle}Nt(s(m))\backslash Sf
\using {\backslash}R
\endprooftree
\prooftree
\prooftree
\prooftree
\prooftree
\justifies
\mbox{\fbox{$Nt(s(m))$}}\ \Rightarrow\ Nt(s(m))
\endprooftree
\justifies
\mbox{\fbox{${\blacksquare}Nt(s(m))$}}\ \Rightarrow\ Nt(s(m))
\using {\blacksquare}L
\endprooftree
\justifies
[{\blacksquare}Nt(s(m))]\ \Rightarrow\ \fbox{${\langle\rangle}Nt(s(m))$}
\using {\langle\rangle}R
\endprooftree
\prooftree
\justifies
\mbox{\fbox{$Sf$}}\ \Rightarrow\ Sf
\endprooftree
\justifies
[{\blacksquare}Nt(s(m))], \mbox{\fbox{${\langle\rangle}Nt(s(m))\backslash Sf$}}\ \Rightarrow\ Sf
\using {\backslash}L
\endprooftree
\justifies
[{\blacksquare}Nt(s(m))], \mbox{\fbox{$({\langle\rangle}Nt(s(m))\backslash Sf){{}{\uparrow}{}}({\langle\rangle}Nt(s(m))\backslash Sf)\{{\square}({\langle\rangle}{\exists}gNt(s(g))\backslash Sf)\}$}}\ \Rightarrow\ Sf
\using {\uparrow}L
\endprooftree
\justifies
[{\blacksquare}Nt(s(m))], {\square}({\langle\rangle}{\exists}gNt(s(g))\backslash Sf), {\blacksquare}({\forall}a{\forall}f(({\langle\rangle}Na\backslash Sf)\backslash ({\langle\rangle}Na\backslash Sf))/Sf), [{\blacksquare}Nt(s(f))], \mbox{\fbox{$((({\langle\rangle}Nt(s(m))\backslash Sf){{}{\uparrow}{}}({\langle\rangle}Nt(s(m))\backslash Sf))/({\exists}c{\langle\rangle}Nc\backslash Sf))\backslash (({\langle\rangle}Nt(s(m))\backslash Sf){{}{\uparrow}{}}({\langle\rangle}Nt(s(m))\backslash Sf))$}}\ \Rightarrow\ Sf
\using {\backslash}L
\endprooftree
\justifies
[{\blacksquare}Nt(s(m))], {\square}({\langle\rangle}{\exists}gNt(s(g))\backslash Sf), {\blacksquare}({\forall}a{\forall}f(({\langle\rangle}Na\backslash Sf)\backslash ({\langle\rangle}Na\backslash Sf))/Sf), [{\blacksquare}Nt(s(f))], \mbox{\fbox{${\forall}h(((({\langle\rangle}Nt(s(m))\backslash Sf){{}{\uparrow}{}}({\langle\rangle}Nt(s(m))\backslash Sh))/({\exists}c{\langle\rangle}Nc\backslash Sf))\backslash (({\langle\rangle}Nt(s(m))\backslash Sf){{}{\uparrow}{}}({\langle\rangle}Nt(s(m))\backslash Sh)))$}}\ \Rightarrow\ Sf
\using {\forall}L
\endprooftree
\justifies
[{\blacksquare}Nt(s(m))], {\square}({\langle\rangle}{\exists}gNt(s(g))\backslash Sf), {\blacksquare}({\forall}a{\forall}f(({\langle\rangle}Na\backslash Sf)\backslash ({\langle\rangle}Na\backslash Sf))/Sf), [{\blacksquare}Nt(s(f))], \mbox{\fbox{${\forall}b{\forall}h(((({\langle\rangle}Nt(s(m))\backslash Sf){{}{\uparrow}{}}({\langle\rangle}Nb\backslash Sh))/({\exists}c{\langle\rangle}Nc\backslash Sf))\backslash (({\langle\rangle}Nt(s(m))\backslash Sf){{}{\uparrow}{}}({\langle\rangle}Nb\backslash Sh)))$}}\ \Rightarrow\ Sf
\using {\forall}L
\endprooftree
\justifies
[{\blacksquare}Nt(s(m))], {\square}({\langle\rangle}{\exists}gNt(s(g))\backslash Sf), {\blacksquare}({\forall}a{\forall}f(({\langle\rangle}Na\backslash Sf)\backslash ({\langle\rangle}Na\backslash Sf))/Sf), [{\blacksquare}Nt(s(f))], \mbox{\fbox{${\forall}g{\forall}b{\forall}h(((({\langle\rangle}Nt(s(m))\backslash Sg){{}{\uparrow}{}}({\langle\rangle}Nb\backslash Sh))/({\exists}c{\langle\rangle}Nc\backslash Sf))\backslash (({\langle\rangle}Nt(s(m))\backslash Sg){{}{\uparrow}{}}({\langle\rangle}Nb\backslash Sh)))$}}\ \Rightarrow\ Sf
\using {\forall}L
\endprooftree
\justifies
[{\blacksquare}Nt(s(m))], {\square}({\langle\rangle}{\exists}gNt(s(g))\backslash Sf), {\blacksquare}({\forall}a{\forall}f(({\langle\rangle}Na\backslash Sf)\backslash ({\langle\rangle}Na\backslash Sf))/Sf), [{\blacksquare}Nt(s(f))], \mbox{\fbox{${\forall}a{\forall}g{\forall}b{\forall}h(((({\langle\rangle}Na\backslash Sg){{}{\uparrow}{}}({\langle\rangle}Nb\backslash Sh))/({\exists}c{\langle\rangle}Nc\backslash Sf))\backslash (({\langle\rangle}Na\backslash Sg){{}{\uparrow}{}}({\langle\rangle}Nb\backslash Sh)))$}}\ \Rightarrow\ Sf
\using {\forall}L
\endprooftree
\justifies
[{\blacksquare}Nt(s(m))], {\square}({\langle\rangle}{\exists}gNt(s(g))\backslash Sf), {\blacksquare}({\forall}a{\forall}f(({\langle\rangle}Na\backslash Sf)\backslash ({\langle\rangle}Na\backslash Sf))/Sf), [{\blacksquare}Nt(s(f))], \mbox{\fbox{${\blacksquare}{\forall}a{\forall}g{\forall}b{\forall}h(((({\langle\rangle}Na\backslash Sg){{}{\uparrow}{}}({\langle\rangle}Nb\backslash Sh))/({\exists}c{\langle\rangle}Nc\backslash Sf))\backslash (({\langle\rangle}Na\backslash Sg){{}{\uparrow}{}}({\langle\rangle}Nb\backslash Sh)))$}}\ \Rightarrow\ Sf
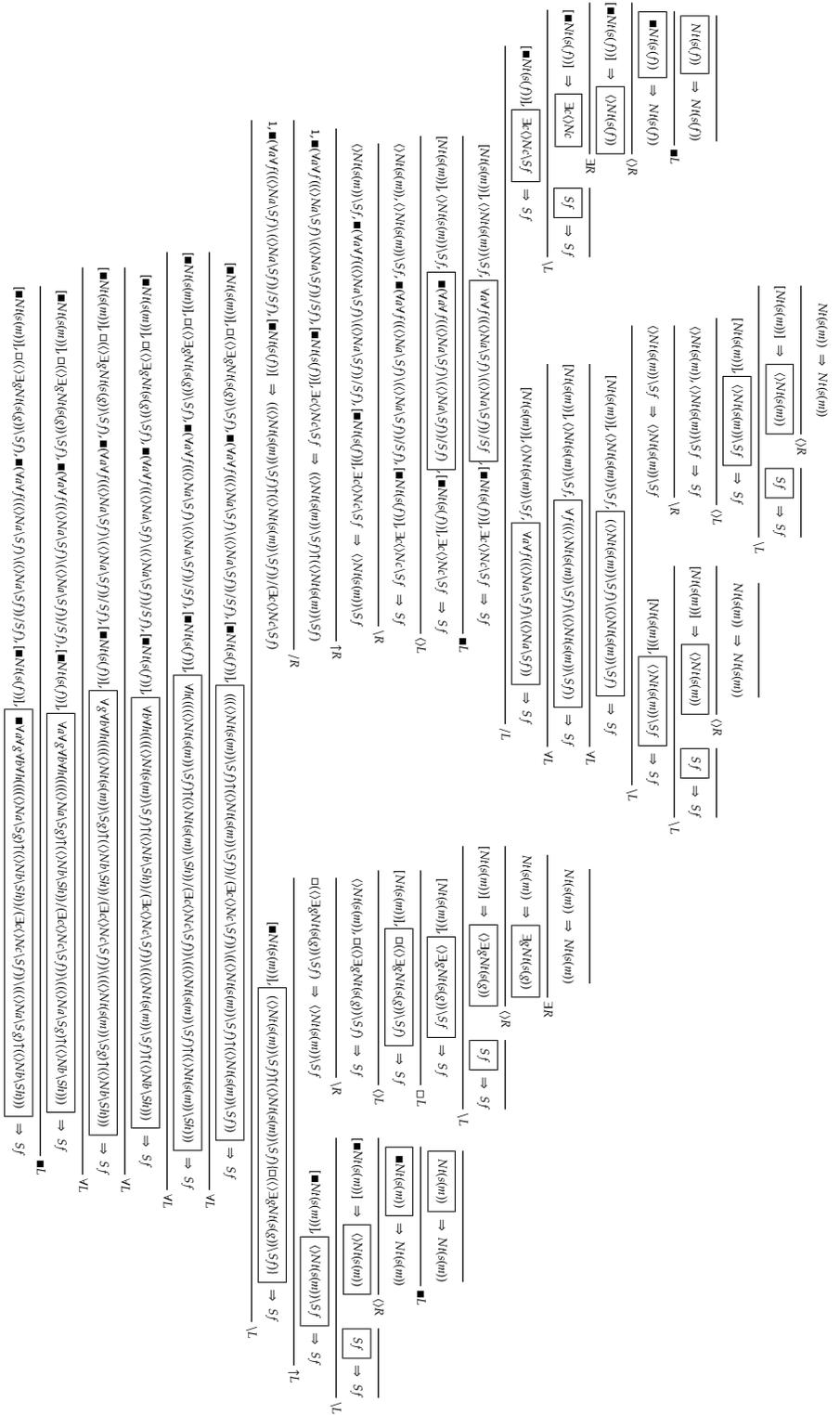
\using {\blacksquare}L
\endprooftree}
\end{center}

\vspace{0.15in}

\noindent
This delivers the semantics:
\disp{
$(({\it before}\ ({\it Past}\ (\mbox{\v{}}{\it sleep}\ {\it m})))\ ({\it Past}\ (\mbox{\v{}}{\it sleep}\ {\it j})))$}

In~tdc(64) there is medial pied-piping:
\disp{
(tdc(64)) ${\bf mountain}{+}[[{\bf the}{+}{\bf painting}{+}{\bf of}{+}{\bf which}{+}{\bf by}{+}{\bf cezanne}{+}[{\bf john}]{+}{\bf sold}{+}$\\ ${\bf for}{+}{\bf tenmilliondollars}]]: {\it CN}{\it s(n)}$
\label{piedpipingex}}
Lexical lookup yields:
\disp{
${\square}{\it CN}{\it s(n)}: {\it mountain}, [[{\blacksquare}{\forall}n(Nt(n)/{\it CN}{\it n}): \iota , {\square}({\it CN}{\it s(n)}/{\it PP}{\it of}): \mbox{\^{}}\lambda A((\mbox{\v{}}{\it of}\ {\it A})\ \mbox{\v{}}{\it painting}), \\
{\square}(({\forall}n({\it CN}{\it n}\backslash {\it CN}{\it n})/{\blacksquare}{\exists}bNb){\&}({\it PP}{\it of}/{\exists}aNa)): \mbox{\^{}}(\mbox{\v{}}{\it of}, \lambda B{\it B}), {\blacksquare}{\forall}n{\forall}m((Nt(n){{}{\uparrow}{}}Nt(m)){{}{\downarrow}{}}\\
({[]^{-1}}{[]^{-1}}({\it CN}{\it m}\backslash {\it CN}{\it m})/{\blacksquare}(({\langle\rangle}Nt(n){\sqcap}!{\blacksquare}Nt(n))\backslash Sf))): \lambda C\lambda D\lambda E\lambda F[({\it E}\ {\it F})\wedge ({\it D}\ ({\it C}\ {\it F}))], \\{\square}({\forall}n({\it CN}{\it n}\backslash {\it CN}{\it n})/{\exists}aNa): \mbox{\^{}}\lambda G\lambda H((\mbox{\v{}}{\it by}\ {\it G})\ {\it H}), {\blacksquare}Nt(s(m)): {\it c}, [{\blacksquare}Nt(s(m)): {\it j}], \\{\square}(({\langle\rangle}{\exists}aNa\backslash Sf)/({\exists}bNb{\bullet}{\it PP}{\it for})): \mbox{\^{}}\lambda I\lambda J({\it Past}\ (((\mbox{\v{}}{\it sell}\ \pi_2{\it I})\ \pi_1{\it I})\ {\it J})), \\
{\blacksquare}({\it PP}{\it for}/{\exists}aNa): \lambda K{\it K}, {\square}Nt(s(n)): {\it tenmilliondollars}]]\ \Rightarrow\ {\it CN}{\it s(n)}$}
There is the derivation:
\vspace{0.15in}
$$
{\tiny
\prooftree
\prooftree
\prooftree
\prooftree
\prooftree
\prooftree
\prooftree
\prooftree
\prooftree
\justifies
\mbox{\fbox{$Nt(s(m))$}}\ \Rightarrow\ Nt(s(m))
\endprooftree
\justifies
\mbox{\fbox{${\blacksquare}Nt(s(m))$}}\ \Rightarrow\ Nt(s(m))
\using {\blacksquare}L
\endprooftree
\justifies
{\blacksquare}Nt(s(m))\ \Rightarrow\ \fbox{${\exists}aNa$}
\using {\exists}R
\endprooftree
\prooftree
\prooftree
\prooftree
\prooftree
\prooftree
\prooftree
\prooftree
\prooftree
\prooftree
\justifies
Nt(s(n))\ \Rightarrow\ Nt(s(n))
\endprooftree
\justifies
Nt(s(n))\ \Rightarrow\ \fbox{${\exists}aNa$}
\using {\exists}R
\endprooftree
\prooftree
\justifies
\mbox{\fbox{${\it PP}{\it of}$}}\ \Rightarrow\ {\it PP}{\it of}
\endprooftree
\justifies
\mbox{\fbox{${\it PP}{\it of}/{\exists}aNa$}}, Nt(s(n))\ \Rightarrow\ {\it PP}{\it of}
\using {/}L
\endprooftree
\justifies
\mbox{\fbox{$({\forall}n({\it CN}{\it n}\backslash {\it CN}{\it n})/{\blacksquare}{\exists}bNb){\&}({\it PP}{\it of}/{\exists}aNa)$}}, Nt(s(n))\ \Rightarrow\ {\it PP}{\it of}
\using {\&}L
\endprooftree
\justifies
\mbox{\fbox{${\square}(({\forall}n({\it CN}{\it n}\backslash {\it CN}{\it n})/{\blacksquare}{\exists}bNb){\&}({\it PP}{\it of}/{\exists}aNa))$}}, Nt(s(n))\ \Rightarrow\ {\it PP}{\it of}
\using {\Box}L
\endprooftree
\prooftree
\justifies
\mbox{\fbox{${\it CN}{\it s(n)}$}}\ \Rightarrow\ {\it CN}{\it s(n)}
\endprooftree
\justifies
\mbox{\fbox{${\it CN}{\it s(n)}/{\it PP}{\it of}$}}, {\square}(({\forall}n({\it CN}{\it n}\backslash {\it CN}{\it n})/{\blacksquare}{\exists}bNb){\&}({\it PP}{\it of}/{\exists}aNa)), Nt(s(n))\ \Rightarrow\ {\it CN}{\it s(n)}
\using {/}L
\endprooftree
\justifies
\mbox{\fbox{${\square}({\it CN}{\it s(n)}/{\it PP}{\it of})$}}, {\square}(({\forall}n({\it CN}{\it n}\backslash {\it CN}{\it n})/{\blacksquare}{\exists}bNb){\&}({\it PP}{\it of}/{\exists}aNa)), Nt(s(n))\ \Rightarrow\ {\it CN}{\it s(n)}
\using {\Box}L
\endprooftree
\prooftree
\justifies
\mbox{\fbox{${\it CN}{\it s(n)}$}}\ \Rightarrow\ {\it CN}{\it s(n)}
\endprooftree
\justifies
{\square}({\it CN}{\it s(n)}/{\it PP}{\it of}), {\square}(({\forall}n({\it CN}{\it n}\backslash {\it CN}{\it n})/{\blacksquare}{\exists}bNb){\&}({\it PP}{\it of}/{\exists}aNa)), Nt(s(n)), \mbox{\fbox{${\it CN}{\it s(n)}\backslash {\it CN}{\it s(n)}$}}\ \Rightarrow\ {\it CN}{\it s(n)}
\using {\backslash}L
\endprooftree
\justifies
{\square}({\it CN}{\it s(n)}/{\it PP}{\it of}), {\square}(({\forall}n({\it CN}{\it n}\backslash {\it CN}{\it n})/{\blacksquare}{\exists}bNb){\&}({\it PP}{\it of}/{\exists}aNa)), Nt(s(n)), \mbox{\fbox{${\forall}n({\it CN}{\it n}\backslash {\it CN}{\it n})$}}\ \Rightarrow\ {\it CN}{\it s(n)}
\using {\forall}L
\endprooftree
\justifies
{\square}({\it CN}{\it s(n)}/{\it PP}{\it of}), {\square}(({\forall}n({\it CN}{\it n}\backslash {\it CN}{\it n})/{\blacksquare}{\exists}bNb){\&}({\it PP}{\it of}/{\exists}aNa)), Nt(s(n)), \mbox{\fbox{${\forall}n({\it CN}{\it n}\backslash {\it CN}{\it n})/{\exists}aNa$}}, {\blacksquare}Nt(s(m))\ \Rightarrow\ {\it CN}{\it s(n)}
\using {/}L
\endprooftree
\justifies
{\square}({\it CN}{\it s(n)}/{\it PP}{\it of}), {\square}(({\forall}n({\it CN}{\it n}\backslash {\it CN}{\it n})/{\blacksquare}{\exists}bNb){\&}({\it PP}{\it of}/{\exists}aNa)), Nt(s(n)), \mbox{\fbox{${\square}({\forall}n({\it CN}{\it n}\backslash {\it CN}{\it n})/{\exists}aNa)$}}, {\blacksquare}Nt(s(m))\ \Rightarrow\ {\it CN}{\it s(n)}
\using {\Box}L
\endprooftree
\prooftree
\justifies
\mbox{\fbox{$Nt(s(n))$}}\ \Rightarrow\ Nt(s(n))
\endprooftree
\justifies
\mbox{\fbox{$Nt(s(n))/{\it CN}{\it s(n)}$}}, {\square}({\it CN}{\it s(n)}/{\it PP}{\it of}), {\square}(({\forall}n({\it CN}{\it n}\backslash {\it CN}{\it n})/{\blacksquare}{\exists}bNb){\&}({\it PP}{\it of}/{\exists}aNa)), Nt(s(n)), {\square}({\forall}n({\it CN}{\it n}\backslash {\it CN}{\it n})/{\exists}aNa), {\blacksquare}Nt(s(m))\ \Rightarrow\ Nt(s(n))
\using {/}L
\endprooftree
\justifies
\mbox{\fbox{${\forall}n(Nt(n)/{\it CN}{\it n})$}}, {\square}({\it CN}{\it s(n)}/{\it PP}{\it of}), {\square}(({\forall}n({\it CN}{\it n}\backslash {\it CN}{\it n})/{\blacksquare}{\exists}bNb){\&}({\it PP}{\it of}/{\exists}aNa)), Nt(s(n)), {\square}({\forall}n({\it CN}{\it n}\backslash {\it CN}{\it n})/{\exists}aNa), {\blacksquare}Nt(s(m))\ \Rightarrow\ Nt(s(n))
\using {\forall}L
\endprooftree
\justifies
\mbox{\fbox{${\blacksquare}{\forall}n(Nt(n)/{\it CN}{\it n})$}}, {\square}({\it CN}{\it s(n)}/{\it PP}{\it of}), {\square}(({\forall}n({\it CN}{\it n}\backslash {\it CN}{\it n})/{\blacksquare}{\exists}bNb){\&}({\it PP}{\it of}/{\exists}aNa)), Nt(s(n)), {\square}({\forall}n({\it CN}{\it n}\backslash {\it CN}{\it n})/{\exists}aNa), {\blacksquare}Nt(s(m))\ \Rightarrow\ Nt(s(n))
\using {\blacksquare}L
\endprooftree
\justifies
\begin{array}{c}
{\blacksquare}{\forall}n(Nt(n)/{\it CN}{\it n}), {\square}({\it CN}{\it s(n)}/{\it PP}{\it of}), {\square}(({\forall}n({\it CN}{\it n}\backslash {\it CN}{\it n})/{\blacksquare}{\exists}bNb){\&}({\it PP}{\it of}/{\exists}aNa)), {\tt 1}, {\square}({\forall}n({\it CN}{\it n}\backslash {\it CN}{\it n})/{\exists}aNa), {\blacksquare}Nt(s(m))\ \Rightarrow\ Nt(s(n)){{}{\uparrow}{}}Nt(s(n))\\
\mbox{\footnotesize\textcircled{1}}
\end{array}
\using {\uparrow}R
\endprooftree}
$$
$$
{\tiny
\prooftree
\prooftree
\prooftree
\prooftree
\prooftree
\prooftree
\prooftree
\prooftree
\prooftree
\prooftree
\prooftree
\justifies
\mbox{\fbox{$Nt(s(n))$}}\ \Rightarrow\ Nt(s(n))
\endprooftree
\justifies
\mbox{\fbox{${\blacksquare}Nt(s(n))$}}\ \Rightarrow\ Nt(s(n))
\using {\blacksquare}L
\endprooftree
\justifies
\mbox{\fbox{${\blacksquare}Nt(s(n))$}};\ \ \Rightarrow\ Nt(s(n))
\using {!}P
\endprooftree
\justifies
{\blacksquare}Nt(s(n));\ \ \Rightarrow\ \fbox{${\exists}bNb$}
\using {\exists}R
\endprooftree
\prooftree
\prooftree
\prooftree
\prooftree
\prooftree
\justifies
\mbox{\fbox{$Nt(s(n))$}}\ \Rightarrow\ Nt(s(n))
\endprooftree
\justifies
\mbox{\fbox{${\square}Nt(s(n))$}}\ \Rightarrow\ Nt(s(n))
\using {\Box}L
\endprooftree
\justifies
{\square}Nt(s(n))\ \Rightarrow\ \fbox{${\exists}aNa$}
\using {\exists}R
\endprooftree
\prooftree
\justifies
\mbox{\fbox{${\it PP}{\it for}$}}\ \Rightarrow\ {\it PP}{\it for}
\endprooftree
\justifies
\mbox{\fbox{${\it PP}{\it for}/{\exists}aNa$}}, {\square}Nt(s(n))\ \Rightarrow\ {\it PP}{\it for}
\using {/}L
\endprooftree
\justifies
\mbox{\fbox{${\blacksquare}({\it PP}{\it for}/{\exists}aNa)$}}, {\square}Nt(s(n))\ \Rightarrow\ {\it PP}{\it for}
\using {\blacksquare}L
\endprooftree
\justifies
{\blacksquare}Nt(s(n));\ {\blacksquare}({\it PP}{\it for}/{\exists}aNa), {\square}Nt(s(n))\ \Rightarrow\ \fbox{${\exists}bNb{\bullet}{\it PP}{\it for}$}
\using {\bullet}R
\endprooftree
\prooftree
\prooftree
\prooftree
\prooftree
\prooftree
\justifies
\mbox{\fbox{$Nt(s(m))$}}\ \Rightarrow\ Nt(s(m))
\endprooftree
\justifies
\mbox{\fbox{${\blacksquare}Nt(s(m))$}}\ \Rightarrow\ Nt(s(m))
\using {\blacksquare}L
\endprooftree
\justifies
{\blacksquare}Nt(s(m))\ \Rightarrow\ \fbox{${\exists}aNa$}
\using {\exists}R
\endprooftree
\justifies
[{\blacksquare}Nt(s(m))]\ \Rightarrow\ \fbox{${\langle\rangle}{\exists}aNa$}
\using {\langle\rangle}R
\endprooftree
\prooftree
\justifies
\mbox{\fbox{$Sf$}}\ \Rightarrow\ Sf
\endprooftree
\justifies
[{\blacksquare}Nt(s(m))], \mbox{\fbox{${\langle\rangle}{\exists}aNa\backslash Sf$}}\ \Rightarrow\ Sf
\using {\backslash}L
\endprooftree
\justifies
{\blacksquare}Nt(s(n));\ [{\blacksquare}Nt(s(m))], \mbox{\fbox{$({\langle\rangle}{\exists}aNa\backslash Sf)/({\exists}bNb{\bullet}{\it PP}{\it for})$}}, {\blacksquare}({\it PP}{\it for}/{\exists}aNa), {\square}Nt(s(n))\ \Rightarrow\ Sf
\using {/}L
\endprooftree
\justifies
{\blacksquare}Nt(s(n));\ [{\blacksquare}Nt(s(m))], \mbox{\fbox{${\square}(({\langle\rangle}{\exists}aNa\backslash Sf)/({\exists}bNb{\bullet}{\it PP}{\it for}))$}}, {\blacksquare}({\it PP}{\it for}/{\exists}aNa), {\square}Nt(s(n))\ \Rightarrow\ Sf
\using {\Box}L
\endprooftree
\justifies
!{\blacksquare}Nt(s(n)), [{\blacksquare}Nt(s(m))], {\square}(({\langle\rangle}{\exists}aNa\backslash Sf)/({\exists}bNb{\bullet}{\it PP}{\it for})), {\blacksquare}({\it PP}{\it for}/{\exists}aNa), {\square}Nt(s(n))\ \Rightarrow\ Sf
\using {!}L
\endprooftree
\justifies
\mbox{\fbox{${\langle\rangle}Nt(s(n)){\sqcap}!{\blacksquare}Nt(s(n))$}}, [{\blacksquare}Nt(s(m))], {\square}(({\langle\rangle}{\exists}aNa\backslash Sf)/({\exists}bNb{\bullet}{\it PP}{\it for})), {\blacksquare}({\it PP}{\it for}/{\exists}aNa), {\square}Nt(s(n))\ \Rightarrow\ Sf
\using {\sqcap}L
\endprooftree
\justifies
[{\blacksquare}Nt(s(m))], {\square}(({\langle\rangle}{\exists}aNa\backslash Sf)/({\exists}bNb{\bullet}{\it PP}{\it for})), {\blacksquare}({\it PP}{\it for}/{\exists}aNa), {\square}Nt(s(n))\ \Rightarrow\ ({\langle\rangle}Nt(s(n)){\sqcap}!{\blacksquare}Nt(s(n)))\backslash Sf
\using {\backslash}R
\endprooftree
\justifies
\begin{array}{c}
{}[{\blacksquare}Nt(s(m))], {\square}(({\langle\rangle}{\exists}aNa\backslash Sf)/({\exists}bNb{\bullet}{\it PP}{\it for})), {\blacksquare}({\it PP}{\it for}/{\exists}aNa), {\square}Nt(s(n))\ \Rightarrow\ {\blacksquare}(({\langle\rangle}Nt(s(n)){\sqcap}!{\blacksquare}Nt(s(n)))\backslash Sf)\\
\mbox{\footnotesize\textcircled{2}}
\end{array}
\using {\blacksquare}R
\endprooftree}
$$

\begin{center}
\rotatebox{-90}{\tiny
\resizebox{\textheight}{!}{
\prooftree
\prooftree
\prooftree
\prooftree
\mbox{\footnotesize\textcircled{1}}\tab\tab\tab\tab\tab
\prooftree
\mbox{\footnotesize\textcircled{2}}\tab
\prooftree
\prooftree
\prooftree
\prooftree
\prooftree
\justifies
\mbox{\fbox{${\it CN}{\it s(n)}$}}\ \Rightarrow\ {\it CN}{\it s(n)}
\endprooftree
\justifies
\mbox{\fbox{${\square}{\it CN}{\it s(n)}$}}\ \Rightarrow\ {\it CN}{\it s(n)}
\using {\Box}L
\endprooftree
\prooftree
\justifies
\mbox{\fbox{${\it CN}{\it s(n)}$}}\ \Rightarrow\ {\it CN}{\it s(n)}
\endprooftree
\justifies
{\square}{\it CN}{\it s(n)}, \mbox{\fbox{${\it CN}{\it s(n)}\backslash {\it CN}{\it s(n)}$}}\ \Rightarrow\ {\it CN}{\it s(n)}
\using {\backslash}L
\endprooftree
\justifies
{\square}{\it CN}{\it s(n)}, [\mbox{\fbox{${[]^{-1}}({\it CN}{\it s(n)}\backslash {\it CN}{\it s(n)})$}}]\ \Rightarrow\ {\it CN}{\it s(n)}
\using {[]^{-1}}L
\endprooftree
\justifies
{\square}{\it CN}{\it s(n)}, [[\mbox{\fbox{${[]^{-1}}{[]^{-1}}({\it CN}{\it s(n)}\backslash {\it CN}{\it s(n)})$}}]]\ \Rightarrow\ {\it CN}{\it s(n)}
\using {[]^{-1}}L
\endprooftree
\justifies
{\square}{\it CN}{\it s(n)}, [[\mbox{\fbox{${[]^{-1}}{[]^{-1}}({\it CN}{\it s(n)}\backslash {\it CN}{\it s(n)})/{\blacksquare}(({\langle\rangle}Nt(s(n)){\sqcap}!{\blacksquare}Nt(s(n)))\backslash Sf)$}}, [{\blacksquare}Nt(s(m))], {\square}(({\langle\rangle}{\exists}aNa\backslash Sf)/({\exists}bNb{\bullet}{\it PP}{\it for})), {\blacksquare}({\it PP}{\it for}/{\exists}aNa), {\square}Nt(s(n))]]\ \Rightarrow\ {\it CN}{\it s(n)}
\using {/}L
\endprooftree
\justifies
{\square}{\it CN}{\it s(n)}, [[{\blacksquare}{\forall}n(Nt(n)/{\it CN}{\it n}), {\square}({\it CN}{\it s(n)}/{\it PP}{\it of}), {\square}(({\forall}n({\it CN}{\it n}\backslash {\it CN}{\it n})/{\blacksquare}{\exists}bNb){\&}({\it PP}{\it of}/{\exists}aNa)), \mbox{\fbox{$(Nt(s(n)){{}{\uparrow}{}}Nt(s(n))){{}{\downarrow}{}}({[]^{-1}}{[]^{-1}}({\it CN}{\it s(n)}\backslash {\it CN}{\it s(n)})/{\blacksquare}(({\langle\rangle}Nt(s(n)){\sqcap}!{\blacksquare}Nt(s(n)))\backslash Sf))$}}, {\square}({\forall}n({\it CN}{\it n}\backslash {\it CN}{\it n})/{\exists}aNa), {\blacksquare}Nt(s(m)), [{\blacksquare}Nt(s(m))], {\square}(({\langle\rangle}{\exists}aNa\backslash Sf)/({\exists}bNb{\bullet}{\it PP}{\it for})), {\blacksquare}({\it PP}{\it for}/{\exists}aNa), {\square}Nt(s(n))]]\ \Rightarrow\ {\it CN}{\it s(n)}
\using {\downarrow}L
\endprooftree
\justifies
{\square}{\it CN}{\it s(n)}, [[{\blacksquare}{\forall}n(Nt(n)/{\it CN}{\it n}), {\square}({\it CN}{\it s(n)}/{\it PP}{\it of}), {\square}(({\forall}n({\it CN}{\it n}\backslash {\it CN}{\it n})/{\blacksquare}{\exists}bNb){\&}({\it PP}{\it of}/{\exists}aNa)), \mbox{\fbox{${\forall}m((Nt(s(n)){{}{\uparrow}{}}Nt(m)){{}{\downarrow}{}}({[]^{-1}}{[]^{-1}}({\it CN}{\it m}\backslash {\it CN}{\it m})/{\blacksquare}(({\langle\rangle}Nt(s(n)){\sqcap}!{\blacksquare}Nt(s(n)))\backslash Sf)))$}}, {\square}({\forall}n({\it CN}{\it n}\backslash {\it CN}{\it n})/{\exists}aNa), {\blacksquare}Nt(s(m)), [{\blacksquare}Nt(s(m))], {\square}(({\langle\rangle}{\exists}aNa\backslash Sf)/({\exists}bNb{\bullet}{\it PP}{\it for})), {\blacksquare}({\it PP}{\it for}/{\exists}aNa), {\square}Nt(s(n))]]\ \Rightarrow\ {\it CN}{\it s(n)}
\using {\forall}L
\endprooftree
\justifies
{\square}{\it CN}{\it s(n)}, [[{\blacksquare}{\forall}n(Nt(n)/{\it CN}{\it n}), {\square}({\it CN}{\it s(n)}/{\it PP}{\it of}), {\square}(({\forall}n({\it CN}{\it n}\backslash {\it CN}{\it n})/{\blacksquare}{\exists}bNb){\&}({\it PP}{\it of}/{\exists}aNa)), \mbox{\fbox{${\forall}n{\forall}m((Nt(n){{}{\uparrow}{}}Nt(m)){{}{\downarrow}{}}({[]^{-1}}{[]^{-1}}({\it CN}{\it m}\backslash {\it CN}{\it m})/{\blacksquare}(({\langle\rangle}Nt(n){\sqcap}!{\blacksquare}Nt(n))\backslash Sf)))$}}, {\square}({\forall}n({\it CN}{\it n}\backslash {\it CN}{\it n})/{\exists}aNa), {\blacksquare}Nt(s(m)), [{\blacksquare}Nt(s(m))], {\square}(({\langle\rangle}{\exists}aNa\backslash Sf)/({\exists}bNb{\bullet}{\it PP}{\it for})), {\blacksquare}({\it PP}{\it for}/{\exists}aNa), {\square}Nt(s(n))]]\ \Rightarrow\ {\it CN}{\it s(n)}
\using {\forall}L
\endprooftree
\justifies
{\square}{\it CN}{\it s(n)}, [[{\blacksquare}{\forall}n(Nt(n)/{\it CN}{\it n}), {\square}({\it CN}{\it s(n)}/{\it PP}{\it of}), {\square}(({\forall}n({\it CN}{\it n}\backslash {\it CN}{\it n})/{\blacksquare}{\exists}bNb){\&}({\it PP}{\it of}/{\exists}aNa)), \mbox{\fbox{${\blacksquare}{\forall}n{\forall}m((Nt(n){{}{\uparrow}{}}Nt(m)){{}{\downarrow}{}}({[]^{-1}}{[]^{-1}}({\it CN}{\it m}\backslash {\it CN}{\it m})/{\blacksquare}(({\langle\rangle}Nt(n){\sqcap}!{\blacksquare}Nt(n))\backslash Sf)))$}}, {\square}({\forall}n({\it CN}{\it n}\backslash {\it CN}{\it n})/{\exists}aNa), {\blacksquare}Nt(s(m)), [{\blacksquare}Nt(s(m))], {\square}(({\langle\rangle}{\exists}aNa\backslash Sf)/({\exists}bNb{\bullet}{\it PP}{\it for})), {\blacksquare}({\it PP}{\it for}/{\exists}aNa), {\square}Nt(s(n))]]\ \Rightarrow\ {\it CN}{\it s(n)}
\using {\blacksquare}L
\endprooftree}}
\end{center}

\vspace{0.15in}

This delivers semantics:
\disp{
$\lambda D[(\mbox{\v{}}{\it mountain}\ {\it D})\wedge ({\it Past}\ (((\mbox{\v{}}{\it sell}\ \mbox{\v{}}{\it tenmilliondollars})\ (\iota \ ((\mbox{\v{}}{\it by}\ {\it c})\ ((\mbox{\v{}}{\it of}\ {\it D})\ \mbox{\v{}}{\it painting}))))\ {\it j}))]$}

In appositive relativisation a relative clause, marked off by a prosodic phrase,
modifies a full noun phrase:
\disp{
(tdc(67)) $[{\bf john}{+}[[{\bf who}{+}{\bf jogs}]]]{+}{\bf sneezed}: Sf$}
The lexical lookup yields:
\disp{
$[{\blacksquare}Nt(s(m)): {\it j}, [[{\blacksquare}{\forall}h{\forall}n({[]^{-1}}{[]^{-1}}(Nt(n)\backslash ((Sh{{}{\uparrow}{}}Nt(n)){{}{\downarrow}{}}Sh))/{\blacksquare}(({\langle\rangle}Nt(n){\sqcap}!{\blacksquare}Nt(n))\backslash Sf)):\\ \lambda A\lambda B\lambda C[({\it A}\ {\it B})\wedge ({\it C}\ {\it B})], {\square}({\langle\rangle}{\exists}gNt(s(g))\backslash Sf): \mbox{\^{}}\lambda D({\it Pres}\ (\mbox{\v{}}{\it jog}\ {\it D}))]]], {\square}({\langle\rangle}{\exists}gNt(s(g))\backslash Sf):\\\mbox{\^{}}\lambda E({\it Past}\ (\mbox{\v{}}{\it sneeze}\ {\it E}))\ \Rightarrow\ Sf$}
There is the derivation:
\vspace{0.15in}
$$
{\tiny
\prooftree
\prooftree
\prooftree
\prooftree
\prooftree
\prooftree
\prooftree
\prooftree
\prooftree
\prooftree
\prooftree
\prooftree
\prooftree
\justifies
Nt(s(m))\ \Rightarrow\ Nt(s(m))
\endprooftree
\justifies
Nt(s(m))\ \Rightarrow\ \fbox{${\exists}gNt(s(g))$}
\using {\exists}R
\endprooftree
\justifies
[Nt(s(m))]\ \Rightarrow\ \fbox{${\langle\rangle}{\exists}gNt(s(g))$}
\using {\langle\rangle}R
\endprooftree
\prooftree
\justifies
\mbox{\fbox{$Sf$}}\ \Rightarrow\ Sf
\endprooftree
\justifies
[Nt(s(m))], \mbox{\fbox{${\langle\rangle}{\exists}gNt(s(g))\backslash Sf$}}\ \Rightarrow\ Sf
\using {\backslash}L
\endprooftree
\justifies
[Nt(s(m))], \mbox{\fbox{${\square}({\langle\rangle}{\exists}gNt(s(g))\backslash Sf)$}}\ \Rightarrow\ Sf
\using {\Box}L
\endprooftree
\justifies
{\langle\rangle}Nt(s(m)), {\square}({\langle\rangle}{\exists}gNt(s(g))\backslash Sf)\ \Rightarrow\ Sf
\using {\langle\rangle}L
\endprooftree
\justifies
\mbox{\fbox{${\langle\rangle}Nt(s(m)){\sqcap}!{\blacksquare}Nt(s(m))$}}, {\square}({\langle\rangle}{\exists}gNt(s(g))\backslash Sf)\ \Rightarrow\ Sf
\using {\sqcap}L
\endprooftree
\justifies
{\square}({\langle\rangle}{\exists}gNt(s(g))\backslash Sf)\ \Rightarrow\ ({\langle\rangle}Nt(s(m)){\sqcap}!{\blacksquare}Nt(s(m)))\backslash Sf
\using {\backslash}R
\endprooftree
\justifies
{\square}({\langle\rangle}{\exists}gNt(s(g))\backslash Sf)\ \Rightarrow\ {\blacksquare}(({\langle\rangle}Nt(s(m)){\sqcap}!{\blacksquare}Nt(s(m)))\backslash Sf)
\using {\blacksquare}R
\endprooftree
\prooftree
\prooftree
\prooftree
\prooftree
\prooftree
\justifies
\mbox{\fbox{$Nt(s(m))$}}\ \Rightarrow\ Nt(s(m))
\endprooftree
\justifies
\mbox{\fbox{${\blacksquare}Nt(s(m))$}}\ \Rightarrow\ Nt(s(m))
\using {\blacksquare}L
\endprooftree
\prooftree
\prooftree
\prooftree
\prooftree
\prooftree
\prooftree
\prooftree
\justifies
Nt(s(m))\ \Rightarrow\ Nt(s(m))
\endprooftree
\justifies
Nt(s(m))\ \Rightarrow\ \fbox{${\exists}gNt(s(g))$}
\using {\exists}R
\endprooftree
\justifies
[Nt(s(m))]\ \Rightarrow\ \fbox{${\langle\rangle}{\exists}gNt(s(g))$}
\using {\langle\rangle}R
\endprooftree
\prooftree
\justifies
\mbox{\fbox{$Sf$}}\ \Rightarrow\ Sf
\endprooftree
\justifies
[Nt(s(m))], \mbox{\fbox{${\langle\rangle}{\exists}gNt(s(g))\backslash Sf$}}\ \Rightarrow\ Sf
\using {\backslash}L
\endprooftree
\justifies
[Nt(s(m))], \mbox{\fbox{${\square}({\langle\rangle}{\exists}gNt(s(g))\backslash Sf)$}}\ \Rightarrow\ Sf
\using {\Box}L
\endprooftree
\justifies
[{\tt 1}], {\square}({\langle\rangle}{\exists}gNt(s(g))\backslash Sf)\ \Rightarrow\ Sf{{}{\uparrow}{}}Nt(s(m))
\using {\uparrow}R
\endprooftree
\prooftree
\justifies
\mbox{\fbox{$Sf$}}\ \Rightarrow\ Sf
\endprooftree
\justifies
[\mbox{\fbox{$(Sf{{}{\uparrow}{}}Nt(s(m))){{}{\downarrow}{}}Sf$}}], {\square}({\langle\rangle}{\exists}gNt(s(g))\backslash Sf)\ \Rightarrow\ Sf
\using {\downarrow}L
\endprooftree
\justifies
[{\blacksquare}Nt(s(m)), \mbox{\fbox{$Nt(s(m))\backslash ((Sf{{}{\uparrow}{}}Nt(s(m))){{}{\downarrow}{}}Sf)$}}], {\square}({\langle\rangle}{\exists}gNt(s(g))\backslash Sf)\ \Rightarrow\ Sf
\using {\backslash}L
\endprooftree
\justifies
[{\blacksquare}Nt(s(m)), [\mbox{\fbox{${[]^{-1}}(Nt(s(m))\backslash ((Sf{{}{\uparrow}{}}Nt(s(m))){{}{\downarrow}{}}Sf))$}}]], {\square}({\langle\rangle}{\exists}gNt(s(g))\backslash Sf)\ \Rightarrow\ Sf
\using {[]^{-1}}L
\endprooftree
\justifies
[{\blacksquare}Nt(s(m)), [[\mbox{\fbox{${[]^{-1}}{[]^{-1}}(Nt(s(m))\backslash ((Sf{{}{\uparrow}{}}Nt(s(m))){{}{\downarrow}{}}Sf))$}}]]], {\square}({\langle\rangle}{\exists}gNt(s(g))\backslash Sf)\ \Rightarrow\ Sf
\using {[]^{-1}}L
\endprooftree
\justifies
[{\blacksquare}Nt(s(m)), [[\mbox{\fbox{${[]^{-1}}{[]^{-1}}(Nt(s(m))\backslash ((Sf{{}{\uparrow}{}}Nt(s(m))){{}{\downarrow}{}}Sf))/{\blacksquare}(({\langle\rangle}Nt(s(m)){\sqcap}!{\blacksquare}Nt(s(m)))\backslash Sf)$}}, {\square}({\langle\rangle}{\exists}gNt(s(g))\backslash Sf)]]], {\square}({\langle\rangle}{\exists}gNt(s(g))\backslash Sf)\ \Rightarrow\ Sf
\using {/}L
\endprooftree
\justifies
[{\blacksquare}Nt(s(m)), [[\mbox{\fbox{${\forall}n({[]^{-1}}{[]^{-1}}(Nt(n)\backslash ((Sf{{}{\uparrow}{}}Nt(n)){{}{\downarrow}{}}Sf))/{\blacksquare}(({\langle\rangle}Nt(n){\sqcap}!{\blacksquare}Nt(n))\backslash Sf))$}}, {\square}({\langle\rangle}{\exists}gNt(s(g))\backslash Sf)]]], {\square}({\langle\rangle}{\exists}gNt(s(g))\backslash Sf)\ \Rightarrow\ Sf
\using {\forall}L
\endprooftree
\justifies
[{\blacksquare}Nt(s(m)), [[\mbox{\fbox{${\forall}h{\forall}n({[]^{-1}}{[]^{-1}}(Nt(n)\backslash ((Sh{{}{\uparrow}{}}Nt(n)){{}{\downarrow}{}}Sh))/{\blacksquare}(({\langle\rangle}Nt(n){\sqcap}!{\blacksquare}Nt(n))\backslash Sf))$}}, {\square}({\langle\rangle}{\exists}gNt(s(g))\backslash Sf)]]], {\square}({\langle\rangle}{\exists}gNt(s(g))\backslash Sf)\ \Rightarrow\ Sf
\using {\forall}L
\endprooftree
\justifies
[{\blacksquare}Nt(s(m)), [[\mbox{\fbox{${\blacksquare}{\forall}h{\forall}n({[]^{-1}}{[]^{-1}}(Nt(n)\backslash ((Sh{{}{\uparrow}{}}Nt(n)){{}{\downarrow}{}}Sh))/{\blacksquare}(({\langle\rangle}Nt(n){\sqcap}!{\blacksquare}Nt(n))\backslash Sf))$}}, {\square}({\langle\rangle}{\exists}gNt(s(g))\backslash Sf)]]], {\square}({\langle\rangle}{\exists}gNt(s(g))\backslash Sf)\ \Rightarrow\ Sf
\using {\blacksquare}L
\endprooftree}
$$
\vspace{0.15in}
\noindent
This delivers semantics:
\disp{
$[({\it Pres}\ (\mbox{\v{}}{\it jog}\ {\it j}))\wedge ({\it Past}\ (\mbox{\v{}}{\it sneeze}\ {\it j}))]$}

There follow four placements of a parenthetical adverbial within a sentence
all derived from the same types and all assigning the same semantics.
First:
\disp{
(tdc(70a)) ${\bf fortunately}{+}[{\bf john}]{+}{\bf has}{+}{\bf perseverance}: Sf$}

\disp{
${\square}{\forall}f(\mbox{$\check{\ }$}Sf{{}{\downarrow}{}}Sf): {\it fortunately}, [{\blacksquare}Nt(s(m)): {\it j}], {\square}(({\langle\rangle}{\exists}gNt(s(g))\backslash Sf)/{\exists}aNa):\\ \mbox{\^{}}\lambda A\lambda B({\it Pres}\ ((\mbox{\v{}}{\it have}\ {\it A})\ {\it B})),
 {\square}(Nt(s(n)){\&}{\it CN}{\it s(n)}): \mbox{\^{}}(({\it gen}\ \mbox{\v{}}{\it perseverance}), \mbox{\v{}}{\it perseverance})\ \\\Rightarrow\ Sf$}
\vspace{0.15in}
$$
{\tiny
\prooftree
\prooftree
\prooftree
\prooftree
\prooftree
\prooftree
\prooftree
\prooftree
\prooftree
\prooftree
\justifies
\mbox{\fbox{$Nt(s(n))$}}\ \Rightarrow\ Nt(s(n))
\endprooftree
\justifies
\mbox{\fbox{$Nt(s(n)){\&}{\it CN}{\it s(n)}$}}\ \Rightarrow\ Nt(s(n))
\using {\&}L
\endprooftree
\justifies
\mbox{\fbox{${\square}(Nt(s(n)){\&}{\it CN}{\it s(n)})$}}\ \Rightarrow\ Nt(s(n))
\using {\Box}L
\endprooftree
\justifies
{\square}(Nt(s(n)){\&}{\it CN}{\it s(n)})\ \Rightarrow\ \fbox{${\exists}aNa$}
\using {\exists}R
\endprooftree
\prooftree
\prooftree
\prooftree
\prooftree
\prooftree
\justifies
\mbox{\fbox{$Nt(s(m))$}}\ \Rightarrow\ Nt(s(m))
\endprooftree
\justifies
\mbox{\fbox{${\blacksquare}Nt(s(m))$}}\ \Rightarrow\ Nt(s(m))
\using {\blacksquare}L
\endprooftree
\justifies
{\blacksquare}Nt(s(m))\ \Rightarrow\ \fbox{${\exists}gNt(s(g))$}
\using {\exists}R
\endprooftree
\justifies
[{\blacksquare}Nt(s(m))]\ \Rightarrow\ \fbox{${\langle\rangle}{\exists}gNt(s(g))$}
\using {\langle\rangle}R
\endprooftree
\prooftree
\justifies
\mbox{\fbox{$Sf$}}\ \Rightarrow\ Sf
\endprooftree
\justifies
[{\blacksquare}Nt(s(m))], \mbox{\fbox{${\langle\rangle}{\exists}gNt(s(g))\backslash Sf$}}\ \Rightarrow\ Sf
\using {\backslash}L
\endprooftree
\justifies
[{\blacksquare}Nt(s(m))], \mbox{\fbox{$({\langle\rangle}{\exists}gNt(s(g))\backslash Sf)/{\exists}aNa$}}, {\square}(Nt(s(n)){\&}{\it CN}{\it s(n)})\ \Rightarrow\ Sf
\using {/}L
\endprooftree
\justifies
[{\blacksquare}Nt(s(m))], \mbox{\fbox{${\square}(({\langle\rangle}{\exists}gNt(s(g))\backslash Sf)/{\exists}aNa)$}}, {\square}(Nt(s(n)){\&}{\it CN}{\it s(n)})\ \Rightarrow\ Sf
\using {\Box}L
\endprooftree
\justifies
{\tt 1}, [{\blacksquare}Nt(s(m))], {\square}(({\langle\rangle}{\exists}gNt(s(g))\backslash Sf)/{\exists}aNa), {\square}(Nt(s(n)){\&}{\it CN}{\it s(n)})\ \Rightarrow\ \mbox{$\check{\ }$}Sf
\using \mbox{$\check{\ }$}R
\endprooftree
\prooftree
\justifies
\mbox{\fbox{$Sf$}}\ \Rightarrow\ Sf
\endprooftree
\justifies
\mbox{\fbox{$\mbox{$\check{\ }$}Sf{{}{\downarrow}{}}Sf$}}, [{\blacksquare}Nt(s(m))], {\square}(({\langle\rangle}{\exists}gNt(s(g))\backslash Sf)/{\exists}aNa), {\square}(Nt(s(n)){\&}{\it CN}{\it s(n)})\ \Rightarrow\ Sf
\using {\downarrow}L
\endprooftree
\justifies
\mbox{\fbox{${\forall}f(\mbox{$\check{\ }$}Sf{{}{\downarrow}{}}Sf)$}}, [{\blacksquare}Nt(s(m))], {\square}(({\langle\rangle}{\exists}gNt(s(g))\backslash Sf)/{\exists}aNa), {\square}(Nt(s(n)){\&}{\it CN}{\it s(n)})\ \Rightarrow\ Sf
\using {\forall}L
\endprooftree
\justifies
\mbox{\fbox{${\square}{\forall}f(\mbox{$\check{\ }$}Sf{{}{\downarrow}{}}Sf)$}}, [{\blacksquare}Nt(s(m))], {\square}(({\langle\rangle}{\exists}gNt(s(g))\backslash Sf)/{\exists}aNa), {\square}(Nt(s(n)){\&}{\it CN}{\it s(n)})\ \Rightarrow\ Sf
\using {\Box}L
\endprooftree}
$$
\vspace{0.15in}
\disp{
$(\mbox{\v{}}{\it fortunately}\ ({\it Pres}\ ((\mbox{\v{}}{\it have}\ ({\it gen}\ \mbox{\v{}}{\it perseverance}))\ {\it j})))$}

\noindent
Second:
\disp{
(tdc(70b)) $[{\bf john}]{+}{\bf fortunately}{+}{\bf has}{+}{\bf perseverance}: Sf$}

\disp{
$[{\blacksquare}Nt(s(m)): {\it j}], {\square}{\forall}f(\mbox{$\check{\ }$}Sf{{}{\downarrow}{}}Sf): {\it fortunately}, {\square}(({\langle\rangle}{\exists}gNt(s(g))\backslash Sf)/{\exists}aNa):\\ \mbox{\^{}}\lambda A\lambda B({\it Pres}\ ((\mbox{\v{}}{\it have}\ {\it A})\ {\it B})),
 {\square}(Nt(s(n)){\&}{\it CN}{\it s(n)}): \mbox{\^{}}(({\it gen}\ \mbox{\v{}}{\it perseverance}), \mbox{\v{}}{\it perseverance})\ \\\Rightarrow\ Sf$}
\vspace{0.15in}
$$
{\tiny
\prooftree
\prooftree
\prooftree
\prooftree
\prooftree
\prooftree
\prooftree
\prooftree
\prooftree
\prooftree
\justifies
\mbox{\fbox{$Nt(s(n))$}}\ \Rightarrow\ Nt(s(n))
\endprooftree
\justifies
\mbox{\fbox{$Nt(s(n)){\&}{\it CN}{\it s(n)}$}}\ \Rightarrow\ Nt(s(n))
\using {\&}L
\endprooftree
\justifies
\mbox{\fbox{${\square}(Nt(s(n)){\&}{\it CN}{\it s(n)})$}}\ \Rightarrow\ Nt(s(n))
\using {\Box}L
\endprooftree
\justifies
{\square}(Nt(s(n)){\&}{\it CN}{\it s(n)})\ \Rightarrow\ \fbox{${\exists}aNa$}
\using {\exists}R
\endprooftree
\prooftree
\prooftree
\prooftree
\prooftree
\prooftree
\justifies
\mbox{\fbox{$Nt(s(m))$}}\ \Rightarrow\ Nt(s(m))
\endprooftree
\justifies
\mbox{\fbox{${\blacksquare}Nt(s(m))$}}\ \Rightarrow\ Nt(s(m))
\using {\blacksquare}L
\endprooftree
\justifies
{\blacksquare}Nt(s(m))\ \Rightarrow\ \fbox{${\exists}gNt(s(g))$}
\using {\exists}R
\endprooftree
\justifies
[{\blacksquare}Nt(s(m))]\ \Rightarrow\ \fbox{${\langle\rangle}{\exists}gNt(s(g))$}
\using {\langle\rangle}R
\endprooftree
\prooftree
\justifies
\mbox{\fbox{$Sf$}}\ \Rightarrow\ Sf
\endprooftree
\justifies
[{\blacksquare}Nt(s(m))], \mbox{\fbox{${\langle\rangle}{\exists}gNt(s(g))\backslash Sf$}}\ \Rightarrow\ Sf
\using {\backslash}L
\endprooftree
\justifies
[{\blacksquare}Nt(s(m))], \mbox{\fbox{$({\langle\rangle}{\exists}gNt(s(g))\backslash Sf)/{\exists}aNa$}}, {\square}(Nt(s(n)){\&}{\it CN}{\it s(n)})\ \Rightarrow\ Sf
\using {/}L
\endprooftree
\justifies
[{\blacksquare}Nt(s(m))], \mbox{\fbox{${\square}(({\langle\rangle}{\exists}gNt(s(g))\backslash Sf)/{\exists}aNa)$}}, {\square}(Nt(s(n)){\&}{\it CN}{\it s(n)})\ \Rightarrow\ Sf
\using {\Box}L
\endprooftree
\justifies
[{\blacksquare}Nt(s(m))], {\tt 1}, {\square}(({\langle\rangle}{\exists}gNt(s(g))\backslash Sf)/{\exists}aNa), {\square}(Nt(s(n)){\&}{\it CN}{\it s(n)})\ \Rightarrow\ \mbox{$\check{\ }$}Sf
\using \mbox{$\check{\ }$}R
\endprooftree
\prooftree
\justifies
\mbox{\fbox{$Sf$}}\ \Rightarrow\ Sf
\endprooftree
\justifies
[{\blacksquare}Nt(s(m))], \mbox{\fbox{$\mbox{$\check{\ }$}Sf{{}{\downarrow}{}}Sf$}}, {\square}(({\langle\rangle}{\exists}gNt(s(g))\backslash Sf)/{\exists}aNa), {\square}(Nt(s(n)){\&}{\it CN}{\it s(n)})\ \Rightarrow\ Sf
\using {\downarrow}L
\endprooftree
\justifies
[{\blacksquare}Nt(s(m))], \mbox{\fbox{${\forall}f(\mbox{$\check{\ }$}Sf{{}{\downarrow}{}}Sf)$}}, {\square}(({\langle\rangle}{\exists}gNt(s(g))\backslash Sf)/{\exists}aNa), {\square}(Nt(s(n)){\&}{\it CN}{\it s(n)})\ \Rightarrow\ Sf
\using {\forall}L
\endprooftree
\justifies
[{\blacksquare}Nt(s(m))], \mbox{\fbox{${\square}{\forall}f(\mbox{$\check{\ }$}Sf{{}{\downarrow}{}}Sf)$}}, {\square}(({\langle\rangle}{\exists}gNt(s(g))\backslash Sf)/{\exists}aNa), {\square}(Nt(s(n)){\&}{\it CN}{\it s(n)})\ \Rightarrow\ Sf
\using {\Box}L
\endprooftree}
$$
\vspace{0.15in}

\disp{
$(\mbox{\v{}}{\it fortunately}\ ({\it Pres}\ ((\mbox{\v{}}{\it have}\ ({\it gen}\ \mbox{\v{}}{\it perseverance}))\ {\it j})))$}

\noindent
Third:
\disp{
(tdc(70c)) $[{\bf john}]{+}{\bf has}{+}{\bf fortunately}{+}{\bf perseverance}: Sf$
}

\disp{
$[{\blacksquare}Nt(s(m)): {\it j}], {\square}(({\langle\rangle}{\exists}gNt(s(g))\backslash Sf)/{\exists}aNa): \mbox{\^{}}\lambda A\lambda B({\it Pres}\ ((\mbox{\v{}}{\it have}\ {\it A})\ {\it B})), {\square}{\forall}f(\mbox{$\check{\ }$}Sf{{}{\downarrow}{}}Sf):\\{\it fortunately},
 {\square}(Nt(s(n)){\&}{\it CN}{\it s(n)}): \mbox{\^{}}(({\it gen}\ \mbox{\v{}}{\it perseverance}), \mbox{\v{}}{\it perseverance})\ \Rightarrow\ Sf$}
\vspace{0.15in}
$$
{\tiny 
\prooftree
\prooftree
\prooftree
\prooftree
\prooftree
\prooftree
\prooftree
\prooftree
\prooftree
\prooftree
\justifies
\mbox{\fbox{$Nt(s(n))$}}\ \Rightarrow\ Nt(s(n))
\endprooftree
\justifies
\mbox{\fbox{$Nt(s(n)){\&}{\it CN}{\it s(n)}$}}\ \Rightarrow\ Nt(s(n))
\using {\&}L
\endprooftree
\justifies
\mbox{\fbox{${\square}(Nt(s(n)){\&}{\it CN}{\it s(n)})$}}\ \Rightarrow\ Nt(s(n))
\using {\Box}L
\endprooftree
\justifies
{\square}(Nt(s(n)){\&}{\it CN}{\it s(n)})\ \Rightarrow\ \fbox{${\exists}aNa$}
\using {\exists}R
\endprooftree
\prooftree
\prooftree
\prooftree
\prooftree
\prooftree
\justifies
\mbox{\fbox{$Nt(s(m))$}}\ \Rightarrow\ Nt(s(m))
\endprooftree
\justifies
\mbox{\fbox{${\blacksquare}Nt(s(m))$}}\ \Rightarrow\ Nt(s(m))
\using {\blacksquare}L
\endprooftree
\justifies
{\blacksquare}Nt(s(m))\ \Rightarrow\ \fbox{${\exists}gNt(s(g))$}
\using {\exists}R
\endprooftree
\justifies
[{\blacksquare}Nt(s(m))]\ \Rightarrow\ \fbox{${\langle\rangle}{\exists}gNt(s(g))$}
\using {\langle\rangle}R
\endprooftree
\prooftree
\justifies
\mbox{\fbox{$Sf$}}\ \Rightarrow\ Sf
\endprooftree
\justifies
[{\blacksquare}Nt(s(m))], \mbox{\fbox{${\langle\rangle}{\exists}gNt(s(g))\backslash Sf$}}\ \Rightarrow\ Sf
\using {\backslash}L
\endprooftree
\justifies
[{\blacksquare}Nt(s(m))], \mbox{\fbox{$({\langle\rangle}{\exists}gNt(s(g))\backslash Sf)/{\exists}aNa$}}, {\square}(Nt(s(n)){\&}{\it CN}{\it s(n)})\ \Rightarrow\ Sf
\using {/}L
\endprooftree
\justifies
[{\blacksquare}Nt(s(m))], \mbox{\fbox{${\square}(({\langle\rangle}{\exists}gNt(s(g))\backslash Sf)/{\exists}aNa)$}}, {\square}(Nt(s(n)){\&}{\it CN}{\it s(n)})\ \Rightarrow\ Sf
\using {\Box}L
\endprooftree
\justifies
[{\blacksquare}Nt(s(m))], {\square}(({\langle\rangle}{\exists}gNt(s(g))\backslash Sf)/{\exists}aNa), {\tt 1}, {\square}(Nt(s(n)){\&}{\it CN}{\it s(n)})\ \Rightarrow\ \mbox{$\check{\ }$}Sf
\using \mbox{$\check{\ }$}R
\endprooftree
\prooftree
\justifies
\mbox{\fbox{$Sf$}}\ \Rightarrow\ Sf
\endprooftree
\justifies
[{\blacksquare}Nt(s(m))], {\square}(({\langle\rangle}{\exists}gNt(s(g))\backslash Sf)/{\exists}aNa), \mbox{\fbox{$\mbox{$\check{\ }$}Sf{{}{\downarrow}{}}Sf$}}, {\square}(Nt(s(n)){\&}{\it CN}{\it s(n)})\ \Rightarrow\ Sf
\using {\downarrow}L
\endprooftree
\justifies
[{\blacksquare}Nt(s(m))], {\square}(({\langle\rangle}{\exists}gNt(s(g))\backslash Sf)/{\exists}aNa), \mbox{\fbox{${\forall}f(\mbox{$\check{\ }$}Sf{{}{\downarrow}{}}Sf)$}}, {\square}(Nt(s(n)){\&}{\it CN}{\it s(n)})\ \Rightarrow\ Sf
\using {\forall}L
\endprooftree
\justifies
[{\blacksquare}Nt(s(m))], {\square}(({\langle\rangle}{\exists}gNt(s(g))\backslash Sf)/{\exists}aNa), \mbox{\fbox{${\square}{\forall}f(\mbox{$\check{\ }$}Sf{{}{\downarrow}{}}Sf)$}}, {\square}(Nt(s(n)){\&}{\it CN}{\it s(n)})\ \Rightarrow\ Sf
\using {\Box}L
\endprooftree}
$$
\vspace{0.15in}
\disp{
$(\mbox{\v{}}{\it fortunately}\ ({\it Pres}\ ((\mbox{\v{}}{\it have}\ ({\it gen}\ \mbox{\v{}}{\it perseverance}))\ {\it j})))$}

\noindent
And fourth:

\disp{
(tdc(70d)) $[{\bf john}]{+}{\bf has}{+}{\bf perseverance}{+}{\bf fortunately}: Sf$}

\disp{
$[{\blacksquare}Nt(s(m)): {\it j}], {\square}(({\langle\rangle}{\exists}gNt(s(g))\backslash Sf)/{\exists}aNa): \mbox{\^{}}\lambda A\lambda B({\it Pres}\ ((\mbox{\v{}}{\it have}\ {\it A})\ {\it B})),\\{\square}(Nt(s(n)){\&}{\it CN}{\it s(n)}): \mbox{\^{}}(({\it gen}\ \mbox{\v{}}{\it perseverance}), \mbox{\v{}}{\it perseverance}), {\square}{\forall}f(\mbox{$\check{\ }$}Sf{{}{\downarrow}{}}Sf): {\it fortunately}\\\Rightarrow\ Sf$}
\vspace{0.15in}
$$
{\tiny
\prooftree
\prooftree
\prooftree
\prooftree
\prooftree
\prooftree
\prooftree
\prooftree
\prooftree
\prooftree
\justifies
\mbox{\fbox{$Nt(s(n))$}}\ \Rightarrow\ Nt(s(n))
\endprooftree
\justifies
\mbox{\fbox{$Nt(s(n)){\&}{\it CN}{\it s(n)}$}}\ \Rightarrow\ Nt(s(n))
\using {\&}L
\endprooftree
\justifies
\mbox{\fbox{${\square}(Nt(s(n)){\&}{\it CN}{\it s(n)})$}}\ \Rightarrow\ Nt(s(n))
\using {\Box}L
\endprooftree
\justifies
{\square}(Nt(s(n)){\&}{\it CN}{\it s(n)})\ \Rightarrow\ \fbox{${\exists}aNa$}
\using {\exists}R
\endprooftree
\prooftree
\prooftree
\prooftree
\prooftree
\prooftree
\justifies
\mbox{\fbox{$Nt(s(m))$}}\ \Rightarrow\ Nt(s(m))
\endprooftree
\justifies
\mbox{\fbox{${\blacksquare}Nt(s(m))$}}\ \Rightarrow\ Nt(s(m))
\using {\blacksquare}L
\endprooftree
\justifies
{\blacksquare}Nt(s(m))\ \Rightarrow\ \fbox{${\exists}gNt(s(g))$}
\using {\exists}R
\endprooftree
\justifies
[{\blacksquare}Nt(s(m))]\ \Rightarrow\ \fbox{${\langle\rangle}{\exists}gNt(s(g))$}
\using {\langle\rangle}R
\endprooftree
\prooftree
\justifies
\mbox{\fbox{$Sf$}}\ \Rightarrow\ Sf
\endprooftree
\justifies
[{\blacksquare}Nt(s(m))], \mbox{\fbox{${\langle\rangle}{\exists}gNt(s(g))\backslash Sf$}}\ \Rightarrow\ Sf
\using {\backslash}L
\endprooftree
\justifies
[{\blacksquare}Nt(s(m))], \mbox{\fbox{$({\langle\rangle}{\exists}gNt(s(g))\backslash Sf)/{\exists}aNa$}}, {\square}(Nt(s(n)){\&}{\it CN}{\it s(n)})\ \Rightarrow\ Sf
\using {/}L
\endprooftree
\justifies
[{\blacksquare}Nt(s(m))], \mbox{\fbox{${\square}(({\langle\rangle}{\exists}gNt(s(g))\backslash Sf)/{\exists}aNa)$}}, {\square}(Nt(s(n)){\&}{\it CN}{\it s(n)})\ \Rightarrow\ Sf
\using {\Box}L
\endprooftree
\justifies
[{\blacksquare}Nt(s(m))], {\square}(({\langle\rangle}{\exists}gNt(s(g))\backslash Sf)/{\exists}aNa), {\square}(Nt(s(n)){\&}{\it CN}{\it s(n)}), {\tt 1}\ \Rightarrow\ \mbox{$\check{\ }$}Sf
\using \mbox{$\check{\ }$}R
\endprooftree
\prooftree
\justifies
\mbox{\fbox{$Sf$}}\ \Rightarrow\ Sf
\endprooftree
\justifies
[{\blacksquare}Nt(s(m))], {\square}(({\langle\rangle}{\exists}gNt(s(g))\backslash Sf)/{\exists}aNa), {\square}(Nt(s(n)){\&}{\it CN}{\it s(n)}), \mbox{\fbox{$\mbox{$\check{\ }$}Sf{{}{\downarrow}{}}Sf$}}\ \Rightarrow\ Sf
\using {\downarrow}L
\endprooftree
\justifies
[{\blacksquare}Nt(s(m))], {\square}(({\langle\rangle}{\exists}gNt(s(g))\backslash Sf)/{\exists}aNa), {\square}(Nt(s(n)){\&}{\it CN}{\it s(n)}), \mbox{\fbox{${\forall}f(\mbox{$\check{\ }$}Sf{{}{\downarrow}{}}Sf)$}}\ \Rightarrow\ Sf
\using {\forall}L
\endprooftree
\justifies
[{\blacksquare}Nt(s(m))], {\square}(({\langle\rangle}{\exists}gNt(s(g))\backslash Sf)/{\exists}aNa), {\square}(Nt(s(n)){\&}{\it CN}{\it s(n)}), \mbox{\fbox{${\square}{\forall}f(\mbox{$\check{\ }$}Sf{{}{\downarrow}{}}Sf)$}}\ \Rightarrow\ Sf
\using {\Box}L
\endprooftree}
$$
\vspace{0.15in}
\disp{
$(\mbox{\v{}}{\it fortunately}\ ({\it Pres}\ ((\mbox{\v{}}{\it have}\ ({\it gen}\ \mbox{\v{}}{\it perseverance}))\ {\it j})))$}

The next example is gapping:
\disp{
(tdc(73)) $[[[{\bf john}]{+}{\bf studies}{+}{\bf logic}{+}{\bf and}{+}[{\bf charles}]{+}{\bf phonetics}]]: Sf$}
(Note the double brackets for the coordinate structure strong island.)
The treatment of the example, 
however,
is modified according to  Morrill and Valent\'{\i}n (2014\cite{mv:words}) 
in view of the observations of Kubota and Levine (2012\cite{htlggapping}).
Lexical lookup yields:

\disp{
$[[[{\blacksquare}Nt(s(m)): {\it j}], {\square}(({\langle\rangle}{\exists}gNt(s(g))\backslash Sf)/{\exists}aNa): \mbox{\^{}}\lambda A\lambda B({\it Pres}\ ((\mbox{\v{}}{\it study}\ {\it A})\ {\it B})),\\ {\square}(Nt(s(n)){\&}{\it CN}{\it s(n)}): \mbox{\^{}}(({\it gen}\ \mbox{\v{}}{\it logic}), \mbox{\v{}}{\it logic}), {\blacksquare}{\forall}w{\forall}a{\forall}b{\forall}f(({\blacksquare}((Sf{{}{\uparrow}{}}((({\langle\rangle}Na\backslash Sf){\multimapinv}Ww)/\\Nb)){{}{\begin{picture}(7,7)(0,-2)\put(1.5,-3){\rotatebox{90}{$\multimap$}}\end{picture}}{_{_{2}}}}Ww)
\backslash {[]^{-1}}{[]^{-1}}((Sf{{}{\uparrow}{}}((({\langle\rangle}Na\backslash Sf){\multimapinv}Ww)/Nb)){{}{\begin{picture}(7,7)(0,-2)\put(1.5,-3){\rotatebox{90}{$\multimap$}}\end{picture}}{_{_{2}}}}Ww))/\mbox{$\hat{\ }$}\mbox{$\hat{\ }$}{\blacksquare}((Sf{{}{\uparrow}{}}((({\langle\rangle}Na\backslash Sf){\multimapinv}Ww)/\\Nb)){{}{\begin{picture}(7,7)(0,-2)\put(1.5,-3){\rotatebox{90}{$\multimap$}}\end{picture}}{_{_{2}}}}Ww)): \lambda C\lambda D\lambda E[({\it D}\ {\it E})\wedge ({\it C}\ {\it E})], [{\blacksquare}Nt(s(m)): {\it c}], {\square}(Nt(s(n)){\&}{\it CN}{\it s(n)}): \\\mbox{\^{}}(({\it gen}\ \mbox{\v{}}{\it phonetics}), \mbox{\v{}}{\it phonetics})]]\ \Rightarrow\ Sf$}
There is the derivation:
\vspace{0.15in}
$$
{\tiny
\prooftree
\prooftree
\prooftree
\prooftree
\prooftree
\prooftree
\prooftree
\prooftree
\prooftree
\prooftree
\justifies
\mbox{\fbox{$Nt(s(n))$}}\ \Rightarrow\ Nt(s(n))
\endprooftree
\justifies
\mbox{\fbox{$Nt(s(n)){\&}{\it CN}{\it s(n)}$}}\ \Rightarrow\ Nt(s(n))
\using {\&}L
\endprooftree
\justifies
\mbox{\fbox{${\square}(Nt(s(n)){\&}{\it CN}{\it s(n)})$}}\ \Rightarrow\ Nt(s(n))
\using {\Box}L
\endprooftree
\prooftree
\prooftree
\justifies
\ \Rightarrow\ \fbox{$W[]$}
\using {\it W}R
\endprooftree
\prooftree
\prooftree
\prooftree
\prooftree
\justifies
\mbox{\fbox{$Nt(s(m))$}}\ \Rightarrow\ Nt(s(m))
\endprooftree
\justifies
\mbox{\fbox{${\blacksquare}Nt(s(m))$}}\ \Rightarrow\ Nt(s(m))
\using {\blacksquare}L
\endprooftree
\justifies
[{\blacksquare}Nt(s(m))]\ \Rightarrow\ \fbox{${\langle\rangle}Nt(s(m))$}
\using {\langle\rangle}R
\endprooftree
\prooftree
\justifies
\mbox{\fbox{$Sf$}}\ \Rightarrow\ Sf
\endprooftree
\justifies
[{\blacksquare}Nt(s(m))], \mbox{\fbox{${\langle\rangle}Nt(s(m))\backslash Sf$}}\ \Rightarrow\ Sf
\using {\backslash}L
\endprooftree
\justifies
[{\blacksquare}Nt(s(m))], \mbox{\fbox{$({\langle\rangle}Nt(s(m))\backslash Sf){\multimapinv}W[]$}}\ \Rightarrow\ Sf
\using {\multimapinv}L
\endprooftree
\justifies
[{\blacksquare}Nt(s(m))], \mbox{\fbox{$(({\langle\rangle}Nt(s(m))\backslash Sf){\multimapinv}W[])/Nt(s(n))$}}, {\square}(Nt(s(n)){\&}{\it CN}{\it s(n)})\ \Rightarrow\ Sf
\using {/}L
\endprooftree
\justifies
[{\blacksquare}Nt(s(m))], {\tt 1}, {\square}(Nt(s(n)){\&}{\it CN}{\it s(n)})\ \Rightarrow\ Sf{{}{\uparrow}{}}((({\langle\rangle}Nt(s(m))\backslash Sf){\multimapinv}W[])/Nt(s(n)))
\using {\uparrow}R
\endprooftree
\justifies
[{\blacksquare}Nt(s(m))], {\tt 1}, W[], {\square}(Nt(s(n)){\&}{\it CN}{\it s(n)})\ \Rightarrow\ Sf{{}{\uparrow}{}}((({\langle\rangle}Nt(s(m))\backslash Sf){\multimapinv}W[])/Nt(s(n)))
\using {\it W}L
\endprooftree
\justifies
[{\blacksquare}Nt(s(m))], {\tt 1}, {\tt 1}, {\square}(Nt(s(n)){\&}{\it CN}{\it s(n)})\ \Rightarrow\ (Sf{{}{\uparrow}{}}((({\langle\rangle}Nt(s(m))\backslash Sf){\multimapinv}W[])/Nt(s(n)))){{}{\begin{picture}(7,7)(0,-2)\put(1.5,-3){\rotatebox{90}{$\multimap$}}\end{picture}}{_{_{2}}}}W[]
\using {{}{\begin{picture}(7,7)(0,-2)\put(1.5,-3){\rotatebox{90}{$\multimap$}}\end{picture}}{_{2}}}R
\endprooftree
\justifies
[{\blacksquare}Nt(s(m))], {\tt 1}, {\tt 1}, {\square}(Nt(s(n)){\&}{\it CN}{\it s(n)})\ \Rightarrow\ {\blacksquare}((Sf{{}{\uparrow}{}}((({\langle\rangle}Nt(s(m))\backslash Sf){\multimapinv}W[])/Nt(s(n)))){{}{\begin{picture}(7,7)(0,-2)\put(1.5,-3){\rotatebox{90}{$\multimap$}}\end{picture}}{_{_{2}}}}W[])
\using {\blacksquare}R
\endprooftree
\justifies
[{\blacksquare}Nt(s(m))], {\tt 1}, {\square}(Nt(s(n)){\&}{\it CN}{\it s(n)})\ \Rightarrow\ \fbox{$\mbox{$\hat{\ }$}{\blacksquare}((Sf{{}{\uparrow}{}}((({\langle\rangle}Nt(s(m))\backslash Sf){\multimapinv}W[])/Nt(s(n)))){{}{\begin{picture}(7,7)(0,-2)\put(1.5,-3){\rotatebox{90}{$\multimap$}}\end{picture}}{_{_{2}}}}W[])$}
\using \mbox{$\hat{\ }$}R
\endprooftree
\justifies
\begin{array}{c}
{}[{\blacksquare}Nt(s(m))], {\square}(Nt(s(n)){\&}{\it CN}{\it s(n)})\ \Rightarrow\ \fbox{$\mbox{$\hat{\ }$}\mbox{$\hat{\ }$}{\blacksquare}((Sf{{}{\uparrow}{}}((({\langle\rangle}Nt(s(m))\backslash Sf){\multimapinv}W[])/Nt(s(n)))){{}{\begin{picture}(7,7)(0,-2)\put(1.5,-3){\rotatebox{90}{$\multimap$}}\end{picture}}{_{_{2}}}}W[])$}\\
\mbox{\footnotesize\textcircled{1}}
\end{array}
\using \mbox{$\hat{\ }$}R
\endprooftree}
$$
$$
{\tiny
\prooftree
\prooftree
\prooftree
\prooftree
\prooftree
\prooftree
\prooftree
\prooftree
\justifies
\mbox{\fbox{$Nt(s(n))$}}\ \Rightarrow\ Nt(s(n))
\endprooftree
\justifies
\mbox{\fbox{$Nt(s(n)){\&}{\it CN}{\it s(n)}$}}\ \Rightarrow\ Nt(s(n))
\using {\&}L
\endprooftree
\justifies
\mbox{\fbox{${\square}(Nt(s(n)){\&}{\it CN}{\it s(n)})$}}\ \Rightarrow\ Nt(s(n))
\using {\Box}L
\endprooftree
\prooftree
\prooftree
\justifies
\ \Rightarrow\ \fbox{$W[]$}
\using {\it W}R
\endprooftree
\prooftree
\prooftree
\prooftree
\prooftree
\justifies
\mbox{\fbox{$Nt(s(m))$}}\ \Rightarrow\ Nt(s(m))
\endprooftree
\justifies
\mbox{\fbox{${\blacksquare}Nt(s(m))$}}\ \Rightarrow\ Nt(s(m))
\using {\blacksquare}L
\endprooftree
\justifies
[{\blacksquare}Nt(s(m))]\ \Rightarrow\ \fbox{${\langle\rangle}Nt(s(m))$}
\using {\langle\rangle}R
\endprooftree
\prooftree
\justifies
\mbox{\fbox{$Sf$}}\ \Rightarrow\ Sf
\endprooftree
\justifies
[{\blacksquare}Nt(s(m))], \mbox{\fbox{${\langle\rangle}Nt(s(m))\backslash Sf$}}\ \Rightarrow\ Sf
\using {\backslash}L
\endprooftree
\justifies
[{\blacksquare}Nt(s(m))], \mbox{\fbox{$({\langle\rangle}Nt(s(m))\backslash Sf){\multimapinv}W[]$}}\ \Rightarrow\ Sf
\using {\multimapinv}L
\endprooftree
\justifies
[{\blacksquare}Nt(s(m))], \mbox{\fbox{$(({\langle\rangle}Nt(s(m))\backslash Sf){\multimapinv}W[])/Nt(s(n))$}}, {\square}(Nt(s(n)){\&}{\it CN}{\it s(n)})\ \Rightarrow\ Sf
\using {/}L
\endprooftree
\justifies
[{\blacksquare}Nt(s(m))], {\tt 1}, {\square}(Nt(s(n)){\&}{\it CN}{\it s(n)})\ \Rightarrow\ Sf{{}{\uparrow}{}}((({\langle\rangle}Nt(s(m))\backslash Sf){\multimapinv}W[])/Nt(s(n)))
\using {\uparrow}R
\endprooftree
\justifies
[{\blacksquare}Nt(s(m))], {\tt 1}, W[], {\square}(Nt(s(n)){\&}{\it CN}{\it s(n)})\ \Rightarrow\ Sf{{}{\uparrow}{}}((({\langle\rangle}Nt(s(m))\backslash Sf){\multimapinv}W[])/Nt(s(n)))
\using {\it W}L
\endprooftree
\justifies
[{\blacksquare}Nt(s(m))], {\tt 1}, {\tt 1}, {\square}(Nt(s(n)){\&}{\it CN}{\it s(n)})\ \Rightarrow\ (Sf{{}{\uparrow}{}}((({\langle\rangle}Nt(s(m))\backslash Sf){\multimapinv}W[])/Nt(s(n)))){{}{\begin{picture}(7,7)(0,-2)\put(1.5,-3){\rotatebox{90}{$\multimap$}}\end{picture}}{_{_{2}}}}W[]
\using {{}{\begin{picture}(7,7)(0,-2)\put(1.5,-3){\rotatebox{90}{$\multimap$}}\end{picture}}{_{2}}}R
\endprooftree
\justifies
\begin{array}{c}
[{\blacksquare}Nt(s(m))], {\tt 1}, {\tt 1}, {\square}(Nt(s(n)){\&}{\it CN}{\it s(n)})\ \Rightarrow\ {\blacksquare}((Sf{{}{\uparrow}{}}((({\langle\rangle}Nt(s(m))\backslash Sf){\multimapinv}W[])/Nt(s(n)))){{}{\begin{picture}(7,7)(0,-2)\put(1.5,-3){\rotatebox{90}{$\multimap$}}\end{picture}}{_{_{2}}}}W[])\\
\mbox{\footnotesize\textcircled{2}}
\end{array}
\using {\blacksquare}R
\endprooftree}
$$

\begin{center}
\rotatebox{-90}{\tiny
\prooftree
\prooftree
\prooftree
\prooftree
\prooftree
\prooftree
\mbox{\footnotesize\textcircled{1}}\tab\tab
\prooftree
\mbox{\footnotesize\textcircled{2}}\tab
\prooftree
\prooftree
\prooftree
\prooftree
\justifies
\ \Rightarrow\ \fbox{$W[]$}
\using {\it W}R
\endprooftree
\prooftree
\prooftree
\prooftree
\prooftree
\prooftree
\prooftree
\prooftree
\prooftree
\prooftree
\prooftree
\justifies
Nt(s(n))\ \Rightarrow\ Nt(s(n))
\endprooftree
\justifies
Nt(s(n))\ \Rightarrow\ \fbox{${\exists}aNa$}
\using {\exists}R
\endprooftree
\prooftree
\prooftree
\prooftree
\prooftree
\justifies
Nt(s(m))\ \Rightarrow\ Nt(s(m))
\endprooftree
\justifies
Nt(s(m))\ \Rightarrow\ \fbox{${\exists}gNt(s(g))$}
\using {\exists}R
\endprooftree
\justifies
[Nt(s(m))]\ \Rightarrow\ \fbox{${\langle\rangle}{\exists}gNt(s(g))$}
\using {\langle\rangle}R
\endprooftree
\prooftree
\justifies
\mbox{\fbox{$Sf$}}\ \Rightarrow\ Sf
\endprooftree
\justifies
[Nt(s(m))], \mbox{\fbox{${\langle\rangle}{\exists}gNt(s(g))\backslash Sf$}}\ \Rightarrow\ Sf
\using {\backslash}L
\endprooftree
\justifies
[Nt(s(m))], \mbox{\fbox{$({\langle\rangle}{\exists}gNt(s(g))\backslash Sf)/{\exists}aNa$}}, Nt(s(n))\ \Rightarrow\ Sf
\using {/}L
\endprooftree
\justifies
[Nt(s(m))], \mbox{\fbox{${\square}(({\langle\rangle}{\exists}gNt(s(g))\backslash Sf)/{\exists}aNa)$}}, Nt(s(n))\ \Rightarrow\ Sf
\using {\Box}L
\endprooftree
\justifies
{\langle\rangle}Nt(s(m)), {\square}(({\langle\rangle}{\exists}gNt(s(g))\backslash Sf)/{\exists}aNa), Nt(s(n))\ \Rightarrow\ Sf
\using {\langle\rangle}L
\endprooftree
\justifies
{\square}(({\langle\rangle}{\exists}gNt(s(g))\backslash Sf)/{\exists}aNa), Nt(s(n))\ \Rightarrow\ {\langle\rangle}Nt(s(m))\backslash Sf
\using {\backslash}R
\endprooftree
\justifies
{\square}(({\langle\rangle}{\exists}gNt(s(g))\backslash Sf)/{\exists}aNa), Nt(s(n)), W[]\ \Rightarrow\ {\langle\rangle}Nt(s(m))\backslash Sf
\using {\it W}L
\endprooftree
\justifies
{\square}(({\langle\rangle}{\exists}gNt(s(g))\backslash Sf)/{\exists}aNa), Nt(s(n))\ \Rightarrow\ ({\langle\rangle}Nt(s(m))\backslash Sf){\multimapinv}W[]
\using {\multimapinv}R
\endprooftree
\justifies
{\square}(({\langle\rangle}{\exists}gNt(s(g))\backslash Sf)/{\exists}aNa)\ \Rightarrow\ (({\langle\rangle}Nt(s(m))\backslash Sf){\multimapinv}W[])/Nt(s(n))
\using {/}R
\endprooftree
\prooftree
\justifies
\mbox{\fbox{$Sf$}}\ \Rightarrow\ Sf
\endprooftree
\justifies
\mbox{\fbox{$Sf{{}{\uparrow}{}}((({\langle\rangle}Nt(s(m))\backslash Sf){\multimapinv}W[])/Nt(s(n)))\{{\square}(({\langle\rangle}{\exists}gNt(s(g))\backslash Sf)/{\exists}aNa)\}$}}\ \Rightarrow\ Sf
\using {\uparrow}L
\endprooftree
\justifies
\mbox{\fbox{$(Sf{{}{\uparrow}{}}((({\langle\rangle}Nt(s(m))\backslash Sf){\multimapinv}W[])/Nt(s(n)))){{}{\begin{picture}(7,7)(0,-2)\put(1.5,-3){\rotatebox{90}{$\multimap$}}\end{picture}}{_{_{2}}}}W[]\{{\square}(({\langle\rangle}{\exists}gNt(s(g))\backslash Sf)/{\exists}aNa): \}$}}\ \Rightarrow\ Sf
\using {{}{\begin{picture}(7,7)(0,-2)\put(1.5,-3){\rotatebox{90}{$\multimap$}}\end{picture}}{_{2}}}L
\endprooftree
\justifies
[\mbox{\fbox{${[]^{-1}}((Sf{{}{\uparrow}{}}((({\langle\rangle}Nt(s(m))\backslash Sf){\multimapinv}W[])/Nt(s(n)))){{}{\begin{picture}(7,7)(0,-2)\put(1.5,-3){\rotatebox{90}{$\multimap$}}\end{picture}}{_{_{2}}}}W[])\{{\square}(({\langle\rangle}{\exists}gNt(s(g))\backslash Sf)/{\exists}aNa): \}$}}]\ \Rightarrow\ Sf
\using {[]^{-1}}L
\endprooftree
\justifies
[[\mbox{\fbox{${[]^{-1}}{[]^{-1}}((Sf{{}{\uparrow}{}}((({\langle\rangle}Nt(s(m))\backslash Sf){\multimapinv}W[])/Nt(s(n)))){{}{\begin{picture}(7,7)(0,-2)\put(1.5,-3){\rotatebox{90}{$\multimap$}}\end{picture}}{_{_{2}}}}W[])\{{\square}(({\langle\rangle}{\exists}gNt(s(g))\backslash Sf)/{\exists}aNa): \}$}}]]\ \Rightarrow\ Sf
\using {[]^{-1}}L
\endprooftree
\justifies
[[[{\blacksquare}Nt(s(m))], {\square}(({\langle\rangle}{\exists}gNt(s(g))\backslash Sf)/{\exists}aNa), {\square}(Nt(s(n)){\&}{\it CN}{\it s(n)}), \mbox{\fbox{${\blacksquare}((Sf{{}{\uparrow}{}}((({\langle\rangle}Nt(s(m))\backslash Sf){\multimapinv}W[])/Nt(s(n)))){{}{\begin{picture}(7,7)(0,-2)\put(1.5,-3){\rotatebox{90}{$\multimap$}}\end{picture}}{_{_{2}}}}W[])\backslash {[]^{-1}}{[]^{-1}}((Sf{{}{\uparrow}{}}((({\langle\rangle}Nt(s(m))\backslash Sf){\multimapinv}W[])/Nt(s(n)))){{}{\begin{picture}(7,7)(0,-2)\put(1.5,-3){\rotatebox{90}{$\multimap$}}\end{picture}}{_{_{2}}}}W[])$}}]]\ \Rightarrow\ Sf
\using {\backslash}L
\endprooftree
\justifies
[[[{\blacksquare}Nt(s(m))], {\square}(({\langle\rangle}{\exists}gNt(s(g))\backslash Sf)/{\exists}aNa), {\square}(Nt(s(n)){\&}{\it CN}{\it s(n)}), \mbox{\fbox{$({\blacksquare}((Sf{{}{\uparrow}{}}((({\langle\rangle}Nt(s(m))\backslash Sf){\multimapinv}W[])/Nt(s(n)))){{}{\begin{picture}(7,7)(0,-2)\put(1.5,-3){\rotatebox{90}{$\multimap$}}\end{picture}}{_{_{2}}}}W[])\backslash {[]^{-1}}{[]^{-1}}((Sf{{}{\uparrow}{}}((({\langle\rangle}Nt(s(m))\backslash Sf){\multimapinv}W[])/Nt(s(n)))){{}{\begin{picture}(7,7)(0,-2)\put(1.5,-3){\rotatebox{90}{$\multimap$}}\end{picture}}{_{_{2}}}}W[]))/\mbox{$\hat{\ }$}\mbox{$\hat{\ }$}{\blacksquare}((Sf{{}{\uparrow}{}}((({\langle\rangle}Nt(s(m))\backslash Sf){\multimapinv}W[])/Nt(s(n)))){{}{\begin{picture}(7,7)(0,-2)\put(1.5,-3){\rotatebox{90}{$\multimap$}}\end{picture}}{_{_{2}}}}W[])$}}, [{\blacksquare}Nt(s(m))], {\square}(Nt(s(n)){\&}{\it CN}{\it s(n)})]]\ \Rightarrow\ Sf
\using {/}L
\endprooftree
\justifies
[[[{\blacksquare}Nt(s(m))], {\square}(({\langle\rangle}{\exists}gNt(s(g))\backslash Sf)/{\exists}aNa), {\square}(Nt(s(n)){\&}{\it CN}{\it s(n)}), \mbox{\fbox{${\forall}f(({\blacksquare}((Sf{{}{\uparrow}{}}((({\langle\rangle}Nt(s(m))\backslash Sf){\multimapinv}W[])/Nt(s(n)))){{}{\begin{picture}(7,7)(0,-2)\put(1.5,-3){\rotatebox{90}{$\multimap$}}\end{picture}}{_{_{2}}}}W[])\backslash {[]^{-1}}{[]^{-1}}((Sf{{}{\uparrow}{}}((({\langle\rangle}Nt(s(m))\backslash Sf){\multimapinv}W[])/Nt(s(n)))){{}{\begin{picture}(7,7)(0,-2)\put(1.5,-3){\rotatebox{90}{$\multimap$}}\end{picture}}{_{_{2}}}}W[]))/\mbox{$\hat{\ }$}\mbox{$\hat{\ }$}{\blacksquare}((Sf{{}{\uparrow}{}}((({\langle\rangle}Nt(s(m))\backslash Sf){\multimapinv}W[])/Nt(s(n)))){{}{\begin{picture}(7,7)(0,-2)\put(1.5,-3){\rotatebox{90}{$\multimap$}}\end{picture}}{_{_{2}}}}W[]))$}}, [{\blacksquare}Nt(s(m))], {\square}(Nt(s(n)){\&}{\it CN}{\it s(n)})]]\ \Rightarrow\ Sf
\using {\forall}L
\endprooftree
\justifies
[[[{\blacksquare}Nt(s(m))], {\square}(({\langle\rangle}{\exists}gNt(s(g))\backslash Sf)/{\exists}aNa), {\square}(Nt(s(n)){\&}{\it CN}{\it s(n)}), \mbox{\fbox{${\forall}b{\forall}f(({\blacksquare}((Sf{{}{\uparrow}{}}((({\langle\rangle}Nt(s(m))\backslash Sf){\multimapinv}W[])/Nb)){{}{\begin{picture}(7,7)(0,-2)\put(1.5,-3){\rotatebox{90}{$\multimap$}}\end{picture}}{_{_{2}}}}W[])\backslash {[]^{-1}}{[]^{-1}}((Sf{{}{\uparrow}{}}((({\langle\rangle}Nt(s(m))\backslash Sf){\multimapinv}W[])/Nb)){{}{\begin{picture}(7,7)(0,-2)\put(1.5,-3){\rotatebox{90}{$\multimap$}}\end{picture}}{_{_{2}}}}W[]))/\mbox{$\hat{\ }$}\mbox{$\hat{\ }$}{\blacksquare}((Sf{{}{\uparrow}{}}((({\langle\rangle}Nt(s(m))\backslash Sf){\multimapinv}W[])/Nb)){{}{\begin{picture}(7,7)(0,-2)\put(1.5,-3){\rotatebox{90}{$\multimap$}}\end{picture}}{_{_{2}}}}W[]))$}}, [{\blacksquare}Nt(s(m))], {\square}(Nt(s(n)){\&}{\it CN}{\it s(n)})]]\ \Rightarrow\ Sf
\using {\forall}L
\endprooftree
\justifies
[[[{\blacksquare}Nt(s(m))], {\square}(({\langle\rangle}{\exists}gNt(s(g))\backslash Sf)/{\exists}aNa), {\square}(Nt(s(n)){\&}{\it CN}{\it s(n)}), \mbox{\fbox{${\forall}a{\forall}b{\forall}f(({\blacksquare}((Sf{{}{\uparrow}{}}((({\langle\rangle}Na\backslash Sf){\multimapinv}W[])/Nb)){{}{\begin{picture}(7,7)(0,-2)\put(1.5,-3){\rotatebox{90}{$\multimap$}}\end{picture}}{_{_{2}}}}W[])\backslash {[]^{-1}}{[]^{-1}}((Sf{{}{\uparrow}{}}((({\langle\rangle}Na\backslash Sf){\multimapinv}W[])/Nb)){{}{\begin{picture}(7,7)(0,-2)\put(1.5,-3){\rotatebox{90}{$\multimap$}}\end{picture}}{_{_{2}}}}W[]))/\mbox{$\hat{\ }$}\mbox{$\hat{\ }$}{\blacksquare}((Sf{{}{\uparrow}{}}((({\langle\rangle}Na\backslash Sf){\multimapinv}W[])/Nb)){{}{\begin{picture}(7,7)(0,-2)\put(1.5,-3){\rotatebox{90}{$\multimap$}}\end{picture}}{_{_{2}}}}W[]))$}}, [{\blacksquare}Nt(s(m))], {\square}(Nt(s(n)){\&}{\it CN}{\it s(n)})]]\ \Rightarrow\ Sf
\using {\forall}L
\endprooftree
\justifies
[[[{\blacksquare}Nt(s(m))], {\square}(({\langle\rangle}{\exists}gNt(s(g))\backslash Sf)/{\exists}aNa), {\square}(Nt(s(n)){\&}{\it CN}{\it s(n)}), \mbox{\fbox{${\forall}w{\forall}a{\forall}b{\forall}f(({\blacksquare}((Sf{{}{\uparrow}{}}((({\langle\rangle}Na\backslash Sf){\multimapinv}Ww)/Nb)){{}{\begin{picture}(7,7)(0,-2)\put(1.5,-3){\rotatebox{90}{$\multimap$}}\end{picture}}{_{_{2}}}}Ww)\backslash {[]^{-1}}{[]^{-1}}((Sf{{}{\uparrow}{}}((({\langle\rangle}Na\backslash Sf){\multimapinv}Ww)/Nb)){{}{\begin{picture}(7,7)(0,-2)\put(1.5,-3){\rotatebox{90}{$\multimap$}}\end{picture}}{_{_{2}}}}Ww))/\mbox{$\hat{\ }$}\mbox{$\hat{\ }$}{\blacksquare}((Sf{{}{\uparrow}{}}((({\langle\rangle}Na\backslash Sf){\multimapinv}Ww)/Nb)){{}{\begin{picture}(7,7)(0,-2)\put(1.5,-3){\rotatebox{90}{$\multimap$}}\end{picture}}{_{_{2}}}}Ww))$}}, [{\blacksquare}Nt(s(m))], {\square}(Nt(s(n)){\&}{\it CN}{\it s(n)})]]\ \Rightarrow\ Sf
\using {\forall}L
\endprooftree
\justifies
[[[{\blacksquare}Nt(s(m))], {\square}(({\langle\rangle}{\exists}gNt(s(g))\backslash Sf)/{\exists}aNa), {\square}(Nt(s(n)){\&}{\it CN}{\it s(n)}), \mbox{\fbox{${\blacksquare}{\forall}w{\forall}a{\forall}b{\forall}f(({\blacksquare}((Sf{{}{\uparrow}{}}((({\langle\rangle}Na\backslash Sf){\multimapinv}Ww)/Nb)){{}{\begin{picture}(7,7)(0,-2)\put(1.5,-3){\rotatebox{90}{$\multimap$}}\end{picture}}{_{_{2}}}}Ww)\backslash {[]^{-1}}{[]^{-1}}((Sf{{}{\uparrow}{}}((({\langle\rangle}Na\backslash Sf){\multimapinv}Ww)/Nb)){{}{\begin{picture}(7,7)(0,-2)\put(1.5,-3){\rotatebox{90}{$\multimap$}}\end{picture}}{_{_{2}}}}Ww))/\mbox{$\hat{\ }$}\mbox{$\hat{\ }$}{\blacksquare}((Sf{{}{\uparrow}{}}((({\langle\rangle}Na\backslash Sf){\multimapinv}Ww)/Nb)){{}{\begin{picture}(7,7)(0,-2)\put(1.5,-3){\rotatebox{90}{$\multimap$}}\end{picture}}{_{_{2}}}}Ww))$}}, [{\blacksquare}Nt(s(m))], {\square}(Nt(s(n)){\&}{\it CN}{\it s(n)})]]\ \Rightarrow\ Sf
\using {\blacksquare}L
\endprooftree}
\end{center}

\vspace{0.15in}

This delivers semantics:
\disp{
$[({\it Pres}\ ((\mbox{\v{}}{\it study}\ ({\it gen}\ \mbox{\v{}}{\it logic}))\ {\it j}))\wedge ({\it Pres}\ ((\mbox{\v{}}{\it study}\ ({\it gen}\ \mbox{\v{}}{\it phonetics}))\ {\it c}))]$}

Example~tdc(75) contains comparative subdeletion:
\disp{
(tdc(75)) $[{\bf john}]{+}{\bf ate}{+}{\bf more}{+}{\bf donuts}{+}{\bf than}{+}[{\bf mary}]{+}{\bf bought}{+}{\bf bagels}: Sf$}
Lexical lookup yields:
\disp{
\vspace{0.15in}
$[{\blacksquare}Nt(s(m)): {\it j}], {\square}(({\langle\rangle}{\exists}aNa\backslash Sf)/{\exists}aNa): \mbox{\^{}}\lambda A\lambda B({\it Past}\ ((\mbox{\v{}}{\it eat}\ {\it A})\ {\it B})),\\
 {\blacksquare}{\forall}h{\forall}g{\forall}f((Sf{{}{\uparrow}{}}(((Sh{{}{\uparrow}{}}Nt(p(g))){{}{\downarrow}{}}Sh)/{\it CN}{\it p(g)})){{}{\downarrow}{}}(Sf/\mbox{$\hat{\ }$}({\it CP}than{{}{\uparrow}{}}\\{\blacksquare}(((Sh{{}{\uparrow}{}}Nt(p(g))){{}{\downarrow}{}}Sh)/{\it CN}{\it p(g)})))):
  \lambda C\lambda D[|\lambda E({\it C}\ \lambda F\lambda G[({\it F}\ {\it E})\wedge ({\it G}\ {\it E})])|>\\|\lambda H\mbox{\v{}}({\it D}\ \lambda I\lambda J[({\it I}\ {\it H})\wedge ({\it J}\ {\it H})])|], {\square}(Nt(p(n)){\&}{\it CN}{\it p(n)}): \mbox{\^{}}(({\it gen}\ \mbox{\v{}}{\it donuts}), \mbox{\v{}}{\it donuts}),\\{\blacksquare}({\it CP}than/{\square}Sf): \lambda K{\it K}, [{\blacksquare}Nt(s(f)): {\it m}], {\square}(({\langle\rangle}{\exists}aNa\backslash Sf)/{\exists}aNa):\\
   \mbox{\^{}}\lambda L\lambda M({\it Past}\ ((\mbox{\v{}}{\it buy}\ {\it L})\ {\it M})), {\square}(Nt(p(n)){\&}{\it CN}{\it p(n)}): \mbox{\^{}}(({\it gen}\ \mbox{\v{}}{\it bagels}), \mbox{\v{}}{\it bagels})\ \Rightarrow\ Sf$}
There is the derivation:
\vspace{0.15in}
$$
{\tiny
\prooftree
\prooftree
\prooftree
\prooftree
\prooftree
\justifies
\mbox{\fbox{${\it CN}{\it p(n)}$}}\ \Rightarrow\ {\it CN}{\it p(n)}
\endprooftree
\justifies
\mbox{\fbox{$Nt(p(n)){\&}{\it CN}{\it p(n)}$}}\ \Rightarrow\ {\it CN}{\it p(n)}
\using {\&}L
\endprooftree
\justifies
\mbox{\fbox{${\square}(Nt(p(n)){\&}{\it CN}{\it p(n)})$}}\ \Rightarrow\ {\it CN}{\it p(n)}
\using {\Box}L
\endprooftree
\prooftree
\prooftree
\prooftree
\prooftree
\prooftree
\prooftree
\justifies
Nt(p(n))\ \Rightarrow\ Nt(p(n))
\endprooftree
\justifies
Nt(p(n))\ \Rightarrow\ \fbox{${\exists}aNa$}
\using {\exists}R
\endprooftree
\prooftree
\prooftree
\prooftree
\prooftree
\prooftree
\justifies
\mbox{\fbox{$Nt(s(m))$}}\ \Rightarrow\ Nt(s(m))
\endprooftree
\justifies
\mbox{\fbox{${\blacksquare}Nt(s(m))$}}\ \Rightarrow\ Nt(s(m))
\using {\blacksquare}L
\endprooftree
\justifies
{\blacksquare}Nt(s(m))\ \Rightarrow\ \fbox{${\exists}aNa$}
\using {\exists}R
\endprooftree
\justifies
[{\blacksquare}Nt(s(m))]\ \Rightarrow\ \fbox{${\langle\rangle}{\exists}aNa$}
\using {\langle\rangle}R
\endprooftree
\prooftree
\justifies
\mbox{\fbox{$Sf$}}\ \Rightarrow\ Sf
\endprooftree
\justifies
[{\blacksquare}Nt(s(m))], \mbox{\fbox{${\langle\rangle}{\exists}aNa\backslash Sf$}}\ \Rightarrow\ Sf
\using {\backslash}L
\endprooftree
\justifies
[{\blacksquare}Nt(s(m))], \mbox{\fbox{$({\langle\rangle}{\exists}aNa\backslash Sf)/{\exists}aNa$}}, Nt(p(n))\ \Rightarrow\ Sf
\using {/}L
\endprooftree
\justifies
[{\blacksquare}Nt(s(m))], \mbox{\fbox{${\square}(({\langle\rangle}{\exists}aNa\backslash Sf)/{\exists}aNa)$}}, Nt(p(n))\ \Rightarrow\ Sf
\using {\Box}L
\endprooftree
\justifies
[{\blacksquare}Nt(s(m))], {\square}(({\langle\rangle}{\exists}aNa\backslash Sf)/{\exists}aNa), {\tt 1}\ \Rightarrow\ Sf{{}{\uparrow}{}}Nt(p(n))
\using {\uparrow}R
\endprooftree
\prooftree
\justifies
\mbox{\fbox{$Sf$}}\ \Rightarrow\ Sf
\endprooftree
\justifies
[{\blacksquare}Nt(s(m))], {\square}(({\langle\rangle}{\exists}aNa\backslash Sf)/{\exists}aNa), \mbox{\fbox{$(Sf{{}{\uparrow}{}}Nt(p(n))){{}{\downarrow}{}}Sf$}}\ \Rightarrow\ Sf
\using {\downarrow}L
\endprooftree
\justifies
[{\blacksquare}Nt(s(m))], {\square}(({\langle\rangle}{\exists}aNa\backslash Sf)/{\exists}aNa), \mbox{\fbox{$((Sf{{}{\uparrow}{}}Nt(p(n))){{}{\downarrow}{}}Sf)/{\it CN}{\it p(n)}$}}, {\square}(Nt(p(n)){\&}{\it CN}{\it p(n)})\ \Rightarrow\ Sf
\using {/}L
\endprooftree
\justifies
\begin{array}{c}
{}[{\blacksquare}Nt(s(m))], {\square}(({\langle\rangle}{\exists}aNa\backslash Sf)/{\exists}aNa), {\tt 1}, {\square}(Nt(p(n)){\&}{\it CN}{\it p(n)})\ \Rightarrow\ Sf{{}{\uparrow}{}}(((Sf{{}{\uparrow}{}}Nt(p(n))){{}{\downarrow}{}}Sf)/{\it CN}{\it p(n)})\\
\mbox{\footnotesize\textcircled{1}}
\end{array}
\using {\uparrow}R
\endprooftree}
$$

\begin{center}
\rotatebox{-90}{\tiny
\prooftree
\prooftree
\prooftree
\prooftree
\prooftree
\mbox{\footnotesize\textcircled{1}}\tab
\prooftree
\prooftree
\prooftree
\prooftree
\prooftree
\prooftree
\prooftree
\prooftree
\prooftree
\prooftree
\prooftree
\justifies
\mbox{\fbox{${\it CN}{\it p(n)}$}}\ \Rightarrow\ {\it CN}{\it p(n)}
\endprooftree
\justifies
\mbox{\fbox{$Nt(p(n)){\&}{\it CN}{\it p(n)}$}}\ \Rightarrow\ {\it CN}{\it p(n)}
\using {\&}L
\endprooftree
\justifies
\mbox{\fbox{${\square}(Nt(p(n)){\&}{\it CN}{\it p(n)})$}}\ \Rightarrow\ {\it CN}{\it p(n)}
\using {\Box}L
\endprooftree
\prooftree
\prooftree
\prooftree
\prooftree
\prooftree
\prooftree
\justifies
Nt(p(n))\ \Rightarrow\ Nt(p(n))
\endprooftree
\justifies
Nt(p(n))\ \Rightarrow\ \fbox{${\exists}aNa$}
\using {\exists}R
\endprooftree
\prooftree
\prooftree
\prooftree
\prooftree
\prooftree
\justifies
\mbox{\fbox{$Nt(s(f))$}}\ \Rightarrow\ Nt(s(f))
\endprooftree
\justifies
\mbox{\fbox{${\blacksquare}Nt(s(f))$}}\ \Rightarrow\ Nt(s(f))
\using {\blacksquare}L
\endprooftree
\justifies
{\blacksquare}Nt(s(f))\ \Rightarrow\ \fbox{${\exists}aNa$}
\using {\exists}R
\endprooftree
\justifies
[{\blacksquare}Nt(s(f))]\ \Rightarrow\ \fbox{${\langle\rangle}{\exists}aNa$}
\using {\langle\rangle}R
\endprooftree
\prooftree
\justifies
\mbox{\fbox{$Sf$}}\ \Rightarrow\ Sf
\endprooftree
\justifies
[{\blacksquare}Nt(s(f))], \mbox{\fbox{${\langle\rangle}{\exists}aNa\backslash Sf$}}\ \Rightarrow\ Sf
\using {\backslash}L
\endprooftree
\justifies
[{\blacksquare}Nt(s(f))], \mbox{\fbox{$({\langle\rangle}{\exists}aNa\backslash Sf)/{\exists}aNa$}}, Nt(p(n))\ \Rightarrow\ Sf
\using {/}L
\endprooftree
\justifies
[{\blacksquare}Nt(s(f))], \mbox{\fbox{${\square}(({\langle\rangle}{\exists}aNa\backslash Sf)/{\exists}aNa)$}}, Nt(p(n))\ \Rightarrow\ Sf
\using {\Box}L
\endprooftree
\justifies
[{\blacksquare}Nt(s(f))], {\square}(({\langle\rangle}{\exists}aNa\backslash Sf)/{\exists}aNa), {\tt 1}\ \Rightarrow\ Sf{{}{\uparrow}{}}Nt(p(n))
\using {\uparrow}R
\endprooftree
\prooftree
\justifies
\mbox{\fbox{$Sf$}}\ \Rightarrow\ Sf
\endprooftree
\justifies
[{\blacksquare}Nt(s(f))], {\square}(({\langle\rangle}{\exists}aNa\backslash Sf)/{\exists}aNa), \mbox{\fbox{$(Sf{{}{\uparrow}{}}Nt(p(n))){{}{\downarrow}{}}Sf$}}\ \Rightarrow\ Sf
\using {\downarrow}L
\endprooftree
\justifies
[{\blacksquare}Nt(s(f))], {\square}(({\langle\rangle}{\exists}aNa\backslash Sf)/{\exists}aNa), \mbox{\fbox{$((Sf{{}{\uparrow}{}}Nt(p(n))){{}{\downarrow}{}}Sf)/{\it CN}{\it p(n)}$}}, {\square}(Nt(p(n)){\&}{\it CN}{\it p(n)})\ \Rightarrow\ Sf
\using {/}L
\endprooftree
\justifies
[{\blacksquare}Nt(s(f))], {\square}(({\langle\rangle}{\exists}aNa\backslash Sf)/{\exists}aNa), \mbox{\fbox{${\blacksquare}(((Sf{{}{\uparrow}{}}Nt(p(n))){{}{\downarrow}{}}Sf)/{\it CN}{\it p(n)})$}}, {\square}(Nt(p(n)){\&}{\it CN}{\it p(n)})\ \Rightarrow\ Sf
\using {\blacksquare}L
\endprooftree
\justifies
[{\blacksquare}Nt(s(f))], {\square}(({\langle\rangle}{\exists}aNa\backslash Sf)/{\exists}aNa), {\blacksquare}(((Sf{{}{\uparrow}{}}Nt(p(n))){{}{\downarrow}{}}Sf)/{\it CN}{\it p(n)}), {\square}(Nt(p(n)){\&}{\it CN}{\it p(n)})\ \Rightarrow\ {\square}Sf
\using {\Box}R
\endprooftree
\prooftree
\justifies
\mbox{\fbox{${\it CP}than$}}\ \Rightarrow\ {\it CP}than
\endprooftree
\justifies
\mbox{\fbox{${\it CP}than/{\square}Sf$}}, [{\blacksquare}Nt(s(f))], {\square}(({\langle\rangle}{\exists}aNa\backslash Sf)/{\exists}aNa), {\blacksquare}(((Sf{{}{\uparrow}{}}Nt(p(n))){{}{\downarrow}{}}Sf)/{\it CN}{\it p(n)}), {\square}(Nt(p(n)){\&}{\it CN}{\it p(n)})\ \Rightarrow\ {\it CP}than
\using {/}L
\endprooftree
\justifies
\mbox{\fbox{${\blacksquare}({\it CP}than/{\square}Sf)$}}, [{\blacksquare}Nt(s(f))], {\square}(({\langle\rangle}{\exists}aNa\backslash Sf)/{\exists}aNa), {\blacksquare}(((Sf{{}{\uparrow}{}}Nt(p(n))){{}{\downarrow}{}}Sf)/{\it CN}{\it p(n)}), {\square}(Nt(p(n)){\&}{\it CN}{\it p(n)})\ \Rightarrow\ {\it CP}than
\using {\blacksquare}L
\endprooftree
\justifies
{\blacksquare}({\it CP}than/{\square}Sf), [{\blacksquare}Nt(s(f))], {\square}(({\langle\rangle}{\exists}aNa\backslash Sf)/{\exists}aNa), {\tt 1}, {\square}(Nt(p(n)){\&}{\it CN}{\it p(n)})\ \Rightarrow\ {\it CP}than{{}{\uparrow}{}}{\blacksquare}(((Sf{{}{\uparrow}{}}Nt(p(n))){{}{\downarrow}{}}Sf)/{\it CN}{\it p(n)})
\using {\uparrow}R
\endprooftree
\justifies
{\blacksquare}({\it CP}than/{\square}Sf), [{\blacksquare}Nt(s(f))], {\square}(({\langle\rangle}{\exists}aNa\backslash Sf)/{\exists}aNa), {\square}(Nt(p(n)){\&}{\it CN}{\it p(n)})\ \Rightarrow\ \fbox{$\mbox{$\hat{\ }$}({\it CP}than{{}{\uparrow}{}}{\blacksquare}(((Sf{{}{\uparrow}{}}Nt(p(n))){{}{\downarrow}{}}Sf)/{\it CN}{\it p(n)}))$}
\using \mbox{$\hat{\ }$}R
\endprooftree
\prooftree
\justifies
\mbox{\fbox{$Sf$}}\ \Rightarrow\ Sf
\endprooftree
\justifies
\mbox{\fbox{$Sf/\mbox{$\hat{\ }$}({\it CP}than{{}{\uparrow}{}}{\blacksquare}(((Sf{{}{\uparrow}{}}Nt(p(n))){{}{\downarrow}{}}Sf)/{\it CN}{\it p(n)}))$}}, {\blacksquare}({\it CP}than/{\square}Sf), [{\blacksquare}Nt(s(f))], {\square}(({\langle\rangle}{\exists}aNa\backslash Sf)/{\exists}aNa), {\square}(Nt(p(n)){\&}{\it CN}{\it p(n)})\ \Rightarrow\ Sf
\using {/}L
\endprooftree
\justifies
[{\blacksquare}Nt(s(m))], {\square}(({\langle\rangle}{\exists}aNa\backslash Sf)/{\exists}aNa), \mbox{\fbox{$(Sf{{}{\uparrow}{}}(((Sf{{}{\uparrow}{}}Nt(p(n))){{}{\downarrow}{}}Sf)/{\it CN}{\it p(n)})){{}{\downarrow}{}}(Sf/\mbox{$\hat{\ }$}({\it CP}than{{}{\uparrow}{}}{\blacksquare}(((Sf{{}{\uparrow}{}}Nt(p(n))){{}{\downarrow}{}}Sf)/{\it CN}{\it p(n)})))$}}, {\square}(Nt(p(n)){\&}{\it CN}{\it p(n)}), {\blacksquare}({\it CP}than/{\square}Sf), [{\blacksquare}Nt(s(f))], {\square}(({\langle\rangle}{\exists}aNa\backslash Sf)/{\exists}aNa), {\square}(Nt(p(n)){\&}{\it CN}{\it p(n)})\ \Rightarrow\ Sf
\using {\downarrow}L
\endprooftree
\justifies
[{\blacksquare}Nt(s(m))], {\square}(({\langle\rangle}{\exists}aNa\backslash Sf)/{\exists}aNa), \mbox{\fbox{${\forall}f((Sf{{}{\uparrow}{}}(((Sf{{}{\uparrow}{}}Nt(p(n))){{}{\downarrow}{}}Sf)/{\it CN}{\it p(n)})){{}{\downarrow}{}}(Sf/\mbox{$\hat{\ }$}({\it CP}than{{}{\uparrow}{}}{\blacksquare}(((Sf{{}{\uparrow}{}}Nt(p(n))){{}{\downarrow}{}}Sf)/{\it CN}{\it p(n)}))))$}}, {\square}(Nt(p(n)){\&}{\it CN}{\it p(n)}), {\blacksquare}({\it CP}than/{\square}Sf), [{\blacksquare}Nt(s(f))], {\square}(({\langle\rangle}{\exists}aNa\backslash Sf)/{\exists}aNa), {\square}(Nt(p(n)){\&}{\it CN}{\it p(n)})\ \Rightarrow\ Sf
\using {\forall}L
\endprooftree
\justifies
[{\blacksquare}Nt(s(m))], {\square}(({\langle\rangle}{\exists}aNa\backslash Sf)/{\exists}aNa), \mbox{\fbox{${\forall}g{\forall}f((Sf{{}{\uparrow}{}}(((Sf{{}{\uparrow}{}}Nt(p(g))){{}{\downarrow}{}}Sf)/{\it CN}{\it p(g)})){{}{\downarrow}{}}(Sf/\mbox{$\hat{\ }$}({\it CP}than{{}{\uparrow}{}}{\blacksquare}(((Sf{{}{\uparrow}{}}Nt(p(g))){{}{\downarrow}{}}Sf)/{\it CN}{\it p(g)}))))$}}, {\square}(Nt(p(n)){\&}{\it CN}{\it p(n)}), {\blacksquare}({\it CP}than/{\square}Sf), [{\blacksquare}Nt(s(f))], {\square}(({\langle\rangle}{\exists}aNa\backslash Sf)/{\exists}aNa), {\square}(Nt(p(n)){\&}{\it CN}{\it p(n)})\ \Rightarrow\ Sf
\using {\forall}L
\endprooftree
\justifies
[{\blacksquare}Nt(s(m))], {\square}(({\langle\rangle}{\exists}aNa\backslash Sf)/{\exists}aNa), \mbox{\fbox{${\forall}h{\forall}g{\forall}f((Sf{{}{\uparrow}{}}(((Sh{{}{\uparrow}{}}Nt(p(g))){{}{\downarrow}{}}Sh)/{\it CN}{\it p(g)})){{}{\downarrow}{}}(Sf/\mbox{$\hat{\ }$}({\it CP}than{{}{\uparrow}{}}{\blacksquare}(((Sh{{}{\uparrow}{}}Nt(p(g))){{}{\downarrow}{}}Sh)/{\it CN}{\it p(g)}))))$}}, {\square}(Nt(p(n)){\&}{\it CN}{\it p(n)}), {\blacksquare}({\it CP}than/{\square}Sf), [{\blacksquare}Nt(s(f))], {\square}(({\langle\rangle}{\exists}aNa\backslash Sf)/{\exists}aNa), {\square}(Nt(p(n)){\&}{\it CN}{\it p(n)})\ \Rightarrow\ Sf
\using {\forall}L
\endprooftree
\justifies
[{\blacksquare}Nt(s(m))], {\square}(({\langle\rangle}{\exists}aNa\backslash Sf)/{\exists}aNa), \mbox{\fbox{${\blacksquare}{\forall}h{\forall}g{\forall}f((Sf{{}{\uparrow}{}}(((Sh{{}{\uparrow}{}}Nt(p(g))){{}{\downarrow}{}}Sh)/{\it CN}{\it p(g)})){{}{\downarrow}{}}(Sf/\mbox{$\hat{\ }$}({\it CP}than{{}{\uparrow}{}}{\blacksquare}(((Sh{{}{\uparrow}{}}Nt(p(g))){{}{\downarrow}{}}Sh)/{\it CN}{\it p(g)}))))$}}, {\square}(Nt(p(n)){\&}{\it CN}{\it p(n)}), {\blacksquare}({\it CP}than/{\square}Sf), [{\blacksquare}Nt(s(f))], {\square}(({\langle\rangle}{\exists}aNa\backslash Sf)/{\exists}aNa), {\square}(Nt(p(n)){\&}{\it CN}{\it p(n)})\ \Rightarrow\ Sf
\using {\blacksquare}L
\endprooftree}
\end{center}

\vspace{0.15in}

This delivers semantics:
\disp{
$[|\lambda C[(\mbox{\v{}}{\it donuts}\ {\it C})\wedge ({\it Past}\ ((\mbox{\v{}}{\it eat}\ {\it C})\ {\it j}))]|>|\lambda F[(\mbox{\v{}}{\it bagels}\ {\it F})\wedge ({\it Past}\ ((\mbox{\v{}}{\it buy}\ {\it F})\ {\it m}))]|]$}

Finally, 
there is the medial reflexivisation:
\disp{
(tdc(86a)) $[{\bf john}]{+}{\bf bought}{+}{\bf himself}{+}{\bf coffee}: Sf$}
The lexical lookup yields:
\disp{
$[{\blacksquare}Nt(s(m)): {\it j}], {\square}(({\langle\rangle}{\exists}aNa\backslash Sf)/({\exists}aNa{\bullet}{\exists}aNa)): \mbox{\^{}}\lambda A\lambda B({\it Past}\ (((\mbox{\v{}}{\it buy}\ \pi_1{\it A})\ \pi_2{\it A})\ {\it B})),\\
 {\blacksquare}{\forall}f((({\langle\rangle}Nt(s(m))\backslash Sf){{}{\uparrow}{}}Nt(s(m))){{}{\downarrow}{}}({\langle\rangle}Nt(s(m))\backslash Sf)): \lambda C\lambda D(({\it C}\ {\it D})\ {\it D}),\\{\square}(Nt(s(n)){\&}{\it CN}{\it s(n)}):\mbox{\^{}}(({\it gen}\ \mbox{\v{}}{\it coffee}), \mbox{\v{}}{\it coffee})\ \Rightarrow\ Sf$}
There is the derivation:

\vspace{0.15in}

\noindent
{\tiny
\prooftree
\prooftree
\prooftree
\prooftree
\prooftree
\prooftree
\prooftree
\prooftree
\prooftree
\prooftree
\prooftree
\justifies
Nt(s(m))\ \Rightarrow\ Nt(s(m))
\endprooftree
\justifies
Nt(s(m))\ \Rightarrow\ \fbox{${\exists}aNa$}
\using {\exists}R
\endprooftree
\prooftree
\prooftree
\prooftree
\prooftree
\justifies
\mbox{\fbox{$Nt(s(n))$}}\ \Rightarrow\ Nt(s(n))
\endprooftree
\justifies
\mbox{\fbox{$Nt(s(n)){\&}{\it CN}{\it s(n)}$}}\ \Rightarrow\ Nt(s(n))
\using {\&}L
\endprooftree
\justifies
\mbox{\fbox{${\square}(Nt(s(n)){\&}{\it CN}{\it s(n)})$}}\ \Rightarrow\ Nt(s(n))
\using {\Box}L
\endprooftree
\justifies
{\square}(Nt(s(n)){\&}{\it CN}{\it s(n)})\ \Rightarrow\ \fbox{${\exists}aNa$}
\using {\exists}R
\endprooftree
\justifies
Nt(s(m)), {\square}(Nt(s(n)){\&}{\it CN}{\it s(n)})\ \Rightarrow\ \fbox{${\exists}aNa{\bullet}{\exists}aNa$}
\using {\bullet}R
\endprooftree
\prooftree
\prooftree
\prooftree
\prooftree
\justifies
Nt(s(m))\ \Rightarrow\ Nt(s(m))
\endprooftree
\justifies
Nt(s(m))\ \Rightarrow\ \fbox{${\exists}aNa$}
\using {\exists}R
\endprooftree
\justifies
[Nt(s(m))]\ \Rightarrow\ \fbox{${\langle\rangle}{\exists}aNa$}
\using {\langle\rangle}R
\endprooftree
\prooftree
\justifies
\mbox{\fbox{$Sf$}}\ \Rightarrow\ Sf
\endprooftree
\justifies
[Nt(s(m))], \mbox{\fbox{${\langle\rangle}{\exists}aNa\backslash Sf$}}\ \Rightarrow\ Sf
\using {\backslash}L
\endprooftree
\justifies
[Nt(s(m))], \mbox{\fbox{$({\langle\rangle}{\exists}aNa\backslash Sf)/({\exists}aNa{\bullet}{\exists}aNa)$}}, Nt(s(m)), {\square}(Nt(s(n)){\&}{\it CN}{\it s(n)})\ \Rightarrow\ Sf
\using {/}L
\endprooftree
\justifies
[Nt(s(m))], \mbox{\fbox{${\square}(({\langle\rangle}{\exists}aNa\backslash Sf)/({\exists}aNa{\bullet}{\exists}aNa))$}}, Nt(s(m)), {\square}(Nt(s(n)){\&}{\it CN}{\it s(n)})\ \Rightarrow\ Sf
\using {\Box}L
\endprooftree
\justifies
{\langle\rangle}Nt(s(m)), {\square}(({\langle\rangle}{\exists}aNa\backslash Sf)/({\exists}aNa{\bullet}{\exists}aNa)), Nt(s(m)), {\square}(Nt(s(n)){\&}{\it CN}{\it s(n)})\ \Rightarrow\ Sf
\using {\langle\rangle}L
\endprooftree
\justifies
{\square}(({\langle\rangle}{\exists}aNa\backslash Sf)/({\exists}aNa{\bullet}{\exists}aNa)), Nt(s(m)), {\square}(Nt(s(n)){\&}{\it CN}{\it s(n)})\ \Rightarrow\ {\langle\rangle}Nt(s(m))\backslash Sf
\using {\backslash}R
\endprooftree
\justifies
{\square}(({\langle\rangle}{\exists}aNa\backslash Sf)/({\exists}aNa{\bullet}{\exists}aNa)), {\tt 1}, {\square}(Nt(s(n)){\&}{\it CN}{\it s(n)})\ \Rightarrow\ ({\langle\rangle}Nt(s(m))\backslash Sf){{}{\uparrow}{}}Nt(s(m))
\using {\uparrow}R
\endprooftree
\prooftree
\prooftree
\prooftree
\prooftree
\justifies
\mbox{\fbox{$Nt(s(m))$}}\ \Rightarrow\ Nt(s(m))
\endprooftree
\justifies
\mbox{\fbox{${\blacksquare}Nt(s(m))$}}\ \Rightarrow\ Nt(s(m))
\using {\blacksquare}L
\endprooftree
\justifies
[{\blacksquare}Nt(s(m))]\ \Rightarrow\ \fbox{${\langle\rangle}Nt(s(m))$}
\using {\langle\rangle}R
\endprooftree
\prooftree
\justifies
\mbox{\fbox{$Sf$}}\ \Rightarrow\ Sf
\endprooftree
\justifies
[{\blacksquare}Nt(s(m))], \mbox{\fbox{${\langle\rangle}Nt(s(m))\backslash Sf$}}\ \Rightarrow\ Sf
\using {\backslash}L
\endprooftree
\justifies
[{\blacksquare}Nt(s(m))], {\square}(({\langle\rangle}{\exists}aNa\backslash Sf)/({\exists}aNa{\bullet}{\exists}aNa)), \mbox{\fbox{$(({\langle\rangle}Nt(s(m))\backslash Sf){{}{\uparrow}{}}Nt(s(m))){{}{\downarrow}{}}({\langle\rangle}Nt(s(m))\backslash Sf)$}}, {\square}(Nt(s(n)){\&}{\it CN}{\it s(n)})\ \Rightarrow\ Sf
\using {\downarrow}L
\endprooftree
\justifies
[{\blacksquare}Nt(s(m))], {\square}(({\langle\rangle}{\exists}aNa\backslash Sf)/({\exists}aNa{\bullet}{\exists}aNa)), \mbox{\fbox{${\forall}f((({\langle\rangle}Nt(s(m))\backslash Sf){{}{\uparrow}{}}Nt(s(m))){{}{\downarrow}{}}({\langle\rangle}Nt(s(m))\backslash Sf))$}}, {\square}(Nt(s(n)){\&}{\it CN}{\it s(n)})\ \Rightarrow\ Sf
\using {\forall}L
\endprooftree
\justifies
[{\blacksquare}Nt(s(m))], {\square}(({\langle\rangle}{\exists}aNa\backslash Sf)/({\exists}aNa{\bullet}{\exists}aNa)), \mbox{\fbox{${\blacksquare}{\forall}f((({\langle\rangle}Nt(s(m))\backslash Sf){{}{\uparrow}{}}Nt(s(m))){{}{\downarrow}{}}({\langle\rangle}Nt(s(m))\backslash Sf))$}}, {\square}(Nt(s(n)){\&}{\it CN}{\it s(n)})\ \Rightarrow\ Sf
\using {\blacksquare}L
\endprooftree}

\vspace{0.15in}

\noindent
This delivers semantics:
\disp{
$({\it Past}\ (((\mbox{\v{}}{\it buy}\ {\it j})\ ({\it gen}\ \mbox{\v{}}{\it coffee}))\ {\it j}))$}

\section{Dutch verb raising and cross-serial dependency}

\setlength{\parindent}{0in}

$
{\bf kan}: (NA\backslash Si){{}{\downarrow}{}}(NA\backslash Sf): \lambda B\lambda C(({\it isable}\ ({\it B}\ {\it C}))\ {\it C})\\
{\bf las}: NA\backslash (Nt(s(B))\backslash Sf): {\it read}\\
{\bf wil}: (NA\backslash Si){{}{\downarrow}{}}(NA\backslash Sf): \lambda B\lambda C(({\it want}\ ({\it B}\ {\it C}))\ {\it C})\\
{\bf kan}: Q/\mbox{$\hat{\ }$}(Sf{{}{\uparrow}{}}((NA\backslash Si){{}{\downarrow}{}}(NA\backslash Sf))): \lambda B({\it B}\ \lambda C\lambda D(({\it isable}\ ({\it C}\ {\it D}))\ {\it D}))\\
{\bf las}: Q/\mbox{$\hat{\ }$}(Sf{{}{\uparrow}{}}(NA\backslash (Nt(s(B))\backslash Sf))): \lambda C({\it C}\ {\it read})\\
{\bf wil}: Q/\mbox{$\hat{\ }$}(Sf{{}{\uparrow}{}}((NA\backslash Si){{}{\downarrow}{}}(NA\backslash Sf))): \lambda B({\it B}\ \lambda C\lambda D(({\it want}\ ({\it C}\ {\it D}))\ {\it D}))\\
{\bf alles}: (SA{{}{\uparrow}{}}Nt(s(n))){{}{\downarrow}{}}SA: \lambda B\forall C[({\it thing}\ {\it C})\rightarrow ({\it B}\ {\it C})]\\
{\bf boeken}: Np(n): {\it books}\\
{\bf cecilia}: Nt(s(f)): {\it c}\\
{\bf de}: Nt(s(A))/{\it CN}{\it A}: {\it the}\\
{\bf helpen}: \mbox{$\rhd^{-1}$}((NA\backslash Si){{}{\downarrow}{}}(NB\backslash (NA\backslash Si))): \lambda C\lambda D(({\it help}\ ({\it C}\ {\it D}))\ {\it D})\\
{\bf henk}: Nt(s(m)): {\it h}\\
{\bf jan}: Nt(s(m)): {\it j}\\
{\bf kunnen}: \mbox{$\rhd^{-1}$}((NA\backslash Si){{}{\downarrow}{}}(NA\backslash Si)): \lambda B\lambda C(({\it isable}\ ({\it B}\ {\it C}))\ {\it C})\\
{\bf lezen}: \mbox{$\rhd^{-1}$}(NA\backslash (NB\backslash Si)): {\it read}\\
{\bf nijlpaarden}: {\it CN}{\it p(n)}: {\it hippos}\\
{\bf voeren}: \mbox{$\rhd^{-1}$}(NA\backslash (NB\backslash Si)): {\it feed}\\
{\bf zag}: (Nt(s(A))\backslash Si){{}{\downarrow}{}}(NB\backslash (Nt(s(A))\backslash Sf)): \lambda C\lambda D(({\it saw}\ ({\it C}\ {\it D}))\ {\it D})\\$

{\scriptsize

\vspace{0.15in}
(d(1)) ${\bf jan}{+}{\bf boeken}{+}{\bf las}: Sf$

\vspace{0.15in}
$Nt(s(m)): {\it j}, Np(n): {\it books}, NA\backslash (Nt(s(B))\backslash Sf): {\it read}\ \Rightarrow\ Sf$

\vspace{0.15in}

\prooftree
\prooftree
\justifies
Np(n)\ \Rightarrow\ Np(n)
\endprooftree
\prooftree
\prooftree
\justifies
Nt(s(m))\ \Rightarrow\ Nt(s(m))
\endprooftree
\prooftree
\justifies
\mbox{\fbox{$Sf$}}\ \Rightarrow\ Sf
\endprooftree
\justifies
Nt(s(m)), \mbox{\fbox{$Nt(s(m))\backslash Sf$}}\ \Rightarrow\ Sf
\using {\backslash}L
\endprooftree
\justifies
Nt(s(m)), Np(n), \mbox{\fbox{$Np(n)\backslash (Nt(s(m))\backslash Sf)$}}\ \Rightarrow\ Sf
\using {\backslash}L
\endprooftree

\vspace{0.15in}

$(({\it read}\ {\it books})\ {\it j})$

\vspace{0.15in}
$Nt(s(m)): {\it j}, Np(n): {\it books}, Q/\mbox{$\hat{\ }$}(Sf{{}{\uparrow}{}}(NA\backslash (Nt(s(B))\backslash Sf))): \lambda C({\it C}\ {\it read})\ \Rightarrow\ Sf$

\vspace{0.15in}
(d(2)) ${\bf jan}{+}{\bf boeken}{+}{\bf kan}{+}{\bf lezen}: Sf$

\vspace{0.15in}
$Nt(s(m)): {\it j}, Np(n): {\it books}, (NA\backslash Si){{}{\downarrow}{}}(NA\backslash Sf): \lambda B\lambda C(({\it isable}\ ({\it B}\ {\it C}))\ {\it C}), \mbox{$\rhd^{-1}$}(ND\backslash (NE\backslash Si)): {\it read}\ \Rightarrow\ Sf$

\vspace{0.15in}

\prooftree
\prooftree
\prooftree
\prooftree
\prooftree
\justifies
Np(n)\ \Rightarrow\ Np(n)
\endprooftree
\prooftree
\prooftree
\justifies
Nt(s(m))\ \Rightarrow\ Nt(s(m))
\endprooftree
\prooftree
\justifies
\mbox{\fbox{$Si\{{\tt 1}\}$}}\ \Rightarrow\ Si
\endprooftree
\justifies
Nt(s(m)), \mbox{\fbox{$Nt(s(m))\backslash Si\{{\tt 1}\}$}}\ \Rightarrow\ Si
\using {\backslash}L
\endprooftree
\justifies
Nt(s(m)), Np(n), \mbox{\fbox{$Np(n)\backslash (Nt(s(m))\backslash Si)\{{\tt 1}\}$}}\ \Rightarrow\ Si
\using {\backslash}L
\endprooftree
\justifies
Nt(s(m)), Np(n), {\tt 1}, \mbox{\fbox{$\mbox{$\rhd^{-1}$}(Np(n)\backslash (Nt(s(m))\backslash Si))$}}\ \Rightarrow\ Si
\using {\rhd^{-1}}L
\endprooftree
\justifies
Np(n), {\tt 1}, \mbox{$\rhd^{-1}$}(Np(n)\backslash (Nt(s(m))\backslash Si))\ \Rightarrow\ Nt(s(m))\backslash Si
\using {\backslash}R
\endprooftree
\prooftree
\prooftree
\justifies
Nt(s(m))\ \Rightarrow\ Nt(s(m))
\endprooftree
\prooftree
\justifies
\mbox{\fbox{$Sf$}}\ \Rightarrow\ Sf
\endprooftree
\justifies
Nt(s(m)), \mbox{\fbox{$Nt(s(m))\backslash Sf$}}\ \Rightarrow\ Sf
\using {\backslash}L
\endprooftree
\justifies
Nt(s(m)), Np(n), \mbox{\fbox{$(Nt(s(m))\backslash Si){{}{\downarrow}{}}(Nt(s(m))\backslash Sf)$}}, \mbox{$\rhd^{-1}$}(Np(n)\backslash (Nt(s(m))\backslash Si))\ \Rightarrow\ Sf
\using {\downarrow}L
\endprooftree

\vspace{0.15in}

$(({\it isable}\ (({\it read}\ {\it books})\ {\it j}))\ {\it j})$

\vspace{0.15in}
$Nt(s(m)): {\it j}, Np(n): {\it books}, Q/\mbox{$\hat{\ }$}(Sf{{}{\uparrow}{}}((NA\backslash Si){{}{\downarrow}{}}(NA\backslash Sf))): \lambda B({\it B}\ \lambda C\lambda D(({\it isable}\ ({\it C}\ {\it D}))\ {\it D})), \mbox{$\rhd^{-1}$}(NE\backslash (NF\backslash Si)): {\it read}\ \Rightarrow\ Sf$

\vspace{0.15in}
(d(3)) ${\bf jan}{+}{\bf boeken}{+}{\bf wil}{+}{\bf kunnen}{+}{\bf lezen}: Sf$

\vspace{0.15in}
$Nt(s(m)): {\it j}, Np(n): {\it books}, (NA\backslash Si){{}{\downarrow}{}}(NA\backslash Sf): \lambda B\lambda C(({\it want}\ ({\it B}\ {\it C}))\ {\it C}), \mbox{$\rhd^{-1}$}((ND\backslash Si){{}{\downarrow}{}}(ND\backslash Si)): \lambda E\lambda F(({\it isable}\ ({\it E}\ {\it F}))\ {\it F}), \mbox{$\rhd^{-1}$}(NG\backslash (NH\backslash Si)): {\it read}\ \Rightarrow\ Sf$

\vspace{0.15in}

\prooftree
\prooftree
\prooftree
\prooftree
\prooftree
\prooftree
\prooftree
\prooftree
\justifies
Np(n)\ \Rightarrow\ Np(n)
\endprooftree
\prooftree
\prooftree
\justifies
Nt(s(m))\ \Rightarrow\ Nt(s(m))
\endprooftree
\prooftree
\justifies
\mbox{\fbox{$Si\{{\tt 1}\}$}}\ \Rightarrow\ Si
\endprooftree
\justifies
Nt(s(m)), \mbox{\fbox{$Nt(s(m))\backslash Si\{{\tt 1}\}$}}\ \Rightarrow\ Si
\using {\backslash}L
\endprooftree
\justifies
Nt(s(m)), Np(n), \mbox{\fbox{$Np(n)\backslash (Nt(s(m))\backslash Si)\{{\tt 1}\}$}}\ \Rightarrow\ Si
\using {\backslash}L
\endprooftree
\justifies
Nt(s(m)), Np(n), {\tt 1}, \mbox{\fbox{$\mbox{$\rhd^{-1}$}(Np(n)\backslash (Nt(s(m))\backslash Si))$}}\ \Rightarrow\ Si
\using {\rhd^{-1}}L
\endprooftree
\justifies
Np(n), {\tt 1}, \mbox{$\rhd^{-1}$}(Np(n)\backslash (Nt(s(m))\backslash Si))\ \Rightarrow\ Nt(s(m))\backslash Si
\using {\backslash}R
\endprooftree
\prooftree
\prooftree
\justifies
Nt(s(m))\ \Rightarrow\ Nt(s(m))
\endprooftree
\prooftree
\justifies
\mbox{\fbox{$Si\{{\tt 1}\}$}}\ \Rightarrow\ Si
\endprooftree
\justifies
Nt(s(m)), \mbox{\fbox{$Nt(s(m))\backslash Si\{{\tt 1}\}$}}\ \Rightarrow\ Si
\using {\backslash}L
\endprooftree
\justifies
Nt(s(m)), Np(n), \mbox{\fbox{$(Nt(s(m))\backslash Si){{}{\downarrow}{}}(Nt(s(m))\backslash Si)\{{\tt 1}\}$}}, \mbox{$\rhd^{-1}$}(Np(n)\backslash (Nt(s(m))\backslash Si))\ \Rightarrow\ Si
\using {\downarrow}L
\endprooftree
\justifies
Nt(s(m)), Np(n), {\tt 1}, \mbox{\fbox{$\mbox{$\rhd^{-1}$}((Nt(s(m))\backslash Si){{}{\downarrow}{}}(Nt(s(m))\backslash Si))$}}, \mbox{$\rhd^{-1}$}(Np(n)\backslash (Nt(s(m))\backslash Si))\ \Rightarrow\ Si
\using {\rhd^{-1}}L
\endprooftree
\justifies
Np(n), {\tt 1}, \mbox{$\rhd^{-1}$}((Nt(s(m))\backslash Si){{}{\downarrow}{}}(Nt(s(m))\backslash Si)), \mbox{$\rhd^{-1}$}(Np(n)\backslash (Nt(s(m))\backslash Si))\ \Rightarrow\ Nt(s(m))\backslash Si
\using {\backslash}R
\endprooftree
\prooftree
\prooftree
\justifies
Nt(s(m))\ \Rightarrow\ Nt(s(m))
\endprooftree
\prooftree
\justifies
\mbox{\fbox{$Sf$}}\ \Rightarrow\ Sf
\endprooftree
\justifies
Nt(s(m)), \mbox{\fbox{$Nt(s(m))\backslash Sf$}}\ \Rightarrow\ Sf
\using {\backslash}L
\endprooftree
\justifies
Nt(s(m)), Np(n), \mbox{\fbox{$(Nt(s(m))\backslash Si){{}{\downarrow}{}}(Nt(s(m))\backslash Sf)$}}, \mbox{$\rhd^{-1}$}((Nt(s(m))\backslash Si){{}{\downarrow}{}}(Nt(s(m))\backslash Si)), \mbox{$\rhd^{-1}$}(Np(n)\backslash (Nt(s(m))\backslash Si))\ \Rightarrow\ Sf
\using {\downarrow}L
\endprooftree

\vspace{0.15in}

$(({\it want}\ (({\it isable}\ (({\it read}\ {\it books})\ {\it j}))\ {\it j}))\ {\it j})$

\vspace{0.15in}
$Nt(s(m)): {\it j}, Np(n): {\it books}, Q/\mbox{$\hat{\ }$}(Sf{{}{\uparrow}{}}((NA\backslash Si){{}{\downarrow}{}}(NA\backslash Sf))): \lambda B({\it B}\ \lambda C\lambda D(({\it want}\ ({\it C}\ {\it D}))\ {\it D})), \mbox{$\rhd^{-1}$}((NE\backslash Si){{}{\downarrow}{}}(NE\backslash Si)): \lambda F\lambda G(({\it isable}\ ({\it F}\ {\it G}))\ {\it G}), \mbox{$\rhd^{-1}$}(NH\backslash (NI\backslash Si)): {\it read}\ \Rightarrow\ Sf$

\vspace{0.15in}
(d(4)) ${\bf jan}{+}{\bf alles}{+}{\bf las}: Sf$

\vspace{0.15in}
$Nt(s(m)): {\it j}, (SA{{}{\uparrow}{}}Nt(s(n))){{}{\downarrow}{}}SA: \lambda B\forall C[({\it thing}\ {\it C})\rightarrow ({\it B}\ {\it C})], ND\backslash (Nt(s(E))\backslash Sf): {\it read}\ \Rightarrow\ Sf$

\vspace{0.15in}

\prooftree
\prooftree
\prooftree
\prooftree
\justifies
Nt(s(n))\ \Rightarrow\ Nt(s(n))
\endprooftree
\prooftree
\prooftree
\justifies
Nt(s(m))\ \Rightarrow\ Nt(s(m))
\endprooftree
\prooftree
\justifies
\mbox{\fbox{$Sf$}}\ \Rightarrow\ Sf
\endprooftree
\justifies
Nt(s(m)), \mbox{\fbox{$Nt(s(m))\backslash Sf$}}\ \Rightarrow\ Sf
\using {\backslash}L
\endprooftree
\justifies
Nt(s(m)), Nt(s(n)), \mbox{\fbox{$Nt(s(n))\backslash (Nt(s(m))\backslash Sf)$}}\ \Rightarrow\ Sf
\using {\backslash}L
\endprooftree
\justifies
Nt(s(m)), {\tt 1}, Nt(s(n))\backslash (Nt(s(m))\backslash Sf)\ \Rightarrow\ Sf{{}{\uparrow}{}}Nt(s(n))
\using {\uparrow}R
\endprooftree
\prooftree
\justifies
\mbox{\fbox{$Sf$}}\ \Rightarrow\ Sf
\endprooftree
\justifies
Nt(s(m)), \mbox{\fbox{$(Sf{{}{\uparrow}{}}Nt(s(n))){{}{\downarrow}{}}Sf$}}, Nt(s(n))\backslash (Nt(s(m))\backslash Sf)\ \Rightarrow\ Sf
\using {\downarrow}L
\endprooftree

\vspace{0.15in}

$\forall B[({\it thing}\ {\it B})\rightarrow (({\it read}\ {\it B})\ {\it j})]$

\vspace{0.15in}
$Nt(s(m)): {\it j}, (SA{{}{\uparrow}{}}Nt(s(n))){{}{\downarrow}{}}SA: \lambda B\forall C[({\it thing}\ {\it C})\rightarrow ({\it B}\ {\it C})], Q/\mbox{$\hat{\ }$}(Sf{{}{\uparrow}{}}(ND\backslash (Nt(s(E))\backslash Sf))): \lambda F({\it F}\ {\it read})\ \Rightarrow\ Sf$

\vspace{0.15in}
(d(5)) ${\bf jan}{+}{\bf alles}{+}{\bf kan}{+}{\bf lezen}: Sf$

\vspace{0.15in}
$Nt(s(m)): {\it j}, (SA{{}{\uparrow}{}}Nt(s(n))){{}{\downarrow}{}}SA: \lambda B\forall C[({\it thing}\ {\it C})\rightarrow ({\it B}\ {\it C})], (ND\backslash Si){{}{\downarrow}{}}(ND\backslash Sf): \lambda E\lambda F(({\it isable}\ ({\it E}\ {\it F}))\ {\it F}), \mbox{$\rhd^{-1}$}(NG\backslash (NH\backslash Si)): {\it read}\ \Rightarrow\ Sf$

\vspace{0.15in}

\prooftree
\prooftree
\prooftree
\prooftree
\prooftree
\prooftree
\prooftree
\justifies
Nt(s(n))\ \Rightarrow\ Nt(s(n))
\endprooftree
\prooftree
\prooftree
\justifies
Nt(s(m))\ \Rightarrow\ Nt(s(m))
\endprooftree
\prooftree
\justifies
\mbox{\fbox{$Si\{{\tt 1}\}$}}\ \Rightarrow\ Si
\endprooftree
\justifies
Nt(s(m)), \mbox{\fbox{$Nt(s(m))\backslash Si\{{\tt 1}\}$}}\ \Rightarrow\ Si
\using {\backslash}L
\endprooftree
\justifies
Nt(s(m)), Nt(s(n)), \mbox{\fbox{$Nt(s(n))\backslash (Nt(s(m))\backslash Si)\{{\tt 1}\}$}}\ \Rightarrow\ Si
\using {\backslash}L
\endprooftree
\justifies
Nt(s(m)), Nt(s(n)), {\tt 1}, \mbox{\fbox{$\mbox{$\rhd^{-1}$}(Nt(s(n))\backslash (Nt(s(m))\backslash Si))$}}\ \Rightarrow\ Si
\using {\rhd^{-1}}L
\endprooftree
\justifies
Nt(s(n)), {\tt 1}, \mbox{$\rhd^{-1}$}(Nt(s(n))\backslash (Nt(s(m))\backslash Si))\ \Rightarrow\ Nt(s(m))\backslash Si
\using {\backslash}R
\endprooftree
\prooftree
\prooftree
\justifies
Nt(s(m))\ \Rightarrow\ Nt(s(m))
\endprooftree
\prooftree
\justifies
\mbox{\fbox{$Sf$}}\ \Rightarrow\ Sf
\endprooftree
\justifies
Nt(s(m)), \mbox{\fbox{$Nt(s(m))\backslash Sf$}}\ \Rightarrow\ Sf
\using {\backslash}L
\endprooftree
\justifies
Nt(s(m)), Nt(s(n)), \mbox{\fbox{$(Nt(s(m))\backslash Si){{}{\downarrow}{}}(Nt(s(m))\backslash Sf)$}}, \mbox{$\rhd^{-1}$}(Nt(s(n))\backslash (Nt(s(m))\backslash Si))\ \Rightarrow\ Sf
\using {\downarrow}L
\endprooftree
\justifies
Nt(s(m)), {\tt 1}, (Nt(s(m))\backslash Si){{}{\downarrow}{}}(Nt(s(m))\backslash Sf), \mbox{$\rhd^{-1}$}(Nt(s(n))\backslash (Nt(s(m))\backslash Si))\ \Rightarrow\ Sf{{}{\uparrow}{}}Nt(s(n))
\using {\uparrow}R
\endprooftree
\prooftree
\justifies
\mbox{\fbox{$Sf$}}\ \Rightarrow\ Sf
\endprooftree
\justifies
Nt(s(m)), \mbox{\fbox{$(Sf{{}{\uparrow}{}}Nt(s(n))){{}{\downarrow}{}}Sf$}}, (Nt(s(m))\backslash Si){{}{\downarrow}{}}(Nt(s(m))\backslash Sf), \mbox{$\rhd^{-1}$}(Nt(s(n))\backslash (Nt(s(m))\backslash Si))\ \Rightarrow\ Sf
\using {\downarrow}L
\endprooftree

\vspace{0.15in}

$\forall B[({\it thing}\ {\it B})\rightarrow (({\it isable}\ (({\it read}\ {\it B})\ {\it j}))\ {\it j})]$

\vspace{0.15in}
$Nt(s(m)): {\it j}, (SA{{}{\uparrow}{}}Nt(s(n))){{}{\downarrow}{}}SA: \lambda B\forall C[({\it thing}\ {\it C})\rightarrow ({\it B}\ {\it C})], Q/\mbox{$\hat{\ }$}(Sf{{}{\uparrow}{}}((ND\backslash Si){{}{\downarrow}{}}(ND\backslash Sf))): \lambda E({\it E}\ \lambda F\lambda G(({\it isable}\ ({\it F}\ {\it G}))\ {\it G})), \mbox{$\rhd^{-1}$}(NH\backslash (NI\backslash Si)): {\it read}\ \Rightarrow\ Sf$

\vspace{0.15in}
(d(6)) ${\bf jan}{+}{\bf cecilia}{+}{\bf henk}{+}{\bf de}{+}{\bf nijlpaarden}{+}{\bf zag}{+}{\bf helpen}{+}{\bf voeren}: Sf$

\vspace{0.15in}
$Nt(s(m)): {\it j}, Nt(s(f)): {\it c}, Nt(s(m)): {\it h}, Nt(s(A))/{\it CN}{\it A}: {\it the}, {\it CN}{\it p(n)}: {\it hippos}, (Nt(s(B))\backslash Si){{}{\downarrow}{}}(NC\backslash (Nt(s(B))\backslash Sf)): \lambda D\lambda E(({\it saw}\ ({\it D}\ {\it E}))\ {\it E}), \mbox{$\rhd^{-1}$}((NF\backslash Si){{}{\downarrow}{}}(NG\backslash (NF\backslash Si))): \lambda H\lambda I(({\it help}\ ({\it H}\ {\it I}))\ {\it I}), \mbox{$\rhd^{-1}$}(NJ\backslash (NK\backslash Si)): {\it feed}\ \Rightarrow\ Sf$

\vspace{0.15in}

\begin{center}
\rotatebox{-90}{
\resizebox{\textheight}{!}{
\prooftree
\prooftree
\prooftree
\prooftree
\prooftree
\prooftree
\prooftree
\prooftree
\prooftree
\justifies
{\it CN}{\it p(n)}\ \Rightarrow\ {\it CN}{\it p(n)}
\endprooftree
\prooftree
\justifies
\mbox{\fbox{$Nt(s(p(n)))$}}\ \Rightarrow\ Nt(s(p(n)))
\endprooftree
\justifies
\mbox{\fbox{$Nt(s(p(n)))/{\it CN}{\it p(n)}$}}, {\it CN}{\it p(n)}\ \Rightarrow\ Nt(s(p(n)))
\using {/}L
\endprooftree
\prooftree
\prooftree
\justifies
Nt(s(m))\ \Rightarrow\ Nt(s(m))
\endprooftree
\prooftree
\justifies
\mbox{\fbox{$Si\{{\tt 1}\}$}}\ \Rightarrow\ Si
\endprooftree
\justifies
Nt(s(m)), \mbox{\fbox{$Nt(s(m))\backslash Si\{{\tt 1}\}$}}\ \Rightarrow\ Si
\using {\backslash}L
\endprooftree
\justifies
Nt(s(m)), Nt(s(p(n)))/{\it CN}{\it p(n)}, {\it CN}{\it p(n)}, \mbox{\fbox{$Nt(s(p(n)))\backslash (Nt(s(m))\backslash Si)\{{\tt 1}\}$}}\ \Rightarrow\ Si
\using {\backslash}L
\endprooftree
\justifies
Nt(s(m)), Nt(s(p(n)))/{\it CN}{\it p(n)}, {\it CN}{\it p(n)}, {\tt 1}, \mbox{\fbox{$\mbox{$\rhd^{-1}$}(Nt(s(p(n)))\backslash (Nt(s(m))\backslash Si))$}}\ \Rightarrow\ Si
\using {\rhd^{-1}}L
\endprooftree
\justifies
Nt(s(p(n)))/{\it CN}{\it p(n)}, {\it CN}{\it p(n)}, {\tt 1}, \mbox{$\rhd^{-1}$}(Nt(s(p(n)))\backslash (Nt(s(m))\backslash Si))\ \Rightarrow\ Nt(s(m))\backslash Si
\using {\backslash}R
\endprooftree
\prooftree
\prooftree
\justifies
Nt(s(m))\ \Rightarrow\ Nt(s(m))
\endprooftree
\prooftree
\prooftree
\justifies
Nt(s(m))\ \Rightarrow\ Nt(s(m))
\endprooftree
\prooftree
\justifies
\mbox{\fbox{$Si\{{\tt 1}\}$}}\ \Rightarrow\ Si
\endprooftree
\justifies
Nt(s(m)), \mbox{\fbox{$Nt(s(m))\backslash Si\{{\tt 1}\}$}}\ \Rightarrow\ Si
\using {\backslash}L
\endprooftree
\justifies
Nt(s(m)), Nt(s(m)), \mbox{\fbox{$Nt(s(m))\backslash (Nt(s(m))\backslash Si)\{{\tt 1}\}$}}\ \Rightarrow\ Si
\using {\backslash}L
\endprooftree
\justifies
Nt(s(m)), Nt(s(m)), Nt(s(p(n)))/{\it CN}{\it p(n)}, {\it CN}{\it p(n)}, \mbox{\fbox{$(Nt(s(m))\backslash Si){{}{\downarrow}{}}(Nt(s(m))\backslash (Nt(s(m))\backslash Si))\{{\tt 1}\}$}}, \mbox{$\rhd^{-1}$}(Nt(s(p(n)))\backslash (Nt(s(m))\backslash Si))\ \Rightarrow\ Si
\using {\downarrow}L
\endprooftree
\justifies
Nt(s(m)), Nt(s(m)), Nt(s(p(n)))/{\it CN}{\it p(n)}, {\it CN}{\it p(n)}, {\tt 1}, \mbox{\fbox{$\mbox{$\rhd^{-1}$}((Nt(s(m))\backslash Si){{}{\downarrow}{}}(Nt(s(m))\backslash (Nt(s(m))\backslash Si)))$}}, \mbox{$\rhd^{-1}$}(Nt(s(p(n)))\backslash (Nt(s(m))\backslash Si))\ \Rightarrow\ Si
\using {\rhd^{-1}}L
\endprooftree
\justifies
Nt(s(m)), Nt(s(p(n)))/{\it CN}{\it p(n)}, {\it CN}{\it p(n)}, {\tt 1}, \mbox{$\rhd^{-1}$}((Nt(s(m))\backslash Si){{}{\downarrow}{}}(Nt(s(m))\backslash (Nt(s(m))\backslash Si))), \mbox{$\rhd^{-1}$}(Nt(s(p(n)))\backslash (Nt(s(m))\backslash Si))\ \Rightarrow\ Nt(s(m))\backslash Si
\using {\backslash}R
\endprooftree
\prooftree
\prooftree
\justifies
Nt(s(f))\ \Rightarrow\ Nt(s(f))
\endprooftree
\prooftree
\prooftree
\justifies
Nt(s(m))\ \Rightarrow\ Nt(s(m))
\endprooftree
\prooftree
\justifies
\mbox{\fbox{$Sf$}}\ \Rightarrow\ Sf
\endprooftree
\justifies
Nt(s(m)), \mbox{\fbox{$Nt(s(m))\backslash Sf$}}\ \Rightarrow\ Sf
\using {\backslash}L
\endprooftree
\justifies
Nt(s(m)), Nt(s(f)), \mbox{\fbox{$Nt(s(f))\backslash (Nt(s(m))\backslash Sf)$}}\ \Rightarrow\ Sf
\using {\backslash}L
\endprooftree
\justifies
Nt(s(m)), Nt(s(f)), Nt(s(m)), Nt(s(p(n)))/{\it CN}{\it p(n)}, {\it CN}{\it p(n)}, \mbox{\fbox{$(Nt(s(m))\backslash Si){{}{\downarrow}{}}(Nt(s(f))\backslash (Nt(s(m))\backslash Sf))$}}, \mbox{$\rhd^{-1}$}((Nt(s(m))\backslash Si){{}{\downarrow}{}}(Nt(s(m))\backslash (Nt(s(m))\backslash Si))), \mbox{$\rhd^{-1}$}(Nt(s(p(n)))\backslash (Nt(s(m))\backslash Si))\ \Rightarrow\ Sf
\using {\downarrow}L
\endprooftree}}
\end{center}

\vspace{0.15in}

$((({\it saw}\ ((({\it help}\ (({\it feed}\ ({\it the}\ {\it hippos}))\ {\it h}))\ {\it h})\ {\it c}))\ {\it c})\ {\it j})$

\vspace{0.15in}
(d(7)) ${\bf wil}{+}{\bf jan}{+}{\bf boeken}{+}{\bf lezen}: Q$

\vspace{0.15in}
$(NA\backslash Si){{}{\downarrow}{}}(NA\backslash Sf): \lambda B\lambda C(({\it want}\ ({\it B}\ {\it C}))\ {\it C}), Nt(s(m)): {\it j}, Np(n): {\it books}, \mbox{$\rhd^{-1}$}(ND\backslash (NE\backslash Si)): {\it read}\ \Rightarrow\ Q$

\vspace{0.15in}
$Q/\mbox{$\hat{\ }$}(Sf{{}{\uparrow}{}}((NA\backslash Si){{}{\downarrow}{}}(NA\backslash Sf))): \lambda B({\it B}\ \lambda C\lambda D(({\it want}\ ({\it C}\ {\it D}))\ {\it D})), Nt(s(m)): {\it j}, Np(n): {\it books}, \mbox{$\rhd^{-1}$}(NE\backslash (NF\backslash Si)): {\it read}\ \Rightarrow\ Q$

\vspace{0.15in}

\prooftree
\prooftree
\prooftree
\prooftree
\prooftree
\prooftree
\prooftree
\prooftree
\justifies
Np(n)\ \Rightarrow\ Np(n)
\endprooftree
\prooftree
\prooftree
\justifies
Nt(s(m))\ \Rightarrow\ Nt(s(m))
\endprooftree
\prooftree
\justifies
\mbox{\fbox{$Si\{{\tt 1}\}$}}\ \Rightarrow\ Si
\endprooftree
\justifies
Nt(s(m)), \mbox{\fbox{$Nt(s(m))\backslash Si\{{\tt 1}\}$}}\ \Rightarrow\ Si
\using {\backslash}L
\endprooftree
\justifies
Nt(s(m)), Np(n), \mbox{\fbox{$Np(n)\backslash (Nt(s(m))\backslash Si)\{{\tt 1}\}$}}\ \Rightarrow\ Si
\using {\backslash}L
\endprooftree
\justifies
Nt(s(m)), Np(n), {\tt 1}, \mbox{\fbox{$\mbox{$\rhd^{-1}$}(Np(n)\backslash (Nt(s(m))\backslash Si))$}}\ \Rightarrow\ Si
\using {\rhd^{-1}}L
\endprooftree
\justifies
Np(n), {\tt 1}, \mbox{$\rhd^{-1}$}(Np(n)\backslash (Nt(s(m))\backslash Si))\ \Rightarrow\ Nt(s(m))\backslash Si
\using {\backslash}R
\endprooftree
\prooftree
\prooftree
\justifies
Nt(s(m))\ \Rightarrow\ Nt(s(m))
\endprooftree
\prooftree
\justifies
\mbox{\fbox{$Sf$}}\ \Rightarrow\ Sf
\endprooftree
\justifies
Nt(s(m)), \mbox{\fbox{$Nt(s(m))\backslash Sf$}}\ \Rightarrow\ Sf
\using {\backslash}L
\endprooftree
\justifies
Nt(s(m)), Np(n), \mbox{\fbox{$(Nt(s(m))\backslash Si){{}{\downarrow}{}}(Nt(s(m))\backslash Sf)$}}, \mbox{$\rhd^{-1}$}(Np(n)\backslash (Nt(s(m))\backslash Si))\ \Rightarrow\ Sf
\using {\downarrow}L
\endprooftree
\justifies
Nt(s(m)), Np(n), {\tt 1}, \mbox{$\rhd^{-1}$}(Np(n)\backslash (Nt(s(m))\backslash Si))\ \Rightarrow\ Sf{{}{\uparrow}{}}((Nt(s(m))\backslash Si){{}{\downarrow}{}}(Nt(s(m))\backslash Sf))
\using {\uparrow}R
\endprooftree
\justifies
Nt(s(m)), Np(n), \mbox{$\rhd^{-1}$}(Np(n)\backslash (Nt(s(m))\backslash Si))\ \Rightarrow\ \fbox{$\mbox{$\hat{\ }$}(Sf{{}{\uparrow}{}}((Nt(s(m))\backslash Si){{}{\downarrow}{}}(Nt(s(m))\backslash Sf)))$}
\using \mbox{$\hat{\ }$}R
\endprooftree
\prooftree
\justifies
\mbox{\fbox{$Q$}}\ \Rightarrow\ Q
\endprooftree
\justifies
\mbox{\fbox{$Q/\mbox{$\hat{\ }$}(Sf{{}{\uparrow}{}}((Nt(s(m))\backslash Si){{}{\downarrow}{}}(Nt(s(m))\backslash Sf)))$}}, Nt(s(m)), Np(n), \mbox{$\rhd^{-1}$}(Np(n)\backslash (Nt(s(m))\backslash Si))\ \Rightarrow\ Q
\using {/}L
\endprooftree

\vspace{0.15in}

$(({\it want}\ (({\it read}\ {\it books})\ {\it j}))\ {\it j})$

\vspace{0.15in}
(d(8)) ${\bf jan}{+}{\bf wil}{+}{\bf boeken}{+}{\bf lezen}: Nt(s(m)){\bullet}\mbox{$\hat{\ }$}(Q{{}{\uparrow}{}}Nt(s(m)))$

\vspace{0.15in}
$Nt(s(m)): {\it j}, (NA\backslash Si){{}{\downarrow}{}}(NA\backslash Sf): \lambda B\lambda C(({\it want}\ ({\it B}\ {\it C}))\ {\it C}), Np(n): {\it books}, \mbox{$\rhd^{-1}$}(ND\backslash (NE\backslash Si)): {\it read}\ \Rightarrow\ Nt(s(m)){\bullet}\mbox{$\hat{\ }$}(Q{{}{\uparrow}{}}Nt(s(m)))$

\vspace{0.15in}
$Nt(s(m)): {\it j}, Q/\mbox{$\hat{\ }$}(Sf{{}{\uparrow}{}}((NA\backslash Si){{}{\downarrow}{}}(NA\backslash Sf))): \lambda B({\it B}\ \lambda C\lambda D(({\it want}\ ({\it C}\ {\it D}))\ {\it D})), Np(n): {\it books}, \mbox{$\rhd^{-1}$}(NE\backslash (NF\backslash Si)): {\it read}\ \Rightarrow\ Nt(s(m)){\bullet}\mbox{$\hat{\ }$}(Q{{}{\uparrow}{}}Nt(s(m)))$

\vspace{0.15in}

\prooftree
\prooftree
\justifies
Nt(s(m))\ \Rightarrow\ Nt(s(m))
\endprooftree
\prooftree
\prooftree
\prooftree
\prooftree
\prooftree
\prooftree
\prooftree
\prooftree
\prooftree
\prooftree
\justifies
Nt(s(m))\ \Rightarrow\ Nt(s(m))
\endprooftree
\prooftree
\prooftree
\justifies
Np(n)\ \Rightarrow\ Np(n)
\endprooftree
\prooftree
\justifies
\mbox{\fbox{$Si\{{\tt 1}\}$}}\ \Rightarrow\ Si
\endprooftree
\justifies
Np(n), \mbox{\fbox{$Np(n)\backslash Si\{{\tt 1}\}$}}\ \Rightarrow\ Si
\using {\backslash}L
\endprooftree
\justifies
Np(n), Nt(s(m)), \mbox{\fbox{$Nt(s(m))\backslash (Np(n)\backslash Si)\{{\tt 1}\}$}}\ \Rightarrow\ Si
\using {\backslash}L
\endprooftree
\justifies
Np(n), Nt(s(m)), {\tt 1}, \mbox{\fbox{$\mbox{$\rhd^{-1}$}(Nt(s(m))\backslash (Np(n)\backslash Si))$}}\ \Rightarrow\ Si
\using {\rhd^{-1}}L
\endprooftree
\justifies
Nt(s(m)), {\tt 1}, \mbox{$\rhd^{-1}$}(Nt(s(m))\backslash (Np(n)\backslash Si))\ \Rightarrow\ Np(n)\backslash Si
\using {\backslash}R
\endprooftree
\prooftree
\prooftree
\justifies
Np(n)\ \Rightarrow\ Np(n)
\endprooftree
\prooftree
\justifies
\mbox{\fbox{$Sf$}}\ \Rightarrow\ Sf
\endprooftree
\justifies
Np(n), \mbox{\fbox{$Np(n)\backslash Sf$}}\ \Rightarrow\ Sf
\using {\backslash}L
\endprooftree
\justifies
Np(n), Nt(s(m)), \mbox{\fbox{$(Np(n)\backslash Si){{}{\downarrow}{}}(Np(n)\backslash Sf)$}}, \mbox{$\rhd^{-1}$}(Nt(s(m))\backslash (Np(n)\backslash Si))\ \Rightarrow\ Sf
\using {\downarrow}L
\endprooftree
\justifies
Np(n), Nt(s(m)), {\tt 1}, \mbox{$\rhd^{-1}$}(Nt(s(m))\backslash (Np(n)\backslash Si))\ \Rightarrow\ Sf{{}{\uparrow}{}}((Np(n)\backslash Si){{}{\downarrow}{}}(Np(n)\backslash Sf))
\using {\uparrow}R
\endprooftree
\justifies
Np(n), Nt(s(m)), \mbox{$\rhd^{-1}$}(Nt(s(m))\backslash (Np(n)\backslash Si))\ \Rightarrow\ \fbox{$\mbox{$\hat{\ }$}(Sf{{}{\uparrow}{}}((Np(n)\backslash Si){{}{\downarrow}{}}(Np(n)\backslash Sf)))$}
\using \mbox{$\hat{\ }$}R
\endprooftree
\prooftree
\justifies
\mbox{\fbox{$Q$}}\ \Rightarrow\ Q
\endprooftree
\justifies
\mbox{\fbox{$Q/\mbox{$\hat{\ }$}(Sf{{}{\uparrow}{}}((Np(n)\backslash Si){{}{\downarrow}{}}(Np(n)\backslash Sf)))$}}, Np(n), Nt(s(m)), \mbox{$\rhd^{-1}$}(Nt(s(m))\backslash (Np(n)\backslash Si))\ \Rightarrow\ Q
\using {/}L
\endprooftree
\justifies
Q/\mbox{$\hat{\ }$}(Sf{{}{\uparrow}{}}((Np(n)\backslash Si){{}{\downarrow}{}}(Np(n)\backslash Sf))), Np(n), {\tt 1}, \mbox{$\rhd^{-1}$}(Nt(s(m))\backslash (Np(n)\backslash Si))\ \Rightarrow\ Q{{}{\uparrow}{}}Nt(s(m))
\using {\uparrow}R
\endprooftree
\justifies
Q/\mbox{$\hat{\ }$}(Sf{{}{\uparrow}{}}((Np(n)\backslash Si){{}{\downarrow}{}}(Np(n)\backslash Sf))), Np(n), \mbox{$\rhd^{-1}$}(Nt(s(m))\backslash (Np(n)\backslash Si))\ \Rightarrow\ \fbox{$\mbox{$\hat{\ }$}(Q{{}{\uparrow}{}}Nt(s(m)))$}
\using \mbox{$\hat{\ }$}R
\endprooftree
\justifies
Nt(s(m)), Q/\mbox{$\hat{\ }$}(Sf{{}{\uparrow}{}}((Np(n)\backslash Si){{}{\downarrow}{}}(Np(n)\backslash Sf))), Np(n), \mbox{$\rhd^{-1}$}(Nt(s(m))\backslash (Np(n)\backslash Si))\ \Rightarrow\ \fbox{$Nt(s(m)){\bullet}\mbox{$\hat{\ }$}(Q{{}{\uparrow}{}}Nt(s(m)))$}
\using {\bullet}R
\endprooftree

\vspace{0.15in}

$({\it j}, \lambda A(({\it want}\ (({\it read}\ {\it A})\ {\it books}))\ {\it books}))$

\vspace{0.15in}

\prooftree
\prooftree
\justifies
Nt(s(m))\ \Rightarrow\ Nt(s(m))
\endprooftree
\prooftree
\prooftree
\prooftree
\prooftree
\prooftree
\prooftree
\prooftree
\prooftree
\prooftree
\prooftree
\justifies
Np(n)\ \Rightarrow\ Np(n)
\endprooftree
\prooftree
\prooftree
\justifies
Nt(s(m))\ \Rightarrow\ Nt(s(m))
\endprooftree
\prooftree
\justifies
\mbox{\fbox{$Si\{{\tt 1}\}$}}\ \Rightarrow\ Si
\endprooftree
\justifies
Nt(s(m)), \mbox{\fbox{$Nt(s(m))\backslash Si\{{\tt 1}\}$}}\ \Rightarrow\ Si
\using {\backslash}L
\endprooftree
\justifies
Nt(s(m)), Np(n), \mbox{\fbox{$Np(n)\backslash (Nt(s(m))\backslash Si)\{{\tt 1}\}$}}\ \Rightarrow\ Si
\using {\backslash}L
\endprooftree
\justifies
Nt(s(m)), Np(n), {\tt 1}, \mbox{\fbox{$\mbox{$\rhd^{-1}$}(Np(n)\backslash (Nt(s(m))\backslash Si))$}}\ \Rightarrow\ Si
\using {\rhd^{-1}}L
\endprooftree
\justifies
Np(n), {\tt 1}, \mbox{$\rhd^{-1}$}(Np(n)\backslash (Nt(s(m))\backslash Si))\ \Rightarrow\ Nt(s(m))\backslash Si
\using {\backslash}R
\endprooftree
\prooftree
\prooftree
\justifies
Nt(s(m))\ \Rightarrow\ Nt(s(m))
\endprooftree
\prooftree
\justifies
\mbox{\fbox{$Sf$}}\ \Rightarrow\ Sf
\endprooftree
\justifies
Nt(s(m)), \mbox{\fbox{$Nt(s(m))\backslash Sf$}}\ \Rightarrow\ Sf
\using {\backslash}L
\endprooftree
\justifies
Nt(s(m)), Np(n), \mbox{\fbox{$(Nt(s(m))\backslash Si){{}{\downarrow}{}}(Nt(s(m))\backslash Sf)$}}, \mbox{$\rhd^{-1}$}(Np(n)\backslash (Nt(s(m))\backslash Si))\ \Rightarrow\ Sf
\using {\downarrow}L
\endprooftree
\justifies
Nt(s(m)), Np(n), {\tt 1}, \mbox{$\rhd^{-1}$}(Np(n)\backslash (Nt(s(m))\backslash Si))\ \Rightarrow\ Sf{{}{\uparrow}{}}((Nt(s(m))\backslash Si){{}{\downarrow}{}}(Nt(s(m))\backslash Sf))
\using {\uparrow}R
\endprooftree
\justifies
Nt(s(m)), Np(n), \mbox{$\rhd^{-1}$}(Np(n)\backslash (Nt(s(m))\backslash Si))\ \Rightarrow\ \fbox{$\mbox{$\hat{\ }$}(Sf{{}{\uparrow}{}}((Nt(s(m))\backslash Si){{}{\downarrow}{}}(Nt(s(m))\backslash Sf)))$}
\using \mbox{$\hat{\ }$}R
\endprooftree
\prooftree
\justifies
\mbox{\fbox{$Q$}}\ \Rightarrow\ Q
\endprooftree
\justifies
\mbox{\fbox{$Q/\mbox{$\hat{\ }$}(Sf{{}{\uparrow}{}}((Nt(s(m))\backslash Si){{}{\downarrow}{}}(Nt(s(m))\backslash Sf)))$}}, Nt(s(m)), Np(n), \mbox{$\rhd^{-1}$}(Np(n)\backslash (Nt(s(m))\backslash Si))\ \Rightarrow\ Q
\using {/}L
\endprooftree
\justifies
Q/\mbox{$\hat{\ }$}(Sf{{}{\uparrow}{}}((Nt(s(m))\backslash Si){{}{\downarrow}{}}(Nt(s(m))\backslash Sf))), {\tt 1}, Np(n), \mbox{$\rhd^{-1}$}(Np(n)\backslash (Nt(s(m))\backslash Si))\ \Rightarrow\ Q{{}{\uparrow}{}}Nt(s(m))
\using {\uparrow}R
\endprooftree
\justifies
Q/\mbox{$\hat{\ }$}(Sf{{}{\uparrow}{}}((Nt(s(m))\backslash Si){{}{\downarrow}{}}(Nt(s(m))\backslash Sf))), Np(n), \mbox{$\rhd^{-1}$}(Np(n)\backslash (Nt(s(m))\backslash Si))\ \Rightarrow\ \fbox{$\mbox{$\hat{\ }$}(Q{{}{\uparrow}{}}Nt(s(m)))$}
\using \mbox{$\hat{\ }$}R
\endprooftree
\justifies
Nt(s(m)), Q/\mbox{$\hat{\ }$}(Sf{{}{\uparrow}{}}((Nt(s(m))\backslash Si){{}{\downarrow}{}}(Nt(s(m))\backslash Sf))), Np(n), \mbox{$\rhd^{-1}$}(Np(n)\backslash (Nt(s(m))\backslash Si))\ \Rightarrow\ \fbox{$Nt(s(m)){\bullet}\mbox{$\hat{\ }$}(Q{{}{\uparrow}{}}Nt(s(m)))$}
\using {\bullet}R
\endprooftree

\vspace{0.15in}

$({\it j}, \lambda A(({\it want}\ (({\it read}\ {\it books})\ {\it A}))\ {\it A}))$

}

\setlength{\parindent}{0.22in}

\chapter{Relativization}

\label{relchap}

In this chapter we present an account of relativisation.

\section{Routes we do not take}

\label{others}

Szaboloczi (1983\cite{szab83}) and Steedman (1987\cite{steedman:cgpg}) aim to
account for parasitic gaps in combinatory categorial grammar (CCG) by means of the
combinator {\bf S} such that ${\bf S}\ x\ y\ z\ =\ (x\ z)\ (y\ z)$,
for example positing a combinatory schema:
\disp{$
y\ass Y/Z, x\ass (Y\bsl X)/Z\yields {\bf S}\ x\ y\ass X/Z
$}
Such a schema makes no sense from the point of view of the logicisation
of grammar pursued here. The rule is not Lambek-valid and any semantics
validating it would also validate schemata which overgenerate massively.
So much the worse, the proponents of CCG would say, for grammar as logic:
grammar is a formal system but not a logic, and one should not care about
things like soundness and completeness.

CCG and type logical grammar agree on the task of defining syntax and
semantics of the (object) natural language. 
What is curious about CCG is that at the same time it declines to consider
syntax (proof theory) and semantics (model theory) of the (meta-)linguistic
formalism. 
A CCG account of parasitic gaps, which employs just the directional slashes and a minimum
of combinatory schemata, must capture the effects of structural inhibition
(islands) and structural facilitation (parasiticy) by good fortune in the interaction of the 
combinatory schemata chosen and the categorial types occurring in grammar.
Here control of structural inhibition by bracket modalities
and control of structural facilitation by subexponentials are separated in an analysis
recognising the distinct algebraic roles of variation from an associative and 
from a linear
regime. This type logical approach lets us state our analysis with clarity
in the knowledge that whatever the empirical adequacy,
the metatheoretical facts are what they are. 
In CCG the metatheory is not logically investigated.

It is interesting to ask why we treat medial extraction here with $\univexp$ rather than with
$\circum$ as illustrated in Section~\ref{dismultsect} of Chapter~\ref{slsc}
(cf.~also Moortgat 1988\cite{moortgat:phd};
Muskens 2003\cite{muskens:03};
Mihali\v{c}ek and Pollard 2012\cite{mihpoll:fg};
Barker and Shan 2015\cite{barkershan:oxford}; and
Kubota and Levine 2015\cite{kl:lap}).
The answer is that, on the one hand, $\circum$ as defined
does not respect island constraints and, on the other hand,
$\circum$ does not extend to parasitic gaps:
it is unclear how a single local inference rule can account for
unbounded recursive nesting of parasitic gaps in subislands. 
Our treatment in terms of $\univexp$
both respects islands, and extends to (unbounded numbers of) parasitic gaps
through iteration of contraction.\footnote{We note that the discontinuity operators
serve to account for the pied-piping aspect of relativisation
(e.g.\ Morrill, Valent\'{\i}n and Fadda 2011\cite{mvf:tdc}),
see example~(\ref{piedpipingex}) in Chapter~\ref{mvfchap}.}

An option available in both CCG and type logical grammar is to attempt to
analyse the nonlinearity of parasitic extraction not syntactically
but lexically. Thus for example Jansche and Vasishth (2002\cite{steedrev}) propose induction of
parasitic gaps in adverbial clauses by a lexicalised gap-duplicating effect in the
adverbial head. All contexts allowing parasitic gaps would require
a corresponding gap-duplicating lexical ambiguity. The appeal to lexical ambiguity
in lexical grammar formalisms is as frequent as it is untenable. Every ambiguity
of every item doubles the lexical insertion search space. And in the case
in hand there is to our knowledge no independent evidence,
such as difference in meaning,
for positing lexical ambiguity underlying parasitic extraction.
We continue on the assumption that it is indeed a syntactic phenomenon.

\section{Relativisation}

\label{relsec}

Our account of relativisation
rests on the lexical projection of islands by argument bracketing
(\mybrack{}) and value antibracketing (\abrack{}),
and a single relative pronoun type of overall shape
$R/((\mybrack{}N\iaconj\univexp N)\bsl S)$ for both subject and object relativisation.
In analysis of the body of relative clauses the higher order succedent argument essentially of form
$\mybrack{}N\iaconj\univexp N$ is lowered into the antecedent according to the
deduction theorem;
in subject relativisation $\mybrack{}N$ is selected by conjunction left, 
and satisfies the (bracketed) subject valency.

In object relativisation $\univexp N$ is selected.
When the $\univexp L$ rule is applied to $\univexp N$,
the hypothetical subtype $N$ moves into the stoup,
from where it can move by $\univexp P$ to any (nonisland) position in its zone,
realising nonparasitic extraction.

However, in addition it can be copied by $\univexp C$ to the stoup of a newly
created weak island domain, realising parasitic extraction.
The $N$ in the outer stoop can be copied by $\univexp C$ repeatedly,
capturing that there may be parasitic gaps in any number of local weak
islands; at the end of this process it moves by $\univexp P$ to a host
position in its zone.
The $N$ in an inner stoup can also be copied by $\univexp C$ to the stoup of any number
of newly created weak subislands,
and so on recursively,
capturing that parasitic gaps can also be hosts to further parasitic gaps;
finally the stoup contents are copied by $\univexp P$ to an extraction site in their zone.

In this section we analyse examples illustrating the account of relativisation.
(cf.\ Morrill 2011, Chapter~5).
The first example is a minimal subject relativisation;
note that the relative clause is doubly bracketed,
corresponding to the fact that relative clauses are strong islands:\footnote{As we will see
relative clauses themselves, being doubly bracketed, will not allow parasitic gaps.
\label{relnopar}}
\disp{
${\bf man}{+}[[{\bf that}{+}{\bf walks}]]: {\it CN}{\it s(m)}$
\label{ex11}}
Lexical lookup yields the following,
where there is semantically inactive additive conjunction of the hypothetical subtypes
$\langle\rangle N$ for subject relativisation and $!\blacksquare N$ for object relativisation;
the (semantically inactive) modality on the object gap subtype is to permit object relativisation from
embedded modal/intensional domains:\footnote{The body of the relative clause is marked as a (semantically inactive) modal domain
in order to make it a scope island.
Thus where, say, \lingform{everyone\/} has a type $\mymod((S\scircum{}N)\sinfix{}S)$
the unmodalised hypothetical subtype $N$ cannot be bound outside the modal domain of the
body of a relative clause in which \lingform{everyone\/} occurs.}
\disp{
${\square}{\it CN}{\it s(m)}: {\it man}, [[{\blacksquare}{\forall}n({[]^{-1}}{[]^{-1}}({\it CN}{\it n}\backslash {\it CN}{\it n})/{\blacksquare}(({\langle\rangle}Nt(n){\sqcap}!{\blacksquare}Nt(n))\backslash Sf)):\\ \lambda A\lambda B\lambda C[({\it B}\ {\it C})\wedge ({\it A}\ {\it C})], {\square}({\langle\rangle}{\exists}gNt(s(g))\backslash Sf): \mbox{\^{}}\lambda D({\it Pres}\ (\mbox{\v{}}{\it walk}\ {\it D}))]]\ \Rightarrow\ {\it CN}{\it s(m)}$}
There is the derivation in Figure~\ref{mtw}, which starts with the relative clause doubly bracketed
(this will always be the case for relativisation). 
After elimination of the outer (semantically inactive)
modality of the relative pronoun, universal left instantiates it to agree with
masculine singular. Then $/L$ partitions in such a way as to select 
the intransitive verb body of the relative
clause as argument of the relative pronoun. 
In the righthand, 
value, 
subtree two antibracket eliminations cancel the double brackets before the head common
noun is modified. 
In the lefthand, 
argument,
subtree (inactive) box right is enabled since the antecedent is modalised,
and under right then lowers the additively conjoined hypothetical subtype
into the antecedent.
Observe how in the lefthand subtree
$\iaconj$L selects the subject relativisation
hypothetical subtype $\mybrack Nt(s(m))$;
the remaining subderivation is the usual intransitive sentence analysis.
\begin{figure}
\begin{center}
\scriptsize
\prooftree
\prooftree
\prooftree
\prooftree
\prooftree
\prooftree
\prooftree
\prooftree
\prooftree
\prooftree
\prooftree
\prooftree
\justifies
Nt(s(m))\ \Rightarrow\ Nt(s(m))
\endprooftree
\justifies
Nt(s(m))\ \Rightarrow\ \fbox{${\exists}gNt(s(g))$}
\using {\exists}R
\endprooftree
\justifies
[Nt(s(m))]\ \Rightarrow\ \fbox{${\langle\rangle}{\exists}gNt(s(g))$}
\using {\langle\rangle}R
\endprooftree
\prooftree
\justifies
\mbox{\fbox{$Sf$}}\ \Rightarrow\ Sf
\endprooftree
\justifies
[Nt(s(m))], \mbox{\fbox{${\langle\rangle}{\exists}gNt(s(g))\backslash Sf$}}\ \Rightarrow\ Sf
\using {\backslash}L
\endprooftree
\justifies
[Nt(s(m))], \mbox{\fbox{${\square}({\langle\rangle}{\exists}gNt(s(g))\backslash Sf)$}}\ \Rightarrow\ Sf
\using {\Box}L
\endprooftree
\justifies
{\langle\rangle}Nt(s(m)), {\square}({\langle\rangle}{\exists}gNt(s(g))\backslash Sf)\ \Rightarrow\ Sf
\using {\langle\rangle}L
\endprooftree
\justifies
\mbox{\fbox{${\langle\rangle}Nt(s(m)){\sqcap}!{\blacksquare}Nt(s(m))$}}, {\square}({\langle\rangle}{\exists}gNt(s(g))\backslash Sf)\ \Rightarrow\ Sf
\using {\sqcap}L
\endprooftree
\justifies
{\square}({\langle\rangle}{\exists}gNt(s(g))\backslash Sf)\ \Rightarrow\ ({\langle\rangle}Nt(s(m)){\sqcap}!{\blacksquare}Nt(s(m)))\backslash Sf
\using {\backslash}R
\endprooftree
\justifies
{\square}({\langle\rangle}{\exists}gNt(s(g))\backslash Sf)\ \Rightarrow\ {\blacksquare}(({\langle\rangle}Nt(s(m)){\sqcap}!{\blacksquare}Nt(s(m)))\backslash Sf)
\using {\blacksquare}R
\endprooftree
\prooftree
\prooftree
\prooftree
\prooftree
\prooftree
\justifies
\mbox{\fbox{${\it CN}{\it s(m)}$}}\ \Rightarrow\ {\it CN}{\it s(m)}
\endprooftree
\justifies
\mbox{\fbox{${\square}{\it CN}{\it s(m)}$}}\ \Rightarrow\ {\it CN}{\it s(m)}
\using {\Box}L
\endprooftree
\prooftree
\justifies
\mbox{\fbox{${\it CN}{\it s(m)}$}}\ \Rightarrow\ {\it CN}{\it s(m)}
\endprooftree
\justifies
{\square}{\it CN}{\it s(m)}, \mbox{\fbox{${\it CN}{\it s(m)}\backslash {\it CN}{\it s(m)}$}}\ \Rightarrow\ {\it CN}{\it s(m)}
\using {\backslash}L
\endprooftree
\justifies
{\square}{\it CN}{\it s(m)}, [\mbox{\fbox{${[]^{-1}}({\it CN}{\it s(m)}\backslash {\it CN}{\it s(m)})$}}]\ \Rightarrow\ {\it CN}{\it s(m)}
\using {[]^{-1}}L
\endprooftree
\justifies
{\square}{\it CN}{\it s(m)}, [[\mbox{\fbox{${[]^{-1}}{[]^{-1}}({\it CN}{\it s(m)}\backslash {\it CN}{\it s(m)})$}}]]\ \Rightarrow\ {\it CN}{\it s(m)}
\using {[]^{-1}}L
\endprooftree
\justifies
{\square}{\it CN}{\it s(m)}, [[\mbox{\fbox{${[]^{-1}}{[]^{-1}}({\it CN}{\it s(m)}\backslash {\it CN}{\it s(m)})/{\blacksquare}(({\langle\rangle}Nt(s(m)){\sqcap}!{\blacksquare}Nt(s(m)))\backslash Sf)$}}, {\square}({\langle\rangle}{\exists}gNt(s(g))\backslash Sf)]]\ \Rightarrow\ {\it CN}{\it s(m)}
\using {/}L
\endprooftree
\justifies
{\square}{\it CN}{\it s(m)}, [[\mbox{\fbox{${\forall}n({[]^{-1}}{[]^{-1}}({\it CN}{\it n}\backslash {\it CN}{\it n})/{\blacksquare}(({\langle\rangle}Nt(n){\sqcap}!{\blacksquare}Nt(n))\backslash Sf))$}}, {\square}({\langle\rangle}{\exists}gNt(s(g))\backslash Sf)]]\ \Rightarrow\ {\it CN}{\it s(m)}
\using {\forall}L
\endprooftree
\justifies
{\square}{\it CN}{\it s(m)}, [[\mbox{\fbox{${\blacksquare}{\forall}n({[]^{-1}}{[]^{-1}}({\it CN}{\it n}\backslash {\it CN}{\it n})/{\blacksquare}(({\langle\rangle}Nt(n){\sqcap}!{\blacksquare}Nt(n))\backslash Sf))$}}, {\square}({\langle\rangle}{\exists}gNt(s(g))\backslash Sf)]]\ \Rightarrow\ {\it CN}{\it s(m)}
\using {\blacksquare}L
\endprooftree
\end{center}
\caption{Derivation for \lingform{man that walks}}
\label{mtw}
\end{figure}
This delivers the required semantics:
\disp{
$\lambda C[(\mbox{\v{}}{\it man}\ {\it C})\wedge ({\it Pres}\ (\mbox{\v{}}{\it walk}\ {\it C}))]$}

\commentout{

The next example comprises a (subject) relative clause in the context
of an entire sentence:
\disp{
$[{\bf the}{+}{\bf man}{+}[[{\bf that}{+}{\bf walks}]]]{+}{\bf talks}: Sf$}
Lexical lookup yields:
\disp{
$[{\blacksquare}{\forall}n(Nt(n)/{\it CN}{\it n}): \iota , {\square}{\it CN}{\it s(m)}: {\it man}, [[{\blacksquare}{\forall}n({[]^{-1}}{[]^{-1}}({\it CN}{\it n}\backslash {\it CN}{\it n})/{\blacksquare}(({\langle\rangle}Nt(n){\sqcap}!{\blacksquare}Nt(n))\backslash Sf)): \lambda A\lambda B\lambda C[({\it B}\ {\it C})\wedge ({\it A}\ {\it C})], {\square}({\langle\rangle}{\exists}gNt(s(g))\backslash Sf): \mbox{\^{}}\lambda D({\it Pres}\ (\mbox{\v{}}{\it walk}\ {\it D}))]]], {\square}({\langle\rangle}{\exists}gNt(s(g))\backslash Sf): \mbox{\^{}}\lambda E({\it Pres}\ (\mbox{\v{}}{\it talk}\ {\it E}))\ \Rightarrow\ Sf$}
There is the derivation given in Figure~\ref{tmtwt}.
\begin{figure}
\begin{center}
\tiny
\prooftree
\prooftree
\prooftree
\prooftree
\prooftree
\prooftree
\prooftree
\prooftree
\prooftree
\prooftree
\prooftree
\prooftree
\prooftree
\prooftree
\prooftree
\prooftree
\prooftree
\prooftree
\prooftree
\justifies
Nt(s(m))\ \Rightarrow\ Nt(s(m))
\endprooftree
\justifies
Nt(s(m))\ \Rightarrow\ \fbox{${\exists}gNt(s(g))$}
\using {\exists}R
\endprooftree
\justifies
[Nt(s(m))]\ \Rightarrow\ \fbox{${\langle\rangle}{\exists}gNt(s(g))$}
\using {\langle\rangle}R
\endprooftree
\prooftree
\justifies
\mbox{\fbox{$Sf$}}\ \Rightarrow\ Sf
\endprooftree
\justifies
[Nt(s(m))], \mbox{\fbox{${\langle\rangle}{\exists}gNt(s(g))\backslash Sf$}}\ \Rightarrow\ Sf
\using {\backslash}L
\endprooftree
\justifies
[Nt(s(m))], \mbox{\fbox{${\square}({\langle\rangle}{\exists}gNt(s(g))\backslash Sf)$}}\ \Rightarrow\ Sf
\using {\Box}L
\endprooftree
\justifies
{\langle\rangle}Nt(s(m)), {\square}({\langle\rangle}{\exists}gNt(s(g))\backslash Sf)\ \Rightarrow\ Sf
\using {\langle\rangle}L
\endprooftree
\justifies
\mbox{\fbox{${\langle\rangle}Nt(s(m)){\sqcap}!{\blacksquare}Nt(s(m))$}}, {\square}({\langle\rangle}{\exists}gNt(s(g))\backslash Sf)\ \Rightarrow\ Sf
\using {\sqcap}L
\endprooftree
\justifies
{\square}({\langle\rangle}{\exists}gNt(s(g))\backslash Sf)\ \Rightarrow\ ({\langle\rangle}Nt(s(m)){\sqcap}!{\blacksquare}Nt(s(m)))\backslash Sf
\using {\backslash}R
\endprooftree
\justifies
{\square}({\langle\rangle}{\exists}gNt(s(g))\backslash Sf)\ \Rightarrow\ {\blacksquare}(({\langle\rangle}Nt(s(m)){\sqcap}!{\blacksquare}Nt(s(m)))\backslash Sf)
\using {\blacksquare}R
\endprooftree
\prooftree
\prooftree
\prooftree
\prooftree
\prooftree
\justifies
\mbox{\fbox{${\it CN}{\it s(m)}$}}\ \Rightarrow\ {\it CN}{\it s(m)}
\endprooftree
\justifies
\mbox{\fbox{${\square}{\it CN}{\it s(m)}$}}\ \Rightarrow\ {\it CN}{\it s(m)}
\using {\Box}L
\endprooftree
\prooftree
\justifies
\mbox{\fbox{${\it CN}{\it s(m)}$}}\ \Rightarrow\ {\it CN}{\it s(m)}
\endprooftree
\justifies
{\square}{\it CN}{\it s(m)}, \mbox{\fbox{${\it CN}{\it s(m)}\backslash {\it CN}{\it s(m)}$}}\ \Rightarrow\ {\it CN}{\it s(m)}
\using {\backslash}L
\endprooftree
\justifies
{\square}{\it CN}{\it s(m)}, [\mbox{\fbox{${[]^{-1}}({\it CN}{\it s(m)}\backslash {\it CN}{\it s(m)})$}}]\ \Rightarrow\ {\it CN}{\it s(m)}
\using {[]^{-1}}L
\endprooftree
\justifies
{\square}{\it CN}{\it s(m)}, [[\mbox{\fbox{${[]^{-1}}{[]^{-1}}({\it CN}{\it s(m)}\backslash {\it CN}{\it s(m)})$}}]]\ \Rightarrow\ {\it CN}{\it s(m)}
\using {[]^{-1}}L
\endprooftree
\justifies
{\square}{\it CN}{\it s(m)}, [[\mbox{\fbox{${[]^{-1}}{[]^{-1}}({\it CN}{\it s(m)}\backslash {\it CN}{\it s(m)})/{\blacksquare}(({\langle\rangle}Nt(s(m)){\sqcap}!{\blacksquare}Nt(s(m)))\backslash Sf)$}}, {\square}({\langle\rangle}{\exists}gNt(s(g))\backslash Sf)]]\ \Rightarrow\ {\it CN}{\it s(m)}
\using {/}L
\endprooftree
\justifies
{\square}{\it CN}{\it s(m)}, [[\mbox{\fbox{${\forall}n({[]^{-1}}{[]^{-1}}({\it CN}{\it n}\backslash {\it CN}{\it n})/{\blacksquare}(({\langle\rangle}Nt(n){\sqcap}!{\blacksquare}Nt(n))\backslash Sf))$}}, {\square}({\langle\rangle}{\exists}gNt(s(g))\backslash Sf)]]\ \Rightarrow\ {\it CN}{\it s(m)}
\using {\forall}L
\endprooftree
\justifies
{\square}{\it CN}{\it s(m)}, [[\mbox{\fbox{${\blacksquare}{\forall}n({[]^{-1}}{[]^{-1}}({\it CN}{\it n}\backslash {\it CN}{\it n})/{\blacksquare}(({\langle\rangle}Nt(n){\sqcap}!{\blacksquare}Nt(n))\backslash Sf))$}}, {\square}({\langle\rangle}{\exists}gNt(s(g))\backslash Sf)]]\ \Rightarrow\ {\it CN}{\it s(m)}
\using {\blacksquare}L
\endprooftree
\prooftree
\justifies
\mbox{\fbox{$Nt(s(m))$}}\ \Rightarrow\ Nt(s(m))
\endprooftree
\justifies
\mbox{\fbox{$Nt(s(m))/{\it CN}{\it s(m)}$}}, {\square}{\it CN}{\it s(m)}, [[{\blacksquare}{\forall}n({[]^{-1}}{[]^{-1}}({\it CN}{\it n}\backslash {\it CN}{\it n})/{\blacksquare}(({\langle\rangle}Nt(n){\sqcap}!{\blacksquare}Nt(n))\backslash Sf)), {\square}({\langle\rangle}{\exists}gNt(s(g))\backslash Sf)]]\ \Rightarrow\ Nt(s(m))
\using {/}L
\endprooftree
\justifies
\mbox{\fbox{${\forall}n(Nt(n)/{\it CN}{\it n})$}}, {\square}{\it CN}{\it s(m)}, [[{\blacksquare}{\forall}n({[]^{-1}}{[]^{-1}}({\it CN}{\it n}\backslash {\it CN}{\it n})/{\blacksquare}(({\langle\rangle}Nt(n){\sqcap}!{\blacksquare}Nt(n))\backslash Sf)), {\square}({\langle\rangle}{\exists}gNt(s(g))\backslash Sf)]]\ \Rightarrow\ Nt(s(m))
\using {\forall}L
\endprooftree
\justifies
\mbox{\fbox{${\blacksquare}{\forall}n(Nt(n)/{\it CN}{\it n})$}}, {\square}{\it CN}{\it s(m)}, [[{\blacksquare}{\forall}n({[]^{-1}}{[]^{-1}}({\it CN}{\it n}\backslash {\it CN}{\it n})/{\blacksquare}(({\langle\rangle}Nt(n){\sqcap}!{\blacksquare}Nt(n))\backslash Sf)), {\square}({\langle\rangle}{\exists}gNt(s(g))\backslash Sf)]]\ \Rightarrow\ Nt(s(m))
\using {\blacksquare}L
\endprooftree
\justifies
{\blacksquare}{\forall}n(Nt(n)/{\it CN}{\it n}), {\square}{\it CN}{\it s(m)}, [[{\blacksquare}{\forall}n({[]^{-1}}{[]^{-1}}({\it CN}{\it n}\backslash {\it CN}{\it n})/{\blacksquare}(({\langle\rangle}Nt(n){\sqcap}!{\blacksquare}Nt(n))\backslash Sf)), {\square}({\langle\rangle}{\exists}gNt(s(g))\backslash Sf)]]\ \Rightarrow\ \fbox{${\exists}gNt(s(g))$}
\using {\exists}R
\endprooftree
\justifies
[{\blacksquare}{\forall}n(Nt(n)/{\it CN}{\it n}), {\square}{\it CN}{\it s(m)}, [[{\blacksquare}{\forall}n({[]^{-1}}{[]^{-1}}({\it CN}{\it n}\backslash {\it CN}{\it n})/{\blacksquare}(({\langle\rangle}Nt(n){\sqcap}!{\blacksquare}Nt(n))\backslash Sf)), {\square}({\langle\rangle}{\exists}gNt(s(g))\backslash Sf)]]]\ \Rightarrow\ \fbox{${\langle\rangle}{\exists}gNt(s(g))$}
\using {\langle\rangle}R
\endprooftree
\prooftree
\justifies
\mbox{\fbox{$Sf$}}\ \Rightarrow\ Sf
\endprooftree
\justifies
[{\blacksquare}{\forall}n(Nt(n)/{\it CN}{\it n}), {\square}{\it CN}{\it s(m)}, [[{\blacksquare}{\forall}n({[]^{-1}}{[]^{-1}}({\it CN}{\it n}\backslash {\it CN}{\it n})/{\blacksquare}(({\langle\rangle}Nt(n){\sqcap}!{\blacksquare}Nt(n))\backslash Sf)), {\square}({\langle\rangle}{\exists}gNt(s(g))\backslash Sf)]]], \mbox{\fbox{${\langle\rangle}{\exists}gNt(s(g))\backslash Sf$}}\ \Rightarrow\ Sf
\using {\backslash}L
\endprooftree
\justifies
[{\blacksquare}{\forall}n(Nt(n)/{\it CN}{\it n}), {\square}{\it CN}{\it s(m)}, [[{\blacksquare}{\forall}n({[]^{-1}}{[]^{-1}}({\it CN}{\it n}\backslash {\it CN}{\it n})/{\blacksquare}(({\langle\rangle}Nt(n){\sqcap}!{\blacksquare}Nt(n))\backslash Sf)), {\square}({\langle\rangle}{\exists}gNt(s(g))\backslash Sf)]]], \mbox{\fbox{${\square}({\langle\rangle}{\exists}gNt(s(g))\backslash Sf)$}}\ \Rightarrow\ Sf
\using {\Box}L
\endprooftree
\end{center}
\caption{Derivation for \lingform{The man that walks talks}},
\label{tmtwt}
\end{figure}
The lowest four inferences prepare the subject of the intransitive matrix verb
and the next three prepare the relative clause modification itself, argument to the subject definite
article. 
After analysis of the outer sentence at the root of the derivation tree, 
the left subtree starting at depth seven is as for the analysis of example~(\ref{ex11}).
All this delivers semantics:
\disp{
$({\it Pres}\ (\mbox{\v{}}{\it talk}\ (\iota \ \lambda D[(\mbox{\v{}}{\it man}\ {\it D})\wedge ({\it Pres}\ (\mbox{\v{}}{\it walk}\ {\it D}))])))$}

}

\begin{figure}
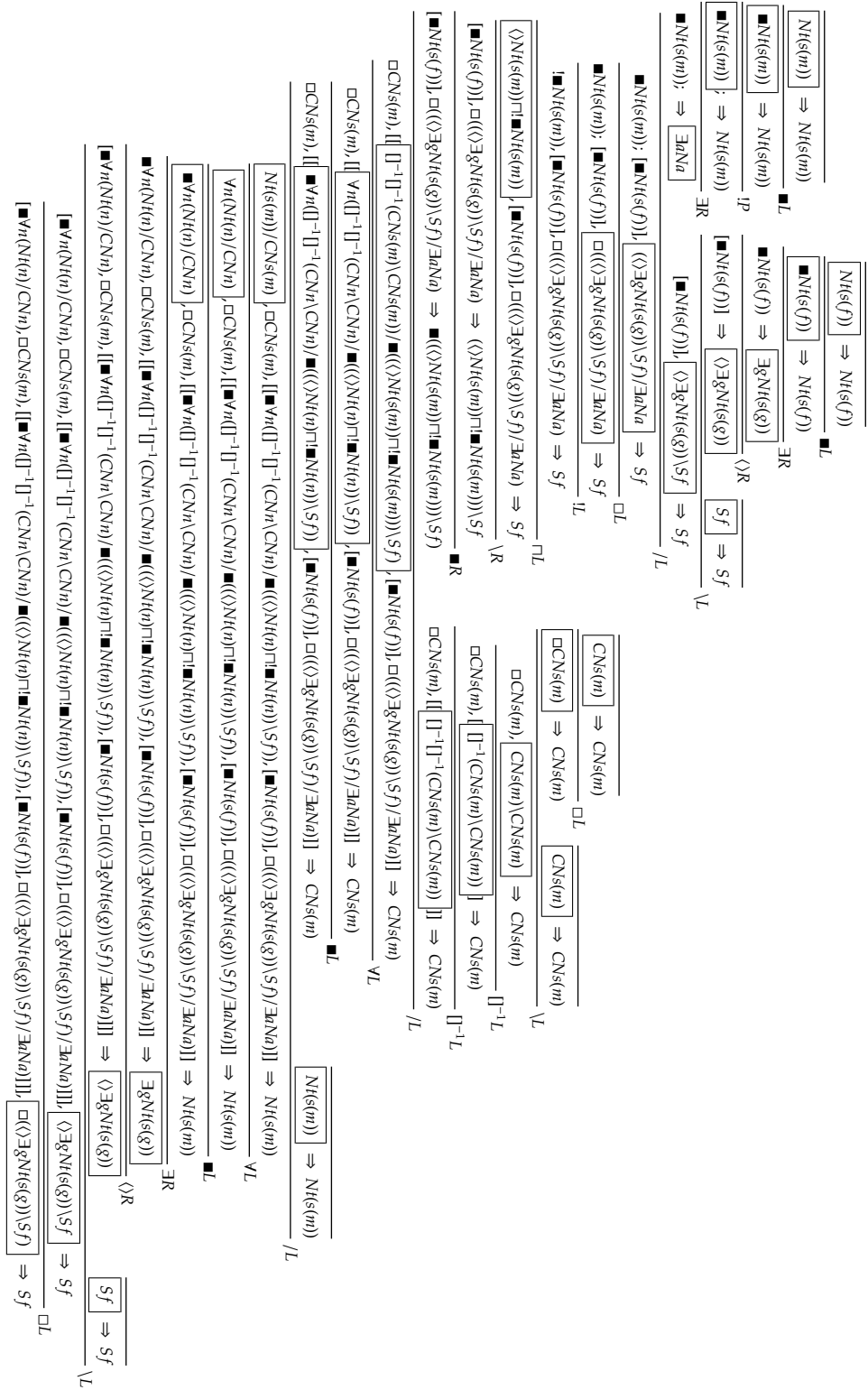

\rotatebox{-90}{
\scriptsize
\prooftree
\prooftree
\prooftree
\prooftree
\prooftree
\prooftree
\prooftree
\prooftree
\prooftree
\prooftree
\prooftree
\prooftree
\prooftree
\prooftree
\prooftree
\prooftree
\prooftree
\prooftree
\prooftree
\prooftree
\justifies
\mbox{\fbox{$Nt(s(m))$}}\ \Rightarrow\ Nt(s(m))
\endprooftree
\justifies
\mbox{\fbox{${\blacksquare}Nt(s(m))$}}\ \Rightarrow\ Nt(s(m))
\using {\blacksquare}L
\endprooftree
\justifies
\mbox{\fbox{${\blacksquare}Nt(s(m))$}};\ \ \Rightarrow\ Nt(s(m))
\using {!}P
\endprooftree
\justifies
{\blacksquare}Nt(s(m));\ \ \Rightarrow\ \fbox{${\exists}aNa$}
\using {\exists}R
\endprooftree
\prooftree
\prooftree
\prooftree
\prooftree
\prooftree
\justifies
\mbox{\fbox{$Nt(s(f))$}}\ \Rightarrow\ Nt(s(f))
\endprooftree
\justifies
\mbox{\fbox{${\blacksquare}Nt(s(f))$}}\ \Rightarrow\ Nt(s(f))
\using {\blacksquare}L
\endprooftree
\justifies
{\blacksquare}Nt(s(f))\ \Rightarrow\ \fbox{${\exists}gNt(s(g))$}
\using {\exists}R
\endprooftree
\justifies
[{\blacksquare}Nt(s(f))]\ \Rightarrow\ \fbox{${\langle\rangle}{\exists}gNt(s(g))$}
\using {\langle\rangle}R
\endprooftree
\prooftree
\justifies
\mbox{\fbox{$Sf$}}\ \Rightarrow\ Sf
\endprooftree
\justifies
[{\blacksquare}Nt(s(f))], \mbox{\fbox{${\langle\rangle}{\exists}gNt(s(g))\backslash Sf$}}\ \Rightarrow\ Sf
\using {\backslash}L
\endprooftree
\justifies
{\blacksquare}Nt(s(m));\ [{\blacksquare}Nt(s(f))], \mbox{\fbox{$({\langle\rangle}{\exists}gNt(s(g))\backslash Sf)/{\exists}aNa$}}\ \Rightarrow\ Sf
\using {/}L
\endprooftree
\justifies
{\blacksquare}Nt(s(m));\ [{\blacksquare}Nt(s(f))], \mbox{\fbox{${\square}(({\langle\rangle}{\exists}gNt(s(g))\backslash Sf)/{\exists}aNa)$}}\ \Rightarrow\ Sf
\using {\Box}L
\endprooftree
\justifies
!{\blacksquare}Nt(s(m)), [{\blacksquare}Nt(s(f))], {\square}(({\langle\rangle}{\exists}gNt(s(g))\backslash Sf)/{\exists}aNa)\ \Rightarrow\ Sf
\using {!}L
\endprooftree
\justifies
\mbox{\fbox{${\langle\rangle}Nt(s(m)){\sqcap}!{\blacksquare}Nt(s(m))$}}, [{\blacksquare}Nt(s(f))], {\square}(({\langle\rangle}{\exists}gNt(s(g))\backslash Sf)/{\exists}aNa)\ \Rightarrow\ Sf
\using {\sqcap}L
\endprooftree
\justifies
[{\blacksquare}Nt(s(f))], {\square}(({\langle\rangle}{\exists}gNt(s(g))\backslash Sf)/{\exists}aNa)\ \Rightarrow\ ({\langle\rangle}Nt(s(m)){\sqcap}!{\blacksquare}Nt(s(m)))\backslash Sf
\using {\backslash}R
\endprooftree
\justifies
[{\blacksquare}Nt(s(f))], {\square}(({\langle\rangle}{\exists}gNt(s(g))\backslash Sf)/{\exists}aNa)\ \Rightarrow\ {\blacksquare}(({\langle\rangle}Nt(s(m)){\sqcap}!{\blacksquare}Nt(s(m)))\backslash Sf)
\using {\blacksquare}R
\endprooftree
\prooftree
\prooftree
\prooftree
\prooftree
\prooftree
\justifies
\mbox{\fbox{${\it CN}{\it s(m)}$}}\ \Rightarrow\ {\it CN}{\it s(m)}
\endprooftree
\justifies
\mbox{\fbox{${\square}{\it CN}{\it s(m)}$}}\ \Rightarrow\ {\it CN}{\it s(m)}
\using {\Box}L
\endprooftree
\prooftree
\justifies
\mbox{\fbox{${\it CN}{\it s(m)}$}}\ \Rightarrow\ {\it CN}{\it s(m)}
\endprooftree
\justifies
{\square}{\it CN}{\it s(m)}, \mbox{\fbox{${\it CN}{\it s(m)}\backslash {\it CN}{\it s(m)}$}}\ \Rightarrow\ {\it CN}{\it s(m)}
\using {\backslash}L
\endprooftree
\justifies
{\square}{\it CN}{\it s(m)}, [\mbox{\fbox{${[]^{-1}}({\it CN}{\it s(m)}\backslash {\it CN}{\it s(m)})$}}]\ \Rightarrow\ {\it CN}{\it s(m)}
\using {[]^{-1}}L
\endprooftree
\justifies
{\square}{\it CN}{\it s(m)}, [[\mbox{\fbox{${[]^{-1}}{[]^{-1}}({\it CN}{\it s(m)}\backslash {\it CN}{\it s(m)})$}}]]\ \Rightarrow\ {\it CN}{\it s(m)}
\using {[]^{-1}}L
\endprooftree
\justifies
{\square}{\it CN}{\it s(m)}, [[\mbox{\fbox{${[]^{-1}}{[]^{-1}}({\it CN}{\it s(m)}\backslash {\it CN}{\it s(m)})/{\blacksquare}(({\langle\rangle}Nt(s(m)){\sqcap}!{\blacksquare}Nt(s(m)))\backslash Sf)$}}, [{\blacksquare}Nt(s(f))], {\square}(({\langle\rangle}{\exists}gNt(s(g))\backslash Sf)/{\exists}aNa)]]\ \Rightarrow\ {\it CN}{\it s(m)}
\using {/}L
\endprooftree
\justifies
{\square}{\it CN}{\it s(m)}, [[\mbox{\fbox{${\forall}n({[]^{-1}}{[]^{-1}}({\it CN}{\it n}\backslash {\it CN}{\it n})/{\blacksquare}(({\langle\rangle}Nt(n){\sqcap}!{\blacksquare}Nt(n))\backslash Sf))$}}, [{\blacksquare}Nt(s(f))], {\square}(({\langle\rangle}{\exists}gNt(s(g))\backslash Sf)/{\exists}aNa)]]\ \Rightarrow\ {\it CN}{\it s(m)}
\using {\forall}L
\endprooftree
\justifies
{\square}{\it CN}{\it s(m)}, [[\mbox{\fbox{${\blacksquare}{\forall}n({[]^{-1}}{[]^{-1}}({\it CN}{\it n}\backslash {\it CN}{\it n})/{\blacksquare}(({\langle\rangle}Nt(n){\sqcap}!{\blacksquare}Nt(n))\backslash Sf))$}}, [{\blacksquare}Nt(s(f))], {\square}(({\langle\rangle}{\exists}gNt(s(g))\backslash Sf)/{\exists}aNa)]]\ \Rightarrow\ {\it CN}{\it s(m)}
\using {\blacksquare}L
\endprooftree
\prooftree
\justifies
\mbox{\fbox{$Nt(s(m))$}}\ \Rightarrow\ Nt(s(m))
\endprooftree
\justifies
\mbox{\fbox{$Nt(s(m))/{\it CN}{\it s(m)}$}}, {\square}{\it CN}{\it s(m)}, [[{\blacksquare}{\forall}n({[]^{-1}}{[]^{-1}}({\it CN}{\it n}\backslash {\it CN}{\it n})/{\blacksquare}(({\langle\rangle}Nt(n){\sqcap}!{\blacksquare}Nt(n))\backslash Sf)), [{\blacksquare}Nt(s(f))], {\square}(({\langle\rangle}{\exists}gNt(s(g))\backslash Sf)/{\exists}aNa)]]\ \Rightarrow\ Nt(s(m))
\using {/}L
\endprooftree
\justifies
\mbox{\fbox{${\forall}n(Nt(n)/{\it CN}{\it n})$}}, {\square}{\it CN}{\it s(m)}, [[{\blacksquare}{\forall}n({[]^{-1}}{[]^{-1}}({\it CN}{\it n}\backslash {\it CN}{\it n})/{\blacksquare}(({\langle\rangle}Nt(n){\sqcap}!{\blacksquare}Nt(n))\backslash Sf)), [{\blacksquare}Nt(s(f))], {\square}(({\langle\rangle}{\exists}gNt(s(g))\backslash Sf)/{\exists}aNa)]]\ \Rightarrow\ Nt(s(m))
\using {\forall}L
\endprooftree
\justifies
\mbox{\fbox{${\blacksquare}{\forall}n(Nt(n)/{\it CN}{\it n})$}}, {\square}{\it CN}{\it s(m)}, [[{\blacksquare}{\forall}n({[]^{-1}}{[]^{-1}}({\it CN}{\it n}\backslash {\it CN}{\it n})/{\blacksquare}(({\langle\rangle}Nt(n){\sqcap}!{\blacksquare}Nt(n))\backslash Sf)), [{\blacksquare}Nt(s(f))], {\square}(({\langle\rangle}{\exists}gNt(s(g))\backslash Sf)/{\exists}aNa)]]\ \Rightarrow\ Nt(s(m))
\using {\blacksquare}L
\endprooftree
\justifies
{\blacksquare}{\forall}n(Nt(n)/{\it CN}{\it n}), {\square}{\it CN}{\it s(m)}, [[{\blacksquare}{\forall}n({[]^{-1}}{[]^{-1}}({\it CN}{\it n}\backslash {\it CN}{\it n})/{\blacksquare}(({\langle\rangle}Nt(n){\sqcap}!{\blacksquare}Nt(n))\backslash Sf)), [{\blacksquare}Nt(s(f))], {\square}(({\langle\rangle}{\exists}gNt(s(g))\backslash Sf)/{\exists}aNa)]]\ \Rightarrow\ \fbox{${\exists}gNt(s(g))$}
\using {\exists}R
\endprooftree
\justifies
[{\blacksquare}{\forall}n(Nt(n)/{\it CN}{\it n}), {\square}{\it CN}{\it s(m)}, [[{\blacksquare}{\forall}n({[]^{-1}}{[]^{-1}}({\it CN}{\it n}\backslash {\it CN}{\it n})/{\blacksquare}(({\langle\rangle}Nt(n){\sqcap}!{\blacksquare}Nt(n))\backslash Sf)), [{\blacksquare}Nt(s(f))], {\square}(({\langle\rangle}{\exists}gNt(s(g))\backslash Sf)/{\exists}aNa)]]]\ \Rightarrow\ \fbox{${\langle\rangle}{\exists}gNt(s(g))$}
\using {\langle\rangle}R
\endprooftree
\prooftree
\justifies
\mbox{\fbox{$Sf$}}\ \Rightarrow\ Sf
\endprooftree
\justifies
[{\blacksquare}{\forall}n(Nt(n)/{\it CN}{\it n}), {\square}{\it CN}{\it s(m)}, [[{\blacksquare}{\forall}n({[]^{-1}}{[]^{-1}}({\it CN}{\it n}\backslash {\it CN}{\it n})/{\blacksquare}(({\langle\rangle}Nt(n){\sqcap}!{\blacksquare}Nt(n))\backslash Sf)), [{\blacksquare}Nt(s(f))], {\square}(({\langle\rangle}{\exists}gNt(s(g))\backslash Sf)/{\exists}aNa)]]], \mbox{\fbox{${\langle\rangle}{\exists}gNt(s(g))\backslash Sf$}}\ \Rightarrow\ Sf
\using {\backslash}L
\endprooftree
\justifies
[{\blacksquare}{\forall}n(Nt(n)/{\it CN}{\it n}), {\square}{\it CN}{\it s(m)}, [[{\blacksquare}{\forall}n({[]^{-1}}{[]^{-1}}({\it CN}{\it n}\backslash {\it CN}{\it n})/{\blacksquare}(({\langle\rangle}Nt(n){\sqcap}!{\blacksquare}Nt(n))\backslash Sf)), [{\blacksquare}Nt(s(f))], {\square}(({\langle\rangle}{\exists}gNt(s(g))\backslash Sf)/{\exists}aNa)]]], \mbox{\fbox{${\square}({\langle\rangle}{\exists}gNt(s(g))\backslash Sf)$}}\ \Rightarrow\ Sf
\using {\Box}L
\endprooftree
}
\caption{Derivation for \lingform{The man that Mary loves walks}},
\label{tmtmlw}
\end{figure}

The next sentence contains a minimal example of object relativisation:
\disp{
 $[{\bf the}{+}{\bf man}{+}[[{\bf that}{+}[{\bf mary}]{+}{\bf loves}]]]{+}{\bf walks}: Sf$}
Lexical lookup yields:
\disp{
$[{\blacksquare}{\forall}n(Nt(n)/{\it CN}{\it n}): \iota , {\square}{\it CN}{\it s(m)}: {\it man}, [[{\blacksquare}{\forall}n({[]^{-1}}{[]^{-1}}({\it CN}{\it n}\backslash {\it CN}{\it n})/\\{\blacksquare}(({\langle\rangle}Nt(n){\sqcap}!{\blacksquare}Nt(n))\backslash Sf)): \lambda A\lambda B\lambda C[({\it B}\ {\it C})\wedge ({\it A}\ {\it C})], [{\blacksquare}Nt(s(f)): {\it m}],\\ {\square}(({\langle\rangle}{\exists}gNt(s(g))\backslash Sf)/{\exists}aNa):
 \mbox{\^{}}\lambda D\lambda E({\it Pres}\ ((\mbox{\v{}}{\it love}\ {\it D})\ {\it E}))]]], {\square}({\langle\rangle}{\exists}gNt(s(g))\backslash Sf):\\ \mbox{\^{}}\lambda F({\it Pres}\ (\mbox{\v{}}{\it walk}\ {\it F}))\ \Rightarrow\ Sf$}
There is the derivation given in Figure~\ref{tmtmlw}.
The lowest four inferences prepare the subject of the intransitive matrix verb
and the next three prepare the relative clause modification itself, argument to the subject definite
article. The analysis of the complex common noun phrase starts in the minor premise
of the lowest $/L$ with (semantically inactive) modality left, and for all left
instantiating agreement to masculine singular.
At the middle $/L$, the righthand subtree cancels the double brackets with the relative pronoun
value antibrackets and the lefthand subtree selects the body of the relative clause
as the semantically inactive modalised higher-order subject-and-object polymorphic relative pronoun argument
type. After (semantically inactive) modality right, licensed since the antecedent
types are modalised, the conjoined hypothetical subject is lowered by $\bsl R$
into the antecedent.
Observe how $\iaconj L$ selects the object relativisation hypothetical
subtype $\univexp\imod Nt(s(m))$ and how this subsequently percolates in the stoup,
passing in particular into the minor premise branch of the upper $/L$ inference
and hence satisfying the object valency of the transitive verb \lingform{love};
subject and intransitive verb phrase are analysed as usual.
This delivers the required semantics:
\disp{
$({\it Pres}\ (\mbox{\v{}}{\it walk}\ (\iota \ \lambda D[(\mbox{\v{}}{\it man}\ {\it D})\wedge ({\it Pres}\ ((\mbox{\v{}}{\it love}\ {\it D})\ {\it m}))])))$}

An example with longer-distance object relativisation,
in the context of an entire sentence, is:
\disp{
$[{\bf the}{+}{\bf man}{+}[[{\bf that}{+}[{\bf john}]{+}{\bf thinks}{+}[{\bf mary}]{+}{\bf loves}]]]{+}{\bf walks}: Sf$}
Lexical lookup yields the following; note how the propositional attitude verb
is polymorphic between a complementised and an uncomplementised sentential
argument,
expressed with a semantically inactive additive disjunction:
\disp{
$[{\blacksquare}{\forall}n(Nt(n)/{\it CN}{\it n}): \iota , {\square}{\it CN}{\it s(m)}: {\it man}, [[{\blacksquare}{\forall}n({[]^{-1}}{[]^{-1}}({\it CN}{\it n}\backslash {\it CN}{\it n})/\\{\blacksquare}(({\langle\rangle}Nt(n){\sqcap}!{\blacksquare}Nt(n))\backslash Sf)): \lambda A\lambda B\lambda C[({\it B}\ {\it C})\wedge ({\it A}\ {\it C})], [{\blacksquare}Nt(s(m)): {\it j}],\\ {\square}(({\langle\rangle}{\exists}gNt(s(g))\backslash Sf)/({\it CP}that{\sqcup}{\square}Sf)): \mbox{\^{}}\lambda D\lambda E({\it Pres}\ ((\mbox{\v{}}{\it think}\ {\it D})\ {\it E})),\\{} [{\blacksquare}Nt(s(f)): {\it m}], {\square}(({\langle\rangle}{\exists}gNt(s(g))\backslash Sf)/{\exists}aNa): \mbox{\^{}}\lambda F\lambda G({\it Pres}\ ((\mbox{\v{}}{\it love}\ {\it F})\ {\it G}))]]],\\ {\square}({\langle\rangle}{\exists}gNt(s(g))\backslash Sf): \mbox{\^{}}\lambda H({\it Pres}\ (\mbox{\v{}}{\it walk}\ {\it H}))\ \Rightarrow\ Sf$}
There is the derivation given in Figure~\ref{tmtjtmlw}.
Inference up as far as $\textcircled{1}$ brings us to analysis of the complex
common noun phrase in the lefthand subtree.
The following preparation of the relative pronoun and double bracket cancellation
of its value are as usual. 
After modality right and under right on the relative pronoun higher-order
argument, $\iaconj L$ selects the object relativisation hypothetical subtype
and $\univexp L$ moves this into the stoup. 
In the stoup it percolates to the subordinate clause, 
(observe how $\iadisj R$ selects the uncomplementised sentential argument type
of the propositional attitude verb)
and there $\univexp P$ moves it into position to satisfy the embedded clause object valency.
\begin{figure}
\begin{center}
\rotatebox{-90}{\tiny
\prooftree
\prooftree
\prooftree
\prooftree
\prooftree
\prooftree
\prooftree
\prooftree
\prooftree
\prooftree
\prooftree
\prooftree
\prooftree
\prooftree
\prooftree
\prooftree
\prooftree
\prooftree
\prooftree
\prooftree
\vdots
\justifies
[{\blacksquare}Nt(s(f))], \mbox{\fbox{${\square}(({\langle\rangle}{\exists}gNt(s(g))\backslash Sf)/{\exists}aNa)$}}, {\blacksquare}Nt(s(m))\ \Rightarrow\ Sf
\using {\Box}L
\endprooftree
\justifies
[{\blacksquare}Nt(s(f))], {\square}(({\langle\rangle}{\exists}gNt(s(g))\backslash Sf)/{\exists}aNa), {\blacksquare}Nt(s(m))\ \Rightarrow\ {\square}Sf
\using {\Box}R
\endprooftree
\justifies
\mbox{\fbox{${\blacksquare}Nt(s(m))$}};\ [{\blacksquare}Nt(s(f))], {\square}(({\langle\rangle}{\exists}gNt(s(g))\backslash Sf)/{\exists}aNa)\ \Rightarrow\ {\square}Sf
\using {!}P
\endprooftree
\justifies
{\blacksquare}Nt(s(m));\ [{\blacksquare}Nt(s(f))], {\square}(({\langle\rangle}{\exists}gNt(s(g))\backslash Sf)/{\exists}aNa)\ \Rightarrow\ \fbox{${\it CP}that{\sqcup}{\square}Sf$}
\using {\sqcup}R
\endprooftree
\prooftree
\prooftree
\prooftree
\prooftree
\prooftree
\justifies
\mbox{\fbox{$Nt(s(m))$}}\ \Rightarrow\ Nt(s(m))
\endprooftree
\justifies
\mbox{\fbox{${\blacksquare}Nt(s(m))$}}\ \Rightarrow\ Nt(s(m))
\using {\blacksquare}L
\endprooftree
\justifies
{\blacksquare}Nt(s(m))\ \Rightarrow\ \fbox{${\exists}gNt(s(g))$}
\using {\exists}R
\endprooftree
\justifies
[{\blacksquare}Nt(s(m))]\ \Rightarrow\ \fbox{${\langle\rangle}{\exists}gNt(s(g))$}
\using {\langle\rangle}R
\endprooftree
\prooftree
\justifies
\mbox{\fbox{$Sf$}}\ \Rightarrow\ Sf
\endprooftree
\justifies
[{\blacksquare}Nt(s(m))], \mbox{\fbox{${\langle\rangle}{\exists}gNt(s(g))\backslash Sf$}}\ \Rightarrow\ Sf
\using {\backslash}L
\endprooftree
\justifies
{\blacksquare}Nt(s(m));\ [{\blacksquare}Nt(s(m))], \mbox{\fbox{$({\langle\rangle}{\exists}gNt(s(g))\backslash Sf)/({\it CP}that{\sqcup}{\square}Sf)$}}, [{\blacksquare}Nt(s(f))], {\square}(({\langle\rangle}{\exists}gNt(s(g))\backslash Sf)/{\exists}aNa)\ \Rightarrow\ Sf
\using {/}L
\endprooftree
\justifies
{\blacksquare}Nt(s(m));\ [{\blacksquare}Nt(s(m))], \mbox{\fbox{${\square}(({\langle\rangle}{\exists}gNt(s(g))\backslash Sf)/({\it CP}that{\sqcup}{\square}Sf))$}}, [{\blacksquare}Nt(s(f))], {\square}(({\langle\rangle}{\exists}gNt(s(g))\backslash Sf)/{\exists}aNa)\ \Rightarrow\ Sf
\using {\Box}L
\endprooftree
\justifies
!{\blacksquare}Nt(s(m)), [{\blacksquare}Nt(s(m))], {\square}(({\langle\rangle}{\exists}gNt(s(g))\backslash Sf)/({\it CP}that{\sqcup}{\square}Sf)), [{\blacksquare}Nt(s(f))], {\square}(({\langle\rangle}{\exists}gNt(s(g))\backslash Sf)/{\exists}aNa)\ \Rightarrow\ Sf
\using {!}L
\endprooftree
\justifies
\mbox{\fbox{${\langle\rangle}Nt(s(m)){\sqcap}!{\blacksquare}Nt(s(m))$}}, [{\blacksquare}Nt(s(m))], {\square}(({\langle\rangle}{\exists}gNt(s(g))\backslash Sf)/({\it CP}that{\sqcup}{\square}Sf)), [{\blacksquare}Nt(s(f))], {\square}(({\langle\rangle}{\exists}gNt(s(g))\backslash Sf)/{\exists}aNa)\ \Rightarrow\ Sf
\using {\sqcap}L
\endprooftree
\justifies
[{\blacksquare}Nt(s(m))], {\square}(({\langle\rangle}{\exists}gNt(s(g))\backslash Sf)/({\it CP}that{\sqcup}{\square}Sf)), [{\blacksquare}Nt(s(f))], {\square}(({\langle\rangle}{\exists}gNt(s(g))\backslash Sf)/{\exists}aNa)\ \Rightarrow\ ({\langle\rangle}Nt(s(m)){\sqcap}!{\blacksquare}Nt(s(m)))\backslash Sf
\using {\backslash}R
\endprooftree
\justifies
[{\blacksquare}Nt(s(m))], {\square}(({\langle\rangle}{\exists}gNt(s(g))\backslash Sf)/({\it CP}that{\sqcup}{\square}Sf)), [{\blacksquare}Nt(s(f))], {\square}(({\langle\rangle}{\exists}gNt(s(g))\backslash Sf)/{\exists}aNa)\ \Rightarrow\ {\blacksquare}(({\langle\rangle}Nt(s(m)){\sqcap}!{\blacksquare}Nt(s(m)))\backslash Sf)
\using {\blacksquare}R
\endprooftree
\prooftree
\prooftree
\prooftree
\prooftree
\prooftree
\justifies
\mbox{\fbox{${\it CN}{\it s(m)}$}}\ \Rightarrow\ {\it CN}{\it s(m)}
\endprooftree
\justifies
\mbox{\fbox{${\square}{\it CN}{\it s(m)}$}}\ \Rightarrow\ {\it CN}{\it s(m)}
\using {\Box}L
\endprooftree
\prooftree
\justifies
\mbox{\fbox{${\it CN}{\it s(m)}$}}\ \Rightarrow\ {\it CN}{\it s(m)}
\endprooftree
\justifies
{\square}{\it CN}{\it s(m)}, \mbox{\fbox{${\it CN}{\it s(m)}\backslash {\it CN}{\it s(m)}$}}\ \Rightarrow\ {\it CN}{\it s(m)}
\using {\backslash}L
\endprooftree
\justifies
{\square}{\it CN}{\it s(m)}, [\mbox{\fbox{${[]^{-1}}({\it CN}{\it s(m)}\backslash {\it CN}{\it s(m)})$}}]\ \Rightarrow\ {\it CN}{\it s(m)}
\using {[]^{-1}}L
\endprooftree
\justifies
{\square}{\it CN}{\it s(m)}, [[\mbox{\fbox{${[]^{-1}}{[]^{-1}}({\it CN}{\it s(m)}\backslash {\it CN}{\it s(m)})$}}]]\ \Rightarrow\ {\it CN}{\it s(m)}
\using {[]^{-1}}L
\endprooftree
\justifies
{\square}{\it CN}{\it s(m)}, [[\mbox{\fbox{${[]^{-1}}{[]^{-1}}({\it CN}{\it s(m)}\backslash {\it CN}{\it s(m)})/{\blacksquare}(({\langle\rangle}Nt(s(m)){\sqcap}!{\blacksquare}Nt(s(m)))\backslash Sf)$}}, [{\blacksquare}Nt(s(m))], {\square}(({\langle\rangle}{\exists}gNt(s(g))\backslash Sf)/({\it CP}that{\sqcup}{\square}Sf)), [{\blacksquare}Nt(s(f))], {\square}(({\langle\rangle}{\exists}gNt(s(g))\backslash Sf)/{\exists}aNa)]]\ \Rightarrow\ {\it CN}{\it s(m)}
\using {/}L
\endprooftree
\justifies
{\square}{\it CN}{\it s(m)}, [[\mbox{\fbox{${\forall}n({[]^{-1}}{[]^{-1}}({\it CN}{\it n}\backslash {\it CN}{\it n})/{\blacksquare}(({\langle\rangle}Nt(n){\sqcap}!{\blacksquare}Nt(n))\backslash Sf))$}}, [{\blacksquare}Nt(s(m))], {\square}(({\langle\rangle}{\exists}gNt(s(g))\backslash Sf)/({\it CP}that{\sqcup}{\square}Sf)), [{\blacksquare}Nt(s(f))], {\square}(({\langle\rangle}{\exists}gNt(s(g))\backslash Sf)/{\exists}aNa)]]\ \Rightarrow\ {\it CN}{\it s(m)}
\using {\forall}L
\endprooftree
\justifies
\begin{array}{c}
{\square}{\it CN}{\it s(m)}, [[\mbox{\fbox{${\blacksquare}{\forall}n({[]^{-1}}{[]^{-1}}({\it CN}{\it n}\backslash {\it CN}{\it n})/{\blacksquare}(({\langle\rangle}Nt(n){\sqcap}!{\blacksquare}Nt(n))\backslash Sf))$}}, [{\blacksquare}Nt(s(m))], {\square}(({\langle\rangle}{\exists}gNt(s(g))\backslash Sf)/({\it CP}that{\sqcup}{\square}Sf)), [{\blacksquare}Nt(s(f))], {\square}(({\langle\rangle}{\exists}gNt(s(g))\backslash Sf)/{\exists}aNa)]]\ \Rightarrow\ {\it CN}{\it s(m)}\\
\textcircled{1}
\end{array}
\using {\blacksquare}L
\endprooftree
\prooftree
\justifies
\mbox{\fbox{$Nt(s(m))$}}\ \Rightarrow\ Nt(s(m))
\endprooftree
\justifies
\mbox{\fbox{$Nt(s(m))/{\it CN}{\it s(m)}$}}, {\square}{\it CN}{\it s(m)}, [[{\blacksquare}{\forall}n({[]^{-1}}{[]^{-1}}({\it CN}{\it n}\backslash {\it CN}{\it n})/{\blacksquare}(({\langle\rangle}Nt(n){\sqcap}!{\blacksquare}Nt(n))\backslash Sf)), [{\blacksquare}Nt(s(m))], {\square}(({\langle\rangle}{\exists}gNt(s(g))\backslash Sf)/({\it CP}that{\sqcup}{\square}Sf)), [{\blacksquare}Nt(s(f))], {\square}(({\langle\rangle}{\exists}gNt(s(g))\backslash Sf)/{\exists}aNa)]]\ \Rightarrow\ Nt(s(m))
\using {/}L
\endprooftree
\justifies
\mbox{\fbox{${\forall}n(Nt(n)/{\it CN}{\it n})$}}, {\square}{\it CN}{\it s(m)}, [[{\blacksquare}{\forall}n({[]^{-1}}{[]^{-1}}({\it CN}{\it n}\backslash {\it CN}{\it n})/{\blacksquare}(({\langle\rangle}Nt(n){\sqcap}!{\blacksquare}Nt(n))\backslash Sf)), [{\blacksquare}Nt(s(m))], {\square}(({\langle\rangle}{\exists}gNt(s(g))\backslash Sf)/({\it CP}that{\sqcup}{\square}Sf)), [{\blacksquare}Nt(s(f))], {\square}(({\langle\rangle}{\exists}gNt(s(g))\backslash Sf)/{\exists}aNa)]]\ \Rightarrow\ Nt(s(m))
\using {\forall}L
\endprooftree
\justifies
\mbox{\fbox{${\blacksquare}{\forall}n(Nt(n)/{\it CN}{\it n})$}}, {\square}{\it CN}{\it s(m)}, [[{\blacksquare}{\forall}n({[]^{-1}}{[]^{-1}}({\it CN}{\it n}\backslash {\it CN}{\it n})/{\blacksquare}(({\langle\rangle}Nt(n){\sqcap}!{\blacksquare}Nt(n))\backslash Sf)), [{\blacksquare}Nt(s(m))], {\square}(({\langle\rangle}{\exists}gNt(s(g))\backslash Sf)/({\it CP}that{\sqcup}{\square}Sf)), [{\blacksquare}Nt(s(f))], {\square}(({\langle\rangle}{\exists}gNt(s(g))\backslash Sf)/{\exists}aNa)]]\ \Rightarrow\ Nt(s(m))
\using {\blacksquare}L
\endprooftree
\justifies
{\blacksquare}{\forall}n(Nt(n)/{\it CN}{\it n}), {\square}{\it CN}{\it s(m)}, [[{\blacksquare}{\forall}n({[]^{-1}}{[]^{-1}}({\it CN}{\it n}\backslash {\it CN}{\it n})/{\blacksquare}(({\langle\rangle}Nt(n){\sqcap}!{\blacksquare}Nt(n))\backslash Sf)), [{\blacksquare}Nt(s(m))], {\square}(({\langle\rangle}{\exists}gNt(s(g))\backslash Sf)/({\it CP}that{\sqcup}{\square}Sf)), [{\blacksquare}Nt(s(f))], {\square}(({\langle\rangle}{\exists}gNt(s(g))\backslash Sf)/{\exists}aNa)]]\ \Rightarrow\ \fbox{${\exists}gNt(s(g))$}
\using {\exists}R
\endprooftree
\justifies
[{\blacksquare}{\forall}n(Nt(n)/{\it CN}{\it n}), {\square}{\it CN}{\it s(m)}, [[{\blacksquare}{\forall}n({[]^{-1}}{[]^{-1}}({\it CN}{\it n}\backslash {\it CN}{\it n})/{\blacksquare}(({\langle\rangle}Nt(n){\sqcap}!{\blacksquare}Nt(n))\backslash Sf)), [{\blacksquare}Nt(s(m))], {\square}(({\langle\rangle}{\exists}gNt(s(g))\backslash Sf)/({\it CP}that{\sqcup}{\square}Sf)), [{\blacksquare}Nt(s(f))], {\square}(({\langle\rangle}{\exists}gNt(s(g))\backslash Sf)/{\exists}aNa)]]]\ \Rightarrow\ \fbox{${\langle\rangle}{\exists}gNt(s(g))$}
\using {\langle\rangle}R
\endprooftree
\prooftree
\justifies
\mbox{\fbox{$Sf$}}\ \Rightarrow\ Sf
\endprooftree
\justifies
[{\blacksquare}{\forall}n(Nt(n)/{\it CN}{\it n}), {\square}{\it CN}{\it s(m)}, [[{\blacksquare}{\forall}n({[]^{-1}}{[]^{-1}}({\it CN}{\it n}\backslash {\it CN}{\it n})/{\blacksquare}(({\langle\rangle}Nt(n){\sqcap}!{\blacksquare}Nt(n))\backslash Sf)), [{\blacksquare}Nt(s(m))], {\square}(({\langle\rangle}{\exists}gNt(s(g))\backslash Sf)/({\it CP}that{\sqcup}{\square}Sf)), [{\blacksquare}Nt(s(f))], {\square}(({\langle\rangle}{\exists}gNt(s(g))\backslash Sf)/{\exists}aNa)]]], \mbox{\fbox{${\langle\rangle}{\exists}gNt(s(g))\backslash Sf$}}\ \Rightarrow\ Sf
\using {\backslash}L
\endprooftree
\justifies
[{\blacksquare}{\forall}n(Nt(n)/{\it CN}{\it n}), {\square}{\it CN}{\it s(m)}, [[{\blacksquare}{\forall}n({[]^{-1}}{[]^{-1}}({\it CN}{\it n}\backslash {\it CN}{\it n})/{\blacksquare}(({\langle\rangle}Nt(n){\sqcap}!{\blacksquare}Nt(n))\backslash Sf)), [{\blacksquare}Nt(s(m))], {\square}(({\langle\rangle}{\exists}gNt(s(g))\backslash Sf)/({\it CP}that{\sqcup}{\square}Sf)), [{\blacksquare}Nt(s(f))], {\square}(({\langle\rangle}{\exists}gNt(s(g))\backslash Sf)/{\exists}aNa)]]], \mbox{\fbox{${\square}({\langle\rangle}{\exists}gNt(s(g))\backslash Sf)$}}\ \Rightarrow\ Sf
\using {\Box}L
\endprooftree}
\end{center}
\caption{Derivation for \lingform{The man that John thinks Mary loves walks}}
\label{tmtjtmlw}
\end{figure}
This delivers the correct semantics:
\disp{
$({\it Pres}\ (\mbox{\v{}}{\it walk}\ (\iota \ \lambda D[(\mbox{\v{}}{\it man}\ {\it D})\wedge ({\it Pres}\ ((\mbox{\v{}}{\it think}\ \mbox{\^{}}({\it Pres}\ ((\mbox{\v{}}{\it love}\ {\it D})\ {\it m})))\ {\it j}))])))$}

\begin{figure}
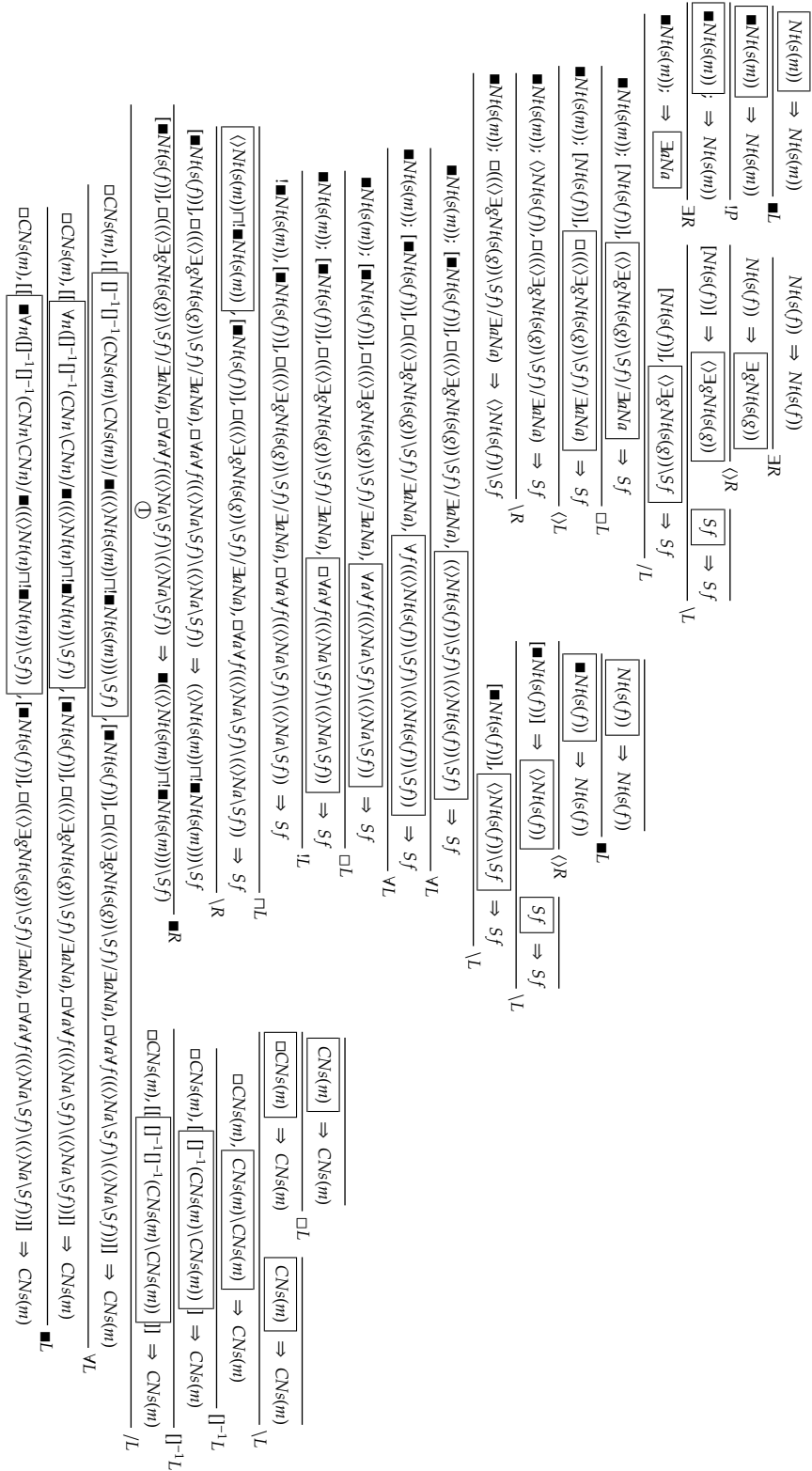

\begin{center}
\rotatebox{-90}{
$
{\scriptsize
\prooftree
\prooftree
\prooftree
\prooftree
\prooftree
\prooftree
\prooftree
\prooftree
\prooftree
\prooftree
\prooftree
\prooftree
\prooftree
\prooftree
\prooftree
\prooftree
\prooftree
\prooftree
\prooftree
\justifies
\mbox{\fbox{$Nt(s(m))$}}\ \Rightarrow\ Nt(s(m))
\endprooftree
\justifies
\mbox{\fbox{${\blacksquare}Nt(s(m))$}}\ \Rightarrow\ Nt(s(m))
\using {\blacksquare}L
\endprooftree
\justifies
\mbox{\fbox{${\blacksquare}Nt(s(m))$}};\ \ \Rightarrow\ Nt(s(m))
\using {!}P
\endprooftree
\justifies
{\blacksquare}Nt(s(m));\ \ \Rightarrow\ \fbox{${\exists}aNa$}
\using {\exists}R
\endprooftree
\prooftree
\prooftree
\prooftree
\prooftree
\justifies
Nt(s(f))\ \Rightarrow\ Nt(s(f))
\endprooftree
\justifies
Nt(s(f))\ \Rightarrow\ \fbox{${\exists}gNt(s(g))$}
\using {\exists}R
\endprooftree
\justifies
[Nt(s(f))]\ \Rightarrow\ \fbox{${\langle\rangle}{\exists}gNt(s(g))$}
\using {\langle\rangle}R
\endprooftree
\prooftree
\justifies
\mbox{\fbox{$Sf$}}\ \Rightarrow\ Sf
\endprooftree
\justifies
[Nt(s(f))], \mbox{\fbox{${\langle\rangle}{\exists}gNt(s(g))\backslash Sf$}}\ \Rightarrow\ Sf
\using {\backslash}L
\endprooftree
\justifies
{\blacksquare}Nt(s(m));\ [Nt(s(f))], \mbox{\fbox{$({\langle\rangle}{\exists}gNt(s(g))\backslash Sf)/{\exists}aNa$}}\ \Rightarrow\ Sf
\using {/}L
\endprooftree
\justifies
{\blacksquare}Nt(s(m));\ [Nt(s(f))], \mbox{\fbox{${\square}(({\langle\rangle}{\exists}gNt(s(g))\backslash Sf)/{\exists}aNa)$}}\ \Rightarrow\ Sf
\using {\Box}L
\endprooftree
\justifies
{\blacksquare}Nt(s(m));\ {\langle\rangle}Nt(s(f)), {\square}(({\langle\rangle}{\exists}gNt(s(g))\backslash Sf)/{\exists}aNa)\ \Rightarrow\ Sf
\using {\langle\rangle}L
\endprooftree
\justifies
{\blacksquare}Nt(s(m));\ {\square}(({\langle\rangle}{\exists}gNt(s(g))\backslash Sf)/{\exists}aNa)\ \Rightarrow\ {\langle\rangle}Nt(s(f))\backslash Sf
\using {\backslash}R
\endprooftree
\prooftree
\prooftree
\prooftree
\prooftree
\justifies
\mbox{\fbox{$Nt(s(f))$}}\ \Rightarrow\ Nt(s(f))
\endprooftree
\justifies
\mbox{\fbox{${\blacksquare}Nt(s(f))$}}\ \Rightarrow\ Nt(s(f))
\using {\blacksquare}L
\endprooftree
\justifies
[{\blacksquare}Nt(s(f))]\ \Rightarrow\ \fbox{${\langle\rangle}Nt(s(f))$}
\using {\langle\rangle}R
\endprooftree
\prooftree
\justifies
\mbox{\fbox{$Sf$}}\ \Rightarrow\ Sf
\endprooftree
\justifies
[{\blacksquare}Nt(s(f))], \mbox{\fbox{${\langle\rangle}Nt(s(f))\backslash Sf$}}\ \Rightarrow\ Sf
\using {\backslash}L
\endprooftree
\justifies
{\blacksquare}Nt(s(m));\ [{\blacksquare}Nt(s(f))], {\square}(({\langle\rangle}{\exists}gNt(s(g))\backslash Sf)/{\exists}aNa), \mbox{\fbox{$({\langle\rangle}Nt(s(f))\backslash Sf)\backslash ({\langle\rangle}Nt(s(f))\backslash Sf)$}}\ \Rightarrow\ Sf
\using {\backslash}L
\endprooftree
\justifies
{\blacksquare}Nt(s(m));\ [{\blacksquare}Nt(s(f))], {\square}(({\langle\rangle}{\exists}gNt(s(g))\backslash Sf)/{\exists}aNa), \mbox{\fbox{${\forall}f(({\langle\rangle}Nt(s(f))\backslash Sf)\backslash ({\langle\rangle}Nt(s(f))\backslash Sf))$}}\ \Rightarrow\ Sf
\using {\forall}L
\endprooftree
\justifies
{\blacksquare}Nt(s(m));\ [{\blacksquare}Nt(s(f))], {\square}(({\langle\rangle}{\exists}gNt(s(g))\backslash Sf)/{\exists}aNa), \mbox{\fbox{${\forall}a{\forall}f(({\langle\rangle}Na\backslash Sf)\backslash ({\langle\rangle}Na\backslash Sf))$}}\ \Rightarrow\ Sf
\using {\forall}L
\endprooftree
\justifies
{\blacksquare}Nt(s(m));\ [{\blacksquare}Nt(s(f))], {\square}(({\langle\rangle}{\exists}gNt(s(g))\backslash Sf)/{\exists}aNa), \mbox{\fbox{${\square}{\forall}a{\forall}f(({\langle\rangle}Na\backslash Sf)\backslash ({\langle\rangle}Na\backslash Sf))$}}\ \Rightarrow\ Sf
\using {\Box}L
\endprooftree
\justifies
!{\blacksquare}Nt(s(m)), [{\blacksquare}Nt(s(f))], {\square}(({\langle\rangle}{\exists}gNt(s(g))\backslash Sf)/{\exists}aNa), {\square}{\forall}a{\forall}f(({\langle\rangle}Na\backslash Sf)\backslash ({\langle\rangle}Na\backslash Sf))\ \Rightarrow\ Sf
\using {!}L
\endprooftree
\justifies
\mbox{\fbox{${\langle\rangle}Nt(s(m)){\sqcap}!{\blacksquare}Nt(s(m))$}}, [{\blacksquare}Nt(s(f))], {\square}(({\langle\rangle}{\exists}gNt(s(g))\backslash Sf)/{\exists}aNa), {\square}{\forall}a{\forall}f(({\langle\rangle}Na\backslash Sf)\backslash ({\langle\rangle}Na\backslash Sf))\ \Rightarrow\ Sf
\using {\sqcap}L
\endprooftree
\justifies
[{\blacksquare}Nt(s(f))], {\square}(({\langle\rangle}{\exists}gNt(s(g))\backslash Sf)/{\exists}aNa), {\square}{\forall}a{\forall}f(({\langle\rangle}Na\backslash Sf)\backslash ({\langle\rangle}Na\backslash Sf))\ \Rightarrow\ ({\langle\rangle}Nt(s(m)){\sqcap}!{\blacksquare}Nt(s(m)))\backslash Sf
\using {\backslash}R
\endprooftree
\justifies
\begin{array}{c}
[{\blacksquare}Nt(s(f))], {\square}(({\langle\rangle}{\exists}gNt(s(g))\backslash Sf)/{\exists}aNa), {\square}{\forall}a{\forall}f(({\langle\rangle}Na\backslash Sf)\backslash ({\langle\rangle}Na\backslash Sf))\ \Rightarrow\ {\blacksquare}(({\langle\rangle}Nt(s(m)){\sqcap}!{\blacksquare}Nt(s(m)))\backslash Sf)\\
\textcircled{1}
\end{array}
\using {\blacksquare}R
\endprooftree
\prooftree
\prooftree
\prooftree
\prooftree
\prooftree
\justifies
\mbox{\fbox{${\it CN}{\it s(m)}$}}\ \Rightarrow\ {\it CN}{\it s(m)}
\endprooftree
\justifies
\mbox{\fbox{${\square}{\it CN}{\it s(m)}$}}\ \Rightarrow\ {\it CN}{\it s(m)}
\using {\Box}L
\endprooftree
\prooftree
\justifies
\mbox{\fbox{${\it CN}{\it s(m)}$}}\ \Rightarrow\ {\it CN}{\it s(m)}
\endprooftree
\justifies
{\square}{\it CN}{\it s(m)}, \mbox{\fbox{${\it CN}{\it s(m)}\backslash {\it CN}{\it s(m)}$}}\ \Rightarrow\ {\it CN}{\it s(m)}
\using {\backslash}L
\endprooftree
\justifies
{\square}{\it CN}{\it s(m)}, [\mbox{\fbox{${[]^{-1}}({\it CN}{\it s(m)}\backslash {\it CN}{\it s(m)})$}}]\ \Rightarrow\ {\it CN}{\it s(m)}
\using {[]^{-1}}L
\endprooftree
\justifies
{\square}{\it CN}{\it s(m)}, [[\mbox{\fbox{${[]^{-1}}{[]^{-1}}({\it CN}{\it s(m)}\backslash {\it CN}{\it s(m)})$}}]]\ \Rightarrow\ {\it CN}{\it s(m)}
\using {[]^{-1}}L
\endprooftree
\justifies
{\square}{\it CN}{\it s(m)}, [[\mbox{\fbox{${[]^{-1}}{[]^{-1}}({\it CN}{\it s(m)}\backslash {\it CN}{\it s(m)})/{\blacksquare}(({\langle\rangle}Nt(s(m)){\sqcap}!{\blacksquare}Nt(s(m)))\backslash Sf)$}}, [{\blacksquare}Nt(s(f))], {\square}(({\langle\rangle}{\exists}gNt(s(g))\backslash Sf)/{\exists}aNa), {\square}{\forall}a{\forall}f(({\langle\rangle}Na\backslash Sf)\backslash ({\langle\rangle}Na\backslash Sf))]]\ \Rightarrow\ {\it CN}{\it s(m)}
\using {/}L
\endprooftree
\justifies
{\square}{\it CN}{\it s(m)}, [[\mbox{\fbox{${\forall}n({[]^{-1}}{[]^{-1}}({\it CN}{\it n}\backslash {\it CN}{\it n})/{\blacksquare}(({\langle\rangle}Nt(n){\sqcap}!{\blacksquare}Nt(n))\backslash Sf))$}}, [{\blacksquare}Nt(s(f))], {\square}(({\langle\rangle}{\exists}gNt(s(g))\backslash Sf)/{\exists}aNa), {\square}{\forall}a{\forall}f(({\langle\rangle}Na\backslash Sf)\backslash ({\langle\rangle}Na\backslash Sf))]]\ \Rightarrow\ {\it CN}{\it s(m)}
\using {\forall}L
\endprooftree
\justifies
{\square}{\it CN}{\it s(m)}, [[\mbox{\fbox{${\blacksquare}{\forall}n({[]^{-1}}{[]^{-1}}({\it CN}{\it n}\backslash {\it CN}{\it n})/{\blacksquare}(({\langle\rangle}Nt(n){\sqcap}!{\blacksquare}Nt(n))\backslash Sf))$}}, [{\blacksquare}Nt(s(f))], {\square}(({\langle\rangle}{\exists}gNt(s(g))\backslash Sf)/{\exists}aNa), {\square}{\forall}a{\forall}f(({\langle\rangle}Na\backslash Sf)\backslash ({\langle\rangle}Na\backslash Sf))]]\ \Rightarrow\ {\it CN}{\it s(m)}
\using {\blacksquare}L
\endprooftree}
$}
\end{center}
\caption{Derivation of medial relativisation: \syncnst{man that Mary likes today}}
\label{mtmlt}
\end{figure}

There follows an example of medial object relativisation (the gap is in a non-peripheral
position left of the adverb):
\disp{
${\bf man}{+}[[{\bf that}{+}[{\bf mary}]{+}{\bf likes}{+}{\bf today}]]: {\it CN}{\it s(m)}$}
Appropriate lexical lookup yields:
\disp{
${\square}{\it CN}{\it s(m)}: {\it man}, [[{\blacksquare}{\forall}n({[]^{-1}}{[]^{-1}}({\it CN}{\it n}\backslash {\it CN}{\it n})/{\blacksquare}(({\langle\rangle}Nt(n){\sqcap}!{\blacksquare}Nt(n))\backslash Sf)):\\ \lambda A\lambda B\lambda C[({\it B}\ {\it C})\wedge ({\it A}\ {\it C})], [{\blacksquare}Nt(s(f)): {\it m}], {\square}(({\langle\rangle}{\exists}gNt(s(g))\backslash Sf)/{\exists}aNa):\\ \mbox{\^{}}\lambda D\lambda E({\it Pres}\ ((\mbox{\v{}}{\it like}\ {\it D})\ {\it E})), {\square}{\forall}a{\forall}f(({\langle\rangle}Na\backslash Sf)\backslash ({\langle\rangle}Na\backslash Sf)): \mbox{\^{}}\lambda F\lambda G(\mbox{\v{}}{\it today}\ ({\it F}\ {\it G}))]]\ \Rightarrow\ {\it CN}{\it s(m)}$}
There is the derivation in Figure~\ref{mtmlt}. 
Analysis of the complex common noun phrase begins at the lefthand
subtree $\textcircled{1}$. After modality right and conditionalisation
of the conjoined hypothetical subtype, 
additive conjunction left applies to this latter to select the object relativisation
subtype, 
which then moves into the stoup.
After preparation of the adverb the stoup contents pass into its
argument subbranch. 
Note how the object relativisation hypothetical gap subtype
percolates in the stoup to satisfy the transitive verb object valency.
The  semantics delivered is:
\disp{
$\lambda C[(\mbox{\v{}}{\it man}\ {\it C})\wedge (\mbox{\v{}}{\it today}\ ({\it Pres}\ ((\mbox{\v{}}{\it like}\ {\it C})\ {\it m})))]$}

As we see in (\ref{obligext}), \lingform{assure\/} has an obligatory extraction
valency: its second object cannot be realized canonically,
by lexical material, but must correspond to an extraction dependency
(Kayne 1984\cite{kayne:bin84}):
\disp{\begin{tabular}[t]{ll}
a. & $\unacc$ John assures Mary Bill to be reliable.\\
b. & the man that John assures Mary to be reliable
\end{tabular}\label{obligext}}
This can be captured by assigning the verb a type in which that extraction valency
is marked by the universal subexponential modality:
$\lingform{assure}\ass (((N\bsl S)/(N\bsl S))/\univexp N)/N$.
Then (\ref{obligext}a) is correctly blocked because $N\yields \univexp N$ is not derivable
whereas (\ref{obligext}b) is correctly derived because $\univexp N\yields \univexp N$
is derivable:

\disp{\prooftree
\prooftree
N\yields N
\justifies
\univexp N\yields N
\using \univexp L
\endprooftree
\justifies
\univexp N\yields \univexp N
\using \univexp R
\endprooftree}

As we remarked as the beginning of Section~\ref{initexchap}
subjects are weak islands (the Subject Condition of Chomsky 1973\cite{chomsky:73});
accordingly in our CatLog2 fragment there is no derivation of simple relativisation
from a subject such as:
\disp{
${\bf man}{+}[[{\bf that}{+}[{\bf the}{+}{\bf friends}{+}{\bf of}]{+}{\bf walk}]]: {\it CN}{\it s(m)}$}
This is because \lingform{walk\/} projects brackets around its subject, but the
permutation of the $\univexp$ hypothetical gap subtype issued by the relative
pronoun is limited to its zone
and cannot penetrate a bracketed subzone.
Roughly, the derivation blocks at~~~\unacc~in:
\disp{
\prooftree
\prooftree
\prooftree
{}[N/\CN, \CN/\PP, \PP/N, N], N\bsl S\yields S
\justifies
N; [N/\CN, \CN/\PP, \PP/N], N\bsl S\yields S
\using~\unacc\univexp P
\endprooftree
\justifies
\univexp N, [N/\CN, \CN/\PP, \PP/N], N\bsl S\yields S
\using \univexp L
\endprooftree
\justifies
{}[N/\CN, \CN/\PP, \PP/N], N\bsl S\yields \univexp N\bsl S
\using \bsl R
\endprooftree
}

However, a weak island `parasitic' gap 
can be licensed by a host gap:
\disp{
${\bf man}{+}[[{\bf that}{+}{\bf the}{+}{\bf friends}{+}{\bf of}{+}{\bf admire}]]: {\it CN}{\it s(m)}$}
Lexical lookup yields:\footnote{We gloss over the use of `difference' here to mark
non-third person singular; its use depends on \emph{absence\/} of derivability
(negation as failure) which of course cannot easily be displayed.}
\disp{
${\square}{\it CN}{\it s(m)}: {\it man}, [[{\blacksquare}{\forall}n({[]^{-1}}{[]^{-1}}({\it CN}{\it n}\backslash {\it CN}{\it n})/{\blacksquare}(({\langle\rangle}Nt(n){\sqcap}!{\blacksquare}Nt(n))\backslash Sf)):\\ \lambda A\lambda B\lambda C[({\it B}\ {\it C})\wedge ({\it A}\ {\it C})], {\blacksquare}{\forall}n(Nt(n)/{\it CN}{\it n}): \iota , {\square}({\it CN}{\it p}/{\it PP}{\it of}): {\it friends},\\ {\square}(({\forall}n({\it CN}{\it n}\backslash {\it CN}{\it n})/{\blacksquare}{\exists}bNb){\&}({\it PP}{\it of}/{\exists}aNa)): \mbox{\^{}}(\mbox{\v{}}{\it of}, \lambda D{\it D}),\\
 {\square}(({\langle\rangle}({\exists}aNa{-}{\exists}gNt(s(g)))\backslash Sf)/{\exists}aNa): \mbox{\^{}}\lambda E\lambda F({\it Pres}\ ((\mbox{\v{}}{\it admire}\ {\it E})\ {\it F}))]]\ \Rightarrow\ {\it CN}{\it s(m)}$}
There is the derivation given in Figure~\ref{mttfoa},
where the use of contraction $\univexp C$, involving brackets and 
stoups,
corresponds to generating the parasitic gap.
The object relativisation hypothetical subtype moves into the stoup
at depth  seven in the lefthand subtree (before this the analysis is standard).
Contraction then applies copying the gap type into the stoup of a newly
created bracketed domain around the subordinate subject.
Applications of $\univexp P$ then move the stoup contents
into the object position of \lingform{admire\/} (host) and
\lingform{of\/} (parasitic). 
\begin{figure}
\begin{center}
\rotatebox{-90}{
\resizebox{\textheight}{!}{
\prooftree
\prooftree
\prooftree
\prooftree
\prooftree
\prooftree
\prooftree
\prooftree
\prooftree
\prooftree
\prooftree
\prooftree
\prooftree
\prooftree
\prooftree
\justifies
\mbox{\fbox{$Nt(s(m))$}}\ \Rightarrow\ Nt(s(m))
\endprooftree
\justifies
\mbox{\fbox{${\blacksquare}Nt(s(m))$}}\ \Rightarrow\ Nt(s(m))
\using {\blacksquare}L
\endprooftree
\justifies
{\blacksquare}Nt(s(m))\ \Rightarrow\ \fbox{${\exists}aNa$}
\using {\exists}R
\endprooftree
\prooftree
\prooftree
\prooftree
\prooftree
\prooftree
\prooftree
\prooftree
\prooftree
\vdots
\justifies
\mbox{\fbox{${\square}({\it CN}{\it p}/{\it PP}{\it of})$}}, {\square}(({\forall}n({\it CN}{\it n}\backslash {\it CN}{\it n})/{\blacksquare}{\exists}bNb){\&}({\it PP}{\it of}/{\exists}aNa)), {\blacksquare}Nt(s(m))\ \Rightarrow\ {\it CN}{\it p}
\using {\Box}L
\endprooftree
\prooftree
\justifies
\mbox{\fbox{$Nt(p)$}}\ \Rightarrow\ Nt(p)
\endprooftree
\justifies
\mbox{\fbox{$Nt(p)/{\it CN}{\it p}$}}, {\square}({\it CN}{\it p}/{\it PP}{\it of}), {\square}(({\forall}n({\it CN}{\it n}\backslash {\it CN}{\it n})/{\blacksquare}{\exists}bNb){\&}({\it PP}{\it of}/{\exists}aNa)), {\blacksquare}Nt(s(m))\ \Rightarrow\ Nt(p)
\using {/}L
\endprooftree
\justifies
\mbox{\fbox{${\forall}n(Nt(n)/{\it CN}{\it n})$}}, {\square}({\it CN}{\it p}/{\it PP}{\it of}), {\square}(({\forall}n({\it CN}{\it n}\backslash {\it CN}{\it n})/{\blacksquare}{\exists}bNb){\&}({\it PP}{\it of}/{\exists}aNa)), {\blacksquare}Nt(s(m))\ \Rightarrow\ Nt(p)
\using {\forall}L
\endprooftree
\justifies
\mbox{\fbox{${\blacksquare}{\forall}n(Nt(n)/{\it CN}{\it n})$}}, {\square}({\it CN}{\it p}/{\it PP}{\it of}), {\square}(({\forall}n({\it CN}{\it n}\backslash {\it CN}{\it n})/{\blacksquare}{\exists}bNb){\&}({\it PP}{\it of}/{\exists}aNa)), {\blacksquare}Nt(s(m))\ \Rightarrow\ Nt(p)
\using {\blacksquare}L
\endprooftree
\justifies
{\blacksquare}{\forall}n(Nt(n)/{\it CN}{\it n}), {\square}({\it CN}{\it p}/{\it PP}{\it of}), {\square}(({\forall}n({\it CN}{\it n}\backslash {\it CN}{\it n})/{\blacksquare}{\exists}bNb){\&}({\it PP}{\it of}/{\exists}aNa)), {\blacksquare}Nt(s(m))\ \Rightarrow\ \fbox{${\exists}aNa$}
\using {\exists}R
\endprooftree
\justifies
{\blacksquare}{\forall}n(Nt(n)/{\it CN}{\it n}), {\square}({\it CN}{\it p}/{\it PP}{\it of}), {\square}(({\forall}n({\it CN}{\it n}\backslash {\it CN}{\it n})/{\blacksquare}{\exists}bNb){\&}({\it PP}{\it of}/{\exists}aNa)), {\blacksquare}Nt(s(m))\ \Rightarrow\ \fbox{${\exists}aNa{-}{\exists}gNt(s(g))$}
\using {-}R
\endprooftree
\justifies
[{\blacksquare}{\forall}n(Nt(n)/{\it CN}{\it n}), {\square}({\it CN}{\it p}/{\it PP}{\it of}), {\square}(({\forall}n({\it CN}{\it n}\backslash {\it CN}{\it n})/{\blacksquare}{\exists}bNb){\&}({\it PP}{\it of}/{\exists}aNa)), {\blacksquare}Nt(s(m))]\ \Rightarrow\ \fbox{${\langle\rangle}({\exists}aNa{-}{\exists}gNt(s(g)))$}
\using {\langle\rangle}R
\endprooftree
\prooftree
\justifies
\mbox{\fbox{$Sf$}}\ \Rightarrow\ Sf
\endprooftree
\justifies
[{\blacksquare}{\forall}n(Nt(n)/{\it CN}{\it n}), {\square}({\it CN}{\it p}/{\it PP}{\it of}), {\square}(({\forall}n({\it CN}{\it n}\backslash {\it CN}{\it n})/{\blacksquare}{\exists}bNb){\&}({\it PP}{\it of}/{\exists}aNa)), {\blacksquare}Nt(s(m))], \mbox{\fbox{${\langle\rangle}({\exists}aNa{-}{\exists}gNt(s(g)))\backslash Sf$}}\ \Rightarrow\ Sf
\using {\backslash}L
\endprooftree
\justifies
[{\blacksquare}{\forall}n(Nt(n)/{\it CN}{\it n}), {\square}({\it CN}{\it p}/{\it PP}{\it of}), {\square}(({\forall}n({\it CN}{\it n}\backslash {\it CN}{\it n})/{\blacksquare}{\exists}bNb){\&}({\it PP}{\it of}/{\exists}aNa)), {\blacksquare}Nt(s(m))], \mbox{\fbox{$({\langle\rangle}({\exists}aNa{-}{\exists}gNt(s(g)))\backslash Sf)/{\exists}aNa$}}, {\blacksquare}Nt(s(m))\ \Rightarrow\ Sf
\using {/}L
\endprooftree
\justifies
[{\blacksquare}{\forall}n(Nt(n)/{\it CN}{\it n}), {\square}({\it CN}{\it p}/{\it PP}{\it of}), {\square}(({\forall}n({\it CN}{\it n}\backslash {\it CN}{\it n})/{\blacksquare}{\exists}bNb){\&}({\it PP}{\it of}/{\exists}aNa)), {\blacksquare}Nt(s(m))], \mbox{\fbox{${\square}(({\langle\rangle}({\exists}aNa{-}{\exists}gNt(s(g)))\backslash Sf)/{\exists}aNa)$}}, {\blacksquare}Nt(s(m))\ \Rightarrow\ Sf
\using {\Box}L
\endprooftree
\justifies
[\mbox{\fbox{${\blacksquare}Nt(s(m))$}};{\blacksquare}{\forall}n(Nt(n)/{\it CN}{\it n}), {\square}({\it CN}{\it p}/{\it PP}{\it of}), {\square}(({\forall}n({\it CN}{\it n}\backslash {\it CN}{\it n})/{\blacksquare}{\exists}bNb){\&}({\it PP}{\it of}/{\exists}aNa))], {\square}(({\langle\rangle}({\exists}aNa{-}{\exists}gNt(s(g)))\backslash Sf)/{\exists}aNa), {\blacksquare}Nt(s(m))\ \Rightarrow\ Sf
\using {!}P
\endprooftree
\justifies
\mbox{\fbox{${\blacksquare}Nt(s(m))$}};\ [{\blacksquare}Nt(s(m));{\blacksquare}{\forall}n(Nt(n)/{\it CN}{\it n}), {\square}({\it CN}{\it p}/{\it PP}{\it of}), {\square}(({\forall}n({\it CN}{\it n}\backslash {\it CN}{\it n})/{\blacksquare}{\exists}bNb){\&}({\it PP}{\it of}/{\exists}aNa))], {\square}(({\langle\rangle}({\exists}aNa{-}{\exists}gNt(s(g)))\backslash Sf)/{\exists}aNa)\ \Rightarrow\ Sf
\using {!}P
\endprooftree
\justifies
\mbox{\fbox{${\blacksquare}Nt(s(m))$}};\ {\blacksquare}{\forall}n(Nt(n)/{\it CN}{\it n}), {\square}({\it CN}{\it p}/{\it PP}{\it of}), {\square}(({\forall}n({\it CN}{\it n}\backslash {\it CN}{\it n})/{\blacksquare}{\exists}bNb){\&}({\it PP}{\it of}/{\exists}aNa)), {\square}(({\langle\rangle}({\exists}aNa{-}{\exists}gNt(s(g)))\backslash Sf)/{\exists}aNa)\ \Rightarrow\ Sf
\using {!}C
\endprooftree
\justifies
!{\blacksquare}Nt(s(m)), {\blacksquare}{\forall}n(Nt(n)/{\it CN}{\it n}), {\square}({\it CN}{\it p}/{\it PP}{\it of}), {\square}(({\forall}n({\it CN}{\it n}\backslash {\it CN}{\it n})/{\blacksquare}{\exists}bNb){\&}({\it PP}{\it of}/{\exists}aNa)), {\square}(({\langle\rangle}({\exists}aNa{-}{\exists}gNt(s(g)))\backslash Sf)/{\exists}aNa)\ \Rightarrow\ Sf
\using {!}L
\endprooftree
\justifies
\mbox{\fbox{${\langle\rangle}Nt(s(m)){\sqcap}!{\blacksquare}Nt(s(m))$}}, {\blacksquare}{\forall}n(Nt(n)/{\it CN}{\it n}), {\square}({\it CN}{\it p}/{\it PP}{\it of}), {\square}(({\forall}n({\it CN}{\it n}\backslash {\it CN}{\it n})/{\blacksquare}{\exists}bNb){\&}({\it PP}{\it of}/{\exists}aNa)), {\square}(({\langle\rangle}({\exists}aNa{-}{\exists}gNt(s(g)))\backslash Sf)/{\exists}aNa)\ \Rightarrow\ Sf
\using {\sqcap}L
\endprooftree
\justifies
{\blacksquare}{\forall}n(Nt(n)/{\it CN}{\it n}), {\square}({\it CN}{\it p}/{\it PP}{\it of}), {\square}(({\forall}n({\it CN}{\it n}\backslash {\it CN}{\it n})/{\blacksquare}{\exists}bNb){\&}({\it PP}{\it of}/{\exists}aNa)), {\square}(({\langle\rangle}({\exists}aNa{-}{\exists}gNt(s(g)))\backslash Sf)/{\exists}aNa)\ \Rightarrow\ ({\langle\rangle}Nt(s(m)){\sqcap}!{\blacksquare}Nt(s(m)))\backslash Sf
\using {\backslash}R
\endprooftree
\justifies
{\blacksquare}{\forall}n(Nt(n)/{\it CN}{\it n}), {\square}({\it CN}{\it p}/{\it PP}{\it of}), {\square}(({\forall}n({\it CN}{\it n}\backslash {\it CN}{\it n})/{\blacksquare}{\exists}bNb){\&}({\it PP}{\it of}/{\exists}aNa)), {\square}(({\langle\rangle}({\exists}aNa{-}{\exists}gNt(s(g)))\backslash Sf)/{\exists}aNa)\ \Rightarrow\ {\blacksquare}(({\langle\rangle}Nt(s(m)){\sqcap}!{\blacksquare}Nt(s(m)))\backslash Sf)
\using {\blacksquare}R
\endprooftree
\prooftree
\prooftree
\prooftree
\prooftree
\prooftree
\justifies
\mbox{\fbox{${\it CN}{\it s(m)}$}}\ \Rightarrow\ {\it CN}{\it s(m)}
\endprooftree
\justifies
\mbox{\fbox{${\square}{\it CN}{\it s(m)}$}}\ \Rightarrow\ {\it CN}{\it s(m)}
\using {\Box}L
\endprooftree
\prooftree
\justifies
\mbox{\fbox{${\it CN}{\it s(m)}$}}\ \Rightarrow\ {\it CN}{\it s(m)}
\endprooftree
\justifies
{\square}{\it CN}{\it s(m)}, \mbox{\fbox{${\it CN}{\it s(m)}\backslash {\it CN}{\it s(m)}$}}\ \Rightarrow\ {\it CN}{\it s(m)}
\using {\backslash}L
\endprooftree
\justifies
{\square}{\it CN}{\it s(m)}, [\mbox{\fbox{${[]^{-1}}({\it CN}{\it s(m)}\backslash {\it CN}{\it s(m)})$}}]\ \Rightarrow\ {\it CN}{\it s(m)}
\using {[]^{-1}}L
\endprooftree
\justifies
{\square}{\it CN}{\it s(m)}, [[\mbox{\fbox{${[]^{-1}}{[]^{-1}}({\it CN}{\it s(m)}\backslash {\it CN}{\it s(m)})$}}]]\ \Rightarrow\ {\it CN}{\it s(m)}
\using {[]^{-1}}L
\endprooftree
\justifies
{\square}{\it CN}{\it s(m)}, [[\mbox{\fbox{${[]^{-1}}{[]^{-1}}({\it CN}{\it s(m)}\backslash {\it CN}{\it s(m)})/{\blacksquare}(({\langle\rangle}Nt(s(m)){\sqcap}!{\blacksquare}Nt(s(m)))\backslash Sf)$}}, {\blacksquare}{\forall}n(Nt(n)/{\it CN}{\it n}), {\square}({\it CN}{\it p}/{\it PP}{\it of}), {\square}(({\forall}n({\it CN}{\it n}\backslash {\it CN}{\it n})/{\blacksquare}{\exists}bNb){\&}({\it PP}{\it of}/{\exists}aNa)), {\square}(({\langle\rangle}({\exists}aNa{-}{\exists}gNt(s(g)))\backslash Sf)/{\exists}aNa)]]\ \Rightarrow\ {\it CN}{\it s(m)}
\using {/}L
\endprooftree
\justifies
{\square}{\it CN}{\it s(m)}, [[\mbox{\fbox{${\forall}n({[]^{-1}}{[]^{-1}}({\it CN}{\it n}\backslash {\it CN}{\it n})/{\blacksquare}(({\langle\rangle}Nt(n){\sqcap}!{\blacksquare}Nt(n))\backslash Sf))$}}, {\blacksquare}{\forall}n(Nt(n)/{\it CN}{\it n}), {\square}({\it CN}{\it p}/{\it PP}{\it of}), {\square}(({\forall}n({\it CN}{\it n}\backslash {\it CN}{\it n})/{\blacksquare}{\exists}bNb){\&}({\it PP}{\it of}/{\exists}aNa))]], {\square}(({\langle\rangle}({\exists}aNa{-}{\exists}gNt(s(g)))\backslash Sf)/{\exists}aNa)]]\ \Rightarrow\ {\it CN}{\it s(m)}
\using {\forall}L
\endprooftree
\justifies
{\square}{\it CN}{\it s(m)}, [[\mbox{\fbox{${\blacksquare}{\forall}n({[]^{-1}}{[]^{-1}}({\it CN}{\it n}\backslash {\it CN}{\it n})/{\blacksquare}(({\langle\rangle}Nt(n){\sqcap}!{\blacksquare}Nt(n))\backslash Sf))$}}, {\blacksquare}{\forall}n(Nt(n)/{\it CN}{\it n}), {\square}({\it CN}{\it p}/{\it PP}{\it of}), {\square}(({\forall}n({\it CN}{\it n}\backslash {\it CN}{\it n})/{\blacksquare}{\exists}bNb){\&}({\it PP}{\it of}/{\exists}aNa))]], {\square}(({\langle\rangle}({\exists}aNa{-}{\exists}gNt(s(g)))\backslash Sf)/{\exists}aNa)]]\ \Rightarrow\ {\it CN}{\it s(m)}
\using {\blacksquare}L
\endprooftree}}
\end{center}
\caption{Derivation of \lingform{man that the friends of admire}}
\label{mttfoa}
\end{figure}\clearpage
This delivers the following semantics in which the gap variable is multiply bound:
\disp{
$\lambda C[(\mbox{\v{}}{\it man}\ {\it C})\wedge ({\it Pres}\ ((\mbox{\v{}}{\it admire}\ {\it C})\ (\iota \ (\mbox{\v{}}{\it friends}\ {\it C}))))]$}

Parasitic extraction from strong islands such as coordinate structures is not acceptable:
\disp{
\unacc that$_i$ Mary showed [[John and the friends of $t_i$]] to $t_i$}
This is successfully blocked because strong islands are doubly bracketed.
Although contraction could apply twice to introduce two bracketings, a copy of the hypothetical
gap subtype would remain trapped in the stoup at the intermediate level of bracketing,
blocking overall derivation.
Likewise, 
as we remarked in footnote~\ref{relnopar}, 
parasitic extraction is not possible from relative clauses themselves,
for the same reason: a superfluous gap subtype would remain in between
the double brackets required for the strong island. 

\begin{figure}
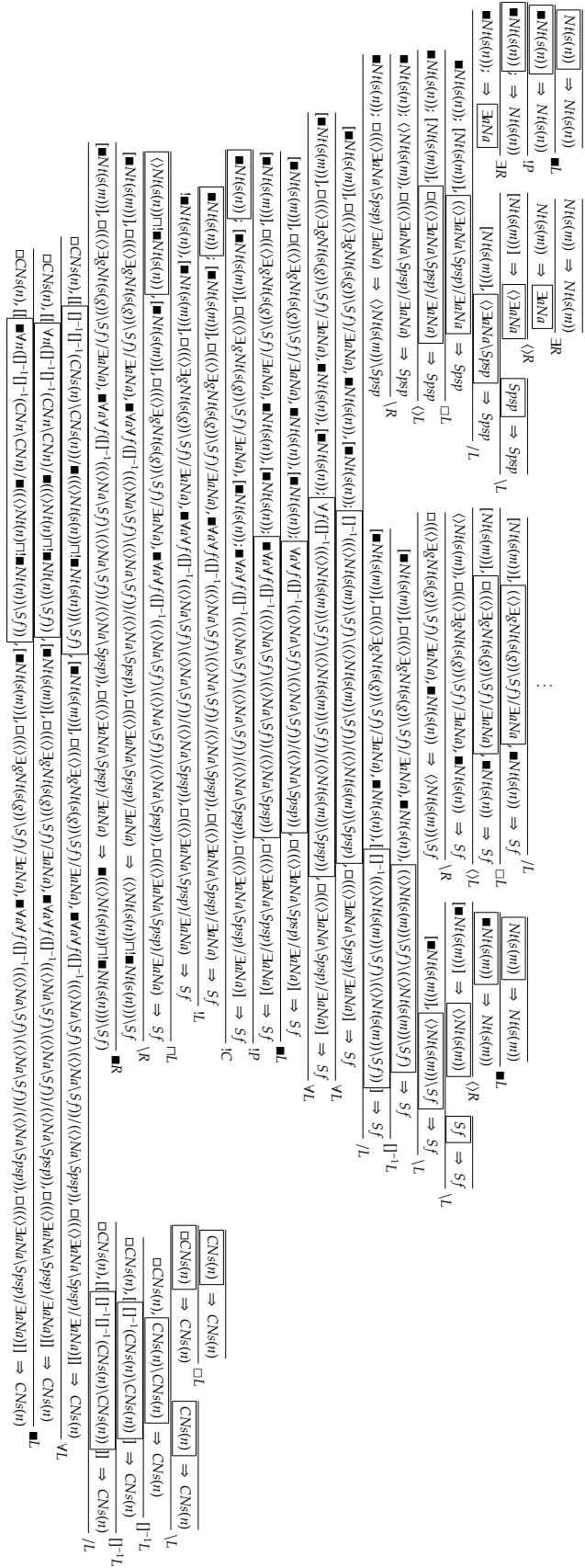

\begin{center}
\rotatebox{-90}{\small
\resizebox{\textheight}{!}{
\prooftree
\prooftree
\prooftree
\prooftree
\prooftree
\prooftree
\prooftree
\prooftree
\prooftree
\prooftree
\prooftree
\prooftree
\prooftree
\prooftree
\prooftree
\prooftree
\prooftree
\prooftree
\prooftree
\prooftree
\prooftree
\justifies
\mbox{\fbox{$Nt(s(n))$}}\ \Rightarrow\ Nt(s(n))
\endprooftree
\justifies
\mbox{\fbox{${\blacksquare}Nt(s(n))$}}\ \Rightarrow\ Nt(s(n))
\using {\blacksquare}L
\endprooftree
\justifies
\mbox{\fbox{${\blacksquare}Nt(s(n))$}};\ \ \Rightarrow\ Nt(s(n))
\using {!}P
\endprooftree
\justifies
{\blacksquare}Nt(s(n));\ \ \Rightarrow\ \fbox{${\exists}aNa$}
\using {\exists}R
\endprooftree
\prooftree
\prooftree
\prooftree
\prooftree
\justifies
Nt(s(m))\ \Rightarrow\ Nt(s(m))
\endprooftree
\justifies
Nt(s(m))\ \Rightarrow\ \fbox{${\exists}aNa$}
\using {\exists}R
\endprooftree
\justifies
[Nt(s(m))]\ \Rightarrow\ \fbox{${\langle\rangle}{\exists}aNa$}
\using {\langle\rangle}R
\endprooftree
\prooftree
\justifies
\mbox{\fbox{$Spsp$}}\ \Rightarrow\ Spsp
\endprooftree
\justifies
[Nt(s(m))], \mbox{\fbox{${\langle\rangle}{\exists}aNa\backslash Spsp$}}\ \Rightarrow\ Spsp
\using {\backslash}L
\endprooftree
\justifies
{\blacksquare}Nt(s(n));\ [Nt(s(m))], \mbox{\fbox{$({\langle\rangle}{\exists}aNa\backslash Spsp)/{\exists}aNa$}}\ \Rightarrow\ Spsp
\using {/}L
\endprooftree
\justifies
{\blacksquare}Nt(s(n));\ [Nt(s(m))], \mbox{\fbox{${\square}(({\langle\rangle}{\exists}aNa\backslash Spsp)/{\exists}aNa)$}}\ \Rightarrow\ Spsp
\using {\Box}L
\endprooftree
\justifies
{\blacksquare}Nt(s(n));\ {\langle\rangle}Nt(s(m)), {\square}(({\langle\rangle}{\exists}aNa\backslash Spsp)/{\exists}aNa)\ \Rightarrow\ Spsp
\using {\langle\rangle}L
\endprooftree
\justifies
{\blacksquare}Nt(s(n));\ {\square}(({\langle\rangle}{\exists}aNa\backslash Spsp)/{\exists}aNa)\ \Rightarrow\ {\langle\rangle}Nt(s(m))\backslash Spsp
\using {\backslash}R
\endprooftree
\prooftree
\prooftree
\prooftree
\prooftree
\prooftree
\prooftree
\vdots
\justifies
[Nt(s(m))], \mbox{\fbox{$({\langle\rangle}{\exists}gNt(s(g))\backslash Sf)/{\exists}aNa$}}, {\blacksquare}Nt(s(n))\ \Rightarrow\ Sf
\using {/}L
\endprooftree
\justifies
[Nt(s(m))], \mbox{\fbox{${\square}(({\langle\rangle}{\exists}gNt(s(g))\backslash Sf)/{\exists}aNa)$}}, {\blacksquare}Nt(s(n))\ \Rightarrow\ Sf
\using {\Box}L
\endprooftree
\justifies
{\langle\rangle}Nt(s(m)), {\square}(({\langle\rangle}{\exists}gNt(s(g))\backslash Sf)/{\exists}aNa), {\blacksquare}Nt(s(n))\ \Rightarrow\ Sf
\using {\langle\rangle}L
\endprooftree
\justifies
{\square}(({\langle\rangle}{\exists}gNt(s(g))\backslash Sf)/{\exists}aNa), {\blacksquare}Nt(s(n))\ \Rightarrow\ {\langle\rangle}Nt(s(m))\backslash Sf
\using {\backslash}R
\endprooftree
\prooftree
\prooftree
\prooftree
\prooftree
\justifies
\mbox{\fbox{$Nt(s(m))$}}\ \Rightarrow\ Nt(s(m))
\endprooftree
\justifies
\mbox{\fbox{${\blacksquare}Nt(s(m))$}}\ \Rightarrow\ Nt(s(m))
\using {\blacksquare}L
\endprooftree
\justifies
[{\blacksquare}Nt(s(m))]\ \Rightarrow\ \fbox{${\langle\rangle}Nt(s(m))$}
\using {\langle\rangle}R
\endprooftree
\prooftree
\justifies
\mbox{\fbox{$Sf$}}\ \Rightarrow\ Sf
\endprooftree
\justifies
[{\blacksquare}Nt(s(m))], \mbox{\fbox{${\langle\rangle}Nt(s(m))\backslash Sf$}}\ \Rightarrow\ Sf
\using {\backslash}L
\endprooftree
\justifies
[{\blacksquare}Nt(s(m))], {\square}(({\langle\rangle}{\exists}gNt(s(g))\backslash Sf)/{\exists}aNa), {\blacksquare}Nt(s(n)), \mbox{\fbox{$({\langle\rangle}Nt(s(m))\backslash Sf)\backslash ({\langle\rangle}Nt(s(m))\backslash Sf)$}}\ \Rightarrow\ Sf
\using {\backslash}L
\endprooftree
\justifies
[{\blacksquare}Nt(s(m))], {\square}(({\langle\rangle}{\exists}gNt(s(g))\backslash Sf)/{\exists}aNa), {\blacksquare}Nt(s(n)), [\mbox{\fbox{${[]^{-1}}(({\langle\rangle}Nt(s(m))\backslash Sf)\backslash ({\langle\rangle}Nt(s(m))\backslash Sf))$}}]\ \Rightarrow\ Sf
\using {[]^{-1}}L
\endprooftree
\justifies
[{\blacksquare}Nt(s(m))], {\square}(({\langle\rangle}{\exists}gNt(s(g))\backslash Sf)/{\exists}aNa), {\blacksquare}Nt(s(n)), [{\blacksquare}Nt(s(n));\mbox{\fbox{${[]^{-1}}(({\langle\rangle}Nt(s(m))\backslash Sf)\backslash ({\langle\rangle}Nt(s(m))\backslash Sf))/({\langle\rangle}Nt(s(m))\backslash Spsp)$}}, {\square}(({\langle\rangle}{\exists}aNa\backslash Spsp)/{\exists}aNa)]\ \Rightarrow\ Sf
\using {/}L
\endprooftree
\justifies
[{\blacksquare}Nt(s(m))], {\square}(({\langle\rangle}{\exists}gNt(s(g))\backslash Sf)/{\exists}aNa), {\blacksquare}Nt(s(n)), [{\blacksquare}Nt(s(n));\mbox{\fbox{${\forall}f({[]^{-1}}(({\langle\rangle}Nt(s(m))\backslash Sf)\backslash ({\langle\rangle}Nt(s(m))\backslash Sf))/({\langle\rangle}Nt(s(m))\backslash Spsp))$}}, {\square}(({\langle\rangle}{\exists}aNa\backslash Spsp)/{\exists}aNa)]\ \Rightarrow\ Sf
\using {\forall}L
\endprooftree
\justifies
[{\blacksquare}Nt(s(m))], {\square}(({\langle\rangle}{\exists}gNt(s(g))\backslash Sf)/{\exists}aNa), {\blacksquare}Nt(s(n)), [{\blacksquare}Nt(s(n));\mbox{\fbox{${\forall}a{\forall}f({[]^{-1}}(({\langle\rangle}Na\backslash Sf)\backslash ({\langle\rangle}Na\backslash Sf))/({\langle\rangle}Na\backslash Spsp))$}}, {\square}(({\langle\rangle}{\exists}aNa\backslash Spsp)/{\exists}aNa)]\ \Rightarrow\ Sf
\using {\forall}L
\endprooftree
\justifies
[{\blacksquare}Nt(s(m))], {\square}(({\langle\rangle}{\exists}gNt(s(g))\backslash Sf)/{\exists}aNa), {\blacksquare}Nt(s(n)), [{\blacksquare}Nt(s(n));\mbox{\fbox{${\blacksquare}{\forall}a{\forall}f({[]^{-1}}(({\langle\rangle}Na\backslash Sf)\backslash ({\langle\rangle}Na\backslash Sf))/({\langle\rangle}Na\backslash Spsp))$}}, {\square}(({\langle\rangle}{\exists}aNa\backslash Spsp)/{\exists}aNa)]\ \Rightarrow\ Sf
\using {\blacksquare}L
\endprooftree
\justifies
\mbox{\fbox{${\blacksquare}Nt(s(n))$}};\ [{\blacksquare}Nt(s(m))], {\square}(({\langle\rangle}{\exists}gNt(s(g))\backslash Sf)/{\exists}aNa), [{\blacksquare}Nt(s(n));{\blacksquare}{\forall}a{\forall}f({[]^{-1}}(({\langle\rangle}Na\backslash Sf)\backslash ({\langle\rangle}Na\backslash Sf))/({\langle\rangle}Na\backslash Spsp)), {\square}(({\langle\rangle}{\exists}aNa\backslash Spsp)/{\exists}aNa)]\ \Rightarrow\ Sf
\using {!}P
\endprooftree
\justifies
\mbox{\fbox{${\blacksquare}Nt(s(n))$}};\ [{\blacksquare}Nt(s(m))], {\square}(({\langle\rangle}{\exists}gNt(s(g))\backslash Sf)/{\exists}aNa), {\blacksquare}{\forall}a{\forall}f({[]^{-1}}(({\langle\rangle}Na\backslash Sf)\backslash ({\langle\rangle}Na\backslash Sf))/({\langle\rangle}Na\backslash Spsp)), {\square}(({\langle\rangle}{\exists}aNa\backslash Spsp)/{\exists}aNa)\ \Rightarrow\ Sf
\using {!}C
\endprooftree
\justifies
!{\blacksquare}Nt(s(n)), [{\blacksquare}Nt(s(m))], {\square}(({\langle\rangle}{\exists}gNt(s(g))\backslash Sf)/{\exists}aNa), {\blacksquare}{\forall}a{\forall}f({[]^{-1}}(({\langle\rangle}Na\backslash Sf)\backslash ({\langle\rangle}Na\backslash Sf))/({\langle\rangle}Na\backslash Spsp)), {\square}(({\langle\rangle}{\exists}aNa\backslash Spsp)/{\exists}aNa)\ \Rightarrow\ Sf
\using {!}L
\endprooftree
\justifies
\mbox{\fbox{${\langle\rangle}Nt(s(n)){\sqcap}!{\blacksquare}Nt(s(n))$}}, [{\blacksquare}Nt(s(m))], {\square}(({\langle\rangle}{\exists}gNt(s(g))\backslash Sf)/{\exists}aNa), {\blacksquare}{\forall}a{\forall}f({[]^{-1}}(({\langle\rangle}Na\backslash Sf)\backslash ({\langle\rangle}Na\backslash Sf))/({\langle\rangle}Na\backslash Spsp)), {\square}(({\langle\rangle}{\exists}aNa\backslash Spsp)/{\exists}aNa)\ \Rightarrow\ Sf
\using {\sqcap}L
\endprooftree
\justifies
[{\blacksquare}Nt(s(m))], {\square}(({\langle\rangle}{\exists}gNt(s(g))\backslash Sf)/{\exists}aNa), {\blacksquare}{\forall}a{\forall}f({[]^{-1}}(({\langle\rangle}Na\backslash Sf)\backslash ({\langle\rangle}Na\backslash Sf))/({\langle\rangle}Na\backslash Spsp)), {\square}(({\langle\rangle}{\exists}aNa\backslash Spsp)/{\exists}aNa)\ \Rightarrow\ ({\langle\rangle}Nt(s(n)){\sqcap}!{\blacksquare}Nt(s(n)))\backslash Sf
\using {\backslash}R
\endprooftree
\justifies
[{\blacksquare}Nt(s(m))], {\square}(({\langle\rangle}{\exists}gNt(s(g))\backslash Sf)/{\exists}aNa), {\blacksquare}{\forall}a{\forall}f({[]^{-1}}(({\langle\rangle}Na\backslash Sf)\backslash ({\langle\rangle}Na\backslash Sf))/({\langle\rangle}Na\backslash Spsp)), {\square}(({\langle\rangle}{\exists}aNa\backslash Spsp)/{\exists}aNa)\ \Rightarrow\ {\blacksquare}(({\langle\rangle}Nt(s(n)){\sqcap}!{\blacksquare}Nt(s(n)))\backslash Sf)
\using {\blacksquare}R
\endprooftree
\prooftree
\prooftree
\prooftree
\prooftree
\prooftree
\justifies
\mbox{\fbox{${\it CN}{\it s(n)}$}}\ \Rightarrow\ {\it CN}{\it s(n)}
\endprooftree
\justifies
\mbox{\fbox{${\square}{\it CN}{\it s(n)}$}}\ \Rightarrow\ {\it CN}{\it s(n)}
\using {\Box}L
\endprooftree
\prooftree
\justifies
\mbox{\fbox{${\it CN}{\it s(n)}$}}\ \Rightarrow\ {\it CN}{\it s(n)}
\endprooftree
\justifies
{\square}{\it CN}{\it s(n)}, \mbox{\fbox{${\it CN}{\it s(n)}\backslash {\it CN}{\it s(n)}$}}\ \Rightarrow\ {\it CN}{\it s(n)}
\using {\backslash}L
\endprooftree
\justifies
{\square}{\it CN}{\it s(n)}, [\mbox{\fbox{${[]^{-1}}({\it CN}{\it s(n)}\backslash {\it CN}{\it s(n)})$}}]\ \Rightarrow\ {\it CN}{\it s(n)}
\using {[]^{-1}}L
\endprooftree
\justifies
{\square}{\it CN}{\it s(n)}, [[\mbox{\fbox{${[]^{-1}}{[]^{-1}}({\it CN}{\it s(n)}\backslash {\it CN}{\it s(n)})$}}]]\ \Rightarrow\ {\it CN}{\it s(n)}
\using {[]^{-1}}L
\endprooftree
\justifies
{\square}{\it CN}{\it s(n)}, [[\mbox{\fbox{${[]^{-1}}{[]^{-1}}({\it CN}{\it s(n)}\backslash {\it CN}{\it s(n)})/{\blacksquare}(({\langle\rangle}Nt(s(n)){\sqcap}!{\blacksquare}Nt(s(n)))\backslash Sf)$}}, [{\blacksquare}Nt(s(m))], {\square}(({\langle\rangle}{\exists}gNt(s(g))\backslash Sf)/{\exists}aNa), {\blacksquare}{\forall}a{\forall}f({[]^{-1}}(({\langle\rangle}Na\backslash Sf)\backslash ({\langle\rangle}Na\backslash Sf))/({\langle\rangle}Na\backslash Spsp)), {\square}(({\langle\rangle}{\exists}aNa\backslash Spsp)/{\exists}aNa)]]\ \Rightarrow\ {\it CN}{\it s(n)}
\using {/}L
\endprooftree
\justifies
{\square}{\it CN}{\it s(n)}, [[\mbox{\fbox{${\forall}n({[]^{-1}}{[]^{-1}}({\it CN}{\it n}\backslash {\it CN}{\it n})/{\blacksquare}(({\langle\rangle}Nt(n){\sqcap}!{\blacksquare}Nt(n))\backslash Sf))$}}, [{\blacksquare}Nt(s(m))], {\square}(({\langle\rangle}{\exists}gNt(s(g))\backslash Sf)/{\exists}aNa), {\blacksquare}{\forall}a{\forall}f({[]^{-1}}(({\langle\rangle}Na\backslash Sf)\backslash ({\langle\rangle}Na\backslash Sf))/({\langle\rangle}Na\backslash Spsp)), {\square}(({\langle\rangle}{\exists}aNa\backslash Spsp)/{\exists}aNa)]]\ \Rightarrow\ {\it CN}{\it s(n)}
\using {\forall}L
\endprooftree
\justifies
{\square}{\it CN}{\it s(n)}, [[\mbox{\fbox{${\blacksquare}{\forall}n({[]^{-1}}{[]^{-1}}({\it CN}{\it n}\backslash {\it CN}{\it n})/{\blacksquare}(({\langle\rangle}Nt(n){\sqcap}!{\blacksquare}Nt(n))\backslash Sf))$}}, [{\blacksquare}Nt(s(m))], {\square}(({\langle\rangle}{\exists}gNt(s(g))\backslash Sf)/{\exists}aNa), {\blacksquare}{\forall}a{\forall}f({[]^{-1}}(({\langle\rangle}Na\backslash Sf)\backslash ({\langle\rangle}Na\backslash Sf))/({\langle\rangle}Na\backslash Spsp)), {\square}(({\langle\rangle}{\exists}aNa\backslash Spsp)/{\exists}aNa)]]\ \Rightarrow\ {\it CN}{\it s(n)}
\using {\blacksquare}L
\endprooftree}}
\end{center}
\caption{Derivation for \lingform{paper that John filed without reading}},
\label{ptjfwr}
\end{figure}

A parasitic gap can also appear in an adverbial weak island:
\disp{
${\bf paper}{+}[[{\bf that}{+}[{\bf john}]{+}{\bf filed}{+}{\bf without}{+}{\bf reading}]]: {\it CN}{\it s(n)}$}
Lexical lookup for this example yields:
\disp{
${\square}{\it CN}{\it s(n)}: {\it paper}, [[{\blacksquare}{\forall}n({[]^{-1}}{[]^{-1}}({\it CN}{\it n}\backslash {\it CN}{\it n})/{\blacksquare}(({\langle\rangle}Nt(n){\sqcap}!{\blacksquare}Nt(n))\backslash Sf)):\\ \lambda A\lambda B\lambda C[({\it B}\ {\it C})\wedge ({\it A}\ {\it C})], [{\blacksquare}Nt(s(m)): {\it j}], {\square}(({\langle\rangle}{\exists}gNt(s(g))\backslash Sf)/{\exists}aNa):\\
\mbox{\^{}}\lambda D\lambda E({\it Past}\ ((\mbox{\v{}}{\it file}\ {\it D})\ {\it E})), {\blacksquare}{\forall}a{\forall}f({[]^{-1}}(({\langle\rangle}Na\backslash Sf)\backslash ({\langle\rangle}Na\backslash Sf))/({\langle\rangle}Na\backslash Spsp)):\\ \lambda F\lambda G\lambda H[({\it G}\ {\it H})\wedge \neg ({\it F}\ {\it H})], {\square}(({\langle\rangle}{\exists}aNa\backslash Spsp)/{\exists}aNa): \mbox{\^{}}\lambda I\lambda J((\mbox{\v{}}{\it read}\ {\it I})\ {\it J})]]\ \Rightarrow\ {\it CN}{\it s(n)}$}
There is the derivation given in Figure~\ref{ptjfwr}.
This time at depth eight contraction copies the host stoup gap into the stoup of
a newly created bracketed domain around the subordinate adverbial phrase.
This delivers semantics:
\disp{
$\lambda C[(\mbox{\v{}}{\it paper}\ {\it C})\wedge [({\it Past}\ ((\mbox{\v{}}{\it file}\ {\it C})\ {\it j}))\wedge \neg ((\mbox{\v{}}{\it read}\ {\it C})\ {\it j})]]$}

\begin{figure}
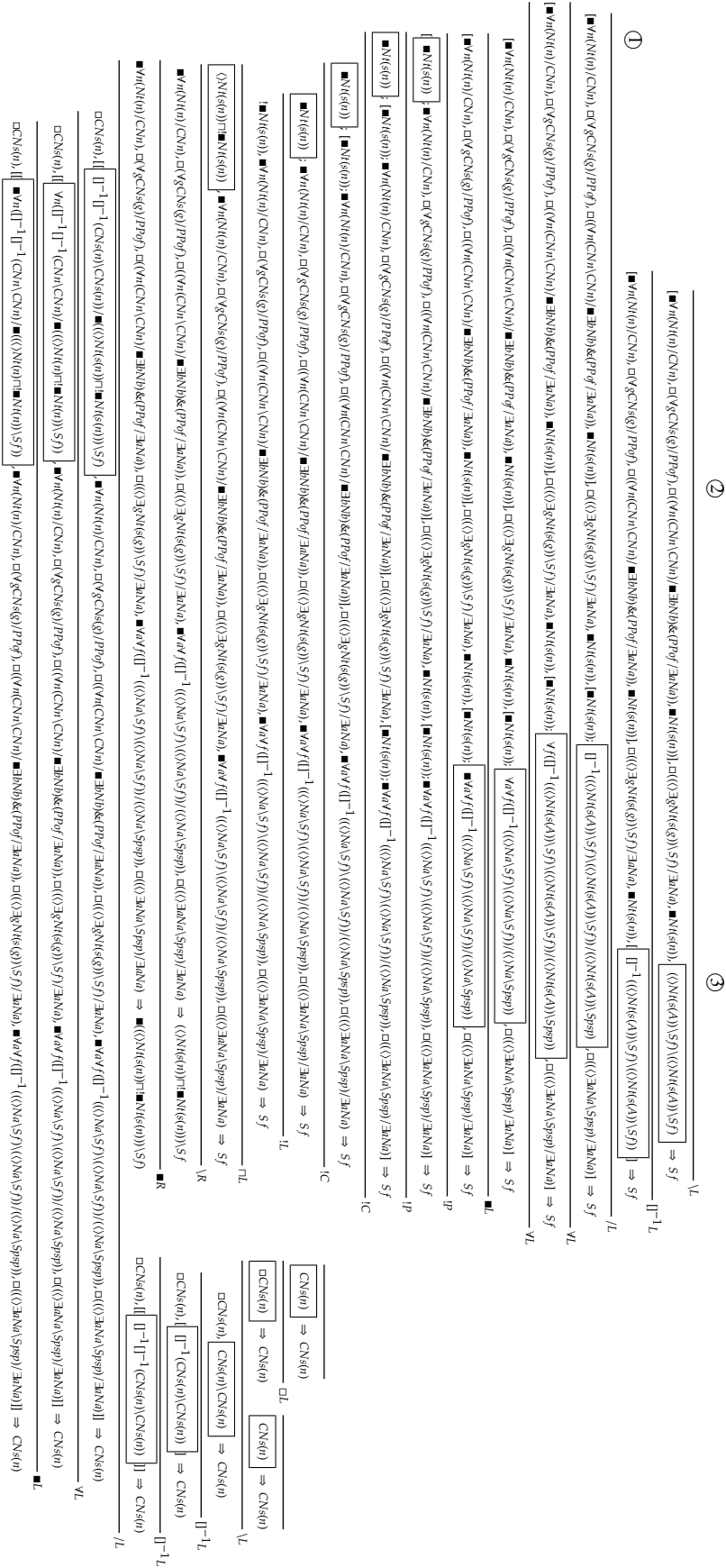

\begin{center}
\rotatebox{-90}{\tiny
\prooftree
\prooftree
\prooftree
\prooftree
\prooftree
\prooftree
\prooftree
\prooftree
\prooftree
\prooftree
\prooftree
\prooftree
\prooftree
\prooftree
\prooftree
\mbox{\footnotesize\textcircled{1}}\tab\tab\tb
\prooftree
\prooftree
\mbox{\footnotesize\textcircled{2}}\tab\tab\tab\tab\tab\tb
\mbox{\footnotesize\textcircled{3}}
\justifies
[{\blacksquare}{\forall}n(Nt(n)/{\it CN}{\it n}), {\square}({\forall}g{\it CN}{\it s(g)}/{\it PP}{\it of}), {\square}(({\forall}n({\it CN}{\it n}\backslash {\it CN}{\it n})/{\blacksquare}{\exists}bNb){\&}({\it PP}{\it of}/{\exists}aNa)), {\blacksquare}Nt(s(n))], {\square}(({\langle\rangle}{\exists}gNt(s(g))\backslash Sf)/{\exists}aNa), {\blacksquare}Nt(s(n)), \mbox{\fbox{$({\langle\rangle}Nt(s(A))\backslash Sf)\backslash ({\langle\rangle}Nt(s(A))\backslash Sf)$}}\ \Rightarrow\ Sf
\using {\backslash}L
\endprooftree
\justifies
[{\blacksquare}{\forall}n(Nt(n)/{\it CN}{\it n}), {\square}({\forall}g{\it CN}{\it s(g)}/{\it PP}{\it of}), {\square}(({\forall}n({\it CN}{\it n}\backslash {\it CN}{\it n})/{\blacksquare}{\exists}bNb){\&}({\it PP}{\it of}/{\exists}aNa)), {\blacksquare}Nt(s(n))], {\square}(({\langle\rangle}{\exists}gNt(s(g))\backslash Sf)/{\exists}aNa), {\blacksquare}Nt(s(n)), [\mbox{\fbox{${[]^{-1}}(({\langle\rangle}Nt(s(A))\backslash Sf)\backslash ({\langle\rangle}Nt(s(A))\backslash Sf))$}}]\ \Rightarrow\ Sf
\using {[]^{-1}}L
\endprooftree
\justifies
[{\blacksquare}{\forall}n(Nt(n)/{\it CN}{\it n}), {\square}({\forall}g{\it CN}{\it s(g)}/{\it PP}{\it of}), {\square}(({\forall}n({\it CN}{\it n}\backslash {\it CN}{\it n})/{\blacksquare}{\exists}bNb){\&}({\it PP}{\it of}/{\exists}aNa)), {\blacksquare}Nt(s(n))], {\square}(({\langle\rangle}{\exists}gNt(s(g))\backslash Sf)/{\exists}aNa), {\blacksquare}Nt(s(n)), [{\blacksquare}Nt(s(n));\mbox{\fbox{${[]^{-1}}(({\langle\rangle}Nt(s(A))\backslash Sf)\backslash ({\langle\rangle}Nt(s(A))\backslash Sf))/({\langle\rangle}Nt(s(A))\backslash Spsp)$}}, {\square}(({\langle\rangle}{\exists}aNa\backslash Spsp)/{\exists}aNa)]\ \Rightarrow\ Sf
\using {/}L
\endprooftree
\justifies
[{\blacksquare}{\forall}n(Nt(n)/{\it CN}{\it n}), {\square}({\forall}g{\it CN}{\it s(g)}/{\it PP}{\it of}), {\square}(({\forall}n({\it CN}{\it n}\backslash {\it CN}{\it n})/{\blacksquare}{\exists}bNb){\&}({\it PP}{\it of}/{\exists}aNa)), {\blacksquare}Nt(s(n))], {\square}(({\langle\rangle}{\exists}gNt(s(g))\backslash Sf)/{\exists}aNa), {\blacksquare}Nt(s(n)), [{\blacksquare}Nt(s(n));\mbox{\fbox{${\forall}f({[]^{-1}}(({\langle\rangle}Nt(s(A))\backslash Sf)\backslash ({\langle\rangle}Nt(s(A))\backslash Sf))/({\langle\rangle}Nt(s(A))\backslash Spsp))$}}, {\square}(({\langle\rangle}{\exists}aNa\backslash Spsp)/{\exists}aNa)]\ \Rightarrow\ Sf
\using {\forall}L
\endprooftree
\justifies
[{\blacksquare}{\forall}n(Nt(n)/{\it CN}{\it n}), {\square}({\forall}g{\it CN}{\it s(g)}/{\it PP}{\it of}), {\square}(({\forall}n({\it CN}{\it n}\backslash {\it CN}{\it n})/{\blacksquare}{\exists}bNb){\&}({\it PP}{\it of}/{\exists}aNa)), {\blacksquare}Nt(s(n))], {\square}(({\langle\rangle}{\exists}gNt(s(g))\backslash Sf)/{\exists}aNa), {\blacksquare}Nt(s(n)), [{\blacksquare}Nt(s(n));\mbox{\fbox{${\forall}a{\forall}f({[]^{-1}}(({\langle\rangle}Na\backslash Sf)\backslash ({\langle\rangle}Na\backslash Sf))/({\langle\rangle}Na\backslash Spsp))$}}, {\square}(({\langle\rangle}{\exists}aNa\backslash Spsp)/{\exists}aNa)]\ \Rightarrow\ Sf
\using {\forall}L
\endprooftree
\justifies
[{\blacksquare}{\forall}n(Nt(n)/{\it CN}{\it n}), {\square}({\forall}g{\it CN}{\it s(g)}/{\it PP}{\it of}), {\square}(({\forall}n({\it CN}{\it n}\backslash {\it CN}{\it n})/{\blacksquare}{\exists}bNb){\&}({\it PP}{\it of}/{\exists}aNa)), {\blacksquare}Nt(s(n))], {\square}(({\langle\rangle}{\exists}gNt(s(g))\backslash Sf)/{\exists}aNa), {\blacksquare}Nt(s(n)), [{\blacksquare}Nt(s(n));\mbox{\fbox{${\blacksquare}{\forall}a{\forall}f({[]^{-1}}(({\langle\rangle}Na\backslash Sf)\backslash ({\langle\rangle}Na\backslash Sf))/({\langle\rangle}Na\backslash Spsp))$}}, {\square}(({\langle\rangle}{\exists}aNa\backslash Spsp)/{\exists}aNa)]\ \Rightarrow\ Sf
\using {\blacksquare}L
\endprooftree
\justifies
[\mbox{\fbox{${\blacksquare}Nt(s(n))$}};{\blacksquare}{\forall}n(Nt(n)/{\it CN}{\it n}), {\square}({\forall}g{\it CN}{\it s(g)}/{\it PP}{\it of}), {\square}(({\forall}n({\it CN}{\it n}\backslash {\it CN}{\it n})/{\blacksquare}{\exists}bNb){\&}({\it PP}{\it of}/{\exists}aNa))], {\square}(({\langle\rangle}{\exists}gNt(s(g))\backslash Sf)/{\exists}aNa), {\blacksquare}Nt(s(n)), [{\blacksquare}Nt(s(n));{\blacksquare}{\forall}a{\forall}f({[]^{-1}}(({\langle\rangle}Na\backslash Sf)\backslash ({\langle\rangle}Na\backslash Sf))/({\langle\rangle}Na\backslash Spsp)), {\square}(({\langle\rangle}{\exists}aNa\backslash Spsp)/{\exists}aNa)]\ \Rightarrow\ Sf
\using {!}P
\endprooftree
\justifies
\mbox{\fbox{${\blacksquare}Nt(s(n))$}};\ [{\blacksquare}Nt(s(n));{\blacksquare}{\forall}n(Nt(n)/{\it CN}{\it n}), {\square}({\forall}g{\it CN}{\it s(g)}/{\it PP}{\it of}), {\square}(({\forall}n({\it CN}{\it n}\backslash {\it CN}{\it n})/{\blacksquare}{\exists}bNb){\&}({\it PP}{\it of}/{\exists}aNa))], {\square}(({\langle\rangle}{\exists}gNt(s(g))\backslash Sf)/{\exists}aNa), [{\blacksquare}Nt(s(n));{\blacksquare}{\forall}a{\forall}f({[]^{-1}}(({\langle\rangle}Na\backslash Sf)\backslash ({\langle\rangle}Na\backslash Sf))/({\langle\rangle}Na\backslash Spsp)), {\square}(({\langle\rangle}{\exists}aNa\backslash Spsp)/{\exists}aNa)]\ \Rightarrow\ Sf
\using {!}P
\endprooftree
\justifies
\mbox{\fbox{${\blacksquare}Nt(s(n))$}};\ [{\blacksquare}Nt(s(n));{\blacksquare}{\forall}n(Nt(n)/{\it CN}{\it n}), {\square}({\forall}g{\it CN}{\it s(g)}/{\it PP}{\it of}), {\square}(({\forall}n({\it CN}{\it n}\backslash {\it CN}{\it n})/{\blacksquare}{\exists}bNb){\&}({\it PP}{\it of}/{\exists}aNa))], {\square}(({\langle\rangle}{\exists}gNt(s(g))\backslash Sf)/{\exists}aNa), {\blacksquare}{\forall}a{\forall}f({[]^{-1}}(({\langle\rangle}Na\backslash Sf)\backslash ({\langle\rangle}Na\backslash Sf))/({\langle\rangle}Na\backslash Spsp)), {\square}(({\langle\rangle}{\exists}aNa\backslash Spsp)/{\exists}aNa)\ \Rightarrow\ Sf
\using {!}C
\endprooftree
\justifies
\mbox{\fbox{${\blacksquare}Nt(s(n))$}};\ {\blacksquare}{\forall}n(Nt(n)/{\it CN}{\it n}), {\square}({\forall}g{\it CN}{\it s(g)}/{\it PP}{\it of}), {\square}(({\forall}n({\it CN}{\it n}\backslash {\it CN}{\it n})/{\blacksquare}{\exists}bNb){\&}({\it PP}{\it of}/{\exists}aNa)), {\square}(({\langle\rangle}{\exists}gNt(s(g))\backslash Sf)/{\exists}aNa), {\blacksquare}{\forall}a{\forall}f({[]^{-1}}(({\langle\rangle}Na\backslash Sf)\backslash ({\langle\rangle}Na\backslash Sf))/({\langle\rangle}Na\backslash Spsp)), {\square}(({\langle\rangle}{\exists}aNa\backslash Spsp)/{\exists}aNa)\ \Rightarrow\ Sf
\using {!}C
\endprooftree
\justifies
!{\blacksquare}Nt(s(n)), {\blacksquare}{\forall}n(Nt(n)/{\it CN}{\it n}), {\square}({\forall}g{\it CN}{\it s(g)}/{\it PP}{\it of}), {\square}(({\forall}n({\it CN}{\it n}\backslash {\it CN}{\it n})/{\blacksquare}{\exists}bNb){\&}({\it PP}{\it of}/{\exists}aNa)), {\square}(({\langle\rangle}{\exists}gNt(s(g))\backslash Sf)/{\exists}aNa), {\blacksquare}{\forall}a{\forall}f({[]^{-1}}(({\langle\rangle}Na\backslash Sf)\backslash ({\langle\rangle}Na\backslash Sf))/({\langle\rangle}Na\backslash Spsp)), {\square}(({\langle\rangle}{\exists}aNa\backslash Spsp)/{\exists}aNa)\ \Rightarrow\ Sf
\using {!}L
\endprooftree
\justifies
\mbox{\fbox{${\langle\rangle}Nt(s(n)){\sqcap}!{\blacksquare}Nt(s(n))$}}, {\blacksquare}{\forall}n(Nt(n)/{\it CN}{\it n}), {\square}({\forall}g{\it CN}{\it s(g)}/{\it PP}{\it of}), {\square}(({\forall}n({\it CN}{\it n}\backslash {\it CN}{\it n})/{\blacksquare}{\exists}bNb){\&}({\it PP}{\it of}/{\exists}aNa)), {\square}(({\langle\rangle}{\exists}gNt(s(g))\backslash Sf)/{\exists}aNa), {\blacksquare}{\forall}a{\forall}f({[]^{-1}}(({\langle\rangle}Na\backslash Sf)\backslash ({\langle\rangle}Na\backslash Sf))/({\langle\rangle}Na\backslash Spsp)), {\square}(({\langle\rangle}{\exists}aNa\backslash Spsp)/{\exists}aNa)\ \Rightarrow\ Sf
\using {\sqcap}L
\endprooftree
\justifies
{\blacksquare}{\forall}n(Nt(n)/{\it CN}{\it n}), {\square}({\forall}g{\it CN}{\it s(g)}/{\it PP}{\it of}), {\square}(({\forall}n({\it CN}{\it n}\backslash {\it CN}{\it n})/{\blacksquare}{\exists}bNb){\&}({\it PP}{\it of}/{\exists}aNa)), {\square}(({\langle\rangle}{\exists}gNt(s(g))\backslash Sf)/{\exists}aNa), {\blacksquare}{\forall}a{\forall}f({[]^{-1}}(({\langle\rangle}Na\backslash Sf)\backslash ({\langle\rangle}Na\backslash Sf))/({\langle\rangle}Na\backslash Spsp)), {\square}(({\langle\rangle}{\exists}aNa\backslash Spsp)/{\exists}aNa)\ \Rightarrow\ ({\langle\rangle}Nt(s(n)){\sqcap}!{\blacksquare}Nt(s(n)))\backslash Sf
\using {\backslash}R
\endprooftree
\justifies
{\blacksquare}{\forall}n(Nt(n)/{\it CN}{\it n}), {\square}({\forall}g{\it CN}{\it s(g)}/{\it PP}{\it of}), {\square}(({\forall}n({\it CN}{\it n}\backslash {\it CN}{\it n})/{\blacksquare}{\exists}bNb){\&}({\it PP}{\it of}/{\exists}aNa)), {\square}(({\langle\rangle}{\exists}gNt(s(g))\backslash Sf)/{\exists}aNa), {\blacksquare}{\forall}a{\forall}f({[]^{-1}}(({\langle\rangle}Na\backslash Sf)\backslash ({\langle\rangle}Na\backslash Sf))/({\langle\rangle}Na\backslash Spsp)), {\square}(({\langle\rangle}{\exists}aNa\backslash Spsp)/{\exists}aNa)\ \Rightarrow\ {\blacksquare}(({\langle\rangle}Nt(s(n)){\sqcap}!{\blacksquare}Nt(s(n)))\backslash Sf)
\using {\blacksquare}R
\endprooftree
\prooftree
\prooftree
\prooftree
\prooftree
\prooftree
\justifies
\mbox{\fbox{${\it CN}{\it s(n)}$}}\ \Rightarrow\ {\it CN}{\it s(n)}
\endprooftree
\justifies
\mbox{\fbox{${\square}{\it CN}{\it s(n)}$}}\ \Rightarrow\ {\it CN}{\it s(n)}
\using {\Box}L
\endprooftree
\prooftree
\justifies
\mbox{\fbox{${\it CN}{\it s(n)}$}}\ \Rightarrow\ {\it CN}{\it s(n)}
\endprooftree
\justifies
{\square}{\it CN}{\it s(n)}, \mbox{\fbox{${\it CN}{\it s(n)}\backslash {\it CN}{\it s(n)}$}}\ \Rightarrow\ {\it CN}{\it s(n)}
\using {\backslash}L
\endprooftree
\justifies
{\square}{\it CN}{\it s(n)}, [\mbox{\fbox{${[]^{-1}}({\it CN}{\it s(n)}\backslash {\it CN}{\it s(n)})$}}]\ \Rightarrow\ {\it CN}{\it s(n)}
\using {[]^{-1}}L
\endprooftree
\justifies
{\square}{\it CN}{\it s(n)}, [[\mbox{\fbox{${[]^{-1}}{[]^{-1}}({\it CN}{\it s(n)}\backslash {\it CN}{\it s(n)})$}}]]\ \Rightarrow\ {\it CN}{\it s(n)}
\using {[]^{-1}}L
\endprooftree
\justifies
{\square}{\it CN}{\it s(n)}, [[\mbox{\fbox{${[]^{-1}}{[]^{-1}}({\it CN}{\it s(n)}\backslash {\it CN}{\it s(n)})/{\blacksquare}(({\langle\rangle}Nt(s(n)){\sqcap}!{\blacksquare}Nt(s(n)))\backslash Sf)$}}, {\blacksquare}{\forall}n(Nt(n)/{\it CN}{\it n}), {\square}({\forall}g{\it CN}{\it s(g)}/{\it PP}{\it of}), {\square}(({\forall}n({\it CN}{\it n}\backslash {\it CN}{\it n})/{\blacksquare}{\exists}bNb){\&}({\it PP}{\it of}/{\exists}aNa)), {\square}(({\langle\rangle}{\exists}gNt(s(g))\backslash Sf)/{\exists}aNa), {\blacksquare}{\forall}a{\forall}f({[]^{-1}}(({\langle\rangle}Na\backslash Sf)\backslash ({\langle\rangle}Na\backslash Sf))/({\langle\rangle}Na\backslash Spsp)), {\square}(({\langle\rangle}{\exists}aNa\backslash Spsp)/{\exists}aNa)]]\ \Rightarrow\ {\it CN}{\it s(n)}
\using {/}L
\endprooftree
\justifies
{\square}{\it CN}{\it s(n)}, [[\mbox{\fbox{${\forall}n({[]^{-1}}{[]^{-1}}({\it CN}{\it n}\backslash {\it CN}{\it n})/{\blacksquare}(({\langle\rangle}Nt(n){\sqcap}!{\blacksquare}Nt(n))\backslash Sf))$}}, {\blacksquare}{\forall}n(Nt(n)/{\it CN}{\it n}), {\square}({\forall}g{\it CN}{\it s(g)}/{\it PP}{\it of}), {\square}(({\forall}n({\it CN}{\it n}\backslash {\it CN}{\it n})/{\blacksquare}{\exists}bNb){\&}({\it PP}{\it of}/{\exists}aNa)), {\square}(({\langle\rangle}{\exists}gNt(s(g))\backslash Sf)/{\exists}aNa), {\blacksquare}{\forall}a{\forall}f({[]^{-1}}(({\langle\rangle}Na\backslash Sf)\backslash ({\langle\rangle}Na\backslash Sf))/({\langle\rangle}Na\backslash Spsp)), {\square}(({\langle\rangle}{\exists}aNa\backslash Spsp)/{\exists}aNa)]]\ \Rightarrow\ {\it CN}{\it s(n)}
\using {\forall}L
\endprooftree
\justifies
{\square}{\it CN}{\it s(n)}, [[\mbox{\fbox{${\blacksquare}{\forall}n({[]^{-1}}{[]^{-1}}({\it CN}{\it n}\backslash {\it CN}{\it n})/{\blacksquare}(({\langle\rangle}Nt(n){\sqcap}!{\blacksquare}Nt(n))\backslash Sf))$}}, {\blacksquare}{\forall}n(Nt(n)/{\it CN}{\it n}), {\square}({\forall}g{\it CN}{\it s(g)}/{\it PP}{\it of}), {\square}(({\forall}n({\it CN}{\it n}\backslash {\it CN}{\it n})/{\blacksquare}{\exists}bNb){\&}({\it PP}{\it of}/{\exists}aNa)), {\square}(({\langle\rangle}{\exists}gNt(s(g))\backslash Sf)/{\exists}aNa), {\blacksquare}{\forall}a{\forall}f({[]^{-1}}(({\langle\rangle}Na\backslash Sf)\backslash ({\langle\rangle}Na\backslash Sf))/({\langle\rangle}Na\backslash Spsp)), {\square}(({\langle\rangle}{\exists}aNa\backslash Spsp)/{\exists}aNa)]]\ \Rightarrow\ {\it CN}{\it s(n)}
\using {\blacksquare}L
\endprooftree}
\end{center}
\caption{Main derivation for \lingform{paper that the editor of filed without reading}}
\label{ptteofwr2}
\end{figure}

In our final relativisation example the host gap licences two parasitic gaps,
in the subject noun phrase and in an adverbial phrase:
\disp{
${\bf paper}{+}[[{\bf that}{+}{\bf the}{+}{\bf editor}{+}{\bf of}{+}{\bf filed}{+}{\bf without}{+}{\bf reading}]]: {\it CN}{\it s(n)}$}
Lexical lookup yields:
\disp{
${\square}{\it CN}{\it s(n)}: {\it paper}, [[{\blacksquare}{\forall}n({[]^{-1}}{[]^{-1}}({\it CN}{\it n}\backslash {\it CN}{\it n})/{\blacksquare}(({\langle\rangle}Nt(n){\sqcap}!{\blacksquare}Nt(n))\backslash Sf)):\\
\lambda A\lambda B\lambda C[({\it B}\ {\it C})\wedge ({\it A}\ {\it C})], {\blacksquare}{\forall}n(Nt(n)/{\it CN}{\it n}): \iota , {\square}({\forall}g{\it CN}{\it s(g)}/{\it PP}{\it of}): {\it editor},\\ {\square}(({\forall}n({\it CN}{\it n}\backslash {\it CN}{\it n})/{\blacksquare}{\exists}bNb){\&}({\it PP}{\it of}/{\exists}aNa)): \mbox{\^{}}(\mbox{\v{}}{\it of}, \lambda D{\it D}),\\
{\square}(({\langle\rangle}{\exists}gNt(s(g))\backslash Sf)/{\exists}aNa): \mbox{\^{}}\lambda E\lambda F({\it Past}\ ((\mbox{\v{}}{\it file}\ {\it E})\ {\it F})),\\
 {\blacksquare}{\forall}a{\forall}f({[]^{-1}}(({\langle\rangle}Na\backslash Sf)\backslash ({\langle\rangle}Na\backslash Sf))/({\langle\rangle}Na\backslash Spsp)): \lambda G\lambda H\lambda I[({\it H}\ {\it I})\wedge \neg ({\it G}\ {\it I})],\\
  {\square}(({\langle\rangle}{\exists}aNa\backslash Spsp)/{\exists}aNa): \mbox{\^{}}\lambda J\lambda K((\mbox{\v{}}{\it read}\ {\it J})\ {\it K})]]\ \Rightarrow\ {\it CN}{\it s(n)}$}
There is the derivation fragmented into Figures~\ref{ptteofwr1} and~\ref{ptteofwr2}.
There are two applications of contraction,
at depth nine and ten, 
projecting brackets around the subordinate subject and adverbial phrase
and giving rise to two parasitic gaps.
\begin{figure}
\begin{center}
\scriptsize
\prooftree
\prooftree
\prooftree
\prooftree
\prooftree
\prooftree
\prooftree
\prooftree
\justifies
\mbox{\fbox{$Nt(s(n))$}}\ \Rightarrow\ Nt(s(n))
\endprooftree
\justifies
\mbox{\fbox{${\blacksquare}Nt(s(n))$}}\ \Rightarrow\ Nt(s(n))
\using {\blacksquare}L
\endprooftree
\justifies
\mbox{\fbox{${\blacksquare}Nt(s(n))$}};\ \ \Rightarrow\ Nt(s(n))
\using {!}P
\endprooftree
\justifies
{\blacksquare}Nt(s(n));\ \ \Rightarrow\ \fbox{${\exists}aNa$}
\using {\exists}R
\endprooftree
\prooftree
\prooftree
\prooftree
\prooftree
\justifies
Nt(s(A))\ \Rightarrow\ Nt(s(A))
\endprooftree
\justifies
Nt(s(A))\ \Rightarrow\ \fbox{${\exists}aNa$}
\using {\exists}R
\endprooftree
\justifies
[Nt(s(A))]\ \Rightarrow\ \fbox{${\langle\rangle}{\exists}aNa$}
\using {\langle\rangle}R
\endprooftree
\prooftree
\justifies
\mbox{\fbox{$Spsp$}}\ \Rightarrow\ Spsp
\endprooftree
\justifies
[Nt(s(A))], \mbox{\fbox{${\langle\rangle}{\exists}aNa\backslash Spsp$}}\ \Rightarrow\ Spsp
\using {\backslash}L
\endprooftree
\justifies
{\blacksquare}Nt(s(n));\ [Nt(s(A))], \mbox{\fbox{$({\langle\rangle}{\exists}aNa\backslash Spsp)/{\exists}aNa$}}\ \Rightarrow\ Spsp
\using {/}L
\endprooftree
\justifies
{\blacksquare}Nt(s(n));\ [Nt(s(A))], \mbox{\fbox{${\square}(({\langle\rangle}{\exists}aNa\backslash Spsp)/{\exists}aNa)$}}\ \Rightarrow\ Spsp
\using {\Box}L
\endprooftree
\justifies
{\blacksquare}Nt(s(n));\ {\langle\rangle}Nt(s(A)), {\square}(({\langle\rangle}{\exists}aNa\backslash Spsp)/{\exists}aNa)\ \Rightarrow\ Spsp
\using {\langle\rangle}L
\endprooftree
\justifies
\begin{array}{c}
{\blacksquare}Nt(s(n));\ {\square}(({\langle\rangle}{\exists}aNa\backslash Spsp)/{\exists}aNa)\ \Rightarrow\ {\langle\rangle}Nt(s(A))\backslash Spsp\\
\mbox{\footnotesize\textcircled{1}}
\end{array}
\using {\backslash}R
\endprooftree

\prooftree
\prooftree
\prooftree
\prooftree
\prooftree
\prooftree
\prooftree
\justifies
\mbox{\fbox{$Nt(s(n))$}}\ \Rightarrow\ Nt(s(n))
\endprooftree
\justifies
\mbox{\fbox{${\blacksquare}Nt(s(n))$}}\ \Rightarrow\ Nt(s(n))
\using {\blacksquare}L
\endprooftree
\justifies
{\blacksquare}Nt(s(n))\ \Rightarrow\ \fbox{${\exists}aNa$}
\using {\exists}R
\endprooftree
\prooftree
\prooftree
\prooftree
\prooftree
\justifies
Nt(s(A))\ \Rightarrow\ Nt(s(A))
\endprooftree
\justifies
Nt(s(A))\ \Rightarrow\ \fbox{${\exists}gNt(s(g))$}
\using {\exists}R
\endprooftree
\justifies
[Nt(s(A))]\ \Rightarrow\ \fbox{${\langle\rangle}{\exists}gNt(s(g))$}
\using {\langle\rangle}R
\endprooftree
\prooftree
\justifies
\mbox{\fbox{$Sf$}}\ \Rightarrow\ Sf
\endprooftree
\justifies
[Nt(s(A))], \mbox{\fbox{${\langle\rangle}{\exists}gNt(s(g))\backslash Sf$}}\ \Rightarrow\ Sf
\using {\backslash}L
\endprooftree
\justifies
[Nt(s(A))], \mbox{\fbox{$({\langle\rangle}{\exists}gNt(s(g))\backslash Sf)/{\exists}aNa$}}, {\blacksquare}Nt(s(n))\ \Rightarrow\ Sf
\using {/}L
\endprooftree
\justifies
[Nt(s(A))], \mbox{\fbox{${\square}(({\langle\rangle}{\exists}gNt(s(g))\backslash Sf)/{\exists}aNa)$}}, {\blacksquare}Nt(s(n))\ \Rightarrow\ Sf
\using {\Box}L
\endprooftree
\justifies
{\langle\rangle}Nt(s(A)), {\square}(({\langle\rangle}{\exists}gNt(s(g))\backslash Sf)/{\exists}aNa), {\blacksquare}Nt(s(n))\ \Rightarrow\ Sf
\using {\langle\rangle}L
\endprooftree
\justifies
\begin{array}{c}
{\square}(({\langle\rangle}{\exists}gNt(s(g))\backslash Sf)/{\exists}aNa), {\blacksquare}Nt(s(n))\ \Rightarrow\ {\langle\rangle}Nt(s(A))\backslash Sf\\
\mbox{\footnotesize\textcircled{2}}
\end{array}
\using {\backslash}R
\endprooftree

\prooftree
\prooftree
\prooftree
\prooftree
\prooftree
\prooftree
\prooftree
\prooftree
\prooftree
\prooftree
\prooftree
\prooftree
\prooftree
\justifies
\mbox{\fbox{$Nt(s(n))$}}\ \Rightarrow\ Nt(s(n))
\endprooftree
\justifies
\mbox{\fbox{${\blacksquare}Nt(s(n))$}}\ \Rightarrow\ Nt(s(n))
\using {\blacksquare}L
\endprooftree
\justifies
{\blacksquare}Nt(s(n))\ \Rightarrow\ \fbox{${\exists}aNa$}
\using {\exists}R
\endprooftree
\prooftree
\justifies
\mbox{\fbox{${\it PP}{\it of}$}}\ \Rightarrow\ {\it PP}{\it of}
\endprooftree
\justifies
\mbox{\fbox{${\it PP}{\it of}/{\exists}aNa$}}, {\blacksquare}Nt(s(n))\ \Rightarrow\ {\it PP}{\it of}
\using {/}L
\endprooftree
\justifies
\mbox{\fbox{$({\forall}n({\it CN}{\it n}\backslash {\it CN}{\it n})/{\blacksquare}{\exists}bNb){\&}({\it PP}{\it of}/{\exists}aNa)$}}, {\blacksquare}Nt(s(n))\ \Rightarrow\ {\it PP}{\it of}
\using {\&}L
\endprooftree
\justifies
\mbox{\fbox{${\square}(({\forall}n({\it CN}{\it n}\backslash {\it CN}{\it n})/{\blacksquare}{\exists}bNb){\&}({\it PP}{\it of}/{\exists}aNa))$}}, {\blacksquare}Nt(s(n))\ \Rightarrow\ {\it PP}{\it of}
\using {\Box}L
\endprooftree
\prooftree
\prooftree
\justifies
\mbox{\fbox{${\it CN}{\it s(A)}$}}\ \Rightarrow\ {\it CN}{\it s(A)}
\endprooftree
\justifies
\mbox{\fbox{${\forall}g{\it CN}{\it s(g)}$}}\ \Rightarrow\ {\it CN}{\it s(A)}
\using {\forall}L
\endprooftree
\justifies
\mbox{\fbox{${\forall}g{\it CN}{\it s(g)}/{\it PP}{\it of}$}}, {\square}(({\forall}n({\it CN}{\it n}\backslash {\it CN}{\it n})/{\blacksquare}{\exists}bNb){\&}({\it PP}{\it of}/{\exists}aNa)), {\blacksquare}Nt(s(n))\ \Rightarrow\ {\it CN}{\it s(A)}
\using {/}L
\endprooftree
\justifies
\mbox{\fbox{${\square}({\forall}g{\it CN}{\it s(g)}/{\it PP}{\it of})$}}, {\square}(({\forall}n({\it CN}{\it n}\backslash {\it CN}{\it n})/{\blacksquare}{\exists}bNb){\&}({\it PP}{\it of}/{\exists}aNa)), {\blacksquare}Nt(s(n))\ \Rightarrow\ {\it CN}{\it s(A)}
\using {\Box}L
\endprooftree
\prooftree
\justifies
\mbox{\fbox{$Nt(s(A))$}}\ \Rightarrow\ Nt(s(A))
\endprooftree
\justifies
\mbox{\fbox{$Nt(s(A))/{\it CN}{\it s(A)}$}}, {\square}({\forall}g{\it CN}{\it s(g)}/{\it PP}{\it of}), {\square}(({\forall}n({\it CN}{\it n}\backslash {\it CN}{\it n})/{\blacksquare}{\exists}bNb){\&}({\it PP}{\it of}/{\exists}aNa)), {\blacksquare}Nt(s(n))\ \Rightarrow\ Nt(s(A))
\using {/}L
\endprooftree
\justifies
\mbox{\fbox{${\forall}n(Nt(n)/{\it CN}{\it n})$}}, {\square}({\forall}g{\it CN}{\it s(g)}/{\it PP}{\it of}), {\square}(({\forall}n({\it CN}{\it n}\backslash {\it CN}{\it n})/{\blacksquare}{\exists}bNb){\&}({\it PP}{\it of}/{\exists}aNa)), {\blacksquare}Nt(s(n))\ \Rightarrow\ Nt(s(A))
\using {\forall}L
\endprooftree
\justifies
\mbox{\fbox{${\blacksquare}{\forall}n(Nt(n)/{\it CN}{\it n})$}}, {\square}({\forall}g{\it CN}{\it s(g)}/{\it PP}{\it of}), {\square}(({\forall}n({\it CN}{\it n}\backslash {\it CN}{\it n})/{\blacksquare}{\exists}bNb){\&}({\it PP}{\it of}/{\exists}aNa)), {\blacksquare}Nt(s(n))\ \Rightarrow\ Nt(s(A))
\using {\blacksquare}L
\endprooftree
\justifies
[{\blacksquare}{\forall}n(Nt(n)/{\it CN}{\it n}), {\square}({\forall}g{\it CN}{\it s(g)}/{\it PP}{\it of}), {\square}(({\forall}n({\it CN}{\it n}\backslash {\it CN}{\it n})/{\blacksquare}{\exists}bNb){\&}({\it PP}{\it of}/{\exists}aNa)), {\blacksquare}Nt(s(n))]\ \Rightarrow\ \fbox{${\langle\rangle}Nt(s(A))$}
\using {\langle\rangle}R
\endprooftree
\prooftree
\justifies
\mbox{\fbox{$Sf$}}\ \Rightarrow\ Sf
\endprooftree
\justifies
\begin{array}{c}
{}[{\blacksquare}{\forall}n(Nt(n)/{\it CN}{\it n}), {\square}({\forall}g{\it CN}{\it s(g)}/{\it PP}{\it of}), {\square}(({\forall}n({\it CN}{\it n}\backslash {\it CN}{\it n})/{\blacksquare}{\exists}bNb){\&}({\it PP}{\it of}/{\exists}aNa)), {\blacksquare}Nt(s(n))], \mbox{\fbox{${\langle\rangle}Nt(s(A))\backslash Sf$}}\ \Rightarrow\ Sf\\
\mbox{\footnotesize\textcircled{3}}
\end{array}
\using {\backslash}L
\endprooftree
\end{center}
\caption{Auxiliary derivations for \lingform{paper that the editor of filed without reading}},
\label{ptteofwr1}
\end{figure}
This delivers the correct semantics:
\disp{
$\lambda C[(\mbox{\v{}}{\it paper}\ {\it C})\wedge [({\it Past}\ ((\mbox{\v{}}{\it file}\ {\it C})\ (\iota \ (\mbox{\v{}}{\it editor}\ {\it C}))))\wedge \neg ((\mbox{\v{}}{\it read}\ {\it C})\ (\iota \ (\mbox{\v{}}{\it editor}\ {\it C})))]]$}

\section{Apparent exceptions}

\label{appexcept}

In this section we address three kinds of apparent exceptions to the account
of relativisation given here.

First, there are examples in which there appears to be a parasitic gap
which is not in an island. 
The following is example~(8a) from Postal (1993\cite{postal:93}):
\disp{
man who$_i$ Mary convinced $t_i$ that John wanted to visit $t_i$
\label{convince}}
And an anonymous L\&P referee pointed out:
\disp{
people whom$_i$ you sent pictures of $t_i$ to $t_i$
\label{pict}}
In respect of such examples we suggest that although there \emph{seems\/} to be
no island, there \emph{could\/} be one.\footnote{Tom Roeper, p.c.}
This is effected as follows for~(\ref{convince}). 
Instead of a type of the form $((N\bsl S)/\CP)/N$ for \lingform{convince\/}
we assume  $((N\bsl S)/\CP)/(N\iadisj\mybrack N)$ where the semantically inactive
additive disjunction disjunct $N$ will be selected ordinarily, and $\mybrack N$ when
there is parasitic extraction, as in~(\ref{convince}). 
Similarly for~(\ref{pict}) we assume for \lingform{picture} type $\CN/(\PP\iadisj\mybrack\PP)$
where the second disjunct projects the brackets of a weak island.\footnote{The argument
pattern $X\iadisj\mybrack X$ is a general mechanism for an argument optional
island $X$. 
Likewise the dual value pattern $X\iaconj\abrack X$ is a general mechanism
for a value optional island $X$.}
Thus in examples such as the following the semantically inactive additive disjunction
inference for \lingform{convince\/}
of type  $((N\bsl S)/$ $\CP)/(N\iadisj\mybrack N)$ will select $N$:
\disp{
\begin{tabular}[t]{ll}
a. & man who$_i$ Mary convinced $t_i$ that John wanted to visit Suzy\\
b. & man who$_i$ Mary convinced the friends of $t_i$ that John wanted to visit Suzy
\end{tabular}}
But for (\ref{convince})  the semantically inactive additive disjunction
inference for \lingform{convince\/}
of type  $((N\bsl S)/\CP)/$ $(N\iadisj\mybrack N)$ will select $\mybrack N$.
Similarly for the picture noun case (\ref{pict}).
This account is not explanatory, but it captures the facts.

Second, 
under certain processing conditions island violations are grammatical
(Lakoff 1986\cite{lakoff:csc};
Deane 1991\cite{deane:91}; Kluender 1992\cite{kluender:92}, 1998\cite{kluender:98};
Kehler 2002\cite{kehler:02};
Hofmeister and Sag 2010\cite{hofsag:10}).\footnote{Kubota
and Levine (2015\cite{kl:lap}, Section 4.6.2) argue for a version
of TLG that freely overgenerates island constraint `violations'
in the syntax.} We cannot undertake to offer a full account here of which processing
conditions these are nor how the processing and combinatoric modules
interact to produce this effect, but we do suggest how the weak island violation is combinatorially
possible without changing the grammar. This is to assume a variant
$\%\univexp P$ of $\univexp P$ as follows applicable under the right
processing conditions:
\disp{$
 \prooftree
 \Xi(\zeta; \Gamma_1, [\{A\ass x\}; \Gamma_2], \Gamma_3)\yields B\ass\psi
 \justifies
 \Xi(\zeta\mun\{A\ass x\}; \Gamma_1, \Gamma_2, \Gamma_3)\yields B\ass\psi
 \using \%\univexp P
 \endprooftree
$\label{partial}}
This allows extraction from weak islands, thus for example:
\disp{
\prooftree
\prooftree
\prooftree
\ldots, [\ldots, A, \ldots], \ldots\yields B
\justifies
\ldots, [A; \ldots, \ldots], \ldots\yields B
\using \univexp P
\endprooftree
\justifies
A; \ldots, \ldots, \ldots, \ldots\yields B
\using \%\univexp P
\endprooftree
\justifies
\univexp A, \ldots, \ldots, \ldots, \ldots\yields B
\using \univexp L
\endprooftree
}

Thirdly, Levine and Hukari (2006\cite{levhuk:06}) cite an apparent example of `symbiotic'
extraction without a host gap:
\disp{
people that$_i$ fans of $t_i$ gather from every continent
just to listen to $t_i$
\label{fans}}
It is interesting that our response to the second issue predicts a possibility
such as~(\ref{fans}). An analysis can begin with the hypothetical
subtype in the stoup emitting a parasitic gap by $\univexp C$ in, 
say, 
the subject in the usual way.
Then $\%\univexp P$ can move the hypothetical subtype in the outer stoup into the
stoup of the adjunt island.
The two gap types in the two island stoups are then moved into their positions
by $\univexp P$'s.

\commentout{

\chapter{Coordination}

\label{coordchap}

\label{coordchap}

In this chapter we analyse examples of coordination,
cf.~Morrill~(2011\cite{morrill:oxford}, Chapter~3, Section~10).

\section{Constituent and `non-constituent'  coordination}

\label{crd:1}

To express the lexical semantics of coordination, including iterated coordination
(e.g.~\lingform{Bill, Mary,} \lingform{Suzy and Fred})
and various arities (zeroary e.g.~sentence,
unary e.g.~verb phrase, binary e.g.~transitive verb, \ldots), we use combinators: a non-empty list
map apply $\alphaplus$ and a non-empty list map $\mathbf{\Phi^n}$ combinator $\Phinplus$. 

The
combinator $\mathbf{\Phi}$ is such that $\mathbf{\Phi}\ x\ y\ z\ w= x\ (y\ w)\ (z\ w)$ (Curry and Feys
1958\cite{curryfeys}).
The non-empty list
map apply combinator $\alphaplus$ is as follows:
\disp{
$
\begin{array}[t]{rcl}
(\alphaplus\ [x]\ y) & = & [(x\ y)]\\
(\alphaplus\ [x, y| z]\ w) & = & [(x\ w)|(\alphaplus\ [y|z]\ w)]
\end{array}
$}
The non-empty list map $\mathbf{\Phi^n}$ combinator $\Phinplus$ is thus:
\disp{
$
\begin{array}[t]{rcl}
(((\Phinplus\ 0\ {\it and})\ x)\ [y]) & = & [y\wedge x]\\
(((\Phinplus\ 0\ {\it or})\ x)\ [y]) & = & [y\vee x]\\
(((\Phinplus\ 0\ {\it and})\ x)\ [y, z|w]) & = & [y\wedge(((\Phinplus\ 0\ {\it and})\ x)\ [z|w])]\\
(((\Phinplus\ 0\ {\it or})\ x)\ [y, z|w]) & = & [y\vee(((\Phinplus\ 0\ {\it or})\ x)\ [z|w])]\\
((((\Phinplus\ (s\ n)\ c)\ x)\ y)\ z) & = & (((\Phinplus\ n\ c)\ (x\ z))\ (\alphaplus\ y\ z))
\end{array}
$}
These equations mean that in semantic evaluation any subterm of the form
on the left is to be replaced by that on the right, successively.

\subsection{Sentence coordination}

The first example is simple sentential conjunction; as we have said subjects
are bracketed since they are (weak) islands and
coordinate structures are doubly bracketed corresponding to the fact that they are
strong islands; these brackets are given in the input:
\disp{
$[[[{\bf john}]{+}{\bf praises}{+}{\bf mary}{+}{\bf and}{+}[{\bf john}]{+}{\bf laughs}]]: Sf$}
Lexical lookup yields  the following where the coordinator type is
essentially $(\exstexp X\bsl\abrack\abrack X)/X$ with $X=S$.
\disp{
$\begin{array}[t]{l}
[[[{\blacksquare}Nt(s(m)): {\it j}], {\square}(({\langle\rangle}{\exists}gNt(s(g))\backslash Sf)/{\exists}aNa): \mbox{\^{}}\lambda A\lambda B({\it Pres}\ ((\mbox{\v{}}{\it praise}\ {\it A})\ {\it B})),\\ {\blacksquare}Nt(s(f)): {\it m},
 {\blacksquare}{\forall}f((?{\blacksquare}Sf\backslash {[]^{-1}}{[]^{-1}}Sf)/{\blacksquare}Sf): (\Phinplus\ {\it 0}\ {\it and}), [{\blacksquare}Nt(s(m)): {\it j}],\\ {\square}({\langle\rangle}{\exists}gNt(s(g))\backslash Sf): \mbox{\^{}}\lambda C({\it Pres}\ (\mbox{\v{}}{\it laugh}\ {\it C}))]]\ \Rightarrow\ Sf
 \end{array}$}
The left conjunt is marked with the existential exponential to allow iterated coordination,
which we will illustrate later;
the conjunts are marked with semantically inactive normal modalities to make coordinate
structures scope islands to quantifiers other than indefinites.
There is the derivation:

\vspace{0.15in}

\begin{center}
\noindent\rotatebox{-90}{\scriptsize
\prooftree
\prooftree
\prooftree
\prooftree
\prooftree
\prooftree
\prooftree
\prooftree
\prooftree
\prooftree
\justifies
\mbox{\fbox{$Nt(s(m))$}}\ \Rightarrow\ Nt(s(m))
\endprooftree
\justifies
\mbox{\fbox{${\blacksquare}Nt(s(m))$}}\ \Rightarrow\ Nt(s(m))
\using {\blacksquare}L
\endprooftree
\justifies
{\blacksquare}Nt(s(m))\ \Rightarrow\ \fbox{${\exists}gNt(s(g))$}
\using {\exists}R
\endprooftree
\justifies
[{\blacksquare}Nt(s(m))]\ \Rightarrow\ \fbox{${\langle\rangle}{\exists}gNt(s(g))$}
\using {\langle\rangle}R
\endprooftree
\prooftree
\justifies
\mbox{\fbox{$Sf$}}\ \Rightarrow\ Sf
\endprooftree
\justifies
[{\blacksquare}Nt(s(m))], \mbox{\fbox{${\langle\rangle}{\exists}gNt(s(g))\backslash Sf$}}\ \Rightarrow\ Sf
\using {\backslash}L
\endprooftree
\justifies
[{\blacksquare}Nt(s(m))], \mbox{\fbox{${\square}({\langle\rangle}{\exists}gNt(s(g))\backslash Sf)$}}\ \Rightarrow\ Sf
\using {\Box}L
\endprooftree
\justifies
\begin{array}{c}
[{\blacksquare}Nt(s(m))], {\square}({\langle\rangle}{\exists}gNt(s(g))\backslash Sf)\ \Rightarrow\ {\blacksquare}Sf\\
\mbox{\footnotesize\textcircled{1}}
\end{array}
\using {\blacksquare}R
\endprooftree
\prooftree
\prooftree
\prooftree
\prooftree
\prooftree
\prooftree
\prooftree
\prooftree
\justifies
\mbox{\fbox{$Nt(s(f))$}}\ \Rightarrow\ Nt(s(f))
\endprooftree
\justifies
\mbox{\fbox{${\blacksquare}Nt(s(f))$}}\ \Rightarrow\ Nt(s(f))
\using {\blacksquare}L
\endprooftree
\justifies
{\blacksquare}Nt(s(f))\ \Rightarrow\ \fbox{${\exists}aNa$}
\using {\exists}R
\endprooftree
\prooftree
\prooftree
\prooftree
\prooftree
\prooftree
\justifies
\mbox{\fbox{$Nt(s(m))$}}\ \Rightarrow\ Nt(s(m))
\endprooftree
\justifies
\mbox{\fbox{${\blacksquare}Nt(s(m))$}}\ \Rightarrow\ Nt(s(m))
\using {\blacksquare}L
\endprooftree
\justifies
{\blacksquare}Nt(s(m))\ \Rightarrow\ \fbox{${\exists}gNt(s(g))$}
\using {\exists}R
\endprooftree
\justifies
[{\blacksquare}Nt(s(m))]\ \Rightarrow\ \fbox{${\langle\rangle}{\exists}gNt(s(g))$}
\using {\langle\rangle}R
\endprooftree
\prooftree
\justifies
\mbox{\fbox{$Sf$}}\ \Rightarrow\ Sf
\endprooftree
\justifies
[{\blacksquare}Nt(s(m))], \mbox{\fbox{${\langle\rangle}{\exists}gNt(s(g))\backslash Sf$}}\ \Rightarrow\ Sf
\using {\backslash}L
\endprooftree
\justifies
[{\blacksquare}Nt(s(m))], \mbox{\fbox{$({\langle\rangle}{\exists}gNt(s(g))\backslash Sf)/{\exists}aNa$}}, {\blacksquare}Nt(s(f))\ \Rightarrow\ Sf
\using {/}L
\endprooftree
\justifies
[{\blacksquare}Nt(s(m))], \mbox{\fbox{${\square}(({\langle\rangle}{\exists}gNt(s(g))\backslash Sf)/{\exists}aNa)$}}, {\blacksquare}Nt(s(f))\ \Rightarrow\ Sf
\using {\Box}L
\endprooftree
\justifies
[{\blacksquare}Nt(s(m))], {\square}(({\langle\rangle}{\exists}gNt(s(g))\backslash Sf)/{\exists}aNa), {\blacksquare}Nt(s(f))\ \Rightarrow\ {\blacksquare}Sf
\using {\blacksquare}R
\endprooftree
\justifies
\begin{array}{c}
[{\blacksquare}Nt(s(m))], {\square}(({\langle\rangle}{\exists}gNt(s(g))\backslash Sf)/{\exists}aNa), {\blacksquare}Nt(s(f))\ \Rightarrow\ \fbox{$?{\blacksquare}Sf$}\\
\mbox{\footnotesize\textcircled{2}}
\end{array}
\using {?}R
\endprooftree
\prooftree
\prooftree
\prooftree
\justifies
\mbox{\fbox{$Sf$}}\ \Rightarrow\ Sf
\endprooftree
\justifies
[\mbox{\fbox{${[]^{-1}}Sf$}}]\ \Rightarrow\ Sf
\using {[]^{-1}}L
\endprooftree
\justifies
\begin{array}{c}
{}[[\mbox{\fbox{${[]^{-1}}{[]^{-1}}Sf$}}]]\ \Rightarrow\ Sf\\
\mbox{\footnotesize\textcircled{3}}
\end{array}
\using {[]^{-1}}L
\endprooftree
\justifies
[[[{\blacksquare}Nt(s(m))], {\square}(({\langle\rangle}{\exists}gNt(s(g))\backslash Sf)/{\exists}aNa), {\blacksquare}Nt(s(f)), \mbox{\fbox{$?{\blacksquare}Sf\backslash {[]^{-1}}{[]^{-1}}Sf$}}]]\ \Rightarrow\ Sf
\using {\backslash}L
\endprooftree
\justifies
[[[{\blacksquare}Nt(s(m))], {\square}(({\langle\rangle}{\exists}gNt(s(g))\backslash Sf)/{\exists}aNa), {\blacksquare}Nt(s(f)), \mbox{\fbox{$(?{\blacksquare}Sf\backslash {[]^{-1}}{[]^{-1}}Sf)/{\blacksquare}Sf$}}, [{\blacksquare}Nt(s(m))], {\square}({\langle\rangle}{\exists}gNt(s(g))\backslash Sf)]]\ \Rightarrow\ Sf
\using {/}L
\endprooftree
\justifies
[[[{\blacksquare}Nt(s(m))], {\square}(({\langle\rangle}{\exists}gNt(s(g))\backslash Sf)/{\exists}aNa), {\blacksquare}Nt(s(f)), \mbox{\fbox{${\forall}f((?{\blacksquare}Sf\backslash {[]^{-1}}{[]^{-1}}Sf)/{\blacksquare}Sf)$}}, [{\blacksquare}Nt(s(m))], {\square}({\langle\rangle}{\exists}gNt(s(g))\backslash Sf)]]\ \Rightarrow\ Sf
\using {\forall}L
\endprooftree
\justifies
[[[{\blacksquare}Nt(s(m))], {\square}(({\langle\rangle}{\exists}gNt(s(g))\backslash Sf)/{\exists}aNa), {\blacksquare}Nt(s(f)), \mbox{\fbox{${\blacksquare}{\forall}f((?{\blacksquare}Sf\backslash {[]^{-1}}{[]^{-1}}Sf)/{\blacksquare}Sf)$}}, [{\blacksquare}Nt(s(m))], {\square}({\langle\rangle}{\exists}gNt(s(g))\backslash Sf)]]\ \Rightarrow\ Sf
\using {\blacksquare}L
\endprooftree}
\end{center}

\vspace{0.15in}

\noindent
In the conclusions of rules boxes mark the active type of the rule application, i.e.~the
type which is decomposed in the premises; as we have seen
the boxes mark \techterm{focused\/} types (Andreoli 1992\cite{andreoli:92}), 
which are the active types in
the conclusions of noninvertible rules, subject to focusing rule application;
the active types in conclusions of invertible rules are not marked; 
whenever a type is boxed it is the active type
--- this improves readability.
Reading from the root, the type of the coordinator projecting the construction
is decomposed, eliminating the semantically inactive outermost modality
and instantiating the tense feature to finite ($f$). 
The result then applies forwards to the righthand conjunct,
the analysis of which is shown in the subtree marked \textcircled{1}:
after removal of the semantically inactive modality in the succedent
(which is licensed by the fact that the antecedents are modalised),
left box elimination is applied to the intransitive verb;
this then applies to its subject and in the minor premise the subject
bracket modality is removed, then the agreement instantiated to
third person singular masculine ($t(s(m))$), and then the semantically inactive subject
modality is removed and the axiom is matched.
The coordinator then applies to the lefthand conjunct,
in the subtree marked \textcircled{2}:
the existential exponential is removed directly from the succedent, since
the coordination is not iterative, and then the semantically inactive succedent
modality is eliminated (as required the antecedents are modalised), and the outer
modality of the transitive verb is removed;
the transitive verb then applies to its object, where the agreement is instantiated to
third person singular feminine ($t(s(f))$), and the analysis of the resulting
verb phrase is the same as in \textcircled{1}.
Finally, in the subderivation \textcircled{3} the coordinator checks off
the doubly bracketed context which it requires/projects.
All this delivers the correct reading under:
\disp{
$[({\it Pres}\ ((\mbox{\v{}}{\it praise}\ {\it m})\ {\it j}))\wedge ({\it Pres}\ (\mbox{\v{}}{\it laugh}\ {\it j}))]$}

\subsection{Verb phrase coordination}

The next example is one of verb phrase conjunction:
\disp{
$[{\bf john}]{+}[[{\bf praises}{+}{\bf mary}{+}{\bf and}{+}{\bf laughs}]]: Sf$}
Lexical lookup yields the coordinator type
basically $(\exstexp X\bsl\abrack\abrack X)/X$ with $X=N\bsl S$.
\disp{
$
\begin{array}[t]{l}
[{\blacksquare}Nt(s(m)): {\it j}], [[{\square}(({\langle\rangle}{\exists}gNt(s(g))\backslash Sf)/{\exists}aNa): \mbox{\^{}}\lambda A\lambda B({\it Pres}\ ((\mbox{\v{}}{\it praise}\ {\it A})\ {\it B})),\\ {\blacksquare}Nt(s(f)): {\it m},
 {\blacksquare}{\forall}a{\forall}f((?{\blacksquare}({\langle\rangle}Na\backslash Sf)\backslash {[]^{-1}}{[]^{-1}}({\langle\rangle}Na\backslash Sf))/{\blacksquare}({\langle\rangle}Na\backslash Sf)):\\ (\Phinplus\ ({\it s}\ {\it 0})\ {\it and}), {\square}({\langle\rangle}{\exists}gNt(s(g))\backslash Sf): \mbox{\^{}}\lambda C({\it Pres}\ (\mbox{\v{}}{\it laugh}\ {\it C}))]]\ \Rightarrow\ Sf
 \end{array}$}
The coordination combinator semantics
$(\Phinplus\ (s\ 0)\ {\it and})$ is such that:
\disp{$
\begin{array}[t]{l}
((((\Phinplus\ (s\ 0)\ {\it and})\ x)\ [y])\ z) =\\
(((\Phinplus\ 0\ {\it and})\ (x\ z))\ (\alphaplus\ [y]\ z)) =\\
(((\Phinplus\ 0\ {\it and})\ (x\ z))\ [(y\ z)]) =\\
{}[(y\ z)\wedge(x\ z)]
\end{array}$}
There is the derivation:

\vspace{0.15in}

\begin{center}
\rotatebox{-90}{\scriptsize
\prooftree
\prooftree
\prooftree
\prooftree
\prooftree
\prooftree
\prooftree
\prooftree
\prooftree
\prooftree
\prooftree
\prooftree
\justifies
Nt(s(m))\ \Rightarrow\ Nt(s(m))
\endprooftree
\justifies
Nt(s(m))\ \Rightarrow\ \fbox{${\exists}gNt(s(g))$}
\using {\exists}R
\endprooftree
\justifies
[Nt(s(m))]\ \Rightarrow\ \fbox{${\langle\rangle}{\exists}gNt(s(g))$}
\using {\langle\rangle}R
\endprooftree
\prooftree
\justifies
\mbox{\fbox{$Sf$}}\ \Rightarrow\ Sf
\endprooftree
\justifies
[Nt(s(m))], \mbox{\fbox{${\langle\rangle}{\exists}gNt(s(g))\backslash Sf$}}\ \Rightarrow\ Sf
\using {\backslash}L
\endprooftree
\justifies
[Nt(s(m))], \mbox{\fbox{${\square}({\langle\rangle}{\exists}gNt(s(g))\backslash Sf)$}}\ \Rightarrow\ Sf
\using {\Box}L
\endprooftree
\justifies
{\langle\rangle}Nt(s(m)), {\square}({\langle\rangle}{\exists}gNt(s(g))\backslash Sf)\ \Rightarrow\ Sf
\using {\langle\rangle}L
\endprooftree
\justifies
{\square}({\langle\rangle}{\exists}gNt(s(g))\backslash Sf)\ \Rightarrow\ {\langle\rangle}Nt(s(m))\backslash Sf
\using {\backslash}R
\endprooftree
\justifies
\begin{array}{c}
{\square}({\langle\rangle}{\exists}gNt(s(g))\backslash Sf)\ \Rightarrow\ {\blacksquare}({\langle\rangle}Nt(s(m))\backslash Sf)\\
\mbox{\footnotesize\textcircled{1}}
\end{array}
\using {\blacksquare}R
\endprooftree
\prooftree
\prooftree
\prooftree
\prooftree
\prooftree
\prooftree
\prooftree
\prooftree
\prooftree
\prooftree
\justifies
\mbox{\fbox{$Nt(s(f))$}}\ \Rightarrow\ Nt(s(f))
\endprooftree
\justifies
\mbox{\fbox{${\blacksquare}Nt(s(f))$}}\ \Rightarrow\ Nt(s(f))
\using {\blacksquare}L
\endprooftree
\justifies
{\blacksquare}Nt(s(f))\ \Rightarrow\ \fbox{${\exists}aNa$}
\using {\exists}R
\endprooftree
\prooftree
\prooftree
\prooftree
\prooftree
\justifies
Nt(s(m))\ \Rightarrow\ Nt(s(m))
\endprooftree
\justifies
Nt(s(m))\ \Rightarrow\ \fbox{${\exists}gNt(s(g))$}
\using {\exists}R
\endprooftree
\justifies
[Nt(s(m))]\ \Rightarrow\ \fbox{${\langle\rangle}{\exists}gNt(s(g))$}
\using {\langle\rangle}R
\endprooftree
\prooftree
\justifies
\mbox{\fbox{$Sf$}}\ \Rightarrow\ Sf
\endprooftree
\justifies
[Nt(s(m))], \mbox{\fbox{${\langle\rangle}{\exists}gNt(s(g))\backslash Sf$}}\ \Rightarrow\ Sf
\using {\backslash}L
\endprooftree
\justifies
[Nt(s(m))], \mbox{\fbox{$({\langle\rangle}{\exists}gNt(s(g))\backslash Sf)/{\exists}aNa$}}, {\blacksquare}Nt(s(f))\ \Rightarrow\ Sf
\using {/}L
\endprooftree
\justifies
[Nt(s(m))], \mbox{\fbox{${\square}(({\langle\rangle}{\exists}gNt(s(g))\backslash Sf)/{\exists}aNa)$}}, {\blacksquare}Nt(s(f))\ \Rightarrow\ Sf
\using {\Box}L
\endprooftree
\justifies
{\langle\rangle}Nt(s(m)), {\square}(({\langle\rangle}{\exists}gNt(s(g))\backslash Sf)/{\exists}aNa), {\blacksquare}Nt(s(f))\ \Rightarrow\ Sf
\using {\langle\rangle}L
\endprooftree
\justifies
{\square}(({\langle\rangle}{\exists}gNt(s(g))\backslash Sf)/{\exists}aNa), {\blacksquare}Nt(s(f))\ \Rightarrow\ {\langle\rangle}Nt(s(m))\backslash Sf
\using {\backslash}R
\endprooftree
\justifies
{\square}(({\langle\rangle}{\exists}gNt(s(g))\backslash Sf)/{\exists}aNa), {\blacksquare}Nt(s(f))\ \Rightarrow\ {\blacksquare}({\langle\rangle}Nt(s(m))\backslash Sf)
\using {\blacksquare}R
\endprooftree
\justifies
\begin{array}{c}
{\square}(({\langle\rangle}{\exists}gNt(s(g))\backslash Sf)/{\exists}aNa), {\blacksquare}Nt(s(f))\ \Rightarrow\ \fbox{$?{\blacksquare}({\langle\rangle}Nt(s(m))\backslash Sf)$}\\
\mbox{\footnotesize\textcircled{2}}
\end{array}
\using {?}R
\endprooftree
\prooftree
\prooftree
\prooftree
\prooftree
\prooftree
\prooftree
\justifies
\mbox{\fbox{$Nt(s(m))$}}\ \Rightarrow\ Nt(s(m))
\endprooftree
\justifies
\mbox{\fbox{${\blacksquare}Nt(s(m))$}}\ \Rightarrow\ Nt(s(m))
\using {\blacksquare}L
\endprooftree
\justifies
[{\blacksquare}Nt(s(m))]\ \Rightarrow\ \fbox{${\langle\rangle}Nt(s(m))$}
\using {\langle\rangle}R
\endprooftree
\prooftree
\justifies
\mbox{\fbox{$Sf$}}\ \Rightarrow\ Sf
\endprooftree
\justifies
[{\blacksquare}Nt(s(m))], \mbox{\fbox{${\langle\rangle}Nt(s(m))\backslash Sf$}}\ \Rightarrow\ Sf
\using {\backslash}L
\endprooftree
\justifies
[{\blacksquare}Nt(s(m))], [\mbox{\fbox{${[]^{-1}}({\langle\rangle}Nt(s(m))\backslash Sf)$}}]\ \Rightarrow\ Sf
\using {[]^{-1}}L
\endprooftree
\justifies
\begin{array}{c}
{}[{\blacksquare}Nt(s(m))], [[\mbox{\fbox{${[]^{-1}}{[]^{-1}}({\langle\rangle}Nt(s(m))\backslash Sf)$}}]]\ \Rightarrow\ Sf\\
\mbox{\footnotesize\textcircled{3}}
\end{array}
\using {[]^{-1}}L
\endprooftree
\justifies
[{\blacksquare}Nt(s(m))], [[{\square}(({\langle\rangle}{\exists}gNt(s(g))\backslash Sf)/{\exists}aNa), {\blacksquare}Nt(s(f)), \mbox{\fbox{$?{\blacksquare}({\langle\rangle}Nt(s(m))\backslash Sf)\backslash {[]^{-1}}{[]^{-1}}({\langle\rangle}Nt(s(m))\backslash Sf)$}}]]\ \Rightarrow\ Sf
\using {\backslash}L
\endprooftree
\justifies
[{\blacksquare}Nt(s(m))], [[{\square}(({\langle\rangle}{\exists}gNt(s(g))\backslash Sf)/{\exists}aNa), {\blacksquare}Nt(s(f)), \mbox{\fbox{$(?{\blacksquare}({\langle\rangle}Nt(s(m))\backslash Sf)\backslash {[]^{-1}}{[]^{-1}}({\langle\rangle}Nt(s(m))\backslash Sf))/{\blacksquare}({\langle\rangle}Nt(s(m))\backslash Sf)$}}, {\square}({\langle\rangle}{\exists}gNt(s(g))\backslash Sf)]]\ \Rightarrow\ Sf
\using {/}L
\endprooftree
\justifies
[{\blacksquare}Nt(s(m))], [[{\square}(({\langle\rangle}{\exists}gNt(s(g))\backslash Sf)/{\exists}aNa), {\blacksquare}Nt(s(f)), \mbox{\fbox{${\forall}f((?{\blacksquare}({\langle\rangle}Nt(s(m))\backslash Sf)\backslash {[]^{-1}}{[]^{-1}}({\langle\rangle}Nt(s(m))\backslash Sf))/{\blacksquare}({\langle\rangle}Nt(s(m))\backslash Sf))$}}, {\square}({\langle\rangle}{\exists}gNt(s(g))\backslash Sf)]]\ \Rightarrow\ Sf
\using {\forall}L
\endprooftree
\justifies
[{\blacksquare}Nt(s(m))], [[{\square}(({\langle\rangle}{\exists}gNt(s(g))\backslash Sf)/{\exists}aNa), {\blacksquare}Nt(s(f)), \mbox{\fbox{${\forall}a{\forall}f((?{\blacksquare}({\langle\rangle}Na\backslash Sf)\backslash {[]^{-1}}{[]^{-1}}({\langle\rangle}Na\backslash Sf))/{\blacksquare}({\langle\rangle}Na\backslash Sf))$}}, {\square}({\langle\rangle}{\exists}gNt(s(g))\backslash Sf)]]\ \Rightarrow\ Sf
\using {\forall}L
\endprooftree
\justifies
[{\blacksquare}Nt(s(m))], [[{\square}(({\langle\rangle}{\exists}gNt(s(g))\backslash Sf)/{\exists}aNa), {\blacksquare}Nt(s(f)), \mbox{\fbox{${\blacksquare}{\forall}a{\forall}f((?{\blacksquare}({\langle\rangle}Na\backslash Sf)\backslash {[]^{-1}}{[]^{-1}}({\langle\rangle}Na\backslash Sf))/{\blacksquare}({\langle\rangle}Na\backslash Sf))$}}, {\square}({\langle\rangle}{\exists}gNt(s(g))\backslash Sf)]]\ \Rightarrow\ Sf
\using {\blacksquare}L
\endprooftree}
\end{center}

\vspace{0.15in}

\noindent
Reading from the root, first the outermost semantically inactive modality of the coordinator
is removed and the agreement features of the subject, $t(s(m))$, and the value of the tense
feature, $f$, are instantiated. 
The coordinator then applies to the righthand conjunct, for which the subderivation
is \textcircled{1}: the succedent semantically inactive modality is removed (the antecedent
is modalised) and the argument modally bracketed nominal is lowered into the
antecedent where its bracket modality is unfolded; the modality of the intransitive verb is removed,
and the subderivation finishes applying to the subject with bracket modality right
and existential quantifier right rules.
The coordinator then applies to the transitive verb plus object lefthand conjunct
in the subderivation rooted at \textcircled{2}: after existential exponential right,
semantically inactive modality right (the antecedents are modalised), and under right
lowering the subject subtype into the antecedent, the bracket modality of the subject is
unfolded; the transitive verb type is then selected, its modality removed,
and applied with modality rules and instantiation of existentially quantified
features to the object (left subsubderivation) and subject (middle subsubderivation).
In the subderivation rooted at \textcircled{3} the context of the coordinate structure
is recognised: double bracketing, and a (bracketed) proper name subject.
All this delivers semantics:
\disp{
$[({\it Pres}\ ((\mbox{\v{}}{\it praise}\ {\it m})\ {\it j}))\wedge ({\it Pres}\ (\mbox{\v{}}{\it laugh}\ {\it j}))]$}

\subsection{Transitive verb coordination}

The next example is of transitive verb coordination, with a complex non-standard
constituent transitive verb in the right-hand conjunt.
\disp{
$[{\bf john}]{+}[[{\bf likes}{+}{\bf and}{+}{\bf will}{+}{\bf love}]]{+}{\bf london}: Sf$}
Appropriate lexical lookup yields the following where the coordinator type is
essentially $(\exstexp X\bsl$ $\abrack\abrack X)/X$ with $X=(N\bsl S)/N$.
\disp{
$\begin{array}[t]{l}
[{\blacksquare}Nt(s(m)): {\it j}], [[{\square}(({\langle\rangle}{\exists}gNt(s(g))\backslash Sf)/{\exists}aNa): \mbox{\^{}}\lambda A\lambda B({\it Pres}\ ((\mbox{\v{}}{\it like}\ {\it A})\ {\it B})),\\
 {\blacksquare}{\forall}f{\forall}a((?{\blacksquare}(({\langle\rangle}Na\backslash Sf)/{\exists}bNb)\backslash {[]^{-1}}{[]^{-1}}(({\langle\rangle}Na\backslash Sf)/{\exists}bNb))/{\blacksquare}(({\langle\rangle}Na\backslash Sf)/{\exists}bNb)):\\ (\Phinplus\ ({\it s}\ ({\it s}\ {\it 0}))\ {\it and}),
  {\blacksquare}{\forall}a(({\langle\rangle}Na\backslash Sf)/({\langle\rangle}Na\backslash Sb)): \lambda C\lambda D({\it Fut}\ ({\it C}\ {\it D})),\\ {\square}(({\langle\rangle}{\exists}aNa\backslash Sb)/{\exists}aNa): \mbox{\^{}}\lambda E\lambda F((\mbox{\v{}}{\it love}\ {\it E})\ {\it F})]], {\blacksquare}Nt(s(n)): {\it l}\ \Rightarrow\ Sf
\end{array}$}
There is the following derivation.
Again, 
after eliminating the modality of the coordinator,
and instantiating the left universal quantifiers for tense and subject agreement,
at the root,
there are three main subderivations marked \textcircled{1},
\textcircled{2}
and \textcircled{3} deriving the right conjunct,
the left conjunct and the coordinate structure context respectively.
In \textcircled{1},
after removing the modality on the right and lowering the object and subject
arguments into the antecedent we have to analyse a sequence which is
essentially S-Aux-TV-O. This involves analysing TV-O as VP
(left subsubderivation),
and S-VP as S (right subsubderivation).
Subtree \textcircled{2} involves essentially derivation of the identity
TV yields TV.
Subderivation \textcircled{3} checks the double bracketing of the coordinate
structure and recognises the right object context and left subject context.
\vspace{0.15in}
$$
\rotatebox{-90}{\tiny
\prooftree
\prooftree
\prooftree
\prooftree
\prooftree
\prooftree
\prooftree
\prooftree
\prooftree
\prooftree
\prooftree
\prooftree
\prooftree
\prooftree
\prooftree
\prooftree
\prooftree
\prooftree
\justifies
N2\ \Rightarrow\ N2
\endprooftree
\justifies
N2\ \Rightarrow\ \fbox{${\exists}aNa$}
\using {\exists}R
\endprooftree
\prooftree
\prooftree
\prooftree
\prooftree
\justifies
Nt(s(m))\ \Rightarrow\ Nt(s(m))
\endprooftree
\justifies
Nt(s(m))\ \Rightarrow\ \fbox{${\exists}aNa$}
\using {\exists}R
\endprooftree
\justifies
[Nt(s(m))]\ \Rightarrow\ \fbox{${\langle\rangle}{\exists}aNa$}
\using {\langle\rangle}R
\endprooftree
\prooftree
\justifies
\mbox{\fbox{$Sb$}}\ \Rightarrow\ Sb
\endprooftree
\justifies
[Nt(s(m))], \mbox{\fbox{${\langle\rangle}{\exists}aNa\backslash Sb$}}\ \Rightarrow\ Sb
\using {\backslash}L
\endprooftree
\justifies
[Nt(s(m))], \mbox{\fbox{$({\langle\rangle}{\exists}aNa\backslash Sb)/{\exists}aNa$}}, N2\ \Rightarrow\ Sb
\using {/}L
\endprooftree
\justifies
[Nt(s(m))], \mbox{\fbox{${\square}(({\langle\rangle}{\exists}aNa\backslash Sb)/{\exists}aNa)$}}, N2\ \Rightarrow\ Sb
\using {\Box}L
\endprooftree
\justifies
{\langle\rangle}Nt(s(m)), {\square}(({\langle\rangle}{\exists}aNa\backslash Sb)/{\exists}aNa), N2\ \Rightarrow\ Sb
\using {\langle\rangle}L
\endprooftree
\justifies
{\square}(({\langle\rangle}{\exists}aNa\backslash Sb)/{\exists}aNa), N2\ \Rightarrow\ {\langle\rangle}Nt(s(m))\backslash Sb
\using {\backslash}R
\endprooftree
\prooftree
\prooftree
\prooftree
\justifies
Nt(s(m))\ \Rightarrow\ Nt(s(m))
\endprooftree
\justifies
[Nt(s(m))]\ \Rightarrow\ \fbox{${\langle\rangle}Nt(s(m))$}
\using {\langle\rangle}R
\endprooftree
\prooftree
\justifies
\mbox{\fbox{$Sf$}}\ \Rightarrow\ Sf
\endprooftree
\justifies
[Nt(s(m))], \mbox{\fbox{${\langle\rangle}Nt(s(m))\backslash Sf$}}\ \Rightarrow\ Sf
\using {\backslash}L
\endprooftree
\justifies
[Nt(s(m))], \mbox{\fbox{$({\langle\rangle}Nt(s(m))\backslash Sf)/({\langle\rangle}Nt(s(m))\backslash Sb)$}}, {\square}(({\langle\rangle}{\exists}aNa\backslash Sb)/{\exists}aNa), N2\ \Rightarrow\ Sf
\using {/}L
\endprooftree
\justifies
[Nt(s(m))], \mbox{\fbox{${\forall}a(({\langle\rangle}Na\backslash Sf)/({\langle\rangle}Na\backslash Sb))$}}, {\square}(({\langle\rangle}{\exists}aNa\backslash Sb)/{\exists}aNa), N2\ \Rightarrow\ Sf
\using {\forall}L
\endprooftree
\justifies
[Nt(s(m))], \mbox{\fbox{${\blacksquare}{\forall}a(({\langle\rangle}Na\backslash Sf)/({\langle\rangle}Na\backslash Sb))$}}, {\square}(({\langle\rangle}{\exists}aNa\backslash Sb)/{\exists}aNa), N2\ \Rightarrow\ Sf
\using {\blacksquare}L
\endprooftree
\justifies
[Nt(s(m))], {\blacksquare}{\forall}a(({\langle\rangle}Na\backslash Sf)/({\langle\rangle}Na\backslash Sb)), {\square}(({\langle\rangle}{\exists}aNa\backslash Sb)/{\exists}aNa), {\exists}bNb\ \Rightarrow\ Sf
\using {\exists}L
\endprooftree
\justifies
{\langle\rangle}Nt(s(m)), {\blacksquare}{\forall}a(({\langle\rangle}Na\backslash Sf)/({\langle\rangle}Na\backslash Sb)), {\square}(({\langle\rangle}{\exists}aNa\backslash Sb)/{\exists}aNa), {\exists}bNb\ \Rightarrow\ Sf
\using {\langle\rangle}L
\endprooftree
\justifies
{\blacksquare}{\forall}a(({\langle\rangle}Na\backslash Sf)/({\langle\rangle}Na\backslash Sb)), {\square}(({\langle\rangle}{\exists}aNa\backslash Sb)/{\exists}aNa), {\exists}bNb\ \Rightarrow\ {\langle\rangle}Nt(s(m))\backslash Sf
\using {\backslash}R
\endprooftree
\justifies
{\blacksquare}{\forall}a(({\langle\rangle}Na\backslash Sf)/({\langle\rangle}Na\backslash Sb)), {\square}(({\langle\rangle}{\exists}aNa\backslash Sb)/{\exists}aNa)\ \Rightarrow\ ({\langle\rangle}Nt(s(m))\backslash Sf)/{\exists}bNb
\using {/}R
\endprooftree
\justifies
\begin{array}{c}
{\blacksquare}{\forall}a(({\langle\rangle}Na\backslash Sf)/({\langle\rangle}Na\backslash Sb)), {\square}(({\langle\rangle}{\exists}aNa\backslash Sb)/{\exists}aNa)\ \Rightarrow\ {\blacksquare}(({\langle\rangle}Nt(s(m))\backslash Sf)/{\exists}bNb)\\
\mbox{\footnotesize\textcircled{1}}
\end{array}
\using {\blacksquare}R
\endprooftree
\prooftree
\prooftree
\prooftree
\prooftree
\prooftree
\prooftree
\prooftree
\prooftree
\prooftree
\prooftree
\prooftree
\justifies
N3\ \Rightarrow\ N3
\endprooftree
\justifies
N3\ \Rightarrow\ \fbox{${\exists}aNa$}
\using {\exists}R
\endprooftree
\prooftree
\prooftree
\prooftree
\prooftree
\justifies
Nt(s(m))\ \Rightarrow\ Nt(s(m))
\endprooftree
\justifies
Nt(s(m))\ \Rightarrow\ \fbox{${\exists}gNt(s(g))$}
\using {\exists}R
\endprooftree
\justifies
[Nt(s(m))]\ \Rightarrow\ \fbox{${\langle\rangle}{\exists}gNt(s(g))$}
\using {\langle\rangle}R
\endprooftree
\prooftree
\justifies
\mbox{\fbox{$Sf$}}\ \Rightarrow\ Sf
\endprooftree
\justifies
[Nt(s(m))], \mbox{\fbox{${\langle\rangle}{\exists}gNt(s(g))\backslash Sf$}}\ \Rightarrow\ Sf
\using {\backslash}L
\endprooftree
\justifies
[Nt(s(m))], \mbox{\fbox{$({\langle\rangle}{\exists}gNt(s(g))\backslash Sf)/{\exists}aNa$}}, N3\ \Rightarrow\ Sf
\using {/}L
\endprooftree
\justifies
[Nt(s(m))], \mbox{\fbox{${\square}(({\langle\rangle}{\exists}gNt(s(g))\backslash Sf)/{\exists}aNa)$}}, N3\ \Rightarrow\ Sf
\using {\Box}L
\endprooftree
\justifies
[Nt(s(m))], {\square}(({\langle\rangle}{\exists}gNt(s(g))\backslash Sf)/{\exists}aNa), {\exists}bNb\ \Rightarrow\ Sf
\using {\exists}L
\endprooftree
\justifies
{\langle\rangle}Nt(s(m)), {\square}(({\langle\rangle}{\exists}gNt(s(g))\backslash Sf)/{\exists}aNa), {\exists}bNb\ \Rightarrow\ Sf
\using {\langle\rangle}L
\endprooftree
\justifies
{\square}(({\langle\rangle}{\exists}gNt(s(g))\backslash Sf)/{\exists}aNa), {\exists}bNb\ \Rightarrow\ {\langle\rangle}Nt(s(m))\backslash Sf
\using {\backslash}R
\endprooftree
\justifies
{\square}(({\langle\rangle}{\exists}gNt(s(g))\backslash Sf)/{\exists}aNa)\ \Rightarrow\ ({\langle\rangle}Nt(s(m))\backslash Sf)/{\exists}bNb
\using {/}R
\endprooftree
\justifies
{\square}(({\langle\rangle}{\exists}gNt(s(g))\backslash Sf)/{\exists}aNa)\ \Rightarrow\ {\blacksquare}(({\langle\rangle}Nt(s(m))\backslash Sf)/{\exists}bNb)
\using {\blacksquare}R
\endprooftree
\justifies
\begin{array}{c}
{\square}(({\langle\rangle}{\exists}gNt(s(g))\backslash Sf)/{\exists}aNa)\ \Rightarrow\ \fbox{$?{\blacksquare}(({\langle\rangle}Nt(s(m))\backslash Sf)/{\exists}bNb)$}\\
\mbox{\footnotesize\textcircled{2}}
\end{array}
\using {?}R
\endprooftree
\prooftree
\prooftree
\prooftree
\prooftree
\prooftree
\prooftree
\justifies
\mbox{\fbox{$Nt(s(n))$}}\ \Rightarrow\ Nt(s(n))
\endprooftree
\justifies
\mbox{\fbox{${\blacksquare}Nt(s(n))$}}\ \Rightarrow\ Nt(s(n))
\using {\blacksquare}L
\endprooftree
\justifies
{\blacksquare}Nt(s(n))\ \Rightarrow\ \fbox{${\exists}bNb$}
\using {\exists}R
\endprooftree
\prooftree
\prooftree
\prooftree
\prooftree
\justifies
\mbox{\fbox{$Nt(s(m))$}}\ \Rightarrow\ Nt(s(m))
\endprooftree
\justifies
\mbox{\fbox{${\blacksquare}Nt(s(m))$}}\ \Rightarrow\ Nt(s(m))
\using {\blacksquare}L
\endprooftree
\justifies
[{\blacksquare}Nt(s(m))]\ \Rightarrow\ \fbox{${\langle\rangle}Nt(s(m))$}
\using {\langle\rangle}R
\endprooftree
\prooftree
\justifies
\mbox{\fbox{$Sf$}}\ \Rightarrow\ Sf
\endprooftree
\justifies
[{\blacksquare}Nt(s(m))], \mbox{\fbox{${\langle\rangle}Nt(s(m))\backslash Sf$}}\ \Rightarrow\ Sf
\using {\backslash}L
\endprooftree
\justifies
[{\blacksquare}Nt(s(m))], \mbox{\fbox{$({\langle\rangle}Nt(s(m))\backslash Sf)/{\exists}bNb$}}, {\blacksquare}Nt(s(n))\ \Rightarrow\ Sf
\using {/}L
\endprooftree
\justifies
[{\blacksquare}Nt(s(m))], [\mbox{\fbox{${[]^{-1}}(({\langle\rangle}Nt(s(m))\backslash Sf)/{\exists}bNb)$}}], {\blacksquare}Nt(s(n))\ \Rightarrow\ Sf
\using {[]^{-1}}L
\endprooftree
\justifies
\begin{array}{c}
[{\blacksquare}Nt(s(m))], [[\mbox{\fbox{${[]^{-1}}{[]^{-1}}(({\langle\rangle}Nt(s(m))\backslash Sf)/{\exists}bNb)$}}]], {\blacksquare}Nt(s(n))\ \Rightarrow\ Sf\\
\mbox{\footnotesize\textcircled{3}}
\end{array}
\using {[]^{-1}}L
\endprooftree
\justifies
[{\blacksquare}Nt(s(m))], [[{\square}(({\langle\rangle}{\exists}gNt(s(g))\backslash Sf)/{\exists}aNa), \mbox{\fbox{$?{\blacksquare}(({\langle\rangle}Nt(s(m))\backslash Sf)/{\exists}bNb)\backslash {[]^{-1}}{[]^{-1}}(({\langle\rangle}Nt(s(m))\backslash Sf)/{\exists}bNb)$}}]], {\blacksquare}Nt(s(n))\ \Rightarrow\ Sf
\using {\backslash}L
\endprooftree
\justifies
[{\blacksquare}Nt(s(m))], [[{\square}(({\langle\rangle}{\exists}gNt(s(g))\backslash Sf)/{\exists}aNa), \mbox{\fbox{$(?{\blacksquare}(({\langle\rangle}Nt(s(m))\backslash Sf)/{\exists}bNb)\backslash {[]^{-1}}{[]^{-1}}(({\langle\rangle}Nt(s(m))\backslash Sf)/{\exists}bNb))/{\blacksquare}(({\langle\rangle}Nt(s(m))\backslash Sf)/{\exists}bNb)$}}, {\blacksquare}{\forall}a(({\langle\rangle}Na\backslash Sf)/({\langle\rangle}Na\backslash Sb)), {\square}(({\langle\rangle}{\exists}aNa\backslash Sb)/{\exists}aNa)]], {\blacksquare}Nt(s(n))\ \Rightarrow\ Sf
\using {/}L
\endprooftree
\justifies
[{\blacksquare}Nt(s(m))], [[{\square}(({\langle\rangle}{\exists}gNt(s(g))\backslash Sf)/{\exists}aNa), \mbox{\fbox{${\forall}a((?{\blacksquare}(({\langle\rangle}Na\backslash Sf)/{\exists}bNb)\backslash {[]^{-1}}{[]^{-1}}(({\langle\rangle}Na\backslash Sf)/{\exists}bNb))/{\blacksquare}(({\langle\rangle}Na\backslash Sf)/{\exists}bNb))$}}, {\blacksquare}{\forall}a(({\langle\rangle}Na\backslash Sf)/({\langle\rangle}Na\backslash Sb)), {\square}(({\langle\rangle}{\exists}aNa\backslash Sb)/{\exists}aNa)]], {\blacksquare}Nt(s(n))\ \Rightarrow\ Sf
\using {\forall}L
\endprooftree
\justifies
[{\blacksquare}Nt(s(m))], [[{\square}(({\langle\rangle}{\exists}gNt(s(g))\backslash Sf)/{\exists}aNa), \mbox{\fbox{${\forall}f{\forall}a((?{\blacksquare}(({\langle\rangle}Na\backslash Sf)/{\exists}bNb)\backslash {[]^{-1}}{[]^{-1}}(({\langle\rangle}Na\backslash Sf)/{\exists}bNb))/{\blacksquare}(({\langle\rangle}Na\backslash Sf)/{\exists}bNb))$}}, {\blacksquare}{\forall}a(({\langle\rangle}Na\backslash Sf)/({\langle\rangle}Na\backslash Sb)), {\square}(({\langle\rangle}{\exists}aNa\backslash Sb)/{\exists}aNa)]], {\blacksquare}Nt(s(n))\ \Rightarrow\ Sf
\using {\forall}L
\endprooftree
\justifies
[{\blacksquare}Nt(s(m))], [[{\square}(({\langle\rangle}{\exists}gNt(s(g))\backslash Sf)/{\exists}aNa), \mbox{\fbox{${\blacksquare}{\forall}f{\forall}a((?{\blacksquare}(({\langle\rangle}Na\backslash Sf)/{\exists}bNb)\backslash {[]^{-1}}{[]^{-1}}(({\langle\rangle}Na\backslash Sf)/{\exists}bNb))/{\blacksquare}(({\langle\rangle}Na\backslash Sf)/{\exists}bNb))$}}, {\blacksquare}{\forall}a(({\langle\rangle}Na\backslash Sf)/({\langle\rangle}Na\backslash Sb)), {\square}(({\langle\rangle}{\exists}aNa\backslash Sb)/{\exists}aNa)]], {\blacksquare}Nt(s(n))\ \Rightarrow\ Sf
\using {\blacksquare}L
\endprooftree}
$$
\vspace{0.15in}
\noindent
The coordination combinator semantics is such that
\disp{
$\begin{array}[t]{l}
(((((\Phinplus\ (s\ (s\ 0))\ {\it and})\ x)\ [y])\ z)\ w) =\\
((((\Phinplus\ (s\ 0)\ {\it and})\ (x\ z))\ (\alphaplus\ [y]\ z))\ w) =\\
((((\Phinplus\ (s\ 0)\ {\it and})\ (x\ z))\ [(y\ z)])\ w) =\\
(((\Phinplus\ 0\ {\it and})\ ((x\ z)\ w))\ (\alphaplus\ [(y\ z)]\ w)) =\\
(((\Phinplus\ 0\ {\it and})\ ((x\ z)\ w))\ [((y\ z)\ w)]) =\\
{}[((y\ z)\ w)\wedge((x\ z)\ w)]
\end{array}$}
All this delivers semantics:
\disp{
$[({\it Pres}\ ((\mbox{\v{}}{\it like}\ {\it l})\ {\it j}))\wedge ({\it Fut}\ ((\mbox{\v{}}{\it love}\ {\it l})\ {\it j}))]$}

\subsection{Ditransitive verb coordination}

The next example is of ditransitive verb disjunction:
\disp{
$[{\bf john}]{+}[[{\bf gave}{+}{\bf or}{+}{\bf sent}]]{+}{\bf mary}{+}{\bf the}{+}{\bf book}: Sf$}
Appropriate lexical lookup yields the semantically labelled sequent where the coordinator type is
essentially $(\exstexp X\bsl\abrack\abrack X)/X$ with $X=((N\bsl S)/N)/N$
(using the curried ditransitive verb type):
\disp{
$\begin{array}[t]{l}
[{\blacksquare}Nt(s(m)): {\it j}], [[{\square}((({\langle\rangle}{\exists}aNa\backslash Sf)/{\exists}aNa)/{\exists}aNa): \mbox{\^{}}\lambda A\lambda B\lambda C({\it Past}\ (((\mbox{\v{}}{\it give}\ {\it A})\ {\it B})\ {\it C})),\\
 {\blacksquare}{\forall}a{\forall}f((?{\blacksquare}((({\langle\rangle}Na\backslash Sf)/{\exists}bNb)/{\exists}bNb)\backslash {[]^{-1}}{[]^{-1}}((({\langle\rangle}Na\backslash Sf)/{\exists}bNb)/{\exists}bNb))/\\{\blacksquare}((({\langle\rangle}Na\backslash Sf)/{\exists}bNb)/{\exists}bNb)): (\Phinplus\ ({\it s}\ ({\it s}\ ({\it s}\ {\it 0})))\ {\it or}), {\square}((({\langle\rangle}{\exists}aNa\backslash Sf)/{\exists}aNa)/{\exists}aNa): \\\mbox{\^{}}\lambda D\lambda E\lambda F({\it Past}\ (((\mbox{\v{}}{\it send}\ {\it D})\ {\it E})\ {\it F}))]],{\blacksquare}Nt(s(f)): {\it m}, {\blacksquare}{\forall}n(Nt(n)/{\it CN}{\it n}): \iota,\\ {\square}{\it CN}{\it s(n)}: {\it book}\ \Rightarrow\ Sf
 \end{array}$}
There is the derivation:
\vspace{0.15in}
$${\tiny
\prooftree
\prooftree
\prooftree
\prooftree
\prooftree
\prooftree
\prooftree
\prooftree
\prooftree
\prooftree
\prooftree
\justifies
N1\ \Rightarrow\ N1
\endprooftree
\justifies
N1\ \Rightarrow\ \fbox{${\exists}aNa$}
\using {\exists}R
\endprooftree
\prooftree
\prooftree
\prooftree
\justifies
N2\ \Rightarrow\ N2
\endprooftree
\justifies
N2\ \Rightarrow\ \fbox{${\exists}aNa$}
\using {\exists}R
\endprooftree
\prooftree
\prooftree
\prooftree
\prooftree
\justifies
Nt(s(m))\ \Rightarrow\ Nt(s(m))
\endprooftree
\justifies
Nt(s(m))\ \Rightarrow\ \fbox{${\exists}aNa$}
\using {\exists}R
\endprooftree
\justifies
[Nt(s(m))]\ \Rightarrow\ \fbox{${\langle\rangle}{\exists}aNa$}
\using {\langle\rangle}R
\endprooftree
\prooftree
\justifies
\mbox{\fbox{$Sf$}}\ \Rightarrow\ Sf
\endprooftree
\justifies
[Nt(s(m))], \mbox{\fbox{${\langle\rangle}{\exists}aNa\backslash Sf$}}\ \Rightarrow\ Sf
\using {\backslash}L
\endprooftree
\justifies
[Nt(s(m))], \mbox{\fbox{$({\langle\rangle}{\exists}aNa\backslash Sf)/{\exists}aNa$}}, N2\ \Rightarrow\ Sf
\using {/}L
\endprooftree
\justifies
[Nt(s(m))], \mbox{\fbox{$(({\langle\rangle}{\exists}aNa\backslash Sf)/{\exists}aNa)/{\exists}aNa$}}, N1, N2\ \Rightarrow\ Sf
\using {/}L
\endprooftree
\justifies
[Nt(s(m))], \mbox{\fbox{${\square}((({\langle\rangle}{\exists}aNa\backslash Sf)/{\exists}aNa)/{\exists}aNa)$}}, N1, N2\ \Rightarrow\ Sf
\using {\Box}L
\endprooftree
\justifies
[Nt(s(m))], {\square}((({\langle\rangle}{\exists}aNa\backslash Sf)/{\exists}aNa)/{\exists}aNa), N1, {\exists}bNb\ \Rightarrow\ Sf
\using {\exists}L
\endprooftree
\justifies
[Nt(s(m))], {\square}((({\langle\rangle}{\exists}aNa\backslash Sf)/{\exists}aNa)/{\exists}aNa), {\exists}bNb, {\exists}bNb\ \Rightarrow\ Sf
\using {\exists}L
\endprooftree
\justifies
{\langle\rangle}Nt(s(m)), {\square}((({\langle\rangle}{\exists}aNa\backslash Sf)/{\exists}aNa)/{\exists}aNa), {\exists}bNb, {\exists}bNb\ \Rightarrow\ Sf
\using {\langle\rangle}L
\endprooftree
\justifies
{\square}((({\langle\rangle}{\exists}aNa\backslash Sf)/{\exists}aNa)/{\exists}aNa), {\exists}bNb, {\exists}bNb\ \Rightarrow\ {\langle\rangle}Nt(s(m))\backslash Sf
\using {\backslash}R
\endprooftree
\justifies
{\square}((({\langle\rangle}{\exists}aNa\backslash Sf)/{\exists}aNa)/{\exists}aNa), {\exists}bNb\ \Rightarrow\ ({\langle\rangle}Nt(s(m))\backslash Sf)/{\exists}bNb
\using {/}R
\endprooftree
\justifies
{\square}((({\langle\rangle}{\exists}aNa\backslash Sf)/{\exists}aNa)/{\exists}aNa)\ \Rightarrow\ (({\langle\rangle}Nt(s(m))\backslash Sf)/{\exists}bNb)/{\exists}bNb
\using {/}R
\endprooftree
\justifies
\begin{array}{c}
{\square}((({\langle\rangle}{\exists}aNa\backslash Sf)/{\exists}aNa)/{\exists}aNa)\ \Rightarrow\ {\blacksquare}((({\langle\rangle}Nt(s(m))\backslash Sf)/{\exists}bNb)/{\exists}bNb)\\
\mbox{\footnotesize\textcircled{1}}
\end{array}
\using {\blacksquare}R
\endprooftree
\prooftree
\prooftree
\vdots
\justifies
{\square}((({\langle\rangle}{\exists}aNa\backslash Sf)/{\exists}aNa)/{\exists}aNa)\ \Rightarrow\ {\blacksquare}((({\langle\rangle}Nt(s(m))\backslash Sf)/{\exists}bNb)/{\exists}bNb)
\using {\blacksquare}R
\endprooftree
\justifies
\begin{array}{c}
{\square}((({\langle\rangle}{\exists}aNa\backslash Sf)/{\exists}aNa)/{\exists}aNa)\ \Rightarrow\ \fbox{$?{\blacksquare}((({\langle\rangle}Nt(s(m))\backslash Sf)/{\exists}bNb)/{\exists}bNb)$}\\
\mbox{\footnotesize\textcircled{2}}
\end{array}
\using {?}R
\endprooftree}
$$

\begin{center}
\rotatebox{-90}{\tiny
\prooftree
\prooftree
\prooftree
\prooftree
\mbox{\footnotesize\textcircled{1}}\tab
\prooftree
\mbox{\footnotesize\textcircled{2}}\tab
\prooftree
\prooftree
\prooftree
\prooftree
\prooftree
\prooftree
\justifies
\mbox{\fbox{$Nt(s(f))$}}\ \Rightarrow\ Nt(s(f))
\endprooftree
\justifies
\mbox{\fbox{${\blacksquare}Nt(s(f))$}}\ \Rightarrow\ Nt(s(f))
\using {\blacksquare}L
\endprooftree
\justifies
{\blacksquare}Nt(s(f))\ \Rightarrow\ \fbox{${\exists}bNb$}
\using {\exists}R
\endprooftree
\prooftree
\prooftree
\prooftree
\prooftree
\prooftree
\prooftree
\prooftree
\justifies
\mbox{\fbox{${\it CN}{\it s(n)}$}}\ \Rightarrow\ {\it CN}{\it s(n)}
\endprooftree
\justifies
\mbox{\fbox{${\square}{\it CN}{\it s(n)}$}}\ \Rightarrow\ {\it CN}{\it s(n)}
\using {\Box}L
\endprooftree
\prooftree
\justifies
\mbox{\fbox{$Nt(s(n))$}}\ \Rightarrow\ Nt(s(n))
\endprooftree
\justifies
\mbox{\fbox{$Nt(s(n))/{\it CN}{\it s(n)}$}}, {\square}{\it CN}{\it s(n)}\ \Rightarrow\ Nt(s(n))
\using {/}L
\endprooftree
\justifies
\mbox{\fbox{${\forall}n(Nt(n)/{\it CN}{\it n})$}}, {\square}{\it CN}{\it s(n)}\ \Rightarrow\ Nt(s(n))
\using {\forall}L
\endprooftree
\justifies
\mbox{\fbox{${\blacksquare}{\forall}n(Nt(n)/{\it CN}{\it n})$}}, {\square}{\it CN}{\it s(n)}\ \Rightarrow\ Nt(s(n))
\using {\blacksquare}L
\endprooftree
\justifies
{\blacksquare}{\forall}n(Nt(n)/{\it CN}{\it n}), {\square}{\it CN}{\it s(n)}\ \Rightarrow\ \fbox{${\exists}bNb$}
\using {\exists}R
\endprooftree
\prooftree
\prooftree
\prooftree
\prooftree
\justifies
\mbox{\fbox{$Nt(s(m))$}}\ \Rightarrow\ Nt(s(m))
\endprooftree
\justifies
\mbox{\fbox{${\blacksquare}Nt(s(m))$}}\ \Rightarrow\ Nt(s(m))
\using {\blacksquare}L
\endprooftree
\justifies
[{\blacksquare}Nt(s(m))]\ \Rightarrow\ \fbox{${\langle\rangle}Nt(s(m))$}
\using {\langle\rangle}R
\endprooftree
\prooftree
\justifies
\mbox{\fbox{$Sf$}}\ \Rightarrow\ Sf
\endprooftree
\justifies
[{\blacksquare}Nt(s(m))], \mbox{\fbox{${\langle\rangle}Nt(s(m))\backslash Sf$}}\ \Rightarrow\ Sf
\using {\backslash}L
\endprooftree
\justifies
[{\blacksquare}Nt(s(m))], \mbox{\fbox{$({\langle\rangle}Nt(s(m))\backslash Sf)/{\exists}bNb$}}, {\blacksquare}{\forall}n(Nt(n)/{\it CN}{\it n}), {\square}{\it CN}{\it s(n)}\ \Rightarrow\ Sf
\using {/}L
\endprooftree
\justifies
[{\blacksquare}Nt(s(m))], \mbox{\fbox{$(({\langle\rangle}Nt(s(m))\backslash Sf)/{\exists}bNb)/{\exists}bNb$}}, {\blacksquare}Nt(s(f)), {\blacksquare}{\forall}n(Nt(n)/{\it CN}{\it n}), {\square}{\it CN}{\it s(n)}\ \Rightarrow\ Sf
\using {/}L
\endprooftree
\justifies
[{\blacksquare}Nt(s(m))], [\mbox{\fbox{${[]^{-1}}((({\langle\rangle}Nt(s(m))\backslash Sf)/{\exists}bNb)/{\exists}bNb)$}}], {\blacksquare}Nt(s(f)), {\blacksquare}{\forall}n(Nt(n)/{\it CN}{\it n}), {\square}{\it CN}{\it s(n)}\ \Rightarrow\ Sf
\using {[]^{-1}}L
\endprooftree
\justifies
[{\blacksquare}Nt(s(m))], [[\mbox{\fbox{${[]^{-1}}{[]^{-1}}((({\langle\rangle}Nt(s(m))\backslash Sf)/{\exists}bNb)/{\exists}bNb)$}}]], {\blacksquare}Nt(s(f)), {\blacksquare}{\forall}n(Nt(n)/{\it CN}{\it n}), {\square}{\it CN}{\it s(n)}\ \Rightarrow\ Sf
\using {[]^{-1}}L
\endprooftree\justifies
[{\blacksquare}Nt(s(m))], [[{\square}((({\langle\rangle}{\exists}aNa\backslash Sf)/{\exists}aNa)/{\exists}aNa), \mbox{\fbox{$?{\blacksquare}((({\langle\rangle}Nt(s(m))\backslash Sf)/{\exists}bNb)/{\exists}bNb)\backslash {[]^{-1}}{[]^{-1}}((({\langle\rangle}Nt(s(m))\backslash Sf)/{\exists}bNb)/{\exists}bNb)$}}]], {\blacksquare}Nt(s(f)), {\blacksquare}{\forall}n(Nt(n)/{\it CN}{\it n}), {\square}{\it CN}{\it s(n)}\ \Rightarrow\ Sf
\using {\backslash}L
\endprooftree
\justifies
[{\blacksquare}Nt(s(m))], [[{\square}((({\langle\rangle}{\exists}aNa\backslash Sf)/{\exists}aNa)/{\exists}aNa), \mbox{\fbox{$(?{\blacksquare}((({\langle\rangle}Nt(s(m))\backslash Sf)/{\exists}bNb)/{\exists}bNb)\backslash {[]^{-1}}{[]^{-1}}((({\langle\rangle}Nt(s(m))\backslash Sf)/{\exists}bNb)/{\exists}bNb))/{\blacksquare}((({\langle\rangle}Nt(s(m))\backslash Sf)/{\exists}bNb)/{\exists}bNb)$}}, {\square}((({\langle\rangle}{\exists}aNa\backslash Sf)/{\exists}aNa)/{\exists}aNa)]], {\blacksquare}Nt(s(f)), {\blacksquare}{\forall}n(Nt(n)/{\it CN}{\it n}), {\square}{\it CN}{\it s(n)}\ \Rightarrow\ Sf
\using {/}L
\endprooftree
\justifies
[{\blacksquare}Nt(s(m))], [[{\square}((({\langle\rangle}{\exists}aNa\backslash Sf)/{\exists}aNa)/{\exists}aNa), \mbox{\fbox{${\forall}f((?{\blacksquare}((({\langle\rangle}Nt(s(m))\backslash Sf)/{\exists}bNb)/{\exists}bNb)\backslash {[]^{-1}}{[]^{-1}}((({\langle\rangle}Nt(s(m))\backslash Sf)/{\exists}bNb)/{\exists}bNb))/{\blacksquare}((({\langle\rangle}Nt(s(m))\backslash Sf)/{\exists}bNb)/{\exists}bNb))$}}, {\square}((({\langle\rangle}{\exists}aNa\backslash Sf)/{\exists}aNa)/{\exists}aNa)]], {\blacksquare}Nt(s(f)), {\blacksquare}{\forall}n(Nt(n)/{\it CN}{\it n}), {\square}{\it CN}{\it s(n)}\ \Rightarrow\ Sf
\using {\forall}L
\endprooftree
\justifies
[{\blacksquare}Nt(s(m))], [[{\square}((({\langle\rangle}{\exists}aNa\backslash Sf)/{\exists}aNa)/{\exists}aNa), \mbox{\fbox{${\forall}a{\forall}f((?{\blacksquare}((({\langle\rangle}Na\backslash Sf)/{\exists}bNb)/{\exists}bNb)\backslash {[]^{-1}}{[]^{-1}}((({\langle\rangle}Na\backslash Sf)/{\exists}bNb)/{\exists}bNb))/{\blacksquare}((({\langle\rangle}Na\backslash Sf)/{\exists}bNb)/{\exists}bNb))$}}, {\square}((({\langle\rangle}{\exists}aNa\backslash Sf)/{\exists}aNa)/{\exists}aNa)]], {\blacksquare}Nt(s(f)), {\blacksquare}{\forall}n(Nt(n)/{\it CN}{\it n}), {\square}{\it CN}{\it s(n)}\ \Rightarrow\ Sf
\using {\forall}L
\endprooftree
\justifies
[{\blacksquare}Nt(s(m))], [[{\square}((({\langle\rangle}{\exists}aNa\backslash Sf)/{\exists}aNa)/{\exists}aNa), \mbox{\fbox{${\blacksquare}{\forall}a{\forall}f((?{\blacksquare}((({\langle\rangle}Na\backslash Sf)/{\exists}bNb)/{\exists}bNb)\backslash {[]^{-1}}{[]^{-1}}((({\langle\rangle}Na\backslash Sf)/{\exists}bNb)/{\exists}bNb))/{\blacksquare}((({\langle\rangle}Na\backslash Sf)/{\exists}bNb)/{\exists}bNb))$}}, {\square}((({\langle\rangle}{\exists}aNa\backslash Sf)/{\exists}aNa)/{\exists}aNa)]], {\blacksquare}Nt(s(f)), {\blacksquare}{\forall}n(Nt(n)/{\it CN}{\it n}), {\square}{\it CN}{\it s(n)}\ \Rightarrow\ Sf
\using {\blacksquare}L
\endprooftree}
\end{center}

\vspace{0.15in}

\noindent
As usual the coordinator modality is removed and the subject agreement and
tense feature are instantiated at the root.
Then the subderivations \textcircled{1} and \textcircled{2} derive the
right and left conjuncts respectively by essentially
deriving the identity TTV yields TTV on the ditransitive verb type TTV:
they are the same except for the initial $?R$ in \textcircled{2},
hence the latter is elided.
The remaining context derivation checks the double bracketing as usual
and for the rest amounts to analysis of a sequence S-TTV-O1-O2
where S and O1 are proper names and O2 is a definite noun phrase.

The example is interesting in that it illustrates coordination of arity three.
The coordination combinator semantics is such that:
\disp{
$\begin{array}[t]{l}
((((((\Phinplus\ (s\ (s\ (s\ 0)))\ {\it or})\ x)\ [y])\ z)\ w)\ u) =\\
(((((\Phinplus\ (s\ (s\ 0))\ {\it or})\ (x\ z))\ (\alphaplus\ [y]\ z))\ w)\ u) =\\
(((((\Phinplus\ (s\ (s\ 0))\ {\it or})\ (x\ z))\ [(y\ z)])\ w)\ u) =\\
((((\Phinplus\ (s\ 0)\ {\it or})\ ((x\ z)\ w))\ (\alphaplus\ [(y\ z)]\ w))\ u) =\\
((((\Phinplus\ (s\ 0)\ {\it or})\ ((x\ z)\ w))\ [((y\ z)\ w])\ u) =\\
(((\Phinplus\ 0\ {\it or})\ (((x\ z)\ w)\ u))\ (\alphaplus\ [((y\ z)\ w)]\ u)) =\\
(((\Phinplus\ 0\ {\it or})\ (((x\ z)\ w)\ u))\ [(((y\ z)\ w)\ u)]) =\\
{}[(((y\ z)\ w)\ u)\wedge(((x\ z)\ w)\ u)]
\end{array}$}
All this delivers semantics:
\disp{
$[({\it Past}\ (((\mbox{\v{}}{\it give}\ {\it m})\ (\iota \ \mbox{\v{}}{\it book}))\ {\it j}))\vee ({\it Past}\ (((\mbox{\v{}}{\it send}\ {\it m})\ (\iota \ \mbox{\v{}}{\it book}))\ {\it j}))]$}

\subsection{Subject coordination}

We continue with subject disjunction:
\disp{
$[[[{\bf john}]{+}{\bf or}{+}[{\bf mary}]]]{+}{\bf sings}: Sf$}
Appropriate lexical lookup inserts a coordinator over lifted subject noun phrases
(cf. Montague 1973\cite{montague:ptq}), 
essentially $(\exstexp X\bsl\abrack\abrack X)/X$ with $X=$ $S/(N\bsl S)$.
\disp{
$\begin{array}[t]{l}
[[[{\blacksquare}Nt(s(m)): {\it j}],
{\blacksquare}{\forall}f((?{\blacksquare}(Sf/({\langle\rangle}{\exists}gNt(s(g))\backslash Sf))\backslash {[]^{-1}}{[]^{-1}}(Sf/({\langle\rangle}{\exists}gNt(s(g))\backslash Sf)))/\\{\blacksquare}(Sf/({\langle\rangle}{\exists}gNt(s(g))\backslash Sf))): (\Phinplus\ ({\it s}\ {\it 0})\ {\it or}), [{\blacksquare}Nt(s(f)): {\it m}]]],\\ {\square}({\langle\rangle}{\exists}gNt(s(g))\backslash Sf): \mbox{\^{}}\lambda A({\it Pres}\ (\mbox{\v{}}{\it sing}\ {\it A}))\ \Rightarrow\ Sf
\end{array}$}
There is the derivation:

\vspace{0.15in}

\begin{center}
\rotatebox{-90}{\tiny
\prooftree
\prooftree
\prooftree
\prooftree
\prooftree
\prooftree
\prooftree
\prooftree
\prooftree
\prooftree
\justifies
\mbox{\fbox{$Nt(s(f))$}}\ \Rightarrow\ Nt(s(f))
\endprooftree
\justifies
\mbox{\fbox{${\blacksquare}Nt(s(f))$}}\ \Rightarrow\ Nt(s(f))
\using {\blacksquare}L
\endprooftree
\justifies
{\blacksquare}Nt(s(f))\ \Rightarrow\ \fbox{${\exists}gNt(s(g))$}
\using {\exists}R
\endprooftree
\justifies
[{\blacksquare}Nt(s(f))]\ \Rightarrow\ \fbox{${\langle\rangle}{\exists}gNt(s(g))$}
\using {\langle\rangle}R
\endprooftree
\prooftree
\justifies
\mbox{\fbox{$Sf$}}\ \Rightarrow\ Sf
\endprooftree
\justifies
[{\blacksquare}Nt(s(f))], \mbox{\fbox{${\langle\rangle}{\exists}gNt(s(g))\backslash Sf$}}\ \Rightarrow\ Sf
\using {\backslash}L
\endprooftree
\justifies
[{\blacksquare}Nt(s(f))]\ \Rightarrow\ Sf/({\langle\rangle}{\exists}gNt(s(g))\backslash Sf)
\using {/}R
\endprooftree
\justifies
[{\blacksquare}Nt(s(f))]\ \Rightarrow\ {\blacksquare}(Sf/({\langle\rangle}{\exists}gNt(s(g))\backslash Sf))
\using {\blacksquare}R
\endprooftree
\prooftree
\prooftree
\prooftree
\prooftree
\prooftree
\prooftree
\prooftree
\prooftree
\prooftree
\justifies
\mbox{\fbox{$Nt(s(m))$}}\ \Rightarrow\ Nt(s(m))
\endprooftree
\justifies
\mbox{\fbox{${\blacksquare}Nt(s(m))$}}\ \Rightarrow\ Nt(s(m))
\using {\blacksquare}L
\endprooftree
\justifies
{\blacksquare}Nt(s(m))\ \Rightarrow\ \fbox{${\exists}gNt(s(g))$}
\using {\exists}R
\endprooftree
\justifies
[{\blacksquare}Nt(s(m))]\ \Rightarrow\ \fbox{${\langle\rangle}{\exists}gNt(s(g))$}
\using {\langle\rangle}R
\endprooftree
\prooftree
\justifies
\mbox{\fbox{$Sf$}}\ \Rightarrow\ Sf
\endprooftree
\justifies
[{\blacksquare}Nt(s(m))], \mbox{\fbox{${\langle\rangle}{\exists}gNt(s(g))\backslash Sf$}}\ \Rightarrow\ Sf
\using {\backslash}L
\endprooftree
\justifies
[{\blacksquare}Nt(s(m))]\ \Rightarrow\ Sf/({\langle\rangle}{\exists}gNt(s(g))\backslash Sf)
\using {/}R
\endprooftree
\justifies
[{\blacksquare}Nt(s(m))]\ \Rightarrow\ {\blacksquare}(Sf/({\langle\rangle}{\exists}gNt(s(g))\backslash Sf))
\using {\blacksquare}R
\endprooftree
\justifies
[{\blacksquare}Nt(s(m))]\ \Rightarrow\ \fbox{$?{\blacksquare}(Sf/({\langle\rangle}{\exists}gNt(s(g))\backslash Sf))$}
\using {?}R
\endprooftree
\prooftree
\prooftree
\prooftree
\prooftree
\prooftree
\prooftree
\prooftree
\prooftree
\prooftree
\prooftree
\prooftree
\justifies
Nt(s(1))\ \Rightarrow\ Nt(s(1))
\endprooftree
\justifies
Nt(s(1))\ \Rightarrow\ \fbox{${\exists}gNt(s(g))$}
\using {\exists}R
\endprooftree
\justifies
[Nt(s(1))]\ \Rightarrow\ \fbox{${\langle\rangle}{\exists}gNt(s(g))$}
\using {\langle\rangle}R
\endprooftree
\prooftree
\justifies
\mbox{\fbox{$Sf$}}\ \Rightarrow\ Sf
\endprooftree
\justifies
[Nt(s(1))], \mbox{\fbox{${\langle\rangle}{\exists}gNt(s(g))\backslash Sf$}}\ \Rightarrow\ Sf
\using {\backslash}L
\endprooftree
\justifies
[Nt(s(1))], \mbox{\fbox{${\square}({\langle\rangle}{\exists}gNt(s(g))\backslash Sf)$}}\ \Rightarrow\ Sf
\using {\Box}L
\endprooftree
\justifies
[{\exists}gNt(s(g))], {\square}({\langle\rangle}{\exists}gNt(s(g))\backslash Sf)\ \Rightarrow\ Sf
\using {\exists}L
\endprooftree
\justifies
{\langle\rangle}{\exists}gNt(s(g)), {\square}({\langle\rangle}{\exists}gNt(s(g))\backslash Sf)\ \Rightarrow\ Sf
\using {\langle\rangle}L
\endprooftree
\justifies
{\square}({\langle\rangle}{\exists}gNt(s(g))\backslash Sf)\ \Rightarrow\ {\langle\rangle}{\exists}gNt(s(g))\backslash Sf
\using {\backslash}R
\endprooftree
\prooftree
\justifies
\mbox{\fbox{$Sf$}}\ \Rightarrow\ Sf
\endprooftree
\justifies
\mbox{\fbox{$Sf/({\langle\rangle}{\exists}gNt(s(g))\backslash Sf)$}}, {\square}({\langle\rangle}{\exists}gNt(s(g))\backslash Sf)\ \Rightarrow\ Sf
\using {/}L
\endprooftree
\justifies
[\mbox{\fbox{${[]^{-1}}(Sf/({\langle\rangle}{\exists}gNt(s(g))\backslash Sf))$}}], {\square}({\langle\rangle}{\exists}gNt(s(g))\backslash Sf)\ \Rightarrow\ Sf
\using {[]^{-1}}L
\endprooftree
\justifies
[[\mbox{\fbox{${[]^{-1}}{[]^{-1}}(Sf/({\langle\rangle}{\exists}gNt(s(g))\backslash Sf))$}}]], {\square}({\langle\rangle}{\exists}gNt(s(g))\backslash Sf)\ \Rightarrow\ Sf
\using {[]^{-1}}L
\endprooftree
\justifies
[[[{\blacksquare}Nt(s(m))], \mbox{\fbox{$?{\blacksquare}(Sf/({\langle\rangle}{\exists}gNt(s(g))\backslash Sf))\backslash {[]^{-1}}{[]^{-1}}(Sf/({\langle\rangle}{\exists}gNt(s(g))\backslash Sf))$}}]], {\square}({\langle\rangle}{\exists}gNt(s(g))\backslash Sf)\ \Rightarrow\ Sf
\using {\backslash}L
\endprooftree
\justifies
[[[{\blacksquare}Nt(s(m))], \mbox{\fbox{$(?{\blacksquare}(Sf/({\langle\rangle}{\exists}gNt(s(g))\backslash Sf))\backslash {[]^{-1}}{[]^{-1}}(Sf/({\langle\rangle}{\exists}gNt(s(g))\backslash Sf)))/{\blacksquare}(Sf/({\langle\rangle}{\exists}gNt(s(g))\backslash Sf))$}}, [{\blacksquare}Nt(s(f))]]], {\square}({\langle\rangle}{\exists}gNt(s(g))\backslash Sf)\ \Rightarrow\ Sf
\using {/}L
\endprooftree
\justifies
[[[{\blacksquare}Nt(s(m))], \mbox{\fbox{${\forall}f((?{\blacksquare}(Sf/({\langle\rangle}{\exists}gNt(s(g))\backslash Sf))\backslash {[]^{-1}}{[]^{-1}}(Sf/({\langle\rangle}{\exists}gNt(s(g))\backslash Sf)))/{\blacksquare}(Sf/({\langle\rangle}{\exists}gNt(s(g))\backslash Sf)))$}}, [{\blacksquare}Nt(s(f))]]], {\square}({\langle\rangle}{\exists}gNt(s(g))\backslash Sf)\ \Rightarrow\ Sf
\using {\forall}L
\endprooftree
\justifies
[[[{\blacksquare}Nt(s(m))], \mbox{\fbox{${\blacksquare}{\forall}f((?{\blacksquare}(Sf/({\langle\rangle}{\exists}gNt(s(g))\backslash Sf))\backslash {[]^{-1}}{[]^{-1}}(Sf/({\langle\rangle}{\exists}gNt(s(g))\backslash Sf)))/{\blacksquare}(Sf/({\langle\rangle}{\exists}gNt(s(g))\backslash Sf)))$}}, [{\blacksquare}Nt(s(f))]]], {\square}({\langle\rangle}{\exists}gNt(s(g))\backslash Sf)\ \Rightarrow\ Sf
\using {\blacksquare}L
\endprooftree}
\end{center}

\vspace{0.15in}

\noindent
In the same manner that we have seen before,
in the left subderivation the righthand conjunct is analysed
as of the lifted type, and in the middle subderivation the lefthand
conjunct is analysed as of the lifted type;
these are the same except for the eventual $?R$ of the latter,
and they centre on the over right lowering
of the higher order verb phrase argument into the antecedent
where it subsequently applies as a functor.
In the righthand derivation the brackets are checked and then application
of the coordinate structure to its verb phrase context involves essentially
derivation of the identity VP yields VP.
This delivers the semantics:
\disp{
$[({\it Pres}\ (\mbox{\v{}}{\it sing}\ {\it j}))\vee ({\it Pres}\ (\mbox{\v{}}{\it sing}\ {\it m}))]$}

\subsection{Object coordination}

Object conjunction, including an object reflexive, is illustrated by:
\disp{
$[{\bf john}]{+}{\bf loves}{+}[[{\bf mary}{+}{\bf and}{+}{\bf himself}]]: Sf$}
The coordination is
basically $(\exstexp X\bsl\abrack\abrack X)/X$ with $X=((N\bsl S)/N)\bsl(N\bsl S)$.
\disp{
\vspace{0.15in}
$\begin{array}[t]{l}
[{\blacksquare}Nt(s(m)): {\it j}], {\square}(({\langle\rangle}{\exists}gNt(s(g))\backslash Sf)/{\exists}aNa): \mbox{\^{}}\lambda A\lambda B({\it Pres}\ ((\mbox{\v{}}{\it love}\ {\it A})\ {\it B})),
{}\mbox{$[[{\blacksquare}Nt(s(f)):{\it m},$}\\
 {\blacksquare}{\forall}f{\forall}a((?{\blacksquare}((({\langle\rangle}Na\backslash Sf)/{\exists}bNb)\backslash ({\langle\rangle}Na\backslash Sf))\backslash {[]^{-1}}{[]^{-1}}((({\langle\rangle}Na\backslash Sf)/{\exists}bNb)\backslash ({\langle\rangle}Na\backslash Sf)))/\\
 {\blacksquare}((({\langle\rangle}Na\backslash Sf)/{\exists}bNb)\backslash ({\langle\rangle}Na\backslash Sf))): (\Phinplus\ ({\it s}\ ({\it s}\ {\it 0}))\ {\it and}),\\ {\blacksquare}{\forall}f((({\langle\rangle}Nt(s(m))\backslash Sf){{\uparrow}{}}Nt(s(m))){{\downarrow}{}}({\langle\rangle}Nt(s(m))\backslash Sf)): \lambda C\lambda D(({\it C}\ {\it D})\ {\it D})]]\ \Rightarrow\ Sf
 \end{array}$}
There is the derivation:


{\tiny
\begin{center}
\prooftree
\prooftree
\prooftree
\prooftree
\prooftree
\prooftree
\prooftree
\prooftree
\prooftree
\prooftree
\prooftree
\prooftree
\prooftree
\justifies
Nt(s(m))\ \Rightarrow\ Nt(s(m))
\endprooftree
\justifies
Nt(s(m))\ \Rightarrow\ \fbox{${\exists}bNb$}
\using {\exists}R
\endprooftree
\prooftree
\prooftree
\prooftree
\justifies
Nt(s(m))\ \Rightarrow\ Nt(s(m))
\endprooftree
\justifies
[Nt(s(m))]\ \Rightarrow\ \fbox{${\langle\rangle}Nt(s(m))$}
\using {\langle\rangle}R
\endprooftree
\prooftree
\justifies
\mbox{\fbox{$Sf$}}\ \Rightarrow\ Sf
\endprooftree
\justifies
[Nt(s(m))], \mbox{\fbox{${\langle\rangle}Nt(s(m))\backslash Sf$}}\ \Rightarrow\ Sf
\using {\backslash}L
\endprooftree
\justifies
[Nt(s(m))], \mbox{\fbox{$({\langle\rangle}Nt(s(m))\backslash Sf)/{\exists}bNb$}}, Nt(s(m))\ \Rightarrow\ Sf
\using {/}L
\endprooftree
\justifies
{\langle\rangle}Nt(s(m)), ({\langle\rangle}Nt(s(m))\backslash Sf)/{\exists}bNb, Nt(s(m))\ \Rightarrow\ Sf
\using {\langle\rangle}L
\endprooftree
\justifies
({\langle\rangle}Nt(s(m))\backslash Sf)/{\exists}bNb, Nt(s(m))\ \Rightarrow\ {\langle\rangle}Nt(s(m))\backslash Sf
\using {\backslash}R
\endprooftree
\justifies
({\langle\rangle}Nt(s(m))\backslash Sf)/{\exists}bNb, {\tt 1}\ \Rightarrow\ ({\langle\rangle}Nt(s(m))\backslash Sf){{\uparrow}{}}Nt(s(m))
\using {\uparrow}R
\endprooftree
\prooftree
\prooftree
\prooftree
\justifies
Nt(s(m))\ \Rightarrow\ Nt(s(m))
\endprooftree
\justifies
[Nt(s(m))]\ \Rightarrow\ \fbox{${\langle\rangle}Nt(s(m))$}
\using {\langle\rangle}R
\endprooftree
\prooftree
\justifies
\mbox{\fbox{$Sf$}}\ \Rightarrow\ Sf
\endprooftree
\justifies
[Nt(s(m))], \mbox{\fbox{${\langle\rangle}Nt(s(m))\backslash Sf$}}\ \Rightarrow\ Sf
\using {\backslash}L
\endprooftree
\justifies
[Nt(s(m))], ({\langle\rangle}Nt(s(m))\backslash Sf)/{\exists}bNb, \mbox{\fbox{$(({\langle\rangle}Nt(s(m))\backslash Sf){{\uparrow}{}}Nt(s(m))){{\downarrow}{}}({\langle\rangle}Nt(s(m))\backslash Sf)$}}\ \Rightarrow\ Sf
\using {\downarrow}L
\endprooftree
\justifies
[Nt(s(m))], ({\langle\rangle}Nt(s(m))\backslash Sf)/{\exists}bNb, \mbox{\fbox{${\forall}f((({\langle\rangle}Nt(s(m))\backslash Sf){{\uparrow}{}}Nt(s(m))){{\downarrow}{}}({\langle\rangle}Nt(s(m))\backslash Sf))$}}\ \Rightarrow\ Sf
\using {\forall}L
\endprooftree
\justifies
[Nt(s(m))], ({\langle\rangle}Nt(s(m))\backslash Sf)/{\exists}bNb, \mbox{\fbox{${\blacksquare}{\forall}f((({\langle\rangle}Nt(s(m))\backslash Sf){{\uparrow}{}}Nt(s(m))){{\downarrow}{}}({\langle\rangle}Nt(s(m))\backslash Sf))$}}\ \Rightarrow\ Sf
\using {\blacksquare}L
\endprooftree
\justifies
{\langle\rangle}Nt(s(m)), ({\langle\rangle}Nt(s(m))\backslash Sf)/{\exists}bNb, {\blacksquare}{\forall}f((({\langle\rangle}Nt(s(m))\backslash Sf){{\uparrow}{}}Nt(s(m))){{\downarrow}{}}({\langle\rangle}Nt(s(m))\backslash Sf))\ \Rightarrow\ Sf
\using {\langle\rangle}L
\endprooftree
\justifies
({\langle\rangle}Nt(s(m))\backslash Sf)/{\exists}bNb, {\blacksquare}{\forall}f((({\langle\rangle}Nt(s(m))\backslash Sf){{\uparrow}{}}Nt(s(m))){{\downarrow}{}}({\langle\rangle}Nt(s(m))\backslash Sf))\ \Rightarrow\ {\langle\rangle}Nt(s(m))\backslash Sf
\using {\backslash}R
\endprooftree
\justifies
{\blacksquare}{\forall}f((({\langle\rangle}Nt(s(m))\backslash Sf){{\uparrow}{}}Nt(s(m))){{\downarrow}{}}({\langle\rangle}Nt(s(m))\backslash Sf))\ \Rightarrow\ (({\langle\rangle}Nt(s(m))\backslash Sf)/{\exists}bNb)\backslash ({\langle\rangle}Nt(s(m))\backslash Sf)
\using {\backslash}R
\endprooftree
\justifies
\begin{array}{c}
{\blacksquare}{\forall}f((({\langle\rangle}Nt(s(m))\backslash Sf){{\uparrow}{}}Nt(s(m))){{\downarrow}{}}({\langle\rangle}Nt(s(m))\backslash Sf))\ \Rightarrow\ {\blacksquare}((({\langle\rangle}Nt(s(m))\backslash Sf)/{\exists}bNb)\backslash ({\langle\rangle}Nt(s(m))\backslash Sf))\\
\mbox{\footnotesize\textcircled{1}}
\end{array}
\using {\blacksquare}R
\endprooftree

\prooftree
\prooftree
\prooftree
\prooftree
\prooftree
\prooftree
\prooftree
\prooftree
\prooftree
\justifies
\mbox{\fbox{$Nt(s(f))$}}\ \Rightarrow\ Nt(s(f))
\endprooftree
\justifies
\mbox{\fbox{${\blacksquare}Nt(s(f))$}}\ \Rightarrow\ Nt(s(f))
\using {\blacksquare}L
\endprooftree
\justifies
{\blacksquare}Nt(s(f))\ \Rightarrow\ \fbox{${\exists}bNb$}
\using {\exists}R
\endprooftree
\prooftree
\prooftree
\prooftree
\justifies
Nt(s(m))\ \Rightarrow\ Nt(s(m))
\endprooftree
\justifies
[Nt(s(m))]\ \Rightarrow\ \fbox{${\langle\rangle}Nt(s(m))$}
\using {\langle\rangle}R
\endprooftree
\prooftree
\justifies
\mbox{\fbox{$Sf$}}\ \Rightarrow\ Sf
\endprooftree
\justifies
[Nt(s(m))], \mbox{\fbox{${\langle\rangle}Nt(s(m))\backslash Sf$}}\ \Rightarrow\ Sf
\using {\backslash}L
\endprooftree
\justifies
[Nt(s(m))], \mbox{\fbox{$({\langle\rangle}Nt(s(m))\backslash Sf)/{\exists}bNb$}}, {\blacksquare}Nt(s(f))\ \Rightarrow\ Sf
\using {/}L
\endprooftree
\justifies
{\langle\rangle}Nt(s(m)), ({\langle\rangle}Nt(s(m))\backslash Sf)/{\exists}bNb, {\blacksquare}Nt(s(f))\ \Rightarrow\ Sf
\using {\langle\rangle}L
\endprooftree
\justifies
({\langle\rangle}Nt(s(m))\backslash Sf)/{\exists}bNb, {\blacksquare}Nt(s(f))\ \Rightarrow\ {\langle\rangle}Nt(s(m))\backslash Sf
\using {\backslash}R
\endprooftree
\justifies
{\blacksquare}Nt(s(f))\ \Rightarrow\ (({\langle\rangle}Nt(s(m))\backslash Sf)/{\exists}bNb)\backslash ({\langle\rangle}Nt(s(m))\backslash Sf)
\using {\backslash}R
\endprooftree
\justifies
{\blacksquare}Nt(s(f))\ \Rightarrow\ {\blacksquare}((({\langle\rangle}Nt(s(m))\backslash Sf)/{\exists}bNb)\backslash ({\langle\rangle}Nt(s(m))\backslash Sf))
\using {\blacksquare}R
\endprooftree
\justifies
\begin{array}{c}
{\blacksquare}Nt(s(f))\ \Rightarrow\ \fbox{$?{\blacksquare}((({\langle\rangle}Nt(s(m))\backslash Sf)/{\exists}bNb)\backslash ({\langle\rangle}Nt(s(m))\backslash Sf))$}\\
\mbox{\footnotesize\textcircled{2}}
\end{array}
\using {?}R
\endprooftree
\end{center}}

\begin{center}
\rotatebox{-90}{\tiny
\prooftree
\prooftree
\prooftree
\prooftree
\mbox{\footnotesize\textcircled{1}}\tab
\prooftree
\mbox{\footnotesize\textcircled{2}}\tab
\prooftree
\prooftree
\prooftree
\prooftree
\prooftree
\prooftree
\prooftree
\prooftree
\prooftree
\prooftree
\prooftree
\justifies
N1\ \Rightarrow\ N1
\endprooftree
\justifies
N1\ \Rightarrow\ \fbox{${\exists}aNa$}
\using {\exists}R
\endprooftree
\prooftree
\prooftree
\prooftree
\prooftree
\justifies
Nt(s(m))\ \Rightarrow\ Nt(s(m))
\endprooftree
\justifies
Nt(s(m))\ \Rightarrow\ \fbox{${\exists}gNt(s(g))$}
\using {\exists}R
\endprooftree
\justifies
[Nt(s(m))]\ \Rightarrow\ \fbox{${\langle\rangle}{\exists}gNt(s(g))$}
\using {\langle\rangle}R
\endprooftree
\prooftree
\justifies
\mbox{\fbox{$Sf$}}\ \Rightarrow\ Sf
\endprooftree
\justifies
[Nt(s(m))], \mbox{\fbox{${\langle\rangle}{\exists}gNt(s(g))\backslash Sf$}}\ \Rightarrow\ Sf
\using {\backslash}L
\endprooftree
\justifies
[Nt(s(m))], \mbox{\fbox{$({\langle\rangle}{\exists}gNt(s(g))\backslash Sf)/{\exists}aNa$}}, N1\ \Rightarrow\ Sf
\using {/}L
\endprooftree
\justifies
[Nt(s(m))], \mbox{\fbox{${\square}(({\langle\rangle}{\exists}gNt(s(g))\backslash Sf)/{\exists}aNa)$}}, N1\ \Rightarrow\ Sf
\using {\Box}L
\endprooftree
\justifies
[Nt(s(m))], {\square}(({\langle\rangle}{\exists}gNt(s(g))\backslash Sf)/{\exists}aNa), {\exists}bNb\ \Rightarrow\ Sf
\using {\exists}L
\endprooftree
\justifies
{\langle\rangle}Nt(s(m)), {\square}(({\langle\rangle}{\exists}gNt(s(g))\backslash Sf)/{\exists}aNa), {\exists}bNb\ \Rightarrow\ Sf
\using {\langle\rangle}L
\endprooftree
\justifies
{\square}(({\langle\rangle}{\exists}gNt(s(g))\backslash Sf)/{\exists}aNa), {\exists}bNb\ \Rightarrow\ {\langle\rangle}Nt(s(m))\backslash Sf
\using {\backslash}R
\endprooftree
\justifies
{\square}(({\langle\rangle}{\exists}gNt(s(g))\backslash Sf)/{\exists}aNa)\ \Rightarrow\ ({\langle\rangle}Nt(s(m))\backslash Sf)/{\exists}bNb
\using {/}R
\endprooftree
\prooftree
\prooftree
\prooftree
\prooftree
\justifies
\mbox{\fbox{$Nt(s(m))$}}\ \Rightarrow\ Nt(s(m))
\endprooftree
\justifies
\mbox{\fbox{${\blacksquare}Nt(s(m))$}}\ \Rightarrow\ Nt(s(m))
\using {\blacksquare}L
\endprooftree
\justifies
[{\blacksquare}Nt(s(m))]\ \Rightarrow\ \fbox{${\langle\rangle}Nt(s(m))$}
\using {\langle\rangle}R
\endprooftree
\prooftree
\justifies
\mbox{\fbox{$Sf$}}\ \Rightarrow\ Sf
\endprooftree
\justifies
[{\blacksquare}Nt(s(m))], \mbox{\fbox{${\langle\rangle}Nt(s(m))\backslash Sf$}}\ \Rightarrow\ Sf
\using {\backslash}L
\endprooftree
\justifies
[{\blacksquare}Nt(s(m))], {\square}(({\langle\rangle}{\exists}gNt(s(g))\backslash Sf)/{\exists}aNa), \mbox{\fbox{$(({\langle\rangle}Nt(s(m))\backslash Sf)/{\exists}bNb)\backslash ({\langle\rangle}Nt(s(m))\backslash Sf)$}}\ \Rightarrow\ Sf
\using {\backslash}L
\endprooftree
\justifies
[{\blacksquare}Nt(s(m))], {\square}(({\langle\rangle}{\exists}gNt(s(g))\backslash Sf)/{\exists}aNa), [\mbox{\fbox{${[]^{-1}}((({\langle\rangle}Nt(s(m))\backslash Sf)/{\exists}bNb)\backslash ({\langle\rangle}Nt(s(m))\backslash Sf))$}}]\ \Rightarrow\ Sf
\using {[]^{-1}}L
\endprooftree
\justifies
[{\blacksquare}Nt(s(m))], {\square}(({\langle\rangle}{\exists}gNt(s(g))\backslash Sf)/{\exists}aNa), [[\mbox{\fbox{${[]^{-1}}{[]^{-1}}((({\langle\rangle}Nt(s(m))\backslash Sf)/{\exists}bNb)\backslash ({\langle\rangle}Nt(s(m))\backslash Sf))$}}]]\ \Rightarrow\ Sf
\using {[]^{-1}}L
\endprooftree\justifies
[{\blacksquare}Nt(s(m))], {\square}(({\langle\rangle}{\exists}gNt(s(g))\backslash Sf)/{\exists}aNa), [[{\blacksquare}Nt(s(f)), \mbox{\fbox{$?{\blacksquare}((({\langle\rangle}Nt(s(m))\backslash Sf)/{\exists}bNb)\backslash ({\langle\rangle}Nt(s(m))\backslash Sf))\backslash {[]^{-1}}{[]^{-1}}((({\langle\rangle}Nt(s(m))\backslash Sf)/{\exists}bNb)\backslash ({\langle\rangle}Nt(s(m))\backslash Sf))$}}]]\ \Rightarrow\ Sf
\using {\backslash}L
\endprooftree
\justifies
[{\blacksquare}Nt(s(m))], {\square}(({\langle\rangle}{\exists}gNt(s(g))\backslash Sf)/{\exists}aNa), [[{\blacksquare}Nt(s(f)), \mbox{\fbox{$(?{\blacksquare}((({\langle\rangle}Nt(s(m))\backslash Sf)/{\exists}bNb)\backslash ({\langle\rangle}Nt(s(m))\backslash Sf))\backslash {[]^{-1}}{[]^{-1}}((({\langle\rangle}Nt(s(m))\backslash Sf)/{\exists}bNb)\backslash ({\langle\rangle}Nt(s(m))\backslash Sf)))/{\blacksquare}((({\langle\rangle}Nt(s(m))\backslash Sf)/{\exists}bNb)\backslash ({\langle\rangle}Nt(s(m))\backslash Sf))$}}, {\blacksquare}{\forall}f((({\langle\rangle}Nt(s(m))\backslash Sf){{\uparrow}{}}Nt(s(m))){{\downarrow}{}}({\langle\rangle}Nt(s(m))\backslash Sf))]]\ \Rightarrow\ Sf
\using {/}L
\endprooftree
\justifies
[{\blacksquare}Nt(s(m))], {\square}(({\langle\rangle}{\exists}gNt(s(g))\backslash Sf)/{\exists}aNa), [[{\blacksquare}Nt(s(f)), \mbox{\fbox{${\forall}a((?{\blacksquare}((({\langle\rangle}Na\backslash Sf)/{\exists}bNb)\backslash ({\langle\rangle}Na\backslash Sf))\backslash {[]^{-1}}{[]^{-1}}((({\langle\rangle}Na\backslash Sf)/{\exists}bNb)\backslash ({\langle\rangle}Na\backslash Sf)))/{\blacksquare}((({\langle\rangle}Na\backslash Sf)/{\exists}bNb)\backslash ({\langle\rangle}Na\backslash Sf)))$}}, {\blacksquare}{\forall}f((({\langle\rangle}Nt(s(m))\backslash Sf){{\uparrow}{}}Nt(s(m))){{\downarrow}{}}({\langle\rangle}Nt(s(m))\backslash Sf))]]\ \Rightarrow\ Sf
\using {\forall}L
\endprooftree
\justifies
[{\blacksquare}Nt(s(m))], {\square}(({\langle\rangle}{\exists}gNt(s(g))\backslash Sf)/{\exists}aNa), [[{\blacksquare}Nt(s(f)), \mbox{\fbox{${\forall}f{\forall}a((?{\blacksquare}((({\langle\rangle}Na\backslash Sf)/{\exists}bNb)\backslash ({\langle\rangle}Na\backslash Sf))\backslash {[]^{-1}}{[]^{-1}}((({\langle\rangle}Na\backslash Sf)/{\exists}bNb)\backslash ({\langle\rangle}Na\backslash Sf)))/{\blacksquare}((({\langle\rangle}Na\backslash Sf)/{\exists}bNb)\backslash ({\langle\rangle}Na\backslash Sf)))$}}, {\blacksquare}{\forall}f((({\langle\rangle}Nt(s(m))\backslash Sf){{\uparrow}{}}Nt(s(m))){{\downarrow}{}}({\langle\rangle}Nt(s(m))\backslash Sf))]]\ \Rightarrow\ Sf
\using {\forall}L
\endprooftree
\justifies
[{\blacksquare}Nt(s(m))], {\square}(({\langle\rangle}{\exists}gNt(s(g))\backslash Sf)/{\exists}aNa), [[{\blacksquare}Nt(s(f)), \mbox{\fbox{${\blacksquare}{\forall}f{\forall}a((?{\blacksquare}((({\langle\rangle}Na\backslash Sf)/{\exists}bNb)\backslash ({\langle\rangle}Na\backslash Sf))\backslash {[]^{-1}}{[]^{-1}}((({\langle\rangle}Na\backslash Sf)/{\exists}bNb)\backslash ({\langle\rangle}Na\backslash Sf)))/{\blacksquare}((({\langle\rangle}Na\backslash Sf)/{\exists}bNb)\backslash ({\langle\rangle}Na\backslash Sf)))$}}, {\blacksquare}{\forall}f((({\langle\rangle}Nt(s(m))\backslash Sf){{\uparrow}{}}Nt(s(m))){{\downarrow}{}}({\langle\rangle}Nt(s(m))\backslash Sf))]]\ \Rightarrow\ Sf
\using {\blacksquare}L
\endprooftree}
\end{center}

\vspace{0.15in}

\noindent
Again the main left (\textcircled{1}), main middle (\textcircled{2}) and main right
subderivations are for the right conjunt, left conjunct, and context respectively.
The subtree \textcircled{1} essentially derives that a subject-oriented reflexive
$((N\bsl S)\circum N)\infix(N\bsl S)$ yields a lifted object type $((N\bsl S)/N)\bsl(N\bsl S)$:
\disp{\mini
\prooftree
\prooftree
\prooftree
\prooftree
(N\bsl S)/N, N\yields N\bsl S
\justifies
(N\bsl S)/N, \sep\yields(N\bsl S)\circum N
\using \circum R
\endprooftree\tb
N, N\bsl S\yields S
\justifies
N, (N\bsl S)/N, ((N\bsl S)\circum N)\infix(N\bsl S)\yields S
\using \infix L
\endprooftree
\justifies
(N\bsl S)/N, ((N\bsl S)\circum N)\infix(N\bsl S)\yields N\bsl S
\using \bsl R
\endprooftree
\justifies
((N\bsl S)\circum N)\infix(N\bsl S)\yields ((N\bsl S)/N)\bsl(N\bsl S)
\using \bsl R
\endprooftree}
The derivation centres on the successive $\infix L$ and $\circum R$ half way up.
The subtree \textcircled{2} essentially derives that a nominal $N$ yields a lifted object type
$((N\bsl S)/N)\bsl(N\bsl S)$:
\disp{$
N\yields ((N\bsl S)/N)\bsl(N\bsl S)$}
The remaining main subtree checks the brackets and applies the coordinate structure
to the transitive verb (deriving TV\yields TV) and subject (deriving N\yields N)
contexts.
All this delivers the correct semantics:
\disp{
$[({\it Pres}\ ((\mbox{\v{}}{\it love}\ {\it m})\ {\it j}))\wedge ({\it Pres}\ ((\mbox{\v{}}{\it love}\ {\it j})\ {\it j}))]$}

\subsection{Right node raising coordination}

The next example is an instance of right node raising:
\disp{
$[[[{\bf john}]{+}{\bf likes}{+}{\bf and}{+}[{\bf mary}]{+}{\bf loves}]]{+}{\bf london}: Sf$}
Appropriate lexical lookup yields  the following where the coordinator type is
essentially $(\exstexp X\bsl$ $\abrack\abrack X)/X$ with $X=S/N$.
\disp{
$\begin{array}[t]{l}
[[[{\blacksquare}Nt(s(m)): {\it j}], {\square}(({\langle\rangle}{\exists}gNt(s(g))\backslash Sf)/{\exists}aNa): \mbox{\^{}}\lambda A\lambda B({\it Pres}\ ((\mbox{\v{}}{\it like}\ {\it A})\ {\it B})),\\
 {\blacksquare}{\forall}f((?{\blacksquare}(Sf/{\exists}aNa)\backslash {[]^{-1}}{[]^{-1}}(Sf/{\exists}aNa))/{\blacksquare}(Sf/{\exists}aNa)): (\Phinplus\ ({\it s}\ {\it 0})\ {\it and}),\\{} [{\blacksquare}Nt(s(f)): {\it m}], {\square}(({\langle\rangle}{\exists}gNt(s(g))\backslash Sf)/{\exists}aNa): \mbox{\^{}}\lambda C\lambda D({\it Pres}\ ((\mbox{\v{}}{\it love}\ {\it C})\ {\it D}))]],\\ {\blacksquare}Nt(s(n)): {\it l}\ \Rightarrow\ Sf
 \end{array}$}
There is the derivation below.

\vspace*{0.15in}

\begin{center}
\rotatebox{-90}{\tiny
\prooftree
\prooftree
\prooftree
\prooftree
\prooftree
\prooftree
\prooftree
\prooftree
\prooftree
\prooftree
\justifies
N2\ \Rightarrow\ N2
\endprooftree
\justifies
N2\ \Rightarrow\ \fbox{${\exists}aNa$}
\using {\exists}R
\endprooftree
\prooftree
\prooftree
\prooftree
\prooftree
\prooftree
\justifies
\mbox{\fbox{$Nt(s(f))$}}\ \Rightarrow\ Nt(s(f))
\endprooftree
\justifies
\mbox{\fbox{${\blacksquare}Nt(s(f))$}}\ \Rightarrow\ Nt(s(f))
\using {\blacksquare}L
\endprooftree
\justifies
{\blacksquare}Nt(s(f))\ \Rightarrow\ \fbox{${\exists}gNt(s(g))$}
\using {\exists}R
\endprooftree
\justifies
[{\blacksquare}Nt(s(f))]\ \Rightarrow\ \fbox{${\langle\rangle}{\exists}gNt(s(g))$}
\using {\langle\rangle}R
\endprooftree
\prooftree
\justifies
\mbox{\fbox{$Sf$}}\ \Rightarrow\ Sf
\endprooftree
\justifies
[{\blacksquare}Nt(s(f))], \mbox{\fbox{${\langle\rangle}{\exists}gNt(s(g))\backslash Sf$}}\ \Rightarrow\ Sf
\using {\backslash}L
\endprooftree
\justifies
[{\blacksquare}Nt(s(f))], \mbox{\fbox{$({\langle\rangle}{\exists}gNt(s(g))\backslash Sf)/{\exists}aNa$}}, N2\ \Rightarrow\ Sf
\using {/}L
\endprooftree
\justifies
[{\blacksquare}Nt(s(f))], \mbox{\fbox{${\square}(({\langle\rangle}{\exists}gNt(s(g))\backslash Sf)/{\exists}aNa)$}}, N2\ \Rightarrow\ Sf
\using {\Box}L
\endprooftree
\justifies
[{\blacksquare}Nt(s(f))], {\square}(({\langle\rangle}{\exists}gNt(s(g))\backslash Sf)/{\exists}aNa), {\exists}aNa\ \Rightarrow\ Sf
\using {\exists}L
\endprooftree
\justifies
[{\blacksquare}Nt(s(f))], {\square}(({\langle\rangle}{\exists}gNt(s(g))\backslash Sf)/{\exists}aNa)\ \Rightarrow\ Sf/{\exists}aNa
\using {/}R
\endprooftree
\justifies
[{\blacksquare}Nt(s(f))], {\square}(({\langle\rangle}{\exists}gNt(s(g))\backslash Sf)/{\exists}aNa)\ \Rightarrow\ {\blacksquare}(Sf/{\exists}aNa)
\using {\blacksquare}R
\endprooftree
\prooftree
\prooftree
\prooftree
\prooftree
\prooftree
\prooftree
\prooftree
\prooftree
\prooftree
\justifies
N3\ \Rightarrow\ N3
\endprooftree
\justifies
N3\ \Rightarrow\ \fbox{${\exists}aNa$}
\using {\exists}R
\endprooftree
\prooftree
\prooftree
\prooftree
\prooftree
\prooftree
\justifies
\mbox{\fbox{$Nt(s(m))$}}\ \Rightarrow\ Nt(s(m))
\endprooftree
\justifies
\mbox{\fbox{${\blacksquare}Nt(s(m))$}}\ \Rightarrow\ Nt(s(m))
\using {\blacksquare}L
\endprooftree
\justifies
{\blacksquare}Nt(s(m))\ \Rightarrow\ \fbox{${\exists}gNt(s(g))$}
\using {\exists}R
\endprooftree
\justifies
[{\blacksquare}Nt(s(m))]\ \Rightarrow\ \fbox{${\langle\rangle}{\exists}gNt(s(g))$}
\using {\langle\rangle}R
\endprooftree
\prooftree
\justifies
\mbox{\fbox{$Sf$}}\ \Rightarrow\ Sf
\endprooftree
\justifies
[{\blacksquare}Nt(s(m))], \mbox{\fbox{${\langle\rangle}{\exists}gNt(s(g))\backslash Sf$}}\ \Rightarrow\ Sf
\using {\backslash}L
\endprooftree
\justifies
[{\blacksquare}Nt(s(m))], \mbox{\fbox{$({\langle\rangle}{\exists}gNt(s(g))\backslash Sf)/{\exists}aNa$}}, N3\ \Rightarrow\ Sf
\using {/}L
\endprooftree
\justifies
[{\blacksquare}Nt(s(m))], \mbox{\fbox{${\square}(({\langle\rangle}{\exists}gNt(s(g))\backslash Sf)/{\exists}aNa)$}}, N3\ \Rightarrow\ Sf
\using {\Box}L
\endprooftree
\justifies
[{\blacksquare}Nt(s(m))], {\square}(({\langle\rangle}{\exists}gNt(s(g))\backslash Sf)/{\exists}aNa), {\exists}aNa\ \Rightarrow\ Sf
\using {\exists}L
\endprooftree
\justifies
[{\blacksquare}Nt(s(m))], {\square}(({\langle\rangle}{\exists}gNt(s(g))\backslash Sf)/{\exists}aNa)\ \Rightarrow\ Sf/{\exists}aNa
\using {/}R
\endprooftree
\justifies
[{\blacksquare}Nt(s(m))], {\square}(({\langle\rangle}{\exists}gNt(s(g))\backslash Sf)/{\exists}aNa)\ \Rightarrow\ {\blacksquare}(Sf/{\exists}aNa)
\using {\blacksquare}R
\endprooftree
\justifies
[{\blacksquare}Nt(s(m))], {\square}(({\langle\rangle}{\exists}gNt(s(g))\backslash Sf)/{\exists}aNa)\ \Rightarrow\ \fbox{$?{\blacksquare}(Sf/{\exists}aNa)$}
\using {?}R
\endprooftree
\prooftree
\prooftree
\prooftree
\prooftree
\prooftree
\prooftree
\justifies
\mbox{\fbox{$Nt(s(n))$}}\ \Rightarrow\ Nt(s(n))
\endprooftree
\justifies
\mbox{\fbox{${\blacksquare}Nt(s(n))$}}\ \Rightarrow\ Nt(s(n))
\using {\blacksquare}L
\endprooftree
\justifies
{\blacksquare}Nt(s(n))\ \Rightarrow\ \fbox{${\exists}aNa$}
\using {\exists}R
\endprooftree
\prooftree
\justifies
\mbox{\fbox{$Sf$}}\ \Rightarrow\ Sf
\endprooftree
\justifies
\mbox{\fbox{$Sf/{\exists}aNa$}}, {\blacksquare}Nt(s(n))\ \Rightarrow\ Sf
\using {/}L
\endprooftree
\justifies
[\mbox{\fbox{${[]^{-1}}(Sf/{\exists}aNa)$}}], {\blacksquare}Nt(s(n))\ \Rightarrow\ Sf
\using {[]^{-1}}L
\endprooftree
\justifies
[[\mbox{\fbox{${[]^{-1}}{[]^{-1}}(Sf/{\exists}aNa)$}}]], {\blacksquare}Nt(s(n))\ \Rightarrow\ Sf
\using {[]^{-1}}L
\endprooftree
\justifies
[[[{\blacksquare}Nt(s(m))], {\square}(({\langle\rangle}{\exists}gNt(s(g))\backslash Sf)/{\exists}aNa), \mbox{\fbox{$?{\blacksquare}(Sf/{\exists}aNa)\backslash {[]^{-1}}{[]^{-1}}(Sf/{\exists}aNa)$}}]], {\blacksquare}Nt(s(n))\ \Rightarrow\ Sf
\using {\backslash}L
\endprooftree
\justifies
[[[{\blacksquare}Nt(s(m))], {\square}(({\langle\rangle}{\exists}gNt(s(g))\backslash Sf)/{\exists}aNa), \mbox{\fbox{$(?{\blacksquare}(Sf/{\exists}aNa)\backslash {[]^{-1}}{[]^{-1}}(Sf/{\exists}aNa))/{\blacksquare}(Sf/{\exists}aNa)$}}, [{\blacksquare}Nt(s(f))], {\square}(({\langle\rangle}{\exists}gNt(s(g))\backslash Sf)/{\exists}aNa)]], {\blacksquare}Nt(s(n))\ \Rightarrow\ Sf
\using {/}L
\endprooftree
\justifies
[[[{\blacksquare}Nt(s(m))], {\square}(({\langle\rangle}{\exists}gNt(s(g))\backslash Sf)/{\exists}aNa), \mbox{\fbox{${\forall}f((?{\blacksquare}(Sf/{\exists}aNa)\backslash {[]^{-1}}{[]^{-1}}(Sf/{\exists}aNa))/{\blacksquare}(Sf/{\exists}aNa))$}}, [{\blacksquare}Nt(s(f))], {\square}(({\langle\rangle}{\exists}gNt(s(g))\backslash Sf)/{\exists}aNa)]], {\blacksquare}Nt(s(n))\ \Rightarrow\ Sf
\using {\forall}L
\endprooftree
\justifies
[[[{\blacksquare}Nt(s(m))], {\square}(({\langle\rangle}{\exists}gNt(s(g))\backslash Sf)/{\exists}aNa), \mbox{\fbox{${\blacksquare}{\forall}f((?{\blacksquare}(Sf/{\exists}aNa)\backslash {[]^{-1}}{[]^{-1}}(Sf/{\exists}aNa))/{\blacksquare}(Sf/{\exists}aNa))$}}, [{\blacksquare}Nt(s(f))], {\square}(({\langle\rangle}{\exists}gNt(s(g))\backslash Sf)/{\exists}aNa)]], {\blacksquare}Nt(s(n))\ \Rightarrow\ Sf
\using {\blacksquare}L
\endprooftree}
\end{center}

\vspace*{0.15in}

As ever the left and middle subderivations are for the right and left conjuncts
respectively, and involve in this case essentially the derivation of
$N, \TV \yields S/N$ which proceeds:
\disp{$
\prooftree
N, \TV, N\yields S
\justifies
N, \TV\yields S/N
\using /R
\endprooftree$}
The right, context, subderivation, applies the coordinate structure of type, basically,
$S/N$ to its right node raised $N$.
All this assigns semantics:
\disp{
$[({\it Pres}\ ((\mbox{\v{}}{\it like}\ {\it l})\ {\it j}))\wedge ({\it Pres}\ ((\mbox{\v{}}{\it love}\ {\it l})\ {\it m}))]$}

\commentout{

There follows conjunction of complex prepositional transitive verb phrases:
\disp{
(crd(21)) $[{\bf john}]{+}[[{\bf gave}{+}{\bf the}{+}{\bf book}{+}{\bf and}{+}{\bf sent}{+}{\bf the}{+}{\bf cd}]]{+}{\bf to}{+}{\bf mary}: Sf$}
Appropriate lexical lookup yields the following where the
coordinator is essentially of the form $(\exstexp X\bsl$ $\abrack\abrack X)/X$ where
$X=$.
\disp{
$[{\blacksquare}Nt(s(m)): {\it j}], [[{\square}(({\langle\rangle}{\exists}aNa\backslash Sf)/({\exists}bNb{\bullet}{\it PP}{\it to})): \mbox{\^{}}\lambda A\lambda B({\it Past}\ (((\mbox{\v{}}{\it give}\ \pi_2{\it A})\ \pi_1{\it A})\ {\it B})), {\blacksquare}{\forall}n(Nt(n)/{\it CN}{\it n}): \iota , {\square}{\it CN}{\it s(n)}: {\it book}, {\blacksquare}{\forall}a{\forall}b{\forall}f((?{\blacksquare}(({\langle\rangle}Na\backslash Sf)/{\it PP}{\it b})\backslash {[]^{-1}}{[]^{-1}}(({\langle\rangle}Na\backslash Sf)/{\it PP}{\it b}))/{\blacksquare}(({\langle\rangle}Na\backslash Sf)/{\it PP}{\it b})): (\Phinplus\ ({\it s}\ ({\it s}\ {\it 0}))\ {\it and}), {\square}(({\langle\rangle}{\exists}aNa\backslash Sf)/({\exists}bNb{\bullet}{\it PP}{\it to})): \mbox{\^{}}\lambda C\lambda D({\it Past}\ (((\mbox{\v{}}{\it sent}\ \pi_2{\it C})\ \pi_1{\it C})\ {\it D})), {\blacksquare}{\forall}n(Nt(n)/{\it CN}{\it n}): \iota , {\square}{\it CN}{\it s(n)}: {\it cd}]], {\blacksquare}(({\it PP}{\it to}/{\exists}aNa){\sqcap}{\forall}n(({\langle\rangle}Nn\backslash Si)/({\langle\rangle}Nn\backslash Sb))): \lambda E{\it E}, {\blacksquare}Nt(s(f)): {\it m}\ \Rightarrow\ Sf$}
There is the derivation:

\clearpage

\vspace{0.15in}

{\tiny
\prooftree
\prooftree
\prooftree
\prooftree
\prooftree
\prooftree
\prooftree
\prooftree
\prooftree
\prooftree
\prooftree
\prooftree
\prooftree
\justifies
\mbox{\fbox{${\it CN}{\it s(n)}$}}\ \Rightarrow\ {\it CN}{\it s(n)}
\endprooftree
\justifies
\mbox{\fbox{${\square}{\it CN}{\it s(n)}$}}\ \Rightarrow\ {\it CN}{\it s(n)}
\using {\Box}L
\endprooftree
\prooftree
\justifies
\mbox{\fbox{$Nt(s(n))$}}\ \Rightarrow\ Nt(s(n))
\endprooftree
\justifies
\mbox{\fbox{$Nt(s(n))/{\it CN}{\it s(n)}$}}, {\square}{\it CN}{\it s(n)}\ \Rightarrow\ Nt(s(n))
\using {/}L
\endprooftree
\justifies
\mbox{\fbox{${\forall}n(Nt(n)/{\it CN}{\it n})$}}, {\square}{\it CN}{\it s(n)}\ \Rightarrow\ Nt(s(n))
\using {\forall}L
\endprooftree
\justifies
\mbox{\fbox{${\blacksquare}{\forall}n(Nt(n)/{\it CN}{\it n})$}}, {\square}{\it CN}{\it s(n)}\ \Rightarrow\ Nt(s(n))
\using {\blacksquare}L
\endprooftree
\justifies
{\blacksquare}{\forall}n(Nt(n)/{\it CN}{\it n}), {\square}{\it CN}{\it s(n)}\ \Rightarrow\ \fbox{${\exists}bNb$}
\using {\exists}R
\endprooftree
\prooftree
\justifies
{\it PP}{\it to}\ \Rightarrow\ {\it PP}{\it to}
\endprooftree
\justifies
{\blacksquare}{\forall}n(Nt(n)/{\it CN}{\it n}), {\square}{\it CN}{\it s(n)}, {\it PP}{\it to}\ \Rightarrow\ \fbox{${\exists}bNb{\bullet}{\it PP}{\it to}$}
\using {\bullet}R
\endprooftree
\prooftree
\prooftree
\prooftree
\prooftree
\justifies
Nt(s(m))\ \Rightarrow\ Nt(s(m))
\endprooftree
\justifies
Nt(s(m))\ \Rightarrow\ \fbox{${\exists}aNa$}
\using {\exists}R
\endprooftree
\justifies
[Nt(s(m))]\ \Rightarrow\ \fbox{${\langle\rangle}{\exists}aNa$}
\using {\langle\rangle}R
\endprooftree
\prooftree
\justifies
\mbox{\fbox{$Sf$}}\ \Rightarrow\ Sf
\endprooftree
\justifies
[Nt(s(m))], \mbox{\fbox{${\langle\rangle}{\exists}aNa\backslash Sf$}}\ \Rightarrow\ Sf
\using {\backslash}L
\endprooftree
\justifies
[Nt(s(m))], \mbox{\fbox{$({\langle\rangle}{\exists}aNa\backslash Sf)/({\exists}bNb{\bullet}{\it PP}{\it to})$}}, {\blacksquare}{\forall}n(Nt(n)/{\it CN}{\it n}), {\square}{\it CN}{\it s(n)}, {\it PP}{\it to}\ \Rightarrow\ Sf
\using {/}L
\endprooftree
\justifies
[Nt(s(m))], \mbox{\fbox{${\square}(({\langle\rangle}{\exists}aNa\backslash Sf)/({\exists}bNb{\bullet}{\it PP}{\it to}))$}}, {\blacksquare}{\forall}n(Nt(n)/{\it CN}{\it n}), {\square}{\it CN}{\it s(n)}, {\it PP}{\it to}\ \Rightarrow\ Sf
\using {\Box}L
\endprooftree
\justifies
{\langle\rangle}Nt(s(m)), {\square}(({\langle\rangle}{\exists}aNa\backslash Sf)/({\exists}bNb{\bullet}{\it PP}{\it to})), {\blacksquare}{\forall}n(Nt(n)/{\it CN}{\it n}), {\square}{\it CN}{\it s(n)}, {\it PP}{\it to}\ \Rightarrow\ Sf
\using {\langle\rangle}L
\endprooftree
\justifies
{\square}(({\langle\rangle}{\exists}aNa\backslash Sf)/({\exists}bNb{\bullet}{\it PP}{\it to})), {\blacksquare}{\forall}n(Nt(n)/{\it CN}{\it n}), {\square}{\it CN}{\it s(n)}, {\it PP}{\it to}\ \Rightarrow\ {\langle\rangle}Nt(s(m))\backslash Sf
\using {\backslash}R
\endprooftree
\justifies
{\square}(({\langle\rangle}{\exists}aNa\backslash Sf)/({\exists}bNb{\bullet}{\it PP}{\it to})), {\blacksquare}{\forall}n(Nt(n)/{\it CN}{\it n}), {\square}{\it CN}{\it s(n)}\ \Rightarrow\ ({\langle\rangle}Nt(s(m))\backslash Sf)/{\it PP}{\it to}
\using {/}R
\endprooftree
\justifies
\begin{array}{c}
{\square}(({\langle\rangle}{\exists}aNa\backslash Sf)/({\exists}bNb{\bullet}{\it PP}{\it to})), {\blacksquare}{\forall}n(Nt(n)/{\it CN}{\it n}), {\square}{\it CN}{\it s(n)}\ \Rightarrow\ {\blacksquare}(({\langle\rangle}Nt(s(m))\backslash Sf)/{\it PP}{\it to})\\
\mbox{\footnotesize\textcircled{1}}
\end{array}
\using {\blacksquare}R
\endprooftree

\prooftree
\prooftree
\prooftree
\prooftree
\prooftree
\prooftree
\prooftree
\prooftree
\prooftree
\prooftree
\prooftree
\prooftree
\prooftree
\prooftree
\justifies
\mbox{\fbox{${\it CN}{\it s(n)}$}}\ \Rightarrow\ {\it CN}{\it s(n)}
\endprooftree
\justifies
\mbox{\fbox{${\square}{\it CN}{\it s(n)}$}}\ \Rightarrow\ {\it CN}{\it s(n)}
\using {\Box}L
\endprooftree
\prooftree
\justifies
\mbox{\fbox{$Nt(s(n))$}}\ \Rightarrow\ Nt(s(n))
\endprooftree
\justifies
\mbox{\fbox{$Nt(s(n))/{\it CN}{\it s(n)}$}}, {\square}{\it CN}{\it s(n)}\ \Rightarrow\ Nt(s(n))
\using {/}L
\endprooftree
\justifies
\mbox{\fbox{${\forall}n(Nt(n)/{\it CN}{\it n})$}}, {\square}{\it CN}{\it s(n)}\ \Rightarrow\ Nt(s(n))
\using {\forall}L
\endprooftree
\justifies
\mbox{\fbox{${\blacksquare}{\forall}n(Nt(n)/{\it CN}{\it n})$}}, {\square}{\it CN}{\it s(n)}\ \Rightarrow\ Nt(s(n))
\using {\blacksquare}L
\endprooftree
\justifies
{\blacksquare}{\forall}n(Nt(n)/{\it CN}{\it n}), {\square}{\it CN}{\it s(n)}\ \Rightarrow\ \fbox{${\exists}bNb$}
\using {\exists}R
\endprooftree
\prooftree
\justifies
{\it PP}{\it to}\ \Rightarrow\ {\it PP}{\it to}
\endprooftree
\justifies
{\blacksquare}{\forall}n(Nt(n)/{\it CN}{\it n}), {\square}{\it CN}{\it s(n)}, {\it PP}{\it to}\ \Rightarrow\ \fbox{${\exists}bNb{\bullet}{\it PP}{\it to}$}
\using {\bullet}R
\endprooftree
\prooftree
\prooftree
\prooftree
\prooftree
\justifies
Nt(s(m))\ \Rightarrow\ Nt(s(m))
\endprooftree
\justifies
Nt(s(m))\ \Rightarrow\ \fbox{${\exists}aNa$}
\using {\exists}R
\endprooftree
\justifies
[Nt(s(m))]\ \Rightarrow\ \fbox{${\langle\rangle}{\exists}aNa$}
\using {\langle\rangle}R
\endprooftree
\prooftree
\justifies
\mbox{\fbox{$Sf$}}\ \Rightarrow\ Sf
\endprooftree
\justifies
[Nt(s(m))], \mbox{\fbox{${\langle\rangle}{\exists}aNa\backslash Sf$}}\ \Rightarrow\ Sf
\using {\backslash}L
\endprooftree
\justifies
[Nt(s(m))], \mbox{\fbox{$({\langle\rangle}{\exists}aNa\backslash Sf)/({\exists}bNb{\bullet}{\it PP}{\it to})$}}, {\blacksquare}{\forall}n(Nt(n)/{\it CN}{\it n}), {\square}{\it CN}{\it s(n)}, {\it PP}{\it to}\ \Rightarrow\ Sf
\using {/}L
\endprooftree
\justifies
[Nt(s(m))], \mbox{\fbox{${\square}(({\langle\rangle}{\exists}aNa\backslash Sf)/({\exists}bNb{\bullet}{\it PP}{\it to}))$}}, {\blacksquare}{\forall}n(Nt(n)/{\it CN}{\it n}), {\square}{\it CN}{\it s(n)}, {\it PP}{\it to}\ \Rightarrow\ Sf
\using {\Box}L
\endprooftree
\justifies
{\langle\rangle}Nt(s(m)), {\square}(({\langle\rangle}{\exists}aNa\backslash Sf)/({\exists}bNb{\bullet}{\it PP}{\it to})), {\blacksquare}{\forall}n(Nt(n)/{\it CN}{\it n}), {\square}{\it CN}{\it s(n)}, {\it PP}{\it to}\ \Rightarrow\ Sf
\using {\langle\rangle}L
\endprooftree
\justifies
{\square}(({\langle\rangle}{\exists}aNa\backslash Sf)/({\exists}bNb{\bullet}{\it PP}{\it to})), {\blacksquare}{\forall}n(Nt(n)/{\it CN}{\it n}), {\square}{\it CN}{\it s(n)}, {\it PP}{\it to}\ \Rightarrow\ {\langle\rangle}Nt(s(m))\backslash Sf
\using {\backslash}R
\endprooftree
\justifies
{\square}(({\langle\rangle}{\exists}aNa\backslash Sf)/({\exists}bNb{\bullet}{\it PP}{\it to})), {\blacksquare}{\forall}n(Nt(n)/{\it CN}{\it n}), {\square}{\it CN}{\it s(n)}\ \Rightarrow\ ({\langle\rangle}Nt(s(m))\backslash Sf)/{\it PP}{\it to}
\using {/}R
\endprooftree
\justifies
{\square}(({\langle\rangle}{\exists}aNa\backslash Sf)/({\exists}bNb{\bullet}{\it PP}{\it to})), {\blacksquare}{\forall}n(Nt(n)/{\it CN}{\it n}), {\square}{\it CN}{\it s(n)}\ \Rightarrow\ {\blacksquare}(({\langle\rangle}Nt(s(m))\backslash Sf)/{\it PP}{\it to})
\using {\blacksquare}R
\endprooftree
\justifies
\begin{array}{c}
{\square}(({\langle\rangle}{\exists}aNa\backslash Sf)/({\exists}bNb{\bullet}{\it PP}{\it to})), {\blacksquare}{\forall}n(Nt(n)/{\it CN}{\it n}), {\square}{\it CN}{\it s(n)}\ \Rightarrow\ \fbox{$?{\blacksquare}(({\langle\rangle}Nt(s(m))\backslash Sf)/{\it PP}{\it to})$}\\
\mbox{\footnotesize\textcircled{2}}
\end{array}
\using {?}R
\endprooftree

\rotatebox{-90}{\scriptsize
\prooftree
\prooftree
\prooftree
\prooftree
\prooftree
\mbox{\footnotesize\textcircled{1}}\tab
\prooftree
\mbox{\footnotesize\textcircled{2}}\tab
\prooftree
\prooftree
\prooftree
\prooftree
\prooftree
\prooftree
\prooftree
\prooftree
\prooftree
\justifies
\mbox{\fbox{$Nt(s(f))$}}\ \Rightarrow\ Nt(s(f))
\endprooftree
\justifies
\mbox{\fbox{${\blacksquare}Nt(s(f))$}}\ \Rightarrow\ Nt(s(f))
\using {\blacksquare}L
\endprooftree
\justifies
{\blacksquare}Nt(s(f))\ \Rightarrow\ \fbox{${\exists}aNa$}
\using {\exists}R
\endprooftree
\prooftree
\justifies
\mbox{\fbox{${\it PP}{\it to}$}}\ \Rightarrow\ {\it PP}{\it to}
\endprooftree
\justifies
\mbox{\fbox{${\it PP}{\it to}/{\exists}aNa$}}, {\blacksquare}Nt(s(f))\ \Rightarrow\ {\it PP}{\it to}
\using {/}L
\endprooftree
\justifies
\mbox{\fbox{$({\it PP}{\it to}/{\exists}aNa){\sqcap}{\forall}n(({\langle\rangle}Nn\backslash Si)/({\langle\rangle}Nn\backslash Sb))$}}, {\blacksquare}Nt(s(f))\ \Rightarrow\ {\it PP}{\it to}
\using {\sqcap}L
\endprooftree
\justifies
\mbox{\fbox{${\blacksquare}(({\it PP}{\it to}/{\exists}aNa){\sqcap}{\forall}n(({\langle\rangle}Nn\backslash Si)/({\langle\rangle}Nn\backslash Sb)))$}}, {\blacksquare}Nt(s(f))\ \Rightarrow\ {\it PP}{\it to}
\using {\blacksquare}L
\endprooftree
\prooftree
\prooftree
\prooftree
\prooftree
\justifies
\mbox{\fbox{$Nt(s(m))$}}\ \Rightarrow\ Nt(s(m))
\endprooftree
\justifies
\mbox{\fbox{${\blacksquare}Nt(s(m))$}}\ \Rightarrow\ Nt(s(m))
\using {\blacksquare}L
\endprooftree
\justifies
[{\blacksquare}Nt(s(m))]\ \Rightarrow\ \fbox{${\langle\rangle}Nt(s(m))$}
\using {\langle\rangle}R
\endprooftree
\prooftree
\justifies
\mbox{\fbox{$Sf$}}\ \Rightarrow\ Sf
\endprooftree
\justifies
[{\blacksquare}Nt(s(m))], \mbox{\fbox{${\langle\rangle}Nt(s(m))\backslash Sf$}}\ \Rightarrow\ Sf
\using {\backslash}L
\endprooftree
\justifies
[{\blacksquare}Nt(s(m))], \mbox{\fbox{$({\langle\rangle}Nt(s(m))\backslash Sf)/{\it PP}{\it to}$}}, {\blacksquare}(({\it PP}{\it to}/{\exists}aNa){\sqcap}{\forall}n(({\langle\rangle}Nn\backslash Si)/({\langle\rangle}Nn\backslash Sb))), {\blacksquare}Nt(s(f))\ \Rightarrow\ Sf
\using {/}L
\endprooftree
\justifies
[{\blacksquare}Nt(s(m))], [\mbox{\fbox{${[]^{-1}}(({\langle\rangle}Nt(s(m))\backslash Sf)/{\it PP}{\it to})$}}], {\blacksquare}(({\it PP}{\it to}/{\exists}aNa){\sqcap}{\forall}n(({\langle\rangle}Nn\backslash Si)/({\langle\rangle}Nn\backslash Sb))), {\blacksquare}Nt(s(f))\ \Rightarrow\ Sf
\using {[]^{-1}}L
\endprooftree
\justifies
[{\blacksquare}Nt(s(m))], [[\mbox{\fbox{${[]^{-1}}{[]^{-1}}(({\langle\rangle}Nt(s(m))\backslash Sf)/{\it PP}{\it to})$}}]], {\blacksquare}(({\it PP}{\it to}/{\exists}aNa){\sqcap}{\forall}n(({\langle\rangle}Nn\backslash Si)/({\langle\rangle}Nn\backslash Sb))), {\blacksquare}Nt(s(f))\ \Rightarrow\ Sf
\using {[]^{-1}}L
\endprooftree
\justifies
[{\blacksquare}Nt(s(m))], [[{\square}(({\langle\rangle}{\exists}aNa\backslash Sf)/({\exists}bNb{\bullet}{\it PP}{\it to})), {\blacksquare}{\forall}n(Nt(n)/{\it CN}{\it n}), {\square}{\it CN}{\it s(n)}, \mbox{\fbox{$?{\blacksquare}(({\langle\rangle}Nt(s(m))\backslash Sf)/{\it PP}{\it to})\backslash {[]^{-1}}{[]^{-1}}(({\langle\rangle}Nt(s(m))\backslash Sf)/{\it PP}{\it to})$}}]], {\blacksquare}(({\it PP}{\it to}/{\exists}aNa){\sqcap}{\forall}n(({\langle\rangle}Nn\backslash Si)/({\langle\rangle}Nn\backslash Sb))), {\blacksquare}Nt(s(f))\ \Rightarrow\ Sf
\using {\backslash}L
\endprooftree
\justifies
[{\blacksquare}Nt(s(m))], [[{\square}(({\langle\rangle}{\exists}aNa\backslash Sf)/({\exists}bNb{\bullet}{\it PP}{\it to})), {\blacksquare}{\forall}n(Nt(n)/{\it CN}{\it n}), {\square}{\it CN}{\it s(n)}, \mbox{\fbox{$(?{\blacksquare}(({\langle\rangle}Nt(s(m))\backslash Sf)/{\it PP}{\it to})\backslash {[]^{-1}}{[]^{-1}}(({\langle\rangle}Nt(s(m))\backslash Sf)/{\it PP}{\it to}))/{\blacksquare}(({\langle\rangle}Nt(s(m))\backslash Sf)/{\it PP}{\it to})$}}, {\square}(({\langle\rangle}{\exists}aNa\backslash Sf)/({\exists}bNb{\bullet}{\it PP}{\it to})), {\blacksquare}{\forall}n(Nt(n)/{\it CN}{\it n}), {\square}{\it CN}{\it s(n)}]], {\blacksquare}(({\it PP}{\it to}/{\exists}aNa){\sqcap}{\forall}n(({\langle\rangle}Nn\backslash Si)/({\langle\rangle}Nn\backslash Sb))), {\blacksquare}Nt(s(f))\ \Rightarrow\ Sf
\using {/}L
\endprooftree
\justifies
[{\blacksquare}Nt(s(m))], [[{\square}(({\langle\rangle}{\exists}aNa\backslash Sf)/({\exists}bNb{\bullet}{\it PP}{\it to})), {\blacksquare}{\forall}n(Nt(n)/{\it CN}{\it n}), {\square}{\it CN}{\it s(n)}, \mbox{\fbox{${\forall}f((?{\blacksquare}(({\langle\rangle}Nt(s(m))\backslash Sf)/{\it PP}{\it to})\backslash {[]^{-1}}{[]^{-1}}(({\langle\rangle}Nt(s(m))\backslash Sf)/{\it PP}{\it to}))/{\blacksquare}(({\langle\rangle}Nt(s(m))\backslash Sf)/{\it PP}{\it to}))$}}, {\square}(({\langle\rangle}{\exists}aNa\backslash Sf)/({\exists}bNb{\bullet}{\it PP}{\it to})), {\blacksquare}{\forall}n(Nt(n)/{\it CN}{\it n}), {\square}{\it CN}{\it s(n)}]], {\blacksquare}(({\it PP}{\it to}/{\exists}aNa){\sqcap}{\forall}n(({\langle\rangle}Nn\backslash Si)/({\langle\rangle}Nn\backslash Sb))), {\blacksquare}Nt(s(f))\ \Rightarrow\ Sf
\using {\forall}L
\endprooftree
\justifies
[{\blacksquare}Nt(s(m))], [[{\square}(({\langle\rangle}{\exists}aNa\backslash Sf)/({\exists}bNb{\bullet}{\it PP}{\it to})), {\blacksquare}{\forall}n(Nt(n)/{\it CN}{\it n}), {\square}{\it CN}{\it s(n)}, \mbox{\fbox{${\forall}b{\forall}f((?{\blacksquare}(({\langle\rangle}Nt(s(m))\backslash Sf)/{\it PP}{\it b})\backslash {[]^{-1}}{[]^{-1}}(({\langle\rangle}Nt(s(m))\backslash Sf)/{\it PP}{\it b}))/{\blacksquare}(({\langle\rangle}Nt(s(m))\backslash Sf)/{\it PP}{\it b}))$}}, {\square}(({\langle\rangle}{\exists}aNa\backslash Sf)/({\exists}bNb{\bullet}{\it PP}{\it to})), {\blacksquare}{\forall}n(Nt(n)/{\it CN}{\it n}), {\square}{\it CN}{\it s(n)}]], {\blacksquare}(({\it PP}{\it to}/{\exists}aNa){\sqcap}{\forall}n(({\langle\rangle}Nn\backslash Si)/({\langle\rangle}Nn\backslash Sb))), {\blacksquare}Nt(s(f))\ \Rightarrow\ Sf
\using {\forall}L
\endprooftree
\justifies
[{\blacksquare}Nt(s(m))], [[{\square}(({\langle\rangle}{\exists}aNa\backslash Sf)/({\exists}bNb{\bullet}{\it PP}{\it to})), {\blacksquare}{\forall}n(Nt(n)/{\it CN}{\it n}), {\square}{\it CN}{\it s(n)}, \mbox{\fbox{${\forall}a{\forall}b{\forall}f((?{\blacksquare}(({\langle\rangle}Na\backslash Sf)/{\it PP}{\it b})\backslash {[]^{-1}}{[]^{-1}}(({\langle\rangle}Na\backslash Sf)/{\it PP}{\it b}))/{\blacksquare}(({\langle\rangle}Na\backslash Sf)/{\it PP}{\it b}))$}}, {\square}(({\langle\rangle}{\exists}aNa\backslash Sf)/({\exists}bNb{\bullet}{\it PP}{\it to})), {\blacksquare}{\forall}n(Nt(n)/{\it CN}{\it n}), {\square}{\it CN}{\it s(n)}]], {\blacksquare}(({\it PP}{\it to}/{\exists}aNa){\sqcap}{\forall}n(({\langle\rangle}Nn\backslash Si)/({\langle\rangle}Nn\backslash Sb))), {\blacksquare}Nt(s(f))\ \Rightarrow\ Sf
\using {\forall}L
\endprooftree
\justifies
[{\blacksquare}Nt(s(m))], [[{\square}(({\langle\rangle}{\exists}aNa\backslash Sf)/({\exists}bNb{\bullet}{\it PP}{\it to})), {\blacksquare}{\forall}n(Nt(n)/{\it CN}{\it n}), {\square}{\it CN}{\it s(n)}, \mbox{\fbox{${\blacksquare}{\forall}a{\forall}b{\forall}f((?{\blacksquare}(({\langle\rangle}Na\backslash Sf)/{\it PP}{\it b})\backslash {[]^{-1}}{[]^{-1}}(({\langle\rangle}Na\backslash Sf)/{\it PP}{\it b}))/{\blacksquare}(({\langle\rangle}Na\backslash Sf)/{\it PP}{\it b}))$}}, {\square}(({\langle\rangle}{\exists}aNa\backslash Sf)/({\exists}bNb{\bullet}{\it PP}{\it to})), {\blacksquare}{\forall}n(Nt(n)/{\it CN}{\it n}), {\square}{\it CN}{\it s(n)}]], {\blacksquare}(({\it PP}{\it to}/{\exists}aNa){\sqcap}{\forall}n(({\langle\rangle}Nn\backslash Si)/({\langle\rangle}Nn\backslash Sb))), {\blacksquare}Nt(s(f))\ \Rightarrow\ Sf
\using {\blacksquare}L
\endprooftree}}

\vspace{0.15in}

\noindent
This delivers semantics:
\disp{
$[({\it Past}\ (((\mbox{\v{}}{\it give}\ {\it m})\ (\iota \ \mbox{\v{}}{\it book}))\ {\it j}))\wedge ({\it Past}\ (((\mbox{\v{}}{\it sent}\ {\it m})\ (\iota \ \mbox{\v{}}{\it cd}))\ {\it j}))]$}

}

\subsection{Argument-cluster left node raising coordination}

The following example is of non-standard constituent argument-cluster coordination,
or, left node raising: 
\disp{
$[{\bf john}]{+}{\bf gave}{+}[[{\bf the}{+}{\bf book}{+}{\bf to}{+}{\bf mary}{+}{\bf and}{+}{\bf the}{+}{\bf cd}{+}{\bf to}{+}{\bf suzy}]]: Sf$}
Appropriate lexical lookup yields the following where the coordinator type is
essentially $(\exstexp X\bsl$ $\abrack\abrack X)/X$ with $X=((N\bsl S)/(N\product\PP))\bsl(N\bsl S)$
(using the uncurried prepositional ditransitive verb type):
\disp{
$\begin{array}[t]{l}
[{\blacksquare}Nt(s(m)): {\it j}], {\square}(({\langle\rangle}{\exists}aNa\backslash Sf)/({\exists}bNb{\bullet}{\it PP}{\it to})): \mbox{\^{}}\lambda A\lambda B({\it Past}\ (((\mbox{\v{}}{\it give}\ \pi_2{\it A})\ \pi_1{\it A})\ {\it B})), \\
{}[[{\blacksquare}{\forall}n(Nt(n)/{\it CN}{\it n}): \iota , {\square}{\it CN}{\it s(n)}: {\it book}, {\blacksquare}(({\it PP}{\it to}/{\exists}aNa){\sqcap}{\forall}n(({\langle\rangle}Nn\backslash Si)/({\langle\rangle}Nn\backslash Sb))):\\ \lambda C{\it C}, {\blacksquare}Nt(s(f)): {\it m},
 {\blacksquare}{\forall}a{\forall}b{\forall}f((?{\blacksquare}((({\langle\rangle}Na\backslash Sf)/({\exists}cNc{\bullet}{\it PP}{\it b}))\backslash ({\langle\rangle}Na\backslash Sf))\backslash\\ {[]^{-1}}{[]^{-1}}((({\langle\rangle}Na\backslash Sf)/({\exists}cNc{\bullet}{\it PP}{\it b}))\backslash ({\langle\rangle}Na\backslash Sf)))/
 {\blacksquare}((({\langle\rangle}Na\backslash Sf)/({\exists}cNc{\bullet}{\it PP}{\it b}))\backslash ({\langle\rangle}Na\backslash Sf))):\\ (\Phinplus\ ({\it s}\ ({\it s}\ {\it 0}))\ {\it and}), {\blacksquare}{\forall}n(Nt(n)/{\it CN}{\it n}): \iota , {\square}{\it CN}{\it s(n)}: {\it cd},\\
 {\blacksquare}(({\it PP}{\it to}/{\exists}aNa){\sqcap}{\forall}n(({\langle\rangle}Nn\backslash Si)/({\langle\rangle}Nn\backslash Sb))):\lambda D{\it D},{\blacksquare}Nt(s(f)): {\it s}]]\ \Rightarrow\ Sf
 \end{array}$}
This has the following derivation.
In \textcircled{1} the righthand conjunct is analysed as essentially of the shape
$((N\bsl S)/(N\product \PP))\bsl(N\bsl S)$.
The main action is in the initial unfolding of this succedent to yield a `canonical' sequent:
\disp{$\mini
\prooftree
\prooftree
N, (N\bsl S)/(N\product \PP), N/\CN, \CN, \PP, N\yields S
\justifies
(N\bsl S)/(N\product \PP), N/\CN, \CN, \PP, N\yields N\bsl S
\using \bsl R
\endprooftree
\justifies
N/\CN, \CN, \PP, N\yields ((N\bsl S)/(N\product \PP))\bsl(N\bsl S)
\using \bsl R
\endprooftree
$}
Subderivation \textcircled{2} is exactly the same --- except for the additional
bottommost existential
exponential right rule.
Reading upwards in the main derivation,
after modality elimination and instantiation of features on the coordinator type
and application to the two conjuncts, the brackets are checked and there is
application of the whole coordinate structure to the left node raised verb
$(N\bsl S)/(N\product \PP)$ (left subsubderivation) and to the subject $N$
(right subsubderivation).
\vspace{0.15in}
$${\tiny
\prooftree
\prooftree
\prooftree
\prooftree
\prooftree
\prooftree
\prooftree
\prooftree
\prooftree
\prooftree
\prooftree
\prooftree
\justifies
\mbox{\fbox{${\it CN}{\it s(n)}$}}\ \Rightarrow\ {\it CN}{\it s(n)}
\endprooftree
\justifies
\mbox{\fbox{${\square}{\it CN}{\it s(n)}$}}\ \Rightarrow\ {\it CN}{\it s(n)}
\using {\Box}L
\endprooftree
\prooftree
\justifies
\mbox{\fbox{$Nt(s(n))$}}\ \Rightarrow\ Nt(s(n))
\endprooftree
\justifies
\mbox{\fbox{$Nt(s(n))/{\it CN}{\it s(n)}$}}, {\square}{\it CN}{\it s(n)}\ \Rightarrow\ Nt(s(n))
\using {/}L
\endprooftree
\justifies
\mbox{\fbox{${\forall}n(Nt(n)/{\it CN}{\it n})$}}, {\square}{\it CN}{\it s(n)}\ \Rightarrow\ Nt(s(n))
\using {\forall}L
\endprooftree
\justifies
\mbox{\fbox{${\blacksquare}{\forall}n(Nt(n)/{\it CN}{\it n})$}}, {\square}{\it CN}{\it s(n)}\ \Rightarrow\ Nt(s(n))
\using {\blacksquare}L
\endprooftree
\justifies
{\blacksquare}{\forall}n(Nt(n)/{\it CN}{\it n}), {\square}{\it CN}{\it s(n)}\ \Rightarrow\ \fbox{${\exists}cNc$}
\using {\exists}R
\endprooftree
\prooftree
\prooftree
\prooftree
\prooftree
\prooftree
\prooftree
\justifies
\mbox{\fbox{$Nt(s(f))$}}\ \Rightarrow\ Nt(s(f))
\endprooftree
\justifies
\mbox{\fbox{${\blacksquare}Nt(s(f))$}}\ \Rightarrow\ Nt(s(f))
\using {\blacksquare}L
\endprooftree
\justifies
{\blacksquare}Nt(s(f))\ \Rightarrow\ \fbox{${\exists}aNa$}
\using {\exists}R
\endprooftree
\prooftree
\justifies
\mbox{\fbox{${\it PP}{\it to}$}}\ \Rightarrow\ {\it PP}{\it to}
\endprooftree
\justifies
\mbox{\fbox{${\it PP}{\it to}/{\exists}aNa$}}, {\blacksquare}Nt(s(f))\ \Rightarrow\ {\it PP}{\it to}
\using {/}L
\endprooftree
\justifies
\mbox{\fbox{$({\it PP}{\it to}/{\exists}aNa){\sqcap}{\forall}n(({\langle\rangle}Nn\backslash Si)/({\langle\rangle}Nn\backslash Sb))$}}, {\blacksquare}Nt(s(f))\ \Rightarrow\ {\it PP}{\it to}
\using {\sqcap}L
\endprooftree
\justifies
\mbox{\fbox{${\blacksquare}(({\it PP}{\it to}/{\exists}aNa){\sqcap}{\forall}n(({\langle\rangle}Nn\backslash Si)/({\langle\rangle}Nn\backslash Sb)))$}}, {\blacksquare}Nt(s(f))\ \Rightarrow\ {\it PP}{\it to}
\using {\blacksquare}L
\endprooftree
\justifies
{\blacksquare}{\forall}n(Nt(n)/{\it CN}{\it n}), {\square}{\it CN}{\it s(n)}, {\blacksquare}(({\it PP}{\it to}/{\exists}aNa){\sqcap}{\forall}n(({\langle\rangle}Nn\backslash Si)/({\langle\rangle}Nn\backslash Sb))), {\blacksquare}Nt(s(f))\ \Rightarrow\ \fbox{${\exists}cNc{\bullet}{\it PP}{\it to}$}
\using {\bullet}R
\endprooftree
\prooftree
\prooftree
\prooftree
\justifies
Nt(s(m))\ \Rightarrow\ Nt(s(m))
\endprooftree
\justifies
[Nt(s(m))]\ \Rightarrow\ \fbox{${\langle\rangle}Nt(s(m))$}
\using {\langle\rangle}R
\endprooftree
\prooftree
\justifies
\mbox{\fbox{$Sf$}}\ \Rightarrow\ Sf
\endprooftree
\justifies
[Nt(s(m))], \mbox{\fbox{${\langle\rangle}Nt(s(m))\backslash Sf$}}\ \Rightarrow\ Sf
\using {\backslash}L
\endprooftree
\justifies
[Nt(s(m))], \mbox{\fbox{$({\langle\rangle}Nt(s(m))\backslash Sf)/({\exists}cNc{\bullet}{\it PP}{\it to})$}}, {\blacksquare}{\forall}n(Nt(n)/{\it CN}{\it n}), {\square}{\it CN}{\it s(n)}, {\blacksquare}(({\it PP}{\it to}/{\exists}aNa){\sqcap}{\forall}n(({\langle\rangle}Nn\backslash Si)/({\langle\rangle}Nn\backslash Sb))), {\blacksquare}Nt(s(f))\ \Rightarrow\ Sf
\using {/}L
\endprooftree
\justifies
{\langle\rangle}Nt(s(m)), ({\langle\rangle}Nt(s(m))\backslash Sf)/({\exists}cNc{\bullet}{\it PP}{\it to}), {\blacksquare}{\forall}n(Nt(n)/{\it CN}{\it n}), {\square}{\it CN}{\it s(n)}, {\blacksquare}(({\it PP}{\it to}/{\exists}aNa){\sqcap}{\forall}n(({\langle\rangle}Nn\backslash Si)/({\langle\rangle}Nn\backslash Sb))), {\blacksquare}Nt(s(f))\ \Rightarrow\ Sf
\using {\langle\rangle}L
\endprooftree
\justifies
({\langle\rangle}Nt(s(m))\backslash Sf)/({\exists}cNc{\bullet}{\it PP}{\it to}), {\blacksquare}{\forall}n(Nt(n)/{\it CN}{\it n}), {\square}{\it CN}{\it s(n)}, {\blacksquare}(({\it PP}{\it to}/{\exists}aNa){\sqcap}{\forall}n(({\langle\rangle}Nn\backslash Si)/({\langle\rangle}Nn\backslash Sb))), {\blacksquare}Nt(s(f))\ \Rightarrow\ {\langle\rangle}Nt(s(m))\backslash Sf
\using {\backslash}R
\endprooftree
\justifies
{\blacksquare}{\forall}n(Nt(n)/{\it CN}{\it n}), {\square}{\it CN}{\it s(n)}, {\blacksquare}(({\it PP}{\it to}/{\exists}aNa){\sqcap}{\forall}n(({\langle\rangle}Nn\backslash Si)/({\langle\rangle}Nn\backslash Sb))), {\blacksquare}Nt(s(f))\ \Rightarrow\ (({\langle\rangle}Nt(s(m))\backslash Sf)/({\exists}cNc{\bullet}{\it PP}{\it to}))\backslash ({\langle\rangle}Nt(s(m))\backslash Sf)
\using {\backslash}R
\endprooftree
\justifies
\begin{array}{c}
{\blacksquare}{\forall}n(Nt(n)/{\it CN}{\it n}), {\square}{\it CN}{\it s(n)}, {\blacksquare}(({\it PP}{\it to}/{\exists}aNa){\sqcap}{\forall}n(({\langle\rangle}Nn\backslash Si)/({\langle\rangle}Nn\backslash Sb))), {\blacksquare}Nt(s(f))\ \Rightarrow\ {\blacksquare}((({\langle\rangle}Nt(s(m))\backslash Sf)/({\exists}cNc{\bullet}{\it PP}{\it to}))\backslash ({\langle\rangle}Nt(s(m))\backslash Sf))\\
\mbox{\footnotesize\textcircled{1}}
\end{array}
\using {\blacksquare}R
\endprooftree}
$$
$$
{\tiny
\prooftree
\prooftree
\prooftree
\prooftree
\prooftree
\prooftree
\prooftree
\prooftree
\prooftree
\prooftree
\prooftree
\prooftree
\prooftree
\justifies
\mbox{\fbox{${\it CN}{\it s(n)}$}}\ \Rightarrow\ {\it CN}{\it s(n)}
\endprooftree
\justifies
\mbox{\fbox{${\square}{\it CN}{\it s(n)}$}}\ \Rightarrow\ {\it CN}{\it s(n)}
\using {\Box}L
\endprooftree
\prooftree
\justifies
\mbox{\fbox{$Nt(s(n))$}}\ \Rightarrow\ Nt(s(n))
\endprooftree
\justifies
\mbox{\fbox{$Nt(s(n))/{\it CN}{\it s(n)}$}}, {\square}{\it CN}{\it s(n)}\ \Rightarrow\ Nt(s(n))
\using {/}L
\endprooftree
\justifies
\mbox{\fbox{${\forall}n(Nt(n)/{\it CN}{\it n})$}}, {\square}{\it CN}{\it s(n)}\ \Rightarrow\ Nt(s(n))
\using {\forall}L
\endprooftree
\justifies
\mbox{\fbox{${\blacksquare}{\forall}n(Nt(n)/{\it CN}{\it n})$}}, {\square}{\it CN}{\it s(n)}\ \Rightarrow\ Nt(s(n))
\using {\blacksquare}L
\endprooftree
\justifies
{\blacksquare}{\forall}n(Nt(n)/{\it CN}{\it n}), {\square}{\it CN}{\it s(n)}\ \Rightarrow\ \fbox{${\exists}cNc$}
\using {\exists}R
\endprooftree
\prooftree
\prooftree
\prooftree
\prooftree
\prooftree
\prooftree
\justifies
\mbox{\fbox{$Nt(s(f))$}}\ \Rightarrow\ Nt(s(f))
\endprooftree
\justifies
\mbox{\fbox{${\blacksquare}Nt(s(f))$}}\ \Rightarrow\ Nt(s(f))
\using {\blacksquare}L
\endprooftree
\justifies
{\blacksquare}Nt(s(f))\ \Rightarrow\ \fbox{${\exists}aNa$}
\using {\exists}R
\endprooftree
\prooftree
\justifies
\mbox{\fbox{${\it PP}{\it to}$}}\ \Rightarrow\ {\it PP}{\it to}
\endprooftree
\justifies
\mbox{\fbox{${\it PP}{\it to}/{\exists}aNa$}}, {\blacksquare}Nt(s(f))\ \Rightarrow\ {\it PP}{\it to}
\using {/}L
\endprooftree
\justifies
\mbox{\fbox{$({\it PP}{\it to}/{\exists}aNa){\sqcap}{\forall}n(({\langle\rangle}Nn\backslash Si)/({\langle\rangle}Nn\backslash Sb))$}}, {\blacksquare}Nt(s(f))\ \Rightarrow\ {\it PP}{\it to}
\using {\sqcap}L
\endprooftree
\justifies
\mbox{\fbox{${\blacksquare}(({\it PP}{\it to}/{\exists}aNa){\sqcap}{\forall}n(({\langle\rangle}Nn\backslash Si)/({\langle\rangle}Nn\backslash Sb)))$}}, {\blacksquare}Nt(s(f))\ \Rightarrow\ {\it PP}{\it to}
\using {\blacksquare}L
\endprooftree
\justifies
{\blacksquare}{\forall}n(Nt(n)/{\it CN}{\it n}), {\square}{\it CN}{\it s(n)}, {\blacksquare}(({\it PP}{\it to}/{\exists}aNa){\sqcap}{\forall}n(({\langle\rangle}Nn\backslash Si)/({\langle\rangle}Nn\backslash Sb))), {\blacksquare}Nt(s(f))\ \Rightarrow\ \fbox{${\exists}cNc{\bullet}{\it PP}{\it to}$}
\using {\bullet}R
\endprooftree
\prooftree
\prooftree
\prooftree
\justifies
Nt(s(m))\ \Rightarrow\ Nt(s(m))
\endprooftree
\justifies
[Nt(s(m))]\ \Rightarrow\ \fbox{${\langle\rangle}Nt(s(m))$}
\using {\langle\rangle}R
\endprooftree
\prooftree
\justifies
\mbox{\fbox{$Sf$}}\ \Rightarrow\ Sf
\endprooftree
\justifies
[Nt(s(m))], \mbox{\fbox{${\langle\rangle}Nt(s(m))\backslash Sf$}}\ \Rightarrow\ Sf
\using {\backslash}L
\endprooftree
\justifies
[Nt(s(m))], \mbox{\fbox{$({\langle\rangle}Nt(s(m))\backslash Sf)/({\exists}cNc{\bullet}{\it PP}{\it to})$}}, {\blacksquare}{\forall}n(Nt(n)/{\it CN}{\it n}), {\square}{\it CN}{\it s(n)}, {\blacksquare}(({\it PP}{\it to}/{\exists}aNa){\sqcap}{\forall}n(({\langle\rangle}Nn\backslash Si)/({\langle\rangle}Nn\backslash Sb))), {\blacksquare}Nt(s(f))\ \Rightarrow\ Sf
\using {/}L
\endprooftree
\justifies
{\langle\rangle}Nt(s(m)), ({\langle\rangle}Nt(s(m))\backslash Sf)/({\exists}cNc{\bullet}{\it PP}{\it to}), {\blacksquare}{\forall}n(Nt(n)/{\it CN}{\it n}), {\square}{\it CN}{\it s(n)}, {\blacksquare}(({\it PP}{\it to}/{\exists}aNa){\sqcap}{\forall}n(({\langle\rangle}Nn\backslash Si)/({\langle\rangle}Nn\backslash Sb))), {\blacksquare}Nt(s(f))\ \Rightarrow\ Sf
\using {\langle\rangle}L
\endprooftree
\justifies
({\langle\rangle}Nt(s(m))\backslash Sf)/({\exists}cNc{\bullet}{\it PP}{\it to}), {\blacksquare}{\forall}n(Nt(n)/{\it CN}{\it n}), {\square}{\it CN}{\it s(n)}, {\blacksquare}(({\it PP}{\it to}/{\exists}aNa){\sqcap}{\forall}n(({\langle\rangle}Nn\backslash Si)/({\langle\rangle}Nn\backslash Sb))), {\blacksquare}Nt(s(f))\ \Rightarrow\ {\langle\rangle}Nt(s(m))\backslash Sf
\using {\backslash}R
\endprooftree
\justifies
{\blacksquare}{\forall}n(Nt(n)/{\it CN}{\it n}), {\square}{\it CN}{\it s(n)}, {\blacksquare}(({\it PP}{\it to}/{\exists}aNa){\sqcap}{\forall}n(({\langle\rangle}Nn\backslash Si)/({\langle\rangle}Nn\backslash Sb))), {\blacksquare}Nt(s(f))\ \Rightarrow\ (({\langle\rangle}Nt(s(m))\backslash Sf)/({\exists}cNc{\bullet}{\it PP}{\it to}))\backslash ({\langle\rangle}Nt(s(m))\backslash Sf)
\using {\backslash}R
\endprooftree
\justifies
{\blacksquare}{\forall}n(Nt(n)/{\it CN}{\it n}), {\square}{\it CN}{\it s(n)}, {\blacksquare}(({\it PP}{\it to}/{\exists}aNa){\sqcap}{\forall}n(({\langle\rangle}Nn\backslash Si)/({\langle\rangle}Nn\backslash Sb))), {\blacksquare}Nt(s(f))\ \Rightarrow\ {\blacksquare}((({\langle\rangle}Nt(s(m))\backslash Sf)/({\exists}cNc{\bullet}{\it PP}{\it to}))\backslash ({\langle\rangle}Nt(s(m))\backslash Sf))
\using {\blacksquare}R
\endprooftree
\justifies
\begin{array}{c}
{\blacksquare}{\forall}n(Nt(n)/{\it CN}{\it n}), {\square}{\it CN}{\it s(n)}, {\blacksquare}(({\it PP}{\it to}/{\exists}aNa){\sqcap}{\forall}n(({\langle\rangle}Nn\backslash Si)/({\langle\rangle}Nn\backslash Sb))), {\blacksquare}Nt(s(f))\ \Rightarrow\ \fbox{$?{\blacksquare}((({\langle\rangle}Nt(s(m))\backslash Sf)/({\exists}cNc{\bullet}{\it PP}{\it to}))\backslash ({\langle\rangle}Nt(s(m))\backslash Sf))$}\\
\mbox{\footnotesize\textcircled{2}}
\end{array}
\using {?}R
\endprooftree}
$$
$$
\rotatebox{0}{\tiny
\prooftree
\prooftree
\prooftree
\prooftree
\prooftree
\mbox{\footnotesize\textcircled{1}}\tab
\prooftree
\mbox{\footnotesize\textcircled{2}}\tab
\prooftree
\prooftree
\prooftree
\prooftree
\prooftree
\prooftree
\prooftree
\prooftree
\prooftree
\prooftree
\prooftree
\prooftree
\prooftree
\justifies
N5\ \Rightarrow\ N5
\endprooftree
\justifies
N5\ \Rightarrow\ \fbox{${\exists}bNb$}
\using {\exists}R
\endprooftree
\prooftree
\justifies
{\it PP}{\it to}\ \Rightarrow\ {\it PP}{\it to}
\endprooftree
\justifies
N5, {\it PP}{\it to}\ \Rightarrow\ \fbox{${\exists}bNb{\bullet}{\it PP}{\it to}$}
\using {\bullet}R
\endprooftree
\prooftree
\prooftree
\prooftree
\prooftree
\justifies
Nt(s(m))\ \Rightarrow\ Nt(s(m))
\endprooftree
\justifies
Nt(s(m))\ \Rightarrow\ \fbox{${\exists}aNa$}
\using {\exists}R
\endprooftree
\justifies
[Nt(s(m))]\ \Rightarrow\ \fbox{${\langle\rangle}{\exists}aNa$}
\using {\langle\rangle}R
\endprooftree
\prooftree
\justifies
\mbox{\fbox{$Sf$}}\ \Rightarrow\ Sf
\endprooftree
\justifies
[Nt(s(m))], \mbox{\fbox{${\langle\rangle}{\exists}aNa\backslash Sf$}}\ \Rightarrow\ Sf
\using {\backslash}L
\endprooftree
\justifies
[Nt(s(m))], \mbox{\fbox{$({\langle\rangle}{\exists}aNa\backslash Sf)/({\exists}bNb{\bullet}{\it PP}{\it to})$}}, N5, {\it PP}{\it to}\ \Rightarrow\ Sf
\using {/}L
\endprooftree
\justifies
[Nt(s(m))], \mbox{\fbox{${\square}(({\langle\rangle}{\exists}aNa\backslash Sf)/({\exists}bNb{\bullet}{\it PP}{\it to}))$}}, N5, {\it PP}{\it to}\ \Rightarrow\ Sf
\using {\Box}L
\endprooftree
\justifies
[Nt(s(m))], {\square}(({\langle\rangle}{\exists}aNa\backslash Sf)/({\exists}bNb{\bullet}{\it PP}{\it to})), {\exists}cNc, {\it PP}{\it to}\ \Rightarrow\ Sf
\using {\exists}L
\endprooftree
\justifies
{\langle\rangle}Nt(s(m)), {\square}(({\langle\rangle}{\exists}aNa\backslash Sf)/({\exists}bNb{\bullet}{\it PP}{\it to})), {\exists}cNc, {\it PP}{\it to}\ \Rightarrow\ Sf
\using {\langle\rangle}L
\endprooftree
\justifies
{\langle\rangle}Nt(s(m)), {\square}(({\langle\rangle}{\exists}aNa\backslash Sf)/({\exists}bNb{\bullet}{\it PP}{\it to})), {\exists}cNc{\bullet}{\it PP}{\it to}\ \Rightarrow\ Sf
\using {\bullet}L
\endprooftree
\justifies
{\square}(({\langle\rangle}{\exists}aNa\backslash Sf)/({\exists}bNb{\bullet}{\it PP}{\it to})), {\exists}cNc{\bullet}{\it PP}{\it to}\ \Rightarrow\ {\langle\rangle}Nt(s(m))\backslash Sf
\using {\backslash}R
\endprooftree
\justifies
{\square}(({\langle\rangle}{\exists}aNa\backslash Sf)/({\exists}bNb{\bullet}{\it PP}{\it to}))\ \Rightarrow\ ({\langle\rangle}Nt(s(m))\backslash Sf)/({\exists}cNc{\bullet}{\it PP}{\it to})
\using {/}R
\endprooftree
\prooftree
\prooftree
\prooftree
\prooftree
\justifies
\mbox{\fbox{$Nt(s(m))$}}\ \Rightarrow\ Nt(s(m))
\endprooftree
\justifies
\mbox{\fbox{${\blacksquare}Nt(s(m))$}}\ \Rightarrow\ Nt(s(m))
\using {\blacksquare}L
\endprooftree
\justifies
[{\blacksquare}Nt(s(m))]\ \Rightarrow\ \fbox{${\langle\rangle}Nt(s(m))$}
\using {\langle\rangle}R
\endprooftree
\prooftree
\justifies
\mbox{\fbox{$Sf$}}\ \Rightarrow\ Sf
\endprooftree
\justifies
[{\blacksquare}Nt(s(m))], \mbox{\fbox{${\langle\rangle}Nt(s(m))\backslash Sf$}}\ \Rightarrow\ Sf
\using {\backslash}L
\endprooftree
\justifies
[{\blacksquare}Nt(s(m))], {\square}(({\langle\rangle}{\exists}aNa\backslash Sf)/({\exists}bNb{\bullet}{\it PP}{\it to})), \mbox{\fbox{$(({\langle\rangle}Nt(s(m))\backslash Sf)/({\exists}cNc{\bullet}{\it PP}{\it to}))\backslash ({\langle\rangle}Nt(s(m))\backslash Sf)$}}\ \Rightarrow\ Sf
\using {\backslash}L
\endprooftree
\justifies
[{\blacksquare}Nt(s(m))], {\square}(({\langle\rangle}{\exists}aNa\backslash Sf)/({\exists}bNb{\bullet}{\it PP}{\it to})), [\mbox{\fbox{${[]^{-1}}((({\langle\rangle}Nt(s(m))\backslash Sf)/({\exists}cNc{\bullet}{\it PP}{\it to}))\backslash ({\langle\rangle}Nt(s(m))\backslash Sf))$}}]\ \Rightarrow\ Sf
\using {[]^{-1}}L
\endprooftree
\justifies
[{\blacksquare}Nt(s(m))], {\square}(({\langle\rangle}{\exists}aNa\backslash Sf)/({\exists}bNb{\bullet}{\it PP}{\it to})), [[\mbox{\fbox{${[]^{-1}}{[]^{-1}}((({\langle\rangle}Nt(s(m))\backslash Sf)/({\exists}cNc{\bullet}{\it PP}{\it to}))\backslash ({\langle\rangle}Nt(s(m))\backslash Sf))$}}]]\ \Rightarrow\ Sf
\using {[]^{-1}}L
\endprooftree
\justifies
\begin{array}{c}
{}[{\blacksquare}Nt(s(m))], {\square}(({\langle\rangle}{\exists}aNa\backslash Sf)/({\exists}bNb{\bullet}{\it PP}{\it to})), [[{\blacksquare}{\forall}n(Nt(n)/{\it CN}{\it n}), {\square}{\it CN}{\it s(n)}, {\blacksquare}(({\it PP}{\it to}/{\exists}aNa){\sqcap}{\forall}n(({\langle\rangle}Nn\backslash Si)/({\langle\rangle}Nn\backslash Sb))), {\blacksquare}Nt(s(f)),\\
 \mbox{\fbox{$?{\blacksquare}((({\langle\rangle}Nt(s(m))\backslash Sf)/({\exists}cNc{\bullet}{\it PP}{\it to}))\backslash ({\langle\rangle}Nt(s(m))\backslash Sf))\backslash {[]^{-1}}{[]^{-1}}((({\langle\rangle}Nt(s(m))\backslash Sf)/({\exists}cNc{\bullet}{\it PP}{\it to}))\backslash ({\langle\rangle}Nt(s(m))\backslash Sf))$}}]]\ \Rightarrow\ Sf
\end{array}
\using {\backslash}L
\endprooftree
\justifies
\begin{array}{c}
{}[{\blacksquare}Nt(s(m))], {\square}(({\langle\rangle}{\exists}aNa\backslash Sf)/({\exists}bNb{\bullet}{\it PP}{\it to})), [[{\blacksquare}{\forall}n(Nt(n)/{\it CN}{\it n}), {\square}{\it CN}{\it s(n)}, {\blacksquare}(({\it PP}{\it to}/{\exists}aNa){\sqcap}{\forall}n(({\langle\rangle}Nn\backslash Si)/({\langle\rangle}Nn\backslash Sb))), {\blacksquare}Nt(s(f)),\\
\mbox{\fbox{$(?{\blacksquare}((({\langle\rangle}Nt(s(m))\backslash Sf)/({\exists}cNc{\bullet}{\it PP}{\it to}))\backslash ({\langle\rangle}Nt(s(m))\backslash Sf))\backslash {[]^{-1}}{[]^{-1}}((({\langle\rangle}Nt(s(m))\backslash Sf)/({\exists}cNc{\bullet}{\it PP}{\it to}))\backslash ({\langle\rangle}Nt(s(m))\backslash Sf)))/{\blacksquare}((({\langle\rangle}Nt(s(m))\backslash Sf)/({\exists}cNc{\bullet}{\it PP}{\it to}))\backslash ({\langle\rangle}Nt(s(m))\backslash Sf))$}},\\
{\blacksquare}{\forall}n(Nt(n)/{\it CN}{\it n}), {\square}{\it CN}{\it s(n)}, {\blacksquare}(({\it PP}{\it to}/{\exists}aNa){\sqcap}{\forall}n(({\langle\rangle}Nn\backslash Si)/({\langle\rangle}Nn\backslash Sb))), {\blacksquare}Nt(s(f))]]\ \Rightarrow\ Sf
\end{array}
\using {/}L
\endprooftree
\justifies
\begin{array}{c}
{}[{\blacksquare}Nt(s(m))], {\square}(({\langle\rangle}{\exists}aNa\backslash Sf)/({\exists}bNb{\bullet}{\it PP}{\it to})), [[{\blacksquare}{\forall}n(Nt(n)/{\it CN}{\it n}), {\square}{\it CN}{\it s(n)}, {\blacksquare}(({\it PP}{\it to}/{\exists}aNa){\sqcap}{\forall}n(({\langle\rangle}Nn\backslash Si)/({\langle\rangle}Nn\backslash Sb))), {\blacksquare}Nt(s(f)),\\
\mbox{\fbox{${\forall}f((?{\blacksquare}((({\langle\rangle}Nt(s(m))\backslash Sf)/({\exists}cNc{\bullet}{\it PP}{\it to}))\backslash ({\langle\rangle}Nt(s(m))\backslash Sf))\backslash {[]^{-1}}{[]^{-1}}((({\langle\rangle}Nt(s(m))\backslash Sf)/({\exists}cNc{\bullet}{\it PP}{\it to}))\backslash ({\langle\rangle}Nt(s(m))\backslash Sf)))/{\blacksquare}((({\langle\rangle}Nt(s(m))\backslash Sf)/({\exists}cNc{\bullet}{\it PP}{\it to}))\backslash ({\langle\rangle}Nt(s(m))\backslash Sf)))$}},\\
{\blacksquare}{\forall}n(Nt(n)/{\it CN}{\it n}), {\square}{\it CN}{\it s(n)}, {\blacksquare}(({\it PP}{\it to}/{\exists}aNa){\sqcap}{\forall}n(({\langle\rangle}Nn\backslash Si)/({\langle\rangle}Nn\backslash Sb))), {\blacksquare}Nt(s(f))]]\ \Rightarrow\ Sf
\end{array}
\using {\forall}L
\endprooftree
\justifies
\begin{array}{c}
{}[{\blacksquare}Nt(s(m))], {\square}(({\langle\rangle}{\exists}aNa\backslash Sf)/({\exists}bNb{\bullet}{\it PP}{\it to})), [[{\blacksquare}{\forall}n(Nt(n)/{\it CN}{\it n}), {\square}{\it CN}{\it s(n)}, {\blacksquare}(({\it PP}{\it to}/{\exists}aNa){\sqcap}{\forall}n(({\langle\rangle}Nn\backslash Si)/({\langle\rangle}Nn\backslash Sb))), {\blacksquare}Nt(s(f)),\\
\mbox{\fbox{${\forall}b{\forall}f((?{\blacksquare}((({\langle\rangle}Nt(s(m))\backslash Sf)/({\exists}cNc{\bullet}{\it PP}{\it b}))\backslash ({\langle\rangle}Nt(s(m))\backslash Sf))\backslash {[]^{-1}}{[]^{-1}}((({\langle\rangle}Nt(s(m))\backslash Sf)/({\exists}cNc{\bullet}{\it PP}{\it b}))\backslash ({\langle\rangle}Nt(s(m))\backslash Sf)))/{\blacksquare}((({\langle\rangle}Nt(s(m))\backslash Sf)/({\exists}cNc{\bullet}{\it PP}{\it b}))\backslash ({\langle\rangle}Nt(s(m))\backslash Sf)))$}},\\
{\blacksquare}{\forall}n(Nt(n)/{\it CN}{\it n}), {\square}{\it CN}{\it s(n)}, {\blacksquare}(({\it PP}{\it to}/{\exists}aNa){\sqcap}{\forall}n(({\langle\rangle}Nn\backslash Si)/({\langle\rangle}Nn\backslash Sb))), {\blacksquare}Nt(s(f))]]\ \Rightarrow\ Sf
\end{array}
\using {\forall}L
\endprooftree
\justifies
\begin{array}{c}
{}[{\blacksquare}Nt(s(m))], {\square}(({\langle\rangle}{\exists}aNa\backslash Sf)/({\exists}bNb{\bullet}{\it PP}{\it to})), [[{\blacksquare}{\forall}n(Nt(n)/{\it CN}{\it n}), {\square}{\it CN}{\it s(n)}, {\blacksquare}(({\it PP}{\it to}/{\exists}aNa){\sqcap}{\forall}n(({\langle\rangle}Nn\backslash Si)/({\langle\rangle}Nn\backslash Sb))), {\blacksquare}Nt(s(f)),\\
\mbox{\fbox{${\forall}a{\forall}b{\forall}f((?{\blacksquare}((({\langle\rangle}Na\backslash Sf)/({\exists}cNc{\bullet}{\it PP}{\it b}))\backslash ({\langle\rangle}Na\backslash Sf))\backslash {[]^{-1}}{[]^{-1}}((({\langle\rangle}Na\backslash Sf)/({\exists}cNc{\bullet}{\it PP}{\it b}))\backslash ({\langle\rangle}Na\backslash Sf)))/{\blacksquare}((({\langle\rangle}Na\backslash Sf)/({\exists}cNc{\bullet}{\it PP}{\it b}))\backslash ({\langle\rangle}Na\backslash Sf)))$}},\\
{\blacksquare}{\forall}n(Nt(n)/{\it CN}{\it n}), {\square}{\it CN}{\it s(n)}, {\blacksquare}(({\it PP}{\it to}/{\exists}aNa){\sqcap}{\forall}n(({\langle\rangle}Nn\backslash Si)/({\langle\rangle}Nn\backslash Sb))), {\blacksquare}Nt(s(f))]]\ \Rightarrow\ Sf
\end{array}
\using {\forall}L
\endprooftree
\justifies
\begin{array}{c}
{}[{\blacksquare}Nt(s(m))], {\square}(({\langle\rangle}{\exists}aNa\backslash Sf)/({\exists}bNb{\bullet}{\it PP}{\it to})), [[{\blacksquare}{\forall}n(Nt(n)/{\it CN}{\it n}), {\square}{\it CN}{\it s(n)}, {\blacksquare}(({\it PP}{\it to}/{\exists}aNa){\sqcap}{\forall}n(({\langle\rangle}Nn\backslash Si)/({\langle\rangle}Nn\backslash Sb))), {\blacksquare}Nt(s(f)),\\
\mbox{\fbox{${\blacksquare}{\forall}a{\forall}b{\forall}f((?{\blacksquare}((({\langle\rangle}Na\backslash Sf)/({\exists}cNc{\bullet}{\it PP}{\it b}))\backslash ({\langle\rangle}Na\backslash Sf))\backslash {[]^{-1}}{[]^{-1}}((({\langle\rangle}Na\backslash Sf)/({\exists}cNc{\bullet}{\it PP}{\it b}))\backslash ({\langle\rangle}Na\backslash Sf)))/{\blacksquare}((({\langle\rangle}Na\backslash Sf)/({\exists}cNc{\bullet}{\it PP}{\it b}))\backslash ({\langle\rangle}Na\backslash Sf)))$}},\\
{\blacksquare}{\forall}n(Nt(n)/{\it CN}{\it n}), {\square}{\it CN}{\it s(n)}, {\blacksquare}(({\it PP}{\it to}/{\exists}aNa){\sqcap}{\forall}n(({\langle\rangle}Nn\backslash Si)/({\langle\rangle}Nn\backslash Sb))), {\blacksquare}Nt(s(f))]]\ \Rightarrow\ Sf
\end{array}
\using {\blacksquare}L
\endprooftree}
$$
\vspace{0.15in}
\noindent
All this correctly assigns semantics:
\disp{
$[({\it Past}\ (((\mbox{\v{}}{\it give}\ {\it m})\ (\iota \ \mbox{\v{}}{\it book}))\ {\it j}))\wedge ({\it Past}\ (((\mbox{\v{}}{\it give}\ {\it s})\ (\iota \ \mbox{\v{}}{\it cd}))\ {\it j}))]$}

\subsection{Argument plus modifier left node raising coordination}

The next example has LNR with arguments and adverbs in the conjuncts:
\disp{
$[{\bf john}]{+}{\bf saw}{+}[[{\bf mary}{+}{\bf today}{+}{\bf and}{+}{\bf bill}{+}{\bf yesterday}]]: Sf$}
Appropriate lexical lookup yields the following where the
coordinator is essentially of the form $(\exstexp X\bsl \abrack\abrack X)/X$ where
$X=((N\bsl S)/N)\bsl(N\bsl S)$.\footnote{Here `saw` is polymorphic between seeing an entity ({\it seee})
and seeing a proposition ({\it seet}).
We shall see in Section~\ref{unlikesect} how this allows coordination over `unlike' types.
The coordination here is over regular transitive verbs,
but we wish to show now the integration of node raising with the propensity
for such other features.}
\disp{
$\begin{array}[t]{l}
[{\blacksquare}Nt(s(m)): {\it j}], {\square}(({\langle\rangle}{\exists}aNa\backslash Sf)/({\exists}aNa{\oplus}{\it CP}that)):\\
\mbox{\^{}}\lambda A\lambda B({\it Past}\ (({\it A}\casearrow C.(\mbox{\v{}}{\it seee}\ {\it C}); D.(\mbox{\v{}}{\it seet}\ {\it D}))\ {\it B})),\\
{} [[{\blacksquare}Nt(s(f)): {\it m}, {\square}{\forall}a{\forall}f(({\langle\rangle}Na\backslash Sf)\backslash ({\langle\rangle}Na\backslash Sf)): \mbox{\^{}}\lambda E\lambda F(\mbox{\v{}}{\it today}\ ({\it E}\ {\it F})), \\{\blacksquare}{\forall}f{\forall}a((?{\blacksquare}((({\langle\rangle}Na\backslash Sf)/{\exists}bNb)\backslash ({\langle\rangle}Na\backslash Sf))\backslash
 {[]^{-1}}{[]^{-1}}((({\langle\rangle}Na\backslash Sf)/{\exists}bNb)\backslash ({\langle\rangle}Na\backslash Sf)))/\\{\blacksquare}((({\langle\rangle}Na\backslash Sf)/{\exists}bNb)\backslash ({\langle\rangle}Na\backslash Sf))): (\Phinplus\ ({\it s}\ ({\it s}\ {\it 0}))\ {\it and}), {\blacksquare}Nt(s(m)): {\it b}, \\{\square}{\forall}a{\forall}f(({\langle\rangle}Na\backslash Sf)\backslash ({\langle\rangle}Na\backslash Sf)): \mbox{\^{}}\lambda G\lambda H(\mbox{\v{}}{\it yesterday}\ ({\it G}\ {\it H}))]]\ \Rightarrow\ Sf
 \end{array}$}
The example has the following derivation.
In \textcircled{1} the righthand conjunct is analysed as essentially of the shape
$((N\bsl S)/N)\bsl(N\bsl S)$. The main action is in the initial unfolding of this succedent
to yield a `canonical' sequent:
\disp{$\mini
\prooftree
\prooftree
N, (N\bsl S)/N, N, (N\bsl S)\bsl(N\bsl S)\yields S
\justifies
(N\bsl S)/N, N, (N\bsl S)\bsl(N\bsl S)\yields N\bsl S
\using \bsl R
\endprooftree
\justifies
N, (N\bsl S)\bsl(N\bsl S)\yields ((N\bsl S)/N)\bsl(N\bsl S)
\using \bsl R
\endprooftree$}
The subtree \textcircled{2} for the lefthand conjunct is exactly the same --- except
for the bottommost existential exponential right rule and the gender of the object ---
hence it has been elided.
The trunk of the main derivation and the checking of the bracket context
are fairly standard by now. In the left and right subsubderivations
the (polymorphic) verb type is shown to yield the left node raised
transitive verb $(N\bsl S)/N$ and subject $N$ coordinate structure arguments.


{\tiny
\begin{center}
\prooftree
\prooftree
\prooftree
\prooftree
\prooftree
\prooftree
\prooftree
\prooftree
\prooftree
\prooftree
\prooftree
\prooftree
\prooftree
\prooftree
\justifies
\mbox{\fbox{$Nt(s(m))$}}\ \Rightarrow\ Nt(s(m))
\endprooftree
\justifies
\mbox{\fbox{${\blacksquare}Nt(s(m))$}}\ \Rightarrow\ Nt(s(m))
\using {\blacksquare}L
\endprooftree
\justifies
{\blacksquare}Nt(s(m))\ \Rightarrow\ \fbox{${\exists}bNb$}
\using {\exists}R
\endprooftree
\prooftree
\prooftree
\prooftree
\justifies
Nt(s(m))\ \Rightarrow\ Nt(s(m))
\endprooftree
\justifies
[Nt(s(m))]\ \Rightarrow\ \fbox{${\langle\rangle}Nt(s(m))$}
\using {\langle\rangle}R
\endprooftree
\prooftree
\justifies
\mbox{\fbox{$Sf$}}\ \Rightarrow\ Sf
\endprooftree
\justifies
[Nt(s(m))], \mbox{\fbox{${\langle\rangle}Nt(s(m))\backslash Sf$}}\ \Rightarrow\ Sf
\using {\backslash}L
\endprooftree
\justifies
[Nt(s(m))], \mbox{\fbox{$({\langle\rangle}Nt(s(m))\backslash Sf)/{\exists}bNb$}}, {\blacksquare}Nt(s(m))\ \Rightarrow\ Sf
\using {/}L
\endprooftree
\justifies
{\langle\rangle}Nt(s(m)), ({\langle\rangle}Nt(s(m))\backslash Sf)/{\exists}bNb, {\blacksquare}Nt(s(m))\ \Rightarrow\ Sf
\using {\langle\rangle}L
\endprooftree
\justifies
({\langle\rangle}Nt(s(m))\backslash Sf)/{\exists}bNb, {\blacksquare}Nt(s(m))\ \Rightarrow\ {\langle\rangle}Nt(s(m))\backslash Sf
\using {\backslash}R
\endprooftree
\prooftree
\prooftree
\prooftree
\justifies
Nt(s(m))\ \Rightarrow\ Nt(s(m))
\endprooftree
\justifies
[Nt(s(m))]\ \Rightarrow\ \fbox{${\langle\rangle}Nt(s(m))$}
\using {\langle\rangle}R
\endprooftree
\prooftree
\justifies
\mbox{\fbox{$Sf$}}\ \Rightarrow\ Sf
\endprooftree
\justifies
[Nt(s(m))], \mbox{\fbox{${\langle\rangle}Nt(s(m))\backslash Sf$}}\ \Rightarrow\ Sf
\using {\backslash}L
\endprooftree
\justifies
[Nt(s(m))], ({\langle\rangle}Nt(s(m))\backslash Sf)/{\exists}bNb, {\blacksquare}Nt(s(m)), \mbox{\fbox{$({\langle\rangle}Nt(s(m))\backslash Sf)\backslash ({\langle\rangle}Nt(s(m))\backslash Sf)$}}\ \Rightarrow\ Sf
\using {\backslash}L
\endprooftree
\justifies
[Nt(s(m))], ({\langle\rangle}Nt(s(m))\backslash Sf)/{\exists}bNb, {\blacksquare}Nt(s(m)), \mbox{\fbox{${\forall}f(({\langle\rangle}Nt(s(m))\backslash Sf)\backslash ({\langle\rangle}Nt(s(m))\backslash Sf))$}}\ \Rightarrow\ Sf
\using {\forall}L
\endprooftree
\justifies
[Nt(s(m))], ({\langle\rangle}Nt(s(m))\backslash Sf)/{\exists}bNb, {\blacksquare}Nt(s(m)), \mbox{\fbox{${\forall}a{\forall}f(({\langle\rangle}Na\backslash Sf)\backslash ({\langle\rangle}Na\backslash Sf))$}}\ \Rightarrow\ Sf
\using {\forall}L
\endprooftree
\justifies
[Nt(s(m))], ({\langle\rangle}Nt(s(m))\backslash Sf)/{\exists}bNb, {\blacksquare}Nt(s(m)), \mbox{\fbox{${\square}{\forall}a{\forall}f(({\langle\rangle}Na\backslash Sf)\backslash ({\langle\rangle}Na\backslash Sf))$}}\ \Rightarrow\ Sf
\using {\Box}L
\endprooftree
\justifies
{\langle\rangle}Nt(s(m)), ({\langle\rangle}Nt(s(m))\backslash Sf)/{\exists}bNb, {\blacksquare}Nt(s(m)), {\square}{\forall}a{\forall}f(({\langle\rangle}Na\backslash Sf)\backslash ({\langle\rangle}Na\backslash Sf))\ \Rightarrow\ Sf
\using {\langle\rangle}L
\endprooftree
\justifies
({\langle\rangle}Nt(s(m))\backslash Sf)/{\exists}bNb, {\blacksquare}Nt(s(m)), {\square}{\forall}a{\forall}f(({\langle\rangle}Na\backslash Sf)\backslash ({\langle\rangle}Na\backslash Sf))\ \Rightarrow\ {\langle\rangle}Nt(s(m))\backslash Sf
\using {\backslash}R
\endprooftree
\justifies
{\blacksquare}Nt(s(m)), {\square}{\forall}a{\forall}f(({\langle\rangle}Na\backslash Sf)\backslash ({\langle\rangle}Na\backslash Sf))\ \Rightarrow\ (({\langle\rangle}Nt(s(m))\backslash Sf)/{\exists}bNb)\backslash ({\langle\rangle}Nt(s(m))\backslash Sf)
\using {\backslash}R
\endprooftree
\justifies
\begin{array}{c}
{\blacksquare}Nt(s(m)), {\square}{\forall}a{\forall}f(({\langle\rangle}Na\backslash Sf)\backslash ({\langle\rangle}Na\backslash Sf))\ \Rightarrow\ {\blacksquare}((({\langle\rangle}Nt(s(m))\backslash Sf)/{\exists}bNb)\backslash ({\langle\rangle}Nt(s(m))\backslash Sf))\\
\mbox{\footnotesize\textcircled{1}}
\end{array}
\using {\blacksquare}R
\endprooftree

\rotatebox{-90}{
\resizebox{\textheight}{!}{
\prooftree
\prooftree
\prooftree
\prooftree
\mbox{\footnotesize\textcircled{1}}\tab
\prooftree
\prooftree
\vdots
\justifies
\begin{array}{c}
{\blacksquare}Nt(s(f)), {\square}{\forall}a{\forall}f(({\langle\rangle}Na\backslash Sf)\backslash ({\langle\rangle}Na\backslash Sf))\ \Rightarrow\ \fbox{$?{\blacksquare}((({\langle\rangle}Nt(s(m))\backslash Sf)/{\exists}bNb)\backslash ({\langle\rangle}Nt(s(m))\backslash Sf))$}\\
\end{array}
\using {?}R
\endprooftree
\prooftree
\prooftree
\prooftree
\prooftree
\prooftree
\prooftree
\prooftree
\prooftree
\prooftree
\prooftree
\prooftree
\prooftree
\justifies
N1\ \Rightarrow\ N1
\endprooftree
\justifies
N1\ \Rightarrow\ \fbox{${\exists}aNa$}
\using {\exists}R
\endprooftree
\justifies
N1\ \Rightarrow\ \fbox{${\exists}aNa{\oplus}{\it CP}that$}
\using {\oplus}R
\endprooftree
\prooftree
\prooftree
\prooftree
\prooftree
\justifies
Nt(s(m))\ \Rightarrow\ Nt(s(m))
\endprooftree
\justifies
Nt(s(m))\ \Rightarrow\ \fbox{${\exists}aNa$}
\using {\exists}R
\endprooftree
\justifies
[Nt(s(m))]\ \Rightarrow\ \fbox{${\langle\rangle}{\exists}aNa$}
\using {\langle\rangle}R
\endprooftree
\prooftree
\justifies
\mbox{\fbox{$Sf$}}\ \Rightarrow\ Sf
\endprooftree
\justifies
[Nt(s(m))], \mbox{\fbox{${\langle\rangle}{\exists}aNa\backslash Sf$}}\ \Rightarrow\ Sf
\using {\backslash}L
\endprooftree
\justifies
[Nt(s(m))], \mbox{\fbox{$({\langle\rangle}{\exists}aNa\backslash Sf)/({\exists}aNa{\oplus}{\it CP}that)$}}, N1\ \Rightarrow\ Sf
\using {/}L
\endprooftree
\justifies
[Nt(s(m))], \mbox{\fbox{${\square}(({\langle\rangle}{\exists}aNa\backslash Sf)/({\exists}aNa{\oplus}{\it CP}that))$}}, N1\ \Rightarrow\ Sf
\using {\Box}L
\endprooftree
\justifies
[Nt(s(m))], {\square}(({\langle\rangle}{\exists}aNa\backslash Sf)/({\exists}aNa{\oplus}{\it CP}that)), {\exists}bNb\ \Rightarrow\ Sf
\using {\exists}L
\endprooftree
\justifies
{\langle\rangle}Nt(s(m)), {\square}(({\langle\rangle}{\exists}aNa\backslash Sf)/({\exists}aNa{\oplus}{\it CP}that)), {\exists}bNb\ \Rightarrow\ Sf
\using {\langle\rangle}L
\endprooftree
\justifies
{\square}(({\langle\rangle}{\exists}aNa\backslash Sf)/({\exists}aNa{\oplus}{\it CP}that)), {\exists}bNb\ \Rightarrow\ {\langle\rangle}Nt(s(m))\backslash Sf
\using {\backslash}R
\endprooftree
\justifies
{\square}(({\langle\rangle}{\exists}aNa\backslash Sf)/({\exists}aNa{\oplus}{\it CP}that))\ \Rightarrow\ ({\langle\rangle}Nt(s(m))\backslash Sf)/{\exists}bNb
\using {/}R
\endprooftree
\prooftree
\prooftree
\prooftree
\prooftree
\justifies
\mbox{\fbox{$Nt(s(m))$}}\ \Rightarrow\ Nt(s(m))
\endprooftree
\justifies
\mbox{\fbox{${\blacksquare}Nt(s(m))$}}\ \Rightarrow\ Nt(s(m))
\using {\blacksquare}L
\endprooftree
\justifies
[{\blacksquare}Nt(s(m))]\ \Rightarrow\ \fbox{${\langle\rangle}Nt(s(m))$}
\using {\langle\rangle}R
\endprooftree
\prooftree
\justifies
\mbox{\fbox{$Sf$}}\ \Rightarrow\ Sf
\endprooftree
\justifies
[{\blacksquare}Nt(s(m))], \mbox{\fbox{${\langle\rangle}Nt(s(m))\backslash Sf$}}\ \Rightarrow\ Sf
\using {\backslash}L
\endprooftree
\justifies
[{\blacksquare}Nt(s(m))], {\square}(({\langle\rangle}{\exists}aNa\backslash Sf)/({\exists}aNa{\oplus}{\it CP}that)), \mbox{\fbox{$(({\langle\rangle}Nt(s(m))\backslash Sf)/{\exists}bNb)\backslash ({\langle\rangle}Nt(s(m))\backslash Sf)$}}\ \Rightarrow\ Sf
\using {\backslash}L
\endprooftree
\justifies
[{\blacksquare}Nt(s(m))], {\square}(({\langle\rangle}{\exists}aNa\backslash Sf)/({\exists}aNa{\oplus}{\it CP}that)), [\mbox{\fbox{${[]^{-1}}((({\langle\rangle}Nt(s(m))\backslash Sf)/{\exists}bNb)\backslash ({\langle\rangle}Nt(s(m))\backslash Sf))$}}]\ \Rightarrow\ Sf
\using {[]^{-1}}L
\endprooftree
\justifies
[{\blacksquare}Nt(s(m))], {\square}(({\langle\rangle}{\exists}aNa\backslash Sf)/({\exists}aNa{\oplus}{\it CP}that)), [[\mbox{\fbox{${[]^{-1}}{[]^{-1}}((({\langle\rangle}Nt(s(m))\backslash Sf)/{\exists}bNb)\backslash ({\langle\rangle}Nt(s(m))\backslash Sf))$}}]]\ \Rightarrow\ Sf
\using {[]^{-1}}L
\endprooftree
\justifies
[{\blacksquare}Nt(s(m))], {\square}(({\langle\rangle}{\exists}aNa\backslash Sf)/({\exists}aNa{\oplus}{\it CP}that)), [[{\blacksquare}Nt(s(f)), {\square}{\forall}a{\forall}f(({\langle\rangle}Na\backslash Sf)\backslash ({\langle\rangle}Na\backslash Sf)), \mbox{\fbox{$?{\blacksquare}((({\langle\rangle}Nt(s(m))\backslash Sf)/{\exists}bNb)\backslash ({\langle\rangle}Nt(s(m))\backslash Sf))\backslash {[]^{-1}}{[]^{-1}}((({\langle\rangle}Nt(s(m))\backslash Sf)/{\exists}bNb)\backslash ({\langle\rangle}Nt(s(m))\backslash Sf))$}}]]\ \Rightarrow\ Sf
\using {\backslash}L
\endprooftree
\justifies
[{\blacksquare}Nt(s(m))], {\square}(({\langle\rangle}{\exists}aNa\backslash Sf)/({\exists}aNa{\oplus}{\it CP}that)), [[{\blacksquare}Nt(s(f)), {\square}{\forall}a{\forall}f(({\langle\rangle}Na\backslash Sf)\backslash ({\langle\rangle}Na\backslash Sf)), \mbox{\fbox{$(?{\blacksquare}((({\langle\rangle}Nt(s(m))\backslash Sf)/{\exists}bNb)\backslash ({\langle\rangle}Nt(s(m))\backslash Sf))\backslash {[]^{-1}}{[]^{-1}}((({\langle\rangle}Nt(s(m))\backslash Sf)/{\exists}bNb)\backslash ({\langle\rangle}Nt(s(m))\backslash Sf)))/{\blacksquare}((({\langle\rangle}Nt(s(m))\backslash Sf)/{\exists}bNb)\backslash ({\langle\rangle}Nt(s(m))\backslash Sf))$}}, {\blacksquare}Nt(s(m)), {\square}{\forall}a{\forall}f(({\langle\rangle}Na\backslash Sf)\backslash ({\langle\rangle}Na\backslash Sf))]]\ \Rightarrow\ Sf
\using {/}L
\endprooftree
\justifies
[{\blacksquare}Nt(s(m))], {\square}(({\langle\rangle}{\exists}aNa\backslash Sf)/({\exists}aNa{\oplus}{\it CP}that)), [[{\blacksquare}Nt(s(f)), {\square}{\forall}a{\forall}f(({\langle\rangle}Na\backslash Sf)\backslash ({\langle\rangle}Na\backslash Sf)), \mbox{\fbox{${\forall}a((?{\blacksquare}((({\langle\rangle}Na\backslash Sf)/{\exists}bNb)\backslash ({\langle\rangle}Na\backslash Sf))\backslash {[]^{-1}}{[]^{-1}}((({\langle\rangle}Na\backslash Sf)/{\exists}bNb)\backslash ({\langle\rangle}Na\backslash Sf)))/{\blacksquare}((({\langle\rangle}Na\backslash Sf)/{\exists}bNb)\backslash ({\langle\rangle}Na\backslash Sf)))$}}, {\blacksquare}Nt(s(m)), {\square}{\forall}a{\forall}f(({\langle\rangle}Na\backslash Sf)\backslash ({\langle\rangle}Na\backslash Sf))]]\ \Rightarrow\ Sf
\using {\forall}L
\endprooftree
\justifies
[{\blacksquare}Nt(s(m))], {\square}(({\langle\rangle}{\exists}aNa\backslash Sf)/({\exists}aNa{\oplus}{\it CP}that)), [[{\blacksquare}Nt(s(f)), {\square}{\forall}a{\forall}f(({\langle\rangle}Na\backslash Sf)\backslash ({\langle\rangle}Na\backslash Sf)), \mbox{\fbox{${\forall}f{\forall}a((?{\blacksquare}((({\langle\rangle}Na\backslash Sf)/{\exists}bNb)\backslash ({\langle\rangle}Na\backslash Sf))\backslash {[]^{-1}}{[]^{-1}}((({\langle\rangle}Na\backslash Sf)/{\exists}bNb)\backslash ({\langle\rangle}Na\backslash Sf)))/{\blacksquare}((({\langle\rangle}Na\backslash Sf)/{\exists}bNb)\backslash ({\langle\rangle}Na\backslash Sf)))$}}, {\blacksquare}Nt(s(m)), {\square}{\forall}a{\forall}f(({\langle\rangle}Na\backslash Sf)\backslash ({\langle\rangle}Na\backslash Sf))]]\ \Rightarrow\ Sf
\using {\forall}L
\endprooftree
\justifies
[{\blacksquare}Nt(s(m))], {\square}(({\langle\rangle}{\exists}aNa\backslash Sf)/({\exists}aNa{\oplus}{\it CP}that)), [[{\blacksquare}Nt(s(f)), {\square}{\forall}a{\forall}f(({\langle\rangle}Na\backslash Sf)\backslash ({\langle\rangle}Na\backslash Sf)), \mbox{\fbox{${\blacksquare}{\forall}f{\forall}a((?{\blacksquare}((({\langle\rangle}Na\backslash Sf)/{\exists}bNb)\backslash ({\langle\rangle}Na\backslash Sf))\backslash {[]^{-1}}{[]^{-1}}((({\langle\rangle}Na\backslash Sf)/{\exists}bNb)\backslash ({\langle\rangle}Na\backslash Sf)))/{\blacksquare}((({\langle\rangle}Na\backslash Sf)/{\exists}bNb)\backslash ({\langle\rangle}Na\backslash Sf)))$}}, {\blacksquare}Nt(s(m)), {\square}{\forall}a{\forall}f(({\langle\rangle}Na\backslash Sf)\backslash ({\langle\rangle}Na\backslash Sf))]]\ \Rightarrow\ Sf
\using {\blacksquare}L
\endprooftree}}
\end{center}}

\vspace{0.15in}

\noindent
This delivers the semantics:
\disp{
$[(\mbox{\v{}}{\it today}\ ({\it Past}\ ((\mbox{\v{}}{\it seee}\ {\it m})\ {\it j})))\wedge (\mbox{\v{}}{\it yesterday}\ ({\it Past}\ ((\mbox{\v{}}{\it seee}\ {\it b})\ {\it j})))]$}

\commentout{

Again, another result of lexical consultation is:
\disp{
$[{\blacksquare}Nt(s(m)): {\it j}], {\square}(({\langle\rangle}{\exists}aNa\backslash Sf)/({\exists}aNa{\oplus}{\it CP}that)): \mbox{\^{}}\lambda A\lambda B({\it Past}\ (({\it A}\casearrow C.(\mbox{\v{}}{\it seee}\ {\it C}); D.(\mbox{\v{}}{\it seet}\ {\it D}))\ {\it B})), [[{\blacksquare}Nt(s(f)): {\it m}, {\square}{\forall}a{\forall}f(({\langle\rangle}Na\backslash Sf)\backslash ({\langle\rangle}Na\backslash Sf)): \mbox{\^{}}\lambda E\lambda F(\mbox{\v{}}{\it today}\ ({\it E}\ {\it F})), {\blacksquare}{\forall}a{\forall}b{\forall}f((?{\blacksquare}((({\langle\rangle}Na\backslash Sf)/({\exists}cNc{\oplus}{\it CP}b))\backslash ({\langle\rangle}Na\backslash Sf))\backslash\\
 {[]^{-1}}{[]^{-1}}((({\langle\rangle}Na\backslash Sf)/({\exists}cNc{\oplus}{\it CP}b))\backslash ({\langle\rangle}Na\backslash Sf)))/{\blacksquare}((({\langle\rangle}Na\backslash Sf)/({\exists}cNc{\oplus}{\it CP}b))\backslash ({\langle\rangle}Na\backslash Sf))): (\Phinplus\ ({\it s}\ ({\it s}\ {\it 0}))\ {\it and}), \\{\blacksquare}Nt(s(m)): {\it b}, {\square}{\forall}a{\forall}f(({\langle\rangle}Na\backslash Sf)\backslash ({\langle\rangle}Na\backslash Sf)): \mbox{\^{}}\lambda G\lambda H(\mbox{\v{}}{\it yesterday}\ ({\it G}\ {\it H}))]]\ \Rightarrow\ Sf$}
Here the coordination is over the polymorphic verb type.
This has the derivation:

\vspace{0.15in}

{\tiny

\prooftree
\prooftree
\prooftree
\prooftree
\prooftree
\prooftree
\prooftree
\prooftree
\prooftree
\prooftree
\prooftree
\prooftree
\prooftree
\prooftree
\prooftree
\justifies
\mbox{\fbox{$Nt(s(m))$}}\ \Rightarrow\ Nt(s(m))
\endprooftree
\justifies
\mbox{\fbox{${\blacksquare}Nt(s(m))$}}\ \Rightarrow\ Nt(s(m))
\using {\blacksquare}L
\endprooftree
\justifies
{\blacksquare}Nt(s(m))\ \Rightarrow\ \fbox{${\exists}cNc$}
\using {\exists}R
\endprooftree
\justifies
{\blacksquare}Nt(s(m))\ \Rightarrow\ \fbox{${\exists}cNc{\oplus}{\it CP}that$}
\using {\oplus}R
\endprooftree
\prooftree
\prooftree
\prooftree
\justifies
Nt(s(m))\ \Rightarrow\ Nt(s(m))
\endprooftree
\justifies
[Nt(s(m))]\ \Rightarrow\ \fbox{${\langle\rangle}Nt(s(m))$}
\using {\langle\rangle}R
\endprooftree
\prooftree
\justifies
\mbox{\fbox{$Sf$}}\ \Rightarrow\ Sf
\endprooftree
\justifies
[Nt(s(m))], \mbox{\fbox{${\langle\rangle}Nt(s(m))\backslash Sf$}}\ \Rightarrow\ Sf
\using {\backslash}L
\endprooftree
\justifies
[Nt(s(m))], \mbox{\fbox{$({\langle\rangle}Nt(s(m))\backslash Sf)/({\exists}cNc{\oplus}{\it CP}that)$}}, {\blacksquare}Nt(s(m))\ \Rightarrow\ Sf
\using {/}L
\endprooftree
\justifies
{\langle\rangle}Nt(s(m)), ({\langle\rangle}Nt(s(m))\backslash Sf)/({\exists}cNc{\oplus}{\it CP}that), {\blacksquare}Nt(s(m))\ \Rightarrow\ Sf
\using {\langle\rangle}L
\endprooftree
\justifies
({\langle\rangle}Nt(s(m))\backslash Sf)/({\exists}cNc{\oplus}{\it CP}that), {\blacksquare}Nt(s(m))\ \Rightarrow\ {\langle\rangle}Nt(s(m))\backslash Sf
\using {\backslash}R
\endprooftree
\prooftree
\prooftree
\prooftree
\justifies
Nt(s(m))\ \Rightarrow\ Nt(s(m))
\endprooftree
\justifies
[Nt(s(m))]\ \Rightarrow\ \fbox{${\langle\rangle}Nt(s(m))$}
\using {\langle\rangle}R
\endprooftree
\prooftree
\justifies
\mbox{\fbox{$Sf$}}\ \Rightarrow\ Sf
\endprooftree
\justifies
[Nt(s(m))], \mbox{\fbox{${\langle\rangle}Nt(s(m))\backslash Sf$}}\ \Rightarrow\ Sf
\using {\backslash}L
\endprooftree
\justifies
[Nt(s(m))], ({\langle\rangle}Nt(s(m))\backslash Sf)/({\exists}cNc{\oplus}{\it CP}that), {\blacksquare}Nt(s(m)), \mbox{\fbox{$({\langle\rangle}Nt(s(m))\backslash Sf)\backslash ({\langle\rangle}Nt(s(m))\backslash Sf)$}}\ \Rightarrow\ Sf
\using {\backslash}L
\endprooftree
\justifies
[Nt(s(m))], ({\langle\rangle}Nt(s(m))\backslash Sf)/({\exists}cNc{\oplus}{\it CP}that), {\blacksquare}Nt(s(m)), \mbox{\fbox{${\forall}f(({\langle\rangle}Nt(s(m))\backslash Sf)\backslash ({\langle\rangle}Nt(s(m))\backslash Sf))$}}\ \Rightarrow\ Sf
\using {\forall}L
\endprooftree
\justifies
[Nt(s(m))], ({\langle\rangle}Nt(s(m))\backslash Sf)/({\exists}cNc{\oplus}{\it CP}that), {\blacksquare}Nt(s(m)), \mbox{\fbox{${\forall}a{\forall}f(({\langle\rangle}Na\backslash Sf)\backslash ({\langle\rangle}Na\backslash Sf))$}}\ \Rightarrow\ Sf
\using {\forall}L
\endprooftree
\justifies
[Nt(s(m))], ({\langle\rangle}Nt(s(m))\backslash Sf)/({\exists}cNc{\oplus}{\it CP}that), {\blacksquare}Nt(s(m)), \mbox{\fbox{${\square}{\forall}a{\forall}f(({\langle\rangle}Na\backslash Sf)\backslash ({\langle\rangle}Na\backslash Sf))$}}\ \Rightarrow\ Sf
\using {\Box}L
\endprooftree
\justifies
{\langle\rangle}Nt(s(m)), ({\langle\rangle}Nt(s(m))\backslash Sf)/({\exists}cNc{\oplus}{\it CP}that), {\blacksquare}Nt(s(m)), {\square}{\forall}a{\forall}f(({\langle\rangle}Na\backslash Sf)\backslash ({\langle\rangle}Na\backslash Sf))\ \Rightarrow\ Sf
\using {\langle\rangle}L
\endprooftree
\justifies
({\langle\rangle}Nt(s(m))\backslash Sf)/({\exists}cNc{\oplus}{\it CP}that), {\blacksquare}Nt(s(m)), {\square}{\forall}a{\forall}f(({\langle\rangle}Na\backslash Sf)\backslash ({\langle\rangle}Na\backslash Sf))\ \Rightarrow\ {\langle\rangle}Nt(s(m))\backslash Sf
\using {\backslash}R
\endprooftree
\justifies
{\blacksquare}Nt(s(m)), {\square}{\forall}a{\forall}f(({\langle\rangle}Na\backslash Sf)\backslash ({\langle\rangle}Na\backslash Sf))\ \Rightarrow\ (({\langle\rangle}Nt(s(m))\backslash Sf)/({\exists}cNc{\oplus}{\it CP}that))\backslash ({\langle\rangle}Nt(s(m))\backslash Sf)
\using {\backslash}R
\endprooftree
\justifies
\begin{array}{c}
{\blacksquare}Nt(s(m)), {\square}{\forall}a{\forall}f(({\langle\rangle}Na\backslash Sf)\backslash ({\langle\rangle}Na\backslash Sf))\ \Rightarrow\ {\blacksquare}((({\langle\rangle}Nt(s(m))\backslash Sf)/({\exists}cNc{\oplus}{\it CP}that))\backslash ({\langle\rangle}Nt(s(m))\backslash Sf))\\
\mbox{\footnotesize\textcircled{1}}
\end{array}
\using {\blacksquare}R
\endprooftree

\prooftree
\prooftree
\prooftree
\prooftree
\prooftree
\prooftree
\prooftree
\prooftree
\prooftree
\prooftree
\prooftree
\prooftree
\prooftree
\prooftree
\prooftree
\prooftree
\justifies
\mbox{\fbox{$Nt(s(f))$}}\ \Rightarrow\ Nt(s(f))
\endprooftree
\justifies
\mbox{\fbox{${\blacksquare}Nt(s(f))$}}\ \Rightarrow\ Nt(s(f))
\using {\blacksquare}L
\endprooftree
\justifies
{\blacksquare}Nt(s(f))\ \Rightarrow\ \fbox{${\exists}cNc$}
\using {\exists}R
\endprooftree
\justifies
{\blacksquare}Nt(s(f))\ \Rightarrow\ \fbox{${\exists}cNc{\oplus}{\it CP}that$}
\using {\oplus}R
\endprooftree
\prooftree
\prooftree
\prooftree
\justifies
Nt(s(m))\ \Rightarrow\ Nt(s(m))
\endprooftree
\justifies
[Nt(s(m))]\ \Rightarrow\ \fbox{${\langle\rangle}Nt(s(m))$}
\using {\langle\rangle}R
\endprooftree
\prooftree
\justifies
\mbox{\fbox{$Sf$}}\ \Rightarrow\ Sf
\endprooftree
\justifies
[Nt(s(m))], \mbox{\fbox{${\langle\rangle}Nt(s(m))\backslash Sf$}}\ \Rightarrow\ Sf
\using {\backslash}L
\endprooftree
\justifies
[Nt(s(m))], \mbox{\fbox{$({\langle\rangle}Nt(s(m))\backslash Sf)/({\exists}cNc{\oplus}{\it CP}that)$}}, {\blacksquare}Nt(s(f))\ \Rightarrow\ Sf
\using {/}L
\endprooftree
\justifies
{\langle\rangle}Nt(s(m)), ({\langle\rangle}Nt(s(m))\backslash Sf)/({\exists}cNc{\oplus}{\it CP}that), {\blacksquare}Nt(s(f))\ \Rightarrow\ Sf
\using {\langle\rangle}L
\endprooftree
\justifies
({\langle\rangle}Nt(s(m))\backslash Sf)/({\exists}cNc{\oplus}{\it CP}that), {\blacksquare}Nt(s(f))\ \Rightarrow\ {\langle\rangle}Nt(s(m))\backslash Sf
\using {\backslash}R
\endprooftree
\prooftree
\prooftree
\prooftree
\justifies
Nt(s(m))\ \Rightarrow\ Nt(s(m))
\endprooftree
\justifies
[Nt(s(m))]\ \Rightarrow\ \fbox{${\langle\rangle}Nt(s(m))$}
\using {\langle\rangle}R
\endprooftree
\prooftree
\justifies
\mbox{\fbox{$Sf$}}\ \Rightarrow\ Sf
\endprooftree
\justifies
[Nt(s(m))], \mbox{\fbox{${\langle\rangle}Nt(s(m))\backslash Sf$}}\ \Rightarrow\ Sf
\using {\backslash}L
\endprooftree
\justifies
[Nt(s(m))], ({\langle\rangle}Nt(s(m))\backslash Sf)/({\exists}cNc{\oplus}{\it CP}that), {\blacksquare}Nt(s(f)), \mbox{\fbox{$({\langle\rangle}Nt(s(m))\backslash Sf)\backslash ({\langle\rangle}Nt(s(m))\backslash Sf)$}}\ \Rightarrow\ Sf
\using {\backslash}L
\endprooftree
\justifies
[Nt(s(m))], ({\langle\rangle}Nt(s(m))\backslash Sf)/({\exists}cNc{\oplus}{\it CP}that), {\blacksquare}Nt(s(f)), \mbox{\fbox{${\forall}f(({\langle\rangle}Nt(s(m))\backslash Sf)\backslash ({\langle\rangle}Nt(s(m))\backslash Sf))$}}\ \Rightarrow\ Sf
\using {\forall}L
\endprooftree
\justifies
[Nt(s(m))], ({\langle\rangle}Nt(s(m))\backslash Sf)/({\exists}cNc{\oplus}{\it CP}that), {\blacksquare}Nt(s(f)), \mbox{\fbox{${\forall}a{\forall}f(({\langle\rangle}Na\backslash Sf)\backslash ({\langle\rangle}Na\backslash Sf))$}}\ \Rightarrow\ Sf
\using {\forall}L
\endprooftree
\justifies
[Nt(s(m))], ({\langle\rangle}Nt(s(m))\backslash Sf)/({\exists}cNc{\oplus}{\it CP}that), {\blacksquare}Nt(s(f)), \mbox{\fbox{${\square}{\forall}a{\forall}f(({\langle\rangle}Na\backslash Sf)\backslash ({\langle\rangle}Na\backslash Sf))$}}\ \Rightarrow\ Sf
\using {\Box}L
\endprooftree
\justifies
{\langle\rangle}Nt(s(m)), ({\langle\rangle}Nt(s(m))\backslash Sf)/({\exists}cNc{\oplus}{\it CP}that), {\blacksquare}Nt(s(f)), {\square}{\forall}a{\forall}f(({\langle\rangle}Na\backslash Sf)\backslash ({\langle\rangle}Na\backslash Sf))\ \Rightarrow\ Sf
\using {\langle\rangle}L
\endprooftree
\justifies
({\langle\rangle}Nt(s(m))\backslash Sf)/({\exists}cNc{\oplus}{\it CP}that), {\blacksquare}Nt(s(f)), {\square}{\forall}a{\forall}f(({\langle\rangle}Na\backslash Sf)\backslash ({\langle\rangle}Na\backslash Sf))\ \Rightarrow\ {\langle\rangle}Nt(s(m))\backslash Sf
\using {\backslash}R
\endprooftree
\justifies
{\blacksquare}Nt(s(f)), {\square}{\forall}a{\forall}f(({\langle\rangle}Na\backslash Sf)\backslash ({\langle\rangle}Na\backslash Sf))\ \Rightarrow\ (({\langle\rangle}Nt(s(m))\backslash Sf)/({\exists}cNc{\oplus}{\it CP}that))\backslash ({\langle\rangle}Nt(s(m))\backslash Sf)
\using {\backslash}R
\endprooftree
\justifies
{\blacksquare}Nt(s(f)), {\square}{\forall}a{\forall}f(({\langle\rangle}Na\backslash Sf)\backslash ({\langle\rangle}Na\backslash Sf))\ \Rightarrow\ {\blacksquare}((({\langle\rangle}Nt(s(m))\backslash Sf)/({\exists}cNc{\oplus}{\it CP}that))\backslash ({\langle\rangle}Nt(s(m))\backslash Sf))
\using {\blacksquare}R
\endprooftree
\justifies
\begin{array}{c}
{\blacksquare}Nt(s(f)), {\square}{\forall}a{\forall}f(({\langle\rangle}Na\backslash Sf)\backslash ({\langle\rangle}Na\backslash Sf))\ \Rightarrow\ \fbox{$?{\blacksquare}((({\langle\rangle}Nt(s(m))\backslash Sf)/({\exists}cNc{\oplus}{\it CP}that))\backslash ({\langle\rangle}Nt(s(m))\backslash Sf))$}\\
\mbox{\footnotesize\textcircled{2}}
\end{array}
\using {?}R
\endprooftree

\rotatebox{-90}{
\prooftree
\prooftree
\prooftree
\prooftree
\prooftree
\mbox{\footnotesize\textcircled{1}}\tab
\prooftree
\mbox{\footnotesize\textcircled{2}}\tab
\prooftree
\prooftree
\prooftree
\prooftree
\prooftree
\prooftree
\prooftree
\prooftree
\prooftree
\prooftree
\prooftree
\prooftree
\prooftree
\justifies
N6\ \Rightarrow\ N6
\endprooftree
\justifies
N6\ \Rightarrow\ \fbox{${\exists}aNa$}
\using {\exists}R
\endprooftree
\justifies
N6\ \Rightarrow\ \fbox{${\exists}aNa{\oplus}{\it CP}that$}
\using {\oplus}R
\endprooftree
\prooftree
\prooftree
\prooftree
\prooftree
\justifies
Nt(s(m))\ \Rightarrow\ Nt(s(m))
\endprooftree
\justifies
Nt(s(m))\ \Rightarrow\ \fbox{${\exists}aNa$}
\using {\exists}R
\endprooftree
\justifies
[Nt(s(m))]\ \Rightarrow\ \fbox{${\langle\rangle}{\exists}aNa$}
\using {\langle\rangle}R
\endprooftree
\prooftree
\justifies
\mbox{\fbox{$Sf$}}\ \Rightarrow\ Sf
\endprooftree
\justifies
[Nt(s(m))], \mbox{\fbox{${\langle\rangle}{\exists}aNa\backslash Sf$}}\ \Rightarrow\ Sf
\using {\backslash}L
\endprooftree
\justifies
[Nt(s(m))], \mbox{\fbox{$({\langle\rangle}{\exists}aNa\backslash Sf)/({\exists}aNa{\oplus}{\it CP}that)$}}, N6\ \Rightarrow\ Sf
\using {/}L
\endprooftree
\justifies
[Nt(s(m))], \mbox{\fbox{${\square}(({\langle\rangle}{\exists}aNa\backslash Sf)/({\exists}aNa{\oplus}{\it CP}that))$}}, N6\ \Rightarrow\ Sf
\using {\Box}L
\endprooftree
\justifies
[Nt(s(m))], {\square}(({\langle\rangle}{\exists}aNa\backslash Sf)/({\exists}aNa{\oplus}{\it CP}that)), {\exists}cNc\ \Rightarrow\ Sf
\using {\exists}L
\endprooftree
\justifies
{\langle\rangle}Nt(s(m)), {\square}(({\langle\rangle}{\exists}aNa\backslash Sf)/({\exists}aNa{\oplus}{\it CP}that)), {\exists}cNc\ \Rightarrow\ Sf
\using {\langle\rangle}L
\endprooftree
\prooftree
\prooftree
\prooftree
\prooftree
\prooftree
\justifies
{\it CP}that\ \Rightarrow\ {\it CP}that
\endprooftree
\justifies
{\it CP}that\ \Rightarrow\ \fbox{${\exists}aNa{\oplus}{\it CP}that$}
\using {\oplus}R
\endprooftree
\prooftree
\prooftree
\prooftree
\prooftree
\justifies
Nt(s(m))\ \Rightarrow\ Nt(s(m))
\endprooftree
\justifies
Nt(s(m))\ \Rightarrow\ \fbox{${\exists}aNa$}
\using {\exists}R
\endprooftree
\justifies
[Nt(s(m))]\ \Rightarrow\ \fbox{${\langle\rangle}{\exists}aNa$}
\using {\langle\rangle}R
\endprooftree
\prooftree
\justifies
\mbox{\fbox{$Sf$}}\ \Rightarrow\ Sf
\endprooftree
\justifies
[Nt(s(m))], \mbox{\fbox{${\langle\rangle}{\exists}aNa\backslash Sf$}}\ \Rightarrow\ Sf
\using {\backslash}L
\endprooftree
\justifies
[Nt(s(m))], \mbox{\fbox{$({\langle\rangle}{\exists}aNa\backslash Sf)/({\exists}aNa{\oplus}{\it CP}that)$}}, {\it CP}that\ \Rightarrow\ Sf
\using {/}L
\endprooftree
\justifies
[Nt(s(m))], \mbox{\fbox{${\square}(({\langle\rangle}{\exists}aNa\backslash Sf)/({\exists}aNa{\oplus}{\it CP}that))$}}, {\it CP}that\ \Rightarrow\ Sf
\using {\Box}L
\endprooftree
\justifies
{\langle\rangle}Nt(s(m)), {\square}(({\langle\rangle}{\exists}aNa\backslash Sf)/({\exists}aNa{\oplus}{\it CP}that)), {\it CP}that\ \Rightarrow\ Sf
\using {\langle\rangle}L
\endprooftree
\justifies
{\langle\rangle}Nt(s(m)), {\square}(({\langle\rangle}{\exists}aNa\backslash Sf)/({\exists}aNa{\oplus}{\it CP}that)), {\exists}cNc{\oplus}{\it CP}that\ \Rightarrow\ Sf
\using {\oplus}L
\endprooftree
\justifies
{\square}(({\langle\rangle}{\exists}aNa\backslash Sf)/({\exists}aNa{\oplus}{\it CP}that)), {\exists}cNc{\oplus}{\it CP}that\ \Rightarrow\ {\langle\rangle}Nt(s(m))\backslash Sf
\using {\backslash}R
\endprooftree
\justifies
{\square}(({\langle\rangle}{\exists}aNa\backslash Sf)/({\exists}aNa{\oplus}{\it CP}that))\ \Rightarrow\ ({\langle\rangle}Nt(s(m))\backslash Sf)/({\exists}cNc{\oplus}{\it CP}that)
\using {/}R
\endprooftree
\prooftree
\prooftree
\prooftree
\prooftree
\justifies
\mbox{\fbox{$Nt(s(m))$}}\ \Rightarrow\ Nt(s(m))
\endprooftree
\justifies
\mbox{\fbox{${\blacksquare}Nt(s(m))$}}\ \Rightarrow\ Nt(s(m))
\using {\blacksquare}L
\endprooftree
\justifies
[{\blacksquare}Nt(s(m))]\ \Rightarrow\ \fbox{${\langle\rangle}Nt(s(m))$}
\using {\langle\rangle}R
\endprooftree
\prooftree
\justifies
\mbox{\fbox{$Sf$}}\ \Rightarrow\ Sf
\endprooftree
\justifies
[{\blacksquare}Nt(s(m))], \mbox{\fbox{${\langle\rangle}Nt(s(m))\backslash Sf$}}\ \Rightarrow\ Sf
\using {\backslash}L
\endprooftree
\justifies
[{\blacksquare}Nt(s(m))], {\square}(({\langle\rangle}{\exists}aNa\backslash Sf)/({\exists}aNa{\oplus}{\it CP}that)), \mbox{\fbox{$(({\langle\rangle}Nt(s(m))\backslash Sf)/({\exists}cNc{\oplus}{\it CP}that))\backslash ({\langle\rangle}Nt(s(m))\backslash Sf)$}}\ \Rightarrow\ Sf
\using {\backslash}L
\endprooftree
\justifies
[{\blacksquare}Nt(s(m))], {\square}(({\langle\rangle}{\exists}aNa\backslash Sf)/({\exists}aNa{\oplus}{\it CP}that)), [\mbox{\fbox{${[]^{-1}}((({\langle\rangle}Nt(s(m))\backslash Sf)/({\exists}cNc{\oplus}{\it CP}that))\backslash ({\langle\rangle}Nt(s(m))\backslash Sf))$}}]\ \Rightarrow\ Sf
\using {[]^{-1}}L
\endprooftree
\justifies
[{\blacksquare}Nt(s(m))], {\square}(({\langle\rangle}{\exists}aNa\backslash Sf)/({\exists}aNa{\oplus}{\it CP}that)), [[\mbox{\fbox{${[]^{-1}}{[]^{-1}}((({\langle\rangle}Nt(s(m))\backslash Sf)/({\exists}cNc{\oplus}{\it CP}that))\backslash ({\langle\rangle}Nt(s(m))\backslash Sf))$}}]]\ \Rightarrow\ Sf
\using {[]^{-1}}L
\endprooftree
\justifies
[{\blacksquare}Nt(s(m))], {\square}(({\langle\rangle}{\exists}aNa\backslash Sf)/({\exists}aNa{\oplus}{\it CP}that)), [[{\blacksquare}Nt(s(f)), {\square}{\forall}a{\forall}f(({\langle\rangle}Na\backslash Sf)\backslash ({\langle\rangle}Na\backslash Sf)), \mbox{\fbox{$?{\blacksquare}((({\langle\rangle}Nt(s(m))\backslash Sf)/({\exists}cNc{\oplus}{\it CP}that))\backslash ({\langle\rangle}Nt(s(m))\backslash Sf))\backslash {[]^{-1}}{[]^{-1}}((({\langle\rangle}Nt(s(m))\backslash Sf)/({\exists}cNc{\oplus}{\it CP}that))\backslash ({\langle\rangle}Nt(s(m))\backslash Sf))$}}]]\ \Rightarrow\ Sf
\using {\backslash}L
\endprooftree
\justifies
[{\blacksquare}Nt(s(m))], {\square}(({\langle\rangle}{\exists}aNa\backslash Sf)/({\exists}aNa{\oplus}{\it CP}that)), [[{\blacksquare}Nt(s(f)), {\square}{\forall}a{\forall}f(({\langle\rangle}Na\backslash Sf)\backslash ({\langle\rangle}Na\backslash Sf)), \mbox{\fbox{$(?{\blacksquare}((({\langle\rangle}Nt(s(m))\backslash Sf)/({\exists}cNc{\oplus}{\it CP}that))\backslash ({\langle\rangle}Nt(s(m))\backslash Sf))\backslash {[]^{-1}}{[]^{-1}}((({\langle\rangle}Nt(s(m))\backslash Sf)/({\exists}cNc{\oplus}{\it CP}that))\backslash ({\langle\rangle}Nt(s(m))\backslash Sf)))/{\blacksquare}((({\langle\rangle}Nt(s(m))\backslash Sf)/({\exists}cNc{\oplus}{\it CP}that))\backslash ({\langle\rangle}Nt(s(m))\backslash Sf))$}}, {\blacksquare}Nt(s(m)), {\square}{\forall}a{\forall}f(({\langle\rangle}Na\backslash Sf)\backslash ({\langle\rangle}Na\backslash Sf))]]\ \Rightarrow\ Sf
\using {/}L
\endprooftree
\justifies
[{\blacksquare}Nt(s(m))], {\square}(({\langle\rangle}{\exists}aNa\backslash Sf)/({\exists}aNa{\oplus}{\it CP}that)), [[{\blacksquare}Nt(s(f)), {\square}{\forall}a{\forall}f(({\langle\rangle}Na\backslash Sf)\backslash ({\langle\rangle}Na\backslash Sf)), \mbox{\fbox{${\forall}f((?{\blacksquare}((({\langle\rangle}Nt(s(m))\backslash Sf)/({\exists}cNc{\oplus}{\it CP}that))\backslash ({\langle\rangle}Nt(s(m))\backslash Sf))\backslash {[]^{-1}}{[]^{-1}}((({\langle\rangle}Nt(s(m))\backslash Sf)/({\exists}cNc{\oplus}{\it CP}that))\backslash ({\langle\rangle}Nt(s(m))\backslash Sf)))/{\blacksquare}((({\langle\rangle}Nt(s(m))\backslash Sf)/({\exists}cNc{\oplus}{\it CP}that))\backslash ({\langle\rangle}Nt(s(m))\backslash Sf)))$}}, {\blacksquare}Nt(s(m)), {\square}{\forall}a{\forall}f(({\langle\rangle}Na\backslash Sf)\backslash ({\langle\rangle}Na\backslash Sf))]]\ \Rightarrow\ Sf
\using {\forall}L
\endprooftree
\justifies
[{\blacksquare}Nt(s(m))], {\square}(({\langle\rangle}{\exists}aNa\backslash Sf)/({\exists}aNa{\oplus}{\it CP}that)), [[{\blacksquare}Nt(s(f)), {\square}{\forall}a{\forall}f(({\langle\rangle}Na\backslash Sf)\backslash ({\langle\rangle}Na\backslash Sf)), \mbox{\fbox{${\forall}b{\forall}f((?{\blacksquare}((({\langle\rangle}Nt(s(m))\backslash Sf)/({\exists}cNc{\oplus}{\it CP}b))\backslash ({\langle\rangle}Nt(s(m))\backslash Sf))\backslash {[]^{-1}}{[]^{-1}}((({\langle\rangle}Nt(s(m))\backslash Sf)/({\exists}cNc{\oplus}{\it CP}b))\backslash ({\langle\rangle}Nt(s(m))\backslash Sf)))/{\blacksquare}((({\langle\rangle}Nt(s(m))\backslash Sf)/({\exists}cNc{\oplus}{\it CP}b))\backslash ({\langle\rangle}Nt(s(m))\backslash Sf)))$}}, {\blacksquare}Nt(s(m)), {\square}{\forall}a{\forall}f(({\langle\rangle}Na\backslash Sf)\backslash ({\langle\rangle}Na\backslash Sf))]]\ \Rightarrow\ Sf
\using {\forall}L
\endprooftree
\justifies
[{\blacksquare}Nt(s(m))], {\square}(({\langle\rangle}{\exists}aNa\backslash Sf)/({\exists}aNa{\oplus}{\it CP}that)), [[{\blacksquare}Nt(s(f)), {\square}{\forall}a{\forall}f(({\langle\rangle}Na\backslash Sf)\backslash ({\langle\rangle}Na\backslash Sf)), \mbox{\fbox{${\forall}a{\forall}b{\forall}f((?{\blacksquare}((({\langle\rangle}Na\backslash Sf)/({\exists}cNc{\oplus}{\it CP}b))\backslash ({\langle\rangle}Na\backslash Sf))\backslash {[]^{-1}}{[]^{-1}}((({\langle\rangle}Na\backslash Sf)/({\exists}cNc{\oplus}{\it CP}b))\backslash ({\langle\rangle}Na\backslash Sf)))/{\blacksquare}((({\langle\rangle}Na\backslash Sf)/({\exists}cNc{\oplus}{\it CP}b))\backslash ({\langle\rangle}Na\backslash Sf)))$}}, {\blacksquare}Nt(s(m)), {\square}{\forall}a{\forall}f(({\langle\rangle}Na\backslash Sf)\backslash ({\langle\rangle}Na\backslash Sf))]]\ \Rightarrow\ Sf
\using {\forall}L
\endprooftree
\justifies
[{\blacksquare}Nt(s(m))], {\square}(({\langle\rangle}{\exists}aNa\backslash Sf)/({\exists}aNa{\oplus}{\it CP}that)), [[{\blacksquare}Nt(s(f)), {\square}{\forall}a{\forall}f(({\langle\rangle}Na\backslash Sf)\backslash ({\langle\rangle}Na\backslash Sf)), \mbox{\fbox{${\blacksquare}{\forall}a{\forall}b{\forall}f((?{\blacksquare}((({\langle\rangle}Na\backslash Sf)/({\exists}cNc{\oplus}{\it CP}b))\backslash ({\langle\rangle}Na\backslash Sf))\backslash {[]^{-1}}{[]^{-1}}((({\langle\rangle}Na\backslash Sf)/({\exists}cNc{\oplus}{\it CP}b))\backslash ({\langle\rangle}Na\backslash Sf)))/{\blacksquare}((({\langle\rangle}Na\backslash Sf)/({\exists}cNc{\oplus}{\it CP}b))\backslash ({\langle\rangle}Na\backslash Sf)))$}}, {\blacksquare}Nt(s(m)), {\square}{\forall}a{\forall}f(({\langle\rangle}Na\backslash Sf)\backslash ({\langle\rangle}Na\backslash Sf))]]\ \Rightarrow\ Sf
\using {\blacksquare}L
\endprooftree}}

\vspace{0.15in}

\noindent
This delivers the same semantics:
\disp{
$[(\mbox{\v{}}{\it today}\ ({\it Past}\ ((\mbox{\v{}}{\it seee}\ {\it m})\ {\it j})))\wedge (\mbox{\v{}}{\it yesterday}\ ({\it Past}\ ((\mbox{\v{}}{\it seee}\ {\it b})\ {\it j})))]$}

}

\subsection{Across the board extraction}

This example is (medial) sentential across the board extraction:
\disp{
${\bf man}{+}[[{\bf that}{+}[[[{\bf john}]{+}{\bf saw}{+}{\bf yesterday}{+}{\bf and}{+}[{\bf bill}]{+}{\bf saw}{+}{\bf today}]]]]: {\it CN}{\it s(m)}$}
Appropriate lexical lookup yields the semantically annotated sequent where the
coordinator is essentially of the form $(\exstexp X\bsl \abrack\abrack X)/X$ where
$X=S/\univexp N$.
\disp{
$\begin{array}[t]{l}
{\square}{\it CN}{\it s(m)}: {\it man}, [[{\blacksquare}{\forall}n({[]^{-1}}{[]^{-1}}({\it CN}{\it n}\backslash {\it CN}{\it n})/{\blacksquare}(({\langle\rangle}Nt(n){\sqcap}!{\blacksquare}Nt(n))\backslash Sf)):\\ \lambda A\lambda B\lambda C[({\it B}\ {\it C})\wedge ({\it A}\ {\it C})], [[[{\blacksquare}Nt(s(m)): {\it j}], {\square}(({\langle\rangle}{\exists}aNa\backslash Sf)/({\exists}aNa{\oplus}{\it CP}that)):\\ \mbox{\^{}}\lambda D\lambda E({\it Past}\ (({\it D}\casearrow F.(\mbox{\v{}}{\it seee}\ {\it F}); G.(\mbox{\v{}}{\it seet}\ {\it G}))\ {\it E})), {\square}{\forall}a{\forall}f(({\langle\rangle}Na\backslash Sf)\backslash ({\langle\rangle}Na\backslash Sf)):\\ \mbox{\^{}}\lambda H\lambda I(\mbox{\v{}}{\it yesterday}\ ({\it H}\ {\it I})),
{\blacksquare}{\forall}a{\forall}f((?{\blacksquare}(Sf/!Na)\backslash {[]^{-1}}{[]^{-1}}(Sf/!Na))/{\blacksquare}(Sf/!Na)):\\ (\Phinplus\ ({\it s}\ {\it 0})\ {\it and}), [{\blacksquare}Nt(s(m)): {\it b}], {\square}(({\langle\rangle}{\exists}aNa\backslash Sf)/({\exists}aNa{\oplus}{\it CP}that)):\\ \mbox{\^{}}\lambda J\lambda K({\it Past}\ (({\it J}\casearrow L.(\mbox{\v{}}{\it seee}\ {\it L}); M.(\mbox{\v{}}{\it seet}\ {\it M}))\ {\it K})),\\ {\square}{\forall}a{\forall}f(({\langle\rangle}Na\backslash Sf)\backslash ({\langle\rangle}Na\backslash Sf)): \mbox{\^{}}\lambda N\lambda O(\mbox{\v{}}{\it today}\ ({\it N}\ {\it O}))]]]]\ \Rightarrow\ {\it CN}{\it s(m)}
\end{array}$}
The relative pronoun is essentially of the form $(\CN\bsl\CN)/((\mybrack N\iaconj\univexp N)\bsl S)$
where the semantically inactively conjoined nominals $\mybrack N$
and $\univexp N$
are for subject relativisation
and object relativisation respectively.
There is the following derivation.
In \textcircled{1} the righthand conjunct is derived as a type of the shape $S/\univexp N$.
The universal exponential argument is lowered into the antecedent and into the stoup:
\disp{$\mini
\prooftree
\prooftree
N; \ldots\yields S
\justifies
\ldots, \univexp N\yields S
\using \univexp L
\endprooftree
\justifies
\ldots\yields S/\univexp N
\using /R
\endprooftree$} 
The nominal percolates leftwards in the stoup into the minor premise
at the application of the adverb to the verb phrase and leftwards again in the stoup 
into the minor premise at the application of the (polymorphic) transitive verb to its
object: at the top leftmost subderivation $\univexp P$ brings the nominal
out of the stoup to fulfil the role of this object.
Subtree \textcircled{2} which analyses the lefthand conjunct
is exactly the same as \textcircled{1} --- except for the bottommost existential
exponential right rule --- hence it has been elided.
The principal new action in the main derivation occurs above \textcircled{3}
where the coordinate structure is supplied as the higher order
relative pronoun argument.
The succession has the form:
\disp{$\mini
\prooftree
\prooftree
\prooftree
N; \ldots\yields S
\justifies
\univexp N, \ldots\yields S
\using \univexp L
\endprooftree
\justifies
\mybrack N\iaconj\univexp N, \ldots\yields S
\using \iaconj L
\endprooftree
\justifies
\ldots\yields (\mybrack N\iaconj\univexp N)\bsl S
\using \bsl R
\endprooftree$}
Once the coordinate structure (here `\ldots') is analysed as $S/\univexp N$
the hypothetical nominal subtype which has entered the stoup yields the $\univexp N$
in the subderivation at the top, which has the form:
\disp{$\mini
\prooftree
\prooftree
\prooftree
N\yields N
\justifies
N; \yields N
\using \univexp P
\endprooftree
\justifies
N; \yields \univexp N
\using \univexp R
\endprooftree
S\yields S
\justifies
N; S/\univexp N\yields S
\using /L
\endprooftree$}

\vspace{0.15in}

\begin{center}{\tiny
\prooftree
\prooftree
\prooftree
\prooftree
\prooftree
\prooftree
\prooftree
\prooftree
\prooftree
\prooftree
\prooftree
\prooftree
\prooftree
\prooftree
\prooftree
\justifies
Nt(s(m))\ \Rightarrow\ Nt(s(m))
\endprooftree
\justifies
\mbox{\fbox{$Nt(s(m))$}};\ \ \Rightarrow\ Nt(s(m))
\using {!}P
\endprooftree
\justifies
Nt(s(m));\ \ \Rightarrow\ \fbox{${\exists}aNa$}
\using {\exists}R
\endprooftree
\justifies
Nt(s(m));\ \ \Rightarrow\ \fbox{${\exists}aNa{\oplus}{\it CP}that$}
\using {\oplus}R
\endprooftree
\prooftree
\prooftree
\prooftree
\prooftree
\justifies
Nt(s(m))\ \Rightarrow\ Nt(s(m))
\endprooftree
\justifies
Nt(s(m))\ \Rightarrow\ \fbox{${\exists}aNa$}
\using {\exists}R
\endprooftree
\justifies
[Nt(s(m))]\ \Rightarrow\ \fbox{${\langle\rangle}{\exists}aNa$}
\using {\langle\rangle}R
\endprooftree
\prooftree
\justifies
\mbox{\fbox{$Sf$}}\ \Rightarrow\ Sf
\endprooftree
\justifies
[Nt(s(m))], \mbox{\fbox{${\langle\rangle}{\exists}aNa\backslash Sf$}}\ \Rightarrow\ Sf
\using {\backslash}L
\endprooftree
\justifies
Nt(s(m));\ [Nt(s(m))], \mbox{\fbox{$({\langle\rangle}{\exists}aNa\backslash Sf)/({\exists}aNa{\oplus}{\it CP}that)$}}\ \Rightarrow\ Sf
\using {/}L
\endprooftree
\justifies
Nt(s(m));\ [Nt(s(m))], \mbox{\fbox{${\square}(({\langle\rangle}{\exists}aNa\backslash Sf)/({\exists}aNa{\oplus}{\it CP}that))$}}\ \Rightarrow\ Sf
\using {\Box}L
\endprooftree
\justifies
Nt(s(m));\ {\langle\rangle}Nt(s(m)), {\square}(({\langle\rangle}{\exists}aNa\backslash Sf)/({\exists}aNa{\oplus}{\it CP}that))\ \Rightarrow\ Sf
\using {\langle\rangle}L
\endprooftree
\justifies
Nt(s(m));\ {\square}(({\langle\rangle}{\exists}aNa\backslash Sf)/({\exists}aNa{\oplus}{\it CP}that))\ \Rightarrow\ {\langle\rangle}Nt(s(m))\backslash Sf
\using {\backslash}R
\endprooftree
\prooftree
\prooftree
\prooftree
\prooftree
\justifies
\mbox{\fbox{$Nt(s(m))$}}\ \Rightarrow\ Nt(s(m))
\endprooftree
\justifies
\mbox{\fbox{${\blacksquare}Nt(s(m))$}}\ \Rightarrow\ Nt(s(m))
\using {\blacksquare}L
\endprooftree
\justifies
[{\blacksquare}Nt(s(m))]\ \Rightarrow\ \fbox{${\langle\rangle}Nt(s(m))$}
\using {\langle\rangle}R
\endprooftree
\prooftree
\justifies
\mbox{\fbox{$Sf$}}\ \Rightarrow\ Sf
\endprooftree
\justifies
[{\blacksquare}Nt(s(m))], \mbox{\fbox{${\langle\rangle}Nt(s(m))\backslash Sf$}}\ \Rightarrow\ Sf
\using {\backslash}L
\endprooftree
\justifies
Nt(s(m));\ [{\blacksquare}Nt(s(m))], {\square}(({\langle\rangle}{\exists}aNa\backslash Sf)/({\exists}aNa{\oplus}{\it CP}that)), \mbox{\fbox{$({\langle\rangle}Nt(s(m))\backslash Sf)\backslash ({\langle\rangle}Nt(s(m))\backslash Sf)$}}\ \Rightarrow\ Sf
\using {\backslash}L
\endprooftree
\justifies
Nt(s(m));\ [{\blacksquare}Nt(s(m))], {\square}(({\langle\rangle}{\exists}aNa\backslash Sf)/({\exists}aNa{\oplus}{\it CP}that)), \mbox{\fbox{${\forall}f(({\langle\rangle}Nt(s(m))\backslash Sf)\backslash ({\langle\rangle}Nt(s(m))\backslash Sf))$}}\ \Rightarrow\ Sf
\using {\forall}L
\endprooftree
\justifies
Nt(s(m));\ [{\blacksquare}Nt(s(m))], {\square}(({\langle\rangle}{\exists}aNa\backslash Sf)/({\exists}aNa{\oplus}{\it CP}that)), \mbox{\fbox{${\forall}a{\forall}f(({\langle\rangle}Na\backslash Sf)\backslash ({\langle\rangle}Na\backslash Sf))$}}\ \Rightarrow\ Sf
\using {\forall}L
\endprooftree
\justifies
Nt(s(m));\ [{\blacksquare}Nt(s(m))], {\square}(({\langle\rangle}{\exists}aNa\backslash Sf)/({\exists}aNa{\oplus}{\it CP}that)), \mbox{\fbox{${\square}{\forall}a{\forall}f(({\langle\rangle}Na\backslash Sf)\backslash ({\langle\rangle}Na\backslash Sf))$}}\ \Rightarrow\ Sf
\using {\Box}L
\endprooftree
\justifies
[{\blacksquare}Nt(s(m))], {\square}(({\langle\rangle}{\exists}aNa\backslash Sf)/({\exists}aNa{\oplus}{\it CP}that)), {\square}{\forall}a{\forall}f(({\langle\rangle}Na\backslash Sf)\backslash ({\langle\rangle}Na\backslash Sf)), !Nt(s(m))\ \Rightarrow\ Sf
\using {!}L
\endprooftree
\justifies
[{\blacksquare}Nt(s(m))], {\square}(({\langle\rangle}{\exists}aNa\backslash Sf)/({\exists}aNa{\oplus}{\it CP}that)), {\square}{\forall}a{\forall}f(({\langle\rangle}Na\backslash Sf)\backslash ({\langle\rangle}Na\backslash Sf))\ \Rightarrow\ Sf/!Nt(s(m))
\using {/}R
\endprooftree
\justifies
\begin{array}{c}
[{\blacksquare}Nt(s(m))], {\square}(({\langle\rangle}{\exists}aNa\backslash Sf)/({\exists}aNa{\oplus}{\it CP}that)), {\square}{\forall}a{\forall}f(({\langle\rangle}Na\backslash Sf)\backslash ({\langle\rangle}Na\backslash Sf))\ \Rightarrow\ {\blacksquare}(Sf/!Nt(s(m)))\\
\mbox{\footnotesize\textcircled{1}}
\end{array}
\using {\blacksquare}R
\endprooftree

\prooftree
\vdots
\justifies
\begin{array}{c}
[{\blacksquare}Nt(s(m))], {\square}(({\langle\rangle}{\exists}aNa\backslash Sf)/({\exists}aNa{\oplus}{\it CP}that)), {\square}{\forall}a{\forall}f(({\langle\rangle}Na\backslash Sf)\backslash ({\langle\rangle}Na\backslash Sf))\ \Rightarrow\ \fbox{$?{\blacksquare}(Sf/!Nt(s(m)))$}\\
\mbox{\footnotesize\textcircled{2}}
\end{array}
\using {?}R
\endprooftree}
\end{center}

\begin{center}
\rotatebox{-90}{\tiny
\prooftree
\prooftree
\prooftree
\prooftree
\prooftree
\prooftree
\prooftree
\prooftree
\prooftree
\prooftree
\prooftree
\mbox{\footnotesize\textcircled{1}}\tab
\prooftree
\mbox{\footnotesize\textcircled{2}}\tab
\prooftree
\prooftree
\prooftree
\prooftree
\prooftree
\prooftree
\prooftree
\justifies
\mbox{\fbox{$Nt(s(m))$}}\ \Rightarrow\ Nt(s(m))
\endprooftree
\justifies
\mbox{\fbox{${\blacksquare}Nt(s(m))$}}\ \Rightarrow\ Nt(s(m))
\using {\blacksquare}L
\endprooftree
\justifies
\mbox{\fbox{${\blacksquare}Nt(s(m))$}};\ \ \Rightarrow\ Nt(s(m))
\using {!}P
\endprooftree
\justifies
{\blacksquare}Nt(s(m));\ \ \Rightarrow\ !Nt(s(m))
\using {!}R
\endprooftree
\prooftree
\justifies
\mbox{\fbox{$Sf$}}\ \Rightarrow\ Sf
\endprooftree
\justifies
{\blacksquare}Nt(s(m));\ \mbox{\fbox{$Sf/!Nt(s(m))$}}\ \Rightarrow\ Sf
\using {/}L
\endprooftree
\justifies
{\blacksquare}Nt(s(m));\ [\mbox{\fbox{${[]^{-1}}(Sf/!Nt(s(m)))$}}]\ \Rightarrow\ Sf
\using {[]^{-1}}L
\endprooftree
\justifies
{\blacksquare}Nt(s(m));\ [[\mbox{\fbox{${[]^{-1}}{[]^{-1}}(Sf/!Nt(s(m)))$}}]]\ \Rightarrow\ Sf
\using {[]^{-1}}L
\endprooftree
\justifies
{\blacksquare}Nt(s(m));\ [[[{\blacksquare}Nt(s(m))], {\square}(({\langle\rangle}{\exists}aNa\backslash Sf)/({\exists}aNa{\oplus}{\it CP}that)), {\square}{\forall}a{\forall}f(({\langle\rangle}Na\backslash Sf)\backslash ({\langle\rangle}Na\backslash Sf)), \mbox{\fbox{$?{\blacksquare}(Sf/!Nt(s(m)))\backslash {[]^{-1}}{[]^{-1}}(Sf/!Nt(s(m)))$}}]]\ \Rightarrow\ Sf
\using {\backslash}L
\endprooftree
\justifies
{\blacksquare}Nt(s(m));\ [[[{\blacksquare}Nt(s(m))], {\square}(({\langle\rangle}{\exists}aNa\backslash Sf)/({\exists}aNa{\oplus}{\it CP}that)), {\square}{\forall}a{\forall}f(({\langle\rangle}Na\backslash Sf)\backslash ({\langle\rangle}Na\backslash Sf)), \mbox{\fbox{$(?{\blacksquare}(Sf/!Nt(s(m)))\backslash {[]^{-1}}{[]^{-1}}(Sf/!Nt(s(m))))/{\blacksquare}(Sf/!Nt(s(m)))$}}, [{\blacksquare}Nt(s(m))], {\square}(({\langle\rangle}{\exists}aNa\backslash Sf)/({\exists}aNa{\oplus}{\it CP}that)), {\square}{\forall}a{\forall}f(({\langle\rangle}Na\backslash Sf)\backslash ({\langle\rangle}Na\backslash Sf))]]\ \Rightarrow\ Sf
\using {/}L
\endprooftree
\justifies
{\blacksquare}Nt(s(m));\ [[[{\blacksquare}Nt(s(m))], {\square}(({\langle\rangle}{\exists}aNa\backslash Sf)/({\exists}aNa{\oplus}{\it CP}that)), {\square}{\forall}a{\forall}f(({\langle\rangle}Na\backslash Sf)\backslash ({\langle\rangle}Na\backslash Sf)), \mbox{\fbox{${\forall}f((?{\blacksquare}(Sf/!Nt(s(m)))\backslash {[]^{-1}}{[]^{-1}}(Sf/!Nt(s(m))))/{\blacksquare}(Sf/!Nt(s(m))))$}}, [{\blacksquare}Nt(s(m))], {\square}(({\langle\rangle}{\exists}aNa\backslash Sf)/({\exists}aNa{\oplus}{\it CP}that)), {\square}{\forall}a{\forall}f(({\langle\rangle}Na\backslash Sf)\backslash ({\langle\rangle}Na\backslash Sf))]]\ \Rightarrow\ Sf
\using {\forall}L
\endprooftree
\justifies
{\blacksquare}Nt(s(m));\ [[[{\blacksquare}Nt(s(m))], {\square}(({\langle\rangle}{\exists}aNa\backslash Sf)/({\exists}aNa{\oplus}{\it CP}that)), {\square}{\forall}a{\forall}f(({\langle\rangle}Na\backslash Sf)\backslash ({\langle\rangle}Na\backslash Sf)), \mbox{\fbox{${\forall}a{\forall}f((?{\blacksquare}(Sf/!Na)\backslash {[]^{-1}}{[]^{-1}}(Sf/!Na))/{\blacksquare}(Sf/!Na))$}}, [{\blacksquare}Nt(s(m))], {\square}(({\langle\rangle}{\exists}aNa\backslash Sf)/({\exists}aNa{\oplus}{\it CP}that)), {\square}{\forall}a{\forall}f(({\langle\rangle}Na\backslash Sf)\backslash ({\langle\rangle}Na\backslash Sf))]]\ \Rightarrow\ Sf
\using {\forall}L
\endprooftree
\justifies
{\blacksquare}Nt(s(m));\ [[[{\blacksquare}Nt(s(m))], {\square}(({\langle\rangle}{\exists}aNa\backslash Sf)/({\exists}aNa{\oplus}{\it CP}that)), {\square}{\forall}a{\forall}f(({\langle\rangle}Na\backslash Sf)\backslash ({\langle\rangle}Na\backslash Sf)), \mbox{\fbox{${\blacksquare}{\forall}a{\forall}f((?{\blacksquare}(Sf/!Na)\backslash {[]^{-1}}{[]^{-1}}(Sf/!Na))/{\blacksquare}(Sf/!Na))$}}, [{\blacksquare}Nt(s(m))], {\square}(({\langle\rangle}{\exists}aNa\backslash Sf)/({\exists}aNa{\oplus}{\it CP}that)), {\square}{\forall}a{\forall}f(({\langle\rangle}Na\backslash Sf)\backslash ({\langle\rangle}Na\backslash Sf))]]\ \Rightarrow\ Sf
\using {\blacksquare}L
\endprooftree
\justifies
!{\blacksquare}Nt(s(m)), [[[{\blacksquare}Nt(s(m))], {\square}(({\langle\rangle}{\exists}aNa\backslash Sf)/({\exists}aNa{\oplus}{\it CP}that)), {\square}{\forall}a{\forall}f(({\langle\rangle}Na\backslash Sf)\backslash ({\langle\rangle}Na\backslash Sf)), {\blacksquare}{\forall}a{\forall}f((?{\blacksquare}(Sf/!Na)\backslash {[]^{-1}}{[]^{-1}}(Sf/!Na))/{\blacksquare}(Sf/!Na)), [{\blacksquare}Nt(s(m))], {\square}(({\langle\rangle}{\exists}aNa\backslash Sf)/({\exists}aNa{\oplus}{\it CP}that)), {\square}{\forall}a{\forall}f(({\langle\rangle}Na\backslash Sf)\backslash ({\langle\rangle}Na\backslash Sf))]]\ \Rightarrow\ Sf
\using {!}L
\endprooftree
\justifies
\mbox{\fbox{${\langle\rangle}Nt(s(m)){\sqcap}!{\blacksquare}Nt(s(m))$}}, [[[{\blacksquare}Nt(s(m))], {\square}(({\langle\rangle}{\exists}aNa\backslash Sf)/({\exists}aNa{\oplus}{\it CP}that)), {\square}{\forall}a{\forall}f(({\langle\rangle}Na\backslash Sf)\backslash ({\langle\rangle}Na\backslash Sf)), {\blacksquare}{\forall}a{\forall}f((?{\blacksquare}(Sf/!Na)\backslash {[]^{-1}}{[]^{-1}}(Sf/!Na))/{\blacksquare}(Sf/!Na)), [{\blacksquare}Nt(s(m))], {\square}(({\langle\rangle}{\exists}aNa\backslash Sf)/({\exists}aNa{\oplus}{\it CP}that)), {\square}{\forall}a{\forall}f(({\langle\rangle}Na\backslash Sf)\backslash ({\langle\rangle}Na\backslash Sf))]]\ \Rightarrow\ Sf
\using {\sqcap}L
\endprooftree
\justifies
\begin{array}{c}
{}[[[{\blacksquare}Nt(s(m))], {\square}(({\langle\rangle}{\exists}aNa\backslash Sf)/({\exists}aNa{\oplus}{\it CP}that)), {\square}{\forall}a{\forall}f(({\langle\rangle}Na\backslash Sf)\backslash ({\langle\rangle}Na\backslash Sf)), {\blacksquare}{\forall}a{\forall}f((?{\blacksquare}(Sf/!Na)\backslash {[]^{-1}}{[]^{-1}}(Sf/!Na))/{\blacksquare}(Sf/!Na)), [{\blacksquare}Nt(s(m))], {\square}(({\langle\rangle}{\exists}aNa\backslash Sf)/({\exists}aNa{\oplus}{\it CP}that)), {\square}{\forall}a{\forall}f(({\langle\rangle}Na\backslash Sf)\backslash ({\langle\rangle}Na\backslash Sf))]]\ \Rightarrow\ ({\langle\rangle}Nt(s(m)){\sqcap}!{\blacksquare}Nt(s(m)))\backslash Sf\\
\mbox{\footnotesize\textcircled{3}}
\end{array}
\using {\backslash}R
\endprooftree
\justifies
[[[{\blacksquare}Nt(s(m))], {\square}(({\langle\rangle}{\exists}aNa\backslash Sf)/({\exists}aNa{\oplus}{\it CP}that)), {\square}{\forall}a{\forall}f(({\langle\rangle}Na\backslash Sf)\backslash ({\langle\rangle}Na\backslash Sf)), {\blacksquare}{\forall}a{\forall}f((?{\blacksquare}(Sf/!Na)\backslash {[]^{-1}}{[]^{-1}}(Sf/!Na))/{\blacksquare}(Sf/!Na)), [{\blacksquare}Nt(s(m))], {\square}(({\langle\rangle}{\exists}aNa\backslash Sf)/({\exists}aNa{\oplus}{\it CP}that)), {\square}{\forall}a{\forall}f(({\langle\rangle}Na\backslash Sf)\backslash ({\langle\rangle}Na\backslash Sf))]]\ \Rightarrow\ {\blacksquare}(({\langle\rangle}Nt(s(m)){\sqcap}!{\blacksquare}Nt(s(m)))\backslash Sf)
\using {\blacksquare}R
\endprooftree
\prooftree
\prooftree
\prooftree
\prooftree
\prooftree
\justifies
\mbox{\fbox{${\it CN}{\it s(m)}$}}\ \Rightarrow\ {\it CN}{\it s(m)}
\endprooftree
\justifies
\mbox{\fbox{${\square}{\it CN}{\it s(m)}$}}\ \Rightarrow\ {\it CN}{\it s(m)}
\using {\Box}L
\endprooftree
\prooftree
\justifies
\mbox{\fbox{${\it CN}{\it s(m)}$}}\ \Rightarrow\ {\it CN}{\it s(m)}
\endprooftree
\justifies
{\square}{\it CN}{\it s(m)}, \mbox{\fbox{${\it CN}{\it s(m)}\backslash {\it CN}{\it s(m)}$}}\ \Rightarrow\ {\it CN}{\it s(m)}
\using {\backslash}L
\endprooftree
\justifies
{\square}{\it CN}{\it s(m)}, [\mbox{\fbox{${[]^{-1}}({\it CN}{\it s(m)}\backslash {\it CN}{\it s(m)})$}}]\ \Rightarrow\ {\it CN}{\it s(m)}
\using {[]^{-1}}L
\endprooftree
\justifies
{\square}{\it CN}{\it s(m)}, [[\mbox{\fbox{${[]^{-1}}{[]^{-1}}({\it CN}{\it s(m)}\backslash {\it CN}{\it s(m)})$}}]]\ \Rightarrow\ {\it CN}{\it s(m)}
\using {[]^{-1}}L
\endprooftree
\justifies
{\square}{\it CN}{\it s(m)}, [[\mbox{\fbox{${[]^{-1}}{[]^{-1}}({\it CN}{\it s(m)}\backslash {\it CN}{\it s(m)})/{\blacksquare}(({\langle\rangle}Nt(s(m)){\sqcap}!{\blacksquare}Nt(s(m)))\backslash Sf)$}}, [[[{\blacksquare}Nt(s(m))], {\square}(({\langle\rangle}{\exists}aNa\backslash Sf)/({\exists}aNa{\oplus}{\it CP}that)), {\square}{\forall}a{\forall}f(({\langle\rangle}Na\backslash Sf)\backslash ({\langle\rangle}Na\backslash Sf)), {\blacksquare}{\forall}a{\forall}f((?{\blacksquare}(Sf/!Na)\backslash {[]^{-1}}{[]^{-1}}(Sf/!Na))/{\blacksquare}(Sf/!Na)), [{\blacksquare}Nt(s(m))], {\square}(({\langle\rangle}{\exists}aNa\backslash Sf)/({\exists}aNa{\oplus}{\it CP}that)), {\square}{\forall}a{\forall}f(({\langle\rangle}Na\backslash Sf)\backslash ({\langle\rangle}Na\backslash Sf))]]]]\ \Rightarrow\ {\it CN}{\it s(m)}
\using {/}L
\endprooftree
\justifies
{\square}{\it CN}{\it s(m)}, [[\mbox{\fbox{${\forall}n({[]^{-1}}{[]^{-1}}({\it CN}{\it n}\backslash {\it CN}{\it n})/{\blacksquare}(({\langle\rangle}Nt(n){\sqcap}!{\blacksquare}Nt(n))\backslash Sf))$}}, [[[{\blacksquare}Nt(s(m))], {\square}(({\langle\rangle}{\exists}aNa\backslash Sf)/({\exists}aNa{\oplus}{\it CP}that)), {\square}{\forall}a{\forall}f(({\langle\rangle}Na\backslash Sf)\backslash ({\langle\rangle}Na\backslash Sf)), {\blacksquare}{\forall}a{\forall}f((?{\blacksquare}(Sf/!Na)\backslash {[]^{-1}}{[]^{-1}}(Sf/!Na))/{\blacksquare}(Sf/!Na)), [{\blacksquare}Nt(s(m))], {\square}(({\langle\rangle}{\exists}aNa\backslash Sf)/({\exists}aNa{\oplus}{\it CP}that)), {\square}{\forall}a{\forall}f(({\langle\rangle}Na\backslash Sf)\backslash ({\langle\rangle}Na\backslash Sf))]]]]\ \Rightarrow\ {\it CN}{\it s(m)}
\using {\forall}L
\endprooftree
\justifies
{\square}{\it CN}{\it s(m)}, [[\mbox{\fbox{${\blacksquare}{\forall}n({[]^{-1}}{[]^{-1}}({\it CN}{\it n}\backslash {\it CN}{\it n})/{\blacksquare}(({\langle\rangle}Nt(n){\sqcap}!{\blacksquare}Nt(n))\backslash Sf))$}}, [[[{\blacksquare}Nt(s(m))], {\square}(({\langle\rangle}{\exists}aNa\backslash Sf)/({\exists}aNa{\oplus}{\it CP}that)), {\square}{\forall}a{\forall}f(({\langle\rangle}Na\backslash Sf)\backslash ({\langle\rangle}Na\backslash Sf)), {\blacksquare}{\forall}a{\forall}f((?{\blacksquare}(Sf/!Na)\backslash {[]^{-1}}{[]^{-1}}(Sf/!Na))/{\blacksquare}(Sf/!Na)), [{\blacksquare}Nt(s(m))], {\square}(({\langle\rangle}{\exists}aNa\backslash Sf)/({\exists}aNa{\oplus}{\it CP}that)), {\square}{\forall}a{\forall}f(({\langle\rangle}Na\backslash Sf)\backslash ({\langle\rangle}Na\backslash Sf))]]]]\ \Rightarrow\ {\it CN}{\it s(m)}
\using {\blacksquare}L
\endprooftree}
\end{center}

\vspace{0.15in}

\noindent
This delivers semantics:
\disp{
$\lambda C[(\mbox{\v{}}{\it man}\ {\it C})\wedge [(\mbox{\v{}}{\it yesterday}\ ({\it Past}\ ((\mbox{\v{}}{\it seee}\ {\it C})\ {\it j})))\wedge (\mbox{\v{}}{\it today}\ ({\it Past}\ ((\mbox{\v{}}{\it seee}\ {\it C})\ {\it b})))]]$}

Note that (with bracket modalities) the CSC is respected in TLG as in 
phrase structure grammar and categorial grammar 
formalisms assuming like type coordination schemata,
because there is no coordinator type instance conjoining
sentences only one of which contains a gap.

\commentout{

And~crd(25) is verb phrase (medial) Across-The-Board extraction:
\disp{
(crd(25)) ${\bf man}{+}[[{\bf that}{+}[{\bf john}]{+}[[{\bf saw}{+}{\bf yesterday}{+}{\bf and}{+}{\bf met}{+}{\bf today}]]]]: {\it CN}{\it s(m)}$}
Appropriate lexical insertion yields the sequent:
\disp{
${\square}{\it CN}{\it s(m)}: {\it man}, [[{\blacksquare}{\forall}n({[]^{-1}}{[]^{-1}}({\it CN}{\it n}\backslash {\it CN}{\it n})/{\blacksquare}(({\langle\rangle}Nt(n){\sqcap}!{\blacksquare}Nt(n))\backslash Sf)): \lambda A\lambda B\lambda C[({\it B}\ {\it C})\wedge ({\it A}\ {\it C})], [{\blacksquare}Nt(s(m)): {\it j}], [[{\square}(({\langle\rangle}{\exists}aNa\backslash Sf)/({\exists}aNa{\oplus}{\it CP}that)): \mbox{\^{}}\lambda D\lambda E({\it Past}\ (({\it D}\casearrow F.(\mbox{\v{}}{\it seee}\ {\it F}); G.(\mbox{\v{}}{\it seet}\ {\it G}))\ {\it E})), {\square}{\forall}a{\forall}f(({\langle\rangle}Na\backslash Sf)\backslash ({\langle\rangle}Na\backslash Sf)): \mbox{\^{}}\lambda H\lambda I(\mbox{\v{}}{\it yesterday}\ ({\it H}\ {\it I})),\\
{\blacksquare}{\forall}a{\forall}b{\forall}f((?{\blacksquare}(({\langle\rangle}Na\backslash Sf)/!Nb)\backslash {[]^{-1}}{[]^{-1}}(({\langle\rangle}Na\backslash Sf)/!Nb))/{\blacksquare}(({\langle\rangle}Na\backslash Sf)/!Nb)): (\Phinplus\ ({\it s}\ ({\it s}\ {\it 0}))\ {\it and}), {\square}(({\langle\rangle}{\exists}aNa\backslash Sf)/{\exists}aNa): \mbox{\^{}}\lambda J\lambda K({\it Past}\ ((\mbox{\v{}}{\it meet}\ {\it J})\ {\it K})), {\square}{\forall}a{\forall}f(({\langle\rangle}Na\backslash Sf)\backslash ({\langle\rangle}Na\backslash Sf)): \mbox{\^{}}\lambda L\lambda M(\mbox{\v{}}{\it today}\ ({\it L}\ {\it M}))]]]]\ \Rightarrow\ {\it CN}{\it s(m)}$}
There is the derivation:

\vspace{0.15in}

{\tiny
\prooftree
\prooftree
\prooftree
\prooftree
\prooftree
\prooftree
\prooftree
\prooftree
\prooftree
\prooftree
\prooftree
\prooftree
\prooftree
\prooftree
\justifies
\mbox{\fbox{$Nt(s(m))$}}\ \Rightarrow\ Nt(s(m))
\endprooftree
\justifies
\mbox{\fbox{${\blacksquare}Nt(s(m))$}}\ \Rightarrow\ Nt(s(m))
\using {\blacksquare}L
\endprooftree
\justifies
{\blacksquare}Nt(s(m))\ \Rightarrow\ \fbox{${\exists}bNb$}
\using {\exists}R
\endprooftree
\prooftree
\prooftree
\prooftree
\justifies
Nt(s(m))\ \Rightarrow\ Nt(s(m))
\endprooftree
\justifies
[Nt(s(m))]\ \Rightarrow\ \fbox{${\langle\rangle}Nt(s(m))$}
\using {\langle\rangle}R
\endprooftree
\prooftree
\justifies
\mbox{\fbox{$Sf$}}\ \Rightarrow\ Sf
\endprooftree
\justifies
[Nt(s(m))], \mbox{\fbox{${\langle\rangle}Nt(s(m))\backslash Sf$}}\ \Rightarrow\ Sf
\using {\backslash}L
\endprooftree
\justifies
[Nt(s(m))], \mbox{\fbox{$({\langle\rangle}Nt(s(m))\backslash Sf)/{\exists}bNb$}}, {\blacksquare}Nt(s(m))\ \Rightarrow\ Sf
\using {/}L
\endprooftree
\justifies
{\langle\rangle}Nt(s(m)), ({\langle\rangle}Nt(s(m))\backslash Sf)/{\exists}bNb, {\blacksquare}Nt(s(m))\ \Rightarrow\ Sf
\using {\langle\rangle}L
\endprooftree
\justifies
({\langle\rangle}Nt(s(m))\backslash Sf)/{\exists}bNb, {\blacksquare}Nt(s(m))\ \Rightarrow\ {\langle\rangle}Nt(s(m))\backslash Sf
\using {\backslash}R
\endprooftree
\prooftree
\prooftree
\prooftree
\justifies
Nt(s(m))\ \Rightarrow\ Nt(s(m))
\endprooftree
\justifies
[Nt(s(m))]\ \Rightarrow\ \fbox{${\langle\rangle}Nt(s(m))$}
\using {\langle\rangle}R
\endprooftree
\prooftree
\justifies
\mbox{\fbox{$Sf$}}\ \Rightarrow\ Sf
\endprooftree
\justifies
[Nt(s(m))], \mbox{\fbox{${\langle\rangle}Nt(s(m))\backslash Sf$}}\ \Rightarrow\ Sf
\using {\backslash}L
\endprooftree
\justifies
[Nt(s(m))], ({\langle\rangle}Nt(s(m))\backslash Sf)/{\exists}bNb, {\blacksquare}Nt(s(m)), \mbox{\fbox{$({\langle\rangle}Nt(s(m))\backslash Sf)\backslash ({\langle\rangle}Nt(s(m))\backslash Sf)$}}\ \Rightarrow\ Sf
\using {\backslash}L
\endprooftree
\justifies
[Nt(s(m))], ({\langle\rangle}Nt(s(m))\backslash Sf)/{\exists}bNb, {\blacksquare}Nt(s(m)), \mbox{\fbox{${\forall}f(({\langle\rangle}Nt(s(m))\backslash Sf)\backslash ({\langle\rangle}Nt(s(m))\backslash Sf))$}}\ \Rightarrow\ Sf
\using {\forall}L
\endprooftree
\justifies
[Nt(s(m))], ({\langle\rangle}Nt(s(m))\backslash Sf)/{\exists}bNb, {\blacksquare}Nt(s(m)), \mbox{\fbox{${\forall}a{\forall}f(({\langle\rangle}Na\backslash Sf)\backslash ({\langle\rangle}Na\backslash Sf))$}}\ \Rightarrow\ Sf
\using {\forall}L
\endprooftree
\justifies
[Nt(s(m))], ({\langle\rangle}Nt(s(m))\backslash Sf)/{\exists}bNb, {\blacksquare}Nt(s(m)), \mbox{\fbox{${\square}{\forall}a{\forall}f(({\langle\rangle}Na\backslash Sf)\backslash ({\langle\rangle}Na\backslash Sf))$}}\ \Rightarrow\ Sf
\using {\Box}L
\endprooftree
\justifies
{\langle\rangle}Nt(s(m)), ({\langle\rangle}Nt(s(m))\backslash Sf)/{\exists}bNb, {\blacksquare}Nt(s(m)), {\square}{\forall}a{\forall}f(({\langle\rangle}Na\backslash Sf)\backslash ({\langle\rangle}Na\backslash Sf))\ \Rightarrow\ Sf
\using {\langle\rangle}L
\endprooftree
\justifies
({\langle\rangle}Nt(s(m))\backslash Sf)/{\exists}bNb, {\blacksquare}Nt(s(m)), {\square}{\forall}a{\forall}f(({\langle\rangle}Na\backslash Sf)\backslash ({\langle\rangle}Na\backslash Sf))\ \Rightarrow\ {\langle\rangle}Nt(s(m))\backslash Sf
\using {\backslash}R
\endprooftree
\justifies
{\blacksquare}Nt(s(m)), {\square}{\forall}a{\forall}f(({\langle\rangle}Na\backslash Sf)\backslash ({\langle\rangle}Na\backslash Sf))\ \Rightarrow\ (({\langle\rangle}Nt(s(m))\backslash Sf)/{\exists}bNb)\backslash ({\langle\rangle}Nt(s(m))\backslash Sf)
\using {\backslash}R
\endprooftree
\justifies
\begin{array}{c}
{\blacksquare}Nt(s(m)), {\square}{\forall}a{\forall}f(({\langle\rangle}Na\backslash Sf)\backslash ({\langle\rangle}Na\backslash Sf))\ \Rightarrow\ {\blacksquare}((({\langle\rangle}Nt(s(m))\backslash Sf)/{\exists}bNb)\backslash ({\langle\rangle}Nt(s(m))\backslash Sf))\\
\mbox{\footnotesize\textcircled{1}}
\end{array}
\using {\blacksquare}R
\endprooftree

\prooftree
\prooftree
\prooftree
\prooftree
\prooftree
\prooftree
\prooftree
\prooftree
\prooftree
\prooftree
\prooftree
\prooftree
\prooftree
\prooftree
\prooftree
\justifies
\mbox{\fbox{$Nt(s(f))$}}\ \Rightarrow\ Nt(s(f))
\endprooftree
\justifies
\mbox{\fbox{${\blacksquare}Nt(s(f))$}}\ \Rightarrow\ Nt(s(f))
\using {\blacksquare}L
\endprooftree
\justifies
{\blacksquare}Nt(s(f))\ \Rightarrow\ \fbox{${\exists}bNb$}
\using {\exists}R
\endprooftree
\prooftree
\prooftree
\prooftree
\justifies
Nt(s(m))\ \Rightarrow\ Nt(s(m))
\endprooftree
\justifies
[Nt(s(m))]\ \Rightarrow\ \fbox{${\langle\rangle}Nt(s(m))$}
\using {\langle\rangle}R
\endprooftree
\prooftree
\justifies
\mbox{\fbox{$Sf$}}\ \Rightarrow\ Sf
\endprooftree
\justifies
[Nt(s(m))], \mbox{\fbox{${\langle\rangle}Nt(s(m))\backslash Sf$}}\ \Rightarrow\ Sf
\using {\backslash}L
\endprooftree
\justifies
[Nt(s(m))], \mbox{\fbox{$({\langle\rangle}Nt(s(m))\backslash Sf)/{\exists}bNb$}}, {\blacksquare}Nt(s(f))\ \Rightarrow\ Sf
\using {/}L
\endprooftree
\justifies
{\langle\rangle}Nt(s(m)), ({\langle\rangle}Nt(s(m))\backslash Sf)/{\exists}bNb, {\blacksquare}Nt(s(f))\ \Rightarrow\ Sf
\using {\langle\rangle}L
\endprooftree
\justifies
({\langle\rangle}Nt(s(m))\backslash Sf)/{\exists}bNb, {\blacksquare}Nt(s(f))\ \Rightarrow\ {\langle\rangle}Nt(s(m))\backslash Sf
\using {\backslash}R
\endprooftree
\prooftree
\prooftree
\prooftree
\justifies
Nt(s(m))\ \Rightarrow\ Nt(s(m))
\endprooftree
\justifies
[Nt(s(m))]\ \Rightarrow\ \fbox{${\langle\rangle}Nt(s(m))$}
\using {\langle\rangle}R
\endprooftree
\prooftree
\justifies
\mbox{\fbox{$Sf$}}\ \Rightarrow\ Sf
\endprooftree
\justifies
[Nt(s(m))], \mbox{\fbox{${\langle\rangle}Nt(s(m))\backslash Sf$}}\ \Rightarrow\ Sf
\using {\backslash}L
\endprooftree
\justifies
[Nt(s(m))], ({\langle\rangle}Nt(s(m))\backslash Sf)/{\exists}bNb, {\blacksquare}Nt(s(f)), \mbox{\fbox{$({\langle\rangle}Nt(s(m))\backslash Sf)\backslash ({\langle\rangle}Nt(s(m))\backslash Sf)$}}\ \Rightarrow\ Sf
\using {\backslash}L
\endprooftree
\justifies
[Nt(s(m))], ({\langle\rangle}Nt(s(m))\backslash Sf)/{\exists}bNb, {\blacksquare}Nt(s(f)), \mbox{\fbox{${\forall}f(({\langle\rangle}Nt(s(m))\backslash Sf)\backslash ({\langle\rangle}Nt(s(m))\backslash Sf))$}}\ \Rightarrow\ Sf
\using {\forall}L
\endprooftree
\justifies
[Nt(s(m))], ({\langle\rangle}Nt(s(m))\backslash Sf)/{\exists}bNb, {\blacksquare}Nt(s(f)), \mbox{\fbox{${\forall}a{\forall}f(({\langle\rangle}Na\backslash Sf)\backslash ({\langle\rangle}Na\backslash Sf))$}}\ \Rightarrow\ Sf
\using {\forall}L
\endprooftree
\justifies
[Nt(s(m))], ({\langle\rangle}Nt(s(m))\backslash Sf)/{\exists}bNb, {\blacksquare}Nt(s(f)), \mbox{\fbox{${\square}{\forall}a{\forall}f(({\langle\rangle}Na\backslash Sf)\backslash ({\langle\rangle}Na\backslash Sf))$}}\ \Rightarrow\ Sf
\using {\Box}L
\endprooftree
\justifies
{\langle\rangle}Nt(s(m)), ({\langle\rangle}Nt(s(m))\backslash Sf)/{\exists}bNb, {\blacksquare}Nt(s(f)), {\square}{\forall}a{\forall}f(({\langle\rangle}Na\backslash Sf)\backslash ({\langle\rangle}Na\backslash Sf))\ \Rightarrow\ Sf
\using {\langle\rangle}L
\endprooftree
\justifies
({\langle\rangle}Nt(s(m))\backslash Sf)/{\exists}bNb, {\blacksquare}Nt(s(f)), {\square}{\forall}a{\forall}f(({\langle\rangle}Na\backslash Sf)\backslash ({\langle\rangle}Na\backslash Sf))\ \Rightarrow\ {\langle\rangle}Nt(s(m))\backslash Sf
\using {\backslash}R
\endprooftree
\justifies
{\blacksquare}Nt(s(f)), {\square}{\forall}a{\forall}f(({\langle\rangle}Na\backslash Sf)\backslash ({\langle\rangle}Na\backslash Sf))\ \Rightarrow\ (({\langle\rangle}Nt(s(m))\backslash Sf)/{\exists}bNb)\backslash ({\langle\rangle}Nt(s(m))\backslash Sf)
\using {\backslash}R
\endprooftree
\justifies
{\blacksquare}Nt(s(f)), {\square}{\forall}a{\forall}f(({\langle\rangle}Na\backslash Sf)\backslash ({\langle\rangle}Na\backslash Sf))\ \Rightarrow\ {\blacksquare}((({\langle\rangle}Nt(s(m))\backslash Sf)/{\exists}bNb)\backslash ({\langle\rangle}Nt(s(m))\backslash Sf))
\using {\blacksquare}R
\endprooftree
\justifies
\begin{array}{c}
{\blacksquare}Nt(s(f)), {\square}{\forall}a{\forall}f(({\langle\rangle}Na\backslash Sf)\backslash ({\langle\rangle}Na\backslash Sf))\ \Rightarrow\ \fbox{$?{\blacksquare}((({\langle\rangle}Nt(s(m))\backslash Sf)/{\exists}bNb)\backslash ({\langle\rangle}Nt(s(m))\backslash Sf))$}\\
\mbox{\footnotesize\textcircled{2}}
\end{array}
\using {?}R
\endprooftree

\rotatebox{-90}{
\prooftree
\prooftree
\prooftree
\prooftree
\mbox{\footnotesize\textcircled{1}}\tab
\prooftree
\mbox{\footnotesize\textcircled{2}}\tab
\prooftree
\prooftree
\prooftree
\prooftree
\prooftree
\prooftree
\prooftree
\prooftree
\prooftree
\prooftree
\prooftree
\prooftree
\justifies
N1\ \Rightarrow\ N1
\endprooftree
\justifies
N1\ \Rightarrow\ \fbox{${\exists}aNa$}
\using {\exists}R
\endprooftree
\justifies
N1\ \Rightarrow\ \fbox{${\exists}aNa{\oplus}{\it CP}that$}
\using {\oplus}R
\endprooftree
\prooftree
\prooftree
\prooftree
\prooftree
\justifies
Nt(s(m))\ \Rightarrow\ Nt(s(m))
\endprooftree
\justifies
Nt(s(m))\ \Rightarrow\ \fbox{${\exists}aNa$}
\using {\exists}R
\endprooftree
\justifies
[Nt(s(m))]\ \Rightarrow\ \fbox{${\langle\rangle}{\exists}aNa$}
\using {\langle\rangle}R
\endprooftree
\prooftree
\justifies
\mbox{\fbox{$Sf$}}\ \Rightarrow\ Sf
\endprooftree
\justifies
[Nt(s(m))], \mbox{\fbox{${\langle\rangle}{\exists}aNa\backslash Sf$}}\ \Rightarrow\ Sf
\using {\backslash}L
\endprooftree
\justifies
[Nt(s(m))], \mbox{\fbox{$({\langle\rangle}{\exists}aNa\backslash Sf)/({\exists}aNa{\oplus}{\it CP}that)$}}, N1\ \Rightarrow\ Sf
\using {/}L
\endprooftree
\justifies
[Nt(s(m))], \mbox{\fbox{${\square}(({\langle\rangle}{\exists}aNa\backslash Sf)/({\exists}aNa{\oplus}{\it CP}that))$}}, N1\ \Rightarrow\ Sf
\using {\Box}L
\endprooftree
\justifies
[Nt(s(m))], {\square}(({\langle\rangle}{\exists}aNa\backslash Sf)/({\exists}aNa{\oplus}{\it CP}that)), {\exists}bNb\ \Rightarrow\ Sf
\using {\exists}L
\endprooftree
\justifies
{\langle\rangle}Nt(s(m)), {\square}(({\langle\rangle}{\exists}aNa\backslash Sf)/({\exists}aNa{\oplus}{\it CP}that)), {\exists}bNb\ \Rightarrow\ Sf
\using {\langle\rangle}L
\endprooftree
\justifies
{\square}(({\langle\rangle}{\exists}aNa\backslash Sf)/({\exists}aNa{\oplus}{\it CP}that)), {\exists}bNb\ \Rightarrow\ {\langle\rangle}Nt(s(m))\backslash Sf
\using {\backslash}R
\endprooftree
\justifies
{\square}(({\langle\rangle}{\exists}aNa\backslash Sf)/({\exists}aNa{\oplus}{\it CP}that))\ \Rightarrow\ ({\langle\rangle}Nt(s(m))\backslash Sf)/{\exists}bNb
\using {/}R
\endprooftree
\prooftree
\prooftree
\prooftree
\prooftree
\justifies
\mbox{\fbox{$Nt(s(m))$}}\ \Rightarrow\ Nt(s(m))
\endprooftree
\justifies
\mbox{\fbox{${\blacksquare}Nt(s(m))$}}\ \Rightarrow\ Nt(s(m))
\using {\blacksquare}L
\endprooftree
\justifies
[{\blacksquare}Nt(s(m))]\ \Rightarrow\ \fbox{${\langle\rangle}Nt(s(m))$}
\using {\langle\rangle}R
\endprooftree
\prooftree
\justifies
\mbox{\fbox{$Sf$}}\ \Rightarrow\ Sf
\endprooftree
\justifies
[{\blacksquare}Nt(s(m))], \mbox{\fbox{${\langle\rangle}Nt(s(m))\backslash Sf$}}\ \Rightarrow\ Sf
\using {\backslash}L
\endprooftree
\justifies
[{\blacksquare}Nt(s(m))], {\square}(({\langle\rangle}{\exists}aNa\backslash Sf)/({\exists}aNa{\oplus}{\it CP}that)), \mbox{\fbox{$(({\langle\rangle}Nt(s(m))\backslash Sf)/{\exists}bNb)\backslash ({\langle\rangle}Nt(s(m))\backslash Sf)$}}\ \Rightarrow\ Sf
\using {\backslash}L
\endprooftree
\justifies
[{\blacksquare}Nt(s(m))], {\square}(({\langle\rangle}{\exists}aNa\backslash Sf)/({\exists}aNa{\oplus}{\it CP}that)), [\mbox{\fbox{${[]^{-1}}((({\langle\rangle}Nt(s(m))\backslash Sf)/{\exists}bNb)\backslash ({\langle\rangle}Nt(s(m))\backslash Sf))$}}]\ \Rightarrow\ Sf
\using {[]^{-1}}L
\endprooftree
\justifies
[{\blacksquare}Nt(s(m))], {\square}(({\langle\rangle}{\exists}aNa\backslash Sf)/({\exists}aNa{\oplus}{\it CP}that)), [[\mbox{\fbox{${[]^{-1}}{[]^{-1}}((({\langle\rangle}Nt(s(m))\backslash Sf)/{\exists}bNb)\backslash ({\langle\rangle}Nt(s(m))\backslash Sf))$}}]]\ \Rightarrow\ Sf
\using {[]^{-1}}L
\endprooftree
\justifies
[{\blacksquare}Nt(s(m))], {\square}(({\langle\rangle}{\exists}aNa\backslash Sf)/({\exists}aNa{\oplus}{\it CP}that)), [[{\blacksquare}Nt(s(f)), {\square}{\forall}a{\forall}f(({\langle\rangle}Na\backslash Sf)\backslash ({\langle\rangle}Na\backslash Sf)), \mbox{\fbox{$?{\blacksquare}((({\langle\rangle}Nt(s(m))\backslash Sf)/{\exists}bNb)\backslash ({\langle\rangle}Nt(s(m))\backslash Sf))\backslash {[]^{-1}}{[]^{-1}}((({\langle\rangle}Nt(s(m))\backslash Sf)/{\exists}bNb)\backslash ({\langle\rangle}Nt(s(m))\backslash Sf))$}}]]\ \Rightarrow\ Sf
\using {\backslash}L
\endprooftree
\justifies
[{\blacksquare}Nt(s(m))], {\square}(({\langle\rangle}{\exists}aNa\backslash Sf)/({\exists}aNa{\oplus}{\it CP}that)), [[{\blacksquare}Nt(s(f)), {\square}{\forall}a{\forall}f(({\langle\rangle}Na\backslash Sf)\backslash ({\langle\rangle}Na\backslash Sf)), \mbox{\fbox{$(?{\blacksquare}((({\langle\rangle}Nt(s(m))\backslash Sf)/{\exists}bNb)\backslash ({\langle\rangle}Nt(s(m))\backslash Sf))\backslash {[]^{-1}}{[]^{-1}}((({\langle\rangle}Nt(s(m))\backslash Sf)/{\exists}bNb)\backslash ({\langle\rangle}Nt(s(m))\backslash Sf)))/{\blacksquare}((({\langle\rangle}Nt(s(m))\backslash Sf)/{\exists}bNb)\backslash ({\langle\rangle}Nt(s(m))\backslash Sf))$}}, {\blacksquare}Nt(s(m)), {\square}{\forall}a{\forall}f(({\langle\rangle}Na\backslash Sf)\backslash ({\langle\rangle}Na\backslash Sf))]]\ \Rightarrow\ Sf
\using {/}L
\endprooftree
\justifies
[{\blacksquare}Nt(s(m))], {\square}(({\langle\rangle}{\exists}aNa\backslash Sf)/({\exists}aNa{\oplus}{\it CP}that)), [[{\blacksquare}Nt(s(f)), {\square}{\forall}a{\forall}f(({\langle\rangle}Na\backslash Sf)\backslash ({\langle\rangle}Na\backslash Sf)), \mbox{\fbox{${\forall}a((?{\blacksquare}((({\langle\rangle}Na\backslash Sf)/{\exists}bNb)\backslash ({\langle\rangle}Na\backslash Sf))\backslash {[]^{-1}}{[]^{-1}}((({\langle\rangle}Na\backslash Sf)/{\exists}bNb)\backslash ({\langle\rangle}Na\backslash Sf)))/{\blacksquare}((({\langle\rangle}Na\backslash Sf)/{\exists}bNb)\backslash ({\langle\rangle}Na\backslash Sf)))$}}, {\blacksquare}Nt(s(m)), {\square}{\forall}a{\forall}f(({\langle\rangle}Na\backslash Sf)\backslash ({\langle\rangle}Na\backslash Sf))]]\ \Rightarrow\ Sf
\using {\forall}L
\endprooftree
\justifies
[{\blacksquare}Nt(s(m))], {\square}(({\langle\rangle}{\exists}aNa\backslash Sf)/({\exists}aNa{\oplus}{\it CP}that)), [[{\blacksquare}Nt(s(f)), {\square}{\forall}a{\forall}f(({\langle\rangle}Na\backslash Sf)\backslash ({\langle\rangle}Na\backslash Sf)), \mbox{\fbox{${\forall}f{\forall}a((?{\blacksquare}((({\langle\rangle}Na\backslash Sf)/{\exists}bNb)\backslash ({\langle\rangle}Na\backslash Sf))\backslash {[]^{-1}}{[]^{-1}}((({\langle\rangle}Na\backslash Sf)/{\exists}bNb)\backslash ({\langle\rangle}Na\backslash Sf)))/{\blacksquare}((({\langle\rangle}Na\backslash Sf)/{\exists}bNb)\backslash ({\langle\rangle}Na\backslash Sf)))$}}, {\blacksquare}Nt(s(m)), {\square}{\forall}a{\forall}f(({\langle\rangle}Na\backslash Sf)\backslash ({\langle\rangle}Na\backslash Sf))]]\ \Rightarrow\ Sf
\using {\forall}L
\endprooftree
\justifies
[{\blacksquare}Nt(s(m))], {\square}(({\langle\rangle}{\exists}aNa\backslash Sf)/({\exists}aNa{\oplus}{\it CP}that)), [[{\blacksquare}Nt(s(f)), {\square}{\forall}a{\forall}f(({\langle\rangle}Na\backslash Sf)\backslash ({\langle\rangle}Na\backslash Sf)), \mbox{\fbox{${\blacksquare}{\forall}f{\forall}a((?{\blacksquare}((({\langle\rangle}Na\backslash Sf)/{\exists}bNb)\backslash ({\langle\rangle}Na\backslash Sf))\backslash {[]^{-1}}{[]^{-1}}((({\langle\rangle}Na\backslash Sf)/{\exists}bNb)\backslash ({\langle\rangle}Na\backslash Sf)))/{\blacksquare}((({\langle\rangle}Na\backslash Sf)/{\exists}bNb)\backslash ({\langle\rangle}Na\backslash Sf)))$}}, {\blacksquare}Nt(s(m)), {\square}{\forall}a{\forall}f(({\langle\rangle}Na\backslash Sf)\backslash ({\langle\rangle}Na\backslash Sf))]]\ \Rightarrow\ Sf
\using {\blacksquare}L
\endprooftree}}

\vspace{0.15in}

\noindent
This delivers semantics:
\disp{
$\lambda C[(\mbox{\v{}}{\it man}\ {\it C})\wedge [(\mbox{\v{}}{\it yesterday}\ ({\it Past}\ ((\mbox{\v{}}{\it seee}\ {\it C})\ {\it j})))\wedge (\mbox{\v{}}{\it today}\ ({\it Past}\ ((\mbox{\v{}}{\it meet}\ {\it C})\ {\it j})))]]$}

}

\section{Iterated coordination}

We consider examples of iterated coordination.

\subsection{Iterated coordination of addicity zero}

Minimally we have the example:
\disp{
$[[[{\bf john}]{+}{\bf walks}{+}[{\bf mary}]{+}{\bf talks}{+}{\bf and}{+}[{\bf bill}]{+}{\bf sings}]]: Sf$}
Appropriate lexical lookup yields:
\disp{
$\begin{array}[t]{l}
[[[{\blacksquare}Nt(s(m)): {\it j}], {\square}({\langle\rangle}{\exists}gNt(s(g))\backslash Sf): \mbox{\^{}}\lambda A({\it Pres}\ (\mbox{\v{}}{\it walk}\ {\it A})),\\{} [{\blacksquare}Nt(s(f)): {\it m}], {\square}({\langle\rangle}{\exists}gNt(s(g))\backslash Sf): \mbox{\^{}}\lambda B({\it Pres}\ (\mbox{\v{}}{\it talk}\ {\it B})), \\{\blacksquare}{\forall}f((?{\blacksquare}Sf\backslash {[]^{-1}}{[]^{-1}}Sf)/{\blacksquare}Sf): (\Phinplus\ {\it 0}\ {\it and}), [{\blacksquare}Nt(s(m)): {\it b}],\\ {\square}({\langle\rangle}{\exists}gNt(s(g))\backslash Sf): \mbox{\^{}}\lambda C({\it Pres}\ (\mbox{\v{}}{\it sing}\ {\it C}))]]\ \Rightarrow\ Sf
\end{array}$}
The coordination combinator $(\Phinplus\ 0\ {\it and})$ is such that:
\disp{
$\begin{array}[t]{l}
(((\Phinplus\ 0\ {\it and})\ x)\ [y, z]) =\\
{}[y\wedge(((\Phinplus\ 0\ {\it and})\ x)\ [z])] =\\
{}[y\wedge[z\wedge x]]
\end{array}$}
There is the derivation:

\vspace{0.15in}

\begin{center}
\rotatebox{-90}{\tiny
\prooftree
\prooftree
\prooftree
\prooftree
\prooftree
\prooftree
\prooftree
\prooftree
\prooftree
\prooftree
\justifies
\mbox{\fbox{$Nt(s(m))$}}\ \Rightarrow\ Nt(s(m))
\endprooftree
\justifies
\mbox{\fbox{${\blacksquare}Nt(s(m))$}}\ \Rightarrow\ Nt(s(m))
\using {\blacksquare}L
\endprooftree
\justifies
{\blacksquare}Nt(s(m))\ \Rightarrow\ \fbox{${\exists}gNt(s(g))$}
\using {\exists}R
\endprooftree
\justifies
[{\blacksquare}Nt(s(m))]\ \Rightarrow\ \fbox{${\langle\rangle}{\exists}gNt(s(g))$}
\using {\langle\rangle}R
\endprooftree
\prooftree
\justifies
\mbox{\fbox{$Sf$}}\ \Rightarrow\ Sf
\endprooftree
\justifies
[{\blacksquare}Nt(s(m))], \mbox{\fbox{${\langle\rangle}{\exists}gNt(s(g))\backslash Sf$}}\ \Rightarrow\ Sf
\using {\backslash}L
\endprooftree
\justifies
[{\blacksquare}Nt(s(m))], \mbox{\fbox{${\square}({\langle\rangle}{\exists}gNt(s(g))\backslash Sf)$}}\ \Rightarrow\ Sf
\using {\Box}L
\endprooftree
\justifies
[{\blacksquare}Nt(s(m))], {\square}({\langle\rangle}{\exists}gNt(s(g))\backslash Sf)\ \Rightarrow\ {\blacksquare}Sf
\using {\blacksquare}R
\endprooftree
\prooftree
\prooftree
\prooftree
\prooftree
\prooftree
\prooftree
\prooftree
\prooftree
\prooftree
\justifies
\mbox{\fbox{$Nt(s(m))$}}\ \Rightarrow\ Nt(s(m))
\endprooftree
\justifies
\mbox{\fbox{${\blacksquare}Nt(s(m))$}}\ \Rightarrow\ Nt(s(m))
\using {\blacksquare}L
\endprooftree
\justifies
{\blacksquare}Nt(s(m))\ \Rightarrow\ \fbox{${\exists}gNt(s(g))$}
\using {\exists}R
\endprooftree
\justifies
[{\blacksquare}Nt(s(m))]\ \Rightarrow\ \fbox{${\langle\rangle}{\exists}gNt(s(g))$}
\using {\langle\rangle}R
\endprooftree
\prooftree
\justifies
\mbox{\fbox{$Sf$}}\ \Rightarrow\ Sf
\endprooftree
\justifies
[{\blacksquare}Nt(s(m))], \mbox{\fbox{${\langle\rangle}{\exists}gNt(s(g))\backslash Sf$}}\ \Rightarrow\ Sf
\using {\backslash}L
\endprooftree
\justifies
[{\blacksquare}Nt(s(m))], \mbox{\fbox{${\square}({\langle\rangle}{\exists}gNt(s(g))\backslash Sf)$}}\ \Rightarrow\ Sf
\using {\Box}L
\endprooftree
\justifies
[{\blacksquare}Nt(s(m))], {\square}({\langle\rangle}{\exists}gNt(s(g))\backslash Sf)\ \Rightarrow\ {\blacksquare}Sf
\using {\blacksquare}R
\endprooftree
\prooftree
\prooftree
\prooftree
\prooftree
\prooftree
\prooftree
\prooftree
\prooftree
\justifies
\mbox{\fbox{$Nt(s(f))$}}\ \Rightarrow\ Nt(s(f))
\endprooftree
\justifies
\mbox{\fbox{${\blacksquare}Nt(s(f))$}}\ \Rightarrow\ Nt(s(f))
\using {\blacksquare}L
\endprooftree
\justifies
{\blacksquare}Nt(s(f))\ \Rightarrow\ \fbox{${\exists}gNt(s(g))$}
\using {\exists}R
\endprooftree
\justifies
[{\blacksquare}Nt(s(f))]\ \Rightarrow\ \fbox{${\langle\rangle}{\exists}gNt(s(g))$}
\using {\langle\rangle}R
\endprooftree
\prooftree
\justifies
\mbox{\fbox{$Sf$}}\ \Rightarrow\ Sf
\endprooftree
\justifies
[{\blacksquare}Nt(s(f))], \mbox{\fbox{${\langle\rangle}{\exists}gNt(s(g))\backslash Sf$}}\ \Rightarrow\ Sf
\using {\backslash}L
\endprooftree
\justifies
[{\blacksquare}Nt(s(f))], \mbox{\fbox{${\square}({\langle\rangle}{\exists}gNt(s(g))\backslash Sf)$}}\ \Rightarrow\ Sf
\using {\Box}L
\endprooftree
\justifies
[{\blacksquare}Nt(s(f))], {\square}({\langle\rangle}{\exists}gNt(s(g))\backslash Sf)\ \Rightarrow\ {\blacksquare}Sf
\using {\blacksquare}R
\endprooftree
\justifies
[{\blacksquare}Nt(s(f))], {\square}({\langle\rangle}{\exists}gNt(s(g))\backslash Sf)\ \Rightarrow\ \fbox{$?{\blacksquare}Sf$}
\using {?}R
\endprooftree
\justifies
[{\blacksquare}Nt(s(m))], {\square}({\langle\rangle}{\exists}gNt(s(g))\backslash Sf), [{\blacksquare}Nt(s(f))], {\square}({\langle\rangle}{\exists}gNt(s(g))\backslash Sf)\ \Rightarrow\ \fbox{$?{\blacksquare}Sf$}
\using {?}E
\endprooftree
\prooftree
\prooftree
\prooftree
\justifies
\mbox{\fbox{$Sf$}}\ \Rightarrow\ Sf
\endprooftree
\justifies
[\mbox{\fbox{${[]^{-1}}Sf$}}]\ \Rightarrow\ Sf
\using {[]^{-1}}L
\endprooftree
\justifies
[[\mbox{\fbox{${[]^{-1}}{[]^{-1}}Sf$}}]]\ \Rightarrow\ Sf
\using {[]^{-1}}L
\endprooftree
\justifies
[[[{\blacksquare}Nt(s(m))], {\square}({\langle\rangle}{\exists}gNt(s(g))\backslash Sf), [{\blacksquare}Nt(s(f))], {\square}({\langle\rangle}{\exists}gNt(s(g))\backslash Sf), \mbox{\fbox{$?{\blacksquare}Sf\backslash {[]^{-1}}{[]^{-1}}Sf$}}]]\ \Rightarrow\ Sf
\using {\backslash}L
\endprooftree
\justifies
[[[{\blacksquare}Nt(s(m))], {\square}({\langle\rangle}{\exists}gNt(s(g))\backslash Sf), [{\blacksquare}Nt(s(f))], {\square}({\langle\rangle}{\exists}gNt(s(g))\backslash Sf), \mbox{\fbox{$(?{\blacksquare}Sf\backslash {[]^{-1}}{[]^{-1}}Sf)/{\blacksquare}Sf$}}, [{\blacksquare}Nt(s(m))], {\square}({\langle\rangle}{\exists}gNt(s(g))\backslash Sf)]]\ \Rightarrow\ Sf
\using {/}L
\endprooftree
\justifies
[[[{\blacksquare}Nt(s(m))], {\square}({\langle\rangle}{\exists}gNt(s(g))\backslash Sf), [{\blacksquare}Nt(s(f))], {\square}({\langle\rangle}{\exists}gNt(s(g))\backslash Sf), \mbox{\fbox{${\forall}f((?{\blacksquare}Sf\backslash {[]^{-1}}{[]^{-1}}Sf)/{\blacksquare}Sf)$}}, [{\blacksquare}Nt(s(m))], {\square}({\langle\rangle}{\exists}gNt(s(g))\backslash Sf)]]\ \Rightarrow\ Sf
\using {\forall}L
\endprooftree
\justifies
[[[{\blacksquare}Nt(s(m))], {\square}({\langle\rangle}{\exists}gNt(s(g))\backslash Sf), [{\blacksquare}Nt(s(f))], {\square}({\langle\rangle}{\exists}gNt(s(g))\backslash Sf), \mbox{\fbox{${\blacksquare}{\forall}f((?{\blacksquare}Sf\backslash {[]^{-1}}{[]^{-1}}Sf)/{\blacksquare}Sf)$}}, [{\blacksquare}Nt(s(m))], {\square}({\langle\rangle}{\exists}gNt(s(g))\backslash Sf)]]\ \Rightarrow\ Sf
\using {\blacksquare}L
\endprooftree}
\end{center}

\vspace{0.15in}

\noindent
This delivers semantics:
\disp{
$[({\it Pres}\ (\mbox{\v{}}{\it walk}\ {\it j}))\wedge [({\it Pres}\ (\mbox{\v{}}{\it talk}\ {\it m}))\wedge ({\it Pres}\ (\mbox{\v{}}{\it sing}\ {\it b}))]]$}

\subsection{Iterated coordination of addicity one}

There is the example of verb phrase iterated coordination:
\disp{
$[{\bf john}]{+}[[{\bf walks}{+}{\bf talks}{+}{\bf and}{+}{\bf sings}]]: Sf$}
Appropriate lexical insertion yields:
\disp{
$\begin{array}[t]{l}
[{\blacksquare}Nt(s(m)): {\it j}], [[{\square}({\langle\rangle}{\exists}gNt(s(g))\backslash Sf): \mbox{\^{}}\lambda A({\it Pres}\ (\mbox{\v{}}{\it walk}\ {\it A})),\\ {\square}({\langle\rangle}{\exists}gNt(s(g))\backslash Sf): \mbox{\^{}}\lambda B({\it Pres}\ (\mbox{\v{}}{\it talk}\ {\it B})),\\
 {\blacksquare}{\forall}a{\forall}f((?{\blacksquare}({\langle\rangle}Na\backslash Sf)\backslash {[]^{-1}}{[]^{-1}}({\langle\rangle}Na\backslash Sf))/{\blacksquare}({\langle\rangle}Na\backslash Sf)): (\Phinplus\ ({\it s}\ {\it 0})\ {\it and}),\\ {\square}({\langle\rangle}{\exists}gNt(s(g))\backslash Sf): \mbox{\^{}}\lambda C({\it Pres}\ (\mbox{\v{}}{\it sing}\ {\it C}))]]\ \Rightarrow\ Sf
 \end{array}$}
The coordinator lexical semantics $(\Phinplus\ (s\ 0)\ {\it and})$ is such that:
\disp{
$\begin{array}[t]{l}
((((\Phinplus\ (s\ 0)\ {\it and})\ x)\ [y, z])\ w) =\\
(((\Phinplus\ 0\ {\it and})\ (x\ w))\ (\alphaplus\ [y, z]\ w)) =\\
(((\Phinplus\ 0\ {\it and})\ (x\ w))\ [(y\ w)|(\alphaplus\ [z]\ w)]) =\\
(((\Phinplus\ 0\ {\it and})\ (x\ w))\ [(y\ w), (z\ w)]) =\\
{}[(y\ w)\wedge(((\Phinplus\ 0\ {\it and})\ (x\ w))\ [(z\ w)])] =\\
{}[(y\ w)\wedge[(z\ w)\wedge(x\ w)]]
\end{array}$}
There is the derivation:

\vspace{0.15in}

\begin{center}
\rotatebox{-90}{\tiny
\prooftree
\prooftree
\prooftree
\prooftree
\prooftree
\prooftree
\prooftree
\prooftree
\prooftree
\prooftree
\prooftree
\prooftree
\justifies
Nt(s(m))\ \Rightarrow\ Nt(s(m))
\endprooftree
\justifies
Nt(s(m))\ \Rightarrow\ \fbox{${\exists}gNt(s(g))$}
\using {\exists}R
\endprooftree
\justifies
[Nt(s(m))]\ \Rightarrow\ \fbox{${\langle\rangle}{\exists}gNt(s(g))$}
\using {\langle\rangle}R
\endprooftree
\prooftree
\justifies
\mbox{\fbox{$Sf$}}\ \Rightarrow\ Sf
\endprooftree
\justifies
[Nt(s(m))], \mbox{\fbox{${\langle\rangle}{\exists}gNt(s(g))\backslash Sf$}}\ \Rightarrow\ Sf
\using {\backslash}L
\endprooftree
\justifies
[Nt(s(m))], \mbox{\fbox{${\square}({\langle\rangle}{\exists}gNt(s(g))\backslash Sf)$}}\ \Rightarrow\ Sf
\using {\Box}L
\endprooftree
\justifies
{\langle\rangle}Nt(s(m)), {\square}({\langle\rangle}{\exists}gNt(s(g))\backslash Sf)\ \Rightarrow\ Sf
\using {\langle\rangle}L
\endprooftree
\justifies
{\square}({\langle\rangle}{\exists}gNt(s(g))\backslash Sf)\ \Rightarrow\ {\langle\rangle}Nt(s(m))\backslash Sf
\using {\backslash}R
\endprooftree
\justifies
{\square}({\langle\rangle}{\exists}gNt(s(g))\backslash Sf)\ \Rightarrow\ {\blacksquare}({\langle\rangle}Nt(s(m))\backslash Sf)
\using {\blacksquare}R
\endprooftree
\prooftree
\prooftree
\prooftree
\prooftree
\prooftree
\prooftree
\prooftree
\prooftree
\prooftree
\prooftree
\justifies
Nt(s(m))\ \Rightarrow\ Nt(s(m))
\endprooftree
\justifies
Nt(s(m))\ \Rightarrow\ \fbox{${\exists}gNt(s(g))$}
\using {\exists}R
\endprooftree
\justifies
[Nt(s(m))]\ \Rightarrow\ \fbox{${\langle\rangle}{\exists}gNt(s(g))$}
\using {\langle\rangle}R
\endprooftree
\prooftree
\justifies
\mbox{\fbox{$Sf$}}\ \Rightarrow\ Sf
\endprooftree
\justifies
[Nt(s(m))], \mbox{\fbox{${\langle\rangle}{\exists}gNt(s(g))\backslash Sf$}}\ \Rightarrow\ Sf
\using {\backslash}L
\endprooftree
\justifies
[Nt(s(m))], \mbox{\fbox{${\square}({\langle\rangle}{\exists}gNt(s(g))\backslash Sf)$}}\ \Rightarrow\ Sf
\using {\Box}L
\endprooftree
\justifies
{\langle\rangle}Nt(s(m)), {\square}({\langle\rangle}{\exists}gNt(s(g))\backslash Sf)\ \Rightarrow\ Sf
\using {\langle\rangle}L
\endprooftree
\justifies
{\square}({\langle\rangle}{\exists}gNt(s(g))\backslash Sf)\ \Rightarrow\ {\langle\rangle}Nt(s(m))\backslash Sf
\using {\backslash}R
\endprooftree
\justifies
{\square}({\langle\rangle}{\exists}gNt(s(g))\backslash Sf)\ \Rightarrow\ {\blacksquare}({\langle\rangle}Nt(s(m))\backslash Sf)
\using {\blacksquare}R
\endprooftree
\prooftree
\prooftree
\prooftree
\prooftree
\prooftree
\prooftree
\prooftree
\prooftree
\prooftree
\justifies
Nt(s(m))\ \Rightarrow\ Nt(s(m))
\endprooftree
\justifies
Nt(s(m))\ \Rightarrow\ \fbox{${\exists}gNt(s(g))$}
\using {\exists}R
\endprooftree
\justifies
[Nt(s(m))]\ \Rightarrow\ \fbox{${\langle\rangle}{\exists}gNt(s(g))$}
\using {\langle\rangle}R
\endprooftree
\prooftree
\justifies
\mbox{\fbox{$Sf$}}\ \Rightarrow\ Sf
\endprooftree
\justifies
[Nt(s(m))], \mbox{\fbox{${\langle\rangle}{\exists}gNt(s(g))\backslash Sf$}}\ \Rightarrow\ Sf
\using {\backslash}L
\endprooftree
\justifies
[Nt(s(m))], \mbox{\fbox{${\square}({\langle\rangle}{\exists}gNt(s(g))\backslash Sf)$}}\ \Rightarrow\ Sf
\using {\Box}L
\endprooftree
\justifies
{\langle\rangle}Nt(s(m)), {\square}({\langle\rangle}{\exists}gNt(s(g))\backslash Sf)\ \Rightarrow\ Sf
\using {\langle\rangle}L
\endprooftree
\justifies
{\square}({\langle\rangle}{\exists}gNt(s(g))\backslash Sf)\ \Rightarrow\ {\langle\rangle}Nt(s(m))\backslash Sf
\using {\backslash}R
\endprooftree
\justifies
{\square}({\langle\rangle}{\exists}gNt(s(g))\backslash Sf)\ \Rightarrow\ {\blacksquare}({\langle\rangle}Nt(s(m))\backslash Sf)
\using {\blacksquare}R
\endprooftree
\justifies
{\square}({\langle\rangle}{\exists}gNt(s(g))\backslash Sf)\ \Rightarrow\ \fbox{$?{\blacksquare}({\langle\rangle}Nt(s(m))\backslash Sf)$}
\using {?}R
\endprooftree
\justifies
{\square}({\langle\rangle}{\exists}gNt(s(g))\backslash Sf), {\square}({\langle\rangle}{\exists}gNt(s(g))\backslash Sf)\ \Rightarrow\ \fbox{$?{\blacksquare}({\langle\rangle}Nt(s(m))\backslash Sf)$}
\using {?}E
\endprooftree
\prooftree
\prooftree
\prooftree
\prooftree
\prooftree
\prooftree
\justifies
\mbox{\fbox{$Nt(s(m))$}}\ \Rightarrow\ Nt(s(m))
\endprooftree
\justifies
\mbox{\fbox{${\blacksquare}Nt(s(m))$}}\ \Rightarrow\ Nt(s(m))
\using {\blacksquare}L
\endprooftree
\justifies
[{\blacksquare}Nt(s(m))]\ \Rightarrow\ \fbox{${\langle\rangle}Nt(s(m))$}
\using {\langle\rangle}R
\endprooftree
\prooftree
\justifies
\mbox{\fbox{$Sf$}}\ \Rightarrow\ Sf
\endprooftree
\justifies
[{\blacksquare}Nt(s(m))], \mbox{\fbox{${\langle\rangle}Nt(s(m))\backslash Sf$}}\ \Rightarrow\ Sf
\using {\backslash}L
\endprooftree
\justifies
[{\blacksquare}Nt(s(m))], [\mbox{\fbox{${[]^{-1}}({\langle\rangle}Nt(s(m))\backslash Sf)$}}]\ \Rightarrow\ Sf
\using {[]^{-1}}L
\endprooftree
\justifies
[{\blacksquare}Nt(s(m))], [[\mbox{\fbox{${[]^{-1}}{[]^{-1}}({\langle\rangle}Nt(s(m))\backslash Sf)$}}]]\ \Rightarrow\ Sf
\using {[]^{-1}}L
\endprooftree
\justifies
[{\blacksquare}Nt(s(m))], [[{\square}({\langle\rangle}{\exists}gNt(s(g))\backslash Sf), {\square}({\langle\rangle}{\exists}gNt(s(g))\backslash Sf), \mbox{\fbox{$?{\blacksquare}({\langle\rangle}Nt(s(m))\backslash Sf)\backslash {[]^{-1}}{[]^{-1}}({\langle\rangle}Nt(s(m))\backslash Sf)$}}]]\ \Rightarrow\ Sf
\using {\backslash}L
\endprooftree
\justifies
[{\blacksquare}Nt(s(m))], [[{\square}({\langle\rangle}{\exists}gNt(s(g))\backslash Sf), {\square}({\langle\rangle}{\exists}gNt(s(g))\backslash Sf), \mbox{\fbox{$(?{\blacksquare}({\langle\rangle}Nt(s(m))\backslash Sf)\backslash {[]^{-1}}{[]^{-1}}({\langle\rangle}Nt(s(m))\backslash Sf))/{\blacksquare}({\langle\rangle}Nt(s(m))\backslash Sf)$}}, {\square}({\langle\rangle}{\exists}gNt(s(g))\backslash Sf)]]\ \Rightarrow\ Sf
\using {/}L
\endprooftree
\justifies
[{\blacksquare}Nt(s(m))], [[{\square}({\langle\rangle}{\exists}gNt(s(g))\backslash Sf), {\square}({\langle\rangle}{\exists}gNt(s(g))\backslash Sf), \mbox{\fbox{${\forall}f((?{\blacksquare}({\langle\rangle}Nt(s(m))\backslash Sf)\backslash {[]^{-1}}{[]^{-1}}({\langle\rangle}Nt(s(m))\backslash Sf))/{\blacksquare}({\langle\rangle}Nt(s(m))\backslash Sf))$}}, {\square}({\langle\rangle}{\exists}gNt(s(g))\backslash Sf)]]\ \Rightarrow\ Sf
\using {\forall}L
\endprooftree
\justifies
[{\blacksquare}Nt(s(m))], [[{\square}({\langle\rangle}{\exists}gNt(s(g))\backslash Sf), {\square}({\langle\rangle}{\exists}gNt(s(g))\backslash Sf), \mbox{\fbox{${\forall}a{\forall}f((?{\blacksquare}({\langle\rangle}Na\backslash Sf)\backslash {[]^{-1}}{[]^{-1}}({\langle\rangle}Na\backslash Sf))/{\blacksquare}({\langle\rangle}Na\backslash Sf))$}}, {\square}({\langle\rangle}{\exists}gNt(s(g))\backslash Sf)]]\ \Rightarrow\ Sf
\using {\forall}L
\endprooftree
\justifies
[{\blacksquare}Nt(s(m))], [[{\square}({\langle\rangle}{\exists}gNt(s(g))\backslash Sf), {\square}({\langle\rangle}{\exists}gNt(s(g))\backslash Sf), \mbox{\fbox{${\blacksquare}{\forall}a{\forall}f((?{\blacksquare}({\langle\rangle}Na\backslash Sf)\backslash {[]^{-1}}{[]^{-1}}({\langle\rangle}Na\backslash Sf))/{\blacksquare}({\langle\rangle}Na\backslash Sf))$}}, {\square}({\langle\rangle}{\exists}gNt(s(g))\backslash Sf)]]\ \Rightarrow\ Sf
\using {\blacksquare}L
\endprooftree}
\end{center}

\vspace{0.15in}

\noindent
This delivers semantics:
\disp{
$[({\it Pres}\ (\mbox{\v{}}{\it walk}\ {\it j}))\wedge [({\it Pres}\ (\mbox{\v{}}{\it talk}\ {\it j}))\wedge ({\it Pres}\ (\mbox{\v{}}{\it sing}\ {\it j}))]]$}

\subsection{Iterated coordination of addicity two}

There is the example of transitive verb phrase iterated coordination:
\disp{
$[{\bf john}]{+}[[{\bf praises}{+}{\bf likes}{+}{\bf and}{+}{\bf will}{+}{\bf love}]]{+}{\bf london}: Sf$}
Lexical insertion yields:
\disp{
$\begin{array}[t]{l}
[{\blacksquare}Nt(s(m)): {\it j}], [[{\square}(({\langle\rangle}{\exists}gNt(s(g))\backslash Sf)/{\exists}aNa): \mbox{\^{}}\lambda A\lambda B({\it Pres}\ ((\mbox{\v{}}{\it praise}\ {\it A})\ {\it B})),\\
 {\square}(({\langle\rangle}{\exists}gNt(s(g))\backslash Sf)/{\exists}aNa): \mbox{\^{}}\lambda C\lambda D({\it Pres}\ ((\mbox{\v{}}{\it like}\ {\it C})\ {\it D})), \\{\blacksquare}{\forall}f{\forall}a((?{\blacksquare}(({\langle\rangle}Na\backslash Sf)/{\exists}bNb)\backslash {[]^{-1}}{[]^{-1}}(({\langle\rangle}Na\backslash Sf)/{\exists}bNb))/{\blacksquare}(({\langle\rangle}Na\backslash Sf)/{\exists}bNb)):\\ (\Phinplus\ ({\it s}\ ({\it s}\ {\it 0}))\ {\it and}), {\blacksquare}{\forall}a(({\langle\rangle}Na\backslash Sf)/({\langle\rangle}Na\backslash Sb)): \lambda E\lambda F({\it Fut}\ ({\it E}\ {\it F})),\\ {\square}(({\langle\rangle}{\exists}aNa\backslash Sb)/{\exists}aNa): \mbox{\^{}}\lambda G\lambda H((\mbox{\v{}}{\it love}\ {\it G})\ {\it H})]],
 {\blacksquare}Nt(s(n)): {\it l}\ \Rightarrow\ Sf
 \end{array}$}
The coordination combinator semantics is such that:
\disp{
$\begin{array}[t]{l}
(((((\Phinplus\ (s\ (s\ 0))\ {\it and})\ x)\ [y, z])\ w)\ u) =\\
((((\Phinplus\ (s\ 0)\ {\it and})\ (x\ w))\ (\alphaplus\ [y, z]\ w))\ u) =\\
((((\Phinplus\ (s\ 0)\ {\it and})\ (x\ w))\ [(y\ w), (z\ w)])\ u) =\\
(((\Phinplus\ 0\ {\it and})\ ((x\ w)\ u))\ (\alphaplus\ [(y\ w), (z\ w)]\ u)) =\\
(((\Phinplus\ 0\ {\it and})\ ((x\ w)\ u))\ [((y\ w)\ u), ((z\ w)\ u)]) =\\
{}[((y\ w)\ u)\wedge[((z\ w)\ u)\wedge((x\ w)\ u)]]
\end{array}$}
There is the derivation:

\vspace{0.15in}

\begin{center}
{\scriptsize
\prooftree
\prooftree
\prooftree
\prooftree
\prooftree
\prooftree
\prooftree
\prooftree
\prooftree
\prooftree
\prooftree
\prooftree
\prooftree
\prooftree
\justifies
N1\ \Rightarrow\ N1
\endprooftree
\justifies
N1\ \Rightarrow\ \fbox{${\exists}aNa$}
\using {\exists}R
\endprooftree
\prooftree
\prooftree
\prooftree
\prooftree
\justifies
Nt(s(m))\ \Rightarrow\ Nt(s(m))
\endprooftree
\justifies
Nt(s(m))\ \Rightarrow\ \fbox{${\exists}aNa$}
\using {\exists}R
\endprooftree
\justifies
[Nt(s(m))]\ \Rightarrow\ \fbox{${\langle\rangle}{\exists}aNa$}
\using {\langle\rangle}R
\endprooftree
\prooftree
\justifies
\mbox{\fbox{$Sb$}}\ \Rightarrow\ Sb
\endprooftree
\justifies
[Nt(s(m))], \mbox{\fbox{${\langle\rangle}{\exists}aNa\backslash Sb$}}\ \Rightarrow\ Sb
\using {\backslash}L
\endprooftree
\justifies
[Nt(s(m))], \mbox{\fbox{$({\langle\rangle}{\exists}aNa\backslash Sb)/{\exists}aNa$}}, N1\ \Rightarrow\ Sb
\using {/}L
\endprooftree
\justifies
[Nt(s(m))], \mbox{\fbox{${\square}(({\langle\rangle}{\exists}aNa\backslash Sb)/{\exists}aNa)$}}, N1\ \Rightarrow\ Sb
\using {\Box}L
\endprooftree
\justifies
{\langle\rangle}Nt(s(m)), {\square}(({\langle\rangle}{\exists}aNa\backslash Sb)/{\exists}aNa), N1\ \Rightarrow\ Sb
\using {\langle\rangle}L
\endprooftree
\justifies
{\square}(({\langle\rangle}{\exists}aNa\backslash Sb)/{\exists}aNa), N1\ \Rightarrow\ {\langle\rangle}Nt(s(m))\backslash Sb
\using {\backslash}R
\endprooftree
\prooftree
\prooftree
\prooftree
\justifies
Nt(s(m))\ \Rightarrow\ Nt(s(m))
\endprooftree
\justifies
[Nt(s(m))]\ \Rightarrow\ \fbox{${\langle\rangle}Nt(s(m))$}
\using {\langle\rangle}R
\endprooftree
\prooftree
\justifies
\mbox{\fbox{$Sf$}}\ \Rightarrow\ Sf
\endprooftree
\justifies
[Nt(s(m))], \mbox{\fbox{${\langle\rangle}Nt(s(m))\backslash Sf$}}\ \Rightarrow\ Sf
\using {\backslash}L
\endprooftree
\justifies
[Nt(s(m))], \mbox{\fbox{$({\langle\rangle}Nt(s(m))\backslash Sf)/({\langle\rangle}Nt(s(m))\backslash Sb)$}}, {\square}(({\langle\rangle}{\exists}aNa\backslash Sb)/{\exists}aNa), N1\ \Rightarrow\ Sf
\using {/}L
\endprooftree
\justifies
[Nt(s(m))], \mbox{\fbox{${\forall}a(({\langle\rangle}Na\backslash Sf)/({\langle\rangle}Na\backslash Sb))$}}, {\square}(({\langle\rangle}{\exists}aNa\backslash Sb)/{\exists}aNa), N1\ \Rightarrow\ Sf
\using {\forall}L
\endprooftree
\justifies
[Nt(s(m))], \mbox{\fbox{${\blacksquare}{\forall}a(({\langle\rangle}Na\backslash Sf)/({\langle\rangle}Na\backslash Sb))$}}, {\square}(({\langle\rangle}{\exists}aNa\backslash Sb)/{\exists}aNa), N1\ \Rightarrow\ Sf
\using {\blacksquare}L
\endprooftree
\justifies
[Nt(s(m))], {\blacksquare}{\forall}a(({\langle\rangle}Na\backslash Sf)/({\langle\rangle}Na\backslash Sb)), {\square}(({\langle\rangle}{\exists}aNa\backslash Sb)/{\exists}aNa), {\exists}bNb\ \Rightarrow\ Sf
\using {\exists}L
\endprooftree
\justifies
{\langle\rangle}Nt(s(m)), {\blacksquare}{\forall}a(({\langle\rangle}Na\backslash Sf)/({\langle\rangle}Na\backslash Sb)), {\square}(({\langle\rangle}{\exists}aNa\backslash Sb)/{\exists}aNa), {\exists}bNb\ \Rightarrow\ Sf
\using {\langle\rangle}L
\endprooftree
\justifies
{\blacksquare}{\forall}a(({\langle\rangle}Na\backslash Sf)/({\langle\rangle}Na\backslash Sb)), {\square}(({\langle\rangle}{\exists}aNa\backslash Sb)/{\exists}aNa), {\exists}bNb\ \Rightarrow\ {\langle\rangle}Nt(s(m))\backslash Sf
\using {\backslash}R
\endprooftree
\justifies
{\blacksquare}{\forall}a(({\langle\rangle}Na\backslash Sf)/({\langle\rangle}Na\backslash Sb)), {\square}(({\langle\rangle}{\exists}aNa\backslash Sb)/{\exists}aNa)\ \Rightarrow\ ({\langle\rangle}Nt(s(m))\backslash Sf)/{\exists}bNb
\using {/}R
\endprooftree
\justifies
\begin{array}{c}
{\blacksquare}{\forall}a(({\langle\rangle}Na\backslash Sf)/({\langle\rangle}Na\backslash Sb)), {\square}(({\langle\rangle}{\exists}aNa\backslash Sb)/{\exists}aNa)\ \Rightarrow\ {\blacksquare}(({\langle\rangle}Nt(s(m))\backslash Sf)/{\exists}bNb)\\
\mbox{\footnotesize\textcircled{1}}
\end{array}
\using {\blacksquare}R
\endprooftree}
\end{center}

\begin{center}
\rotatebox{-90}{\scriptsize
\prooftree
\prooftree
\prooftree
\prooftree
\prooftree
\prooftree
\prooftree
\prooftree
\prooftree
\prooftree
\justifies
N1\ \Rightarrow\ N1
\endprooftree
\justifies
N1\ \Rightarrow\ \fbox{${\exists}aNa$}
\using {\exists}R
\endprooftree
\prooftree
\prooftree
\prooftree
\prooftree
\justifies
Nt(s(m))\ \Rightarrow\ Nt(s(m))
\endprooftree
\justifies
Nt(s(m))\ \Rightarrow\ \fbox{${\exists}gNt(s(g))$}
\using {\exists}R
\endprooftree
\justifies
[Nt(s(m))]\ \Rightarrow\ \fbox{${\langle\rangle}{\exists}gNt(s(g))$}
\using {\langle\rangle}R
\endprooftree
\prooftree
\justifies
\mbox{\fbox{$Sf$}}\ \Rightarrow\ Sf
\endprooftree
\justifies
[Nt(s(m))], \mbox{\fbox{${\langle\rangle}{\exists}gNt(s(g))\backslash Sf$}}\ \Rightarrow\ Sf
\using {\backslash}L
\endprooftree
\justifies
[Nt(s(m))], \mbox{\fbox{$({\langle\rangle}{\exists}gNt(s(g))\backslash Sf)/{\exists}aNa$}}, N1\ \Rightarrow\ Sf
\using {/}L
\endprooftree
\justifies
[Nt(s(m))], \mbox{\fbox{${\square}(({\langle\rangle}{\exists}gNt(s(g))\backslash Sf)/{\exists}aNa)$}}, N1\ \Rightarrow\ Sf
\using {\Box}L
\endprooftree
\justifies
[Nt(s(m))], {\square}(({\langle\rangle}{\exists}gNt(s(g))\backslash Sf)/{\exists}aNa), {\exists}bNb\ \Rightarrow\ Sf
\using {\exists}L
\endprooftree
\justifies
{\langle\rangle}Nt(s(m)), {\square}(({\langle\rangle}{\exists}gNt(s(g))\backslash Sf)/{\exists}aNa), {\exists}bNb\ \Rightarrow\ Sf
\using {\langle\rangle}L
\endprooftree
\justifies
{\square}(({\langle\rangle}{\exists}gNt(s(g))\backslash Sf)/{\exists}aNa), {\exists}bNb\ \Rightarrow\ {\langle\rangle}Nt(s(m))\backslash Sf
\using {\backslash}R
\endprooftree
\justifies
{\square}(({\langle\rangle}{\exists}gNt(s(g))\backslash Sf)/{\exists}aNa)\ \Rightarrow\ ({\langle\rangle}Nt(s(m))\backslash Sf)/{\exists}bNb
\using {/}R
\endprooftree
\justifies
{\square}(({\langle\rangle}{\exists}gNt(s(g))\backslash Sf)/{\exists}aNa)\ \Rightarrow\ {\blacksquare}(({\langle\rangle}Nt(s(m))\backslash Sf)/{\exists}bNb)
\using {\blacksquare}R
\endprooftree
\prooftree
\prooftree
\prooftree
\prooftree
\prooftree
\prooftree
\prooftree
\prooftree
\prooftree
\prooftree
\justifies
N2\ \Rightarrow\ N2
\endprooftree
\justifies
N2\ \Rightarrow\ \fbox{${\exists}aNa$}
\using {\exists}R
\endprooftree
\prooftree
\prooftree
\prooftree
\prooftree
\justifies
Nt(s(m))\ \Rightarrow\ Nt(s(m))
\endprooftree
\justifies
Nt(s(m))\ \Rightarrow\ \fbox{${\exists}gNt(s(g))$}
\using {\exists}R
\endprooftree
\justifies
[Nt(s(m))]\ \Rightarrow\ \fbox{${\langle\rangle}{\exists}gNt(s(g))$}
\using {\langle\rangle}R
\endprooftree
\prooftree
\justifies
\mbox{\fbox{$Sf$}}\ \Rightarrow\ Sf
\endprooftree
\justifies
[Nt(s(m))], \mbox{\fbox{${\langle\rangle}{\exists}gNt(s(g))\backslash Sf$}}\ \Rightarrow\ Sf
\using {\backslash}L
\endprooftree
\justifies
[Nt(s(m))], \mbox{\fbox{$({\langle\rangle}{\exists}gNt(s(g))\backslash Sf)/{\exists}aNa$}}, N2\ \Rightarrow\ Sf
\using {/}L
\endprooftree
\justifies
[Nt(s(m))], \mbox{\fbox{${\square}(({\langle\rangle}{\exists}gNt(s(g))\backslash Sf)/{\exists}aNa)$}}, N2\ \Rightarrow\ Sf
\using {\Box}L
\endprooftree
\justifies
[Nt(s(m))], {\square}(({\langle\rangle}{\exists}gNt(s(g))\backslash Sf)/{\exists}aNa), {\exists}bNb\ \Rightarrow\ Sf
\using {\exists}L
\endprooftree
\justifies
{\langle\rangle}Nt(s(m)), {\square}(({\langle\rangle}{\exists}gNt(s(g))\backslash Sf)/{\exists}aNa), {\exists}bNb\ \Rightarrow\ Sf
\using {\langle\rangle}L
\endprooftree
\justifies
{\square}(({\langle\rangle}{\exists}gNt(s(g))\backslash Sf)/{\exists}aNa), {\exists}bNb\ \Rightarrow\ {\langle\rangle}Nt(s(m))\backslash Sf
\using {\backslash}R
\endprooftree
\justifies
{\square}(({\langle\rangle}{\exists}gNt(s(g))\backslash Sf)/{\exists}aNa)\ \Rightarrow\ ({\langle\rangle}Nt(s(m))\backslash Sf)/{\exists}bNb
\using {/}R
\endprooftree
\justifies
{\square}(({\langle\rangle}{\exists}gNt(s(g))\backslash Sf)/{\exists}aNa)\ \Rightarrow\ {\blacksquare}(({\langle\rangle}Nt(s(m))\backslash Sf)/{\exists}bNb)
\using {\blacksquare}R
\endprooftree
\justifies
{\square}(({\langle\rangle}{\exists}gNt(s(g))\backslash Sf)/{\exists}aNa)\ \Rightarrow\ \fbox{$?{\blacksquare}(({\langle\rangle}Nt(s(m))\backslash Sf)/{\exists}bNb)$}
\using {?}R
\endprooftree
\justifies
\begin{array}{c}
{\square}(({\langle\rangle}{\exists}gNt(s(g))\backslash Sf)/{\exists}aNa), {\square}(({\langle\rangle}{\exists}gNt(s(g))\backslash Sf)/{\exists}aNa)\ \Rightarrow\ \fbox{$?{\blacksquare}(({\langle\rangle}Nt(s(m))\backslash Sf)/{\exists}bNb)$}\\
\mbox{\footnotesize\textcircled{2}}
\end{array}
\using {?}E
\endprooftree}
\end{center}

\begin{center}
\rotatebox{-90}{\tiny
\prooftree
\prooftree
\prooftree
\prooftree
\mbox{\footnotesize\textcircled{1}}\tab
\prooftree
\mbox{\footnotesize\textcircled{2}}\tab
\prooftree
\prooftree
\prooftree
\prooftree
\prooftree
\prooftree
\justifies
\mbox{\fbox{$Nt(s(n))$}}\ \Rightarrow\ Nt(s(n))
\endprooftree
\justifies
\mbox{\fbox{${\blacksquare}Nt(s(n))$}}\ \Rightarrow\ Nt(s(n))
\using {\blacksquare}L
\endprooftree
\justifies
{\blacksquare}Nt(s(n))\ \Rightarrow\ \fbox{${\exists}bNb$}
\using {\exists}R
\endprooftree
\prooftree
\prooftree
\prooftree
\prooftree
\justifies
\mbox{\fbox{$Nt(s(m))$}}\ \Rightarrow\ Nt(s(m))
\endprooftree
\justifies
\mbox{\fbox{${\blacksquare}Nt(s(m))$}}\ \Rightarrow\ Nt(s(m))
\using {\blacksquare}L
\endprooftree
\justifies
[{\blacksquare}Nt(s(m))]\ \Rightarrow\ \fbox{${\langle\rangle}Nt(s(m))$}
\using {\langle\rangle}R
\endprooftree
\prooftree
\justifies
\mbox{\fbox{$Sf$}}\ \Rightarrow\ Sf
\endprooftree
\justifies
[{\blacksquare}Nt(s(m))], \mbox{\fbox{${\langle\rangle}Nt(s(m))\backslash Sf$}}\ \Rightarrow\ Sf
\using {\backslash}L
\endprooftree
\justifies
[{\blacksquare}Nt(s(m))], \mbox{\fbox{$({\langle\rangle}Nt(s(m))\backslash Sf)/{\exists}bNb$}}, {\blacksquare}Nt(s(n))\ \Rightarrow\ Sf
\using {/}L
\endprooftree
\justifies
[{\blacksquare}Nt(s(m))], [\mbox{\fbox{${[]^{-1}}(({\langle\rangle}Nt(s(m))\backslash Sf)/{\exists}bNb)$}}], {\blacksquare}Nt(s(n))\ \Rightarrow\ Sf
\using {[]^{-1}}L
\endprooftree
\justifies
[{\blacksquare}Nt(s(m))], [[\mbox{\fbox{${[]^{-1}}{[]^{-1}}(({\langle\rangle}Nt(s(m))\backslash Sf)/{\exists}bNb)$}}]], {\blacksquare}Nt(s(n))\ \Rightarrow\ Sf
\using {[]^{-1}}L
\endprooftree
\justifies
[{\blacksquare}Nt(s(m))], [[{\square}(({\langle\rangle}{\exists}gNt(s(g))\backslash Sf)/{\exists}aNa), {\square}(({\langle\rangle}{\exists}gNt(s(g))\backslash Sf)/{\exists}aNa), \mbox{\fbox{$?{\blacksquare}(({\langle\rangle}Nt(s(m))\backslash Sf)/{\exists}bNb)\backslash {[]^{-1}}{[]^{-1}}(({\langle\rangle}Nt(s(m))\backslash Sf)/{\exists}bNb)$}}]], {\blacksquare}Nt(s(n))\ \Rightarrow\ Sf
\using {\backslash}L
\endprooftree
\justifies
[{\blacksquare}Nt(s(m))], [[{\square}(({\langle\rangle}{\exists}gNt(s(g))\backslash Sf)/{\exists}aNa), {\square}(({\langle\rangle}{\exists}gNt(s(g))\backslash Sf)/{\exists}aNa), \mbox{\fbox{$(?{\blacksquare}(({\langle\rangle}Nt(s(m))\backslash Sf)/{\exists}bNb)\backslash {[]^{-1}}{[]^{-1}}(({\langle\rangle}Nt(s(m))\backslash Sf)/{\exists}bNb))/{\blacksquare}(({\langle\rangle}Nt(s(m))\backslash Sf)/{\exists}bNb)$}}, {\blacksquare}{\forall}a(({\langle\rangle}Na\backslash Sf)/({\langle\rangle}Na\backslash Sb)), {\square}(({\langle\rangle}{\exists}aNa\backslash Sb)/{\exists}aNa)]], {\blacksquare}Nt(s(n))\ \Rightarrow\ Sf
\using {/}L
\endprooftree
\justifies
[{\blacksquare}Nt(s(m))], [[{\square}(({\langle\rangle}{\exists}gNt(s(g))\backslash Sf)/{\exists}aNa), {\square}(({\langle\rangle}{\exists}gNt(s(g))\backslash Sf)/{\exists}aNa), \mbox{\fbox{${\forall}a((?{\blacksquare}(({\langle\rangle}Na\backslash Sf)/{\exists}bNb)\backslash {[]^{-1}}{[]^{-1}}(({\langle\rangle}Na\backslash Sf)/{\exists}bNb))/{\blacksquare}(({\langle\rangle}Na\backslash Sf)/{\exists}bNb))$}}, {\blacksquare}{\forall}a(({\langle\rangle}Na\backslash Sf)/({\langle\rangle}Na\backslash Sb)), {\square}(({\langle\rangle}{\exists}aNa\backslash Sb)/{\exists}aNa)]], {\blacksquare}Nt(s(n))\ \Rightarrow\ Sf
\using {\forall}L
\endprooftree
\justifies
[{\blacksquare}Nt(s(m))], [[{\square}(({\langle\rangle}{\exists}gNt(s(g))\backslash Sf)/{\exists}aNa), {\square}(({\langle\rangle}{\exists}gNt(s(g))\backslash Sf)/{\exists}aNa), \mbox{\fbox{${\forall}f{\forall}a((?{\blacksquare}(({\langle\rangle}Na\backslash Sf)/{\exists}bNb)\backslash {[]^{-1}}{[]^{-1}}(({\langle\rangle}Na\backslash Sf)/{\exists}bNb))/{\blacksquare}(({\langle\rangle}Na\backslash Sf)/{\exists}bNb))$}}, {\blacksquare}{\forall}a(({\langle\rangle}Na\backslash Sf)/({\langle\rangle}Na\backslash Sb)), {\square}(({\langle\rangle}{\exists}aNa\backslash Sb)/{\exists}aNa)]], {\blacksquare}Nt(s(n))\ \Rightarrow\ Sf
\using {\forall}L
\endprooftree
\justifies
[{\blacksquare}Nt(s(m))], [[{\square}(({\langle\rangle}{\exists}gNt(s(g))\backslash Sf)/{\exists}aNa), {\square}(({\langle\rangle}{\exists}gNt(s(g))\backslash Sf)/{\exists}aNa), \mbox{\fbox{${\blacksquare}{\forall}f{\forall}a((?{\blacksquare}(({\langle\rangle}Na\backslash Sf)/{\exists}bNb)\backslash {[]^{-1}}{[]^{-1}}(({\langle\rangle}Na\backslash Sf)/{\exists}bNb))/{\blacksquare}(({\langle\rangle}Na\backslash Sf)/{\exists}bNb))$}}, {\blacksquare}{\forall}a(({\langle\rangle}Na\backslash Sf)/({\langle\rangle}Na\backslash Sb)), {\square}(({\langle\rangle}{\exists}aNa\backslash Sb)/{\exists}aNa)]], {\blacksquare}Nt(s(n))\ \Rightarrow\ Sf
\using {\blacksquare}L
\endprooftree}
\end{center}

\vspace{0.15in}

\noindent
All this assigns the correct semantics:
\disp{
$[({\it Pres}\ ((\mbox{\v{}}{\it praise}\ {\it l})\ {\it j}))\wedge [({\it Pres}\ ((\mbox{\v{}}{\it like}\ {\it l})\ {\it j}))\wedge ({\it Fut}\ ((\mbox{\v{}}{\it love}\ {\it l})\ {\it j}))]]$}

\section{Coordination of `unlike' types}

\label{unlikesect}

In the following we have coordinate unlike types with nominal and
adjectival complementation of \emph{is}.
\disp{
$[{\bf bond}]{+}{\bf is}{+}[[{\bf 007}{+}{\bf and}{+}{\bf teetotal}]]: Sf$}
Together with a suitable coordinator type, a polymorphic assignment to the copula
of the form $(N\bsl S)/(N\adisj(\CN/\CN))$ predicts such coordination under a \emph{like\/}
type scheme (Morrill 1990\cite{morrill:galt}; Johnson and Bayer 1995\cite{johnsonbayer:95};
Bayer 1996\cite{bayer:96}).
Lexical lookup of types yields the following, where the coordinator is of the form
$(X\bsl \abrack\abrack X)/X$ where $X=(N\bsl S)/(N\adisj((\CN/\CN)\iadisj(\CN\bsl\CN)))$.
\disp{
$\begin{array}[t]{l}
[{\blacksquare}Nt(s(m)): {\it b}], {\blacksquare}(({\langle\rangle}{\exists}gNt(s(g))\backslash Sf)/({\exists}aNa{\oplus}({\exists}g(({\it CN}{\it g}/{\it CN}{\it g}){\sqcup}({\it CN}{\it g}\backslash {\it CN}{\it g})){-}I))):\\ \lambda A\lambda B({\it Pres}\ ({\it A}\casearrow C.[{\it B}={\it C}]; D.(({\it D}\ \lambda E[{\it E}={\it B}])\ {\it B}))), [[{\blacksquare}{\forall}gNt(s(g)): {\it 007},\\ {\blacksquare}{\forall}f{\forall}a(({\blacksquare}((({\langle\rangle}Na\backslash Sf)/({\exists}bNb{\oplus}{\exists}g(({\it CN}{\it g}/{\it CN}{\it g}){\sqcup}({\it CN}{\it g}\backslash {\it CN}{\it g}))))\backslash ({\langle\rangle}Na\backslash Sf))\backslash\\ {[]^{-1}}{[]^{-1}}((({\langle\rangle}Na\backslash Sf)/({\exists}bNb{\oplus}{\exists}g(({\it CN}{\it g}/{\it CN}{\it g}){\sqcup}({\it CN}{\it g}\backslash {\it CN}{\it g}))))\backslash ({\langle\rangle}Na\backslash Sf)))/\\{\blacksquare}((({\langle\rangle}Na\backslash Sf)/({\exists}bNb{\oplus}{\exists}g(({\it CN}{\it g}/{\it CN}{\it g}){\sqcup}({\it CN}{\it g}\backslash {\it CN}{\it g}))))\backslash ({\langle\rangle}Na\backslash Sf))): \\\lambda F\lambda G\lambda H\lambda I[(({\it G}\ {\it H})\ {\it I})\wedge (({\it F}\ {\it H})\ {\it I})], {\square}{\forall}n({\it CN}{\it n}/{\it CN}{\it n}): \mbox{\^{}}\lambda J\lambda K[({\it J}\ {\it K})\wedge (\mbox{\v{}}{\it teetotal}\ {\it K})]]]\\ \Rightarrow~Sf
\end{array}$}
The derivation is as shown below.

\begin{center}\scriptsize
\prooftree
\prooftree
\prooftree
\prooftree
\prooftree
\prooftree
\prooftree
\prooftree
\prooftree
\prooftree
\prooftree
\prooftree
\justifies
{\it CN}{\it A}\ \Rightarrow\ {\it CN}{\it A}
\endprooftree
\prooftree
\justifies
\mbox{\fbox{${\it CN}{\it A}$}}\ \Rightarrow\ {\it CN}{\it A}
\endprooftree
\justifies
\mbox{\fbox{${\it CN}{\it A}/{\it CN}{\it A}$}}, {\it CN}{\it A}\ \Rightarrow\ {\it CN}{\it A}
\using {/}L
\endprooftree
\justifies
\mbox{\fbox{${\forall}n({\it CN}{\it n}/{\it CN}{\it n})$}}, {\it CN}{\it A}\ \Rightarrow\ {\it CN}{\it A}
\using {\forall}L
\endprooftree
\justifies
\mbox{\fbox{${\square}{\forall}n({\it CN}{\it n}/{\it CN}{\it n})$}}, {\it CN}{\it A}\ \Rightarrow\ {\it CN}{\it A}
\using {\Box}L
\endprooftree
\justifies
{\square}{\forall}n({\it CN}{\it n}/{\it CN}{\it n})\ \Rightarrow\ \fbox{${\it CN}{\it A}/{\it CN}{\it A}$}
\using {\sqcup}R
\endprooftree
\justifies
{\square}{\forall}n({\it CN}{\it n}/{\it CN}{\it n})\ \Rightarrow\ \fbox{${\exists}g({\it CN}{\it g}/{\it CN}{\it g})$}
\using {\exists}R
\endprooftree
\justifies
{\square}{\forall}n({\it CN}{\it n}/{\it CN}{\it n})\ \Rightarrow\ \fbox{${\exists}bNb{\oplus}{\exists}g({\it CN}{\it g}/{\it CN}{\it g})$}
\using {\oplus}R
\endprooftree
\prooftree
\prooftree
\prooftree
\justifies
Nt(s(m))\ \Rightarrow\ Nt(s(m))
\endprooftree
\justifies
[Nt(s(m))]\ \Rightarrow\ \fbox{${\langle\rangle}Nt(s(m))$}
\using {\langle\rangle}R
\endprooftree
\prooftree
\justifies
\mbox{\fbox{$Sf$}}\ \Rightarrow\ Sf
\endprooftree
\justifies
[Nt(s(m))], \mbox{\fbox{${\langle\rangle}Nt(s(m))\backslash Sf$}}\ \Rightarrow\ Sf
\using {\backslash}L
\endprooftree
\justifies
[Nt(s(m))], \mbox{\fbox{$({\langle\rangle}Nt(s(m))\backslash Sf)/({\exists}bNb{\oplus}{\exists}g({\it CN}{\it g}/{\it CN}{\it g}))$}}, {\square}{\forall}n({\it CN}{\it n}/{\it CN}{\it n})\ \Rightarrow\ Sf
\using {/}L
\endprooftree
\justifies
{\langle\rangle}Nt(s(m)), ({\langle\rangle}Nt(s(m))\backslash Sf)/({\exists}bNb{\oplus}{\exists}g({\it CN}{\it g}/{\it CN}{\it g})), {\square}{\forall}n({\it CN}{\it n}/{\it CN}{\it n})\ \Rightarrow\ Sf
\using {\langle\rangle}L
\endprooftree
\justifies
({\langle\rangle}Nt(s(m))\backslash Sf)/({\exists}bNb{\oplus}{\exists}g({\it CN}{\it g}/{\it CN}{\it g})), {\square}{\forall}n({\it CN}{\it n}/{\it CN}{\it n})\ \Rightarrow\ {\langle\rangle}Nt(s(m))\backslash Sf
\using {\backslash}R
\endprooftree
\justifies
{\square}{\forall}n({\it CN}{\it n}/{\it CN}{\it n})\ \Rightarrow\ (({\langle\rangle}Nt(s(m))\backslash Sf)/({\exists}bNb{\oplus}{\exists}g({\it CN}{\it g}/{\it CN}{\it g})))\backslash ({\langle\rangle}Nt(s(m))\backslash Sf)
\using {\backslash}R
\endprooftree
\justifies
\begin{array}{c}
{\square}{\forall}n({\it CN}{\it n}/{\it CN}{\it n})\ \Rightarrow\ {\blacksquare}((({\langle\rangle}Nt(s(m))\backslash Sf)/({\exists}bNb{\oplus}{\exists}g({\it CN}{\it g}/{\it CN}{\it g})))\backslash ({\langle\rangle}Nt(s(m))\backslash Sf))\\
\mbox{\footnotesize\textcircled{1}}
\end{array}
\using {\blacksquare}R
\endprooftree

\prooftree
\prooftree
\prooftree
\prooftree
\prooftree
\prooftree
\prooftree
\prooftree
\prooftree
\prooftree
\justifies
\mbox{\fbox{$Nt(s(A))$}}\ \Rightarrow\ Nt(s(A))
\endprooftree
\justifies
\mbox{\fbox{${\forall}gNt(s(g))$}}\ \Rightarrow\ Nt(s(A))
\using {\forall}L
\endprooftree
\justifies
\mbox{\fbox{${\blacksquare}{\forall}gNt(s(g))$}}\ \Rightarrow\ Nt(s(A))
\using {\blacksquare}L
\endprooftree
\justifies
{\blacksquare}{\forall}gNt(s(g))\ \Rightarrow\ \fbox{${\exists}bNb$}
\using {\exists}R
\endprooftree
\justifies
{\blacksquare}{\forall}gNt(s(g))\ \Rightarrow\ \fbox{${\exists}bNb{\oplus}{\exists}g({\it CN}{\it g}/{\it CN}{\it g})$}
\using {\oplus}R
\endprooftree
\prooftree
\prooftree
\prooftree
\justifies
Nt(s(m))\ \Rightarrow\ Nt(s(m))
\endprooftree
\justifies
[Nt(s(m))]\ \Rightarrow\ \fbox{${\langle\rangle}Nt(s(m))$}
\using {\langle\rangle}R
\endprooftree
\prooftree
\justifies
\mbox{\fbox{$Sf$}}\ \Rightarrow\ Sf
\endprooftree
\justifies
[Nt(s(m))], \mbox{\fbox{${\langle\rangle}Nt(s(m))\backslash Sf$}}\ \Rightarrow\ Sf
\using {\backslash}L
\endprooftree
\justifies
[Nt(s(m))], \mbox{\fbox{$({\langle\rangle}Nt(s(m))\backslash Sf)/({\exists}bNb{\oplus}{\exists}g({\it CN}{\it g}/{\it CN}{\it g}))$}}, {\blacksquare}{\forall}gNt(s(g))\ \Rightarrow\ Sf
\using {/}L
\endprooftree
\justifies
{\langle\rangle}Nt(s(m)), ({\langle\rangle}Nt(s(m))\backslash Sf)/({\exists}bNb{\oplus}{\exists}g({\it CN}{\it g}/{\it CN}{\it g})), {\blacksquare}{\forall}gNt(s(g))\ \Rightarrow\ Sf
\using {\langle\rangle}L
\endprooftree
\justifies
({\langle\rangle}Nt(s(m))\backslash Sf)/({\exists}bNb{\oplus}{\exists}g({\it CN}{\it g}/{\it CN}{\it g})), {\blacksquare}{\forall}gNt(s(g))\ \Rightarrow\ {\langle\rangle}Nt(s(m))\backslash Sf
\using {\backslash}R
\endprooftree
\justifies
{\blacksquare}{\forall}gNt(s(g))\ \Rightarrow\ (({\langle\rangle}Nt(s(m))\backslash Sf)/({\exists}bNb{\oplus}{\exists}g({\it CN}{\it g}/{\it CN}{\it g})))\backslash ({\langle\rangle}Nt(s(m))\backslash Sf)
\using {\backslash}R
\endprooftree
\justifies
\begin{array}{c}
{\blacksquare}{\forall}gNt(s(g))\ \Rightarrow\ {\blacksquare}((({\langle\rangle}Nt(s(m))\backslash Sf)/({\exists}bNb{\oplus}{\exists}g({\it CN}{\it g}/{\it CN}{\it g})))\backslash ({\langle\rangle}Nt(s(m))\backslash Sf))\\
\mbox{\footnotesize\textcircled{2}}
\end{array}
\using {\blacksquare}R
\endprooftree

\prooftree
\prooftree
\prooftree
\prooftree
\prooftree
\prooftree
\prooftree
\justifies
N1\ \Rightarrow\ N1
\endprooftree
\justifies
N1\ \Rightarrow\ \fbox{${\exists}aNa$}
\using {\exists}R
\endprooftree
\justifies
N1\ \Rightarrow\ \fbox{${\exists}aNa{\oplus}{\exists}g({\it CN}{\it g}/{\it CN}{\it g})$}
\using {\oplus}R
\endprooftree
\prooftree
\prooftree
\prooftree
\prooftree
\justifies
Nt(s(m))\ \Rightarrow\ Nt(s(m))
\endprooftree
\justifies
Nt(s(m))\ \Rightarrow\ \fbox{${\exists}gNt(s(g))$}
\using {\exists}R
\endprooftree
\justifies
[Nt(s(m))]\ \Rightarrow\ \fbox{${\langle\rangle}{\exists}gNt(s(g))$}
\using {\langle\rangle}R
\endprooftree
\prooftree
\justifies
\mbox{\fbox{$Sf$}}\ \Rightarrow\ Sf
\endprooftree
\justifies
[Nt(s(m))], \mbox{\fbox{${\langle\rangle}{\exists}gNt(s(g))\backslash Sf$}}\ \Rightarrow\ Sf
\using {\backslash}L
\endprooftree
\justifies
[Nt(s(m))], \mbox{\fbox{$({\langle\rangle}{\exists}gNt(s(g))\backslash Sf)/({\exists}aNa{\oplus}{\exists}g({\it CN}{\it g}/{\it CN}{\it g})$}}, N1\ \Rightarrow\ Sf
\using {/}L
\endprooftree
\justifies
[Nt(s(m))], \mbox{\fbox{${\blacksquare}(({\langle\rangle}{\exists}gNt(s(g))\backslash Sf)/({\exists}aNa{\oplus}{\exists}g({\it CN}{\it g}/{\it CN}{\it g})))$}}, N1\ \Rightarrow\ Sf
\using {\blacksquare}L
\endprooftree
\justifies
[Nt(s(m))], {\blacksquare}(({\langle\rangle}{\exists}gNt(s(g))\backslash Sf)/({\exists}aNa{\oplus}{\exists}g({\it CN}{\it g}/{\it CN}{\it g}))), {\exists}bNb\ \Rightarrow\ Sf
\using {\exists}L
\endprooftree
\justifies
\begin{array}{c}
{\langle\rangle}Nt(s(m)), {\blacksquare}(({\langle\rangle}{\exists}gNt(s(g))\backslash Sf)/({\exists}aNa{\oplus}{\exists}g({\it CN}{\it g}/{\it CN}{\it g}))), {\exists}bNb\ \Rightarrow\ Sf\\
\mbox{\footnotesize\textcircled{3}}
\end{array}
\using {\langle\rangle}L
\endprooftree

\rotatebox{-0}{\tiny
\prooftree
\prooftree
\prooftree
\prooftree
\prooftree
\prooftree
\prooftree
\vdots
\justifies
{\it CN}{\it 5581}/{\it CN}{\it 5581}\ \Rightarrow\ \fbox{${\exists}g(({\it CN}{\it g}/{\it CN}{\it g}){\sqcup}({\it CN}{\it g}\backslash {\it CN}{\it g}))$}
\using {\exists}R
\endprooftree
\justifies
{\it CN}{\it 5581}/{\it CN}{\it 5581}\ \Rightarrow\ \fbox{${\exists}g(({\it CN}{\it g}/{\it CN}{\it g}){\sqcup}({\it CN}{\it g}\backslash {\it CN}{\it g}))-I$}
\using {-}R
\endprooftree
\justifies
{\it CN}{\it 5581}/{\it CN}{\it 5581}\ \Rightarrow\ \fbox{${\exists}aNa{\oplus}({\exists}g(({\it CN}{\it g}/{\it CN}{\it g}){\sqcup}({\it CN}{\it g}\backslash {\it CN}{\it g})){-}I)$}
\using {\oplus}R
\endprooftree
\prooftree
\prooftree
\prooftree
\prooftree
\justifies
Nt(s(m))\ \Rightarrow\ Nt(s(m))
\endprooftree
\justifies
Nt(s(m))\ \Rightarrow\ \fbox{${\exists}gNt(s(g))$}
\using {\exists}R
\endprooftree
\justifies
[Nt(s(m))]\ \Rightarrow\ \fbox{${\langle\rangle}{\exists}gNt(s(g))$}
\using {\langle\rangle}R
\endprooftree
\prooftree
\justifies
\mbox{\fbox{$Sf$}}\ \Rightarrow\ Sf
\endprooftree
\justifies
[Nt(s(m))], \mbox{\fbox{${\langle\rangle}{\exists}gNt(s(g))\backslash Sf$}}\ \Rightarrow\ Sf
\using {\backslash}L
\endprooftree
\justifies
[Nt(s(m))], \mbox{\fbox{$({\langle\rangle}{\exists}gNt(s(g))\backslash Sf)/({\exists}aNa{\oplus}({\exists}g(({\it CN}{\it g}/{\it CN}{\it g}){\sqcup}({\it CN}{\it g}\backslash {\it CN}{\it g})){-}I))$}}, {\it CN}{\it 5581}/{\it CN}{\it 5581}\ \Rightarrow\ Sf
\using {/}L
\endprooftree
\justifies
[Nt(s(m))], \mbox{\fbox{${\blacksquare}(({\langle\rangle}{\exists}gNt(s(g))\backslash Sf)/({\exists}aNa{\oplus}{\exists}g(({\it CN}{\it g}/{\it CN}{\it g})))$}}, {\it CN}{\it 5581}/{\it CN}{\it 5581}\ \Rightarrow\ Sf
\using {\blacksquare}L
\endprooftree
\justifies
[Nt(s(m))], {\blacksquare}(({\langle\rangle}{\exists}gNt(s(g))\backslash Sf)/({\exists}aNa{\oplus}{\exists}g({\it CN}{\it g}/{\it CN}{\it g}))), {\exists}g({\it CN}{\it g}/{\it CN}{\it g})\ \Rightarrow\ Sf
\using {\exists}L
\endprooftree
\justifies
\begin{array}{c}
{\langle\rangle}Nt(s(m)), {\blacksquare}(({\langle\rangle}{\exists}gNt(s(g))\backslash Sf)/({\exists}aNa{\oplus}{\exists}g({\it CN}{\it g}/{\it CN}{\it g}))), {\exists}g({\it CN}{\it g}/{\it CN}{\it g})\ \Rightarrow\ Sf\\
\mbox{\footnotesize\textcircled{4}}
\end{array}
\using {\langle\rangle}L
\endprooftree}
\end{center}
\begin{center}
\rotatebox{-90}{\tiny
\prooftree
\prooftree
\prooftree
\prooftree
\mbox{\footnotesize\textcircled{1}}\tab\tab
\prooftree
\mbox{\footnotesize\textcircled{2}}\tab
\prooftree
\prooftree
\prooftree
\prooftree
\prooftree
\prooftree
\mbox{\footnotesize\textcircled{3}}\tab\tab\tab
\mbox{\footnotesize\textcircled{4}}
\justifies
{\langle\rangle}Nt(s(m)), {\blacksquare}(({\langle\rangle}{\exists}gNt(s(g))\backslash Sf)/({\exists}aNa{\oplus}{\exists}g({\it CN}{\it g}/{\it CN}{\it g}))), {\exists}bNb{\oplus}{\exists}g({\it CN}{\it g}/{\it CN}{\it g})\ \Rightarrow\ Sf
\using {\oplus}L
\endprooftree
\justifies
{\blacksquare}(({\langle\rangle}{\exists}gNt(s(g))\backslash Sf)/({\exists}aNa{\oplus}{\exists}g({\it CN}{\it g}/{\it CN}{\it g}))), {\exists}bNb{\oplus}{\exists}g({\it CN}{\it g}/{\it CN}{\it g})\ \Rightarrow\ {\langle\rangle}Nt(s(m))\backslash Sf
\using {\backslash}R
\endprooftree
\justifies
{\blacksquare}(({\langle\rangle}{\exists}gNt(s(g))\backslash Sf)/({\exists}aNa{\oplus}{\exists}g({\it CN}{\it g}/{\it CN}{\it g}))\ \Rightarrow\ ({\langle\rangle}Nt(s(m))\backslash Sf)/({\exists}bNb{\oplus}{\exists}g({\it CN}{\it g}/{\it CN}{\it g}))
\using {/}R
\endprooftree
\prooftree
\prooftree
\prooftree
\prooftree
\justifies
\mbox{\fbox{$Nt(s(m))$}}\ \Rightarrow\ Nt(s(m))
\endprooftree
\justifies
\mbox{\fbox{${\blacksquare}Nt(s(m))$}}\ \Rightarrow\ Nt(s(m))
\using {\blacksquare}L
\endprooftree
\justifies
[{\blacksquare}Nt(s(m))]\ \Rightarrow\ \fbox{${\langle\rangle}Nt(s(m))$}
\using {\langle\rangle}R
\endprooftree
\prooftree
\justifies
\mbox{\fbox{$Sf$}}\ \Rightarrow\ Sf
\endprooftree
\justifies
[{\blacksquare}Nt(s(m))], \mbox{\fbox{${\langle\rangle}Nt(s(m))\backslash Sf$}}\ \Rightarrow\ Sf
\using {\backslash}L
\endprooftree
\justifies
[{\blacksquare}Nt(s(m))], {\blacksquare}(({\langle\rangle}{\exists}gNt(s(g))\backslash Sf)/({\exists}aNa{\oplus}{\exists}g({\it CN}{\it g}/{\it CN}{\it g}))), \mbox{\fbox{$(({\langle\rangle}Nt(s(m))\backslash Sf)/({\exists}bNb{\oplus}{\exists}g({\it CN}{\it g}/{\it CN}{\it g})))\backslash ({\langle\rangle}Nt(s(m))\backslash Sf)$}}\ \Rightarrow\ Sf
\using {\backslash}L
\endprooftree
\justifies
[{\blacksquare}Nt(s(m))], {\blacksquare}(({\langle\rangle}{\exists}gNt(s(g))\backslash Sf)/({\exists}aNa{\oplus}{\exists}g({\it CN}{\it g}/{\it CN}{\it g}))), [\mbox{\fbox{${[]^{-1}}((({\langle\rangle}Nt(s(m))\backslash Sf)/({\exists}bNb{\oplus}{\exists}g({\it CN}{\it g}/{\it CN}{\it g}))))\backslash ({\langle\rangle}Nt(s(m))\backslash Sf))$}}]\ \Rightarrow\ Sf
\using {[]^{-1}}L
\endprooftree
\justifies
[{\blacksquare}Nt(s(m))], {\blacksquare}(({\langle\rangle}{\exists}gNt(s(g))\backslash Sf)/({\exists}aNa{\oplus}{\exists}g({\it CN}{\it g}/{\it CN}{\it g}))), [[\mbox{\fbox{${[]^{-1}}{[]^{-1}}((({\langle\rangle}Nt(s(m))\backslash Sf)/({\exists}bNb{\oplus}{\exists}g({\it CN}{\it g}/{\it CN}{\it g}))))\backslash ({\langle\rangle}Nt(s(m))\backslash Sf))$}}]]\ \Rightarrow\ Sf
\using {[]^{-1}}L
\endprooftree
\justifies
\begin{array}{c}
{}[{\blacksquare}Nt(s(m))], {\blacksquare}(({\langle\rangle}{\exists}gNt(s(g))\backslash Sf)/({\exists}aNa{\oplus}{\exists}g({\it CN}{\it g}/{\it CN}{\it g}))), [[{\blacksquare}{\forall}gNt(s(g)),\\
\mbox{\fbox{${\blacksquare}((({\langle\rangle}Nt(s(m))\backslash Sf)/({\exists}bNb{\oplus}{\exists}g({\it CN}{\it g}/{\it CN}{\it g})))\backslash ({\langle\rangle}Nt(s(m))\backslash Sf))\backslash {[]^{-1}}{[]^{-1}}((({\langle\rangle}Nt(s(m))\backslash Sf)/({\exists}bNb{\oplus}{\exists}g({\it CN}{\it g}/{\it CN}{\it g})))\backslash ({\langle\rangle}Nt(s(m))\backslash Sf))$}}]]\ \Rightarrow\ Sf
\end{array}
\using {\backslash}L
\endprooftree
\justifies
\begin{array}{c}
{}[{\blacksquare}Nt(s(m))], {\blacksquare}(({\langle\rangle}{\exists}gNt(s(g))\backslash Sf)/({\exists}aNa{\oplus}{\exists}g({\it CN}{\it g}/{\it CN}{\it g}))), [[{\blacksquare}{\forall}gNt(s(g)),\\
\mbox{\fbox{$({\blacksquare}((({\langle\rangle}Nt(s(m))\backslash Sf)/({\exists}bNb{\oplus}{\exists}g({\it CN}{\it g}/{\it CN}{\it g})))\backslash ({\langle\rangle}Nt(s(m))\backslash Sf))\backslash {[]^{-1}}{[]^{-1}}((({\langle\rangle}Nt(s(m))\backslash Sf)/({\exists}bNb{\oplus}{\exists}g({\it CN}{\it g}/{\it CN}{\it g})))\backslash ({\langle\rangle}Nt(s(m))\backslash Sf)))/{\blacksquare}((({\langle\rangle}Nt(s(m))\backslash Sf)/({\exists}bNb{\oplus}{\exists}g({\it CN}{\it g}/{\it CN}{\it g})))\backslash ({\langle\rangle}Nt(s(m))\backslash Sf))$}},\\
{\square}{\forall}n({\it CN}{\it n}/{\it CN}{\it n})]]\ \Rightarrow\ Sf
\end{array}
\using {/}L
\endprooftree
\justifies
\begin{array}{c}
{}[{\blacksquare}Nt(s(m))], {\blacksquare}(({\langle\rangle}{\exists}gNt(s(g))\backslash Sf)/({\exists}aNa{\oplus}{\exists}g({\it CN}{\it g}/{\it CN}{\it g}))), [[{\blacksquare}{\forall}gNt(s(g)),\\
\mbox{\fbox{${\forall}a(({\blacksquare}((({\langle\rangle}Na\backslash Sf)/({\exists}bNb{\oplus}{\exists}g({\it CN}{\it g}/{\it CN}{\it g})))\backslash ({\langle\rangle}Na\backslash Sf))\backslash {[]^{-1}}{[]^{-1}}((({\langle\rangle}Na\backslash Sf)/({\exists}bNb{\oplus}{\exists}g({\it CN}{\it g}/{\it CN}{\it g})))\backslash ({\langle\rangle}Na\backslash Sf)))/{\blacksquare}((({\langle\rangle}Na\backslash Sf)/({\exists}bNb{\oplus}{\exists}g({\it CN}{\it g}/{\it CN}{\it g})))\backslash ({\langle\rangle}Na\backslash Sf)))$}},\\
{\square}{\forall}n({\it CN}{\it n}/{\it CN}{\it n})]]\ \Rightarrow\ Sf
\end{array}
\using {\forall}L
\endprooftree
\justifies
\begin{array}{c}
{}[{\blacksquare}Nt(s(m))], {\blacksquare}(({\langle\rangle}{\exists}gNt(s(g))\backslash Sf)/({\exists}aNa{\oplus}{\exists}g({\it CN}{\it g}/{\it CN}{\it g}))), [[{\blacksquare}{\forall}gNt(s(g)),\\
\mbox{\fbox{${\forall}f{\forall}a(({\blacksquare}((({\langle\rangle}Na\backslash Sf)/({\exists}bNb{\oplus}{\exists}g({\it CN}{\it g}/{\it CN}{\it g})))\backslash ({\langle\rangle}Na\backslash Sf))\backslash {[]^{-1}}{[]^{-1}}((({\langle\rangle}Na\backslash Sf)/({\exists}bNb{\oplus}{\exists}g({\it CN}{\it g}/{\it CN}{\it g}))))\backslash ({\langle\rangle}Na\backslash Sf)))/{\blacksquare}((({\langle\rangle}Na\backslash Sf)/({\exists}bNb{\oplus}{\exists}g({\it CN}{\it g}/{\it CN}{\it g})))\backslash ({\langle\rangle}Na\backslash Sf)))$}},\\
{\square}{\forall}n({\it CN}{\it n}/{\it CN}{\it n})]]\ \Rightarrow\ Sf
\end{array}
\using {\forall}L
\endprooftree
\justifies
\begin{array}{c}
{}[{\blacksquare}Nt(s(m))], {\blacksquare}(({\langle\rangle}{\exists}gNt(s(g))\backslash Sf)/({\exists}aNa{\oplus}{\exists}g({\it CN}{\it g}/{\it CN}{\it g}))), [[{\blacksquare}{\forall}gNt(s(g)),\\
\mbox{\fbox{${\blacksquare}{\forall}f{\forall}a(({\blacksquare}((({\langle\rangle}Na\backslash Sf)/({\exists}bNb{\oplus}{\exists}g({\it CN}{\it g}/{\it CN}{\it g})))\backslash ({\langle\rangle}Na\backslash Sf))\backslash {[]^{-1}}{[]^{-1}}((({\langle\rangle}Na\backslash Sf)/({\exists}bNb{\oplus}{\exists}g({\it CN}{\it g}/{\it CN}{\it g})))\backslash ({\langle\rangle}Na\backslash Sf)))/{\blacksquare}((({\langle\rangle}Na\backslash Sf)/({\exists}bNb{\oplus}{\exists}g({\it CN}{\it g}/{\it CN}{\it g})))\backslash ({\langle\rangle}Na\backslash Sf)))$}},\\
{\square}{\forall}n({\it CN}{\it n}/{\it CN}{\it n})]]\ \Rightarrow\ Sf
\end{array}
\using {\blacksquare}L
\endprooftree}
\end{center}


\noindent
This yields semantics:
\disp{
$[({\it Pres}\ [{\it b}={\it 007}])\wedge ({\it Pres}\ (\mbox{\v{}}{\it teetotal}\ {\it b}))]$.}
The same account can be given for other verbs, for example our polymorphic
type for \lingform{saw} will allow under a suitable coordinator type
\scare{unlike} coordination such as \lingform{John saw the facts and that}
\lingform{Mary had been right}.

\section{Annex: Gapping}

In a series of papers Kubota and Levine give an account of gapping and determiner
gapping in terms of hybrid type logical grammar, including anomalous
scopal interactions with auxiliaries and negative quantifiers.\footnote{The contents
of this section are to appear in {\it Natural Language and Linguistic Theory}.} We make three
observations:
i) under the counterpart assumptions that Kubota and Levine make, the 
existent displacement type logical grammar account of gapping already accounts
for the scopal interactions, 
ii) Kubota and Levine overgenerate determiner-verb order inconsistencies in
determiner gapping conjuncts whereas the immediate adaptation of their
proposal to displacement type logical grammar does not do so, and
iii) Kubota and Levine do not capture simplex gapping
as a special case of complex gapping, but require
distinct lexical entries for the two cases; we show how a generalisation of displacement type logical grammar
allows both simplex and discontinuous gapping under a single type assignment.

Consider the following four types of gapping:
simple gapping, discontinuous gapping,
determiner gapping, and discontinuous determiner gapping.
\disp{\begin{tabular}[t]{ll}
a. & Leslie met Sandy and Robin (met) Bill.\\
b. & John wants Watford to win and Daniel (wants) Chelsea (to win).
\label{gappingex}
\end{tabular}}

\disp{\begin{tabular}[t]{ll}
a. & Some dogs like Whiskas and (some) cats (like) Alpo.\\
b. & Every cook wants Bar\c{c}a to win and (every) waiter (wants)\\& Madrid (to win).
\label{detgappingex}
\end{tabular}}
Kubota and Levine (2012\cite{htlggapping}; 2013\cite{kublev:detgap}; 2015\cite{htlgnllt:gapping}; henceforth K\&L)
develop an account of gapping in hybrid type logical grammar (HTLG),
an extension of Lambek calculus admitting functional expressions in the phonological component.\footnote{
An anonymous referee questions whether gapping is after all a purely combinatoric phenomenon citing split
antecedent gapping (i), and non-ATB gapping (ii):

\ 

\noindent
i) Sue goes running $6$ times a week, and Alex lifts weights $3$ times a week, but neither every day. \\
ii) Either Pat came with Chris and Sandy came with Kim, or Pat with Kim and the others were alone.

\

We cannot enter fully into this question here except to note that such examples do not show that there is
\emph{no\/} combinatoric component to gapping, but rather that it is a more generalised phenomenon in the case of iterated coordination, which we do not address here. 
}
K\&L (2015\cite{htlgnllt:gapping}) provide a review of the literature
and argue in broad terms the advantages of an analysis of gapping
as hypothetical reasoning,
to which we have nothing to add; we in turn review their type logical proposal.
The Lambek rules of HTLG are as follows:\footnote{We
make some notational adjustments in order to smoothen comparison of hybrid
TLG and displacement TLG.}
\disp{
\prooftree
\prooftree
\leadsto
\alpha\ass C/B\ass \phi
\endprooftree
\prooftree
\leadsto
\beta\ass B\ass \psi
\endprooftree
\justifies
\alpha\add\beta\ass C\ass (\phi\ \psi)
\using
/E
\endprooftree
\tb
\prooftree
\prooftree
\leadsto
\alpha\ass A\ass \phi
\endprooftree
\prooftree
\leadsto
\beta\ass A\bsl C\ass \psi
\endprooftree
\justifies
\alpha\add\beta\ass C\ass (\psi\ \phi)
\using
\bsl E
\endprooftree
\vtab
\prooftree
\prooftree
\prooftree
\justifies
{b}\ass B\ass y
\using n
\endprooftree
\leadsto
\alpha\add{b}\ass C\ass \chi
\endprooftree
\justifies
\alpha\ass C/B\ass \lambda y\chi
\using /I^n
\endprooftree
\tb
\prooftree
\prooftree
\prooftree
\justifies
{a}\ass A\ass x
\using n
\endprooftree
\leadsto
{a}\add\beta\ass C\ass \chi
\endprooftree
\justifies
\beta\ass A\bsl C\ass \lambda x\chi
\using \bsl I^n
\endprooftree
}
In these rules $\alpha\ass A\ass \phi$ signifies a sign with phonology/prosodics $\alpha$,
syntactic type $A$, and semantics $\phi$. The elimination $(E)$ rules combine signs by 
concatenation prosodically, \emph{modus ponens\/} type-logically, and functional application semantically;
the introduction $(I)$ rules of hypothetical reasoning conditionalise assumptions: when passing from premise to
conclusion the assumption coindexed with the rule name $n$ becomes closed. Every rule instance has a fresh index. 
The introduction rules are semantically interpreted
by functional abstraction. To these directional rules HTLG adds inference rules for a nondirectional type constructor $|$
interpreted by phonological functional application and phonological functional abstraction:
\disp{
\prooftree
\prooftree
\leadsto
\alpha\ass C|B\ass \phi
\endprooftree
\prooftree
\leadsto
\beta\ass B\ass \psi
\endprooftree
\justifies
\alpha(\beta)\ass C\ass (\phi\ \psi)
\using
|E
\endprooftree
\tb
\prooftree
\prooftree
\prooftree
\justifies
\phi\ass B\ass y
\using n
\endprooftree
\leadsto
\alpha\ass C\ass \chi
\endprooftree
\justifies
\lambda \phi.\alpha\ass C|B\ass \lambda y\chi
\using |I^n
\endprooftree
}
These are the characteristic rules of HTLG (for more details we refer the reader to the papers of Kubota and Levine). 

In relation to gapping, K\&L (2015\cite{htlgnllt:gapping})~(53) present a type assignment which is in essence:\footnote{Throughout, $\VP$ abbreviates $N\bsl S$. We limit attention to gapping of the transitive
verb category; so far as we are aware gapping in other categories raises no new issues
differentiating between hybrid TLG and displacement TLG.} 
\disp{\begin{tabular}[t]{l}$
\lambda\sigma_2\lambda\sigma_1\lambda\phi.\sigma_1(\phi)\add\syncnst{and}\add
\sigma_2(\nil)\ass (X|X)|X\ass \lambda x\lambda y\lambda z[(y\ z)\wedge(x\ z)]$\\
where $X=S|(\VP/N)$
\end{tabular}}
And they derive from this
simple gapping such as (\ref{gappingex}a);
see {\it ibid}~(52) and~(55).
In our notation, their~(52) is:
\disp{
\prooftree
\prooftree
\syncnst{robin}\ass N\ass \semcnst{r}
\prooftree
\prooftree
\justifies
\phi\ass \VP/N\ass x
\using i
\endprooftree
\syncnst{bill}\ass N\ass \semcnst{b}
\justifies
\phi\add\syncnst{bill}\ass\VP\ass(x\ \semcnst{b})
\using E/
\endprooftree
\justifies
\syncnst{robin}\add\phi\add\syncnst{bill}\ass S\ass((x\ \semcnst{b})\ \semcnst{r})
\using E\bsl
\endprooftree
\justifies
\lambda\phi.\syncnst{robin}\add\phi\add\syncnst{bill}\ass S|(\VP/N)\ass\lambda x((x\ \semcnst{b})\ \semcnst{r})
\using I|^i
\endprooftree
}
Continuing the derivation as in their~(55) yields:
\disp{$\syncnst{leslie}+\syncnst{met}+\syncnst{sandy}+\syncnst{and}+\syncnst{robin}+\syncnst{bill}\ass
S\ass((\semcnst{met\ s})\ \semcnst{l})\wedge((\semcnst{met\ b})\ \semcnst{r})$}
K\&L also assume in their~(56) a raised type for an auxiliary:
\disp{$\lambda\sigma.\sigma(\syncnst{must})\ass S|(S|(VP/VP))\ass \lambda x(\semcnst{Nec}\ (x\ \lambda yy))$}
They show that such assignments to auxiliaries license the auxiliary wide-scope
reading of, say, 
$$lingform{John must eat steak and Mary pizza}.$$

Likewise, K\&L (2015\cite{htlgnllt:gapping})~(66) present a type assignment
for determiner gapping which is:
\disp{\begin{tabular}[t]{l}
$\lambda\rho_2\lambda\rho_1\lambda\phi\lambda\sigma.\rho_1(\phi)(\sigma)\add\syncnst{and}\add
\rho_2(\nil)(\lambda\chi\lambda\psi.\psi(\chi))\ass (X|X)|X\ass$\\$ \lambda x\lambda y\lambda z\lambda w[((y\ z)\ w)\wedge((x\ z)\ w)]$\\
where $X=(S|((S|(S|N))|\CN))|(\VP/N)$
\end{tabular}
}
And they assume, K\&L (2015\cite{htlgnllt:gapping})~(65),
 a raised type for a negative determiner of the form:
\disp{$
\lambda\rho.\rho(\lambda\phi\lambda\sigma.\sigma(\syncnst{no}\add\phi))\ass
S|((S|(S|(S|N)))|\CN)\ass
\lambda x(\neg(x\ \lambda y\lambda z\exists w[(y\ w)\wedge(z\ w)]))$}
These enable derivation of the split scope reading of:
\disp{
No dog eats whiskas or cat alpo.
}

\section{The scopal anomalies are already available in displacement TLG}

K\&L (2015\cite{htlgnllt:gapping}) state that `our own analysis
resembles most closely Morrill et al.'s (2011) (which is a refinement of
Hendriks 1995)'. The displacement type logical grammar (DTLG)
gapping assignment is as follows, where $\wprod$ is discontinuous product, which is semantically
interpreted by ordered pairing, and `$\pi_1 x$' selects the first component of $x$:\footnote{The notation
$\pi_1$ (and $\pi_2$) represents first (and second) projection of an ordered pair,
so that e.g.\ $\pi_1(\phi, \psi) = \phi$.}
 \disp{\begin{tabular}[t]{l}
$\syncnst{and}\ass (X\bsl X)/(X\wprod I)\ass \lambda x\lambda y\lambda z[(y\ z)\wedge(\pi_1x\ z)]$\\
where $X=S\circum(\VP/N)$
\end{tabular}}
Here, the symbols such as $I$ and $\circum{}$ are connectives of the displacement calculus, which we define
later.
The corresponding determiner gapping assignment would be:
\disp{\begin{tabular}[t]{l}$
\syncnst{and}\ass (X\bsl X)/((X\wprod I)\wprod I)\ass\lambda x\lambda y\lambda z\lambda w[((y\ z)\ w)\wedge((\pi_1\pi_1x\ z)\ w)]$\\
where $X=(S \circum(\VP/N))\circum(((S\circum N)\infix S)/\CN)$
\end{tabular}}

The auxiliary and negative determiner assignments would be:
\disp{$\syncnst{must}\ass (S\circum(VP/VP))\infix S\ass \lambda x(\semcnst{Nec}\ (x\ \lambda yy))$}
\disp{$
\syncnst{no}\ass
(S\circum(((S\circum N)\infix S)/\CN))\infix S\ass
\lambda x(\neg(x\ \lambda y\lambda z\exists w[(y\ w)\wedge(z\ w)]))$}
Thus the analyses of gapping scope anomalies are already available in
DTLG. Indeed K\&L (2015\cite{htlgnllt:gapping}) state, 'To be fair,
the core of our empirical results, so far as we can tell, seems to straightforwardly
carry over to Morrill et al.'s (2011) system.'
So K\&L's type logical contribution is the determiner gapping,
and the observation that raised auxiliary and negative
determiner assignments capture the scopal anomalies in HTLG,
but also in DTLG. 
Thus, although K\&L couch their solution in terms of HTLG,
HTLG and DTLG are on a par in respect of gapping and the scopal anomalies.
Since this point appears to have been granted we do not elaborate on it further.
But we will see a respect in which DTLG improves on HTLG,
and another respect in which DTLG can be made to improve further on HTLG.

\section{The HTLG determiner gapping assignment overgenerates}

Kubota (p.c.) points out that the HTLG analysis of
determiner gapping incorrectly predicts such examples as the following,
where the determiner and the transitive verb orders are not
consistent in the conjuncts:
\disp{\begin{tabular}[t]{ll}
a. & \unacc Some dogs like Whiskas and I (like) (some) cats.\\
b. &\unacc  I like some cats and (some) dogs (like) Whiskas.
\end{tabular}\label{perm}}
This order inconsistency overgeneration
arises because the $|((S|(S|N))|\CN)$ and $|(\VP/N)$ arguments in the types for
determiner gapping do not distinguish the linear order of the phonological
variables they bind. Thus, for example, in relation to (\ref{perm}a), both 
$\lambda v\lambda q.q\add\syncnst{dogs}\add v\add\syncnst{Whiskas}$
and $\lambda v\lambda q.\syncnst{I}\add v\add q\add\syncnst{cats}$
have type $(S|((S|(S|N))|\CN))|(\VP/N)$.
Note that the issue has nothing to do with the order in which the quantifier
and transitive verb arguments are abstracted,
but rather with the left to right position of the variables they bind in the body of the phonological term,
to which the type constructor $|$ is insensitive.
DTLG does not overgenerate in this way because the positions of
discontinuity are indexed for left to right order.

\subsection{HTLG requires distinct type assignments for simplex and discontinuous gapping}

Our analysis is inspired by that of
Kubota and Levine, 
which assigns coordinator conjunct types and prosody as follows:

\noindent
{\footnotesize
\begin{tabular}[t]{c|c|c}
 &&\\
conjunt type  & gapping & determiner gapping\\
\& &&\\
coordinator &&\\
phonology
  &&\\
\hline
 &&\\
simplex & $S|(\VP/N)$ & $(S|(\VP/N))|((S|(S|N))|\CN)$\\
 & $\lambda\sigma_2\lambda\sigma_1\lambda\phi.\sigma_1(\phi)\add\syncnst{and}\add
\sigma_2(\nil)$ & $\lambda\rho_2\lambda\rho_1\lambda\phi\lambda\sigma.\rho_1(\phi)(\sigma)\add\syncnst{and}\add
\rho_2(\lambda\chi\lambda\psi.\psi(\chi))(\nil)$\\
 &K\&L (2012\cite{htlggapping})~(5) K\&L (2013\cite{kublev:detgap})~(13)&
 K\&L (2013\cite{kublev:detgap})~(24) K\&L (2015\cite{htlgnllt:gapping})~(66)~(83)\\
& K\&L (2015\cite{htlgnllt:gapping})~(53)\\
 \hline
 &&\\
discont. & $S|(\VP|N)$ & $(S|(\VP|N))|((S|(S|N))|\CN)$\\
 & $\lambda\rho_2\lambda\rho_1\lambda\phi.\rho_1(\phi)\add\syncnst{and}\add\rho_2(\lambda\pi.\pi)$ & $\lambda\rho_2\lambda\rho_1\lambda\phi\lambda\sigma.\rho_1(\phi)(\sigma)\add\syncnst{and}\add
\rho_2(\lambda\chi\lambda\psi.\psi(\chi))(\lambda\pi.\pi)$\\
 & K\&L (2012\cite{htlggapping})~(20) & (by extrapolation)\\
 \end{tabular}}
 
\noindent
However, our analysis of gapping represents an improvement
on the HTLG analysis in that we require only a single type for
simple and discontinuous gapping.
HTLG requires two types because simplex gapping conditionalises $\VP/N$ of sort
\emph{string\/} whereas discontinuous gapping conditionalises $\VP|N$ of sort \emph{function},
and these require two distinct prosodic operations: application to the empty string
$\nil$ (of sort \emph{string}) and to the identity function $\lambda\pi.\pi$ (of sort \emph{function})
respectively.

\subsection{Our account}

We give an account of simple and discontinuous gapping as in~(\ref{gappingex})
and simplex and discontinuous determiner gapping as in~(\ref{detgappingex})
which is optimal in that a single coordinator type
generates discontinuous gapping with simple gapping as a special case
and a single coordinator type generates discontinuous determiner gapping with
simplex determiner gapping as a special case.
The account is expressed in an extension of displacement calculus
({\bf D}; Morrill et al.\ 2011\cite{mvf:tdc}) which we call second-order displacement calculus
{\bf D$^2$}.

\subsubsection{Second-order displacement calculus}

Displacement calculus is a logic of discontinuous strings.
By discontinuous strings we mean strings punctuated by a distinguished
vocabulary item `$\one$' called the \techterm{separator}.
In contrast to HTLG, 
the phonological terms of displacement calculus have no lambda abstraction
and,
instead of a set of variable placeholders,
DTLG has a single placeholder, the separator.
The sort of a discontinuous string is the number of separators
it contains.
We notate by $L_i, i\ge 0,$ the set of all strings of sort $i$ with respect to
some alphabet.
We consider the operations concatenation, intercalation, and
adjunction on discontinuous strings. Concatenation is represented in (\ref{concop}).

\disp{$
\begin{array}[t]{c}
\fbox{$\tb\alpha\tb$\strut}\ \add\ \fbox{$\tb\beta\tb$\strut}\\
=\\
\fbox{$\tb\alpha\tb$\strut}\fbox{$\tb\beta\tb$\strut}\\\\
\mbox{concatenation\ } +: L_i, L_j\rightarrow L_{i{+}j}
\end{array}
$\label{concop}}
For example, the concatenation of $\syncnst{Leslie}\add\one\add\syncnst{Sandy}$ and
$\syncnst{and}\add\syncnst{Robin}\add\syncnst{Bill}$ is:

\disp{$\begin{array}[t]{c}
\syncnst{Leslie}\add\one\add\syncnst{Sandy} + 
\syncnst{and}\add\syncnst{Robin}\add\syncnst{Bill}
\\ =
\\
\syncnst{Leslie}\add\one\add\syncnst{Sandy}\add\syncnst{and}\add\syncnst{Robin}\add\syncnst{Bill}
\end{array}
$}

Intercalation is represented in (\ref{intercop}):
\disp{$
\begin{array}[t]{c}
\fbox{$\tb\alpha\tb$\strut}\fbox{$1$\strut}\fbox{$\tb\gamma\tb$\strut}
\ \swrap{k}\ 
\fbox{$\tb\beta\tb$\strut}\\
=\\
\fbox{$\tb\alpha\tb$\strut}\fbox{$\tb\beta\tb$\strut}\fbox{$\tb\gamma\tb$\strut}\\\\
\mbox{intercalation\ } \swrap{k}: L_{i'}, L_j\rightarrow L_{i{+}j}
\end{array}
$\label{intercop}}
For example, the intercalation at the second separator of $\one\add\syncnst{dogs}\add\one\add\syncnst{Whiskas}$ 
$\add\syncnst{and}\add\syncnst{cats}\add\syncnst{Alpo}$ and $\syncnst{like}$ is:

\disp{$\begin{array}[t]{c}
\one\add\syncnst{dogs}\add\one\add\syncnst{Whiskas}\add\syncnst{and}\add\syncnst{cats}\add\syncnst{Alpo}\ \swrap{2}\ \syncnst{like}
\\ =
\\
\sep\add\syncnst{dogs}\add\syncnst{like}\add\syncnst{Whiskas}\add\syncnst{and}\add\syncnst{cats}\add\syncnst{Alpo}
\end{array}
$}

Finally, adjunction is represented in (\ref{adjunctop}):
\disp{$
\begin{array}[t]{c}
\fbox{$\tb\alpha\tb$\strut}\fbox{$\one$\strut}\fbox{$\tb\gamma\tb$\strut}\fbox{$\one$\strut}\fbox{$\tb\epsilon\tb$\strut}
\ \swraptwo{k, l}\ 
\fbox{$\tb\beta\tb$\strut}\fbox{$1$\strut}\fbox{$\tb\delta\tb$\strut}\\
=\\
\fbox{$\tb\alpha\tb$\strut}\fbox{$\tb\beta\tb$\strut}\fbox{$\tb\gamma\tb$\strut}\fbox{$\tb\delta\tb$\strut}\fbox{$\tb\epsilon\tb$\strut}\\\\
\mbox{adjunction\ } \swraptwo{k, l}: L_{i{+}2}, L_{j{+}1}\rightarrow L_{i{+}j}
\end{array}
$\label{adjunctop}}
For example, the adjunction at the first separators of
$$\syncnst{John}\add\one\add\syncnst{W.}\add\one\add\syncnst{and}\add$ $\syncnst{Daniel}\add\syncnst{C.}$$
and
$$\syncnst{wants}\add\one\add\syncnst{to}\add\syncnst{win}$ is:

\disp{$\begin{array}[t]{c}
\syncnst{John}\add\one\add\syncnst{Watford}\add\one\add\syncnst{and}\add\syncnst{Daniel}\add\syncnst{Chelsea}\ \swraptwo{1, 1}\ \syncnst{wants}\add\one\add\syncnst{to}\add\syncnst{win}
\\ =
\\
\syncnst{John}\add\syncnst{wants}\add\syncnst{Watford}\add\syncnst{to}\add\syncnst{win}\add\syncnst{and}\add\syncnst{Daniel}\add\syncnst{Chelsea}\end{array}
$}

We will have three families of type-constructors defined in relation to the three prosodic operations of concatenation, intercalation, and adjunction. The syntactic types are sorted $\Tp_0, \Tp_1, \Tp_2, \ldots$
according to the number of points of discontinuity $0, 1, 2, \ldots$ their expressions contain.
The sets $\Tp_i$ of types of sort $i$ are defined by mutual recursion in terms of sets ${\cal P}_i$
of primitive types of sort $i$ as follows:\\

\noindent
{\small$
\begin{array}{rclrclll}
\Tp_i & ::= & {\cal P}_i\\\\
\Tp_i & ::= & \Tp_{i{+}j}/\Tp_j & T(C/B) & = & T(B){\rightarrow}T(C)
 & \mbox{over}\\
\Tp_j & ::= & \Tp_i\bsl\Tp_{i{+}j} & T(A\bsl C) & = & T(A){\rightarrow}T(C)
 & \mbox{under}\\
\Tp_{i{+}j} & ::= & \Tp_i\product\Tp_j & T(A\product B) & = & T(A)\&T(B) &
\mbox{continuous product}\\
\Tp_0 & ::= & I & T(I) & = & \top & \mbox{cont.\ unit}\\\\
\Tp_{i{+}1} & ::= & \Tp_{i{+}j}\scircum{k}\Tp_j, 1\le k\le i{+}1
& T(C\scircum{k} B) & = & T(B){\rightarrow}T(C) & \mbox{extract}\\
\Tp_j & ::= & \Tp_{i{+}1}\sinfix{k}\Tp_{i{+}j}, 1\le k\le i{+}1 
& T(A\sinfix{k} C) & = & T(A){\rightarrow}T(C) & \mbox{infix}\\
\Tp_{i{+}j} & ::= & \Tp_{i{+}1}\swprod{k}\Tp_j, 1\le k\le i{+}1 
& T(A\swprod{k} B) & = & T(A)\&T(B)& \mbox{disc.\ product}\\
\Tp_1 & ::= & J & T(J) & = & \top & \mbox{disc.\ unit}\\\\
\Tp_{i{+}2} & ::= & \Tp_{i{+}j}\scircumtwo{k, l}\Tp_{j{+}1}, 1\le k\le i{+}1, 1\le l\le j{+}1
& T(C\scircumtwo{k, l} B) & = & T(B){\rightarrow}T(C) & \mbox{2nd order extract}\\
\Tp_{j{+}1} & ::= & \Tp_{i{+}2}\sinfixtwo{k, l}\Tp_{i{+}j}, 1\le k\le i{+}1, 1\le l\le j{+}1 
& T(A\sinfixtwo{k, l} C) & = & T(A){\rightarrow}T(C) & \mbox{2nd order infix}\\
\Tp_{i{+}j} & ::= & \Tp_{i{+}2}\sdprodtwo{k, l}\Tp_{j{+}1}, 1\le k\le i{+}1, 1\le l\le j{+}1 
& T(A\swprodtwo{k, l} B) & = & T(A)\&T(B)& \mbox{2nd order disc.\ product}\\
\Tp_2 & ::= & K & T(K) & = & \top & \mbox{2nd order disc.\ unit}
\end{array}
$}\\

\noindent The second column of this table shows the standard categorial semantic type map 
for the connectives.\footnote{For example, if $T(N)=e$ where $e$ is the semantic type for individuals,
and $T(S)=t$ where $t$ is the semantic type corresponding to truth-values, we have then that 
$T(N\bsl S)=e\rightarrow t$.} Each type of sort $i$ is interpreted as a set of (discontinuous) strings
of sort $i$.
The prosodic interpretation is as follows:\\

$
{\small\begin{array}{rcl}
\oscott C/B\cscott & = & \{s_1|\ \forall s_2\in\oscott B\cscott, s_1\add s_2\in\oscott C\cscott\}\\
\oscott A\bsl C\cscott & = & \{s_2|\ \forall s_1\in\oscott A\cscott, s_1\add s_2\in\oscott C\cscott\}\\
\oscott A\product B\cscott & = & \{s_1\add s_2|\ s_1\in\oscott A\cscott\ \&\ s_2\in\oscott B\cscott\}\\
\oscott I\cscott & = & \{\nil\}
\\\\
\oscott C\scircum{k}B\cscott & = & \{s_1|\ \forall s_2\in\oscott B\cscott, s_1\swrap{k} s_2\in\oscott C\cscott\}\\
\oscott A\sinfix{k} C\cscott & = & \{s_2|\ \forall s_1\in\oscott A\cscott, s_1\swrap{k} s_2\in\oscott C\cscott\}\\
\oscott A\swprod{k} B\cscott & = & \{s_1\swrap{k} s_2|\ s_1\in\oscott A\cscott\ \&\ s_2\in\oscott B\cscott\}\\
\oscott J\cscott & = & \{\one\}
\\\\
\oscott C\scircumtwo{k, l}B\cscott & = & \{s_1|\ \forall s_2\in\oscott B\cscott, s_1\swraptwo{k, l} s_2\in\oscott C\cscott\}\\
\oscott A\sinfixtwo{k, l} C\cscott & = & \{s_2|\ \forall s_1\in\oscott A\cscott, s_1\swraptwo{k, l} s_2\in\oscott C\cscott\}\\
\oscott A\swprodtwo{k,l} B\cscott & = & \{s_1\swraptwo{k, l} s_2|\ s_1\in\oscott A\cscott\ \&\ s_2\in\oscott B\cscott\}\\
\oscott K\cscott & = & \{\one\add\one\add\one\}
\end{array}}
$\\

Although linguistically only some of the power, and hence only some of the rules, are necessary here,
the framework of DTLG complies with the modern logical paradigm of logic as an interpreted formal
language, and aspiration to \emph{soundness\/} (that everything said is true) and \emph{completeness\/}
(that everything true is said). This, to us, is the rock on which type \emph{logical\/} grammar is founded.
Thus, we present here all the rules so that the reader has the complete picture
of which the gapping analysis uses just a part.

The rules for second-order displacement calculus fall into three groups for the concatenative,
intercalative, and adjunctive connective families. Each family contains four connectives:
the two implicational residuals, the conjunctive product, and product unit. Each connective
has two rules, namely a rule of elimination (E), eliminating the connective reading from premise
to conclusion, and a rule of introduction (I) introducing the connective reading from premise to
conclusion.\footnote{Except we omit the elimination rules for product units which,
as well as being unmotivated linguistically, are awkward to formulate in the format used here.} Although there are many
rules, the reader should be aware of the high degree
of symmetry between them, and that the rules simply formalise the necessary and sufficient conditions
for membership of syntactic types: the rules are essentially the result of restating the interpretation clauses
given above.

The labelled natural deduction for second-order displacement calculus is as follows,
where we use three conventions. Firstly,
where $\syncnst{a}$ is a prosodic constant of sort $i$,
the \techterm{vector} $\vect{\syncnst{a}}$ is $\syncnst{a}_0\add\one\add\syncnst{a}_1\add\one\add\cdots\add\one\add\syncnst{a}_i$; for example,
if \syncnst{a} is of sort 1, $\vect{\syncnst{a}}=\syncnst{a}_0\add\one\add\syncnst{a}_1$. Secondly,
where $\alpha$ is a discontinuous string of sort $i>0$,
$\alpha \smwrap{k} \beta$, $1\le k\le i$, is the result of replacing the $k$th separator in $\alpha$
by $\beta$ (counting from the left);
for example
\syncnst{John}\add\one\add\syncnst{W.}\add\one\add\syncnst{and}\add\syncnst{Daniel}\add\syncnst{C.}\ \smwrap{2}\ \syncnst{to}\add\syncnst{win} = 
\syncnst{John}\add\one\add\syncnst{W.}\add\syncnst{to}\add\syncnst{win}\add\syncnst{and}\add\syncnst{Daniel}\add\syncnst{C.}. Thirdly,
$\alpha \smwwrap{k} \beta$ abbreviates $\alpha \smwrap{k} \one\add\beta\add\one$;
for example
\syncnst{John}\add\one\add\syncnst{and}\add\syncnst{Daniel}\add\syncnst{C.}\ \smwwrap{1}\
\syncnst{W.} =
\syncnst{John}\add\one\add\syncnst{W.}\add\one\add\syncnst{and}\add\syncnst{Daniel}\add\syncnst{C.}.\\

Continuous family:
\begin{itemize}
\item
Elimination rules for
implications\\

\prooftree
\prooftree
\leadsto
\alpha\ass C/B\ass \phi
\endprooftree
\prooftree
\leadsto
\beta\ass B\ass \psi
\endprooftree
\justifies
\alpha\add\beta\ass C\ass (\phi\ \psi)
\using
/E
\endprooftree
\tab
\prooftree
\prooftree
\leadsto
\alpha\ass A\ass \phi
\endprooftree
\prooftree
\leadsto
\beta\ass A\bsl C\ass \psi
\endprooftree
\justifies
\alpha\add\beta\ass C\ass (\psi\ \phi)
\using
\bsl E
\endprooftree\\\ \\

\item
Elimination rule for product\\

\prooftree
\prooftree
\leadsto
\gamma\ass A\product B\ass\chi
\endprooftree
\prooftree
\prooftree
\justifies
\vect{a}\ass A\ass x
\using n
\endprooftree
\prooftree
\justifies
\vect{b}\ass B\ass y
\using n
\endprooftree
\leadsto
\delta(\vect{a}\add\vect{b})\ass D\ass \omega(x, y)
\endprooftree
\justifies
\delta(\gamma)\ass D\ass\omega(\pi_1\chi, \pi_2\chi)
\using \product E^n
\endprooftree\\\ \\

\item
Introduction rules for implications\\

\prooftree
\prooftree
\prooftree
\justifies
\vect{b}\ass B\ass y
\using n
\endprooftree
\leadsto
\alpha\add\vect{b}\ass C\ass \chi
\endprooftree
\justifies
\alpha\ass C/B\ass \lambda y\chi
\using /I^n
\endprooftree
\tab
\prooftree
\prooftree
\prooftree
\justifies
\vect{a}\ass A\ass x
\using n
\endprooftree
\leadsto
\vect{a}\add\beta\ass C\ass \chi
\endprooftree
\justifies
\beta\ass A\bsl C\ass \lambda x\chi
\using \bsl I^n
\endprooftree\\\ \\

\item
Introduction rules for product and product unit\\

\prooftree
\prooftree
\leadsto
\alpha\ass A\ass \phi
\endprooftree
\tb
\prooftree
\leadsto
\beta\ass B\ass \psi
\endprooftree
\justifies
\alpha\add\beta\ass A\product B\ass (\phi, \psi)
\using
\product I
\endprooftree
\tab
\prooftree
\justifies
\nil\ass I\ass\zero
\using II
\endprooftree\\
\end{itemize}

Discontinuous family:
\begin{itemize}
\item
Elimination rules for
implications\\

\prooftree
\prooftree
\leadsto
\alpha\ass C\scircum{k}B\ass \phi
\endprooftree
\prooftree
\leadsto
\beta\ass B\ass \psi
\endprooftree
\justifies
\alpha\smwrap{k}\beta\ass C\ass (\phi\ \psi)
\using
\circum E
\endprooftree
\tab
\prooftree
\prooftree
\leadsto
\alpha\ass A\ass \phi
\endprooftree
\prooftree
\leadsto
\beta\ass A\sinfix{k} C\ass \psi
\endprooftree
\justifies
\alpha\smwrap{k}\beta\ass C\ass (\psi\ \phi)
\using
\infix E
\endprooftree\\\ \\

\item 
Elimination rule for product\\

\prooftree
\prooftree
\leadsto
\gamma\ass A\wprod B\ass\chi
\endprooftree
\prooftree
\prooftree
\justifies
\vect{a}\ass A\ass x
\using n
\endprooftree
\prooftree
\justifies
\vect{b}\ass B\ass y
\using n
\endprooftree
\leadsto
\delta(\vect{a}\smwrap{k}\vect{b})\ass D\ass \omega(x, y)
\endprooftree
\justifies
\delta(\gamma)\ass D\ass\omega(\pi_1\chi, \pi_2\chi)
\using \wprod E^n
\endprooftree\\\ \\

\item
Introduction rules for implications\\

\prooftree
\prooftree
\prooftree
\justifies
\vect{b}\ass B\ass y
\using n
\endprooftree
\leadsto
\alpha\smwrap{k}\vect{b}\ass C\ass \chi
\endprooftree
\justifies
\alpha\ass C\scircum{k}B\ass \lambda y\chi
\using \circum I^n
\endprooftree
\tab
\prooftree
\prooftree
\prooftree
\justifies
\vect{a}\ass A\ass x
\using n
\endprooftree
\leadsto
\vect{a}\smwrap{k}\beta\ass C\ass \chi
\endprooftree
\justifies
\beta\ass A\sinfix{k} C\ass \lambda x\chi
\using \infix I^n
\endprooftree\\\ \\

\item
Introduction rules for product and product unit\\

\prooftree
\prooftree
\leadsto
\alpha\ass A\ass \phi
\endprooftree
\tb
\prooftree
\leadsto
\beta\ass B\ass \psi
\endprooftree
\justifies
\alpha\smwrap{k}\beta\ass A\swprod{k} B\ass (\phi, \psi)
\using
\product I
\endprooftree
\tab
\prooftree
\justifies
\one\ass J\ass\zero
\using IJ
\endprooftree
\end{itemize}

Second-order discontinuous family:
\begin{itemize}
\item
Elimination rules for
implications\\

\prooftree
\prooftree
\leadsto
\alpha\smwwrap{k}\gamma\ass C\scircumtwo{k, l}B\ass \phi
\endprooftree
\prooftree
\leadsto
\beta\ass B\ass \psi
\endprooftree
\justifies
\alpha\smwrap{k}(\beta\smwrap{l}\gamma)\ass C\ass (\phi\ \psi)
\using
\circumtwo E
\endprooftree
\tab
\prooftree
\prooftree
\leadsto
\alpha\smwwrap{k}\gamma\ass A\ass \phi
\endprooftree
\prooftree
\leadsto
\beta\ass A\sinfixtwo{k, l} C\ass \psi
\endprooftree
\justifies
\alpha\smwrap{k}(\beta\smwrap{l}\gamma)\ass C\ass (\psi\ \phi)
\using
\infixtwo E
\endprooftree\\\ \\

\item
Elimination rule for product\\

\prooftree
\prooftree
\leadsto
\gamma\ass A\swprodtwo{k, l} B\ass\chi
\endprooftree
\prooftree
\prooftree
\justifies
\vect{a}\smwwrap{k}\vect{c}\ass A\ass x
\using n
\endprooftree
\prooftree
\justifies
\vect{b}\ass B\ass y
\using n
\endprooftree
\leadsto
\delta(\vect{a}\smwrap{k}(\vect{b}\smwrap{l}\vect{c}))\ass D\ass \omega(x, y)
\endprooftree
\justifies
\delta(\gamma)\ass D\ass\omega(\pi_1\chi, \pi_2\chi)
\using \wprodtwo E^n
\endprooftree\\\ \\

\item
Introduction rules for implications\\

\prooftree
\prooftree
\prooftree
\justifies
\vect{b}\ass B\ass y
\using n
\endprooftree
\leadsto
\alpha\smwrap{k}(\vect{b}\smwrap{l}\gamma)\ass C\ass \chi
\endprooftree
\justifies
\alpha\smwwrap{k}\gamma\ass C\scircumtwo{k, l}B\ass \lambda y\chi
\using \circumtwo I^n
\endprooftree
\tab
\prooftree
\prooftree
\prooftree
\justifies
\vect{a}\smwrap{l}\vect{c}\ass A\ass x
\using n
\endprooftree
\leadsto
\vect{a}\smwrap{k}(\beta\smwrap{l}\vect{c})\ass C\ass \chi
\endprooftree
\justifies
\beta\ass A\sinfixtwo{k, l} C\ass \lambda x\chi
\using \infixtwo I^n
\endprooftree\\\ \\

\item
Introduction rules for product and product unit\\

\prooftree
\prooftree
\leadsto
\alpha\smwwrap{k}\gamma\ass A\ass \phi
\endprooftree
\tb
\prooftree
\leadsto
\beta\ass B\ass \psi
\endprooftree
\justifies
\alpha\smwrap{k}(\beta\smwrap{l}\gamma)\ass A\swprodtwo{k, l} B\ass (\phi, \psi)
\using
\product E
\endprooftree
\tab
\prooftree
\justifies
\one\add\one\add\one\ass K\ass\zero
\using IK
\endprooftree
\end{itemize}
\ \\

\noindent
We adopt the convention that when subscripts $k$ and $l$ are omitted they
are $1$, i.e.\ they default to $1$.
By way of example, the following auxiliary derivation shows that a subject followed
by an object has type $S\circumtwo(\VP\circum N)$:
\disp{
 \prooftree
 \prooftree
 \syncnst{sbj}\ass N\ass\semcnst{sbj}
 \prooftree
 \prooftree
 \justifies
 a_0\add\sep\add a_1\ass\VP\circum N\ass x
 \using i
 \endprooftree
 \syncnst{obj}\ass N\ass\semcnst{obj}
 \justifies
 a_0\add\syncnst{obj}\add a_1\ass\VP\ass(x\ \semcnst{obj})
 \using \circum E
 \endprooftree
 \justifies
 \syncnst{sbj}\add a_0\add\syncnst{obj}\add a_1\ass S\ass((x\ \semcnst{obj})\ \semcnst{sbj})
 \using \bsl E
 \endprooftree
 \justifies
 \syncnst{sbj}\add\sep\add\syncnst{obj}\add\sep\ass S\circumtwo(\VP\circum N)\ass\lambda x((x\ \semcnst{obj})\ \semcnst{sbj})
 \using \circumtwo I^i
 \endprooftree
 \label{subjobjder}
}

\subsubsection{Analyses}

The account of gapping consists in an assignment to the
coordinator of the following type:
\disp{\begin{tabular}[t]{l}
$\syncnst{and}\ass (X\bsl X)/(X\wprodtwo J)\ass \lambda x\lambda y\lambda z[(y\ z)\wedge(\pi_1 x\ z)]$\\
where $X= S\circumtwo(\VP\circum N)$
\end{tabular}\label{gappingtype}}
Consider example (\ref{gappingex}a) of simple gapping:
\lingform{Leslie met Sandy and Robin Bill}.
Then there is the following derivation of (\ref{gappingex}a) using (\ref{subjobjder}):

\noindent{\footnotesize
\prooftree
\prooftree
 \prooftree
 {\rm Leslie}\ {\rm Sandy}
 \justifies
 \begin{array}{c}
 \syncnst{L}\add\sep\add\syncnst{S}\add\sep\ass\\
 S\circumtwo(\VP\circum N)\ass\\
 \lambda x((x\ \semcnst{s})\ \semcnst{l})
 \end{array}
 \endprooftree 
 \prooftree
 \begin{array}[b]{c}
 \syncnst{and}\ass\\
 ((S\circumtwo(\VP\circum N))\bsl(S\circumtwo(\VP\circum N)))/\\
 /((S\circumtwo (\VP\circum N))\wprodtwo J)\ass\\
 \lambda x\lambda y\lambda z[(y\ z)\wedge(\pi_1 x\ z)]
 \end{array}
 \prooftree
 \prooftree
 {\rm Robin}\ {\rm Bill}
 \justifies
\begin{array}{c}
\syncnst{R}\add\sep\add\syncnst{B}\add\sep\ass\\
S\circumtwo(\VP\circum N)\ass\\
\lambda x((x\ \semcnst{b})\ \semcnst{r})
\end{array}
 \endprooftree
 \prooftree
 \justifies
 \sep\ass J\ass \zero
 \using JR
 \endprooftree
 \justifies
 \syncnst{R}\add\syncnst{B}\ass (S\circum(\VP/N))\wprodtwo J\ass (\lambda x((x\ \semcnst{b})\ \semcnst{r}), \zero)
 \using \wprodtwo I
 \endprooftree
\justifies
\syncnst{and}\add\syncnst{R}\add\syncnst{B}\ass
 (S\circumtwo(\VP\circum N))\bsl(S\circumtwo(\VP\circum N))\ass
 \lambda y\lambda z[(y\ z)\wedge((z\ \semcnst{b})\ \semcnst{r})]
 \using /E
 \endprooftree
\justifies
\syncnst{L}\add\sep\add\syncnst{S}\add\sep\add\syncnst{and}\add\syncnst{R}\add\syncnst{B}\ass
 S\circumtwo(\VP\circum N)\ass
 \lambda z[((z\ \semcnst{s})\ \semcnst{l})\wedge((z\ \semcnst{b})\ \semcnst{r})]
 \using \bsl E
 \endprooftree
 \prooftree
 \prooftree
 \syncnst{met}\ass\VP/N\ass\semcnst{meet}
 \prooftree
\justifies
 a\ass N\ass x
 \using i
 \endprooftree
 \justifies
 \syncnst{met}\add a\ass\VP\ass(\semcnst{meet}\ x)
 \using /E
 \endprooftree
\justifies
 \syncnst{met}\add \sep\ass\VP\circum N\ass\lambda x(\semcnst{meet}\ x)
 \using \circum I^i
 \endprooftree
 \justifies
\syncnst{L}\add\syncnst{met}\add\syncnst{S}\add\syncnst{and}\add\syncnst{R}\add\syncnst{B}\ass
 S\ass
 {}[((\semcnst{meet}\ \semcnst{s})\ \semcnst{l})\wedge((\semcnst{meet}\ \semcnst{b})\ \semcnst{r})]
\using \circumtwo E
\endprooftree 
 
 }
 
\ 

\noindent
And from the same coordinator type assignment (\ref{gappingtype})
 there is the following derivation of the discontinuous
 gapping (\ref{gappingex}b)
 \lingform{John wants Watford to win and} \lingform{Daniel Chelsea}:

\ 

\noindent
 {\footnotesize
\prooftree
\prooftree
 \prooftree
 {\rm John}\ {\rm Watford}
 \justifies
 \begin{array}{c}
 \syncnst{J}\add\sep\add\syncnst{W}\add\sep\ass\\
 S\circumtwo(\VP\circum N)\ass\\
 \lambda x((x\ \semcnst{w})\ \semcnst{j})
 \end{array}
 \endprooftree 
 \prooftree
 \begin{array}[b]{c}
 \syncnst{and}\ass\\
 ((S\circumtwo(\VP\circum N))\bsl(S\circumtwo(\VP\circum N)))/\\
 /((S\circumtwo (\VP\circum N))\wprodtwo J)\ass\\
 \lambda x\lambda y\lambda z[(y\ z)\wedge(\pi_1 x\ z)]
 \end{array}
 \prooftree
 \prooftree
 {\rm Daniel}\ {\rm Chelsea}
 \justifies
 \begin{array}{c}
 \syncnst{D}\add\sep\add\syncnst{S}\add\sep\ass\\
 S\circumtwo(\VP\circum N)\ass\\
 \lambda x((x\ \semcnst{s})\ \semcnst{d})
 \end{array}
 \endprooftree
 \prooftree
 \justifies
 \begin{array}{c}
 \sep\ass\\
 J\ass\\
 \zero
 \end{array}
 \using JR
 \endprooftree
 \justifies
 \begin{array}{c}
 \syncnst{D}\add\syncnst{C}\ass\\
 (S\circum(\VP/N))\wprodtwo J\ass\\
 (\lambda x((x\ \semcnst{c})\ \semcnst{d}), \zero)
 \end{array}
 \using \wprodtwo I
 \endprooftree
\justifies
\begin{array}{c}
\syncnst{and}\add\syncnst{D}\add\syncnst{C}\ass\\
 (S\circumtwo(\VP\circum N))\bsl(S\circumtwo(\VP\circum N))\ass\\
 \lambda y\lambda z[(y\ z)\wedge((z\ \semcnst{s})\ \semcnst{d})]
 \end{array}
 \using /E
 \endprooftree
\justifies
\begin{array}{c}
\syncnst{J}\add\sep\add\syncnst{W}\add\sep\add\syncnst{and}\add\syncnst{D}\add\syncnst{C}\ass\\
 S\circumtwo(\VP\circum N)\ass\\
 \lambda z[((z\ \semcnst{w})\ \semcnst{j})\wedge((z\ \semcnst{c})\ \semcnst{d})]
 \end{array}
 \using \bsl E
 \endprooftree
 \prooftree
 \prooftree
 \prooftree
 \begin{array}{c}
 \syncnst{wants}\ass\\
 (\VP/\VP)/N\ass\\
 \semcnst{want}
 \end{array}
 \prooftree
\justifies
\begin{array}{c}
 a\ass\\
 N\ass\\
 x
\end{array}
 \using i
 \endprooftree
 \justifies
 \begin{array}{c}
 \syncnst{wants}\add a\ass\\
 \VP/\VP\ass\\
 (\semcnst{want}\ x)
 \end{array}
 \using /E
 \endprooftree
 \prooftree
 \mbox{to win}
 \justifies
 \begin{array}{c}
\syncnst{to}\add\syncnst{win}\ass\\
\VP\ass\\
\semcnst{win}
\end{array}
 \endprooftree
 \justifies
 \begin{array}{c}
 \syncnst{wants}\add a\add\syncnst{to}\add\syncnst{win}\ass\\
 \VP\ass\\
 ((\semcnst{want}\ x)\ \semcnst{win})
 \end{array}
 \using /E
 \endprooftree
\justifies
\begin{array}{c}
 \syncnst{wants}\add \sep\add\syncnst{to}\add\syncnst{win}\ass\\
 \VP\circum N\ass\\
 \lambda x((\semcnst{want}\ x)\ \semcnst{win})
 \end{array}
 \using \circum I^i
 \endprooftree
 \justifies
\syncnst{J}\add\syncnst{wants}\add\syncnst{W}\add\syncnst{to}\add\syncnst{win}\add\syncnst{and}\add\syncnst{D}\add\syncnst{C}\ass
 S\ass
 {}[(((\semcnst{want}\ \semcnst{w})\ \semcnst{win})\ \semcnst{j})\wedge(((\semcnst{want}\ \semcnst{c})\ \semcnst{win})\ \semcnst{d})]
\using \circumtwo E
\endprooftree 
 }
 
 \ 
 
\noindent Observe how in the last step of both of the above derivations adjunction combines a string
 of sort $2$ and a string of sort $1$. But, in the first, simplex, case the separator of the second operand
 is right peripheral, whereas in the second, complex, case the separator of the second operand is medial.
 This is how the account unifies simplex and complex gapping under a single coordinator type.
 
 The account of determiner gapping consists in an assignment to the
coordinator of the following type ($Q$ is $((S\circum N)\infix S)/\CN$):
\disp{\begin{tabular}[t]{l}
$\syncnst{and}\ass (X\bsl X)/((X\wprodtwo J)\wprod I)\ass \lambda x\lambda y\lambda z\lambda w[((y\ z)\ w)\wedge((\pi_1\pi_1 x\ z)\ w)]$\\
where $X= (S\circumtwo (\VP\circum N))\circum Q$
\end{tabular}
\label{detgappingtype}}
Consider example (\ref{detgappingex}a) of simplex determiner gapping:
\lingform{Some dogs like Whiskas and} \lingform{cats Alpo}.
We use the following auxiliary derivation showing that a common noun followed
by an object has type $(S\circumtwo (\VP\circum N))\circum Q$:
\disp{
 \prooftree
 \prooftree
 \prooftree
\prooftree
 \prooftree
 \prooftree
\justifies 
 a\ass N\ass x
 \using i
 \endprooftree
 \prooftree
 \prooftree
 \justifies
 b_0\add\sep\add b_1\ass\VP\circum N\ass y
 \using j
 \endprooftree
 \syncnst{obj}\ass N\ass\semcnst{obj}
 \justifies
 b_0\add\syncnst{obj}\add b_1\ass\VP\ass(y\ \semcnst{obj})
 \using \circum E
 \endprooftree
 \justifies
 a\add b_0\add\syncnst{obj}\add b_1\ass S\ass((y\ \semcnst{obj})\ x)
 \using \bsl E
 \endprooftree 
 \justifies
 \sep\add b_0\add\syncnst{obj}\add b_1\ass S\circum N\ass\lambda x((y\ \semcnst{obj})\ x)
 \using \circum I^i
 \endprooftree 
  \prooftree
 \prooftree
\justifies
 c\ass Q\ass z
\using k
\endprooftree
\syncnst{cn}\ass\CN\ass\semcnst{cn}
\justifies 
c\add\syncnst{cn}\ass (S\circum N)\infix S\ass(z\ \semcnst{cn})
\using /E
\endprooftree
\justifies
c\add\syncnst{cn}\add b_0\add\syncnst{obj}\add b_1\ass S\ass((z\ \semcnst{cn})\ \lambda x((y\ \semcnst{obj})\ x))
\using \infix E
\endprooftree
\justifies
c\add\syncnst{cn}\add \sep\add\syncnst{obj}\add\sep\ass S\circumtwo(\VP\circum N)\ass\lambda y((z\ \semcnst{cn})\ \lambda x((y\ \semcnst{obj})\ x))
\using \circumtwo I^j 
\endprooftree
\justifies
\sep\add\syncnst{cn}\add \sep\add\syncnst{obj}\add\sep\ass (S\circumtwo(\VP\circum N))\circum Q\ass\lambda z\lambda y((z\ \semcnst{cn})\ \lambda x((y\ \semcnst{obj})\ x))
\using \circum I^k
\endprooftree
}
The derivation of (\ref{detgappingex}a) is given in Figure~\ref{detgappingfig}.
\begin{figure}
\begin{center}
\rotatebox{-90}{\scriptsize
\prooftree
\prooftree
\prooftree
 \prooftree
 \syncnst{dogs}\ass\CN\ass\semcnst{dogs}\tb
 \syncnst{W}\ass N\ass\semcnst{w}
 \justifies
 \begin{array}{c}
 \sep\add\syncnst{dogs}\add \sep\add\syncnst{W}\add\sep\ass\\
 (S\circumtwo(\VP\circum N))\circum Q\ass\\
 \lambda z\lambda y((z\ \semcnst{dogs})\ \lambda x((y\ \semcnst{w})\ x))
 \end{array}
\endprooftree
\prooftree
 \begin{array}[b]{c}
 \syncnst{and}\ass\\
 (((S\circumtwo(\VP\circum N))\circum Q)\bsl((S\circumtwo(\VP\circum N))\circum Q))/\\
 /((((S\circumtwo(\VP\circum N))\circum Q)\wprod I)\wprodtwo J)\ass\\
 \lambda x\lambda y\lambda z\lambda w[((y\ z)\ w)\wedge((\pi_1\pi_1 x\ z)\ w)]
 \end{array}
 \prooftree
 \prooftree
 \prooftree
 \syncnst{cats}\ass\CN\ass\semcnst{cats}\tab\tab
 \syncnst{A}\ass N\ass\semcnst{a}
 \justifies
 \sep\add\syncnst{cats}\add \sep\add\syncnst{A}\add\sep\ass (S\circumtwo(\VP\circum N))\circum Q\ass\lambda z\lambda y((z\ \semcnst{cats})\ \lambda x((y\ \semcnst{a})\ x))
\endprooftree
\prooftree
\justifies
\nil\ass I\ass\zero
\using II
\endprooftree
\justifies
\syncnst{cats}\add \sep\add\syncnst{A}\add\one\ass ((S\circum(\VP/N))\circum Q)\wprod I\ass(\lambda z\lambda y((z\ \semcnst{cats})\ \lambda x((y\ \semcnst{a})\ x)), \zero)
\using \wprod I
\endprooftree
\prooftree
\justifies
\one\ass J\ass\zero
\using JI
\endprooftree
\justifies
\syncnst{cats}\add\syncnst{A}\ass (((S\circum(\VP/N))\circum Q)\wprod I)\wprodtwo J\ass((\lambda z\lambda y((z\ \semcnst{cats})\ \lambda x((y\ \semcnst{a})\ x)), \zero), \zero)
\using \wprodtwo I
\endprooftree
 \justifies
 \syncnst{and}\add\syncnst{cats}\add\syncnst{A}\ass
 ((S\circumtwo(\VP\circum N))\circum Q)\bsl((S\circumtwo(\VP\circum N))\circum Q)
   \ass\lambda y\lambda z\lambda w[((y\ z)\ w)\wedge((z\ \semcnst{cats})\ \lambda x((w\ \semcnst{a})\ x)]
 \using /E
 \endprooftree
 \justifies
 \sep\add\syncnst{dogs}\add \sep\add\syncnst{W}\add\one\add\syncnst{and}\add\syncnst{cats}\add\syncnst{A}\ass
 (S\circumtwo(\VP\circum N))\circum Q
   \ass\lambda z\lambda w[((z\ \semcnst{dogs})\ \lambda x((w\ \semcnst{w})\ x))\wedge((z\ \semcnst{cats})\ \lambda x((w\ \semcnst{a})\ x)]
 \using \bsl E
 \endprooftree
 \syncnst{some}\ass Q\ass\exists
\justifies
\begin{array}{c}
\syncnst{some}\add\syncnst{dogs}\add \sep\add\syncnst{W}\add\one\add\syncnst{and}\add\syncnst{cats}\add\syncnst{A}\ass\\
S\circumtwo(\VP\circum N)\ass\\
\lambda w[((\exists\ \semcnst{dogs})\ \lambda x((w\ \semcnst{w})\ x))\wedge((\exists\ \semcnst{cats})\ \lambda x((w\ \semcnst{a})\ x)]
\end{array}
 \using \circum E
 \endprooftree
 \prooftree
\prooftree
 \syncnst{like}\ass\VP/N\ass\semcnst{like}
 \prooftree
\justifies
 a\ass N\ass x
 \using i
 \endprooftree
 \justifies
 \syncnst{like}\add a\ass\VP\ass(\semcnst{like}\ x)
 \using /E
 \endprooftree
\justifies
 \syncnst{like}\add \sep\ass\VP\circum N\ass\lambda x(\semcnst{like}\ x)
 \using \circum I^i
 \endprooftree
 \justifies
 \syncnst{some}\add\syncnst{dogs}\add\syncnst{like}\add\syncnst{W}\add\syncnst{and}\add\syncnst{cats}\add\syncnst{A}\ass
 S
   \ass[((\exists\ \semcnst{dogs})\ \lambda x((\semcnst{like}\ \semcnst{w})\ x))\wedge((\exists\ \semcnst{cats})\ \lambda x((\semcnst{like}\ \semcnst{a})\ x)]
 \using \circumtwo E
 \endprooftree}
 \end{center}
 \caption{Determiner gapping}
 \label{detgappingfig}
 \end{figure}
 Finally, from the same coordinator type assignment (\ref{detgappingtype})
 we can derive the case of discontinuous determiner gapping (\ref{detgappingex}b)
 $$\lingform{Every cook wants Bar\c{c}a} \lingform{to win and waiter Madrid}$$
 in Figure~\ref{comdetgappingfig}.

\begin{figure}
\begin{center}
\rotatebox{-90}{\scriptsize
\prooftree
\prooftree
\prooftree
 \prooftree
 \syncnst{cook}\ass\CN\ass\semcnst{cook}\tb
 \syncnst{B}\ass N\ass\semcnst{b}
 \justifies
 \begin{array}{c}
 \sep\add\syncnst{cook}\add \sep\add\syncnst{B}\add\sep\ass\\
 (S\circumtwo(\VP\circum N))\circum Q\ass\\
 \lambda z\lambda y((z\ \semcnst{cook})\ \lambda x((y\ \semcnst{b})\ x))
 \end{array}
\endprooftree
\prooftree
 \begin{array}[b]{c}
 \syncnst{and}\ass\\
 (((S\circumtwo(\VP\circum N))\circum Q)\bsl((S\circumtwo(\VP\circum N))\circum Q))/\\
 /((((S\circumtwo(\VP\circum N))\circum Q)\wprod I)\wprodtwo J)\ass\\
 \lambda x\lambda y\lambda z\lambda w[((y\ z)\ w)\wedge((\pi_1\pi_1 x\ z)\ w)]
 \end{array}
 \prooftree
 \prooftree
 \prooftree
 \syncnst{waiter}\ass\CN\ass\semcnst{waiter}\tb
 \syncnst{M}\ass N\ass\semcnst{m}
 \justifies
\begin{array}{c}
\sep\add\syncnst{waiter}\add \sep\add\syncnst{M}\add\sep\ass\\
(S\circumtwo(\VP\circum N))\circum Q\ass\\
\lambda z\lambda y((z\ \semcnst{waiter})\ \lambda x((y\ \semcnst{m})\ x))
\end{array}
\endprooftree
\prooftree
\justifies
\nil\ass I\ass\zero
\using II
\endprooftree
\justifies
\begin{array}{c}
\syncnst{waiter}\add \sep\add\syncnst{M}\add\one\ass\\
((S\circum(\VP/N))\circum Q)\wprod I\ass\\
(\lambda z\lambda y((z\ \semcnst{waiter})\ \lambda x((y\ \semcnst{m})\ x)), \zero)
\end{array}
\using \wprod I
\endprooftree
\prooftree
\justifies
\one\ass J\ass\zero
\using JI
\endprooftree
\justifies
\begin{array}{c}
\syncnst{waiter}\add\syncnst{M}\ass\\
(((S\circum(\VP/N))\circum Q)\wprod I)\wprodtwo J\ass\\
((\lambda z\lambda y((z\ \semcnst{waiter})\ \lambda x((y\ \semcnst{m})\ x)), \zero), \zero)
\end{array}
\using \wprodtwo I
\endprooftree
 \justifies
 \syncnst{and}\add\syncnst{waiter}\add\syncnst{M}\ass
 ((S\circumtwo(\VP\circum N))\circum Q)\bsl((S\circumtwo(\VP\circum N))\circum Q)
   \ass\lambda y\lambda z\lambda w[((y\ z)\ w)\wedge((z\ \semcnst{waiter})\ \lambda x((w\ \semcnst{m})\ x)]
 \using /E
 \endprooftree
 \justifies
 \sep\add\syncnst{cook}\add \sep\add\syncnst{B}\add\one\add\syncnst{and}\add\syncnst{waiter}\add\syncnst{M}\ass
 (S\circumtwo(\VP\circum N))\circum Q
   \ass\lambda z\lambda w[((z\ \semcnst{cook})\ \lambda x((w\ \semcnst{b})\ x))\wedge((z\ \semcnst{waiter})\ \lambda x((w\ \semcnst{m})\ x)]
 \using \bsl E
 \endprooftree
 \syncnst{every}\ass Q\ass\forall
\justifies
 \syncnst{every}\add\syncnst{cook}\add \sep\add\syncnst{B}\add\one\add\syncnst{and}\add\syncnst{waiter}\add\syncnst{M}\ass
 S\circumtwo(\VP\circum N)
   \ass\lambda w[((\forall\ \semcnst{cook})\ \lambda x((w\ \semcnst{b})\ x))\wedge((\forall\ \semcnst{waiter})\ \lambda x((w\ \semcnst{m})\ x)]
 \using \circum E
 \endprooftree
 \prooftree
 \prooftree
 \prooftree
 \begin{array}{c}
 \syncnst{wants}\ass\\
 (\VP/\VP)/N\ass\\
 \semcnst{want}
 \end{array}
 \tb
 \prooftree
\justifies
 a\ass N\ass x
 \using i
 \endprooftree
 \justifies
 \syncnst{wants}\add a\ass\VP/\VP\ass(\semcnst{want}\ x)
 \using /E
 \endprooftree
 \prooftree
 \mbox{to win}
 \justifies
\syncnst{to}\add\syncnst{win}\ass\VP\ass\semcnst{win}
 \endprooftree
 \justifies
 \syncnst{wants}\add a\add\syncnst{to}\add\syncnst{win}\ass\VP\ass((\semcnst{want}\ x)\ \semcnst{win})
 \using /E
 \endprooftree
\justifies
 \syncnst{wants}\add \sep\add\syncnst{to}\add\syncnst{win}\ass\VP\circum N\ass\lambda x((\semcnst{want}\ x)\ \semcnst{win})
 \using \circum I^i
 \endprooftree
 \justifies
 \syncnst{every}\add\syncnst{cook}\add\syncnst{wants}\add\syncnst{B}\add\syncnst{to}\add\syncnst{win}\add\syncnst{and}\add\syncnst{waiter}\add\syncnst{M}\ass
 S
   \ass[((\forall\ \semcnst{cook})\ \lambda x(((\semcnst{want}\ \semcnst{b})\ \semcnst{win})\ x))\wedge((\forall\ \semcnst{waiter})\ \lambda x(((\semcnst{want}\ \semcnst{m})\ \semcnst{win})\ x)]
 \using \circumtwo E
 \endprooftree}
 \end{center}
 \caption{Discontinuous determiner gapping}
 \label{comdetgappingfig}
 \end{figure}
 
 }
 
\section{Conclusion}

The categorial analysis of gapping as like-type coordination was
established in Steedman (1990\cite{steed:gapping90}) and
Hendriks (1995\cite{hendriksP:phd}). In the framework of HTLG
Kubota and Levine (2012\cite{htlggapping}) go further in that they provide
like-type coordination for discontinuous gapping.
Our analysis is inspired by that of
Kubota and Levine (2012\cite{htlggapping}; 2013\cite{kublev:detgap}).
However, our analysis of gapping represents an improvement
on the HTLG analysis because we do not require two types for
simple and discontinuous gapping: a single type suffices.

Finally, we have noted that the HTLG account of gapping suffers
from determiner-transitive verb
order inconsistency overgeneration; 
the same problem would arise for HTLG in relation to discontinuous determiner gapping:
\disp{\begin{tabular}[t]{ll}
a. & \unacc Some boy wants Everton to win and Mary (wants) (some) London club (to win)\\
 b. & \unacc Mary wants some London club to win and (some) boy (wants)  
 Everton (to win)
 \end{tabular}}
 In addition to capturing simplex gapping (and determiner gapping)
 as a special case of complex gapping (and determiner gapping),
 our DTLG account does not have the order inconsistency overgeneration problem
 of determiner gapping and discontinuous determiner gapping.

\part{CONCLUSION}

\chapter{Going further}

\label{concchap}

We give a sketch of second-order and higher-order
displacement calculus.

\section{Gapping}

In a series of papers Kubota and Levine give an account of gapping and determiner
gapping in terms of hybrid type logical grammar, including anomalous
scopal interactions with auxiliaries and negative quantifiers.\footnote{The contents
of this section are to appear in {\it Natural Language and Linguistic Theory}.} We make three
observations:
i) under the counterpart assumptions that Kubota and Levine make, the 
existent displacement type logical grammar account of gapping already accounts
for the scopal interactions, 
ii) Kubota and Levine overgenerate determiner-verb order inconsistencies in
determiner gapping conjuncts whereas the immediate adaptation of their
proposal to displacement type logical grammar does not do so, and
iii) Kubota and Levine do not capture simplex gapping
as a special case of complex gapping, but require
distinct lexical entries for the two cases; we show how a generalisation of displacement type logical grammar
allows both simplex and discontinuous gapping under a single type assignment.

Consider the following four types of gapping:
simple gapping, discontinuous gapping,
determiner gapping, and discontinuous determiner gapping.
\disp{\begin{tabular}[t]{ll}
a. & Leslie met Sandy and Robin (met) Bill.\\
b. & John wants Watford to win and Daniel (wants) Chelsea (to win).
\label{gappingex}
\end{tabular}}

\disp{\begin{tabular}[t]{ll}
a. & Some dogs like Whiskas and (some) cats (like) Alpo.\\
b. & Every cook wants Bar\c{c}a to win and (every) waiter (wants)\\& Madrid (to win).
\label{detgappingex}
\end{tabular}}
Kubota and Levine (2012\cite{htlggapping}; 2013\cite{kublev:detgap}; 2015\cite{htlgnllt:gapping}; henceforth K\&L)
develop an account of gapping in hybrid type logical grammar (HTLG),
an extension of Lambek calculus admitting functional expressions in the phonological component.\footnote{
An anonymous NLLT referee questioned whether gapping is after all a purely combinatoric phenomenon citing split
antecedent gapping (i), and non-ATB gapping (ii):

\ 

\noindent
i) Sue goes running $6$ times a week, and Alex lifts weights $3$ times a week, but neither every day. \\
ii) Either Pat came with Chris and Sandy came with Kim, or Pat with Kim and the others were alone.

\

We cannot enter fully into this question here except to note that such examples do not show that there is
\emph{no\/} combinatoric component to gapping, but rather that it is a more generalised phenomenon in the case of iterated coordination, which we do not address here. 
}
K\&L (2015\cite{htlgnllt:gapping}) provide a review of the literature
and argue in broad terms the advantages of an analysis of gapping
as hypothetical reasoning,
to which we have nothing to add; we in turn review their type logical proposal.
The Lambek rules of HTLG are as follows:\footnote{We
make some notational adjustments in order to smoothen comparison of hybrid
TLG and displacement TLG.}
\disp{
\prooftree
\prooftree
\leadsto
\alpha\ass C/B\ass \phi
\endprooftree
\prooftree
\leadsto
\beta\ass B\ass \psi
\endprooftree
\justifies
\alpha\add\beta\ass C\ass (\phi\ \psi)
\using
/E
\endprooftree
\tb
\prooftree
\prooftree
\leadsto
\alpha\ass A\ass \phi
\endprooftree
\prooftree
\leadsto
\beta\ass A\bsl C\ass \psi
\endprooftree
\justifies
\alpha\add\beta\ass C\ass (\psi\ \phi)
\using
\bsl E
\endprooftree
\vtab
\prooftree
\prooftree
\prooftree
\justifies
{b}\ass B\ass y
\using n
\endprooftree
\leadsto
\alpha\add{b}\ass C\ass \chi
\endprooftree
\justifies
\alpha\ass C/B\ass \lambda y\chi
\using /I^n
\endprooftree
\tb
\prooftree
\prooftree
\prooftree
\justifies
{a}\ass A\ass x
\using n
\endprooftree
\leadsto
{a}\add\beta\ass C\ass \chi
\endprooftree
\justifies
\beta\ass A\bsl C\ass \lambda x\chi
\using \bsl I^n
\endprooftree
}
In these rules $\alpha\ass A\ass \phi$ signifies a sign with phonology/prosodics $\alpha$,
syntactic type $A$, and semantics $\phi$. The elimination $(E)$ rules combine signs by 
concatenation prosodically, \emph{modus ponens\/} type-logically, and functional application semantically;
the introduction $(I)$ rules of hypothetical reasoning conditionalise assumptions: when passing from premise to
conclusion the assumption coindexed with the rule name $n$ becomes closed. Every rule instance has a fresh index. 
The introduction rules are semantically interpreted
by functional abstraction. To these directional rules HTLG adds inference rules for a nondirectional type constructor $|$
interpreted by phonological functional application and phonological functional abstraction:
\disp{
\prooftree
\prooftree
\leadsto
\alpha\ass C|B\ass \phi
\endprooftree
\prooftree
\leadsto
\beta\ass B\ass \psi
\endprooftree
\justifies
\alpha(\beta)\ass C\ass (\phi\ \psi)
\using
|E
\endprooftree
\tb
\prooftree
\prooftree
\prooftree
\justifies
\phi\ass B\ass y
\using n
\endprooftree
\leadsto
\alpha\ass C\ass \chi
\endprooftree
\justifies
\lambda \phi.\alpha\ass C|B\ass \lambda y\chi
\using |I^n
\endprooftree
}
These are the characteristic rules of HTLG (for more details we refer the reader to the papers of Kubota and Levine). 

In relation to gapping, K\&L (2015\cite{htlgnllt:gapping})~(53) present a type assignment which is in essence:\footnote{Throughout, $\VP$ abbreviates $N\bsl S$. We limit attention to gapping of the transitive
verb category; so far as we are aware gapping in other categories raises no new issues
differentiating between hybrid TLG and displacement TLG.} 
\disp{\begin{tabular}[t]{l}$
\lambda\sigma_2\lambda\sigma_1\lambda\phi.\sigma_1(\phi)\add\syncnst{and}\add
\sigma_2(\nil)\ass (X|X)|X\ass \lambda x\lambda y\lambda z[(y\ z)\wedge(x\ z)]$\\
where $X=S|(\VP/N)$
\end{tabular}}
And they derive from this
simple gapping such as (\ref{gappingex}a);
see {\it ibid}~(52) and~(55).
In our notation, their~(52) is:
\disp{
\prooftree
\prooftree
\syncnst{robin}\ass N\ass \semcnst{r}
\prooftree
\prooftree
\justifies
\phi\ass \VP/N\ass x
\using i
\endprooftree
\syncnst{bill}\ass N\ass \semcnst{b}
\justifies
\phi\add\syncnst{bill}\ass\VP\ass(x\ \semcnst{b})
\using E/
\endprooftree
\justifies
\syncnst{robin}\add\phi\add\syncnst{bill}\ass S\ass((x\ \semcnst{b})\ \semcnst{r})
\using E\bsl
\endprooftree
\justifies
\lambda\phi.\syncnst{robin}\add\phi\add\syncnst{bill}\ass S|(\VP/N)\ass\lambda x((x\ \semcnst{b})\ \semcnst{r})
\using I|^i
\endprooftree
}
Continuing the derivation as in their~(55) yields:
\disp{$\syncnst{leslie}+\syncnst{met}+\syncnst{sandy}+\syncnst{and}+\syncnst{robin}+\syncnst{bill}\ass
S\ass((\semcnst{met\ s})\ \semcnst{l})\wedge((\semcnst{met\ b})\ \semcnst{r})$}
K\&L also assume in their~(56) a raised type for an auxiliary:
\disp{$\lambda\sigma.\sigma(\syncnst{must})\ass S|(S|(VP/VP))\ass \lambda x(\semcnst{Nec}\ (x\ \lambda yy))$}
They show that such assignments to auxiliaries license the auxiliary wide-scope
reading of, say, \lingform{John must}
\lingform{eat steak and Mary pizza}.

Likewise, K\&L (2015\cite{htlgnllt:gapping})~(66) present a type assignment
for determiner gapping which is:
\disp{\begin{tabular}[t]{l}
$\lambda\rho_2\lambda\rho_1\lambda\phi\lambda\sigma.\rho_1(\phi)(\sigma)\add\syncnst{and}\add
\rho_2(\nil)(\lambda\chi\lambda\psi.\psi(\chi))\ass (X|X)|X\ass$\\$ \lambda x\lambda y\lambda z\lambda w[((y\ z)\ w)\wedge((x\ z)\ w)]$\\
where $X=(S|((S|(S|N))|\CN))|(\VP/N)$
\end{tabular}
}
And they assume, K\&L (2015\cite{htlgnllt:gapping})~(65),
 a raised type for a negative determiner of the form:
\disp{$
\lambda\rho.\rho(\lambda\phi\lambda\sigma.\sigma(\syncnst{no}\add\phi))\ass
S|((S|(S|(S|N)))|\CN)\ass
\lambda x(\neg(x\ \lambda y\lambda z\exists w[(y\ w)\wedge(z\ w)]))$}
These enable derivation of the split scope reading of:
\disp{
No dog eats whiskas or cat alpo.
}

\section{The scopal anomalies are already available in displacement TLG}

K\&L (2015\cite{htlgnllt:gapping}) state that `our own analysis
resembles most closely Morrill et al.'s (2011) (which is a refinement of
Hendriks 1995)'. The displacement type logical grammar (DTLG)
gapping assignment is as follows, where $\wprod$ is discontinuous product, which is semantically
interpreted by ordered pairing, and `$\pi_1 x$' selects the first component of $x$:\footnote{The notation
$\pi_1$ (and $\pi_2$) represents first (and second) projection of an ordered pair,
so that e.g.\ $\pi_1(\phi, \psi) = \phi$.}
 \disp{\begin{tabular}[t]{l}
$\syncnst{and}\ass (X\bsl X)/(X\wprod I)\ass \lambda x\lambda y\lambda z[(y\ z)\wedge(\pi_1x\ z)]$\\
where $X=S\circum(\VP/N)$
\end{tabular}}
Here, the symbols such as $I$ and $\circum{}$ are connectives of the displacement calculus, which we define
later.
The corresponding determiner gapping assignment would be:
\disp{\begin{tabular}[t]{l}$
\syncnst{and}\ass (X\bsl X)/((X\wprod I)\wprod I)\ass\lambda x\lambda y\lambda z\lambda w[((y\ z)\ w)\wedge((\pi_1\pi_1x\ z)\ w)]$\\
where $X=(S \circum(\VP/N))\circum(((S\circum N)\infix S)/\CN)$
\end{tabular}}

The auxiliary and negative determiner assignments would be:
\disp{$\syncnst{must}\ass (S\circum(VP/VP))\infix S\ass \lambda x(\semcnst{Nec}\ (x\ \lambda yy))$}
\disp{$
\syncnst{no}\ass
(S\circum(((S\circum N)\infix S)/\CN))\infix S\ass
\lambda x(\neg(x\ \lambda y\lambda z\exists w[(y\ w)\wedge(z\ w)]))$}
Thus the analyses of gapping scope anomalies are already available in
DTLG. Indeed K\&L (2015\cite{htlgnllt:gapping}) state, 'To be fair,
the core of our empirical results, so far as we can tell, seems to straightforwardly
carry over to Morrill et al.'s (2011) system.'
So K\&L's type logical contribution is the determiner gapping,
and the observation that raised auxiliary and negative
determiner assignments capture the scopal anomalies in HTLG,
but also in DTLG. 
Thus, although K\&L couch their solution in terms of HTLG,
HTLG and DTLG are on a par in respect of gapping and the scopal anomalies.
Since this point appears to have been granted we do not elaborate on it further.
But we will see a respect in which DTLG improves on HTLG,
and another respect in which DTLG can be made to improve further on HTLG.

\section{The HTLG determiner gapping assignment overgenerates}

Kubota (p.c.) points out that the HTLG analysis of
determiner gapping incorrectly predicts examples such as the following,
where the determiner and the transitive verb orders are not
consistent in the conjuncts:
\disp{\begin{tabular}[t]{ll}
a. & \unacc Some dogs like Whiskas and I (like) (some) cats.\\
b. &\unacc  I like some cats and (some) dogs (like) Whiskas.
\end{tabular}\label{perm}}
This order inconsistency overgeneration
arises because the $|((S|(S|N))|\CN)$ and $|(\VP/N)$ arguments in the types for
determiner gapping do not distinguish the linear order of the phonological
variables they bind. Thus, for example, in relation to (\ref{perm}a), both 
$\lambda v\lambda q.q\add\syncnst{dogs}\add v\add\syncnst{Whiskas}$
and $\lambda v\lambda q.\syncnst{I}\add v\add q\add\syncnst{cats}$
have type $(S|((S|(S|N))|\CN))|(\VP/N)$.
Note that the issue has nothing to do with the order in which the quantifier
and transitive verb arguments are abstracted,
but rather with the left to right position of the variables they bind in the body of the phonological term,
to which the type constructor $|$ is insensitive.
DTLG does not overgenerate in this way because the positions of
discontinuity are indexed for left to right order.

\subsection{HTLG requires distinct type assignments for simplex and discontinuous gapping}

Our analysis is inspired by that of
Kubota and Levine, 
which assigns coordinator conjunct types and prosody as follows:

\noindent
{\footnotesize
\begin{tabular}[t]{c|c|c}
 &&\\
conjunt type  & gapping & determiner gapping\\
\& &&\\
coordinator &&\\
phonology
  &&\\
\hline
 &&\\
simplex & $S|(\VP/N)$ & $(S|(\VP/N))|((S|(S|N))|\CN)$\\
 & $\lambda\sigma_2\lambda\sigma_1\lambda\phi.\sigma_1(\phi)\add\syncnst{and}\add
\sigma_2(\nil)$ & $\lambda\rho_2\lambda\rho_1\lambda\phi\lambda\sigma.\rho_1(\phi)(\sigma)\add\syncnst{and}\add
\rho_2(\lambda\chi\lambda\psi.\psi(\chi))(\nil)$\\
 &K\&L (2012\cite{htlggapping})~(5) K\&L (2013\cite{kublev:detgap})~(13)&
 K\&L (2013\cite{kublev:detgap})~(24) K\&L (2015\cite{htlgnllt:gapping})~(66)~(83)\\
& K\&L (2015\cite{htlgnllt:gapping})~(53)\\
 \hline
 &&\\
discont. & $S|(\VP|N)$ & $(S|(\VP|N))|((S|(S|N))|\CN)$\\
 & $\lambda\rho_2\lambda\rho_1\lambda\phi.\rho_1(\phi)\add\syncnst{and}\add\rho_2(\lambda\pi.\pi)$ & $\lambda\rho_2\lambda\rho_1\lambda\phi\lambda\sigma.\rho_1(\phi)(\sigma)\add\syncnst{and}\add
\rho_2(\lambda\chi\lambda\psi.\psi(\chi))(\lambda\pi.\pi)$\\
 & K\&L (2012\cite{htlggapping})~(20) & (by extrapolation)\\
 \end{tabular}}
 
\noindent
However, our analysis of gapping represents an improvement
on the HTLG analysis in that we require only a single type for
simple and discontinuous gapping.
HTLG requires two types because simplex gapping conditionalises $\VP/N$ of sort
\emph{string\/} whereas discontinuous gapping conditionalises $\VP|N$ of sort \emph{function},
and these require two distinct prosodic operations: application to the empty string
$\nil$ (of sort \emph{string}) and to the identity function $\lambda\pi.\pi$ (of sort \emph{function})
respectively.

\subsection{Our account}

We give an account of simple and discontinuous gapping as in~(\ref{gappingex})
and simplex and discontinuous determiner gapping as in~(\ref{detgappingex})
which is optimal in that a single coordinator type
generates discontinuous gapping with simple gapping as a special case
and a single coordinator type generates discontinuous determiner gapping with
simplex determiner gapping as a special case.
The account is expressed in an extension of displacement calculus
({\bf D}; Morrill et al.\ 2011\cite{mvf:tdc}) which we call second-order displacement calculus
{\bf D$^2$}.

\subsubsection{Second-order displacement calculus}

Displacement calculus is a logic of discontinuous strings.
By discontinuous strings we mean strings punctuated by a distinguished
vocabulary item `$\one$' called the \techterm{separator}.
In contrast to HTLG, 
the phonological terms of displacement calculus have no lambda abstraction
and,
instead of a set of variable placeholders,
DTLG has a single placeholder, the separator.
The sort of a discontinuous string is the number of separators
it contains.
We notate by $L_i, i\ge 0,$ the set of all strings of sort $i$ with respect to
some alphabet.
We consider the operations concatenation, intercalation, and
adjunction on discontinuous strings. Concatenation is represented in (\ref{concop}).

\disp{$
\begin{array}[t]{c}
\fbox{$\tb\alpha\tb$\strut}\ \add\ \fbox{$\tb\beta\tb$\strut}\\
=\\
\fbox{$\tb\alpha\tb$\strut}\fbox{$\tb\beta\tb$\strut}\\\\
\mbox{concatenation\ } +: L_i, L_j\rightarrow L_{i{+}j}
\end{array}
$\label{concop}}
For example, the concatenation of $\syncnst{Leslie}\add\one\add\syncnst{Sandy}$ and
$\syncnst{and}\add\syncnst{Robin}\add\syncnst{Bill}$ is:

\disp{$\begin{array}[t]{c}
\syncnst{Leslie}\add\one\add\syncnst{Sandy} + 
\syncnst{and}\add\syncnst{Robin}\add\syncnst{Bill}
\\ =
\\
\syncnst{Leslie}\add\one\add\syncnst{Sandy}\add\syncnst{and}\add\syncnst{Robin}\add\syncnst{Bill}
\end{array}
$}

Intercalation is represented in (\ref{intercop}):
\disp{$
\begin{array}[t]{c}
\fbox{$\tb\alpha\tb$\strut}\fbox{$1$\strut}\fbox{$\tb\gamma\tb$\strut}
\ \swrap{k}\ 
\fbox{$\tb\beta\tb$\strut}\\
=\\
\fbox{$\tb\alpha\tb$\strut}\fbox{$\tb\beta\tb$\strut}\fbox{$\tb\gamma\tb$\strut}\\\\
\mbox{intercalation\ } \swrap{k}: L_{i{+}1}, L_j\rightarrow L_{i{+}j}
\end{array}
$\label{intercop}}
For example, the intercalation at the second separator of $\one\add\syncnst{dogs}\add\one\add\syncnst{Whiskas}$ 
$\add\syncnst{and}\add\syncnst{cats}\add\syncnst{Alpo}$ and $\syncnst{like}$ is:

\disp{$\begin{array}[t]{c}
\one\add\syncnst{dogs}\add\one\add\syncnst{Whiskas}\add\syncnst{and}\add\syncnst{cats}\add\syncnst{Alpo}\ \swrap{2}\ \syncnst{like}
\\ =
\\
\sep\add\syncnst{dogs}\add\syncnst{like}\add\syncnst{Whiskas}\add\syncnst{and}\add\syncnst{cats}\add\syncnst{Alpo}
\end{array}
$}

Finally, adjunction is represented in (\ref{adjunctop}):
\disp{$
\begin{array}[t]{c}
\fbox{$\tb\alpha\tb$\strut}\fbox{$\one$\strut}\fbox{$\tb\gamma\tb$\strut}\fbox{$\one$\strut}\fbox{$\tb\epsilon\tb$\strut}
\ \swraptwo{k, l}\ 
\fbox{$\tb\beta\tb$\strut}\fbox{$1$\strut}\fbox{$\tb\delta\tb$\strut}\\
=\\
\fbox{$\tb\alpha\tb$\strut}\fbox{$\tb\beta\tb$\strut}\fbox{$\tb\gamma\tb$\strut}\fbox{$\tb\delta\tb$\strut}\fbox{$\tb\epsilon\tb$\strut}\\\\
\mbox{adjunction\ } \swraptwo{k, l}: L_{i{+}2}, L_{j{+}1}\rightarrow L_{i{+}j}
\end{array}
$\label{adjunctop}}
For example, the adjunction at the first separators of $\syncnst{John}\add\one\add\syncnst{Watford}\add\one\add\syncnst{and}\add$ $\syncnst{Daniel}\add$ $\syncnst{Chelsea}$ and $\syncnst{wants}\add\one\add\syncnst{to}\add\syncnst{win}$ is:

\disp{$\begin{array}[t]{c}
\syncnst{John}\add\one\add\syncnst{Watford}\add\one\add\syncnst{and}\add\syncnst{Daniel}\add\syncnst{Chelsea}\ \swraptwo{1, 1}\ \syncnst{wants}\add\one\add\syncnst{to}\add\syncnst{win}
\\ =
\\
\syncnst{John}\add\syncnst{wants}\add\syncnst{Watford}\add\syncnst{to}\add\syncnst{win}\add\syncnst{and}\add\syncnst{Daniel}\add\syncnst{Chelsea}\end{array}
$}

We will have three families of type-constructors defined in relation to the three prosodic operations of concatenation, intercalation, and adjunction. The syntactic types are sorted $\Tp_0, \Tp_1, \Tp_2, \ldots$
according to the number of points of discontinuity $0, 1, 2, \ldots$ their expressions contain.
The sets $\Tp_i$ of types of sort $i$ are defined by mutual recursion in terms of sets ${\cal P}_i$
of primitive types of sort $i$ as follows:\\

\noindent
{\small$
\begin{array}{rclrclll}
\Tp_i & ::= & {\cal P}_i\\\\
\Tp_i & ::= & \Tp_{i{+}j}/\Tp_j & T(C/B) & = & T(B){\rightarrow}T(C)
 & \mbox{over}\\
\Tp_j & ::= & \Tp_i\bsl\Tp_{i{+}j} & T(A\bsl C) & = & T(A){\rightarrow}T(C)
 & \mbox{under}\\
\Tp_{i{+}j} & ::= & \Tp_i\product\Tp_j & T(A\product B) & = & T(A)\&T(B) &
\mbox{continuous product}\\
\Tp_0 & ::= & I & T(I) & = & \top & \mbox{cont.\ unit}\\\\
\Tp_{i{+}1} & ::= & \Tp_{i{+}j}\scircum{k}\Tp_j, 1\le k\le i{+}1
& T(C\scircum{k} B) & = & T(B){\rightarrow}T(C) & \mbox{extract}\\
\Tp_j & ::= & \Tp_{i{+}1}\sinfix{k}\Tp_{i{+}j}, 1\le k\le i{+}1 
& T(A\sinfix{k} C) & = & T(A){\rightarrow}T(C) & \mbox{infix}\\
\Tp_{i{+}j} & ::= & \Tp_{i{+}1}\swprod{k}\Tp_j, 1\le k\le i{+}1 
& T(A\swprod{k} B) & = & T(A)\&T(B)& \mbox{disc.\ product}\\
\Tp_1 & ::= & J & T(J) & = & \top & \mbox{disc.\ unit}\\\\
\Tp_{i{+}2} & ::= & \Tp_{i{+}j}\scircumtwo{k, l}\Tp_{j{+}1}, 1\le k\le i{+}1, 1\le l\le j{+}1
& T(C\scircumtwo{k, l} B) & = & T(B){\rightarrow}T(C) & \mbox{2nd order extract}\\
\Tp_{j{+}1} & ::= & \Tp_{i{+}2}\sinfixtwo{k, l}\Tp_{i{+}j}, 1\le k\le i{+}1, 1\le l\le j{+}1 
& T(A\sinfixtwo{k, l} C) & = & T(A){\rightarrow}T(C) & \mbox{2nd order infix}\\
\Tp_{i{+}j} & ::= & \Tp_{i{+}2}\sdprodtwo{k, l}\Tp_{j{+}1}, 1\le k\le i{+}1, 1\le l\le j{+}1 
& T(A\swprodtwo{k, l} B) & = & T(A)\&T(B)& \mbox{2nd order disc.\ product}\\
\Tp_2 & ::= & K & T(K) & = & \top & \mbox{2nd order disc.\ unit}
\end{array}
$}\\

\noindent The second column of this table shows the standard categorial semantic type map 
for the connectives.\footnote{For example, if $T(N)=e$ where $e$ is the semantic type for individuals,
and $T(S)=t$ where $t$ is the semantic type corresponding to truth-values, we have then that 
$T(N\bsl S)=e\rightarrow t$.} Each type of sort $i$ is interpreted as a set of (discontinuous) strings
of sort $i$.
The prosodic interpretation is as follows:\\

$
{\small\begin{array}{rcl}
\oscott C/B\cscott & = & \{s_1|\ \forall s_2\in\oscott B\cscott, s_1\add s_2\in\oscott C\cscott\}\\
\oscott A\bsl C\cscott & = & \{s_2|\ \forall s_1\in\oscott A\cscott, s_1\add s_2\in\oscott C\cscott\}\\
\oscott A\product B\cscott & = & \{s_1\add s_2|\ s_1\in\oscott A\cscott\ \&\ s_2\in\oscott B\cscott\}\\
\oscott I\cscott & = & \{\nil\}
\\\\
\oscott C\scircum{k}B\cscott & = & \{s_1|\ \forall s_2\in\oscott B\cscott, s_1\swrap{k} s_2\in\oscott C\cscott\}\\
\oscott A\sinfix{k} C\cscott & = & \{s_2|\ \forall s_1\in\oscott A\cscott, s_1\swrap{k} s_2\in\oscott C\cscott\}\\
\oscott A\swprod{k} B\cscott & = & \{s_1\swrap{k} s_2|\ s_1\in\oscott A\cscott\ \&\ s_2\in\oscott B\cscott\}\\
\oscott J\cscott & = & \{\one\}
\\\\
\oscott C\scircumtwo{k, l}B\cscott & = & \{s_1|\ \forall s_2\in\oscott B\cscott, s_1\swraptwo{k, l} s_2\in\oscott C\cscott\}\\
\oscott A\sinfixtwo{k, l} C\cscott & = & \{s_2|\ \forall s_1\in\oscott A\cscott, s_1\swraptwo{k, l} s_2\in\oscott C\cscott\}\\
\oscott A\swprodtwo{k,l} B\cscott & = & \{s_1\swraptwo{k, l} s_2|\ s_1\in\oscott A\cscott\ \&\ s_2\in\oscott B\cscott\}\\
\oscott K\cscott & = & \{\one\add\one\add\one\}
\end{array}}
$\\

Although linguistically only some of the power, and hence only some of the rules, are necessary here,
the framework of DTLG complies with the modern logical paradigm of logic as an interpreted formal
language, and aspiration to \emph{soundness\/} (that everything said is true) and \emph{completeness\/}
(that everything true is said). This, to us, is the rock on which type \emph{logical\/} grammar is founded.
Thus, we present here all the rules so that the reader has the complete picture
of which the gapping analysis uses just a part.

The rules for second-order displacement calculus fall into three groups for the concatenative,
intercalative, and adjunctive connective families. Each family contains four connectives:
the two implicational residuals, the conjunctive product, and product unit. Each connective
has two rules, namely a rule of elimination (E), eliminating the connective reading from premise
to conclusion, and a rule of introduction (I) introducing the connective reading from premise to
conclusion.\footnote{Except we omit the elimination rules for product units which,
as well as being unmotivated linguistically, are awkward to formulate in the format used here.} Although there are many
rules, the reader should be aware of the high degree
of symmetry between them, and that the rules simply formalise the necessary and sufficient conditions
for membership of syntactic types: the rules are essentially the result of restating the interpretation clauses
given above.

The labelled natural deduction for second-order displacement calculus is as follows,
where we use three conventions. Firstly,
where $\syncnst{a}$ is a prosodic constant of sort $i$,
the \techterm{vector} $\vect{\syncnst{a}}$ is $\syncnst{a}_0\add\one\add\syncnst{a}_1\add\one\add\cdots\add\one\add\syncnst{a}_i$; for example,
if \syncnst{a} is of sort 1, $\vect{\syncnst{a}}=\syncnst{a}_0\add\one\add\syncnst{a}_1$. Secondly,
where $\alpha$ is a discontinuous string of sort $i>0$,
$\alpha \smwrap{k} \beta$, $1\le k\le i$, is the result of replacing the $k$th separator in $\alpha$
by $\beta$ (counting from the left);
for example
\syncnst{John}\add\one\add\syncnst{Watford}\add\one\add\syncnst{and}\add\syncnst{Daniel}\add\syncnst{Chelsea}\ \smwrap{2}\ \syncnst{to}\add\syncnst{win} = 
\syncnst{John}\add\one\add\syncnst{Watford}\add\syncnst{to}\add\syncnst{win}\add\syncnst{and}\add\syncnst{Daniel}\add\\
$\syncnst{Chelsea}$. Thirdly,
$\alpha \smwwrap{k} \beta$ abbreviates $\alpha \smwrap{k} \one\add\beta\add\one$;
for example
\syncnst{John}\add\one\add\syncnst{and}\add\syncnst{Daniel}\add\syncnst{Chelsea}\ \smwwrap{1}\
\syncnst{Watford} =
\syncnst{John}\add\one\add\syncnst{Watford}\add\one\add\syncnst{and}\add\syncnst{Daniel}\add\syncnst{Chelsea}.\\

Continuous family:
\begin{itemize}
\item
Elimination rules for
implications\\

\prooftree
\prooftree
\leadsto
\alpha\ass C/B\ass \phi
\endprooftree
\prooftree
\leadsto
\beta\ass B\ass \psi
\endprooftree
\justifies
\alpha\add\beta\ass C\ass (\phi\ \psi)
\using
/E
\endprooftree
\tab
\prooftree
\prooftree
\leadsto
\alpha\ass A\ass \phi
\endprooftree
\prooftree
\leadsto
\beta\ass A\bsl C\ass \psi
\endprooftree
\justifies
\alpha\add\beta\ass C\ass (\psi\ \phi)
\using
\bsl E
\endprooftree\\\ \\

\item
Elimination rule for product\\

\prooftree
\prooftree
\leadsto
\gamma\ass A\product B\ass\chi
\endprooftree
\prooftree
\prooftree
\justifies
\vect{a}\ass A\ass x
\using n
\endprooftree
\prooftree
\justifies
\vect{b}\ass B\ass y
\using n
\endprooftree
\leadsto
\delta(\vect{a}\add\vect{b})\ass D\ass \omega(x, y)
\endprooftree
\justifies
\delta(\gamma)\ass D\ass\omega(\pi_1\chi, \pi_2\chi)
\using \product E^n
\endprooftree\\\ \\

\item
Introduction rules for implications\\

\prooftree
\prooftree
\prooftree
\justifies
\vect{b}\ass B\ass y
\using n
\endprooftree
\leadsto
\alpha\add\vect{b}\ass C\ass \chi
\endprooftree
\justifies
\alpha\ass C/B\ass \lambda y\chi
\using /I^n
\endprooftree
\tab
\prooftree
\prooftree
\prooftree
\justifies
\vect{a}\ass A\ass x
\using n
\endprooftree
\leadsto
\vect{a}\add\beta\ass C\ass \chi
\endprooftree
\justifies
\beta\ass A\bsl C\ass \lambda x\chi
\using \bsl I^n
\endprooftree\\\ \\

\item
Introduction rules for product and product unit\\

\prooftree
\prooftree
\leadsto
\alpha\ass A\ass \phi
\endprooftree
\tb
\prooftree
\leadsto
\beta\ass B\ass \psi
\endprooftree
\justifies
\alpha\add\beta\ass A\product B\ass (\phi, \psi)
\using
\product I
\endprooftree
\tab
\prooftree
\justifies
\nil\ass I\ass\zero
\using II
\endprooftree\\
\end{itemize}

Discontinuous family:
\begin{itemize}
\item
Elimination rules for
implications\\

\prooftree
\prooftree
\leadsto
\alpha\ass C\scircum{k}B\ass \phi
\endprooftree
\prooftree
\leadsto
\beta\ass B\ass \psi
\endprooftree
\justifies
\alpha\smwrap{k}\beta\ass C\ass (\phi\ \psi)
\using
\circum E
\endprooftree
\tab
\prooftree
\prooftree
\leadsto
\alpha\ass A\ass \phi
\endprooftree
\prooftree
\leadsto
\beta\ass A\sinfix{k} C\ass \psi
\endprooftree
\justifies
\alpha\smwrap{k}\beta\ass C\ass (\psi\ \phi)
\using
\infix E
\endprooftree\\\ \\

\item 
Elimination rule for product\\

\prooftree
\prooftree
\leadsto
\gamma\ass A\wprod B\ass\chi
\endprooftree
\prooftree
\prooftree
\justifies
\vect{a}\ass A\ass x
\using n
\endprooftree
\prooftree
\justifies
\vect{b}\ass B\ass y
\using n
\endprooftree
\leadsto
\delta(\vect{a}\smwrap{k}\vect{b})\ass D\ass \omega(x, y)
\endprooftree
\justifies
\delta(\gamma)\ass D\ass\omega(\pi_1\chi, \pi_2\chi)
\using \wprod E^n
\endprooftree\\\ \\

\item
Introduction rules for implications\\

\prooftree
\prooftree
\prooftree
\justifies
\vect{b}\ass B\ass y
\using n
\endprooftree
\leadsto
\alpha\smwrap{k}\vect{b}\ass C\ass \chi
\endprooftree
\justifies
\alpha\ass C\scircum{k}B\ass \lambda y\chi
\using \circum I^n
\endprooftree
\tab
\prooftree
\prooftree
\prooftree
\justifies
\vect{a}\ass A\ass x
\using n
\endprooftree
\leadsto
\vect{a}\smwrap{k}\beta\ass C\ass \chi
\endprooftree
\justifies
\beta\ass A\sinfix{k} C\ass \lambda x\chi
\using \infix I^n
\endprooftree\\\ \\

\item
Introduction rules for product and product unit\\

\prooftree
\prooftree
\leadsto
\alpha\ass A\ass \phi
\endprooftree
\tb
\prooftree
\leadsto
\beta\ass B\ass \psi
\endprooftree
\justifies
\alpha\smwrap{k}\beta\ass A\swprod{k} B\ass (\phi, \psi)
\using
\product I
\endprooftree
\tab
\prooftree
\justifies
\one\ass J\ass\zero
\using IJ
\endprooftree
\end{itemize}

Second-order discontinuous family:
\begin{itemize}
\item
Elimination rules for
implications\\

\prooftree
\prooftree
\leadsto
\alpha\smwwrap{k}\gamma\ass C\scircumtwo{k, l}B\ass \phi
\endprooftree
\prooftree
\leadsto
\beta\ass B\ass \psi
\endprooftree
\justifies
\alpha\smwrap{k}(\beta\smwrap{l}\gamma)\ass C\ass (\phi\ \psi)
\using
\circumtwo E
\endprooftree
\tab
\prooftree
\prooftree
\leadsto
\alpha\smwwrap{k}\gamma\ass A\ass \phi
\endprooftree
\prooftree
\leadsto
\beta\ass A\sinfixtwo{k, l} C\ass \psi
\endprooftree
\justifies
\alpha\smwrap{k}(\beta\smwrap{l}\gamma)\ass C\ass (\psi\ \phi)
\using
\infixtwo E
\endprooftree\\\ \\

\item
Elimination rule for product\\

\prooftree
\prooftree
\leadsto
\gamma\ass A\swprodtwo{k, l} B\ass\chi
\endprooftree
\prooftree
\prooftree
\justifies
\vect{a}\smwwrap{k}\vect{c}\ass A\ass x
\using n
\endprooftree
\prooftree
\justifies
\vect{b}\ass B\ass y
\using n
\endprooftree
\leadsto
\delta(\vect{a}\smwrap{k}(\vect{b}\smwrap{l}\vect{c}))\ass D\ass \omega(x, y)
\endprooftree
\justifies
\delta(\gamma)\ass D\ass\omega(\pi_1\chi, \pi_2\chi)
\using \wprodtwo E^n
\endprooftree\\\ \\

\item
Introduction rules for implications\\

\prooftree
\prooftree
\prooftree
\justifies
\vect{b}\ass B\ass y
\using n
\endprooftree
\leadsto
\alpha\smwrap{k}(\vect{b}\smwrap{l}\gamma)\ass C\ass \chi
\endprooftree
\justifies
\alpha\smwwrap{k}\gamma\ass C\scircumtwo{k, l}B\ass \lambda y\chi
\using \circumtwo I^n
\endprooftree
\tab
\prooftree
\prooftree
\prooftree
\justifies
\vect{a}\smwrap{l}\vect{c}\ass A\ass x
\using n
\endprooftree
\leadsto
\vect{a}\smwrap{k}(\beta\smwrap{l}\vect{c})\ass C\ass \chi
\endprooftree
\justifies
\beta\ass A\sinfixtwo{k, l} C\ass \lambda x\chi
\using \infixtwo I^n
\endprooftree\\\ \\

\item
Introduction rules for product and product unit\\

\prooftree
\prooftree
\leadsto
\alpha\smwwrap{k}\gamma\ass A\ass \phi
\endprooftree
\tb
\prooftree
\leadsto
\beta\ass B\ass \psi
\endprooftree
\justifies
\alpha\smwrap{k}(\beta\smwrap{l}\gamma)\ass A\swprodtwo{k, l} B\ass (\phi, \psi)
\using
\product E
\endprooftree
\tab
\prooftree
\justifies
\one\add\one\add\one\ass K\ass\zero
\using IK
\endprooftree
\end{itemize}
\ \\

\noindent
We adopt the convention that when subscripts $k$ and $l$ are omitted they
are $1$, i.e.\ they default to $1$.

By way of example, the following auxiliary derivation shows that a subject followed
by an object has type $S\circumtwo(\VP\circum N)$:
\disp{
 \prooftree
 \prooftree
 \syncnst{sbj}\ass N\ass\semcnst{sbj}
 \prooftree
 \prooftree
 \justifies
 a_0\add\sep\add a_1\ass\VP\circum N\ass x
 \using i
 \endprooftree
 \syncnst{obj}\ass N\ass\semcnst{obj}
 \justifies
 a_0\add\syncnst{obj}\add a_1\ass\VP\ass(x\ \semcnst{obj})
 \using \circum E
 \endprooftree
 \justifies
 \syncnst{sbj}\add a_0\add\syncnst{obj}\add a_1\ass S\ass((x\ \semcnst{obj})\ \semcnst{sbj})
 \using \bsl E
 \endprooftree
 \justifies
 \syncnst{sbj}\add\sep\add\syncnst{obj}\add\sep\ass S\circumtwo(\VP\circum N)\ass\lambda x((x\ \semcnst{obj})\ \semcnst{sbj})
 \using \circumtwo I^i
 \endprooftree
 \label{subjobjder}
}

\subsubsection{Analyses}

The account of gapping consists in an assignment to the
coordinator of the following type:
\disp{\begin{tabular}[t]{l}
$\syncnst{and}\ass (X\bsl X)/(X\wprodtwo J)\ass \lambda x\lambda y\lambda z[(y\ z)\wedge(\pi_1 x\ z)]$\\
where $X= S\circumtwo(\VP\circum N)$
\end{tabular}\label{gappingtype}}
Consider example (\ref{gappingex}a) of simple gapping:
\lingform{Leslie met Sandy and Robin Bill}.
Then there is the following derivation of (\ref{gappingex}a) using (\ref{subjobjder}):

\noindent{\footnotesize
\prooftree
\prooftree
 \prooftree
 {\rm Leslie}\ {\rm Sandy}
 \justifies
 \begin{array}{c}
 \syncnst{L}\add\sep\add\syncnst{S}\add\sep\ass\\
 S\circumtwo(\VP\circum N)\ass\\
 \lambda x((x\ \semcnst{s})\ \semcnst{l})
 \end{array}
 \endprooftree 
 \prooftree
 \begin{array}[b]{c}
 \syncnst{and}\ass\\
 ((S\circumtwo(\VP\circum N))\bsl(S\circumtwo(\VP\circum N)))/\\
 /((S\circumtwo (\VP\circum N))\wprodtwo J)\ass\\
 \lambda x\lambda y\lambda z[(y\ z)\wedge(\pi_1 x\ z)]
 \end{array}
 \prooftree
 \prooftree
 {\rm Robin}\ {\rm Bill}
 \justifies
\begin{array}{c}
\syncnst{R}\add\sep\add\syncnst{B}\add\sep\ass\\
S\circumtwo(\VP\circum N)\ass\\
\lambda x((x\ \semcnst{b})\ \semcnst{r})
\end{array}
 \endprooftree
 \prooftree
 \justifies
 \sep\ass J\ass \zero
 \using JR
 \endprooftree
 \justifies
 \syncnst{R}\add\syncnst{B}\ass (S\circum(\VP/N))\wprodtwo J\ass (\lambda x((x\ \semcnst{b})\ \semcnst{r}), \zero)
 \using \wprodtwo I
 \endprooftree
\justifies
\syncnst{and}\add\syncnst{R}\add\syncnst{B}\ass
 (S\circumtwo(\VP\circum N))\bsl(S\circumtwo(\VP\circum N))\ass
 \lambda y\lambda z[(y\ z)\wedge((z\ \semcnst{b})\ \semcnst{r})]
 \using /E
 \endprooftree
\justifies
\syncnst{L}\add\sep\add\syncnst{S}\add\sep\add\syncnst{and}\add\syncnst{R}\add\syncnst{B}\ass
 S\circumtwo(\VP\circum N)\ass
 \lambda z[((z\ \semcnst{s})\ \semcnst{l})\wedge((z\ \semcnst{b})\ \semcnst{r})]
 \using \bsl E
 \endprooftree
 \prooftree
 \prooftree
 \syncnst{met}\ass\VP/N\ass\semcnst{meet}
 \prooftree
\justifies
 a\ass N\ass x
 \using i
 \endprooftree
 \justifies
 \syncnst{met}\add a\ass\VP\ass(\semcnst{meet}\ x)
 \using /E
 \endprooftree
\justifies
 \syncnst{met}\add \sep\ass\VP\circum N\ass\lambda x(\semcnst{meet}\ x)
 \using \circum I^i
 \endprooftree
 \justifies
\syncnst{L}\add\syncnst{met}\add\syncnst{S}\add\syncnst{and}\add\syncnst{R}\add\syncnst{B}\ass
 S\ass
 {}[((\semcnst{meet}\ \semcnst{s})\ \semcnst{l})\wedge((\semcnst{meet}\ \semcnst{b})\ \semcnst{r})]
\using \circumtwo E
\endprooftree 
 
 }
 
\ 

\noindent
And from the same coordinator type assignment (\ref{gappingtype})
 there is the following derivation of the discontinuous
 gapping (\ref{gappingex}b)
 \lingform{John wants Watford to win and} \lingform{Daniel Chelsea}:

\ 

\noindent
 {\footnotesize
\prooftree
\prooftree
 \prooftree
 {\rm John}\ {\rm Watford}
 \justifies
 \begin{array}{c}
 \syncnst{J}\add\sep\add\syncnst{W}\add\sep\ass\\
 S\circumtwo(\VP\circum N)\ass\\
 \lambda x((x\ \semcnst{w})\ \semcnst{j})
 \end{array}
 \endprooftree 
 \prooftree
 \begin{array}[b]{c}
 \syncnst{and}\ass\\
 ((S\circumtwo(\VP\circum N))\bsl(S\circumtwo(\VP\circum N)))/\\
 /((S\circumtwo (\VP\circum N))\wprodtwo J)\ass\\
 \lambda x\lambda y\lambda z[(y\ z)\wedge(\pi_1 x\ z)]
 \end{array}
 \prooftree
 \prooftree
 {\rm Daniel}\ {\rm Chelsea}
 \justifies
 \begin{array}{c}
 \syncnst{D}\add\sep\add\syncnst{S}\add\sep\ass\\
 S\circumtwo(\VP\circum N)\ass\\
 \lambda x((x\ \semcnst{s})\ \semcnst{d})
 \end{array}
 \endprooftree
 \prooftree
 \justifies
 \begin{array}{c}
 \sep\ass\\
 J\ass\\
 \zero
 \end{array}
 \using JR
 \endprooftree
 \justifies
 \begin{array}{c}
 \syncnst{D}\add\syncnst{C}\ass\\
 (S\circum(\VP/N))\wprodtwo J\ass\\
 (\lambda x((x\ \semcnst{c})\ \semcnst{d}), \zero)
 \end{array}
 \using \wprodtwo I
 \endprooftree
\justifies
\begin{array}{c}
\syncnst{and}\add\syncnst{D}\add\syncnst{C}\ass\\
 (S\circumtwo(\VP\circum N))\bsl(S\circumtwo(\VP\circum N))\ass\\
 \lambda y\lambda z[(y\ z)\wedge((z\ \semcnst{s})\ \semcnst{d})]
 \end{array}
 \using /E
 \endprooftree
\justifies
\begin{array}{c}
\syncnst{J}\add\sep\add\syncnst{W}\add\sep\add\syncnst{and}\add\syncnst{D}\add\syncnst{C}\ass\\
 S\circumtwo(\VP\circum N)\ass\\
 \lambda z[((z\ \semcnst{w})\ \semcnst{j})\wedge((z\ \semcnst{c})\ \semcnst{d})]
 \end{array}
 \using \bsl E
 \endprooftree
 \prooftree
 \prooftree
 \prooftree
 \begin{array}{c}
 \syncnst{wants}\ass\\
 (\VP/\VP)/N\ass\\
 \semcnst{want}
 \end{array}
 \prooftree
\justifies
\begin{array}{c}
 a\ass\\
 N\ass\\
 x
\end{array}
 \using i
 \endprooftree
 \justifies
 \begin{array}{c}
 \syncnst{wants}\add a\ass\\
 \VP/\VP\ass\\
 (\semcnst{want}\ x)
 \end{array}
 \using /E
 \endprooftree
 \prooftree
 \mbox{to win}
 \justifies
 \begin{array}{c}
\syncnst{to}\add\syncnst{win}\ass\\
\VP\ass\\
\semcnst{win}
\end{array}
 \endprooftree
 \justifies
 \begin{array}{c}
 \syncnst{wants}\add a\add\syncnst{to}\add\syncnst{win}\ass\\
 \VP\ass\\
 ((\semcnst{want}\ x)\ \semcnst{win})
 \end{array}
 \using /E
 \endprooftree
\justifies
\begin{array}{c}
 \syncnst{wants}\add \sep\add\syncnst{to}\add\syncnst{win}\ass\\
 \VP\circum N\ass\\
 \lambda x((\semcnst{want}\ x)\ \semcnst{win})
 \end{array}
 \using \circum I^i
 \endprooftree
 \justifies
\syncnst{J}\add\syncnst{wants}\add\syncnst{W}\add\syncnst{to}\add\syncnst{win}\add\syncnst{and}\add\syncnst{D}\add\syncnst{C}\ass
 S\ass
 {}[(((\semcnst{want}\ \semcnst{w})\ \semcnst{win})\ \semcnst{j})\wedge(((\semcnst{want}\ \semcnst{c})\ \semcnst{win})\ \semcnst{d})]
\using \circumtwo E
\endprooftree 
 }
 
 \ 
 
\noindent Observe how in the last step of both of the above derivations adjunction combines a string
 of sort $2$ and a string of sort $1$. But, in the first, simplex, case the separator of the second operand
 is right peripheral, whereas in the second, complex, case the separator of the second operand is medial.
 This is how the account unifies simplex and complex gapping under a single coordinator type.
 
 The account of determiner gapping consists in an assignment to the
coordinator of the following type ($Q$ is $((S\circum N)\infix S)/\CN$):
\disp{\begin{tabular}[t]{l}
$\syncnst{and}\ass (X\bsl X)/((X\wprodtwo J)\wprod I)\ass \lambda x\lambda y\lambda z\lambda w[((y\ z)\ w)\wedge((\pi_1\pi_1 x\ z)\ w)]$\\
where $X= (S\circumtwo (\VP\circum N))\circum Q$
\end{tabular}
\label{detgappingtype}}
Consider example (\ref{detgappingex}a) of simplex determiner gapping:
\lingform{Some dogs like Whiskas and} \lingform{cats Alpo}.
We use the following auxiliary derivation showing that a common noun followed
by an object has type $(S\circumtwo (\VP\circum$ $N))\circum Q$:
\disp{
 \prooftree
 \prooftree
 \prooftree
\prooftree
 \prooftree
 \prooftree
\justifies 
 a\ass N\ass x
 \using i
 \endprooftree
 \prooftree
 \prooftree
 \justifies
 b_0\add\sep\add b_1\ass\VP\circum N\ass y
 \using j
 \endprooftree
 \syncnst{obj}\ass N\ass\semcnst{obj}
 \justifies
 b_0\add\syncnst{obj}\add b_1\ass\VP\ass(y\ \semcnst{obj})
 \using \circum E
 \endprooftree
 \justifies
 a\add b_0\add\syncnst{obj}\add b_1\ass S\ass((y\ \semcnst{obj})\ x)
 \using \bsl E
 \endprooftree 
 \justifies
 \sep\add b_0\add\syncnst{obj}\add b_1\ass S\circum N\ass\lambda x((y\ \semcnst{obj})\ x)
 \using \circum I^i
 \endprooftree 
  \prooftree
 \prooftree
\justifies
 c\ass Q\ass z
\using k
\endprooftree
\syncnst{cn}\ass\CN\ass\semcnst{cn}
\justifies 
c\add\syncnst{cn}\ass (S\circum N)\infix S\ass(z\ \semcnst{cn})
\using /E
\endprooftree
\justifies
c\add\syncnst{cn}\add b_0\add\syncnst{obj}\add b_1\ass S\ass((z\ \semcnst{cn})\ \lambda x((y\ \semcnst{obj})\ x))
\using \infix E
\endprooftree
\justifies
c\add\syncnst{cn}\add \sep\add\syncnst{obj}\add\sep\ass S\circumtwo(\VP\circum N)\ass\lambda y((z\ \semcnst{cn})\ \lambda x((y\ \semcnst{obj})\ x))
\using \circumtwo I^j 
\endprooftree
\justifies
\sep\add\syncnst{cn}\add \sep\add\syncnst{obj}\add\sep\ass (S\circumtwo(\VP\circum N))\circum Q\ass\lambda z\lambda y((z\ \semcnst{cn})\ \lambda x((y\ \semcnst{obj})\ x))
\using \circum I^k
\endprooftree
}
The derivation of (\ref{detgappingex}a) is given in Figure~\ref{detgappingfig}.
\begin{figure}
\begin{center}
\rotatebox{-90}{\scriptsize
\prooftree
\prooftree
\prooftree
 \prooftree
 \syncnst{dogs}\ass\CN\ass\semcnst{dogs}\tb
 \syncnst{W}\ass N\ass\semcnst{w}
 \justifies
 \begin{array}{c}
 \sep\add\syncnst{dogs}\add \sep\add\syncnst{W}\add\sep\ass\\
 (S\circumtwo(\VP\circum N))\circum Q\ass\\
 \lambda z\lambda y((z\ \semcnst{dogs})\ \lambda x((y\ \semcnst{w})\ x))
 \end{array}
\endprooftree
\prooftree
 \begin{array}[b]{c}
 \syncnst{and}\ass\\
 (((S\circumtwo(\VP\circum N))\circum Q)\bsl((S\circumtwo(\VP\circum N))\circum Q))/\\
 /((((S\circumtwo(\VP\circum N))\circum Q)\wprod I)\wprodtwo J)\ass\\
 \lambda x\lambda y\lambda z\lambda w[((y\ z)\ w)\wedge((\pi_1\pi_1 x\ z)\ w)]
 \end{array}
 \prooftree
 \prooftree
 \prooftree
 \syncnst{cats}\ass\CN\ass\semcnst{cats}\tab\tab
 \syncnst{A}\ass N\ass\semcnst{a}
 \justifies
 \sep\add\syncnst{cats}\add \sep\add\syncnst{A}\add\sep\ass (S\circumtwo(\VP\circum N))\circum Q\ass\lambda z\lambda y((z\ \semcnst{cats})\ \lambda x((y\ \semcnst{a})\ x))
\endprooftree
\prooftree
\justifies
\nil\ass I\ass\zero
\using II
\endprooftree
\justifies
\syncnst{cats}\add \sep\add\syncnst{A}\add\one\ass ((S\circum(\VP/N))\circum Q)\wprod I\ass(\lambda z\lambda y((z\ \semcnst{cats})\ \lambda x((y\ \semcnst{a})\ x)), \zero)
\using \wprod I
\endprooftree
\prooftree
\justifies
\one\ass J\ass\zero
\using JI
\endprooftree
\justifies
\syncnst{cats}\add\syncnst{A}\ass (((S\circum(\VP/N))\circum Q)\wprod I)\wprodtwo J\ass((\lambda z\lambda y((z\ \semcnst{cats})\ \lambda x((y\ \semcnst{a})\ x)), \zero), \zero)
\using \wprodtwo I
\endprooftree
 \justifies
 \syncnst{and}\add\syncnst{cats}\add\syncnst{A}\ass
 ((S\circumtwo(\VP\circum N))\circum Q)\bsl((S\circumtwo(\VP\circum N))\circum Q)
   \ass\lambda y\lambda z\lambda w[((y\ z)\ w)\wedge((z\ \semcnst{cats})\ \lambda x((w\ \semcnst{a})\ x)]
 \using /E
 \endprooftree
 \justifies
 \sep\add\syncnst{dogs}\add \sep\add\syncnst{W}\add\one\add\syncnst{and}\add\syncnst{cats}\add\syncnst{A}\ass
 (S\circumtwo(\VP\circum N))\circum Q
   \ass\lambda z\lambda w[((z\ \semcnst{dogs})\ \lambda x((w\ \semcnst{w})\ x))\wedge((z\ \semcnst{cats})\ \lambda x((w\ \semcnst{a})\ x)]
 \using \bsl E
 \endprooftree
 \syncnst{some}\ass Q\ass\exists
\justifies
\begin{array}{c}
\syncnst{some}\add\syncnst{dogs}\add \sep\add\syncnst{W}\add\one\add\syncnst{and}\add\syncnst{cats}\add\syncnst{A}\ass\\
S\circumtwo(\VP\circum N)\ass\\
\lambda w[((\exists\ \semcnst{dogs})\ \lambda x((w\ \semcnst{w})\ x))\wedge((\exists\ \semcnst{cats})\ \lambda x((w\ \semcnst{a})\ x)]
\end{array}
 \using \circum E
 \endprooftree
 \prooftree
\prooftree
 \syncnst{like}\ass\VP/N\ass\semcnst{like}
 \prooftree
\justifies
 a\ass N\ass x
 \using i
 \endprooftree
 \justifies
 \syncnst{like}\add a\ass\VP\ass(\semcnst{like}\ x)
 \using /E
 \endprooftree
\justifies
 \syncnst{like}\add \sep\ass\VP\circum N\ass\lambda x(\semcnst{like}\ x)
 \using \circum I^i
 \endprooftree
 \justifies
 \syncnst{some}\add\syncnst{dogs}\add\syncnst{like}\add\syncnst{W}\add\syncnst{and}\add\syncnst{cats}\add\syncnst{A}\ass
 S
   \ass[((\exists\ \semcnst{dogs})\ \lambda x((\semcnst{like}\ \semcnst{w})\ x))\wedge((\exists\ \semcnst{cats})\ \lambda x((\semcnst{like}\ \semcnst{a})\ x)]
 \using \circumtwo E
 \endprooftree}
 \end{center}
 \caption{Determiner gapping}
 \label{detgappingfig}
 \end{figure}
 Finally, from the same coordinator type assignment (\ref{detgappingtype})
 we can derive the case of discontinuous determiner gapping (\ref{detgappingex}b):
$$\lingform{Every cook wants Bar\c{c}a} \lingform{to win and waiter Madrid}$$
 in Figure~\ref{comdetgappingfig}.

\begin{figure}
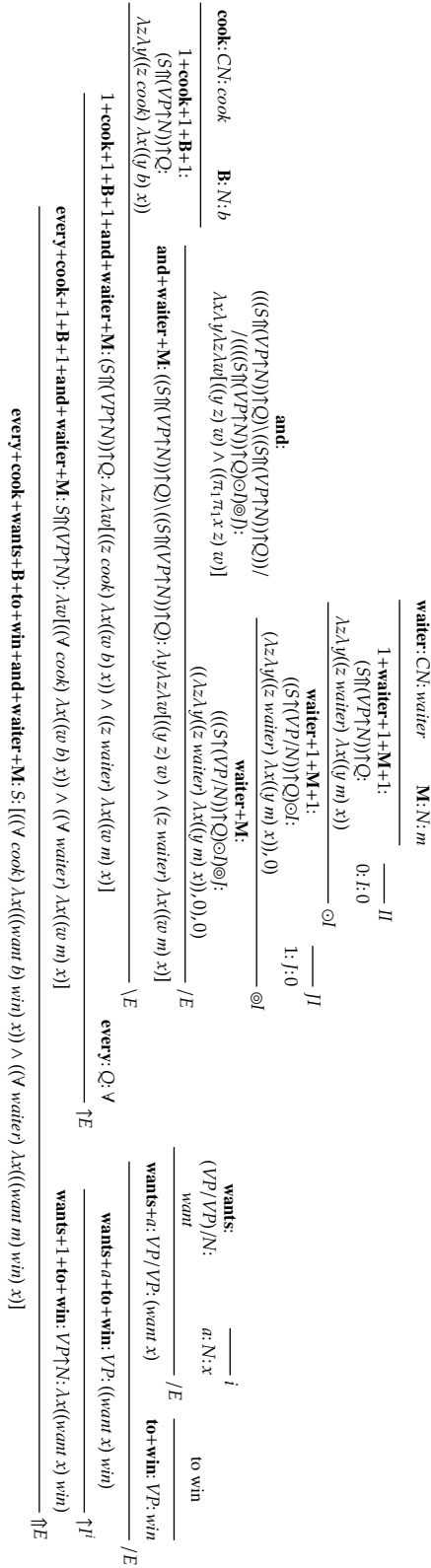

\begin{center}
\rotatebox{-90}{\scriptsize
\prooftree
\prooftree
\prooftree
 \prooftree
 \syncnst{cook}\ass\CN\ass\semcnst{cook}\tb
 \syncnst{B}\ass N\ass\semcnst{b}
 \justifies
 \begin{array}{c}
 \sep\add\syncnst{cook}\add \sep\add\syncnst{B}\add\sep\ass\\
 (S\circumtwo(\VP\circum N))\circum Q\ass\\
 \lambda z\lambda y((z\ \semcnst{cook})\ \lambda x((y\ \semcnst{b})\ x))
 \end{array}
\endprooftree
\prooftree
 \begin{array}[b]{c}
 \syncnst{and}\ass\\
 (((S\circumtwo(\VP\circum N))\circum Q)\bsl((S\circumtwo(\VP\circum N))\circum Q))/\\
 /((((S\circumtwo(\VP\circum N))\circum Q)\wprod I)\wprodtwo J)\ass\\
 \lambda x\lambda y\lambda z\lambda w[((y\ z)\ w)\wedge((\pi_1\pi_1 x\ z)\ w)]
 \end{array}
 \prooftree
 \prooftree
 \prooftree
 \syncnst{waiter}\ass\CN\ass\semcnst{waiter}\tb
 \syncnst{M}\ass N\ass\semcnst{m}
 \justifies
\begin{array}{c}
\sep\add\syncnst{waiter}\add \sep\add\syncnst{M}\add\sep\ass\\
(S\circumtwo(\VP\circum N))\circum Q\ass\\
\lambda z\lambda y((z\ \semcnst{waiter})\ \lambda x((y\ \semcnst{m})\ x))
\end{array}
\endprooftree
\prooftree
\justifies
\nil\ass I\ass\zero
\using II
\endprooftree
\justifies
\begin{array}{c}
\syncnst{waiter}\add \sep\add\syncnst{M}\add\one\ass\\
((S\circum(\VP/N))\circum Q)\wprod I\ass\\
(\lambda z\lambda y((z\ \semcnst{waiter})\ \lambda x((y\ \semcnst{m})\ x)), \zero)
\end{array}
\using \wprod I
\endprooftree
\prooftree
\justifies
\one\ass J\ass\zero
\using JI
\endprooftree
\justifies
\begin{array}{c}
\syncnst{waiter}\add\syncnst{M}\ass\\
(((S\circum(\VP/N))\circum Q)\wprod I)\wprodtwo J\ass\\
((\lambda z\lambda y((z\ \semcnst{waiter})\ \lambda x((y\ \semcnst{m})\ x)), \zero), \zero)
\end{array}
\using \wprodtwo I
\endprooftree
 \justifies
 \syncnst{and}\add\syncnst{waiter}\add\syncnst{M}\ass
 ((S\circumtwo(\VP\circum N))\circum Q)\bsl((S\circumtwo(\VP\circum N))\circum Q)
   \ass\lambda y\lambda z\lambda w[((y\ z)\ w)\wedge((z\ \semcnst{waiter})\ \lambda x((w\ \semcnst{m})\ x)]
 \using /E
 \endprooftree
 \justifies
 \sep\add\syncnst{cook}\add \sep\add\syncnst{B}\add\one\add\syncnst{and}\add\syncnst{waiter}\add\syncnst{M}\ass
 (S\circumtwo(\VP\circum N))\circum Q
   \ass\lambda z\lambda w[((z\ \semcnst{cook})\ \lambda x((w\ \semcnst{b})\ x))\wedge((z\ \semcnst{waiter})\ \lambda x((w\ \semcnst{m})\ x)]
 \using \bsl E
 \endprooftree
 \syncnst{every}\ass Q\ass\forall
\justifies
 \syncnst{every}\add\syncnst{cook}\add \sep\add\syncnst{B}\add\one\add\syncnst{and}\add\syncnst{waiter}\add\syncnst{M}\ass
 S\circumtwo(\VP\circum N)
   \ass\lambda w[((\forall\ \semcnst{cook})\ \lambda x((w\ \semcnst{b})\ x))\wedge((\forall\ \semcnst{waiter})\ \lambda x((w\ \semcnst{m})\ x)]
 \using \circum E
 \endprooftree
 \prooftree
 \prooftree
 \prooftree
 \begin{array}{c}
 \syncnst{wants}\ass\\
 (\VP/\VP)/N\ass\\
 \semcnst{want}
 \end{array}
 \tb
 \prooftree
\justifies
 a\ass N\ass x
 \using i
 \endprooftree
 \justifies
 \syncnst{wants}\add a\ass\VP/\VP\ass(\semcnst{want}\ x)
 \using /E
 \endprooftree
 \prooftree
 \mbox{to win}
 \justifies
\syncnst{to}\add\syncnst{win}\ass\VP\ass\semcnst{win}
 \endprooftree
 \justifies
 \syncnst{wants}\add a\add\syncnst{to}\add\syncnst{win}\ass\VP\ass((\semcnst{want}\ x)\ \semcnst{win})
 \using /E
 \endprooftree
\justifies
 \syncnst{wants}\add \sep\add\syncnst{to}\add\syncnst{win}\ass\VP\circum N\ass\lambda x((\semcnst{want}\ x)\ \semcnst{win})
 \using \circum I^i
 \endprooftree
 \justifies
 \syncnst{every}\add\syncnst{cook}\add\syncnst{wants}\add\syncnst{B}\add\syncnst{to}\add\syncnst{win}\add\syncnst{and}\add\syncnst{waiter}\add\syncnst{M}\ass
 S
   \ass[((\forall\ \semcnst{cook})\ \lambda x(((\semcnst{want}\ \semcnst{b})\ \semcnst{win})\ x))\wedge((\forall\ \semcnst{waiter})\ \lambda x(((\semcnst{want}\ \semcnst{m})\ \semcnst{win})\ x)]
 \using \circumtwo E
 \endprooftree}
 \end{center}
 \caption{Discontinuous determiner gapping}
 \label{comdetgappingfig}
 \end{figure}
 
 \clearpage
 
\section{Summary}

The categorial analysis of gapping as like-type coordination was
established in Steedman (1990\cite{steed:gapping90}) and
Hendriks (1995\cite{hendriksP:phd}). In the framework of HTLG
Kubota and Levine (2012\cite{htlggapping}) go further in that they provide
like-type coordination for discontinuous gapping.
Our analysis is inspired by that of
Kubota and Levine (2012\cite{htlggapping}; 2013\cite{kublev:detgap}).
However, our analysis of gapping represents an improvement
on the HTLG analysis because we do not require two types for
simple and discontinuous gapping: a single type suffices.

Finally, we have noted that the HTLG account of gapping suffers
from determiner-transitive verb
order inconsistency overgeneration; 
the same problem would arise for HTLG in relation to discontinuous determiner gapping:
\disp{\begin{tabular}[t]{ll}
a. & \unacc Some boy wants Everton to win and Mary (wants) (some) London club (to win)\\
 b. & \unacc Mary wants some London club to win and (some) boy (wants)  
 Everton (to win)
 \end{tabular}}
 In addition to capturing simplex gapping (and determiner gapping)
 as a special case of complex gapping (and determiner gapping),
 our DTLG account does not have the order inconsistency overgeneration problem
 of determiner gapping and discontinuous determiner gapping.
 
\noindent
$$\rotatebox{-90}{$
		\begin{array}{r|ccc|ccc|c|c|c|c|c|c|c|c|}
			\cline{2-14}
			& \multicolumn{3}{c|}{\begin{tabular}{c}cont.\\mult.\end{tabular}}&
			\multicolumn{3}{c|}{\begin{tabular}{c}disc.\\ mult.\end{tabular}}  &
			\mathrm{add.} & \mathrm{qu.} & \begin{tabular}{c}norm.\\ mod.\end{tabular} & 
			\begin{tabular}{c}brack.\\ mod.\end{tabular} & \mathrm{exp.} 
			&\multicolumn{1}{c|}{\begin{tabular}{c}limited\\contr.\\\& limited\\expan.\end{tabular}} & \mathrm{diff.}
			\\
			\cline{2-14}
			&&&&&&&&&&&&&\\
			&/ & & \bsl & \scircum{} & & \sinfix{} & \aconj_{} & \bigwedge_{} & \mymod_{} & \abrack_{} & \univexp_{} &|_{}&\\
			& & \product  & & & \swprod{}{} &&&&&& && -  \\
			& & I &&& J_{}&& \adisj_{} & \bigvee_{} & \emod_{} & \mybrack_{} & \exstexp_{} &+ & \\
			&&&&&&&&&&&&&\\
			\cline{2-14}
			&\multicolumn{3}{c|}{\rotatebox{90}{\begin{tabular}[t]{l}Continuous\\subcategorisation\end{tabular}}}  &
			\multicolumn{3}{c|}{\rotatebox{90}{\begin{tabular}[t]{l}Disontinuous\\subcategorisation\end{tabular}}}&
			\rotatebox{90}{\begin{tabular}[t]{l}Weak\\polymorphism\end{tabular}}&
			\rotatebox{90}{\begin{tabular}[t]{l}Features\end{tabular}}&
			\rotatebox{90}{\begin{tabular}[t]{l}Intensionality\end{tabular}}&
			\rotatebox{90}{\begin{tabular}[t]{l}Syntactic\\domains\end{tabular}}&
			\rotatebox{90}{\begin{tabular}[t]{l}Nonlinearity\end{tabular}}&
			\rotatebox{90}{\begin{tabular}[t]{l}Anaphora\\\& words as types\end{tabular}}&
			\rotatebox{90}{\begin{tabular}[t]{l}Exceptions\end{tabular}}\\
			\cline{2-14}
		\end{array}$}
$$

\noindent
$$\rotatebox{-90}{$\scriptsize
\begin{array}{r|ccc|ccc|ccc|c|c|c|c|c|c|c|}
& \multicolumn{3}{c|}{\begin{tabular}{c}cont.\\mult.\end{tabular}}&
\multicolumn{3}{c|}{\begin{tabular}{c}disc.\\ mult.\end{tabular}}  &
\multicolumn{3}{c|}{\begin{tabular}{c}secon.\\ disc.\\ mult.\end{tabular}}  &
\mathrm{add.} & \mathrm{qu.} & \begin{tabular}{c}norm.\\ mod.\end{tabular} & 
\begin{tabular}{c}brack.\\ mod.\end{tabular} & \mathrm{exp.} 
 &\multicolumn{1}{c|}{\begin{tabular}{c}limited\\contr.\\\& weak.\end{tabular}}
\\
 \cline{1-16}
&&&&&&&&&&&&&&&\\
&/_{1} & & \bsl_{2} & \sextract{}_{5} & & \sinfix{}_{6} & \sextracttwo{}_{9} && \sinfixtwo{}_{10} & \aconj_{13} & \bigwedge_{15} & \mymod_{17} & \abrack_{19} & \univexp_{21} &|_{23}\\
\mathrm{primitive} & & \product_{3}  & & & \sdprod{}_{7} &&& \sdprodtwo{}_{11} &&&&&& &   \\
& & I_{4} &&& J_{8} &&& K_{12} && \adisj_{14} & \bigvee_{16} & \emod_{18} & \mybrack_{20} & \exstexp_{22} &W_{24}\\
&&&&&&&&&&&&&&&\\
 \cline{1-16}
  &&&&&&&&&&&&\\
 \mathrm{sem.} &\argover_{25}\ \fununder_{26}&&\funover_{28}\ \argunder_{27}&\funextract{}_{31}\ \arginfix{}_{32} &&\argextract{}_{34}\ \funinfix{}_{35}&\funextracttwo{}_{37}\ \arginfixtwo{}_{38} &&\argextracttwo{}_{40}\ \funinfixtwo{}_{41} & \iaconj_{43} & \forall_{45} & \imod_{47} 
 \\
 \mathrm{inactive}&&&&&&&&&&&&\\
  \mathrm{variants}&\leftiprod{}_{29}&&\rightiprod{}_{30}&\upperiwprod{}_{35}&&\loweriwprod{}_{36}&\upperiwprodtwo{}_{41}&&\loweriwprodtwo{}_{42}&\iadisj_{44}&\exists_{46}&\iemod_{48}&\multicolumn{1}{c}{}
 \\
 &&&&&&&&&&&&\\
 \cline{1-13}
 & &&&&&&&&\\
\mathrm{det.}&\leftproj_{49} && \rightproj_{50} && \ssplit{}_{53} &&& \ssplittwo{}_{55} &\\
&&&&&&&&&&\multicolumn{5}{c|}{}&\\
\mathrm{synth.}&\leftinj_{51} && \rightinj_{52} && \sbridge{}_{54} &&& \sbridgetwo{}_{56} &&\multicolumn{4}{c}{}&&\mbox{diff.}
\\
 &&&&&&&&&&\multicolumn{1}{l}{}&\multicolumn{4}{c|}{}&\mbox{}\\
 \cline{1-10}\cline{16-16}
 &&&&&&&\multicolumn{4}{l}{}&\multicolumn{4}{c|}{}&\\
\mathrm{non-det.} &  & \ndiv_{57} && \nextract_{59} && \ninfix_{60} &\multicolumn{8}{c|}{}&\\
 &&&&&&& \multicolumn{1}{l}{}&\multicolumn{7}{c|}{}&\diff_{62} \\
\mathrm{synth.}  && \nprod_{58} &&& \ndprod_{61} &&\multicolumn{8}{c|}{}&\\
 &&&&&&&\multicolumn{8}{c|}{}&\\
 \cline{1-7}\cline{16-16}
\end{array}$}
$$

\chapter{Scripture}

Parsing the first sentence of the Bible with parser/theorem-prover {\it CatLog3\/} (available on GitHub;
Morrill 2019\cite{catlog3}), the analysis and semantics delivered are:

\disp{
(gen(one(1))) ${\bf in}{+}{\bf the}{+}{\bf beginning}{+}[{\bf God}]{+}{\bf created}{+}[[{\bf the}{+}{\bf heaven}{+}{\bf and}{+}{\bf the}{+}{\bf earth}]]: Sf$}

\disp{\footnotesize
${\square}({\forall}f(Sf{\div}Sf)/{\exists}aNa): {\it in}, {\blacksquare}{\forall}n(Nt(n)/{\it CN}{\it n}): \iota , {\square}{\it CN}{\it s(n)}: {\it beginning}, [{\blacksquare}Nt(s(m)): {\it God}],\\
{\square}(({\langle\rangle}{\exists}aNa\backslash Sf)/{\exists}aNa): \mbox{\^{}}\lambda A\lambda B({\it Past}\ ((\mbox{\v{}}{\it create}\ {\it A})\ {\it B})), [[{\blacksquare}{\forall}n(Nt(n)/{\it CN}{\it n}): \iota , {\square}{\it CN}{\it s(n)}: {\it heaven},\\
{\blacksquare}{\forall}f{\forall}a((?{\blacksquare}((({\langle\rangle}Na\backslash Sf)/{\exists}bNb)\backslash ({\langle\rangle}Na\backslash Sf))\backslash {[]^{-1}}{[]^{-1}}((({\langle\rangle}Na\backslash Sf)/{\exists}bNb)\backslash ({\langle\rangle}Na\backslash Sf)))/{\blacksquare}((({\langle\rangle}Na\backslash Sf)/{\exists}bNb)\backslash ({\langle\rangle}Na\backslash Sf))): (\Phinplus{}\ ({\it s}\ ({\it s}\ {\it 0}))\ {\it and}), {\blacksquare}{\forall}n(Nt(n)/{\it CN}{\it n}): \iota , {\square}{\it CN}{\it s(n)}: {\it earth}]]\ \Rightarrow\ Sf$}
\vspace{0.15in}
$${\tiny
\prooftree
\prooftree
\prooftree
\prooftree
\prooftree
\prooftree
\justifies
\mbox{\fbox{${\it CN}{\it s(n)}$}}\ \Rightarrow\ {\it CN}{\it s(n)}
\endprooftree
\justifies
\mbox{\fbox{${\square}{\it CN}{\it s(n)}$}}\ \Rightarrow\ {\it CN}{\it s(n)}
\using {\Box}L
\endprooftree
\prooftree
\justifies
\mbox{\fbox{$Nt(s(n))$}}\ \Rightarrow\ Nt(s(n))
\endprooftree
\justifies
\mbox{\fbox{$Nt(s(n))/{\it CN}{\it s(n)}$}}, {\square}{\it CN}{\it s(n)}\ \Rightarrow\ Nt(s(n))
\using {/}L
\endprooftree
\justifies
\mbox{\fbox{${\forall}n(Nt(n)/{\it CN}{\it n})$}}, {\square}{\it CN}{\it s(n)}\ \Rightarrow\ Nt(s(n))
\using {\forall}L
\endprooftree
\justifies
\mbox{\fbox{${\blacksquare}{\forall}n(Nt(n)/{\it CN}{\it n})$}}, {\square}{\it CN}{\it s(n)}\ \Rightarrow\ Nt(s(n))
\using {\blacksquare}L
\endprooftree
\justifies
\begin{array}{c}
{\blacksquare}{\forall}n(Nt(n)/{\it CN}{\it n}), {\square}{\it CN}{\it s(n)}\ \Rightarrow\ \fbox{${\exists}aNa$}\\
\mbox{\footnotesize\textcircled{1}}
\end{array}
\using {\exists}R
\endprooftree
\tb
\prooftree
\prooftree
\prooftree
\prooftree
\prooftree
\prooftree
\prooftree
\prooftree
\prooftree
\prooftree
\prooftree
\justifies
\mbox{\fbox{${\it CN}{\it s(n)}$}}\ \Rightarrow\ {\it CN}{\it s(n)}
\endprooftree
\justifies
\mbox{\fbox{${\square}{\it CN}{\it s(n)}$}}\ \Rightarrow\ {\it CN}{\it s(n)}
\using {\Box}L
\endprooftree
\prooftree
\justifies
\mbox{\fbox{$Nt(s(n))$}}\ \Rightarrow\ Nt(s(n))
\endprooftree
\justifies
\mbox{\fbox{$Nt(s(n))/{\it CN}{\it s(n)}$}}, {\square}{\it CN}{\it s(n)}\ \Rightarrow\ Nt(s(n))
\using {/}L
\endprooftree
\justifies
\mbox{\fbox{${\forall}n(Nt(n)/{\it CN}{\it n})$}}, {\square}{\it CN}{\it s(n)}\ \Rightarrow\ Nt(s(n))
\using {\forall}L
\endprooftree
\justifies
\mbox{\fbox{${\blacksquare}{\forall}n(Nt(n)/{\it CN}{\it n})$}}, {\square}{\it CN}{\it s(n)}\ \Rightarrow\ Nt(s(n))
\using {\blacksquare}L
\endprooftree
\justifies
{\blacksquare}{\forall}n(Nt(n)/{\it CN}{\it n}), {\square}{\it CN}{\it s(n)}\ \Rightarrow\ \fbox{${\exists}bNb$}
\using {\exists}R
\endprooftree
\prooftree
\prooftree
\prooftree
\justifies
Nt(s(m))\ \Rightarrow\ Nt(s(m))
\endprooftree
\justifies
[Nt(s(m))]\ \Rightarrow\ \fbox{${\langle\rangle}Nt(s(m))$}
\using {\langle\rangle}R
\endprooftree
\prooftree
\justifies
\mbox{\fbox{$Sf$}}\ \Rightarrow\ Sf
\endprooftree
\justifies
[Nt(s(m))], \mbox{\fbox{${\langle\rangle}Nt(s(m))\backslash Sf$}}\ \Rightarrow\ Sf
\using {\backslash}L
\endprooftree
\justifies
[Nt(s(m))], \mbox{\fbox{$({\langle\rangle}Nt(s(m))\backslash Sf)/{\exists}bNb$}}, {\blacksquare}{\forall}n(Nt(n)/{\it CN}{\it n}), {\square}{\it CN}{\it s(n)}\ \Rightarrow\ Sf
\using {/}L
\endprooftree
\justifies
{\langle\rangle}Nt(s(m)), ({\langle\rangle}Nt(s(m))\backslash Sf)/{\exists}bNb, {\blacksquare}{\forall}n(Nt(n)/{\it CN}{\it n}), {\square}{\it CN}{\it s(n)}\ \Rightarrow\ Sf
\using {\langle\rangle}L
\endprooftree
\justifies
({\langle\rangle}Nt(s(m))\backslash Sf)/{\exists}bNb, {\blacksquare}{\forall}n(Nt(n)/{\it CN}{\it n}), {\square}{\it CN}{\it s(n)}\ \Rightarrow\ {\langle\rangle}Nt(s(m))\backslash Sf
\using {\backslash}R
\endprooftree
\justifies
{\blacksquare}{\forall}n(Nt(n)/{\it CN}{\it n}), {\square}{\it CN}{\it s(n)}\ \Rightarrow\ (({\langle\rangle}Nt(s(m))\backslash Sf)/{\exists}bNb)\backslash ({\langle\rangle}Nt(s(m))\backslash Sf)
\using {\backslash}R
\endprooftree
\justifies
\begin{array}{c}
{\blacksquare}{\forall}n(Nt(n)/{\it CN}{\it n}), {\square}{\it CN}{\it s(n)}\ \Rightarrow\ {\blacksquare}((({\langle\rangle}Nt(s(m))\backslash Sf)/{\exists}bNb)\backslash ({\langle\rangle}Nt(s(m))\backslash Sf))\\
\mbox{\footnotesize\textcircled{2}}
\end{array}
\using {\blacksquare}R
\endprooftree}
$$
$${
\prooftree
\prooftree
\prooftree
\prooftree
\prooftree
\prooftree
\prooftree
\prooftree
\prooftree
\prooftree
\prooftree
\prooftree
\justifies
\mbox{\fbox{${\it CN}{\it s(n)}$}}\ \Rightarrow\ {\it CN}{\it s(n)}
\endprooftree
\justifies
\mbox{\fbox{${\square}{\it CN}{\it s(n)}$}}\ \Rightarrow\ {\it CN}{\it s(n)}
\using {\Box}L
\endprooftree
\prooftree
\justifies
\mbox{\fbox{$Nt(s(n))$}}\ \Rightarrow\ Nt(s(n))
\endprooftree
\justifies
\mbox{\fbox{$Nt(s(n))/{\it CN}{\it s(n)}$}}, {\square}{\it CN}{\it s(n)}\ \Rightarrow\ Nt(s(n))
\using {/}L
\endprooftree
\justifies
\mbox{\fbox{${\forall}n(Nt(n)/{\it CN}{\it n})$}}, {\square}{\it CN}{\it s(n)}\ \Rightarrow\ Nt(s(n))
\using {\forall}L
\endprooftree
\justifies
\mbox{\fbox{${\blacksquare}{\forall}n(Nt(n)/{\it CN}{\it n})$}}, {\square}{\it CN}{\it s(n)}\ \Rightarrow\ Nt(s(n))
\using {\blacksquare}L
\endprooftree
\justifies
{\blacksquare}{\forall}n(Nt(n)/{\it CN}{\it n}), {\square}{\it CN}{\it s(n)}\ \Rightarrow\ \fbox{${\exists}bNb$}
\using {\exists}R
\endprooftree
\prooftree
\prooftree
\prooftree
\justifies
Nt(s(m))\ \Rightarrow\ Nt(s(m))
\endprooftree
\justifies
[Nt(s(m))]\ \Rightarrow\ \fbox{${\langle\rangle}Nt(s(m))$}
\using {\langle\rangle}R
\endprooftree
\prooftree
\justifies
\mbox{\fbox{$Sf$}}\ \Rightarrow\ Sf
\endprooftree
\justifies
[Nt(s(m))], \mbox{\fbox{${\langle\rangle}Nt(s(m))\backslash Sf$}}\ \Rightarrow\ Sf
\using {\backslash}L
\endprooftree
\justifies
[Nt(s(m))], \mbox{\fbox{$({\langle\rangle}Nt(s(m))\backslash Sf)/{\exists}bNb$}}, {\blacksquare}{\forall}n(Nt(n)/{\it CN}{\it n}), {\square}{\it CN}{\it s(n)}\ \Rightarrow\ Sf
\using {/}L
\endprooftree
\justifies
{\langle\rangle}Nt(s(m)), ({\langle\rangle}Nt(s(m))\backslash Sf)/{\exists}bNb, {\blacksquare}{\forall}n(Nt(n)/{\it CN}{\it n}), {\square}{\it CN}{\it s(n)}\ \Rightarrow\ Sf
\using {\langle\rangle}L
\endprooftree
\justifies
({\langle\rangle}Nt(s(m))\backslash Sf)/{\exists}bNb, {\blacksquare}{\forall}n(Nt(n)/{\it CN}{\it n}), {\square}{\it CN}{\it s(n)}\ \Rightarrow\ {\langle\rangle}Nt(s(m))\backslash Sf
\using {\backslash}R
\endprooftree
\justifies
{\blacksquare}{\forall}n(Nt(n)/{\it CN}{\it n}), {\square}{\it CN}{\it s(n)}\ \Rightarrow\ (({\langle\rangle}Nt(s(m))\backslash Sf)/{\exists}bNb)\backslash ({\langle\rangle}Nt(s(m))\backslash Sf)
\using {\backslash}R
\endprooftree
\justifies
{\blacksquare}{\forall}n(Nt(n)/{\it CN}{\it n}), {\square}{\it CN}{\it s(n)}\ \Rightarrow\ {\blacksquare}((({\langle\rangle}Nt(s(m))\backslash Sf)/{\exists}bNb)\backslash ({\langle\rangle}Nt(s(m))\backslash Sf))
\using {\blacksquare}R
\endprooftree
\justifies
\begin{array}{c}
{\blacksquare}{\forall}n(Nt(n)/{\it CN}{\it n}), {\square}{\it CN}{\it s(n)}\ \Rightarrow\ \fbox{$?{\blacksquare}((({\langle\rangle}Nt(s(m))\backslash Sf)/{\exists}bNb)\backslash ({\langle\rangle}Nt(s(m))\backslash Sf))$}\\
\mbox{\footnotesize\textcircled{3}}
\end{array}
\using {?}R
\endprooftree}
$$
$$
\rotatebox{-90}{\tiny
\prooftree
\prooftree
\mbox{\footnotesize\textcircled{1}}\tab
\prooftree
\prooftree
\prooftree
\prooftree
\prooftree
\prooftree
\mbox{\footnotesize\textcircled{2}}\tab
\prooftree
\mbox{\footnotesize\textcircled{3}}\tab
\prooftree
\prooftree
\prooftree
\prooftree
\prooftree
\prooftree
\prooftree
\prooftree
\prooftree
\prooftree
\prooftree
\justifies
N7\ \Rightarrow\ N7
\endprooftree
\justifies
N7\ \Rightarrow\ \fbox{${\exists}aNa$}
\using {\exists}R
\endprooftree
\prooftree
\prooftree
\prooftree
\prooftree
\justifies
Nt(s(m))\ \Rightarrow\ Nt(s(m))
\endprooftree
\justifies
Nt(s(m))\ \Rightarrow\ \fbox{${\exists}aNa$}
\using {\exists}R
\endprooftree
\justifies
[Nt(s(m))]\ \Rightarrow\ \fbox{${\langle\rangle}{\exists}aNa$}
\using {\langle\rangle}R
\endprooftree
\prooftree
\justifies
\mbox{\fbox{$Sf$}}\ \Rightarrow\ Sf
\endprooftree
\justifies
[Nt(s(m))], \mbox{\fbox{${\langle\rangle}{\exists}aNa\backslash Sf$}}\ \Rightarrow\ Sf
\using {\backslash}L
\endprooftree
\justifies
[Nt(s(m))], \mbox{\fbox{$({\langle\rangle}{\exists}aNa\backslash Sf)/{\exists}aNa$}}, N7\ \Rightarrow\ Sf
\using {/}L
\endprooftree
\justifies
[Nt(s(m))], \mbox{\fbox{${\square}(({\langle\rangle}{\exists}aNa\backslash Sf)/{\exists}aNa)$}}, N7\ \Rightarrow\ Sf
\using {\Box}L
\endprooftree
\justifies
[Nt(s(m))], {\square}(({\langle\rangle}{\exists}aNa\backslash Sf)/{\exists}aNa), {\exists}bNb\ \Rightarrow\ Sf
\using {\exists}L
\endprooftree
\justifies
{\langle\rangle}Nt(s(m)), {\square}(({\langle\rangle}{\exists}aNa\backslash Sf)/{\exists}aNa), {\exists}bNb\ \Rightarrow\ Sf
\using {\langle\rangle}L
\endprooftree
\justifies
{\square}(({\langle\rangle}{\exists}aNa\backslash Sf)/{\exists}aNa), {\exists}bNb\ \Rightarrow\ {\langle\rangle}Nt(s(m))\backslash Sf
\using {\backslash}R
\endprooftree
\justifies
{\square}(({\langle\rangle}{\exists}aNa\backslash Sf)/{\exists}aNa)\ \Rightarrow\ ({\langle\rangle}Nt(s(m))\backslash Sf)/{\exists}bNb
\using {/}R
\endprooftree
\prooftree
\prooftree
\prooftree
\prooftree
\justifies
\mbox{\fbox{$Nt(s(m))$}}\ \Rightarrow\ Nt(s(m))
\endprooftree
\justifies
\mbox{\fbox{${\blacksquare}Nt(s(m))$}}\ \Rightarrow\ Nt(s(m))
\using {\blacksquare}L
\endprooftree
\justifies
[{\blacksquare}Nt(s(m))]\ \Rightarrow\ \fbox{${\langle\rangle}Nt(s(m))$}
\using {\langle\rangle}R
\endprooftree
\prooftree
\justifies
\mbox{\fbox{$Sf$}}\ \Rightarrow\ Sf
\endprooftree
\justifies
[{\blacksquare}Nt(s(m))], \mbox{\fbox{${\langle\rangle}Nt(s(m))\backslash Sf$}}\ \Rightarrow\ Sf
\using {\backslash}L
\endprooftree
\justifies
[{\blacksquare}Nt(s(m))], {\square}(({\langle\rangle}{\exists}aNa\backslash Sf)/{\exists}aNa), \mbox{\fbox{$(({\langle\rangle}Nt(s(m))\backslash Sf)/{\exists}bNb)\backslash ({\langle\rangle}Nt(s(m))\backslash Sf)$}}\ \Rightarrow\ Sf
\using {\backslash}L
\endprooftree
\justifies
[{\blacksquare}Nt(s(m))], {\square}(({\langle\rangle}{\exists}aNa\backslash Sf)/{\exists}aNa), [\mbox{\fbox{${[]^{-1}}((({\langle\rangle}Nt(s(m))\backslash Sf)/{\exists}bNb)\backslash ({\langle\rangle}Nt(s(m))\backslash Sf))$}}]\ \Rightarrow\ Sf
\using {[]^{-1}}L
\endprooftree
\justifies
[{\blacksquare}Nt(s(m))], {\square}(({\langle\rangle}{\exists}aNa\backslash Sf)/{\exists}aNa), [[\mbox{\fbox{${[]^{-1}}{[]^{-1}}((({\langle\rangle}Nt(s(m))\backslash Sf)/{\exists}bNb)\backslash ({\langle\rangle}Nt(s(m))\backslash Sf))$}}]]\ \Rightarrow\ Sf
\using {[]^{-1}}L
\endprooftree
\justifies
[{\blacksquare}Nt(s(m))], {\square}(({\langle\rangle}{\exists}aNa\backslash Sf)/{\exists}aNa), [[{\blacksquare}{\forall}n(Nt(n)/{\it CN}{\it n}), {\square}{\it CN}{\it s(n)}, \mbox{\fbox{$?{\blacksquare}((({\langle\rangle}Nt(s(m))\backslash Sf)/{\exists}bNb)\backslash ({\langle\rangle}Nt(s(m))\backslash Sf))\backslash {[]^{-1}}{[]^{-1}}((({\langle\rangle}Nt(s(m))\backslash Sf)/{\exists}bNb)\backslash ({\langle\rangle}Nt(s(m))\backslash Sf))$}}]]\ \Rightarrow\ Sf
\using {\backslash}L
\endprooftree
\justifies
[{\blacksquare}Nt(s(m))], {\square}(({\langle\rangle}{\exists}aNa\backslash Sf)/{\exists}aNa), [[{\blacksquare}{\forall}n(Nt(n)/{\it CN}{\it n}), {\square}{\it CN}{\it s(n)}, \mbox{\fbox{$(?{\blacksquare}((({\langle\rangle}Nt(s(m))\backslash Sf)/{\exists}bNb)\backslash ({\langle\rangle}Nt(s(m))\backslash Sf))\backslash {[]^{-1}}{[]^{-1}}((({\langle\rangle}Nt(s(m))\backslash Sf)/{\exists}bNb)\backslash ({\langle\rangle}Nt(s(m))\backslash Sf)))/{\blacksquare}((({\langle\rangle}Nt(s(m))\backslash Sf)/{\exists}bNb)\backslash ({\langle\rangle}Nt(s(m))\backslash Sf))$}}, {\blacksquare}{\forall}n(Nt(n)/{\it CN}{\it n}), {\square}{\it CN}{\it s(n)}]]\ \Rightarrow\ Sf
\using {/}L
\endprooftree
\justifies
[{\blacksquare}Nt(s(m))], {\square}(({\langle\rangle}{\exists}aNa\backslash Sf)/{\exists}aNa), [[{\blacksquare}{\forall}n(Nt(n)/{\it CN}{\it n}), {\square}{\it CN}{\it s(n)}, \mbox{\fbox{${\forall}a((?{\blacksquare}((({\langle\rangle}Na\backslash Sf)/{\exists}bNb)\backslash ({\langle\rangle}Na\backslash Sf))\backslash {[]^{-1}}{[]^{-1}}((({\langle\rangle}Na\backslash Sf)/{\exists}bNb)\backslash ({\langle\rangle}Na\backslash Sf)))/{\blacksquare}((({\langle\rangle}Na\backslash Sf)/{\exists}bNb)\backslash ({\langle\rangle}Na\backslash Sf)))$}}, {\blacksquare}{\forall}n(Nt(n)/{\it CN}{\it n}), {\square}{\it CN}{\it s(n)}]]\ \Rightarrow\ Sf
\using {\forall}L
\endprooftree
\justifies
[{\blacksquare}Nt(s(m))], {\square}(({\langle\rangle}{\exists}aNa\backslash Sf)/{\exists}aNa), [[{\blacksquare}{\forall}n(Nt(n)/{\it CN}{\it n}), {\square}{\it CN}{\it s(n)}, \mbox{\fbox{${\forall}f{\forall}a((?{\blacksquare}((({\langle\rangle}Na\backslash Sf)/{\exists}bNb)\backslash ({\langle\rangle}Na\backslash Sf))\backslash {[]^{-1}}{[]^{-1}}((({\langle\rangle}Na\backslash Sf)/{\exists}bNb)\backslash ({\langle\rangle}Na\backslash Sf)))/{\blacksquare}((({\langle\rangle}Na\backslash Sf)/{\exists}bNb)\backslash ({\langle\rangle}Na\backslash Sf)))$}}, {\blacksquare}{\forall}n(Nt(n)/{\it CN}{\it n}), {\square}{\it CN}{\it s(n)}]]\ \Rightarrow\ Sf
\using {\forall}L
\endprooftree
\justifies
[{\blacksquare}Nt(s(m))], {\square}(({\langle\rangle}{\exists}aNa\backslash Sf)/{\exists}aNa), [[{\blacksquare}{\forall}n(Nt(n)/{\it CN}{\it n}), {\square}{\it CN}{\it s(n)}, \mbox{\fbox{${\blacksquare}{\forall}f{\forall}a((?{\blacksquare}((({\langle\rangle}Na\backslash Sf)/{\exists}bNb)\backslash ({\langle\rangle}Na\backslash Sf))\backslash {[]^{-1}}{[]^{-1}}((({\langle\rangle}Na\backslash Sf)/{\exists}bNb)\backslash ({\langle\rangle}Na\backslash Sf)))/{\blacksquare}((({\langle\rangle}Na\backslash Sf)/{\exists}bNb)\backslash ({\langle\rangle}Na\backslash Sf)))$}}, {\blacksquare}{\forall}n(Nt(n)/{\it CN}{\it n}), {\square}{\it CN}{\it s(n)}]]\ \Rightarrow\ Sf
\using {\blacksquare}L
\endprooftree
\prooftree
\justifies
\mbox{\fbox{$Sf$}}\ \Rightarrow\ Sf
\endprooftree
\justifies
\mbox{\fbox{$Sf{\div}Sf$}}, [{\blacksquare}Nt(s(m))], {\square}(({\langle\rangle}{\exists}aNa\backslash Sf)/{\exists}aNa), [[{\blacksquare}{\forall}n(Nt(n)/{\it CN}{\it n}), {\square}{\it CN}{\it s(n)}, {\blacksquare}{\forall}f{\forall}a((?{\blacksquare}((({\langle\rangle}Na\backslash Sf)/{\exists}bNb)\backslash ({\langle\rangle}Na\backslash Sf))\backslash {[]^{-1}}{[]^{-1}}((({\langle\rangle}Na\backslash Sf)/{\exists}bNb)\backslash ({\langle\rangle}Na\backslash Sf)))/{\blacksquare}((({\langle\rangle}Na\backslash Sf)/{\exists}bNb)\backslash ({\langle\rangle}Na\backslash Sf))), {\blacksquare}{\forall}n(Nt(n)/{\it CN}{\it n}), {\square}{\it CN}{\it s(n)}]]\ \Rightarrow\ Sf
\using {\div}L
\endprooftree
\justifies
\mbox{\fbox{${\forall}f(Sf{\div}Sf)$}}, [{\blacksquare}Nt(s(m))], {\square}(({\langle\rangle}{\exists}aNa\backslash Sf)/{\exists}aNa), [[{\blacksquare}{\forall}n(Nt(n)/{\it CN}{\it n}), {\square}{\it CN}{\it s(n)}, {\blacksquare}{\forall}f{\forall}a((?{\blacksquare}((({\langle\rangle}Na\backslash Sf)/{\exists}bNb)\backslash ({\langle\rangle}Na\backslash Sf))\backslash {[]^{-1}}{[]^{-1}}((({\langle\rangle}Na\backslash Sf)/{\exists}bNb)\backslash ({\langle\rangle}Na\backslash Sf)))/{\blacksquare}((({\langle\rangle}Na\backslash Sf)/{\exists}bNb)\backslash ({\langle\rangle}Na\backslash Sf))), {\blacksquare}{\forall}n(Nt(n)/{\it CN}{\it n}), {\square}{\it CN}{\it s(n)}]]\ \Rightarrow\ Sf
\using {\forall}L
\endprooftree
\justifies
\mbox{\fbox{${\forall}f(Sf{\div}Sf)/{\exists}aNa$}}, {\blacksquare}{\forall}n(Nt(n)/{\it CN}{\it n}), {\square}{\it CN}{\it s(n)}, [{\blacksquare}Nt(s(m))], {\square}(({\langle\rangle}{\exists}aNa\backslash Sf)/{\exists}aNa), [[{\blacksquare}{\forall}n(Nt(n)/{\it CN}{\it n}), {\square}{\it CN}{\it s(n)}, {\blacksquare}{\forall}f{\forall}a((?{\blacksquare}((({\langle\rangle}Na\backslash Sf)/{\exists}bNb)\backslash ({\langle\rangle}Na\backslash Sf))\backslash {[]^{-1}}{[]^{-1}}((({\langle\rangle}Na\backslash Sf)/{\exists}bNb)\backslash ({\langle\rangle}Na\backslash Sf)))/{\blacksquare}((({\langle\rangle}Na\backslash Sf)/{\exists}bNb)\backslash ({\langle\rangle}Na\backslash Sf))), {\blacksquare}{\forall}n(Nt(n)/{\it CN}{\it n}), {\square}{\it CN}{\it s(n)}]]\ \Rightarrow\ Sf
\using {/}L
\endprooftree
\justifies
\mbox{\fbox{${\square}({\forall}f(Sf{\div}Sf)/{\exists}aNa)$}}, {\blacksquare}{\forall}n(Nt(n)/{\it CN}{\it n}), {\square}{\it CN}{\it s(n)}, [{\blacksquare}Nt(s(m))], {\square}(({\langle\rangle}{\exists}aNa\backslash Sf)/{\exists}aNa), [[{\blacksquare}{\forall}n(Nt(n)/{\it CN}{\it n}), {\square}{\it CN}{\it s(n)}, {\blacksquare}{\forall}f{\forall}a((?{\blacksquare}((({\langle\rangle}Na\backslash Sf)/{\exists}bNb)\backslash ({\langle\rangle}Na\backslash Sf))\backslash {[]^{-1}}{[]^{-1}}((({\langle\rangle}Na\backslash Sf)/{\exists}bNb)\backslash ({\langle\rangle}Na\backslash Sf)))/{\blacksquare}((({\langle\rangle}Na\backslash Sf)/{\exists}bNb)\backslash ({\langle\rangle}Na\backslash Sf))), {\blacksquare}{\forall}n(Nt(n)/{\it CN}{\it n}), {\square}{\it CN}{\it s(n)}]]\ \Rightarrow\ Sf
\using {\Box}L
\endprooftree}
$$
\vspace{0.15in}
\disp{
$((\mbox{\v{}}{\it in}\ (\iota \ \mbox{\v{}}{\it beginning}))\ [({\it Past}\ ((\mbox{\v{}}{\it create}\ (\iota \ \mbox{\v{}}{\it heaven}))\ {\it God}))\wedge ({\it Past}\ ((\mbox{\v{}}{\it create}\ (\iota \ \mbox{\v{}}{\it earth}))\ {\it God}))])$}

\bibliographystyle{plain}
\bibliography{bib260417}

\end{document}